\documentclass{article}

\PassOptionsToPackage{numbers, sort&compress}{natbib}

\usepackage[preprint, final]{neurips_2026}

\setcitestyle{numbers,square}

\usepackage[utf8]{inputenc} %
\usepackage[T1]{fontenc}    %
\usepackage{hyperref}       %
\usepackage{url}            %
\usepackage{booktabs}       %
\usepackage{amsfonts}       %
\usepackage{nicefrac}       %
\usepackage{microtype}      %
\usepackage{xcolor}         %

\usepackage{multirow}

\usepackage{fancyvrb}
\DefineVerbatimEnvironment{commands}{Verbatim}{%
  frame=single, framesep=2mm, rulecolor=\color{black!35},
  fontsize=\small, xleftmargin=1em, xrightmargin=1em}

\usepackage{pifont}
\usepackage{forest}

\usepackage{mathrsfs}

\usepackage{caption}

\definecolor{neuripsblue}{HTML}{1F4E79}
\hypersetup{
    colorlinks=true,
    allcolors=neuripsblue
}

\definecolor{plotblue}{HTML}{4477AA}
\definecolor{plotcyan}{HTML}{66CCEE}
\definecolor{plotgreen}{HTML}{228833}
\definecolor{plotorange}{HTML}{EE7733}
\definecolor{plotpurple}{HTML}{AA3377}

\usepackage{tabularx}
\usepackage{placeins}
\usepackage{siunitx}

\usepackage{tcolorbox}
\tcbuselibrary{skins}
\usepackage{etoolbox}
\usepackage{xparse}          %

\usepackage{stackengine}

\usepackage{array}

\newcolumntype{T}{%
  S[table-format=2.1]@{ }%
  S[table-format=2.1]@{ }%
  S[table-format=2.1]%
}
\newcolumntype{B}{%
  S[table-format=2.1]@{ }%
  S[table-format=2.1]@{ }%
  S[table-format=2.1(2)]%
}

\newcolumntype{U}{S[table-format=2.1(2)]} %
\newcolumntype{V}{S[table-format=2.2(3)]} %

\newcolumntype{A}{S[table-format=2.1]}

\usepackage{capt-of}

\usepackage{subcaption}

\usepackage{pgfplots}
\pgfplotsset{compat=1.18}
\usepgfplotslibrary{fillbetween}
\usepgfplotslibrary{groupplots}
\usepackage{tikz}
\usetikzlibrary{positioning, arrows.meta, calc}
\usetikzlibrary{external}
\usetikzlibrary{backgrounds}
\usetikzlibrary{plotmarks}
\usetikzlibrary{calc}
\usetikzlibrary{shapes.geometric,shapes.misc,shapes.symbols,
                shapes.arrows,shapes.multipart,shapes.callouts}
\usepgfplotslibrary{fillbetween}
\usepackage{amsmath}
\usepackage{amssymb}
\usepackage{bm}
\usepackage{stmaryrd}
\SetSymbolFont{stmry}{bold}{U}{stmry}{m}{n}
\usepackage{mathtools}
\usepackage{accents}
\usepackage{amsthm}
\usepackage{thm-restate}

\newtheorem{theorem}{Theorem}
\newtheorem{closingtheorem}[theorem]{Theorem} %
\newtheorem{lemma}{Lemma}
\newtheorem{closinglemma}[lemma]{Lemma} %
\newtheorem{proposition}{Proposition}
\newtheorem{closingproposition}[proposition]{Proposition} %
\newtheorem{corollary}{Corollary}
\newtheorem{closingcorollary}[corollary]{Corollary} %

\newtheorem{context}{Context}
\newtheorem{convention}{Convention}
\newtheorem{closingassumption}{Assumption}
\newtheorem{fact}{Fact}
\newtheorem{closingfact}[fact]{Fact}

\theoremstyle{definition}
\newtheorem{definition}{Definition}

\newtheorem{closingdefinition}[definition]{Definition} %
\newtheorem{simpleclosingdefinition}[definition]{Definition} %
\newtheorem{example}{Example}
\newtheorem{simpleexample}[example]{Example} %
\newtheorem{construction}{Construction}

\theoremstyle{remark}
\newtheorem{remark}{Remark}
\newtheorem{closingremark}[remark]{Remark}

\newcommand{\environmentEndofItalicStyle}[1]{%
  \textit{\small{}#1}%
}
\newcommand{\environmentEndofNormalStyle}[1]{%
  {\small\normalfont #1}%
}

\newcommand{\environmentEndof}[3]{%
  \penalty0\hbox{}\nobreak
  \hskip .75em plus 1fill\relax
  \hbox{#1{End of #2~#3.}}%
}

\newcommand{\definitionclosingsymbol}{\resizebox{5pt}{6pt}{$\blacksquare$}}

\newcommand{\exampleclosingsymbol}{\resizebox{3pt}{4pt}{$\blacksquare$}}

\makeatletter
\def\endclosingdefinition{%
  \environmentEndof{\environmentEndofItalicStyle}{Definition}{\thedefinition}%
  \@endtheorem
}
\def\endsimpleclosingdefinition{%
  \hfill\definitionclosingsymbol%
  \@endtheorem
}
\def\endsimpleexample{%
  \hfill\exampleclosingsymbol%
  \@endtheorem
}
\def\endexample{%
  \environmentEndof{\environmentEndofItalicStyle}{Example}{\theexample}%
  \@endtheorem
}
\def\endclosingremark{%
  \environmentEndof{\environmentEndofItalicStyle}{Remark}{\theremark}%
  \@endtheorem
}

\def\endclosingtheorem{%
  \environmentEndof{\environmentEndofNormalStyle}{Theorem}{\thetheorem}%
  \@endtheorem
}
\def\endclosinglemma{%
  \environmentEndof{\environmentEndofNormalStyle}{Lemma}{\thelemma}%
  \@endtheorem
}
\def\endclosingproposition{%
  \environmentEndof{\environmentEndofNormalStyle}{Proposition}{\theproposition}%
  \@endtheorem
}
\def\endclosingcorollary{%
  \environmentEndof{\environmentEndofNormalStyle}{Corollary}{\thecorollary}%
  \@endtheorem
}
\def\endclosingfact{%
  \environmentEndof{\environmentEndofNormalStyle}{Fact}{\thefact}%
  \@endtheorem
}
\def\endclosingassumption{%
  \environmentEndof{\environmentEndofNormalStyle}{Assumption}{\theclosingassumption}%
  \@endtheorem
}
\makeatother

\makeatletter
\DeclareFontFamily{U}{mathb}{\hyphenchar\font45}
\DeclareFontShape{U}{mathb}{m}{n}{
  <5> <6> <7> <8> <9> <10> gen * mathb
  <10.95> mathb10
  <12> <14.4> <17.28> <20.74> <24.88> mathb12
}{}
\DeclareSymbolFont{mathabxmathb}{U}{mathb}{m}{n}

\DeclareMathSymbol{\drsh}{\mathrel}{mathabxmathb}{"EB}
\makeatother

\newsavebox{\sidealignmainbox}
\newsavebox{\sidealignsidebox}

\newlength{\sidealignshift}
\newlength{\sidealignfreespace}
\newlength{\sidealigndesiredshift}

\newcommand{\sidealignbias}{0.5}

\NewDocumentCommand{\sidealignrow}{m m}{%
  \sbox{\sidealignmainbox}{%
    $\displaystyle
      \begin{aligned}[c]
        #1
      \end{aligned}
    $%
  }%
  \sbox{\sidealignsidebox}{%
    $\displaystyle
      #2%
    $%
  }%
  \setlength{\sidealigndesiredshift}{%
    \sidealignbias\wd\sidealignsidebox
  }%
  \setlength{\sidealignfreespace}{%
    \dimexpr
      \displaywidth
      -\wd\sidealignmainbox
      -\wd\sidealignsidebox
    \relax
  }%
  \ifdim\sidealignfreespace<0pt
    \setlength{\sidealignshift}{0pt}%
  \else
    \setlength{\sidealignshift}{\sidealigndesiredshift}%
    \ifdim\sidealignshift>\sidealignfreespace
      \setlength{\sidealignshift}{\sidealignfreespace}%
    \fi
  \fi
  \makebox{%
    \hspace*{\sidealignshift}%
    \usebox{\sidealignmainbox}%
    \usebox{\sidealignsidebox}%
  }%
}

\NewDocumentCommand{\sideeq}{m m m o}{%
  \sidealignrow{#1}{%
    \qquad\quad
    \text{#2}
    \qquad
    #3%
    \IfValueT{#4}{%
      \qquad\quad
      #4%
    }%
  }%
}

\NewDocumentCommand{\quanteq}{m m}{%
  \sidealignrow{#1}{%
    \qquad\quad
    #2%
  }%
}

\NewDocumentEnvironment{sidealign}{}
  {\begin{equation}}
  {\end{equation}}

\NewDocumentEnvironment{sidealign*}{}
  {\begin{equation*}}
  {\end{equation*}}

\newenvironment{proofsketch}[1][noqed]{%
  \begin{proof}[Proof sketch]%
  \def\proofsketchoption{#1}%
  \def\proofsketchqed{qed}%
  \ifx\proofsketchoption\proofsketchqed
  \else
    
  \fi
  \ignorespaces
}{%
  \end{proof}%
}

\newcommand{\partenvironmentOneletterabbreviation}{}

\newcommand{\partenvironmentEndofText}[2]{%
  End~of~\partenvironmentOneletterabbreviation.#2%
}

\newlength{\partenvironmentheaderindent}
\newlength{\partenvironmentbodyindent}
\newlength{\partenvironmentlightpostskip}
\newcommand{\partenvironmentOpeningSymbol}{%
  \raisebox{1pt}{%
    \resizebox{4pt}{4pt}{%
      $\blacktriangleright$%
    }%
  } \ 
}

\newcommand{\partenvironmentClosingSymbol}{%
  \hspace{2pt}%
  \raisebox{.1em}{%
    \resizebox{4pt}{4pt}{%
      $\blacktriangleleft$%
    }%
  }%
}

\ExplSyntaxOn

\tl_new:N   \l_partenvironment_indent_tl
\tl_new:N   \l_partenvironment_header_style_tl
\tl_new:N   \l_partenvironment_body_style_tl
\tl_new:N   \l_partenvironment_title_tl
\bool_new:N \l_partenvironment_body_italic_bool
\bool_new:N \l_partenvironment_light_bool

\keys_define:nn { mathenvs / part }
  {
    header .choice:,
    header / bold .code:n =
      { \tl_set:Nn \l_partenvironment_header_style_tl { \bfseries } },
    header / bf .meta:n = { header = bold },
    header / boldface .meta:n = { header = bold },
    header / italic .code:n =
      { \tl_set:Nn \l_partenvironment_header_style_tl { \itshape } },
    header / it .meta:n = { header = italic },
    header / bolditalic .code:n =
      { \tl_set:Nn \l_partenvironment_header_style_tl { \bfseries \itshape } },
    header / normal .code:n =
      { \tl_set:Nn \l_partenvironment_header_style_tl { \normalfont } },
    header / roman .meta:n = { header = normal },

    body .choice:,
    body / bold .code:n =
      {
        \tl_set:Nn \l_partenvironment_body_style_tl { \bfseries }
        \bool_set_false:N \l_partenvironment_body_italic_bool
      },
    body / bf .meta:n = { body = bold },
    body / boldface .meta:n = { body = bold },
    body / italic .code:n =
      {
        \tl_set:Nn \l_partenvironment_body_style_tl { \itshape }
        \bool_set_true:N \l_partenvironment_body_italic_bool
      },
    body / it .meta:n = { body = italic },
    body / bolditalic .code:n =
      {
        \tl_set:Nn \l_partenvironment_body_style_tl { \bfseries \itshape }
        \bool_set_true:N \l_partenvironment_body_italic_bool
      },
    body / normal .code:n =
      {
        \tl_set:Nn \l_partenvironment_body_style_tl { \normalfont }
        \bool_set_false:N \l_partenvironment_body_italic_bool
      },
    body / roman .meta:n = { body = normal },

    indent .choice:,
    indent / all .code:n =
      { \tl_set:Nn \l_partenvironment_indent_tl { all } },
    indent / body .code:n =
      { \tl_set:Nn \l_partenvironment_indent_tl { body } },
    indent / none .code:n =
      { \tl_set:Nn \l_partenvironment_indent_tl { none } },
    indent .default:n = all,
    noindent .meta:n = { indent = none },

    light .bool_set:N = \l_partenvironment_light_bool,
    light .default:n = true,

    title .tl_set:N = \l_partenvironment_title_tl,
  }

\cs_new_protected:Npn \partenvironment_options_reset:
  {
    \tl_set:Nn \l_partenvironment_indent_tl { none }
    \tl_set:Nn \l_partenvironment_header_style_tl { \bfseries }
    \tl_set:Nn \l_partenvironment_body_style_tl { \normalfont }
    \bool_set_false:N \l_partenvironment_body_italic_bool
    \bool_set_false:N \l_partenvironment_light_bool
    \tl_clear:N \l_partenvironment_title_tl
  }

\cs_new_protected:Npn \partenvironment_options_set:n #1
  {
    \partenvironment_options_reset:
    \tl_if_blank:nF { #1 }
      {
        \str_if_eq:eeTF { \tl_trim_spaces:n { #1 } } { indent }
          { \keys_set:nn { mathenvs / part } { indent } }
          {
            \str_if_eq:eeTF { \tl_trim_spaces:n { #1 } } { noindent }
              { \keys_set:nn { mathenvs / part } { noindent } }
              {
                \str_if_eq:eeTF { \tl_trim_spaces:n { #1 } } { light }
                  { \keys_set:nn { mathenvs / part } { light } }
                  {
                    \regex_match:nnTF { [,=] } { #1 }
                      { \keys_set:nn { mathenvs / part } { #1 } }
                      { \keys_set:nn { mathenvs / part } { title = {#1} } }
                  }
              }
          }
      }
    \bool_if:NT \l_partenvironment_light_bool
      { \tl_set:Nn \l_partenvironment_indent_tl { none } }
  }

\ExplSyntaxOff

\newcounter{definitionpart}[definition]

\renewcommand{\thedefinitionpart}{%
  \thedefinition.\arabic{definitionpart}%
}

\newcommand{\definitionpartname}{Sub-definition}

\ExplSyntaxOn

\cs_new_protected:Npn \partenvironment_header:
  {
    {%
      \tl_use:N \l_partenvironment_header_style_tl
      \bool_if:NF \l_partenvironment_light_bool
        { \partenvironmentOpeningSymbol{} }
      \definitionpartname{}~\thedefinitionpart
    }%
    \tl_if_empty:NF \l_partenvironment_title_tl
      {%
        {\normalfont\ (\tl_use:N \l_partenvironment_title_tl)}%
      }%
    .%
  }

\cs_new_protected:Npn \partenvironment_first_paragraph_begin:n #1
  {
    \leftskip=#1\relax
    \hangindent=\dimexpr\partenvironmentbodyindent-#1\relax
    \hangafter=1\relax
    \noindent
  }

\cs_new_eq:NN \partenvironment_saved_par: \par

\cs_new_protected:Npn \partenvironment_subsequent_body_paragraphs:
  {
    \cs_set_eq:NN \partenvironment_saved_par: \par
    \cs_set_protected:Npn \par
      {
        \partenvironment_saved_par:
        \leftskip=\partenvironmentbodyindent\relax
        \parindent=0pt\relax
        \hangindent=0pt\relax
        \hangafter=1\relax
      }
  }

\cs_new_protected:Npn \partenvironment_closing:nn #1#2
  {
    \mode_if_horizontal:TF
      {
        \penalty0\hbox{}\nobreak
        \hskip .75em plus 1fill\relax
        \hbox{\partenvironment_closing_text:nn {#1}{#2}}
      }
      {
        \noindent\hfill
        \hbox{\partenvironment_closing_text:nn {#1}{#2}}
      }
  }

\cs_new_protected:Npn \partenvironment_closing_text:nn #1#2
  {
    \bool_if:NTF \l_partenvironment_body_italic_bool
      { \environmentEndofNormalStyle{\partenvironmentEndofText{#1}{#2}} }
      { \environmentEndofItalicStyle{\partenvironmentEndofText{#1}{#2}} }
    \partenvironmentClosingSymbol
  }

\NewDocumentEnvironment{definitionpart}{O{}}
  {
    \partenvironment_options_set:n { #1 }
    \par\smallskip
    \refstepcounter{definitionpart}

    \group_begin:
    \renewcommand{\partenvironmentOneletterabbreviation}{SD}

    \tl_if_eq:NnTF \l_partenvironment_indent_tl { all }
      {
        \partenvironment_first_paragraph_begin:n
          { \partenvironmentheaderindent }
        \partenvironment_header:
        \hspace{5pt plus 1pt minus 1pt}%
        \partenvironment_subsequent_body_paragraphs:
      }
      {
        \tl_if_eq:NnTF \l_partenvironment_indent_tl { body }
          {
            \partenvironment_first_paragraph_begin:n { 0pt }
            \partenvironment_header:
            \hspace{5pt plus 1pt minus 1pt}%
            \partenvironment_subsequent_body_paragraphs:
          }
          {
            \noindent
            \partenvironment_header:
            \hspace{5pt plus 1pt minus 1pt}%
          }
      }

    \tl_use:N \l_partenvironment_body_style_tl
    \ignorespaces
  }
  {
    \bool_if:NTF \l_partenvironment_light_bool
      {
        \par
        \addvspace{\partenvironmentlightpostskip}
      }
      {
        \partenvironment_closing:nn {\definitionpartname{}} {\thedefinitionpart}
        \par
      }
    \group_end:
  }
\ExplSyntaxOff

\NewDocumentEnvironment{definitionpartsimple}{o}
{%
  \IfValueTF{#1}
    {\begin{definitionpart}[title={#1}]}
    {\begin{definitionpart}}
}
{%
  \end{definitionpart}
}

\usepackage{scalerel}
\usepackage[ruled,vlined,linesnumbered]{algorithm2e}
\SetAlFnt{\fontsize{10pt}{13pt}\selectfont} %
\SetProcNameFnt{\ttfamily}

\SetKwData{VarErr}{err}
\SetKwData{VarRes}{res}
\SetKwData{VarIP}{ip}
\SetKwData{VarSD}{sd}
\SetKwData{VarV}{v}
\SetKwData{VarX}{x}
\SetKwData{VarY}{y}
\SetKwData{FunctionF}{F}
\SetKwData{VarA}{a}
\SetKwData{VarB}{b}
\SetKwData{VarN}{n}
\SetKwData{VarI}{i}
\SetKwData{VarT}{t}
\SetKwData{ConstC}{c}
\SetKwData{TensorV}{V}
\SetKwData{TensorY}{Y}
\SetKwData{TensorX}{X}
\SetKwData{TensorA}{A}
\SetKwData{TensorB}{B}
\SetKwData{TensorT}{T}
\SetKwData{TensorN}{N}
\SetDataSty{texttt}   %

\SetKwProg{Fn}{Function}{:}{}
\SetKwFor{ParFor}{in parallel for}{do}{}
\SetKwInOut{Input}{Input}
\SetKwInOut{Output}{Output}
\SetKwComment{Comment}{$\triangleright$\ }{}
\SetKw{KwDownTo}{down to}
\SetKw{Allocate}{allocate}

\newcommand{\dState}{d_\textup{\textsc{s}}}
\newcommand{\nLayers}{n_\textup{\textsc{l}}}

\newcommand{\mmconf}[2]{\ensuremath{\code{MinMax}({#1},{#2})}}  %

\newcommand{\dpconf}[1]{\ensuremath{\code{DeltaProduct}(#1)}}   %

\newcommand{\smallplus}{%
  \mathbin{\vcenter{\hbox{\scalebox{0.85}{$+$}}}}%
}

\newcommand{\smalltimes}{%
  \mathbin{\vcenter{\hbox{\scalebox{0.85}{$\times$}}}}%
}

\newcommand{\cin}{\mkern-2mu\in\mkern-2mu}
\newcommand{\ctimes}{\mkern-2mu\times\mkern-2mu}

\newcommand{\newxmin}{x_\textup{\textsc{l}}}
\newcommand{\newxmax}{x_\textup{\textsc{h}}}



\newcommand{\BigO}[1]{%
  O({#1})%
}
\newcommand{\BigTheta}[1]{%
  \Theta({#1})%
}

\newcommand{\bigBigO}[1]{%
  O\bigl({#1}\bigr)%
}

\newcommand{\initialstate}[1]{\dot{#1}}

\newcommand{\timeslice}[2]{{#1}\mkern-3mu:\mkern-3mu{#2}}
\newcommand{\ambientExtensionSymb}{\uparrow}
\newcommand{\ambientExtension}[1]{{#1}^{\ambientExtensionSymb}}
\newcommand{\ambientClosureSymb}{\uparrow}
\newcommand{\ambientClosure}[1]{{#1}^{\ambientClosureSymb}}

\newcommand{\convAbbr}{\mathrm{cv}}
\newcommand{\convOperator}{\mathrm{conv}}

\newcommand{\parClassSymb}{\mathrm{p}}

\definecolor{sectionmarkblue}{RGB}{235,245,255}
\newcommand{\UpdateSectionHEAD}{BEGIN --- UPDATE ------------------------------}
\newcommand{\UpdateSectionTAIL}{END --- UPDATE ------------------------------}

\newbool{UpdateSecNumbered}
\boolfalse{UpdateSecNumbered}

\ifdef{\linenumbers}{%
  \let\UpdateSecOrigLineNumbers\linenumbers
  \let\UpdateSecOrigNoLineNumbers\nolinenumbers
  \renewcommand{\linenumbers}{\UpdateSecOrigLineNumbers\booltrue{UpdateSecNumbered}}
  \renewcommand{\nolinenumbers}{\UpdateSecOrigNoLineNumbers\boolfalse{UpdateSecNumbered}}
}{}

\newcommand{\UpdateSecSuspend}{%
  \ifdef{\UpdateSecOrigNoLineNumbers}{%
    \ifbool{UpdateSecNumbered}{\UpdateSecOrigNoLineNumbers}{}%
    \let\linenomath\relax
    \let\endlinenomath\relax
    \let\linenomathWithnumbers\relax
    \let\endlinenomathWithnumbers\relax
    \let\linenomathNonumbers\relax
    \let\endlinenomathNonumbers\relax
  }{}%
}
\newcommand{\UpdateSecResume}{%
  \ifdef{\UpdateSecOrigLineNumbers}{%
    \ifbool{UpdateSecNumbered}{\UpdateSecOrigLineNumbers}{}%
  }{}%
}

\newcommand{\UpdateSecTikzExtOff}{%
  \ifdef{\tikzexternaldisable}{\tikzexternaldisable}{}%
}
\newcommand{\UpdateSecTikzExtOn}{%
  \ifdef{\tikzexternalenable}{\tikzexternalenable}{}%
}

\NewDocumentEnvironment{updatesection}{o}
 {%
  \par\begingroup
  \UpdateSecSuspend
  \UpdateSecTikzExtOff
  \begin{tcolorbox}[
    enhanced,
    colback=sectionmarkblue,
    colframe=sectionmarkblue,
    boxrule=0pt,
    arc=0pt,
    outer arc=0pt,
    left=6pt, right=6pt, top=4pt, bottom=4pt,
    before skip=6pt, after skip=6pt,
  ]%
  \noindent\textbf{\UpdateSectionHEAD\IfValueT{#1}{ #1}}%
  \par\medskip
 }
 {%
  \par\medskip
  \noindent\textbf{\UpdateSectionTAIL\IfValueT{#1}{ #1}}%
  \end{tcolorbox}%
  \UpdateSecTikzExtOn
  \endgroup
  \UpdateSecResume
  \par
 }

\makeatletter
\newcommand{\mBig}{\bBigg@{1.25}}
\newcommand{\mBigm}{\mathrel\mBig}
\newcommand{\mbig}{\bBigg@{1.1}}
\newcommand{\mmbig}{\bBigg@{1.15}}
\makeatother

\newcommand{\minmaxdomainMainSymb}{V}

\newcommand{\minmaxdomain}{\minmaxdomainMainSymb}

\newcommand{\minmaxdatadomainSubscriptSymb}{{\scriptscriptstyle\adaoplustimes}}
\NewDocumentCommand{\minmaxdatadomain}{o}{%
  \minmaxdomainMainSymb_{%
    \mkern-2mu\minmaxdatadomainSubscriptSymb,\IfValueT{#1}{{#1}}%
  }%
}

\NewDocumentCommand{\vmin}{o o}{%
  \IfValueT{#1}{v_{\mathrm{min},#1}}%
  \IfValueF{#1}{v_{\mathrm{min}}}%
  \IfValueT{#2}{^{#2}}%
}
\NewDocumentCommand{\vmax}{o o}{%
  \IfValueT{#1}{v_{\mathrm{max},#1}}%
  \IfValueF{#1}{v_{\mathrm{max}}}%
  \IfValueT{#2}{^{#2}}%
}

\newcommand{\minmaxdatamonoid}{%
  \mathcal{M}_{\scriptscriptstyle\adaoplustimes}%
}

\newcommand{\minmaxdatatransformations}{%
  \mathcal{T}_{\scriptscriptstyle\adaoplustimes}%
}

\newcommand{\minmaxdataidentitySymbol}{E}
\newcommand{\minmaxdataidentity}{%
  \minmaxdataidentitySymbol_{\scriptscriptstyle\adaoplustimes}%
}

\newcommand{\computationInputDefaultSymbol}{I}
\newcommand{\computationInputStyle}[1]{\mathcal{#1}}
\NewDocumentCommand{\computationInput}{o}{%
  \IfValueF{#1}{\computationInputStyle{\computationInputDefaultSymbol}}%
  \IfValueT{#1}{\computationInputStyle{#1}}%
}

\newcommand{\cmark}{\ding{51}}

\newcommand{\dlatching}[1]{\texttt{Latching}(\ensuremath{#1})}

\newcommand{\dlatchingSmall}[1]{{\small\texttt{Latching}(\ensuremath{#1})}}
\newcommand{\dsequences}[1]{\texttt{Sequences}(\ensuremath{#1})}
\newcommand{\dsequencesLongRange}[1]{\texttt{SequencesLR}(\ensuremath{#1})}

\newcommand{\dsequencesSmall}[1]{{\small \texttt{Sequences}(\ensuremath{#1})}}
\newcommand{\dheads}[1]{\texttt{Heads}(\ensuremath{#1})}

\newcommand{\dheadsSmall}[1]{{\small \texttt{InductionHeads}(\ensuremath{#1})}}

\newcommand{\dword}[1]{\texttt{Word}(\ensuremath{#1})}

\newcommand{\mypar}[1]{\vspace{0em}\paragraph{#1}}

\newcommand{\compactParagraphBefore}{0.6ex}
\newcommand{\compactParagraphAfter}{0.4em}

\makeatletter
\newcommand{\compactParagraph}[2][]{%
  \def\compactParagraph@prefix{}%
  \if\relax\detokenize{#1}\relax
  \else
    \@ifundefined{c@compactParagraphGroup@#1}{%
      \newcounter{compactParagraphGroup@#1}%
    }{}%
    \refstepcounter{compactParagraphGroup@#1}%
    \edef\compactParagraph@prefix{%
      \arabic{compactParagraphGroup@#1}.\noexpand\nobreakspace
    }%
  \fi
  \@startsection{paragraph}{4}{\z@}%
    {\compactParagraphBefore}%
    {-\compactParagraphAfter}%
    {\normalfont\normalsize\bfseries}%
    *{\compactParagraph@prefix#2}%
}
\makeatother

\newcommand{\smallminus}{\mathord{\raisebox{.1em}{\scalebox{0.7}{$-$}}}}

\newcommand{\msmallminus}{\mathord{\raisebox{.1em}{\scalebox{0.8}{$-$}}}}
\newcommand{\msmallplus}{\mathord{\raisebox{.0em}{\scalebox{0.8}{$+$}}}}

\newcommand{\dvSymbol}{\mathsf{D}}
\newcommand{\dv}{\dvSymbol}

\newcommand{\dvopSeparator}{\mkern0mu}
\newcommand{\dvop}[1]{\dvSymbol\dvopSeparator #1}

\newcommand{\pdvop}[2]{\ensuremath{\dvSymbol_{#2}{#1}}}

\newcommand{\blockindex}[1]{{[#1]}}
\newcommand{\inputblock}[2]{#1_{\blockindex{#2}}}
\newcommand{\bdvop}[2]{\dvSymbol_{\blockindex{#2}} #1}

\newcommand{\sdvStyle}[1]{\tilde{#1}}
\NewDocumentCommand{\sdv}{o}{%
  \sdvStyle{\dvSymbol}%
  \IfValueT{#1}{^{#1}}%
}
\NewDocumentCommand{\sjacob}{o}{%
  \sdvStyle{\mathrm{J}}%
  \IfValueT{#1}{^{#1}}%
}
\NewDocumentCommand{\snabla}{o}{%
  \sdvStyle{\nabla}%
  \IfValueT{#1}{^{#1}}%
}

\NewDocumentCommand{\sdvop}{o m}{%
  \IfValueTF{#1}{%
    \sdv[#1]%
  }{%
    \sdv%
  }%
  #2%
}

\NewDocumentCommand{\bsdvop}{o m m}{%
  \IfValueTF{#1}{%
    \sdv[#1]_{#3}%
  }{%
    \sdv_{#3}%
  }%
  #2%
}

\NewDocumentCommand{\psdvop}{o m m}{%
  \IfValueTF{#1}{%
    \bsdvop[#1]{#2}{#3}%
  }{%
    \bsdvop{#2}{#3}%
  }%
}

\newcommand{\subdiff}{\partial_\circ}
\newcommand{\supdiff}{\partial^{\scriptscriptstyle +}_\circ}

\NewDocumentCommand{\intensionalderivatives}{o}{%
  \partial_\bullet%
  \IfValueT{#1}{%
    \mkern1mu%
    {#1}%
  }%
}

\newcommand{\intensionalderivativesop}[1]{\intensionalderivatives #1}

\newcommand{\targetspace}[1]{\hat{#1}}

\newcommand{\lossfnsty}[1]{#1}
\NewDocumentCommand{\lossfn}{o o}{%
  \ensuremath{%
    \lossfnsty{\ell}%
    \IfValueT{#1}{%
      _{#1\IfValueT{#2}{\text{-}#2}}%
    }%
  }
}

\newcommand{\lossNameSty}[1]{\textup{\textsc{#1}}}

\newcommand{\selossName}{\lossNameSty{se}}
\newcommand{\seloss}{%
  \lossfn[\selossName]%
}

\newcommand{\bcelossName}{\lossNameSty{bce}}
\newcommand{\bceloss}{%
  \lossfn[\bcelossName]%
}

\newcommand{\softmaxcelossName}{\lossNameSty{ce}}
\NewDocumentCommand{\softmaxceloss}{o}{%
  \IfValueF{#1}{\lossfn[\softmaxcelossName]}%
  \IfValueT{#1}{\lossfn[\softmaxcelossName][#1]}%
}

\newcommand{\Lossfnsty}[1]{\mathcal{#1}}
\newcommand{\LossfnSymb}{\Lossfnsty{L}}
\NewDocumentCommand{\Lossfn}{m o}{%
  \ensuremath{%
    \LossfnSymb_{#1}%
    \IfValueT{#2}{^{#2}}%
  }
}
\newcommand{\LossGrowth}[3]{%
  \ensuremath{%
    \operatorname{%
      \mathsf{G}_{{\nabla\mkern-3mu\LossfnSymb}}%
      ^{#1,#2,#3}%
    }
  }
}
\NewDocumentCommand{\cgnode}{m o o}{%
  \mathring{#1}%
  \IfValueT{#2}{_{#2}}%
  \IfValueT{#3}{^{#3}}%
}

\newcommand{\matgraph}[2]{%
  {#1}^{#2}%
}
\newcommand{\vgraphext}[3]{%
  #1
  \xRightarrow[\mathrm{v}]{#2}
  #3%
}

\NewDocumentCommand{\cglabel}{m o o}{%
  \check{#1}%
  \IfValueT{#2}{_{#2}}%
  \IfValueT{#3}{^{#3}}%
}
\NewDocumentCommand{\cgval}{m o o}{%
  \hat{#1}%
  \IfValueT{#2}{_{#2}}%
  \IfValueT{#3}{^{#3}}%
}
\NewDocumentCommand{\cgvalparop}{m m m o o}{%
  \hat{#1}\!\left(#2;#3\right)%
  \IfValueT{#4}{_{#4}}%
  \IfValueT{#5}{^{#5}}%
}
\NewDocumentCommand{\cgvalconstparop}{m m m m}{%
  \hat{G}_\mathrm{c}^\parClassSymb[#1, #2](#3;#4)%
}
\NewDocumentCommand{\widecgval}{m o o}{%
  \widehat{#1}%
  \IfValueT{#2}{_{#2}}%
  \IfValueT{#3}{^{#3}}%
}

\newcommand{\mathraisebox}[2]{\raisebox{#1}{\ensuremath{#2}}}
\newcommand{\mathscalebox}[2]{\scalebox{#1}[#1]{\ensuremath{#2}}}
\newcommand{\idx}[2]{%
  \ensuremath{#1\mathraisebox{0.2ex}{%
      \mathscalebox{0.8}{%
      \scalebox{1.1}[1]{\ensuremath{[}}\mathscalebox{1.1}{\mathraisebox{-0.15ex}{#2}}\scalebox{1.1}[1]{\ensuremath{]}}
      }%
  }}%
}%

\newcommand{\oplustimes}{%
  \mathbin{%
    \ooalign{$\oplus$\cr\hidewidth$\otimes$\hidewidth\cr}%
  }%
}
\NewDocumentCommand{\minmax}{o}{%
  \mathbin{%
    \ooalign{%
      $\oplus$\cr
      \hidewidth$\otimes$\hidewidth\cr
    }%
  }%
  \IfValueT{#1}{(#1)}%
}
\NewDocumentCommand{\bigminmax}{o}{%
  \mathbin{%
    \ooalign{%
      $\oplus$\cr
      \hidewidth$\otimes$\hidewidth\cr
    }%
  }%
  \IfValueT{#1}{\big(#1\big)}%
}
\newcommand{\minmaxopsymb}{%
  \mathbin{%
    \ooalign{%
      $\oplus$\cr
      \hidewidth$\otimes$\hidewidth\cr
    }%
  }
}
\newcommand{\minmaxop}[3]{%
  #1 \minmaxopsymb (#2,#3)
}

\newcommand{\minmaxopfunc}[4]{%
  \minmaxopsymb_{#1}(#2,#3,#4)
}

\newcommand{\minmaxcomp}{\oplustimes}
\newcommand{\mathsidecomment}[1]{&&\text{\footnotesize | #1}}

\newcommand{\transposeop}{\ensuremath{\top}}

\newcommand{\ceta}[1]{\mathcal{C}_\eta(#1)}
\newcommand{\mealy}[1]{\mathcal{M}(#1)}
\newcommand{\mceta}[1]{\mealy{\ceta{#1}}}

\newcommand{\groupsty}[1]{\mathsf{#1}}
\newcommand{\gop}[1]{\hspace{.15ex}\cdot_{\scaleobj{0.7}{#1}}\hspace{-.1ex}}
\newcommand{\gsym}[1]{\mathrm{Sym}(#1)}

\newcommand{\ol}{\overline}

\newcommand{\dec}{\ensuremath{\mathrm{dec}}}
\newcommand{\enc}{\ensuremath{\mathrm{enc}}}

\newcommand{\todo}[1]{}

\newcommand{\im}{\ensuremath{\operatorname{Im}}}

\newcommand{\powersetplus}[1]{\operatorname{\ensuremath{\mathcal{P}_+}}(#1)}

\newcommand{\defeq}{\vcentcolon=}
\newcommand{\eqdef}{=\vcentcolon}
\renewcommand{\vec}{\mathbf} 

\newcommand{\ptuple}[1]{(#1)} 
\newcommand{\bigptuple}[1]{\big(#1\big)} 
\newcommand{\mbigptuple}[1]{\mbig(#1\mbig)} 
\newcommand{\mmbigptuple}[1]{\mmbig(#1\mmbig)} 
\newcommand{\Bigptuple}[1]{\Big(#1\Big)} 
\newcommand{\mBigptuple}[1]{\mBig(#1\mBig)} 
\newcommand{\tuple}[1]{\langle #1 \rangle} 
\newcommand{\bigtuple}[1]{\big\langle #1 \big\rangle}

\newcommand{\mtuplesep}{\,|\,} 
\newcommand{\mtuple}[2]{\langle #1 \mtuplesep #2 \rangle}

\newcommand{\cptuplesep}{;\mkern1mu} 
\newcommand{\bigcptuplesep}{;\mkern2mu} 
 
\newcommand{\Bigcptuplesep}{;\mkern4mu} 
\newcommand{\cptuple}[2]{\ptuple{#1\cptuplesep #2}} 
\newcommand{\bigcptuple}[2]{\bigptuple{#1\bigcptuplesep #2}} 
\newcommand{\mbigcptuple}[2]{\mbigptuple{#1\Bigcptuplesep #2}} 
\newcommand{\mBigcptuple}[2]{\mBigptuple{#1\Bigcptuplesep #2}} 
\newcommand{\Bigcptuple}[2]{\Bigptuple{#1\Bigcptuplesep #2}} 

\newcommand{\ctuplesep}{;\mkern1mu} 
\newcommand{\bigctuplesep}{;\mkern2mu} 
\newcommand{\ctuple}[2]{\tuple{#1\ctuplesep #2}} 
\newcommand{\bigctuple}[2]{\bigtuple{#1\bigctuplesep #2}} 

\newcommand{\mctuple}[3]{\langle #1 \cptuplesep #2 \mtuplesep #3 \rangle} 
\newcommand{\bigmctuple}[3]{\big\langle #1 \cptuplesep #2 \mtuplesep #3 \big\rangle}

\newcommand{\ivalsep}{..}
\newcommand{\subivalsep}{\mathinner{\ldotp\!\ldotp}}
\newcommand{\ival}[1]{[{#1}]}
\newcommand{\sival}[2]{{[#2]}}
\newcommand{\iival}[2]{{[{#1}\ivalsep{#2}]}}

\newcommand{\subiival}[2]{[{#1}\subivalsep{#2}]}
\newcommand{\iivallo}[2]{{({#1}\ivalsep{#2}]}}
\newcommand{\iivalro}[2]{{[{#1}\ivalsep{#2})}}
\newcommand{\iivalo}[2]{{({#1}\ivalsep{#2})}}

\newcommand{\seq}[3]{#1_{\scriptscriptstyle\subiival{#2}{#3}}}

\newcommand{\parallelop}{\parallel}

\newcommand{\restr}[2]{{#1}\raisebox{-0.2em}{\ensuremath{|_{#2}}}}
\newcommand{\graphextrestr}[2]{\restr{#1}{\mathrm{ext}=#2}}
\newcommand{\graphindexrestr}[2]{\restr{#1}{#2}}

\newcommand{\naturals}{\mathbb{N}}
\newcommand{\positivesuperscript}{\scalebox{0.55}{$>\!0$}}

\newcommand{\nonnegativesuperscript}{\scalebox{0.55}{$\geq\!0$}}
\newcommand{\posnaturals}{{\mathbb{N}^{\positivesuperscript}}}
\newcommand{\reals}{\mathbb{R}}
\newcommand{\posreals}{{\mathbb{R}^{\positivesuperscript}}}
\newcommand{\nonnegreals}{{\mathbb{R}^{\nonnegativesuperscript}}}

\newcommand{\finitestate}{\ensuremath{\eta}-finite}

\newcommand{\etafinite}{\finitestate}

\newcommand{\Repval}[1]{%
  \mathopen{\langle\mkern2mu\kern-3.5pt\langle}%
  #1%
  \mathclose{\rangle\mkern2mu\kern-3.5pt\rangle}%
}
\newcommand{\repval}[1]{%
  {\check{#1}}%
}

\newcommand{\dynop}[1]{\mathrm{dyn}(#1)}
\newcommand{\bigdynop}[1]{\mathrm{dyn}\big(#1\big)}
\newcommand{\pinst}[2]{#1[#2]}
\newcommand{\pinstLight}[2]{#1_{#2}}
\newcommand{\ppinst}[2]{#1[#2]}
\newcommand{\Pinst}[2]{\pinst{(#1)}{#2}}

\newcommand{\fpinst}[3]{#1\cptuple{#2}{#3}}
\newcommand{\bigfpinst}[3]{#1\bigcptuple{#2}{#3}}
\newcommand{\mbigfpinst}[3]{#1\mbigcptuple{#2}{#3}}
\newcommand{\mBigfpinst}[3]{#1\mBigcptuple{#2}{#3}}
\newcommand{\Bigfpinst}[3]{#1\Bigcptuple{#2}{#3}}

\newcommand{\timedomain}{\mathbb{T}}

\newcommand{\statefuncSymb}{\chi}
\newcommand{\statefunc}{\statefuncSymb}
\newcommand{\indexedStateToStateFuncfamilySuperscriptVerticalOffset}{-0.15ex}
\NewDocumentCommand{\indexedStateToStateFuncfamily}{m o o o}{%
  \repval{#1}%
  \IfValueT{#3}{{%
      ^{%
        \raisebox{%
          \indexedStateToStateFuncfamilySuperscriptVerticalOffset%
          }{%
          \ensuremath{%
            \scriptstyle #2%
          }%
        }%
      }%
      _{#3}%
  }}%
  \IfValueT{#4}{%
    (%
      {#4}%
    )%
  }%
}
\NewDocumentCommand{\indexedParStateToStateFuncfamily}{m o o o o}{%
  \repval{#1}%
  \IfValueT{#4}{%
    [  %
      {#3}, 
      {#4};
      \mkern1mu
      {#2} 
    ]  %
  }%
  \IfValueT{#5}{%
    (  %
      {#5}
    )  %
  }%
}
\NewDocumentCommand{\indexedDynParamToStateFuncfamily}{m o o o o o}{%
  \repval{#1}%
  \IfValueT{#3}{%
    [  %
      {#2}, 
      {#3},
      {#4} 
    ]  %
  }%
  \IfValueT{#6}{%
    \IfValueT{#5}{%
      \big(  %
        {#5}
      \big)  %
    }%
  }%
  \IfValueF{#6}{%
    \IfValueT{#5}{%
      (  %
        {#5}
      )  %
    }%
  }%
}
\NewDocumentCommand{\indexedDynParamToLatestStateFuncfamily}{m o o o}{%
  \repval{#1}%
  \IfValueT{#3}{%
    [  %
      {#2}, 
      {#3}
    ]  %
  }%
  \IfValueT{#4}{%
    (  %
      {#4}
    )  %
  }%
}
\NewDocumentCommand{\indexedParamToStateFuncfamily}{m o o o}{%
  \repval{#1}%
  \IfValueT{#3}{%
    [  %
      {#2}, 
      {#3} 
    ]  %
  }%
  \IfValueT{#4}{%
    (  %
      {#4}
    )  %
  }%
}
\NewDocumentCommand{\indexedParamToLatestStateFuncfamily}{m o o}{%
  \repval{#1}%
  \IfValueT{#2}{%
    [  %
      {#2}
    ]  %
  }%
  \IfValueT{#3}{%
    (  %
      {#3}
    )  %
  }%
}

\newcommand{\infseqOperatorSymb}{\omega}
\newcommand{\infseqs}[1]{#1^\infseqOperatorSymb}

\newcommand{\binfseqOperatorSymb}{\omega\mathrm{b}}
\newcommand{\binfseqs}[1]{#1^{\binfseqOperatorSymb}}

\newcommand{\narroweq}{%
  \mathrel{\scalebox{0.8}[1]{$\scriptstyle=$}}%
}
\newcommand{\compactSeqSpec}[4]{(#1_{#2})_{#2 \narroweq #3}^{\smash{\raisebox{-0.15ex}{\ensuremath{\scriptstyle #4}}}}}

\newcommand{\boundsubscript}[1]{\scalebox{0.55}{$\!\leq\!#1$}}
\newcommand{\truncation}[2]{#1_{\boundsubscript{#2}}}

\newcommand{\linop}{\mathbb{L}}
\newcommand{\flattenop}{\mathrm{flatten}}
\newcommand{\closureopp}[2]{\mathrm{CL}_{#2}(#1)}
\newcommand{\closureop}[1]{\mathrm{CL}(#1)}
\newcommand{\openballopp}[2]{\mathrm{Ball}_{#2}(#1)}
\newcommand{\openballop}[1]{\mathrm{Ball}(#1)}

\newcommand{\closedballop}[1]{\overline{\mathrm{Ball}}(#1)}

\newcommand{\classHyperparmSeparator}{\mkern1mu|\mkern1mu}
\newcommand{\classNameStyle}[1]{%
  \raisebox{-.15ex}{\ensuremath{\mathrm{#1}}}%
}
\newcommand{\classBrackets}[1]{%
  [\mkern1mu#1\mkern1mu]%
}
\newcommand{\classTypeStyle}[1]{%
  \raisebox{-.15ex}{\ensuremath{#1}}%
}
\NewDocumentCommand{\classGeneric}{m m m}{%
  \classTypeStyle{#1}%
  \IfValueTF{#3}{%
    \classBrackets{%
      \classNameStyle{#2}%
      \classHyperparmSeparator
      #3%
    }%
  }{%
    \classBrackets{%
      \classNameStyle{#2}%
    }%
  }%
}

\newcommand{\ClassSymbStyle}[1]{\mathcal{#1}}

\newcommand{\classUnitsSymb}{\ClassSymbStyle{U}}
\NewDocumentCommand{\classunits}{m o}{%
  \classGeneric{\classUnitsSymb}{#1}{#2}%
}
\newcommand{\classParUnitsSymb}{\classUnitsSymb^\parClassSymb}
\NewDocumentCommand{\classparunits}{m o}{%
  \classGeneric{\classParUnitsSymb}{#1}{#2}%
}

\newcommand{\classParNeuralUnitsSymb}{\classUnitsSymb^\parClassSymb_\analyticClassScriptSymb}
\NewDocumentCommand{\classparneuralunits}{m o}{%
  \classGeneric{\classParNeuralUnitsSymb}{#1}{#2}%
}

\newcommand{\classModelsSymb}{\ClassSymbStyle{M}}
\NewDocumentCommand{\classmodels}{m o}{%
  \classGeneric{\classModelsSymb}{#1}{#2}%
}

\newcommand{\classDynamicsSymb}{\ClassSymbStyle{D}}
\NewDocumentCommand{\classdynamics}{m o}{%
  \classGeneric{\classDynamicsSymb}{#1}{#2}%
}

\newcommand{\classParDynamicsSymb}{\classDynamicsSymb^\parClassSymb}
\NewDocumentCommand{\classpardynamics}{m o}{%
  \classGeneric{\classParDynamicsSymb}{#1}{#2}%
}

\newcommand{\classSystemsSymb}{\ClassSymbStyle{S}}
\NewDocumentCommand{\classSystems}{m o}{%
  \classGeneric{\classSystemsSymb}{#1}{#2}%
}

\newcommand{\classParSystemsSymb}{\classSystemsSymb^\parClassSymb}
\NewDocumentCommand{\classParSystems}{m o}{%
  \classGeneric{\classParSystemsSymb}{#1}{#2}%
}

\newcommand{\classneuralunits}[1]{\ClassSymbStyle{U}^{\mathrm{n}}_\mathrm{#1}}

\newcommand{\rncdynop}{\operatorname{RNC}_{\mathrm{dyn}}}
\newcommand{\rncdynparop}{\operatorname{RNC}^\parClassSymb_{\mathrm{dyn}}}
\newcommand{\rncop}{\operatorname{RNC}}
\newcommand{\rncparop}{\operatorname{RNC}^\parClassSymb}
\newcommand{\paroutfunc}{\gamma}
\newcommand{\paroutfuncp}[1]{\gamma_{#1}}

\newcommand{\samplespace}{\mathcal{Z}}
\newcommand{\samplespacep}[1]{\samplespace^{#1}}
\newcommand{\truncsamplespace}[1]{\truncation{\samplespace}{#1}}
\newcommand{\truncsamplespacep}[2]{\truncation{\samplespace^{#1}}{#2}}

\newcommand{\horizontruncsamplespacep}[3]{%
  \truncation{%
    \samplespace^{%
      #1,%
      \boundseqsuperscript{#3}%
    }%
  }{%
    #2%
  }%
}

\NewDocumentCommand{\admissiblelossfunctions}{o}{%
  \mathfrak{L}%
  \IfValueT{#1}{^{#1}}%
}
\NewDocumentCommand{\outputadmissiblelossfunctions}{o}{%
  \mathfrak{L}_{\mathrm{out}}%
  \IfValueT{#1}{^{#1}}%
}

\makeatletter
\newcommand{\adaoplustimes}{%
  \mathbin{\mathpalette\oplustimesauxss\relax}%
}
\newcommand{\oplustimesauxss}[2]{%
  \ooalign{%
    $\m@th#1\oplus$\cr
    \hidewidth$\m@th#1\otimes$\hidewidth\cr
  }%
}
\makeatother

\newcommand{\analyticlatticeabbreviation}{%
  \mathord{\mathpalette\analyticlatticeabbreviationaux\relax}%
}

\newcommand{\analyticlatticeabbreviationaux}[2]{%
  \vcenter{\hbox{%
    \setbox0=\hbox{$#1\mathrm{A}$}%
    \tikzset{external/export next=false}%
    \tikz[
      x=\wd0,
      y=\ht0,
      baseline=0pt,
      line cap=round,
      line join=round,
      line width=.11\wd0
    ]{%
      \draw (0.07,0) -- (0.56,1);

      \draw (0.56,0) -- (0.56,1);

      \draw (0.75,0) -- (0.75,1);

      \draw (0.25,0.3) -- (0.56,0.3);

      \draw (0.75,0) -- (1.1,0);
    }%
  }}%
}

\newcommand{\minmaxsdvSubscriptOffset}{\mkern-2mu}
\newcommand{\minmaxsdvSubscript}{{\minmaxsdvSubscriptOffset\scriptscriptstyle\adaoplustimes}}
\NewDocumentCommand{\minmaxsdv}{o}{%
  \IfValueTF{#1}{%
    \sdv[#1]_\minmaxsdvSubscript%
  }{%
  \sdv_\minmaxsdvSubscript%
  }%
}

\NewDocumentCommand{\minmaxsdvop}{o m}{%
  \IfValueTF{#1}{%
    \minmaxsdv[#1]\dvopSeparator #2%
  }{%
    \minmaxsdv\dvopSeparator #2%
  }%
}

\NewDocumentCommand{\analyticlatticesdv}{o}{%
  \IfValueTF{#1}{%
    \sdv[#1]_{\mkern-4mu\analyticlatticeabbreviation}%
  }{%
  \sdv_{\mkern-4mu\analyticlatticeabbreviation}%
  }%
}

\NewDocumentCommand{\analyticlatticesdvop}{o m}{%
  \IfValueTF{#1}{%
    \analyticlatticesdv[#1] #2%
  }{%
    \analyticlatticesdv #2%
  }%
}

\newcommand{\classprimitivesymb}{\mathcal{P}}
\newcommand{\classfuncsymb}{\mathcal{F}}

\newcommand{\minmaxclassprimitive}{%
  \classprimitivesymb_{\scriptscriptstyle\adaoplustimes}%
}

\newcommand{\minmaxprimitivefuncclass}{%
  \minmaxclassprimitive%
}

\newcommand{\minmaxfuncclass}{%
  \classfuncsymb_{\scriptscriptstyle\adaoplustimes}%
}

\newcommand{\analyticClassScriptSymb}{{\scriptscriptstyle\mathrm{AN}}}
\NewDocumentCommand{\analyticfuncclass}{o o}{%
  \classfuncsymb_\analyticClassScriptSymb%
  \IfValueT{#1}{(#1,#2)}%
}
\NewDocumentCommand{\analyticparamfuncclass}{o o o}{%
  \classfuncsymb^\parClassSymb_\analyticClassScriptSymb%
  \IfValueT{#1}{(#1,#2; #3)}%
}
\newcommand{\neuralFuncClass}{\neuralFuncClass}

\newcommand{\pureprimitiveminmaxfuncclass}{%
  \classprimitivesymb_\mathrm{MM}%
}

\newcommand{\pureminmaxfuncclass}{%
  \classfuncsymb_\mathrm{MM}%
}

\newcommand{\classparfunc}{\classfuncsymb^\parClassSymb}

\newcommand{\ind}[1]{\mathbf{1}[\mkern2mu #1 \mkern2mu]}

\newcommand{\getentry}[2]{(#1)_{#2}}
\newcommand{\sgetentry}[2]{#1_{#2}}
\newcommand{\biggetentry}[2]{\big(#1\big)_{#2}}

\newcommand{\getrow}[2]{\getentry{#1}{#2,:}}

\NewDocumentCommand{\constgraph}{m m}{%
  G_{\mathrm{c}}[#1, #2]%
}
\NewDocumentCommand{\constgraphval}{m m}{%
  \hat{G}_{\mathrm{c}}[#1, #2]%
}

\NewDocumentCommand{\constparamgraph}{m m}{%
  G_\mathrm{c}^{\parClassSymb}[#1, #2]%
}
\NewDocumentCommand{\constpargraph}{m m}{%
  \constparamgraph{#1}{#2}%
}

\NewDocumentCommand{\indexedcompgraphfamily}{o}{%
  \mathcal{G}%
  \IfValueT{#1}{\!\left[#1\right]}%
}
\NewDocumentCommand{\bigIndexedcompgraphfamily}{o}{%
  \mathcal{G}%
  \IfValueT{#1}{\mBig[#1\mBig]}%
}
\NewDocumentCommand{\mBigIndexedcompgraphfamily}{o}{%
  \mathcal{G}%
  \IfValueT{#1}{\mBig[#1\mBig]}%
}

\NewDocumentCommand{\customIndexedcompgraphfamily}{m o}{%
  #1%
  \IfValueT{#2}{\!\left[#2\right]}%
}
\NewDocumentCommand{\bigCustomIndexedcompgraphfamily}{m o}{%
  #1%
  \IfValueT{#2}{\mBig[#2\mBig]}%
}
\NewDocumentCommand{\mBigCustomIndexedcompgraphfamily}{m o}{%
  #1%
  \IfValueT{#2}{\mBig[#2\mBig]}%
}

\newcommand{\compgraphfamilysymb}{\mathcal{G}}
\NewDocumentCommand{\compgraphfamily}{m o}{%
  \compgraphfamilysymb^{\mathrm{#1}}%
  \IfValueT{#2}{\!\left[#2\right]}%
}

\NewDocumentCommand{\analyticlatticegraphclass}{o}{%
  \mathcal{G}_{\mkern-3mu\analyticlatticeabbreviation}%
  \IfValueT{#1}{\!\left[#1\right]}%
}

\NewDocumentCommand{\analyticlatticestategraphclass}{m o o o}{%
  \mathcal{G}_{\mkern-5mu\analyticlatticeabbreviation}%
  \!\left(
    #1%
    \IfValueT{#2}{;#2}%
    \IfValueT{#3}{%
      \IfValueTF{#2}{,}{;}%
      #3%
    }%
    \IfValueT{#4}{%
      \IfValueTF{#2}{%
        ,#4%
      }{%
        \IfValueTF{#3}{,#4}{;#4}%
      }%
    }%
  \right)%
}

\NewDocumentCommand{\analyticlatticerecgraphclass}{o}{%
  \mathcal{G}_{%
    \mathrm{rec},
    \mkern-5mu\analyticlatticeabbreviation%
  }%
  \IfValueT{#1}{\!\left(#1\right)}%
}

\NewDocumentCommand{\analyticlatticeobjectivegraphclass}{m o O{\lossfn}}{%
  \analyticlatticegraphclass^{\textup{\textsc{obj}}}%
  \!\left(
    #1,#3%
    \IfValueT{#2}{;#2}%
  \right)%
}

\NewDocumentCommand{\minmaxgraphclass}{o}{%
  \mathcal{G}_{{\scriptscriptstyle\adaoplustimes}}%
  \IfValueT{#1}{\!\left[#1\right]}%
}

\newcommand{\minmaxstategraphclasssym}{\mathcal{G}}
\NewDocumentCommand{\minmaxstategraphclass}{m o o o}{%
  \minmaxstategraphclasssym_{{\scriptscriptstyle\adaoplustimes}}%
  \!\left(
    #1%
    \IfValueT{#2}{;#2}%
    \IfValueT{#3}{%
      \IfValueTF{#2}{,}{;}%
      #3%
    }%
    \IfValueT{#4}{%
      \IfValueTF{#2}{%
        ,#4%
      }{%
        \IfValueTF{#3}{,#4}{;#4}%
      }%
    }%
  \right)%
}

\NewDocumentCommand{\minmaxrecgraphclass}{o}{%
  \mathcal{G}_{%
    \mathrm{rec},
    {\scriptscriptstyle\adaoplustimes}%
  }%
  \IfValueT{#1}{\!\left(#1\right)}%
}

\NewDocumentCommand{\minmaxobjectivegraphclass}{m o o}{%
  \minmaxgraphclass^{\textup{\textsc{obj}}}%
  \!\left(
  #1       %
  \IfValueT{#3}{%
    ;      %
    #3     %
    \IfValueT{#2}{%
      ,    %
      #2   %
    }%
  }%
  \right)%
}

\NewDocumentCommand{\minmaxconstructedobjectivegraphclass}{m o O{\lossfn}}{%
  \mathcal{C}^{\textup{\textsc{obj}}}%
  \!\left(
    #1,#3%
    \IfValueT{#2}{;#2}%
  \right)%
}

\NewDocumentCommand{\minmaxextensionclass}{o}{%
  \mathcal{E}_{{\scriptscriptstyle\adaoplustimes}}%
  \IfValueT{#1}{%
    \!\left(
      #1
    \right)%
  }%
}

\NewDocumentCommand{\stategraphclass}{m o m o o o}{%
  \mathcal{#1}%
  \IfValueT{#2}{^{#2}}%
  \!\left(
    #3%
    \IfValueT{#4}{;#4}%
    \IfValueT{#5}{%
      \IfValueTF{#4}{,}{;}%
      #5%
    }%
    \IfValueT{#6}{%
      \IfValueTF{#4}{%
        ,#6%
      }{%
        \IfValueTF{#5}{,#6}{;#6}%
      }%
    }%
  \right)%
}

\NewDocumentCommand{\selgraphsymb}{o}{%
  \Gamma%
  \IfValueT{#1}{\!\left(#1\right)}%
}
\NewDocumentCommand{\mBigSelgraphsymb}{o}{%
  \Gamma%
  \IfValueT{#1}{\mBig(#1\mBig)}%
}
\NewDocumentCommand{\bigSelgraphsymb}{o}{%
  \Gamma%
  \IfValueT{#1}{\big(#1\big)}%
}

\newcommand{\minmaxselgraphsymbSubscriptOffset}{\mkern-4mu}
\newcommand{\minmaxselgraphsymbSubscriptScale}{0.8}

\newcommand{\minmaxselgraphsymbSubscript}{%
  {%
    \minmaxselgraphsymbSubscriptOffset
    \mathchoice
      {\scriptscriptstyle\adaoplustimes}%
      {\scriptscriptstyle\adaoplustimes}%
      {\scriptscriptstyle\adaoplustimes}%
      {%
        \scalebox{\minmaxselgraphsymbSubscriptScale}{%
          $\scriptscriptstyle\adaoplustimes$%
        }%
      }%
  }%
}

\NewDocumentCommand{\minmaxselgraphsymb}{o}{%
  \Gamma%
  _\minmaxselgraphsymbSubscript%
  \IfValueT{#1}{\!\left(#1\right)}%
}

\NewDocumentCommand{\selgraphoperator}{o}{%
  \Gamma_{\IfValueTF{#1}{#1}{\sdv}}%
}

\NewDocumentCommand{\selgraph}{m o}{%
  \Gamma_{\IfValueTF{#2}{#2}{\sdv}}%
  \!\left(#1\right)%
}

\newcommand{\actfun}[1]{\mathcal{A}_{#1}}

\NewDocumentCommand{\arity}{m o}{%
  \mathrm{ar}\IfValueT{#2}{_{#2}}(#1)%
}
\newcommand{\arityin}[1]{%
  \arity{#1}[\mathrm{in}]%
}

\newcommand{\arityout}[1]{%
  \arity{#1}[\mathrm{out}]%
}

\newcommand{\minmaxscalaralgebra}{%
\mathbb{M}%
}

\NewDocumentCommand{\id}{o}{%
  \mathrm{id}%
  \IfValueT{#1}{_{#1}}%
}

\NewDocumentCommand{\classparfeedforwardreps}{o}{%
  \mathfrak{R}^{\parClassSymb}_{\neuralFuncClass}%
  \IfValueT{#1}{\!\left(#1\right)}%
}

\NewDocumentCommand{\classparfeedforwardexprs}{o}{%
  \mathfrak{E}^{\parClassSymb}_{\neuralFuncClass}%
  \IfValueT{#1}{\!\left(#1\right)}%
}

\NewDocumentCommand{\classanalyticparfeedforwardexprs}{o}{%
  \mathfrak{E}^{\parClassSymb,\omega}_{\neuralFuncClass}%
  \IfValueT{#1}{\!\left(#1\right)}%
}

\NewDocumentCommand{\classparfeedforwardmaps}{o}{%
  \classfuncsymb^{\parClassSymb}_{\neuralFuncClass}%
  \IfValueT{#1}{\!\left(#1\right)}%
}

\NewDocumentCommand{\classfeedforwardmaps}{o}{%
  \classfuncsymb_{\neuralFuncClass}%
  \IfValueT{#1}{\!\left(#1\right)}%
}

\NewDocumentCommand{\classanalyticparfeedforwardmaps}{o}{%
  \classfuncsymb^{\parClassSymb,\omega}_{\neuralFuncClass}%
  \IfValueT{#1}{\!\left(#1\right)}%
}

\NewDocumentCommand{\classanalyticfeedforwardmaps}{o}{%
  \classfuncsymb^\analyticClassScriptSymb_{\neuralFuncClass}%
  \IfValueT{#1}{\!\left(#1\right)}%
}

\NewDocumentCommand{\classparfeedforwardgraphs}{o}{%
  \mathcal{G}_{\mathrm{ff}}%
  \IfValueT{#1}{[#1]}%
}

\NewDocumentCommand{\classanalyticparfeedforwardgraphs}{o}{%
  \mathcal{G}^\analyticClassScriptSymb_{\mathrm{ff}}%
  \IfValueT{#1}{[#1]}%
}

\newcommand{\minreductionsymb}{\mathsf{min}}
\newcommand{\maxreductionsymb}{\mathsf{max}}
\newcommand{\minreduction}[1]{\minreductionsymb_{#1}}
\newcommand{\maxreduction}[1]{\maxreductionsymb_{#1}}

\NewDocumentCommand{\constructedgraphvar}{m o o}{%
  \mathsf{#1}%
  \IfValueT{#2}{_{#2}}%
  \IfValueT{#3}{^{#3}}%
}

\NewDocumentCommand{\constructionwitnessrel}{o}{%
  \mathrel{%
    \Vdash%
    \IfValueT{#1}{_{#1}}%
  }%
}

\NewDocumentCommand{\constructionwitnesses}{o m}{%
  \mathsf{CW}%
  \IfValueT{#1}{_{#1}}%
  \!\left(#2\right)%
}

\newcommand{\entries}[1]{\operatorname{entries}\!\left(#1\right)}

\newcommand{\papfullclass}{\mathcal{F}_\mathrm{PAP}}
\newcommand{\papsubclass}{\mathcal{P}_\mathrm{PAP}}

\NewDocumentCommand{\minmaxscanvgraph}{m m m}{%
  \operatorname{scan}_{%
    {\scriptscriptstyle\adaoplustimes},\mathrm{v}%
  }%
  \!\left(
    #1,#2,#3
  \right)%
}

\NewDocumentCommand{\minmaxscangraph}{m m m}{%
  \operatorname{scan}_{%
    {\scriptscriptstyle\adaoplustimes}%
  }%
  \!\left(
    #1,#2,#3
  \right)%
}

\NewDocumentCommand{\minmaxscangraphclass}{m m m}{%
  \minmaxgraphclass^{\mathrm{scan}}%
  \!\left(
    #1,#2,#3
  \right)%
}

\NewDocumentCommand{\minmaxseqvgraph}{m m m}{%
  \operatorname{seq}_{%
    {\scriptscriptstyle\adaoplustimes},\mathrm{v}%
  }%
  \!\left(
    #1,#2,#3
  \right)%
}

\NewDocumentCommand{\minmaxseqgraph}{m m m}{%
  \operatorname{seq}_{%
    {\scriptscriptstyle\adaoplustimes}%
  }%
  \!\left(
    #1,#2,#3
  \right)%
}

\NewDocumentCommand{\minmaxseqvgraphclass}{m m m}{%
  \minmaxgraphclass^{\mathrm{seq},\mathrm{v}}%
  \!\left(
    #1,#2,#3
  \right)%
}

\NewDocumentCommand{\minmaxseqgraphclass}{m m m}{%
  \minmaxgraphclass^{\mathrm{seq}}%
  \!\left(
    #1,#2,#3
  \right)%
}

\newcommand{\dynfuncCompSymb}{%
   \triangleright
}
\newcommand{\analyticextension}{\dynfuncCompSymb}

\newcommand{\outputextensionNostate}[2]{{#1} \dynfuncCompSymb {#2}}
\newcommand{\outputextensionSystem}[2]{\outputextensionNostate{#1}{#2}}
\newcommand{\outputextension}[3]{\outputextensionNostate{\tuple{#1,#2}}{#3}}
\newcommand{\inputprecomposition}[2]{%
  #1 \dynfuncCompSymb{} #2%
}

\NewDocumentCommand{\interior}{o}{%
  \operatorname{int}%
  \IfValueT{#1}{_{#1}}%
}

\makeatletter
\newcommand{\blackparallel}{%
  \mathrel{\mathpalette\blackparallel@\relax}%
}
\newcommand{\blackparallel@}[2]{%
  \begingroup
  \tikzexternaldisable
  \tikz[baseline=(b.base)]{%
    \node[inner sep=0pt,outer sep=0pt] (b)
      {$#1\phantom{\parallel}$};

    \fill
      ($(b.south west)!0.18!(b.south east)$)
      rectangle
      ($(b.north west)!0.36!(b.north east)$);

    \fill
      ($(b.south west)!0.64!(b.south east)$)
      rectangle
      ($(b.north west)!0.82!(b.north east)$);
  }%
  \endgroup
}
\makeatother

\makeatletter
\newcommand{\blackltimes}{%
  \mathbin{\mathpalette\blackltimes@\relax}%
}
\newcommand{\blackltimes@}[2]{%
  \begingroup
  \tikzexternaldisable
  \tikz[baseline=(b.base)]{%
    \node[inner sep=0pt,outer sep=0pt] (b)
      {$#1\phantom{\ltimes}$};

    \coordinate (tl) at
      ($(b.north west)!0.14!(b.north east)$);
    \coordinate (bl) at
      ($(b.south west)!0.14!(b.south east)$);
    \coordinate (lm) at
      ($(b.north west)!0.49!(b.south west)$);
    \coordinate (rm) at
      ($(b.north east)!0.49!(b.south east)$);

    \begin{scope}
      \clip
        (b.south west)
        rectangle
        ($(b.north west)!0.50!(b.north east)$);

      \fill
        ($(tl)!0.04!(bl)$)
        --
        ($(lm)!0.54!(rm)$)
        --
        ($(bl)!0.04!(tl)$)
        -- cycle;
    \end{scope}

    \node[inner sep=0pt,outer sep=0pt] at (b.center)
      {$#1\ltimes$};
  }%
  \endgroup
}
\makeatother

\newcommand{\rv}[1]{\mathcal{#1}}
\newcommand{\probop}{\mathrm{P}}
\newcommand{\tsc}[1]{\textup{\textsc{#1}}}

\newcommand{\tout}{\ensuremath{^\ddagger}}
\newcommand{\cmarko}{~~\cmark \tout}

\newcommand{\notexecuted}{{---}}

\newcommand{\code}[1]{\texttt{\detokenize{#1}}}

\newcommand{\minmaxversion}{\texttt{v0.4.1}}

\NewDocumentCommand{\activevar}{o o o}{%
  \operatorname{activeVariable}%
  \IfValueT{#1}{_{#1}}%
  \IfValueT{#2}{^{#2}}%
  \IfValueT{#3}{({#3})}%
}

\newcommand{\activevarmapsymb}{I}
\NewDocumentCommand{\activevarmap}{m o}{%
  \activevarmapsymb_{#1}%
  \IfValueT{#2}{(#2)}%
}

\newcommand{\selectedvarmapsymb}{\sigma}
\NewDocumentCommand{\selectedvarmap}{o o o}{%
  \selectedvarmapsymb%
  \IfValueT{#1}{_{#1}}%
  \IfValueT{#2}{^{#2}}%
  \IfValueT{#3}{(#3)}%
}

\newcommand{\tiebreakingselectorsymb}{\rho}
\NewDocumentCommand{\tiebreakingselector}{o o}{%
  \tiebreakingselectorsymb%
  \IfValueT{#1}{_{#1}}%
  \IfValueT{#2}{(#2)}%
}
\NewDocumentCommand{\tiebreakingselectorscheme}{o}{%
  \tiebreakingselectorsymb%
  \IfValueT{#1}{[{#1}]}%
}

\newcommand{\functionclassclosure}[1]{\mathrm{cl}(#1)}

\newcommand{\partitionstyle}[1]{\mathcal{#1}}

\newcommand{\partitionselect}[2]{#1[#2]}

\NewDocumentCommand{\partition}{m o o}{%
  \IfValueF{#2}{%
    \partitionstyle{#1}%
  }%
  \IfValueT{#2}{%
    \IfValueF{#3}{%
      \partitionselect{\partitionstyle{#1}}{#2}%
    }%
  }%
  \IfValueT{#3}{%
    \partitionselect{\partitionstyle{#1}_{#3}}{#2}%
  }%
}

\NewDocumentCommand{\genericpartition}{o}{%
  \IfValueF{#1}{\partition{P}}%
  \IfValueT{#1}{\partition{P}[#1]}%
}

\NewDocumentCommand{\activevarpattern}{o}{%
  \alpha%
  \IfValueT{#1}{_{#1}}%
}

\newcommand{\sourcevar}{\mathsf{sourceVariable}}
\newcommand{\activeSource}{\mathsf{activeSourceVariable}}

\newcommand{\sourceValueSymb}{\mathsf{sourceValue}}
\NewDocumentCommand{\sourceValue}{o o o}{%
  \sourceValueSymb%
  \IfValueT{#1}{_{#1}}%
  \IfValueT{#2}{^{#2}}%
  \IfValueT{#3}{({#3})}%
}

\newcommand{\flatminmaxfunc}{\mathsf{mm}}

\newcommand{\sourceCompSymbol}{\diamond}

\NewDocumentCommand{\sourceComp}{m m o o o o}{%
  #1
  \IfValueT{#6}{\ensuremath{#6}}
  \mathrel{%
    \sourceCompSymbol
    \IfValueT{#3}{%
      _{#3%
        \IfValueT{#4}{;#4}%
        \IfValueT{#5}{,#5}%
      }%
    }%
  }
  \IfValueT{#6}{\ensuremath{#6}}
  #2%
}

\NewDocumentCommand{\affinehyperplane}{m m}{%
  H\!\left(#1,#2\right)%
}

\NewDocumentCommand{\coordinateEqualityHyperplane}{m m}{%
  H^{\mathrm{eq}}_{#1,#2}%
}

\NewDocumentCommand{\coordinateThresholdHyperplane}{m m}{%
  H^{\mathrm{th}}_{#1}\!\left(#2\right)%
}

\newcommand{\boundseqsuperscript}[1]{%
  \raisebox{1pt}{%
    \scalebox{0.53}{%
      \ensuremath{%
        \mkern1mu \leq \mkern-4mu #1%
        }%
    }%
  }%
}

\newcommand{\seqsOperatorStyle}[4]{%
  \raisebox{#4}{%
    \scalebox{#2}{%
      \ensuremath{%
        \mkern#3\mathord{#1}%
      }%
    }%
  }%
}

\newcommand{\allseqsOperatorSymbol}{*}
\newcommand{\allseqsOperatorScaleFactor}{0.7}
\newcommand{\allseqsOperatorShiftX}{0mu}
\newcommand{\allseqsOperatorShiftY}{1pt}
\NewDocumentCommand{\allseqs}{m o}{%
  #1^{%
    \seqsOperatorStyle{%
      \allseqsOperatorSymbol%
    }{%
      \allseqsOperatorScaleFactor%
    }{%
      \allseqsOperatorShiftX%
    }{%
      \allseqsOperatorShiftY%
    }
    \IfValueT{#2}{\boundseqsuperscript{#2}}%
  }%
}

\newcommand{\allnonemptyseqsOperatorSymbol}{\boldsymbol{+}}
\newcommand{\allnonemptyseqsOperatorScaleFactor}{0.55}
\newcommand{\allnonemptyseqsOperatorShiftX}{-1mu}
\newcommand{\allnonemptyseqsOperatorShiftY}{1pt}
\newcommand{\allnonemptyseqsOperator}{%
  \seqsOperatorStyle{%
        \allnonemptyseqsOperatorSymbol%
      }{%
        \allnonemptyseqsOperatorScaleFactor%
      }{%
        \allnonemptyseqsOperatorShiftX%
      }{%
        \allnonemptyseqsOperatorShiftY%
      }
}
\NewDocumentCommand{\allnonemptyseqs}{m o}{%
  #1^{%
    \allnonemptyseqsOperator%
    \IfValueT{#2}{\boundseqsuperscript{#2}}%
  }%
}

\makeatletter

\newcommand{\repblacktriangleright}{%
  \mathrel{\mathpalette\repblacktriangleright@\relax}%
}

\newcommand{\repblacktriangleright@}[2]{%
  \begingroup
  \tikzexternaldisable
  \tikz[baseline=(b.base)]{%
    \node[inner sep=0pt,outer sep=0pt] (b)
      {$#1\phantom{\triangleright}$};

    \coordinate (tl) at
      ($(b.north west)!0.14!(b.north east)$);
    \coordinate (bl) at
      ($(b.south west)!0.14!(b.south east)$);
    \coordinate (r) at
      ($(b.west)!0.92!(b.east)$);

    \fill
      (tl)
      --
      (r)
      --
      (bl)
      --
      cycle;
  }%
  \endgroup
}

\makeatother

\newcommand{\presentationSty}[1]{\mathsf{#1}}
\newcommand{\presentationVal}[1]{\llbracket {#1} \rrbracket}

\newcommand{\replayer}{\blackparallel}

\newcommand{\repcascade}{\blackltimes}

\newcommand{\repinputprecomposition}[2]{%
  {#1}\repblacktriangleright{#2}%
}

\newcommand{\repoutoutextension}[3]{%
  \tuple{#1,#2} \repblacktriangleright {#3}%
}

\newcommand{\characteristicFunction}[1]{\mathbf{1}_{#1}}

\usepackage[inline]{enumitem}
\usepackage{calc}   %
\newlist{inlineenum}{enumerate*}{1}
\setlist[inlineenum,1]{label=(\roman*), afterlabel=~}
\newlist{inlineenumc}{enumerate*}{1}
\setlist[inlineenumc,1]{label=\textnormal{(\Roman*)}, afterlabel=~}
\newcommand{\smallbullet}{\raisebox{0.2ex}{\scalebox{0.65}{$\bullet$}}}
\setlist[itemize,1]{
  leftmargin=9pt,
  labelindent=0pt,
  labelsep=.5em,
  label=\smallbullet
}

\setlist[itemize,2]{
  leftmargin=11pt,
  labelindent=0pt,
  labelsep=.5em,
  label=\(\circ\)
}
\setlist[enumerate]{label=(\arabic*), leftmargin=1.7em, labelindent=0pt, labelsep=.4em}

\newlist{compactitemize}{itemize}{1}
\setlist[compactitemize]{
  label={},
  leftmargin=0pt,
  labelsep=0pt,
  itemindent=0pt,
  topsep=.2em,
  partopsep=0pt,
  parsep=0pt,
  itemsep=.2em
}

\newlist{compactenumerate}{enumerate}{1}
\setlist[compactenumerate]{
  label=\arabic*.,
  ref=\arabic*,
  leftmargin=*,
  labelsep=.4em,
  itemindent=0pt,
  topsep=0em,
  partopsep=0pt,
  parsep=0pt,
  itemsep=.1em
}

\newlist{romanenumerate}{enumerate}{1}
\setlist[romanenumerate]{label=(\roman*), leftmargin=1.8em, labelindent=0pt, labelsep=.5em}

\newlist{inlineenuml}{enumerate*}{1}
\setlist[inlineenuml,1]{
  label={},
  afterlabel=\nobreakspace
}

\makeatletter
\newcommand{\nameditem}[1]{%
  \item[\emph{(#1)}]%
  \@ifundefined{phantomsection}{}{\phantomsection}%
  \def\@currentlabel{#1}%
}
\makeatother

\usepackage{etoc}

\title{MinMax Recurrent Neural Cascades}

\author{%
  Alessandro Ronca \\[.6em]
  IRIS-AI, Montefiascone, Italy \\[.2em]
  \texttt{alessandro.ronca@iris-ai.org}
}

\begin{document}

\maketitle

\begin{abstract}

We introduce \emph{MinMax Recurrent Neural Cascades (MinMax RNCs)}, 
a class of recurrent neural networks
based on a novel form of recurrence in the MinMax algebra. 
MinMax RNCs are a first in 
simultaneously enjoying the key properties for truly long-context processing: 
strong formal expressivity with long-term memory, 
efficient inference both in recurrent and parallel form, 
stable dynamics and gradients, 
and 
long-range sensitivity ensured by non-vanishing state gradients. 
First, they can store and retain information arbitrarily long, 
and manipulate it in a rich way,
with a formal expressivity corresponding to the regular languages---%
arguably the maximal expressivity for finite-memory systems.
Second, in addition to linear-time inference via recurrence,
they also admit log-depth--linear-work inference via parallel scan.
Third, their states and activations are uniformly bounded independently of sequence length. 
Fourth, although not everywhere-differentiable, 
they admit pseudo-gradients that are uniformly bounded independently of sequence length and  
coincide with ordinary gradients almost everywhere.
Fifth, they do not suffer from vanishing state gradients:
the pseudo-gradient of a state with respect to an arbitrarily-distant past state can retain unit
norm, while coinciding with the ordinary gradient. 
Empirically, we find that the theoretical properties translate to strong practical performance. 
We evaluate MinMax RNCs on a suite of fundamental synthetic tasks,
and observe they solve all tasks perfectly, 
generalising seamlessly on longer sequences when tested up to length~1M,
and 
outperforming a selection of baselines that well-represent the state of the art. 
We train a 112M-parameter MinMax RNC for next-token prediction in OWT, 
obtaining competitive performance for its size, 
and 
hence initial evidence of the potential of MinMax RNCs for real-world sequence-modelling tasks.

 \end{abstract}

\addtocontents{toc}{\protect\setcounter{tocdepth}{-1}}

\section{Introduction}

Sequence models face a tension between 
computational efficiency,
expressivity, 
stability, 
and 
long-range sensitivity. 
\emph{Transformers}~\citep{vaswanispujgkp17} enable highly parallel computation, but standard full
self-attention has quadratic cost in the sequence length, and it is unable to
detect basic patterns such as interleaving of designated tokens, as it is insensitive to
inputs occurring early on~\citep{chiangcp23}.
\emph{Nonlinear recurrence}
as in~\citep{elman1990,hochreiters97,cho2014gru,beck2024xlstm}
avoids the quadratic cost 
and 
yields an expressivity including all regular languages~\citep{knorozovar24},
but it is inherently sequential 
and 
can suffer from vanishing state gradients~\citep{bengio13vanishing}.

\emph{Linear recurrence} underlies a broad range of recent architectures~%
\citep{behrouz2026titans,benkish2025decimamba,dubinin2026improved,fanetal2024advancing,gu2024mamba,guderr20,dao2024transformers,peng2021random,peng2025rwkv,pengetal2023rwkv,pmlrv119katharopoulos20a,pmlrv202orvieto23a,pmlrv235orvieto24a,polimnfdbber23,qin2024hgrn,schlagis21,siems2025deltaproduct,sun2025learning,walker2026structured,yang2025gated,yu2026blockbiased,zhang2024gated}
and offers a tradeoff:
linear state updates compose associatively 
and therefore admit evaluation via \emph{parallel scan}~\citep{blelloch1990prefix}, 
but they introduce a challenge in ensuring stability.
A common solution is to adopt \emph{contractive parametrisations}---%
as in~%
Mamba~\citep{guderr20}, 
mLSTM~\citep{beck2024xlstm}, 
xLSTM-L~\citep{beck2025xlstm}, 
Gated DeltaNet~\citep{yang2025gated},
and
Gated DeltaProduct~\citep{siems2025deltaproduct}---%
which ensure stability but introduce their own limitations.
Although existing results show that models like Mamba have an expressivity including all star-free
regular languages, such results rely on decoding outputs with a separation margin that vanishes with
sequence length. We introduce a notion of \emph{positive-margin expressivity} that requires outputs
to have a length-independent positive separation margin, 
and show that all contractive Linear RNNs have a positive-margin expressivity within 
the so-called \emph{definite languages}, %
reflecting the fact that sufficiently long suffixes erase the influence of earlier prefixes%
~(see~\S\ref{appendix:linear-rnns},~\S\ref{app:mamba},~\S\ref{app:mLSTM}).
Abandoning contraction avoids the expressivity limitations,
as states can propagate with no attenuation.
However, it generally results into instability:
\emph{expansive parametrisations} as in~RWKV-7~\citep{pengetal2023rwkv} are unstable as they can amplify
states at each update step~(cf.~\citep{siems2025deltaproduct});
and
even \emph{non-expansive parametrisations} as in~DeltaNet~\citep{schlagis21,yang2024parallelizing}
and its generalisation DeltaProduct~\citep{siems2025deltaproduct}
are unstable as their additive inputs can accumulate across state updates yielding
divergent state trajectories,
and 
also linear growth of loss gradients for increasing sequence length%
~(see~\S\ref{app:delta-net}).
What is needed is therefore a mechanism that can propagate states without
attenuation, amplification, or additive accumulation.

\par\smallskip
\emph{MinMax recurrence$\mkern1mu$}~provides such a mechanism.
It is the affine recurrence in the MinMax algebra, 
obtained from ordinary affine recurrence by replacing sum and product $({\smallplus},\mkern-2mu{\smalltimes})$ 
with their MinMax counterparts~$({\oplus},{\otimes})$ that  
use max/min in place of addition/multiplication at the level of scalars,
\begin{equation*}
  \qquad\quad
  x_t 
  \mkern2mu
  = 
  \mkern1mu
  (A_t \otimes x_{t-1}) \oplus b_t 
  \qquad\qquad\;\;
  \text{ for }\;\;\,
  A_t \cin \reals^{d \times d}, 
  \;\;
  b_t, x_t \cin \reals^d
  \mkern2mu
  .
\end{equation*}
MinMax recurrence retains the algebraic properties that enable parallel scan,
while yielding the desired form of state propagation 
since min/max operations simply select coordinates among the ones of the previous
state~$x_{t-1}$ and of the inputs~$\tuple{A_t,b_t}$, 
thus avoiding the above issues. %
Additionally, unlike ordinary affine recurrences, MinMax recurrence is numerically exact in
floating-point arithmetic as its updates involve only min and max, further ensuring
stability of implementations on long sequences.

\paragraph{Overview.}
We present \emph{MinMax Recurrent Neural Cascades (MinMax RNCs)}, 
that adopt a general architecture obtained by cascading layers of \emph{MinMax Neurons},
that employ MinMax recurrence %
with feedforward networks~$\tuple{R,s}$ yielding 
$A_t = R(u_t)$ and $b_t = s(u_t)$ on input~$u_t$. %
We show they enjoy the desired theoretical properties jointly, and we assess them experimentally
observing strong performance in line with the theoretical analyses, hence obtaining an experimental
validation of the faithfulness of such analyses in characterising the practical capabilities of
MinMax RNCs.
\begin{itemize}
  \item \emph{Expressivity~(\S\ref{sec:mainbody-expressivity}).} 
    MinMax Neurons can realise all fundamental semiautomata%
    ~(Theorems~\ref{theorem:expressivity-main-ir},~\ref{theorem:expressivity-main-perm}),
    making MinMax RNCs capable of recognising all regular languages, 
    also when requiring a positive margin on output decoding~%
    (Theorem~\ref{thm:main-body-expressivity}). 
    The state dimension of a MinMax Neuron corresponds to 
    the \emph{permutation degree} of the semiautomata it can realise. 
    Thus, scalar states suffice for all star-free languages,
    and $d$-dimensional states suffice for all languages whose canonical
    automaton admits a cascade decomposition into semiautomata of permutation degree at most~$d$.

  \item \emph{Complexity~(\S\ref{sec:main-body-complexity}).} 
    The computational complexity of MinMax RNCs stems from the core task of evaluating
    MinMax recurrence.
    For~$T$ the sequence length and $d$~the state dimension,
    MinMax recurrence can be evaluated sequentially in time~$\BigO{Td^2}$ 
    (Theorem~\ref{thm:complexity-minmax-recurrence-main-body}), 
    and it admits an associative recurrence-composition operator enabling
    parallel-scan evaluation 
    with depth~$\BigO{\log T \log d}$ and work~$\BigO{Td^3}$, 
    hence runtime~$\BigO{\log T \log d}$ given $\BigTheta{T}$~processors~%
    (Theorem~\ref{thm:complexity-minmax-recurrence-main-body}).

  \item \emph{Stability and gradients~(\S\ref{sec:main-body-stability}).} 
    MinMax RNCs have states, activations, and outputs bounded uniformly on bounded inputs and
    parameters independently of sequence length%
    ~(Theorem~\ref{th:stability-simpler}). 
    Although they are not everywhere-differentiable due to min/max operations,
    we identity a pseudo-derivative operator that is 
    in line with PyTorch's AutoDiff~\citep{pytorchautograd}
    and
    coincides with ordinary differentiation almost everywhere;
    the corresponding state pseudo-Jacobians and loss pseudo-gradients are bounded uniformly on
    bounded inputs and parameters independently of sequence length%
    ~(Theorem~\ref{thm:gradient-bounded-main}).
    Furthermore,
    all MinMax Neurons underlying the expressivity of MinMax RNCs
    admit input sequences over which their state pseudo-Jacobians retain norm one at all
    timesteps while coinciding with the ordinary Jacobian eveywhere%
    ~(Theorem~\ref{thm:gradient-nonvanishing}). 

    On way to see how MinMax recurrence is able to propagate states 
    with no attenuation while ensuring stability 
    is to note that its state updates are \emph{piecewise affine} 
    with a special form~(\S\ref{sec:minmax-piecewise-linear}).
    The state-input space admits a finite partition into regions~$Z$ 
    with affine state update %
    \begin{equation*}
      x_t = A_Z \cdot x_{t-1} + b_Z(u_t)
      \qquad\text{with}\quad
      (A_Z)_{i,:} \mkern-2mu = \mathbf{0}%
      \quad\text{or}\quad 
      (A_Z)_{i,:} \mkern-2mu = \mathbf{e}_\ell, \; (b_Z)_i \mkern-1mu  = 0 
      \mkern2mu,
    \end{equation*}
    for $\mathbf{e}_\ell$ a standard basis row-vector.
    Thus, the state updates are non-expansive since $\|A_Z\| \leq 1$, 
    with $\| A_Z \| = 1$ when $A_Z$ is a permutation matrix, with identity as a special case;
    at the same time, 
    $(b_Z)_i$ being identically zero whenever $(A_Z)_{i,:}$ is nonzero 
    ensures no additive accumulation.

  \item \emph{%
      Empirical evaluation on star-free tasks~(\S\ref{subsec:experiments-synth--star-free}).%
    } 
    We evaluate MinMax RNCs with scalar-state neurons, 
    on a suite of star-free tasks,
    comparing it against the following baselines:
    Mamba~\cite{guderr20}, mLSTM~\cite{beck2024xlstm}, and xLSTM-L~\citep{beck2025xlstm},
    based on contractive linear recurrence;
    DeltaNet~\citep{schlagis21,yang2024parallelizing} 
    based on non-expansive linear recurrence;
    sLSTM~\citep{beck2024xlstm} 
    based on nonlinear recurrence.
    MinMax RNCs achieve perfect validation accuracy on all tasks, 
    retaining the same accuracy when evaluated on longer sequences up to the considered maximum length~1M.
    In contrast, all baselines show an evaluation accuracy that decreases with sequence length,
    dropping to random guess already at length 100--200k.
    MinMax RNCs have competitive throughput compared to the baselines.

  \item \emph{%
      Empirical evaluation on tasks beyond star-free~(\S\ref{subsec:experiments-synth--groups}).%
    } 
    We evaluate MinMax RNCs on the \emph{word problems} of the groups $S_3$ and
    $S_4$. %
    They achieve perfect validation accuracy, 
    and retaining the same accuracy on the considered max
    length~1M,
    in contrast with DeltaProduct whose accuracy begins to 
    deteriorate already at length~200--500~\citep{siems2025deltaproduct}.

  \item \emph{Empirical evaluation on next-token prediction~(\S\ref{sec:experiments-llm}).} 
    We train a MinMax RNC of 112M parameters on next-token prediction in OpenWebText~(OWT),
    achieving a loss comparable to GPT-2, 
    and hence gaining evidence of the potential of MinMax RNCs 
    in real-world applications. 
\end{itemize}
The paper presents the results along with the key supporting arguments. 
The appendices provide
extended background~(\S\ref{app:add-prel}),
auxiliary notions and results%
~(\S\ref{sec:additional-on-dynamics-and-systems}--\S\ref{app:additional-rnc}),
full proofs~(\S\ref{app:expr}--\S\ref{sec:minmax-piecewise-linear}),
details of the implementation~(\S\ref{sec:minmax-implementation})
and of the experiments%
~(\S\ref{app:experiments}--\S\ref{app:details-of-next-token-prediction-on-openwebtext}),
technical analysis of related work~(\S\ref{appendix:linear-rnns}--\S\ref{app:mLSTM}).

\section{Preliminaries}
\label{sec:preliminaries}

For $i,j \in \naturals$ with $i \leq j$,
we define 
$\iival{i}{j} \defeq \{i, i+1, \ldots, j \}$ and $\sival{1}{j} \defeq \iival{1}{j}$.
For $X$ a set, 
$\allseqs{X}$ is the set of all non-empty finite sequences
of elements of $X$,
and
$\allnonemptyseqs{X}$ is $\allseqs{X}$ excluding the empty sequence.
For $X$ a normed space and $r \in \posreals$,
$\truncation{X}{r}$ is the restriction of $X$ to elements having norm at most $r$.

\mypar{MinMax algebra.}
For $a,b \in \reals$, we define the operators %
$a \oplus b \defeq \max(a,b)$ and %
$a \odot b \defeq \min(a,b)$;
they apply component-wise to vectors and matrices;
they are associative, commutative, and $\odot$ distributes over $\oplus$. %
For $A \in \reals^{m \times p},B \in \reals^{p \times n}$,
the operator
$A \otimes B$ yields $C \in \reals^{m \times n}$ with
$C_{i,j} = \bigoplus_{k = 1}^p A_{i,k} \odot B_{k,j}$; 
it is associative and distributes over $\oplus$;
$B \in \reals^n$ is seen as $B \in \reals^{n \times 1}$.

\mypar{Dynamical systems.}
A \emph{dynamical system} is a tuple 
$S = \tuple{ X, U, f, x_0, Y, h }$
where 
$X,U,Y$ are metric spaces called the  
\emph{state}, \emph{input}, and \emph{output space}, respectively;
$f: X \times U \to X$ and $h: X \times U \times X \to Y$ are continuous functions
called the \emph{dynamics} and \emph{output function}, respectively;
and $x_0 \in X$ is the \emph{initial state}.
We refer to $D = \tuple{ X,U,f }$ as \emph{dynamics}, of $S$ in this case.
An \emph{input sequence} is $u = \compactSeqSpec{u}{t}{1}{T} \in \allnonemptyseqs{U}$.
The \emph{state sequence} of $D$ on~$u$ from state~$x_0$ is  
$\compactSeqSpec{x}{t}{1}{T}$ with \emph{step-$t$ state}~$x_t \defeq f(x_{t-1}, u_t)$,
and
the \emph{output sequence} of $S$ is
$\compactSeqSpec{y}{t}{1}{T}$ with \emph{step-$t$ output}~$y_t \defeq h(x_{t-1},u_t,x_t)$,
with associated maps
$\repval{D}(x_0,u) \defeq x_T$,
$\mkern2mu\indexedStateToStateFuncfamily{D}[u][t][x_0] \defeq x_t\mkern1mu$,
and
$\repval{S}(u) \defeq y_T$.

\mypar{Compositions.}
Given systems 
$S_1 = \tuple{X_1,U,f_1, \initialstate{x}_1, Z, h_1}$ and
$S_2 = \tuple{X_2, Z,f_2, \initialstate{x}_2, Y, h_2}$,
their \emph{cascade composition} $S_1 \ltimes S_2$
is the system $\tuple{X ,U,f, \initialstate{x}, Y, h}$ 
having state space $X \defeq X_1 \times X_2$,
initial state $\initialstate{x} \defeq \tuple{\initialstate{x}_1, \initialstate{x}_2}$,
dynamics function 
$f(\tuple{x_1,x_2}, u) \defeq \tuple{x_1', f_2(x_2, h_1(x_1,u,x_1'))}$ for $x_1' = f_1(x_1, u)$,
and
output function $h(\tuple{x_1,x_2}, u, \tuple{x'_1,x'_2}) = h_2(x_2, h_1(x_1,u,x'_1), x'_2)$.
Similarly,
dynamics can be cascaded as $D_1 \ltimes D_2$ by seeing $D_1$ as having an identity input function,
and they can be composed in parallel as $D_1 \parallel D_2$ 
by seeing $D_1$ as having a void output function.

\mypar{Algebraic Automata Theory (AAT), cf.~\citep{arbib,hartmanis}.}
An \emph{automaton} is a system $A = \tuple{Q, \Sigma, \delta, q_0, \Gamma, \theta}$ where
$Q,\Sigma,\Gamma$ are finite spaces under the discrete metric, with $\Sigma$ and $\Gamma$ called
\emph{alphabets} and $\tuple{Q, \Sigma, \delta}$ called a \emph{semiautomaton}.
For $\sigma \in \Sigma$,
the \emph{state transformation} $\delta_\sigma$ of $D$ is 
$\delta_\sigma(q) = \delta(q,\sigma)$.
Semiautomaton $D$ is an \emph{identity-reset semiautomaton} if each transformation
$\delta_\sigma$ is identity or constant, %
and it is a \emph{permutation semiautomaton} if each 
$\delta_\sigma$ is a
bijection.
A semiautomaton $D$ \emph{realises} a semiautomaton $D'$ when there is a way to assign the states
and inputs of $D'$ to the ones of $D$ in a consistent way;
the realisability relation is transitive and closed under cascading.
By the \emph{Krohn-Rhodes theorem}, every semiautomaton $D$ can be realised by a cascade
$D_1 \ltimes \cdots \ltimes D_n$ where each $D_i$ is an identity-reset or permutation semiautomaton
closely connected to $D$. %
A semiautomaton is \emph{group-free} if realised by a cascade of identity-reset semiautomata.
A function $F : \Sigma^+ \to \Gamma$ over alphabets
is \emph{(group-free) regular} if it is defined by some (group-free) automaton.
A \emph{language} is a subset of $\Sigma^+$,
and it is \emph{(star-free) regular} if its characteristic function is (group-free) regular.

\mypar{Metric Automata Theory (MAT)~\citep{mat}.}
To relate a system $S$ to a function $F: \Sigma^+ \to \Gamma$ over alphabets,
one says that $S$ \emph{can-implement} $F$ if there are a continuous encoder $\enc$ and decoder
$\dec$ such that $F(w) = \dec(S(\enc(w)))$ with $\enc$ applied element-wise.
Automata theory can be used to establish whether $S$ can implement $F$ 
whenever $S$ is $\eta$-finite. 
System $S$ is \emph{$\eta$-finite} if so are its input and state space.
For our purposes,
a space $X \subseteq \reals$ is \emph{$\eta$-finite} if it is the union of a finite number of
compact disjoint intervals, called \emph{$\eta$-components}, and every space 
$X_1\times \cdots \times X_n$ is $\eta$-finite if so is each $X_i \subseteq \reals$. 
Function $F$ can be implemented by
an $\eta$-finite system $S$ iff $F$ can be implemented by an automaton whose
semiautomaton is realised by the \emph{$\eta$-canonical semiautomaton} of the dynamics of $S$, 
obtained from the dynamics of $S$ by seeing each input and state $\eta$-component as a single
element.
A system~$S$ \emph{can-recognise} a language language if it can-implement its characteristic function. 

\section{Model} 
\label{sec:model}

\tikzsetnextfilename{layer_diagram}
\begin{figure}[t]
  \centering
\begin{tikzpicture}[
  >=Latex,
  block/.style={
  draw=gray!60,
  fill=gray!6,
  line width=1.2pt,
  rectangle,
  minimum width=14mm,
  minimum height=8mm,
  inner xsep=3mm,
  inner ysep=1.5mm,
  align=center,
  line join=round
},
  mlp/.style={
  draw=gray!60,
  fill=gray!6,
  line width=1.2pt,
  trapezium,
  trapezium stretches=true,
  trapezium left angle=70,
  trapezium right angle=70,
  minimum width=20mm,
  minimum height=7mm,
  inner xsep=3mm,
  inner ysep=1.0mm,
  align=center,
  line join=round
},
  io/.style={
    align=center
  },
  connector/.style={
    ->,
    line width=0.5pt
  }
]

\draw[dashed] (6mm,8.5mm) rectangle (65mm,40mm);
\draw[dashed] (-6mm,8.5mm) rectangle (-65mm,40mm);

\node[io] (u) {$u_t$};

\node[mlp, above=10mm of u, xshift=-47mm] (R) {$R^1(u_t)$};
\node[mlp, above=10mm of u, xshift=-19mm] (S) {$s^1(u_t)$};
\node[mlp, above=10mm of u, xshift=19mm] (R2) {$R^2(u_t)$};
\node[mlp, above=10mm of u, xshift=47mm] (S2) {$s^2(u_t)$};

\coordinate (midRS) at ($(R)!0.5!(S)$);
\coordinate (midRS2) at ($(R2)!0.5!(S2)$);

\node[block, above=12mm of midRS]  (F) {$(R^1_t \otimes x^1_{t-1}) \oplus s^1_t$};
\node[block, above=12mm of midRS2] (F2) {$(R^2_t \otimes x^2_{t-1}) \oplus s^2_t$};

\coordinate (midF) at ($(F)!0.5!(F2)$);

\node[io, left=5mm of F]  (x)  {$x^1_{t-1}$};
\node[io, right=5mm of F2] (x2) {$x^2_{t-1}$};

\node[mlp, above=15mm of midF] (H) {$h\big((x^1_{t-1},x^2_{t-1}),\,u_t,\,(x^1_t,x^2_t)\big)$};

\node[io, above=5mm of H] (y) {$y_t$};

\draw[connector] (u) -- ++(0,6mm) -| (R.south);
\draw[connector] (u) -- ++(0,6mm) -| (S.south);
\draw[connector] (u) -- ++(0,6mm) -| (R2.south);
\draw[connector] (u) -- ++(0,6mm) -| (S2.south);

\draw[connector] (x)  -- (F.west);
\draw[connector] (x2) -- (F2.east);

\draw[connector] (H.north) -- (y);

\draw[connector] (R.north) -- node[left,yshift=2mm]  {$R^1_t$} ++(0,4mm) -| ($(F.south) + (-2mm,0)$);
\draw[connector] (S.north) -- node[right,yshift=2mm] {$s^1_t$} ++(0,4mm) -| ($(F.south) + (2mm,0)$);

\draw[connector] (R2.north) -- node[left,yshift=2mm]  {$R^2_t$} ++(0,4mm) -| ($(F2.south) + (-2mm,0)$);
\draw[connector] (S2.north) -- node[right,yshift=2mm] {$s^2_t$} ++(0,4mm) -| ($(F2.south) + (2mm,0)$);

\draw[connector] (F.north) --  node[left, xshift=-0.0mm,yshift=5mm] {$x^1_t$} ++(0,7.0mm) -| 
  ($(H.south) + (-12mm,0)$);
\draw[connector] (F2.north) -- node[right, xshift=0.5mm,yshift=5mm] {$x^2_t$} ++(0,7.0mm) -| 
  ($(H.south) + (12mm,0)$);

\draw[connector] (u) -- (H.south);

\draw[connector] ($(x.north) + (1mm,-0.5mm)$)   |- ($(H.west) + (0mm,0mm)$);
\draw[connector] ($(x2.north) + (-1mm,0mm)$) |- ($(H.east) + (0mm,0mm)$);

\end{tikzpicture}
\vspace{-1em}
\caption{Diagram of a MinMax Recurrent Layer consisting of two units $D_1,D_2$, where $(R^i,s^i)$ are the
input functions of $D_i$, and $h$ is the output function of the layer. 
  The state $(x^1_t,x^2_t)$ and output $y_t$ at time $t$ are computed from the input $u_t$ at time
  $t$ and the previous state $(x^1_{t-1},x^2_{t-1})$.}
\label{fig:rnc-diagram}
\end{figure}

We define MinMax Recurrent Neurons as a specialisation of MinMax Recurrent Units,
which formalise arbitrary dynamics based on MinMax recurrence.
To favour simplicity, here we characterise Neurons in terms of MLPs~\citep{goodfellow2016deep}, 
but in fact all our results hold under a more general characterisation that covers most feedforward
neural networks---see~\S\ref{subsec:analytic-euclidean-maps} and~\S\ref{subsec:parametric-minmax-rncs}.

\begin{simpleclosingdefinition}
  We define
  a \emph{MinMax Recurrent Unit} with input functions
  $R : U \subseteq \reals^m \to \reals^{n \times n}$ 
  and
  $s : U \subseteq \reals^m \to \reals^n$
  as dynamics $D = \tuple{X, U, f}$ 
  having 
  state space $X \subseteq \reals^n$
  and 
  dynamics function:
  \begin{equation*}
    f(x,u) = \big(R(u) \otimes x\big) \oplus s(u).
  \end{equation*}
  We refer to $R$ as the \emph{reset function}, to~$s$ as the \emph{set function}, and to each of
  them as an \emph{input function}, also writing~$D = \mtuple{X,U,f}{R,s}$.
  Then, unit~$D$ is a \emph{MinMax Recurrent Neuron} if $\tuple{R,s}$ are MLPs.
\end{simpleclosingdefinition}

We present two examples to provide intuition on the mechanics of MinMax Neurons.
\begin{simpleexample} \label{ex:model-1}
  For the MinMax Neuron $D = \mtuple{\reals, \reals^2, f}{R,s}$ 
  having reset function $R(a,b) = a$ and set function $s(a,b) = b$,
  for an initial state $x_0 = 0$ and input sequence $u = (0,2),(7,0),(5,0),(0,1)$,
  if we write $u_{t,1},u_{t,2}$ for the first and second input component at time $t$,
  the state sequence is:
  \begin{align*}
    & 
    x_1 
    = 
    (u_{1,1} \otimes x_0) \oplus u_{1,2} 
    = 
    (0 \odot 0) \oplus 2 
    = 2,
    && 
    x_2 
    = 
    (u_{2,1} \otimes x_1) \oplus u_{2,2} 
    = 
    (7 \odot 2) \oplus 0 
    = 2,
    \\[.2em]
    & 
    x_3 
    = 
    (u_{3,1} \otimes x_2) \oplus u_{3,2}
    = 
    (5 \odot 2) \oplus 0 
    = 
    2,
    && 
    x_4 
    = 
    (u_{4,1} \otimes x_3) \oplus u_{4,2} 
    = 
    (0 \odot 2) \oplus 1 
    = 
    1.
  \end{align*}
  At step $t$,
  the neuron stores $s(u_t)$ if $R(u_t) \leq s(u_t)$,
  and keeps $x_{t-1}$ if $s(u_t) \leq x_{t-1} \leq R(u_t)$.
\end{simpleexample}

\begin{simpleexample} \label{ex:model-2}
  Let $D = \mtuple{\reals^2, \reals, f}{R,s}$ be the MinMax Neuron with 
    $R(u)=
    \begin{psmallmatrix}
      1\smallminus u & u\\
      u & 1\smallminus u
    \end{psmallmatrix}$,
    $s(u)=
    \begin{psmallmatrix}
      0\\0
    \end{psmallmatrix}$.
    Note that $R(0) = \begin{psmallmatrix}0&1\\1&0\end{psmallmatrix}$ is the identity matrix,
  and
$R(1) = \begin{psmallmatrix}1&0\\0&1\end{psmallmatrix}$ the swap matrix.
From $x_0=\begin{psmallmatrix}1\\0\end{psmallmatrix}$, on the input sequence 
  $u = 1,0,1$, the state sequence is
$x_1 = \begin{psmallmatrix}0\\1\end{psmallmatrix}, 
x_2 = \begin{psmallmatrix}0\\1\end{psmallmatrix},
x_3= \begin{psmallmatrix}1\\0\end{psmallmatrix}$, obtained as:
  \begin{align*}
    x_0
    \;\leadsto\;
  \begin{psmallmatrix}0&1\\1&0\end{psmallmatrix}
    \otimes
  \begin{psmallmatrix}1\\0\end{psmallmatrix}
    \oplus
  \begin{psmallmatrix}0\\0\end{psmallmatrix}
    = x_1
    \;\leadsto\;
  \begin{psmallmatrix}1&0\\0&1\end{psmallmatrix}
    \otimes
  \begin{psmallmatrix}0\\1\end{psmallmatrix}
    \oplus
  \begin{psmallmatrix}0\\0\end{psmallmatrix}
    = x_2
    \;\leadsto\;
  \begin{psmallmatrix}0&1\\1&0\end{psmallmatrix}
    \otimes
  \begin{psmallmatrix}0\\1\end{psmallmatrix}
    \oplus
  \begin{psmallmatrix}0\\0\end{psmallmatrix}
    =x_3.
  \end{align*}
  State components are swapped on input $1$ and left unchanged on $0$. Hence $D$ implements parity.
\end{simpleexample}

MinMax RNCs are obtained by cascading layers of MinMax Recurrent Neurons. 
The diagram of a two-unit layer is shown in Figure~\ref{fig:rnc-diagram}, and cascading corresponds
to stacking such diagrams.

\begin{simpleclosingdefinition} \label{def:main--minmax-rnc}
  A \emph{MinMax Recurrent Cascade} is a cascade composition
  $S = S_1 \mkern-2mu \ltimes \mkern-2mu \cdots \mkern-1mu \ltimes \mkern-1mu S_n$
  where each $S_i$ is a \emph{MinMax Recurrent Layer}, defined as a system with dynamics $D_i$
  given by a parallel composition of MinMax Recurrent Units, namely 
  $D_i = D_{i,1} \parallel \cdots \parallel D_{i,n_i}$; 
  each $D_{i,j}$ is a \emph{unit} of $S_i$, and in turn of $S$;
  the \emph{state degree} of $S$ is 
  the maximum state dimension among its units $D_{i,j}$.
  Then $S$ is a \emph{MinMax Recurrent Neural Cascade}, abbreviated as \emph{MinMax RNC}, 
  if
  \begin{inlineenum}
  \item
    each unit $D_{i,j}$ is a MinMax Recurrent Neuron, 
    and
  \item
    the output function of each layer $S_i$ is an MLP.
  \end{inlineenum}
\end{simpleclosingdefinition}

\section{Expressivity}
\label{sec:mainbody-expressivity}

We characterise the expressivity of MinMax RNCs in terms of the functions 
over finite alphabets they can implement, using AAT and MAT as presented
in~\S\ref{sec:preliminaries} and further in \S\ref{app:add-prel}.
We first provide a group-theoretic notion of degree, 
which we then show to provide a fine-grained characterisation of the expressivity of MinMax RNCs 
in terms of their state degree.
\begin{definition} \label{def:mainbody-perm-degree}
  The \emph{(permutation) degree} of a group $G$ is the minimum number $n$
  such that $G$ is isomorphic to a subgroup of the symmetric group $\mathsf{S}_n$.
  The \emph{degree} of a permutation semiautomaton is the degree of its transition group.
  The \emph{degree} of a regular function $F$ is the maximum between one and the 
  degree %
  of any simple group that is a factor of a subgroup of the transition monoid of $F$.
\end{definition}

We state the main theorem regarding the expressivity of MinMax RNCs. 
\begin{restatable}{theorem}{thmmainbodyexpressivity} \label{thm:main-body-expressivity}
  Every regular function of degree~$n$ can be implemented by 
  an \etafinite{} MinMax RNC of state degree~$n$.
  Every~function that is not regular cannot be implemented by a MinMax RNC.
  Every~function that is not group-free cannot be implemented by a MinMax RNC of state degree one.
\end{restatable}
\begin{proofsketch}[qed]
  For the first result,
  implementing a regular function amounts to realising the semiautomaton $A$ of an automaton that
  implements it. 
  Then $A$ can be realised by a cascade of identity-reset and permutation semiautomata of degree
  bounded by the degree of the function; in particular, they are all identity-reset in the
  group-free case.
  The key step is to show \etafinite{} MinMax Neurons that can realise such semiautomata, as
  we do in the next section (Theorems~\ref{theorem:expressivity-main-ir}
  and~\ref{theorem:expressivity-main-perm}).
  Such Neurons can be composed into a cascade that can realise $A$ and hence implement the
  given regular function.
  The second result follows from the fact that, for every 
  MinMax RNC implementing a function over a finite alphabet, there is an equivalent one having a
  finite state space, implying that the function is regular.
  For the third result,
  we observe that, for any fixed scalar values $r,s$ and state $x_0$, the recurrence 
  $x_t = (r \otimes x_{t-1}) \oplus s$ yields a state sequence that converges to a constant in two
  steps, meaning that MinMax Units of state degree one cannot realise permutation semiautomata, 
  which prevents them, and cascades thereof, from implementing any function that is not group-free.
\end{proofsketch}

We present a novel, stricter notion of expressivity tha excludes those cases where outputs
of a system can only be decoded with a separation margin that vanishes as sequence length increases.
\begin{definition} 
  A decoder $\dec \mkern-3mu:\mkern-2mu Y \mkern-5mu\to\mkern-2mu \Gamma$
  \emph{admits a margin $\gamma \mkern-3mu\in\mkern-2mu \posreals$ on $Y' \mkern-3mu\subseteq\mkern-2mu Y$}
  if 
  $d_Y(y',y) \mkern-1mu < \mkern-1mu \gamma$
  implies
  $\dec(y') \mkern-1mu = \mkern-1mu \dec(y)$ 
  for all $y' \mkern-3mu\cin Y'\mkern-1mu, \mkern2mu y \cin Y$.
  A system~$S$ 
  \emph{%
    can-implement a function 
    $F : \allnonemptyseqs{\Sigma} \mkern-3mu \to \Gamma$ 
    with positive margin%
  }
  if it can-implement $F$ 
  with 
  some~$\enc$ and 
  some~$\dec$ decoder admitting a margin $\gamma \in \posreals$ 
  on 
  the set~%
  $Y_S \defeq \{ S(\enc(w)) : w \in \allnonemptyseqs{\Sigma} \}$;
  and it \emph{can-recognise a language~$L \subseteq \allnonemptyseqs{\Sigma}$ with positive margin}
  if it can-implement the characteristic function~$\mathbf{1}_L$ with positive margin.
\end{definition} 

For \etafinite{} systems, positive-margin implementation coincides with the ordinary
implementation---see~\S\ref{subsec:aux--expressivity}.
Thus, we obtain the following exact characterisation of the expressitivity of MinMax RNCs and 
of their subclass with state degree one, 
also under $\eta$-finiteness and positive~margin requirements.
\begin{theorem}
  \label{thm:main-body--expressivity--overall}
  The functions that can be implemented by MinMax RNCs are the regular functions,
  and 
  the group-free functions for MinMax RNCs of state degree one;
  the languages that can be recognised by MinMax RNCs are the regular languages,
  and the star-free languages for MinMax RNCs of state degree one.
  The two claims hold the same when requiring $\eta$-finiteness or a positive margin.
\end{theorem}

\subsection{Capabilities of MinMax Recurrent Neurons to Realise Fundamental Semiautomata}
\label{sec:mainbody-semiautomata}

We show the core abilities underlying the expressivity of MinMax RNCs.
First we present Theorem~\ref{theorem:expressivity-main-ir}, focusing on realising
\emph{identity-reset}
semiautomata---Example~\ref{ex:model-1} corresponds to this case.
Intuitively, it requires storing information and keeping it unchanged as long as needed.
More formally, it requires dynamics to have a state transformation $f(\cdot, u)$ acting as an identity,
and a set of state transformations $\{f(\cdot,u_i)\}_i$ acting as constant functions.
For MinMax Neurons, when $s \leq r$ the transformation
$f_{r,s}(x) = (r \odot x) \oplus s$ is identity for all $x \in [s,r]$,
and it is constant $f_{r,s}(x) = s$ when $s > r$.

\begin{theorem}\label{theorem:expressivity-main-ir}
  Every identity-reset semiautomaton can be realised by an \etafinite{}
  MinMax Recurrent Neuron having state dimension one.
\end{theorem}
\begin{proofsketch}[qed]
  Every identity-reset semiautomaton
  $A = \tuple{Q, \Sigma, \delta}$,
  with states $Q = \{ q_i \}_{i=1}^n$ and inputs $\Sigma = \{ \sigma_j \}_{j=1}^m$,
  can be realised by a MinMax Unit~$D = \mtuple{X, U, f}{R, s}$
  with
  states $X = \bigcup_{i=1}^n X_i$ 
  for pairwise-disjoint intervals 
  $X_i = [x_i \,\msmallminus\, \epsilon, x_i \,\msmallplus\, \epsilon]$,
  scalar inputs $U = \{ u_j \}_{j=1}^m$,
  and input functions
  \begin{align*}
    R(u_j) \defeq 
    \begin{cases}
      \newxmax & \text{if } \delta_{\sigma_j} \mkern-3mu = \mkern-1mu \id_Q,
      \\
      \newxmin & \text{if } \delta_{\sigma_j} \mkern-3mu = \mkern-1mu q_\ell,
    \end{cases}
    \qquad
    s(u_j) \defeq 
    \begin{cases}
      \newxmin & \text{if } \delta_{\sigma_j} \mkern-3mu = \mkern-1mu \id_Q,
      \\
      x_\ell   & \text{if } \delta_{\sigma_j} \mkern-3mu = \mkern-1mu q_\ell,
    \end{cases}
    \qquad\text{for }\quad
    \begin{aligned}
      \newxmin \mkern-2mu & < \mkern-1mu x_\mathrm{min} \defeq \mkern-2mu \min X,
    \\
      \newxmax \mkern-2mu & < \mkern-1mu x_\mathrm{max} \mkern-3mu\defeq \mkern-2mu\max X.
    \end{aligned}
  \end{align*}
  Unit~$D$ can realise $A$ 
  as it can represent
  each state-input pair $\tuple{q_i,\sigma_j}$ by the pair of $\eta$-state and
  $\eta$-input~$\tuple{X_i,u_j}$, 
  over which its state transformation $f(\cdot\mkern1mu,u_j)$ 
  is the same as $\delta_{\sigma_j}$.
  In fact,
  \begin{align*}
    f(x,u_j) =
    \begin{cases}
      (\newxmax \otimes x) \oplus \newxmin \mkern2mu = \mkern2mu x  
      & \text{if } \delta_{\sigma_j} \mkern-3mu = \mkern-1mu \id_Q,
      \\
      (\newxmin \otimes x) \oplus x_\ell  \mkern2mu = \mkern2mu x_\ell   
      & \text{if } \delta_{\sigma_j} \mkern-3mu = \mkern-1mu q_\ell,
    \end{cases}
    \quad
    \leadsto
    \quad
    f(\cdot\mkern1mu,u_j) 
    =
    \begin{cases}
      \id_X & \text{if } \delta_{\sigma_j} \mkern-3mu = \mkern-1mu \id_Q,
      \\
      x_\ell   & \text{if } \delta_{\sigma_j} \mkern-3mu = \mkern-1mu q_\ell.
    \end{cases}
  \end{align*}
  The same holds as long as the outputs $\newxmin$, $\newxmax$, and $\{x_\ell\}_{\ell=1}^n$ 
  of $\tuple{R,s}$ do not change by more 
  than~$\hat\epsilon \defeq \min\{ \epsilon,\, x_\mathrm{min} \mkern-1mu\,\msmallminus\,\newxmin,\,
  \newxmax\mkern-1mu\,\msmallminus\,x_\mathrm{min} \}$,
  and hence the Unit~$D$ can be turned into a MinMax Neuron~$\hat D$ 
  by replacing~$\tuple{R,s}$
  with $\hat{\epsilon}$-approximating MLPs~$\tuple{\hat R, \hat s}$,
  that exist since $U$ is compact.
\end{proofsketch}

Next we present Theorem~\ref{theorem:expressivity-main-perm}, that focuses on realising any 
permutation semiautomaton~$A$---such as the one in Example~\ref{ex:model-2}.
It requires dynamics to admit state transformations $\{f(\cdot, u_i)\}_i$ acting as bijections. 
This could be shown directly using a state dimension equal to the number of states of~$A$---we show
it in the appendix. Here we adopt an intermediate characterisation of permutation semiautomata,
to show a stronger result connecting the state dimension of a MinMax Neuron to the degree $n$
of~$A$---always bounded by its number of states, but as large as $n!$ in some cases.
Semiautomaton~$A$ can be realised by a semiautomaton~$A_\mathrm{S}$
where every state is an arrangement $\mathbf{i} = \tuple{i_1, \ldots, i_n}$ of the indices 
$\sival{1}{n}$, and each transformation $\delta_\sigma(\mathbf{i})$ rearranges $\mathbf{i}$ into
$\tuple{i_{\pi(1)}, \ldots, i_{\pi(n)}}$ for $\pi$ a bijection determined by $\sigma$.
Thus, we can realise $A_\mathrm{S}$ in place of $A$, and we can do so by representing directly its
$n$-dimensional states using vectors of $\reals^n$.
In a MinMax recurrence, the multiplication $x' = R \otimes x$ allows for assigning any coordinate of 
$x$ to any of $x'$. For instance, in~$x'_\ell = \bigoplus_{k=1}^n (R_{\ell,k} \odot x_k)$, 
we can obtain $x'_\ell = x_{k^\star}$ by setting $R_{\ell,k^\star}$ to a high value and 
every other $R_{\ell,k}$ to a low value.

\begin{theorem} \label{theorem:expressivity-main-perm}
  Every permutation semiautomaton $A$ can be realised by an $\eta$-finite MinMax
  Recurrent Neuron having state dimension equal to the degree of $A$.
\end{theorem}
\begin{proofsketch}[qed]
  Every semiautomaton $A$ having degree~$n$ and inputs 
  $\Sigma = \{ \sigma_j \}_{j=1}^m$ %
  is realised by a semiautomaton $A_\mathrm{S} = \tuple{\sival{1}{n}^n, \Sigma, \delta}$
  with
  $\delta(\tuple{i_1, \ldots, i_n}, \sigma_j) = \tuple{\pi_j(i_1), \ldots, \pi_j(i_n)}$
  where $\pi_j: \sival{1}{n} \to \sival{1}{n}$ is a bijection determined by $\sigma_j\mkern1mu$.
  Semiautomaton~$A_\mathrm{S}$, and hence $A$,
  can be realised by
  the MinMax Unit $D = \mtuple{X^n, U, f}{R, s}$
  having state components
  $X = \bigcup_{i=1}^n X_i$ 
  for pairwise-disjoint intervals $X_i = [x_i \,\msmallminus\, \epsilon, x_i \,\msmallplus\, \epsilon]$,
  inputs $U = \{ u_j \}_{j=1}^m$,
  and 
  input functions given component-wise by
  \begin{align*}
    \quad
    R(u_j)_{\ell,k} 
    \defeq
    \begin{cases}
      \newxmax & \text{if } \pi_j(\ell) \mkern-2mu =\mkern-2mu k,
      \\
      \newxmin & \text{otherwise},
    \end{cases}
    \qquad\quad
    s(u_j)_\ell \defeq \newxmin \mkern2mu,
    \qquad\quad\;\text{ for }\quad
    \begin{aligned}
      \newxmin \mkern-2mu & < \mkern-1mu x_\mathrm{min} \defeq \mkern-2mu \min X,
    \\
      \newxmax \mkern-2mu & < \mkern-1mu x_\mathrm{max} \mkern-3mu\defeq \mkern-2mu\max X.
    \end{aligned}
  \end{align*}
  Unit~$D$ can realise $A$ 
  as it can represent
  each state-input pair $\tuple{\mathbf{i},\sigma_j}$ 
  with $\mathbf{i} = \tuple{i_1, \ldots, i_n}$ 
  by the $\eta$-pair~$\tuple{X_{\mathbf{i}},u_j}$ 
  with $X_{\mathbf{i}} = X_{i_1} \mkern-2mu \ctimes \cdots \ctimes X_{i_n}$,
  over which its state transformation $f(\cdot\mkern1mu,u_j)$ permutes coordinates in the same
  way as~$\delta_{\sigma_j}$.
  In fact, each transformation $f(\cdot\mkern1mu,u_j)$ amounts to the linear operator given by the
  permutation matrix~$P_{\pi_j}$ representing $\pi_j$, with 
  entries~$(P_{\pi_j})_{\ell,k} = \ind{\pi_j(\ell) \mkern-2mu =\mkern-2mu k}$.
  \begin{align} \label{eq:theorem:expressivity-main-perm}
    f(x,u_j)_\ell
    \mkern-1mu= \mkern-2mu
    \big(\newxmax\odot x_{\pi_j(\ell)}\big)
    \oplus
    \big(\!\textstyle\bigoplus_{k\neq\pi_j(\ell)}
    \newxmin\odot x_k\big)
    \oplus \newxmin
    \mkern-2mu \,=\,
    x_{\pi_j(\ell)}
    \quad\leadsto\quad
    f(\cdot\mkern1mu,u_j)
    \mkern1mu = \mkern1mu P_{\pi_j} \mkern-3mu \cdot x 
  \end{align}
  The same holds as long as the output components~$\newxmin$ and~$\newxmax$
  of~$\tuple{R,s}$ do not change by more 
  than~$\hat\epsilon \defeq \min\{ \epsilon,\, x_\mathrm{min} \mkern-1mu\,\msmallminus\,\newxmin,\,
  \newxmax\mkern-1mu\,\msmallminus\,x_\mathrm{min} \}$,
  and hence the Unit~$D$ can be turned into a MinMax Neuron~$\hat D$ 
  by replacing~$\tuple{R,s}$
  with $\hat{\epsilon}$-approximating MLPs~$\tuple{\hat R, \hat s}$,
  that exist since $U$ is compact.
\end{proofsketch}

\section{Complexity}
\label{sec:main-body-complexity}

We analyse the sequential and parallel complexity of evaluating MinMax RNCs.
First we analyse the complexity of evaluating MinMax recurrence for $T$ steps, which amounts to
computing all states $x_t = (R_t \otimes x_{t-1}) \oplus s_t$ given
$x_0 \in \reals^{\dState}$,
$R_t \in \reals^{{\dState} \times {\dState}}$, and
$s_t \in \reals^{\dState}$ for all $t \in \iival{1}{T}$.

\begin{theorem} \label{thm:complexity-minmax-recurrence-main-body}
  MinMax recurrence can be evaluated sequentially in time $\BigO{T \cdot  \dState^2}$,
  and also in parallel 
  with depth~$\BigO{\log T \cdot  \log \dState}$ and work~$\BigO{T \cdot \dState^3}$.
\end{theorem}
\begin{proof}[Proof sketch]
  The sequential algorithm that computes states by evaluating
  $x_t = (R_t \otimes x_{t-1}) \oplus s_t$ for $t$ from $1$ to $T$ runs in the specified time
  since the most expensive operation at each step is the matrix-vector multiplication
  $R_t \otimes x_{t-1}$ which can be done in $\BigO{\dState^2}$.
  To obtain a parallel algorithm, we observe that letting
  $\tau_t(x) \defeq (R_t \otimes x) \oplus s_t$, we have $x_t = (\tau_t \circ \cdots \circ \tau_1)(x_0)$,
  and each composition $\tau_t \circ \cdots \circ \tau_1$ can be computed via \emph{parallel scan},
  cf.~\cite{parallelscan,blelloch1990prefix,kogge2009parallel}. Specifically, parallel scan
  evaluates all prefixes of an expression over any given monoid---i.e., expressions where several
  elements are combined via an associative operator, for which some element acts as identity.
  In our case, the relevant monoid is the finite set of matrix-vector pairs~$\tuple{R,s}$ over the
  scalars of $x_0, R_t, s_t$, equipped with the associative operator
  $\tuple{R,s} \oplustimes \tuple{R',s'} \defeq \tuple{R' \otimes R,\, (R' \otimes s) \oplus s'}$.
  Then
  we have $(\tau_t \circ \cdots \circ \tau_1)(x) = (R_{1:t} \otimes x) \oplus s_{1:t}$
  with $\tuple{R_{1:t},s_{1:t}} = \tuple{R_1, s_1} \oplustimes \cdots \oplustimes \tuple{R_t, s_t}$
  that can be computed via parallel scan.
  The complexity bound holds
  since
  parallel scan performs $\BigO{\log T}$ rounds of depth $\BigO{\log \dState}$,
  and it carries out $\BigO{T}$ compositions each of work $\BigO{\dState^3}$.
\end{proof}

The above results for MinMax recurrence make it straightforward to derive algorithms and bounds 
for of \emph{evaluating MinMax RNCs}, that is to compute the output sequence of
a MinMax RNC $S$ over an input $u$. 
We provide bounds in terms of 
the input-sequence length~$T$,
number~$\nLayers$ of layers,
state degree$~\dState$ ,
and
maximum dimension~$d$ across the input space, output space, hidden layers of its MLPs,
and state space of any layer.
The bounds consider the number of hidden MLP layers as fixed, 
in order to be more intelligible---general bounds in~\S\ref{appendix:proofs-complexity}.

\begin{theorem} \label{thm:complexity-main-sequential} \label{thm:complexity-main-parallel}
  MinMax RNCs
  can be evaluated sequentially in time
  $\bigBigO{T \mkern-2mu \cdot \nLayers \mkern-2mu \cdot d \cdot \mkern-1mu ( d + \dState^2 )}$,
  and also in parallel with 
  work~$\bigBigO{T  \mkern-1mu \cdot \nLayers \mkern-2mu \cdot d \cdot\mkern-1mu ( d + \dState^2 )}$
  and 
  depth~$\bigBigO{\nLayers\mkern-3mu \cdot \mkern-1mu (\log T \cdot \log \dState \mkern-1mu+ \log d )}$.
\end{theorem}

The following bound highlights the dependency on the sequence length~$T$, 
obtained immediately from the above since work and depth translate to runtime 
via Brent's Scheduling Principle, cf.~\citep{blellochmaggs10}.
\begin{corollary} \label{corollary:complexity-main-2}
  MinMax RNCs of fixed size can be evaluated in time $\BigO{\log T}$ over $\BigTheta{T}$ processors.
\end{corollary}

\section{Stability and Gradients}
\label{sec:main-body-stability}
\label{sec:main-body-gradient}

We study stability and gradient properties of parametric MinMax RNCs where parameters are the ones
of their MLPs---see~\S\ref{subsec:parametric-minmax-rncs}. 
Parametric MinMax RNCs have states and outputs that are bounded uniformly over bounded input and
parameter values, independently of input-sequence length.
\begin{restatable}{theorem}{thstabilitysimpler}
  \label{th:stability-simpler}
  Let $S$ be a parametric MinMax RNC with $U$ and $\Theta$ its input and parameter spaces.
  For every
  $r\cin\nonnegreals$,
  there is
  $B_r \cin\reals$
  such that, 
  for every~$\theta \in \truncation{\Theta}{r}$,
  on every input sequence 
  $\compactSeqSpec{u}{t}{1}{T} \in \allnonemptyseqs{\truncation{U}{r}}\mkern2mu$,
  the instance~$\pinstLight{S}{\theta}$ 
  has 
  states~$\compactSeqSpec{x}{t}{1}{T}$ 
  and outputs~$\compactSeqSpec{y}{t}{1}{T}$ 
  satisfying
  $\|x_t\|,\|y_t\| \leq B_r$
  for all $t \in \ival{T}$.
\end{restatable}
\begin{proofsketch}[qed]
  The key observation is that each MinMax recurrence step
  $x_t = (R_t\otimes x_{t-1}) \oplus s_t$
  satisfies the one-step max-norm bound 
  $\|x_t\|_\mathrm{max} \leq \max\left\{ \|R_t\|_\mathrm{max}, \|s_t\|_\mathrm{max} \right\}$,
  which is independent of $\|x_{t-1}\|_\mathrm{max}$.
  The reset and set values are produced by MLPs, which are bounded on bounded inputs.
  Boundedness in Euclidean or max norm implies boundedness in the other.
  Thus,
  we obtain
  a bound on the states~$x_t$ that is uniform in~$t$
  by
  applying the above one-step bound to each recurrent unit 
  and 
  proceeding inductively through the layers of the RNC, 
  with base case given by the input sequence $u = \compactSeqSpec{u}{t}{1}{T}$ 
  satisfying $\max_t \|u_t\| \leq r$,
  Then,
  a bound on the outputs~$y_t$ follows from boundedness of the MLPs
  on the resulting bounded states, inputs, and parameters.
\end{proofsketch}

To study gradient properties,
we note that ordinary derivatives may not exist everywhere due to the min and max operations in the
recurrence.
Such operations consist of \emph{MinMax reductions}~$\min_n \text{/} \max_n$ that map 
scalars $x_1, \ldots, x_n$ to the corresponding minimum$\mkern2mu\text{/}\mkern2mu$maximum
value~$x_\star$.
We extend the ordinary derivative operator to
the \emph{pseudo-derivative operator~$\minmaxsdv$}, 
that assigns~$\frac{1}{k}$ %
to each the $k$~\emph{active variables~$x_i = x_\star$} of a MinMax reduction
and zero to the others---thus mapping $\min_n \text{/} \max_n$ to 
its \emph{minimum-norm sub$\mkern2mu\text{/}\mkern2mu$supergradient},
cf.~\citep{rockafellar1997convex}.
Then, the pseudo-derivative operator~$\minmaxsdv$ is extended to the other relevant functions 
via the \emph{chain rule}.
However, different chain-rule derivations may yield different pseudo-derivatives.
We formalise this dependency by parametrising the operator~$\minmaxsdv$ with a choice of a 
\emph{selector~$\selgraphsymb$} over \emph{computation graphs}, 
obtaining what we call a \emph{selected pseudo-derivative operator}~$\minmaxsdv^{\selgraphsymb}$,
that applies the chain rule according to 
the graph selected by~$\selgraphsymb$, 
and applies the operator~$\minmaxsdv$ locally at the level of nodes.
Computation graphs are taken from a class~$\minmaxgraphclass$ we identify, 
that includes the graphs arising from a broad variety of evaluation strategies, and in particular
from sequential and parallel-scan evaluation---thus covering the algorithms underlying our
complexity results.
Notably, our operator~$\minmaxsdv^{\selgraphsymb}$ is in line with 
of \emph{Pytorch’s~AutoDiff}~\citep{pytorchautograd} when applied to MinMax RNCs,
and hence we may expect our results to provide a reasonably-faithful characterisation of the
empirical behaviour of MinMax RNCs implemented in PyTorch.

We show next that the pseudo-derivatives of loss functions of MinMax RNCs are bounded uniformly on
bounded inputs and parameters, and they coincide with the ordinary
derivative almost everywhere. 
For~$S$ a parametric MinMax RNC with $U$ and $Y$ its input and output spaces,
its \emph{sample space}~$\samplespace \defeq \allnonemptyseqs{U}\ctimes Y$
its \emph{$r$-bounded sample space}~%
$\truncsamplespace{r} 
\defeq 
\truncation{\allnonemptyseqs{U}}{r} \ctimes \truncation{\targetspace{Y}}{r}$,
and 
the \emph{loss function}~%
$\Lossfn{z}[\lossfn](\theta) \defeq \lossfn (\fpinst{\repval{S}}{u}{\theta}, y)$
for a given sample~$z = \ptuple{u,y} \in \samplespace$,
under a choice of $\lossfn : Y^2 \to \reals$.

\begin{restatable}[Boundedness and correctness of loss pseudo-gradients]{theorem}{thmgradientboundedmain}
  \label{thm:gradient-bounded-main}
  Let $S$ be a MinMax RNC with parameter space~$\Theta$, 
  let $\lossfn:Y^2 \mkern-2mu\to \mkern-2mu\reals$ be analytic,
  and 
  let $\selgraphsymb$ be a graph selector over~$\minmaxgraphclass$. 
  \emph{(Boundedness):}~%
  for every $r \cin \nonnegreals$,
  there is $B_r\cin \reals$ such that
  $\|\minmaxsdv^{\selgraphsymb}\mkern-1mu \Lossfn{z}[\lossfn](\theta)\| 
  \mkern-2mu \leq \mkern-2mu 
  B_r$
  for all
  $z \cin \truncsamplespace{r}\mkern1mu,\mkern1mu\theta \cin \truncation{\Theta}{r}$.
  \emph{(Correctness):}~%
  for every $z \cin \samplespace$,
  the equality~$\minmaxsdv^{\selgraphsymb}\mkern-1mu \Lossfn{z}[\lossfn](\theta)
  = \dv\Lossfn{z}[\lossfn](\theta)$
  holds for almost every $\theta \cin \Theta$.
\end{restatable}
\begin{proofsketch}[qed]
  \emph{(Boundedness):}~%
  the key observation is that
  MinMax recurrence propagates states exclusively via min/max reductions,
  whose sub/supergradients have non-negative entries summing to one,
  and hence amount to non-expansive linear operators 
  that do not contribute in increasing derivative norms.
  Thus, derivative norms do not increase across time and can only increase across layers, 
  where growth is determined by derivatives of MLPs, that are bounded on bounded inputs as they are
  continuously-differentiable, and boundedness of inputs holds by Theorem~\ref{th:stability-simpler}.
  \emph{(Correctness):}~%
  the pseudo-derivative $\minmaxsdv^{\selgraphsymb}\mkern-1mu \Lossfn{z}[\lossfn]$
  is an \emph{intensional derivative}~\citep{lee2020correctness},
  guaranteed to coincide with the ordinary derivative $\dv\Lossfn{z}[\lossfn]$ 
  almost everyhere.
  It is an intensional derivative since so are 
  minimum-norm sub/supergradients of min/max reductions
  and
  the ordinary derivatives of MLPs,
  and consequently the pseudo-derivatives 
  obtained by $\minmaxsdv^{\selgraphsymb}$ 
  via chain-rule derivations.
\end{proofsketch}

We next show that pseudo-derivatives of MinMax RNCs need not vanish over arbitrarily long horizons,
while coinciding with ordinary derivatives.
Notably, this holds for the MinMax Neurons underlying our expressivity results,
since they have exact identity and permutation state transformations, 
which correspond to linear isometries.
\begin{restatable}[Non-vanishing state-to-state derivatives]{theorem}{thmgradientnonvanishing}
  \label{thm:gradient-nonvanishing}
  For every $\epsilon \in \posreals$,
  there exists $(\mathcal{D}_n)_{n \geq 1}$ satisfying the following jointly:
  \begin{inlineenum}
  \item
    each $D \in \mathcal{D}_n$ is a MinMax Neuron whose state~space contains an $n$-dimension
    hypercube~$\tilde{X}_D$ of side length~$\epsilon$;
  \item
    every identity-reset semiautomaton can be realised by a Neuron of~$\mathcal{D}_1$;
  \item
    every $n$-degree permutation semiautomaton can be realised by a Neuron of~$\mathcal{D}_n$;
  \item
    each $D \in \mathcal{D}_n$ 
    admits an infinite input squence $u$ such that, for every selector~$\selgraphsymb$
    over~$\minmaxgraphclass$,
  \end{inlineenum}
  \begin{align*}
    \minmaxsdvop[\selgraphsymb]{\indexedStateToStateFuncfamily{D}[u][t][x_0]}
    = \dvop{\indexedStateToStateFuncfamily{D}[u][t][x_0]}
    \quad\text{and}\quad
    \|\minmaxsdvop[\selgraphsymb]{\indexedStateToStateFuncfamily{D}[u][t][x_0]} \| = 1
    \qquad\quad\forall 
    x_0 \in \tilde{X}_D \quad \forall t \in \posnaturals \mkern2mu .
  \end{align*}
\end{restatable}
\begin{proofsketch}[qed]
  Neurons $D \in \mathcal{D}_1$ that can realise identity-reset semiautomata
  are from Theorem~\ref{theorem:expressivity-main-ir}.
  We choose $u = u_j \cdots u_j$ 
  for $u_j$ inducing an identity state transformation $f(\cdot\mkern1mu,u_j)$,
  and so have that 
  $\indexedStateToStateFuncfamily{D}[u][t]$ is identity for every $t$,
  with derivative
  $\dvop{\indexedStateToStateFuncfamily{D}[u][t][x_0]} = 1$.
  Neurons $D \in \mathcal{D}_n$ that can realise $n$-degree permutation semiautomata
  are from Theorem~\ref{theorem:expressivity-main-perm}.
  We choose~$\tilde{X}_D = X_1 \ctimes \cdots \ctimes X_n$ with $X_i$ pairwise-disjoint 
  intervals,
  and so have that every $f(\cdot\mkern1mu,u_t)$
  is a coordinate permutation represented by a permutation matrix~$P_t$ as
  per~\eqref{eq:theorem:expressivity-main-perm}.
  Then also $\dvop{\indexedStateToStateFuncfamily{D}[u][t]}$ is a coordinate permutation,
  represented by the permutation matrix $P_t \cdots P_1$, that has unit norm
  and gives to the derivative $\dvop{\indexedStateToStateFuncfamily{D}[u][t]} = P_t \cdots P_1$.
  In both cases, 
  the derivative coincides with~$\minmaxsdvop[\selgraphsymb]{\indexedStateToStateFuncfamily{D}[u][t][x_0]}$
  since the constructions ensure that no ties occur among the arguments of min/max operators
  and, in the permutation case, 
  each coordinate of~$\indexedStateToStateFuncfamily{D}[u][t][x_0]$
  is a unique coordinate of $x_0$ by the choice of $\tilde{X}_D$.
\end{proofsketch}

\section{Experimental Evaluation}
\label{sec:main-body--experiments}

We evaluate MinMax RNCs~\citep{minmaxgithub} implemented in PyTorch~\citep{paszke2019pytorch} run
on a single A100~GPU---see~\S\ref{sec:minmax-implementation},~\ref{sec:experimental_setup}.
An overview of the evaluation is given by Figure~\ref{fig:results-charts}.

\subsection{Evaluation on group-free tasks}
\label{subsec:experiments-synth--star-free}

Our first goal is to assess the capabilities of MinMax RNCs of degree one to learn group-free tasks.

\textbf{Benchmarks.}
We consider three families of group-free tasks that test long-range capabilities.
\dlatching{n} is based on~\citep{bengio1994learning} and consists in returning the first element of the input
sequence, generated from a set of $20n$ tokens, with $n$ that can appear first.
It focuses on storing and retaining information over long horizons, with $n$ increasing the amount
of information.
\dsequences{n} is novel and consists in simultaneously matching two patterns~$E_1$ and~$E_2$
consisting of~$n$ events to be matched in sequence, 
with the $j$-th event of~$E_i$ being a token from a given set~$E_{i,j}$, 
and restarting to match~$E_i$ when encountering a token from a given set~$I_i$.
It focuses on the ability to solve inherently sequential tasks, requiring to combine $n$ pieces of
information scattered across the input sequence, and hence to maintain intermediate information over
long horizons.
\dheads{n} is from \citep{olsson2022context,gu2024mamba} and consists in recalling, at the second
occurrence of a designated marker, the token that has appeared right after the first occurrence of
the marker; with $n$ possible tokens to recall, the marker's first occurrence within the first $30$
steps for training, and $50$ for validation and evaluation.
It focuses on the ability to store, retain, and recall information over long horizons,
with $n$ increasing the amount of information.

\textbf{Baselines.}
We include a representative selection of SOTA baselines 
covering different forms of recurrence:
Mamba~\cite{guderr20}, mLSTM~\cite{beck2024xlstm}, xLSTM-L~\citep{beck2025xlstm},
based on contractive linear recurrence;
DeltaNet~\citep{schlagis21} %
based on non-expansive linear recurrence;
sLSTM~\citep{beck2024xlstm}
based on nonlinear recurrence.

\textbf{Setup.}
We train on sequences of length up to~512, validate up to~2048,
and evaluate on significantly longer sequences up to length~1M;
with the rationale that the ability of a model to generalise from length~512 to~1M 
should reasonably indicate whether it has learned to solve the task in general, 
or only its restriction to sequences of a bounded length.
The training configuration for MinMax is:
Adam optimiser with weight decay~$1\mathrm{e}{-4}$, 
constant learning rate~$1\mathrm{e}{-3}$, 
batch size~$64$. %
MinMax uses AMP and PyTorch's compile.
The baselines are trained according to the respective papers.
In the case of multiple hyperparameter configurations,
we evaluate the model having the highest validation accuracy.
The time limit is~4h for each model-benchmark pair.
Datasets have size~20k for training, 1k~for validation, and~1k for evaluation.
Models are set to comparable architectural hyperparameters:
fixed for \dlatching{n} and \dheads{n} with~2 layers and \textasciitilde125k~params; 
the same for \dsequences{n} but with $n+1$~layers---all models fail to learn with fewer layers.
\dlatching{n} and \dsequences{n} are seen as seq-to-seq tasks, 
averaging loss and accuracy across sequence steps. 
For \dheads{n} we measure loss and accuracy at the recall step.

\usetikzlibrary{patterns.meta}

\definecolor{mcMinMax}      {HTML}{43863E}
\definecolor{mcMamba}       {HTML}{00A0D5}
\definecolor{mcXLSTM}       {HTML}{EA3323}
\definecolor{mcMLSTM}       {HTML}{B6824B}
\definecolor{mcSLSTM}       {HTML}{9D3C75}
\definecolor{mcDeltaNet}    {HTML}{EF8633}
\definecolor{mcDeltaProduct}{HTML}{1F7A8C}
 \colorlet{rcMinMax}      {mcMinMax}
\colorlet{rcMamba}       {mcMamba}
\colorlet{rcXLSTM}       {mcXLSTM}
\colorlet{rcMLSTM}       {mcMLSTM}
\colorlet{rcSLSTM}       {mcSLSTM}
\colorlet{rcDeltaNet}    {mcDeltaNet}
\colorlet{rcDeltaProduct}{mcDeltaProduct}
\colorlet{rcText}{black!93}
\colorlet{rcMuted}{rcText!65}

\def\rcAxis{0.7pt}
\def\rcAxisHalf{0.35pt}
\def\rcBW{0.96}\def\rcHW{0.48}

\pgfplotsset{
  rc results/.style={
    width=0.5\linewidth-0.88cm, height=2.92cm, scale only axis,
    ybar, bar width=\rcBW,
    font=\footnotesize,
    xmin=-3.25, xmax=23.25,
    xtick={0,7,14,19,22},
    xticklabels={{\texttt{Latching}},{\texttt{Sequences}},{\texttt{Heads}},{$S_3$},{$S_4$}},
    xticklabel style={font=\footnotesize, yshift=-1.4mm},
    axis x line*=bottom, axis y line*=left, y axis line style={draw=none},
    x axis line style={line width=\rcAxis, rcText!70, yshift=\rcAxisHalf},
    major tick length=0pt,
    ymajorgrids, major grid style={line width=0.4pt, rcText!14},
    yticklabel style={font=\fontsize{6}{7}\selectfont, color=rcMuted,
                      xshift=0.7mm},%
    title style={font=\footnotesize, color=rcText, yshift=-0.5mm},
    every axis plot/.append style={draw=none},
    clip=false,
  },
}

\def\rcA{-2.5}\def\rcB{-1.5}\def\rcC{-0.5}\def\rcD{0.5}\def\rcE{1.5}\def\rcF{2.5}
\def\rcL{0}\def\rcQ{7}\def\rcH{14}\def\rcSthree{19}\def\rcSfour{22}

\newcommand\rcstub[1]{%
  \fill[pattern={Lines[angle=45, distance=0.5mm, line width=0.4pt]}, pattern color=rcSLSTM]
    (axis cs:#1-\rcHW,0) rectangle ([yshift=2.2pt]axis cs:#1+\rcHW,0);
  \draw[rcSLSTM, line width=0.4pt] ([xshift=0.2pt]axis cs:#1-\rcHW,0) |-
    ([yshift=2pt]axis cs:#1,0) -| ([xshift=-0.2pt]axis cs:#1+\rcHW,0);
  \fill[rcSLSTM] (axis cs:#1-\rcHW,0) rectangle ([yshift=\rcAxis]axis cs:#1+\rcHW,0);}

\newcommand\rcbracket[2]{%
  \draw[rcText!45, line width=0.5pt]
    ([yshift=-0.6mm]axis cs:#1-#2,0) -- ([yshift=-1.2mm]axis cs:#1-#2,0) --
    ([yshift=-1.2mm]axis cs:#1+#2,0) -- ([yshift=-0.6mm]axis cs:#1+#2,0);}
\newcommand\rcbrackets{%
  \rcbracket{\rcL}{2.98}\rcbracket{\rcQ}{2.98}\rcbracket{\rcH}{2.98}%
  \rcbracket{\rcSthree}{0.98}\rcbracket{\rcSfour}{0.98}}

\newlength\rcOverflow\newlength\rcTmp
\settowidth\rcOverflow{\fontsize{6}{7}\selectfont$100$\%}
\settowidth\rcTmp{\fontsize{6}{7}\selectfont$75$\%}
\addtolength\rcOverflow{-\rcTmp}

\begin{figure}[t]
\centering
\tikzexternaldisable
\begin{tikzpicture}
  \begin{axis}[hide axis, xmin=0, xmax=1, ymin=0, ymax=1, width=4cm, height=2cm,
      legend columns=7,
      legend style={draw=none, fill=none, font=\footnotesize,
        /tikz/every odd column/.append style={column sep=0.8mm},
        /tikz/every even column/.append style={column sep=2.2mm}},
      legend image code/.code={\fill[#1] (0mm,-1.1mm) rectangle (2.6mm,1.1mm);}]
    \addlegendimage{draw=none, fill=rcMinMax}      \addlegendentry{MinMax}
    \addlegendimage{draw=none, fill=rcMamba}       \addlegendentry{Mamba}
    \addlegendimage{draw=none, fill=rcXLSTM}       \addlegendentry{xLSTM-L}
    \addlegendimage{draw=none, fill=rcMLSTM}       \addlegendentry{mLSTM}
    \addlegendimage{draw=none, fill=rcSLSTM}       \addlegendentry{sLSTM}
    \addlegendimage{draw=none, fill=rcDeltaNet}    \addlegendentry{DeltaNet}
    \addlegendimage{draw=none, fill=rcDeltaProduct}\addlegendentry{DeltaProduct}
  \end{axis}
\end{tikzpicture}
\tikzexternalenable

\vspace{1mm}

\noindent\hspace*{-\rcOverflow}%
\tikzsetnextfilename{plot__results__accuracy}%
\begin{tikzpicture}
  \begin{axis}[rc results, ymin=0, ymax=100, ytick={25,50,75,100},
      yticklabel={\pgfmathprintnumber{\tick}\%},
      extra y ticks={0}, extra y tick labels={0\%}, extra y tick style={grid=none},
      title={Evaluation accuracy up to length 1M}]
    \addplot[fill=rcMinMax, bar shift=\rcA] coordinates {(\rcL,100) (\rcQ,100) (\rcH,100)};
    \addplot[fill=rcMamba,  bar shift=\rcB] coordinates {(\rcL,54)  (\rcQ,47)  (\rcH,52)};
    \addplot[fill=rcXLSTM,  bar shift=\rcC] coordinates {(\rcL,8)   (\rcQ,47)  (\rcH,5)};
    \addplot[fill=rcMLSTM,  bar shift=\rcD] coordinates {(\rcL,7)   (\rcQ,45)  (\rcH,5)};
    \addplot[fill=rcSLSTM,  bar shift=\rcE] coordinates {(\rcL,28)  (\rcQ,47)  (\rcH,4.5)};
    \addplot[fill=rcDeltaNet, bar shift=\rcF] coordinates {(\rcL,8) (\rcQ,17)  (\rcH,4.5)};
    \addplot[fill=rcMinMax, bar shift=-0.5] coordinates {(\rcSthree,100) (\rcSfour,100)};
    \addplot[fill=rcDeltaProduct, bar shift=0.5] coordinates {(\rcSthree,17.16) (\rcSfour,4.33)};
    \rcbrackets
  \end{axis}
\end{tikzpicture}\hfill%
\tikzsetnextfilename{plot__results__throughput}%
\begin{tikzpicture}
  \begin{axis}[rc results, ymin=0, ymax=2, ytick={0.5,1.5,2},
      yticklabel={\pgfmathprintnumber{\tick}$\mkern-2mu\times$},
      extra y ticks={0,1},
      extra y tick labels={{0$\mkern-2mu\times$},{\textcolor{rcMinMax}{\textbf{1$\mkern-2mu\times$}}}},
      extra y tick style={grid=none},
      title={Throughput relative to MinMax}]
    \addplot[fill=rcMinMax, bar shift=\rcA] coordinates {(\rcL,1) (\rcQ,1) (\rcH,1)};
    \addplot[fill=rcMamba,  bar shift=\rcB] coordinates {(\rcL,0.297) (\rcQ,0.908) (\rcH,0.279)};
    \addplot[fill=rcXLSTM,  bar shift=\rcC] coordinates {(\rcL,1.090) (\rcQ,1.881) (\rcH,1.210)};
    \addplot[fill=rcMLSTM,  bar shift=\rcD] coordinates {(\rcL,0.103) (\rcQ,0.186) (\rcH,0.084)};
    \rcstub{\rcL+\rcE}\rcstub{\rcQ+\rcE}\rcstub{\rcH+\rcE}
    \addplot[fill=rcDeltaNet, bar shift=\rcF] coordinates {(\rcL,0.615) (\rcQ,1.190) (\rcH,1.157)};
    \addplot[fill=rcMinMax, bar shift=-0.5] coordinates {(\rcSthree,1) (\rcSfour,1)};
    \addplot[fill=rcDeltaProduct, bar shift=0.5] coordinates {(\rcSthree,0.888) (\rcSfour,0.888)};
    \rcbrackets
    \draw[rcMinMax, line width=0.8pt, dash pattern=on 1.4mm off 0.8mm]
      ({rel axis cs:0,0}|-{axis cs:0,1}) -- ({rel axis cs:1,0}|-{axis cs:0,1});
  \end{axis}
\end{tikzpicture}

\caption{Accuracy (left) and throughput (right);
  group-free benchmarks aggregated as in Table~\ref{tab:experiments-mainbody-compiled-avgthr}.
}
\label{fig:results-charts}
\end{figure}

\textbf{Results.}
Table~\ref{tab:experiments-mainbody-compiled-avgthr} and Figure~\ref{fig:results-charts} summarise the results on group-free tasks.
MinMax always achieves perfect validation accuracy and retains it at evaluation time,
whereas
the other models can achieve perfect or near-perfect validation accuracy but fail to retain it at
evaluation time as it decreases over sequence length,
as shown in Figure~\ref{fig:accuracy-over-length-skill};
this validates the formal-expressivity results for MinMax RNCs of state degree one~%
(Theorem~\ref{thm:main-body--expressivity--overall}), 
and indirectly the results on their favourable stability and gradient properties~%
(Theorems~\ref{th:stability-simpler}--\ref{thm:gradient-nonvanishing}).
MinMax has the second-highest training throughput after xLSTM-L,
and the highest throughput at evaluation time. 
This validates the computational-complexity results for MinMax RNCs of state-degree one~%
(Theorem~\ref{thm:complexity-minmax-recurrence-main-body}). 
We note that the throughput of MinMax is significantly lower 
when not using AMP and PyTorch's~compile,
which suggests that MinMax recurrence is effectively 
able to take advantage of hardware acceleration in a significant manner.

\newcommand{\tableLabelSize}{\small}
\newcommand{\tableLabelAccuracy}{{\tableLabelSize Acc}}
\newcommand{\tableLabelThroughput}{{\tableLabelSize Thr}}
\newlength{\groupsep}\setlength{\groupsep}{16pt}
\newlength{\accsep}\setlength{\accsep}{6pt}
\newcommand{\spanhead}[3]{\multicolumn{#1}{@{}c@{\hspace{#2}}}{#3}}
\newcommand{\intentry}[1]{\num[round-mode=none]{#1}}
\newcommand{\failsym}{\dag}
\newcommand{\failmark}{\textsuperscript{\failsym}}
\newcommand{\failed}{{\rlap{\failmark}}}
\newcommand{\timeoutsym}{\ddag}
\newcommand{\timeoutmark}{\textsuperscript{\timeoutsym}}
\newcommand{\timedout}{{\rlap{\timeoutmark}}}
\newcommand{\smallbatchsym}{\S}
\newcommand{\smallbatchmark}{\textsuperscript{\smallbatchsym}}
\newcommand{\smallbatch}{{\rlap{\smallbatchmark}}}
\newcommand{\tableLabelFailed}{\failsym~no training seed succeeded (val.~acc.~<50\%)}
\newcommand{\tableLabelTimeout}{\timeoutsym~training timed out}
\newcommand{\tableLabelSmallBatch}{\smallbatchsym~training succeeded only with batch size~8}

\begin{table}[t]
\centering
\caption{\emph{Evaluation accuracy}~($\%$): means of 99\% CIs with max margin~1.7.
  \emph{Relative throughput}~(tok/s, ratio w.r.t.\ MinMax): avg.\ of training and evaluation
  ratios of the means of 99\% CIs with max margin~0.4.
  For \dlatching{2^n} and \dsequences{m}, avg.\ over
  $n \cin \{2,\ldots,9\}$ and $m \cin \{2,3,4,8,16\}$.
}
\vspace{.2em}
\label{tab:experiments-mainbody-compiled-avgthr}
{\setlength{\tabcolsep}{2pt}%
\sisetup{round-mode=places}%
\begin{tabular}{@{}l@{\hspace{\groupsep}}
  *{5}{S[table-format=3.0, round-precision=0]@{\hspace{\accsep}}
       S[table-format=1.1, round-precision=1]@{\hspace{\groupsep}}}
  S[table-format=3.0, round-precision=0]@{\hspace{\accsep}}
  S[table-format=1.1, round-precision=1]@{}}
\toprule
& \spanhead{2}{\groupsep}{MinMax} & \spanhead{2}{\groupsep}{Mamba} & \spanhead{2}{\groupsep}{xLSTM-L}
  & \spanhead{2}{\groupsep}{mLSTM} & \spanhead{2}{\groupsep}{sLSTM} & \spanhead{2}{0pt}{DeltaNet} \\
\cmidrule(l{0pt}r{\groupsep}){2-3} \cmidrule(l{0pt}r{\groupsep}){4-5} \cmidrule(l{0pt}r{\groupsep}){6-7}
\cmidrule(l{0pt}r{\groupsep}){8-9} \cmidrule(l{0pt}r{\groupsep}){10-11} \cmidrule(l{0pt}){12-13}
& {\tableLabelAccuracy} & {\tableLabelThroughput}
  & {\tableLabelAccuracy} & {\tableLabelThroughput}
  & {\tableLabelAccuracy} & {\tableLabelThroughput}
  & {\tableLabelAccuracy} & {\tableLabelThroughput}
  & {\tableLabelAccuracy} & {\tableLabelThroughput}
  & {\tableLabelAccuracy} & {\tableLabelThroughput} \\
\midrule
\dlatching{2^n}
  & 100 & {\intentry{1}} & 54  & 0.297 & 8  & 1.090 & 7  & 0.103 & 28\timedout & {\intentry{0}} & 8  & 0.615 \\
\dsequences{m}
  & 100 & {\intentry{1}} & 47  & 0.908 & 47 & 1.881 & 45 & 0.186 & 47\timedout & {\intentry{0}} & 17 & 1.190 \\
\dheads{16}
  & 100 & {\intentry{1}} & 100\smallbatch & 0.261 & 7\failed & 1.216 & 7  & 0.075 & 6  & {\intentry{0}} & 5  & 0.964 \\
\dheads{30}
  & 100 & {\intentry{1}} & 4\failed   & 0.297 & 3\failed & 1.204 & 3  & 0.093 & 3  & {\intentry{0}} & 4  & 1.349 \\
\bottomrule
\addlinespace[3pt]
\multicolumn{13}{@{}l@{}}{\footnotesize \tableLabelFailed \quad \tableLabelTimeout\quad \tableLabelSmallBatch} \\
\end{tabular}}
\end{table}

\definecolor{mcMinMax}      {HTML}{43863E}
\definecolor{mcMamba}       {HTML}{00A0D5}
\definecolor{mcXLSTM}       {HTML}{EA3323}
\definecolor{mcMLSTM}       {HTML}{B6824B}
\definecolor{mcSLSTM}       {HTML}{9D3C75}
\definecolor{mcDeltaNet}    {HTML}{EF8633}
\definecolor{mcDeltaProduct}{HTML}{1F7A8C}

\begin{figure}[!h]
  \centering
  \tikzexternaldisable
  \begin{tikzpicture}
  \begin{axis}[
    hide axis,
    xmin=0, xmax=1,
    ymin=0, ymax=1,
    width=5cm,
    height=2cm,
    legend cell align=left,
    legend columns=6,
    legend style={
      font=\footnotesize,
      draw=none,
      /tikz/every even column/.append style={column sep=0.4cm},
    },
  ]

    \addlegendimage{mcMinMax, line width=1.0pt, mark=none}
    \addlegendentry{MinMax}

    \addlegendimage{mcMamba, line width=1.0pt, mark=none}
    \addlegendentry{Mamba}

    \addlegendimage{mcXLSTM, line width=1.0pt, mark=none}
    \addlegendentry{xLSTM-L}

    \addlegendimage{mcMLSTM, line width=1.0pt, mark=none}
    \addlegendentry{mLSTM}

    \addlegendimage{mcSLSTM, line width=1.0pt, mark=none}
    \addlegendentry{sLSTM}

    \addlegendimage{mcDeltaNet, line width=1.0pt, mark=none}
    \addlegendentry{DeltaNet}

  \end{axis}
  \end{tikzpicture}
  \tikzexternalenable

  \noindent\begin{minipage}[t]{0.5\textwidth}
    \centering

    \tikzsetnextfilename{plot__latching_2__skill}
    \begin{tikzpicture}
      \tikzpicturedependsonfile{data/evaluations_skill/latching_2/mamba_lm__layers_2__small__trainer_mamba.csv}
      \tikzpicturedependsonfile{data/evaluations_skill/latching_2/maxoutrnn_lm__layers_2__small__trainer_maxout.csv}
      \tikzpicturedependsonfile{data/evaluations_skill/latching_2/xlstm_lm__layers_2__small__trainer_xlstm.csv}
      \tikzpicturedependsonfile{data/evaluations_skill/latching_2/slstm_lm__layers_2__small__trainer_xlstm.csv}
      \tikzpicturedependsonfile{data/evaluations_skill/latching_2/xlstmlarge_lm__layers_2__small__trainer_xlstm.csv}
      \tikzpicturedependsonfile{data/evaluations_skill/latching_2/deltanet_lm__layers_2__small__trainer_deltanet.csv}

      \begin{axis}[
        width=\linewidth+12pt,
        height=4cm,
        title={\dlatching{8}},
        title style={font=\footnotesize, yshift=-1.3ex},
        extra x ticks={1024}, extra x tick labels={Step},
        extra x tick style={grid=none, major tick length=0pt,
          xticklabel style={anchor=base west, xshift=-0.3em, font=\footnotesize}},
        xticklabel style={anchor=base, yshift=-2ex},
        xticklabels={2e5,4e5,6e5,8e5,1e6},
        after end axis/.code={\node[anchor=south east, font=\scriptsize, inner sep=0pt]
          at ([xshift=-0.333em, yshift=7pt]axis cs:1024,100) {Gain};},
        yticklabel={\pgfmathprintnumber{\tick}\%},
        grid=both,
        ymin=-8, ymax=110,
        xmin=1024, xmax=1000000,
        legend cell align=left,
        legend columns=5,
        legend style={font=\footnotesize},
        legend to name=commonlegend,
        label style={font=\footnotesize},
        tick label style={font=\scriptsize},
        ytick distance=25,
        scaled x ticks=false,
        xtick={200000,400000,600000,800000,1000000},
      ]

        \addplot[mcMamba, line width=1.0pt,mark=none]
          table[x=step, y expr=\thisrow{acc_mean}*100, col sep=comma]
          {data/evaluations_skill/latching_2/mamba_lm__layers_2__small__trainer_mamba.csv};

        \addplot[mcMLSTM, line width=1.0pt, mark=none]
          table[x=step, y expr=\thisrow{acc_mean}*100, col sep=comma]
          {data/evaluations_skill/latching_2/xlstm_lm__layers_2__small__trainer_xlstm.csv};

        \addplot[mcXLSTM, line width=1.0pt, mark=none]
          table[x=step, y expr=\thisrow{acc_mean}*100, col sep=comma]
          {data/evaluations_skill/latching_2/xlstmlarge_lm__layers_2__small__trainer_xlstm.csv};

        \addplot[mcMinMax, line width=1.0pt, mark=none]
          table[x=step, y expr=\thisrow{acc_mean}*100, col sep=comma]
          {data/evaluations_skill/latching_2/maxoutrnn_lm__layers_2__small__trainer_maxout.csv};

        \addplot[mcSLSTM, line width=1.0pt,mark=none]
          table[x=step, y expr=\thisrow{acc_mean}*100, col sep=comma]
          {data/evaluations_skill/latching_2/slstm_lm__layers_2__small__trainer_xlstm.csv};

        \addplot[mcDeltaNet, line width=1.0pt, mark=none]
          table[x=step, y expr=\thisrow{acc_mean}*100, col sep=comma]
          {data/evaluations_skill/latching_2/deltanet_lm__layers_2__small__trainer_deltanet.csv};

        \addplot[name path=hi1, draw=none]
          table[x=step, y expr=\thisrow{acc_high}*100, col sep=comma]
          {data/evaluations_skill/latching_2/mamba_lm__layers_2__small__trainer_mamba.csv};
        \addplot[name path=lo1, draw=none]
          table[x=step, y expr=\thisrow{acc_low}*100, col sep=comma]
          {data/evaluations_skill/latching_2/mamba_lm__layers_2__small__trainer_mamba.csv};
        \addplot[fill=mcMamba!20, draw=none] fill between[of=hi1 and lo1];

        \addplot[name path=hi2, draw=none]
          table[x=step, y expr=\thisrow{acc_high}*100, col sep=comma]
          {data/evaluations_skill/latching_2/maxoutrnn_lm__layers_2__small__trainer_maxout.csv};
        \addplot[name path=lo2, draw=none]
          table[x=step, y expr=\thisrow{acc_low}*100, col sep=comma]
          {data/evaluations_skill/latching_2/maxoutrnn_lm__layers_2__small__trainer_maxout.csv};
        \addplot[fill=mcMinMax!20, draw=none] fill between[of=hi2 and lo2];

        \addplot[name path=hi3, draw=none]
          table[x=step, y expr=\thisrow{acc_high}*100, col sep=comma]
          {data/evaluations_skill/latching_2/xlstm_lm__layers_2__small__trainer_xlstm.csv};
        \addplot[name path=lo3, draw=none]
          table[x=step, y expr=\thisrow{acc_low}*100, col sep=comma]
          {data/evaluations_skill/latching_2/xlstm_lm__layers_2__small__trainer_xlstm.csv};
        \addplot[fill=mcMLSTM!20, draw=none] fill between[of=hi3 and lo3];

        \addplot[name path=hi4, draw=none]
          table[x=step, y expr=\thisrow{acc_high}*100, col sep=comma]
          {data/evaluations_skill/latching_2/xlstmlarge_lm__layers_2__small__trainer_xlstm.csv};
        \addplot[name path=lo4, draw=none]
          table[x=step, y expr=\thisrow{acc_low}*100, col sep=comma]
          {data/evaluations_skill/latching_2/xlstmlarge_lm__layers_2__small__trainer_xlstm.csv};
        \addplot[fill=mcXLSTM!20, draw=none] fill between[of=hi4 and lo4];

        \addplot[name path=hi5, draw=none]
          table[x=step, y expr=\thisrow{acc_high}*100, col sep=comma]
          {data/evaluations_skill/latching_2/deltanet_lm__layers_2__small__trainer_deltanet.csv};
        \addplot[name path=lo5, draw=none]
          table[x=step, y expr=\thisrow{acc_low}*100, col sep=comma]
          {data/evaluations_skill/latching_2/deltanet_lm__layers_2__small__trainer_deltanet.csv};
        \addplot[fill=mcDeltaNet!20, draw=none] fill between[of=hi5 and lo5];

      \end{axis}
    \end{tikzpicture}
  \end{minipage}%
  \begin{minipage}[t]{0.5\textwidth}
    \centering
    \tikzsetnextfilename{plot__sequences_1__skill}
    \begin{tikzpicture}
      \tikzpicturedependsonfile{data/evaluations_skill/sequences_1/mamba_lm__layers_2__small__trainer_mamba.csv}
      \tikzpicturedependsonfile{data/evaluations_skill/sequences_1/maxoutrnn_lm__layers_2__small__trainer_maxout.csv}
      \tikzpicturedependsonfile{data/evaluations_skill/sequences_1/xlstm_lm__layers_2__small__trainer_xlstm.csv}
      \tikzpicturedependsonfile{data/evaluations_skill/sequences_1/slstm_lm__layers_2__small__trainer_xlstm.csv}
      \tikzpicturedependsonfile{data/evaluations_skill/sequences_1/xlstmlarge_lm__layers_2__small__trainer_xlstm.csv}
      \tikzpicturedependsonfile{data/evaluations_skill/sequences_1/deltanet_lm__layers_2__small__trainer_deltanet.csv}

      \begin{axis}[
        width=\linewidth+12pt,
        height=4cm,
        title={\dsequences{2}},
        title style={font=\footnotesize, yshift=-1.3ex},
        extra x ticks={1024}, extra x tick labels={Step},
        extra x tick style={grid=none, major tick length=0pt,
          xticklabel style={anchor=base west, xshift=-0.3em, font=\footnotesize}},
        xticklabel style={anchor=base, yshift=-2ex},
        xticklabels={2e5,4e5,6e5,8e5,1e6},
        after end axis/.code={\node[anchor=south east, font=\scriptsize, inner sep=0pt]
          at ([xshift=-0.333em, yshift=7pt]axis cs:1024,100) {Gain};},
        yticklabel={\pgfmathprintnumber{\tick}\%},
        grid=both,
        ymin=-8, ymax=110,
        xmin=1024, xmax=1000000,
        label style={font=\footnotesize},
        tick label style={font=\scriptsize},
        ytick distance=25,
        scaled x ticks=false,
        xtick={200000,400000,600000,800000,1000000},
      ]

        \addplot[mcMamba, line width=1.0pt, mark=none]
          table[x=step, y expr=\thisrow{acc_mean}*100, col sep=comma]
          {data/evaluations_skill/sequences_1/mamba_lm__layers_2__small__trainer_mamba.csv};

        \addplot[mcMLSTM, line width=1.0pt, mark=none]
          table[x=step, y expr=\thisrow{acc_mean}*100, col sep=comma]
          {data/evaluations_skill/sequences_1/xlstm_lm__layers_2__small__trainer_xlstm.csv};

        \addplot[mcXLSTM, line width=1.0pt, mark=none]
          table[x=step, y expr=\thisrow{acc_mean}*100, col sep=comma]
          {data/evaluations_skill/sequences_1/xlstmlarge_lm__layers_2__small__trainer_xlstm.csv};

        \addplot[mcMinMax, line width=1.0pt, mark=none]
          table[x=step, y expr=\thisrow{acc_mean}*100, col sep=comma]
          {data/evaluations_skill/sequences_1/maxoutrnn_lm__layers_2__small__trainer_maxout.csv};

        \addplot[mcSLSTM, line width=1.0pt, mark=none]
          table[x=step, y expr=\thisrow{acc_mean}*100, col sep=comma]
          {data/evaluations_skill/sequences_1/slstm_lm__layers_2__small__trainer_xlstm.csv};

        \addplot[mcDeltaNet, line width=1.0pt, mark=none]
          table[x=step, y expr=\thisrow{acc_mean}*100, col sep=comma]
          {data/evaluations_skill/sequences_1/deltanet_lm__layers_2__small__trainer_deltanet.csv};

        \addplot[name path=hi1, draw=none]
          table[x=step, y expr=\thisrow{acc_high}*100, col sep=comma]
          {data/evaluations_skill/sequences_1/mamba_lm__layers_2__small__trainer_mamba.csv};
        \addplot[name path=lo1, draw=none]
          table[x=step, y expr=\thisrow{acc_low}*100, col sep=comma]
          {data/evaluations_skill/sequences_1/mamba_lm__layers_2__small__trainer_mamba.csv};
        \addplot[fill=mcMamba!20, draw=none] fill between[of=hi1 and lo1];

        \addplot[name path=hi2, draw=none]
          table[x=step, y expr=\thisrow{acc_high}*100, col sep=comma]
          {data/evaluations_skill/sequences_1/maxoutrnn_lm__layers_2__small__trainer_maxout.csv};
        \addplot[name path=lo2, draw=none]
          table[x=step, y expr=\thisrow{acc_low}*100, col sep=comma]
          {data/evaluations_skill/sequences_1/maxoutrnn_lm__layers_2__small__trainer_maxout.csv};
        \addplot[fill=mcMinMax!20, draw=none] fill between[of=hi2 and lo2];

        \addplot[name path=hi3, draw=none]
          table[x=step, y expr=\thisrow{acc_high}*100, col sep=comma]
          {data/evaluations_skill/sequences_1/xlstm_lm__layers_2__small__trainer_xlstm.csv};
        \addplot[name path=lo3, draw=none]
          table[x=step, y expr=\thisrow{acc_low}*100, col sep=comma]
          {data/evaluations_skill/sequences_1/xlstm_lm__layers_2__small__trainer_xlstm.csv};
        \addplot[fill=mcMLSTM!20, draw=none] fill between[of=hi3 and lo3];

        \addplot[name path=hi4, draw=none]
          table[x=step, y expr=\thisrow{acc_high}*100, col sep=comma]
          {data/evaluations_skill/sequences_1/xlstmlarge_lm__layers_2__small__trainer_xlstm.csv};
        \addplot[name path=lo4, draw=none]
          table[x=step, y expr=\thisrow{acc_low}*100, col sep=comma]
          {data/evaluations_skill/sequences_1/xlstmlarge_lm__layers_2__small__trainer_xlstm.csv};
        \addplot[fill=mcXLSTM!20, draw=none] fill between[of=hi4 and lo4];

        \addplot[name path=hi5, draw=none]
          table[x=step, y expr=\thisrow{acc_high}*100, col sep=comma]
          {data/evaluations_skill/sequences_1/deltanet_lm__layers_2__small__trainer_deltanet.csv};
        \addplot[name path=lo5, draw=none]
          table[x=step, y expr=\thisrow{acc_low}*100, col sep=comma]
          {data/evaluations_skill/sequences_1/deltanet_lm__layers_2__small__trainer_deltanet.csv};
        \addplot[fill=mcDeltaNet!20, draw=none] fill between[of=hi5 and lo5];

      \end{axis}
    \end{tikzpicture}
  \end{minipage}
  \vspace{-0.5em} %
  \caption{Accuracy gain 
    $\max\bigl(0,
    \frac{\mathrm{acc}-\mathrm{acc}_{\mathsf{naive}}}{1-\mathrm{acc}_{\mathsf{naive}}}\bigr)$ over a
    naive baseline that ignores inputs and predicts the most likely label at each step:
    $100\%$ is perfect accuracy, and $0\%$ is no better than the baseline.%
  }%
  \label{fig:accuracy-over-length-skill}
\end{figure}

\textbf{Assessment of non-vanishing gradient.}
Solving~\dsequences{n} requires storing the first event of a sequence until the
second one arrives; the error signal to learn doing so arises from the second
event onwards, and has to propagate back across the \emph{event gap}---the number of steps
between the two events.
In the training data of~\dsequences{2}, the event gap is 4~steps on average.
We train on a long-range variant,~\dsequencesLongRange{2}, with the same events but
sequences of length~$2^{17}$, where the event gap is 37k~steps on average.
MinMax achieves near-perfect validation accuracy ($\geq 99.99\%$),
whereas Mamba and DeltaNet do not learn the task: their validation accuracy remains at
the~\textasciitilde94\% of the trivial predictor that never reports a match.
This suggests that the error signal propagates across the event gaps in MinMax, and not
in Mamba and DeltaNet, hence providing empirical support to the non-vanishing gradient
results of Theorem~\ref{thm:gradient-nonvanishing}---see
Appendix~\ref{app:experiments-longrange} for details.

\subsection{Evaluation on group-based tasks}
\label{subsec:experiments-synth--groups}
To evaluate MinMax RNCs beyond group-free tasks,
we consider the \emph{word problems} defined by the symmetry groups~$S_3$ and~$S_4$, 
included in the evaluation of DeltaProduct by~\citet{siems2025deltaproduct},
which thus provides us with a baseline.
We set MinMax to have 
a single layer 
and 
state degree equal to~$3$ and~$4$ on~$S_3$ and~$S_4$, respectively, 
which suffices according to Theorem~\ref{thm:main-body-expressivity}.

We train on 20k~sequences of length~128 on~$S_3$ and 200k on~$S_4$, validate on 1k~sequences of
length~512, and evaluate on 1k~sequences of length~1M.
MinMax achieves perfect validation and evaluation accuracies.
In contrast, the single-layer DeltaProduct results reported in
Fig.~5~of~\citep{siems2025deltaproduct} show an accuracy that begins to deteriorate already at
length~400--500 on~$S_3$ and length~200--400 on~$S_4$,
despite training over datasets of 2M~examples.
For a closer comparison,
we train single-layer DeltaProduct using their training configuration but on datasets as 
above: its best validation accuracy is~97\% on~$S_3$ and 99\%~on~$S_4$ 
(in line with~\citep{siems2025deltaproduct}); when evaluated on length up to~1M, its
accuracy falls to chance level within the first~$10^5$ steps on~$S_3$ and~$10^4$ steps
on~$S_4$---see Appendix~\ref{app:experiments-groups} and Figure~\ref{fig:groups_eval}.

\subsection{Next-token prediction on OpenWebText (OWT)} 
\label{sec:experiments-llm}
We train a degree-one MinMax RNC of 112M parameters for next-token prediction 
following \citep{karpathy2022nanogpt}, which reproduces GPT-2 on OWT~\citep{Gokaslan2019OpenWeb}.
It achieves the same validation loss of~$3.12$ as the one reported in
\citep{karpathy2022nanogpt} for the released OpenAI GPT-2 124M checkpoint when evaluated on OWT.

\section{Conclusion}

The theoretical analysis and empirical evaluation suggest that MinMax recurrence is a relevant 
alternative to the existing forms of recurrence.
Our future work will focus on better understanding the effectiveness of MinMax RNCs 
in real-world applications, and also on the potential of MinMax recurrence as a 
mechanism to integrate existing architectures in order to improve their capabilities.

\mypar{Limitations.}
We acknowledge that
\begin{inlineenum}
\item
  our empirical evaluation is relatively limited compared to the ones included in
related research papers;
\item
  we do not provide precise answers regarding the depth and
  width required for cascades to recognise a given language, due to the corresponding questions
  being open in automata theory.
\end{inlineenum}

\bibliographystyle{unsrtnat}
\bibliography{bibliography}    

\makeatletter
\newif\ifpartlocaltoc \partlocaltoctrue
\newcommand{\nextpartlabel}{}
\newcommand{\appendixpart}{\@ifstar
  {\global\partlocaltocfalse\@appendixpart}{\@appendixpart}}
\newcommand{\@appendixpart}[2]{\gdef\nextpartlabel{#1}\part{#2}}

\renewcommand\part{%
  \clearpage
  \thispagestyle{empty}%
  \null\vfill
  \secdef\@part\@spart
}
\renewcommand\@part[2][]{%
  \refstepcounter{part}%
  \ifx\nextpartlabel\@empty\else
    \label{\nextpartlabel}\global\let\nextpartlabel\@empty
  \fi
  \addcontentsline{toc}{part}{\thepart\hspace{1em}#1}%
  \begin{center}
    {\huge\bfseries Part \thepart\par}
    \vspace{1.5em}
    {\Huge\bfseries #2\par}
  \end{center}
  \ifpartlocaltoc
    \vspace{2.5em}%
    \begingroup
      \etocsettocstyle{}{}%
      \etocsetlevel{subsection}{6}%
      \localtableofcontents
    \endgroup
  \fi
  \global\partlocaltoctrue
  \vfill\null
  \clearpage
}
\renewcommand\@spart[1]{%
  \begin{center}
    {\Huge\bfseries #1\par}
  \end{center}
  \vfill\null
  \clearpage
}
\makeatother

\clearpage
\appendix
\addtocontents{toc}{\protect\setcounter{tocdepth}{2}}
\renewcommand{\contentsname}{Appendices}
\etocsettocstyle{\textbf{Table of Contents}}{}

\thispagestyle{empty}
\begin{center}
  {\Huge\bfseries Appendices\par}
\end{center}
\vspace{2em}

\noindent
The appendices contain the formal development behind the main body (the notions it builds on,
the proofs of its results), the implementation of MinMax RNCs, the details of every experiment,
and an analysis of the related architectures we compare with.
They are organised in five parts.
\begin{description}
  \item[Part~\ref{part:background}$\mkern2mu$:$\mkern2mu$ Background]
    extends the preliminaries of the main body.
  \item[Part~\ref{part:notions}: Preliminary Notions and Results]
    introduces dynamics and systems and their compositions, generic architectures, the MinMax
    algebra, selected intensional derivatives, and the properties we study, with
    auxiliary architecture-agnostic results for them.
  \item[Part~\ref{part:theory}: Theoretical Results for MinMax RNCs]
    proves the results of the main body on expressivity, algorithms and complexity, stability,
    and gradients, and describes the piecewise-affine structure of MinMax recurrences.
  \item[Part~\ref{part:experiments}: Implementation and Experiments]
    describes the implementation of MinMax RNCs and gives the details of the experiments on
    group-free benchmarks, long-range dependencies, group-based benchmarks and OpenWebText.
  \item[Part~\ref{part:related}: Technical Analysis of Related Architectures]
    analyses the formal properties of Linear RNNs, DeltaNet, Mamba, and mLSTM.
\end{description}

\section*{Where to find the details of the main body}
\begin{center}
\begin{tabularx}{0.6\textwidth}{@{}Xl@{}}
  \toprule
  \textbf{Main body} & \textbf{Appendix} \\
  \midrule
  Expressivity (Theorems~\ref{thm:main-body-expressivity},
    \ref{theorem:expressivity-main-ir} and~\ref{theorem:expressivity-main-perm})
    & \ref{app:expr} \\
  Complexity (Theorems~\ref{thm:complexity-minmax-recurrence-main-body}
    and~\ref{thm:complexity-main-sequential})
    & \ref{appendix:algorithms}, \ref{appendix:proofs-complexity} \\
  Stability (Theorem~\ref{th:stability-simpler})
    & \ref{sec:appendix:stability} \\
  Gradients (Theorems~\ref{thm:gradient-bounded-main} and~\ref{thm:gradient-nonvanishing})
    & \ref{app:gradient}, \ref{appendix:sec:gradient-aux} \\
  Model (Section~\ref{sec:model})
    & \ref{sec:minmax-implementation} \\
  Experiments (Section~\ref{sec:main-body--experiments})
    & \ref{app:experiments}--\ref{app:details-of-next-token-prediction-on-openwebtext} \\
  \bottomrule
\end{tabularx}
\end{center}

\clearpage
\begingroup
\renewcommand{\baselinestretch}{0.9}\selectfont
\tableofcontents
\endgroup

\appendixpart*{part:background}{Background}

\clearpage{}%

\section{Extended Background}
\label{app:add-prel}

\localtableofcontents

\subsection{Notation}

We write $\naturals$ for the natural numbers, 
we write $\posnaturals$ for the natural numbers exlcuding zero,
we write $\reals$ for the real numbers,
we write $\posreals$ for the non-negative real numbers,
and
we write $\posreals$ for the positive real numbers---zero is excluded.

We write $\timedomain$ for $\posnaturals$ when it is meant as the set of all timepoints.

For $X$ a set, we write $\powersetplus{X}$ for the set of all non-empty subsets of $X$.

Given $i,j \in \naturals$,
we write $\iival{i}{j}$ for the integer interval $\{ i, \ldots, j \}$,
we write $\iivallo{i}{j}$ for the integer interval $\{ i+1, \ldots, j \}$,
and
we write $\iivalro{i}{j}$ for the integer interval $\{ i, \ldots, j-1 \}$.
Such intervals can be empty, e.g., the interval $\iivallo{1}{1}$ is empty.

For $\ptuple{X,\|\cdot \|}$ a normed space
and 
for $r \in \nonnegreals$, 
we define $\truncation{X}{r} \defeq \{x\in X \mid \|x\|\leq r\}$.

\subsection{Functions}
\label{subsec:function-products-composition}

\begin{closingdefinition}[Injective, surjective, and bijective functions]
  \label{def:injective-surjective-bijective}
  A function $f : X \to Y$ is:
  \begin{itemize}
    \item
      \emph{injective} if, for every
      $x,x'\in X$,
      \begin{equation*}
        f(x)
        =
        f(x')
        \quad\Longrightarrow\quad
        x
        =
        x';
      \end{equation*}

    \item
      \emph{surjective} if, for every
      $y\in Y$,
      there exists
      $x\in X$
      such that $f(x) = y$;

    \item
      \emph{bijective} if it is both injective and surjective---also called a \emph{bijection}.
  \end{itemize}
\end{closingdefinition}

\begin{closingfact}[Inverse of a bijection]
  \label{fact:bijection-inverse}
  A function $f : X \to Y$ is bijective if and only if there exists a unique function
  \begin{equation*}
    f^{-1}
    :
    Y
    \to
    X
  \end{equation*}
  such that
  \begin{align*}
    f^{-1}\circ f
    &=
    \id[X],
    \\
    f\circ f^{-1}
    &=
    \id[Y],
  \end{align*}
  where
  $\id[X]$
  and
  $\id[Y]$
  denote the identity functions on
  $X$
  and
  $Y$,
  respectively.
  The function $f^{-1}$ is called the \emph{inverse} of $f$.
\end{closingfact}

\begin{closingdefinition}[Identity function]
  \label{def:identity-function}
  The \emph{identity function} on a set $X$ is the function
  \begin{align*}
    \id[X]
    :
    X
    &\to
    X,
    \\
    x
    &\mapsto
    x.
  \end{align*}
  We write $\id$ in place of $\id[X]$ when $X$ is clear from the context.
\end{closingdefinition}

\begin{closingdefinition}[Transformation and permutation]
  \label{def:transformation-permutation}
  A \emph{transformation} on a set $X$ is a function $f : X \to X$.
  A \emph{permutation} on a set $X$ is a bijective transformation on $X$.
\end{closingdefinition}

\begin{closingfact}
  Every identity function $\id[X]$ is a permutation on $X$.
\end{closingfact}

We use the following standard operations on functions.

\begin{closingdefinition}[Product of functions]
  \label{def:function-product}
  Let
  \begin{equation*}
    f : X \to Y,
    \qquad
    g : U \to V.
  \end{equation*}
  The \emph{product} of
  $f$
  and
  $g$
  is the function
  \begin{equation*}
    f \times g
    :
    X \times U
    \to
    Y \times V
  \end{equation*}
  defined by
  \begin{equation*}
    (f \times g)(x,u)
    \defeq
    \ptuple{
      f(x),
      g(u)
    }.
  \end{equation*}

  More generally, for functions
  \begin{equation*}
    f_i : X_i \to Y_i,
    \qquad
    i\in\iival{1}{k},
  \end{equation*}
  their product is the function
  \begin{equation*}
    f_1
    \times
    \cdots
    \times
    f_k
    :
    X_1
    \times
    \cdots
    \times
    X_k
    \to
    Y_1
    \times
    \cdots
    \times
    Y_k
  \end{equation*}
  defined by
  \begin{equation*}
    \big(
      f_1
      \times
      \cdots
      \times
      f_k
    \big)
    \ptuple{
      x_1,
      \ldots,
      x_k
    }
    \defeq
    \ptuple{
      f_1(x_1),
      \ldots,
      f_k(x_k)
    }.
  \end{equation*}
\end{closingdefinition}

\begin{closingdefinition}[Tupling of functions]
  \label{def:function-tupling}
  Let
  \begin{equation*}
    f_i : X \to Y_i,
    \qquad
    i\in\iival{1}{k},
  \end{equation*}
  be functions having a common domain.
  Their \emph{tupling} is the function
  \begin{equation*}
    \ptuple{
      f_1,
      \ldots,
      f_k
    }
    :
    X
    \to
    Y_1
    \times
    \cdots
    \times
    Y_k
  \end{equation*}
  defined by
  \begin{equation*}
    \ptuple{
      f_1,
      \ldots,
      f_k
    }(x)
    \defeq
    \ptuple{
      f_1(x),
      \ldots,
      f_k(x)
    }.
  \end{equation*}
\end{closingdefinition}

\begin{closingremark}
  The product and tupling operations are distinct.
  In particular,
  \begin{equation*}
    f_1\times\cdots\times f_k
  \end{equation*}
  acts on a tuple of independent arguments, whereas
  \begin{equation*}
    \ptuple{f_1,\ldots,f_k}
  \end{equation*}
  evaluates all functions on the same argument.
\end{closingremark}

\begin{closingdefinition}[Function composition]
  \label{def:function-composition}
  Let
  \begin{equation*}
    f : X \to Y,
    \qquad
    g : Y \to Z.
  \end{equation*}
  The \emph{composition} of
  $f$
  followed by
  $g$
  is the function
  \begin{equation*}
    g \circ f
    :
    X
    \to
    Z
  \end{equation*}
  defined by
  \begin{equation*}
    (g\circ f)(x)
    \defeq
    g\big(f(x)\big).
  \end{equation*}

  More generally, for composable functions
  \begin{equation*}
    f_i
    :
    X_{i-1}
    \to
    X_i,
    \qquad
    i\in\iival{1}{k},
  \end{equation*}
  we write
  \begin{equation*}
    f_k
    \circ
    \cdots
    \circ
    f_1
    :
    X_0
    \to
    X_k
  \end{equation*}
  for their iterated composition, defined by
  \begin{equation*}
    \big(
      f_k
      \circ
      \cdots
      \circ
      f_1
    \big)(x)
    \defeq
    f_k
    \big(
      \cdots
      f_2
      \big(
        f_1(x)
      \big)
      \cdots
    \big).
  \end{equation*}
\end{closingdefinition}
\begin{closingremark}
  Function composition is associative whenever the corresponding functions
  are composable.
  Hence parentheses in iterated compositions are omitted.
\end{closingremark}

\begin{closingdefinition}[Closure of class of functions]
  \label{def:function-class-closure}
  For $\mathcal{P}$ a class of functions,
  its \emph{closure} $\functionclassclosure{\mathcal{P}}$ is the smallest class of functions such
  that:
  \begin{itemize}
    \item
      $\mathcal{P} \subseteq \functionclassclosure{\mathcal{P}}$;
    \item
      $\functionclassclosure{\mathcal{P}}$ is closed under function tupling:
      \begin{align*}
        \big\{ 
          \ptuple{f_1, \ldots, f_n} 
          \bigm| 
          f_1, \ldots, f_n \in \functionclassclosure{\mathcal{P}},\, 
          n \in \posnaturals 
        \big\}
        \,\subseteq\,
        \mathcal{F};
      \end{align*}
    \item
      $\functionclassclosure{\mathcal{P}}$ is closed under function composition:
      \begin{align*}
        \big\{ 
          f \circ g
          \bigm| 
          f : \reals^{m_1} \to \reals^{m} \ldots, f_n \in \functionclassclosure{\mathcal{P}},\, 
          n \in \posnaturals 
        \big\}
        \,\subseteq\,
        \functionclassclosure{\mathcal{P}}.
      \end{align*}
  \end{itemize}
\end{closingdefinition}

\subsubsection{Euclidean Functions}

\begin{closingdefinition}
  A \emph{function between finite-dimensional Euclidean spaces} 
  is a function 
  $f : X \to Y$ 
  where
  $X \subseteq \reals^n$
  and
  $Y \subseteq \reals^m$
  for some $n \in \naturals$ and some $m\in\posnaturals$.

  We also call $f$ a \emph{Euclidean function} or \emph{Euclidean map}. 

  The following terminology applies:
  \begin{itemize}
    \item
      $X$ is the \emph{input domain} or \emph{input space};
    \item
      $Y$ is the \emph{codomain} or \emph{output space};
    \item
      $\im(f) \defeq \{ f(x) \mid x \in X \}$ is the \emph{image};
    \item
      $\arityin{f} \defeq n$ is the \emph{input dimension} or \emph{input arity};
    \item
      $\arityout{f} \defeq m$ is the \emph{output dimension} or \emph{output arity}.
  \end{itemize}
\end{closingdefinition}

\begin{closingdefinition}
  A \emph{parametric Euclidean function} is a function
  \begin{equation*}
    f : X \times \Theta \to Y,
  \end{equation*}
  where
  \begin{equation*}
    X \subseteq \reals^n,
    \qquad
    \Theta \subseteq \reals^p,
    \qquad
    Y \subseteq \reals^m,
  \end{equation*}
  for some $n,p \in \naturals$ and some $m\in\posnaturals$.

  We typically write $f(x;\theta)$ in place of $f(x,\theta)$.

  The following terminology applies:
  \begin{itemize}
    \item
      $X$ is the \emph{input domain} or \emph{input space};
    \item
      $\Theta$ is the \emph{parameter domain} or \emph{parameter space};
    \item
      $Y$ is the \emph{codomain} or \emph{output space};

    \item
      $\arityin{f} \defeq n$ is the \emph{input dimension} or \emph{input arity};
    \item
      $\arityout{f} \defeq p$ is the \emph{parameter dimension} or \emph{parameter arity};
    \item
      $\arityout{f} \defeq m$ is the \emph{output dimension} or 
      \emph{output arity}.
  \end{itemize}

  For every $\theta\in\Theta$, the function
  \begin{equation*}
    \pinst{f}{\theta} : X \to Y,
    \qquad
    \pinst{f}{\theta}(x) \defeq f(x;\theta),
  \end{equation*}
  is called the \emph{instance of $f$ with parameter values $\theta$}.
\end{closingdefinition}

\begin{fact}
  We consider every parametric Euclidean function $f : X \times \Theta \to Y$ 
  to also be a Euclidean function.
  Indeed, if
  $X \subseteq \reals^n$,
  $\Theta \subseteq \reals^p$,
  and
  $Y \subseteq \reals^m$,
  then
  the joint input-parameter space $X \times \Theta$ can be identified with a subset of 
  $\reals^{n+p}$.
\end{fact}

\subsection{Sequences}
\label{subsec:sequences}

\subsubsection{Finite Sequences}

\begin{closingdefinition}[Finite sequences]
  Let $X$ be a set
  and
  let $n \in \naturals$.

  A \emph{finite sequence $x$ over $X$} written \emph{in explicit form} is an expression 
  \begin{align*}
    x_{i_1}, \ldots, x_{i_n}
  \end{align*}
  where
  \begin{itemize}
    \item
      $n \in \naturals$ is called the \emph{length} of $x$;
    \item
      $i_1, \ldots, i_n \in \naturals$ are indices satisfying $i_1 < \ldots < i_n$,
      and we call $I = \{ i_1, \ldots, i_n \}$ the index set of $x$
    \item
      for every $j \in \iival{1}{n}$,
      we have $x_{i_j} \in X$, called the \emph{$j$-th element} of $x$.
  \end{itemize}

  Such sequence $x$ can be equivalently written in \emph{indexed form} as 
  \begin{align*}
    (x_i)_{i \in I},
  \end{align*}
  and the following terminology applies:
  \begin{itemize}
    \item
      $x$ is an \emph{empty sequence} if $|x| = 0$;
    \item
      $x$ is a \emph{non-empty sequence} if $|x| > 0$.
  \end{itemize}
\end{closingdefinition}

\begin{closingfact}
  There exists a unique empty sequence, regardless of the choice of an underlying set $X$.
  Such sequence is therefore called \emph{the empty sequence}, and it is denoted by $\varepsilon$.
  Thus, instead of saying that a sequence $x$ is \emph{an empty sequence}, 
  we can say that $x$ is \emph{the empty sequence}.
\end{closingfact}

\begin{closingdefinition}[Sets of finite sequences]
  Let $X$ be a set,
  and
  let $N \in \posnaturals$.

  We define the following notation for sets of finite sequences:
  \begin{itemize}
    \item
      $\allseqs{X}$ 
      denotes 
      the set of all finite sequences over $X$;
    \item
      $\allnonemptyseqs{X}$ 
      denotes 
      the set of all finite \emph{non-empty} sequences over $X$;
    \item
      $X^N$ denotes 
      the set of all sequences over $X$ having length $N$;
    \item
      $\allseqs{X}[N]$ 
      denotes
      the set consisting of each sequence $x$ over $X$ 
      having length $|x| \leq N$;
    \item
      $\allnonemptyseqs{X}[N]$ 
      denotes
      the set consisting of each non-empty sequence $x$ over $X$ 
      having length $|x| \leq N$.
  \end{itemize}
\end{closingdefinition}

\begin{closingdefinition}[Slicing operator]
  For $x = x_{i_1}, \ldots, x_{i_n}$ a finite sequence over a set $X$,
  and 
  for $j,\ell \in \iival{1}{n}$ with $j \leq \ell$,
  we define 
  \begin{align*}
    \seq{x}{i_j}{i_\ell} \defeq x_{i_j}, x_{i_j+1},\ldots,x_{i_\ell} 
  \end{align*}
  which is itself a finite sequence over $X$ having length $\ell-j+1$.
\end{closingdefinition}

\subsubsection{Infinite Sequences}

\begin{closingdefinition}[Infinite sequences]
  Let $X$ be a set.

  An \emph{infinite sequence $x$ over $X$} written in \emph{explicit form} is an expression 
  \begin{align*}
    x_{i_1}, \ldots, x_{i_n}, \ldots 
  \end{align*}
  where:
  \begin{itemize}
    \item
      for every $j \in \posnaturals$, we have $i_j \in \naturals$;
    \item
      for every $j,\ell \in \posnaturals$, 
      we have that $j < \ell$ implies $i_j < i_\ell$;
    \item
      for every $j \in \posnaturals$,
      we have $x_{i_j} \in X$, called the \emph{$j$-th element} of $x$.
  \end{itemize}

  We call $I = \{ i_j \in \naturals \mid j \in \posnaturals \}$ the \emph{index set} of $x$.

  The above sequence $x$ can be equivalently written \emph{in indexed form} as
  \begin{align*}
    (x_{i_j})_{j \in \posnaturals},
  \end{align*}
  and also as
  \begin{align*}
    (x_i)_{i \in I}.
  \end{align*}
\end{closingdefinition}

\begin{closingdefinition}[Sets of finite sequences]
  Let $X$ be a set.

  We define the following notation for sets of infinite sequences:
  \begin{itemize}
    \item
      $\infseqs{X}$ 
      denotes the set of all infinite sequences over $X$;
    \item
      when $X$ is equipped with a norm $\|\cdot\|_X$,
      we write
      $\binfseqs{X}$  
      for the set of all bounded infinite sequences over $X$; namely
      $\binfseqs{X}$ consists of each sequence 
      $x = (x_i)_{i \in I} \in \infseqs{X}$
      satisfying
      \begin{align*}
        \sup_{i \in I} \| x_i \|_X \,\leq\, B_x
      \end{align*}
      for some constant $B_x \in \nonnegreals$.
  \end{itemize}
\end{closingdefinition}

\begin{closingdefinition}[Slicing operator]
For $x = (x_i)_{i \in I}$ an infinite sequence over a set $X$,
and for $j,\ell \in I$ satisfying $j \leq \ell$,
we define 
\begin{align*}
  \seq{x}{i}{\ell} \defeq x_i,x_{i+1},\ldots,x_\ell
\end{align*}
which is a finite sequence over $X$ having length $\ell-i+1$.
\end{closingdefinition}

\subsubsection{String Notation}

A \emph{string} is a sequence, with the only notational difference that commas are omitted when it
is written in explict form.
For instance, 
\begin{gather*}
  x_1, x_2 \ldots, x_n \qquad\leadsto\qquad x_1 x_2 \cdots x_n
\end{gather*}
and
\begin{gather*}
  x_1, x_2 \ldots, x_n, \ldots \qquad\leadsto\qquad x_1 x_2 \cdots x_n \cdots
\end{gather*}
for finite and infinite strings, respectively.

\subsection{Vectors and Matrices}

Let $\mathbb{D}$ be a set.

For $n \in \naturals_{>0}$, a \emph{vector of dimension $n$ over $\mathbb{D}$}
is an element
$x \in \mathbb{D}^n$.
We write vectors as
\begin{align*}
  x
  =
  (x_1,\ldots,x_n)
  =
  \bigl(x_1,\ldots,x_n\bigr),
\end{align*}
where, for every $i \in \{1,\ldots,n\}$,
\begin{align*}
  \getentry{x}{i} = x_i \in \mathbb{D}
\end{align*}
is called the \emph{$i$-th component} of $x$.

For $m,n \in \naturals_{>0}$, a \emph{matrix of dimension $m \times n$ over $\mathbb{D}$}
is an element
$A \in \mathbb{D}^{m \times n}$.
We write matrices as
\begin{align*}
  A
  =
  \begin{pmatrix}
    a_{1,1} & \cdots & a_{1,n}
    \\[.3em]
    \vdots & \ddots & \vdots
    \\[.3em]
    a_{m,1} & \cdots & a_{m,n}
  \end{pmatrix},
\end{align*}
where, for every
$i \in \iival{1}{m}$
and
$j \in \iival{1}{n}$,
\begin{align*}
  \getentry{A}{i,j} = a_{i,j} \in \mathbb{D}
\end{align*}
is called the \emph{$(i,j)$-th entry} of $A$.
For every $i \in \iival{1}{m}$, we denote the \emph{$i$-th row} of $A$ by
\begin{align*}
  \getrow{A}{i}
  \,\defeq\,
  \begin{pmatrix}
    a_{i,1}
    &
    \ldots
    &
    a_{i,n}
  \end{pmatrix}
  \,\in \mathbb{D}^{1 \times n}
\end{align*}
We use \emph{entry} as a generic term for both a component of a vector
and an entry of a matrix.

Whenever vectors are used in a matrix setting, we identify them with column vectors.
Thus, for
$x = (x_1,\ldots,x_n) \in \mathbb{D}^n$,
we identify
\begin{align*}
  x
  \equiv
  \begin{pmatrix}
    x_1\\
    \vdots\\
    x_n
  \end{pmatrix}
  \in \mathbb{D}^{n \times 1}.
\end{align*}
For instance, matrix--vector products such as
$A \cdot x$, for $A \in \mathbb{D}^{m \times n}$,
are understood with $x$ interpreted as a column vector.

\begin{definition}[Set of entries]
  Let $\mathbb{D}$ be a domain.

  For a vector $x \in \mathbb{D}^n$, we define its \emph{set of entries} as
  \begin{equation*}
    \entries{x}
    \defeq
    \left\{
      \getentry{x}{i}
      \,\middle|\,
      i \in \iival{1}{n}
    \right\}.
  \end{equation*}

  For a matrix $A \in \mathbb{D}^{m \times n}$, we define its
  \emph{set of entries} as
  \begin{equation*}
    \entries{A}
    \defeq
    \left\{
      \getentry{A}{i,j}
      \,\middle|\,
      i \in \iival{1}{m},
      \ j \in \iival{1}{n}
    \right\}.
  \end{equation*}
\end{definition}

\begin{closingdefinition}[Deterministic row-substochastic matrix]
  \label{def:deterministic-row-substochastic}
  A matrix $A\in\reals^{n\times n}$ is
  \emph{deterministic row-substochastic} if
  \begin{align*}
    A\in\{0,1\}^{n\times n},
    \qquad
    A \cdot 1_n \,\leq\, 1_n. 
  \end{align*}
  Equivalently, every row is either zero or a standard basis row.
  In particular, $A \geq 0_{n,n}$ and $\|A\|_\infty\leq 1$.
\end{closingdefinition}

\subsubsection{Norms}

We introduce norms of vectors and matrices over $\reals$.

\begin{closingdefinition}
The \emph{Euclidean norm} of a vector $x \in \reals^n$ is
\begin{align*}
  \|x\|
  \defeq
  \sqrt{
    \sum_{i=1}^n
    \bigl(\getentry{x}{i}\bigr)^2
  }.
\end{align*}
\end{closingdefinition}

\begin{closingdefinition}
The \emph{maximum norm} of a vector $x \in \reals^n$ is
\begin{align*}
  \|x\|_\infty
  \defeq
  \|x\|_\mathrm{max}
  \defeq
  \max_{1 \leq i \leq n}
  \big|\getentry{x}{i}\big|.
\end{align*}
\end{closingdefinition}

\begin{closingfact}[Relationship between Euclidean and maximum norms]
  The following inequality holds:
  \begin{align*}
    \|x\|
    \leq
    \sqrt{n} \cdot \|x\|_\infty
    =
    \sqrt{n} \cdot \|x\|_\mathrm{max}
    \qquad
    \forall x \in \reals^n.
  \end{align*}
\end{closingfact}

\begin{closingdefinition}[Spectral norm]
  \label{def:spectral-norm}
  The \emph{spectral norm} of a matrix $A\in\reals^{m\times n}$ is
  \begin{equation*}
    \|A\|_2
    \defeq
    \sup_{x\in\reals^n\setminus\{0\}}
    \frac{
      \|A\cdot x\|
    }{
      \|x\|
    }.
  \end{equation*}
  When we refer to the \emph{norm} of a matrix, 
  we mean its \emph{spectral norm};
  and 
  we may write $\|A\|$ in place of $\|A\|_2$.
\end{closingdefinition}

\begin{closingdefinition}[Frobenius norm]
  \label{def:matrix-frobenius-norm}
  The \emph{Frobenius norm} of a matrix $A\in\reals^{m\times n}$ is
  \begin{equation*}
    \|A\|_{\mathrm{F}}
    \defeq
    \sqrt{
      \sum_{i=1}^{m}
      \sum_{j=1}^{n}
      \bigl(
        \getentry{A}{i,j}
      \bigr)^2
    }.
  \end{equation*}
\end{closingdefinition}

\begin{closingfact}[Relationship between Frobenius and spectral norms]
  \label{fact:frobenius-spectral-norms}
  For every
  $A\in\reals^{m\times n}$,
  \begin{equation*}
    \|A\|
    \,\leq\,
    \|A\|_{\mathrm{F}}
    \,\leq\,
    \sqrt{
      \operatorname{rank}(A)
    }
    \cdot
    \|A\|
    \,\leq\,
    \sqrt{
      \min\{m,n\}
    }
    \cdot
    \|A\|,
  \end{equation*}
  where
  $\|A\|$
  denotes the spectral norm of $A$.
\end{closingfact}

\begin{closingdefinition}[Maximum and row norms]
  The \emph{maximum norm} of a matrix $A \in \reals^{m \times n}$ is
  \begin{align*}
    \|A\|_\mathrm{max}
    \defeq
    \max_{1 \leq i \leq m}
    \max_{1 \leq j \leq n}
    \big|\getentry{A}{i,j}\big|.
  \end{align*}
  The \emph{row-wise maximum Euclidean norm}, or simply \emph{maximum norm}, 
  of a matrix $A \in \reals^{m \times n}$ is
  \begin{align*}
    \|A\|_\mathrm{row}
    \defeq
    \max_{i \in \iival{1}{m}}
    \|\getrow{A}{i}\|.
  \end{align*}
\end{closingdefinition}

\subsection{Notation for Derivatives}
\label{subsec:derivative-notation}

We use
$\dv$
to denote the derivative operator on functions between finite-dimensional Euclidean spaces.

Let
\begin{equation*}
  f
  :
  \reals^{d_\mathrm{in}}
  \to
  \reals^{d_\mathrm{out}}
\end{equation*}
be differentiable.
We denote its derivative by
\begin{equation*}
  \dvop{f}
  :
  \reals^{d_\mathrm{in}}
  \to
  \reals^{d_\mathrm{out}\times d_\mathrm{in}},
\end{equation*}
where
\begin{equation*}
  \dvop{f}(x)
\end{equation*}
is the Jacobian matrix of
$f$
at
$x\in\reals^{d_\mathrm{in}}$.

\subsubsection{Positional partial derivatives}
\label{subsubsec:positional-partial-derivatives}

Let us consider a Euclidean function 
$f : \reals^{d_\mathrm{in}} \to \reals^{d_\mathrm{out}}$.

Let $x\in\reals^{d_\mathrm{in}}$ be an input of $f$, which we write as
\begin{equation*}
  x
  =
  \ptuple{
    x_1,
    \ldots,
    x_{d_\mathrm{in}}
  }.
\end{equation*}

For every
$i\in\iival{1}{d_\mathrm{in}}$,
we denote by
\begin{equation*}
  \pdvop{f}{i}
\end{equation*}
the
\emph{partial derivative of $f$ with respect to its $i$-th input coordinate}.
Thus,
\begin{equation*}
  \pdvop{f}{i}
  :
  \reals^{d_\mathrm{in}}
  \to
  \reals^{d_\mathrm{out}},
\end{equation*}
and
\begin{equation*}
  \pdvop{f}{i}(x)
\end{equation*}
is the $i$-th column of
$\dvop{f}(x)$,
identified with a vector in
$\reals^{d_\mathrm{out}}$.

Accordingly,
\begin{equation*}
  \dvop{f}(x)
  =
  \begin{pmatrix}
    \pdvop{f}{1}(x)
    &
    \cdots
    &
    \pdvop{f}{d_\mathrm{in}}(x)
  \end{pmatrix}.
\end{equation*}

\subsubsection{Blockwise positional partial derivatives}

Let
$f : X \to Y$
be a Euclidean function where
\begin{equation*}
  X
  \,=\,
  \reals^{d_1}
  \times
  \cdots
  \times
  \reals^{d_k},
  \qquad\quad
  Y
  =
  \reals^{d_\mathrm{out}},
\end{equation*}
and let
$x\in X$
be an input of
$f$.

We canonically identify
\begin{equation*}
  X
  \cong
  \reals^{d_1+\cdots+d_k},
\end{equation*}
and accordingly write
\begin{equation*}
  x
  =
  \ptuple{
    x_1,
    \ldots,
    x_{d_1+\cdots+d_k}
  }.
\end{equation*}

For every
$i\in\iival{1}{k}$,
let
\begin{equation*}
  b_i
  \defeq
  \sum_{r=1}^{i-1} d_r,
\end{equation*}
and define the
\emph{$i$-th block-index vector}
by
\begin{equation*}
  \blockindex{i}
  \defeq
  \ptuple{
    b_i+1,
    \ldots,
    b_i+d_i
  }.
\end{equation*}
Thus,
$\blockindex{i}$
is the ordered vector of scalar input-coordinate indices belonging to the
$i$-th input block.

We denote by
\begin{equation*}
  \inputblock{x}{i}
  \defeq
  \ptuple{
    x_{b_i+1},
    \ldots,
    x_{b_i+d_i}
  }
  \in
  \reals^{d_i}
\end{equation*}
the subvector of
$x$
whose coordinates are indexed by
$\blockindex{i}$,
and call it the
\emph{$i$-th input block}
of
$x$.

For every
$i\in\iival{1}{k}$,
we denote by
\begin{equation*}
  \bdvop{f}{i}
\end{equation*}
the
\emph{partial derivative of $f$ with respect to its $i$-th input block}.
Thus,
\begin{equation*}
  \bdvop{f}{i}
  :
  X
  \to
  \reals^{d_\mathrm{out}\times d_i}.
\end{equation*}

More precisely, for every
$x\in X$,
we define
\begin{equation*}
  \bdvop{f}{i}(x)
  \defeq
  \begin{pmatrix}
    \pdvop{f}{b_i+1}(x)
    &
    \cdots
    &
    \pdvop{f}{b_i+d_i}(x)
  \end{pmatrix},
\end{equation*}
where
$\pdvop{f}{j}$
denotes the partial derivative of
$f$
with respect to its
$j$-th scalar input coordinate, as defined in
\S\ref{subsubsec:positional-partial-derivatives}.

Equivalently,
$\bdvop{f}{i}(x)$
is the submatrix of
$\dvop{f}(x)$
formed, in their original order, by the columns whose indices belong to
$\blockindex{i}$.

Consequently,
\begin{equation*}
  \dvop{f}(x)
  =
  \begin{pmatrix}
    \bdvop{f}{1}(x)
    &
    \cdots
    &
    \bdvop{f}{k}(x)
  \end{pmatrix}.
\end{equation*}

\subsubsection{Named partial derivatives}
\label{subsubsec:named-partial-derivatives}

Let us consider a Euclidean function
\begin{equation*}
  f
  :
  \reals^{d_\mathrm{in}}
  \to
  \reals^{d_\mathrm{out}},
\end{equation*}
and let
\begin{equation*}
  x
  =
  \ptuple{
    x_1,
    \ldots,
    x_{d_\mathrm{in}}
  }
  \in
  \reals^{d_\mathrm{in}}.
\end{equation*}

When evaluating a positional partial derivative at
$x$,
we also use a named partial-derivative notation in which the subscript
identifies the corresponding scalar input coordinate of the evaluation point.

More precisely, for every
$i\in\iival{1}{d_\mathrm{in}}$,
we write
\begin{equation*}
  \pdvop{f}{x_i}
  \big(
    x_1,
    \ldots,
    x_{d_\mathrm{in}}
  \big)
  \defeq
  \bdvop{f}{i}
  \big(
    x_1,
    \ldots,
    x_{d_\mathrm{in}}
  \big).
\end{equation*}
Thus,
\begin{equation*}
  \pdvop{f}{x_i}
  \big(
    x_1,
    \ldots,
    x_{d_\mathrm{in}}
  \big)
  \in
  \reals^{d_\mathrm{out}}
\end{equation*}
denotes the partial derivative of
$f$
at
$x$
with respect to the scalar input coordinate
$x_i$.

Hence,
$\bdvop{f}{i}$
is the positional partial-derivative function associated with the
$i$-th scalar input coordinate,
whereas
\begin{equation*}
  \pdvop{f}{x_i}
  \big(
    x_1,
    \ldots,
    x_{d_\mathrm{in}}
  \big)
\end{equation*}
is notation for that partial derivative evaluated at
$x$
and named according to the corresponding scalar input coordinate.

When a scalar input coordinate at the evaluation point is itself given by the
value of a function or expression, we often identify that coordinate by the
corresponding function or expression symbol rather than by its full value.

More precisely, suppose that
\begin{equation*}
  f
  :
  \reals^2
  \to
  \reals^{d_\mathrm{out}},
\end{equation*}
and consider an evaluation point of the form
\begin{equation*}
  \ptuple{
    g(x),
    y
  },
\end{equation*}
where
$g(x),y\in\reals$.
The fully explicit named partial-derivative notation for the first scalar input
coordinate is
\begin{equation*}
  \pdvop{f}{g(x)}
  \big(
    g(x),
    y
  \big).
\end{equation*}
When there is no ambiguity, we abbreviate the name of this scalar input
coordinate by the symbol of the function producing it, and write
\begin{equation*}
  \pdvop{f}{g}
  \big(
    g(x),
    y
  \big)
  \defeq
  \pdvop{f}{g(x)}
  \big(
    g(x),
    y
  \big).
\end{equation*}

More generally, named subscripts may abbreviate the expression defining the
corresponding scalar input coordinate whenever the association between the name
and the coordinate at the evaluation point is clear from context.

\subsubsection{Blockwise named partial derivatives}
\label{subsubsec:blockwise-named-partial-derivatives}

Let
$f : X \to Y$
be a Euclidean function where
\begin{equation*}
  X
  =
  \reals^{d_1}
  \times
  \cdots
  \times
  \reals^{d_k},
  \qquad
  Y
  =
  \reals^{d_\mathrm{out}},
\end{equation*}
and let
\begin{equation*}
  x
  =
  \ptuple{
    x_1,
    \ldots,
    x_{d_1+\cdots+d_k}
  }
  \in X,
\end{equation*}
where, as in the preceding subsection, we canonically identify
\begin{equation*}
  X
  \cong
  \reals^{d_1+\cdots+d_k}.
\end{equation*}

For every
$i\in\iival{1}{k}$,
recall that
$\blockindex{i}$
is the $i$-th block-index vector and
\begin{equation*}
  \inputblock{x}{i}
  \in
  \reals^{d_i}
\end{equation*}
is the corresponding
$i$-th input block of
$x$.

When evaluating a blockwise positional partial derivative at
$x$,
we also use a named partial-derivative notation in which the subscript
identifies the corresponding input block of the evaluation point.

More precisely, for every
$i\in\iival{1}{k}$,
we write
\begin{equation*}
  \pdvop{f}{\inputblock{x}{i}}(x)
  \defeq
  \bdvop{f}{i}(x).
\end{equation*}
Thus,
\begin{equation*}
  \pdvop{f}{\inputblock{x}{i}}(x)
  \in
  \reals^{d_\mathrm{out}\times d_i}
\end{equation*}
denotes the Jacobian block of
$f$
at
$x$
corresponding to variations of the input block
$\inputblock{x}{i}$,
that is, to variations of the scalar input coordinates whose indices belong to
$\blockindex{i}$.

Hence,
$\bdvop{f}{i}$
is the blockwise positional partial-derivative function associated with the
$i$-th input block,
whereas
\begin{equation*}
  \pdvop{f}{\inputblock{x}{i}}(x)
\end{equation*}
is notation for that partial derivative evaluated at
$x$
and named according to the corresponding input block.

For example, if
\begin{equation*}
  f
  :
  \reals^{d_x}
  \times
  \reals^{d_\theta}
  \to
  \reals^{d_\mathrm{out}},
\end{equation*}
then, at a point
\begin{equation*}
  \ptuple{x,\theta},
\end{equation*}
we write
\begin{equation*}
  \pdvop{f}{x}(x,\theta)
  \in
  \reals^{d_\mathrm{out}\times d_x}
  \qquad\text{and}\qquad
  \pdvop{f}{\theta}(x,\theta)
  \in
  \reals^{d_\mathrm{out}\times d_\theta}
\end{equation*}
for the Jacobian blocks corresponding to the
$x$
and
$\theta$
input blocks, respectively.

When an input block at the evaluation point is itself given by the value of a
function or expression, we often identify that input block by the corresponding
function or expression symbol rather than by its full value.

More precisely, suppose that
\begin{equation*}
  f
  :
  \reals^{d_1}
  \times
  \reals^{d_2}
  \to
  \reals^{d_\mathrm{out}},
\end{equation*}
and consider an evaluation point of the form
\begin{equation*}
  \ptuple{
    g(x),
    y
  },
\end{equation*}
where
\begin{equation*}
  g(x)\in\reals^{d_1},
  \qquad
  y\in\reals^{d_2}.
\end{equation*}
The fully explicit named partial-derivative notation for the first input block is
\begin{equation*}
  \pdvop{f}{g(x)}
  \big(
    g(x),
    y
  \big).
\end{equation*}
When there is no ambiguity, we abbreviate the name of this input block by the
symbol of the function producing it, and write
\begin{equation*}
  \pdvop{f}{g}
  \big(
    g(x),
    y
  \big)
  \defeq
  \pdvop{f}{g(x)}
  \big(
    g(x),
    y
  \big).
\end{equation*}

More generally, named subscripts may abbreviate the expression defining the
corresponding input block whenever the association between the name and the
input block at the evaluation point is clear from context.

\subsection{Partitions of a Set}
\label{subsec:partitions}

\begin{definition}[Partition]
  Let
  $X$
  be a set.
  A
  \emph{partition}
  of
  $X$
  is a set
  $\genericpartition$
  of nonempty subsets of
  $X$
  such that
  \begin{equation*}
    \bigcup_{P\in\genericpartition} P
    =
    X,
  \end{equation*}
  and
  \begin{equation*}
    P\cap P'
    =
    \emptyset
    \qquad
    \forall
    P,P'\in \genericpartition
    \text{ with }
    P\neq P'.
  \end{equation*}
  The elements of
  $\genericpartition$
  are called 
  the \emph{regions} of the partition $\genericpartition$,
  or 
  \emph{$\genericpartition$-regions}.
\end{definition}

\begin{definition}[Refinement of a partition]
  Let
  $\genericpartition$
  and
  $\mathcal{Q}$
  be partitions of a set
  $X$.
  We say that
  $\mathcal{Q}$
  is a
  \emph{refinement}
  of
  $\genericpartition$, denoted by
  \begin{align*}
    \mathcal{Q} \preceq \genericpartition,
  \end{align*}
  if, for every
  $Q\in\mathcal{Q}$,
  there exists
  $P\in\genericpartition$
  such that
  \begin{equation*}
    Q
    \subseteq
    P,
  \end{equation*}
  where $P$ is called the corresponding $\genericpartition$-region for $Q$,
  which we write as 
  $\genericpartition[Q] \defeq P$,
  and as 
  $\genericpartition[q] \defeq P$ when $Q = \{ q \}$.
\end{definition}

\subsection{Affine Functions}

\begin{definition}[Affine and linear functions]
  \label{def:affine-linear-function}
  Let 
  $X \subseteq \reals^d$ and 
  let $Y \subseteq \reals^m$.
  A function $g: X \to Y$
  is
  \emph{affine}
  if there exist
  a matrix
  $A\in\reals^{m\times d}$
  and
  a vector
  $b\in\reals^m$
  such that
  \begin{sidealign*}
    \quanteq{
      g(x)
      \,=\,
      A\cdot x+b
    }{
      \forall x \in X;
    }
  \end{sidealign*}
  in this case,
  we refer to $\tuple{A,b}$ as \emph{affine coefficients} for $g$.

  An affine function $g$ 
  is
  \emph{linear}
  if it admits affine coefficients of the form
  $\tuple{A,0_m}$;
  equivalently,
  \begin{sidealign*}
    \quanteq{
      g(x)
      \,=\,
      A\cdot x
    }{
      \forall x\in\reals^d.
    }
  \end{sidealign*}
\end{definition}

\subsection{Piecewise Representations of Functions}

\begin{closingdefinition}[Piecewise representation]
  \label{def:piecewise-representation}
  Let $X$ and $Y$ be spaces.

  \par\medskip
  \textbf{Syntax.}~
  A \emph{piecewise representation} on $(X,Y)$
  is a set
  \begin{equation*}
    \gamma 
    =
    \big\{ \tuple{P,f_P} \big\}_{P \in \partitionstyle{P}},
  \end{equation*}
  such that
  \begin{itemize}
    \item
      $\partitionstyle{P}$ is a finite partition of $X$,
      and
    \item
      $f_P : P \to Y$ for every $P\in\partitionstyle{P}$;
  \end{itemize}
  the following terminology applies:
  \begin{itemize}
    \item
      $\partitionstyle{P}$ 
      is 
      \emph{the (input-)partitition of $\gamma$};
    \item
      every $P \in \partitionstyle{P}$ 
      is 
      \emph{a region of $\gamma$};
    \item
      every function $f_P$ for $P \in \partitionstyle{P}$ 
      is 
      \emph{the piece of $\gamma$ on the region $P$}.
  \end{itemize}

  \par\medskip
  \textbf{Semantics.}~
  The \emph{function represented by $\gamma$} is%
  \footnote{%
    Recall the notation $\partition{P}[x]$ for $x \in X$ introduced in~\S\ref{subsec:partitions}.%
  }
  \begin{align*}
    \Repval{\gamma} : X \to Y,
    \qquad\qquad
    \Repval{\gamma}(x)
    \defeq
    f_{\partition{P}[x]}(x)
    \qquad
    \forall x \in X.
  \end{align*}
  Then,
  for $f$ a function,
  we say that $\gamma$ 
  is 
  a \emph{piecewise representation of $f$} 
  if 
  $\Repval{\gamma} = f$.
  In this case,
  the following terminology applies:
  \begin{itemize}
    \item
      $\partitionstyle{P}$ 
      is 
      \emph{the (input-)partitition of $f$ when represented by $\gamma$};
    \item
      every $P \in \partitionstyle{P}$ 
      is 
      \emph{a region of $f$ when represented by $\gamma$};
    \item
      every function $f_P$ for $P \in \partitionstyle{P}$ 
      is 
      \emph{the piece of $f$ on the region $P$ when represented by $\gamma$}.
  \end{itemize}
\end{closingdefinition}

\subsection{Piecewise-Affine Functions}

\begin{definition}[Piecewise-affine representation]
  \label{def:piecewise-affine-representation}
  A \emph{piecewise representation}
  \begin{equation*}
    \gamma 
    =
    \big\{ \tuple{P,f_P} \big\}_{P \in \partitionstyle{P}},
  \end{equation*}
  is affine if,
  for every $P \in \partitionstyle{P}$,
  the piece $f_P$ is an affine function.
\end{definition}

\begin{definition}[Piecewise-affine function]
  \label{def:piecewise-affine-function}
  A function $f$ is \emph{piecewise affine} 
  if there exists a piecewise-affine representation $\gamma$ that represents $f$.
\end{definition}

\begin{closingdefinition}[Piecewise-affine coefficient-based representation]
  \label{definition:piecewise-affine-coefficients}
  Let $X \subseteq \reals^n$ and $Y \subseteq \reals^m$.

  \par\medskip
  \textbf{Syntax.}~
  A \emph{piecewise-affine coefficient-based representation} on $(X,Y)$
  is a set
  \begin{equation*}
    \gamma 
    =
    \big\{ \tuple{P,A_P,b_P} \big\}_{P \in \partitionstyle{P}},
  \end{equation*}
  such that:
  (i)~$\partitionstyle{P}$ is a finite partition of $X$;
  and
  (ii)~$A_P\in\reals^{m\times d}$ and $b_P\in\reals^m$, for every $P\in\partitionstyle{P}$.
  The following terminology applies:
  \begin{itemize}
    \item
      $\partitionstyle{P}$ 
      is 
      \emph{the (input-)partitition of $\gamma$};
    \item
      every $P \in \partitionstyle{P}$ 
      is 
      \emph{a region of $\gamma$};
    \item
      $\tuple{A_P,b_P}$ for $P \in \partitionstyle{P}$ 
      are
      \emph{the affine coefficients of $\gamma$ on the region $P$}.
  \end{itemize}

  \par\medskip
  \textbf{Semantics.}~
  The \emph{function represented by $\gamma$} is%
  \footnote{%
    Recall the notation $\partition{P}[x]$ for $x \in X$ introduced in~\S\ref{subsec:partitions}.%
  }
  \begin{align*}
    \Repval{\gamma}(x)
    \defeq
    A_{\partition{P}{x}} \cdot x
    +
    b_{\partition{P}{x}}
    \qquad\quad
    \forall x \in X.
  \end{align*}
\end{closingdefinition}

\begin{closingfact}
  \label{fact:piecewise-affine-coefficient-representation}
  We state the following two facts on the interchangeability of 
  piecewise-affine representations
  and 
  their coefficient-based counterpart.

  \par\medskip
  \textbf{Fact~\ref{fact:piecewise-affine-coefficient-representation}.1.}~
  For every piecewise representation $\gamma$ on $(X,Y)$ with
  \begin{equation*}
    \gamma
    =
    \big\{
      \tuple{P,f_P}
    \big\}_{P\in\partitionstyle{P}},
    \quad
    X\subseteq\reals^d,
    \quad 
    Y\subseteq\reals^m,
  \end{equation*}
  the following two claims are equivalent:
  \begin{itemize}
    \item
      $\gamma$ is piecewise affine;
    \item
      there exists a piecewise-affine coefficient-based representation $\hat{\gamma}$ on $(X,Y)$ 
      such that
    \begin{equation*}
      \hat{\gamma} = \big\{ \tuple{P,A_P,b_P} \big\}_{P\in\partitionstyle{P}}
      \qquad\text{ with }\qquad
      f_P(x) \,=\, A_P\cdot x + b_P
      \qquad
      \forall P\in\partitionstyle{P},\,
      \forall x\in X;
    \end{equation*}
    in particular,
    we have $\Repval{\gamma} = \Repval{\hat{\gamma}}$.
  \end{itemize}

  \par\medskip
  \textbf{Fact~\ref{fact:piecewise-affine-coefficient-representation}.2.}~
  For every piecewise-affine coefficient-based representation
  \begin{equation*}
    \hat{\gamma}
    =
    \big\{
      \tuple{P,A_P,b_P}
    \big\}_{P\in\partitionstyle{P}},
  \end{equation*}
  the piecewise-affine representation
  \begin{equation*}
    \gamma
    \defeq
    \big\{
      \tuple{P,f_P}
    \big\}_{P\in\partitionstyle{P}}
    \qquad\text{ having }\qquad
    f_P(x) \,\defeq\, A_P\cdot x + b_P
    \qquad
    \forall P\in\partitionstyle{P},\,
    \forall x\in X
  \end{equation*}
  satisfies
  \begin{align*}
    \Repval{\gamma} = \Repval{\hat{\gamma}}.
  \end{align*}
  We refer to $\gamma$ as 
  the \emph{(canonical) piecewise-affine representation} induced by $\hat{\gamma}$.
\end{closingfact}

\begin{definition}[Piecewise-linear representation]
  \label{def:piecewise-linear-representation}
  A piecewise-affine representation
  \begin{equation*}
    \gamma 
    =
    \big\{ \tuple{P,f_P} \big\}_{P \in \partitionstyle{P}},
  \end{equation*}
  is
  \emph{piecewise linear}
  if, for every $P \in \partitionstyle{P}$, 
  the piece $f_P$ is a linear map.
\end{definition}

\begin{closingfact}
  \label{fact:piecewise-linear-coefficient-representation}
  We state the following two facts on the interchangeability of 
  piecewise-linear representations
  and 
  their coefficient-based counterpart.

  \par\medskip
  \textbf{Fact~\ref{fact:piecewise-linear-coefficient-representation}.1.}~
  For every $\gamma$ a piecewise representation on $(X,Y)$ with
  \begin{equation*}
    \gamma
    =
    \big\{
      \tuple{P,f_P}
    \big\}_{P\in\partitionstyle{P}},
    \quad
    X\subseteq\reals^d,
    \quad
    Y\subseteq\reals^m,
  \end{equation*}
  the following two claims are equivalent:
  \begin{itemize}
    \item
      $\gamma$ is piecewise linear;
    \item
      there exists a piecewise-affine coefficient-based representation
      $\hat{\gamma}$ on $(X,Y)$
      such that
      \begin{equation*}
        \hat{\gamma}
        =
        \big\{
          \tuple{P,A_P,0_m}
        \big\}_{P\in\partitionstyle{P}}
        \qquad\text{ with }\qquad
        f_P(x)
        \,=\,
        A_P\cdot x
        \qquad
        \forall P\in\partitionstyle{P},\,
        \forall x\in X;
      \end{equation*}
      in particular,
      we have $\Repval{\gamma} = \Repval{\hat{\gamma}}$.
  \end{itemize}

  \par\medskip
  \textbf{Fact~\ref{fact:piecewise-linear-coefficient-representation}.2.}~
  For every piecewise-affine coefficient-based representation
  \begin{equation*}
    \hat{\gamma}
    =
    \big\{
      \tuple{P,A_P,0_m}
    \big\}_{P\in\partitionstyle{P}},
  \end{equation*}
  the piecewise-linear representation
  \begin{equation*}
    \gamma
    \defeq
    \big\{
      \tuple{P,f_P}
    \big\}_{P\in\partitionstyle{P}}
    \qquad\text{ having }\qquad
    f_P(x)
    \,\defeq\,
    A_P\cdot x
    \qquad
    \forall P\in\partitionstyle{P},\,
    \forall x\in X
  \end{equation*}
  satisfies
  \begin{align*}
    \Repval{\gamma}
    =
    \Repval{\hat{\gamma}}.
  \end{align*}
  We refer to
  $\gamma$
  as the
  \emph{(canonical) piecewise-linear representation}
  induced by
  $\hat{\gamma}$.
\end{closingfact}

\begin{closingdefinition}[Partially-affine piecewise representations and functions]
  \label{definition:piecewise-uniformly-affine}
  Let
  $X\subseteq\reals^n$,
  $U\subseteq\reals^d$,
  and
  $Y\subseteq\reals^m$.

  A piecewise representation
  \begin{equation*}
    \gamma
    =
    \big\{
      \tuple{P,f_P}
    \big\}_{P\in\partitionstyle{P}}
  \end{equation*}
  on
  $(X\times U,Y)$
  is
  \emph{affine in the first argument}
  if, for every
  $P\in\partitionstyle{P}$,
  there exist a matrix
  \begin{equation*}
    A_P\in\reals^{m\times n}
  \end{equation*}
  and a function
  \begin{equation*}
    b_P:U\to\reals^m
  \end{equation*}
  such that
  \begin{sidealign*}
    \quanteq{
      f_P(x,u)
      \,=\,
      A_P\cdot x+b_P(u)
    }{
      \forall (x,u)\in X\times U.
    }
  \end{sidealign*}

  In this case,
  we call
  $\tuple{A_P,b_P}$
  a
  \emph{coefficient-offset pair of $\gamma$ on the region $P$};
  in particular,
  $A_P$
  is the
  \emph{linear coefficient}
  and
  $b_P$
  is the
  \emph{offset function}
  of
  $\gamma$
  on
  $P$.

  We say that a function
  \begin{equation*}
    f:X\times U\to Y
  \end{equation*}
  is
  \emph{piecewise affine in its first argument}
  if it is represented by some piecewise representation that is affine
  in the first argument.
\end{closingdefinition}

\begin{closingdefinition}[Partially-affine coefficient-based piecewise representation]
  \label{definition:piecewise-uniformly-affine-coefficients}
  Let
  $X\subseteq\reals^n$,
  $U\subseteq\reals^d$,
  and
  $Y\subseteq\reals^m$.

  \par\medskip
  \textbf{Syntax.}~
  A
  \emph{coefficient-based piecewise representation on $(X,U,Y)$ that is affine
  in the first argument}
  is a set
  \begin{equation*}
    \hat{\gamma}
    =
    \big\{
      \tuple{P,A_P,b_P}
    \big\}_{P\in\partitionstyle{P}},
  \end{equation*}
  such that:
  \begin{enumerate}
    \item
      $\partitionstyle{P}$
      is a finite partition of
      $X\times U$;
    \item
      $A_P\in\reals^{m\times n}$
      for every
      $P\in\partitionstyle{P}$;
    \item
      $b_P:U\to\reals^m$
      for every
      $P\in\partitionstyle{P}$.
  \end{enumerate}

  The following terminology applies:
  \begin{itemize}
    \item
      $\partitionstyle{P}$
      is
      \emph{the (input-)partition of $\hat{\gamma}$};
    \item
      every
      $P\in\partitionstyle{P}$
      is
      \emph{a region of $\hat{\gamma}$};
    \item
      $\tuple{A_P,b_P}$
      is
      \emph{the coefficient-offset pair of $\hat{\gamma}$ on the region $P$};
    \item
      $A_P$
      is
      \emph{the linear coefficient of $\hat{\gamma}$ on the region $P$};
    \item
      $b_P$
      is
      \emph{the offset function of $\hat{\gamma}$ on the region $P$}.
  \end{itemize}

  \par\medskip
  \textbf{Semantics.}~
  The
  \emph{function represented by $\hat{\gamma}$}
  is
  \begin{align*}
    \Repval{\hat{\gamma}}(x,u)
    \defeq
    A_{\partition{P}{(x,u)}}\cdot x
    +
    b_{\partition{P}{(x,u)}}(u)
    \qquad\quad
    \forall (x,u)\in X\times U.
  \end{align*}
\end{closingdefinition}

\begin{closingfact}
  \label{fact:piecewise-uniformly-affine-coefficient-representation}
  We state the following two facts on the interchangeability of
  piecewise representations that are affine in the first argument
  and
  their coefficient-based counterparts.

  \par\medskip
  \textbf{Fact~\ref{fact:piecewise-uniformly-affine-coefficient-representation}.1.}~
  For every piecewise representation
  $\gamma$
  on
  $(X\times U,Y)$
  with
  \begin{equation*}
    \gamma
    =
    \big\{
      \tuple{P,f_P}
    \big\}_{P\in\partitionstyle{P}},
    \quad
    X\subseteq\reals^n,
    \quad
    U\subseteq\reals^d,
    \quad
    Y\subseteq\reals^m,
  \end{equation*}
  the following two claims are equivalent:
  \begin{itemize}
    \item
      $\gamma$
      is affine in the first argument;

    \item
      there exists a coefficient-based piecewise representation
      $\hat{\gamma}$
      on
      $(X,U,Y)$
      that is affine in the first argument
      such that
      \begin{equation*}
        \hat{\gamma}
        =
        \big\{
          \tuple{P,A_P,b_P}
        \big\}_{P\in\partitionstyle{P}}
        \qquad\text{with}\qquad
        f_P(x,u)
        =
        A_P\cdot x+b_P(u)
      \end{equation*}
      for every
      $P\in\partitionstyle{P}$
      and every
      $(x,u)\in X\times U$;

      in particular,
      \begin{equation*}
        \Repval{\gamma}
        =
        \Repval{\hat{\gamma}}.
      \end{equation*}
  \end{itemize}

  \par\medskip
  \textbf{Fact~\ref{fact:piecewise-uniformly-affine-coefficient-representation}.2.}~
  For every coefficient-based piecewise representation
  $\hat{\gamma}$
  on
  $(X,U,Y)$
  that is affine in the first argument,
  \begin{equation*}
    \hat{\gamma}
    =
    \big\{
      \tuple{P,A_P,b_P}
    \big\}_{P\in\partitionstyle{P}},
  \end{equation*}
  the piecewise representation
  \begin{equation*}
    \gamma
    \defeq
    \big\{
      \tuple{P,f_P}
    \big\}_{P\in\partitionstyle{P}}
    \qquad\text{having}\qquad
    f_P(x,u)
    \defeq
    A_P\cdot x+b_P(u)
  \end{equation*}
  for every
  $P\in\partitionstyle{P}$
  and every
  $(x,u)\in X\times U$
  is affine in the first argument and satisfies
  \begin{equation*}
    \Repval{\gamma}
    =
    \Repval{\hat{\gamma}}.
  \end{equation*}

  We refer to
  $\gamma$
  as the
  \emph{(canonical) piecewise representation affine in the first argument}
  induced by
  $\hat{\gamma}$.
\end{closingfact}

\begin{closingfact}
  \label{fact:piecewise-uniformly-affine-fixed-second-argument}
  Let
  \begin{equation*}
    \gamma
    =
    \big\{
      \tuple{P,f_P}
    \big\}_{P\in\partitionstyle{P}}
  \end{equation*}
  be a piecewise representation that is affine in the first argument,
  and let
  \begin{equation*}
    \hat{\gamma}
    =
    \big\{
      \tuple{P,A_P,b_P}
    \big\}_{P\in\partitionstyle{P}}
  \end{equation*}
  be a corresponding coefficient-based representation as in
  Fact~\ref{fact:piecewise-uniformly-affine-coefficient-representation}.

  For every
  $u\in U$
  and every
  $P\in\partitionstyle{P}$,
  define
  \begin{equation*}
    P_u
    \defeq
    \left\{
      x\in X
      \;\middle|\;
      (x,u)\in P
    \right\}.
  \end{equation*}
  Then
  \begin{equation*}
    \partitionstyle{P}_u
    \defeq
    \left\{
      P_u
      \;\middle|\;
      P\in\partitionstyle{P},
      \;
      P_u\neq\emptyset
    \right\}
  \end{equation*}
  is a finite partition of
  $X$.

  Furthermore, defining
  \begin{equation*}
    f_u:X\to Y,
    \qquad
    f_u(x)
    \defeq
    \Repval{\gamma}(x,u),
  \end{equation*}
  we have that
  $f_u$
  is piecewise affine.

  More precisely,
  for every nonempty
  $P_u\in\partitionstyle{P}_u$,
  \begin{sidealign*}
    \quanteq{
      f_u(x)
      =
      A_P\cdot x+b_P(u)
    }{
      \forall x\in P_u,
    }
  \end{sidealign*}
  so that
  $\tuple{A_P,b_P(u)}$
  are affine coefficients for
  $f_u$
  on the region
  $P_u$.
\end{closingfact}

\subsection{Analytic Euclidean Maps and Neural Primitives}
\label{subsec:analytic-euclidean-maps}

We first isolate the function class used by the RNC constructions in this
appendix.  The relevant notion is deliberately stronger than ordinary analyticity on the
relative domain of a Euclidean space: we require an analytic map on the entire
Euclidean ambient space.  This is the condition needed below to obtain uniform
value and derivative bounds on arbitrary bounded subsets, without additional
assumptions that such subsets remain a positive distance from the boundary of an
open domain.

We first recall the ordinary notion of analyticity.

\begin{closingdefinition}[Ordinarily-analytic Euclidean maps, cf.~\citep{krantz2002primer}]
  \label{def:ordinarily-analytic-euclidean-map}
  Let $\Omega_X = \reals^m$
  and
  let $X \subseteq \Omega_X$ be open.
  A Euclidean function $f : X \to \reals^n$ is \emph{ordinarily-analytic}
  if,
  for every $c\in X$,
  there exist
  $r\in\posreals$
  and
  coefficients
  $c_\alpha\in\reals^n$, for $\alpha\in\naturals^m$,
  such that,
  letting $\hat X \defeq \openballopp{c,r}{\Omega_X}$, 
  the following three conditions hold jointly:
  \begin{enumerate}
    \item 
      \label{item:def:ordinarily-analytic-euclidean-map--1}
      the inclusion $\hat X \subseteq X$ holds;
    \item
      \label{item:def:ordinarily-analytic-euclidean-map--2}
      function $f$ is given by the power series
      \begin{equation*}
        f(x)
        \,=\,
        \sum_{\alpha\in\naturals^m}
        \mBig(
          c_\alpha
          \cdot
          \textstyle
          \prod_{i=1}^{m}
          \big(
            \sgetentry{x}{i}
            -
            \sgetentry{c}{i}
          \big)^{\sgetentry{\alpha}{i}}
        \mBig)
        \qquad\quad
        \forall
        x \in \hat X
        \,;
      \end{equation*}
    \item
      \label{item:def:ordinarily-analytic-euclidean-map--3}
      the power series of the above Point~\ref{item:def:ordinarily-analytic-euclidean-map--2}
      converges absolutely on $\hat X$.
  \end{enumerate}
\end{closingdefinition}

\begin{closingdefinition}[Analytic Euclidean maps]
  \label{def:analytic-euclidean-map}
  Let $X$ and $Y$ be Euclidean spaces
  having ambient spaces $\Omega_X=\reals^m$ and $\Omega_Y=\reals^n$,
  respectively.

  \begin{definitionpart}[indent=body]
    A Euclidean map
    $f:X\to Y$
    is \emph{ambiently-analytic} if there exists an ordinarily-analytic Euclidean map
    $\tilde f:\Omega_X\to\Omega_Y$
    such that
    $\restr{\tilde f}{X}=f$.
    Any such $\tilde f$ is called an \emph{analytic ambient extension} of $f$.

    We write 
    $\analyticfuncclass[X][Y]$
    for the class of analytic Euclidean maps from $X$ to $Y$.
    We also write $\analyticfuncclass$ for the class of all such maps
    across arbitrary finite-dimensional Euclidean source and target spaces.
  \end{definitionpart}

  \begin{definitionpart}[indent=body]
    In addition to $X$ and $Y$,
    let $\Theta$ be a Euclidean parameter space having ambient space $\Omega_\Theta$.

    A parametric Euclidean map
    $f:X\times\Theta\to Y$
    is a \emph{ambiently-analytic} if it admits an ordinarily-analytic
    ambient extension
    $\tilde f: \Omega_X\times\Omega_\Theta \to \Omega_Y$
    such that
    $\restr{\tilde f}{X\times\Theta}=f$.

    We write 
    $\analyticparamfuncclass[X][Y][\Theta]$
    for the class of analytic parametric Euclidean maps from $X$ to $Y$ 
    with parameter space $\Theta$.
  \end{definitionpart}

  When we say \emph{analytic} we mean \emph{ambiently-analytic} unless specified otherwise.
\end{closingdefinition}

\begin{closingremark}
  The ambient-extension requirement is essential for boundedness arguments used in our
  results---e.g., Proposition~\ref{prop:feedforward-neural-map-boundedness}.
  Ordinary analyticity alone would not suffice: for example,
  $x\mapsto 1/x$ is ordinarily-analytic on $(0,1)$, but maps the bounded set
  $(0,1)$ to an unbounded set.
  By contrast, an analytic ambient extension $\tilde f : \Omega_Z \to \Omega_Y$ is defined on the
  entire ambient Euclidean space $\Omega_Z = \reals^d$ and it is continuous with continuous
  derivatives of every order;
  hence $\tilde f$ is bounded on every bounded subset of its domain $\Omega_Z$,
  and, for every $n \in \posnaturals$, the same holds for its $n$-th derivative $\dv^n \tilde f$.
\end{closingremark}

\begin{closingfact}[Analytic Euclidean maps are smooth up to the closure]
  \label{fact:analytic-euclidean-map-smooth-up-to-closure}
  Every analytic Euclidean map $f:X\to Y$
  admits an infinitely-continuously-differentiable
  ambient extension $\tilde f:\Omega_X\to\Omega_Y$.
  In particular,
  $\restr{\tilde f}{\closureop{X}}$
  is a continuously-differentiable extension of $f$
  to the closure of its domain.
\end{closingfact}
\begin{proof}
  By
  Definition~\ref{def:analytic-euclidean-map},
  there exists an ordinarily-analytic ambient extension
  $\tilde f:\Omega_X\to\Omega_Y$
  such that $\restr{\tilde f}{X}=f$.
  Since every ordinarily-analytic Euclidean map between finite-dimensional Euclidean spaces is
  infinitely-continuously-differentiable,
  $\tilde f$
  is continuously differentiable on
  $\Omega_X$.
  Therefore,
  $\restr{\tilde f}{\closureop{X}}$
  is a continuously-differentiable extension of
  $f$
  to the closure of its domain.
\end{proof}

\begin{proposition}[Analytic maps are closed under tupling and composition]
  \label{prop:analytic-maps-closure}
  The class of analytic Euclidean maps is closed under finite function
  tupling and function composition.
  Consequently, if every primitive in a class $\mathcal{P}$ is analytic, then
  the closure $\mathcal{F}$ of $\mathcal{P}$ satisfies
  \begin{equation*}
    \mathcal{F}
    \subseteq
    \analyticfuncclass.
  \end{equation*}
\end{proposition}
\begin{proof}
  For each $i \in \iival{1}{k}$,
  let
  $f_i:X\to Y_i$
  be analytic Euclidean maps with analytic ambient extensions
  $\tilde f_i$.
  Then
  \begin{equation*}
    \ptuple{\tilde f_1,\ldots,\tilde f_k}
  \end{equation*}
  is ordinarily-analytic on $\Omega_X$ and restricts to
  $\ptuple{f_1,\ldots,f_k}$ on $X$.
  Hence analytic Euclidean maps are closed under finite tupling.

  Similarly, 
  let
  $f:X\to Y$
  and
  $g:Y\to Z$
  be analytic Euclidean maps with analytic ambient extensions
  $\tilde f:\Omega_X\to\Omega_Y$
  and
  $\tilde g:\Omega_Y\to\Omega_Z$.
  Then
  \begin{equation*}
    \tilde g\circ\tilde f:
    \Omega_X\to\Omega_Z
  \end{equation*}
  is ordinarily-analytic and restricts to $g\circ f$ on $X$.
  Thus analytic Euclidean maps are also closed under composition.

  Therefore the class $\analyticfuncclass$ contains
  $\mathcal{P}$ and it is closed under the two operations in
  Definition~\ref{def:function-class-closure}.
  By minimality of the closure of $\mathcal{P}$,
  we conclude
  $\mathcal{F} \subseteq \analyticfuncclass$.
\end{proof}

\begin{closingdefinition}[Coordinate maps]
  \label{def:coordinate-map}
  Let $n,m\in\posnaturals$
  and
  let $\iota : \iival{1}{m} \to \iival{1}{n}$.

  \begin{definitionpart}[indent=body]
    The \emph{coordinate map induced by $\iota$} is the Euclidean function
    \begin{equation*}
      \pi_\iota
      :
      \reals^n
      \to
      \reals^m
      \qquad\qquad
      \pi_\iota(x)
      \defeq
      \mmbigptuple{
        \getentry{x}{\iota(1)},\,
        \ldots,\,
        \getentry{x}{\iota(m)}
      }
      \,.
    \end{equation*}
  \end{definitionpart}

  \begin{definitionpart}[indent=body]
    We have the following special cases.

    A \emph{coordinate projection} is a coordinate map $\pi_\iota$ 
    where $\iota$ is strictly increasing.

    A \emph{coordinate permutation} is a coordinate map $\pi_\iota$ 
    where $\iota$ is a bijection.
  \end{definitionpart}
\end{closingdefinition}

\begin{closingfact}[Coordinate maps are analytic]
  \label{fact:coordinate-maps-analytic}
  Every coordinate map is analytic.
  In particular, every coordinate projection and every coordinate permutation
  is analytic.
\end{closingfact}
\begin{proof}
  Every coordinate map is a linear,
  and hence polynomial,
  map on its full Euclidean ambient space.
  It is therefore ordinarily-analytic on its full Euclidean ambient space,
  and consequently ambiently-analytic.
\end{proof}

\begin{proposition}[Analytic neural primitives]
  \label{prop:analytic-neural-primitives}
  The following ``neural primitives'' are ambiently-analytic, as they are ordinarily-analytic on their
  full Euclidean ambient spaces.
  \begin{itemize}
    \item
      Analyticity is satisfied by
      every affine map
      $x\mapsto A \cdot x+b$
      and 
      every parametric affine map
      $(x,A,b)\mapsto A \cdot x+b$;
      and in particular by every coordinate projection and permutation.

    \item
      For every $d\in\posnaturals$ and every $\varepsilon>0$, 
      analyticity is satisfied by
      the parametric LayerNorm map
      \begin{equation*}
        \operatorname{LN}_{\varepsilon}
        :
        \reals^d\times\reals^d\times\reals^d
        \to
        \reals^d
        \,,
        \qquad\quad
        \operatorname{LN}_{\varepsilon}(x,\gamma,\beta)
        \,\defeq\,
        \gamma \ast
        \frac{x-\mu(x) \cdot \mathbf{1}_d}
        {\sqrt{v(x)+\varepsilon}}
        +\beta
        \,,
      \end{equation*}
      where
      \begin{equation*}
        \mu(x)
        \,\defeq\,
        \frac{1}{d} \cdot 
        \sum_{i=1}^d \sgetentry{x}{i}
        \,,
        \qquad\qquad
        v(x)
        \,\defeq\,
        \frac{1}{d} \cdot 
        \sum_{i=1}^d
        \big(
          \sgetentry{x}{i}-\mu(x)
        \big)^2
        \,,
      \end{equation*}
      with $\ast$ denoting elementwise product.

    \item
      For every $d\in\posnaturals$ and every $\varepsilon>0$, 
      analyticity is satisfied by
      the parametric RMS-normalisation map
      \begin{equation*}
        \operatorname{RMSNorm}_{\varepsilon}
        :
        \reals^d\times\reals^d
        \to
        \reals^d
        \,,
        \qquad\quad
        \operatorname{RMSNorm}_{\varepsilon}(x,\gamma)
        \,\defeq\,
        \gamma
        \ast
        \frac{
          x
        }{
          \sqrt{
          \big(\|x\|^2 / d\big) + \varepsilon
          }
        }
        \,,
      \end{equation*}
      where $\ast$ is elementwise product.

    \item
      The scalar `activation' functions
      \begin{equation*}
        \begin{gathered}
          \operatorname{id},
          \qquad
          \exp,
          \qquad
          \sin,
          \qquad
          \cos,
          \qquad
          \tanh,
          \qquad
          \sigma(x)
          \defeq
          \frac{1}{1+e^{-x}},
          \\[.6em]
          \operatorname{softplus}(x) 
          \defeq
          \log(1+e^x),
          \quad\;\;
          \operatorname{SiLU}(x) 
          \defeq
          x \cdot \sigma(x),
          \quad\;\;
          \operatorname{GELU}(x) 
          \defeq
          \frac{x}{2}
          \cdot
          \left(
            1+
            \operatorname{erf}
            \left(\frac{x}{\sqrt{2}}\right)
          \right)
        \end{gathered}
      \end{equation*}
      are ambiently-analytic on $\reals$.
    \item
      The softmax map
      \begin{equation*}
        \operatorname{softmax}(x)_i
        \defeq
        \frac{e^{\getentry{x}{i}}}{\sum_{j=1}^d e^{\getentry{x}{j}}}
      \end{equation*}
      is ambiently-analytic on $\reals^d$.
  \end{itemize}
\end{proposition}
\begin{proof}
  Every affine map is polynomial and is therefore ordinarily-analytic
  on its full Euclidean ambient space.
  The same applies to every parametric affine map
  $(x,A,b)\mapsto A\cdot x+b$,
  since each of its coordinates is a polynomial in the entries of
  $x$, $A$, and $b$.
  In particular,
  coordinate projections and coordinate permutations are linear maps
  and are therefore ordinarily-analytic on their full Euclidean ambient spaces.

  For LayerNorm, the maps $\mu$ and $v$ are polynomial.
  Since
  \begin{equation*}
    v(x)+\varepsilon
    \geq
    \varepsilon
    >
    0,
  \end{equation*}
  the scalar map
  $z\mapsto z^{-1/2}$
  is evaluated only on the positive half-line,
  where it is ordinarily-analytic.
  Closure of ordinary analyticity under composition, multiplication, and addition
  therefore shows that
  $\operatorname{LN}_{\varepsilon}$
  is ordinarily-analytic jointly in
  $x,\gamma,\beta$
  on its full Euclidean ambient space.

  The RMS-normalisation claim follows analogously from
  \begin{equation*}
    \frac{\|x\|^2}{d}
    +
    \varepsilon
    \geq
    \varepsilon
    >
    0.
  \end{equation*}

  The functions
  $\operatorname{id}$,
  $\exp$,
  $\sin$,
  $\cos$,
  and
  $\tanh$
  are ordinarily-analytic on
  $\reals$.
  The sigmoid
  $\sigma$
  is ordinarily-analytic because
  $1+e^{-x}>0$
  for every
  $x\in\reals$.
  Softplus is ordinarily-analytic because
  $1+e^x>0$
  for every
  $x\in\reals$,
  and the logarithm is ordinarily-analytic on the positive half-line.
  SiLU is a product of ordinarily-analytic functions.
  The error function is ordinarily-analytic on
  $\reals$,
  and therefore
  $\operatorname{GELU}$
  is ordinarily-analytic by composition, multiplication, and addition.

  Finally,
  for every
  $x\in\reals^d$,
  \begin{equation*}
    \sum_{j=1}^d
    e^{\getentry{x}{j}}
    >
    0.
  \end{equation*}
  Hence every coordinate of the softmax map is a quotient of
  ordinarily-analytic functions with a nowhere-vanishing denominator.
  Therefore,
  softmax is ordinarily-analytic on
  $\reals^d$.

  Thus,
  each of the preceding maps is ordinarily-analytic on its full Euclidean
  ambient space,
  and consequently ambiently-analytic by
  Definition~\ref{def:analytic-euclidean-map}.
\end{proof}

\begin{closingcorollary}[Closures of analytic neural primitives]
  \label{cor:closure-analytic-neural-primitives}
  Let $\mathcal{P}$ be any class of analytic Euclidean functions,
  including any of the ``neural primitives'' listed in
  Proposition~\ref{prop:analytic-neural-primitives}.
  Then, every function in the closure of $\mathcal{P}$ is analytic.
\end{closingcorollary}

\begin{remark}[Parametric maps]
  The preceding statements apply directly to parametric neural maps by treating
  parameters as additional Euclidean input coordinates.
  Thus, for example,
  affine weights and biases, LayerNorm scale and bias, and RMSNorm scale may all
  vary as parameters while preserving joint analyticity in the ordinary inputs
  and parameters.
\end{remark}

\subsection{Affine Hyperplanes and Semilinear Sets}
\label{subsec:affine-hyperplanes}

\todo{}

\begin{closingdefinition}[Affine hyperplane]
  \label{def:affine-hyperplane}
  Let
  $d\in\posnaturals$.
  For
  $a\in\reals^d\setminus\{0\}$
  and
  $b\in\reals$,
  we define
  \begin{align*}
    \affinehyperplane{a}{b}
    \defeq
    \big\{
      z\in\reals^d
      \bigm|
      a^\transposeop\cdot z=b
    \big\}.
  \end{align*}
  We call every set of this form an
  \emph{affine hyperplane}
  in
  $\reals^d$.
\end{closingdefinition}

\begin{closingdefinition}[Affine equalities and inequalities]
  \label{def:affine-equalities-inequalities}
  Let
  $d\in\posnaturals$,
  and let
  \begin{align*}
    h:\reals^d\to\reals
  \end{align*}
  be an affine function.

  An expression of the form
  \begin{align*}
    h(z)=0
  \end{align*}
  is called an
  \emph{affine equality}.

  An expression of the form
  \begin{align*}
    h(z)\geq 0
  \end{align*}
  is called an
  \emph{affine inequality},
  and an expression of the form
  \begin{align*}
    h(z)>0
  \end{align*}
  is called a
  \emph{strict affine inequality}.
\end{closingdefinition}

\begin{closingremark}
  Affine inequalities of the forms
  \begin{align*}
    h(z)\leq 0
    \qquad\text{and}\qquad
    h(z)<0
  \end{align*}
  are included in the preceding terminology,
  since they are equivalently written as
  \begin{align*}
    -h(z)\geq 0
    \qquad\text{and}\qquad
    -h(z)>0,
  \end{align*}
  respectively.
\end{closingremark}

\begin{closingdefinition}[Finite affine constraint system]
  \label{def:finite-affine-constraint-system}
  Let
  $d\in\posnaturals$.
  A
  \emph{finite affine constraint system}
  on
  $\reals^d$
  is a finite collection of affine equalities,
  affine inequalities,
  and strict affine inequalities.

  Equivalently,
  it may be represented by finite index sets
  $I,J,K$
  and affine functions
  \begin{align*}
    h_i,g_j,q_k
    :
    \reals^d
    \to
    \reals,
    \qquad
    i\in I,
    \quad
    j\in J,
    \quad
    k\in K,
  \end{align*}
  through the system
  \begin{align*}
    h_i(z)
    &=
    0
    &&
    \forall i\in I,
    \\
    g_j(z)
    &\geq
    0
    &&
    \forall j\in J,
    \\
    q_k(z)
    &>
    0
    &&
    \forall k\in K.
  \end{align*}
\end{closingdefinition}

\begin{closingdefinition}[Affine-constraint set]
  \label{def:affine-constraint-set}
  Let
  $d\in\posnaturals$.
  An
  \emph{affine-constraint set}
  in
  $\reals^d$
  is the solution set of a finite affine constraint system.

  Equivalently,
  it is a set of the form
  \begin{align*}
    P
    =
    \mBig\{
      z\in\reals^d
      \mBigm|
      \textstyle\bigwedge_{i\in I} h_i(z)=0\,,
      \;\,
      \textstyle\bigwedge_{j\in J} g_j(z)\geq0\,,
      \;\,
      \textstyle\bigwedge_{k\in K} q_k(z)>0
    \mBig\},
  \end{align*}
  where
  $I,J,K$
  are finite index sets
  and
  \begin{align*}
    h_i,g_j,q_k
    :
    \reals^d
    \to
    \reals
  \end{align*}
  are affine functions.
\end{closingdefinition}

\begin{closingdefinition}[Semilinear set]
  \label{def:semilinear-set}
  Let
  $d\in\posnaturals$.
  A set
  $S\subseteq\reals^d$
  is
  \emph{semilinear}
  if it is a finite union of affine-constraint sets.

  Equivalently,
  there exist
  $m\in\naturals$
  and affine-constraint sets
  \begin{align*}
    P_1,\ldots,P_m
    \subseteq
    \reals^d
  \end{align*}
  such that
  \begin{align*}
    S
    =
    \bigcup_{\ell=1}^{m}
    P_\ell.
  \end{align*}
\end{closingdefinition}

\begin{closingdefinition}[Coordinate-equality and coordinate-threshold hyperplanes]
  \label{def:coordinate-equality-threshold-hyperplanes}
  Let
  $n\in\posnaturals$.

  For
  $j,k\in\iival{1}{n}$
  with
  $j<k$,
  we define the
  \emph{coordinate-equality hyperplane}
  \begin{align*}
    \coordinateEqualityHyperplane{j}{k}
    \defeq
    \big\{
      x\in\reals^n
      \bigm|
      \getentry{x}{j}
      =
      \getentry{x}{k}
    \big\}.
  \end{align*}

  For
  $j\in\iival{1}{n}$
  and
  $c\in\reals$,
  we define the
  \emph{coordinate-threshold hyperplane}
  \begin{align*}
    \coordinateThresholdHyperplane{j}{c}
    \defeq
    \big\{
      x\in\reals^n
      \bigm|
      \getentry{x}{j}
      =
      c
    \big\}.
  \end{align*}
\end{closingdefinition}

\begin{closingdefinition}[Cells induced by a finite hyperplane family]
  \label{def:affine-hyperplane-cells}
  Let
  $d\in\posnaturals$,
  and let
  $\mathcal H$
  be a finite family of affine hyperplanes in
  $\reals^d$.

  For an affine hyperplane
  \begin{align*}
    H
    =
    \affinehyperplane{a}{b},
  \end{align*}
  we define the relation
  $\equiv_H$
  on
  $\reals^d$
  by
  \begin{align*}
    z\equiv_H z'
    \quad\Longleftrightarrow\quad
    &
    \left(
      a^\transposeop\cdot z-b
    \right)
    \left(
      a^\transposeop\cdot z'-b
    \right)
    >0
    \\
    &{}\quad\text{or}\quad
    \left(
      a^\transposeop\cdot z-b
    \right)
    =
    \left(
      a^\transposeop\cdot z'-b
    \right)
    =
    0.
  \end{align*}
  Thus,
  $z\equiv_H z'$
  if and only if
  $z$ and $z'$
  lie on the same side of
  $H$,
  or both lie on
  $H$.

  We define the relation
  $\equiv_{\mathcal H}$
  on
  $\reals^d$
  by
  \begin{align*}
    z
    \equiv_{\mathcal H}
    z'
    \quad\Longleftrightarrow\quad
    z
    \equiv_H
    z'
    \qquad
    \forall H\in\mathcal H.
  \end{align*}

  The
  \emph{hyperplane cells induced by $\mathcal H$}
  are the equivalence classes of
  $\equiv_{\mathcal H}$.
  We denote the corresponding partition of
  $\reals^d$
  by
  \begin{align*}
    \partitionstyle{C}(\mathcal H)
    \defeq
    \reals^d/\equiv_{\mathcal H}.
  \end{align*}
\end{closingdefinition}

\begin{closingfact}[Finiteness of hyperplane-cell partitions]
  \label{fact:affine-hyperplane-cells-finite}
  Let
  $d\in\posnaturals$,
  and let
  $\mathcal H$
  be a finite family of affine hyperplanes in
  $\reals^d$.
  Then
  \begin{align*}
    \partitionstyle{C}(\mathcal H)
  \end{align*}
  is a finite partition of
  $\reals^d$.
  In particular,
  \begin{align*}
    \big|
      \partitionstyle{C}(\mathcal H)
    \big|
    \,\leq\,
    3^{|\mathcal H|}.
  \end{align*}
\end{closingfact}
\begin{proof}
  By
  Definition~\ref{def:affine-hyperplane-cells},
  every hyperplane
  $H\in\mathcal H$
  partitions
  $\reals^d$
  into three equivalence classes under
  $\equiv_H$:
  the hyperplane
  $H$
  itself and its two open sides.

  Hence, every equivalence class of
  $\equiv_{\mathcal H}$
  is determined by specifying, for every
  $H\in\mathcal H$,
  which of these three classes contains its elements.
  Therefore there are at most
  \begin{align*}
    3^{|\mathcal H|}
  \end{align*}
  equivalence classes.

  Since
  $\partitionstyle{C}(\mathcal H)$
  is, by definition, the set of equivalence classes of
  $\equiv_{\mathcal H}$,
  it is a finite partition of
  $\reals^d$,
  and
  \begin{align*}
    \big|
      \partitionstyle{C}(\mathcal H)
    \big|
    \,\leq\,
    3^{|\mathcal H|}.
  \end{align*}
\end{proof}

\subsection{PAP Functions and Intensional Derivatives}
\label{subsec:pap-intensional-preliminaries}

\todo{}

We introduce the notions of \emph{piecewise-analytic-under-analytic-partition (PAP)} functions and
\emph{intensional derivatives} introduced by \citep{lee2020correctness}.
We restate only the definitions and results needed below;
see Definitions~2--7 and Propositions~6--10 of
\citep{lee2020correctness} for the original development.

\begin{closingdefinition}[PAP functions]
  \label{def:pap-function}
  Let
  $X \subseteq \reals^n$
  and
  $Y \subseteq \reals^m$.

  Following Definition~3 of \citep{lee2020correctness},
  a set $A \subseteq \reals^n$ is \emph{analytic}
  if it can be described by finitely many strict and non-strict
  inequalities of real-analytic functions on open domains; that is,
  if there exist finite families of real-analytic functions
  \begin{equation*}
    g_j^+ : U_j^+ \to \reals,
    \qquad
    g_\ell^- : U_\ell^- \to \reals,
  \end{equation*}
  with open domains $U_j^+,U_\ell^- \subseteq \reals^n$,
  such that
  \begin{equation*}
    A
    =
    \left\{
      x\in\reals^n
      \;\middle|\;
      \bigwedge_j
      \bigl(
        x\in U_j^+
        \land
        g_j^+(x)>0
      \bigr)
      \land
      \bigwedge_\ell
      \bigl(
        x\in U_\ell^-
        \land
        g_\ell^-(x)\leq 0
      \bigr)
    \right\}.
  \end{equation*}
  A partition is called \emph{analytic} if all of its sets are analytic.

  A \emph{piecewise representation} of a function
  $f:X\to Y$
  is a finite or countable family
  \begin{equation*}
    \gamma
    =
    \bigl\{
      \ptuple{A_i,f_i}
    \bigr\}_{i\in I},
  \end{equation*}
  where
  $\{A_i\}_{i\in I}$ is a partition of $X$ and, for every $i\in I$,
  \begin{equation*}
    f_i : U_i \to Y
  \end{equation*}
  is defined on an open set
  $U_i\subseteq\reals^n$
  containing $A_i$.
  Its evaluation is the function
  $\llbracket\gamma\rrbracket:X\to Y$ given by
  \begin{equation*}
    \llbracket\gamma\rrbracket(x)
    =
    f_i(x)
    \qquad
    \text{for the unique $i\in I$ such that $x\in A_i$}.
  \end{equation*}

  The representation $\gamma$ is a
  \emph{piecewise-analytic-under-analytic-partition representation},
  abbreviated as a \emph{PAP representation},
  if
  $\{A_i\}_{i\in I}$ is an analytic partition and every $f_i$ is
  real-analytic.

  A function
  $f:X\to Y$
  is a \emph{PAP function} if it is the evaluation of some
  PAP representation:
  \begin{equation*}
    f
    =
    \llbracket\gamma\rrbracket.
  \end{equation*}

  We write
  \begin{equation*}
    \papfullclass(X,Y)
  \end{equation*}
  for the class of PAP functions from $X$ to $Y$.
\end{closingdefinition}

\begin{closingdefinition}[Intensional derivative]
  \label{def:intensional-derivative}
  Let
  \begin{equation*}
    f:X\to Y,
    \qquad
    X\subseteq\reals^n,
    \qquad
    Y\subseteq\reals^m,
  \end{equation*}
  be a PAP function, and let
  \begin{equation*}
    \gamma
    =
    \bigl\{
      \ptuple{A_i,f_i}
    \bigr\}_{i\in I}
  \end{equation*}
  be a PAP representation of $f$.

  The \emph{intensional derivative of the representation $\gamma$}
  is the representation
  \begin{equation*}
    D\gamma
    \defeq
    \bigl\{
      \ptuple{A_i,Df_i}
    \bigr\}_{i\in I}.
  \end{equation*}
  By Proposition~6 of \citep{lee2020correctness},
  $D\gamma$ is itself a PAP representation.

  A function
  \begin{equation*}
    d f
    :
    X
    \to
    \reals^{m\times n}
  \end{equation*}
  is an \emph{intensional derivative of $f$}
  if there exists a PAP representation $\gamma$ of $f$ such that
  \begin{equation*}
    d f
    =
    \llbracket D\gamma\rrbracket.
  \end{equation*}
  We write
  \begin{equation*}
    \intensionalderivatives f
  \end{equation*}
  for the set of all intensional derivatives of $f$.
\end{closingdefinition}

\begin{proposition}[Almost-everywhere correctness of intensional derivatives]
  \label{prop:pap-intensional-derivative-ae-correct}
  Let
  \begin{equation*}
    f:X\to Y,
    \qquad
    X\subseteq\reals^n,
    \qquad
    Y\subseteq\reals^m,
  \end{equation*}
  be a PAP function,
  and let
  \begin{equation*}
    d f
    \in
    \intensionalderivatives f
  \end{equation*}
  be an intensional derivative of
  $f$.
  Then there exists a full-measure subset
  \begin{equation*}
    \hat X
    \subseteq
    X
  \end{equation*}
  such that
  $f$
  is differentiable at every
  $x\in\hat X$
  and
  \begin{equation*}
    Df(x)
    =
    d f(x)
    \qquad
    \forall x\in\hat X.
  \end{equation*}
\end{proposition}

\begin{proof}
  This is the first-order case of
  Proposition~8 of
  \citet{lee2020correctness}.
\end{proof}

\begin{proposition}[Composition and chain rule for intensional derivatives]
  \label{prop:pap-intensional-chain-rule}
  Let
  \begin{equation*}
    f:X\to Y,
    \qquad
    g:Y\to Z,
  \end{equation*}
  where
  \begin{equation*}
    X\subseteq\reals^n,
    \qquad
    Y\subseteq\reals^m,
    \qquad
    Z\subseteq\reals^\ell,
  \end{equation*}
  and suppose that $f$ and $g$ are PAP functions.
  Then:
  \begin{enumerate}
    \item
      the composition $g\circ f$ is a PAP function;
    \item
      for every
      \begin{equation*}
        d f\in\intensionalderivatives f
        \qquad\text{and}\qquad
        d g\in\intensionalderivatives g,
      \end{equation*}
      the function
      \begin{equation*}
        d h(x)
        \defeq
        d g\bigl(f(x)\bigr)\cdot d f(x)
      \end{equation*}
      is an intensional derivative of
      $h\defeq g\circ f$, that is,
      \begin{equation*}
        d h\in\intensionalderivatives(g\circ f).
      \end{equation*}
  \end{enumerate}
\end{proposition}
\begin{proof}
  The first claim is Proposition~9 of \citep{lee2020correctness},
  and the second claim is Proposition~10 of
  \citep{lee2020correctness}.
\end{proof}

\begin{lemma}[Finite tupling of intensional derivatives]
  \label{lem:pap-intensional-tupling}
  For
  $i\in\iival{1}{k}$,
  let
  \begin{equation*}
    f_i:X\to\reals^{m_i}
  \end{equation*}
  be PAP functions having intensional derivatives
  \begin{equation*}
    d f_i
    \in
    \intensionalderivatives f_i.
  \end{equation*}
  Define
  \begin{equation*}
    f
    \defeq
    \ptuple{f_1,\ldots,f_k}
  \end{equation*}
  and
  \begin{equation*}
    d f(x)
    \defeq
    \begin{pmatrix}
      d f_1(x)
      \\
      \vdots
      \\
      d f_k(x)
    \end{pmatrix}.
  \end{equation*}
  Then $f$ is PAP and
  \begin{equation*}
    d f
    \in
    \intensionalderivatives f.
  \end{equation*}
\end{lemma}

\begin{proof}
  For every $i\in\iival{1}{k}$,
  since $d f_i$ is an intensional derivative of $f_i$,
  there exists a PAP representation
  \begin{equation*}
    \gamma_i
    =
    \bigl\{
      \ptuple{A_{i,j},f_{i,j}}
    \bigr\}_{j\in I_i}
  \end{equation*}
  of $f_i$ such that
  \begin{equation*}
    d f_i
    =
    \llbracket D\gamma_i\rrbracket.
  \end{equation*}

  Consider the finite common refinement of these representations.
  For every
  \begin{equation*}
    \alpha
    =
    \ptuple{j_1,\ldots,j_k}
    \in
    I_1\times\cdots\times I_k,
  \end{equation*}
  let
  \begin{equation*}
    A_\alpha
    \defeq
    \bigcap_{i=1}^k A_{i,j_i},
  \end{equation*}
  and define
  \begin{equation*}
    f_\alpha
    \defeq
    \ptuple{
      f_{1,j_1},
      \ldots,
      f_{k,j_k}
    }
  \end{equation*}
  on the intersection of the corresponding open domains.

  Finite intersections of analytic sets are analytic,
  finite intersections of open sets are open,
  and finite tuples of analytic functions are analytic.
  Furthermore,
  $I_1\times\cdots\times I_k$
  is countable.
  Hence
  \begin{equation*}
    \gamma
    \defeq
    \bigl\{
      \ptuple{A_\alpha,f_\alpha}
    \bigr\}_{\alpha\in I_1\times\cdots\times I_k}
  \end{equation*}
  is a PAP representation of $f$.
  This is the finite-tupling closure of PAP functions also used in
  the proof of Proposition~11 of \citep{lee2020correctness}.

  Furthermore,
  for every $\alpha$,
  \begin{equation*}
    Df_\alpha(x)
    =
    \begin{pmatrix}
      Df_{1,j_1}(x)
      \\
      \vdots
      \\
      Df_{k,j_k}(x)
    \end{pmatrix}.
  \end{equation*}
  It follows that
  \begin{equation*}
    \llbracket D\gamma\rrbracket(x)
    =
    \begin{pmatrix}
      d f_1(x)
      \\
      \vdots
      \\
      d f_k(x)
    \end{pmatrix}
    =
    d f(x).
  \end{equation*}
  Hence $d f\in\intensionalderivatives f$.
\end{proof}

\begin{remark}
  Proposition~7 of \citep{lee2020correctness} shows that every PAP
  function admits intensional derivatives of every order.
  Proposition~8 shows that every first-order intensional derivative
  agrees with the ordinary derivative wherever required outside a
  measure-zero set.
  These results motivate the notion of intensional derivative but are
  not needed for Proposition~\ref{prop:correctness-graph-derivative};
  its proof uses only the definitions above, finite tupling, and the
  composition and chain-rule results of Propositions~9 and~10 of
  \citep{lee2020correctness}.
\end{remark}

\subsection{Lattice Theory}
\label{subsec:lattice-theoretic-preliminaries}

\todo{}

We present the elementary lattice-theoretic notions of
\emph{meet},
\emph{join},
\emph{lattice},
and
\emph{distributive lattice}
used in this work---cf.~\citep{davey2002introduction}.

\begin{closingdefinition}
  A \emph{partially ordered set} (or \emph{poset})
  is a pair
  $\ptuple{X,\preceq}$
  where
  $\preceq$
  is a reflexive, antisymmetric, and transitive binary relation on
  $X$.
\end{closingdefinition}

\begin{closingdefinition}
  Let $\ptuple{X,\preceq}$ be a poset,
  and 
  let $x,y\in X$.

  An element $z\in X$ is:
  \begin{itemize}
    \item
      a \emph{lower bound} of $x$ and $y$
      if
      $z\preceq x$
      and
      $z\preceq y$;
    \item
      an \emph{upper bound} of $x$ and $y$ 
      if
      $x\preceq z$
      and
      $y\preceq z$.
  \end{itemize}

  The \emph{meet} of $x$ and $y$, when it exists, 
  is their greatest lower bound and is denoted by $x\wedge y$.
  Thus
  \begin{equation*}
    x\wedge y\preceq x,
    \qquad
    x\wedge y\preceq y,
  \end{equation*}
  and, for every $z\in X$,
  \begin{equation*}
    z\preceq x
    \;\land\;
    z\preceq y
    \quad\Longrightarrow\quad
    z\preceq x\wedge y.
  \end{equation*}

  Dually, the \emph{join} of $x$ and $y$, when it exists,
  is their least upper bound and is denoted by $x\vee y$.
  Thus
  \begin{equation*}
    x\preceq x\vee y,
    \qquad
    y\preceq x\vee y,
  \end{equation*}
  and, for every $z\in X$,
  \begin{equation*}
    x\preceq z
    \;\land\;
    y\preceq z
    \quad\Longrightarrow\quad
    x\vee y\preceq z.
  \end{equation*}

  Whenever they exist,
  meets and joins are unique by antisymmetry.
\end{closingdefinition}

\begin{closingdefinition}[Lattice]
  \label{def:lattice}
  A poset $\ptuple{X,\preceq}$ is a \emph{lattice}
  if every pair $x,y\in X$ admits both 
  a meet $x\wedge y$ 
  and 
  a join $x\vee y$.
\end{closingdefinition}

\begin{closingfact}
  Meet and join are associative in every lattice.
  Hence finite non-empty meets and joins are unambiguously denoted by
  \begin{equation*}
    \bigwedge_{i=1}^n x_i
    \qquad\text{and}\qquad
    \bigvee_{i=1}^n x_i,
  \end{equation*}
  respectively.
\end{closingfact}

\begin{closingdefinition}[Distributive lattice]
  \label{def:distributive-lattice}
  A lattice
  $\ptuple{X,\preceq}$
  is \emph{distributive}
  if, for every
  $x,y,z\in X$,
  \begin{align*}
    x\wedge(y\vee z)
    &=
    (x\wedge y)\vee(x\wedge z),
    \\
    x\vee(y\wedge z)
    &=
    (x\vee y)\wedge(x\vee z).
  \end{align*}
\end{closingdefinition}

\begin{closingfact}
  In a distributive lattice,
  the binary distributive laws extend to arbitrary finite non-empty
  meets and joins.
  In particular, for every
  $n\in\posnaturals$,
  \begin{align*}
    x
    \wedge
    \bigvee_{i=1}^n y_i
    \,=\,
    \bigvee_{i=1}^n
    (x\wedge y_i),
    \qquad\qquad
    x
    \vee
    \bigwedge_{i=1}^n y_i
    \,=\,
    \bigwedge_{i=1}^n
    (x\vee y_i).
  \end{align*}
\end{closingfact}

\subsection{Algebraic Structures}
\label{subsec:algebraic-structures}

\todo{}

We recall the elementary notions of universal algebra used in this work;
cf.~\citep{burris2011course}.
Here, the term
\emph{algebra}
is used in the universal-algebraic sense of a set equipped with a collection
of finitary operations, rather than in the more specialised sense of an
algebra over a field.

\begin{closingdefinition}[Finitary operation]
  \label{def:finitary-operation}
  Let $X$ be a non-empty set,
  and 
  let
  $n\in\naturals$.

  An \emph{$n$-ary operation on $X$} is a function $f : X^n \to X$.
  The integer $n$ is called the \emph{arity} of $f$.

  An operation of finite arity is called a
  \emph{finitary operation}.
  In particular,
  operations of arity
  $1$
  and
  $2$
  are called
  \emph{unary}
  and
  \emph{binary},
  respectively.
\end{closingdefinition}

\begin{closingdefinition}[Algebraic signature]
  \label{def:algebraic-signature}
  An \emph{algebraic signature} is a pair
  $\Sigma = \ptuple{ F, \arity }$ 
  where
  $F$ is a set of operation symbols 
  and
  $\arity : F \to \naturals$ assigns an \emph{arity} to every operation symbol.

  When the arity of every operation symbol is clear from context,
  we identify the signature with $F$.
\end{closingdefinition}

\begin{closingdefinition}[Algebra]
  \label{def:algebra}
  Let $\Sigma = \ptuple{ F, \arity }$ be an algebraic signature.

  A \emph{$\Sigma$-algebra} is a tuple
  $\mathcal{A} = \ptuple{ A, (f^\mathcal{A})_{f\in F} }$
  where:
  \begin{itemize}
    \item
      $A$ is a non-empty set, 
      called the \emph{carrier} or \emph{underlying set} of $\mathcal{A}$;
    \item
      for every $f\in F$, 
      $f^\mathcal{A} : A^{\arity(f)} \to A$
      is an operation on $A$.
  \end{itemize}

  When no ambiguity arises,
  we omit the superscript
  $\mathcal{A}$
  from the operations.
\end{closingdefinition}

\begin{closingdefinition}[Binary algebra]
  \label{def:binary-algebra}
  A \emph{binary algebra} $\mathcal{A}$ is an algebra whose signature consists only of
  binary operation symbols.

  In particular,
  an algebra having two binary operations
  $\star$
  and
  $\diamond$
  is written as
  \begin{equation*}
    \mathcal{A}
    =
    \ptuple{
      A,
      \star,
      \diamond
    }
    \quad\text{ where }\quad
    \star,
    \diamond
    :
    A\times A
    \to
    A.
  \end{equation*}
  Binary operations are written using \emph{infix notation}.
  Given a binary operation
  $\star : A\times A \to A$,
  and
  elements $a,b\in A$,
  we write
  $a\star b$
  in place of
  $\star(a,b)$.
\end{closingdefinition}

\begin{closingdefinition}[Properties of binary operations]
  \label{def:binary-operation-properties}
  A binary operation $\star$ on a set $A$ is:
  \begin{itemize}
    \item
      \emph{associative} if
      \begin{equation*}
        (x\star y)\star z
        \,=\,
        x\star(y\star z)
        \qquad\quad
        \forall x,y,z\in A;
      \end{equation*}

    \item
      \emph{commutative} if
      \begin{equation*}
        x\star y
        \,=\,
        y\star x
        \qquad\quad
        \forall x,y\in A;
      \end{equation*}

    \item
      \emph{idempotent} if
      \begin{equation*}
        x\star x
        \,=\,
        x
        \qquad\quad
        \forall x\in A.
      \end{equation*}
  \end{itemize}
\end{closingdefinition}

\begin{closingdefinition}[Distributive binary operations]
  \label{def:distributive-binary-operations}
  Let
  $\star$
  and
  $\diamond$
  be binary operations on a set
  $A$.
  Then:
  \begin{itemize}
    \item
      operation
      $\diamond$
      is \emph{left-distributive over}
      $\star$
      if
      \begin{equation*}
        x\diamond(y\star z)
        =
        (x\diamond y)
        \star
        (x\diamond z)
        \qquad\quad
        \forall x,y,z\in A;
      \end{equation*}

    \item
      operation
      $\diamond$
      is \emph{right-distributive over}
      $\star$
      if
      \begin{equation*}
        (x\star y)\diamond z
        =
        (x\diamond z)
        \star
        (y\diamond z)
        \qquad\quad
        \forall x,y,z\in A;
      \end{equation*}

    \item
      operation
      $\diamond$
      \emph{distributes over}
      $\star$
      if it is both left-distributive and right-distributive over
      $\star$;

    \item
      operations
      $\star$
      and
      $\diamond$
      are
      \emph{mutually distributive}
      if
      $\diamond$
      distributes over
      $\star$
      and
      $\star$
      distributes over
      $\diamond$.
  \end{itemize}
\end{closingdefinition}

\begin{closingdefinition}[Subalgebra]
  \label{def:subalgebra}
  Let 
  $\mathcal{A} = \ptuple{ A, (f^\mathcal{A})_{f\in F} }$
  be a $\Sigma$-algebra.

  A \emph{subalgebra} of $\mathcal{A}$ 
  is 
  a $\Sigma$-algebra 
  $\mathcal{B} = \ptuple{ B, (f^\mathcal{B})_{f\in F} }$
  such that
  \begin{equation*}
    B\subseteq A 
    \qquad\;\text{ and }\qquad\;
    f^\mathcal{B}
    =
    f^\mathcal{A}
    \big|_{B^{\arity(f)}}
    \quad
    \forall f\in F.
  \end{equation*}
  Equivalently,
  $B$ 
  is closed under all operations of
  $\mathcal{A}$,
  and
  $\mathcal{B}$
  carries their restrictions to
  $B$.
\end{closingdefinition}

\begin{closingdefinition}[Homomorphism and isomorphism]
  \label{def:algebra-homomorphism}
  Let us consider 
  $\Sigma$-algebras
  \begin{equation*}
    \mathcal{A}
    =
    \ptuple{
      A,
      (f^\mathcal{A})_{f\in F}
    }
    \qquad\text{and}\qquad
    \mathcal{B}
    =
    \ptuple{
      B,
      (f^\mathcal{B})_{f\in F}
    }.
  \end{equation*}

  A function $h : A \to B$
  is a \emph{homomorphism} if,
  for every operation symbol
  $f\in F$ of arity $n=\arity(f)$,
  \begin{equation*}
    h\big(
      f^\mathcal{A}(a_1,\ldots,a_n)
    \big)
    =
    f^\mathcal{B}\big(
      h(a_1),\ldots,h(a_n)
    \big)
    \qquad\quad
    \forall\, \ptuple{a_1,\ldots,a_n}\in A^n.
  \end{equation*}

  A bijective homomorphism is called an \emph{isomorphism}.
  Two algebras are \emph{isomorphic} if there exists an isomorphism between them.
\end{closingdefinition}

\begin{closingdefinition}[Lattice algebra]
  \label{def:lattice-algebra}
  A \emph{lattice algebra} is a binary algebra
  \begin{align*}
    \mathcal{A} = \ptuple{ A, \vee, \wedge }
  \end{align*}
  whose two operations are associative, commutative, and idempotent,
  and satisfy the following identities for every $x,y \in A$:
  \begin{align*}
    x\vee(x\wedge y)
    & \,=\,
    x,
    \\[.4em]
    x\wedge(x\vee y)
    & \,=\,
    x.
  \end{align*}

  A lattice algebra is
  \emph{distributive}
  if
  $\vee$
  and
  $\wedge$
  are mutually distributive.
\end{closingdefinition}

\begin{closingfact}[Order and algebraic presentations of lattices]
  \label{fact:lattice-order-algebra-equivalence}
  The order-theoretic and algebraic notions of lattice are equivalent.

  In particular,
  if
  $\ptuple{A,\preceq}$
  is a lattice in the sense of
  Definition~\ref{def:lattice},
  then
  \begin{equation*}
    \ptuple{
      A,
      \vee,
      \wedge
    }
  \end{equation*}
  is a lattice algebra,
  where
  $\vee$
  and
  $\wedge$
  are respectively the join and meet induced by
  $\preceq$.

  Conversely,
  a lattice algebra
  $\ptuple{A,\vee,\wedge}$
  induces the partial order
  \begin{equation*}
    x\preceq y
    \quad\Longleftrightarrow\quad
    x\vee y=y,
  \end{equation*}
  equivalently,
  \begin{equation*}
    x\preceq y
    \quad\Longleftrightarrow\quad
    x\wedge y=x.
  \end{equation*}

  Under these correspondences,
  distributive lattices correspond exactly to distributive lattice algebras.
\end{closingfact}

\subsection{Semigroup and Group Theory}
\label{app:prel-groups}

\todo{}

We recall the elementary notions of semigroup, group, and permutation-group
theory used in this work;
cf.~\citep{howie1995fundamentals} for semigroup theory and
\citep{dixon1996permutation} for permutation groups.
Semigroups and groups are particular algebraic structures in the sense of
Section~\ref{subsec:algebraic-structures}.

\begin{closingdefinition}[Semigroup]
  \label{def:semigroup}
  A \emph{semigroup} is a binary algebra
  \begin{equation*}
    S
    =
    \ptuple{
      A_S,
      \gop{S}
    }
  \end{equation*}
  whose binary operation
  \begin{equation*}
    \gop{S}
    :
    A_S\times A_S
    \to
    A_S
  \end{equation*}
  is associative:
  \begin{equation*}
    (a\gop{S}b)\gop{S}c
    =
    a\gop{S}(b\gop{S}c)
    \qquad
    \forall a,b,c\in A_S.
  \end{equation*}

  The semigroup
  $S$
  is \emph{finite} if its carrier
  $A_S$
  is finite.
\end{closingdefinition}

\begin{closingdefinition}[Products of subsets of a semigroup]
  \label{def:semigroup-subset-products}
  Let
  \begin{equation*}
    S
    =
    \ptuple{
      A_S,
      \gop{S}
    }
  \end{equation*}
  be a semigroup,
  and let
  $B,C\subseteq A_S$.

  Their product is
  \begin{equation*}
    B\gop{S}C
    \defeq
    \left\{
      b\gop{S}c
      \mid
      b\in B,\,
      c\in C
    \right\}.
  \end{equation*}

  For a non-empty subset
  $B\subseteq A_S$,
  its positive powers are defined recursively by
  \begin{align*}
    B^1
    &\defeq
    B,
    \\
    B^{n+1}
    &\defeq
    B^n\gop{S}B
    \qquad
    \forall n\in\posnaturals,
  \end{align*}
  and
  \begin{equation*}
    B^+
    \defeq
    \bigcup_{n\in\posnaturals} B^n.
  \end{equation*}
\end{closingdefinition}

\begin{closingfact}
  Let
  $S$
  be a semigroup and let
  $B$
  be a non-empty subset of its carrier.
  Then
  $B^+$
  is the smallest subsemigroup of
  $S$
  whose carrier contains
  $B$.
  It is therefore the
  \emph{subsemigroup generated by}
  $B$.
\end{closingfact}

\begin{closingdefinition}[Group]
  \label{def:group}
  A \emph{group} is a semigroup
  \begin{equation*}
    G
    =
    \ptuple{
      A_G,
      \gop{G}
    }
  \end{equation*}
  satisfying the following properties:
  \begin{enumerate}
    \item
      there exists an element
      $e\in A_G$
      such that
      \begin{equation*}
        e\gop{G}a
        =
        a\gop{G}e
        =
        a
        \qquad
        \forall a\in A_G;
      \end{equation*}
      this element is unique and is called the
      \emph{identity element} of
      $G$;

    \item
      for every
      $a\in A_G$,
      there exists an element
      $b\in A_G$
      such that
      \begin{equation*}
        a\gop{G}b
        =
        b\gop{G}a
        =
        e;
      \end{equation*}
      this element is unique and is called the
      \emph{inverse} of
      $a$,
      denoted by
      $a^{-1}$.
  \end{enumerate}

  The group
  $G$
  is \emph{finite} if its carrier
  $A_G$
  is finite.
\end{closingdefinition}

\begin{closingdefinition}[Subsemigroup and subgroup]
  \label{def:subsemigroup-subgroup}
  Let
  \begin{equation*}
    S
    =
    \ptuple{
      A_S,
      \gop{S}
    }
  \end{equation*}
  be a semigroup.

  A \emph{subsemigroup} of
  $S$
  is a subalgebra
  \begin{equation*}
    S'
    =
    \ptuple{
      A_{S'},
      \gop{S'}
    }
  \end{equation*}
  of
  $S$.
  Equivalently,
  $A_{S'}$
  is a non-empty subset of
  $A_S$
  closed under
  $\gop{S}$,
  and
  $\gop{S'}$
  is the restriction of
  $\gop{S}$
  to
  $A_{S'}$.

  Let
  $G$
  be a group.
  A \emph{subgroup} of
  $G$
  is a subsemigroup of
  $G$
  that is itself a group.
\end{closingdefinition}

\begin{closingfact}
  Homomorphisms and isomorphisms between semigroups or groups are precisely
  homomorphisms and isomorphisms in the sense of
  Definition~\ref{def:algebra-homomorphism}
  for their corresponding algebraic signatures.

  In particular,
  an isomorphism between semigroups
  \begin{equation*}
    S_1
    =
    \ptuple{
      A_1,
      \gop{1}
    }
    \qquad\text{and}\qquad
    S_2
    =
    \ptuple{
      A_2,
      \gop{2}
    }
  \end{equation*}
  is a bijection
  \begin{equation*}
    \psi
    :
    A_1
    \to
    A_2
  \end{equation*}
  satisfying
  \begin{equation*}
    \psi(a\gop{1}b)
    =
    \psi(a)\gop{2}\psi(b)
    \qquad
    \forall a,b\in A_1.
  \end{equation*}
\end{closingfact}

\begin{closingdefinition}[Transformation semigroup and group]
  \label{def:transformation-semigroup}
  Let
  $X$
  be a set.

  A \emph{transformation semigroup} on
  $X$
  is a pair
  \begin{equation*}
    \ptuple{
      X,
      T
    }
  \end{equation*}
  where
  $T$
  is a set of transformations on $X$,
  as per Definition~\ref{def:transformation-permutation},
  such that
  \begin{equation*}
    \ptuple{
      T,
      \circ
    }
  \end{equation*}
  is a semigroup under function composition.

  A \emph{transformation group} on
  $X$
  is a transformation semigroup
  $\ptuple{X,T}$
  such that
  \begin{equation*}
    \ptuple{
      T,
      \circ
    }
  \end{equation*}
  is a group.
\end{closingdefinition}

\begin{closingdefinition}[Permutation group]
  \label{def:permutation-group}
  A \emph{permutation group} on a set
  $X$
  is a transformation group
  \begin{equation*}
    \ptuple{
      X,
      P
    }
  \end{equation*}
  such that every element of
  $P$
  is a permutation of
  $X$ as per Definition~\ref{def:transformation-permutation}.

  If
  $X$
  is finite,
  the \emph{degree} of the permutation group is
  \begin{equation*}
    |X|.
  \end{equation*}
\end{closingdefinition}

\begin{closingdefinition}[Symmetric group]
  \label{def:symmetric-group}
  Let
  $X$
  be a set.
  The \emph{symmetric group} on
  $X$,
  denoted by
  $\gsym{X}$,
  is the permutation group
  \begin{equation*}
    \ptuple{
      X,
      P_X
    },
  \end{equation*}
  where
  $P_X$
  is the set of all permutations of
  $X$,
  equipped with function composition.

  Every permutation group on
  $X$
  is a subgroup of
  $\gsym{X}$.
\end{closingdefinition}

\begin{theorem}[Cayley]
  \label{thm:cayley}
  Every group
  \begin{equation*}
    G
    =
    \ptuple{
      A_G,
      \gop{G}
    }
  \end{equation*}
  is isomorphic to a subgroup of
  $\gsym{A_G}$.
\end{theorem}

\subsection{Algebraic Automata Theory}
\label{sec:appendix-aat}

Algebraic Automata Theory (AAT), cf.~\citep{hartmanis,ginzburg,arbib,domosi2005algebraic}, offers a
refined perspective on automata, seeing an automaton not as
a transition system (a graph where one moves from a node to another according to a given input),
but instead as a device that manipulates elements of an underlying set by applying transformations
selected by the inputs.
As as result, it connects automata to semigroup and group theory, allowing for advanced
characterisations automata.

An \emph{automaton} is a system $A = \tuple{Q, \Sigma, \delta, q_0, \Gamma, \theta}$ where
$Q,\Sigma,\Gamma$ are finite spaces under the discrete metric. 
The standard terminology differs from the one of dynamical systems as 
$\Sigma$ is called input alphabet, $\delta$ is called transition function,
$\Gamma$ is called output alphabet, and 
the tuple $D = \tuple{Q, \Sigma, \delta}$ is called a \emph{semiautomaton}.

For $\sigma \in \Sigma$,
the \emph{state transformation} $\delta_\sigma$ of $D$ is the function 
$\delta_\sigma: Q \to Q$ given by $\delta_\sigma(q) = \delta(q,\sigma)$.
Semiautomaton $D$ is an \emph{identity-reset semiautomaton} if each transformation $\delta_\sigma$
is the identity function or a constant function,
and it is a \emph{permutation semiautomaton} if each transformation $\delta_\sigma$ is a bijection.

The \emph{transition monoid} of $D$ is the transformation monoid 
$\groupsty{M}(D) = \ptuple{X,\circ}$ where 
$$X = \{ \mathrm{id} \} \cup 
\{\delta_{\sigma_1} \circ \cdots 
\circ \delta_{\sigma_n} \mid \sigma_1, \ldots, \sigma_n \in \Sigma \}.$$

For $\groupsty{G} = \tuple{G,\gamma}$ a finite group with elements $G$ and operator 
$\gamma : G \times G \to G$, 
the \emph{$\groupsty{G}$-semiautomaton} is $A_\groupsty{G} = \tuple{G,G,\gamma}$.

A \emph{Mealy machine} is a tuple $M = \langle Q,\Sigma,\delta, \Gamma, \theta \rangle$ where $D_M = \langle Q,\Sigma,\delta \rangle$ is a semiautomaton,
$\Gamma$ is an output alphabet, and
$\theta: Q \times \Sigma \to \Gamma$ is an output function.
It defines the mapping $Q\times \Sigma^+\to \Gamma$ given by
\begin{align*}
  M(q,\sigma_1 \cdots \sigma_n) = \theta\big(D_M(q,\sigma_1 \cdots \sigma_n),\sigma_n\big).
\end{align*}

Given an automaton $A= \tuple{Q,\Sigma,\delta,q_0, \Gamma,\theta}$, 
the \emph{associated Mealy machine} $M_A=\tuple{Q,\Sigma,\delta, \Gamma,\theta}$ is obtained by
dropping the initial state from~$A$.

Given a semiautomaton $D_A=\tuple{Q,\Sigma,\delta}$,
its \emph{canonical Mealy machine} is
\begin{align*}
    \mathcal{M}(D) \defeq
    \tuple{Q,\Sigma, \delta, \Gamma, \theta}, 
    \quad
    \text{where } \Gamma \defeq Q \times \Sigma \text{, and } \theta \defeq \mathrm{id}.
\end{align*}

\begin{definition}[Assignments and realisations; Definitions~1.14 and~1.15 of \citep{hartmanis}]
  If $M=\tuple{Q,\Sigma, \delta,\Gamma,\theta}$ and 
  $M'=\tuple{Q',\Sigma',\delta',\Gamma',\theta'}$ are Mealy Machines, then the triple 
  $\ptuple{\alpha, \iota,\zeta}$ is called an \emph{assignment} of $M$ into $M'$ when the functions
    \begin{align*}
      \alpha:Q\to \powersetplus{Q'},
        \quad
        \iota:\Sigma\to \Sigma',
        \quad
        \zeta: \Gamma'\to \Gamma,
    \end{align*}
    satisfy the two conditions below for every $q\in Q$, every $q'\in \alpha(q)$, and 
    every $\sigma\in \Sigma$.
    \begin{align*}
        \text{I)}\quad & \delta'\big(q',\iota(\sigma)\big)\in \alpha\big(\delta(q,\sigma)\big)\\
        \text{II)}\quad&\zeta\circ \theta'\big(q',\iota(\sigma)\big)=\theta\big(q,\sigma\big)
    \end{align*}
    If an assignment of $M$ into $M'$ exists, then we say 
    that $M'$ is a \emph{realisation} of $M$ or, equivalently,
    that $M'$ \emph{realises} $M$.
  \end{definition}

\begin{fact}[\S1.3 of \cite{hartmanis}]\label{realisation-trans}
    If $M$ is a realisation of $M'$ and $M'$ is a realisation of $M''$, then $M$ realies $M''$.
\end{fact}

\begin{theorem}[Theorem 1.5 in \S 1.3 of \cite{hartmanis}]
  \label{thm:realize-factor}
  If $M'=\tuple{Q',\Sigma',\delta',\Gamma',\theta'}$ is a realisation of 
  $M=\tuple{Q,\Sigma, \delta,\Gamma,\theta}$ through an assignment 
  $\ptuple{\alpha,\iota,\zeta}$, then for all $q_0\in Q$, 
  $\vec{w} \defeq \seq{w}{1}{\ell} \in \Sigma^+$,
  and $q_0'\in \alpha(q_0)$, it holds that
    \begin{align*}
      \theta\big(M(q_0,\vec{w}),w_\ell) =
      \zeta\circ \theta'\big( M'(q_0',\iota(\vec{w})), \iota(w_\ell)\big),
    \end{align*}
    i.e., $M(q_0,\vec{w})=\zeta\circ M'\big(q'_0,\iota(\vec{w})\big)$.
\end{theorem}

\begin{theorem}[Krohn-Rhodes, cf.~Theorem~7.9 in \citep{hartmanis}] \label{thm:krohn-rhodes}
  Let $A = \tuple{Q,\Sigma,\delta}$ be a semiautomaton, and 
  let $\mathbf{G}$ be the set of simple groups that are factors of a subgroup of the transition
  monoid of $A$.
  Then, $A$ is realised by a cascade $A_1 \ltimes \cdots \ltimes A_m$ where each $A_i$ 
  is either a two-state identity-reset semiautomaton or a $\groupsty{G}$-semiautomaton for some 
  $\groupsty{G} \in \mathbf{G}$.
\end{theorem}

\begin{corollary}
  Let $A = \tuple{Q,\Sigma,\delta}$ be a semiautomaton.
  Then, $A$ is realised by a cascade $A_1 \ltimes \cdots \ltimes A_m$ where each $A_i$ 
  is an identity-reset or a permutation semiautomaton.
\end{corollary}
\begin{proof}
  Implies by the Krohn-Rhodes theorem since
  the state transformations of any 
  $\groupsty{G}$-semiautomaton $A_\groupsty{G} = \tuple{G,G,\gamma}$ are bijections.
  In fact, any transformation is given by $\gamma_g(g') = \gamma(g',g)$, which is invertible since every
  group element admits an inverse element, and hence it is a bijection.
\end{proof}

\subsection{Metric Spaces and Ambient Spaces}

We follow the standard terminology for metric spaces and topology
(cf.~\citealp{willard2004general}), consistently with
Metric Automata Theory~\citep{mat}.
In particular, spaces are regarded intrinsically as metric spaces.
When useful, we additionally regard a metric space as a subspace of a
designated \emph{ambient space}; this allows us to refer to structures of
the ambient space, such as Euclidean coordinates and norms, without requiring
the space itself to coincide with the ambient space.

\begin{closingdefinition}[Metrics and metric spaces]
  Let $X$ be a set.
  A \emph{metric} on $X$ is a function
  \begin{equation*}
    d_X : X\times X\to\nonnegreals
  \end{equation*}
  such that, for every $x,y,z\in X$,
  \begin{align*}
    d_X(x,y)=0
    &\quad\Longleftrightarrow\quad
    x=y,
    &
    d_X(x,y)
    &=
    d_X(y,x),
    &
    d_X(x,z)
    &\leq
    d_X(x,y)+d_X(y,z).
  \end{align*}
  A \emph{metric space} is a set equipped with a metric.

  We generally identify a metric space with its underlying set and omit its
  metric from the notation whenever the metric is clear from the context.
\end{closingdefinition}

\begin{closingdefinition}[Metric subspaces and ambient spaces]
  \label{def:ambient-metric-space}
  Let $X$ and $\Omega_X$ be metric spaces.
  We say that $X$ is a \emph{metric subspace} of $\Omega_X$ if
  \begin{equation*}
    X\subseteq\Omega_X
    \qquad\text{and}\qquad
    d_X
    =
    \restr{d_{\Omega_X}}{X\times X}.
  \end{equation*}

  When $\Omega_X$ is a designated metric space of which $X$ is a metric
  subspace, we call $\Omega_X$ the \emph{ambient space} of $X$.
  Accordingly, whenever we write
  \begin{equation*}
    X\subseteq\Omega_X,
  \end{equation*}
  with $\Omega_X$ specified as the ambient space of $X$, the metric on $X$
  is understood to be the metric induced by $\Omega_X$.
\end{closingdefinition}

\begin{closingremark}[Intrinsic and ambient structure]
  A metric space need not have a distinguished ambient space, and an ambient
  realisation need not be unique.
  Thus, the ambient space is additional structure associated with the
  presentation of a space rather than part of the abstract notion of metric
  space.

  Metric notions involving only points of $X$, such as distances and
  boundedness, are intrinsic to $X$.
  When an ambient space $\Omega_X$ is specified, additional notions referring
  to the surrounding space---for example Euclidean coordinates, norms,
  derivatives, and Lebesgue measure---are understood with respect to
  $\Omega_X$.
\end{closingremark}

\begin{closingdefinition}[Balls and bounded sets]
  Let $X$ be a metric space.
  For every $x\in X$ and $r\in\nonnegreals$, the \emph{open ball} and
  \emph{closed ball} of radius $r$ centred at $x$ are, respectively,
  \begin{align*}
    \openballop{x,r}{X}
    &\defeq
    \big\{
      y\in X
      :
      d_X(x,y)<r
    \big\},
    &
    \closedballop{x,r}{X}
    &\defeq
    \big\{
      y\in X
      :
      d_X(x,y)\leq r
    \big\}.
  \end{align*}

  A set $K\subseteq X$ is \emph{bounded} if there exist
  $x\in X$ and $r\in\nonnegreals$ such that
  \begin{equation*}
    K\subseteq\overline{B}_X(x,r).
  \end{equation*}
\end{closingdefinition}

\begin{closingdefinition}[Open and closed sets]
  Let $X$ be a metric space and let $A\subseteq X$.
  The set $A$ is \emph{open in $X$} if, for every $x\in A$, there exists
  $r>0$ such that
  \begin{equation*}
    B_X(x,r)\subseteq A.
  \end{equation*}
  The set $A$ is \emph{closed in $X$} if $X\setminus A$ is open in $X$.
\end{closingdefinition}

\begin{closingdefinition}[Interior]
  Let $X$ be a metric space and let $A\subseteq X$.
  The \emph{interior of $A$ in $X$} is the set
  \begin{equation*}
    \interior[X](A)
    \defeq
    \left\{
    x\in A
    ;\middle|;
    \exists r>0:
    B_X(x,r)\subseteq A
    \right\}.
  \end{equation*}

  If $X$ has ambient space $\Omega_X$, we distinguish the intrinsic interior
  $\operatorname{int}*X(A)$ from the \emph{ambient interior}
  \begin{equation*}
    \operatorname{int}*{\Omega_X}(A),
  \end{equation*}
  namely the interior of $A$ when regarded as a subset of $\Omega_X$.
\end{closingdefinition}

\begin{closingremark}[Interior and openness]
  For every $A\subseteq X$,
  \begin{equation*}
    A\text{ is open in }X
    \qquad\Longleftrightarrow\qquad
    \interior[X](A)=A.
  \end{equation*}
  In general, when $X$ is a proper metric subspace of an ambient space
  $\Omega_X$, one may have
  \begin{equation*}
    \operatorname{int}*X(A)
    \neq
    \operatorname{int}*{\Omega_X}(A).
  \end{equation*}
\end{closingremark}

\begin{closingremark}[Subspace and ambient topology]
  Let $\Omega_X$ be the ambient space of $X$.
  Openness and closedness of a subset $A\subseteq X$ are understood relative
  to $X$ unless the ambient space is explicitly mentioned.
  In particular, $A$ may be open or closed in $X$ without being open or
  closed in $\Omega_X$.

  By contrast, since $X$ carries the metric induced by $\Omega_X$, a subset
  $A\subseteq X$ is bounded in $X$ if and only if it is bounded as a subset
  of $\Omega_X$.
\end{closingremark}

\begin{closingdefinition}[Product metric spaces]
  Let $X_1,\ldots,X_k$ be metric spaces.
  Their Cartesian product
  \begin{equation*}
    X_1\times\cdots\times X_k
  \end{equation*}
  is equipped with the product metric
  \begin{equation*}
    d\big(
      \tuple{x_1,\ldots,x_k},
      \tuple{x'_1,\ldots,x'_k}
    \big)
    \defeq
    \sqrt{
      \sum_{i=1}^k
      d_{X_i}(x_i,x'_i)^2
    }.
  \end{equation*}

  If each $X_i$ has ambient space $\Omega_{X_i}$, we take
  \begin{equation*}
    \Omega_{X_1\times\cdots\times X_k}
    \defeq
    \Omega_{X_1}\times\cdots\times\Omega_{X_k}
  \end{equation*}
  as the canonical ambient space of their product.
\end{closingdefinition}

\begin{closingdefinition}[Euclidean spaces]
  For every $n\in\posnaturals$, the \emph{Euclidean $n$-space} is
  $\reals^n$ equipped with the Euclidean metric
  \begin{equation*}
    d(x,y)
    \defeq
    \|x-y\|
    =
    \sqrt{
      \sum_{i=1}^n
      (x_i-y_i)^2
    }.
  \end{equation*}
  As usual, we omit the metric and simply write $\reals^n$.

  A metric space $X$ has a \emph{Euclidean ambient space} if its designated
  ambient space satisfies
  \begin{equation*}
    \Omega_X=\reals^n
  \end{equation*}
  for some $n\in\posnaturals$.
  In this case,
  \begin{equation*}
    X\subseteq\Omega_X=\reals^n,
  \end{equation*}
  and $X$ carries the Euclidean metric restricted to $X$.
\end{closingdefinition}

\begin{closingremark}[Products of Euclidean ambient spaces]
  Under the canonical identification
  \begin{equation*}
    \reals^{n_1}\times\cdots\times\reals^{n_k}
    \cong
    \reals^{n_1+\cdots+n_k},
  \end{equation*}
  the product metric above coincides with the Euclidean metric.
  Hence finite products of spaces having Euclidean ambient spaces again have
  Euclidean ambient spaces.
\end{closingremark}

\subsection{Measure Theory}
\label{subsec:measure-theoretic-preliminaries}

We recall the basic notions from measure theory that we use throughout the paper.
In particular, whenever a space is contained in a Euclidean ambient space,
measurability and almost-everywhere statements are understood with respect to
the Lebesgue measure of that ambient space.

We follow the standard terminology for measurable spaces, Lebesgue measure,
null sets, almost-everywhere properties, and product measures
(cf.~\citealp{folland1999real}).

\begin{closingdefinition}[Measurable spaces and measures]
  A \emph{$\sigma$-algebra} on a set $X$ is a collection
  $\Sigma\subseteq 2^X$ such that:
  \begin{itemize}
    \item
      $X\in\Sigma$;
    \item
      if $A\in\Sigma$, then $X\setminus A\in\Sigma$;
    \item
      if $(A_i)_{i\in\posnaturals}$ is a sequence of sets in $\Sigma$, then
      \begin{equation*}
        \bigcup_{i\in\posnaturals} A_i
        \in
        \Sigma.
      \end{equation*}
  \end{itemize}
  The pair $\tuple{X,\Sigma}$ is called a \emph{measurable space},
  and the elements of $\Sigma$ are called \emph{measurable sets}.

  A \emph{measure} on $\tuple{X,\Sigma}$ is a function
  \begin{equation*}
    \mu:\Sigma\to[0,\infty]
  \end{equation*}
  such that
  $\mu(\varnothing)=0$
  and, for every sequence of pairwise-disjoint sets
  $(A_i)_{i\in\posnaturals}$ in $\Sigma$,
  \begin{equation*}
    \mu\left(
      \bigcup_{i\in\posnaturals}A_i
    \right)
    =
    \sum_{i\in\posnaturals}\mu(A_i).
  \end{equation*}
  The triple
  $\tuple{X,\Sigma,\mu}$
  is called a \emph{measure space}.
\end{closingdefinition}

\begin{closingdefinition}[Measurable functions]
  Let
  $\tuple{X,\Sigma_X}$
  and
  $\tuple{Y,\Sigma_Y}$
  be measurable spaces.
  A function
  \begin{equation*}
    f:X\to Y
  \end{equation*}
  is \emph{measurable} if
  \begin{equation*}
    f^{-1}(A)\in\Sigma_X
    \qquad\qquad
    \forall A\in\Sigma_Y.
  \end{equation*}
\end{closingdefinition}

\begin{closingdefinition}[Lebesgue measure]
  For every $n\in\posnaturals$,
  we denote by
  $\mathcal{L}^n$
  the $\sigma$-algebra of Lebesgue-measurable subsets of
  $\reals^n$
  and by
  \begin{equation*}
    \lambda_n:\mathcal{L}^n\to[0,\infty]
  \end{equation*}
  the $n$-dimensional \emph{Lebesgue measure}.

  Accordingly,
  \begin{equation*}
    \tuple{\reals^n,\mathcal{L}^n,\lambda_n}
  \end{equation*}
  is the $n$-dimensional Lebesgue measure space.
\end{closingdefinition}

\begin{closingdefinition}[Ambient Lebesgue measure]
  \label{def:ambient-lebesgue-measure}
  Let $X$ be a metric space having Euclidean ambient space
  \begin{equation*}
    \Omega_X=\reals^n,
  \end{equation*}
  and suppose that $X$ is Lebesgue-measurable in $\Omega_X$.

  The \emph{Lebesgue $\sigma$-algebra of $X$} is
  \begin{equation*}
    \mathcal{L}_X
    \defeq
    \big\{
      A\subseteq X
      :
      A\in\mathcal{L}^n
    \big\},
  \end{equation*}
  and the \emph{ambient Lebesgue measure on $X$} is the restriction
  \begin{equation*}
    \lambda_X:
    \mathcal{L}_X
    \to
    [0,\infty],
    \qquad
    \lambda_X(A)
    \defeq
    \lambda_n(A).
  \end{equation*}

  When the ambient space is clear from the context, we simply refer to
  Lebesgue measurability and Lebesgue measure on $X$.
\end{closingdefinition}

\begin{closingdefinition}[Null sets and full-measure subsets]
  \label{def:null-full-measure}
  Let $X$ have a Euclidean ambient space and be Lebesgue-measurable in it.

  A measurable set
  $N\subseteq X$
  is \emph{null} or \emph{Lebesgue-null} if
  \begin{equation*}
    \lambda_X(N)=0.
  \end{equation*}

  Given measurable sets
  \begin{equation*}
    A\subseteq B\subseteq X,
  \end{equation*}
  we say that
  $A$ has \emph{full measure in $B$} if
  \begin{equation*}
    \lambda_X(B\setminus A)=0.
  \end{equation*}
\end{closingdefinition}

\begin{closingdefinition}[Almost-everywhere properties]
  \label{def:almost-everywhere}
  Let $B$ be a measurable subset of a space having a Euclidean ambient space,
  and let $\mathcal{P}(x)$ be a property defined for $x\in B$.

  We say that
  $\mathcal{P}$ holds \emph{almost everywhere in $B$},
  abbreviated \emph{a.e. in $B$},
  if the set
  \begin{equation*}
    \big\{
      x\in B
      :
      \mathcal{P}(x)\text{ does not hold}
    \big\}
  \end{equation*}
  is Lebesgue-null.

  Equivalently,
  $\mathcal{P}$ holds a.e. in $B$
  if there exists a full-measure subset
  $\hat B\subseteq B$
  such that
  $\mathcal{P}(x)$ holds for every
  $x\in\hat B$.
\end{closingdefinition}

\begin{closingfact}[Restriction of full-measure subsets]
  \label{fact:full-measure-restriction}
  Let
  $A\subseteq B\subseteq X$
  be measurable sets such that
  $A$ has full measure in $B$.
  Then, for every measurable
  $C\subseteq B$,
  the set
  \begin{equation*}
    A\cap C
  \end{equation*}
  has full measure in $C$.
\end{closingfact}
\begin{proof}
  We have
  \begin{equation*}
    C\setminus(A\cap C)
    =
    C\setminus A
    \subseteq
    B\setminus A.
  \end{equation*}
  Since
  $B\setminus A$
  is Lebesgue-null,
  so is
  $C\setminus(A\cap C)$.
\end{proof}

\begin{closingfact}[Measurability of bounded truncations]
  \label{fact:measurable-truncations}
  Let $X$ be a Lebesgue-measurable subset of a Euclidean ambient space
  $\Omega_X$.
  Then, for every $r\in\nonnegreals$,
  its bounded truncation
  \begin{equation*}
    \truncation{X}{r}
    =
    X
    \cap
    \overline B_{\Omega_X}(0,r)
  \end{equation*}
  is Lebesgue-measurable and has finite Lebesgue measure.
\end{closingfact}
\begin{proof}
  The closed Euclidean ball
  $\overline B_{\Omega_X}(0,r)$
  is Lebesgue-measurable and bounded.
  Hence its intersection with the measurable set $X$ is measurable.
  Furthermore,
  \begin{equation*}
    \truncation{X}{r}
    \subseteq
    \overline B_{\Omega_X}(0,r),
  \end{equation*}
  and every bounded Lebesgue-measurable subset of a finite-dimensional
  Euclidean space has finite Lebesgue measure.
\end{proof}

\begin{closingfact}[Lebesgue measure on product spaces]
  \label{fact:lebesgue-product-measure}
  Let
  \begin{equation*}
    \Omega_X=\reals^m,
    \qquad
    \Omega_Y=\reals^n.
  \end{equation*}
  Under the canonical identification
  \begin{equation*}
    \Omega_X\times\Omega_Y
    \cong
    \reals^{m+n},
  \end{equation*}
  the Lebesgue measure on the product ambient space coincides with the
  product measure
  \begin{equation*}
    \lambda_{m+n}
    =
    \lambda_m\times\lambda_n.
  \end{equation*}

  In particular, if
  $N\subseteq\Omega_X$
  is Lebesgue-null and
  $A\subseteq\Omega_Y$
  is Lebesgue-measurable, then
  \begin{equation*}
    N\times A
  \end{equation*}
  is Lebesgue-null in
  $\Omega_X\times\Omega_Y$.
\end{closingfact}

\begin{closingfact}[Products preserve full measure]
  \label{fact:product-full-measure}
  Let
  $A\subseteq B\subseteq X$
  be measurable sets such that
  $A$ has full measure in $B$, and let
  $C\subseteq Y$
  be measurable.
  Then
  \begin{equation*}
    A\times C
  \end{equation*}
  has full measure in
  \begin{equation*}
    B\times C.
  \end{equation*}
\end{closingfact}
\begin{proof}
  We have
  \begin{equation*}
    (B\times C)\setminus(A\times C)
    =
    (B\setminus A)\times C.
  \end{equation*}
  Since
  $B\setminus A$
  is Lebesgue-null,
  Fact~\ref{fact:lebesgue-product-measure}
  implies that
  $(B\setminus A)\times C$
  is Lebesgue-null.
\end{proof}

\begin{closingremark}[Ambient dimension and almost-everywhere properties]
  \label{rem:ambient-dimension-ae}
  Lebesgue measure is defined with respect to the designated Euclidean
  ambient space.
  Consequently, if
  \begin{equation*}
    X\subseteq\Omega_X=\reals^n
  \end{equation*}
  is itself Lebesgue-null in $\Omega_X$, then every measurable subset of
  $X$ has full measure in $X$.

  For this reason, parameter spaces on which almost-everywhere properties
  are considered are understood to be represented in their independent
  Euclidean coordinates, rather than as lower-dimensional subsets of
  unnecessarily larger ambient spaces.
\end{closingremark}

\subsection{Metric Automata Theory}

Metric Automata Theory (MAT) \citep{mat} extends the applicability of AAT to arbitrary dynamical systems, with
automata as a special case.
We introduce the relevant notions.

A \emph{path} in $X$ from $a$ to $b$ is a continuous map $\gamma: [0,1]\to X$ such that
$\gamma(0)=a$ and $\gamma(1)=b$. We can define a relation $\sim_X$, where $a\sim_X b$ when there is
a path in $X$ from $a$ to $b$. This relation is an equivalence, partitioning $X$ into 
equivalence classes $[x]_{\sim_X} = \{ x' \in X \mid x' \sim_X x \}$ for $x \in X$,  
called \emph{(path-connected) components}. For space $X$, we denote the set of
its equivalence classes by $\ol X$. 
Path-connectedness is preserved by every continuous function $f:X \to Y$, and hence
the function $\ol f : \ol X \to \ol Y$ with $\ol f(\ol x) = \ol{f(x)}$ is well-defined.

A metric space $X$ is \emph{\etafinite{}} if it is a finite union of compact, path-connected
components.
The continuous image of an \etafinite{} space is \etafinite{} (Lemma~18 of \citep{mat}).
The Cartesian-product $X \times Y$ space of \etafinite{} spaces is \etafinite{}. 
The $\eta$-components of $X \times Y$ are the products $\eta$-components of $X$ and
$\eta$-components of $Y$ (Lemma~19 of \citep{mat}).

\begin{lemma}[Lemma~22 of \citep{mat}] \label{lemma:mat-eta-constant-continuous}
  Let $X$ be an \etafinite{} space and $\Sigma$ a finite alphabet. 
  Then a function $f : X \to \Sigma$ is continuous if and only if it is constant on the
  $\eta$-components of $X$.
\end{lemma}

Dynamics $D = \tuple{X,U,f}$ and system $S = \tuple{X,U,f,x_0,Y,h}$ are 
\emph{\etafinite{}} if $X$ and $U$ are \etafinite{}; in such case,
the semiautomaton $\mathcal{C}_\eta(D) \defeq \tuple{\ol{X}, \ol{U}, \ol{f}}$
and the automaton $\mathcal{C}_\eta(S) \defeq \tuple{\ol{X}, \ol{U}, \ol{f}, [x_0]_{\sim_X}, \ol{Y},
\ol{h}}$ are well-defined and they are called the \emph{$\eta$-canonical semiautomaton of $D$} and the
\emph{$\eta$-canonical automaton} of $S$, respectively.

\begin{definition}\label{def:encoder-decoder}
  An \emph{encoder} is a continuous function $\enc: \Sigma\to U$ where $\Sigma$ is a finite alphabet
  called \emph{input alphabet}.
  An \emph{decoder} is a continuous function $\dec: Y \to \Gamma$ where $\Gamma$ is a finite
  alphabet called \emph{output alphabet}.
\end{definition}

\begin{definition}[Definition~1 of \citep{mat}] \label{def:implement}
  Given an encoder
  $\enc: \Sigma\to U$ and a decoder $\dec: Y\to \Gamma$,
  we say that a system $S$ 
  \emph{implements a function $F:\Sigma^+\to \Gamma$ with encoder $\enc$ and decoder $\dec$}
  if
  $F(w)= \dec\!\big(S(\enc(w))\big)$,
  for every $w\in \Sigma^+$,
  where $\enc(w)\in U^+$ applies $\enc$ element-wise.
  We also say that $S$ \emph{can-implement}~$F$ if it implements~$F$ for some choice of $\enc$ and
  $\dec$.
  When $\Gamma = \{ 0, 1\}$, we say that $S$ \emph{recognises} a language $L$ if it implements its
  indicator function $\mathbb{I}_L$, and that $S$ \emph{can-recognise} $L$ if it can-implement
  its characteristic function $\mathbb{I}_L$.
\end{definition}

\begin{theorem}[Theorem~1 of \citep{mat}]\label{thm:canon-implement}
    An \etafinite{} system $S$ can-implement the same functions as its canonical automaton, which are necessarily regular.
\end{theorem}

\begin{closingdefinition}[Definition~23 of \citep{mat}]
  Given \etafinite{} dynamics $D,D'$,
  dynamics $D'$ are a realisation of $D$ if 
  $\mathcal{M}(\ceta{D'})$ is a realisation of $\mathcal{M}(\ceta{D})$ of $D$.
  Given automata $A,A'$,
  automaton $A'$ is a realisation of $A$ if the associated machine $M_{A'}$ is a
  realisation of of the associated machine $M_A$ via an assignment $(\alpha,\iota, \zeta)$, and
  the respective initial states $x_0',x_0$ are such that $x_0'\in \alpha (x_0)$.
  Given an \etafinite{} system $S'$ and a system $S$,
  system $S'$ is a realisation of $S$ if $\ceta{S'}$ is a realisation of $\ceta{S}$.
\end{closingdefinition}

\begin{lemma}[Lemma~29 of \citep{mat}]\label{lemma:realizability-universal}
    Suppose that semiautomaton $D'$ is a realisation of semiautomaton $D$. Then 
    \begin{inlineenum}
      \item
        for any machine $M$ with dynamics $D$, the canonical machine $\mathcal{M}(D')$ of $D'$ is a
        realisation of $M$;
  \item
    for any automaton $A$ with dynamics $D$, an initial state can be picked for $\mathcal{M}(D')$
    such that the resulting automaton is a realisation of $A$.
    \end{inlineenum}
\end{lemma}

\begin{theorem}[Theorem~30 of \citep{mat}]\label{realize-to-implement}
    Let $S$ and $S'$ be \etafinite{} systems. %
    If $\ceta{S'}$ is a realisation of $\ceta{S}$, then $S'$ can implement all the functions
    that $S$ can implement.
\end{theorem}

\begin{theorem}[Theorem~31 of \citep{mat}]\label{thm:realize-dynamics}
    Let $D,D'$ be \etafinite{} dynamics. Suppose that $D'$ is a realisation of $D$. Then any
    function implemented by a system with dynamics $D$ can be implemented by some system $S'$ with
    dynamics $D'$ and an identity output function.
\end{theorem}

In the above Theorem~\ref{thm:realize-dynamics}, the remark on the identity output function is not
present in the original statement of the theorem, but it is clear from its proof.

\subsubsection{Aperiodicity and eta-convergence}
\label{sec:prelim-aperiodicity}

\begin{definition}[Convergence, Definition~31 \citep{mat}]
  For a \etafinite{} space $X$, we say a sequence $(x_n)_{n \geq 1}$ with $x_n \in X$
  $\eta$-converges in $X$, if eventually all its terms lie in the same $\eta$-component of $X$.
\end{definition}

\begin{definition}[Aperiodicity, Definition~32 \citep{mat}]
  We say that \etafinite{} dynamics $D=\tuple{ X, U, f}$ are \emph{aperiodic} if, for every $x_0\in
  X$ and every input sequence $(u_n)_{n\geq 1}$ that is $\eta$-convergent in $U$, we have that the
  corresponding state sequence $(x_n)_{n\geq 1}$ is  $\eta$-convergent in $X$.
\end{definition}

\begin{theorem}[Theorem~50 of \citep{mat}] \label{thm:aperiodic}
    \etafinite{} dynamics are aperiodic if and only if their $\eta$-canonical semiautomaton is
    group-free.
\end{theorem}

\subsection{Definite Languages}
\label{subsec:preliminaries--definite-languages}

\begin{closingdefinition}[Definite languages, cf.~\citep{holzerj14}]
  \label{def:positive-margin-definite-language}
  A language
  $L$
  over a finite alphabet
  $\Sigma$
  is a
  \emph{definite language}
  if
  \begin{equation*}
    L = L_1 \cup \{ uv \mid u \in \Sigma^*,\, v \in L_2 \}
  \end{equation*}
  for some finite languages $L_1$ and $L_2$ over $\Sigma$.
  We denote the class of definite languages by $\mathsf{Def}$.
\end{closingdefinition}

\begin{closinglemma}[Fixed-suffix characterisation of definite languages]
  \label{lemma:positive-margin-definite-suffix-characterisation}
  Let $L$ be a language over a finite alphabet $\Sigma$.
  Then
  $L$
  is definite if and only if there exists
  $N\in\posnaturals$
  such that, for every
  $v,w\in\Sigma^*$
  and
  $z\in\Sigma^N$,
  \begin{equation}
    \label{eq:positive-margin-definite-suffix-characterisation}
    vz\in L
    \quad\Longleftrightarrow\quad
    wz\in L.
  \end{equation}
\end{closinglemma}

\begin{proof}
  Suppose first that
  $L$
  is definite.
  Then there exist finite languages
  $L_1,L_2$
  such that
  \begin{equation*}
    L = L_1 \cup \{ uv \mid u \in \Sigma^*,\, v \in L_2 \}.
  \end{equation*}
  Since
  $L_1$
  and
  $L_2$
  are finite,
  there exists
  $N\in\posnaturals$
  larger than the length of every string in
  $L_1$
  and at least as large as the length of every string in
  $L_2$.
  Let
  $v,w\in\Sigma^*$
  and
  $z\in\Sigma^N$.
  Since
  $|vz|,|wz|\geq N$,
  neither
  $vz$
  nor
  $wz$
  belongs to
  $L_1$.
  Furthermore, membership in
  $\{ ps \mid p \in  \Sigma^*, \, s \in L_2 \}$
  is determined by a suffix $s$ whose length is at most $N$,
  and therefore is determined completely by the common suffix $z$.
  Hence
  \begin{equation*}
    vz\in L
    \quad\Longleftrightarrow\quad
    wz\in L.
  \end{equation*}

  Conversely,
  suppose that
  $N\in\posnaturals$
  satisfies
  Equation~\eqref{eq:positive-margin-definite-suffix-characterisation}.
  Define
  \begin{align*}
    L_1
    \defeq
    L
    \cap
    \allseqs{\Sigma}[N-1]
    \qquad\qquad
    L_2
    \defeq
    L
    \cap
    \Sigma^N.
  \end{align*}
  Since
  $\Sigma$
  is finite,
  both
  $L_1$
  and
  $L_2$
  are finite.

  Let
  $x\in\Sigma^*$
  have length at least
  $N$
  and write
  \begin{equation*}
    x
    =
    vz,
    \qquad
    z\in\Sigma^N.
  \end{equation*}
  Taking
  $w$
  to be the empty string in
  Equation~\eqref{eq:positive-margin-definite-suffix-characterisation}
  gives
  \begin{equation*}
    x\in L
    \quad\Longleftrightarrow\quad
    z\in L
    \quad\Longleftrightarrow\quad
    z\in L_2.
  \end{equation*}
  Therefore
  \begin{equation*}
    x\in L
    \quad\Longleftrightarrow\quad
    x\in\Sigma^*L_2.
  \end{equation*}
  The words of length strictly less than
  $N$
  are accounted for by
  $L_1$.
  Hence
  \begin{equation*}
    L
    =
    L_1
    \cup
    \{ ps \mid p \in \Sigma^*,\, s \in L_2 \},
  \end{equation*}
  and thus
  $L$
  is definite.
\end{proof}

\subsection{Graphs, Isomorphisms, and Renaming of Nodes}
\label{subsec:graph-isomorphisms}

\todo{}

We introduce the notions of graph isomorphism and equality up to a
renaming of nodes that we use for computation graphs.
Since the edges of a computation graph have an ordered vector of source
nodes, we work with directed multi-edge graphs whose edges have the same
form.

\begin{closingdefinition}[Graph]
  \label{def:graph}
  A \emph{graph} is a pair
  \begin{align*}
    G = (V,E),
  \end{align*}
  where:
  \begin{itemize}
    \item
      $V$ is a finite set of \emph{nodes};
    \item
      $E \subseteq V^* \times V$
      is a set of \emph{directed multi-edges}.
  \end{itemize}
  For an edge
  \begin{align*}
    \big(
      (u_1,\ldots,u_k),
      v
    \big)
    \in E,
  \end{align*}
  we write
  \begin{align*}
    (u_1,\ldots,u_k)
    \to
    v.
  \end{align*}
  The vector
  $(u_1,\ldots,u_k)$
  is the \emph{source vector} of the edge,
  $v$ is its \emph{target},
  and
  $k$ is its \emph{arity}.

  The ordering of the source vector is part of the graph structure:
  in general,
  \begin{align*}
    (u_1,u_2)\to v
    \qquad\text{and}\qquad
    (u_2,u_1)\to v
  \end{align*}
  are distinct edges.

  The \emph{dependency relation} of $G$ is the binary relation
  $\prec_G \,\subseteq\, V\times V$ defined by
  \begin{align*}
    u \prec_G v
    \quad\Longleftrightarrow\quad
    \exists
    (u_1,\ldots,u_k)\to v \in E
    \ \text{such that}\
    u=u_i
    \ \text{for some}\
    i\in\iival{1}{k}.
  \end{align*}
  The graph $G$ is \emph{acyclic} if the transitive closure of
  $\prec_G$ is irreflexive.
\end{closingdefinition}

\begin{closingdefinition}[Renaming of nodes]
  \label{def:graph-renaming}
  Let
  \begin{align*}
    G=(V,E)
  \end{align*}
  be a graph and let
  \begin{align*}
    \rho : V \to V'
  \end{align*}
  be a bijection.
  We extend $\rho$ componentwise to node-vectors by setting
  \begin{align*}
    \rho(u_1,\ldots,u_k)
    \defeq
    \big(
      \rho(u_1),
      \ldots,
      \rho(u_k)
    \big).
  \end{align*}

  The \emph{renaming of $G$ induced by $\rho$} is the graph
  \begin{align*}
    \rho(G)
    \defeq
    \big(
      V',
      \rho(E)
    \big),
  \end{align*}
  where
  \begin{align*}
    \rho(E)
    \defeq
    \left\{
      \rho(u_1,\ldots,u_k)
      \to
      \rho(v)
      \;\middle|\;
      (u_1,\ldots,u_k)\to v \in E
    \right\}.
  \end{align*}
  Thus, a renaming changes only the identities of the nodes and leaves
  the complete edge structure, including the ordering of the source nodes
  of every edge, unchanged.
\end{closingdefinition}

\begin{closingdefinition}[Graph isomorphism]
  \label{def:graph-isomorphism}
  Let
  \begin{align*}
    G=(V,E),
    \qquad
    G'=(V',E')
  \end{align*}
  be graphs.
  A \emph{graph isomorphism from $G$ to $G'$} is a bijection
  \begin{align*}
    \rho : V \to V'
  \end{align*}
  such that
  \begin{align*}
    (u_1,\ldots,u_k)\to v \in E
    \quad\Longleftrightarrow\quad
    \big(
      \rho(u_1),
      \ldots,
      \rho(u_k)
    \big)
    \to
    \rho(v)
    \in E'
  \end{align*}
  for every
  $k\in\naturals$,
  $(u_1,\ldots,u_k)\in V^k$,
  and
  $v\in V$.

  The graphs $G$ and $G'$ are \emph{isomorphic}, denoted
  \begin{align*}
    G \cong G',
  \end{align*}
  if there exists a graph isomorphism from $G$ to $G'$.
\end{closingdefinition}

\begin{proposition}
  \label{prop:graph-isomorphism-renaming}
  Let $G$ and $G'$ be graphs.
  Then,
  \begin{align*}
    G \cong G'
    \quad\Longleftrightarrow\quad
    \exists\,\rho
    \text{ a bijection between their node sets such that }
    \rho(G)=G'.
  \end{align*}
\end{proposition}
\begin{proof}
  Immediate by
  Definitions~\ref{def:graph-renaming}
  and~\ref{def:graph-isomorphism}.
\end{proof}

\begin{closingremark}
  Accordingly, we say that two graphs are
  \emph{equal up to a renaming of nodes}
  precisely when they are isomorphic.
  Thus,
  \begin{align*}
    G \cong G'
  \end{align*}
  means that $G$ and $G'$ have exactly the same graph structure and differ
  at most in the names assigned to their nodes.
\end{closingremark}

\clearpage{}%

\appendixpart{part:notions}{Preliminary Notions and Results}

\clearpage{}%

\section{Additional on Dynamics and Systems}
\label{sec:additional-on-dynamics-and-systems}
\label{appendix:additional-on-dynamics-and-systems}

\localtableofcontents

\subsection{Parametric Dynamical Systems}

\begin{closingdefinition}
  \emph{Parametric dynamics} are a tuple
  $D = \ctuple{X, U, f}{\Theta}$
  where:
  \begin{itemize}
    \item
      $X$, $U$, and $\Theta$ are metric spaces called the
      \emph{state space}, \emph{input space}, and \emph{parameter space} of $D$,
      respectively;
    \item
      $f: X \times U \times \Theta \to X$ is a continuous function called
      the \emph{dynamics function} of $D$.
  \end{itemize}

  For $\theta \in \Theta$, the \emph{instance of $D$ with parameter values $\theta$}
  is the dynamics
  \begin{equation*}
    \pinst{D}{\theta} \defeq \tuple{X, U, \ppinst{f}{\theta}},
    \quad\ \text{where}\ \quad
    \ppinst{f}{\theta}(x,u) \defeq \fpinst{f}{x,u}{\theta}.
  \end{equation*}
\end{closingdefinition}

\begin{closingdefinition} \label{def:parametric-system}
  A \emph{parametric (dynamical) system} is a tuple
  $S
  =
  \ctuple{
  X,U,f,x^\mathrm{init},Y,h
  }{
  \Theta_f,\Theta_h
  }
  $
  where:
  \begin{itemize}
    \item
      $X$, $U$, and $Y$ are metric spaces called the
      \emph{state space}, \emph{input space}, and \emph{output space} of $S$,
      respectively;
    \item
      $\Theta_f$ and $\Theta_h$ are metric spaces called the
      \emph{dynamics-function parameter space} and
      \emph{output-specific parameter space} of $S$, respectively;
      and
      $\Theta
      \defeq
      \Theta_f\times\Theta_h$
      is the parameter space of $S$;
    \item
        $f:
        X\times U\times\Theta_f
        \to
        X$
      is a continuous function called the \emph{dynamics function} of $S$;
    \item
      $x^\mathrm{init}\in X$ is called the \emph{initial state} of $S$;
    \item
        $h:
        X\times U\times X\times \Theta_h
        \to
        Y$
      is a continuous function called the \emph{output function} of $S$.
  \end{itemize}

  We call
  $\dynop{S}
  \defeq
  \ctuple{X,U,f}{\Theta_f}$
  the \emph{dynamics of $S$}.

  For
  $\theta_f\in\Theta_f$
  and
  $\theta_h\in\Theta_h$,
  the \emph{instance of $S$ with parameter values
  $\ptuple{\theta_f,\theta_h}$} is
  \begin{equation*}
    \pinst{S}{\theta_f,\theta_h}
    \defeq
    \tuple{
    X,
    U,
    \ppinst{f}{\theta_f},
    x^\mathrm{init},
    Y,
    \ppinst{h}{\theta_h}
    },
  \end{equation*}
  where
  \begin{align*}
    \ppinst{f}{\theta_f}(x,u)
    &\defeq
    \fpinst{f}{x,u}{\theta_f},
    &
    \ppinst{h}{\theta_h}(x,u,x')
    &\defeq
    \bigfpinst{h}{x,u,x'}{\theta_h}.
  \end{align*}
\end{closingdefinition}

\begin{closingfact}
  In the context of Definition~\ref{def:parametric-system},
  we have
  $\dynop{\pinst{S}{\theta_f,\theta_h}}
  =
  \pinst{\dynop{S}}{\theta_f}$.
\end{closingfact}

\begin{closingdefinition}[Input sequences]
  Let $D$ be parametric dynamics and let $S$ be a parametric system.

  An \emph{input sequence for $D$} is an input sequence for its instances $\pinst{D}{\theta}$,
  and
  an \emph{input sequence for $S$} is an input sequence for its instances $\pinst{S}{\theta}$.
\end{closingdefinition}

\begin{closingdefinition}[State sequences]
  Let $D = \ctuple{X,U,f}{\Theta}$ be parametric dynamics,
  let $\theta \in \Theta$,
  let $u$ be an input sequence for $D$,
  and
  let $x_0 \in X$.

  The \emph{state sequence of $D$ on input $u$ when started from $x_0$
  with parameters $\theta$}
  is the state sequence of $\pinst{D}{\theta}$ on input $u$ when started from $x_0$.

  Parametric dynamics $D$ define the function
  \begin{equation*}
    \repval{D} : X \times U^* \times \Theta \to X
  \end{equation*}
  given by
  \begin{equation*}
    \fpinst{\repval{D}}{x_0,u}{\theta}
    \defeq
    \repval{\pinst{D}{\theta}}(x_0,u).
  \end{equation*}

  When $D$ are the dynamics of a parametric system $S$ having $x_0$ as its initial state,
  \emph{the state sequence of $S$ with dynamics-function parameters $\theta$}
  is the state sequence of $D$ on input $u$ when started from $x_0$
  with parameters $\theta$.
\end{closingdefinition}

\begin{closingdefinition}[Output sequences]
  Let $S = \ctuple{X,U,f,x^\mathrm{init},Y,h}{\Theta}$ be a parametric system,
  let $\theta \in \Theta$,
  and
  let $u$ be an input sequence for $S$.
  The \emph{output sequence of $S$ on input $u$ with parameter values $\theta$}
  is the output sequence of $\pinst{S}{\theta}$ on input $u$.
  Parametric system $S$ defines the function
  \begin{equation*}
    \repval{S} : U^+ \times \Theta \to Y,
    \qquad
    \fpinst{\repval{S}}{u}{\theta}
    \defeq
    \repval{\pinst{S}{\theta}}(u).
  \end{equation*}
\end{closingdefinition}

\subsection{Composition Operators for Dynamics and Systems}

\begin{closingdefinition}[Cascade composition of dynamics, Part~I]
  Given a pair of dynamics $D_1$ and $D_2$ along with a continuous function $g$,
  \begin{align*}
    D_1 = \tuple{X_1,U,f_1},
    \qquad
    D_2 = \tuple{X_2,Z,f_2},
    \quad\qquad
    g:
    X_1\times U\times X_1
    \to
    Z,
  \end{align*}
  the \emph{cascade composition of $D_1$ and $D_2$ with connection $g$}
  is the dynamics
  \begin{equation*}
    D_1\ltimes_g D_2
    \defeq
    \tuple{X_1\times X_2,U,f},
  \end{equation*}
  having dynamics function
  \begin{align*}
    f\big(\tuple{x_1,x_2},u\big)
    \defeq
    \bigtuple{
      x'_1,
      f_2(x_2,z)
    }
    \quad\;\ \text{for}\ \quad
    x'_1 \defeq f_1(x_1,u)
    \quad\text{and}\quad
    z \defeq g(x_1,u,x'_1).
  \end{align*}
\end{closingdefinition}

\begin{closingdefinition}[Cascade composition of dynamics, Part~II]
  Given dynamics
  \begin{align*}
    D_1 = \tuple{X_1,U,f_1},
    \qquad
    D_2 = \tuple{X_2,U\times X_1,f_2},
  \end{align*}
  their \emph{cascade composition} are the dynamics
  \begin{equation*}
    D_1\ltimes D_2
    \defeq
    D_1\ltimes_\pi D_2
    \qquad\text{where}\qquad
    \pi(x_1,u,x'_1)
    \defeq
    \tuple{u,x_1}.
  \end{equation*}
\end{closingdefinition}

\begin{closingdefinition}[Parallel composition of dynamics]
  Given dynamics
  \begin{align*}
    D_1 = \tuple{X_1,U,f_1},
    \qquad
    D_2 = \tuple{X_2,U,f_2},
  \end{align*}
  their \emph{parallel composition} are the dynamics
  \begin{equation*}
    D_1\parallel D_2
    \defeq
    \tuple{X_1\times X_2,U,f},
  \end{equation*}
  having dynamics function
  \begin{equation*}
    f\big(\tuple{x_1,x_2},u\big)
    \defeq
    \bigtuple{
    f_1(x_1,u),
    f_2(x_2,u)
    }.
  \end{equation*}
\end{closingdefinition}

\begin{closingfact}[Parallel composition as a cascade]
  \label{fact:parametric-parallel-composition-as-cascade}
  \todo{}
  Let
  \begin{equation*}
    D_1
    =
    \tuple{X_1,U,f_1},
    \qquad
    D_2
    =
    \tuple{X_2,U,f_2}
  \end{equation*}
  be dynamics.

  Define
  \begin{equation*}
    \widetilde D_2
    \defeq
    \tuple{
      X_2,
      U\times X_1,
      \widetilde f_2
    },
  \end{equation*}
  where
  \begin{equation*}
    \widetilde f_2
    \big(
      x_2,
      \tuple{u,x_1}
    \big)
    \defeq
    f_2(x_2,u).
  \end{equation*}

  Then
  \begin{equation*}
    D_1
    \parallel
    D_2
    =
    D_1
    \ltimes
    \widetilde D_2.
  \end{equation*}
\end{closingfact}
\begin{proof}
  Both sides have state space
  $X_1\times X_2$
  and input space
  $U$.
  Moreover,
  for every
  $x_1\in X_1$,
  $x_2\in X_2$,
  and
  $u\in U$,
  their dynamics map
  $\tuple{\tuple{x_1,x_2},u}$
  to
  \begin{equation*}
    \bigtuple{
      f_1(x_1,u),
      f_2(x_2,u)
    }.
  \end{equation*}
  Hence the two dynamics are equal.
\end{proof}

\begin{closingdefinition}[Cascade composition of systems]
  Given systems
  \begin{align*}
    S_1 &= \tuple{X_1,U,f_1,x^\mathrm{init}_1,Z,h_1},
        &
    S_2 &= \tuple{X_2,Z,f_2,x^\mathrm{init}_2,Y,h_2},
  \end{align*}
  their \emph{cascade composition} is the system
  \begin{equation*}
    S_1\ltimes S_2
    \defeq
    \bigtuple{
    X_1\times X_2,
    U,
    f,
    \tuple{x^\mathrm{init}_1,x^\mathrm{init}_2},
    Y,
    h
    },
  \end{equation*}
  having dynamics function
  \begin{align*}
    f\bigptuple{\tuple{x_1,x_2},u}
    \defeq
    \bigtuple{
      x'_1,
      f_2\ptuple{x_2,z}
    }
    \quad\;\text{for}\quad
    x'_1
    \defeq
    f_1\ptuple{x_1,u}
    \quad\text{and}\quad
    z
    \defeq
    h_1\ptuple{x_1,u,x'_1},
  \end{align*}
  and output function
  \begin{equation*}
    h\bigptuple{
    \tuple{x_1,x_2},
    u,
    \tuple{x'_1,x'_2}
    }
    \defeq
    h_2\ptuple{x_2,z,x'_2}
    \quad\;\text{for}\quad
    z
    \defeq
    h_1\ptuple{x_1,u,x'_1}.
  \end{equation*}
\end{closingdefinition}

\begin{closingfact}[Dynamics induced by layer output functions]
  For every pair of systems
  \begin{align*}
    S_1
    =
    \tuple{X_1,U,f_1,x^\mathrm{init}_1,Z,g},
    \qquad
    S_2
    =
    \tuple{X_2,Z,f_2,x^\mathrm{init}_2,Y,h},
  \end{align*}
  we have
  \begin{equation*}
    \dynop{S_1\ltimes S_2}
    =
    D_1\ltimes_g D_2.
  \end{equation*}
\end{closingfact}

\subsection{Composition Operators for Parametric Dynamics and Systems}

\todo{}

\subsubsection{Cascade Composition with Connecting Functions}

\begin{closingdefinition}[Cascade composition of parametric dynamics with a connection]
  \label{def:cascade-composition-parametric-dynamics-connection}
  Given parametric dynamics
  \begin{equation*}
    D_1
    =
    \ctuple{
      X_1,
      U,
      f_1
    }{
      \Theta_1
    },
    \qquad
    D_2
    =
    \ctuple{
      X_2,
      Z,
      f_2
    }{
      \Theta_2
    },
  \end{equation*}
  and a parametric connecting function
  \begin{equation*}
    g
    :
    X_1
    \times
    U
    \times
    X_1
    \times
    \Theta_g
    \to
    Z,
  \end{equation*}
  the
  \emph{cascade composition of $D_1$ and $D_2$ with connection $g$}
  is the parametric dynamics
  \begin{equation*}
    D_1
    \ltimes_g
    D_2
    \defeq
    \ctuple{
      X_1\times X_2,
      U,
      f
    }{
      \Theta_1
      \times
      \Theta_g
      \times
      \Theta_2
    },
  \end{equation*}
  having dynamics function
  \begin{equation*}
    \bigfpinst{f}{
      \tuple{x_1,x_2},
      u
    }{
      \theta_1,
      \theta_g,
      \theta_2
    }
    \defeq
    \bigtuple{
      x'_1,
      x'_2
    },
  \end{equation*}
  where
  \begin{align*}
    x'_1
    &\defeq
    \fpinst{f_1}{x_1,u}{\theta_1},
    \\
    z
    &\defeq
    \bigfpinst{g}{
      x_1,u,x'_1
    }{
      \theta_g
    },
    \\
    x'_2
    &\defeq
    \fpinst{f_2}{x_2,z}{\theta_2}.
  \end{align*}
\end{closingdefinition}

\subsubsection{Simple Cascade Composition}

\begin{closingdefinition}[Simple cascade composition of parametric dynamics]
  \label{def:simple-cascade-composition-parametric-dynamics}
  Given parametric dynamics
  \begin{equation*}
    D_1
    =
    \ctuple{
      X_1,
      U,
      f_1
    }{
      \Theta_1
    },
    \qquad
    D_2
    =
    \ctuple{
      X_2,
      X_1\times U\times X_1,
      f_2
    }{
      \Theta_2
    },
  \end{equation*}
  their
  \emph{simple cascade composition}
  is the parametric dynamics
  \begin{equation*}
    D_1
    \ltimes
    D_2
    \defeq
    \ctuple{
      X_1\times X_2,
      U,
      f
    }{
      \Theta_1\times\Theta_2
    },
  \end{equation*}
  having dynamics function
  \begin{equation*}
    \bigfpinst{f}{
      \tuple{x_1,x_2},
      u
    }{
      \theta_1,\theta_2
    }
    \defeq
    \bigtuple{
      x'_1,
      x'_2
    },
  \end{equation*}
  where
  \begin{align*}
    x'_1
    &\defeq
    \fpinst{f_1}{x_1,u}{\theta_1},
    \\
    x'_2
    &\defeq
    \fpinst{f_2}{
      x_2,
      \tuple{x_1,u,x'_1}
    }{
      \theta_2
    }.
  \end{align*}

  We also write
  $f=f_1\ltimes f_2$.
\end{closingdefinition}

\begin{closingremark}
  A simple cascade is a special case of a cascade with a connection.
  In fact, let
  \begin{equation*}
    \pi
    :
    X_1
    \times
    U
    \times
    X_1
    \to
    X_1
    \times
    U
    \times
    X_1
  \end{equation*}
  be the identity connecting function
  \begin{equation*}
    \pi(x_1,u,x'_1)
    \defeq
    \tuple{x_1,u,x'_1}.
  \end{equation*}
  Then
  \begin{equation*}
    D_1\ltimes D_2
    =
    D_1\ltimes_\pi D_2.
  \end{equation*}
\end{closingremark}

\begin{closingdefinition}[Cascade composition of parametric systems]
  \label{def:cascade-composition-parametric-systems}
  Given a pair of parametric systems
  \begin{equation*}
    S_1
    =
    \ctuple{
      X_1,
      U,
      f_1,
      x^\mathrm{init}_1,
      Z,
      h_1
    }{
      \Theta_{f_1},
      \Theta_{h_1}
    },
    \qquad
    S_2
    =
    \ctuple{
      X_2,
      Z,
      f_2,
      x^\mathrm{init}_2,
      Y,
      h_2
    }{
      \Theta_{f_2},
      \Theta_{h_2}
    },
  \end{equation*}
  and a choice
  \begin{equation*}
    z^\mathrm{init}\in Z,
  \end{equation*}
  their
  \emph{cascade composition}
  is the parametric system
  \begin{equation*}
    S_1
    \ltimes
    S_2
    \defeq
    \bigctuple{
      X_1\times Z\times X_2,
      U,
      f,
      \tuple{x^\mathrm{init}_1,z^\mathrm{init},x^\mathrm{init}_2},
      Y,
      h
    }{
      \Theta_f,
      \Theta_h
    },
  \end{equation*}
  having parameter spaces
  \begin{equation*}
    \Theta_f
    \defeq
    \Theta_{f_1}
    \times
    \Theta_{h_1}
    \times
    \Theta_{f_2},
    \qquad
    \Theta_h
    \defeq
    \Theta_{h_2},
  \end{equation*}
  dynamics function
  \begin{equation*}
    \bigfpinst{f}{
      \tuple{x_1,\bar z,x_2},
      u
    }{
      \theta_{f_1},
      \theta_{h_1},
      \theta_{f_2}
    }
    \defeq
    \bigtuple{
      x'_1,
      z,
      x'_2
    },
  \end{equation*}
  and output function
  \begin{equation*}
    \bigfpinst{h}{
      \tuple{x_1,\bar z,x_2},
      u,
      \tuple{x'_1,z,x'_2}
    }{
      \theta_{h_2}
    }
    \defeq
    \bigfpinst{h_2}{
      x_2,
      z,
      x'_2
    }{
      \theta_{h_2}
    },
  \end{equation*}
  where
  \begin{align*}
    x'_1
    &\defeq
    \fpinst{f_1}{
      x_1,u
    }{
      \theta_{f_1}
    },
    \\
    z
    &\defeq
    \bigfpinst{h_1}{
      x_1,u,x'_1
    }{
      \theta_{h_1}
    },
    \\
    x'_2
    &\defeq
    \fpinst{f_2}{
      x_2,z
    }{
      \theta_{f_2}
    }.
  \end{align*}
\end{closingdefinition}

\begin{closingremark}[Stored connecting value]
  \label{rem:parametric-system-cascade-stored-connection}
  The additional state component in
  \begin{equation*}
    X_1\times Z\times X_2
  \end{equation*}
  stores the connecting value produced during the current transition.
  Its previous value
  $\bar z$
  is not used by the dynamics and is overwritten at every step.

  Consequently,
  the choice of
  $z^\mathrm{init}$
  does not affect any state component other than the unused connecting
  component at time zero, and does not affect the input--output behaviour on
  nonempty input words.

  Storing the connecting value makes the cascade compatible with
  Definition~\ref{def:parametric-system}:
  the composite output function depends only on its output-specific parameters
  $\Theta_h=\Theta_{h_2}$,
  while the parameters
  $\Theta_{h_1}$
  required to generate the connecting value belong only to the composite
  dynamics-function parameter space
  $\Theta_f$.
\end{closingremark}

\begin{closingfact}[Dynamics induced by parametric layer output functions]
  For every pair of parametric systems
  $S_1$
  and
  $S_2$
  as in
  Definition~\ref{def:cascade-composition-parametric-systems},
  the output function
  \begin{equation*}
    h_1:
    X_1\times U\times X_1\times\Theta_{h_1}
    \to
    Z
  \end{equation*}
  is a parametric connecting function with parameter space
  $\Theta_{h_1}$.

  In particular, let
  \begin{equation*}
    \widehat D
    \defeq
    \dynop{S_1}
    \ltimes_{h_1}
    \dynop{S_2}
    =
    \ctuple{
      X_1\times X_2,
      U,
      \widehat f
    }{
      \Theta_{f_1}
      \times
      \Theta_{h_1}
      \times
      \Theta_{f_2}
    }.
  \end{equation*}
  Then, for every
  \begin{equation*}
    x_1\in X_1,
    \qquad
    \bar z\in Z,
    \qquad
    x_2\in X_2,
    \qquad
    u\in U,
  \end{equation*}
  and every admissible parameter value,
  if
  \begin{equation*}
    \bigfpinst{\widehat f}{
      \tuple{x_1,x_2},u
    }{
      \theta_{f_1},
      \theta_{h_1},
      \theta_{f_2}
    }
    =
    \tuple{x'_1,x'_2},
  \end{equation*}
  then
  \begin{equation*}
    \bigfpinst{f}{
      \tuple{x_1,\bar z,x_2},u
    }{
      \theta_{f_1},
      \theta_{h_1},
      \theta_{f_2}
    }
    =
    \bigtuple{
      x'_1,
      \bigfpinst{h_1}{x_1,u,x'_1}{\theta_{h_1}},
      x'_2
    }.
  \end{equation*}
  Thus the first and third state components evolve exactly according to the
  connected cascade
  $\dynop{S_1}\ltimes_{h_1}\dynop{S_2}$,
  while the middle component stores its connecting value.
\end{closingfact}

\subsubsection{Parallel Composition}

\begin{closingdefinition}[Parallel composition of parametric dynamics]
  Given a pair of parametric dynamics
  \begin{align*}
    D_1
    &=
    \ctuple{
      X_1,
      U,
      f_1
    }{
      \Theta_1
    },
    \qquad
    D_2
    =
    \ctuple{
      X_2,
      U,
      f_2
    }{
      \Theta_2
    },
  \end{align*}
  their
  \emph{parallel composition}
  is the parametric dynamics
  \begin{equation*}
    D_1
    \parallel
    D_2
    \defeq
    \ctuple{
      X_1\times X_2,
      U,
      f
    }{
      \Theta_1\times\Theta_2
    },
  \end{equation*}
  having dynamics function
  \begin{equation*}
    \bigfpinst{f}{
      \tuple{x_1,x_2},
      u
    }{
      \theta_1,\theta_2
    }
    \defeq
    \bigtuple{
      \fpinst{f_1}{x_1,u}{\theta_1},
      \fpinst{f_2}{x_2,u}{\theta_2}
    }.
  \end{equation*}
  We also write
  $f=f_1\parallel f_2$.
\end{closingdefinition}

\begin{closingfact}[Parallel composition as a cascade]
  \label{fact:parallel-composition-as-cascade}
  \todo{}
  Let
  \begin{equation*}
    D_1
    =
    \ctuple{
      X_1,
      U,
      f_1
    }{
      \Theta_1
    },
    \qquad
    D_2
    =
    \ctuple{
      X_2,
      U,
      f_2
    }{
      \Theta_2
    }
  \end{equation*}
  be parametric dynamics.

  Define
  \begin{equation*}
    \widetilde D_2
    \defeq
    \ctuple{
      X_2,
      X_1\times U\times X_1,
      \widetilde f_2
    }{
      \Theta_2
    },
  \end{equation*}
  where
  \begin{equation*}
    \fpinst{\widetilde f_2}{
      x_2,
      \tuple{x_1,u,x'_1}
    }{
      \theta_2
    }
    \defeq
    \fpinst{f_2}{
      x_2,
      u
    }{
      \theta_2
    }.
  \end{equation*}
  Then
  \begin{equation*}
    D_1
    \parallel
    D_2
    =
    D_1
    \ltimes
    \widetilde D_2.
  \end{equation*}
\end{closingfact}
\begin{proof}
  Both sides have state space
  $X_1\times X_2$,
  input space
  $U$,
  and parameter space
  $\Theta_1\times\Theta_2$.
  Moreover,
  for every
  $x_1\in X_1$,
  $x_2\in X_2$,
  $u\in U$,
  $\theta_1\in\Theta_1$,
  and
  $\theta_2\in\Theta_2$,
  letting
  \begin{equation*}
    x'_1
    \defeq
    \fpinst{f_1}{x_1,u}{\theta_1},
  \end{equation*}
  the simple cascade has dynamics
  \begin{align*}
    \bigtuple{
      x'_1,
      \fpinst{\widetilde f_2}{
        x_2,
        \tuple{x_1,u,x'_1}
      }{
        \theta_2
      }
    }
    &=
    \bigtuple{
      \fpinst{f_1}{x_1,u}{\theta_1},
      \fpinst{f_2}{x_2,u}{\theta_2}
    },
  \end{align*}
  which is precisely the dynamics of
  $D_1\parallel D_2$.
\end{proof}

\subsubsection{Input Precomposition}

\begin{closingdefinition}[Parametric input precomposition]
  \label{def:parametric-input-precomposition}
  Let
  \begin{equation*}
    D
    =
    \ctuple{
      X,
      Z,
      f
    }{
      \Theta
    }
  \end{equation*}
  be parametric dynamics, and let
  \begin{equation*}
    q
    :
    V
    \times
    \Theta_q
    \to
    Z
  \end{equation*}
  be a parametric continuous function.

  The
  \emph{input precomposition of $D$ by $q$}
  is the parametric dynamics
  \begin{equation*}
    \inputprecomposition{q}{D}
    \defeq
    \ctuple{
      X,
      V,
      f_q
    }{
      \Theta_q
      \times
      \Theta
    },
  \end{equation*}
  having dynamics function
  \begin{equation*}
    \bigfpinst{f_q}{
      x,v
    }{
      \theta_q,\theta
    }
    \defeq
    \fpinst{f}{
      x,
      \fpinst{q}{v}{\theta_q}
    }{
      \theta
    }.
  \end{equation*}
\end{closingdefinition}

\subsubsection{Output Extensions}
\label{subsubsec:output-extension}

\begin{closingdefinition}[Output extension of parametric dynamics]
  \label{def:parametric-output-extension}
  \label{def:parametric-output-extension--dyn-par}
  Let
  \begin{equation*}
    D
    \,=\,
    \bigctuple{
      X,\,
      U,\,
      f
    }{\,
      \Theta_f
    }
  \end{equation*}
  be parametric dynamics,
  let $x_0\in X$,
  let $Y$ and $\Theta_h$ be metric spaces,
  and let
  \begin{equation*}
    h
    \,:\,
    (X \times U \times X)
    \times
    \Theta_h
    \,\to\,
    Y
  \end{equation*}
  be a continuous function.

  The
  \emph{output extension of $\tuple{D,x_0}$ by $h$}
  is the parametric system
  \begin{equation*}
    \outputextension{D}{x_0}{h}
    \,\defeq\,
    \bigctuple{
      X,\,
      U,\,
      f,\,
      x_0,\,
      Y,\,
      h
    }{\,
      \Theta_f,\,
      \Theta_h
    }.
  \end{equation*}
  We call it an \emph{analytic output extension} if $h$ is an analytic Euclidean function.
\end{closingdefinition}

\begin{closingdefinition}[Output extension of systems]
  \label{def:parametric-output-extension--system}
  Given a system and a continuous function, 
  given respectively by
  \begin{equation*}
    S \,=\, \bigtuple{ X,\, U,\, f,\, x_0,\, Y,\, h }\,,
    \qquad\qquad
    \tilde{h} : Y \times \to \tilde{Y}\,,
    \qquad\qquad
  \end{equation*}
  the \emph{output extension of $S$ by $\tilde{h}$}
  is the system
  \begin{equation*}
    \outputextensionSystem{S}{\tilde{h}}
    \,\defeq\, 
    \bigtuple{ X,\, U,\, f,\, x_0,\, Y,\; \tilde{h} \circ h }\,.
  \end{equation*}
  We call it an \emph{analytic output extension} if $\tilde{h}$ is an analytic Euclidean function.
\end{closingdefinition}

\begin{closingdefinition}[Output extension of parametric systems]
  \label{def:parametric-output-extension--par-system}
  Given a parametric system and a continuous parametric function, 
  given respectively by
  \begin{equation*}
    S \,=\, \bigctuple{ X,\, U,\, f,\, x_0,\, Y,\, h }{\,\Theta_f,\mkern2mu \Theta_h}\,,
    \qquad\qquad
    \tilde{h} : Y \times \Theta_{\tilde{h}} \to \tilde{Y}\,,
    \qquad\qquad
  \end{equation*}
  the \emph{output extension of $S$ by $\tilde{h}$}
  is the parametric system
  \begin{equation*}
    \outputextensionSystem{S}{\tilde{h}}
    \,\defeq\, 
    \bigctuple{ 
      X,\, 
      U,\, 
      f,\, 
      x_0,\, 
      Y,\; 
      \mkern1mu{\tilde{h} \mkern-1mu \circ \mkern-1mu h}
      \,\mkern1mu
    }{\,
      \Theta_f, \mkern2mu \Theta_h \mkern-2mu\times\mkern-1mu \Theta_{\tilde{h}}
    }\,.
  \end{equation*}
  We call it an \emph{analytic output extension} if $\tilde{h}$ is an analytic Euclidean function.
\end{closingdefinition}

\begin{closingfact}[Parametric systems as output extensions]
  \label{fact:parametric-system-as-output-extension}
  Every
  parameteric system
  \begin{equation*}
    S \,=\, \bigctuple{ X,\, U,\, f,\, x_0,\, Y,\, h }{\Theta_f,\, \Theta_h}\,,
    \qquad\qquad
    \tilde{h} : Y \times \Theta_{\tilde{h}} \to \tilde{Y}\,.
  \end{equation*}
  can be expressed as the output extension
  \begin{equation*}
    S
    \,=\,
    \outputextension{S}{x_0}{h}\,.
  \end{equation*}
\end{closingfact}

\subsection{Properties of Composition Operators for Parametric Dynamics and Systems}

\todo{}

\begin{restatable}[Dynamics composition commutes with instantiation]
  {proposition}{compositioncommuteswithinstantiation}
  \label{prop:composition-commutes-with-instantiation}
  Let
  \begin{equation*}
    D_1
    =
    \ctuple{
      X_1,
      U,
      f_1
    }{
      \Theta_1
    },
    \qquad
    D_2
    =
    \ctuple{
      X_2,
      Z,
      f_2
    }{
      \Theta_2
    },
  \end{equation*}
  and let
  \begin{equation*}
    g
    :
    X_1
    \times
    U
    \times
    X_1
    \times
    \Theta_g
    \to
    Z
  \end{equation*}
  be a parametric connecting function.

  Then:
  \begin{enumerate}
    \item
      for every
      $\theta_1\in\Theta_1$,
      $\theta_g\in\Theta_g$,
      and
      $\theta_2\in\Theta_2$,
      \begin{equation}
        \label{eq:connected-cascade-instantiation}
        \Pinst{
          D_1\ltimes_gD_2
        }{
          \theta_1,\theta_g,\theta_2
        }
        =
        \pinst{D_1}{\theta_1}
        \ltimes_{\ppinst{g}{\theta_g}}
        \pinst{D_2}{\theta_2};
      \end{equation}

    \item
      if
      \begin{equation*}
        Z
        =
        X_1\times U\times X_1,
      \end{equation*}
      then, for every
      $\theta_1\in\Theta_1$
      and
      $\theta_2\in\Theta_2$,
      \begin{equation}
        \label{eq:simple-cascade-instantiation}
        \Pinst{
          D_1\ltimes D_2
        }{
          \theta_1,\theta_2
        }
        =
        \pinst{D_1}{\theta_1}
        \ltimes
        \pinst{D_2}{\theta_2};
      \end{equation}

    \item
      if
      $Z=U$,
      then, for every
      $\theta_1\in\Theta_1$
      and
      $\theta_2\in\Theta_2$,
      \begin{equation}
        \label{eq:parallel-composition-instantiation}
        \Pinst{
          D_1\parallel D_2
        }{
          \theta_1,\theta_2
        }
        =
        \pinst{D_1}{\theta_1}
        \parallel
        \pinst{D_2}{\theta_2}.
      \end{equation}
  \end{enumerate}
\end{restatable}

\begin{proof}
  For
  Equation~\eqref{eq:connected-cascade-instantiation},
  fix
  \begin{equation*}
    \theta_1\in\Theta_1,
    \qquad
    \theta_g\in\Theta_g,
    \qquad
    \theta_2\in\Theta_2.
  \end{equation*}
  By definition, the dynamics function of
  \begin{equation*}
    \Pinst{
      D_1\ltimes_gD_2
    }{
      \theta_1,\theta_g,\theta_2
    }
  \end{equation*}
  maps
  $\tuple{\tuple{x_1,x_2},u}$
  to
  \begin{equation*}
    \bigtuple{
      x'_1,
      \fpinst{f_2}{x_2,z}{\theta_2}
    },
  \end{equation*}
  where
  \begin{equation*}
    x'_1
    \defeq
    \fpinst{f_1}{x_1,u}{\theta_1},
    \qquad
    z
    \defeq
    \bigfpinst{g}{
      x_1,u,x'_1
    }{
      \theta_g
    }.
  \end{equation*}
  Since
  \begin{equation*}
    \ppinst{g}{\theta_g}(x_1,u,x'_1)
    =
    \bigfpinst{g}{
      x_1,u,x'_1
    }{
      \theta_g
    },
  \end{equation*}
  this is exactly the dynamics function of
  \begin{equation*}
    \pinst{D_1}{\theta_1}
    \ltimes_{\ppinst{g}{\theta_g}}
    \pinst{D_2}{\theta_2}.
  \end{equation*}

  For
  Equation~\eqref{eq:simple-cascade-instantiation},
  both sides map
  $\tuple{\tuple{x_1,x_2},u}$
  to
  \begin{equation*}
    \bigtuple{
      x'_1,
      \fpinst{f_2}{
        x_2,
        \tuple{x_1,u,x'_1}
      }{
        \theta_2
      }
    },
    \qquad
    x'_1
    \defeq
    \fpinst{f_1}{x_1,u}{\theta_1}.
  \end{equation*}

  Finally,
  Equation~\eqref{eq:parallel-composition-instantiation}
  follows because both sides map
  $\tuple{\tuple{x_1,x_2},u}$
  to
  \begin{equation*}
    \bigtuple{
      \fpinst{f_1}{x_1,u}{\theta_1},
      \fpinst{f_2}{x_2,u}{\theta_2}
    }.
  \end{equation*}
\end{proof}

\begin{restatable}[System cascade composition commutes with instantiation]
  {proposition}{systemcompositioncommuteswithinstantiation}
  \label{prop:system-composition-commutes-with-instantiation}
  Let
  \begin{equation*}
    S_1
    =
    \ctuple{
      X_1,
      U,
      f_1,
      x^\mathrm{init}_1,
      Z,
      h_1
    }{
      \Theta_{f_1},
      \Theta_{h_1}
    },
    \qquad
    S_2
    =
    \ctuple{
      X_2,
      Z,
      f_2,
      x^\mathrm{init}_2,
      Y,
      h_2
    }{
      \Theta_{f_2},
      \Theta_{h_2}
    },
  \end{equation*}
  and fix the same
  $z^\mathrm{init}\in Z$
  in the parametric and nonparametric cascade constructions.

  Then, for every
  \begin{equation*}
    \theta_{f_1}\in\Theta_{f_1},
    \qquad
    \theta_{h_1}\in\Theta_{h_1},
    \qquad
    \theta_{f_2}\in\Theta_{f_2},
    \qquad
    \theta_{h_2}\in\Theta_{h_2},
  \end{equation*}
  we have
  \begin{equation}
    \label{eq:system-cascade-instantiation}
    \Pinst{
      S_1\ltimes S_2
    }{
      \tuple{
        \theta_{f_1},
        \theta_{h_1},
        \theta_{f_2}
      },
      \theta_{h_2}
    }
    =
    \pinst{S_1}{
      \theta_{f_1},
      \theta_{h_1}
    }
    \ltimes
    \pinst{S_2}{
      \theta_{f_2},
      \theta_{h_2}
    }.
  \end{equation}
\end{restatable}

\begin{proof}
  Fix admissible parameter values.
  Both sides have state space
  \begin{equation*}
    X_1\times Z\times X_2,
  \end{equation*}
  input space
  $U$,
  initial state
  \begin{equation*}
    \tuple{
      x^\mathrm{init}_1,
      z^\mathrm{init},
      x^\mathrm{init}_2
    },
  \end{equation*}
  and output space
  $Y$.

  Let
  \begin{equation*}
    \tuple{
      x_1,
      \bar z,
      x_2
    }
    \in
    X_1\times Z\times X_2,
    \qquad
    u\in U.
  \end{equation*}
  On both sides,
  \begin{align*}
    x'_1
    &\defeq
    \fpinst{f_1}{
      x_1,u
    }{
      \theta_{f_1}
    },
    \\
    z
    &\defeq
    \bigfpinst{h_1}{
      x_1,u,x'_1
    }{
      \theta_{h_1}
    },
    \\
    x'_2
    &\defeq
    \fpinst{f_2}{
      x_2,z
    }{
      \theta_{f_2}
    },
  \end{align*}
  and hence the dynamics function maps
  \begin{equation*}
    \tuple{
      \tuple{x_1,\bar z,x_2},
      u
    }
  \end{equation*}
  to
  \begin{equation*}
    \tuple{
      x'_1,
      z,
      x'_2
    }.
  \end{equation*}

  Likewise, on both sides the output function maps
  \begin{equation*}
    \bigtuple{
      \tuple{x_1,\bar z,x_2},
      u,
      \tuple{x'_1,z,x'_2}
    }
  \end{equation*}
  to
  \begin{equation*}
    \bigfpinst{h_2}{
      x_2,z,x'_2
    }{
      \theta_{h_2}
    }.
  \end{equation*}
  Therefore the two instantiated systems coincide.
\end{proof}

\begin{restatable}[Underlying dynamics of a system cascade]
  {corollary}{underlyingdynamicsofasystemcascade}
  \label{cor:underlying-dynamics-system-cascade}
  Let
  $S_1$
  and
  $S_2$
  be parametric systems as above, and fix
  \begin{equation*}
    \theta_{f_1}\in\Theta_{f_1},
    \qquad
    \theta_{h_1}\in\Theta_{h_1},
    \qquad
    \theta_{f_2}\in\Theta_{f_2},
    \qquad
    \theta_{h_2}\in\Theta_{h_2}.
  \end{equation*}

  Let
  \begin{equation*}
    \widehat D
    \defeq
    \bigdynop{
      \pinst{S_1}{
        \theta_{f_1},
        \theta_{h_1}
      }
    }
    \ltimes_{
      \ppinst{h_1}{\theta_{h_1}}
    }
    \bigdynop{
      \pinst{S_2}{
        \theta_{f_2},
        \theta_{h_2}
      }
    },
  \end{equation*}
  and let
  $\widehat f$
  denote its dynamics function.

  Let
  $\bar f$
  denote the dynamics function of
  \begin{equation*}
    \bigdynop{
      \Pinst{
        S_1\ltimes S_2
      }{
        \tuple{
          \theta_{f_1},
          \theta_{h_1},
          \theta_{f_2}
        },
        \theta_{h_2}
      }
    }.
  \end{equation*}

  Then, for every
  \begin{equation*}
    x_1\in X_1,
    \qquad
    \bar z\in Z,
    \qquad
    x_2\in X_2,
    \qquad
    u\in U,
  \end{equation*}
  if
  \begin{equation*}
    \widehat f
    \big(
      \tuple{x_1,x_2},
      u
    \big)
    =
    \tuple{x'_1,x'_2},
  \end{equation*}
  then
  \begin{equation}
    \label{eq:underlying-system-cascade-storage}
    \bar f
    \big(
      \tuple{x_1,\bar z,x_2},
      u
    \big)
    =
    \bigtuple{
      x'_1,
      \ppinst{h_1}{\theta_{h_1}}
      \big(
        x_1,u,x'_1
      \big),
      x'_2
    }.
  \end{equation}

  In particular, the first and third state components evolve exactly as
  $\widehat D$.
\end{restatable}

\begin{proof}
  By
  Proposition~\ref{prop:system-composition-commutes-with-instantiation},
  the instantiated system cascade uses
  \begin{align*}
    x'_1
    &=
    \fpinst{f_1}{
      x_1,u
    }{
      \theta_{f_1}
    },
    \\
    z
    &=
    \ppinst{h_1}{\theta_{h_1}}
    \big(
      x_1,u,x'_1
    \big),
    \\
    x'_2
    &=
    \fpinst{f_2}{
      x_2,z
    }{
      \theta_{f_2}
    }.
  \end{align*}
  The pair
  $\tuple{x'_1,x'_2}$
  is precisely the update produced by
  $\widehat D$,
  while the additional middle component stores
  $z$.
  This gives
  Equation~\eqref{eq:underlying-system-cascade-storage}.
\end{proof}

\begin{corollary}[Simple dynamics cascades induced by system cascades]
  \label{cor:simple-dynamics-cascades-as-system-cascades}
  Let
  $S_1$
  and
  $S_2$
  be as above, and suppose that
  \begin{equation}
    \label{eq:system-output-simple-connection}
    \bigfpinst{h_1}{
      x_1,u,x'_1
    }{
      \theta_{h_1}
    }
    =
    \tuple{
      x_1,u,x'_1
    }
  \end{equation}
  for every admissible
  \begin{equation*}
    x_1,
    \qquad
    u,
    \qquad
    x'_1,
    \qquad
    \theta_{h_1}.
  \end{equation*}

  Then the first and third state components of
  \begin{equation*}
    \bigdynop{
      \Pinst{
        S_1\ltimes S_2
      }{
        \tuple{
          \theta_{f_1},
          \theta_{h_1},
          \theta_{f_2}
        },
        \theta_{h_2}
      }
    }
  \end{equation*}
  evolve exactly according to the simple cascade
  \begin{equation*}
    \bigdynop{
      \pinst{S_1}{
        \theta_{f_1},
        \theta_{h_1}
      }
    }
    \ltimes
    \bigdynop{
      \pinst{S_2}{
        \theta_{f_2},
        \theta_{h_2}
      }
    }.
  \end{equation*}
\end{corollary}

\begin{proof}
  Under
  Equation~\eqref{eq:system-output-simple-connection},
  the instantiated connecting function
  $\ppinst{h_1}{\theta_{h_1}}$
  is the identity connecting function
  \begin{equation*}
    \pi(x_1,u,x'_1)
    \defeq
    \tuple{
      x_1,u,x'_1
    }.
  \end{equation*}
  Hence the connected cascade appearing in
  Corollary~\ref{cor:underlying-dynamics-system-cascade}
  is precisely the simple cascade.
\end{proof}

\begin{restatable}[Instantiation of iterated compositions]
  {corollary}{instantiationofiteratedcompositions}
  \label{cor:instantiation-iterated-compositions}
  Let
  \begin{equation*}
    D_1,\ldots,D_n
  \end{equation*}
  be composable parametric dynamics with parameter spaces
  \begin{equation*}
    \Theta_1,\ldots,\Theta_n,
  \end{equation*}
  and let
  \begin{equation*}
    \theta
    =
    \tuple{
      \theta_1,\ldots,\theta_n
    }
    \in
    \Theta_1\times\cdots\times\Theta_n.
  \end{equation*}

  Then
  \begin{equation*}
    \Pinst{
      D_1\ltimes\cdots\ltimes D_n
    }{
      \theta
    }
    =
    \pinst{D_1}{\theta_1}
    \ltimes
    \cdots
    \ltimes
    \pinst{D_n}{\theta_n}.
  \end{equation*}
  Likewise, whenever parallel composition is well-defined,
  \begin{equation*}
    \Pinst{
      D_1\parallel\cdots\parallel D_n
    }{
      \theta
    }
    =
    \pinst{D_1}{\theta_1}
    \parallel
    \cdots
    \parallel
    \pinst{D_n}{\theta_n}.
  \end{equation*}

  The analogous identity holds for iterated cascade compositions of parametric
  systems, using the same initial connecting values in the parametric and
  instantiated cascades and the canonical identification of finite Cartesian
  products.
\end{restatable}

\begin{proof}
  The claims follow by induction on
  $n$
  from associativity,
  Proposition~\ref{prop:composition-commutes-with-instantiation},
  and
  Proposition~\ref{prop:system-composition-commutes-with-instantiation}.
\end{proof}

\begin{closingremark}[Generality of simple cascades]
  Proposition~\ref{prop:parametric-cascade-connection-absorption}
  and
  Theorem~\ref{thm:simple-cascades-generality}
  show that, whenever the considered class of parametric recurrent dynamics is
  closed under parametric input precomposition, allowing arbitrary parametric
  connecting functions does not enlarge the class of cascade dynamics:
  every connected cascade can be represented exactly as a simple cascade.
\end{closingremark}

\subsubsection{Generality of Simple Cascades}

\begin{proposition}[Absorption of connecting functions]
  \label{prop:parametric-cascade-connection-absorption}
  Let
  \begin{equation*}
    D_1
    =
    \ctuple{
      X_1,
      U,
      f_1
    }{
      \Theta_1
    },
    \qquad
    D_2
    =
    \ctuple{
      X_2,
      Z,
      f_2
    }{
      \Theta_2
    },
  \end{equation*}
  and let
  \begin{equation*}
    g
    :
    X_1
    \times
    U
    \times
    X_1
    \times
    \Theta_g
    \to
    Z
  \end{equation*}
  be a parametric connecting function.

  Let
  \begin{equation*}
    \widetilde D_2
    \defeq
    \inputprecomposition{g}{D_2},
  \end{equation*}
  where
  $g$
  is regarded as a parametric map from
  \begin{equation*}
    X_1\times U\times X_1
  \end{equation*}
  to
  $Z$.
  Explicitly,
  \begin{equation*}
    \widetilde D_2
    =
    \ctuple{
      X_2,
      X_1\times U\times X_1,
      \widetilde f_2
    }{
      \Theta_g\times\Theta_2
    },
  \end{equation*}
  where
  \begin{equation}
    \label{eq:absorbed-downstream-dynamics}
    \bigfpinst{\widetilde f_2}{
      x_2,
      \tuple{x_1,u,x'_1}
    }{
      \theta_g,\theta_2
    }
    \defeq
    \fpinst{f_2}{
      x_2,
      \bigfpinst{g}{
        x_1,u,x'_1
      }{
        \theta_g
      }
    }{
      \theta_2
    }.
  \end{equation}

  Then
  \begin{equation}
    \label{eq:parametric-connected-simple-cascade-equivalence}
    D_1
    \ltimes_g
    D_2
    =
    D_1
    \ltimes
    \widetilde D_2.
  \end{equation}
\end{proposition}
\begin{proof}
  Both sides have state space
  \begin{equation*}
    X_1\times X_2,
  \end{equation*}
  input space
  $U$,
  and parameter space
  \begin{equation*}
    \Theta_1
    \times
    \Theta_g
    \times
    \Theta_2.
  \end{equation*}

  For every
  \begin{equation*}
    x_1\in X_1,
    \qquad
    x_2\in X_2,
    \qquad
    u\in U,
  \end{equation*}
  and every
  \begin{equation*}
    \theta_1\in\Theta_1,
    \qquad
    \theta_g\in\Theta_g,
    \qquad
    \theta_2\in\Theta_2,
  \end{equation*}
  let
  \begin{equation*}
    x'_1
    \defeq
    \fpinst{f_1}{x_1,u}{\theta_1}.
  \end{equation*}
  The second component of the connected cascade is
  \begin{equation*}
    \fpinst{f_2}{
      x_2,
      \bigfpinst{g}{
        x_1,u,x'_1
      }{
        \theta_g
      }
    }{
      \theta_2
    }.
  \end{equation*}
  By
  Equation~\eqref{eq:absorbed-downstream-dynamics},
  this is exactly
  \begin{equation*}
    \bigfpinst{\widetilde f_2}{
      x_2,
      \tuple{x_1,u,x'_1}
    }{
      \theta_g,\theta_2
    },
  \end{equation*}
  which is the second component of the simple cascade.
  The first components are identical by definition.
\end{proof}

\begin{closingdefinition}[Closure under parametric input precomposition]
  \label{def:parametric-dynamics-input-precomposition-closure}
  A class
  $\classDynamicsSymb$
  of parametric dynamics is
  \emph{closed under input precomposition} with parametric functions from a class $\classparfunc$
  if, for every
  \begin{equation*}
    D
    \in
    \classDynamicsSymb
  \end{equation*}
  and every parametric continuous function
  $f \in \classparfunc$
  for which
  $\inputprecomposition{f}{D}$
  is well-defined, we have
  \begin{equation*}
    \big(\inputprecomposition{f}{D}\big)
    \in
    \classDynamicsSymb.
  \end{equation*}
\end{closingdefinition}

\begin{theorem}[Generality of simple cascades]
  \label{thm:simple-cascades-generality}
  Let
  $\classDynamicsSymb$
  be a class of parametric dynamics closed under input
  precomposition with parametric functions from a class $\classparfunc$.

  Let $D$ be a cascade given by
  \begin{equation*}
    D \,=\, D_1 \ltimes_g D_2
  \end{equation*}
  with 
  $D_2\in\classDynamicsSymb$
  and 
  connecting function
  $g\in\classparfunc$.
  Then
  $D$ is equal to a simple cascade
  \begin{equation*}
    \tilde{D} 
    \,=\,
    D_1
    \ltimes
    \tilde D_2
  \end{equation*}
  for some
  $\tilde D_2 \in \classDynamicsSymb$.

  Consequently, 
  the class of simple cascades having units from $\classDynamicsSymb$ 
  coincides with 
  the class of cascades having units from $\classDynamicsSymb$ 
  and connecting functions from $\classparfunc$.
\end{theorem}
\begin{proof}
  By
  Proposition~\ref{prop:parametric-cascade-connection-absorption},
  \begin{equation*}
    D_1
    \ltimes_g
    D_2
    =
    D_1
    \ltimes
    \big(
      \inputprecomposition{g}{D_2}
    \big).
  \end{equation*}
  Since
  $\classDynamicsSymb$
  is closed under parametric input precomposition and
  $D_2\in\classDynamicsSymb$,
  we have
  \begin{equation*}
    \big(\inputprecomposition{g}{D_2}\big)
    \in
    \classDynamicsSymb.
  \end{equation*}
  Taking
  \begin{equation*}
    \tilde D_2
    \defeq
    \inputprecomposition{g}{D_2}
  \end{equation*}
  proves the claim.
\end{proof}

\begin{closingfact}[Composition of recurrent-unit input functions]
  \label{fact:simple-cascade-input-function-composition}
  We note that cascade compositions employing connecting functions can be expressed as simple
  cascade compositions---which do not employ connecting functions.

  Consider a class 
  $\classDynamicsSymb^\parClassSymb$ 
  of parametric recurrent units 
  \begin{align*}
    D \,=\, \bigmctuple{X,U,f}{\Theta_f \times \Theta_\varphi}{\varphi}
    \qquad\text{with}\qquad
    \fpinst{f}{x,u}{\theta_f,\theta_\varphi} 
    \;=\; 
    \bigfpinst{\tilde f}{x, \fpinst{\varphi}{u}{\theta_\varphi}}{\theta_f},\;
    \varphi \in \classparfunc.
  \end{align*}
  Namely, the dependence of the dynamics function $f$ on
  inputs $u$ is specified through a parametric function $\varphi \in \classparfunc$.
  For such class $\classDynamicsSymb^\parClassSymb$,
  closure under input precomposition with parametric functions from $\Psi$ 
  amounts to closure of $\classparfunc$ under precomposition with functions 
  from $\Psi$.
  In fact, 
  for every $D \in \classDynamicsSymb^\parClassSymb$
  and every $\psi \in \Psi$,
  we have
  \begin{align*}
    D_\psi 
    \,\defeq\,
    \inputprecomposition{\psi}{D} 
    \,=\,
    \bigmctuple{X,U,f}{\Theta_f \times \Theta_\varphi \times \Theta_\psi}{\tilde \varphi}
    \quad\text{ with }\quad
    \fpinst{\tilde \varphi}{u}{\theta_\varphi,\theta_\psi} 
    \,=\,
    \bigfpinst{\varphi}{\fpinst{\psi}{u}{\theta_\psi}}{\theta_\varphi},
  \end{align*}
  and 
  hence $D_\psi \in \classDynamicsSymb^\parClassSymb$.

  Thus,
  all dynamics $D$ given by a cascade composition of units from 
  $\classDynamicsSymb^\parClassSymb$
  with connecting functions from $\classparfunc$,
  can be expressed as a simple cascade composition of units from
  $\classDynamicsSymb^\parClassSymb$---with no use of connecting functions.
\end{closingfact}

\begin{closingfact}[Associativity]
  Under the canonical identification of finite Cartesian products,
  the cascade and parallel composition operators for parametric dynamics are
  associative.
  The cascade composition operator for parametric systems is likewise
  associative when the initial connecting values are fixed for each
  inter-layer space.

  In particular, the state space of a finite system cascade is canonically
  identified with the interleaved product
  \begin{equation*}
    X_1
    \times
    Z_1
    \times
    X_2
    \times
    \cdots
    \times
    Z_{n-1}
    \times
    X_n,
  \end{equation*}
  independently of parenthesisation.
  Hence one can write
  \begin{equation*}
    D_1\ltimes\cdots\ltimes D_n,
    \qquad
    D_1\parallel\cdots\parallel D_n,
    \qquad
    S_1\ltimes\cdots\ltimes S_n
  \end{equation*}
  without specifying parentheses.
\end{closingfact}

\subsection{Euclidean Dynamics and Systems}

\begin{closingdefinition}[Euclidean parametric dynamics and systems]
  Let us consider parametric dynamics $D=\ctuple{X,U,f}{\Theta}$.
  Then, we say that
  dynamics $D$ are \emph{Euclidean} if
  the ambient spaces
  $\Omega_X$,
  $\Omega_U$, and
  $\Omega_\Theta$
  of
  $X$,
  $U$, and
  $\Theta$
  are Euclidean spaces---namely,
  $X\subseteq\Omega_X=\reals^n$,
  $U\subseteq\Omega_U=\reals^m$,
  and
  $\Theta\subseteq\Omega_\Theta=\reals^p$
  for some $n,m,p\in\posnaturals$;
\end{closingdefinition}

\begin{closingdefinition}[Euclidean and measurable-Euclidean parametric systems]
  Let us consider a parametric system
  $S = \ctuple{ X,U,f,x^\mathrm{init},Y,h }{ \Theta_f,\Theta_h }$.
  Then,
  we say that $S$ is \emph{Euclidean} if its dynamics $\dynop{S}$ are Euclidean 
  and additionally
  the ambient spaces $\Omega_Y$ and $\Omega_{\Theta_h}$ of $Y$ and $\Theta_h$
  are Euclidean spaces---namely,
  $Y\subseteq\Omega_Y=\reals^q$
  and
  $\Theta_h\subseteq\Omega_{\Theta_h}=\reals^r$
  for some $q,r\in\posnaturals$;
\end{closingdefinition}

\subsubsection{Analytic Output Extensions}

\begin{closingdefinition}[Analytic output extension]
  \label{def:analytic-output-extension}
  Let
  \begin{equation*}
    D
    =
    \ctuple{
      X,
      U,
      f
    }{
      \Theta_f
    }
  \end{equation*}
  be Euclidean parametric dynamics,
  let
  $x^\mathrm{init}\in X$,
  let
  $Y$
  and
  $\Theta_h$
  be Euclidean spaces,
  and let
  \begin{equation*}
    h
    \in
    \analyticparamfuncclass[
      X\times U\times X
    ][
      Y
    ][
      \Theta_h
    ].
  \end{equation*}

  The output extension
  \begin{equation*}
    \ptuple{D,x^\mathrm{init}}
    \triangleright
    h
  \end{equation*}
  is called the
  \emph{analytic output extension of
  $\ptuple{D,x^\mathrm{init}}$
  by
  $h$}.
\end{closingdefinition}

\subsection{State Functions of Dynamics and Systems}
\label{subsubsec:auxiliary-functions-for-dynamics-and-systems}

\todo{}

\begin{closingdefinition}
  Given dynamics
  $D = \tuple{X,U,f}$,
  their \emph{state-to-state function}
  for fixed
  $u \in \infseqs{U}$
  and
  $t \in \timedomain$
  is the function 
  $\indexedStateToStateFuncfamily{D}[u][t] : X \to X$
  that maps each state $x_0 \in X$ to the corresponding state at time $t$,
  namely
  \begin{align*}
    \indexedStateToStateFuncfamily{D}[u][t][x_0]
    \,\defeq\, 
    \repval{D}(x_0,\seq{u}{1}{t})\,.
  \end{align*}
\end{closingdefinition}

\begin{closingdefinition}
  Given parametric dynamics
  $D = \ctuple{X,U,f}{\Theta}$,
  their \emph{state-to-state function}
  for fixed
  $u \in \infseqs{U}$,
  $\theta \in \Theta$,
  and
  $t \in \timedomain$
  is the function $\indexedParStateToStateFuncfamily{D}[\theta][u][t] : X \to X$
  that maps each state $x_0 \in X$ to the corresponding state at time $t$,
  \begin{align*}
    \indexedParStateToStateFuncfamily{D}[\theta][u][t][x_0]
    \,\defeq\,
    \fpinst{\repval{D}}{x_0,\seq{u}{1}{t}}{\theta}\,.
  \end{align*}
\end{closingdefinition}

\begin{closingdefinition} \label{def:parameter-to-state-function}
  Given parametric dynamics $D = \ctuple{X,U,f}{\Theta}$,
  \begin{itemize}
    \item
      their \emph{parameter-to-state function}
      at time $t \in \timedomain$
      for a fixed \emph{infinite} input sequence $u \in \infseqs{U}$
      and a fixed initial state $x_0 \in X$,
      is the function 
      $\indexedDynParamToStateFuncfamily{D}[x_0][u][t] : \Theta \to X$
      that maps each parameter value $\theta \in \Theta$ to the state of $D$ at time $t$ when
      executed on input $u$ starting in state $x_0$,
      \begin{align*}
        \indexedDynParamToStateFuncfamily{D}[x_0][u][t][\theta]
        \,\defeq\, 
        \fpinst{\repval{D}}{x_0, \seq{u}{1}{t}}{\theta}\,.
      \end{align*}
    \item
      their \emph{parameter-to-latest-state function}, 
      for a fixed \emph{finite} input sequence $u \in U^+$ of length $t$
      and a fixed initial state $x_0 \in X$,
      is the function 
      $\indexedDynParamToLatestStateFuncfamily{D}[x_0][u] : \Theta \to X$
      that maps each parameter value $\theta \in \Theta$ to the state of $D$ at time $t$ when
      executed on input $u$ starting in state $x_0$,
      \begin{align*}
        \indexedDynParamToLatestStateFuncfamily{D}[x_0][u][\theta]
        \,\defeq\, 
        \fpinst{\repval{D}}{x_0, u}{\theta}\,;
      \end{align*}
  \end{itemize}

  Given a parametric system $S$ having dynamics $D \defeq \dynop{S}$ 
  and initial state $x_0$,
  its \emph{parameter-to-latest-state function}, 
  for a fixed \emph{finite} input sequence $u \in U^+$ 
  is the function 
  \begin{align*}
    \indexedParamToLatestStateFuncfamily{S}[u]
    \,\defeq\,
    \indexedDynParamToLatestStateFuncfamily{D}[x_0][u]\,.
  \end{align*}
\end{closingdefinition}

\begin{closingdefinition}[Intermediate state-to-state function]
  \label{def:intermediate-state-to-state-function}
  Let us consider
  dynamics $D=\tuple{X,U,f}$,
  an input sequence $u\in\infseqs{U}$,
  and indices $s,t\in\naturals$ with $s<t$.
  Then,
  the \emph{state-to-state function of $D$ from time $s$ to time $t$} on input $u$,
  is the function
  \begin{align*}
    \indexedDynParamToStateFuncfamily[u][\timeslice{s}{t}] : X \to X
  \end{align*}
  that maps each state $x\in X$ at time $s$ to the state reached at time $t$
  under the input subsequence $\seq{u}{s+1}{t}$, namely
  \begin{align*}
    \indexedDynParamToStateFuncfamily[u][\timeslice{s}{t}][x]
    \,\defeq\,
    \repval{D}\big(x,\seq{u}{s+1}{t}\big)\,.
  \end{align*}
\end{closingdefinition}

\clearpage{}%

\clearpage{}%

\section{Generic Architectures}
\label{appendix:generic-architectures}

\todo{}

\todo{}

\localtableofcontents

\subsection{Recurrent Layers}

\begin{closingdefinition}[Recurrent layer presentations]
  \label{def:recurrent-layer-presentation}
  We define recurrent layer presentations and their parametric counterparts.

  \begin{definitionpart}[indent=body]
    A \emph{recurrent layer presentation} is an expression
    \begin{equation*}
      \presentationSty{L}
      \,\defeq\,
      D_1
      \replayer
      \cdots
      \replayer
      D_n
    \end{equation*}
    where
    $n\in\posnaturals$ is called the \emph{width} of the presentation,
    and
    $D_1,\ldots,D_n$ are dynamics having a common input space, 
    called the \emph{units} of the presentation. 

    The dynamics \emph{presented} by $\presentationSty{L}$ are
    \begin{equation*}
      \presentationVal{\presentationSty{L}}
      \,\defeq\,
      D_1
      \parallel
      \cdots
      \parallel
      D_n\,.
    \end{equation*}

    We also say that $\presentationSty{L}$ is a presentation of the recurrent layer
    $\presentationVal{\presentationSty{L}}$.

    For $\classDynamicsSymb$ a class of dynamics,
    the presentation $\presentationSty{L}$ is 
    a \emph{recurrent layer of type $\classDynamicsSymb$}
    if its units belong to $\classDynamicsSymb$, 
    namely 
    $\{D_1, \ldots, D_n\} \subseteq \classDynamicsSymb$.
  \end{definitionpart}

  \begin{definitionpart}[indent=body]
    A \emph{parametric recurrent layer presentation} is an expression
    \begin{equation*}
      \presentationSty{L}^\parClassSymb
      \,\defeq\,
      D_1^\parClassSymb
      \replayer
      \cdots
      \replayer
      D_n^\parClassSymb
    \end{equation*}
    where
    $n\in\posnaturals$ is called the \emph{width} of the presentation,
    and
    $D_1^\parClassSymb,\ldots,D_n^\parClassSymb$ are parametric dynamics
    having a common input space, 
    called the \emph{units} of the presentation.

    The parametric dynamics \emph{presented} by 
    $\presentationSty{L}^\parClassSymb$ 
    are
    \begin{equation*}
      \presentationVal{\presentationSty{L}^\parClassSymb}
      \,\defeq\,
      D_1^\parClassSymb
      \parallel
      \cdots
      \parallel
      D_n^\parClassSymb\,.
    \end{equation*}

    We also say that $\presentationSty{L}^\parClassSymb$ is a presentation of the parametric
    recurrent layer $\presentationVal{\presentationSty{L}^\parClassSymb}$.

    For $\classParDynamicsSymb$ a class of parametric dynamics,
    the presentation $\presentationSty{L}^\parClassSymb$ is 
    a \emph{parameteric recurrent layer presentation of type $\classParDynamicsSymb$}
    if its units belong to $\classParDynamicsSymb$---namely 
    $\{D_1^\parClassSymb, \ldots, D_n^\parClassSymb\} 
    \subseteq 
    \classParDynamicsSymb$.
  \end{definitionpart}
\end{closingdefinition}

\begin{closingdefinition}[Recurrent layer dynamics]
  \label{def:recurrent-layer-dynamics}
  We define recurrent layer dynamics along with their parametric counterparts.

  \begin{definitionpart}[indent=body]
    \emph{Recurrent layer dynamics} are dynamics 
    that are presented by some 
    recurrent layer presentation.
  \end{definitionpart}

  \begin{definitionpart}[indent=body]
    \emph{Parametric recurrent layer dynamics} are parametric dynamics 
    that are presented by some
    parametric recurrent layer presentation.
  \end{definitionpart}
\end{closingdefinition}

\begin{closingdefinition}[Classes of recurrent layer dynamics]
  \label{def:recurrent-layer-dynamics--classes}
  We define classes of recurrent layer dynamics along with their parametric counterparts.

  \begin{definitionpart}[indent=body]
    Let $\classDynamicsSymb$ be a class of dynamics.

    We say that
    recurrent layer dynamics are 
    \emph{recurrent layer $\classDynamicsSymb$-dynamics}
    if they are presented by some recurrent layer
    presentation of type~$\classDynamicsSymb$.
  \end{definitionpart}

  \begin{definitionpart}[indent=body]
    Let $\classParDynamicsSymb$ be a class of parametric dynamics.

    We say that
    parametric recurrent layer dynamics are 
    \emph{parametric recurrent layer $\classParDynamicsSymb$-dynamics}
    if they are presented by some parametric recurrent layer presentation 
    of type~$\classParDynamicsSymb$.
  \end{definitionpart}
\end{closingdefinition}

\subsection{Recurrent Cascades}

\begin{closingdefinition}[Recurrent cascade presentations]
  \label{def:recurrent-cascade-presentation}
  We define recurrent cascade presentations and their parametric counterparts.

  \begin{definitionpart}[indent=body]
    A \emph{recurrent cascade presentation} is an expression
    \begin{equation*}
      \presentationSty{C}
      \,\defeq\,
      \repinputprecomposition{g_1}{\presentationSty{L}_1}
      \repcascade_{g_2}
      \presentationSty{L}_2
      \repcascade_{g_3}
      \cdots
      \repcascade_{g_\ell}
      \presentationSty{L}_\ell
    \end{equation*}
    where
    $\ell\in\posnaturals$ is called the \emph{depth} of the presentation,
    $\presentationSty{L}_1,\ldots,\presentationSty{L}_\ell$ are recurrent layer presentations,
    called the \emph{layers} of the presentation, 
    and
    $g_1,\ldots,g_\ell$ are continuous functions,
    called the \emph{connecting functions} of the presentation,
    such that the composition
    \begin{equation*}
      \presentationVal{\presentationSty{C}}
      \,\defeq\,
      \inputprecomposition{g_1}{
        \presentationVal{\presentationSty{L}_1}
      }
      \ltimes_{g_2}
      \presentationVal{\presentationSty{L}_2}
      \ltimes_{g_3}
      \cdots
      \ltimes_{g_\ell}
      \presentationVal{\presentationSty{L}_\ell}
    \end{equation*}
    is well-defined;
    in this case, we call $\presentationVal{\presentationSty{C}}$ 
    the dynamics \emph{presented} by $\presentationSty{C}$.

    We also say that $\presentationSty{C}$ is a presentation of the recurrent cascade dynamics
    $\presentationVal{\presentationSty{C}}$.

    Letting $n_i$ be the width of the layer presentation $\presentationSty{L}_i$, 
    for each $i\in\iival{1}{\ell}$,
    we call
    \begin{equation*}
      \ptuple{
        n_1,
        \ldots,
        n_\ell
      }
    \end{equation*}
    the
    \emph{width profile} of the presentation $\presentationSty{C}$.

    For
    $\classDynamicsSymb$
    a class of dynamics
    and
    $\classfuncsymb$
    a class of functions,
    the presentation $\presentationSty{C}$
    is a
    \emph{recurrent cascade presentation of type $\ptuple{\classDynamicsSymb,\classfuncsymb}$}
    if
    (i)~each of its layers is a recurrent layer presentation of type $\classDynamicsSymb$,
    and
    (ii)~each of its connecting functions belongs to $\classfuncsymb$---namely
    $\{ g_1, \ldots, g_\ell \} \subseteq \classfuncsymb$.
  \end{definitionpart}

  \begin{definitionpart}[indent=body]
    A \emph{parametric recurrent cascade presentation} is an expression
    \begin{equation*}
      \presentationSty{C}^\parClassSymb
      \,\defeq\,
      \repinputprecomposition{
        g_1^\parClassSymb
      }{
        \presentationSty{L}_1^\parClassSymb
      }
      \repcascade_{
        g_2^\parClassSymb
      }
      \presentationSty{L}_2^\parClassSymb
      \repcascade_{
        g_3^\parClassSymb
      }
      \cdots
      \repcascade_{
        g_\ell^\parClassSymb
      }
      \presentationSty{L}_\ell^\parClassSymb
    \end{equation*}
    where
    $\ell\in\posnaturals$ is called the \emph{depth} of the presentation,
    $\presentationSty{L}_1^\parClassSymb, \ldots,
    \presentationSty{L}_\ell^\parClassSymb$
    are parametric recurrent layer presentations,
    called the \emph{layers} of the presentation,
    and
    $g_1^\parClassSymb,
    \ldots,
    g_\ell^\parClassSymb$
    are parametric continuous functions,
    called the \emph{connecting functions} of the presentation,
    such that the composition
    \begin{equation*}
      \presentationVal{
        \presentationSty{C}^\parClassSymb
      }
      \,\defeq\,
      \inputprecomposition{
        g_1^\parClassSymb
      }{
        \presentationVal{
          \presentationSty{L}_1^\parClassSymb
        }
      }
      \ltimes_{
        g_2^\parClassSymb
      }
      \presentationVal{
        \presentationSty{L}_2^\parClassSymb
      }
      \ltimes_{
        g_3^\parClassSymb
      }
      \cdots
      \ltimes_{
        g_\ell^\parClassSymb
      }
      \presentationVal{
        \presentationSty{L}_\ell^\parClassSymb
      }
    \end{equation*}
    is well-defined;
    in this case,
    we call 
    $\presentationVal{\presentationSty{C}^\parClassSymb}$
    the parametric dynamics \emph{presented} by
    $\presentationSty{C}^\parClassSymb$.

    We also say that $\presentationSty{C}^\parClassSymb$ is a presentation of the parametric
    recurrent cascade dynamics $\presentationVal{\presentationSty{C}^\parClassSymb}$.

    Letting $n_i$ be the width of $\presentationSty{L}_i^\parClassSymb$, 
    for each $i\in\iival{1}{\ell}$,
    we call
    \begin{equation*}
      \ptuple{
        n_1,
        \ldots,
        n_\ell
      }
    \end{equation*}
    the \emph{width profile} of the presentation $\presentationSty{C}$.

    For
    $\classParDynamicsSymb$
    a class of parametric dynamics
    and
    $\classparfunc$
    a class of parametric functions,
    the presentation
    $\presentationSty{C}^\parClassSymb$
    is a
    \emph{parametric recurrent cascade presentation of 
    type~$\ptuple{\classParDynamicsSymb,\classparfunc}$}
    if
    (i)~each of its layers is a parametric recurrent layer presentation 
    of type~$\classParDynamicsSymb$,
    and
    (ii)~each of its connecting functions belongs to $\classparfunc$---namely
    $
    \{ g_1^\parClassSymb, \ldots, g_\ell^\parClassSymb \}
    \subseteq
    \classparfunc
    $.
  \end{definitionpart}
\end{closingdefinition}

\begin{closingdefinition}[Recurrent cascade dynamics]
  \label{def:recurrent-cascade-dynamics}
  We define recurrent cascade dynamics along with their parametric counterparts.

  \begin{definitionpart}[indent=body]
    Dynamics are called
    \emph{recurrent cascade dynamics}
    if they are presented by some recurrent cascade presentation.
  \end{definitionpart}

  \begin{definitionpart}[indent=body]
    Parametric dynamics are called
    \emph{parametric recurrent cascade dynamics}
    if they are presented by some parametric recurrent cascade presentation.
  \end{definitionpart}
\end{closingdefinition}

\begin{closingdefinition}[Classes of recurrent cascade dynamics]
  \label{def:recurrent-cascade-dynamics--classes}
  We define classes of recurrent cascade dynamics along with their parametric counterparts.

  \begin{definitionpart}[indent=body]
    Given 
    a class $\classDynamicsSymb$ of dynamics 
    and 
    a class $\classfuncsymb$ of functions,
    recurrent cascade dynamics are called
    \emph{%
      recurrent cascade dynamics with 
      units from~$\classDynamicsSymb$ 
      and 
      connecting functions from~$\classfuncsymb$%
    } 
    if they are presented by some recurrent cascade presentation
    of type $\ptuple{\classDynamicsSymb,\classfuncsymb}$.
  \end{definitionpart}

  \begin{definitionpart}[indent=body]
    Given 
    a class $\classParDynamicsSymb$ of parametric dynamics 
    and
    a class $\classparfunc$ of parametric functions,
    parametric recurrent cascade dynamics are called
    \emph{%
      parametric recurrent cascade dynamics with 
      units from~$\classParDynamicsSymb$ 
      and 
      connecting functions from~$\classparfunc$%
    } 
    if they are presented by some recurrent cascade presentation
    of type $\ptuple{\classParDynamicsSymb,\classparfunc}$.
  \end{definitionpart}
\end{closingdefinition}

\subsection{Recurrent Neural Cascades (RNCs)}

\begin{closingdefinition}[Recurrent Neural Cascade Dynamics]
  \label{def:rnc-dynamics}
  We define recurrent neural cascade dynamics and their parametric counterpart.

  \begin{definitionpart}[indent=body]
    Let $\classDynamicsSymb$ be a class of Euclidean dynamics.

    The \emph{class of Recurrent Neural Cascade dynamics of type $\classDynamicsSymb$}
    consists of all recurrent cascade dynamics 
    whose units are from $\classDynamicsSymb$
    and 
    whose connecting functions are from $\analyticfuncclass$.

    It is abbreviated as \emph{$\classDynamicsSymb$-RNC dynamics}.

    It is denoted by $\rncdynop(\classDynamicsSymb)$.
  \end{definitionpart}

  \begin{definitionpart}[indent=body]
    Let $\classParDynamicsSymb$ be a class of parametric Euclidean dynamics.

    The \emph{class of parametric Recurrent Neural Cascade dynamics of type $\classParDynamicsSymb$}
    consists of all parametric recurrent cascade dynamics 
    whose units are from $\classParDynamicsSymb$
    and 
    whose connecting functions are from $\analyticparamfuncclass$.

    It is abbreviated as \emph{$\classParDynamicsSymb$-RNC dynamics}.

    It is denoted by $\rncdynparop(\classParDynamicsSymb)$.
  \end{definitionpart}
\end{closingdefinition}

\begin{closingdefinition}[Recurrent Neural Cascades]
  \label{def:rnc}
  We define recurrent neural cascades and their parametric counterpart.

  \begin{definitionpart}[indent=body]
    Let $\classDynamicsSymb$ be a class of Euclidean dynamics.

    The \emph{class of Recurrent Neural Cascades (RNCs) of type $\classDynamicsSymb$}
    consists of each system $S$ 
    having
    dynamics $\dynop{S} \in \rncdynop(\classDynamicsSymb)$
    and 
    output function $h \in \analyticfuncclass$.

    It is abbreviated as \emph{$\classDynamicsSymb$-RNCs}.

    It is denoted by $\rncop(\classDynamicsSymb)$.
  \end{definitionpart}

  \begin{definitionpart}[indent=body]
    Let $\classParDynamicsSymb$ be a class of parametric Euclidean dynamics.

    The \emph{class of parametric Recurrent Neural Cascades (RNCs) of type $\classParDynamicsSymb$}
    consists of each parametric system $S^\parClassSymb$ 
    having
    parametric dynamics $\dynop{S^\parClassSymb} \in \rncdynparop(\classParDynamicsSymb)$
    and 
    parametric output function $h^\parClassSymb \in \analyticparamfuncclass$.

    It is abbreviated as \emph{$\classParDynamicsSymb$-RNCs}.

    It is denoted by $\rncparop(\classParDynamicsSymb)$.
  \end{definitionpart}
\end{closingdefinition}

\begin{definition}[RNC-closed dynamics class]
  \label{def:rnc-closed-dynamics-class}
  We define the notion of RNC-closed class.

  A class
  $\classDynamicsSymb$
  of dynamics is
  \emph{RNC-closed}
  if
  $\rncdynop(\classDynamicsSymb{}) \subseteq \classDynamicsSymb{}$.

  A class
  $\classParDynamicsSymb$
  of parametric dynamics is
  \emph{RNC-closed}
  if
  $\rncdynparop(\classParDynamicsSymb{}) \subseteq \classParDynamicsSymb{}$.
\end{definition}

\subsubsection{RNCs as analytic output extension presentations}

\begin{closingdefinition}[Output extension presentations]
  \label{def:output-extension-presentation}
  We define output extension presentations and their parametric counterparts.

  \begin{definitionpart}[indent=body]
    Let $\presentationSty{D}$ be a recurrent layer or cascade presentation,
    let $D \defeq \presentationVal{\presentationSty{D}}$ be the dynamics presented by
    $\presentationSty{D}$,
    let $X$ be the state space of $D$,
    let $U$ be the input space of $D$,
    let $x^\mathrm{init} \in X$,
    and 
    let $h: X \times U \times X \to Y$ be a continuous function.

    The expression
    \begin{equation*}
      \presentationSty{S}
      \,\defeq\,
      \repoutoutextension{
        \presentationSty{D}
      }{
        x^\mathrm{init}
      }{
        h
      }
    \end{equation*}
    is called the
    \emph{output extension presentation of $\ptuple{\presentationSty{D},x^\mathrm{init}}$ by $h$},
    and 
    the system \emph{presented} by $\presentationSty{S}$
    is
    \begin{equation*}
      \presentationVal{\presentationSty{S}}
      \,\defeq\,
      \outputextension{
        \presentationVal{\presentationSty{D}}
      }{
        x^\mathrm{init}
      }{
        h
      }\,.
    \end{equation*}

    We also call $\presentationSty{S}$ a presentation of the system
    $\presentationVal{\presentationSty{S}}$.

    We say that
    $\presentationSty{S}$
    is an
    \emph{analytic output extension presentation}
    if 
    dynamics $D$ are Euclidean
    and 
    $h \in \analyticfuncclass$.
  \end{definitionpart}

  \begin{definitionpart}[indent=body]
    Let $\presentationSty{D}^\parClassSymb$ be a parametric recurrent layer or cascade presentation,
    let $D^\parClassSymb \defeq \presentationVal{\presentationSty{D}^\parClassSymb}$ be the
    parametric dynamics presented by $\presentationSty{D}^\parClassSymb$,
    let $X$ be the state space of $D^\parClassSymb$,
    let $U$ be the input space of $D^\parClassSymb$,
    let $x^\mathrm{init} \in X$,
    and 
    let
    $h^\parClassSymb : (X \times U \times X) \times \Theta_h \to Y$
    be a parametric continuous function---where the parameter space is $\Theta_h$.

    The expression
    \begin{equation*}
      \presentationSty{S}^\parClassSymb
      \,\defeq\,
      \repoutoutextension{
        \presentationSty{D}^\parClassSymb
      }{
        x^\mathrm{init}
      }{
        h^\parClassSymb
      }
    \end{equation*}
    is called the
    \emph{%
      parametric output extension presentation of 
      $\ptuple{ \presentationSty{D}^\parClassSymb, x^\mathrm{init} }$ 
      by $h^\parClassSymb$%
    },
    and 
    the parametric system \emph{presented} by $\presentationSty{S}^\parClassSymb$ is
    \begin{equation*}
      \presentationVal{
        \presentationSty{S}^\parClassSymb
      }
      \,\defeq\,
      \outputextension{
        \presentationVal{\presentationSty{D}^\parClassSymb}
      }{
        x^\mathrm{init}
      }{
        h^\parClassSymb
      }
      \,.
    \end{equation*}

    We also call $\presentationSty{S}^\parClassSymb$ a presentation of 
    the parametric system
    $\presentationVal{\presentationSty{S}^\parClassSymb}$.

    We say that
    $\presentationSty{S}^\parClassSymb$
    is a \emph{parametric analytic output extension presentation}
    if 
    parametric dynamics $D^\parClassSymb$ are Euclidean
    and 
    $h^\parClassSymb \in \analyticparamfuncclass$.
  \end{definitionpart}
\end{closingdefinition}

\begin{closingfact}[RNCs as analytic output extension presentations]
  \label{fact:rnc-as-analytic-output-extension}
  Let $\classDynamicsSymb$ be a class of Euclidean dynamics,
  and
  let $S\in \rncop(\classDynamicsSymb)$ 
  with
  \begin{align*}
    S
    \,=\,
    \bigtuple{X,U,f,x^\mathrm{init},Y,h}
    \,.
  \end{align*}

  There exists
  a recurrent cascade presentation 
  $\presentationSty{C}$ of the dynamics $\dynop{S}$
  such that
  $S$ is presented by the output extension presentation 
  \begin{align*}
    \presentationSty{S}
    \,\defeq\,
    \repoutoutextension{
      \presentationSty{C}
    }{
      x^\mathrm{init}
    }{
      h
    }
  \,;
  \end{align*}
  namely,
  \begin{equation*}
    \dynop{S}
    \,=\,
    \presentationVal{\presentationSty{C}}
    \qquad\text{ and }\qquad
    S
    \,=\,
    \presentationVal{\presentationSty{S}}
    \,=\,
    \outputextension{
      \presentationVal{\presentationSty{C}}
    }{
      x^\mathrm{init}
    }{
      h
    }
    \,;
  \end{equation*}
  in particular,
  the recurrent cascade presentation $\presentationSty{C}$ 
  has type $\ptuple{\classDynamicsSymb,\analyticfuncclass}$,
  and 
  the output extension $\presentationSty{S}$ is an analytic output extension.
\end{closingfact}

\begin{closingfact}[Parametric RNCs as analytic output extension presentations]
  \label{fact:rnc-as-analytic-output-extension--param}
  Let $\classParDynamicsSymb$ be a class of parametric Euclidean dynamics,
  and
  let $S^\parClassSymb\in \rncparop(\classParDynamicsSymb)$ 
  with
  \begin{align*}
    S^\parClassSymb 
    \,=\,
    \bigctuple{X,U,f^\parClassSymb,x^\mathrm{init},Y,h^\parClassSymb}{\Theta_f, \Theta_h}
    \,.
  \end{align*}

  There exists
  a parametric recurrent cascade presentation 
  $\presentationSty{C}^\parClassSymb$ of the parametric dynamics $\dynop{S^\parClassSymb}$
  such that
  $S^\parClassSymb$ is presented by the parametric output extension presentation 
  \begin{align*}
    \presentationSty{S}^\parClassSymb 
    \,\defeq\,
    \repoutoutextension{
      \presentationSty{C}^\parClassSymb
    }{
      x^\mathrm{init}
    }{
      h^\parClassSymb
    }
  \,;
  \end{align*}
  namely,
  \begin{equation*}
    \dynop{S^\parClassSymb}
    \,=\,
    \presentationVal{\presentationSty{C}^\parClassSymb}
    \qquad\text{ and }\qquad
    S
    \,=\,
    \presentationVal{\presentationSty{S}^\parClassSymb}
    \,=\,
    \outputextension{
      \presentationVal{\presentationSty{C}^\parClassSymb}
    }{
      x^\mathrm{init}
    }{
      h^\parClassSymb
    }
    \,;
  \end{equation*}
  in particular,
  the parametric recurrent cascade presentation $\presentationSty{C}^\parClassSymb$ 
  has type $\ptuple{\classParDynamicsSymb,\analyticparamfuncclass}$,
  and 
  the output extension $\presentationSty{S}^\parClassSymb$ is an analytic output extension.
\end{closingfact}

\begin{proof}[%
  Proof of Facts~\ref{fact:rnc-as-analytic-output-extension},%
  ~\ref{fact:rnc-as-analytic-output-extension--param}%
]
  We first prove
  Fact~\ref{fact:rnc-as-analytic-output-extension}.

  Let $S \in \rncop(\classDynamicsSymb)$.
  By Definition~\ref{def:rnc},
  we have
  \begin{equation*}
    \dynop{S} \in \rncdynop(\classDynamicsSymb)
    \qquad\text{and}\qquad
    h \in \analyticfuncclass.
  \end{equation*}
  By
  Definition~\ref{def:rnc-dynamics}
  and
  Definition~\ref{def:recurrent-cascade-dynamics--classes},
  there exists a recurrent cascade presentation
  $\presentationSty{C}$
  of type $\ptuple{ \classDynamicsSymb, \analyticfuncclass }$
  such that
  \begin{equation*}
    \presentationVal{
      \presentationSty{C}
    }
    =
    \dynop{S}.
  \end{equation*}
  Hence,
  by
  Definition~\ref{def:output-extension-presentation},
  the expression
  \begin{equation*}
    \presentationSty{S}
    \,\defeq\,
    \repoutoutextension{
      \presentationSty{C}
    }{
      x^\mathrm{init}
    }{
      h
    }
  \end{equation*}
  is an output extension presentation satisfying
  \begin{align*}
    \presentationVal{
      \presentationSty{S}
    }
    \;=\;
    \outputextension{
      \presentationVal{
        \presentationSty{C}
      }
    }{
      x^\mathrm{init}
    }{
      h
    }
    \;=\;
    \outputextension{
      \dynop{S}
    }{
      x^\mathrm{init}
    }{
      h
    }
    \;=\;
    S
    \,.
  \end{align*}
  Since
  $\dynop{S}$
  is Euclidean
  and
  $h\in\analyticfuncclass$,
  Definition~\ref{def:output-extension-presentation}
  further implies that
  $\presentationSty{S}$
  is an analytic output extension presentation.
  This proves
  Fact~\ref{fact:rnc-as-analytic-output-extension}.

  \par\medskip\noindent
  The parametric claim is analogous.

  Let $S^\parClassSymb \in \rncparop(\classParDynamicsSymb)$.

  By
  Definition~\ref{def:rnc},
  we have
  \begin{equation*}
    \dynop{S^\parClassSymb} \in \rncdynparop(\classParDynamicsSymb)
    \qquad\text{and}\qquad
    h^\parClassSymb \in \analyticparamfuncclass
    \,.
  \end{equation*}
  By
  Definition~\ref{def:rnc-dynamics}
  and
  Definition~\ref{def:recurrent-cascade-dynamics--classes},
  there exists a parametric recurrent cascade presentation
  $\presentationSty{C}^\parClassSymb$
  of type $\ptuple{ \classParDynamicsSymb, \analyticparamfuncclass }$
  such that
  \begin{equation*}
    \presentationVal{
      \presentationSty{C}^\parClassSymb
    }
    =
    \dynop{S^\parClassSymb}
    \,.
  \end{equation*}
  Therefore,
  by
  Definition~\ref{def:output-extension-presentation},
  the expression
  \begin{equation*}
    \presentationSty{S}^\parClassSymb
    \,\defeq\,
    \repoutoutextension{
      \presentationSty{C}^\parClassSymb
    }{
      x^\mathrm{init}
    }{
      h^\parClassSymb
    }
  \end{equation*}
  is a parametric output extension presentation satisfying
  \begin{align*}
    \presentationVal{
      \presentationSty{S}^\parClassSymb
    }
    \;=\;
    \outputextension{
      \presentationVal{
        \presentationSty{C}^\parClassSymb
      }
    }{
      x^\mathrm{init}
    }{
      h^\parClassSymb
    }
    \;=\;
    \outputextension{
      \dynop{S^\parClassSymb}
    }{
      x^\mathrm{init}
    }{
      h^\parClassSymb
    }
    \;=\;
    S^\parClassSymb
    \,.
  \end{align*}
  Since
  $\dynop{S^\parClassSymb}$
  is Euclidean
  and
  $h^\parClassSymb\in\analyticparamfuncclass$,
  Definition~\ref{def:output-extension-presentation}
  further implies that
  $\presentationSty{S}^\parClassSymb$
  is a parametric analytic output extension presentation.
  This proves Fact~\ref{fact:rnc-as-analytic-output-extension--param}.
\end{proof}

\subsection{Remarks on Convolution}
\label{subsec:remarks-on-convolution}

\todo{}

\todo{}

We show that convolutional preprocessing can be expressed naturally through
cascade composition.
More precisely,
a convolution acting on a two-step window can be realised as the output
function of a system having a trivial state,
and prepending this convolution to arbitrary target dynamics is equivalent
to precomposing the target dynamics by the convolution map.

\begin{closingdefinition}[Two-step convolution]
  \label{def:two-step-convolution}
  Let $d_\convAbbr\in\posnaturals$,
  and let
  \begin{equation*}
    U \subseteq \Omega_U = \reals^{d_\mathrm{in}},
    \qquad\qquad
    \Theta_\convAbbr
    \,\defeq\,
    \reals^{d_\convAbbr\times d_\mathrm{in}}
    \times
    \reals^{d_\convAbbr\times d_\mathrm{in}}
    \times
    \reals^{d_\convAbbr}.
  \end{equation*}

  The \emph{two-step convolution} with input space $U$ and parameter space $\Theta_\convAbbr$ is
  the parametric Euclidean map
  $\convOperator_{U,\Theta_\convAbbr} : U^2 \times \Theta_\convAbbr \to \reals^{d_\convAbbr}$ 
  given by
  \begin{align*}
    &
    \bigfpinst{\convOperator_{U,\Theta_\convAbbr}}{
      u^\mathrm{prev},
      u^\mathrm{cur}
    }{
      K_1,
      K_2,
      b
    }
    \,\defeq\,
    K_1\cdot u^\mathrm{cur}
    +
    K_2\cdot u^\mathrm{prev}
    +
    b
    \,.
  \end{align*}
  We write 
  $\convOperator$
  in place of
  $\convOperator_{U,\Theta_\convAbbr}$
  when $U,\Theta_\convAbbr$ are clear from the context.
\end{closingdefinition}

\begin{closingfact}
  \label{fact:two-step-convolution-analytic}
  The two-step convolution
  of Definition~\ref{def:two-step-convolution}
  is a parametric analytic Euclidean map.
\end{closingfact}
\begin{proof}
  The function $\convOperator_{\Omega_U,\Omega_{\Theta_\convAbbr}}$ is an ambient extension 
  of $\convOperator_{U,\Theta_\convAbbr}$, and it is polynomial in all of its arguments,
  and hence ordinarily-analytic.
\end{proof}

\begin{closingdefinition}[Stateless convolution system]
  \label{def:stateless-convolution-system}
  Let $X_0 = \{ x_0 \}$ be a singleton space.

  The \emph{stateless system associated with the two-step convolution}
  of Definition~\ref{def:two-step-convolution}
  is the parametric system
  \begin{equation*}
    S_\convAbbr
    \,\defeq\,
    \ctuple{
      X_0,
      U^2,
      f_0,
      x_0,
      \reals^{d_\convAbbr},
      h_\convAbbr
    }{\,
      X_0,
      \Theta_\convAbbr
    }
  \end{equation*}
  where
  \begin{align*}
    \bigfpinst{f_0}{
      x_0,
      \ptuple{u^\mathrm{prev},u^\mathrm{cur}}
    }{
      x_0
    }
    & \defeq
    x_0
    \,,
    \\[.3em]
    \bigfpinst{h_\convAbbr}{
      x_0,
      \ptuple{u^\mathrm{prev},u^\mathrm{cur}},
      x_0
    }{
      K_1,
      K_2,
      b
    }
    & \defeq
    \bigfpinst{\convOperator}{
      u^\mathrm{prev},
      u^\mathrm{cur}
    }{
      K_1,
      K_2,
      b
    }
    \,.
  \end{align*}
\end{closingdefinition}

\begin{closingproposition}[Prepending convolution by cascade composition]
  \label{prop:convolution-prepending-by-cascade}
  Let
  $D = \ctuple{X, U, f}{\Theta_f}$
  be arbitrary parametric dynamics with $U \subseteq \reals^{d_\convAbbr}$,
  and let
  $S_\convAbbr$ be the stateless convolution system of
  Definition~\ref{def:stateless-convolution-system}.
  Let us define the parametric dynamics
  \begin{align*}
    \hat D
    \,\defeq\,
    {\dynop{S_\convAbbr} \ltimes_{h_\convAbbr} D}
    \qquad\text{and}\qquad
    D_\convAbbr
    \,\defeq\,
    \inputprecomposition{\convOperator}{D}
    \,.
  \end{align*}

  We have that
  dynamics $\hat D$ and $D_\convAbbr$ are isomorphic.

  In particular,
  the above dynamics are of the form
  \begin{align*}
    \hat D
    \,=\,
    \ctuple{
      X_0\times X,
      U^2,
      \hat f
    }{
      X_0,
      \Theta_\convAbbr,
      \Theta_f
    },
    \qquad\quad
    D_\convAbbr
    \,=\,
    \ctuple{
      X,
      U^2,
      f_\convAbbr
    }{
      \Theta_\convAbbr,
      \Theta_f
    },
  \end{align*}
  and 
  dynamics $\hat D$ are isomorphic to dynamics $D_\convAbbr$,
  under the 
  state-space isomorphism 
  $\pi_2(x_0, x)\defeq x$ 
  and 
  parameter-space isomorphism 
  $\pi_{2,3}(x_0, \theta_\convAbbr, \theta_f) \defeq \ptuple{\theta_\convAbbr, \theta_f}$.

  Namely,
  for 
  every $x\in X$,
  every $u^\mathrm{p,c}\in U^2$,
  every $\theta_\convAbbr\in\Theta_\convAbbr$,
  and
  every $\theta_f\in\Theta_f$,
  \begin{align*}
    \pi_2
    \left(
      \bigfpinst{\hat f}{
        \tuple{x_0,x},
        u^\mathrm{p,c}
      }{
        x_0,
        \theta_\convAbbr, 
        \theta_f
      }
    \right)
    & \;=\;
    \bigfpinst{f_\convAbbr}{
      \pi_2(x_0,x),
      u^\mathrm{p,c}
    }{
      \pi_{2,3}(x_0, \theta_\convAbbr, \theta_f)
    }
    \\[.3em]
    & \;=\;
    \bigfpinst{f_\convAbbr}{
      x,
      u^\mathrm{p,c}
    }{
      \theta_\convAbbr,
      \theta_f
    }
    \,.
  \end{align*}
\end{closingproposition}
\begin{proof}
  \todo{}
  By
  Definition~\ref{def:stateless-convolution-system},
  \begin{equation*}
    \dynop{S_\convAbbr}
    =
    \ctuple{
      X_0,
      U^2,
      f_0
    }{
      X_0
    },
  \end{equation*}
  where
  \begin{equation*}
    \bigfpinst{f_0}{
      x_0,
      u^\mathrm{p,c}
    }{
      x_0
    }
    =
    x_0
  \end{equation*}
  for every
  $u^\mathrm{p,c}\in U^2$,
  and
  \begin{equation*}
    \bigfpinst{h_\convAbbr}{
      x_0,
      u^\mathrm{p,c},
      x_0
    }{
      \theta_\convAbbr
    }
    =
    \fpinst{\convOperator}{
      u^\mathrm{p,c}
    }{
      \theta_\convAbbr
    }.
  \end{equation*}

  Hence,
  by
  Definition~\ref{def:cascade-composition-parametric-dynamics-connection},
  \begin{equation*}
    \hat D
    =
    \ctuple{
      X_0\times X,
      U^2,
      \hat f
    }{
      X_0,
      \Theta_\convAbbr,
      \Theta_f
    },
  \end{equation*}
  and, for every
  $x\in X$,
  every
  $u^\mathrm{p,c}\in U^2$,
  every
  $\theta_\convAbbr\in\Theta_\convAbbr$,
  and every
  $\theta_f\in\Theta_f$,
  we have
  \begin{align*}
    \bigfpinst{\hat f}{
      \tuple{x_0,x},
      u^\mathrm{p,c}
    }{
      x_0,
      \theta_\convAbbr,
      \theta_f
    }
    &=
    \bigtuple{
      \bigfpinst{f_0}{
        x_0,
        u^\mathrm{p,c}
      }{
        x_0
      },
      \bigfpinst{f}{
        x,
        \bigfpinst{h_\convAbbr}{
          x_0,
          u^\mathrm{p,c},
          \bigfpinst{f_0}{
            x_0,
            u^\mathrm{p,c}
          }{
            x_0
          }
        }{
          \theta_\convAbbr
        }
      }{
        \theta_f
      }
    }
    \\
    &=
    \bigtuple{
      x_0,
      \bigfpinst{f}{
        x,
        \fpinst{\convOperator}{
          u^\mathrm{p,c}
        }{
          \theta_\convAbbr
        }
      }{
        \theta_f
      }
    }.
  \end{align*}
  Therefore,
  applying
  $\pi_2$
  gives
  \begin{equation}
    \label{eq:proof-convolution-cascade-projected-update}
    \pi_2
    \left(
      \bigfpinst{\hat f}{
        \tuple{x_0,x},
        u^\mathrm{p,c}
      }{
        x_0,
        \theta_\convAbbr,
        \theta_f
      }
    \right)
    =
    \bigfpinst{f}{
      x,
      \fpinst{\convOperator}{
        u^\mathrm{p,c}
      }{
        \theta_\convAbbr
      }
    }{
      \theta_f
    }.
  \end{equation}

  On the other hand,
  by
  Definition~\ref{def:parametric-input-precomposition},
  \begin{equation*}
    D_\convAbbr
    =
    \inputprecomposition{\convOperator}{D}
    =
    \ctuple{
      X,
      U^2,
      f_\convAbbr
    }{
      \Theta_\convAbbr,
      \Theta_f
    },
  \end{equation*}
  with
  \begin{equation}
    \label{eq:proof-convolution-input-precomposition-update}
    \bigfpinst{f_\convAbbr}{
      x,
      u^\mathrm{p,c}
    }{
      \theta_\convAbbr,
      \theta_f
    }
    =
    \bigfpinst{f}{
      x,
      \fpinst{\convOperator}{
        u^\mathrm{p,c}
      }{
        \theta_\convAbbr
      }
    }{
      \theta_f
    }.
  \end{equation}

  Combining
  Equations~\eqref{eq:proof-convolution-cascade-projected-update}
  and~\eqref{eq:proof-convolution-input-precomposition-update},
  and using
  \begin{equation*}
    \pi_2(x_0,x)
    =
    x,
    \qquad
    \pi_{2,3}(
      x_0,
      \theta_\convAbbr,
      \theta_f
    )
    =
    \ptuple{
      \theta_\convAbbr,
      \theta_f
    },
  \end{equation*}
  we obtain
  \begin{align*}
    \pi_2
    \left(
      \bigfpinst{\hat f}{
        \tuple{x_0,x},
        u^\mathrm{p,c}
      }{
        x_0,
        \theta_\convAbbr,
        \theta_f
      }
    \right)
    &=
    \bigfpinst{f_\convAbbr}{
      \pi_2(x_0,x),
      u^\mathrm{p,c}
    }{
      \pi_{2,3}(
        x_0,
        \theta_\convAbbr,
        \theta_f
      )
    }
    \\
    &=
    \bigfpinst{f_\convAbbr}{
      x,
      u^\mathrm{p,c}
    }{
      \theta_\convAbbr,
      \theta_f
    }.
  \end{align*}

  Finally,
  since
  $X_0$
  is a singleton,
  the coordinate projections
  \begin{equation*}
    \pi_2:
    X_0\times X
    \to
    X
  \end{equation*}
  and
  \begin{equation*}
    \pi_{2,3}:
    X_0
    \times
    \Theta_\convAbbr
    \times
    \Theta_f
    \to
    \Theta_\convAbbr
    \times
    \Theta_f
  \end{equation*}
  are isomorphisms.
  Since
  $\hat D$
  and
  $D_\convAbbr$
  moreover have the same input space
  $U^2$,
  the preceding identity shows that their dynamics functions are intertwined by these
  isomorphisms.
  Therefore,
  $\hat D$
  and
  $D_\convAbbr$
  are isomorphic.
\end{proof}

\todo{}

\begin{closingcorollary}[Prepending convolution to a parametric system]
  \label{cor:convolution-prepending-to-system}
  Let
  \begin{equation*}
    S
    =
    \ctuple{
      X,
      \reals^{d_\convAbbr},
      f,
      x^\mathrm{init},
      Y,
      h
    }{
      \Theta_f,
      \Theta_h
    }
  \end{equation*}
  be a parametric system.

  In the cascade
  \begin{equation*}
    S_\convAbbr
    \ltimes
    S,
  \end{equation*}
  the target-state component evolves according to the
  convolution-precomposed dynamics
  \begin{equation*}
    \inputprecomposition{\convOperator}{\dynop{S}}.
  \end{equation*}

  In particular,
  for an input sequence
  \begin{equation*}
    v
    =
    \big(
      u_t^\mathrm{p,c}
    \big)_{t\in\ival{T}}
    =
    \big(
      \ptuple{
        u_{t-1},
        u_t
      }
    \big)_{t\in\ival{T}}
    \in
    \big(U^2\big)^+,
  \end{equation*}
  the input presented to the target system at time
  $t$
  is
  \begin{equation*}
    \fpinst{\convOperator}{
      u_t^\mathrm{p,c}
    }{
      \theta_\convAbbr
    }
    =
    K_1\cdot u_t
    +
    K_2\cdot u_{t-1}
    +
    b.
  \end{equation*}
\end{closingcorollary}
\begin{proof}
  By
  Proposition~\ref{prop:convolution-prepending-by-cascade},
  the connected dynamics
  \begin{equation*}
    \dynop{S_\convAbbr}
    \ltimes_{h_\convAbbr}
    \dynop{S}
  \end{equation*}
  are isomorphic to
  \begin{equation*}
    \inputprecomposition{\convOperator}{\dynop{S}},
  \end{equation*}
  with the isomorphism given by projection onto the target-state component.

  Hence,
  in the system cascade
  $S_\convAbbr\ltimes S$,
  the target-state component evolves according to
  $\inputprecomposition{\convOperator}{\dynop{S}}$.

  For
  \begin{equation*}
    u_t^\mathrm{p,c}
    =
    \ptuple{
      u_{t-1},
      u_t
    },
  \end{equation*}
  Definition~\ref{def:two-step-convolution}
  gives
  \begin{equation*}
    \fpinst{\convOperator}{
      u_t^\mathrm{p,c}
    }{
      \theta_\convAbbr
    }
    =
    K_1\cdot u_t
    +
    K_2\cdot u_{t-1}
    +
    b,
  \end{equation*}
  which proves the claim.
\end{proof}

\begin{closingdefinition}[Repeat dynamics, Definition~24 of~\citep{mat}]
  \label{def:repeat-dynamics}
  The \emph{repeat dynamics on $U$} are the dynamics
  \begin{equation*}
    R_U
    \,\defeq\,
    \tuple{
      U^2,
      U,
      r_U
    },
    \qquad\text{ where }\qquad
    r_U
    \left(
      \ptuple{
        u^\mathrm{prev},
        u^\mathrm{cur}
      },
      u
    \right)
    \,\defeq\,
    \ptuple{
      u^\mathrm{cur},
      u
    }
    \,.
  \end{equation*}
  We denote by
  \begin{equation*}
    R_U^\mathrm{p}
    \,\defeq\,
    \ctuple{
      U^2,
      U,
      r_U^\mathrm{p}
    }{
      X_0
    }
  \end{equation*}
  its trivially-parametric counterpart,
  where
  \begin{equation*}
    \bigfpinst{r_U^\mathrm{p}}{
      \ptuple{
        u^\mathrm{prev},
        u^\mathrm{cur}
      },
      u
    }{
      x_0
    }
    \,\defeq\,
    \ptuple{
      u^\mathrm{cur},
      u
    }
    \,.
  \end{equation*}
\end{closingdefinition}

\begin{closingfact}[State sequence of repeat dynamics]
  \label{fact:repeat-dynamics-state-sequence}
  Let
  \begin{equation*}
    u^\mathrm{init}
    =
    \ptuple{
      u_{-1},
      u_0
    }
    \in
    U^2
  \end{equation*}
  be an initial state of
  $R_U$,
  and let
  \begin{equation*}
    u
    =
    (u_t)_{t\in\ival{T}}
    \in
    U^+
  \end{equation*}
  be an input sequence.

  Then the state sequence
  $(s_t)_{t\in\iival{0}{T}}$
  of
  $R_U$
  satisfies
  \begin{equation*}
    s_0
    =
    \ptuple{
      u_{-1},
      u_0
    },
    \qquad
    s_t
    =
    \ptuple{
      u_{t-1},
      u_t
    }
    \qquad
    \forall
    t\in\ival{T}.
  \end{equation*}
\end{closingfact}
\begin{proof}
  The claim for
  $t=0$
  holds by the choice of initial state.

  For
  $t\in\ival{T}$,
  assuming
  \begin{equation*}
    s_{t-1}
    =
    \ptuple{
      u_{t-2},
      u_{t-1}
    },
  \end{equation*}
  Definition~\ref{def:repeat-dynamics}
  gives
  \begin{align*}
    s_t
    &=
    r_U(
      s_{t-1},
      u_t
    )
    \\
    &=
    r_U
    \left(
      \ptuple{
        u_{t-2},
        u_{t-1}
      },
      u_t
    \right)
    \\
    &=
    \ptuple{
      u_{t-1},
      u_t
    }.
  \end{align*}
  The claim follows by induction.
\end{proof}

\begin{closingdefinition}[Repeat-convolution system]
  \label{def:repeat-convolution-system}
  Fix
  \begin{equation*}
    u_{-1},
    u_0
    \in
    U.
  \end{equation*}

  The \emph{repeat-convolution system} is the parametric system
  \begin{equation*}
    S_{\mathrm{rep},\convAbbr}
    \,\defeq\,
    \ctuple{
      U^2,
      U,
      r_U^\mathrm{p},
      \ptuple{
        u_{-1},
        u_0
      },
      \reals^{d_\convAbbr},
      h_{\mathrm{rep},\convAbbr}
    }{
      X_0,
      \Theta_\convAbbr
    },
  \end{equation*}
  where
  \begin{equation*}
    \bigfpinst{
      h_{\mathrm{rep},\convAbbr}
    }{
      s,
      u,
      s'
    }{
      \theta_\convAbbr
    }
    \,\defeq\,
    \fpinst{\convOperator}{
      s'
    }{
      \theta_\convAbbr
    }.
  \end{equation*}
\end{closingdefinition}

\begin{closingcorollary}[Prepending causal convolution to an unwindowed input sequence]
  \label{cor:convolution-prepending-unwindowed}
  Let
  \begin{equation*}
    S
    =
    \ctuple{
      X,
      \reals^{d_\convAbbr},
      f,
      x^\mathrm{init},
      Y,
      h
    }{
      \Theta_f,
      \Theta_h
    }
  \end{equation*}
  be a parametric system,
  and consider the cascade
  \begin{equation*}
    S_{\mathrm{rep},\convAbbr}
    \ltimes
    S.
  \end{equation*}

  For every input sequence
  \begin{equation*}
    u
    =
    (u_t)_{t\in\ival{T}}
    \in
    U^+,
  \end{equation*}
  the input presented to the target system at time
  $t$
  is
  \begin{equation*}
    \fpinst{\convOperator}{
      u_{t-1},
      u_t
    }{
      \theta_\convAbbr
    }
    =
    K_1\cdot u_t
    +
    K_2\cdot u_{t-1}
    +
    b
    \qquad
    \forall
    t\in\ival{T}.
  \end{equation*}

  Thus,
  \todo{}
  prepending the repeat-convolution system
  $S_{\mathrm{rep},\convAbbr}$
  realises causal two-step convolution directly on the input sequence
  $(u_t)_{t\in\ival{T}}$.
\end{closingcorollary}
\begin{proof}
  Let
  $(s_t)_{t\in\iival{0}{T}}$
  be the state sequence of the repeat-dynamics component of
  $S_{\mathrm{rep},\convAbbr}$.

  By
  Fact~\ref{fact:repeat-dynamics-state-sequence},
  \begin{equation*}
    s_t
    =
    \ptuple{
      u_{t-1},
      u_t
    }
    \qquad
    \forall
    t\in\ival{T}.
  \end{equation*}

  By
  Definition~\ref{def:repeat-convolution-system},
  the connecting value produced by
  $S_{\mathrm{rep},\convAbbr}$
  at time
  $t$
  is therefore
  \begin{align*}
    \bigfpinst{
      h_{\mathrm{rep},\convAbbr}
    }{
      s_{t-1},
      u_t,
      s_t
    }{
      \theta_\convAbbr
    }
    &=
    \fpinst{\convOperator}{
      s_t
    }{
      \theta_\convAbbr
    }
    \\
    &=
    \fpinst{\convOperator}{
      u_{t-1},
      u_t
    }{
      \theta_\convAbbr
    }
    \\
    &=
    K_1\cdot u_t
    +
    K_2\cdot u_{t-1}
    +
    b.
  \end{align*}

  By the definition of cascade composition of parametric systems,
  this connecting value is precisely the input presented to
  the target system
  $S$
  at time
  $t$.
\end{proof}

\begin{closingremark}[Windowed and unwindowed realisations of convolution]
  \label{rem:causal-convolution-unwindowed-input}
  Corollary~\ref{cor:convolution-prepending-to-system}
  and
  Corollary~\ref{cor:convolution-prepending-unwindowed}
  give two equivalent ways of placing a causal two-step convolution before a
  target system.

  In
  Corollary~\ref{cor:convolution-prepending-to-system},
  the external input has already been transformed into the window sequence
  \begin{equation*}
    \big(
      \ptuple{
        u_{t-1},
        u_t
      }
    \big)_{t\in\ival{T}},
  \end{equation*}
  and consequently the convolutional system can have a trivial state.

  In
  Corollary~\ref{cor:convolution-prepending-unwindowed},
  the external input is the original sequence
  $(u_t)_{t\in\ival{T}}$.
  The repeat dynamics
  $R_U$
  provide precisely the one-step memory needed to construct the state
  \begin{equation*}
    \ptuple{
      u_{t-1},
      u_t
    }
  \end{equation*}
  at time
  $t$,
  after which the same convolution map
  $\convOperator$
  is applied.
\end{closingremark}

\clearpage{}%

\clearpage{}%

\section{MinMax Algebra for MinMax Recurrences}
\label{app:minmax-algebra}

\localtableofcontents

This appendix develops the algebraic operations underlying MinMax
recurrences.
They arise by specialising the lattice-theoretic and algebraic notions of
Sections~\ref{subsec:lattice-theoretic-preliminaries}
and~\ref{subsec:algebraic-structures}
to the real line equipped with its usual order.

At the scalar level,
the lattice join and meet become maximum and minimum, respectively.
We denote these operations by
$\oplus$
and
$\odot$,
and extend both operations componentwise to vectors and matrices.

The scalar operations
$\oplus$
and
$\odot$
then induce a MinMax matrix product,
denoted by
$\otimes$.
It is obtained from ordinary matrix multiplication by replacing scalar
addition with
$\oplus$
and scalar multiplication with
$\odot$.
Thus,
$\otimes$
is reserved for the induced matrix--vector and matrix--matrix products,
whereas
$\odot$
denotes scalar or componentwise minimum.

\subsection{Scalar MinMax Algebra}
\label{subsec:minmax-scalar-algebra}

We first specialise the order-theoretic notion of lattice to the real line.

\begin{proposition}[Real distributive lattice]
  \label{prop:real-lattice}
  The ordered set
  $\ptuple{\reals,\leq}$
  is a distributive lattice, with
  \begin{equation*}
    x\wedge y
    =
    \min\{x,y\},
    \qquad
    x\vee y
    =
    \max\{x,y\}
    \qquad
    \forall x,y\in\reals.
  \end{equation*}
\end{proposition}

\begin{proof}
  For
  $x,y\in\reals$,
  $\min\{x,y\}$
  is a lower bound of
  $x$
  and
  $y$,
  and every common lower bound is at most
  $\min\{x,y\}$.
  Hence it is their meet.
  Dually,
  $\max\{x,y\}$
  is their join.

  To show distributivity,
  let
  $x,y,z\in\reals$.
  Since the usual order is total,
  assume without loss of generality that
  $y\leq z$.
  Then
  \begin{equation*}
    x\wedge(y\vee z)
    =
    x\wedge z
    =
    (x\wedge y)\vee(x\wedge z),
  \end{equation*}
  because
  $x\wedge y\leq x\wedge z$.
  The dual distributive identity follows analogously.
\end{proof}

We now introduce the notation used throughout the MinMax constructions.

\begin{closingdefinition}[Scalar MinMax algebra]
  \label{def:minmax-scalar-algebra}
  The \emph{scalar MinMax algebra} is the binary algebra
  \begin{equation*}
    \minmaxscalaralgebra
    \defeq
    \ptuple{
    \reals,
    \oplus,
    \odot
    },
  \end{equation*}
  where, for every
  $x,y\in\reals$,
  \begin{align*}
    x\oplus y
    &\defeq
    x\vee y
    =
    \max\{x,y\},
    \\
    x\odot y
    &\defeq
    x\wedge y
    =
    \min\{x,y\}.
  \end{align*}
\end{closingdefinition}

Thus,
$\oplus$
and
$\odot$
are respectively the join and meet operations of the real lattice,
written using the algebraic notation employed by MinMax recurrences.

\begin{proposition}[Algebraic structure of the scalar MinMax algebra]
  \label{prop:minmax-scalar-distributive-lattice-algebra}
  The scalar MinMax algebra
  $\minmaxscalaralgebra$
  is a distributive lattice algebra.
\end{proposition}

\begin{proof}
  By
  Proposition~\ref{prop:real-lattice},
  $\ptuple{\reals,\leq}$
  is a distributive lattice whose join and meet are respectively
  maximum and minimum.
  Hence,
  by
  Fact~\ref{fact:lattice-order-algebra-equivalence}
  and
  Definition~\ref{def:minmax-scalar-algebra},
  \begin{equation*}
    \ptuple{
    \reals,
    \oplus,
    \odot
    }
  \end{equation*}
  is the corresponding distributive lattice algebra.
\end{proof}

\begin{closingfact}[Scalar MinMax identities]
  \label{fact:minmax-scalar-identities}
  The operations
  $\oplus$
  and
  $\odot$
  are associative, commutative, idempotent, and mutually distributive.

  In particular,
  for every
  $x,y,z\in\reals$,
  \begin{align*}
    (x\oplus y)\oplus z
    &=
    x\oplus(y\oplus z),
    &
    x\oplus y
    &=
    y\oplus x,
    &
    x\oplus x
    &=
    x,
    \\
    (x\odot y)\odot z
    &=
    x\odot(y\odot z),
    &
    x\odot y
    &=
    y\odot x,
    &
    x\odot x
    &=
    x,
  \end{align*}
  and
  \begin{align*}
    x\odot(y\oplus z)
    &=
    (x\odot y)
    \oplus
    (x\odot z),
    \\
    x\oplus(y\odot z)
    &=
    (x\oplus y)
    \odot
    (x\oplus z).
  \end{align*}

  Furthermore,
  the absorption identities give
  \begin{align*}
    x\oplus(x\odot y)
    &=
    x,
    \\
    x\odot(x\oplus y)
    &=
    x.
  \end{align*}
\end{closingfact}

\begin{closingdefinition}[MinMax reductions]
  \label{def:minmax-scalar-reductions}
  Let $n\in\posnaturals$ with $n \geq 2$.

  The \emph{Min reduction} of arity $n$ is
  \begin{align*}
    \minreduction{n}
    &: \reals^n \to \reals,
    &
    \minreduction{n}(x_1,\ldots,x_n)
    &
    \,\defeq\;
    \bigodot_{i=1}^n x_i
    \;=\;
    \min_{i\in\iival{1}{n}} x_i,
  \end{align*}
  The \emph{Max reduction} of arity $n$ is
  \begin{align*}
    \maxreduction{n}
    &: \reals^n \to \reals,
    &
    \maxreduction{n}(x_1,\ldots,x_n)
    &
    \,\defeq\;
    \bigoplus_{i=1}^n x_i
    \;=\;
    \max_{i\in\iival{1}{n}} x_i.
  \end{align*}
  We call both $\minreduction{n}$ or $\maxreduction{n}$ a \emph{MinMax reduction}.

  The \emph{MinMax primitive function family} is
  \begin{equation*}
    \minmaxclassprimitive
    \,\defeq\,
    \big\{
      \minreduction{n}
      \mid
      n\in\naturals \text{ with } n \geq 2
    \big\}
    \cup
    \big\{
      \maxreduction{n}
      \mid
      n\in\naturals \text{ with } n \geq 2
    \big\}.
  \end{equation*}
\end{closingdefinition}

\begin{proposition}[Finite distributivity of the scalar MinMax operations]
  \label{prop:minmax-finite-distributivity}
  For every
  $n\in\posnaturals$,
  every
  $x\in\reals$,
  and every
  $y_1,\ldots,y_n\in\reals$,
  \begin{align}
    \label{eq:minmax-odot-distributes-over-plus}
    x
    \odot
    \bigoplus_{i=1}^n y_i
    &=
    \bigoplus_{i=1}^n
    \big(
    x\odot y_i
    \big),
    \\
    \label{eq:minmax-plus-distributes-over-odot}
    x
    \oplus
    \bigodot_{i=1}^n y_i
    &=
    \bigodot_{i=1}^n
    \big(
    x\oplus y_i
    \big).
  \end{align}
\end{proposition}

\begin{proof}
  By
  Proposition~\ref{prop:minmax-scalar-distributive-lattice-algebra},
  $\oplus$
  and
  $\odot$
  are mutually distributive.
  The finite identities follow by induction on
  $n$,
  using associativity of the two operations.
\end{proof}

\subsection{Componentwise MinMax Operations}
\label{subsec:minmax-componentwise-operations}

For every
$n\in\posnaturals$,
we equip
$\reals^n$
with the componentwise order
$\leq_{\mathrm{cw}}$
defined by
\begin{equation*}
  x\leq_{\mathrm{cw}}y
  \quad\Longleftrightarrow\quad
  \getentry{x}{i}
  \leq
  \getentry{y}{i}
  \qquad
  \forall i\in\iival{1}{n}.
\end{equation*}

\begin{proposition}[Euclidean distributive lattice]
  \label{prop:euclidean-lattice}
  For every
  $n\in\posnaturals$,
  the ordered set
  \begin{equation*}
    \ptuple{\reals^n,\leq_{\mathrm{cw}}}
  \end{equation*}
  is a distributive lattice whose meet and join are computed componentwise:
  \begin{align*}
    \getentry{x\wedge y}{i}
    &=
    \getentry{x}{i}
    \wedge
    \getentry{y}{i}
    =
    \min
    \left\{
      \getentry{x}{i},
      \getentry{y}{i}
    \right\},
    \\
    \getentry{x\vee y}{i}
    &=
    \getentry{x}{i}
    \vee
    \getentry{y}{i}
    =
    \max
    \left\{
      \getentry{x}{i},
      \getentry{y}{i}
    \right\}
  \end{align*}
  for every
  $i\in\iival{1}{n}$.
\end{proposition}

\begin{proof}
  The meet and join properties follow coordinatewise from
  Proposition~\ref{prop:real-lattice}.
  The distributive identities also hold coordinatewise.
\end{proof}

The scalar operations
$\oplus$
and
$\odot$
extend naturally to vectors and matrices by componentwise application.

\begin{closingdefinition}[Componentwise MinMax operations]
  \label{def:minmax-componentwise-operations}
  \label{def:minmax-componentwise-addition}
  Let
  $x,y\in\reals^n$.
  Their \emph{componentwise MinMax operations}
  are defined by
  \begin{align*}
    \getentry{x\oplus y}{i}
    &\defeq
    \getentry{x}{i}
    \oplus
    \getentry{y}{i}
    =
    \max
    \left\{
      \getentry{x}{i},
      \getentry{y}{i}
    \right\},
    \\
    \getentry{x\odot y}{i}
    &\defeq
    \getentry{x}{i}
    \odot
    \getentry{y}{i}
    =
    \min
    \left\{
      \getentry{x}{i},
      \getentry{y}{i}
    \right\}
  \end{align*}
  for every
  $i\in\iival{1}{n}$.

  Analogously,
  for
  $A,B\in\reals^{m\times n}$,
  we define
  \begin{align*}
    \getentry{A\oplus B}{i,j}
    &\defeq
    \getentry{A}{i,j}
    \oplus
    \getentry{B}{i,j},
    \\
    \getentry{A\odot B}{i,j}
    &\defeq
    \getentry{A}{i,j}
    \odot
    \getentry{B}{i,j}
  \end{align*}
  for every
  $i\in\iival{1}{m}$
  and
  $j\in\iival{1}{n}$.
\end{closingdefinition}

\begin{closingfact}[Componentwise lattice operations]
  \label{fact:minmax-componentwise-lattice-operations}
  For every
  $x,y\in\reals^n$,
  \begin{equation*}
    x\odot y
    =
    x\wedge y,
    \qquad
    x\oplus y
    =
    x\vee y,
  \end{equation*}
  where the meet and join are taken with respect to
  $\leq_{\mathrm{cw}}$.
  The analogous identities hold entrywise for matrices of equal dimensions.
\end{closingfact}

\subsection{MinMax Matrix Algebra}
\label{subsec:minmax-matrix-algebra}

The scalar MinMax algebra
$\minmaxscalaralgebra$
induces vector and matrix products analogously to ordinary matrix algebra:
scalar multiplication is replaced by
$\odot$,
finite scalar addition is replaced by
$\oplus$,
and the resulting matrix product is denoted by
$\otimes$.

The use of the word
\emph{algebra}
in the title of this subsection is descriptive.
Because multiplication of rectangular matrices changes dimensions,
the collection of all rectangular matrices is not itself a single algebra
in the sense of
Definition~\ref{def:algebra}.

\begin{closingdefinition}[MinMax matrix products]
  \label{def:minmax-matrix-products}

  Let
  \begin{equation*}
    A\in\reals^{m\times n},
    \qquad
    x\in\reals^n.
  \end{equation*}
  Their \emph{MinMax matrix--vector product}
  is the vector
  \begin{equation*}
    A\otimes x
    \in
    \reals^m
  \end{equation*}
  defined by
  \begin{align*}
    \getentry{A\otimes x}{i}
    &\defeq
    \bigoplus_{j=1}^n
    \left(
    \getentry{A}{i,j}
    \odot
    \getentry{x}{j}
    \right)
    \\
    &=
    \max_{j\in\iival{1}{n}}
    \min
    \left\{
      \getentry{A}{i,j},
      \getentry{x}{j}
    \right\}
  \end{align*}
  for every
  $i\in\iival{1}{m}$.

  Let now
  \begin{equation*}
    A\in\reals^{m\times n},
    \qquad
    B\in\reals^{n\times p}.
  \end{equation*}
  Their \emph{MinMax matrix product}
  is the matrix
  \begin{equation*}
    A\otimes B
    \in
    \reals^{m\times p}
  \end{equation*}
  defined by
  \begin{align*}
    \getentry{A\otimes B}{i,j}
    &\defeq
    \bigoplus_{k=1}^n
    \left(
    \getentry{A}{i,k}
    \odot
    \getentry{B}{k,j}
    \right)
    \\
    &=
    \max_{k\in\iival{1}{n}}
    \min
    \left\{
      \getentry{A}{i,k},
      \getentry{B}{k,j}
    \right\}
  \end{align*}
  for every
  $i\in\iival{1}{m}$
  and
  $j\in\iival{1}{p}$.
\end{closingdefinition}

\begin{proposition}[MinMax matrix identities]
  \label{prop:minmax-algebraic-identities}
  Whenever the expressions below are dimensionally well-defined,
  the MinMax matrix operations satisfy
  \begin{align}
    (A\otimes B)\otimes C
    &=
    A\otimes(B\otimes C),
    \label{eq:minmax-matrix-associativity}
    \\
    (A\otimes B)\otimes x
    &=
    A\otimes(B\otimes x),
    \label{eq:minmax-action-associativity}
    \\
    A\otimes(x\oplus y)
    &=
    (A\otimes x)
    \oplus
    (A\otimes y),
    \label{eq:minmax-action-distributivity}
    \\
    A\otimes(B\oplus C)
    &=
    (A\otimes B)
    \oplus
    (A\otimes C),
    \label{eq:minmax-matrix-right-distributivity}
    \\
    (A\oplus B)\otimes C
    &=
    (A\otimes C)
    \oplus
    (B\otimes C).
    \label{eq:minmax-matrix-left-distributivity}
  \end{align}
\end{proposition}

\begin{proof}
  The claims follow from associativity and commutativity of the scalar
  operations
  $\oplus$
  and
  $\odot$,
  together with distributivity of
  $\odot$
  over
  $\oplus$,
  which hold by
  Proposition~\ref{prop:minmax-scalar-distributive-lattice-algebra}.

  We first prove matrix associativity.
  Let
  \begin{equation*}
    A\in\reals^{m\times n},
    \qquad
    B\in\reals^{n\times p},
    \qquad
    C\in\reals^{p\times q}.
  \end{equation*}
  For every
  $i\in\iival{1}{m}$
  and
  $j\in\iival{1}{q}$,
  \begin{align*}
    \getentry{(A\otimes B)\otimes C}{i,j}
    &=
    \bigoplus_{k=1}^p
    \left[
      \left(
      \bigoplus_{\ell=1}^n
      \big(
      \getentry{A}{i,\ell}
      \odot
      \getentry{B}{\ell,k}
      \big)
      \right)
      \odot
      \getentry{C}{k,j}
    \right]
    \\
    &=
    \bigoplus_{k=1}^p
    \bigoplus_{\ell=1}^n
    \left(
    \getentry{A}{i,\ell}
    \odot
    \getentry{B}{\ell,k}
    \odot
    \getentry{C}{k,j}
    \right)
    \\
    &=
    \bigoplus_{\ell=1}^n
    \bigoplus_{k=1}^p
    \left(
    \getentry{A}{i,\ell}
    \odot
    \getentry{B}{\ell,k}
    \odot
    \getentry{C}{k,j}
    \right)
    \\
    &=
    \bigoplus_{\ell=1}^n
    \left[
      \getentry{A}{i,\ell}
      \odot
      \left(
      \bigoplus_{k=1}^p
      \big(
      \getentry{B}{\ell,k}
      \odot
      \getentry{C}{k,j}
      \big)
      \right)
    \right]
    \\
    &=
    \getentry{A\otimes(B\otimes C)}{i,j}.
  \end{align*}
  This proves
  \eqref{eq:minmax-matrix-associativity}.

  The matrix--vector associativity
  \eqref{eq:minmax-action-associativity}
  follows by the same argument.

  For
  \eqref{eq:minmax-action-distributivity},
  let
  \begin{equation*}
    A\in\reals^{m\times n},
    \qquad
    x,y\in\reals^n.
  \end{equation*}
  Then, for every
  $i\in\iival{1}{m}$,
  \begin{align*}
    \getentry{A\otimes(x\oplus y)}{i}
    &=
    \bigoplus_{j=1}^n
    \left(
    \getentry{A}{i,j}
    \odot
    \big(
    \getentry{x}{j}
    \oplus
    \getentry{y}{j}
    \big)
    \right)
    \\
    &=
    \bigoplus_{j=1}^n
    \left[
      \big(
      \getentry{A}{i,j}
      \odot
      \getentry{x}{j}
      \big)
      \oplus
      \big(
      \getentry{A}{i,j}
      \odot
      \getentry{y}{j}
      \big)
    \right]
    \\
    &=
    \getentry{
      (A\otimes x)
      \oplus
      (A\otimes y)
    }{i}.
  \end{align*}

  The two matrix distributivity identities
  \eqref{eq:minmax-matrix-right-distributivity}
  and
  \eqref{eq:minmax-matrix-left-distributivity}
  follow entrywise by the same reasoning.
\end{proof}

\subsection{MinMax Affine Operators}
\label{subsec:minmax-affine-operators}

The MinMax matrix operations induce a class of affine-like maps that is
closed under composition.

\begin{restatable}[MinMax affine operators]{closingdefinition}{defminmaxaffineoperator}
  \label{def:minmax-affine-operator}
  \label{def:minmax-affine-operator-composition}
  Given
  \begin{equation*}
    A\in\reals^{n\times n},
    \qquad
    b\in\reals^n,
  \end{equation*}
  the pair
  \begin{equation*}
    M \defeq \tuple{A,b}
  \end{equation*}
  defines the \emph{MinMax affine operator}
  \begin{align*}
    \actfun{M}
    :
    \reals^n
    \to
    \reals^n,
    \qquad\quad
    x
    \mapsto
    (A\otimes x)\oplus b.
  \end{align*}

  Given two MinMax affine operators
  \begin{equation*}
    M_1
    =
    \tuple{A_1,b_1},
    \qquad
    M_2
    =
    \tuple{A_2,b_2},
  \end{equation*}
  we define their \emph{MinMax affine composition} by
  \begin{equation*}
    M_1
    \oplustimes
    M_2
    \defeq
    \bigtuple{
      A_2\otimes A_1, 
      (A_2\otimes b_1)\oplus b_2
    }.
  \end{equation*}
\end{restatable}

\begin{lemma}[Composition of MinMax affine operators]
  \label{lem:minmax-recurrence-operator-composition}
  Let
  $M_1 = \tuple{A_1,b_1}$
  and
  $M_2 = \tuple{A_2,b_2}$
  be MinMax affine operators of the same dimension.
  Then
  \begin{equation}
    \label{eq:minmax-affine-composition-correct}
    \actfun{M_1\oplustimes M_2}
    \,=\,
    \actfun{M_2}
    \circ
    \actfun{M_1}.
  \end{equation}

  Furthermore,
  $\oplustimes$
  is associative:
  \begin{equation}
    \label{eq:minmax-affine-composition-associative}
    (M_1\oplustimes M_2)\oplustimes M_3
    \,=\,
    M_1\oplustimes(M_2\oplustimes M_3)
  \end{equation}
  for all MinMax affine operators
  $M_1,M_2,M_3$
  of the same dimension.
\end{lemma}
\begin{proof}
  For every $x\in\reals^n$,
  Proposition~\ref{prop:minmax-algebraic-identities}
  gives
  \begin{align*}
    \actfun{M_2}
    \big(
    \actfun{M_1}(x)
    \big)
    &=
    A_2
    \otimes
    \big(
    (A_1\otimes x)
    \oplus
    b_1
    \big)
    \oplus
    b_2
    \\[.4em]
    &=
    \big(
    A_2\otimes(A_1\otimes x)
    \big)
    \oplus
    (A_2\otimes b_1)
    \oplus
    b_2
    \\[.4em]
    &=
    \big(
    (A_2\otimes A_1)
    \otimes
    x
    \big)
    \oplus
    \big(
    (A_2\otimes b_1)
    \oplus
    b_2
    \big)
    \\[.4em]
    &=
    \actfun{M_1\oplustimes M_2}(x).
  \end{align*}
  This proves
  \eqref{eq:minmax-affine-composition-correct}.

  We prove associativity directly at the level of the representing pairs.
  Let additionally
  \begin{equation*}
    M_3
    =
    \tuple{A_3,b_3}.
  \end{equation*}
  By
  Definition~\ref{def:minmax-affine-operator},
  \begin{align*}
    (M_1\oplustimes M_2)\oplustimes M_3
    &=
    \bigtuple{
      A_3\otimes(A_2\otimes A_1),\;
      A_3\otimes\big((A_2\otimes b_1)\oplus b_2\big)\oplus b_3
    }
    \\[.4em]
    &=
    \bigtuple{
      (A_3\otimes A_2)\otimes A_1,\;
      \big((A_3\otimes A_2)\otimes b_1\big)
      \oplus
      (A_3\otimes b_2)
      \oplus
      b_3
    },
  \end{align*}
  where we used
  Proposition~\ref{prop:minmax-algebraic-identities}.

  On the other hand,
  \begin{align*}
    M_1\oplustimes(M_2\oplustimes M_3)
    &=
    \bigtuple{
      (A_3\otimes A_2)\otimes A_1,\;
      \big((A_3\otimes A_2)\otimes b_1\big)
      \oplus
      \big((A_3\otimes b_2)\oplus b_3\big)
    }
    \\[.4em]
    &=
    \bigtuple{
      (A_3\otimes A_2)\otimes A_1,\;
      \big((A_3\otimes A_2)\otimes b_1\big)
      \oplus
      (A_3\otimes b_2)
      \oplus
      b_3
    },
  \end{align*}
  by associativity of
  $\oplus$.
  Hence
  \eqref{eq:minmax-affine-composition-associative}
  follows.
\end{proof}

\begin{closingremark}
  By associativity,
  finite compositions of MinMax affine operators are unambiguously written as
  \begin{equation*}
    M_1
    \oplustimes
    \cdots
    \oplustimes
    M_T.
  \end{equation*}
\end{closingremark}

\begin{closingproposition}[MinMax affine monoid over a bounded domain]
\label{prop:minmax-affine-monoid}
Let $V\subseteq\reals$ be a set with a minimum $\vmin \defeq \min V$ and a maximum
$\vmax \defeq \max V$---for instance, a nonempty finite set or a closed interval
$[\vmin,\vmax]$---and let $n\in\posnaturals$.
Define
\begin{equation*}
\mathcal{M}_{V,n}
\defeq
V^{n\times n}\times V^n,
\qquad\quad
\mathcal{E}_{V,n}
\defeq
\tuple{E,e},
\end{equation*}
where
\begin{align*}
\getentry{E}{i,j}
\defeq
\begin{cases}
\vmax & \text{ if } i=j,
\\[.3em]
\vmin & \text{ if } i\neq j,
\end{cases}
\qquad\qquad
\getentry{e}{i}
\defeq
\vmin.
\end{align*}
Then:
\begin{enumerate}
\item
\label{item:prop:minmax-affine-monoid-monoid}
$\ptuple{\mathcal{M}_{V,n},\oplustimes,\mathcal{E}_{V,n}}$ is a monoid, the
\emph{MinMax affine monoid over~$V$ of dimension~$n$};
\item
\label{item:prop:minmax-affine-monoid-action}
for every $M\in\mathcal{M}_{V,n}$, the restriction
$\tau_M \defeq \restr{\actfun{M}}{V^n}$ maps $V^n$ into $V^n$, and
\begin{equation}
\label{eq:minmax-affine-monoid-action}
\tau_{M_1\oplustimes M_2}
\,=\,
\tau_{M_2}\circ\tau_{M_1}
\qquad\quad
\forall M_1,M_2\in\mathcal{M}_{V,n},
\end{equation}
and $\tau_{\mathcal{E}_{V,n}}$ is the identity function on~$V^n$.
\end{enumerate}
We write $\mathcal{M}$ and $\mathcal{E}$ when $V$ and $n$ are clear from the context.
\end{closingproposition}
\begin{proof}
\emph{Closure.}
For $A\in V^{m\times k}$ and $x\in V^k$, every entry
$\getentry{A\otimes x}{i} = \bigoplus_{j} \getentry{A}{i,j}\odot\getentry{x}{j}$
is a maximum of minima of entries of $A$ and~$x$, hence one of those entries, and it belongs
to~$V$; the same holds for matrix products and for~$\oplus$. Hence
$M_1\oplustimes M_2\in\mathcal{M}_{V,n}$ for all $M_1,M_2\in\mathcal{M}_{V,n}$, and
$\actfun{M}$ maps $V^n$ into $V^n$ for every $M\in\mathcal{M}_{V,n}$; also
$\mathcal{E}_{V,n}\in\mathcal{M}_{V,n}$.
\emph{Associativity and action.}
Associativity of~$\oplustimes$ and
Eq.~\eqref{eq:minmax-affine-monoid-action} follow from
Lemma~\ref{lem:minmax-recurrence-operator-composition}, restricting
Eq.~\eqref{eq:minmax-affine-composition-correct} to~$V^n$.
\emph{Identity.}
Since $\vmin\le v\le\vmax$ for every $v\in V$, we have
$\vmax\odot v = v$, $\vmin\odot v = \vmin$ and $\vmin\oplus v = v$.
Let $x\in V^n$ and $i\in\iival{1}{n}$. Then
\begin{equation*}
\getentry{(E\otimes x)\oplus e}{i}
\,=\,
\big(\vmax\odot\getentry{x}{i}\big)
\oplus
\bigoplus_{j\neq i}\big(\vmin\odot\getentry{x}{j}\big)
\oplus
\vmin
\,=\,
\getentry{x}{i}\oplus\vmin
\,=\,
\getentry{x}{i};
\end{equation*}
hence $\tau_{\mathcal{E}_{V,n}}$ is the identity on~$V^n$, and, column by column,
$E\otimes A = A$ for every $A\in V^{n\times n}$.
Similarly, $\getentry{A\otimes E}{i,j} = \getentry{A}{i,j}\oplus\vmin = \getentry{A}{i,j}$ and
$\getentry{A\otimes e}{i} = \vmin$.
Therefore, for $M = \tuple{A,b}\in\mathcal{M}_{V,n}$,
\begin{align*}
\mathcal{E}_{V,n}\oplustimes M
&=
\bigtuple{A\otimes E,\,(A\otimes e)\oplus b}
=
\tuple{A,b},
\\[.3em]
M\oplustimes\mathcal{E}_{V,n}
&=
\bigtuple{E\otimes A,\,(E\otimes b)\oplus e}
=
\tuple{A,b}.
\qedhere
\end{align*}
\end{proof}

\subsection{MinMax Recurrences}
\label{subsec:minmax-algebra-recurrences}

A MinMax recurrence is obtained by applying a sequence of MinMax affine
operators to a recurrent state.

\begin{proposition}[Finite-horizon MinMax recurrence]
  \label{prop:minmax-recurrence-affine-composition}
  Let $x_0 \in \reals^n$,
  let $T\in\posnaturals$,
  let
  \begin{align*}
    (R_t)_{t=1}^T, 
    \qquad 
    (s_t)_{t=1}^T,
    \qquad
    (x_t)_{t=1}^T,
    \qquad
    (M_t)_{t=1}^T
  \end{align*}
  be as follows, for every $t \in \iival{1}{T}$:
  \begin{equation*}
    R_t\in\reals^{n\times n},
    \qquad
    s_t\in\reals^n,
    \qquad
    x_t \defeq \big( R_t\otimes x_{t-1} \big) \oplus s_t,
    \qquad
    M_t \defeq \tuple{R_t,s_t}.
  \end{equation*}
  Then,
  \begin{align*}
    x_t = \actfun{M_t}(x_{t-1})
    \qquad\quad
    \forall t\in\iival{1}{T},
  \end{align*}
  and $x_T = \actfun{ M_1 \oplustimes \cdots \oplustimes M_T }(x_0)$.
\end{proposition}
\begin{proof}
  The one-step identity follows directly from
  Definition~\ref{def:minmax-affine-operator}.

  Repeated application of
  Lemma~\ref{lem:minmax-recurrence-operator-composition}
  gives
  \begin{align*}
    x_T
    \,=\,
    \actfun{M_T}
    \circ
    \cdots
    \circ
    \actfun{M_1}(x_0)
    \,=\,
    \actfun{
      M_1
      \oplustimes
      \cdots
      \oplustimes
      M_T
    }(x_0),
  \end{align*}
  where the parenthesisation is immaterial by associativity of $\oplustimes$.
\end{proof}

\begin{corollary}[Finite MinMax-affine representation]
  \label{cor:minmax-recurrence-affine-representation}
  Under the assumptions of
  Proposition~\ref{prop:minmax-recurrence-affine-composition},
  there exist
  \begin{equation*}
    A_T\in\reals^{n\times n},
    \qquad
    b_T\in\reals^n
  \end{equation*}
  such that
  \begin{equation*}
    M_1
    \oplustimes
    \cdots
    \oplustimes
    M_T
    =
    \tuple{A_T,b_T},
  \end{equation*}
  and consequently
  \begin{equation}
    \label{eq:minmax-recurrence-finite-affine-form}
    x_T
    \,=\,
    (A_T\otimes x_0)
    \oplus
    b_T\,.
  \end{equation}
\end{corollary}
\begin{proof}
  The class of MinMax affine operators is closed under
  $\oplustimes$
  by
  Definition~\ref{def:minmax-affine-operator}.
  Hence their finite composition has the form
  $\tuple{A_T,b_T}$.
  The result then follows from
  Proposition~\ref{prop:minmax-recurrence-affine-composition}.
\end{proof}

\subsection{Finite Restrictions of the MinMax Algebra}
\label{subsec:minmax-finite-restrictions}

Moving from $\reals$ to its extension $\reals \cup \{ +\infty, -\infty \}$,
one can extend the scalar MinMax algebra over $\reals$ to the MinMax semiring 
over $\reals \cup \{ +\infty, -\infty \}$,
\begin{align*}
  \ptuple{ \reals \cup \{ +\infty, -\infty \},\, \oplus,\, \odot,\, +\infty,\, -\infty },
\end{align*}
where 
$+\infty$ is the identity element for $\oplus$
and
$-\infty$ is the identity element for $\odot$.

However, in this work we do not adopt such extension, as it seems unnecessary.
In fact, for finite computations, the relevant MinMax algebra is the one over the finite domain
$\minmaxdomain$ of values that occur during a computation. 
This restricted variant of the MinMax algebra retains the key algebraic properties of the one over
$\reals$, and additionally it admits identity elements 
for both $\oplus$ and $\odot$, which are the smallest and largest values of $\minmaxdomain$. 

\begin{closingdefinition}[Finite MinMax restriction]
  \label{def:minmax-finite-restriction}
  Let
  $\minmaxdomain\subseteq\reals$
  be a nonempty finite set, and define
  \begin{equation*}
    \vmin
    \defeq
    \min \minmaxdomain,
    \qquad\quad
    \vmax
    \defeq
    \max \minmaxdomain.
  \end{equation*}
  The \emph{finite MinMax restriction induced by $\minmaxdomain$} is 
  $\ptuple{ \minmaxdomain, \oplus, \odot }$
  where
  $\oplus$
  and
  $\odot$
  are the restrictions to
  $\minmaxdomain$
  of the scalar MinMax operations from
  Definition~\ref{def:minmax-scalar-algebra}.
\end{closingdefinition}

Observe that every subset of a totally ordered set is closed under binary
minimum and maximum.
Hence the above operations are well-defined on
$\minmaxdomain$.

\begin{proposition}[Identities of a finite MinMax restriction]
  \label{prop:minmax-finite-restriction-identities}
  Let
  $\minmaxdomain\subseteq\reals$
  be nonempty and finite.
  Then, 
  for every $x\in\minmaxdomain$,
  \begin{align*}
    \vmin\oplus x
    \,=\,
    x\oplus \vmin
    \,=\,
    x
    \qquad\text{and}\qquad
    \vmax\odot x
    \,=\,
    x\odot \vmax
    \,=\,
    x.
  \end{align*}
  Furthermore,
  for every $x\in\minmaxdomain$,
  \begin{align*}
    \vmin\odot x
    \,=\,
    x\odot \vmin
    \,=\,
    \vmin.
  \end{align*}
  Consequently,
  we have that
  $\ptuple{ \minmaxdomain, \oplus, \odot, \vmin, \vmax }$
  is a finite commutative idempotent semiring, 
  with additive identity $\vmin$ and multiplicative identity $\vmax$.
\end{proposition}
\begin{proof}
  Since
  $\vmin\leq x\leq \vmax$,
  \begin{align*}
    \vmin\oplus x
    &=
    \max\{\vmin,x\}
    =
    x,
    \\
    \vmax\odot x
    &=
    \min\{\vmax,x\}
    =
    x,
    \\
    \vmin\odot x
    &=
    \min\{\vmin,x\}
    =
    \vmin.
  \end{align*}
  Commutativity gives the corresponding right identities.
  The remaining semiring properties follow from
  Fact~\ref{fact:minmax-scalar-identities}.
\end{proof}

The finite restrictions above arise naturally when evaluating a MinMax
recurrence over a finite input sequence.
\begin{closingdefinition}[Data-generated MinMax domain]
  \label{def:minmax-data-generated-domain}
  Let
  $T,n\in\posnaturals$,
  let $x_0\in\reals^n$,
  and 
  let $(R_t)_{t=1}^T$ and $(s_t)_{t=1}^T$ be sequence of functions
  \begin{equation*}
    R_t\in\reals^{n\times n},
    \qquad\qquad
    s_t\in\reals^n.
  \end{equation*}
  The \emph{MinMax domain generated by the data}
  $\bigptuple{x_0,(R_t,s_t)_{t=1}^T}$
  is the finite set
  \begin{equation*}
    \minmaxdatadomain
    \defeq
    \entries{x_0}
    \cup
    \bigcup_{t=1}^T \entries{R_t}
    \cup
    \bigcup_{t=1}^T \entries{s_t}\,.
  \end{equation*}
  We write
  \begin{equation*}
    \vmin
    \defeq
    \min \minmaxdatadomain,
    \qquad
    \vmax
    \defeq
    \max \minmaxdatadomain\,.
  \end{equation*}
\end{closingdefinition}

Since
$\oplus$
and$\odot$
ar selective operations---that is, each returns one of its
arguments---MinMax computations cannot generate scalar values outside the
data-generated domain.

\begin{proposition}[Closure of finite MinMax computations]
  \label{prop:minmax-data-generated-closure}
  Let
  $\minmaxdatadomain$
  be as in
  Definition~\ref{def:minmax-data-generated-domain}.
  Then:
  \begin{itemize}
    \item
      if
      $A\in\minmaxdatadomain^{m\times n}$
      and
      $x\in\minmaxdatadomain^n$,
      then $A\otimes x \in \minmaxdatadomain^m$;
    \item
      if
      $A_1\in\minmaxdatadomain^{m\times n}$
      and
      $A_2\in\minmaxdatadomain^{n\times p}$,
      then $A_1\otimes A_2 \in \minmaxdatadomain^{m\times p}$;
    \item
      if
      $M_1 = \tuple{A_1,b_1}$
      and
      $M_2 = \tuple{A_2,b_2}$
      have all their entries in
      $\minmaxdatadomain$,
      then
      $M_1\oplustimes M_2$
      has all its entries in
      $\minmaxdatadomain$;
    \item
      every state $x_t$
      of the recurrence $x_t = (R_t\otimes x_{t-1}) \oplus s_t$
      belongs to $\minmaxdatadomain^n$.
  \end{itemize}
\end{proposition}
\begin{proof}
  Every application of
  $\oplus$
  or
  $\odot$
  to elements of
  $\minmaxdatadomain$
  returns one of its arguments, and hence again an element of
  $\minmaxdatadomain$.
  The first two claims therefore follow directly from
  Definition~\ref{def:minmax-matrix-products}.
  The third follows from
  Definition~\ref{def:minmax-affine-operator}.
  The last claim follows by induction on
  $t$,
  starting from
  $x_0\in\minmaxdatadomain^n$.
\end{proof}

We can therefore construct an exact identity for MinMax affine composition
using only scalar values appearing in the data.

\begin{proposition}[Data-generated identity for MinMax affine composition]
  \label{prop:minmax-data-generated-identity}
  Let
  $\minmaxdatadomain$
  be as in
  Definition~\ref{def:minmax-data-generated-domain},
  and let
  \begin{equation*}
    E
    \in
    \minmaxdatadomain^{n\times n},
    \qquad\quad
    e
    \in
    \minmaxdatadomain^n,
  \end{equation*}
  be defined by
  \begin{align*}
    \getentry{E}{i,j}
    \defeq
    \begin{cases}
      \vmax & \text{ if } i=j,
      \\[.3em]
      \vmin & \text{ if } i\neq j,
    \end{cases}
    \qquad\qquad
    \getentry{e}{i}
    \defeq
    \vmin.
  \end{align*}
  Equivalently,
  \begin{align*}
    E
    =
    \vmin \cdot \mathbf{1}_{n,n}
    +
    (\vmax-\vmin) \cdot \mathbf{I}_{n,n},
    \qquad\qquad
    e
    = \vmin \cdot \mathbf{1}_n.
  \end{align*}

  Then, for every
  $M
  =
  \tuple{A,b}
  \in
  \minmaxdatadomain^{n\times n}
  \!\times\!
  \minmaxdatadomain^n$,
  we have
  \begin{equation*}
    \tuple{E,e}
    \oplustimes
    M
    \,=\,
    M
    \oplustimes
    \tuple{E,e}
    \,=\,
    M.
  \end{equation*}
  Hence,
  $\bigptuple{\minmaxdatadomain^{n\times n} \!\times\! \minmaxdatadomain^n, \oplustimes, \tuple{E,e}}$
  is a finite monoid.
\end{proposition}
\begin{proof}
  By
  Proposition~\ref{prop:minmax-finite-restriction-identities},
  $\vmin$
  is the identity of
  $\oplus$
  and
  $\vmax$
  is the identity of
  $\odot$
  over
  $\minmaxdatadomain$.
  Therefore, for every
  $A\in\minmaxdatadomain^{n\times n}$
  and
  $b\in\minmaxdatadomain^n$,
  \begin{align*}
    E\otimes A
    &=
    A,
    &
    A\otimes E
    &=
    A,
    \\
    E\otimes b
    &=
    b,
    &
    A\otimes e
    &=
    e.
  \end{align*}
  Hence, by
  Definition~\ref{def:minmax-affine-operator},
  \begin{align*}
    \tuple{E,e}\oplustimes\tuple{A,b}
    &=
    \tuple{
      A\otimes E,
      (A\otimes e)\oplus b
    }
    =
    \tuple{A,b},
    \\
    \tuple{A,b}\oplustimes\tuple{E,e}
    &=
    \bigtuple{
      E\otimes A,\,
      (E\otimes b)\oplus e
    }
    =
    \tuple{A,b}.
  \end{align*}
  Associativity of
  $\oplustimes$
  follows from
  Lemma~\ref{lem:minmax-recurrence-operator-composition}.
  Finiteness follows from finiteness of
  $\minmaxdatadomain$.
\end{proof}

\begin{closingremark}
Proposition~\ref{prop:minmax-data-generated-identity} is the case
$V = \minmaxdatadomain$ of Proposition~\ref{prop:minmax-affine-monoid}.
\end{closingremark}

\todo{}

\begin{proposition}[Data-generated MinMax transformation monoid]
  \label{prop:minmax-data-generated-transformation-monoid}

  Let
  \begin{equation*}
    \minmaxdatamonoid
    \defeq
    \minmaxdatadomain^{n\times n}
    \times
    \minmaxdatadomain^n
  \end{equation*}
  be the finite monoid of MinMax affine operators from
  Proposition~\ref{prop:minmax-data-generated-identity},
  with operation
  $\oplustimes$
  and identity
  \begin{equation*}
    \minmaxdataidentity
    \defeq
    \tuple{E,e}.
  \end{equation*}

  Define the set of induced transformations
  \begin{equation*}
    \minmaxdatatransformations
    \defeq
    \left\{
      \restr{\actfun{M}}{\minmaxdatadomain^n}
      \mid
      M\in\minmaxdatamonoid
    \right\}\,.
  \end{equation*}

  Then every
  \begin{equation*}
    \tau
    \in
    \minmaxdatatransformations
  \end{equation*}
  is a transformation
  \begin{equation*}
    \tau
    :
    \minmaxdatadomain^n
    \to
    \minmaxdatadomain^n,
  \end{equation*}
  and
  \begin{equation*}
    \ptuple{
      \minmaxdatadomain^n,
      \minmaxdatatransformations
    }
  \end{equation*}
  is a finite transformation monoid under standard function composition.

  Furthermore, the action map
  \begin{equation*}
    \rho
    :
    \minmaxdatamonoid
    \to
    \minmaxdatatransformations,
    \qquad
    M
    \mapsto
    \restr{\actfun{M}}{\minmaxdatadomain^n}
  \end{equation*}
  is surjective and satisfies
  \begin{equation}
    \label{eq:minmax-data-generated-action-antihomomorphism}
    \rho(M_1\oplustimes M_2)
    =
    \rho(M_2)\circ\rho(M_1).
  \end{equation}
  Thus,
  $\rho$
  is an anti-homomorphism from
  $\minmaxdatamonoid$
  onto the induced transformation monoid,
  or equivalently a homomorphism from the opposite monoid
  $\minmaxdatamonoid^{\mathrm{op}}$.
\end{proposition}

\begin{proof}
  By
  Proposition~\ref{prop:minmax-data-generated-closure},
  every
  $M=\tuple{A,b}\in\minmaxdatamonoid$
  induces a well-defined transformation
  \begin{equation*}
    \actfun{M}
    :
    \minmaxdatadomain^n
    \to
    \minmaxdatadomain^n.
  \end{equation*}

  By
  Lemma~\ref{lem:minmax-recurrence-operator-composition},
  for every
  $M_1,M_2\in\minmaxdatamonoid$,
  \begin{equation*}
    \actfun{M_1\oplustimes M_2}
    =
    \actfun{M_2}
    \circ
    \actfun{M_1}.
  \end{equation*}
  Hence
  $\minmaxdatatransformations$
  is closed under function composition:
  if
  $\tau_1 = \restr{\actfun{M_1}}{\minmaxdatadomain^n}$
  and
  $\tau_2 = \restr{\actfun{M_2}}{\minmaxdatadomain^n}$,
  then
  \begin{equation*}
    \tau_2\circ\tau_1
    =
    \restr{
    \actfun{M_1\oplustimes M_2}
    }{
    \minmaxdatadomain^n
    }
    \in
    \minmaxdatatransformations\,.
  \end{equation*}

  Proposition~\ref{prop:minmax-data-generated-identity}
  gives
  \begin{equation*}
    \restr{
    \actfun{\minmaxdataidentity}
    }{
    \minmaxdatadomain^n
    }
    =
    \operatorname{id}_{\minmaxdatadomain^n}.
  \end{equation*}
  so the identity transformation belongs to
  $\minmaxdatatransformations$.
  Associativity follows from associativity of standard function composition.
  Finiteness follows from finiteness of
  $\minmaxdatamonoid$.

  Finally,
  surjectivity of
  $\rho$
  follows directly from the definition of
  $\minmaxdatatransformations$,
  while
  Equation~\eqref{eq:minmax-data-generated-action-antihomomorphism}
  is precisely
  Lemma~\ref{lem:minmax-recurrence-operator-composition}.
\end{proof}

\begin{corollary}[Prefix products represent recurrence transformations]
  \label{cor:minmax-data-generated-prefix-transformations}

  Under the assumptions of
  Proposition~\ref{prop:minmax-recurrence-affine-composition},
  let
  \begin{equation*}
    \tau_t
    \defeq
    \restr{\actfun{M_t}}{\minmaxdatadomain^n}\,.
  \end{equation*}
  Then, 
  for every
  $t\in\iival{1}{T}$,
  \begin{equation}
    \label{eq:minmax-prefix-product-transformation}
    \restr{
    \actfun{
    M_1
    \oplustimes
    \cdots
    \oplustimes
    M_t
    }
    }{
    \minmaxdatadomain^n
    }
    =
    \tau_t
    \circ
    \cdots
    \circ
    \tau_1\,,
  \end{equation}
  and consequently
  \begin{equation*}
    x_t
    =
    \actfun{
      M_1
      \oplustimes
      \cdots
      \oplustimes
      M_t
    }(x_0)
    =
    (\tau_t\circ\cdots\circ\tau_1)(x_0)\,.
  \end{equation*}
\end{corollary}
\begin{proof}
  Repeated application of
  Equation~\eqref{eq:minmax-data-generated-action-antihomomorphism}
  gives
  \begin{equation*}
    \rho
    \big(
      M_1
      \oplustimes
      \cdots
      \oplustimes
      M_t
    \big)
    =
    \rho(M_t)
    \circ
    \cdots
    \circ
    \rho(M_1).
  \end{equation*}
  By the definitions of
  $\rho$
  and
  $\tau_t$,
  this is precisely
  Equation~\eqref{eq:minmax-prefix-product-transformation}.
  The state identity then follows from
  Proposition~\ref{prop:minmax-recurrence-affine-composition}.
\end{proof}

\todo{}

Thus, the parallel algorithms do not require extending the scalar domain to
$\reals\cup\{-\infty,+\infty\}$.
For every finite input sequence, they operate exactly over the finite MinMax
restriction generated by the values occurring in that computation.
In particular,
$\vmin$
and
$\vmax$
do not merely approximate
$-\infty$
and
$+\infty$:
within the data-generated domain they are respectively the exact identities
of
$\oplus$
and
$\odot$.
The identity element required by parallel scan is therefore derived entirely
from the finite data of the computation.

\clearpage{}%

\clearpage{}%

\section{Selected Intensional-Derivative Operators}
\label{app:selected-intensional-derivative-operators}
\label{sec:selected-intensional-derivative-operators}

\todo{}

\localtableofcontents

\subsection{Computation Graphs}
\label{subsec:comp-graphs}

\begin{closingdefinition}[Computation graph]
  \label{def:computation-graph}
  Let $\mathcal{P}$ be a class of functions.

  A \emph{$\mathcal{P}$-computation graph}, 
  or simply \emph{computation graph},
  $G$
  of 
  input dimension $d_\mathrm{in} \in \naturals$ and
  output dimension $d_\mathrm{out} \in \posnaturals$,
  is a tuple
  \begin{align*}
    G = \tuple{V, E, \cgnode{v}[][\mathrm{in}], \cgnode{v}[][\mathrm{c}], \cgnode{v}[][\mathrm{out}], \lambda}
  \end{align*}
  consisting of the following components:
  \begin{itemize}
    \item
      a finite set nodes $V$;
    \item
      the \emph{input node-vector}
      \begin{align*}
        \cgnode{v}[][\mathrm{in}] 
        = 
        \ptuple{\cgnode{v}[1][\mathrm{in}], \ldots, \cgnode{v}[d_\mathrm{in}][\mathrm{in}]} 
      \end{align*}
      consisting of of pair-wise distinct nodes from $V$, 
      that we call \emph{input nodes};
    \item
      the \emph{output node-vector}
      \begin{align*}
        \cgnode{v}[][\mathrm{out}] 
        = 
        \ptuple{\cgnode{v}[1][\mathrm{out}], \ldots, \cgnode{v}[d_\mathrm{out}][\mathrm{out}]} 
      \end{align*}
      consisting of of pair-wise distinct nodes from $V \in \entries{\cgnode{v}[][\mathrm{in}]}$, 
      that we call \emph{output nodes};
    \item
      the \emph{computation node-vector} $\cgnode{v}[][\mathrm{c}]$ consisting of pair-wise distinct
      nodes from $V \setminus \entries{\cgnode{v}[][\mathrm{in}]}$, 
      that we call \emph{computation nodes}, 
      and 
      such that $\entries{\cgnode{v}[][\mathrm{in}]} \cup \entries{\cgnode{v}[][\mathrm{c}]} = V$;
    \item
      a finite set of directed multi-edges
      \begin{equation*}
        E
        \subseteq
        V^*
        \times
        V_\mathrm{c},
      \end{equation*}
      such that:
      \begin{itemize}
        \item
          every computation node has exactly one incoming multi-edge;
        \item
          the resulting directed graph is acyclic;
      \end{itemize}
      if the unique incoming multi-edge of a computation node
      $\cgnode{v}$
      is
      \begin{equation*}
        \ptuple{
          \cgnode{u}_1,
          \ldots,
          \cgnode{u}_k
        }
        \to
        \cgnode{v},
      \end{equation*}
      then
      $k$
      is the \emph{arity} of
      $\cgnode{v}$,
      and
      \begin{equation*}
        \cgnode{u}
        \defeq
        \ptuple{
          \cgnode{u}_1,
          \ldots,
          \cgnode{u}_k
        }
      \end{equation*}
      is its \emph{predecessor node-vector},
    \item
      a node-labelling function $\lambda : \entries{V_\mathrm{c}} \to \mathcal{P}$
      such that, 
      for every computation node $\cgnode{v} \in \entries{V_\mathrm{c}}$,
      its arity $k$ matches the input-arity of its label
      $\lambda(\cgnode{v}) : \reals^k \to \reals$.
  \end{itemize}

  For every node $\cgnode{v} \in V$,
  we write $\cglabel{v} \defeq \lambda(\cgnode{v})$ when the labelling function $\lambda$ is clear
  from the context.

  The \emph{value function} of the nodes of $G$ is defined inductively as follows:
  \begin{itemize}
    \item
      for every $i \in \iival{1}{d_\mathrm{in}}$,
      the value of the input node $\cgnode{v}[i][\mathrm{in}]$
      is
      \begin{equation*}
        \cgval{v}[i][{\mathrm{in}}]
        :
        \reals^{d_\mathrm{in}}
        \to
        \reals,
        \qquad\quad
        \cgval{v}[i][\mathrm{in}](x)
        \defeq
        \getentry{x}{i};
      \end{equation*}
    \item
      for every computation node $\cgnode{v} \in V_\mathrm{c}$,
      having predecessor node-vector
      \begin{equation*}
        \cgnode{u}
        =
        \ptuple{
        \cgnode{u}[1],
        \ldots,
        \cgnode{u}[k]
        }
      \end{equation*}
      its value function is defined inductively by
      \begin{equation*}
        \cgval{v}(x)
        \defeq
        \cglabel{v}
        \big(
          \cgval{u}[i](x),
          \ldots,
          \cgval{u}[k](x)
        \big).
      \end{equation*}
  \end{itemize}
  Then,
  the \emph{value function of the graph $G$} is
  \begin{equation*}
    \cgval{G}
    :
    \reals^{d_\mathrm{in}}
    \to
    \reals^{d_\mathrm{out}},
    \qquad\quad
    \cgval{G}(x)
    \defeq
    \bigptuple{
    \cgval{v}[1][\mathrm{out}](x),
      \ldots,
      \cgval{v}[d_\mathrm{out}][\mathrm{out}](x)
    }.
  \end{equation*}

  We say that $G$ \emph{represents} the function $\cgval{G}$.
\end{closingdefinition}

\begin{closingdefinition}[Computation graph with virtual computation nodes]
  \label{def:computation-graph-with-virtual-nodes}
  Let
  $\mathcal{P}$
  be a class of functions,
  let $d_\mathrm{in}\in\naturals$,
  and
  let $d_\mathrm{out}\in\posnaturals$.

  A \emph{$\mathcal{P}$-computation graph with virtual computation nodes},
  also called a \emph{$\mathcal{P}$-virtual-computation graph}
  or simply a \emph{virtual-computation graph},
  is a tuple
  \begin{equation*}
    G
    \defeq
    \tuple{
      V,
      V_\mathrm{v},
      E,
      \cgnode{v}[][\mathrm{in}],
      \cgnode{v}[][\mathrm{c}],
      \cgnode{v}[][\mathrm{out}],
      \lambda
    }
  \end{equation*}
  consisting of the following components:
  \begin{itemize}
    \item
      a finite set $V$ of \emph{concrete nodes};

    \item
      a finite set $V_\mathrm{v}$ of \emph{virtual computation nodes} satisfying
      $V \cap V_\mathrm{v} = \emptyset$;

    \item
      the \emph{input node-vector} $\cgnode{v}[][\mathrm{in}]$
      consisting of pair-wise distinct concrete nodes from $V$,
      called the \emph{input nodes};

    \item
      the \emph{concrete computation node-vector}
      $\cgnode{v}[][\mathrm{c}]$
      consisting of $d_\mathrm{in}$ pair-wise distinct concrete nodes from
      $V \setminus \entries{\cgnode{v}[][\mathrm{in}]}$,
      called the \emph{concrete computation nodes},
      and 
      satisfying 
      $\entries{\cgnode{v}[][\mathrm{in}]} \cup \entries{\cgnode{v}[][\mathrm{c}]} = V$;

    \item
      the \emph{output node-vector} $\cgnode{v}[][\mathrm{out}]$
      consisting of $d_\mathrm{out}$ pair-wise distinct nodes from
      $V \cup V_\mathrm{v}$,
      called the \emph{output nodes};

    \item
      a finite set of directed multi-edges
      \begin{equation*}
        E
        \subseteq
        \left(
          V
          \cup
          V_\mathrm{v}
        \right)^*
        \times
        \left(
          \entries{\cgnode{v}[][\mathrm{c}]}
          \cup
          V_\mathrm{v}
        \right),
      \end{equation*}
      satisfying the following conditions:
      \begin{itemize}
        \item
          every concrete computation node and every virtual computation node
          has exactly one incoming multi-edge;

        \item
          the directed graph induced by
          $E$
          over
          $V\cup V_\mathrm{v}$
          is acyclic;
      \end{itemize}

      for $\cgnode{u} \to \cgnode{v}$ the unique incoming multi-edge of a concrete or virtual computation
      node $\cgnode{v}$,
      the \emph{(input) arity} of $\cgnode{v}$,
      is
      $\arityin{\cgnode{v}} \defeq |\cgnode{u}|$
      and
      its \emph{predecessor node-vector} is $\cgnode{u}$;

    \item
      a node-labelling function
      \begin{equation*}
        \lambda
        :
        \entries{\cgnode{v}[][\mathrm{c}]}
        \cup
        V_\mathrm{v}
        \to
        \mathcal{P}
      \end{equation*}
      such that the arity of every concrete or virtual computation node $\cgnode{v}$ matches the
      input arity of its label, namely $\arityin{\cgnode{v}} = \arityin{\lambda(\cgnode{v})}$.
  \end{itemize}

  We call
  $\arityin{G} \defeq d_\mathrm{in}$
  the \emph{input dimension} or \emph{input arity} of $G$,
  and 
  we call
  $\arityout{G} \defeq d_\mathrm{out}$
  the \emph{output dimension} or \emph{output arity} of $G$.

  When no ambiguity arises,
  we call a virtual computation node simply a
  \emph{virtual node}.

  A \emph{node} of $G$ is a concrete node or a virtual node of $G$.
  A \emph{computation node} of $G$ is a concrete computation node 
  or a virtual computation node of $G$.

  For every computation node $\cgnode{v}$,
  we write
  \begin{equation*}
    \cglabel{v}
    \defeq
    \lambda(\cgnode{v})
  \end{equation*}
  when the labelling function
  $\lambda$
  is clear from the context.

  The \emph{value functions} of the concrete and virtual nodes of $G$
  are defined inductively as follows:
  \begin{itemize}
    \item
      for every
      $i\in\iival{1}{d_\mathrm{in}}$,
      the value function of the input node
      $\cgnode{v}[i][\mathrm{in}]$
      is
      \begin{equation*}
        \cgval{v}[i][\mathrm{in}]
        :
        \reals^{d_\mathrm{in}}
        \to
        \reals,
        \qquad
        \cgval{v}[i][\mathrm{in}](x)
        \defeq
        \getentry{x}{i};
      \end{equation*}

    \item
      for every computation node $\cgnode{v}$
      having predecessor node-vector
      $\cgnode{u}
      =
      \ptuple{
        \cgnode{u}[1],
        \ldots,
        \cgnode{u}[k]
      }$,
      its value function is defined inductively by
      \begin{equation*}
        \cgval{v}(x)
        \defeq
        \cglabel{v}
        \big(
          \cgval{u}[1](x),
          \ldots,
          \cgval{u}[k](x)
        \big).
      \end{equation*}
  \end{itemize}

  The \emph{value function of $G$} is
  \begin{equation*}
    \cgval{G}
    :
    \reals^{d_\mathrm{in}}
    \to
    \reals^{d_\mathrm{out}},
    \qquad
    \cgval{G}(x)
    \defeq
    \bigptuple{
      \cgval{v}[1][\mathrm{out}](x),
      \ldots,
      \cgval{v}[d_\mathrm{out}][\mathrm{out}](x)
    }.
  \end{equation*}

  We say that $G$ \emph{represents} the function $\cgval{G}$.

  A computation graph in the sense of Definition~\ref{def:computation-graph}
  is identified with a virtual-computation graph 
  where $V_\mathrm{v} = \emptyset$.
\end{closingdefinition}

\begin{closingdefinition}[Virtual-node extension]
  \label{def:virtual-node-extension}
  Let
  \begin{equation*}
    G
    =
    \tuple{
      V,
      V_\mathrm{v},
      E,
      \cgnode{v}[][\mathrm{in}],
      \cgnode{v}[][\mathrm{c}],
      \cgnode{v}[][\mathrm{out}],
      \lambda
    }
  \end{equation*}
  be a virtual-computation graph,
  let $\cgnode{u} = \ptuple{ \cgnode{u}[1], \ldots, \cgnode{u}[k] } \in (V\cup V_\mathrm{v})^k$
  be a node-vector presentation of $G$,
  let $f:\reals^k\to\reals$ be a scalar Euclidean function,
  and 
  let
  $\cgnode{w}\notin V\cup V_\mathrm{v}$ be a fresh node.

  The \emph{virtual-node extension of $G$ obtained by adjoining
  $\cgnode{w}$ with predecessor node-vector $\cgnode{u}$ and label $f$}
  is the virtual-computation graph
  \begin{equation*}
    G'
    \defeq
    \tuple{
      V,
      V_\mathrm{v}',
      E',
      \cgnode{v}[][\mathrm{in}],
      \cgnode{v}[][\mathrm{c}],
      \cgnode{v}[][\mathrm{out}],
      \lambda'
    }
  \end{equation*}
  where
  \begin{align*}
    V_\mathrm{v}'
    &\defeq
    V_\mathrm{v}
    \cup
    \{\cgnode{w}\},
    \qquad\quad
    E'
    &\defeq
    E
    \cup
    \{
      \cgnode{u}
      \to
      \cgnode{w}
    \},
    \qquad\quad
    \lambda'
    \defeq
    \lambda
    \cup
    \{
      \cgnode{w}
      \mapsto
      f
    \}.
  \end{align*}

  We say that $\cgnode{w}$ is the \emph{virtual node adjoined by the extension}.

  A \emph{virtual-node extension} of
  $G$
  is any virtual-computation graph obtained from
  $G$
  by finitely many successive applications of this construction.
\end{closingdefinition}

\begin{closingdefinition}[Materialisation]
  \label{def:virtual-graph-materialisation}
  Let $V_\mathrm{fresh}$ be an infinite ordered set of nodes satisfying $V_\mathrm{fresh}\cap V=\emptyset$,
  and
  let $G$ be a virtual-computation graph given by
  \begin{equation*}
    G
    =
    \tuple{
    V,
    V_\mathrm{v},
    E,
    \cgnode{v}[][\mathrm{in}],
    \cgnode{v}[][\mathrm{c}],
    \cgnode{v}[][\mathrm{out}],
    \lambda
    }.
  \end{equation*}

  A \emph{materialisation map} for $G$ with fresh nodes from $V_\mathrm{fresh}$ 
  is an injective map $\mu:V_\mathrm{v}\to V_\mathrm{fresh}$.
  We extend
  $\mu$
  to all concrete and virtual nodes by defining
  \begin{equation*}
    \bar\mu(\cgnode{v})
    \defeq
    \begin{cases}
      \cgnode{v}
      & \text{ if }
      \cgnode{v}\in V,
      \\[.2em]
      \mu(\cgnode{v})
      & \text{ if }
      \cgnode{v}\in V_\mathrm{v},
    \end{cases}
  \end{equation*}
  and we apply $\bar\mu$ components-wise to node-vector and node-matrix presentations.

  The \emph{materialisation of $G$ under $\mu$} is the computation graph
  \begin{equation*}
    \matgraph{G}{\mu}
    \defeq
    \tuple{
    V^\mu,
    E^\mu,
    \cgnode{v}[][\mathrm{in},\mu],
    \cgnode{v}[][\mathrm{c},\mu],
    \cgnode{v}[][\mathrm{out},\mu],
    \lambda^\mu
    }
  \end{equation*}
  having components as follows:
  \begin{itemize}
    \item
      the set of all nodes, the input node-vector, and the output node-vector are as follows:
      \begin{align*}
        V^\mu 
        \defeq 
        V \cup \mu(V_\mathrm{v})
        \qquad\quad
        \cgnode{v}[][\mathrm{in},\mu]
        \defeq
        \cgnode{v}[][\mathrm{in}],
        \qquad\quad
        \cgnode{v}[][\mathrm{out},\mu]
        \defeq
        \bar\mu
        \big(
        \cgnode{v}[][\mathrm{out}]
        \big);
      \end{align*}
    \item
      the computation node-vector
      $\cgnode{v}[][\mathrm{c},\mu]$
      is an ordering of the set
      $\entries{\cgnode{v}[][\mathrm{c}]} \cup \mu(V_\mathrm{v})$ 
      according to the order provided by $V_\mathrm{fresh}$;

    \item
      the multi-edge set is
      \begin{equation*}
        E^\mu
        \defeq
        \big\{
          \bar\mu(\cgnode{u})
          \to
          \bar\mu(\cgnode{v})
          \,\bigm|\,
          \cgnode{u}
          \to
          \cgnode{v}
          \in
          E
        \big\},
      \end{equation*}
      where
      $\bar\mu$
      is applied entrywise to predecessor node-vectors;

    \item
      finally,
      the labelling function is given by
      \begin{equation*}
        \lambda^\mu(\cgnode{v})
        \defeq
        \lambda\big(\bar\mu^{-1}(\cgnode{v})\big)
        \qquad\quad
        \forall \cgnode{v} \in \entries{\cgnode{v}[][\mathrm{c},\mu]}.
      \end{equation*}
  \end{itemize}

  We call
  $\matgraph{G}{\mu}$
  a \emph{materialised graph of $G$}.
  A concrete or virtual node, node-vector, or node-matrix presentation is
  \emph{materialised} by applying $\bar\mu$ entrywise.
\end{closingdefinition}

\begin{closingproposition}[Correctness of materialisation]
  \label{prop:virtual-graph-materialisation-correctness}
  Let
  $G$
  be a
  $\mathcal{P}$-computation graph with virtual computation nodes,
  and let
  $G^\mu
  \defeq
  \matgraph{G}{\mu}$
  be a materialisation of
  $G$.

  Then:
  \begin{itemize}
    \item
      $G^\mu$
      is a
      $\mathcal{P}$-computation graph in the sense of
      Definition~\ref{def:computation-graph};

    \item
      materialisation preserves the value function of every node:
      if
      $\cgnode{v}^\mu
      \defeq
      \bar\mu(\cgnode{v})$,
      then
      $\cgval{v^\mu}
      =
      \cgval{v}$;

    \item
      consequently,
      materialisation preserves the value function of every node-vector and
      node-matrix presentation;

    \item
      in particular, $\cgval{G^\mu} = \cgval{G}$.
  \end{itemize}
  Hence $G$ and every materialised graph of $G$ represent the same function.
\end{closingproposition}
\begin{proof}
  Since
  $\mu$
  is injective and its image is disjoint from
  $V$,
  the map
  $\bar\mu$
  is injective.
  Hence pair-wise distinct nodes remain pair-wise distinct under
  materialisation.

  Every concrete computation node retains its original predecessors,
  while every virtual computation node is replaced by a concrete computation
  node whose predecessors are the materialisations of its original
  predecessors.
  Thus every computation node of
  $G^\mu$
  has exactly one incoming multi-edge.
  Moreover,
  injective relabelling of the nodes preserves acyclicity.
  Therefore
  $G^\mu$
  is a computation graph.

  Since materialisation preserves every node label,
  and every concrete or virtual computation-node label of
  $G$
  belongs to
  $\mathcal{P}$,
  the graph
  $G^\mu$
  is a
  $\mathcal{P}$-computation graph.

  We prove preservation of node-value functions by induction along a
  topological ordering of
  $G$.
  The input nodes are unchanged, so the claim holds for them.

  Let
  $\cgnode{v}$
  be a concrete or virtual computation node with predecessor node-vector
  \begin{equation*}
    \cgnode{u}
    =
    \ptuple{
      \cgnode{u}[1],
      \ldots,
      \cgnode{u}[k]
    },
  \end{equation*}
  and suppose that the claim holds for its predecessors.
  Write
  \begin{equation*}
    \cgnode{v}^\mu
    \defeq
    \bar\mu(\cgnode{v}),
    \qquad
    \cgnode{u}[i]^\mu
    \defeq
    \bar\mu(\cgnode{u}[i]).
  \end{equation*}
  Then
  \begin{align*}
    \cgval{v^\mu}(x)
    &=
    \cglabel{v}
    \big(
      \cgval{u[1]^\mu}(x),
      \ldots,
      \cgval{u[k]^\mu}(x)
    \big)
    \\
    &=
    \cglabel{v}
    \big(
      \cgval{u}[1](x),
      \ldots,
      \cgval{u}[k](x)
    \big)
    \\
    &=
    \cgval{v}(x).
  \end{align*}
  This proves preservation of all node-value functions.

  The claim for node-vector and node-matrix presentations follows
  componentwise.
  Applying it to the output node-vector gives
  \begin{equation*}
    \cgval{G^\mu}
    =
    \cgval{G},
  \end{equation*}
  as required.
\end{proof}

\begin{closingfact}[Uniqueness of materialisation]
  \label{fact:virtual-graph-materialisation-uniqueness}
  Any two materialisations of the same computation graph with virtual
  computation nodes differ only by the names of the fresh computation nodes
  and, possibly, by the ordering chosen for the computation node-vector.

  In particular,
  materialisation is unique up to renaming of the materialised virtual nodes
  and permutation of the computation node-vector.
\end{closingfact}

\begin{closingdefinition}[Node-vector and node-matrix presentations]
  \label{def:computation-graph-vector-matrix-presentations}
  Let us consider a computation graph
  \begin{align*}
    G = \tuple{V, E, \cgnode{v}[][\mathrm{in}], \cgnode{v}[][\mathrm{c}], \cgnode{v}[][\mathrm{out}], \lambda}.
  \end{align*}
  Then,
  \begin{itemize}
    \item
      a \emph{node-vector presentation} for nodes of $G$, having dimension $m\in\naturals$
      is a vector $\cgnode{v} \in V^m$,
      and
      its value function is
      \begin{align*}
        \cgval{v}
        :
        \reals^{d_\mathrm{in}}
        \to
        \reals^m,
        \qquad\quad
        \cgval{v}(x)
        \defeq
        \ptuple{
          \cgval{v}[1](x),
          \ldots,
          \cgval{v}[m](x)
        };
      \end{align*}
    \item
      a \emph{node-matrix presentation} for nodes of $G$, 
      having dimension $m\times n$ for $m,n\in\posnaturals$,
      is a matrix $\cgnode{A} \defeq \in V^{m\times n}$,
      and
      its value function is
      \begin{align*}
        \cgval{A}
        :
        \reals^{d_\mathrm{in}}
        \to
        \reals^{m\times n},
        \qquad\quad
        \cgval{A}(x)
        \defeq
        \big(
        \getentry{\cgval{A}}{i,j}(x)
        \big)_{
          i\in\iival{1}{m},
          \,
          j\in\iival{1}{n}
        }.
      \end{align*}
  \end{itemize}

  Node-vector and node-matrix presentations are structured collections of
  scalar nodes and do not constitute additional kinds of nodes.

  In particular, 
  the input node-vector $\cgnode{v}[][\mathrm{in}]$,
  every predecessor node-vector $\cgnode{u}$,
  and 
  the output node-vector $\cgnode{v}[][\mathrm{out}]$
  of a computation graph are node-vector presentations.
\end{closingdefinition}

\begin{closingdefinition}[Node presentations with virtual nodes]
  \label{def:virtual-node-presentations}
  \todo{}
  Let
  \begin{equation*}
    G
    =
    \tuple{
      V,
      V_\mathrm{v},
      E,
      \cgnode{v}[][\mathrm{in}],
      \cgnode{v}[][\mathrm{c}],
      \cgnode{v}[][\mathrm{out}],
      \lambda
    }
  \end{equation*}
  be a computation graph with virtual computation nodes.

  \emph{Node-vector presentations} and \emph{node-matrix presentations} of $G$
  are defined as in Definition~\ref{def:computation-graph-vector-matrix-presentations},
  with entries taken from
  $V \cup V_\mathrm{v}$.  
  Also their value functions are as in 
  Definition~\ref{def:computation-graph-vector-matrix-presentations}.

  A node-vector or node-matrix presentation is \emph{concrete} if all of its entries belong to $V$,
  and \emph{virtual} otherwise.
\end{closingdefinition}

\begin{closingdefinition}[Independence from graph nodes]
  \label{def:computation-graph-node-independence}
  Let
  $G$
  be a computation graph, and 
  let $\cgnode{u}$ be a node-vector presentation of $G$.

  A node
  $\cgnode{v}$
  of
  $G$
  is \emph{independent of $\cgnode{u}$}
  if it is not reachable by a directed path from any 
  node $\getentry{\cgnode{u}}{i}$ for $i\in\iival{1}{m}$.

  A node-vector or node-matrix presentation is
  \emph{independent of $\cgnode{u}$}
  if each of its scalar nodes is independent of
  $\cgnode{u}$.

  The computation graph
  $G$
  is \emph{independent of $\cgnode{u}$}
  if its output node-vector is independent of
  $\cgnode{u}$.
\end{closingdefinition}

\begin{closingdefinition}[Restriction of a computation graph]
  \label{def:computation-graph-restriction}
  Let
  $G$
  be a computation graph,
  let $\cgnode{u}$ be a node-vector consisting of distinct input nodes of $G$,
  and 
  let $u \in \reals^{|\cgnode{u}|}$.

  The \emph{restriction of $G$ obtained by fixing $\cgnode{u}=u$},
  denoted by
  \begin{equation*}
    \restr{G}{\cgnode{u}=u}\,,
  \end{equation*}
  is obtained by replacing every
  $\getentry{\cgnode{u}}{i}$
  by a zero-arity computation node labelled by the constant $\getentry{u}{i}$.

  Several disjoint input node-vectors may be fixed simultaneously. For instance, 
  for $\cgnode{w}$ an additional node-vector
  consisting of pair-wise distinct input nodes of $G$
  and such that $\entries{\cgnode{w}} \cap \entries{\cgnode{u}} = \emptyset$,
  we may write
  \begin{equation*}
    \restr{G}{
      \cgnode{u}=u,\,
      \cgnode{w}=w
    }
  \end{equation*}
  for $w \in \reals^{|\cgnode{w}|}$.
\end{closingdefinition}

\begin{closingfact}[Correctness of computation-graph restriction]
  \label{fact:computation-graph-restriction}
  Restricting a computation graph amounts to partially evaluating its value
  function at the corresponding fixed input values.

  In particular, 
  for $G$ a computation graph
  with input node-vector split into two node-vectors as
  \begin{equation*}
    \cgnode{v}[][\mathrm{in}] = \ptuple{ \cgnode{u}, \cgnode{w} }
  \end{equation*}
  letting
  \begin{equation*}
    G_u
    \defeq
    \restr{G}{\cgnode{u}=u}
    \qquad\quad 
    \forall u \in \reals^{|\cgnode{u}|},
  \end{equation*}
  we have
  \begin{equation*}
    \cgval{G}[u](w)
    =
    \cgval{G}(u,w)
    \qquad\quad
    \forall u \in \reals^{|\cgnode{u}|},\;
    \forall w \in \reals^{|\cgnode{w}|}.
  \end{equation*}
\end{closingfact}

\begin{closingdefinition}[Output restriction of a computation graph]
  \label{def:computation-graph-output-restriction}
  Let us consider a computation graph
  \begin{equation*}
    G \defeq 
    \bigtuple{ V, \cgnode{v}[][\mathrm{in}], \cgnode{v}[][\mathrm{c}],  \cgnode{v}[][\mathrm{out}], \lambda }
  \end{equation*}
  having 
  input dimension $d_\mathrm{in}$
  and
  output dimension $d_\mathrm{out}$,
  and
  let 
  $I \defeq \ptuple{ i_1, \ldots, i_n }$
  be a vector of pair-wise distinct indices from 
  $\iival{1}{d_\mathrm{out}}$, for some $n \in \iival{1}{d_\mathrm{out}}$.

  The \emph{output-projection of $G$ to coordinates $I$},
  is the computation graph 
  \begin{align*}
    & \graphindexrestr{G}{I} \defeq 
    \bigtuple{V, \cgnode{v}[][\mathrm{in}], \cgnode{v}[][\mathrm{c}], \cgnode{v}[I][\mathrm{out}], \lambda}
    \\[.5em]
    \text{where }\;
    & \cgnode{v}[I][\mathrm{out}]
    \defeq
    \bigptuple{
      \getentry{V^{\mathrm{out}}}{i_1},
      \ldots,
      \getentry{V^{\mathrm{out}}}{i_n}
    }.
  \end{align*}

  For $j \in \iival{1}{d_\mathrm{out}}$,
  the \emph{scalar output-projection of $G$ to its $j$-th output coordinate}
  is the computation graph
  \begin{equation*}
    \graphindexrestr{G}{j} \defeq \graphindexrestr{G}{J}
  \end{equation*}
  for $J$ the singleton tuple $J \defeq \ptuple{j}$.
\end{closingdefinition}

\begin{closingfact}[Correctness of computation-graph output projection]
  \label{fact:computation-graph-output-projection}
  Projecting the output of a computation graph amounts to projecting the
  corresponding coordinates of its value function, as described next.

  Let $G$ represent
  \begin{equation*}
    f
    :
    \reals^{d_\mathrm{in}}
    \to
    \reals^{d_\mathrm{out}},
    \qquad\quad
    f(x)
    =
    \bigptuple{
      f_1(x),
      \ldots,
      f_{d_\mathrm{out}}(x)
    },
  \end{equation*}
  and 
  let $I \defeq \ptuple{ i_1, \ldots, i_n }$
  be a vector of pair-wise distinct indices from
  $\iival{1}{d_\mathrm{out}}$,
  for some
  $n\in\iival{1}{d_\mathrm{out}}$.

  Then,
  the output-projection
  $\graphindexrestr{G}{I}$
  represents the function
  \begin{equation*}
    f_I
    :
    \reals^{d_\mathrm{in}}
    \to
    \reals^n,
    \qquad\quad
    f_I(x)
    =
    \bigptuple{
      f_{i_1}(x),
      \ldots,
      f_{i_n}(x)
    }.
  \end{equation*}
  Equivalently,
  \begin{equation*}
    \graphindexrestr{\cgval{G}}{I}(x)
    =
    \bigptuple{
      \cgval{G}[i_1](x),
      \ldots,
      \cgval{G}[i_n](x)
    }
    \qquad\quad
    \forall x\in\reals^{d_\mathrm{in}}.
  \end{equation*}

  In particular, for every
  $j\in\iival{1}{d_\mathrm{out}}$,
  the scalar output-projection
  $\graphindexrestr{G}{j}$
  represents the scalar function
  \begin{equation*}
    f_j
    :
    \reals^{d_\mathrm{in}}
    \to
    \reals,
  \end{equation*}
  and it satisfies
  \begin{equation*}
    \graphindexrestr{\cgval{G}}{j}(x)
    =
    \cgval{G}[j](x)
    =
    f_j(x)
    \qquad\quad
    \forall x\in\reals^{d_\mathrm{in}}.
  \end{equation*}
\end{closingfact}

\begin{closingdefinition}[Computation-graph isomorphism]
  \label{def:computation-graph-isomorphism}
  Let us consider a pair of computation graphs 
  \begin{align*}
      G_1 
      = 
      \tuple{
        V_1, 
        E_1,
        \cgnode{v}[1][\mathrm{in}], 
        \cgnode{v}[1][\mathrm{c}], 
        \cgnode{v}[1][\mathrm{out}], 
        \lambda_1
      },
      \qquad\qquad
      G_2 
      = 
      \tuple{
        V_2, 
        E_2,
        \cgnode{v}[2][\mathrm{in}], 
        \cgnode{v}[2][\mathrm{c}], 
        \cgnode{v}[2][\mathrm{out}], 
        \lambda_2
    }
  \end{align*}
  where each $G_i$ has input and output dimension $d_i^\mathrm{in}$ and $d_i^\mathrm{out}$,
  respectively.

  A \emph{computation-graph isomorphism from $G_1$ to $G_2$}
  is a graph isomorphism
  \begin{align*}
    \rho : V_1 \to V_2
  \end{align*}
  satisfying the following conditions:
  \begin{itemize}
    \item
      the input and output dimensions agree:
      \begin{align*}
        d_1^\mathrm{in}
        =
        d_2^\mathrm{in},
        \qquad
        d_1^\mathrm{out}
        =
        d_2^\mathrm{out};
      \end{align*}
    \item
      input nodes and their ordering are preserved:
      \begin{align*}
        \rho\big(
          \getentry{\cgnode{v}[1][\mathrm{in}]}{i}
        \big)
        =
        \getentry{\cgnode{v}[2][\mathrm{in}]}{i}
        \qquad\quad
        \forall i\in\iival{1}{d_1^\mathrm{in}};
      \end{align*}
    \item
      computation nodes are preserved:
      \begin{align*}
        \rho(\cgnode{v}[1][\mathrm{c}]) = \cgnode{v}[2][\mathrm{c}];
      \end{align*}
    \item
      output nodes and their ordering are preserved:
      \begin{align*}
        \rho\big(
          \getentry{\cgnode{v}[1][\mathrm{out}]}{i}
        \big)
        =
        \getentry{\cgnode{v}[2][\mathrm{out}]}{i}
        \qquad\quad
        \forall i\in\iival{1}{d_1^\mathrm{out}};
      \end{align*}
    \item
      computation-node labels are preserved:
      \begin{align*}
        \lambda_2(\rho(\cgnode{v}))
        =
        \lambda_1(\cgnode{v})
        \qquad\quad
        \forall \cgnode{v} \in \cgnode{v}[1][\mathrm{c}].
      \end{align*}
  \end{itemize}

  Two computation graphs are called \emph{isomorphic}, 
  or \emph{equal up to a renaming of nodes},
  if there exists a computation-graph isomorphism between them.
  We again denote this relation by
  \begin{align*}
    G_1 \cong G_2.
  \end{align*}
\end{closingdefinition}

\begin{proposition}[Isomorphic graphs are semantically equivalent]
  \label{prop:isomorphic-computation-graphs-same-function}
  Let
  $G_1$
  and
  $G_2$
  be computation graphs.
  If
  \begin{equation*}
    G_1
    \cong
    G_2,
  \end{equation*}
  then they represent the same function:
  \begin{equation*}
    \cgval{G}[1]
    =
    \cgval{G}[2].
  \end{equation*}
\end{proposition}
\begin{proof}
  Let
  \begin{align*}
      G_1
      =
      \tuple{
        V_1,
        E_1,
        \cgnode{v}[1][\mathrm{in}],
        \cgnode{v}[1][\mathrm{c}],
        \cgnode{v}[1][\mathrm{out}],
        \lambda_1
      },
      \qquad
      G_2
      =
      \tuple{
        V_2,
        E_2,
        \cgnode{v}[2][\mathrm{in}],
        \cgnode{v}[2][\mathrm{c}],
        \cgnode{v}[2][\mathrm{out}],
        \lambda_2
      },
  \end{align*}
  and let $\rho : V_1 \to V_2$ be a computation-graph isomorphism from $G_1$ to $G_2$.
  We show, following a topological ordering of the computation nodes of $G_1$,
  that every node
  $\cgnode{v}\in V_1$
  and its corresponding node
  $\rho(\cgnode{v})\in V_2$
  have the same value function.

  The claim holds for the input nodes.
  In fact, for every
  $i\in\iival{1}{d_1^\mathrm{in}}$,
  Definition~\ref{def:computation-graph-isomorphism}
  gives
  \begin{equation*}
    \rho
    \big(
      \getentry{\cgnode{v}[1][\mathrm{in}]}{i}
    \big)
    =
    \getentry{\cgnode{v}[2][\mathrm{in}]}{i},
  \end{equation*}
  and the corresponding input nodes therefore have the same value function.

  Let now
  $\cgnode{v}\in\cgnode{v}[1][\mathrm{c}]$
  be a computation node having predecessor node-vector
  \begin{equation*}
    \ptuple{
      \cgnode{u}_1,
      \ldots,
      \cgnode{u}_k
    },
  \end{equation*}
  and suppose inductively that each
  $\cgnode{u}_j$
  and
  $\rho(\cgnode{u}_j)$
  have the same value function.
  Since
  $\rho$
  is a graph isomorphism,
  the predecessor node-vector of
  $\rho(\cgnode{v})$
  is
  \begin{equation*}
    \ptuple{
      \rho(\cgnode{u}_1),
      \ldots,
      \rho(\cgnode{u}_k)
    }.
  \end{equation*}
  Furthermore, computation-node labels are preserved, so
  \begin{equation*}
    \lambda_2
    \big(
      \rho(\cgnode{v})
    \big)
    =
    \lambda_1(\cgnode{v}).
  \end{equation*}
  Hence, 
  letting 
  $\cgnode{v}[\rho] \defeq \rho(\cgnode{v})$
  and 
  $\cgnode{v}[i,\rho] \defeq \rho(\cgnode{u}[i])$,
  for every input $x\in\reals^{d_1^\mathrm{in}}$,
  we have
  \begin{align*}
    \cgval{v}[\rho](x)
    &=
    \cgval{v}[\rho]
    \big(
      \cgval{u}[i,\rho](x),
      \ldots,
      \cgval{u}[k,\rho](x),
    \big)
    \\[.2em]
    &=
    \cgval{v}[\rho]
    \big(
      \cgval{u}[i](x),
      \ldots,
      \cgval{u}[k](x)
    \big)
    \\[.2em]
    &=
    \cgval{v}(x).
  \end{align*}
  This proves the claim for every computation node.

  Finally, since
  $\rho$
  preserves the output node-vector,
  corresponding output coordinates of
  $G_1$
  and
  $G_2$
  have the same value function.
  Therefore $\cgval{G}[1] = \cgval{G}[2]$, as required.
\end{proof}

\begin{closingdefinition}[Renaming of computation nodes]
  \label{def:computation-graph-equality-up-to-renaming}
  Let us consider a pair of computation graphs
  \begin{align*}
      G_1
      =
      \tuple{
        V_1,
        E_1,
        \cgnode{v}[1][\mathrm{in}],
        \cgnode{v}[1][\mathrm{c}],
        \cgnode{v}[1][\mathrm{out}],
        \lambda_1
      },
      \qquad
      G_2
      =
      \tuple{
        V_2,
        E_2,
        \cgnode{v}[2][\mathrm{in}],
        \cgnode{v}[2][\mathrm{c}],
        \cgnode{v}[2][\mathrm{out}],
        \lambda_2
      }
  \end{align*}
  whose input node-vectors coincide,
  namely $\cgnode{v}[1][\mathrm{in}] = \cgnode{v}[2][\mathrm{in}]$,
  and hence we write 
  $\cgnode{v}[][\mathrm{in}] \defeq 
  \cgnode{v}[1][\mathrm{in}] = \cgnode{v}[2][\mathrm{in}]$.

  We say that
  $G_1$
  and
  $G_2$
  are
  \emph{equal up to a renaming of computation nodes},
  and write
  \begin{equation*}
    G_1
    \cong_\mathrm{c}
    G_2,
  \end{equation*}
  if there exists a computation-graph isomorphism
  $\rho : V_1 \to V_2$
  that acts as the identity function on $\cgnode{v}[][\mathrm{in}]$, namely
  \begin{equation*}
    \rho
    \big(
      \getentry{\cgnode{v}[][\mathrm{in}]}{i}
    \big)
    =
    \getentry{\cgnode{v}[][\mathrm{in}]}{i}
    \qquad\quad
    \forall
    i\in\iival{1}{d^\mathrm{in}}.
  \end{equation*}
  We also say that $G_2$ is \emph{$G_1$ after renaming computation nodes}---and symmetrically that
  $G_1$ is \emph{$G_2$ after renaming computation nodes}.
\end{closingdefinition}

\begin{closingdefinition}[Common-input union of computation graphs]
  \label{def:computation-graph-common-input-union-nonparametric}
  Let $G_1,\ldots,G_m$
  be computation graphs given by 
  \begin{align*}
      G_i 
      = 
      \tuple{
        V_i, 
        E_i,
        \cgnode{v}[][\mathrm{in}], 
        \cgnode{v}[i][\mathrm{c}], 
        \cgnode{v}[i][\mathrm{out}], 
        \lambda_i
      }
      \qquad\qquad \forall i \in \iival{1}{m}
  \end{align*}
  satisfying\footnote{The condition can always be guaranteed via a renaming of computation nodes.}
  \begin{equation*}
    \entries{\cgnode{v}[i][\mathrm{c}]}
    \cap
    \entries{\cgnode{v}[j][\mathrm{c}]}
    =
    \emptyset
    \qquad\quad
    \forall i,j \in \iival{1}{m}\, \text{ with }\, i \neq j.
  \end{equation*}
  Their
  \emph{common-input union} is the computation graph
  \begin{equation*}
    G_1
    \sqcup_{\mathrm{in}}
    \cdots
    \sqcup_{\mathrm{in}}
    G_m
    \,\defeq\,
    G
    = 
    \tuple{
    V, 
    E,
    \cgnode{v}[][\mathrm{in}], 
    \cgnode{v}[][\mathrm{c}], 
    \cgnode{v}[][\mathrm{out}], 
    \lambda
    }
  \end{equation*}
  where:
  \begin{itemize}
    \item
      the computation node-vector is
      \begin{equation*}
        \cgnode{v}[][\mathrm{c}]
        \defeq
        \cgnode{v}[1][\mathrm{c}] \parallel  \cdots \parallel \cgnode{v}[m][\mathrm{c}]
      \end{equation*}
      using $\parallel$ to denote concatenation of vectors;
    \item
      the multi-edge set is $E \defeq \bigcup_{i=1}^m E_i$;
    \item
      the labelling function $\lambda$ is defined for every
      computation node $\cgnode{v}\in \cgnode{v}[][\mathrm{c}]$ as:
      \begin{equation*}
        \lambda(\cgnode{v}) \defeq \lambda_i(\cgnode{v})
      \end{equation*}
      where $i \in \iival{1}{m}$ is the unique index such that
      $\cgnode{v}\in \cgnode{v}[i][\mathrm{c}]$;
    \item
      the output node-vector is
      \begin{equation*}
        \cgnode{v}[][\mathrm{out}]
        \defeq 
        \cgnode{v}[1][\mathrm{out}] \parallel  \cdots \parallel \cgnode{v}[m][\mathrm{out}]
      \end{equation*}
      again using $\parallel$ to denote concatenation of vectors.
  \end{itemize}
\end{closingdefinition}

\begin{closingdefinition}[Parametric computation graph]
  \label{def:parametric-computation-graph}
  Let $\mathcal{P}$ be a class of parametric Euclidean functions.

  A \emph{$\mathcal{P}$-parametric computation graph}
  is a pair
  \begin{equation*}
    \mathcal{G}
    =
    \tuple{
      G,
      \cgnode{v}[][\mathrm{par}]
    },
  \end{equation*}
  where 
  $G$ is a $\mathcal{P}$-computation graph, 
  called the \emph{underlying computation graph} of $\mathcal{G}$, 
  \begin{equation*}
    G
    =
    \tuple{
      V,
      E,
      \cgnode{v}[][\mathrm{in}],
      \cgnode{v}[][\mathrm{c}],
      \cgnode{v}[][\mathrm{out}],
      \lambda
    }
  \end{equation*}
  and 
  $\entries{\cgnode{v}[][\mathrm{par}]}$ 
  is a node-vector of pair-wise distinct nodes from 
  $\entries{\cgnode{v}[][\mathrm{in}]}$;
  we call  
  $\entries{\cgnode{v}[][\mathrm{par}]}$
  the \emph{parameter node-vector},
  and we refer to its components as \emph{parameter nodes}.
  The remaining input nodes form the \emph{external-input-node set} 
  $\cgnode{v}[][\mathrm{ext}]$, 
  obtained from 
  $\cgnode{v}[][\mathrm{in}]$ 
  by removing the nodes occurring in 
  $\cgnode{v}[][\mathrm{par}]$;
  and we refer to the components of $\cgnode{v}[][\mathrm{ext}]$
  as \emph{external-input nodes}.
  The parameter nodes and external-input nodes inherit their ordering from the
  input node-vector
  $\cgnode{v}[][\mathrm{in}]$.
  We denote the corresponding node-vector presentations by 
  $\cgnode{v}[][\mathrm{par}]$ 
  and 
  $\cgnode{v}[][\mathrm{ext}]$ 
  respectively, and their dimensions are
  \begin{equation*}
    d_\mathrm{par}
    \defeq
    |\cgnode{v}[][\mathrm{par}]|\,,
    \qquad\quad
    d_\mathrm{ext}
    \defeq
    |\cgnode{v}[][\mathrm{ext}]|\,.
  \end{equation*}
  The input dimension and output dimension of $\mathcal{G}$ 
  is
  the input dimension $d_\mathrm{in}$ and output dimension $d_\mathrm{out}$ of $G$ respectively.
  Thus, $d_\mathrm{in} = d_\mathrm{ext} + d_\mathrm{par}$.

  By identifying an input
  $x\in\reals^{d_\mathrm{in}}$
  with a pair $\ptuple{u,\theta} \in \reals^{d_\mathrm{ext}} \times \reals^{d_\mathrm{par}}$
  according to the positions of the corresponding nodes in $\cgnode{v}[][\mathrm{in}]$,
  we write the value function of
  $\mathcal{G}$
  as
  \begin{equation*}
    \cgval{\mathcal{G}}
    :
    \reals^{d_\mathrm{ext}}
    \times
    \reals^{d_\mathrm{par}}
    \to
    \reals^{d_\mathrm{out}},
    \qquad\quad
    \cgvalparop{\mathcal{G}}{u}{\theta}
    \defeq
    \cgval{G}(u,\theta).
  \end{equation*}
  Similarly,
  for every node $\cgnode{v}\in V$,
  we write
  $\cgvalparop{v}{u}{\theta}$
  for its value on input $\ptuple{u,\theta}$.

  We say that $\mathcal{G}$ \emph{represents} the parametric function $\cgval{\mathcal{G}}$.
\end{closingdefinition}

\begin{closingdefinition}[Common-input union of parametric computation graphs]
  \label{def:computation-graph-common-input-union}
  Let $m \in \posnaturals$,
  and let us introduce parametric computation graphs
  \begin{equation*}
    \mathcal{G}_i
    =
    \tuple{
      G_i,
      \cgnode{v}[][\mathrm{par}]
    }
    \qquad\quad
    \qquad
    \forall i\in\iival{1}{m},
  \end{equation*}
  having the same input node-vector and the same parameter-node set
  $\cgnode{v}[][\mathrm{par}]$,
  and assume that their underlying computation graphs
  $G_1,\ldots,G_m$
  satisfy the conditions of
  Definition~\ref{def:computation-graph-common-input-union-nonparametric}.

  Their \emph{common-input union}
  is 
  the parametric computation graph
  \begin{equation*}
    \mathcal{G}_1
    \sqcup_{\mathrm{in}}
    \cdots
    \sqcup_{\mathrm{in}}
    \mathcal{G}_m
    \defeq
    \bigtuple{
      G_1
      \sqcup_{\mathrm{in}}
      \cdots
      \sqcup_{\mathrm{in}}
      G_m,
      \cgnode{v}[][\mathrm{par}]
    }.
  \end{equation*}
\end{closingdefinition}

\begin{closingfact}[Correctness of common-input union]
  \label{fact:computation-graph-common-input-union}
  Let
  $Q_1,\ldots,Q_m$
  be computation graphs satisfying the conditions of
  Definition~\ref{def:computation-graph-common-input-union-nonparametric},
  and suppose that
  $Q_i$
  represents
  \begin{equation*}
    q_i
    :
    \reals^{d_\mathrm{in}}
    \to
    \reals^{d_i},
    \qquad
    i\in\iival{1}{m}.
  \end{equation*}
  Then, their common-input union
  \begin{equation*}
    Q_1
    \sqcup_{\mathrm{in}}
    \cdots
    \sqcup_{\mathrm{in}}
    Q_m
  \end{equation*}
  represents the function
  \begin{equation*}
    q \defeq \ptuple{ q_1, \ldots, q_m } 
    :
    \reals^{d_\mathrm{in}}
    \to
    \reals^{d_1+\cdots+d_m},
    \qquad\;
    q(x) \defeq \ptuple{ q_1(x), \ldots, q_m(x) }.
  \end{equation*}

  Similarly, let $\mathcal{Q}_1,\ldots,\mathcal{Q}_m$
  be parametric computation graphs satisfying the conditions of
  Definition~\ref{def:computation-graph-common-input-union},
  and, for every $i \in \iival{1}{m}$,
  let
  \begin{equation*}
    q_i
    :
    \reals^{d_\mathrm{ext}}
    \times
    \reals^{d_\mathrm{par}}
    \to
    \reals^{d_i}
  \end{equation*}
  be the parametric function represented by $\mathcal{Q}_i$.
  Then, their common-input union
  \begin{equation*}
    \mathcal{Q}_1
    \sqcup_{\mathrm{in}}
    \cdots
    \sqcup_{\mathrm{in}}
    \mathcal{Q}_m
  \end{equation*}
  represents the parametric function
  \begin{equation*}
    q \defeq \ptuple{ q_1, \ldots, q_m } :
    \reals^{d_\mathrm{ext}}
    \times
    \reals^{d_\mathrm{par}}
    \to
    \reals^{d_1+\cdots+d_m},
    \qquad\;
    q(u;\theta)
    \defeq
    \ptuple{ q_1(u;\theta), \ldots, q_m(u;\theta) }.
  \end{equation*}
\end{closingfact}

\begin{closingdefinition}[External-input restriction]
  \label{def:parametric-computation-graph-external-input-restriction}
  Let
  $\mathcal{G}
  =
  \tuple{
    G,
    \cgnode{v}[][\mathrm{par}]
  }$
  be a parametric computation graph having external-input node-vector
  $\cgnode{v}[][\mathrm{ext}]$ of dimension $d_\mathrm{ext}$.

  For every $u \in \reals^{d_\mathrm{ext}}$,
  the
  \emph{external-input restriction of $\mathcal{G}$ at $u$}
  is the parametric computation graph
  \begin{equation*}
    \graphextrestr{\mathcal{G}}{u}
    \defeq
    \tuple{
      \restr{G}{\cgnode{v}[][\mathrm{ext}]=u},
      \cgnode{v}[][\mathrm{par}]
    }.
  \end{equation*}
  Thus,
  all external-input nodes are to the have constant values given by the components of $u$,
  while the parameter nodes remain input nodes.

  For 
  every $u \in \reals^{d_\mathrm{ext}}$,
  and 
  every node $\cgnode{v}$ of $G$, 
  we denote by 
  $\graphextrestr{v}{u}$ 
  the corresponding node of 
  $\graphextrestr{\mathcal{G}}{u}$---namely, the same node $\cgnode{v}$ but now seen as a node of
  the graph $\graphextrestr{\mathcal{G}}{u}$.
\end{closingdefinition}

\begin{closingfact}[Correctness of external-input restriction]
  \label{fact:parametric-computation-graph-external-input-restriction}
  Assume the context of
  Definition~\ref{def:parametric-computation-graph-external-input-restriction},

  For every $u\in\reals^{d_\mathrm{ext}}$,
  the following holds:
  \begin{itemize}
    \item
      the restriction $\graphextrestr{\mathcal{G}}{u}$
      has no external-input nodes and has parameter-node vector $\cgnode{v}[][\mathrm{par}]$;
    \item
      for every node $\cgnode{v}$ of $G$,
      the following two identities hold:
      \begin{equation*}
        \graphextrestr{\cgval{v}}{u}(\theta)
        =
        \cgval{v}(u;\theta)
        \quad\text{ and }\quad
        \graphextrestr{\cgval{\mathcal{G}}}{u}(\theta)
        =
        \cgval{\mathcal{G}}(u;\theta)
        \qquad\;
        \forall \theta\in\reals^{d_\mathrm{par}}.
      \end{equation*}
  \end{itemize}
\end{closingfact}

\begin{closingdefinition}[Parameter independence]
  \label{def:parameter-independent-computation-graph}
  Let
  $\mathcal{G} = \tuple{ G, \cgnode{v}[][\mathrm{par}] }$
  be a parametric computation graph.

  A node
  $\cgnode{v}$
  of
  $G$
  is
  \emph{parameter-independent}
  if it is independent of
  $\cgnode{v}[][\mathrm{par}]$
  in the sense of
  Definition~\ref{def:computation-graph-node-independence};
  equivalently,
  we have that $\cgnode{v}$ is not a parameter node and is not reachable by a directed path from any
  parameter node.

  A node-vector or node-matrix presentation of
  $G$
  is
  \emph{parameter-independent}
  if each of its scalar nodes is parameter-independent.

  The parametric computation graph
  $\mathcal{G}$
  is
  \emph{parameter-independent}
  if its output node-vector is parameter-independent.
\end{closingdefinition}

\begin{closingdefinition}[Canonical constant computation graph]
  \label{def:canonical-constant-computation-graph}
  Let $c \in \reals^n$
  and 
  let $\cgnode{v}[][\mathrm{in}]$
  be a vector of $d_\mathrm{in}$ pair-wise distinct input nodes.

  The
  \emph{canonical constant computation graph representing $c$ over $\cgnode{v}[][\mathrm{in}]$}
  is the computation graph
  \begin{equation*}
    \constgraph{c}{\cgnode{v}[][\mathrm{in}]}
    \defeq
    \tuple{
      V,
      E,
      \cgnode{v}[][\mathrm{in}],
      \cgnode{c},
      \cgnode{c},
      \lambda
    }
  \end{equation*}
  having components as follows:
  \begin{itemize}
    \item
      the node-vector $\cgnode{c}$
      is a vector of $n$ pair-wise distinct nodes not occurring in 
      $\entries{\cgnode{v}[][\mathrm{in}]}$;
    \item
      the node set is $V \defeq \entries{\cgnode{v}[][\mathrm{in}]} \cup \entries{\cgnode{c}}$;
    \item
      the edge set is
      $E \defeq \big\{ \ptuple{} \to \cgnode{c}[i] \bigm| i\in\iival{1}{n} \big\}$---it
      consists of multi-edges of zero input-arity;
    \item
      the labelling function is $\lambda : \entries{\cgnode{c}} \to \entries{c}$
      given by
      $\lambda(\cgnode{c}[i]) \defeq \getentry{c}{i}\,$
      for every $i \in \iival{1}{n}$.
  \end{itemize}
  Note that the labelling function $\lambda$ assigns nodes to constant functions, which have zero
  input-arity, hence correctly matching the input-arity of the introduced multi-edges.
\end{closingdefinition}

\begin{closingfact}
  In the context of Definition~\ref{def:canonical-constant-computation-graph},
  \begin{itemize}
    \item
      both the computation-node vector and the output node-vector of 
      $\constgraph{c}{\cgnode{v}[][\mathrm{in}]}$
      coincide with $\cgnode{c}$;
    \item
      the value function of the graph is the following constant function:
      \begin{equation*}
        \constgraphval{c}{\cgnode{v}[][\mathrm{in}]}(x)
        =
        c
        \qquad\quad
        \forall
        x\in\reals^{d_\mathrm{in}}.
      \end{equation*}
  \end{itemize}
\end{closingfact}

\begin{closingfact}[Canonical constant graphs for dynamics]
  Assume the context of Definition~\ref{def:canonical-constant-computation-graph}.

  Let $D$ be parametric dynamics 
  having 
  input space $U \subseteq \Omega_U = \reals^{d_\mathrm{in}}$.
  Let 
  $\cgnode{v}[][\mathrm{in}]$ 
  be a node-vector of $d_\mathrm{in}$ pair-wise distinct input nodes.

  Then, 
  the graph $\constgraph{c}{\cgnode{v}[][\mathrm{in}]}$
  is a canonical constant computation graph representing the constant~$c$
  over the input ambient-space $\Omega_U$ of $D$.
\end{closingfact}

\begin{closingdefinition}[Parametric canonical constant computation graph]
  \label{def:parametric-canonical-constant-computation-graph}
  Let $c \in \reals^n$.
  We define two variants of canonical constant parametric computation graphs.
  \begin{itemize}
    \item
      Let 
      $\mathcal{H} = \tuple{ H, \cgnode{v}[][\mathrm{par}] }$
      be a parametric computation graph,
      and 
      let $\cgnode{v}[][\mathrm{in}]$ be its input node-vector.
      The
      \emph{canonical constant parametric computation graph representing $c$ over $\mathcal{H}$}
      is
      \begin{equation*}
        \constpargraph{c}{\mathcal{H}}
        \defeq
        \bigtuple{
        \constgraph{c}{\cgnode{v}[][\mathrm{in}]},\,
        \cgnode{v}[][\mathrm{par}]
        }.
      \end{equation*}

    \item
      Let $d_\mathrm{in}, d_\mathrm{par} \in \naturals$,
      let $\cgnode{v}[][\mathrm{in}]$ be a node-vector of $d_\mathrm{in}$ pair-wise distinct input nodes,
      and
      let
      $\cgnode{v}[][\mathrm{par}]$ be a node-vector of $d_\mathrm{par}$ pair-wise distinct nodes
      that are not in $\entries{\cgnode{v}[][\mathrm{in}]}$.
      The
      \emph{canonical constant parametric computation graph representing $c$ 
        over the input node-vector 
        $\cgnode{v}[][\mathrm{in}]$ and 
        parameter node-vector 
      $\cgnode{v}[][\mathrm{par}]$}
      is
      \begin{equation*}
        \constparamgraph{c}{
          \cgnode{v}[][\mathrm{in}], 
          \cgnode{v}[][\mathrm{par}] 
          }
        \defeq
        \bigtuple{
        \constgraph{c}{\cgnode{v}[][\mathrm{par}]},\,
        \cgnode{v}[][\mathrm{par}]
        }.
      \end{equation*}
  \end{itemize}
\end{closingdefinition}

\begin{closingfact}
  Assume the context of Definition~\ref{def:parametric-canonical-constant-computation-graph}.
  We have the following facts for canonical constant graphs.

  The canonical graph $\constpargraph{c}{\mathcal{H}}$ satisfies the following:
  \begin{itemize}
    \item
      it has the same external-input nodes and parameter nodes as $\mathcal{H}$;
    \item
      it is parameter-independent;
    \item
      its value function satisfies the following identity:
      \begin{equation*}
        \cgvalconstparop{c}{\mathcal{H}}{u}{\theta} = c
        \qquad\quad
        \forall u\in\reals^{d_\mathrm{ext}},\,
        \forall \theta\in\reals^{d_\mathrm{par}}.
      \end{equation*}
  \end{itemize}

  The canonical constant graph
  $\constparamgraph{c}{\cgnode{v}[][\mathrm{in}], \cgnode{v}[][\mathrm{par}]}$
  satisfies the following:
  \begin{itemize}
    \item
      it has no external-input nodes;
    \item
      it has parameter-node vector $\cgnode{v}[][\mathrm{par}]$;
    \item
      it is parameter-independent;
    \item
      its value function satisfies the following identity:
      \begin{equation*}
        \cgvalconstparop{c}{\cgnode{v}[][\mathrm{in}], \cgnode{v}[][\mathrm{par}]}{u}{\theta} = c
        \qquad\quad
        \forall
        \theta\in\Theta.
        \forall u\in\reals^{d_\mathrm{ext}},\,
        \forall \theta\in \Theta.
      \end{equation*}
  \end{itemize}
\end{closingfact}

\begin{closingfact}[Parametric canonical constant graphs for parametric dynamics]
  Assume the context of Definition~\ref{def:parametric-canonical-constant-computation-graph}.

  Let $D$ be parametric dynamics 
  having 
  input space $U \subseteq \Omega_U = \reals^{d_\mathrm{in}}$
  and
  parameter space $\Theta \subseteq \Omega_\Theta = \reals^{d_\mathrm{par}}$.
  Let 
  $\cgnode{v}[][\mathrm{in}]$ 
  be a node-vector of $d_\mathrm{in}$ pair-wise distinct input nodes,
  and
  let 
  $\cgnode{v}[][\mathrm{par}]$ 
  be a node-vector of $d_\mathrm{par}$ pair-wise distinct nodes that are not in
  $\entries{\cgnode{v}[][\mathrm{in}]}$.

  Then, 
  the graph $\constparamgraph{c}{\cgnode{v}[][\mathrm{in}], \cgnode{v}[][\mathrm{par}]}$
  is a canonical constant parametric computation graph representing the constant~$c$
  over the joint input-parameter ambient-space $\Omega_U \times \Omega_\Theta$ of $D$.
\end{closingfact}

\begin{closingdefinition}[Computation-graph family]
  \label{def:selected-computation-graph-family-new}
  For $X$ a set and $\mathcal{P}$ a class of Euclidean functions,
  a \emph{$\mathcal{P}$-computation-graph family with index set $X$} 
  is map from $X$ to non-empty sets of $\mathcal{P}$-computation graphs, 
  and we write it as
  \begin{equation*}
    \indexedcompgraphfamily
    =
    \big(
      \indexedcompgraphfamily[x]
    \big)_{x \in X},
  \end{equation*}
  meaning that 
  $\indexedcompgraphfamily[x]$ is the set of computation graphs for $x \in X$.

  For $\mathcal{Q}$ and $\mathcal{P}$ classes of Euclidean functions,
  a \emph{$\mathcal{P}$-computation-graph family for $\mathcal{Q}$}
  is a $\mathcal{P}$-computation-graph family with index set $\mathcal{Q}$
  such that,
  for every function $q \in \mathcal{Q}$,
  every computation graph $G \in \indexedcompgraphfamily[q]$
  satisfies the following conditions jointly:
  \begin{enumerate}
    \item
      the input dimension of $G$ equals the input arity of $q$;
    \item
      the output dimension of $G$ equals the output arity of $q$;
    \item
      $G$ represents $q$.
  \end{enumerate}
  In this case,
  for every $q\in\mathcal{Q}$,
  we call
  $\indexedcompgraphfamily[q]$
  the
  \emph{computation-graph family for $q$ induced by $\indexedcompgraphfamily$},
  or simply the
  \emph{computation-graph family for $q$} when $\indexedcompgraphfamily$ is clear from context.

  For $\mathcal{Q}^\mathrm{p}$ 
  and 
  $\mathcal{P}^\mathrm{p}$ classes of parametric Euclidean functions,
  a \emph{$\mathcal{P}^\mathrm{p}$-parametric computation-graph family for $\mathcal{Q}^\mathrm{p}$}
  is a $\mathcal{P}^\mathrm{p}$-computation-graph family with index set $\mathcal{Q}^\mathrm{p}$
  such that,
  for every function $q \in \mathcal{Q}^\mathrm{p}$,
  every computation graph $G \in \indexedcompgraphfamily[q]$
  is a parametric computation graph satisfying the following conditions jointly:
  \begin{enumerate}
    \item
      the input dimension of $G$ equals the input arity of $q$;
    \item
      the parameter dimension of $G$ equals the parameter arity of $q$;
    \item
      the output dimension of $G$ equals the output arity of $q$;
    \item
      $G$ represents $q$.
  \end{enumerate}
  In this case,
  for every
  $q\in\mathcal{Q}^\mathrm{p}$,
  we call
  $\indexedcompgraphfamily[q]$
  the
  \emph{parametric computation-graph family for $q$ induced by $\indexedcompgraphfamily$},
  or simply the \emph{parametric computation-graph family for $q$} when 
  $\indexedcompgraphfamily$ is clear from context.
\end{closingdefinition}

\begin{closingdefinition}[Graph-family admissibility]
  Let $\mathcal{Q}$ be a class of Euclidean functions,
  and
  let $\indexedcompgraphfamily$ be a computation-graph family for $\mathcal{Q}$.
  A pseudo-derivative operator $\sdv$ is
  \emph{admissible for $\indexedcompgraphfamily$}
  if,
  for 
  every function $q \in \mathcal{Q}$ 
  and 
  every graph $G \in \indexedcompgraphfamily[q]$,
  we have that
  $\sdv$ is admissible for $G$.
  The definition applies seamlessly to the parametric case.
\end{closingdefinition}

\begin{closingdefinition}[Graph selectors]
  \label{def:graph-selector}
  Let $\mathcal{Q}$ be a class of Euclidean functions,
  and 
  let $\indexedcompgraphfamily$ be a computation-graph family for $\mathcal{Q}$.
  A \emph{graph selector for $\mathcal{Q}$ over $\indexedcompgraphfamily$}
  is a map
  $\selgraphsymb$
  satisfying
  \begin{equation*}
    \selgraphsymb[q]
    \in
    \indexedcompgraphfamily[q]
    \qquad\quad
    \forall
    q\in\mathcal{Q}.
  \end{equation*}
  The definition applies seamlessly to the parametric case.
\end{closingdefinition}

\subsection{Pseudo-derivative Operators}
\label{subsec:pseudoderivative-operator}

\begin{closingdefinition}[Pseudo-derivative operators]
  \label{def:pseudo-derivative-operator}
  Let $q: X \to Y$ be a Euclidean function 
  having 
  input arity $d_\mathrm{in}$
  and
  output arity $d_\mathrm{out}$.
  A \emph{pseudo-derivative operator} is an operator $\sdv$ that maps 
  Euclidean functions to Euclidean matrix-valued functions.
  The operator $\sdv$ is \emph{applicable} to $q$
  when $\sdv$ is defined for $q$ and it yiels a Euclidean function of the following form:
  \begin{equation*} 
    \sdvop{q} : X \to \reals^{\arityout{q}\times \arityin{q}}.
  \end{equation*}
  In this case,
  the function $\sdvop{q}$ is the \emph{pseudo-derivative function of $q$ under $\sdv$};
  we may also call $\sdvop{q}$ the \emph{pseudo-Jacobian} of $q$ under $\sdv$,
  and 
  the \emph{pseudo-gradient} of $q$ under $\sdv$ when $d_\mathrm{out} = 1$.

  Let $\mathcal{Q}$ be a class of Euclidean functions.
  A \emph{pseudo-derivative operator} applicable to $\mathcal{Q}$ is a 
  pseudo-derivative operator that is applicable to every function $q \in \mathcal{Q}$.
\end{closingdefinition}

\begin{closingdefinition}[Positional partial pseudo-derivatives]
  \label{def:block-pseudo-derivative}
  Let
  $\sdv$
  be a pseudo-derivative operator,
  and let
  \begin{equation*}
    q
    :
    \reals^{d_1}
    \times
    \cdots
    \times
    \reals^{d_k}
    \to
    \reals^{d_\mathrm{out}}
  \end{equation*}
  be a function to which
  $\sdv$
  is applicable.

  For
  \begin{equation*}
    x
    =
    \ptuple{
      x_1,
      \ldots,
      x_k
    },
    \qquad
    x_i\in\reals^{d_i},
  \end{equation*}
  write the pseudo-derivative matrix according to the input-block
  decomposition as
  \begin{equation*}
    \sdvop{q}(x)
    =
    \begin{pmatrix}
      J_1(x)
      &
      \cdots
      &
      J_k(x)
    \end{pmatrix},
    \qquad
    J_i(x)
    \in
    \reals^{d_\mathrm{out}\times d_i}.
  \end{equation*}
  For every
  $i\in\iival{1}{k}$,
  the \emph{pseudo-derivative of $q$ with respect to its $i$-th input block}
  is the function
  \begin{equation*}
    \bsdvop{q}{i}
  \end{equation*}
  defined by
  \begin{equation*}
    \bsdvop{q}{i}(x)
    \defeq
    J_i(x).
  \end{equation*}

  Hence,
  \begin{equation*}
    \sdvop{q}(x)
    =
    \begin{pmatrix}
      \bsdvop{q}{1}(x)
      &
      \cdots
      &
      \bsdvop{q}{k}(x)
    \end{pmatrix}.
  \end{equation*}
\end{closingdefinition}

\begin{closingdefinition}[Named partial pseudo-derivatives]
  \label{def:named-partial-pseudo-derivatives}
  Let
  $\sdv$
  be a pseudo-derivative operator,
  and let
  \begin{equation*}
    q
    :
    \reals^{d_1}
    \times
    \cdots
    \times
    \reals^{d_k}
    \to
    \reals^{d_\mathrm{out}}
  \end{equation*}
  be a function to which
  $\sdv$
  is applicable.

  When evaluating the pseudo-derivative at a point
  \begin{equation*}
    x
    =
    \ptuple{
      x_1,
      \ldots,
      x_k
    },
  \end{equation*}
  we use a \emph{named partial pseudo-derivative notation} in which the subscript
  identifies one of the input blocks of the evaluation point.

  More precisely, for every
  $i\in\iival{1}{k}$,
  we write
  \begin{equation*}
    \psdvop{q}{x_i}
    \big(
      x_1,
      \ldots,
      x_k
    \big)
    \defeq
    \bsdvop{q}{i}
    \big(
      x_1,
      \ldots,
      x_k
    \big).
  \end{equation*}
  Thus,
  \begin{equation*}
    \psdvop{q}{x_i}
    \big(
      x_1,
      \ldots,
      x_k
    \big)
    \in
    \reals^{d_\mathrm{out}\times d_i}
  \end{equation*}
  denotes the block of the pseudo-derivative matrix at
  $\ptuple{x_1,\ldots,x_k}$
  corresponding to variations of the input block
  $x_i$.

  Hence,
  $\bsdvop{q}{i}$
  denotes a block pseudo-derivative function indexed by the position of the
  corresponding input block, whereas
  \begin{equation*}
    \psdvop{q}{x_i}
    \big(
      x_1,
      \ldots,
      x_k
    \big)
  \end{equation*}
  is a pointwise notation in which the differentiated input block is identified
  by its name at the evaluation point.

  When an input block at the evaluation point is itself given by the value of a
  function or expression, we use the same convention.

  More precisely, suppose that
  \begin{equation*}
    q
    :
    \reals^{d_1}
    \times
    \reals^{d_2}
    \to
    \reals^{d_\mathrm{out}},
  \end{equation*}
  and consider an evaluation point of the form
  \begin{equation*}
    \ptuple{
      g(x),
      y
    }.
  \end{equation*}
  The fully explicit named partial pseudo-derivative notation for the first input
  block is
  \begin{equation*}
    \psdvop{q}{g(x)}
    \big(
      g(x),
      y
    \big).
  \end{equation*}
  When there is no ambiguity, we abbreviate the name of this input block by the
  symbol of the function producing it, and write
  \begin{equation*}
    \psdvop{q}{g}
    \big(
      g(x),
      y
    \big)
    \defeq
    \psdvop{q}{g(x)}
    \big(
      g(x),
      y
    \big).
  \end{equation*}

  More generally, named subscripts may abbreviate the expression defining the
  corresponding input block whenever the association between the name and the
  block at the evaluation point is clear from context.
\end{closingdefinition}

\begin{closingdefinition}[Graph-admissible pseudo-derivative operator]
  \label{def:admissible-pseudo-derivative-operator}
  A pseudo-derivative operator $\sdv$ is 
  \emph{admissible for a computation graph}~$G$ if it is applicable to every 
  label of a computation node of $G$.
  A pseudo-derivative operator $\sdv$
  is \emph{admissible for a parametric computation graph}
  $\mathcal{G}=\tuple{G,\cgnode{v}[][\mathrm{par}]}$
  if it is admissible for its underlying computation graph $G$.
\end{closingdefinition}

\begin{closingdefinition}[Graph-induced pseudo-derivatives]
  \label{def:graph-induced-pseudo-derivative}
  Let $G$ be a computation graph having input dimension $d_\mathrm{in}$,
  and 
  let $\sdv$ be a pseudo-derivative operator that is admissible for $G$.

  The 
  \emph{pseudo-derivative operator induced by $G$ with local pseudo-derivative operator $\sdv$}
  is the operator $\sdv[G]$ defined first on the nodes of $G$ and then on $G$ itself, as follows: 
  \begin{itemize}
    \item
      for every input node
      $\cgnode{v}^{\mathrm{in}}_i$,
      we define its pseudo-derivative as:
      \begin{equation*}
        \sdvop[G]{\cgnode{v}^{\mathrm{in}}_i}(x)
        \,\defeq\,
        D\cgval{v}^{\mathrm{in}}_i(x)
        \,=\,
        e_i^\top
        \, \in \reals^{1\times d_\mathrm{in}}
        \qquad\qquad \forall x \in \reals^{\mathrm{in}};
      \end{equation*}
      where $e_i$ is the $i$-th standard basis vector of $\reals^{d_\mathrm{in}}$'

    \item
      for every zero-arity computation node $\cgnode{v}$,
      we define its pseudo-derivative as:
      \begin{equation*}
        \sdvop[G]{\cgnode{v}}(x)
        \,\defeq\,
        0
        \,\in
        \reals^{1\times d_\mathrm{in}}
        \qquad\qquad \forall x \in \reals^{\mathrm{in}};
      \end{equation*}
    \item
      for every node-vector
      $\cgnode{u} = \ptuple{\cgnode{u}_1,\ldots,\cgnode{u}_k}$,
      we define its pseudo-derivative by vertically stacking the
      pseudo-derivatives of its components:
      \begin{equation*}
        \sdvop[G]{\cgnode{u}}(x)
        \,\defeq\,
        \begin{pmatrix}
          \sdvop[G]{\cgnode{u}_1}(x)
          \\
          \vdots
          \\
          \sdvop[G]{\cgnode{u}_k}(x)
        \end{pmatrix}
        \qquad\qquad \forall x \in \reals^{\mathrm{in}},
      \end{equation*}
      where the right-hand side is understood as a block-column matrix.

    \item
      for every positive-arity computation node $\cgnode{v}$
      having input node-vector $\cgnode{u}$,
      we define its pseudo-derivative as:
      \begin{equation*}
        \sdvop[G]{\cgnode{v}}(x)
        \,\defeq\,
        \sdvop{\cglabel{v}}
        \big(
          \cgval{u}(x)
        \big)
        \cdot
        \sdvop[G]{\cgnode{u}}(x)
        \qquad\qquad \forall x \in \reals^{\mathrm{in}};
      \end{equation*}
    \item
      finally, 
      for $\cgnode{v}^{\mathrm{out}}$ the output node-vector of $G$,
      we define the pseudo-derivative of $G$ as:
      \begin{equation*}
        \sdv G
        \defeq
        \sdvop[G]{\cgnode{v}^{\mathrm{out}}}.
      \end{equation*}
  \end{itemize}

  The definition applies to every parametric computation graph
  $\mathcal{G}=\ptuple{G,\cgnode{v}[][\mathrm{par}]}$
  by applying it to its underlying computation graph $G$.
  In particular,
  \begin{equation*}
    \sdv \mathcal{G} \defeq \sdv G.
  \end{equation*}
\end{closingdefinition}

\begin{proposition}[Correctness of graph-induced ordinary derivatives]
  \label{prop:correctness-graph-induced-ordinary-derivative}
  Let $\dv$ be the ordinary derivative operator,
  let $\mathcal{P}_\mathrm{diff}$ be a class of differentiable Euclidean functions,
  let $G$ be an $\mathcal{P}_\mathrm{diff}$-computation graph,
  and 
  let $f \defeq \cgval{G}$ be the function represented by $G$.
  Then,
  \begin{itemize}
    \item
      the operator $\dv$ is a pseudo-derivative operator that is admissible for $G$;
    \item 
      the function $f$ is differentiable and the graph-induced pseudo-derivative of $G$
  coincides with the ordinary derivative of $f$:
  \begin{equation*}
    \dv G = \dv f.
  \end{equation*}
  \end{itemize}
\end{proposition}
\begin{proof}
  Since $G$ is an $\mathcal{P}_\mathrm{diff}$-computation graph and every function in
  $\mathcal{P}_\mathrm{diff}$ is differentiable, the ordinary derivative operator $\dv$
  is admissible for $G$.

  We prove, following a topological ordering of the nodes of $G$, that
  every node $\cgnode{v}$ has a differentiable value function
  $\cgval{v}$ and satisfies
  \begin{equation}
    \label{eq:graph-induced-ordinary-derivative-node}
    \sdvop[G]{\cgnode{v}}(x)
    =
    \dv\cgval{v}(x)
    \qquad
    \forall x \in \reals^{d_\mathrm{in}},
  \end{equation}
  where the local pseudo-derivative operator $\sdv$ is taken to be $\dv$.

  First, let $\cgnode{v}^{\mathrm{in}}_i$ be an input node.
  Its value function is the coordinate projection
  \begin{equation*}
    \cgval{v}^{\mathrm{in}}_i(x)
    =
    \getentry{x}{i},
  \end{equation*}
  which is differentiable, and by
  Definition~\ref{def:graph-induced-pseudo-derivative},
  \begin{equation*}
    \sdvop[G]{\cgnode{v}^{\mathrm{in}}_i}(x)
    =
    e_i^\top
    =
    \dv\cgval{v}^{\mathrm{in}}_i(x).
  \end{equation*}

  Now let $\cgnode{v}$ be a zero-arity computation node.
  Its value function is constant as a function of the input $x$, and is
  therefore differentiable with
  \begin{equation*}
    \dv\cgval{v}(x)=0.
  \end{equation*}
  Again by
  Definition~\ref{def:graph-induced-pseudo-derivative},
  \begin{equation*}
    \sdvop[G]{\cgnode{v}}(x)
    =
    0
    =
    \dv\cgval{v}(x).
  \end{equation*}

  Finally, let $\cgnode{v}$ be a positive-arity computation node with
  predecessor node-vector
  \begin{equation*}
    \cgnode{u}
    =
    \ptuple{\cgnode{u}_1,\ldots,\cgnode{u}_k}.
  \end{equation*}
  By the induction hypothesis, every value function
  $\cgval{u}_i$ is differentiable and
  \begin{equation*}
    \sdvop[G]{\cgnode{u}_i}(x)
    =
    \dv\cgval{u}_i(x).
  \end{equation*}
  Hence the node-vector value function
  \begin{equation*}
    \cgval{u}
    =
    \ptuple{
      \cgval{u}_1,
      \ldots,
      \cgval{u}_k
    }
  \end{equation*}
  is differentiable, with derivative obtained by vertically stacking
  the derivatives of its components:
  \begin{equation*}
    \dv\cgval{u}(x)
    =
    \begin{pmatrix}
      \dv\cgval{u}_1(x)
      \\
      \vdots
      \\
      \dv\cgval{u}_k(x)
    \end{pmatrix}.
  \end{equation*}
  By the induction hypothesis and the definition of the
  pseudo-derivative of a node-vector,
  \begin{equation*}
    \sdvop[G]{\cgnode{u}}(x)
    =
    \dv\cgval{u}(x).
  \end{equation*}

  Since $G$ is an $\mathcal{P}_\mathrm{diff}$-computation graph,
  its node label $\cglabel{v}$ belongs to $\mathcal{P}_\mathrm{diff}$ and is therefore
  differentiable.  Furthermore,
  \begin{equation*}
    \cgval{v}
    =
    \cglabel{v}
    \circ
    \cgval{u}.
  \end{equation*}
  Thus $\cgval{v}$ is differentiable and, by the chain rule,
  \begin{align*}
    \dv\cgval{v}(x)
    &=
    \dv\cglabel{v}
    \big(
      \cgval{u}(x)
    \big)
    \cdot
    \dv\cgval{u}(x)
    \\[.3em]
    &=
    \dv\cglabel{v}
    \big(
      \cgval{u}(x)
    \big)
    \cdot
    \sdvop[G]{\cgnode{u}}(x)
    \\[.3em]
    &=
    \sdvop[G]{\cgnode{v}}(x),
  \end{align*}
  where the last equality follows from
  Definition~\ref{def:graph-induced-pseudo-derivative}.
  This proves
  \eqref{eq:graph-induced-ordinary-derivative-node}
  for every node of $G$.

  Applying the same componentwise argument to the output node-vector
  $\cgnode{v}^{\mathrm{out}}$, we obtain that
  \begin{equation*}
    f
    =
    \cgval{G}
    =
    \cgval{v}[][\mathrm{out}]
  \end{equation*}
  is differentiable and
  \begin{align*}
    \dv f(x)
    &=
    \dv\cgval{v}[\mathrm{out}](x)
    \\
    &=
    \sdvop[G]{\cgnode{v}^{\mathrm{out}}}(x)
    \\
    &=
    \dv G(x).
  \end{align*}
  Therefore
  \begin{equation*}
    \dv G = \dv f,
  \end{equation*}
  as required.
\end{proof}

\begin{lemma}[Parameter-independent graphs have zero parameter derivative]
  \label{lem:parameter-independent-zero-derivative}
  Let $\mathcal{G}$ be a parametric computation graph,
  and
  let $\sdv$ be a pseudo-derivative operator that is admissible for $\mathcal{G}$.

  Let $u$ and $\theta$ be an external input and parameter values for $\mathcal{G}$, respectively.
  \begin{itemize}
    \item
      If $\cgnode{v}$ is a parameter-independent node-vector or node-matrix presentation of
      $\mathcal{G}$,
      then
      \begin{equation*}
        \sdvop{\graphextrestr{\cgnode{v}}{u}}(\theta) = 0.
      \end{equation*}
    \item
      If $\mathcal{G}$ is parameter-independent, then
      \begin{equation*}
        \sdvop{\graphextrestr{\mathcal{G}}{u}}(\theta) = 0.
      \end{equation*}
  \end{itemize}
\end{lemma}
\begin{proof}
  We first prove that every parameter-independent node
  $\cgnode{v}$ of $\mathcal{G}$ satisfies
  \begin{equation}
    \label{eq:parameter-independent-node-zero-derivative}
    \sdvop{\graphextrestr{v}{u}}(\theta)
    =
    0.
  \end{equation}
  We proceed according to a topological ordering of the nodes of
  $\graphextrestr{\mathcal{G}}{u}$.

  Suppose first that
  $\cgnode{v}$
  is an external-input node of
  $\mathcal{G}$.
  By
  Definition~\ref{def:parametric-computation-graph-external-input-restriction},
  $\graphextrestr{v}{u}$
  is a zero-arity computation node.
  Hence, by
  Definition~\ref{def:graph-induced-pseudo-derivative},
  \begin{equation*}
    \sdvop{\graphextrestr{v}{u}}(\theta)
    =
    0.
  \end{equation*}

  Now let
  $\cgnode{v}$
  be a parameter-independent computation node of
  $\mathcal{G}$,
  with input node-vector
  \begin{equation*}
    \cgnode{w}
    =
    \ptuple{
      \cgnode{w}_1,
      \ldots,
      \cgnode{w}_k
    }.
  \end{equation*}
  Every predecessor
  $\cgnode{w}_i$
  is parameter-independent.
  Indeed, if
  $\cgnode{w}_i$
  were a parameter node or reachable from a parameter node, then
  $\cgnode{v}$
  would also be reachable from a parameter node, contradicting its
  parameter-independence.

  Therefore, by the induction hypothesis,
  \begin{equation*}
    \sdvop{\graphextrestr{w_i}{u}}(\theta)
    =
    0
    \qquad
    \forall i\in\iival{1}{k}.
  \end{equation*}
  Thus,
  by the definition of the pseudo-derivative of a node-vector,
  \begin{equation*}
    \sdvop{\graphextrestr{w}{u}}(\theta)
    =
    0.
  \end{equation*}

  Since external-input restriction leaves
  $\cgnode{v}$
  and its label unchanged,
  the recursive definition of the graph-induced pseudo-derivative gives
  \begin{align*}
    \sdvop{\graphextrestr{v}{u}}(\theta)
    \,=\,
    \sdvop{\cglabel{v}}
    \big( \cgval{w}_u(\theta) \big)
    \cdot
    \sdvop{\graphextrestr{w}{u}}(\theta)
    \,=\,
    0
    \qquad\text{ where }\quad
    w_u \defeq \graphextrestr{w}{u}.
  \end{align*}
  This proves
  \eqref{eq:parameter-independent-node-zero-derivative}
  for every parameter-independent node.

  If
  $\cgnode{v}$
  is a parameter-independent node-vector presentation, then every scalar node
  occurring in
  $\cgnode{v}$
  is parameter-independent.
  Hence
  \eqref{eq:parameter-independent-node-zero-derivative}
  holds coordinatewise, and therefore
  \begin{equation*}
    \sdvop{\graphextrestr{\cgnode{v}}{u}}(\theta)
    =
    0.
  \end{equation*}
  The same argument applies entrywise to a parameter-independent node-matrix
  presentation.
  This proves the first claim.

  Finally, if
  $\mathcal{G}$
  is parameter-independent, then its output node-vector
  $\cgnode{v}^{\mathrm{out}}$
  is parameter-independent.
  By the first claim,
  \begin{equation*}
    \sdvop{
      \graphextrestr{\cgnode{v}^{\mathrm{out}}}{u}
    }(\theta)
    =
    0.
  \end{equation*}
  Since
  $\graphextrestr{\cgnode{v}^{\mathrm{out}}}{u}$
  is the output node-vector of
  $\graphextrestr{\mathcal{G}}{u}$,
  it follows that
  \begin{equation*}
    \sdvop{\graphextrestr{\mathcal{G}}{u}}(\theta)
    =
    0.
  \end{equation*}
  This concludes the proof.
\end{proof}

\subsection{Selected Pseudo-Derivative Operators}
\label{subsec:graph-selected-pseudo-derivative-operators}

\begin{closingdefinition}[Selected pseudo-derivative operator]
  \label{def:selected-pseudo-derivative-operator}
  \todo{}
  Let $\mathcal{P}$ be a class of Euclidean functions,
  let $\mathcal{F}$ be the closure of $\mathcal{F}$,
  let $\indexedcompgraphfamily$ be a $\mathcal{P}$-computation graph family for $\mathcal{F}$,
  let $\selgraphsymb$ be a graph selector over $\indexedcompgraphfamily$,
  and 
  let $\sdv$ be a pseudo-derivative operator that is applicable to $\mathcal{P}$.

  The \emph{graph-selected pseudo-derivative operator induced by $\ptuple{\sdv,\selgraphsymb}$},
  is defined by
  \begin{equation*}
    \sdv[\selgraphsymb] q
    \defeq
    \sdvop[\selgraphsymb[q]]{q}
    \qquad
    \forall q\in\mathcal{Q}.
  \end{equation*}
  The following terminology applies:
  \begin{itemize}
    \item
      $\sdv$ 
      is 
      the \emph{local pseudo-derivative operator} of $\sdv[\selgraphsymb]$;
    \item
      $\selgraphsymb$ 
      is 
      the \emph{graph selector} of $\sdv[\selgraphsymb]$, 
      or simply its \emph{selector};
    \item
      $\selgraphsymb[q]$ 
      is 
      the \emph{selected computation graph of $q$ under $\sdv[\selgraphsymb]$};
    \item
      we abbreviate the term \emph{graph-selected pseudo-derivative operator}
      as 
      \emph{selected pseudo-derivative operator}
      and also as 
      \emph{SPD-operator}.
  \end{itemize}
\end{closingdefinition}

\begin{closingdefinition}[Named partial selected pseudo-derivatives]
  \label{def:named-partial-selected-pseudo-derivatives}
  Let
  $\sdv[\selgraphsymb]$
  be a selected pseudo-derivative operator,
  and let
  \begin{equation*}
    q
    :
    \reals^{d_1}
    \times
    \cdots
    \times
    \reals^{d_k}
    \to
    \reals^{d_\mathrm{out}}
  \end{equation*}
  be a function to which
  $\sdv[\selgraphsymb]$
  is applicable.

  We extend the notation of
  Definition~\ref{def:named-partial-pseudo-derivatives}
  to selected pseudo-derivatives as follows.

  For every
  $i\in\iival{1}{k}$,
  \begin{equation*}
    \bsdvop[\selgraphsymb]{q}{i}
  \end{equation*}
  denotes the block selected pseudo-derivative function corresponding to the
  $i$th input block, and
  \begin{equation*}
    \psdvop[\selgraphsymb]{q}{x_i}
    \big(
      x_1,
      \ldots,
      x_k
    \big)
    \defeq
    \bsdvop[\selgraphsymb]{q}{i}
    \big(
      x_1,
      \ldots,
      x_k
    \big)
  \end{equation*}
  denotes its value at the point
  $\ptuple{x_1,\ldots,x_k}$,
  with the differentiated input block identified by its name.

  The same abbreviation convention applies when an input block is given by the
  value of a function or expression.
  In particular, for an evaluation point
  \begin{equation*}
    \ptuple{
      g(x),
      y
    },
  \end{equation*}
  we write
  \begin{equation*}
    \psdvop[\selgraphsymb]{q}{g}
    \big(
      g(x),
      y
    \big)
    \defeq
    \psdvop[\selgraphsymb]{q}{g(x)}
    \big(
      g(x),
      y
    \big).
  \end{equation*}

  More generally, named subscripts may abbreviate the expression defining the
  corresponding input block whenever the association between the name and the
  block at the evaluation point is clear from context.
\end{closingdefinition}

\subsection{Intensional-Derivative Operators}
\label{sec:selected-intensional-derivatives}

\begin{closingdefinition}[Intensional-derivative operators]
  \label{def:intensional-derivative-operator}
  Let $q$ be a Euclidean PAP function 
  having 
  input arity $d_\mathrm{in}$
  and
  output arity $d_\mathrm{out}$.
  An \emph{intensional-derivative operator} applicable to $q$ is 
  a pseudo-derivative operator $\sdv$ such that $\sdvop{q}$ is an intensional derivative of $q$,
  namely $\sdvop{q} \in \intensionalderivativesop{q}$.
  In this case,
  the function $\sdvop{q}$ is the \emph{intensional-derivative function of $q$ under $\sdv$};
  we may also call $\sdvop{q}$ the \emph{intensional-Jacobian} of $q$ under $\sdv$,
  and 
  the \emph{intensional-gradient} of $q$ under $\sdv$ when $d_\mathrm{out} = 1$.

  Let $\mathcal{Q}$ be a class of Euclidean PAP functions.
  An \emph{intensional-derivative operator} applicable to $\mathcal{Q}$ is a 
  pseudo-derivative operator that is an intensional-derivative operator 
  for every function $q \in \mathcal{Q}$.
\end{closingdefinition}

\subsection{Graph-Induced Intensional-Derivative Operators}
\label{subsec:selected-intensional-derivatives}

\begin{closingdefinition}[Intensional graph-admissibility]
  \label{def:intensionally-admissible-pseudo-derivative}
  A pseudo-derivative operator $\sdv$
  is \emph{intensionally-admissible for a computation graph $G$}
  if it is an intensional-derivative operator for every
  computation-node label $\cglabel{v}$ of $G$.

  A pseudo-derivative operator $\sdv$
  is \emph{intensionally-admissible for a parametric computation graph}
  $\mathcal{G}=\ptuple{G,\cgnode{v}[][\mathrm{par}]}$
  if it is intensionally-admissible for $G$.
\end{closingdefinition}

\begin{proposition}[Correctness of graph-induced pseudo-derivatives]
  \label{prop:correctness-graph-derivative}
  Let $\papsubclass$ be a class of PAP Euclidean functions,
  let $G$ be a $\papsubclass$-computation graph,
  and let
  \begin{equation*}
    f \defeq \cgval{G}
  \end{equation*}
  be the function represented by $G$.

  If $\sdv$ is intensionally-admissible for $G$, then
  $f$ is a PAP function and the graph-induced pseudo-derivative
  induced by $G$ is an intensional derivative of $f$,
  namely, $\sdv G \in \intensionalderivatives f$.
\end{proposition}
\begin{proof}
  Since $\sdv$ is intensionally-admissible for $G$,
  it is an intensional-derivative operator for every computation-node
  label of $G$.
  In particular, $\sdv$ is applicable to every computation-node label,
  and hence it is admissible for $G$.
  Therefore, the graph-induced pseudo-derivative $\sdv[G]$ is well-defined.

  We prove, following a topological ordering of the nodes of $G$, that
  every node $\cgnode{v}$ satisfies the following two properties:
  \begin{enumerate}
    \item
      its value function $\cgval{v}$ is PAP;
    \item
      its graph-induced pseudo-derivative is an intensional derivative
      of its value function:
      \begin{equation}
        \label{eq:graph-induced-intensional-derivative-node}
        \sdvop[G]{\cgnode{v}}
        \in
        \intensionalderivatives \cgval{v}.
      \end{equation}
  \end{enumerate}

  First, let $\cgnode{v}^{\mathrm{in}}_i$ be an input node.
  Its value function is the coordinate projection
  \begin{equation*}
    \cgval{v}^{\mathrm{in}}_i(x)
    =
    \getentry{x}{i}.
  \end{equation*}
  This function is analytic and therefore PAP.
  Furthermore, it admits a one-piece PAP representation whose analytic
  representative is the coordinate projection itself.
  The intensional derivative induced by this representation is its
  ordinary derivative
  \begin{equation*}
    x
    \longmapsto
    e_i^\top.
  \end{equation*}
  By
  Definition~\ref{def:graph-induced-pseudo-derivative},
  \begin{equation*}
    \sdvop[G]{\cgnode{v}^{\mathrm{in}}_i}(x)
    =
    e_i^\top.
  \end{equation*}
  Hence
  \begin{equation*}
    \sdvop[G]{\cgnode{v}^{\mathrm{in}}_i}
    \in
    \intensionalderivatives
    \cgval{v}^{\mathrm{in}}_i.
  \end{equation*}

  Now let $\cgnode{v}$ be a zero-arity computation node.
  Its value function is constant as a function of the input of $G$.
  Hence it is analytic and therefore PAP.
  A one-piece PAP representation of this constant function has
  identically-zero derivative, so
  \begin{equation*}
    x
    \longmapsto
    0
  \end{equation*}
  is an intensional derivative of $\cgval{v}$.
  By
  Definition~\ref{def:graph-induced-pseudo-derivative},
  \begin{equation*}
    \sdvop[G]{\cgnode{v}}(x)
    =
    0.
  \end{equation*}
  Thus
  \begin{equation*}
    \sdvop[G]{\cgnode{v}}
    \in
    \intensionalderivatives
    \cgval{v}.
  \end{equation*}

  Finally, let $\cgnode{v}$ be a positive-arity computation node
  having predecessor node-vector
  \begin{equation*}
    \cgnode{u}
    =
    \ptuple{
      \cgnode{u}_1,
      \ldots,
      \cgnode{u}_k
    }.
  \end{equation*}
  By the induction hypothesis, for every
  $i\in\iival{1}{k}$,
  the value function $\cgval{u}_i$ is PAP and
  \begin{equation*}
    \sdvop[G]{\cgnode{u}_i}
    \in
    \intensionalderivatives
    \cgval{u}_i.
  \end{equation*}

  By
  Lemma~\ref{lem:pap-intensional-tupling},
  the node-vector value function
  \begin{equation*}
    \cgval{u}
    =
    \ptuple{
      \cgval{u}_1,
      \ldots,
      \cgval{u}_k
    }
  \end{equation*}
  is PAP, and the vertically stacked function
  \begin{equation*}
    x
    \longmapsto
    \begin{pmatrix}
      \sdvop[G]{\cgnode{u}_1}(x)
      \\
      \vdots
      \\
      \sdvop[G]{\cgnode{u}_k}(x)
    \end{pmatrix}
  \end{equation*}
  is an intensional derivative of $\cgval{u}$.
  By
  Definition~\ref{def:graph-induced-pseudo-derivative},
  this function is exactly
  $\sdvop[G]{\cgnode{u}}$.
  Therefore,
  \begin{equation}
    \label{eq:graph-induced-intensional-predecessors}
    \sdvop[G]{\cgnode{u}}
    \in
    \intensionalderivatives
    \cgval{u}.
  \end{equation}

  Since $G$ is a $\mathcal{P}_\mathrm{PAP}$-computation graph,
  its computation-node label satisfies
  \begin{equation*}
    \cglabel{v}
    \in
    \mathcal{P}_\mathrm{PAP}.
  \end{equation*}
  Hence $\cglabel{v}$ is PAP.

  Furthermore, since $\sdv$ is intensionally-admissible for $G$,
  it is an intensional-derivative operator for $\cglabel{v}$.
  Thus
  \begin{equation}
    \label{eq:graph-induced-intensional-local}
    \sdvop{\cglabel{v}}
    \in
    \intensionalderivatives
    \cglabel{v}.
  \end{equation}

  By the definition of the value function of a computation node,
  \begin{equation*}
    \cgval{v}
    =
    \cglabel{v}
    \circ
    \cgval{u}.
  \end{equation*}
  Since both $\cgval{u}$ and $\cglabel{v}$ are PAP,
  Proposition~9 of \citet{lee2020correctness}
  implies that $\cgval{v}$ is PAP.

  Furthermore, by
  \eqref{eq:graph-induced-intensional-predecessors},
  \eqref{eq:graph-induced-intensional-local},
  and the chain rule for intensional derivatives given by
  Proposition~10 of \citet{lee2020correctness},
  the function
  \begin{equation*}
    x
    \longmapsto
    \sdvop{\cglabel{v}}
    \big(
      \cgval{u}(x)
    \big)
    \cdot
    \sdvop[G]{\cgnode{u}}(x)
  \end{equation*}
  is an intensional derivative of
  \begin{equation*}
    \cglabel{v}
    \circ
    \cgval{u}
    =
    \cgval{v}.
  \end{equation*}
  By
  Definition~\ref{def:graph-induced-pseudo-derivative},
  this function is precisely
  $\sdvop[G]{\cgnode{v}}$.
  Hence
  \begin{equation*}
    \sdvop[G]{\cgnode{v}}
    \in
    \intensionalderivatives
    \cgval{v}.
  \end{equation*}

  This establishes both induction claims for every node of $G$.

  It remains to consider the output node-vector
  \begin{equation*}
    \cgnode{v}^{\mathrm{out}}
    =
    \ptuple{
      \cgnode{v}^{\mathrm{out}}_1,
      \ldots,
      \cgnode{v}^{\mathrm{out}}_{d_\mathrm{out}}
    }.
  \end{equation*}
  By the result just proved, for every
  $i\in\iival{1}{d_\mathrm{out}}$,
  the value function
  $\cgval{v}^{\mathrm{out}}_i$
  is PAP and
  \begin{equation*}
    \sdvop[G]{\cgnode{v}^{\mathrm{out}}_i}
    \in
    \intensionalderivatives
    \cgval{v}^{\mathrm{out}}_i.
  \end{equation*}
  Applying
  Lemma~\ref{lem:pap-intensional-tupling}
  once more shows that
  \begin{equation*}
    f
    =
    \cgval{G}
    =
    \cgval{\cgnode{v}^{\mathrm{out}}}
  \end{equation*}
  is PAP and that
  \begin{equation*}
    \sdvop[G]{\cgnode{v}^{\mathrm{out}}}
    \in
    \intensionalderivatives f.
  \end{equation*}

  Finally, by
  Definition~\ref{def:graph-induced-pseudo-derivative},
  \begin{equation*}
    \sdv G
    =
    \sdvop[G]{\cgnode{v}^{\mathrm{out}}},
  \end{equation*}
  and therefore
  \begin{equation*}
    \sdv G
    \in
    \intensionalderivatives f.
  \end{equation*}
  Hence $f$ is PAP and the graph-induced pseudo-derivative induced by
  $G$ is an intensional derivative of $f$, as required.
\end{proof}

\begin{corollary}[Correctness from an intensional-derivative operator]
  \label{cor:correctness-graph-intensional-derivative-operator}
  Let $\papsubclass$ be a class of PAP Euclidean functions,
  let $G$ be an $\papsubclass$-computation graph,
  and let
  \begin{equation*}
    f \defeq \cgval{G}
  \end{equation*}
  be the function represented by $G$.

  If $\sdv$ is an intensional-derivative operator for
  $\papsubclass$,
  then $f$ is a PAP function and the graph-induced pseudo-derivative
  induced by $G$ is an intensional derivative of $f$,
  namely $\sdv G \in \intensionalderivatives f$.
\end{corollary}
\begin{proof}
  Since $G$ is an $\papsubclass$-computation graph,
  every computation-node label belongs to
  $\papsubclass$.
  Since $\sdv$ is an intensional-derivative operator for
  $\papsubclass$,
  it is an intensional-derivative operator for every computation-node
  label of $G$.
  Hence $\sdv$ is intensionally-admissible for $G$ by
  Definition~\ref{def:intensionally-admissible-pseudo-derivative}.

  The result follows from
  Proposition~\ref{prop:correctness-graph-derivative}.
\end{proof}

\begin{corollary}[Correctness for parametric computation graphs]
  \label{cor:correctness-parametric-graph-derivative}
  Let $\papsubclass$ be a class of PAP Euclidean functions,
  let $\mathcal{G} = \ptuple{ G, \cgnode{v}[][\mathrm{par}] }$
  be a parametric computation graph whose underlying computation graph
  $G$ is a $\papsubclass$-computation graph, 
  and 
  let $f \defeq \cgval{\mathcal{G}}$ be the parametric function represented by $\mathcal{G}$.

  If $\sdv$ is intensionally-admissible for $\mathcal{G}$, 
  then $f$ is a PAP function and the graph-induced pseudo-derivative
  induced by $\mathcal{G}$ is an intensional derivative of $f$, namely
  $\sdv \mathcal{G} \in \intensionalderivatives f$.
\end{corollary}
\begin{proof}
  Since $\sdv$ is intensionally-admissible for $\mathcal{G}$,
  Definition~\ref{def:intensionally-admissible-pseudo-derivative}
  implies that $\sdv$ is intensionally-admissible for its underlying
  computation graph $G$.

  Since $G$ is a $\papsubclass$-computation graph,
  Proposition~\ref{prop:correctness-graph-derivative}
  therefore implies that $\cgval{G}$ is PAP and
  \begin{equation*}
    \sdv G
    \in
    \intensionalderivatives
    \cgval{G}.
  \end{equation*}

  By the definition of a parametric computation graph,
  $\cgval{\mathcal{G}}$ is the value function $\cgval{G}$ with its
  input coordinates distinguished into external-input and parameter
  coordinates.  Hence
  \begin{equation*}
    f
    =
    \cgval{\mathcal{G}}
    =
    \cgval{G},
  \end{equation*}
  under this identification.

  Furthermore, by
  Definition~\ref{def:graph-induced-pseudo-derivative},
  \begin{equation*}
    \sdv \mathcal{G}
    =
    \sdv G.
  \end{equation*}
  Therefore $f$ is PAP and
  \begin{equation*}
    \sdv \mathcal{G}
    \in
    \intensionalderivatives f,
  \end{equation*}
  as required.
\end{proof}

\subsection{Selected Intensional-Derivative Operators}
\label{subsec:selected-intensional-derivative-operators}

\begin{closingdefinition}[Selected intensional-derivative operator]
  \label{def:selected-intensional-derivative-operator}
  Let
  $\mathcal Q$
  be a class of PAP Euclidean functions,
  and let
  $\sdv[\selgraphsymb]$
  be a selected pseudo-derivative operator applicable to $\mathcal Q$.

  The operator
  $\sdv[\selgraphsymb]$
  is a
  \emph{selected intensional-derivative operator}%
  for $\mathcal Q$
  if it is an intensional-derivative operator for $\mathcal Q$.
\end{closingdefinition}

\begin{proposition}[Intensional-admissibility implies SID]
  \label{prop:intensional-admissibility-implies-sid}
  Let
  $\papsubclass$
  be a class of PAP Euclidean functions,
  let
  $\mathcal Q$
  be a class of Euclidean functions,
  and let
  $\sdv[\selgraphsymb]$
  be an SPD-operator applicable to $\mathcal Q$
  such that
  \begin{equation*}
    \selgraphsymb[q]
  \end{equation*}
  is a $\papsubclass$-computation graph for every
  $q\in\mathcal Q$.

  If
  $\sdv$
  is intensionally-admissible for
  $\im(\selgraphsymb)$,
  then every function in $\mathcal Q$ is PAP and
  $\sdv[\selgraphsymb]$
  is an SID-operator for $\mathcal Q$.
\end{proposition}
\begin{proof}
  Let
  $q\in\mathcal Q$
  be arbitrary, and let
  \begin{equation*}
    G_q
    \defeq
    \selgraphsymb[q].
  \end{equation*}
  By hypothesis,
  $G_q$
  is a $\papsubclass$-computation graph.
  Moreover,
  since
  \begin{equation*}
    G_q
    \in
    \im(\selgraphsymb),
  \end{equation*}
  the hypothesis of intensional admissibility implies that
  $\sdv$
  is intensionally-admissible for
  $G_q$.

  Since $G_q$ represents $q$,
  Proposition~\ref{prop:correctness-graph-derivative}
  implies that $q$ is PAP and that
  \begin{equation*}
    \sdv G_q
    \in
    \intensionalderivatives q.
  \end{equation*}

  By
  Definition~\ref{def:selected-pseudo-derivative-operator},
  \begin{equation*}
    \sdv[\selgraphsymb]q
    =
    \sdv G_q
    =
    \sdv\,\selgraphsymb[q].
  \end{equation*}
  Therefore
  \begin{equation*}
    \sdv[\selgraphsymb]q
    \in
    \intensionalderivatives q.
  \end{equation*}

  Since
  $q\in\mathcal Q$
  was arbitrary,
  every function in $\mathcal Q$ is PAP and
  $\sdv[\selgraphsymb]$
  is an intensional-derivative operator for
  $\mathcal Q$.
  Hence, by
  Definition~\ref{def:selected-intensional-derivative-operator},
  it is an SID-operator for
  $\mathcal Q$.
\end{proof}

\begin{closingremark}[Dependence on the selected computation graph]
  \label{remark:dependence-on-the-selected-computation-graph}
  Fix a local pseudo-derivative operator $\sdv$.
  Different computation graphs representing the same function may induce
  different intensional derivatives of that function.

  An SID-operator
  $\sdv[\selgraphsymb]$
  includes a graph selector
  $\selgraphsymb$
  that fixes one computation graph for each function in its domain.
  Consequently,
  $\sdv[\selgraphsymb]$
  uniquely determines a selected intensional derivative for every
  function to which it is applicable.

  At the same time,
  two SID-operators
  $\sdv[\selgraphsymb_1]$
  and
  $\sdv[\selgraphsymb_2]$
  may have the same local pseudo-derivative operator $\sdv$
  but different graph selectors.
  They may therefore produce different intensional derivatives of the
  same function; namely, it is possible that
  \begin{equation*}
    \sdv[\selgraphsymb_1]f
    \,\neq\,
    \sdv[\selgraphsymb_2]f
  \end{equation*}
  for some function $f$.
\end{closingremark}

\begin{example}[Different selected intensional derivatives for min and max functions]
  \label{example:dependence-on-the-selected-computation-graph}
  We exemplify
  Remark~\ref{remark:dependence-on-the-selected-computation-graph}.

  Consider the pointwise maximum functions
  $\max_n : \reals^n \to \reals$
  from
  Definition~\ref{def:minmax-scalar-reductions},
  \begin{equation*}
    \max_n(x_1,\ldots,x_n)
    \defeq
    \bigoplus_{i=1}^n x_i,
  \end{equation*}
  and let $\sdv$ be the MinMax pseudo-derivative operator from
  Definition~\ref{def:minmax-pseudoderivative-operator}.
  In particular,
  \begin{equation*}
    \sdvop{\max_2}(a,a)
    =
    \left(
      \frac12,
      \frac12
    \right)
    \qquad
    \forall a\in\reals.
  \end{equation*}

  The function $\max_3$ admits, among others, computation graphs
  $G_1$ and $G_2$ corresponding respectively to
  \begin{align*}
    \max_3(x_1,x_2,x_3)
    &=
    \max_2\big(
      \max_2(x_1,x_2),
      x_3
    \big),
    \\
    \max_3(x_1,x_2,x_3)
    &=
    \max_2\big(
      x_1,
      \max_2(x_2,x_3)
    \big).
  \end{align*}
  Both are computation graphs whose computation-node labels are
  instances of the PAP function $\max_2$.

  Let
  $\selgraphsymb_1$
  and
  $\selgraphsymb_2$
  be graph selectors satisfying
  \begin{equation*}
    \selgraphsymb_1[\max_3]
    =
    G_1,
    \qquad
    \selgraphsymb_2[\max_3]
    =
    G_2.
  \end{equation*}
  Since $\sdv$ is an intensional-derivative operator for the local
  primitive $\max_2$ (as shown in Appendix~\ref{app:gradient}), it is intensionally-admissible for
  both $G_1$ and $G_2$.
  Hence the corresponding selected pseudo-derivative operators are
  SID-operators by
  Proposition~\ref{prop:intensional-admissibility-implies-sid}.

  Consider now a point
  \begin{equation*}
    x
    =
    \ptuple{a,a,a},
    \qquad
    a\in\reals.
  \end{equation*}
  At this point, every maximum node in both graphs is evaluated at a tie.
  \begin{itemize}
    \item
      For $G_1$, the inner maximum assigns weight $1/2$ to each of
      $x_1$ and $x_2$, while the outer maximum assigns weight $1/2$
      to the inner maximum and $1/2$ to $x_3$.
      Consequently,
      \begin{equation*}
        \sdv[\selgraphsymb_1]\max_3(a,a,a)
        =
        \left(
          \frac14,
          \frac14,
          \frac12
        \right).
      \end{equation*}

    \item
      For $G_2$, the inner maximum assigns weight $1/2$ to each of
      $x_2$ and $x_3$, while the outer maximum assigns weight $1/2$
      to $x_1$ and $1/2$ to the inner maximum.
      Consequently,
      \begin{equation*}
        \sdv[\selgraphsymb_2]\max_3(a,a,a)
        =
        \left(
          \frac12,
          \frac14,
          \frac14
        \right).
      \end{equation*}
  \end{itemize}

  Therefore,
  \begin{equation*}
    \sdv[\selgraphsymb_1]\max_3(a,a,a)
    \neq
    \sdv[\selgraphsymb_2]\max_3(a,a,a).
  \end{equation*}
  Thus, two SID-operators having the same local pseudo-derivative operator
  $\sdv$ but different graph selectors may select different intensional
  derivatives of the same function, even though their selected computation
  graphs represent the same function $\max_3$.
\end{example}

\subsection{Analytic Extensions of Computation Graphs}
\label{subsec:analytic-extensions-of-computation-graphs}
\label{subsec:smooth-extensions-and-smooth-chain-compatibility}

We introduce a construction for extending computation graphs by an analytic
outer function.  It will be used to represent analytic parameter-dependent
maps, output maps, and losses, and to propagate graph-induced
pseudo-derivatives through them.

\begin{closingdefinition}[Analytic extension of computation graphs]
  \label{def:analytic-extension-selected-graph}
  \label{def:smooth-extension-selected-graph}
  Let
  $Q_1,\ldots,Q_k$,
  for $k\in\posnaturals$,
  be computation graphs satisfying the conditions of
  Definition~\ref{def:computation-graph-common-input-union-nonparametric},
  with common input dimension $d_\mathrm{in}$,
  respective output dimensions $d_1,\ldots,d_k$,
  and represented functions
  \begin{equation*}
    q_i:\reals^{d_\mathrm{in}}\to\reals^{d_i},
    \qquad
    i\in\iival{1}{k}.
  \end{equation*}
  Let
  \begin{equation*}
    g
    =
    \ptuple{g_1,\ldots,g_{d_\mathrm{out}}}
    :
    \reals^{d_1}
    \times\cdots\times
    \reals^{d_k}
    \to
    \reals^{d_\mathrm{out}}
  \end{equation*}
  be analytic.

  Let
  \begin{equation*}
    Q^\sqcup
    \defeq
    Q_1
    \sqcup_{\mathrm{in}}
    \cdots
    \sqcup_{\mathrm{in}}
    Q_k,
    \qquad
    \cgnode{u}
    \defeq
    \cgnode{v}^{\mathrm{out},1}
    \parallel\cdots\parallel
    \cgnode{v}^{\mathrm{out},k},
  \end{equation*}
  where
  $\cgnode{v}^{\mathrm{out},i}$
  is the output node-vector inherited from $Q_i$.

  The
  \emph{analytic extension of $\ptuple{Q_1,\ldots,Q_k}$ by $g$},
  denoted by
  \begin{equation*}
    G
    \defeq
    \ptuple{Q_1,\ldots,Q_k}
    \triangleright
    g,
  \end{equation*}
  is obtained from $Q^\sqcup$ by adjoining fresh computation nodes
  \begin{equation*}
    \cgnode{w}
    \defeq
    \ptuple{
      \cgnode{w}_1,
      \ldots,
      \cgnode{w}_{d_\mathrm{out}}
    },
  \end{equation*}
  with predecessor node-vector $\cgnode{u}$ and labels
  \begin{equation*}
    \cglabel{w_j}
    \defeq
    g_j,
    \qquad
    j\in\iival{1}{d_\mathrm{out}},
  \end{equation*}
  and by taking $\cgnode{w}$ as the output node-vector.
  All nodes, multi-edges, labels, and the input node-vector inherited from
  $Q^\sqcup$ are otherwise unchanged.
\end{closingdefinition}

\begin{proposition}[Correctness of analytic graph extensions]
  \label{prop:correctness-analytic-extension-selected-graph}
  \label{prop:correctness-smooth-extension-selected-graph}
  In the context of
  Definition~\ref{def:analytic-extension-selected-graph},
  the graph
  \begin{equation*}
    G
    =
    \ptuple{Q_1,\ldots,Q_k}
    \triangleright
    g
  \end{equation*}
  represents
  \begin{equation*}
    F
    :
    \reals^{d_\mathrm{in}}
    \to
    \reals^{d_\mathrm{out}},
    \qquad
    F(x)
    \defeq
    g
    \big(
      q_1(x),
      \ldots,
      q_k(x)
    \big).
  \end{equation*}
\end{proposition}

\begin{proof}
  For every
  $x\in\reals^{d_\mathrm{in}}$,
  the predecessor node-vector of the newly introduced output nodes satisfies
  \begin{equation*}
    \cgval{u}(x)
    =
    \ptuple{
      q_1(x),
      \ldots,
      q_k(x)
    },
  \end{equation*}
  because the corresponding components are the output node-vectors of
  $Q_1,\ldots,Q_k$.
  Since
  $\cglabel{w_j}=g_j$
  for every
  $j\in\iival{1}{d_\mathrm{out}}$,
  we obtain
  \begin{equation*}
    \cgval{G}(x)
    =
    g
    \big(
      q_1(x),
      \ldots,
      q_k(x)
    \big)
    =
    F(x).
  \end{equation*}
\end{proof}

\begin{closingfact}[Preservation of computation-graph primitive classes]
  \label{fact:analytic-extension-primitive-class}
  In the context of
  Definition~\ref{def:analytic-extension-selected-graph},
  let $\mathcal P$ be a class of Euclidean functions.
  If every $Q_i$ is a $\mathcal P$-computation graph and
  \begin{equation*}
    g_j\in\mathcal P
    \qquad
    \forall j\in\iival{1}{d_\mathrm{out}},
  \end{equation*}
  then
  \begin{equation*}
    \ptuple{Q_1,\ldots,Q_k}
    \triangleright
    g
  \end{equation*}
  is a $\mathcal P$-computation graph.
\end{closingfact}

\begin{proof}
  The extension retains all computation-node labels of
  $Q_1,\ldots,Q_k$
  and introduces only the labels
  $g_1,\ldots,g_{d_\mathrm{out}}$,
  which belong to $\mathcal P$ by hypothesis.
\end{proof}

\begin{proposition}[Analytic chain rule for graph extensions]
  \label{prop:analytic-graph-extension-chain-rule}
  \label{prop:selected-graph-smooth-extension-chain-rule}
  In the context of
  Definition~\ref{def:analytic-extension-selected-graph},
  let
  \begin{equation*}
    G
    \defeq
    \ptuple{Q_1,\ldots,Q_k}
    \triangleright
    g,
  \end{equation*}
  and let $\sdv$ be a pseudo-derivative operator admissible for $G$.
  If
  \begin{equation*}
    \sdvop{g_j}
    =
    \dvop{g_j}
    \qquad
    \forall j\in\iival{1}{d_\mathrm{out}},
  \end{equation*}
  then, for every
  $x\in\reals^{d_\mathrm{in}}$,
  \begin{align}
    \label{eq:analytic-graph-extension-chain-rule}
    \sdv G(x)
    &=
    \dvop{g}
    \big(
      q_1(x),
      \ldots,
      q_k(x)
    \big)
    \begin{pmatrix}
      \sdv Q_1(x)
      \\
      \vdots
      \\
      \sdv Q_k(x)
    \end{pmatrix}
    \notag
    \\
    &=
    \sum_{i=1}^k
    \pdvop{
      g
      \big(
        q_1(x),
        \ldots,
        q_k(x)
      \big)
    }{q_i}
    \sdv Q_i(x).
  \end{align}
\end{proposition}

\begin{proof}
  Fix
  $x\in\reals^{d_\mathrm{in}}$.
  Since the extension does not modify any node inherited from
  $Q_1,\ldots,Q_k$,
  Definition~\ref{def:graph-induced-pseudo-derivative} gives
  \begin{equation*}
    \sdvop[G]{\cgnode{u}}(x)
    =
    \begin{pmatrix}
      \sdv Q_1(x)
      \\
      \vdots
      \\
      \sdv Q_k(x)
    \end{pmatrix}.
  \end{equation*}
  Moreover,
  \begin{equation*}
    \cgval{u}(x)
    =
    \ptuple{
      q_1(x),
      \ldots,
      q_k(x)
    }.
  \end{equation*}
  Since the new output nodes have labels
  $g_1,\ldots,g_{d_\mathrm{out}}$
  and
  $\sdvop{g_j}=\dvop{g_j}$,
  Definition~\ref{def:graph-induced-pseudo-derivative} yields
  \begin{equation*}
    \sdv G(x)
    =
    \dvop{g}
    \big(
      q_1(x),
      \ldots,
      q_k(x)
    \big)
    \begin{pmatrix}
      \sdv Q_1(x)
      \\
      \vdots
      \\
      \sdv Q_k(x)
    \end{pmatrix}.
  \end{equation*}
  Expanding
  $\dvop{g}$
  by input blocks gives the second equality in
  \eqref{eq:analytic-graph-extension-chain-rule}.
\end{proof}

\clearpage{}%

\clearpage{}%

\section{Properties: Expressivity, BIBS, BIBO, BIBSD, BIBLG, and nnVSD}
\label{sec:properties}
\label{appendix:properties}

\localtableofcontents

This appendix introduces the formal properties used throughout the paper to
characterise the expressive power, stability, and gradient behaviour of recurrent
systems and their parametric variants.
The properties are organised around three increasingly derivative-sensitive levels.
At the level of values, BIBS and BIBO formalise boundedness of state and output
trajectories under bounded inputs
(Definitions~\ref{def:bibs} and~\ref{def:bibo}).
At the level of parameter sensitivity, BIBSD requires derivatives of states with
respect to parameters to remain bounded, while BIBLG formulates the corresponding
property directly for gradients of training objectives.
Finally, nnVSD captures a complementary phenomenon: state-to-state derivatives need
not necessarily vanish with the sequence length
(Definition~\ref{def:nnvsd}).

Several variants are required because the results in the paper quantify over different
objects.
For parametric systems, we distinguish weak properties, where bounds may depend on the
particular bounded input trajectory and initial state, from uniform properties, where a
single bound applies on every bounded input, state, and parameter set; see
Definitions~\ref{def:parametric-wbibs-wbibo} and~\ref{def:parametric-ubibs-ubibo}
for BIBS and BIBO, and
Definitions~\ref{def:wbibsd} and~\ref{def:uniform-bibsd}
for BIBSD.
For derivatives that may fail to exist everywhere, we also consider the
almost-everywhere variants, including
Definitions~\ref{def:ae-wbibsd} and~\ref{def:ae-uniform-bibsd}.
In addition, selected intensional derivatives require explicit control of the
computation graph used to represent a function.
This leads to selected and graph-uniform formulations, in particular
Definitions~\ref{def:selected-uniform-bibsd},
\ref{def:graph-uniform-selected-ubibsd},
\ref{def:sdv-biblg}, and
\ref{def:graph-uniform-selected-biblg}.
These distinctions allow the statements in the main text to separate properties of
the represented recurrent system from properties of a particular differentiable or
piecewise-differentiable computational realisation.

The appendix is organised as follows.
We first define formal-language expressivity and its $\eta$-finite variant in
Definitions~\ref{def:expressivity} and~\ref{def:eta-expressivity}.
We then introduce BIBS and BIBO and their weak and uniform parametric variants in
Definitions~\ref{def:bibs}--\ref{def:parametric-ubibs-ubibo}.
Next, we develop BIBSD, progressing from weak and almost-everywhere formulations to
uniform, selected, and graph-uniform variants; see in particular
Definitions~\ref{def:wbibsd},
\ref{def:ae-wbibsd},
\ref{def:uniform-bibsd},
\ref{def:ae-uniform-bibsd},
and~\ref{def:graph-uniform-selected-ubibsd}.
We subsequently define samples, losses, and objective functions in
Definition~\ref{def:sample-space-loss-function}, and use them to formulate BIBLG,
ae-BIBLG, selected BIBLG, and graph-uniform selected BIBLG in
Definitions~\ref{def:biblg},
\ref{def:ae-biblg},
\ref{def:sdv-biblg}, and
\ref{def:graph-uniform-selected-biblg}.
The connection between the uniform state-derivative properties and loss-gradient
bounds is summarised in
Remark~\ref{rem:uniform-bibsd-loss-gradient-bounds}.
Finally, Definition~\ref{def:nnvsd} introduces nnVSD, which formalises the existence
of bounded-input trajectories along which state-to-state derivatives remain bounded
away from zero uniformly in time.

\subsection{Expressivity}
\label{subsec:expressivity}

Definition~\ref{def:implement} gives the underlying notion of exact
implementation and recognition.  
We refine it by requiring the decoder to
separate the outputs produced by the system from differently decoded outputs
by a fixed positive metric margin.

For this section,
let us fix an arbitrary system 
\begin{equation*}
  S \,=\, \bigtuple{X,\,U,\,f,\,x_0,\,Y,\,h}\;.
\end{equation*}

\begin{closingdefinition}[Decoder margin]
  \label{def:positive-margin-decoder-margin}
  Let 
  $\gamma\in\nonnegreals$
  and 
  let 
  $\hat{Y} \subseteq Y$.
  We say that a decoder $\dec : Y \to \Gamma$%
  \footnote{%
    The notion of \emph{decoder} is introduced by
    Definition~\ref{def:encoder-decoder}.
    Recall that $\Gamma$ is a finite alphabet.%
  }
  \emph{admits a margin $\gamma$ on $\tilde{Y}$}
  if, 
  for every pair of $\hat{y} \in \hat{Y}$ and $y\in Y$,%
  \footnote{Recall that $d_Y$ denotes the metric of the output space $Y$.}
  \begin{equation*}
    d_Y(\hat{y},y)
    <
    \gamma
    \quad\Longrightarrow\quad
    \dec(\hat{y})
    =
    \dec(y)\,.
  \end{equation*}
\end{closingdefinition}

\begin{closingdefinition}[Implementation with a margin]
  \label{def:positive-margin-implementation}
  Given a function $F : \Sigma^+ \to \Gamma$, 
  we define the following notions of implementation of $F$ by system $S$ with a margin.

  \begin{definitionpart}[light]
    Given 
    an encoder $\enc:\Sigma \to U$
    and 
    a decoder $\dec:Y \to \Gamma$,%
    \footnote{%
      The notions of \emph{encoder} and \emph{decoder} are introduced by
      Definition~\ref{def:encoder-decoder}.
      Recall that $\Sigma$ and $\Gamma$ are finite alphabets 
      called the 
      \emph{input alphabet}
      and 
      \emph{ouput alphabet}, 
      respectively. %
    }
    system~$S$ 
    \emph{implements $F$ with margin $\gamma \in \posreals$, encoder $\enc$, and decoder $\dec$},
    if 
    (i)~$S$ implements $F$ with encoder $\enc$ and decoder $\dec$,
    and
    (ii)~$\dec$ admits a $\gamma$-margin on the output set $Y_S$ given by
    \begin{equation*}
      Y_S \defeq \big\{ S(\enc(w)) \,:\, w \in \Sigma^+ \big\}.
    \end{equation*}
  \end{definitionpart}

  \begin{definitionpart}[light]
    System~$S$
    \emph{can-implement $F$ with margin $\gamma \in \posreals$}
    if 
    there exist an encoder $\enc$ and decoder $\dec$
    such that
    $S$ implements $F$ with margin $\gamma$,
    encoder $\enc$, and decoder $\dec$.
  \end{definitionpart}

  \begin{definitionpart}[light]
    System~$S$ \emph{can-implement $F$ with positive margin}
    if $S$ can-implement $F$ with some positive margin $\gamma \in \posreals$.
  \end{definitionpart}
\end{closingdefinition}

\begin{closingdefinition}[Recognition with a margin]
  \label{def:positive-margin-recognition}
  Given a language $L \subseteq \allseqs{\Sigma}$, 
  we define the following notions of recognition of $L$ by system $S$ with a margin.

  \begin{definitionpart}[light]
    Given 
    an encoder $\enc:\Sigma \to U$
    and 
    a decoder $\dec:Y \to \{0,1\}$,%
    system~$S$
    \emph{%
      recognises language~$L$ 
      with margin $\gamma \in \posreals$, 
      encoder $\enc$, 
      and 
      decoder $\dec$%
    }
    if $S$ implements
    the characteristic function 
    $\characteristicFunction{L}$
    with margin $\gamma$, encoder $\enc$, and decoder $\dec$.
  \end{definitionpart}

  \begin{definitionpart}[light]
    System~$S$
    \emph{recognises language~$L$ with margin $\gamma \in \posreals$}
    if there exist
    an encoder $\enc:\Sigma \to U$
    and 
    a decoder $\dec:Y \to \{0,1\}$,%
    such that $S$ recognises language $L$
    with margin $\gamma$, encoder $\enc$, and decoder $\dec$.
  \end{definitionpart}

  \begin{definitionpart}[light]
    System~$S$
    \emph{can-recognise language~$L$ with positive margin}
    if $S$ can-recognise $L$ with some positive margin $\gamma \in \posreals$.
  \end{definitionpart}
\end{closingdefinition}

\subsubsection{Expressivity of a Class of Systems}

\begin{closingdefinition}[Ordinary expressivity]
  \label{def:expressivity}
  We define the notion of \emph{formal-language expressivity} for
  classes of systems,
  parameteric systems,
  and
  classes of parametric systems:
  \begin{itemize}
    \item
      the \emph{formal-language expressivity} of a class $\mathcal{S}$ of systems is the set
      consisting of each language $L$ for which that exists 
      a system $S \in \mathcal{S}$ that can-recognise $L$;
    \item
      the \emph{formal-language expressivity} of a parameteric system $S$ is the set consisting of
      each language $L$ for which there exists 
      an instance $\pinst{S}{\theta}$ of $S$ that can-recognise $L$;
    \item
      the \emph{formal-language expressivity} of a class $\mathcal{S}$ of parameteric systems
      is the set consisting of each language $L$ for which there exists a parametric system $S \in
      \mathcal{S}$ admitting an instance $\pinst{S}{\theta}$ that can-recognise $L$.
  \end{itemize}
  We often say \emph{expressivity} instead of \emph{formal-language expressivity}.
\end{closingdefinition}

\begin{closingdefinition}[$\eta$-expressivity]
  \label{def:eta-expressivity}
  We define the notion of \emph{formal-language $\eta$-expressivity} for
  classes of systems,
  parameteric systems,
  and
  classes of parametric systems:
  \begin{itemize}
    \item
      the \emph{formal-language $\eta$-expressivity} of a class $\mathcal{S}$ of systems
      is the set consisting of each language $L$ for which there exists an $\eta$-finite system
      $S \in \mathcal{S}$ 
      that can-recognise $L$;
    \item
      the \emph{formal-language $\eta$-expressivity} of a parameteric system $S$ is the set
      consisting of each language $L$ for which there exists an $\eta$-finite instance
      $\pinst{S}{\theta}$ of $S$
      that can-recognise $L$;
    \item
      the \emph{formal-language $\eta$-expressivity} of a class $\mathcal{S}$ of parameteric
      systems is the set consisting of each language $L$ for which there exists 
      a parameteric system $S \in \mathcal{S}$ admitting an instance $\pinst{S}{\theta}$ 
      that can-recognise $L$.
  \end{itemize}
  We often say \emph{$\eta$-expressivity} instead of \emph{formal-language $\eta$-expressivity}.
\end{closingdefinition}

\begin{closingdefinition}[Positive-margin expressivity]
  \label{def:positive-margin-expressivity}
  We define the notion of \emph{positive-margin formal-language expressivity} for
  classes of systems,
  parameteric systems,
  and
  classes of parametric systems:
  \begin{itemize}
    \item
      the \emph{positive-margin formal-language expressivity} of a class $\mathcal{S}$ of systems
      is the set consisting of each language $L$ for which there exists a system $S \in \mathcal{S}$ 
      that can-recognise $L$ with positive margin;
    \item
      the \emph{positive-margin formal-language expressivity} of a parameteric system $S$ is the set
      consisting of each language $L$ for wichh there exists an instance $\pinst{S}{\theta}$ of $S$ 
      that can-recognise $L$ with positive margin;
    \item
      the \emph{positive-margin formal-language expressivity} of a class $\mathcal{S}$ of parameteric
      systems is the set consisting of each language $L$ for which there exists a parametric system
      $S \in \mathcal{S}$ admitting an instance $\pinst{S}{\theta}$ 
      that can-recognise $L$ with positive margin.
  \end{itemize}
  We often say
  \emph{positive-margin expressivity}
  instead of
  \emph{positive-margin formal-language expressivity}.
\end{closingdefinition}

\subsection{BIBS and BIBO}

\begin{restatable}{closingdefinition}{defbibs}
  \label{def:bibs}
  Euclidean dynamics $D = \tuple{X,U,f}$
  satisfy \emph{bounded-input bounded-state (BIBS)}
  if,
  for every $r \in \nonnegreals$,
  there exists $B_r \in \nonnegreals$
  such that,
  for every choice 
    of an initial state $\initialstate{x} \in \truncation{X}{r}$
    and
    of an input sequence $(u_t)_{t \in \timedomain} \in \infseqs{\truncation{U}{r}}$,
  the corresponding state sequence
  \begin{equation*}
    (x_t)_{t \in \timedomain} 
    \,\defeq\,
    \big(\repval{D}(\initialstate{x},\seq{u}{1}{t})\big)_{t \in \timedomain}
    \qquad\text{satisfies}\qquad
    \sup_{t \in \timedomain}
    \|x_t\|
    \leq
    B\,.
  \end{equation*}

  A Euclidean system $S$ satisfies \emph{BIBS}
  if its dynamics $\dynop{S}$ satisfy BIBS.
\end{restatable}

\begin{restatable}{closingdefinition}{defbibo}
  \label{def:bibo}
  A Euclidean system
  $S = \tuple{X,U,f,x^\mathrm{init},Y,h}$
  satisfies
  \emph{bounded-input bounded-output (BIBO)}
  if,
  for every $r \in \nonnegreals$,
  there exists $B_r\in\nonnegreals$
  such that,
  for every choice 
    of an initial state $\initialstate{x}\in \truncation{X}{r}$
    and
    of an input sequence $(u_t)_{t\in\timedomain}\in\infseqs{\truncation{U}{r}}$,
  the corresponding output sequence
  \begin{align*}
    \hspace{-4em}
    \begin{aligned}
      (y_t)_{t \in \timedomain}
      & 
      \defeq
      \big(h(x_{t-1},u_t,x_t)\big)_{t \in \timedomain}
      \\[.1em]
      \text{with }\;\,
      (x_t)_{t \in \timedomain} 
      & 
      \defeq
      \big( \repval{\dynop{S}}(\initialstate{x},\seq{u}{1}{t}) \big)_{t \in \timedomain}
    \end{aligned}
    \raisebox{.1em}{\ensuremath{
        \qquad\;\text{ satisfies }\qquad
        \sup_{t\in\timedomain}
        \|y_t\|
        \,\leq\,
        B_r\,.
    }}
  \end{align*}
\end{restatable}

\begin{closingdefinition}[Weak parametric and class-wise BIBS and BIBO]
  \label{def:parametric-wbibs-wbibo}
  \label{def:parametric-bibs-bibo}
  The non-parametric BIBS and BIBO properties are extended to parametric objects
  in a weak, instance-wise sense as follows:
  \begin{itemize}
    \item
      Euclidean parametric dynamics $D$ satisfy
      \emph{weak bounded-input bounded-state (wBIBS)}
      if
      \begin{equation*}
        \pinst{D}{\theta}
        \text{ satisfies BIBS}
        \qquad\qquad
        \forall\theta\in\Theta;
      \end{equation*}

    \item
      a Euclidean parametric system $S$ satisfies
      \emph{weak bounded-input bounded-state (wBIBS)}
      if
      \begin{equation*}
        \pinst{S}{\theta}
        \text{ satisfies BIBS}
        \qquad\qquad
        \forall\theta\in\Theta;
      \end{equation*}

    \item
      a Euclidean parametric system $S$ satisfies
      \emph{weak bounded-input bounded-output (wBIBO)}
      if
      \begin{equation*}
        \pinst{S}{\theta}
        \text{ satisfies BIBO}
        \qquad\qquad
        \forall\theta\in\Theta;
      \end{equation*}

    \item
      a class of Euclidean dynamics satisfies BIBS if every
      member of the class satisfies the corresponding property;

    \item
      a class of Euclidean dynamics or systems satisfies BIBS or BIBO if every
      member of the class satisfies the corresponding property;

    \item
      a class of Euclidean parametric dynamics satisfies
      wBIBS if every member of the class satisfies the corresponding
      property.

    \item
      a class of Euclidean parametric systems satisfies
      wBIBS or wBIBO if every member of the class satisfies the corresponding
      property.
  \end{itemize}
\end{closingdefinition}

\begin{remark}
  The qualifier \emph{weak} refers to the absence of uniformity across parameter
  instances and bounded input trajectories.
  In particular, the BIBS or BIBO bound of an instance may depend on the
  parameter value, the initial state, and the bounded input sequence.
\end{remark}

\begin{closingdefinition}[Uniform parametric and class-wise BIBS and BIBO]
  \label{def:parametric-ubibs-ubibo}
  We define the following BIBS and BIBO notions.

  \begin{definitionpart}[light]
    Euclidean parametric dynamics
    $D = \ctuple{X,U,f}{\Theta}$
    satisfy
    \emph{uniform bounded-input bounded-state (uBIBS)}
    if,
    for 
    every $r\in\nonnegreals$,
    there exists $B_r\in\nonnegreals$ such that,
    for
    every choice 
      of parameter values $\theta \in \truncation{\Theta}{r}$,
      of an initial state $\initialstate{x} \in \truncation{X}{r}$,
      and
      of an input sequence $u \in \infseqs{\truncation{U}{r}}$,
    the corresponding state sequence 
    \begin{equation*}
      (x_t)_{t \in \timedomain} 
      \,\defeq \,
      \big( \repval{\pinst{D}{\theta}}(\initialstate{x},\seq{u}{1}{t}) \big)_{t \in \timedomain}
      \qquad\;\text{satisfies}\qquad\;\,
      \sup_{t\in\timedomain}
      \left\|
      x_t
      \right\|
      \,\leq\,
      B_r\,.
    \end{equation*}
  \end{definitionpart}

  \begin{definitionpart}[light]
    A Euclidean parametric system $S$ 
    satisfies \emph{uBIBS}
    if its dynamics $\dynop{S}$ satisfy uBIBS.
  \end{definitionpart}

  \begin{definitionpart}[light]
    A Euclidean parametric system $S = \ctuple{X,U,f,x^\mathrm{init},Y,h}{\Theta^f,\Theta^h}$ 
    satisfies
    \emph{uniform bounded-input bounded-output (uBIBO)}
    if,
    for every $r\in\nonnegreals$,
    there exists $B_r\in\nonnegreals$ such that,
    for 
    every choice 
      of dynamics-parameter values $\theta_f \in \truncation{\Theta^f}{r}$,
      of output-parameter values $\theta_h \in \truncation{\Theta^h}{r}$,
      of an initial state $\initialstate{x} \in \truncation{X}{r}$,
      and
      of an input sequence $u \in \infseqs{\truncation{U}{r}}$,
    the corresponding output sequence 
    \begin{align*}
      \begin{aligned}
        (y_t)_{t \in \timedomain}
        & 
        \defeq
        \fpinst{h}{x_{t-1},u_t,x_t}{\theta_h}
        \\[.1em]
        \text{with}\;\;
        (x_t)_{t \in \timedomain}
        & 
        \defeq
        \big(\repval{\pinst{\dynop{S}}{\theta_f}}(\initialstate{x},\seq{u}{1}{t}) \big)_{t\in\timedomain}
      \end{aligned}
      \raisebox{.1em}{
        \ensuremath{
          \qquad\text{satisfies}\qquad\;
          \sup_{t\in\timedomain}
          \|y_t\|
          \leq
          B_r\,.
        }
      }
    \end{align*}
  \end{definitionpart}

  \begin{definitionpart}[light]
    A class of Euclidean parametric dynamics or parametric systems satisfies
    uBIBS or uBIBO if every member of the class satisfies the corresponding
    property.
  \end{definitionpart}
\end{closingdefinition}

\begin{remark}[Relationship between weak and uniform BIBS/BIBO]
  Uniform boundedness implies weak boundedness:
  \begin{align*}
    \mathrm{uBIBS}
    &\Longrightarrow
    \mathrm{wBIBS},
    \\
    \mathrm{uBIBO}
    &\Longrightarrow
    \mathrm{wBIBO}.
  \end{align*}
  The converses need not hold, since under wBIBS and wBIBO the bound may vary
  with the parameter instance and the bounded input trajectory.
\end{remark}

\subsection{BIBSD: Bounded-Input Bounded-State-Derivative}

\todo{}

\todo{}

\begin{restatable}[wBIBSD]{closingdefinition}{defubibsd}
  \label{def:wbibsd}
  \label{def:ubibsd}
  Euclidean parametric dynamics
  $D = \ctuple{X,U,f}{\Theta}$
  satisfy
  \emph{weak bounded-input bounded-state-derivative (wBIBSD)}
  if,
  for every constant $r\in\nonnegreals$
  and 
  for every choice 
    of an initial state $\initialstate{x}\in\truncation{X}{r}$
    and
    of an input sequence $u\in\infseqs{\truncation{U}{r}}$,
  there exists a constant
  $B_{r,u,\initialstate{x}}\in\nonnegreals$
  such that, 
  for every $t\in\timedomain$,
  \begin{align*}
    \indexedDynParamToStateFuncfamily{D}[\initialstate{x}][u][t]
    \mkern2mu\textit{ is differentiable on }\mkern2mu
    \truncation{\Theta}{r}
    \qquad\
    \text{and}\
    \qquad
    \sup_{\theta\in\truncation{\Theta}{r}}
    \big\|
      \dvop{
        \indexedDynParamToStateFuncfamily{D}[\initialstate{x}][u][t][\theta]
      }
    \big\|
    \leq
    B_{r,u,\initialstate{x}}\,.
  \end{align*}
\end{restatable}

\begin{restatable}[ae-wBIBSD]{closingdefinition}{defaeubibsd}
  \label{def:ae-wbibsd}
  \label{def:ae-ubibsd}
  Euclidean parametric dynamics
  $D = \ctuple{X,U,f}{\Theta}$
  satisfy
  \emph{almost-everywhere weak bounded-input bounded-state-derivative
  (ae-wBIBSD)}
  if,
  for 
  every constant $r\in\nonnegreals$,
  every choice 
    of an initial state $\initialstate{x}\in\truncation{X}{r}$,
    and
    of an input sequence $u\in\infseqs{\truncation{U}{r}}$,
  there exist
  \begin{align*}
    \textit{a constant }\mkern2mu
    B_r^{\initialstate{x},u}\in\nonnegreals\,,
    \qquad\;\;
    \textit{a full-measure subset }\mkern4mu
    \hat{\Theta}_r^{\initialstate{x},u} \subseteq \truncation{\Theta}{r}\,,
  \end{align*}
  such that,
  for every $t\in\timedomain$,
  \begin{align*}
    \indexedDynParamToStateFuncfamily{D}[\initialstate{x}][u][t]
    \mkern2mu\textit{ is differentiable on }\mkern2mu
    \hat{\Theta}_r^{\initialstate{x},u}
    \qquad \text{and} \qquad
    \sup_{\theta\in\hat{\Theta}_r^{\initialstate{x},u}}
    \big\|
      \dvop{
        \indexedDynParamToStateFuncfamily{D}[\initialstate{x}][u][t][\theta]
      }
    \big\|
    \leq
    B_r^{\initialstate{x},u}\;.
  \end{align*}
\end{restatable}

\begin{remark}
  The adjective \emph{weak} indicates that the derivative bound may depend on the
  bounded input trajectory $u$ and the initial state $\initialstate{x}$.
  The bound is nevertheless uniform over the bounded parameter set
  $\truncation{\Theta}{r}$ and over time.
\end{remark}

\begin{restatable}[uBIBSD]{closingdefinition}{defubibsdstronger}
  \label{def:ubibsd-stronger}
  \label{def:uniform-bibsd}
  Euclidean parametric dynamics
  $D = \ctuple{X,U,f}{\Theta}$
  satisfy
  \emph{uniform bounded-input bounded-state-derivative (uBIBSD)}
  if,
  for every constant $r\in\nonnegreals$,
  there exists a constant $B_r\in\nonnegreals$ such that,
  for every choice 
    of an initial state $\initialstate{x}\in\truncation{X}{r}$ 
    and 
    of an input sequence $u\in\infseqs{\truncation{U}{r}}$,
  we have that,
  for every $t\in\timedomain$,
  \begin{align*}
    \indexedDynParamToStateFuncfamily{D}[\initialstate{x}][u][t]
    \mkern2mu\textit{ is differentiable on }\mkern2mu
    \truncation{\Theta}{r}
    \theta
    \qquad\
    \text{and}\
    \qquad
    \sup_{\theta\in\truncation{\Theta}{r}}
    \big\|
      \dvop{
        \indexedDynParamToStateFuncfamily{D}[\initialstate{x}][u][t][\theta]
      }
    \big\|
    \leq
    B_r\,.
  \end{align*}
\end{restatable}

\begin{restatable}[ae-uBIBSD]{closingdefinition}{defaeubibsduniform}
  \label{def:ae-uniform-bibsd}
  Euclidean parametric dynamics
  $D = \ctuple{X,U,f}{\Theta}$
  satisfy
  \emph{almost-everywhere uniform bounded-input bounded-state-derivative (ae-uBIBSD)}
  if,
  for every constant $r\in\nonnegreals$,
  there exists a constant $B_r\in\nonnegreals$ such that,
  for every choice 
    of an initial state $\initialstate{x}\in\truncation{X}{r}$
    and 
    of an input sequence $u\in\infseqs{\truncation{U}{r}}$,
  there exists
  \begin{equation*}
    \textit{a full-measure subset }\mkern2mu 
    \hat{\Theta}_{r,u,\initialstate{x}}
    \subseteq
    \truncation{\Theta}{r}
  \end{equation*}
  such that, 
  for every $t\in\timedomain$,
  \begin{align*}
    \indexedDynParamToStateFuncfamily{D}[\initialstate{x}][u][t]
    \mkern2mu\textit{ is differentiable on }\mkern2mu
    \hat{\Theta}_{r,u,\initialstate{x}}
    \qquad\
    \text{and}\
    \qquad
    \sup_{\theta\in\hat{\Theta}_{r,u,\initialstate{x}}}
    \big\|
      \dvop{
        \indexedDynParamToStateFuncfamily{D}[\initialstate{x}][u][t][\theta]
      }
    \big\|
    \leq
    B_r\,.
  \end{align*}
\end{restatable}

The notions of selected intensional derivative operator, selected computation
graph, computation-graph family, and graph selector are introduced in
Section~\ref{sec:selected-intensional-derivatives}.

\begin{closingdefinition}[Selected uniform BIBSD]
  \label{def:selected-uniform-bibsd}
  Let $\sdv$ be a selected intensional derivative operator.
  Euclidean parametric dynamics
  $D = \ctuple{X,U,f}{\Theta}$
  satisfy
  \emph{$\sdv$-uniform bounded-input bounded-state-derivative ($\sdv$-uBIBSD)}
  if,
  for every $r\in\nonnegreals$,
  there exists $B_r\in\nonnegreals$ such that,
  for choice of
    an initial state every $\initialstate{x}\in\truncation{X}{r}$
    and 
    an input sequence $u\in\infseqs{\truncation{U}{r}}$,
  we have that, 
  for every $t\in\timedomain$,
  \begin{align*}
    \sdv
    \mkern2mu\textit{ is applicable to }\mkern1mu
    \restr{\indexedDynParamToStateFuncfamily{D}[\initialstate{x}][u][t]}{\truncation{\Theta}{r}}
    \qquad\ \text{and}\ \qquad
    \sup_{\theta\in\truncation{\Theta}{r}}
    \big\|
    \sdvop{\indexedDynParamToStateFuncfamily{D}[\initialstate{x}][u][t][\theta]}
    \big\|
    \leq
    B_r\,.
  \end{align*}
\end{closingdefinition}

\begin{closingdefinition}[State computation-graph family]
  \label{def:state-computation-graph-family-for-dynamics}
  For $D$ Euclidean parametric dynamics,
  a \emph{state computation-graph family} for $D$ 
  is a computation-graph family $\indexedcompgraphfamily$ 
  for the family of parameter-to-state functions $\indexedDynParamToStateFuncfamily{D}$.
\end{closingdefinition}

\begin{closingdefinition}[Graph-uniform selected uBIBSD]
  \label{def:graph-uniform-selected-ubibsd}
  Let $D$ be Euclidean parametric dynamics,
  let $\indexedcompgraphfamily$ be a state computation-graph family for $D$,
  and
  let $\sdv$ be a pseudo-derivative operator that is intensionally-admissible for 
  $\indexedcompgraphfamily$.

  We say that $D$ satisfies
  \emph{%
    $\indexedcompgraphfamily$-uniform selected bounded-input 
    bounded-state-derivative 
    ($\indexedcompgraphfamily$-$\sdv$-uBIBSD)%
  }
  if, 
  for every $r\in\nonnegreals$,
  there exists $B_r\in\nonnegreals$
  such that
  \begin{equation*}
    \qquad\qquad
    \sup_{\theta\in\truncation{\Theta}{r}}
    \big\|
      \sdv G(\theta)
    \big\|
    \leq
    B_r
    \qquad\qquad
    \forall
    u\in\infseqs{\truncation{U}{r}},
    \;\,
    \forall
    \initialstate{x}\in\truncation{X}{r},
    \;\,
    \forall
    t\in\timedomain,
    \;\,
    \forall
    G\in\mBigIndexedcompgraphfamily[\indexedDynParamToStateFuncfamily{D}[\initialstate{x}][u][t]]\,.
  \end{equation*}
\end{closingdefinition}

\todo{}

\begin{remark}[Relationship between weak and uniform BIBSD]
  The uniform properties imply their weak counterparts:
  \begin{align*}
    \mathrm{uBIBSD}
    &\Longrightarrow
    \mathrm{wBIBSD},
    \\
    \mathrm{ae\text{-}uBIBSD}
    &\Longrightarrow
    \mathrm{ae\text{-}wBIBSD}.
  \end{align*}
  The converses need not hold because the constants in wBIBSD and ae-wBIBSD
  may depend on the bounded input trajectory and the initial state.
\end{remark}

\begin{closingdefinition}[System-wise BIBSD properties]
  \label{def:system-wise-bibsd-properties}
  The preceding state-derivative properties are extended to an arbitrary Euclidean
  parametric system $S$ as follows:
  \begin{itemize}
    \item
      system~$S$ 
      satisfies 
      wBIBSD or uBIBSD if its dynamics $\dynop{S}$ satisfy the
      corresponding property and its output function $h$ is differentiable
      jointly in its non-parameter and parameter arguments;
    \item
      system~$S$ 
      satisfies 
      ae-wBIBSD or ae-uBIBSD if its dynamics $\dynop{S}$ satisfy
      the corresponding property and its output function $h$ is
      almost-everywhere differentiable jointly in its non-parameter and
      parameter arguments;
    \item
      given $\sdv$ a selected intensional derivative operator,
      we say that 
      system~$S$ 
      satisfies 
      $\sdv$-uBIBSD if 
      its dynamics $\dynop{S}$ 
      satisfy
      $\sdv$-uBIBSD;
    \item
      given a state computation-graph family $\indexedcompgraphfamily$  
      for the dynamics $\dynop{S}$,
      and 
      a pseudo-derivative operator $\sdv$ that is intensionally-admissible for 
      $\indexedcompgraphfamily$,
      we say that 
      system~$S$ 
      satisfies 
      $\indexedcompgraphfamily$-$\sdv$-uBIBSD if 
      its dynamics $\dynop{S}$ 
      satisfy $\indexedcompgraphfamily$-$\sdv$-uBIBSD.
  \end{itemize}
\end{closingdefinition}

\begin{closingdefinition}[Class-wise BIBSD properties]
  \label{def:class-wise-bibsd-properties}
  A class of Euclidean parametric dynamics or parametric systems 
  satisfies
  wBIBSD, 
  ae-wBIBSD, 
  uBIBSD, 
  ae-uBIBSD, 
  $\sdv$-uBIBSD,
  or 
  $\indexedcompgraphfamily$-$\sdv$-uBIBSD 
  if every member of
  the class satisfies the corresponding property.
\end{closingdefinition}

\begin{remark}[Uniform properties for loss-gradient bounds]
  \label{rem:uniform-bibsd-loss-gradient-bounds}
  The uniform properties are the variants 
  requiring a single bound on every bounded input and parameter set.
  This requirement is in line with the loss-gradient notions 
  introduced below in~\S\ref{subsec:properties-biblg}. 
  In particular, uBIBS holding jointly with
  uBIBSD, ae-uBIBSD, $\sdv$-uBIBSD, or $\indexedcompgraphfamily$-$\sdv$-uBIBSD,
  provides the state and state-parameter derivative uniformity
  required by the corresponding 
  BIBLG, 
  ae-BIBLG, 
  $\sdv$-BIBLG,
  or 
  $\indexedcompgraphfamily$-$\sdv$-BIBLG 
  lifting results.
\end{remark}

\subsection{BIBLG: Bounded-Input Bounded-Loss-Gradient}
\label{subsec:properties-biblg}

\begin{closingdefinition}[Samples, loss and objective functions]
  \label{def:sample-space-loss-function}
  Let $S$ be a Euclidean parametric system,\footnote{%
    We recall that $\Omega_Y$ denotes the Euclidean ambient space of $Y$.%
  }
  \begin{align*}
    S = \ctuple{X,U,f,x^\mathrm{init},Y,h}{\Theta}.
  \end{align*}
  Let $\targetspace{Y} \subseteq \reals^{d_\mathrm{target}}$ be a space that we consider associated
  with the output space $Y$ of $S$, and we call \emph{target space}.
  Then,
  \begin{itemize}
    \item
      its \emph{sample space} is $\samplespacep{S} \defeq U^+\times \targetspace{Y}$, and we define 
      the following restricted variants, for $r\in\nonnegreals$ and $T \in \timedomain$:
      \begin{itemize}
        \item
          the \emph{$T$-horizon sample space} is
          $\truncsamplespacep{S}{r} 
          \defeq 
          \allnonemptyseqs{U}[T] \times \targetspace{Y}$;
        \item
          the \emph{$r$-truncated sample space} is
          $\truncsamplespacep{S}{r} 
          \defeq 
          \truncation{\allnonemptyseqs{U}}{r} \times \truncation{\targetspace{Y}}{r}$;
        \item
          the \emph{$T$-horizon $r$-truncated sample space} is
          $\horizontruncsamplespacep{S}{r}{T} 
          \defeq 
          \truncation{\allnonemptyseqs{U}[T]}{r} \times
          \truncation{\targetspace{Y}}{r}$;
      \end{itemize}
    \item
      the set of \emph{admissible loss functions} $\admissiblelossfunctions[S]$ consists of each
      analytic function $\lossfn: Y\times \targetspace{Y} \to \reals$;
    \item
      the \emph{objective function} 
      induced by 
      an admissible loss function $\ell \in \admissiblelossfunctions[S]$ 
      on a sample $z=(u,y)\in \samplespacep{S}$ 
      is 
      \begin{equation*}
        \Lossfn{S,z}[\lossfn]:
        \Theta
        \to
        \reals,
        \qquad\qquad
        \Lossfn{S,z}[\lossfn](\theta)
        \defeq
        \lossfn
        \big(
          \fpinst{\repval{S}}{u}{\theta},\,
          y
        \big).
      \end{equation*}
  \end{itemize}
  When $S$ is clear from the context,
  we write $\samplespace$ in place of $\samplespacep{S}$,
  we write $\truncation{\samplespace}{r}$ in place of $\truncsamplespacep{S}{r}$,
  and 
  we write $\Lossfn{z}[\lossfn]$ in place of $\Lossfn{S,z}[\lossfn]$;
  when also the loss function $\lossfn$ is clear from the context,
  we write $\Lossfn{z}$ in place of $\Lossfn{S,z}[\lossfn]$;
\end{closingdefinition}

\begin{closingdefinition}[Loss-gradient growth]
  \label{def:Loss-gradient-growth}
  Given 
  a parametric Euclidean system $S$, 
  a radius $r \in \nonnegreals$, 
  and
  an admissible loss function $\lossfn \in \admissiblelossfunctions[S]$,
  the corresponding \emph{loss-gradient growth function} 
  is
  \begin{equation*}
    \LossGrowth{S}{\lossfn}{r}: \timedomain \to \nonnegreals\,,
    \qquad\quad
    \LossGrowth{S}{\lossfn}{r}(T)
    \defeq
    \sup
    \mBig\{
      \,
      \big\|
      \dvop{\Lossfn{S,z}[\lossfn]}(\theta)
      \big\|
      \;:\;
      z \in \horizontruncsamplespacep{S}{r}{T},\;
      \theta \in \truncation{\Theta}{r}
    \mBig\}\,.
  \end{equation*}
\end{closingdefinition}

\begin{convention}[Ambient extensions of admissible loss functions]
  \label{convention:derivative-of-loss-functions}
  \todo{}
  Every admissible loss function $\lossfn : Y \times \targetspace{Y} \to \reals$ admits 
  an ambient extension $\bar \lossfn$ by Definition~\ref{def:analytic-euclidean-map}.
  The extension $\bar \lossfn$ is continuously-differentiable since it is analytic.
  Derivatives of $\lossfn$ are understood as the corresponding derivatives of $\bar\lossfn$
  evaluated on $Y\times \targetspace{Y}$.
\end{convention}

\begin{closingdefinition}[BIBLG]
  \label{def:biblg}
  A Euclidean parametric system $S$, having parameter space $\Theta$ and sample space
  $\samplespacep{S}$,
  satisfies
  \emph{bounded-input bounded-loss-gradient (BIBLG)}
  if,
  for every admissible loss function $\lossfn \in \admissiblelossfunctions[S]$ 
  and every $r\in\nonnegreals$, 
  there exists $B_r\in\nonnegreals$ such that, 
  for every $z \in \truncsamplespacep{S}{r}$,
  \begin{align*}
    \Lossfn{S,z}[\lossfn]
    \mkern2mu\textit{ is differentiable on }\mkern2mu 
    \truncation{\Theta}{r}
    \quad\; \text{ and } \quad\;
    \sup_{\theta \in \truncation{\Theta}{r}}
    \big\|
    \dvop{\Lossfn{S,z}[\lossfn]}(\theta)
    \big\|
    \leq
    B_r\,.
  \end{align*}
\end{closingdefinition}

\begin{closingdefinition}[ae-BIBLG]
  \label{def:ae-biblg}
  A measurable Euclidean parametric system $S$ having parameter space $\Theta$
  satisfies
  \emph{almost-everywhere bounded-input bounded-loss-gradient (ae-BIBLG)}
  if
  for 
  every admissible loss function 
  $\lossfn \in \admissiblelossfunctions[S]$ 
  and 
  every $r\in\nonnegreals$, 
  there exists $B_{\lossfn,r}\in\nonnegreals$ 
  such that, 
  for every $z \in \truncsamplespacep{S}{r}$,
  there exists 
  \begin{align*}
    \textit{a full-measure subset }\mkern3mu
    \truncation{\hat{\Theta}}{r} \subseteq \truncation{\Theta}{r}
  \end{align*}
  such that
  \begin{align*}
    \Lossfn{S,z}[\lossfn]
    \mkern2mu\textit{ is differentiable on }\mkern2mu
    \truncation{\hat{\Theta}}{r}
    \quad\; \text{ and } \quad\;
    \sup_{\theta \in \truncation{\hat{\Theta}}{r}}
    \big\|
      \dvop{\Lossfn{S,z}[\lossfn]}(\theta)
    \big\|
    \leq
    B_{\lossfn,r}\,.
  \end{align*}
\end{closingdefinition}

\begin{closingdefinition}[$\sdv$-BIBLG]
  \label{def:sdv-biblg}
  A Euclidean parametric system $S$ having parameter space $\Theta$ 
  satisfies
  \emph{bounded-input bounded-loss-gradient with respect to $\sdv$},
  abbreviated as $\sdv$-BIBLG,
  if, 
  for 
  every loss function $\lossfn \in \admissiblelossfunctions[S]$
  and 
  every $r\in\nonnegreals$, 
  there exists $B_{\lossfn,r}\in\nonnegreals$ such that,
  for every sample $z \in \truncsamplespacep{S}{r}$,
  \begin{align*}
    \sdv
    \mkern2mu\textit{ is applicable to }\mkern1mu
    \restr{\Lossfn{S,z}[\lossfn]}{\truncation{\Theta}{r}}
    \quad\; \text{ and } \quad\;
    \sup_{\theta \in \truncation{\Theta}{r}}
    \big\|
      \dvop{\Lossfn{S,z}[\lossfn]}(\theta)
    \big\|
    \,\leq\,
    B_{\lossfn,r}\,.
  \end{align*}
\end{closingdefinition}

\begin{closingdefinition}[Objective computation-graph family]
  \label{def:loss-computation-graph-family}
  For $S$ a Euclidean parametric system,
  an \emph{objective computation-graph family} for $S$ 
  is a computation-graph family $\indexedcompgraphfamily$ 
  for the family of loss functions $\Lossfn{S}$.
\end{closingdefinition}

\begin{closingdefinition}[Graph-uniform selected BIBLG]
  \label{def:graph-uniform-selected-biblg}
  Let $S$ be a Euclidean parametric system,
  let $\indexedcompgraphfamily$ be an objective computation-graph family for $S$,
  and
  let $\sdv$ be a pseudo-derivative operator that is intensionally-admissible for 
  $\indexedcompgraphfamily$.

  We say that $S$ satisfies
  \emph{%
    $\indexedcompgraphfamily$-uniform selected bounded-input bounded-loss-gradient
    ($\indexedcompgraphfamily$-$\sdv$-BIBLG)}
  if, 
  for 
  every $\lossfn \admissiblelossfunctions[S]$
  and 
  every $r\in\nonnegreals$,
  there exists $B_{\lossfn,r}\in\nonnegreals$
  such that
  \begin{equation*}
    \qquad\qquad\qquad\qquad\qquad
    \sup_{\theta\in\truncation{\Theta}{r}}
    \big\|
      \sdv G(\theta)
    \big\|
    \leq
    B_{\lossfn,r}
    \qquad\qquad\quad
    \forall
    z\in \truncsamplespacep{S}{r},
    \;\,
    \forall
    t\in\timedomain,
    \;\,
    \forall
    G\in\indexedcompgraphfamily[\Lossfn{S,z}[\lossfn]]\,.
  \end{equation*}
\end{closingdefinition}

\begin{closingdefinition}
  A class $\classParDynamicsSymb$ of Euclidean parametric systems satisfies 
  \emph{BIBLG}, 
  \emph{ae-BIBLG},
  \emph{$\sdv$-BIBLG}, 
  or 
  \emph{$\indexedcompgraphfamily$-$\sdv$-BIBLG}
  if
  every system $S \in \classParDynamicsSymb$ satisfies 
  BIBLG, 
  ae-BIBLG, 
  $\sdv$-BIBLG, 
  and
  $\indexedcompgraphfamily$-$\sdv$-BIBLG, 
  respectively.
\end{closingdefinition}

\subsection{nnVSD: Non-necessarily Vanishing State Derivative}

\begin{restatable}{closingdefinition}{defnnvsd}
  \label{def:nnvsd}
  Euclidean dynamics $D = \tuple{X,U,f}$ satisfy the
  \emph{non-necessarily vanishing state-derivative (nnVSD)}
  property if
  there exist
    a state $\initialstate{x} \in X$,
    a bounded sequence $u \in \binfseqs{U}$,
    and
    a positive constant $B \in \posreals$
  such that,
  for every $t \in \timedomain$,
  \begin{align*}
    \indexedStateToStateFuncfamily{D}[u][t]
    \mkern2mu\textit{ is differentiable at }\mkern1mu 
    \initialstate{x}
    \qquad\
    \text{and}\
    \qquad
    \big\|
      \dvop{
        \indexedStateToStateFuncfamily{D}[u][t][\initialstate{x}]
      } 
    \big\| 
    \geq B\,.
  \end{align*}
  Then,
  \begin{itemize}
    \item
      a Euclidean parametric system $S^\parClassSymb$ satisfies \emph{nnVSD}
      if
      its dynamics $\dynop{S^\parClassSymb}$ satisfy nnVSD;
    \item
      a class $\classParDynamicsSymb$ of Euclidean parametric dynamics satisfies \emph{nnVSD}
      if
      some dynamics $D^\parClassSymb \in \classParDynamicsSymb$ satisfy nnVSD;
    \item
      a class $\classParSystemsSymb$ of Euclidean parametric systems satisfies \emph{nnVSD}
      if
      some system $S^\parClassSymb \in \classParSystemsSymb$ satisfies nnVSD.
  \end{itemize}
\end{restatable}

\clearpage{}%

\clearpage{}%

\section{Architecture-agnostic Results: Expressivity, BIBS, BIBO, BIBSD, BIBLG}
\label{app:properties--aux}

\localtableofcontents*

\subsection{Results for Expressivty}
\label{subsec:aux--expressivity}

\subsubsection{\texorpdfstring{$\eta$}{Eta}-finite expressivity implies positive-margin expressivity}
\label{subsec:eta-finite-implies-positive-margin}

We next relate
$\eta$-expressivity
to positive-margin expressivity.
The key observation is that
$\eta$-finiteness
provides compact state and input spaces,
while a continuous decoder into a finite alphabet separates the compact
sets corresponding to distinct output symbols by a positive metric distance.

\begin{closinglemma}[Positive margin of finite-valued decoders on compact sets]
  \label{lemma:positive-margin-compact-decoder}
  Let
  $Y$
  be a metric space,
  let
  $\Gamma$
  be a finite alphabet endowed with the discrete topology,
  let
  \begin{equation*}
    \dec
    :
    Y
    \to
    \Gamma
  \end{equation*}
  be continuous,
  and let
  $K\subseteq Y$
  be compact.

  Then there exists
  $\gamma\in\posreals$
  such that, for every
  $y\in K$
  and
  $y'\in Y$,
  \begin{equation}
    \label{eq:positive-margin-compact-decoder}
    d_Y(y,y')
    <
    \gamma
    \quad\Longrightarrow\quad
    \dec(y')
    =
    \dec(y).
  \end{equation}
\end{closinglemma}

\begin{proof}
  For every
  $a\in\dec(K)$,
  define
  \begin{equation*}
    K_a
    \defeq
    K
    \cap
    \dec^{-1}(\{a\}).
  \end{equation*}
  Since
  $\Gamma$
  is discrete,
  the singleton
  $\{a\}$
  is both open and closed.
  Since
  $\dec$
  is continuous,
  \begin{equation*}
    \dec^{-1}(\{a\})
  \end{equation*}
  is therefore both open and closed in
  $Y$.
  Consequently,
  $K_a$
  is compact and
  \begin{equation*}
    Y
    \setminus
    \dec^{-1}(\{a\})
  \end{equation*}
  is closed.

  If
  \begin{equation*}
    Y
    \setminus
    \dec^{-1}(\{a\})
    \neq
    \emptyset,
  \end{equation*}
  the compact set
  $K_a$
  and the closed set
  \begin{equation*}
    Y
    \setminus
    \dec^{-1}(\{a\})
  \end{equation*}
  are disjoint.
  Hence their distance
  \begin{equation*}
    \delta_a
    \defeq
    \inf
    \mBig\{
      d_Y(y,y')
      \;\mBigm|\;
      y\in K_a,
      \;
      y'
      \in
      Y
      \setminus
      \dec^{-1}(\{a\})
    \mBig\}
  \end{equation*}
  is strictly positive.
  If instead
  \begin{equation*}
    Y
    \setminus
    \dec^{-1}(\{a\})
    =
    \emptyset,
  \end{equation*}
  define
  \begin{equation*}
    \delta_a
    \defeq
    1.
  \end{equation*}

  Since
  $\dec(K)$
  is a nonempty finite set,
  \begin{equation*}
    \gamma
    \defeq
    \min_{a\in\dec(K)}
    \delta_a
  \end{equation*}
  is strictly positive.

  Let now
  $y\in K$
  and
  $y'\in Y$
  satisfy
  \begin{equation*}
    d_Y(y,y')
    <
    \gamma.
  \end{equation*}
  Let
  \begin{equation*}
    a
    \defeq
    \dec(y).
  \end{equation*}
  Then
  $y\in K_a$.
  If
  $\dec(y')\neq a$,
  then
  \begin{equation*}
    y'
    \in
    Y
    \setminus
    \dec^{-1}(\{a\}),
  \end{equation*}
  and hence
  \begin{equation*}
    d_Y(y,y')
    \geq
    \delta_a
    \geq
    \gamma,
  \end{equation*}
  which is a contradiction.
  Therefore
  \begin{equation*}
    \dec(y')
    =
    \dec(y),
  \end{equation*}
  proving
  Equation~\eqref{eq:positive-margin-compact-decoder}.
\end{proof}

\begin{closingtheorem}[$\eta$-finite recognition implies positive-margin recognition]
  \label{thm:eta-finite-implies-positive-margin}
  Let
  \begin{equation*}
    S
    =
    \tuple{
      X,
      U,
      f,
      x_0,
      Y,
      h
    }
  \end{equation*}
  be an
  $\eta$-finite system,
  let
  $\Sigma$
  be a finite alphabet,
  and suppose that
  $S$
  recognises a language
  $L$
  over
  $\Sigma$
  with continuous encoder
  \begin{equation*}
    \enc
    :
    \Sigma
    \to
    U
  \end{equation*}
  and continuous decoder
  \begin{equation*}
    \dec
    :
    Y
    \to
    \{0,1\}.
  \end{equation*}

  Then
  $S$
  recognises
  $L$
  with positive margin,
  with the same encoder
  $\enc$
  and decoder
  $\dec$.
\end{closingtheorem}

\begin{proof}
  Since
  $S$
  is
  $\eta$-finite,
  its state space
  $X$
  and input space
  $U$
  are finite unions of compact
  $\eta$-components.
  Hence
  $X$
  and
  $U$
  are compact.

  Therefore,
  \begin{equation*}
    X
    \times
    U
    \times
    X
  \end{equation*}
  is compact.
  Since the output function
  \begin{equation*}
    h
    :
    X
    \times
    U
    \times
    X
    \to
    Y
  \end{equation*}
  is continuous,
  its image
  \begin{equation*}
    K
    \defeq
    h
    \big(
      X
      \times
      U
      \times
      X
    \big)
  \end{equation*}
  is compact.

  In particular,
  the reachable output set
  \begin{equation*}
    R
    \defeq
    \big\{
      S(\enc(w))
      \mid
      w\in\Sigma^+
    \big\}
  \end{equation*}
  satisfies
  \begin{equation*}
    R
    \subseteq
    K.
  \end{equation*}

  By
  Lemma~\ref{lemma:positive-margin-compact-decoder},
  there exists
  $\gamma\in\posreals$
  such that, for every
  $y\in K$
  and
  $y'\in Y$,
  \begin{equation*}
    d_Y(y,y')
    <
    \gamma
    \quad\Longrightarrow\quad
    \dec(y')
    =
    \dec(y).
  \end{equation*}
  Since
  $R\subseteq K$,
  the same implication holds for every
  $y\in R$.
  Therefore,
  $\dec$
  admits a
  $\gamma$-margin on the reachable output set
  $R$.

  Since
  $S$
  already recognises
  $L$
  with encoder
  $\enc$
  and decoder
  $\dec$,
  Definition~\ref{def:positive-margin-recognition}
  implies that
  $S$
  recognises
  $L$
  with margin
  $\gamma$.
  Since
  $\gamma>0$,
  it recognises
  $L$
  with positive margin.
\end{proof}

\begin{closingcorollary}[$\eta$-expressivity implies positive-margin expressivity]
  \label{cor:eta-expressivity-implies-positive-margin}
  Let
  $\mathcal S$
  be a class of systems.
  Every language in the
  $\eta$-expressivity of
  $\mathcal S$
  belongs to its positive-margin expressivity.
\end{closingcorollary}

\begin{proof}
  Let
  $L$
  belong to the
  $\eta$-expressivity of
  $\mathcal S$.
  By
  Definition~\ref{def:eta-expressivity},
  there exists an
  $\eta$-finite system
  $S\in\mathcal S$
  that recognises
  $L$.
  By
  Theorem~\ref{thm:eta-finite-implies-positive-margin},
  the same system
  $S$
  recognises
  $L$
  with positive margin.
\end{proof}

\subsection{Results for BIBS and BIBO}

\begin{closingdefinition}[Boundedness-preserving functions]
  \label{def:boundedness-preserving}
  Let $X$ and $Y$ be metric spaces.
  A function
  $g:X\to Y$
  is \emph{boundedness-preserving}
  if
  $g(K)$ is bounded for every bounded set
  $K\subseteq X$.
\end{closingdefinition}

\begin{lemma}
  \label{lemma:continuous-closed-euclidean-boundedness-preserving}
  Let $X$ and $Y$ be metric spaces having Euclidean ambient spaces
  $\Omega_X=\reals^m$
  and
  $\Omega_Y=\reals^n$.
  Suppose that $X$ is closed in $\Omega_X$.
  Then every continuous function
  $g:X\to Y$
  is boundedness-preserving.
\end{lemma}
\begin{proof}
  Let $K\subseteq X$ be bounded.
  Since $X$ carries the metric induced by $\Omega_X$,
  $K$ is also bounded as a subset of $\Omega_X$.
  Since $X$ is closed in $\Omega_X$,
  the closure $\ambientClosure{K}$ of $K$ in $\Omega_X$
  is contained in $X$.
  Furthermore, $\ambientClosure{K}$ is closed and bounded in the Euclidean space
  $\Omega_X$, and hence compact.
  Since $g$ is continuous,
  $g(\ambientClosure{K})$ is compact and therefore bounded.
  Since $g(K)\subseteq g(\ambientClosure{K})$, we conclude that $g(K)$ is bounded.
  Thus $g$ is boundedness-preserving.
\end{proof}

\begin{proposition}[Boundedness preservation of analytic Euclidean maps]
  \label{prop:feedforward-neural-map-boundedness}
  Let $f:X\times\Theta\to Y$ be a parametric analytic Euclidean map.
  For every
  $\theta\in\Theta$,
  the function
  \begin{equation*}
    \pinst{f}{\theta}
    :
    X
    \to
    Y
    \qquad\qquad
    \pinst{f}{\theta}(x) 
    = 
    \fpinst{f}{x}{\theta}
  \end{equation*}
  is boundedness-preserving.
\end{proposition}
\begin{proof}
  Fix $\theta\in\Theta$, and let
  $\ambientExtension{f}$ be an analytic ambient extension of $f$.
  Let $K\subseteq X$ be bounded.
  Since $X$ carries the metric induced by $\Omega_X$,
  $K$ is bounded in $\Omega_X$.
  Hence its closure
  $\ambientClosure{K}$
  in $\Omega_X$ is compact.
  The function $x \mapsto \fpinst{\ambientExtension{f}}{x}{\theta}$ is continuous, and hence
  $\ambientExtension{f}(\ambientClosure{K},\theta)$
  is compact and hence bounded.
  Since $f^\theta(K) \subseteq \ambientExtension{f}(\ambientClosure{K},\theta)$,
  we conclude that $f^\theta(K)$ is bounded.
  Thus $f^\theta$ is boundedness-preserving.
\end{proof}

\begin{corollary}
  \label{cor:restricted-feedforward-neural-map-boundedness}
  Let $f : X \times \Theta \to Y$ be a parametric analytic Euclidean map,
  let $\theta \in \Theta$,
  let $f_\theta \defeq \pinst{f}{\theta}$,
  and
  let $\hat{X} \subseteq X$.
  Then,
  the restriction $\restr{f_\theta}{\hat{X}}$ is boundedness-preserving.
\end{corollary}
\begin{proof}
  Let $K\subseteq \hat{X}$ be bounded. 
  Then $K$ is a bounded subset of $X$.
  By
  Proposition~\ref{prop:feedforward-neural-map-boundedness},
  $\fpinst{f}{\theta}{K}$ is bounded.
  Hence the restriction
  $\restr{\pinst{f}{\theta}}{\hat{X}}$
  is boundedness-preserving.
\end{proof}

\begin{corollary}
  \label{cor:restricted-feedforward-neural-map-boundedness--nonparam}
  Let $f : X \times \to Y$ be an analytic Euclidean map,
  and
  let $\hat{X} \subseteq X$.
  Then,
  the restriction $\restr{f}{\hat{X}}$ is boundedness-preserving.
\end{corollary}
\begin{proof}
  Let us define the parametric Euclidean function
  \begin{align*}
    f^\parClassSymb : X \times \{\theta_0\} \to Y \,,
    \qquad\qquad
    \fpinst{f^\parClassSymb}{x}{\theta_0} \defeq f(x)\,.
  \end{align*}
  We have that $f^\parClassSymb$ is analytic, 
  and hence,
  by Corollary~\ref{cor:restricted-feedforward-neural-map-boundedness},
  we have that
  $\restr{\pinst{f^\parClassSymb}{\theta_0}}{\hat{X}}$ is boundedness-preserving.
  Therefore $f$ is boundedness-preserving since
  $f = \pinst{f^\parClassSymb}{\theta_0}$.
\end{proof}

\begin{lemma}[BIBS implies BIBO]
  \label{lemma:bibs-implies-bibo}
  Let $S$ be a Euclidean system
  whose output function $h$ is boundedness-preserving.
  If $S$ satisfies BIBS, then $S$ satisfies BIBO.
\end{lemma}
\begin{proof}
  Let
  $x_0\in X$
  and
  $u = (u_t)_{t\in\timedomain} \in \binfseqs{U}$
  be arbitrary.

  Let $(x_t)_{t\in\naturals}$ be the state sequence of the dynamics $D \defeq \dynop{S}$ 
  on input $u$ when started in $x_0$; namely,
  \begin{align*}
    x_t = \repval{D}(x_0, \seq{u}{1}{t})
    \qquad \forall t \in \timedomain\,.
  \end{align*}
  Since $S$ satisfies BIBS, there exists
  $B_X\in\nonnegreals$
  such that
  \begin{equation*}
    \sup_{t\in\naturals} \|x_t\|
    \leq
    B_X\,.
  \end{equation*}
  Thus,
  the state sequence $(x_t)_{t\in\naturals}$ is bounded.
  Furthermore, $u$ is bounded by assumption.
  Hence the set
  \begin{equation*}
    K
    \defeq
    \big\{
      \ptuple{x_{t-1},u_t,x_t}
      \,:\,
      t\in\timedomain
    \big\}
  \end{equation*}
  is bounded in
  $X\times U\times X$.

  Let $(y_t)_{t\in\timedomain}$ be the output sequence of the $S$ on input $u$; namely,
  \begin{align*}
    y_t = \repval{S}(\seq{u}{1}{t})
    \qquad \forall t \in \timedomain\,.
  \end{align*}
  Since $h$ is boundedness-preserving, we have that $h(K)$ is bounded.
  Therefore, there exists $B_Y\in\nonnegreals$ such that
  \begin{equation*}
    \sup_{t\in\timedomain} \|y_t\| \leq B_Y\,.
  \end{equation*}

  Thus $S$ satisfies BIBO.
\end{proof}

\begin{corollary}[Analytic output functions]
  \label{cor:bibs-neural-output-bibo}
  Let
  $S$ be a Euclidean parametric system whose output function $h$ is analytic.
  If $S$ satisfies wBIBS, then $S$ satisfies wBIBO.
\end{corollary}
\begin{proof}
  Let
  \begin{equation*}
    S
    =
    \ctuple{
    X,U,f,x^\mathrm{init},Y,h
    }{
    \Theta^f,\Theta^h
    },
  \end{equation*}
  and
  $\ptuple{\theta^f,\theta^h} \in \Theta^f\times\Theta^h$
  be arbitrary.

  Since $S$ satisfies wBIBS,
  we have that $\pinst{S}{\theta^f,\theta^h}$ satisfies BIBS.

  By
  Corollary~\ref{cor:restricted-feedforward-neural-map-boundedness},
  the output function $\pinst{h}{\theta_h}$
  is boundedness-preserving.

  By Lemma~\ref{lemma:bibs-implies-bibo}
  it follows that
  $\pinst{S}{\theta^f,\theta^h}$
  satisfies BIBO,
  Since $\ptuple{\theta^f,\theta^h}$ was arbitrary,
  we conclude that $S$ satisfies wBIBO.
\end{proof}

\begin{lemma}[BIBS under parallel composition]
  \label{lemma:bibs-parallel}
  Let $D = D_1 \parallelop \cdots \parallelop D_n$ be a parallel composition of Euclidean dynamics.
  If every $D_i$ satisfies BIBS, then $D$ satisfies BIBS.
\end{lemma}
\begin{proof}
  Let
  $x_0 = \tuple{x_0^1,\ldots,x_0^n}$
  and
  $u = (u_t)_{t\in\timedomain} \in \binfseqs{U}$
  be arbitrary.

  For every
  $i\in\iival{1}{n}$,
  let $(x_t^i)_{t \in \naturals}$ be 
  the state sequence of $D_i$ on input $u$ when started in $x_0^i$; in particular,
  \begin{equation*}
    x_t = \repval{D_i}(x_0^i, \seq{u}{1}{t})
    \qquad
    \forall t \in \timedomain\,;
  \end{equation*}
  since $D_i$ satisfies BIBS,
  there exists
  $B_i\in\nonnegreals$
  such that
  \begin{equation*}
    \sup_{t\in\timedomain} \|x_t^i\| \leq B_i\,,
  \end{equation*}

  By the definition of parallel composition,
  we have
  that the state sequence $(x_t)_{t \in \naturals}$ of $D$ on input $u$ when started in $x_0$
  is
  \begin{equation*}
    x_t = \tuple{x_t^1,\ldots,x_t^n}
    \qquad \forall t \in \naturals\,,
  \end{equation*}
  and hence
  \begin{align*}
    \|x_t\|
    \,=\,
    \sqrt{
      \sum_{i=1}^n
      \|x_t^i\|^2
    }
    \;\leq\;
    \sqrt{
      \sum_{i=1}^n
      B_i^2
    }
    \qquad\qquad
    \forall t\in\naturals\,.
  \end{align*}

  Therefore $D$ satisfies BIBS.
\end{proof}

\begin{corollary}[Parametric wBIBS under parallel composition]
  \label{cor:parametric-bibs-parallel}
  Let $D^\parClassSymb = D_1^\parClassSymb \parallelop \cdots \parallelop D_n^\parClassSymb$
  be a parallel composition of parametric Euclidean dynamics.
  If every $D_i$ satisfies wBIBS, then $D$ satisfies wBIBS.
\end{corollary}
\begin{proof}
  Let
  $\theta = \tuple{\theta_1,\ldots,\theta_n}$
  be arbitrary.
  By
  Corollary~\ref{cor:instantiation-iterated-compositions},
  \begin{equation*}
    \pinst{D^\parClassSymb}{\theta}
    =
    \pinst{D_1^\parClassSymb}{\theta_1}
    \parallelop
    \cdots
    \parallelop
    \pinst{D_n^\parClassSymb}{\theta_n}.
  \end{equation*}
  For every $i \in \iival{1}{n}$,
  since $D_i^\parClassSymb$ satisfies wBIBS,
  its instance $\pinst{D_i^\parClassSymb}{\theta_i}$ satisfies BIBS.
  By Lemma~\ref{lemma:bibs-parallel}, 
  it follows that $\pinst{D^\parClassSymb}{\theta}$ satisfies BIBS.
  Since $\theta$ was arbitrary, 
  we conclude that $D^\parClassSymb$ satisfies wBIBS.
\end{proof}

\begin{closinglemma}[BIBS of recurrent cascade dynamics]
  \label{lemma:recurrent-cascade-bibs}
  Let $\classDynamicsSymb$ be a class of Euclidean dynamics satisfying BIBS,
  let $\classfuncsymb$ be a class of boundedness-preserving functions,
  and 
  let $D$ be recurrent cascade dynamics of type $\ptuple{\classDynamicsSymb,\classfuncsymb}$.
  Then,
  dynamics $D$ satisfy BIBS.
\end{closinglemma}
\begin{proof}
  Let $X$ be the state space of $D$,
  and
  let $U$ be the state space of $D$.

  We have that $D$ admits a recurrent cascade presentation 
  \begin{align*}
    \presentationSty{C}
    \,=\, 
    \repinputprecomposition{g_1}{\presentationSty{L}_1} 
    \repcascade_{g_2} \presentationSty{L}_2
    \cdots 
    \repcascade_{g_\ell} \presentationSty{L}_\ell
  \end{align*}
  where
  every $g_i$ is an analytic Euclidean map,
  and
  every $\presentationSty{L}_i$ is a recurrent layer dynamics presentation 
  \begin{equation*}
    \presentationSty{L}_i
    \,=\,
    D_{i,1}
    \replayer
    \cdots
    \replayer
    D_{i,n_i}
  \end{equation*}
  with every unit $D_{i,n_i}$ belonging to $\classDynamicsSymb$.

  Since $\classDynamicsSymb$ satisfies BIBS,
  for every $i \in \iival{1}{\ell}$ and every $j \in \iival{1}{n_i}$,
  we have that $D_{i,j}$ satisfies BIBS.
  Hence,
  for every $i \in \iival{1}{\ell}$,
  since
  \begin{align*}
    D_i 
    \defeq 
    \presentationVal{\presentationSty{L}_i}
    \,=\,
    D_{i,1}
    \parallel
    \cdots
    \parallel
    D_{i,n_i}\,,
  \end{align*}
  by Lemma~\ref{lemma:bibs-parallel},
  we have that $D_i$ satisfies BIBS.

  Let $x_0 = \tuple{x_0^1,\ldots,x_0^\ell} \in X$ 
  and 
  $u = (u_t)_{t\in\timedomain} \in \binfseqs{U}$ be arbitrary.

  For every $i\in\iival{1}{\ell}$
  let $u^i \defeq (u^i_t)_{t \in \timedomain}$ 
  be the sequence of inputs received by the layer $D_i$ when $D$
  is executed on input $u$ starting from state $x_0$,
  and 
  let $x^i \defeq (x^i_t)_{t \in \timedomain}$ be the state sequence of $D_i$ on input $u^i$ when
  started in state $x_0^i$.

  We proceed by induction on $i$ from $1$ to $\ell$, 
  showing that 
  the sequences $u^i$ and $x^i$ are bounded; namely,
  we show there exist
  $B_1,\ldots,B_i\in\nonnegreals$
  such that
  \begin{equation*}
    \sup_{t\in\timedomain}
    \max\big\{ \|u_t^k\|,\, \|x_t^k\| \big\}
    \,\leq\,
    B_k
    \qquad\qquad
    \forall k\in\iival{1}{i}\,.
  \end{equation*}

  In the base case we have $i=1$.
  We have
  \begin{equation*}
    u^1_t \,=\, g_1(u_t)
    \qquad
    \forall t\in\timedomain\,,
  \end{equation*}
  and hence
  the sequence $u^1$ is bounded, since the input sequence $u$ is bounded and 
  the function $g_1$ is analytic and hence boundedness-preserving
  by Corollary~\ref{cor:restricted-feedforward-neural-map-boundedness--nonparam}.
  Since $D_1$ satisfies BIBS,
  it follows that its state sequence $x^1$ is bounded.
  Thus, there exists 
  $B_1 \in\nonnegreals$
  such that
  \begin{equation*}
    \sup_{t\in\timedomain}
    \max\big\{ \|u_t^1\|,\, \|x_t^1\| \big\}
    \,\leq\,
    B_1 \,.
  \end{equation*}

  In the inductive case we have $i \geq 2$,
  and we assume by induction
  that there exist
  $B_1,\ldots,B_{i-1}\in\nonnegreals$
  such that
  \begin{equation*}
    \sup_{t\in\timedomain}
    \max\big\{ \|u_t^k\|,\, \|x_t^k\| \big\}
    \,\leq\,
    B_k
    \qquad\qquad
    \forall k\in\iival{1}{i-1}\,.
  \end{equation*}

  By the definition of cascade composition,
  \begin{equation*}
    u^i_t
    \,=\,
    g_i\big(x_{t-1}^{i-1},\,u^{i-1}_{i,t},\,x_t^{i-1}\big)
    \qquad\qquad
    \forall t\in\timedomain\,,
  \end{equation*}
  and hence the sequence $u^i$ is bounded,
  since the sequences $x^{i-1}$ and $u^{i-1}$ are bounded by induction and 
  the function $g_i$ is analytic and hence boundedness-preserving
  by Corollary~\ref{cor:restricted-feedforward-neural-map-boundedness--nonparam};
  since $D_i$ satisfies BIBS, it follows that its state sequence $x^i$ is bounded.
  Thus, there exists 
  $B_i \in\nonnegreals$
  such that
  \begin{equation*}
    \sup_{t\in\timedomain}
    \max\big\{ \|u_t^i\|,\, \|x_t^i\| \big\}
    \,\leq\,
    B_i \,,
  \end{equation*}
  and hence
  we have
  $B_1,\ldots,B_i\in\nonnegreals$
  such that
  \begin{equation*}
    \sup_{t\in\timedomain}
    \max\big\{ \|u_t^k\|,\, \|x_t^k\| \big\}
    \,\leq\,
    B_k
    \qquad\qquad
    \forall k\in\iival{1}{i}\,.
  \end{equation*}

  We conclude that the state sequence $(x_t)_{t \in \timedomain}$ of $D$ on input $u$ when started
  in $x_0$ 
  is bounded since its components are components of the sequences 
  $(x^i_t)_{t \in \timedomain}$ for $i \in \iival{1}{\ell}$.

  Since $x_0$ and $u$ were arbitrary,
  we conclude that $D$ satisfies BIBS.
\end{proof}

\begin{closingtheorem}[BIBS and BIBO of Recurrent Neural Cascades]
  \label{theorem:rnc-bibs-bibo}
  Let $\classDynamicsSymb$ be a class of Euclidean dynamics satisfying BIBS.

  Then:
  \begin{itemize}
    \item
      every $\classDynamicsSymb$-RNC dynamics satisfies BIBS;
    \item
      every $\classDynamicsSymb$-RNC satisfies BIBS and BIBO.
  \end{itemize}

  Consequently, the class of $\classDynamicsSymb$-RNC dynamics satisfies BIBS,
  and 
  the class of $\classDynamicsSymb$-RNCs satisfies BIBS and BIBO.
\end{closingtheorem}
\begin{proof}
  Let $D$ be arbitrary $\classDynamicsSymb$-RNC dynamics.
  By definition of RNC, 
  we have that $D$ are recurrent cascade dynamics of type
  $\ptuple{\classDynamicsSymb,\analyticfuncclass}$.
  Thus, they admit a recurrent cascade dynamics presentation
  where 
  \begin{itemize}
    \item
      every unit of belongs to $\classDynamicsSymb$ and therefore satisfies BIBS;
    \item
      every connecting function is an analytic Euclidean map.
  \end{itemize}

  Every analytic Euclidean map is boundedness-preserving by
  Corollary~\ref{cor:restricted-feedforward-neural-map-boundedness--nonparam},
  and hence,
  by Lemma~\ref{lemma:recurrent-cascade-bibs},
  we conclude that
  implies that $D$ satisfies BIBS.

  Now let $S$ be an arbitrary $\classDynamicsSymb$-RNC.
  By definition, its dynamics $\dynop{S}$ are $\classDynamicsSymb$-RNC dynamics,
  and
  its output function $h$ is analytic.

  By the first part of the proof,
  $\dynop{S}$ satisfies BIBS,
  and hence $S$ satisfies BIBS.
  Furthermore, its output function $h$ is boundedness-preserving by 
  Corollary~\ref{cor:restricted-feedforward-neural-map-boundedness--nonparam}.
  Therefore,
  by
  Lemma~\ref{lemma:bibs-implies-bibo},
  we conclude that $S$ satisfies BIBO.

  Since $D$ and $S$ were arbitrary, the corresponding classes satisfy
  the claimed properties.
\end{proof}

\subsubsection{Results for uBIBS and uBIBO}

The BIBS closure results for parallel and recurrent-cascade composition admit
uniform parametric analogues.  

\begin{lemma}[uBIBS under parallel composition]
  \label{lem:parametric-ubibs-parallel}
  Let $D^\parClassSymb$ be parametric Euclidean dynamics given by a parallel composition
  $D^\parClassSymb = D^\parClassSymb_1 \parallelop \cdots \parallelop D^\parClassSymb_n$.
  If every $D^\parClassSymb_i$ satisfies uBIBS, then $D^\parClassSymb$ satisfies uBIBS.
\end{lemma}
\begin{proof}
  Let $D^\parClassSymb = \ctuple{X,U,f}{\Theta}$.
  Fix $r\in\nonnegreals$ arbitrary.
  Let
  \begin{equation*}
    \theta
    =
    \tuple{\theta_1,\ldots,\theta_n}
    \in
    \truncation{\Theta}{r}\,,
    \qquad
    x_0
    =
    \tuple{x_0^1,\ldots,x_0^n}
    \in
    \truncation{X}{r}\,,
    \qquad
    u\in\infseqs{\truncation{U}{r}}
  \end{equation*}
  be arbitrary.
  We have
  \begin{equation*}
    \|\theta_i\|\leq r
    \qquad\text{and}\qquad
    \|x_0^i\|\leq r
    \qquad\qquad
    \forall i\in\iival{1}{n}.
  \end{equation*}

  For every $i \in \iival{1}{n}$,
  since $D_i$ satisfies uBIBS, 
  there exists $B_{i,r}\in\nonnegreals$ such that
  \begin{equation*}
    \sup_{t\in\timedomain}
    \|x_t^i\|
    \leq
    B_{i,r}\,.
  \end{equation*}
  The complete state is
  \begin{equation*}
    x_t
    =
    \tuple{x_t^1,\ldots,x_t^n},
  \end{equation*}
  and hence
  \begin{equation*}
    \|x_t\|
    \leq
    \sqrt{
      \sum_{i=1}^n B_{i,r}^2
    }
    \qquad
    \forall t\in\timedomain.
  \end{equation*}
  Since $r$, $\theta$, $x_0$, and $u$ were arbitrary,
  we conclude that $D$ satisfies uBIBS.
\end{proof}

\begin{theorem}[uBIBS of parametric RNC dynamics]
  \label{thm:rnc-ubibs-closure}
  Let $\classParDynamicsSymb$ be a class of parametric Euclidean dynamics satisfying uBIBS.  
  Then every parametric RNC dynamics in $\rncdynparop(\classParDynamicsSymb)$ satisfies uBIBS.  
  Consequently, the class $\rncdynparop(\classParDynamicsSymb)$ satisfies uBIBS.
\end{theorem}
\begin{proof}
  Let $D \in \rncdynparop(\classParDynamicsSymb)$ be arbitrary.
  We have that $D$ is given by a cascade composition
  \begin{equation*}
    D
    \,=\,
    \inputprecomposition{g_1}{D_1}
    \ltimes_{g_2}
    D_2
    \cdots
    \ltimes_{g_{L}}
    D_L\,.
  \end{equation*}
  Every $D_\ell$ is given by a finite parallel composition of dynamics belonging to
  $\classParDynamicsSymb$, and every function $g_\ell$ is a parametric analytic Euclidean map.
  Thus,
  by Lemma~\ref{lem:parametric-ubibs-parallel}, every $D_\ell$ satisfies uBIBS.

  Fix $r\in\nonnegreals$ arbitrary, 
  and let
  \begin{equation*}
    \theta\in\truncation{\Theta}{r},
    \qquad
    x_0\in\truncation{X}{r},
    \qquad
    u\in\infseqs{\truncation{U}{r}}
  \end{equation*}
  be arbitrary.  

  For every $\ell \in \iival{1}{L}$,
  let $u^\ell \defeq (u_t^\ell)_{t \in \timedomain}$ be the input sequence received by $D_\ell$ when
  $D$ is executed on input $u$ and started in $x_0$,
  and
  and let $(x_t^\ell)_{t\in \timedomain}$ be the state sequence of $D_\ell$ 
  on input $u^\ell$ when started in $x^\ell_0$.

  We proceed by induction on $\ell$ from $1$ to $L$,
  showing there exist
  $B_1, \ldots, B_\ell \in \nonnegreals$ such that
  \begin{align*}
    \sup_{t \in \timedomain} \max\{ u^\ell_t, x^\ell_t \}
    \,\leq\, 
    B_i
    \qquad\quad 
    \forall i \in \iival{1}{\ell}\,.
  \end{align*}

  In the base case we have $\ell = 1$.
  We have
  that the initial state $x^1_0$ and the parameter values of $D_1$ from the 
  joint parameter values $\theta$ are bounded by $r$;
  furthermore,
  \begin{equation*}
    u_t^1=g_1(u_t) \qquad\quad \forall t \in \timedomain\,,
  \end{equation*}
  and hence $u^1$ is bounded, since the input sequence $u$ is bounded by $r$ 
  and $g_1$ is analytic and hence boundedness-preserving 
  by Corollary~\ref{cor:restricted-feedforward-neural-map-boundedness--nonparam}.
  Since $D_1$ satisfies uBIBS, 
  we conclude there exists
  $B_1\in\nonnegreals$ such that
  \begin{equation*}
    \sup_{t \in \timedomain} \max\{ u^1_t, x^1_t \}
    \,\leq\, 
    B_1\,.
  \end{equation*}

  In the inductive case we have $\ell \geq 2$,
  and, by induction,
  there exist
  $B_1, \ldots, B_{\ell-1} \in \nonnegreals$ such that
  \begin{align*}
    \sup_{t \in \timedomain} \max\{ u^i_t, x^i_t \}
    \,\leq\, 
    B_i
    \qquad\quad 
    \forall i \in \iival{1}{\ell-1}\,.
  \end{align*}

  The initial state $x^\ell_0$ and the parameter values of $D_\ell$ from the joint
  parameter values $\theta$ are bounded by $r$;
  furthermore, 
  \begin{equation*}
    u^\ell_t
    \,=\,
    \bigptuple{x_{t-1}^\ell,\,u_t^\ell,\,x_t^\ell}
    \qquad\qquad \forall t \in \timedomain\,,
  \end{equation*}
  and hence $u^\ell$ is bounded, since 
  $x^{\ell-1},u^{\ell-1}$ are bounded by $B_{\ell-1}$ by the inductive hypothesis,
  and $g_\ell$ is analytic and hence boundedness-preserving 
  by Corollary~\ref{cor:restricted-feedforward-neural-map-boundedness--nonparam}.
  Since $D_\ell$ satisfies uBIBS, 
  we conclude there exists
  $B_\ell\in\nonnegreals$ such that
  \begin{align*}
    \sup_{t \in \timedomain} \max\{ u^\ell_t, x^\ell_t \}
    \,\leq\, 
    B_\ell\,.
  \end{align*}
  Thus, by the inductive hypothesis,
  there exist
  $B_1, \ldots, B_{\ell} \in \nonnegreals$ such that
  \begin{align*}
    \sup_{t \in \timedomain} \max\{ u^i_t, x^i_t \}
    \,\leq\, 
    B_i
    \qquad\quad 
    \forall i \in \iival{1}{\ell}\,.
  \end{align*}

  Since the cascade decomposition of $D$ given by dynamics $D_i$ and functions $g_i$ was arbitrary, 
  and 
  since $r$, $\theta$, $x_0$, and $u$ were arbitrary,
  we conclude that $D$ satisfies uBIBS.  
  In turn, since $D$ was arbitrary, 
  we conclude that the corresponding class 
  $\rncdynparop(\classParDynamicsSymb)$ 
  satisfies uBIBS.
\end{proof}

\subsection{Results for BIBSD and BIBLG}

\subsubsection{Auxiliary results}

\begin{lemma}[Graph-uniform selected-derivative bounds under analytic extensions]
  \label{lem:graph-uniform-selected-feedforward-extension-bounds}
  Let $h:X\times\Phi\to Y$
  be an analytic parametric Euclidean map. 

  Let $\Lambda$ be a Euclidean parameter space.

  Let $A \subseteq \posnaturals$ be an arbitrary index set.
  For every $a \in A$,
  let 
  $p_a: \Lambda\to X$ 
  and 
  $q_a : \Lambda\to\Phi$ 
  be functions,
  and
  let
  $\customIndexedcompgraphfamily{\mathcal{P}_a}$
  and
  $\customIndexedcompgraphfamily{\mathcal{Q}_a}$ 
  be indexed computation-graph--families 
  for $p_a$ and $q_a$, respectively.

  For every $a \in A$,
  we define the function
  \begin{equation*}
    h_a(\lambda)
    \,\defeq\,
    \bigfpinst{h}{p_a(\lambda)}{\,q_a(\lambda)},
  \end{equation*}
  and 
  the the computation-graph--family
  \begin{equation*}
    \mathcal{H}_a
    \,\defeq \,
    \mBig\{\,
      \ptuple{P_a,Q_a}
      \analyticextension
      h
      \mkern4mu\mBigm|\mkern3mu
      P_a\in\customIndexedcompgraphfamily{\mathcal{P}}[a],
      \;
      Q_a\in \customIndexedcompgraphfamily{\mathcal{Q}}[a]
    \mBig\}\,.
  \end{equation*}

  Let $\sdv$ be a pseudoderivative operator
  such that,
  for every $a \in A$,
  it is admissible for the graph families $\mathcal{P}_a$,$\mathcal{Q}_a$, and $\mathcal{H}_a$.

  Let $r\in\nonnegreals$,
  and
  suppose there exist
  $B^q_r, \hat{B}^q_r, B^p_r, \hat{B}^p_r \in \nonnegreals$
  such that, 
  for every $a\in A$,
  \begin{align*}
    \sup_{\lambda\in\truncation{\Lambda}{r}}
    \|q_a(\lambda)\|
    &\leq
    B^q_r\,,
    &
    \sup_{Q_a\in\mathcal{Q}_a}
    \sup_{\lambda\in\truncation{\Lambda}{r}}
    \|\sdv Q_a(\lambda)\|
    &\leq
    \hat{B}^q_r\,,
    \\
    \sup_{\lambda\in\truncation{\Lambda}{r}}
    \|p_a(\lambda)\|
    &\leq
    B^p_r\,,
    &
    \sup_{P_a\in\mathcal{P}_a}
    \sup_{\lambda\in\truncation{\Lambda}{r}}
    \|\sdv P_a(\lambda)\|
    &\leq
    \hat{B}^p_r\,.
  \end{align*}

  Then, 
  there exist
  $C_r,\hat{C}_r\in\nonnegreals$
  such that, for every $a \in A$,
  \begin{enumerate}
    \item
      every graph $H_a\in\mathcal{H}_a$ represents $h_a$; 
    \item
      the two following inequalities hold:
      \begin{align*}
        \sup_{\lambda\in\truncation{\Lambda}{r}}
        \|h_a(\lambda)\|
        &\leq
        C_r
        \,,
        &
        \sup_{H_a\in\mathcal{H}_a}
        \sup_{\lambda\in\truncation{\Lambda}{r}}
        \|\sdv H_a(\lambda)\|
        &\leq
        \hat{C}_r 
        \,.
      \end{align*}
  \end{enumerate}
\end{lemma}
\begin{proof}
  \todo{}
  Let $\rho_r \defeq \max ( B^p_r, B^q_r)$.
  By Proposition~\ref{prop:feedforward-neural-map-boundedness},
  there exist
  $M_r,M^p_r,M^q_r\in\nonnegreals$
  such that, whenever
    $x\in\truncation{X}{\rho_r}$
    and
    $\phi\in\truncation{\Phi}{\rho_r}$,
  we have
  \begin{align*}
    \|\bigfpinst{h}{x}{\phi}\|
    \leq
    M_r\,,
    \qquad\;
    \left\|
      \pdvop{\bigfpinst{h}{x}{\phi}}{x}
    \right\|
    \leq
    M^p_r\,,
    \qquad\;
    \left\|
      \pdvop{\bigfpinst{h}{x}{\phi}}{\phi}
    \right\|
    \leq
    M^q_r\,.
  \end{align*}

  Let $a\in A$ be arbitrary.
  For every
  $\lambda\in\truncation{\Lambda}{r}$,
  the assumed value bounds give
  \begin{equation*}
    p_a(\lambda)
    \in
    \truncation{X}{\rho_r}\,,
    \qquad\;
    q_a(\lambda)
    \in
    \truncation{\Phi}{\rho_r}\,.
  \end{equation*}
  Consequently,
  \begin{equation*}
    \|h_a(\lambda)\|
    \,=\,
    \left\|
      \bigfpinst{h}{
        p_a(\lambda)
      }{
        q_a(\lambda)
      }
    \right\|
    \,\leq\,
    M_r\,.
  \end{equation*}
  Hence
  \begin{equation*}
    \sup_{\lambda\in\truncation{\Lambda}{r}}
    \|h_a(\lambda)\|
    \,\leq\,
    M_r\,.
  \end{equation*}

  Now let
  \begin{equation*}
    H_a
    \defeq
    \ptuple{P_a,Q_a}
    \analyticextension
    h
    \in
    \mathcal H_a
  \end{equation*}
  be arbitrary.
  By
  Proposition~\ref{prop:correctness-analytic-extension-selected-graph},
  the computation graph
  $H_a$
  represents the function
  \begin{equation*}
    \lambda
    \mapsto
    \bigfpinst{h}{
      p_a(\lambda)
    }{
      q_a(\lambda)
    }
    \,=\,
    h_a(\lambda)\,.
  \end{equation*}
  This proves the first claim.

  By
  Proposition~\ref{prop:analytic-graph-extension-chain-rule},
  for every
  $\lambda\in\truncation{\Lambda}{r}$,
  \begin{align*}
    \sdv H_a(\lambda)
    \;=\;
    \mBig(
      \pdvop{
        \bigfpinst{h}{
          p_a(\lambda)
        }{
          q_a(\lambda)
        }
      }{x}
      \cdot
      \sdv P_a(\lambda)
    \mBig)
    +
    \mBig(
      \pdvop{
        \bigfpinst{h}{
          p_a(\lambda)
        }{
          q_a(\lambda)
        }
      }{\phi}
      \cdot
      \sdv Q_a(\lambda)
    \mBig)\,.
  \end{align*}
  Therefore,
  \begin{align*}
    \|\sdv H_a(\lambda)\|
    &\,\leq\,
    M^p_r \cdot
    \|\sdv P_a(\lambda)\|
    +
    M^q_r \cdot
    \|\sdv Q_a(\lambda)\|
    \\[.2em]
    &\,\leq\,
    M^p_r \cdot \hat B^p_r
    +
    M^q_r \cdot \hat B^q_r\,.
  \end{align*}
  Since
  $P_a\in\mathcal P_a$
  and
  $Q_a\in\mathcal Q_a$
  were arbitrary, we obtain
  \begin{equation*}
    \sup_{H_a\in\mathcal H_a}
    \sup_{\lambda\in\truncation{\Lambda}{r}}
    \|\sdv H_a(\lambda)\|
    \,\leq\,
    M^p_r \cdot \hat B^p_r
    +
    M^q_r \cdot \hat B^q_r\,.
  \end{equation*}

  Since
  $a\in A$
  was arbitrary and all the constants above are independent of
  $a$,
  the claim follows by taking
  \begin{equation*}
    C_r
    \,\defeq\,
    M_r,
    \qquad
    \hat C_r
    \,\defeq\,
    M^p_r \cdot \hat B^p_r
    +
    M^q_r \cdot \hat B^q_r\,.
  \end{equation*}
\end{proof}

\begin{corollary}[Coordinate-projection case]
  \label{cor:selected-feedforward-coordinate-projection-bounds}
  Assume the setting of
  Lemma~\ref{lem:graph-uniform-selected-feedforward-extension-bounds}.

  Suppose $q:\Lambda\to\Phi$ is a linear map satisfying
  \begin{equation}
    \label{eq:coordinate-projection-nonexpansive}
    \text{(I)~~~}
    \|q(\lambda)\|
    \leq
    \|\lambda\|
    \quad
    \forall\,\lambda\in\Lambda\,,
    \qquad\qquad
    \text{(II)~~~}
    \|q(\lambda)\|
    q_a = q 
    \quad 
    \forall\, a\in A\,,
  \end{equation}
  Suppose $Q$ is a computation graph representing $q$ and satisfying
  \begin{equation}
    \label{eq:coordinate-projection-selected-derivative}
    \text{(III)~~~}
    \sdv Q(\lambda)
    =
    \dvop{q}(\lambda)
    \quad
    \forall\,\lambda\in\Lambda\,,
    \qquad\qquad
    \text{(IV)~~~}
    \customIndexedcompgraphfamily{\mathcal{Q}}[a]
    =
    \{Q\}
    \quad
    \forall\,a\in A\,.
  \end{equation}

  Let
  $r\in\nonnegreals$,
  and
  suppose that there exist
  $B^p_r,\hat B^p_r\in\nonnegreals$
  such that, for every
  $a\in A$,
  \begin{align*}
    \sup_{\lambda\in\truncation{\Lambda}{r}}
    \|p_a(\lambda)\|
    &\leq
    B^p_r\,,
    &
    \sup_{P_a\in\customIndexedcompgraphfamily{\mathcal{P}}[a]}
    \sup_{\lambda\in\truncation{\Lambda}{r}}
    \|\sdv P_a(\lambda)\|
    &\leq
    \hat B^p_r\,.
  \end{align*}

  Then, 
  there exist
  $C_r,\hat C_r\in\nonnegreals$
  such that, 
  for every $a\in A$,
  \begin{align*}
    \sup_{\lambda\in\truncation{\Lambda}{r}}
    \|h_a(\lambda)\|
    &\leq
    C_r\,,
    &
    \sup_{H_a\in\mathcal H_a}
    \sup_{\lambda\in\truncation{\Lambda}{r}}
    \|\sdv H_a(\lambda)\|
    &\leq
    \hat C_r\,.
  \end{align*}
\end{corollary}
\begin{proof}
  \todo{}
  By
  condition~(I) in Equation~\eqref{eq:coordinate-projection-nonexpansive},
  \begin{equation*}
    \sup_{\lambda\in\truncation{\Lambda}{r}}
    \|q_a(\lambda)\|
    =
    \sup_{\lambda\in\truncation{\Lambda}{r}}
    \|q(\lambda)\|
    \leq
    r
    \qquad
    \forall\,a\in A.
  \end{equation*}

  Furthermore, since
  $q$
  is linear,
  its derivative is constant and satisfies
  \begin{equation*}
    \|\dvop{q}(\lambda)\|
    =
    \|q\|_{\mathrm{op}}
    \leq
    1
    \qquad
    \forall\,\lambda\in\Lambda,
  \end{equation*}
  where the final inequality follows from
  condition~(I) in Equation~\eqref{eq:coordinate-projection-nonexpansive}.
  Hence, by
  the condition~(III) in Equation~\eqref{eq:coordinate-projection-selected-derivative},
  \begin{equation*}
    \sup_{Q_a\in\customIndexedcompgraphfamily{\mathcal{Q}}[a]}
    \sup_{\lambda\in\truncation{\Lambda}{r}}
    \|\sdv Q_a(\lambda)\|
    \leq
    1
    \qquad
    \forall\,a\in A.
  \end{equation*}

  Thus, the hypotheses of
  Lemma~\ref{lem:graph-uniform-selected-feedforward-extension-bounds}
  hold with
  \begin{equation*}
    B^q_r
    \defeq
    r,
    \qquad
    \hat B^q_r
    \defeq
    1.
  \end{equation*}
  The claim follows.
\end{proof}

\begin{closingdefinition}[Output-context parameter function]
  \label{def:output-context-parameter-function}
  Let $D = \ctuple{X,U,f}{\Theta}$ be Euclidean parametric dynamics,
  let $x_0\in X$, and let $u=u_1u_2\cdots u_t\in U^+$.

  The \emph{output-context parameter function} for $\ptuple{x_0,u}$ 
  is the function
  $\paroutfunc_u : \Theta \to (X\times U\times X)$
  given by
  \begin{align*}
    \qquad\quad
    \paroutfunc_u(\theta)
    \defeq
    \ptuple{
      x_{t-1},
      u_t,
      x_t
    }
    \qquad\text{for}\qquad
    x_{t-1} \defeq \fpinst{\repval{D}}{x_0,\seq{u}{1}{t-1}}{\theta},
    \;\;
    x_t \defeq \fpinst{\repval{D}}{x_0,u}{\theta}.
  \end{align*}
\end{closingdefinition}

\begin{lemma}[Uniform state bounds imply uniform output-context bounds]
  \label{lem:uniform-output-context-state-bounds}
  Let
  $D=\ctuple{X,U,f}{\Theta}$
  be Euclidean parametric dynamics satisfying uBIBS,
  let
  $x_0\in X$,
  and let
  $\paroutfuncp{u}$
  be the output-context parameter function of
  Definition~\ref{def:output-context-parameter-function}.

  Then, 
  for every 
  $r\in\nonnegreals$,
  there exists 
  $B_r\in\nonnegreals$
  such that
  \begin{equation*}
    \quad
    \sup_{\theta\in\truncation{\Theta}{r}}
    \big\|
      \paroutfuncp{u}(\theta)
    \big\|
    \leq
    B_r
    \qquad\qquad \forall u\in\truncation{U}{r}^+\,.
  \end{equation*}
\end{lemma}
\begin{proof}
  Fix
  $r\in\nonnegreals$,
  and 
  let $\rho_r \defeq \max ( r, \|x_0\|)$.
  Let $u = u_1\cdots u_t \in \truncation{U}{r}^+$ be arbitrary, 
  and 
  extend it to $\bar u \in \infseqs{\truncation{U}{\rho_r}}$ 
  by repeating $u_t$ after time $t$.

  By uBIBS, applied at radius
  $\rho_r$,
  there exists
  $M_{\rho_r}\in\nonnegreals$,
  independent of
  $u$,
  such that, 
  for every 
  $\theta \in \truncation{\Theta}{r} \subseteq \truncation{\Theta}{\rho_r}$,
  every state on input $\bar u$ has norm at most $M_{\rho_r}$.

  Let
  \begin{equation*}
    \bar M_r
    \defeq
    \max
    \big\{
      M_{\rho_r},
      \|x_0\|
    \big\}.
  \end{equation*}
  Both state components of
  $\paroutfuncp{u}(\theta)$
  therefore have norm at most $\bar M_r$,
  and additionally $\|u_t\| \leq r$.
  Hence
  \begin{equation*}
    \sup_{\theta\in\truncation{\Theta}{r}}
    \big\|
      \paroutfuncp{u}(\theta)
    \big\|
    \,\leq\,
    \sqrt{
      2 \cdot \bar M_r^2 + r^2
    }\,.
  \end{equation*}
  Since the right-hand side is independent of
  $u$,
  the claim follows.
\end{proof}

\subsubsection{BIBSD: Graph-Uniform selected-derivative bounds for state gradients}
\label{subsec:graph-families-selected-derivatives}

Recall the following notions:
PAP functions and their intensional derivatives
computation graph~(in~\S\ref{subsec:pap-intensional-preliminaries});
computation graphs and selected intensional
derivatives~(in~\S\ref{app:selected-intensional-derivative-operators});
the relevant BIBSD properties relative to selected derivative operators and 
and the following BIBSD variants:
ae-uBIBSD~(Definition~\ref{def:ae-uniform-bibsd}),
selected-uniform--uBIBSD~(Definition~\ref{def:selected-uniform-bibsd}),
and
graph-uniform--uBIBSD~(Definition~\ref{def:graph-uniform-selected-ubibsd}).

\begin{proposition}[Graph-uniform bounds imply selected-uniform--uBIBSD]
  \label{prop:graph-uniform-implies-selector-ubibsd}
  Let us consider parametric Euclidean dynamics $D = \ctuple{X,U,f}{\Theta}$
  and
  their parameter-to-state function family $\indexedDynParamToStateFuncfamily{D}$.
  Let 
  $\indexedcompgraphfamily$ be a computation-graph family 
  for $\indexedDynParamToStateFuncfamily{D}$,
  and
  let
  $\selgraphsymb$
  be a graph selector for $\indexedDynParamToStateFuncfamily{D}$ over $\indexedcompgraphfamily$,
  and
  let $\sdv$ be a pseudo-derivative operator such that 
  $D$ satisfies $\indexedcompgraphfamily$-$\sdv$-uBIBSD.
  Then,
  \begin{enumerate}
    \item
      the selected intensional derivative operator $\sdv[\selgraphsymb]$
      is applicable to the family $\indexedDynParamToStateFuncfamily{D}\,$;
    \item
      the dynamics $D$ satisfy $\sdv[\selgraphsymb]\mkern-2mu$-uBIBSD.
  \end{enumerate}
\end{proposition}
\begin{proof}
  Since
  $D$
  satisfies
  $\indexedcompgraphfamily$-$\sdv$-uBIBSD,
  Definition~\ref{def:graph-uniform-selected-ubibsd}
  implies, in particular, that
  $\sdv$
  is intensionally-admissible for every computation graph in
  $\indexedcompgraphfamily$.

  Let $q \in \indexedDynParamToStateFuncfamily{D}$ be arbitrary.
  Since
  $\selgraphsymb$
  is a graph selector for
  $\indexedDynParamToStateFuncfamily{D}$
  over
  $\indexedcompgraphfamily$,
  we have
  $\selgraphsymb[q] \in \indexedcompgraphfamily[q]$
  by Definition~\ref{def:graph-selector},
  and hence
  $\sdv$ is intensionally-admissible for $\selgraphsymb[q]$.
  Since $\selgraphsymb[q]$ represents $q$,
  Proposition~\ref{prop:correctness-graph-derivative}
  implies that $q$ is PAP, and also that
  $\sdv\,\selgraphsymb[q] \in \intensionalderivatives q$.%
  \footnote{%
    Recall that the function $\sdv\,\selgraphsymb[q] \in \intensionalderivatives q$ 
    means that $\sdv\,\selgraphsymb[q]$ is an intensional derivative function of $q$.
    See~\S\ref{subsec:pap-intensional-preliminaries}.%
  }
  By
  Definition~\ref{def:selected-pseudo-derivative-operator},
  we have $\sdv[\selgraphsymb]q = \sdv\,\selgraphsymb[q]$,
  and hence
  $\sdv[\selgraphsymb]q \in \intensionalderivatives q$.
  Since
  $q\in\indexedDynParamToStateFuncfamily{D}$
  was arbitrary,
  $\sdv[\selgraphsymb]$
  is applicable to
  $\indexedDynParamToStateFuncfamily{D}$
  and it is a selected intensional derivative operator for this family.

  It remains to establish the uniform bound.
  Let
  $r\in\nonnegreals$
  be arbitrary.
  Since
  $D$
  satisfies
  $\indexedcompgraphfamily$-$\sdv$-uBIBSD,
  there exists
  $B_r\in\nonnegreals$
  such that
  \begin{equation*}
    \qquad
    \sup_{\theta\in\truncation{\Theta}{r}}
    \big\|
      \sdv G(\theta)
    \big\|
    \leq
    B_r
    \qquad\quad
    \forall
    u\in\infseqs{\truncation{U}{r}},
    \;\;
    \forall
    x_0\in\truncation{X}{r},
    \;\;
    \forall
    t\in\timedomain,
    \;\;
    \forall
    G
    \in
    \bigIndexedcompgraphfamily[
      \indexedDynParamToStateFuncfamily{D}[u][x_0][t]
    ]\,.
  \end{equation*}
  In particular, since $\selgraphsymb$ is a graph selector,
  \begin{equation*}
    \bigSelgraphsymb[
      \mkern2mu
      \indexedDynParamToStateFuncfamily{D}[u][x_0][t]
      \mkern2mu
    ]
    \in
    \bigIndexedcompgraphfamily[
      \indexedDynParamToStateFuncfamily{D}[u][x_0][t]
    ]\,.
  \end{equation*}
  Hence
  \begin{equation*}
    \sup_{\theta\in\truncation{\Theta}{r}}
    \left\|
      \sdv
      \bigSelgraphsymb[
        \mkern2mu
        \indexedDynParamToStateFuncfamily{D}[u][x_0][t]
        \mkern2mu
      ](\theta)
    \right\|
    \leq
    B_r\,.
  \end{equation*}
  By
  Definition~\ref{def:selected-pseudo-derivative-operator},
  the above is equivalent to
  \begin{equation*}
    \sup_{\theta\in\truncation{\Theta}{r}}
    \left\|
      \sdvop[\selgraphsymb]{
        \indexedDynParamToStateFuncfamily{D}[u][x_0][t]
      }(\theta)
    \right\|
    \leq
    B_r\,.
  \end{equation*}
  Since
  $r\in\nonnegreals$
  was arbitrary,
  Definition~\ref{def:selected-uniform-bibsd}
  implies that
  $D$
  satisfies
  $\sdv[\selgraphsymb]\mkern-2mu$-uBIBSD.
\end{proof}

\begin{proposition}[Graph-uniform state-derivative bounds imply ae-uBIBSD under a.e~correctness]
  \label{prop:graph-uniform-implies-ae-ubibsd}
  Let us consider 
  measurable-Euclidean parametric dynamics $D = \ctuple{X,U,f}{\Theta}$ 
  and
  their parameter-to-state function family
  $\indexedDynParamToStateFuncfamily{D}$.
  Let
  $\indexedcompgraphfamily$
  be an indexed computation-graph family
  for
  $\indexedDynParamToStateFuncfamily{D}$,
  and 
  let $\sdv$ be a pseudo-derivative operator 
  such that 
  $D$ satisfies
  $\indexedcompgraphfamily$-$\sdv$--uBIBSD.

  Suppose that,
  for 
  every $q \in \indexedDynParamToStateFuncfamily{D}$,
  every $G \in \indexedcompgraphfamily[q]$,
  and
  every $r \in \nonnegreals$,
  there exists 
  a full-measure subset 
  $\hat{\Theta}_r^{q,G} \subseteq \truncation{\Theta}{r}$ 
  such that, 
  for all parameter values $\theta\in \hat{\Theta}_r^{q,G}$,
  \begin{equation*}
    q \ \text{ is differentiable at } \ \theta
    \qquad\text{ and }\qquad
    \dvop{q}(\theta) = \sdv G(\theta) \,.
  \end{equation*}
  Then $D$ satisfies ae-uBIBSD.
\end{proposition}
\begin{proof}
  \todo{}
  Let
  $r\in\nonnegreals$
  be arbitrary.
  Since
  $D$
  satisfies
  $\indexedcompgraphfamily$-$\sdv$-uBIBSD,
  there exists
  $B_r\in\nonnegreals$
  such that
  \begin{equation}
    \label{eq:proof:graph-uniform-implies-ae-ubibsd-bound}
    \sup_{\theta\in\truncation{\Theta}{r}}
    \big\|
      \sdv G(\theta)
    \big\|
    \leq
    B_r
    \qquad\qquad
    \forall\,
    u\in\infseqs{\truncation{U}{r}},
    \;
    \forall\,
    x_0\in\truncation{X}{r},
    \;
    \forall\,
    t\in\timedomain,
    \;
    \forall\,
    G
    \in
    \bigIndexedcompgraphfamily[
      \indexedDynParamToStateFuncfamily{D}[x_0][u][t]
    ]\,.
  \end{equation}
  Now, 
  let
  $u\in\infseqs{\truncation{U}{r}}$
  and
  $x_0\in\truncation{X}{r}$
  be arbitrary,
  and 
  let
  \begin{equation*}
    q_t
    \defeq
    \indexedDynParamToStateFuncfamily{D}[x_0][u][t]
    \,\in \,
    \indexedDynParamToStateFuncfamily{D}
    \qquad\quad
    \forall\, t\in\timedomain\,.
  \end{equation*}
  Since
  $\indexedcompgraphfamily$
  is an indexed computation-graph family for
  $\indexedDynParamToStateFuncfamily{D}$,
  for every $t\in\timedomain$,
  there exists a computation graph
  \begin{equation*}
    G_t
    \in
    \indexedcompgraphfamily[q_t].
  \end{equation*}
  By the assumed a.e.~correctness,
  for every
  $t\in\timedomain$,
  there exists a full-measure subset
  \begin{equation*}
    \hat{\Theta}_r^{q_t,G_t}
    \subseteq
    \truncation{\Theta}{r}
  \end{equation*}
  such that, 
  for every $\theta\in \hat{\Theta}_r^{q_t,G_t}$,
  \begin{equation*}
    q_t
    \text{ is differentiable at }
    \theta
    \qquad\text{and}\qquad
    \dvop{q_t}(\theta)
    =
    \sdv G_t(\theta)\,.
  \end{equation*}

  Let us define
  \begin{equation*}
    \hat{\Theta}_{r,u,x_0}
    \defeq\,
    \bigcap_{t\in\timedomain}
    \hat{\Theta}_r^{q_t,G_t}\,.
  \end{equation*}
  Since
  $\timedomain$
  is countable,
  we have that
  $\hat{\Theta}_{r,u,x_0}$
  is a full-measure subset of
  $\truncation{\Theta}{r}$.

  Let
  $\theta\in\hat{\Theta}_{r,u,x_0}$
  and
  $t\in\timedomain$
  be arbitrary.
  Then
  $\theta\in \hat{\Theta}_r^{q_t,G_t}$,
  and hence
  $\indexedDynParamToStateFuncfamily{D}[x_0][u][t] = q_t$
  is differentiable at
  $\theta$.
  Furthermore,
  by
  Equation~\eqref{eq:proof:graph-uniform-implies-ae-ubibsd-bound},
  \begin{align*}
    \big\|
    \dvop{\indexedDynParamToStateFuncfamily{D}[x_0][u][t][\theta]}
    \big\|
    \,=\,
    \big\|
      \dvop{q_t}(\theta)
    \big\|
    \,=\,
    \big\|
      \sdv G_t(\theta)
    \big\|
    \,\leq\,
    B_r\,.
  \end{align*}

  Thus,
  for the constant
  $B_r$,
  every
  $u\in\infseqs{\truncation{U}{r}}$
  and
  $x_0\in\truncation{X}{r}$
  admit a full-measure subset
  $\hat{\Theta}_{r,u,x_0}$
  on which every parameter-to-state function
  $\indexedDynParamToStateFuncfamily{D}[x_0][u][t]$
  is differentiable and has derivative norm bounded by
  $B_r$,
  uniformly over
  $t\in\timedomain$.

  Since
  $r\in\nonnegreals$
  was arbitrary,
  Definition~\ref{def:ae-uniform-bibsd}
  implies that
  $D$
  satisfies ae-uBIBSD.
\end{proof}

\subsubsection{BIBLG: Graph-uniform selected-derivative bounds for loss gradients}

\begin{proposition}[Graph-uniform bounds imply selector-wise $\sdv$-BIBLG]
  \label{prop:graph-uniform-implies-selector-biblg}
  Let $S$ be a Euclidean parametric system,
  let $\indexedcompgraphfamily$ be an objective computation-graph family for $S$,
  let $\selgraphsymb$ be a graph-selector for $\Lossfn{S}$ over $\indexedcompgraphfamily$,
  and
  let $\sdv$ be a pseudo-derivative operator such that $S$
  satisfies
  $\indexedcompgraphfamily$-$\sdv$-BIBLG.
  Then,
  \begin{enumerate}
    \item
      the selected intensional derivative operator~$\sdv[\selgraphsymb]$
      is applicable to the family of objective functions~$\Lossfn{S}$;
    \item
      system~$S$ satisfies $\sdv[\selgraphsymb]\mkern-2mu$-BIBLG.
  \end{enumerate}
\end{proposition}
\begin{proof}
  \todo{}
  Since
  $S$
  satisfies
  $\indexedcompgraphfamily$-$\sdv$-BIBLG,
  Definition~\ref{def:graph-uniform-selected-biblg}
  implies, in particular, that
  $\sdv$
  is intensionally-admissible for every computation graph in
  $\indexedcompgraphfamily$.

  Let
  $q\in\Lossfn{S}$
  be arbitrary.
  Then, by definition of
  $\Lossfn{S}$,
  there exist
  an admissible loss function
  $\lossfn\in\admissiblelossfunctions[S]$
  and
  a sample
  $z\in\samplespacep{S}$
  such that
  \begin{equation*}
    q
    =
    \Lossfn{z}[\lossfn]\,.
  \end{equation*}
  By the assumption on
  $\selgraphsymb$,
  we have
  \begin{equation*}
    \selgraphsymb[q]
    =
    \selgraphsymb[
      \mkern2mu
      \Lossfn{z}[\lossfn]
      \mkern2mu
    ]
    \in
    \indexedcompgraphfamily[\Lossfn{S,z}[\lossfn]].
  \end{equation*}
  Hence $\sdv$ is intensionally-admissible for $\selgraphsymb[q]$.

  Since $\selgraphsymb[q]$ represents $q$,
  Proposition~\ref{prop:correctness-graph-derivative}
  implies that $q$ is PAP, 
  and also that
  \begin{equation*}
    \sdv\big(\mkern2mu\selgraphsymb[q]\!\big)
    \in
    \mkern1mu
    \intensionalderivatives q\,.
  \end{equation*}
  By
  Definition~\ref{def:selected-pseudo-derivative-operator},
  we have
  \begin{equation*}
    \sdv[\selgraphsymb]q
    \,=\,
    \sdv\big(\mkern2mu\selgraphsymb[q]\!\big),
  \end{equation*}
  and hence
  \begin{equation*}
    \sdv[\selgraphsymb]q
    \in
    \intensionalderivatives q\,.
  \end{equation*}
  Since
  $q\in \Lossfn{S}$
  was arbitrary,
  $\sdv[\selgraphsymb]$
  is applicable to
  $\Lossfn{S}$
  and it is a selected intensional derivative operator for this family.

  \par\medskip\noindent
  It remains to establish the uniform bound.
  Let 
  $\lossfn \in \admissiblelossfunctions[S]$ 
  and 
  $r\in\nonnegreals$
  be arbitrary.

  Since
  $S$
  satisfies
  $\indexedcompgraphfamily$-$\sdv$-BIBLG,
  there exists
  $B_{\lossfn,r}\in\nonnegreals$
  such that
  \begin{equation*}
    \qquad
    \sup_{\theta\in\truncation{\Theta}{r}}
    \big\|
      \sdv G(\theta)
    \big\|
    \leq
    B_{\lossfn,r}
    \qquad\quad
    \forall\,
    z\in\truncsamplespacep{S}{r},
    \;\;
    \forall\,
    G\in \bigIndexedcompgraphfamily[\Lossfn{S,z}[\lossfn]]\,.
  \end{equation*}

  In particular, by the assumption on
  $\selgraphsymb$,
  \begin{equation*}
    \bigSelgraphsymb[
      \mkern2mu
      \Lossfn{S,z}[\lossfn]
      \mkern2mu
    ]
    \in
    \qquad\quad
    \forall\,
    z\in\truncsamplespacep{S}{r},\;
    \forall\,
    \bigIndexedcompgraphfamily[\Lossfn{S,z}[\lossfn]]\,.
  \end{equation*}
  Hence
  \begin{equation*}
    \sup_{\theta\in\truncation{\Theta}{r}}
    \left\|
      \sdv
      \bigSelgraphsymb[
        \mkern2mu
        \Lossfn{z}[\lossfn]
        \mkern2mu
      ](\theta)
    \right\|
    \leq
    B_{\lossfn,r}
    \qquad\quad
    \forall\,
    z\in\truncsamplespacep{S}{r}\,.
  \end{equation*}

  By
  Definition~\ref{def:selected-pseudo-derivative-operator},
  the above is equivalent to
  \begin{equation*}
    \sup_{\theta\in\truncation{\Theta}{r}}
    \left\|
      \sdvop[\selgraphsymb]{
        \Lossfn{z}[\lossfn]
      }(\theta)
    \right\|
    \leq
    B_{\lossfn,r}
    \qquad\quad
    \forall\,
    z\in\truncsamplespacep{S}{r}\,.
  \end{equation*}

  Since
  $\lossfn\in\admissiblelossfunctions[S]$
  and
  $r\in\nonnegreals$
  were arbitrary,
  Definition~\ref{def:sdv-biblg}
  implies that
  $S$
  satisfies
  $\sdv[\selgraphsymb]\mkern-2mu$-BIBLG.
\end{proof}

\begin{proposition}[Graph-uniform loss-gradient bounds imply ae-BIBLG under a.e.~correctness]
  \label{prop:graph-uniform-implies-ae-biblg}
  \todo{}
  Let
  $S$
  be a measurable-Euclidean parametric system,
  let
  $\indexedcompgraphfamily$
  be an objective computation-graph family for
  $S$,
  and
  let
  $\sdv$
  be a pseudo-derivative operator such that
  $S$
  satisfies
  $\indexedcompgraphfamily$-$\sdv$-BIBLG.

  Suppose that,
  for every
  $q\in\Lossfn{S}$,
  every
  $G\in\indexedcompgraphfamily[q]$,
  and
  every
  $r\in\nonnegreals$,
  there exists a full-measure subset
  \begin{equation*}
    \hat{\Theta}_r^{q,G}
    \subseteq
    \truncation{\Theta}{r}
  \end{equation*}
  such that,
  for every
  $\theta\in\hat{\Theta}_r^{q,G}$,
  \begin{equation*}
    q
    \text{ is differentiable at }
    \theta
    \qquad\text{and}\qquad
    \dvop{q}(\theta)
    =
    \sdv G(\theta)\,.
  \end{equation*}

  Then
  $S$
  satisfies ae-BIBLG.
\end{proposition}
\begin{proof}
  Let
  $\lossfn\in\admissiblelossfunctions[S]$
  and
  $r\in\nonnegreals$
  be arbitrary.
  Since
  $S$
  satisfies
  $\indexedcompgraphfamily$-$\sdv$-BIBLG,
  there exists
  $B_{\lossfn,r}\in\nonnegreals$
  such that
  \begin{equation}
    \label{eq:proof:graph-uniform-implies-ae-biblg-bound}
    \sup_{\theta\in\truncation{\Theta}{r}}
    \big\|
      \sdv G(\theta)
    \big\|
    \leq
    B_{\lossfn,r}
    \qquad\quad
    \forall\,
    z\in\truncsamplespacep{S}{r},
    \;\;
    \forall\,
    G
    \in
    \bigIndexedcompgraphfamily[
      \Lossfn{S,z}[\lossfn]
    ]\,.
  \end{equation}

  Let
  $z\in\truncsamplespacep{S}{r}$
  be arbitrary, and let
  \begin{equation*}
    q_z
    \defeq
    \Lossfn{S,z}[\lossfn]
    \,\in\,
    \Lossfn{S}.
  \end{equation*}
  Since
  $\indexedcompgraphfamily$
  is an objective computation-graph family for
  $S$,
  there exists a computation graph
  \begin{equation*}
    G_z
    \in
    \indexedcompgraphfamily[q_z].
  \end{equation*}

  By the assumed a.e.~correctness,
  there exists a full-measure subset
  \begin{equation*}
    \hat{\Theta}_r^{q_z,G_z}
    \subseteq
    \truncation{\Theta}{r}
  \end{equation*}
  such that, for every
  $\theta\in\hat{\Theta}_r^{q_z,G_z}$,
  \begin{equation*}
    q_z
    \text{ is differentiable at }
    \theta
    \qquad\text{and}\qquad
    \dvop{q_z}(\theta)
    =
    \sdv G_z(\theta)\,.
  \end{equation*}

  Hence, for every
  $\theta\in\hat{\Theta}_r^{q_z,G_z}$,
  Equation~\eqref{eq:proof:graph-uniform-implies-ae-biblg-bound}
  gives
  \begin{align*}
    \big\|
      \dvop{\Lossfn{S,z}[\lossfn]}(\theta)
    \big\|
    \,=\,
    \big\|
      \dvop{q_z}(\theta)
    \big\|
    \,=\,
    \big\|
      \sdv G_z(\theta)
    \big\|
    \,\leq\,
    B_{\lossfn,r}\,.
  \end{align*}

  Thus, for the constant
  $B_{\lossfn,r}$,
  every
  $z\in\truncsamplespacep{S}{r}$
  admits a full-measure subset of
  $\truncation{\Theta}{r}$
  on which
  $\Lossfn{S,z}[\lossfn]$
  is differentiable and its derivative norm is bounded by
  $B_{\lossfn,r}$.

  Since
  $\lossfn\in\admissiblelossfunctions[S]$
  and
  $r\in\nonnegreals$
  were arbitrary,
  Definition~\ref{def:ae-biblg}
  implies that
  $S$
  satisfies ae-BIBLG.
\end{proof}

\clearpage{}%

\clearpage{}%
\section{Additional Notions for MinMax RNCs}
\label{app:additional-rnc}
\label{appendix:additional-minmaxrnc}

\localtableofcontents

\subsection{MinMax Recurrence Operator}

We introduce the following notation for the MinMax recurrence.

\begin{closingdefinition}[MinMax recurrence operator]
  \label{def:minmax-recurrrence}
  The \emph{MinMax recurrence operator $\minmaxopsymb$} is the binary operator 
  \begin{align*}
    \minmaxop{x}{A}{b} \,\defeq\, (A\otimes x)\oplus b
  \end{align*}
  where $x \in \reals^n$ and $\ptuple{A,b} \in \reals^{n \times n} \!\times \reals^n$
  for $n \in \posnaturals$.

  The \emph{MinMax recurrence operator function $\minmaxopsymb_n$} of dimension $n \in \posnaturals$
  is the function
  \begin{align*}
    \minmaxopsymb_n \,:\, \reals^n \!\times (\reals^{n \times n} \times \reals^n) \to \reals^n,
    \qquad\qquad
    \minmaxopfunc{n}{x}{A}{b} \defeq \minmaxop{x}{A}{b}.
  \end{align*}
\end{closingdefinition}

\subsection{Parametric MinMax RNCs}
\label{subsec:parametric-minmax-rncs}

\begin{closingdefinition}[Parametric MinMax recurrent units]
  \label{def:parametric-minmax-recurrent-unit}
  A \emph{parametric MinMax recurrent unit} of state dimension $n$ is Euclidean
  parametric dynamics
  \begin{equation*}
    D
    =
    \ctuple{X,U,f}{\Theta}
  \end{equation*}
  having Euclidean ambient spaces
  \begin{equation*}
    X\subseteq\Omega_X=\reals^n,
    \qquad
    U\subseteq\Omega_U=\reals^m,
    \qquad
    \Theta\subseteq\Omega_\Theta=\reals^p,
  \end{equation*}
  for some $m,p\in\posnaturals$, together with continuous parametric functions
  \begin{align*}
    R:U\times\Theta
    &\to
    \reals^{n\times n},
    &
    s:U\times\Theta
    &\to
    \reals^n,
  \end{align*}
  such that
  \begin{equation*}
    \fpinst{f}{x,u}{\theta}
    =
    \bigminmax[
      x,
      \fpinst{R}{u}{\theta},
      \fpinst{s}{u}{\theta}]
    =
    \left(
      \fpinst{R}{u}{\theta}\otimes x
    \right)
    \oplus
    \fpinst{s}{u}{\theta}
  \end{equation*}
  for every
  $x\in X$,
  $u\in U$, and
  $\theta\in\Theta$.

  We call $R$ the \emph{reset function} and $s$ the \emph{set function}, and refer to
  them jointly as the \emph{input functions} of $D$, also writing 
  \begin{equation*}
    D
    =
    \mctuple{X,U,f}{\Theta}{R,s}
  \end{equation*}

  The unit $D$ is a \emph{parametric MinMax recurrent neuron} if its input functions $\ptuple{R,s}$
  are analytic.

  A MinMax recurrent unit is an instance $\pinst{D}{\theta}$ of $D$ for $\theta \in \Theta$,
  and it is a \emph{MinMax recurrent neuron} if its input functions
  $\bigptuple{\pinst{R}{\theta},\pinst{s}{\theta}}$ are analytic.

  We write
  $\classparneuralunits{MinMax}$
  for the class of parametric MinMax recurrent neurons,
  and 
  we write
  $\classneuralunits{MinMax}$
  for the class of MinMax recurrent neurons.
\end{closingdefinition}

\begin{closingdefinition}[Parametric MinMax recurrent neural cascades]
  \label{def:parametric-minmax-rnc}
  The class of \emph{MinMax Recurrent Neural Cascade dynamics
  (MinMax RNC dynamics)} is
  \begin{equation*}
    \rncdynop\big(\classneuralunits{MinMax}\big),
  \end{equation*}
  and the class of \emph{MinMax Recurrent Neural Cascades (MinMax RNCs)} is
  \begin{equation*}
    \rncop\big(\classneuralunits{MinMax}\big).
  \end{equation*}

  The class of \emph{parametric MinMax Recurrent Neural Cascade dynamics
  (parametric MinMax RNC dynamics)} is
  \begin{equation*}
    \rncdynop\big(\classparneuralunits{MinMax}\big),
  \end{equation*}
  and the class of \emph{parametric MinMax Recurrent Neural Cascades (parametric MinMax RNCs)} is
  \begin{equation*}
    \rncop\big(\classparneuralunits{MinMax}\big).
  \end{equation*}
\end{closingdefinition}

\begin{closingdefinition}[Parametric MinMax recurrent layers and cascades]
  \label{def:parametric-minmax-cascades}
  Let
  $\mathfrak{D}^\mathrm{p}_\mathrm{MinMax}$
  denote the class of parametric MinMax recurrent units.

  A \emph{parametric MinMax recurrent layer} is a parametric recurrent layer whose
  units belong to $\mathfrak{D}^\mathrm{p}_\mathrm{MinMax}$, and a
  \emph{parametric MinMax recurrent cascade} is a parametric recurrent cascade whose
  units belong to $\mathfrak{D}^\mathrm{p}_\mathrm{MinMax}$.
\end{closingdefinition}

\subsection{Floating-point MinMax RNCs}
\label{subsec:minmax--floating-point}

We first discuss the background, and then define floating-point MinMax RNCs.

\subsubsection{Floating-point MLPs}

We adopt all definitions from \citep{hwang25floating}.
In particular, we adopt the definition of floating-point domains and operations from
\S2.1 and the definition of floating-point neural networks from \S2.2.
We refer to their floating-point neural networks as $\mathbb{F}$-MLPs.
By their Corollaries~3.8 and~3.11, $\mathbb{F}$-MLPs can represent all functions over
$\mathbb{F}^n$.

Note that $\mathbb{F}$-MLPs are not simply ordinary MLPs restricted to the
floating-point subset of their real-valued domain.  Rather, they are compositions of
primitive functions supplied by the considered floating-point arithmetic.

\subsubsection{Floating-point MinMax RNCs}
\label{sec:float-minmaxrncs}

We next introduce MinMax recurrent units and RNCs evaluated in floating-point
arithmetic over a floating-point domain $\mathbb{F}$.

\begin{closingdefinition}[Floating-point MinMax recurrent units]
  \label{def:floating-point-minmax-recurrent-unit}
  Let $\mathbb{F}$ be a floating-point domain.
  A \emph{floating-point MinMax recurrent unit over $\mathbb{F}$ of state dimension $n$} is dynamics
  \begin{equation*}
    D
    =
    \tuple{X,U,f}
  \end{equation*}
  having
  \begin{equation*}
    X\subseteq\mathbb{F}^n,
    \qquad
    U\subseteq\mathbb{F}^m,
  \end{equation*}
  for some $m\in\posnaturals$, together with input functions
  \begin{align*}
    R:U
    &\to
    \mathbb{F}^{n\times n},
    &
    s:U
    &\to
    \mathbb{F}^n,
  \end{align*}
  such that
  \begin{equation*}
    f(x,u)
    =
    \big(R(u)\otimes x\big)\oplus s(u)
  \end{equation*}
  for every $x\in X$ and $u\in U$.

  We call $R$ the \emph{reset function} and $s$ the \emph{set function}, and refer to
  them jointly as the \emph{input functions} of $D$, also writing 
  \begin{equation*}
    D
    =
    \ctuple{X,U,f}{R,s}
  \end{equation*}

  The unit $D$ is a \emph{floating-point MinMax recurrent neuron} if $R$ and $s$ are
  $\mathbb{F}$-MLPs.
\end{closingdefinition}

\begin{closingdefinition}[Floating-point MinMax recurrent neural cascades]
  \label{def:floating-point-minmax-rnc}
  Let $\mathbb{F}$ be a floating-point domain.

  A \emph{floating-point MinMax recurrent layer over $\mathbb{F}$} is a recurrent
  layer whose units are floating-point MinMax recurrent units over $\mathbb{F}$.
  A \emph{floating-point MinMax recurrent cascade over $\mathbb{F}$} is a recurrent
  cascade whose units are floating-point MinMax recurrent units over $\mathbb{F}$.

  A \emph{floating-point MinMax Recurrent Neural Cascade (MinMax RNC) over
  $\mathbb{F}$} is a floating-point MinMax recurrent cascade such that:
  \begin{itemize}
    \item
      every unit is a floating-point MinMax recurrent neuron over $\mathbb{F}$;
    \item
      every intermediate function and the final output function are
      $\mathbb{F}$-MLPs, under the canonical identification of their finite Cartesian
      product domains with subsets of some $\mathbb{F}^d$.
  \end{itemize}

  The \emph{state degree} of a floating-point MinMax RNC is the maximum state
  dimension of any of its units.
\end{closingdefinition}

We observe that the only functions introduced by a floating-point MinMax RNC in
addition to those appearing in its $\mathbb{F}$-MLPs are the minimum and maximum
operations.  When evaluated on $\mathbb{F}$, these are native floating-point
operations in the sense that they return one of their floating-point arguments and
therefore introduce no additional rounding operation.

\clearpage{}%

\appendixpart{part:theory}{Theoretical Results for MinMax RNCs}

\clearpage{}%

\section{Expressivity of MinMax RNCs}
\label{app:expr}

In this appendix we provide full proofs of the results of Section~\ref{sec:mainbody-expressivity}.
\begin{itemize}
  \item
    Section~\ref{app:expre-prel-res} presents preliminary results.
    \begin{itemize}
      \item
        Subsection~\ref{app:expre-prel-res-perm} develops the preliminary results for characterising
        the state degree of MinMax Neurons in terms of the permutation degree of a semiautomaton.
      \item
        Subsections~\ref{app:comp-inp-f},~\ref{app:cascades-aux} jointly provide
        preliminary results regarding the cascading dynamics while preserving the characteristics
        we are interested in to show results later.
    \end{itemize}
  \item
    Section~\ref{app:realising-sem} proves the theorems of Section~\ref{sec:mainbody-semiautomata},
    also making use of the preliminary results from Subsection~\ref{app:expre-prel-res-perm}.
  \item
    Section~\ref{app:realising-casc} focuses on realising casacades of semiautomata using cascades
    of MinMax Neurons, also making use of the preliminary results from 
    Subsections~\ref{app:comp-inp-f},~\ref{app:cascades-aux}.
  \item
    Section~\ref{sec:appendix-expressivity-languages} proves the main expressivity theorem
    (Theorem~\ref{thm:main-body-expressivity}), relying on all of the above.
\end{itemize}

Additionally, we prove that the expressivity results hold seamlessly for floating-point MinMax RNCs
(Theorem~\ref{th:language-expressivity-overall-float} and Corollary~\ref{cor:language-expressivity-overall-float}).

\subsection{Preliminary Results}
\label{app:expre-prel-res}

\subsubsection{Permutation Degree}
\label{app:expre-prel-res-perm}

We provide the more formally-precise version of Definition~\ref{def:mainbody-perm-degree}.
\begin{definition}[See Def.~\ref{def:mainbody-perm-degree}]
  The \emph{minimal faithful permutation degree} of a group $G$, written $\mu(G)$, is the smallest
  number $n$ such that there is an isomorphism between $G$ and a subgroup of the symmetric group
  $\gsym{\iival{1}{n}}$.
  The \emph{degree} of a permutation semiautomaton is the degree of its transition monoid (which is
  a group).
  The \emph{degree} of a regular function $F$ is the maximum between one and the 
  degree of any simple group that is a factor of a subgroup of the transition monoid of the
  (semiautomaton of) the canonical automaton of $F$.
\end{definition}

\begin{theorem} \label{theorem:realisation-via-minimal-degree}
  Let $G$ be a group having minimal faithful permutation degree $n$, and let
  $A=\tuple{G,G,\delta}$ be the semiautomaton with $\delta(x,a)=xa$.
  There exists $\varphi : G \to \gsym{\iival{1}{n}}$
  such that
  $A$ is realised by the semiautomaton $A' = \tuple{\iival{1}{n}^n, G, \delta'}$ with
  $\delta'(\tuple{x_1, \ldots, x_n}, g) = \tuple{x_{\pi(1)}, \ldots, x_{\pi(n)}}$
  for $\pi$ the permutation $\pi = \varphi(g)$.
\end{theorem}
\begin{proof}
  For each $g\in G$, let
  \begin{equation*}
    \delta_g: G\to G,\qquad \delta_g(x)=xg,
  \end{equation*}
  Then, the transition group of $A$ is $G_A= \{\delta_g\mid g\in G\}$.

  Since $G$ has minimal faithful permutation degree equal to $n$,
  by Cayley's theorem (reported as Theorem~\ref{thm:cayley} in Section~\ref{app:prel-groups}),
  there exists an isomorphism $\alpha: G \to H$
  between $G$ and a subgroup $H$ of the symmetric group $\gsym{\iival{1}{n}}$.

  Let us define the semiautomaton $A'=(H,G,\delta')$
  where
  \begin{equation*}
    \delta'(h,g)= h\alpha(g)
    \qquad
    (h\in H,\ g\in G).
  \end{equation*}
  Noting that the inverse function $\alpha^{-1}: H \to G$ is well-defined since $\alpha$ is bijective,
  we define the function $\zeta: H \times G \to G \times G$ as $\zeta(h,g) =
  \tuple{\alpha^{-1}(h), g}$.
  Then we show that the tuple 
  $(\alpha, \mathrm{id}, \zeta)$ is an assignment of $\mealy{A}$ into $\mealy{A'}$.

  Let $x,\sigma \in G$.
  It suffices to show
  \begin{align*}
    \text{(I)}~~ & \delta'(\alpha(x), \sigma) = \alpha(\delta(x,\sigma)),
    \\
    \text{(II)}~~ & \zeta(\alpha(x), \sigma) = \tuple{x, \sigma}.
  \end{align*}

  We have that (I) holds since
  \begin{align*}
    \delta'(\alpha(x), \sigma)
    & =
    \alpha(x)\alpha(\sigma)
    \\
    & =
    \alpha(x \sigma)
    \\
    & =
    \alpha(\delta(x,\sigma)).
  \end{align*}

  We have that (II) holds since
  \begin{align*}
    \zeta(\alpha(x), \sigma) 
    & = \tuple{\alpha^{-1}(\alpha(x)), \sigma}
    \\
    & = \tuple{x, \sigma}.
  \end{align*}

  Now, let us define the semiautomaton $A''=(\iival{1}{n}^n,G,\delta'')$
  where
  \begin{equation*}
    \delta''(\tuple{i_1, \ldots, i_n},g) \defeq \tuple{i_{\pi(1)}, \ldots, i_{\pi(n)}} 
    \quad\text{ for } \pi = \alpha(g).
  \end{equation*}
  We define the function $\alpha': H \to \iival{1}{n}^n$ as 
  $\alpha'(h) = \tuple{h(1),\ldots, h(n)}$,
  and
  we define the function $\zeta': \iival{1}{n}^n \times G \to G \times G$ as 
  $\zeta'(\tuple{i_1, \ldots, i_n},g) \defeq \tuple{\pi, g}$
  for $\pi:\iival{1}{n} \to \iival{1}{n}$ such that $\pi(j) = i_j$.
  Then we show that the tuple 
  $(\alpha', \mathrm{id}, \zeta')$ is an assignment of $\mealy{A'}$ into $\mealy{A''}$.

  Let $h \in H,g \in G$.
  It suffices to show
  \begin{align*}
    \text{(I)}~~ & \delta''(\alpha'(h), g) = \alpha'(\delta'(h,g)),
    \\
    \text{(II)}~~ & \zeta'(\alpha'(g), g) = \tuple{h, g}.
  \end{align*}

  We have that (I) holds since
  \begin{align*}
    \delta''(\alpha'(h), g)
    & =
    \delta''(\tuple{h(1), \ldots, h(n)}, g)
    \\
    & =
    \tuple{h(\pi(1)), \ldots, h(\pi(n))} \quad\text{ for } \pi = \alpha(g)
    \\
    & =
    \tuple{(h \pi)(1), \ldots, (h \pi)(n)} 
    \\
    & =
    \alpha'(h \pi)
    \\
    & =
    \alpha'(h \alpha(g))
    \\
    \alpha'\big( \delta'(h,g)\big).
  \end{align*}

  We have that (II) holds since
  \begin{align*}
    \zeta'(\alpha'(h), g) 
    & = \zeta'(\tuple{h(1), \ldots, h(n)}, g) 
    \\
    & = \zeta'(\pi, g) 
  \end{align*}
  for $\pi$ such that $\pi(i) = h(i)$ for all $i$, but then $\pi = h$,
  and hence $\zeta'(\alpha'(h), g) = \zeta'(h, g)$ as required.

  Overall, we have shown that $A''$ realises $A'$, that in turn realises $A$.
  Thus, by transitivity, we conclude that $A''$ realises $A$.
  Furthermore, we have that $\varphi \defeq \alpha' \circ \alpha$ is a function 
  $G \to \gsym{\iival{1}{n}}$.
  Thus, $\varphi$ and $A''$ witness the claim of the theorem.
\end{proof}

\begin{theorem} \label{cor:realisation-via-minimal-degree}
  Let $A = \tuple{Q,\Sigma,\delta}$ be a permutation semiautomaton of degree $n$.
  There exists $\varphi : \Sigma \to \gsym{\iival{1}{n}}$
  such that
  $A$ is realised by the semiautomaton $A' = \tuple{\iival{1}{n}^n, \Sigma, \delta'}$ with
  $\delta'(\tuple{x_1, \ldots, x_n}, g) = \tuple{x_{\pi(1)}, \ldots, x_{\pi(n)}}$
  for $\pi$ the permutation $\pi = \varphi(\sigma)$.
\end{theorem}
\begin{proof}
  The transition monoid
  $\groupsty{M}(A) = (G,\circ)$ of $A$ is a group of degree $n$. 
  By Theorem~\ref{theorem:realisation-via-minimal-degree} there is 
  $\varphi' : G \to \gsym{\iival{1}{n}}$
  such that
  the $\groupsty{M}(A)$-semiautomaton is realised by the semiautomaton 
  $A' = \tuple{\iival{1}{n}^n, G, \delta'}$ with
  $\delta'(\tuple{x_1, \ldots, x_n}, g) = \tuple{x_{\pi(1)}, \ldots, x_{\pi(n)}}$
  for $\pi$ the permutation $\pi' = \varphi(g)$.
  Then $A'$ realises $A$ since it realises
  the $\groupsty{M}(A)$-semiautomaton, which realises $A$ since it has a superset of its
  transformations.
  Finally,
  $\varphi \defeq \psi \circ \varphi'$ where $\psi(\sigma) \defeq \delta_\sigma \in G$.
\end{proof}

\subsubsection{Composition with an input function}
\label{app:comp-inp-f}

\todo{}

\begin{definition}
Dynamics $D$ and $D'$ are
\emph{strongly equivalent} if $\mathcal{C}_\eta(D) = \mathcal{C}_\eta(D')$ (which also implies
$X=X'$ and $U=U'$).
\end{definition}

\begin{definition}
  Given dynamics $D = \tuple{X, U, f}$ and a function $\varphi : V \to U$,
  the composition $\varphi \circ D$ yields the dynamics $D_\varphi = \tuple{X, V, f_\varphi}$
  with $f_\varphi(x, v) = f(x, \varphi(v))$.
\end{definition}

\begin{definition}
  A class of functions $\Psi$ satifies the \emph{universal approximation property} if,
  for every continuous function $f: X \subseteq \reals^m \to \reals^n$ with $X$ compact
  and 
  for every $\epsilon > 0$,
  class $\Psi$ contains a function $\hat{f}: X \to \reals^n$ such that
  $\|f(x) - \hat{f}(x)\| \leq \epsilon$ for every $x \in X$.
\end{definition}

\begin{proposition}\label{prop:composition-approximation}
  Let $\Psi$ be a class of functions satisfying the universal approximation property,
  let $D$ be \etafinite{} dynamics with input space $U$,
  let $U'$ be an \etafinite{} space,
  and 
  let $\varphi: U' \to U$ be a continuous function.
  Furthermore,
  suppose that each $\eta$-component $U^\eta$ of $U$ satisfies
  $\bar{B}_{\Omega_U}(u) \subseteq U^\eta$ for some $u \in U$.
  Then,
  the dynamics $\varphi \circ D$ admit strongly equivalent dynamics
  $\psi \circ D$ for some $\psi \in \Psi$.
\end{proposition}
\begin{proof}
  Let $D = \tuple{X, U, f}$.
  Then,
  naming $D_\varphi \defeq \varphi \circ D$,
  we have $D_\varphi = \tuple{X, U', f_\varphi}$ with $f_\varphi(x,u') = f(x,\varphi(u'))$

  Let $U^\eta_1, \ldots, U^\eta_n$ be the $\eta$-components of $U$.
  By assumption, there exist $u_1, \ldots, u_n$ such that
  $B(u_i, \epsilon) \subseteq U^\eta_i$ for every $i \in \iival{1}{n}$.
  Let $\varphi': U' \to \{ u_1, \ldots, u_n \}$ be the function defined as $\varphi'(u') = u_i$
  for $i \in \iival{1}{n}$ such that $\varphi(u') \in U^\eta_i$.
  Fitst, we have that $\varphi'$ is well-defined since $U$ is partitioned by its $\eta$-components.
  Second, we have $\ol{\varphi'} = \ol{\varphi}$ since 
  $\ol{\varphi'}(\ol{u'}) = \ol{\varphi'(u')}$ and $\varphi'(u') \in \ol{\varphi(u')}$ by
  construction.
  Third, we have that $\varphi'$ is continuous;
  in fact,
  by continuity of $\varphi$, we have that $u'_1 \sim_{U'} u'_2$ implies
  $\varphi(u'_1) \sim_U \varphi(u'_2)$, hence $\varphi'$ is constant on the $\eta$-components of
  $U'$, and hence $\varphi'$ is continuous by Lemma~22 of~\cite{mat}.

  Then, since we have shown that $\varphi'$ is continuous,
  there exists $\psi \in \Psi$ that is an $\epsilon$-approximation of $\varphi'$.

  Letting $D_\psi \defeq \psi \circ D$,
  we have $D_\psi = \tuple{X, U', f_\psi}$ with $f_\psi(x,u') = f(x,\psi(u'))$.

  Since $D_\varphi$ and $D_\psi$ have the same state and input space, it suffices to show 
  $\ol{f_\varphi} = \ol{f_\psi}$.

  First we show $\ol{\psi} = \ol{\varphi}$.
  Namely, letting $u' \in U'$, we show that $\ol\psi(\ol{u'}) = \ol\varphi(\ol{u'})$.
  First, by the approximation assumption, we have
  \begin{align*}
    \psi(u') 
    \in 
    \bar{B}_{\Omega_U}\big(\varphi'(u'), \epsilon\big) 
    \subseteq 
    \ol{\varphi'(u')} 
    = 
    \ol{\varphi'}(\ol{u'})
    = 
    \ol{\varphi}(\ol{u'}),
  \end{align*}
  with the last equality argued above.
  Hence
  $\psi(u') \in \ol{\varphi}(\ol{u'})$,
  which implies
  $\ol{\psi(u')} = \ol{\varphi}(\ol{u'})$.
  Then,
  \begin{align*}
    \ol\psi(\ol{u'}) 
    & = 
    \ol{\psi(u')}
    =
    \ol{\varphi}(\ol{u'}),
  \end{align*}
  which concludes the proof of the claim $\ol{\psi} = \ol{\varphi}$.

  Finally we show $\ol{f_\varphi} = \ol{f_\psi}$ as required.
  By the two following chains of equalities,
  \begin{align*}
    & \ol{f_\varphi}(\ol x, \ol{u'}) = \ol{f_\varphi(x, u')} = \ol{f(x, \varphi(u'))}
    = \ol{f}(\ol x, \ol{\varphi(u')}),
    \\[.3em]
    & \ol{f_\psi}(\ol x, \ol{u'}) = \ol{f_\psi(x, u')} = \ol{f(x, \psi(u'))}
    = \ol{f}(\ol x, \ol{\psi(u')}),
  \end{align*}
  we conclude $\ol{f_\varphi} = \ol{f_\psi}$ since we have shown above $\ol\varphi = \ol\psi$.
\end{proof}

\begin{remark}
  Under strong equivalence, we can replace $D$ with $D'$ anywhere, e.g., in a cascade.
\end{remark}

\subsubsection{Cascades}
\label{app:cascades-aux}

\begin{lemma}\label{lemma:realize-parallel}
  Let us consider a parallel composition
  $D = (D_1 \parallelop \cdots \parallelop D_n)$
  where each $D_i$ is \etafinite{}.
  Let $U$ be the input space of $D$, and let $m$ be the number of $\eta$-components of $U$.
  For every $i \in \iival{1}{n}$,
  let $D'_i$ be dynamics that can realise $D_i$.
  Then, for every \etafinite{} space $\hat{U}$ whose number of $\eta$-components is $m$, 
  there exist continuous functions $\phi_1, \ldots, \phi_n$ with domain $\hat{U}$ such that 
  $D$ can be realised by
  the composition
  $D' = (\phi_1 \circ D'_1 \parallelop \cdots \parallelop \phi_n \circ D'_n)$.
\end{lemma}
\begin{proof}
  Let us note that, for each $i \in \iival{1}{n}$, the dynamics $D_i$ are of the form 
  $D_i = \tuple{X_i, U, f_i}$. In particular, all dynamics $D_1, \ldots, D_n$ share the same
  input space as they are part of a parallel composition.
  For each $i \in \iival{1}{n}$, 
  let dynamics $D'_i$ be of the form 
  $D'_i = \tuple{X'_i, U'_i, f'_i}$.
  Then, for each $i \in \iival{1}{n}$,
  since $D'_i$ is a realisation of $D_i$,
  there exists an assignment $(\alpha_i, \iota_i, \zeta_i)$ of
  $\mathcal{M}(\mathcal{C}_\eta(D_i))=\langle \overline X_i, \overline U, \overline f_i,
  \overline X_i \times \overline U_i, \mathrm{id}\rangle$ 
  into 
  $\mathcal{M}(\mathcal{C}_\eta(D_i'))= \langle \overline X'_i, \overline U'_i, \overline 
  f'_i, \overline X'_i \times \overline U'_i, \mathrm{id}\rangle$.

  Let $U^\eta_1, \ldots, U^\eta_m$ be the $\eta$-components of $U$,
  let $\hat{U}$ be an \etafinite{} space whose number of $\eta$-components is $m$,
  and 
  let $\hat{U}^\eta_1, \ldots, \hat{U}^\eta_m$ be its $\eta$-components.
  Let $u_1, \ldots, u_m$ be such that $u_i \in U^\eta_i$ for every $i \in
  \iival{1}{m}$, and hence $\ol{u_i} = U^\eta_i$.
  Similarly,
  let $\hat{u}_1, \ldots, \hat{u}_m$ be such that $\hat{u}_i \in \hat{U}^\eta_i$ for every $i \in
  \iival{1}{m}$, and hence $\ol{\hat{u}_i} = \hat{U}^\eta_i$.

  Then let us define $\iota: \ol{U} \to \ol{\hat{U}}$ as $\iota(\ol{u_j}) = \ol{\hat{u}_j}$, to be
  used below, also noting that it is a bijection and hence it admits an inverse 
  $\iota^{-1}: \ol{\hat{U}} \to \ol{U} $.

  For every $i \in \iival{1}{n}$,
  noting that $\iota_i$ is of the form $\iota_i: \ol U \to \ol{U'_i}$,
  and 
  letting $u'_{i,1}, \ldots, u'_{i,m_i} \in U'_i$ be arbitrarily-chosen elements such that 
  $\ol{u'_{i,1}}, \ldots, \ol{u'_{i,m_i}}$ are the $\eta$-components of $U'_i$,
  we define $\phi_i: \hat{U} \to U_i$ 
  as $\phi_i(\hat{u}) \defeq u'_{i,k}$
  such that
  $u'_{i,k} \in \iota_i\big(\iota^{-1}(\ol{\hat{u}})\big)$.
  We have that each $\phi_i$ is a well-defined continuous function.
  First, we have that
  each $\phi_i$ is well-defined since
  there exists a unique $k \in \iival{1}{m_i}$ such that $u'_{i,k} \in
  \iota_i\big(\iota^{-1}(\ol{\hat{u}})\big)$
  by our choice of $u'_{i,1}, \ldots, u'_{i,m_i}$.
  Second,
  we have that each $\phi_i$ is continuous by Lemma~22 of~\cite{mat}.

  Then, we have that $\ol{\phi_i} = \iota_i \circ \iota^{-1}$;
  in fact, we have
  $\ol{\phi_i}(\ol{\hat{u}}) = \iota_i(\iota^{-1}(\ol{\hat{u}}))$
  since 
  $\phi_i(\hat{u}) = u'_{i,k}$
  with $u'_{i,k} \in \iota_i\big(\iota^{-1}(\ol{\hat{u}})\big)$ by definition,
  and hence
  $\ol{\phi_i}(\ol{\hat{u}}) = \ol{u'_{i,k}} = \iota_i\big(\iota^{-1}(\ol{\hat{u}})\big)$.

  Let us note that $D$ and $D'$ are of the following form
  \begin{align*}
    D & = \tuple{X, U, f} 
    \quad \text{ with } \quad
    X = X_1 \times \cdots \times X_n
    \\[.5em]
    D' & = \tuple{X', U, f'} 
    \quad \text{ with } \quad
    X' = X'_1 \times \cdots \times X'_n.
  \end{align*}

  We show that $D'$ realises $D$,
  by showing an assignment $(\alpha, \iota, \zeta)$ of the $\eta$-canonical semiautomaton
  $\mathcal{C}_\eta(D)$ into the $\eta$-canonical semiautomaton $\mathcal{C}_\eta(D')$.
  Let us note that $\mathcal{C}_\eta(D)$ and $\mathcal{C}_\eta(D')$ are of the following form
  \begin{align*}
    \mathcal{C}_\eta(D) & = \tuple{\overline X, \overline U, \overline f} 
    \quad \text{ with } \quad
    \overline X = \overline X_1 \times \cdots \times \overline X_n,
    \\[.5em]
    \mathcal{C}_\eta(D') & = \tuple{\overline X', \overline U, \overline f'} 
    \quad \text{ with } \quad
    \overline X' = \overline X'_1 \times \cdots \times \overline X'_n.
  \end{align*}

  We have defined above the function $\iota$, and next we define the two remaining functions for
  an assignment $(\alpha, \iota, \zeta)$,
  \begin{align*}
    \alpha:\ & \overline X \to \mathcal{P}_+\big(\overline X'\big)
    && \text{as }\; \alpha \defeq \alpha_1 \times \cdots \times \alpha_n,
    \\[.5em]
    \zeta:\ & \big(\overline X' \times \ol{\hat{U}}\big) \to 
    \big(\overline X \times \overline U \big)
    && \text{as }\;\zeta\big(\langle \overline x_1',
    \ldots, \overline x_n'\rangle, \ol{\hat{u}} \big)
    \defeq \big \langle \tuple{\overline x_1, \ldots, \overline x_n}, \iota^{-1}(\ol{\hat{u}}) \big\rangle
    \\[.2em]
    & && \text{for } 
    \tuple{\overline x_i, \_} =\zeta_i(\overline x_i', \ol{\phi_i}(\hat{u})).
  \end{align*}

  Let us consider 
  \begin{gather*}
    \tuple{\overline x_1, \ldots, \overline x_n} \in \overline X,
    \qquad
     \overline u \in \overline U,
     \qquad
     \tuple{ \overline x_1', \ldots, \overline x_n'} \in 
    \alpha\big(\tuple{\overline x_1, \ldots, \overline x_n}\big),
    \\[.5em]
    \big\langle  \overline x_{1,\mathrm{new}}, \ldots,  \overline x_{m,\mathrm{new}}\big\rangle
    =
    \overline f\big(\langle\overline x_1, \ldots, \overline x_n\big\rangle,
    \overline u\big),
    \qquad
    \big\langle  \overline x_{1,\mathrm{new}}', \ldots,  \overline
    x_{n,\mathrm{new}}'\big\rangle
    =
    \overline f'\big(\langle\overline x_1', \ldots, \overline x_n'\big\rangle,
    \overline u\big).
  \end{gather*}

  By Property~(I) of assignment,
  for every $i \in \iival{1}{n}$,
  \begin{align*}
    \overline x_{i,\mathrm{new}}'
    =\overline f_i' \big(\overline x_i', \iota_i(\overline u)\big)
    \in
    \alpha_i\big(\overline f_i(\overline x_i, \overline u)\big),
  \end{align*}
  and hence
  \begin{align*}
    \bigtuple{\overline x_{1,\mathrm{new}}', \ldots, \overline x_{n,\mathrm{new}}'}
    & =\bigtuple{\overline f_1' \big(\overline x_1', \ol{\phi_1}(\iota(\overline u))\big),\ldots, \overline f_n'
  \big(\overline x_n', \ol{\phi_n}(\iota(\overline u))\big)}
    \\[.3em]
    & =\bigtuple{\overline f_1' \big(\overline x_1', \iota_1(\iota^{-1}(\iota(\overline u)))\big),\ldots, \overline f_n'
  \big(\overline x_n', \iota_n(\iota^{-1}(\iota(\overline u)))\big)}
  \mathsidecomment{Shown above $\ol{\phi_i} = \iota_i \circ \iota^{-1}$}
    \\[.3em]
    & =\bigtuple{\overline f_1' \big(\overline x_1', \iota_1(\overline u)\big),\ldots, \overline f_n'
    \big(\overline x_n', \iota_n(\overline u)\big)}
    \\[.3em]
    & \in 
    \left(\alpha_1\big(\overline f_1' \big(\overline x_1', \iota_1(\overline u)\big) \big)
    \times \cdots \times 
    \alpha_n\big(\overline f_n' \big(\overline x_n', \iota_n(\overline u)\big) \big) \right)
    \\[.3em]
    & =
    \alpha \big(
      \overline f \big( \tuple{\overline x_1, \ldots, \overline x_n}, \overline u \big) 
    \big),
  \end{align*}
  showing that the constructed assignment $(\alpha, \iota, \zeta)$ satisfies Property~(I)
  as an assignment of $\mathcal{C}(D)$ into $\mathcal{C}(D')$.

  For every $i \in \iival{1}{n}$,
  by definition of $\alpha$,
  we have
  that
  $\tuple{\overline x_1', \ldots, \overline x_n'}\in 
  \alpha\big(\tuple{\overline x_1, \ldots, \overline x_n}\big)$
  implies
  $\overline x_i' \in \alpha_i(\overline x_i)$,
  and hence by Property~(II) of assignment,
  \begin{align*}
    \zeta_i(\overline x_i', \iota_i(\overline u)) 
    = 
    \tuple{\overline x_i, \overline u}.
  \end{align*}
  Then,
  by definition of $\zeta$,
  \begin{align*}
    \zeta\big(\langle \overline x_1', \ldots, \overline x_n'\rangle, \iota(\ol u) \big)
    & = 
    \big \langle \tuple{\ol{x''_1}, \ldots, \ol{x''_n}}, \iota^{-1}(\iota(\ol u)) \big\rangle
    \quad\text{ for }\quad
    \tuple{\ol{x''_i}, \_} =\zeta_i(\overline x_i', \ol{\phi_i}(\iota(\ol u))).
  \end{align*}
  We have $\ol{x''}_i = \ol{x}_i$
  since $\ol{\phi_i} = \iota_i \circ \iota^{-1}$,
  and hence 
  $\zeta_i(\overline x_i', \ol{\phi_i}(\iota(\ol u))) = \zeta_i(\overline x_i', \iota_i(\ol u))
  = \tuple{\overline x_i, \overline u}$.
  Also, $\iota^{-1}(\iota(\ol u)) = \ol u$.
  Therefore, the equation above becomes
  \begin{align*}
    \zeta\big(\langle \overline x_1', \ldots, \overline x_n'\rangle, \iota(\ol u) \big)
    & = 
    \big \langle \tuple{\ol{x_1}, \ldots, \ol{x_n}}, \ol u \big\rangle,
  \end{align*}
  as required to show
  that the constructed assignment $(\alpha, \iota, \zeta)$ satisfies Property~(II)
  as an assignment of $\mathcal{C}(D)$ into $\mathcal{C}(D')$.
  Therefore $(\alpha, \iota, \zeta)$ is an assignment of $\mathcal{C}(D)$ into $\mathcal{C}(D')$
  as required.
\end{proof}

\begin{lemma}\label{lemma:realize-cascade}
  Let us consider a cascade
  \begin{align*}
    C & = 
    \big( D_{1,1} \parallelop \cdots \parallelop D_{1,m_1} \big)
    \ltimes \cdots \ltimes
    \big( D_{d,1} \parallelop \cdots \parallelop D_{d,m_d} \big),
  \end{align*}
  where each $D_{i,j}$ is $\eta$-finite.
  Let $U$ be the input space of $D$, and let $m$ be the number of $\eta$-components of $U$.
  For every $i \in \iival{1}{d}$ and every $j \in \iival{1}{m_i}$,
  let $D'_{i,j}$ be dynamics that can realise $D_{i,j}$.
  Then, for every \etafinite{} space $\hat{U}$ whose number of $\eta$-components is $m$, 
  there exist continuous functions $\phi_{i,j}$ such that
  $C$ can be realised by
  \begin{align*}
    C' & = 
    \big( \phi_{1,1} \circ D'_{1,1} \parallelop \cdots \parallelop \phi_{1,m_1} \circ D'_{1,m_1} \big)
    \ltimes \cdots \ltimes
    \big( \phi_{d,1} \circ D'_{d,1} \parallelop \cdots \parallelop \phi_{d,m_d} \circ D'_{d,m_d} \big)
  \end{align*}
  and every $\phi_{1,j}$ has domain $\hat{U}$.
\end{lemma}
\begin{proof}
  For each $i \in \iival{1}{d}$,
  let us name 
  \begin{align*}
    C_i & \defeq 
    \big( D_{1,1} \parallelop \cdots \parallelop D_{1,m_1} \big)
    \ltimes \cdots \ltimes
    \big( D_{i,1} \parallelop \cdots \parallelop D_{i,m_{d-1}} \big),
  \end{align*}
  For each $i \in \iival{1}{d}$ and each $j \in \iival{1}{m_i}$,
  let $D_{i,j} = \tuple{X_{i,j}, U_i, f_{i,j}}$, noting that dynamics occurring in the same parallel
  composition share the same input space.

  We show the claim by induction on $d$.
  In the base case $d=1$ and the claim holds by Lemma~\ref{lemma:realize-parallel}.
  In the inductive case $d>1$ and we assume the claim holds for $d-1$. 

  By Lemma~\ref{lemma:realize-parallel}, there exist continuous functions 
  $\varphi_{d,1}, \ldots, \varphi_{d,m_d}$ with domain $U_{d-1}$ such that
  \begin{align*}
    \hat{D}_d 
    \defeq 
    \big( 
      \varphi_{d,1} \circ D'_{d,1} 
      \parallelop \cdots \parallelop 
      \varphi_{d,m_d} \circ D'_{d,m_d} 
    \big)
  \end{align*}
  can realise
  \begin{align*}
    D_d 
    \defeq 
    \big( 
      D_{d,1} \parallelop \cdots \parallelop D_{d,m_d} 
    \big).
  \end{align*}
  Since $\hat{D}_d$ can realise $D_d$,
  there exists an assignment $(\alpha_d, \iota_d, \zeta_d)$ of
  $\mceta{D_d}$ into $\mceta{\hat{D}_d}$.

  Furthermore,
  by the inductive hypothesis, there exist functions $\phi_{i,j}$ such that the cascade
  \begin{align*}
    C'_{d-1} & = 
    \big( \phi_{1,1} \circ D'_{1,1} \parallelop \cdots \parallelop \phi_{1,m_1} \circ D'_{1,m_1} \big)
    \ltimes \cdots \ltimes
    \big( 
    \phi_{d-1,1} \circ D'_{d-1,1} 
    \parallelop \cdots \parallelop 
    \phi_{d-1,m_{d-1}} \circ D'_{d-1,m_d} 
    \big)
  \end{align*}
  can realise $C_{d-1}$.
  In particular, we can choose each function $\phi_{1,j}$ to have domain $\hat{U}$,
  and each function $\phi_{i,j}$ for $i > 1$ to have domain $U_i$.
  Since 
  $C'_{d-1}$ can realise $C_{d-1}$,
  there exists an assignment $(\alpha_\mathrm{ih}, \iota_\mathrm{ih}, \zeta_\mathrm{ih})$ of
  $\mathcal{M}(\mathcal{C}_\eta(C_{d-1}))$ into $\mathcal{M}(\mathcal{C}_\eta(C'_{d-1}))$,

  Let us choose a continuous function 
  $\varphi: U_d \to U_d$
  and
  $\psi: \mathbf{X}'_{d-1} \times \hat{U} \to U_d$
  such that $\ol{\varphi} = \iota_d$ and $\ol{\psi} = \zeta_\mathrm{ih}$, 
  which exist by Lemma~22 of \citep{mat}.
  Then, for each $i \in \iival{1}{m_d}$, 
  let us define $\phi_{d,i} \defeq  \varphi_{d,i} \circ \varphi \circ \psi$.
  Then, let
  \begin{align*}
    D'_d 
    \defeq 
    \big( 
      \phi_{d,1} \circ D'_{d,1} 
      \parallelop \cdots \parallelop 
      \phi_{d,m_d} \circ D'_{d,m_d} 
    \big),
  \end{align*}
  and let us define
  \begin{align*}
    C'_{d-1} & \defeq 
    \big( D'_{1,1} \parallelop \cdots \parallelop D'_{1,m_1} \big)
    \ltimes \cdots \ltimes
    \big( D'_{d-1,1} \parallelop \cdots \parallelop D'_{d-1,m_{d-1}} \big),
    \\[.3em]
    C' & \defeq C'_{d-1} \ltimes D'_d.
  \end{align*}
  Also, let us note that
  \begin{align*}
    C & = C_{d-1} \ltimes D_d.
  \end{align*}

  The above dynamics are of the following form,
  \begin{align*}
    C_{d-1} & 
    = 
    \tuple{
      \mathbf{X}_{d-1},
      U,\, 
      \mathbf{f}_{d-1}
    },
    \\[.3em]
    C'_{d-1} & 
    = 
    \tuple{
      \mathbf{X}'_{d-1},
      \hat{U},\, 
      \mathbf{f}'_{d-1}
    },
    \\[.3em]
    D_d & = \tuple{X_d,\, U_d,\, f_d}
    \quad\text{ with }\quad
    U_d = U \times \mathbf{X}_{d-1},
    \\[.3em]
    D'_d & = \tuple{X'_d,\, U'_d,\, f'_d}
    \quad\text{ with }\quad
    U'_d = \hat{U} \times \mathbf{X}'_{d-1},
  \end{align*}
  and hence
  \begin{align*}
    C_d & 
    = 
    \tuple{
      \mathbf{X}_{d-1} \times X_d,\;
      U,\; 
      \mathbf{f}_{d-1} \ltimes f_d
    },
    \\[.3em]
    C'_d & 
    = 
    \tuple{
      \mathbf{X}'_{d-1} \times X'_d,\;
      \hat{U},\; 
      \mathbf{f}'_{d-1} \ltimes f'_d
    }.
  \end{align*}
  Finally, let us note
  \begin{align*}
    \mceta{D_d}
    & =
    \bigtuple{
      \ol{X_d},\;
      \ol{U_d},\; 
      \ol{f_d},\;
      \ol{X_d} \times \ol{U_d},\;
      \mathrm{id}
    },
    \\[.3em]
    \mceta{D'_d}
    & =
    \bigtuple{
      \ol{X'_d},\;
      \ol{U'_d},\; 
      \ol{f'_d},\;
      \ol{X'_d} \times \ol{U'_d},\;
      \mathrm{id}
    },
    \\[.3em]
    \mceta{C_{d-1}}
    & =
    \bigtuple{
      \ol{\mathbf{X}_{d-1}} \times \ol{X_d},\;\;
      \ol{U},\; \;
      \ol{\mathbf{f}_{d-1}} \ltimes \ol{f_d},\;\;
      \ol{\mathbf{X}_{d-1}} \times \ol{X_d} \times \ol{U},\;\;
      \mathrm{id}
    },
    \\[.3em]
    \mceta{C'_{d-1}}
    & =
    \bigtuple{
      \ol{\mathbf{X}'_{d-1}} \times \ol{X'_d},\;\;
      \ol{\hat{U}},\; \;
      \ol{\mathbf{f}'_{d-1}} \ltimes \ol{f'_d},\;\;
      \ol{\mathbf{X}'_{d-1}} \times \ol{X'_d} \times \ol{\hat{U}},\;\;
      \mathrm{id}
    }.
  \end{align*}

  We show that $C'$ can realise $C$ by showing that
  the triple $(\alpha, \iota, \zeta)$ defined below is an assignment of
  $\mceta{C_d}$ 
  into
  $\mceta{C'_d}$,
  \begin{align*}
    \alpha:\ & \ol{\mathbf{X}_{d-1}} \times \ol{X_d} \to \mathcal{P}_+\big(\ol{\mathbf{X}'_{d-1}}
    \times \ol{X'_d}\big)
    && \text{as }\; \alpha \defeq \alpha_\mathrm{ih} \times \alpha_d,
    \\[.3em]
    \iota :\ & \ol{U} \to \ol{\hat{U}} && \text{as }\;\iota \defeq \iota_\mathrm{ih},
    \\[.3em]
    \zeta:\ & \big(\ol{\mathbf{X}'_{d-1}} \times \ol{X'_d} \times \ol{\hat{U}} \big) \to 
    \big( \ol{\mathbf{X}_{d-1}} \times \ol{X_d} \times \ol{U} \big)
    && \text{as }\;
    \zeta\big(\tuple{\ol{\mathbf{x}'_{d-1}},  \ol{x'_d}}, \ol{\hat{u}} \big)
    \defeq 
    \tuple{c,b,a}
    \\[.1em]
    & && \text{for }
    \tuple{a,b,c} = \zeta_d\big(\ol{x'_d}, \iota_d(\zeta_\mathrm{ih}(\ol{\mathbf{x}'_{d-1}},
    \ol{\hat{u}}))\big).
  \end{align*}

  Let us consider 
  \begin{equation*}
    \ol{u},\;\;
    \ol{\hat{u}},\;\;
    \ol{\mathbf{x}_{d-1}},\;\;
    \ol{x_d},\;\;
    \ol{\mathbf{x}'_{d-1}},\;\;
    \ol{x'_d}
  \end{equation*}
  such that
  \begin{align*}
    &\ol{u} \in \ol{U},
     \\[.3em]
     & \ol{\hat{u}} = \iota(\ol{u}) 
     && \text{i.e., } \ol{\hat{u}} = \iota_\mathrm{ih}(\ol{u}),
     \\[.3em]
     &\tuple{\ol{\mathbf{x}_{d-1}}, \ol{x_d}} \in \ol{\mathbf{X}_{d-1}} \times \ol{X_d},
     \\[.3em]
     &\tuple{\ol{\mathbf{x}'_{d-1}}, \ol{x'_d}} \in 
    \alpha\big(\tuple{\ol{\mathbf{x}_{d-1}}, \ol{x_d}}\big),
    &&
    \text{i.e., } \ol{\mathbf{x}'_{d-1}} \in \alpha_\mathrm{ih}\big(\ol{\mathbf{x}_{d-1}}\big)
    \;\land\; \ol{x'_d} \in \alpha_d\big(\ol{x_d}\big).
  \end{align*}
  Furthermore,
  let us consider 
  \begin{equation*}
    \ol{\mathbf{x}_{d-1,\mathrm{new}}},\;\;
    \ol{x_{d,\mathrm{new}}},\;\;
    \ol{\mathbf{x}'_{d-1,\mathrm{new}}},\;\;
    \ol{x'_{d,\mathrm{new}}}
  \end{equation*}
  such that
  \begin{align}
    \label{tmp1}
    \bigtuple{\ol{\mathbf{x}_{d-1,\mathrm{new}}}, \ol{x_{d,\mathrm{new}}}}
    & =
    \bigtuple{\ol{\mathbf{f}_{d-1}}\big(\ol{\mathbf{x}_{d-1}},\ol{u}\big),\; 
    \ol{f_d}\big(\ol{x_d}, \tuple{\ol{\mathbf{x}_{d-1}},\ol{u}}\big)},
    \\[.5em]
    \label{tmp2}
    \bigtuple{\ol{\mathbf{x}'_{d-1,\mathrm{new}}}, \ol{x'_{d,\mathrm{new}}}}
    & =
    \bigtuple{\ol{\mathbf{f}'_{d-1}}\big(\ol{\mathbf{x}'_{d-1}},\ol{\hat{u}}\big),\; 
    \ol{f'_d}\big(\ol{x'_d}, \tuple{\ol{\mathbf{x}'_{d-1}},\ol{\hat{u}}}\big)}.
  \end{align}

  To show Property~(I) of assignment, it suffices to show
  \begin{align*}
    &
  \bigtuple{\ol{\mathbf{x}'_{d-1,\mathrm{new}}}, \ol{x'_{d,\mathrm{new}}}}
  \in
  \alpha\big(\bigtuple{\ol{\mathbf{x}_{d-1,\mathrm{new}}}, \ol{x_{d,\mathrm{new}}}}\big),
  \\[.5em]
  \text{i.e., } & \;
  \ol{\mathbf{x}'_{d-1,\mathrm{new}}} \in \alpha_\mathrm{ih}\big(\ol{\mathbf{x}_{d-1,\mathrm{new}}}\big)
  \;\land\;
  \ol{x'_{d,\mathrm{new}}} \in  \alpha_d\big( \ol{x_{d,\mathrm{new}}}\big).
  \end{align*}
  First, we have
  \begin{align*}
  \ol{\mathbf{x}'_{d-1,\mathrm{new}}} 
  & = 
  \ol{\mathbf{f}'_{d-1}}\big(\ol{\mathbf{x}'_{d-1}},\ol{\hat{u}}\big)
  \mathsidecomment{\eqref{tmp2}}
  \\[.3em]
  & = 
  \ol{\mathbf{f}'_{d-1}}\big(\ol{\mathbf{x}'_{d-1}},\ol{\iota_\mathrm{ih}}(\ol{u})\big)
  \mathsidecomment{def.\ of $\iota$}
  \\[.3em]
  & \in
  \alpha_\mathrm{ih}\big(\ol{\mathbf{f}_{d-1}}\big(\ol{\mathbf{x}_{d-1}},\ol{u}\big)\big).
  \mathsidecomment{property~(I) since $\ol{\mathbf{x}'_{d-1}} \in \alpha_\mathrm{ih}(\ol{\mathbf{x}_{d-1}})$}
  \\[.3em]
  & =
  \alpha_\mathrm{ih}\big(\ol{\mathbf{x}_{d-1,\mathrm{new}}}\big).
  \mathsidecomment{\eqref{tmp1}}
  \end{align*}
  Then, we have
  \begin{align*}
    \ol{x'_{d,\mathrm{new}}}
    & =
    \ol{f'_d}\big(\ol{x'_d}, \tuple{\ol{\mathbf{x}'_{d-1}},\ol{\hat{u}}}\big)
    \mathsidecomment{\eqref{tmp2}}
    \\[.3em]
    & =
    \ol{f'_d}\big(\ol{x'_d}, \tuple{\ol{\mathbf{x}'_{d-1}},\iota(\ol{u})}\big)
    \mathsidecomment{def.\ of $\ol{\hat{u}}$}
    \\[.3em]
    & =
    \ol{\hat{f}_d}\big(\ol{x'_d}, \ol{\varphi}(\ol{\psi}(\ol{\mathbf{x}'_{d-1}},\iota(\ol{u})))\big)
    \mathsidecomment{def.\ of $f'_d$, letting $\hat{f}_d$ the dynamics func. of $\hat{D}_d$}
    \\[.3em]
    & =
    \ol{\hat{f}_d}\big(\ol{x'_d}, \iota_d(\zeta_\mathrm{ih}(\ol{\mathbf{x}'_{d-1}},\iota(\ol{u})))\big)
    \mathsidecomment{def.\ of $\varphi$ and $\psi$}
    \\[.3em]
    & =
    \ol{\hat{f}_d}\big(\ol{x'_d},
    \iota_d(\zeta_\mathrm{ih}(\ol{\mathbf{x}'_{d-1}},\iota_\mathrm{ih}(\ol{u})))\big)
    \mathsidecomment{def.\ of $\iota$}
    \\[.3em]
    & =
    \ol{\hat{f}_d}\big(\ol{x'_d}, \iota_d(\ol{\mathbf{x}_{d-1}},\ol{u}))\big)
    \mathsidecomment{property~(II) since $\ol{\mathbf{x}'_{d-1}} \in \alpha_\mathrm{ih}(\mathbf{x}_{d-1})$}
    \\[.3em]
    & \in
    \alpha_d\big(\ol{f_d}\big(\ol{x_d}, \tuple{\ol{\mathbf{x}_{d-1}},\ol{u}}\big)\big)
    \mathsidecomment{property~(I) since $\ol{x'_d} \in \alpha_d(\ol{x_d})$}
    \\[.3em]
    & =
    \alpha_d\big(\ol{x_{d,\mathrm{new}}}\big).
    \mathsidecomment{\eqref{tmp1}}
  \end{align*}
  Thus, Property~(I) holds for the constructed assignment $(\alpha, \iota, \zeta)$.

  Next we show Property~(II).
  By definition of $\zeta$, we have
  \begin{align*}
    & \zeta\big(\tuple{\ol{\mathbf{x}'_{d-1}},  \ol{x'_d}}, \ol{\hat{u}} \big)
      =
    \tuple{c, b, a},
    \\
    &
    \text{where } \tuple{a,b,c} = \zeta_d\big(\ol{x'_d},\, \iota_d(\zeta_\mathrm{ih}(\ol{\mathbf{x}'_{d-1}},
    \ol{\hat{u}}))\big).
  \end{align*}
  Then,
  \begin{align*}
    \zeta_\mathrm{ih}(\ol{\mathbf{x}'_{d-1}}, \ol{\hat{u}})
    & =
    \zeta_\mathrm{ih}(\ol{\mathbf{x}'_{d-1}}, \iota_\mathrm{ih}(\ol{u})) 
    \mathsidecomment{def.~of $\ol{\hat{u}}$}
    \\[.3em]
    & = \tuple{\ol{\mathbf{x}_{d-1}}, \ol{u}},
    \mathsidecomment{Property~(II)}
  \end{align*}
  and 
  \begin{align*}
    \zeta_d\big(\ol{x'_d},\, \iota_d(\ol{\mathbf{x}_{d-1}}, \ol{u})\big)
    =
    \tuple{\ol{x_d}, \ol{u}, \ol{\mathbf{x}_{d-1}}}.
    \mathsidecomment{Property~(II)}
  \end{align*}
  Therefore,
  \begin{align*}
    \zeta\big(\tuple{\ol{\mathbf{x}'_{d-1}},  \ol{x'_d}}, \ol{\hat{u}} \big)
      =
    \tuple{\ol{\mathbf{x}_{d-1}}, \ol{x_d}, \ol{u}},
  \end{align*}
  as required by Property~(II).
\end{proof}

\begin{definition} \label{def:equivalent-dynamics}
  Dynamics $D$ and $D'$ are \emph{equivalent} if
  \begin{inlineenum}
  \item
    for every system $S$ with dynamics $D$, there is 
    a system $S'$ with dynamics $D'$ such that $S$ and $S'$ can implement the same functions,
    and
  \item
    for every system $S'$ with dynamics $D'$, there is 
    a system $S$ with dynamics $D$ such that $S$ and $S'$ can implement the same functions.
  \end{inlineenum}
\end{definition}

\begin{proposition} \label{prop:equiv-same-automaton}
  Given \etafinite{} dynamics $D$ and $D'$,
  if $\ceta{D} = \ceta{D'}$,
  then $D$ and $D'$ are equivalent.
\end{proposition}
\begin{proof}
  Immediate by Theorem~31 of~\citep{mat}.
\end{proof}

\begin{lemma}\label{lemma:replace-cascade}
  Let $\Phi$ be a class of functions satisfying the universal approximation property.
  Let us consider a cascade
  \begin{align*}
    C & = 
    \big( \phi_{1,1} \circ D_{1,1} \parallelop \cdots \parallelop \phi_{1,m_1} \circ D_{1,m_1} \big)
    \ltimes \cdots \ltimes
    \big( \phi_{d,1} \circ D_{d,1} \parallelop \cdots \parallelop \phi_{d,m_d} \circ D_{d,m_d} \big)
  \end{align*}
  where each $D_{i,j}$ is over $\reals$, it is \etafinite{}, and it has an input space $U_i$ such
  that every $\eta$-component $U^\eta$ of $U_i$ satisfies
  $\bar{B}_{\Omega_{U_i}}(u, \epsilon) \subseteq U^\eta$ for some $u \in U^\eta$ and some $\epsilon > 0$.
  Then, 
  there exist $\hat{\phi}_{i,j} \in \Phi$
  such that the cascade
  \begin{align*}
    \hat{C} & = 
    \big( \hat{\phi}_{1,1} \circ D_{1,1} \parallelop \cdots \parallelop \hat{\phi}_{1,m_1} \circ D_{1,m_1} \big)
    \ltimes \cdots \ltimes
    \big( \hat{\phi}_{d,1} \circ D_{d,1} \parallelop \cdots \parallelop \hat{\phi}_{d,m_d} \circ
    D_{d,m_d} \big),
  \end{align*}
  is equivalent to $C$.
\end{lemma}
\begin{proof}
  Let $i \in \iival{1}{d}$
  and
  let $j \in \iival{1}{m_i}$.
  Let $U_i$ be the input space of $D_{i,j}$,
  and
  let $U_{i,1}^\eta,\ldots,U_{i,n_i}^\eta$ be the $\eta$-components of $U_i$.
  By assumption,
  for each $\ell \in \iival{1}{n_i}$,
  there are $\tilde{u}_{i,\ell} \in U_{i,\ell}^\eta$ and $\epsilon > 0$
  such that $\bar{B}_{\Omega_{U_i}}(\tilde{u}_{i,\ell},\epsilon) \subseteq U_{i,\ell}^\eta$.
  Hence,
  by Lemma~22 of~\citep{mat}, there is a continuous function $\phi'_{i,j}$ 
  with image in $\{\tilde{u}_{i,1},\ldots,\tilde{u}_{i,n_i}\}$ satisfying
  $\ol{\phi'_{i,j}} = \ol{\phi_{i,j}}$.
  We choose $\hat{\phi}_{i,j} \in \Phi$ that is an $\epsilon$-approximation
  of $\phi'_{i,j}$.
  It follows that $\hat{\phi}_{i,j}(z) \in \bar{B}_{\Omega_{U_i}}(\phi'_{i,j}(z), \epsilon)
  \subseteq U^\ell_{i,j}$ for all $z \in U_i$,
  hence
  $\ol{\hat{\phi}_{i,j}} = \ol{\phi_{i,j}}$,
  and 
  hence 
  the $\eta$-canonical semiautomaton $\ceta{\hat{D}_{i,j}}$ is the same as the $\eta$-canonical
  semiautomaton $\ceta{D_{i,j}}$.
  Then,
  the same holds for the entire cascade; namely,
  the $\eta$-canonical semiautomaton $\ceta{\hat{C}}$ coincides with the $\eta$-canonical
  semiautomaton $\ceta{C}$.
  Then the result follows immediately by Proposition~\ref{prop:equiv-same-automaton}.
\end{proof}

\subsection{Realising semiautomata using single MinMax Recurrent Units}
\label{app:realising-sem}

In this section we first provide an additional result and then provide full proofs of the results of
Section~\ref{sec:mainbody-semiautomata}.
However, here we adopt a different execution of the proofs, where we first show the existence of
MinMax Units that allow for choosing states and inputs in a flexible way, showing tha the input
functions can be approximated, but replacing them with MLPs at a later step. 

The next theorem is an additional result, showing that a single MinMax Recurrent Unit can realise
any semiautomaton. The idea is to represent the transition matrix of the
semiautomaton.

\begin{restatable}{lemma}{lemmaexpressivitymainbasic}
  \label{lemma:expressivity-main-basic}
  Let $A$ be a semiautomaton having $n$ states and $m$ inputs,
  let $\epsilon \geq 0$,
  let $x_0,x_1 \in \reals$ with $(x_1 - x_0) > 2\epsilon$,
  and 
  let $U \subseteq \reals$ be the union of $m$ pair-wise disjoint closed intervals.
  There exist continuous functions 
  $R: U \to \{ x_0,x_1 \}^{n \times n}$ and $s: U \to \{ x_0,x_1 \}^n$ such that
  semiautomaton $A$ can be realised by the \etafinite{} MinMax Recurrent Unit 
  $D = \tuple{X^n, U, f \mid R, s}$ where $X = [x_0-\epsilon,x_0+\epsilon] \cup [x_1-\epsilon,x_1+\epsilon]$;
  furthermore, for every $\epsilon$-approximations $(\hat{R},\hat{s})$ of $(R,s)$,
  the MinMax Recurrent Unit
  $\hat{D} = \tuple{X^n, U, f \mid \hat{R}, \hat{s}}$ is strongly equivalent to $D$.
\end{restatable}
\begin{proof}
  Let $A = \tuple{Q, \Sigma, \delta}$ be a semiautomaton 
  with 
  states $Q = \{ q_1, \ldots, q_n \}$
  and 
  input alphabet $\Sigma = \{ \sigma_1, \ldots, \sigma_m \}$.

  \paragraph{Construction.}
  We define the MinMax Recurrent Unit $D = \tuple{X^n, U, f \mid R, s}$.

  Let $U_1, \ldots, U_m \subseteq \reals$ be the pair-wise disjount closed intervals such that 
  $U = \bigcup_{i=1}^m U_i$.

  We define the reset function $R: U \to \{ x_0, x_1 \}^{n \times n}$
  and the set function $s: U \to \{ x_0, x_1 \}^n$.
  Let $\Delta \defeq x_1 - x_0$.
  For each $i \in \iival{1}{m}$,
  let $T_i \in \{0,1\}^{n \times n}$ be the 
  matrix such that $(T_i)_{j,k} = 1$ iff $\delta(q_k,\sigma_i) = q_j$.
  For each $i \in \iival{1}{m}$,
  let $a_i: U \to \{ 0, \Delta \}$ be the function defined as 
  $a_i(u) = \Delta$ iff $u \in U_i$.
  Note that the functions $a_1, \ldots, a_m$ are continuous by
  Lemma~\ref{lemma:mat-eta-constant-continuous} since they are
  constant over $\eta$- components of $U$.
  Then,
  \begin{align*}
    R(u) & \defeq x_0 \cdot \mathbf{1}_{n,n} 
    + 
    \sum_{i=1}^m a_i(u) \cdot T_i,
    \\
    s(u) & \defeq x_0 \cdot \mathbf{1}_n.
  \end{align*}
  Note that function $R$ is continuous since it is the composition of continuous functions,
  and function $s$ is continuous since it is constant.
  This completes the construction of the MinMax Recurrent Unit. Next we prove the claims.

  \paragraph{$\eta$-finiteness.}
  The MinMax Recurrent Unit $D$ is $\eta$-finite since $U$ is a finite union of closed
  intervals, and $X^n$ is the cross-product of a finite union of closed intervals.

  \paragraph{Realisation.}
  To show that $D$ can realise $A$, it suffices to show that
  $\mathcal{M}(\mathcal{C}_\eta(D))$ is a realisation of $\mathcal{M}(A)$.
  Note that $\mathcal{M}(A) = \tuple{Q, \Sigma, \delta, \Gamma, \theta}$
  where $\Gamma = Q \times \Sigma$ and $\theta = \mathrm{id}$.

  For each $i \in \iival{1}{m}$,
  let us choose an arbitrary $u_i \in \ol{U_i}$.
  The construction of $D$ allows us to easily derive that its canonical
  semiautomaton has the form
  $\mathcal{C}_\eta(D) = \tuple{\ol{X}^n, \ol U, \ol{f}}$
  where
  \begin{align*}
    \ol{X}^n 
    & =
    \big\{ \tuple{\ol{b_1}, \ldots, \ol{b_n}} \mid b_i \in \{ x_0, x_1 \} \big\},
    \\[.5em]
    \overline U 
    & = 
    \left\{ \ol{u_1}, \ldots, \ol{u_m} \right\},
    \\[.5em]
    \ol{f}(\tuple{\ol{b_1}, \ldots, \ol{b_n}}, \ol u) 
    & =
    \ol{f(\tuple{b_1, \ldots, b_n}, u)},
    \qquad \forall\, b_1, \ldots, b_n \in \{x_0,x_1\},\; \forall\, u \in U.
  \end{align*}
  Then 
  $\mathcal{M}(\mathcal{C}_\eta(D)) = \tuple{\ol{X}^n, \ol U, \ol{f}, \ol{X}^n \times \ol U,
  \mathrm{id}}$.

  To show that
  $\mceta{D}$ is a realisation of $\mealy{A}$,
  it suffices to show an assignment $\ptuple{\alpha, \iota,\zeta}$ 
  of $\mceta{D}$ into $\mealy{A}$,
  satisfying the two conditions~(I) and~(II) below for every $q\in Q$
  and every $\sigma \in \Sigma$:
  \begin{align*}
    \text{(I)}\quad  & \ol{f} \big(\alpha(q),\iota(\sigma)\big) = \alpha\big(\delta(q,\sigma)\big),
    \\
    \text{(II)}\quad & \zeta\big(\alpha(q),\iota(\sigma)\big) = \tuple{q,\sigma}.
  \end{align*}
  Let us define the candidate assignment $\ptuple{\alpha,\iota,\zeta}$ 
  where 
  \begin{align*}
    \alpha\left(q_i\right) 
    & \defeq 
    \tuple{\ol{b_1}, \ldots, \ol{b_n}}
    \text{ with }  
    \ol{b_j} = \ol{x_1} \text{ iff } j = i,
    \\[.5em]
    \iota(\sigma_i) 
    & \defeq 
    \ol{u_i},
    \\[.5em]
    \zeta\left(\tuple{\ol{b_1}, \ldots, \ol{b_n}}, \ol{u_i} \right) 
    & \defeq 
    \begin{cases}
      \tuple{\alpha^{-1}(\ol{b_1}, \ldots, \ol{b_n}), \iota^{-1}(\ol{u_i})} & \text{if }
      \tuple{\ol{b_1}, \ldots, \ol{b_n}} \in \im(\alpha),
      \\[.3em]
      \tuple{q_1, \sigma_1} & \text{otherwise}.
    \end{cases}
  \end{align*}
  Note that the inverse function $\alpha^{-1}$ is well-defined over the image $\im(\alpha)$ since
  $\alpha$ is injective (intuitively, each tuple 
  $\tuple{\ol{b_1}, \ldots, \ol{b_n}}$ is the one-hot encoding of the index of a state $q_i$, 
  with $\ol{x_1}$ standing for one and $\ol{x_0}$ standing for zero).
  Furthermore, the inverse function $\iota^{-1}$ is well-defined for all $\ol{u_i} \in \ol{U}$
  since $\iota$ is a bijection.

  Let $q_i \in Q$, let $\tuple{\ol{b_1}, \ldots, \ol{b_n}} = \alpha(q_i)$, and let $\sigma_j \in \Sigma$.

  We have that property~(II) holds immediately, since by definition of $\zeta$ we have
  $\zeta\big(\alpha(q_i),\iota(\sigma_j)\big) =
  \zeta\big(\alpha^{-1}(\alpha(q_i)),\iota^{-1}(\iota(\sigma_j))\big) =
  \zeta\big(q_i,\sigma_j\big)$.

  Next we show (I).

  Note that
  $\ol{b_i} = \ol{x_1}$ and $\ol{b_\ell} = \ol{x_0}$ for every $\ell \neq i$.
  Let $q_k$ be the next state of $A$, namely
  $q_k = \delta(q_i,\sigma_j)$.
  Thus $\alpha(\delta(q_i,\sigma)) = \tuple{\ol{b'_1}, \ldots, \ol{b'_n}}$
  with 
  $\ol{b'_k} = \ol{x_1}$, and $\ol{b'_\ell} = \ol{x_0}$ for every $\ell \neq k$.
  To show~(I)
  it suffices to show
  $\ol{f}(\tuple{\ol{b_1}, \ldots, \ol{b_n}},\ol{u_j}) = \tuple{\ol{b'_1}, \ldots, \ol{b'_n}}$.
  In turn, 
  letting $\ell \in \iival{1}{n}$,
  it suffices to show
  $(\ol{f}(\tuple{\ol{b_1}, \ldots, \ol{b_n}},\ol{u_j}))_\ell = \ol{x_1}$
  if and only if $\ell = k$.

  First,
  for every $p \in \iival{1}{n}$,
  we have
  \begin{align*}
    \big(R(u_j) \big)_{\ell,p}
    \;=\;
    x_0 + \Delta \cdot T_j
    \;=\;
    \begin{cases}
      x_0 + (x_1- x_0) & \text{if } \delta(q_p,\sigma_j) = q_\ell,
      \\
      x_0  & \text{otherwise,}
    \end{cases}
    \;=\;
    \begin{cases}
      x_1 & \text{if } \delta(q_p,\sigma_j) = q_\ell,
      \\
      x_0  & \text{otherwise.}
    \end{cases}
  \end{align*}
  Then, 
  \begin{align*}
    \big(R(u_j) \otimes \ptuple{b_1, \ldots, b_n} \big)_\ell
    & =
    \bigoplus_{p=1}^n \Big(\big(R(u_j)\big)_{\ell,p} \odot b_p\Big)
    \\
    & =
    \Big(\big(R(u_j)\big)_{\ell,i} \odot x_1 \Big) 
    \oplus \bigoplus_{p\neq i} \Big(\big(R(u_j)\big)_{\ell,p} \odot x_0 \Big)
    \\
    & =
    \Big(\big(R(u_j)\big)_{\ell,i} \odot x_1 \Big) 
    \oplus  
    x_0
    \\
    & =
    \Big(\big(R(u_j)\big)_{\ell,i} \odot x_1 \Big) 
    \\
    & =
    \begin{cases}
      x_1 \odot x_1 & \text{if } \ell = k,
      \\
      x_0 \odot x_1 & \text{otherwise},
    \end{cases}
    \\
    & =
    \begin{cases}
      x_1  & \text{if } \ell = k,
      \\
      x_0  & \text{otherwise}.
    \end{cases}
  \end{align*}
  Then,
  \begin{align*}
    \big(f (\ptuple{b_1, \ldots, b_n}, u_j) \big)_\ell
    & =
    \big( f (\ptuple{b_1, \ldots, b_n}, u_j )\big)_\ell
    \\
    & =
    \big((R(u_j) \otimes \ptuple{b_1, \ldots, b_n}) \big)_\ell \oplus \big(s(u_j)\big)_\ell
    \\
    & =
    \big((R(u_j) \otimes \ptuple{b_1, \ldots, b_n}) \big)_\ell \oplus x_0
    \\
    & =
    \big(R(u_j) \otimes \ptuple{b_1, \ldots, b_n} \big)_\ell
    \\
    & =
    \begin{cases}
      x_1  & \text{if } \ell = k,
      \\
      x_0  & \text{otherwise}.
    \end{cases}
  \end{align*}
  Therefore,
  \begin{align*}
    \big(\ol f (\ptuple{\ol{b_1}, \ldots, \ol{b_n}}, \ol{u_j}) \big)_\ell
    =
    \big(\ol{ f (\ptuple{b_1, \ldots, b_n}, u_j)} \big)_\ell
    =
    \begin{cases}
      \ol{x_1}  & \text{if } \ell = k,
      \\
      \ol{x_0}  & \text{otherwise},
    \end{cases}
  \end{align*}
  as required.

  \paragraph{Strong equivalence of approximations.}
  For arbitrary $x \in X^n$ and $u \in U$, it suffices to show
  $\ol f (\ol x, \ol u) = \ol{\hat{f}} (\ol x, \ol u)$.
  First we show 
  $\ol R(u) = \ol{\hat{R}}(u)$ and $\ol s(u) = \ol{\hat{s}}(u)$.
  For every $\ell, p \in \iival{1}{n}$,
  we have $(\ol{R(u)})_{\ell,p} = (\ol{\hat{R}(u)})_{\ell,p}$
  since $(R(u))_{\ell,p} \in \{x_0,x_1\}$, the intervals $[x_0-\epsilon,x_0+\epsilon]$ and
  $[x_1-\epsilon,x_1+\epsilon]$ are path-connected components, and 
  $(R(u))_{\ell,p} = x_i$ implies 
  $(\hat{R}(u))_{\ell,p} \in [x_i-\epsilon,x_i+\epsilon]$.
  Similarly, 
  for every $\ell \in \iival{1}{n}$,
  we have $(\ol{s(u)})_{\ell} = (\ol{\hat{s}(u)})_{\ell}$
  since $(s(u))_{\ell} = x_0$, the interval $[x_0-\epsilon,x_0+\epsilon]$ is a $\eta$-component,
  and $(s(u))_{\ell} = x_0$ implies 
  $(\hat{s}(u))_{\ell} \in [x_0-\epsilon,x_0+\epsilon]$.
  Then, recalling the notation for MinMax recurrence (Definition~\ref{def:minmax-recurrrence}),
  $\ol f (\ol x, \ol u) = \ol{f(x,u)} = \ol{\minmax(x, R(u), s(u))}$
  and
  $\ol{\hat{f}} (\ol x, \ol u) = \ol{\hat{f}(x,u)} = \ol{\minmax(x, \hat{R}(u), \hat{s}(u))}$.
  Having $\ol R(u) = \ol{\hat{R}}(u)$ and $\ol s(u) = \ol{\hat{s}}(u)$, 
  we obtain
  $\ol{\minmax(x, \hat{R}(u), \hat{s}(u))} =  \ol{\minmax(x, R(u), s(u))}$
  by continuity of the $\minmax$ function, since the continuous image of path-connected components
  is path-connected.
  Therefore,
  $\ol f (\ol x, \ol u) = \ol{\hat{f}} (\ol x, \ol u)$ as required.
\end{proof}

The next lemma is the full version of Theorem~\ref{theorem:expressivity-main-ir} from
Section~\ref{sec:mainbody-semiautomata}, except that we postpone replacing the input functions with
MLPs.

\begin{restatable}{lemma}{lemmaexpressivitymainir}
  \label{lemma:expressivity-main-ir}
  Let $A$ be an identity-reset semiautomaton having $n$ states and $m$ inputs,
  let $\epsilon \geq 0$,
  let $x_1, \ldots, x_n \in \reals$ with $|x_i - x_j| > 2\epsilon$ for all $i\neq j$,
  and 
  let $U \subseteq \reals$ be the union of $m$ pair-wise disjoint closed intervals.
  There exist continuous functions 
  $R,s: U \to \{ x_1, \ldots,x_n \}$ such that
  semiautomaton $A$ can be realised by the \etafinite{} MinMax Recurrent Unit 
  $D = \tuple{X, U, f \mid R, s}$ where 
  $X = \bigcup_{i=1}^n [x_i-\epsilon,x_i+\epsilon]$;
  furthermore, for every $\epsilon$-approximations $(\hat{R},\hat{s})$ of $(R,s)$,
  the MinMax Recurrent Unit
  $\hat{D} = \tuple{X, U, f \mid \hat{R}, \hat{s}}$ is strongly equivalent to $D$.
\end{restatable}
\begin{proof}
  Let $A = \tuple{Q, \Sigma, \delta}$ be an identity-reset semiautomaton with 
  states $Q = \{ q_1, \ldots, q_n \}$
  and 
  inputs $\Sigma = \{ \sigma_1, \ldots, \sigma_m \}$.

  \paragraph{Construction.}
  We define the MinMax Recurrent Unit $D = \tuple{X, U, f \mid R, s}$.
  Let $U_1, \ldots, U_m$ be the pair-wise disjoint closed intervals such that 
  $U = \bigcup_{i=1}^m U_i$.
  Let $x_\mathrm{max} \defeq \max\{ x_1, \ldots, x_n \}$,
  and
  let $x_\mathrm{min} \defeq \min\{ x_1, \ldots, x_n \}$.

  Next we define the input functions $R: U \to \{ x_1, \ldots, x_n \}$ and $s: U \to \{ x_1, \ldots, x_n \}$.
  For $u \in U_i$, letting $\tau_i(q) \defeq \delta(q,\sigma_i)$ be the state transformation induced
  by $\sigma_i$,
  we define
  \begin{align*}
    R(u) & \defeq 
    \begin{cases}
      x_\mathrm{min} & \text{ if } \tau_i \text{ is identity},
      \\
      x_\mathrm{max} & \text{ if } \tau_i \text{ is constant},
    \end{cases}
    \\[.5em]
    s(u) & \defeq 
    \begin{cases}
      x_\mathrm{min} & \text{ if } \tau_i \text{ is identity},
      \\
      x_j & \text{ if } \tau_i \text{ is constant with value } q_j.
    \end{cases}
  \end{align*}
  Note that the functions $R$ and $s$ are continuous by
  Lemma~\ref{lemma:mat-eta-constant-continuous} since they are constant on
  each $\eta$-component $U_i$.

  This completes the construction of the MinMax Recurrent Unit. Next we prove the claims.

  \paragraph{$\eta$-finiteness.}
  The MinMax Recurrent Unit $D$ is $\eta$-finite since $U$ and $X$ are finite unions of closed
  intervals.

  \paragraph{Realisation.}
  To show that $D$ can realise $A$, it suffices to show that
  $\mathcal{M}(\mathcal{C}_\eta(D))$ is a realisation of $\mathcal{M}(A)$.
  Note that $\mathcal{M}(A) = \tuple{Q, \Sigma, \delta, \Gamma, \theta}$
  where $\Gamma = Q \times \Sigma$ and $\theta = \mathrm{id}$.

  For each $i \in \iival{1}{m}$,
  let us choose an arbitrary $u_i \in \ol{U_i}$.
  The construction of $D$ allows us to easily derive that its canonical
  semiautomaton has the form
  $\mathcal{C}_\eta(D) = \tuple{\ol{X}, \ol U, \ol{f}}$
  where
  \begin{align*}
    \ol{X} 
    & =
    \{ \ol x_1, \ldots, \ol x_n \},
    \\[.5em]
    \overline U 
    & = 
    \left\{ \ol u_1, \ldots, \ol u_m \right\},
    \\[.5em]
    \ol{f}(\ol x, \ol u) 
    & =
    \ol{f(x, u)},
    && \forall\, x \in \{x_1,\ldots, x_n\},\; \forall\, u \in U.
  \end{align*}
  Then 
  $\mathcal{M}(\mathcal{C}_\eta(D)) = \tuple{\ol{X}, \ol U, \ol{f}, \ol{X} \times \ol U,
  \mathrm{id}}$.

  To show that
  $\mceta{D}$ is a realisation of $\mealy{A}$,
  it suffices to show an assignment $\ptuple{\alpha, \iota,\zeta}$ 
  of $\mceta{A}$ into $\mealy{A}$,
  satisfying the two conditions (I) and (II) below for every $q_i\in Q$
  and every $\sigma_j \in \Sigma$.
  \begin{align*}
    \text{(I)}\quad  & \ol{f}\big(\alpha(q_i),\iota(\sigma_j)\big) = \alpha\big(\delta(q_i,\sigma_j)\big)\\
    \text{(II)}\quad & \zeta\big(\alpha(q_i),\iota(\sigma_j)\big) = \theta\big(q_i,\sigma_j \big)
  \end{align*}
  Let us define the candidate assignment $\ptuple{\alpha, \iota,\zeta}$ 
  where 
  \begin{align*}
    \alpha(q_i) & \defeq \ol{x_i},
    \\[.3em]
    \iota(\sigma_j) & \defeq \ol{u_j},
    \\[.3em]
    \zeta(\ol{x_i}, \ol{u_j}) & \defeq \tuple{q_i, \sigma_j}.
  \end{align*}

  Let $q_i\in Q$, let $\sigma_j \in \Sigma$, and let us note that $\alpha(q_i) = \ol{x_i}$ and
  $\iota(\sigma_j) = \ol{u_j}$.
  We have that condition~(II) holds by definition of $\zeta$.

  Next we show (I).
  We have
  \begin{align*}
  \ol{f}\big(\alpha(q_i),\iota(\sigma_j)\big)
  =
  \ol{f}\big(\ol{x_i},\ol{u_j}\big)
  =
  \ol{f\big(x_i,u_j\big)}
  \end{align*}

  Let $\tau(q) \defeq \delta(q, \sigma_j)$.
  Let $q_k$ be the next state of $A$, namely
  $q_k = \tau(q_i)$.
  As $\alpha(\delta(q_i,\sigma_j)) = x_k$,
  to show (I), 
  it suffices to show $f(x_i,u_j) = x_k$.

  We have that $\tau$ is a reset-identity transformation, and hence it
  is either identity or constant.

  Say that $\tau$ is identity, hence $q_k = q_i$,
  and hence we need to show $f(x_i, u_j) = x_i$.
  We have
  \begin{align*}
    f(x_i, u_j)
    & = (x_i \odot x_\mathrm{max}) \oplus  x_\mathrm{min} 
    && \text{\small by def.\ of $R(u_j)$},
    \\
    & = x_i \oplus  x_\mathrm{min} 
    && \text{\small since $x_i \leq x_\mathrm{max}$ by construction of $x_\mathrm{max}$},
    \\
    & = x_i
    && \text{\small since $x_i \geq x_\mathrm{min}$ by construction of $x_\mathrm{min}$},
  \end{align*}
  which proves the case when $\tau$ is identity.

  Now say that $\tau$ is constant with value $q_k$,
  and hence it suffices to show $f(x_i,u_j) = x_k$.
  We have
  \begin{align*}
    f(x_i, u_j)
    & = (x_i \odot x_\mathrm{min}) \oplus  x_k
    && \text{\small by def.\ of $R(u_j)$},
    \\
    & = x_\mathrm{min} \oplus  x_k
    && \text{\small since $x_i \geq x_\mathrm{min}$ by construction of $x_\mathrm{min}$},
    \\
    & = x_k
    && \text{\small since $x_k \leq x_\mathrm{max}$ by construction of $x_\mathrm{max}$},
  \end{align*}
  which proves the case when $\tau$ is constant.

  This concludes the proof of (I), and hence the proof that 
  the MinMax Recurrent Unit can realise $A$.

  \paragraph{Strong equivalence of approximations.} 
  Omitted since it is analogous to the one of Lemma~\ref{lemma:expressivity-main-basic}.
\end{proof}

The next lemma is the full version of Theorem~\ref{theorem:expressivity-main-perm} from
Section~\ref{sec:mainbody-semiautomata}, except that we postpone replacing the input functions with
MLPs.

\begin{restatable}{lemma}{lemmaexpressivitymainperm}
  \label{lemma:expressivity-main-perm}
  Let $A$ be a permutation semiautomaton having $n$ inputs and permutation degree $d$.
  Let $\epsilon \geq 0$,
  let $x_1, \ldots, x_d \in \reals$ with $|x_i - x_j| > 2\epsilon$ for all $i\neq j$,
  and 
  let $U \subseteq \reals$ be the union of $n$ pair-wise disjoint closed intervals.
  There exist continuous functions 
  $R: U \to \{ x_1,\ldots, x_d \}^{d \times d}$ and $s: U \to \{ x_1,\dots,x_d \}^d$ such that
  semiautomaton $A$ can be realised by the \etafinite{} MinMax Recurrent Unit 
  $D = \tuple{X^d, U, f \mid R, s}$ where
  $X = \bigcup_{i=1}^d [x_i-\epsilon,x_i+\epsilon]$;
  furthermore, for every $\epsilon$-approximations $(\hat{R},\hat{s})$ of $(R,s)$,
  the MinMax Recurrent Unit
  $\hat{D} = \tuple{X^d, U, f \mid \hat{R}, \hat{s}}$ is strongly equivalent to $D$.
\end{restatable}
\begin{proof}
  Since $A$ has permutation degree $d$,
  by Theorem~\ref{cor:realisation-via-minimal-degree},
  there exists $\varphi: \Sigma \to \gsym{\iival{1}{d}}$
  such that
  the semiautomaton $A' = \tuple{\iival{1}{d}^d, \Sigma, \delta'}$
  with $\delta'(\tuple{q_1, \ldots, q_d}, \sigma) = \tuple{q_{\pi(1)}, \ldots, q_{\pi(n)}}$
  for $\pi: \iival{1}{d} \to \iival{1}{d}$ the permutation given by $\pi = \varphi(\sigma)$.
  Let $U_1, \ldots, U_n$ be the pair-wise disjoint closed intervals such that 
  $U = \bigcup_{i=1}^n U_i$.
  Let us enumerate the elements of $\Sigma$ as $\sigma_1, \ldots, \sigma_n$.
  Let $\pi_i$ be the permutation given by $\pi_i = \varphi(\sigma_i)$.

  \paragraph{Construction.}
  We define the MinMax Recurrent Unit $D = \tuple{X, U, f \mid R, s}$.
  Let $x_\mathrm{max} = \max \{x_1, \ldots, x_d \}$,
  let $x_\mathrm{min} = \min \{x_1, \ldots, x_d \}$,
  and
  let $x_\Delta = x_\mathrm{max} - x_\mathrm{min}$.

  For each $i \in \iival{1}{n}$,
  let $P_i \in \{ 0,1 \}^{n \times n}$ be the (permutation) matrix where $(P_i)_{j,k} = 1$
  iff $\pi_i(k) = j$,
  and
  let $a_i : U \to \{ 0, x_\Delta \}$ be the function defined as $a_i(u) = x_\Delta$ iff 
  $u \in U_i$.
  Then,
  we define the functions $R: U \to \{x_1, \ldots, x_d\}^{d \times d}$ and 
  $s: U \to \{x_1, \ldots, x_d\}^d$ as
  \begin{align*}
    R(u) \defeq x_\mathrm{min} \cdot \mathbf{1}_{d,d} + \sum_{i=1}^n a_i(u) \cdot P_i 
    \quad\qquad
    s(u) \defeq x_\mathrm{min} \cdot \mathbf{1}_d.
  \end{align*}
  Note that each function $a_i$ is continuous since it is constant over $\eta$-components of $U$,
  and hence $R$ is continuous since it is a finite composition of continuous functions.
  Also $s$ is continuous since it is constant.
  This completes the construction of the MinMax Recurrent Unit. Next we prove the claims.

  \paragraph{$\eta$-finiteness.}
  The MinMax Recurrent Unit $D$ is $\eta$-finite since $U$ is a finite union of closed intervals, 
  and $X$ is the cross-product of a finite union of closed intervals.

  \paragraph{Realisation.}
  To show that $D$ can realise $A$, it suffices to show that
  $D$ can realise $A'$ since $A'$ realises $A$.
  In turn, 
  it suffices to show
  that
  $\mceta{D}$ is a realisation of $\mealy{A'}$. 
  Note that $\mealy{A'} = \tuple{\iival{1}{d}^d, \Sigma, \delta', \iival{1}{d}^d \times \Sigma, \mathrm{id}}$.

  The construction of $D$ allows us to easily derive that its canonical
  semiautomaton has the form
  $\ceta{D} = \tuple{\ol{X}^d, \ol U, \ol f}$
  where
  \begin{align*}
    \ol{X}^d & = \left\{ \tuple{\ol{x_{i_1}}, \ldots, \ol{x_{i_d}}} \mid i_1, \ldots, i_d \in
    \iival{1}{d} \right\},
    \\[.3em]
    \ol U & = \left\{ \ol{u_i} \mid i \in \iival{1}{n} \right\},
    \\[.3em]
    \ol f (\tuple{\ol{x_{i_1}}, \ldots, \ol{x_{i_d}}}, \ol{u_i}) 
    & = \ol{f (\tuple{x_{i_1}, \ldots, x_{i_d}}, u_i)}
  \end{align*}
  Thus $\mceta{D} = \tuple{\ol{X}^d, \ol U, \ol f, \ol{X}^d \times \ol U, \mathrm{id}}$.

  To show that
  $\mceta{D}$ is a realisation of $\mealy{A'}$,
  it suffices to show an assignment $\ptuple{\alpha, \iota,\zeta}$ 
  of $\mceta{D}$ into $\mealy{A'}$,
  satisfying the two conditions (I) and (II) below for every 
  $\tuple{q_1, \ldots, q_d}\in \iival{1}{d}^d$ and every $\sigma\in \Sigma$, 
  letting $\tuple{\ol{x_{q_1}}, \ldots, \ol{x_{q_d}}} = \alpha(q_1, \ldots, q_d)$,
  \begin{align*}
    \text{(I)}\quad  & \ol{f}\big(\tuple{\ol{x_{q_1}}, \ldots, \ol{x_{q_d}}},\iota(\sigma)\big) = 
    \alpha\big(\delta'(\tuple{q_1, \ldots, q_d},\sigma)\big)\\
    \text{(II)}\quad & \zeta\big(\tuple{\ol{x_{q_1}}, \ldots, \ol{x_{q_d}}},\iota(\sigma)\big)=
    \tuple{\tuple{q_1, \ldots, q_d},\sigma}
  \end{align*}
  Let us define the candidate assignment $\ptuple{\alpha, \iota,\zeta}$ 
  where 
  \begin{align*}
    \alpha\left(\tuple{q_1, \ldots, q_d}\right) 
    & \defeq 
    \tuple{\ol{x_{q_1}}, \ldots, \ol{x_{q_1}}},
    \\[.3em]
    \iota(\sigma_i) & \defeq \ol{u_i}, 
    \\[.3em]
    \zeta\left(\tuple{\ol{x_{q_1}}, \ldots, \ol{x_{q_d}}}, \ol{u_i} \right) 
    & \defeq 
    \tuple{\tuple{q_1, \ldots, q_d}, \sigma_i}.
  \end{align*}

  Let $\tuple{q_1, \ldots, q_d}\in \iival{1}{d}^d$ and let $\sigma_i \in \Sigma$.
  Note $\alpha(q_1, \ldots, q_d) = \tuple{\ol{x_{q_1}}, \ldots, \ol{x_{q_d}}}$
  and
  $\iota(\sigma_i) = \ol{u_i}$.

  Condition~(II) holds immediately by construction.
  Next we show condition~(I).

  We have 
  \begin{equation*}
    \delta'(\tuple{q_1, \ldots, q_d}, \sigma_i) = \tuple{q_{\pi_i(1)}, \ldots, q_{\pi_i(d)}},
  \end{equation*}
  and hence
  letting $j \in \iival{1}{d}$ and $j' = \pi_i(j)$,
  by the definition of $\alpha$ it suffices to show 
  \begin{equation*}
    \Big(\ol{f}\big(\tuple{\ol{x_{q_1}}, \ldots, \ol{x_{q_d}}},\ol{u_i}\big) \Big)_j
    =
    \ol{x_{q_{j'}}}.
  \end{equation*}

  First, we have 
  \begin{align*}
    R(u_i) & = x_\mathrm{min} \cdot \mathbf{1}_{n,n} + x_\Delta \cdot P_i,
    \\
    s(u_i) & = x_\mathrm{min} \cdot \mathbf{1}_n,
  \end{align*}
  and hence,
  for every $\ell \in \iival{1}{d}$,
  we have
  \begin{align*}
    \big(R(u_i)\big)_{j,\ell} & = x_\mathrm{min} + x_\Delta \cdot \big(P_i\big)_{j,\ell} 
    = 
    \begin{cases}
      x_\mathrm{max} & \text{ if } \pi_i(j) = \ell,
      \\
      x_\mathrm{min} & \text{ otherwise,}
    \end{cases}
    = 
    \begin{cases}
      x_\mathrm{max} & \text{ if } \ell = j',
      \\
      x_\mathrm{min} & \text{ otherwise,}
    \end{cases}
    \\
    \big(s(u_i)\big)_j  & = x_\mathrm{min}.
  \end{align*}
  Thus, 
  \begin{align*}
    \Big(f\big(\tuple{x_{q_1}, \ldots, x_{q_d}},u_i \big) \Big)_j
    & = 
    \left(\bigoplus_{\ell=1}^d \big(R(u_i)\big)_{j,\ell} \odot x_{q_\ell}\right) \oplus \big(s(u_i)\big)_j
    \\[.3em]
    & = 
    (x_\mathrm{max} \odot x_{q_{j'}}) \oplus \left(\bigoplus_{\ell \neq j'} x_\mathrm{min} \odot
    x_{q_\ell}\right) \oplus \big(s(u_i)\big)_j
    \\[.3em]
    & = 
    x_{q_{j'}} \oplus \left(\bigoplus_{\ell \neq j'} x_\mathrm{min} \right) \oplus \big(s(u_i)\big)_j
    \\[.3em]
    & = 
    x_{q_{j'}} \oplus x_\mathrm{min} \oplus \big(s(u_i)\big)_j
    \\[.3em]
    & = 
    x_{q_{j'}} \oplus x_\mathrm{min} \oplus x_\mathrm{min}
    \\[.3em]
    & = 
    x_{q_{j'}},
  \end{align*}
  and hence
  \begin{align*}
    \Big(\ol{f}\big(\tuple{\ol{x_{q_1}}, \ldots, \ol{x_{q_d}}},\ol{u_i}\big) \Big)_j
    & = 
    \Big(\ol{f\big(\tuple{\ol{x_{q_1}}, \ldots, \ol{x_{q_d}}},\ol{u_i}\big)} \Big)_j
    = 
    \ol{x_{q_{j'}}},
  \end{align*}
  as required.
  This proves that $(\alpha,\iota,\zeta)$ is an assignment of $\mealy{A'}$ into $\mceta{D}$,
  and hence overall we have proved that $D$ can realise $A$, by the arguments put forward in the
  course of the proof.

  \paragraph{Strong equivalence of approximations.}
  Omitted since it is analogous to the one of Lemma~\ref{lemma:expressivity-main-basic}.
\end{proof}

\subsection{Realising Cascades}
\label{app:realising-casc}

This section develops part of the proof of Theorem~\ref{thm:main-body-expressivity}.
As discussed in the proof sketch of the theorem, the expressivity results are obtained via realising
cascades of semiautomata. 
In this section we show that the MinMax Units obtained in the previous section can be composed into
cascade to realise cascades of semiautomata.

\begin{theorem} \label{th:expressivity-cascade-real}
  Let us consider a cascade
  \begin{align*}
    C & = 
    \big( A_{1,1} \parallelop \cdots \parallelop A_{1,m_1} \big)
    \ltimes \cdots \ltimes
    \big( A_{d,1} \parallelop \cdots \parallelop A_{d,m_d} \big),
  \end{align*}
  where each $A_{i,j}$ is a semiautomaton.
  We have that $C$ can be realised by the following \etafinite{} MinMax Recurrent Cascade
  \begin{align*}
    C' & = 
    \big( D_{1,1} \parallelop \cdots \parallelop D_{1,m_1} \big)
    \ltimes \cdots \ltimes
    \big( D_{d,1} \parallelop \cdots \parallelop D_{d,m_d} \big),
  \end{align*}
  where the input functions of each unit $D_{i,j}$ are MLPs, 
  and the state dimension of each unit $D_{i,j}$ is
  \begin{inlineenum}
  \item
    one if $A_{i,j}$ is identity-reset,
  \item
    the degree of $A_{i,j}$ if $A_{i,j}$ is a permutation semiautomaton,
  \item
    the number of states of $A_{i,j}$ otherwise.
  \end{inlineenum}
\end{theorem}
\begin{proof}
  We have that each $A_{i,j}$ can be realised by an \etafinite{} MinMax Recurrent Unit 
  $D'_{i,j}$ with MLPs as input functions and state dimension as required. In fact,
  \begin{inlineenum}
  \item
    when $A_{i,j}$ is identity-reset, we can apply Theorem~\ref{lemma:expressivity-main-ir};
  \item
    when $A_{i,j}$ is permutation, we can apply Theorem~\ref{lemma:expressivity-main-perm};
  \item
    we can always apply Theorem~\ref{lemma:expressivity-main-basic}.
  \end{inlineenum}

  Then, by Lemma~\ref{lemma:realize-cascade},
  we have that $C$ can be realised by
  a cascade
  \begin{align*}
    C' & = 
    \big( 
      \phi_{1,1} \circ D'_{1,1} 
      \parallelop \cdots \parallelop 
      \phi_{1,m_1} \circ D'_{1,m_1} 
    \big) 
    \ltimes \cdots \ltimes
    \big( 
      \phi_{d,1} \circ D'_{d,1} 
      \parallelop \cdots \parallelop 
      \phi_{d,m_d} \circ D'_{d,m_d} 
    \big),
  \end{align*}
  where each $\phi_{i,j}$ is a continuous function, and each $\phi_{1,j}$ is over a compact subset
  $\hat{U}$ of $\reals$; and we note that each $\phi_{i,j}$ with $i > 1$ is also over a compact
  real-valued domain, as its domain is given by the cross-product of $U$ and $\eta$-finite state
  spaces, which are compact by definition.
  Then, by Proposition~\ref{prop:composition-approximation} and by the Universal Approximation
  Theorem for MLPs,
  we have that each $\phi_{i,j} \circ D'_{i,j}$ is strongly equivalent to 
  $\psi_{i,j} \circ D'_{i,j}$ for some MLP $\psi_{i,j}$.
  Thus, the following cascade 
  \begin{align*}
    \hat{C}' & = 
    \big( 
      \psi_{1,1} \circ D'_{1,1} 
      \parallelop \cdots \parallelop 
      \psi_{1,m_1} \circ D'_{1,m_1} 
    \big) 
    \ltimes \cdots \ltimes
    \big( 
      \psi_{d,1} \circ D'_{d,1} 
      \parallelop \cdots \parallelop 
      \psi_{d,m_d} \circ D'_{d,m_d} 
    \big)
  \end{align*}
  is strongly equivalent to $C'$; namely, $\ceta{C'} = \ceta{\hat{C}'}$. Hence, $\hat{C}'$ can
  realise $C$, since $C'$ can realise $C$. 
  Then, we note that each $\hat{D}'_{i,j} \defeq \psi_{i,j} \circ D'_{i,j}$ is a MinMax Recurrent
  Unit, since $D'_{i,j}$ is so, and composing it with $\psi_{i,j}$ amounts to composing the
  input functions of $D'_{i,j}$ with $\psi_{i,j}$.
  Therefore, $\hat{C}'$ is a MinMax RNC and the given cascade $C$ can be realised by $\hat{C}'$.
\end{proof}

The following theorem regards floating-point MinMax RNCs, that are defined in
Section~\ref{sec:float-minmaxrncs}.

\begin{theorem} \label{th:expressivity-cascade-float}
  Let us consider a cascade
  \begin{align*}
    C & = 
    \big( A_{1,1} \parallelop \cdots \parallelop A_{1,m_1} \big)
    \ltimes \cdots \ltimes
    \big( A_{d,1} \parallelop \cdots \parallelop A_{d,m_d} \big),
  \end{align*}
  where each $A_{i,j}$ is a semiautomaton.
  We have that $C$ can be realised by a floating-point MinMax Recurrent Cascade over some 
  $\mathbb{F}$, having the form
  \begin{align*}
    C' & = 
    \big( D_{1,1} \parallelop \cdots \parallelop D_{1,m_1} \big)
    \ltimes \cdots \ltimes
    \big( D_{d,1} \parallelop \cdots \parallelop D_{d,m_d} \big),
  \end{align*}
  where the input functions of each unit $D_{i,j}$ are $\mathbb{F}$-MLPs,
  and the state dimension of each unit $D_{i,j}$ is
  \begin{inlineenum}
  \item
    one if $A_{i,j}$ is identity-reset,
  \item
    the degree of $A_{i,j}$ if $A_{i,j}$ is a permutation semiautomaton,
  \item
    the number of states of $A_{i,j}$ otherwise.
  \end{inlineenum}
\end{theorem}
\begin{proof}
  For each $A_{i,j}$, let us define $m_{i,j}$ as follows:
  \begin{itemize}
    \item
      if $A_{i,j}$ is a permutation semiautomaton, then $m_{i,j}$ is the maximum between the degree
      and the number of inputs of $A_{i,j}$;
    \item
      otherwise, $m_{i,j}$ is the maximum between the number of states and
      the number of inputs of $A_{i,j}$.
  \end{itemize}
  Then,
  let $m$ be the largest $m_{i,j}$,
  and
  let $\mathbb{F}$ be a floating-point domain containing at least $m$ elements.

  We have that each $A_{i,j}$ can be realised by a MinMax Recurrent Unit 
  $D'_{i,j}$ over $\mathbb{F}$ with $\mathbb{F}$-MLPs as input functions and
  state dimension as required. In fact,
  \begin{inlineenum}
  \item
    when $A_{i,j}$ is identity-reset, we apply Theorem~\ref{lemma:expressivity-main-ir};
  \item
    when $A_{i,j}$ is permutation, we apply Theorem~\ref{lemma:expressivity-main-perm};
  \item
    otherwise, we apply Theorem~\ref{lemma:expressivity-main-basic}.
  \end{inlineenum}

  Then, by Lemma~\ref{lemma:realize-cascade},
  we have that $C$ can be realised by
  a cascade
  \begin{align*}
    C' & = 
    \big( 
      \phi_{1,1} \circ D'_{1,1} 
      \parallelop \cdots \parallelop 
      \phi_{1,m_1} \circ D'_{1,m_1} 
    \big) 
    \ltimes \cdots \ltimes
    \big( 
      \phi_{d,1} \circ D'_{d,1} 
      \parallelop \cdots \parallelop 
      \phi_{d,m_d} \circ D'_{d,m_d} 
    \big),
  \end{align*}
  where each $\phi_{1,j}$ has a domain $\hat{U} \subseteq \mathbb{F}^k$ for some $k$,
  and hence each $\phi_{i,j}$ is necessarily a function with $\mathbb{F}$-valued domain and
  codomain, since it maps from the cross product of $\hat{U}$ with state spaces of the Units
  $D'_{i,j}$ to the state space of one of the Units, which are all $\mathbb{F}$-valued.
  Thus, each $D_{i,j} \defeq \phi_{i,j} \circ D'_{i,j}$ is a MinMax Recurrent Unit with input
  functions $\phi_{i,j} \circ R_{i,j}$ and each $\phi_{i,j} \circ s_{i,j}$ over $\mathbb{F}$,
  which can be expressed as $\mathbb{F}$-MLPs by any of Corollary~3.8 or Corollary~3.11
  of~\citep{hwang25floating}.
  Therefore, $C'$ can be equivalently rewritten as
  \begin{align*}
    C' & = 
    \big( 
      D_{1,1} 
      \parallelop \cdots \parallelop 
      D_{1,m_1} 
    \big) 
    \ltimes \cdots \ltimes
    \big( 
      D_{d,1} 
      \parallelop \cdots \parallelop 
      D_{d,m_d} 
    \big),
  \end{align*}
  which is as required.
\end{proof}

\begin{restatable}{theorem}{thexpressivitycascade} \label{th:expressivity-cascade}
  Let us consider a cascade
  \begin{align*}
    C & = 
    \big( A_{1,1} \parallelop \cdots \parallelop A_{1,m_1} \big)
    \ltimes \cdots \ltimes
    \big( A_{d,1} \parallelop \cdots \parallelop A_{d,m_d} \big),
  \end{align*}
  where each $A_{i,j}$ is a semiautomaton.
  We have that $C$ can be realised by an \etafinite{} MinMax Recurrent Cascade, and also by a
  floating-point MinMax Recurrent Cascade over some $\mathbb{F}$, both of the form
  \begin{align*}
    C' & = 
    \big( D_{1,1} \parallelop \cdots \parallelop D_{1,m_1} \big)
    \ltimes \cdots \ltimes
    \big( D_{d,1} \parallelop \cdots \parallelop D_{d,m_d} \big),
  \end{align*}
  where the input functions of each unit $D_{i,j}$ are MLPs and $\mathbb{F}$-MLPs, respectively, 
  and the state dimension of each unit $D_{i,j}$ is
  \begin{inlineenum}
  \item
    one if $A_{i,j}$ is identity-reset,
  \item
    the degree of $A_{i,j}$ if $A_{i,j}$ is a permutation semiautomaton,
  \item
    the number of states of $A_{i,j}$ otherwise.
  \end{inlineenum}
\end{restatable}
\begin{proof}
  It holds immediately by Theorem~\ref{th:expressivity-cascade-real} and 
  Theorem~\ref{th:expressivity-cascade-float}, which focus on the real-valued and float-valued case,
  respectively.
\end{proof}

\subsection{Formal Expressivity (Proof of Theorem~\ref{thm:main-body-expressivity})}
\label{sec:appendix-expressivity-languages}

In this section we conclude the proof of Theorem~\ref{thm:main-body-expressivity} putting together
the results obtained in the previous sections.
We prove it by parts, restating the theorem at the end and referring to all the proven sub-theorems.

The next theorem corresponds to the first claim in Theorem~\ref{thm:main-body-expressivity}.

\begin{restatable}{theorem}{thexpressivityfunctions}\label{thm:main-body-expressivity-1}
  Every regular function $F$ (regular language $L$) having degree $d$ can be implemented
  (recognised) by an $\eta$-finite MinMax RNC and by a floating-point MinMax RNC,
  both having state degree $d$.
\end{restatable}
\begin{proof}
  We first note that the case of regular languages is subsumed by the case of regular functions.
  In fact, for every language $L$ over an alphabet $\Sigma$,
  we have that a system can recognise $L$ if it can implement its 
  characteristic function $\mathbb{I}_L: \Sigma^* \to \{0,1\}$, by definition, and such function is
  a regular function.
  Thus, we proceed with the proof for functions.

  Let $F: \Sigma^* \to \Gamma$ be a regular function having degree $n$.
  let $A_F$ be the canonical automaton of $F$ and let $D_F$ be the semiautomaton of $A_F$.
  By Theorem~\ref{thm:krohn-rhodes},
  semiautomaton $D_F$ can be realised by a cascade
  $$
    C = A_1 \ltimes \cdots \ltimes A_n
  $$ 
  where each $A_i$ is either a two-state identity-reset semiautomaton or a
  $\groupsty{G}$-semiautomaton for $G$ a simple group that is a factor of a subgroup of the
  transition monoid of $D_F$.
  We note that, since $F$ has degree $d$ by assumption, the permutation degree $d_i$ of $A_i$ is at
  most $d$.

  The rest of the proof is split into two parts: in the first part we show the existence of an
  \etafinite{} MinMax RNC witnessing the claim, and in the second part we show the existence of a
  floating-point MinMax RNC witnessing the claim---the arguments in the two parts are similar, but
  we keep the two parts separate for the sake of clarity.

  \paragraph{Part I.}
  By Theorem~\ref{th:expressivity-cascade},
  we have that cascade $C$ can be realised by a cascade
  \begin{equation*}
    C' = D_1 \ltimes \cdots \times D_n
  \end{equation*}
  where each $D_i$ is an \etafinite{} MinMax Recurrent Unit with MLPs as input functions.
  Letting $C' = \tuple{X,U,f}$,
  by Theorem~\ref{thm:realize-dynamics}, we have that $C'$ can be extended into a system 
  $S = \tuple{X,U,f,x_0,X \times U,\mathrm{id}}$ that can implement $F$.
  In particular, system $S$ has an identity output function, which can be seen as a trivial MLP.
  Therefore, $S$ is an \etafinite{} MinMax RNC that can implement $F$, as required.

  \paragraph{Part II.}
  By Theorem~\ref{th:expressivity-cascade-float},
  we have that cascade $C$ can be realised by a cascade
  \begin{equation*}
    C' = D_1 \ltimes \cdots \times D_n
  \end{equation*}
  where each $D_i$ is a floating-point MinMax Recurrent Unit over some floating-point domain 
  $\mathbb{F}$ with $\mathbb{F}$-MLPs as input functions, both having state dimension equal to one
  if $A_i$ is
  identity-reset and equal to the degree $d_i$ of $A_i$ otherwise; we note that $d_i \leq d$.
  Letting $C' = \tuple{X,U,f}$,
  by Theorem~\ref{thm:realize-dynamics}, we have that $C'$ can be extended into a system 
  $S = \tuple{X,U,f,x_0,X \times U,\mathrm{id}}$ that can implement $F$.
  In particular, system $S$ has an identity output function, which can be seen as a trivial
  $\mathbb{F}$-MLP.
  Therefore, $S$ is a floating-point MinMax RNC that can implement $F$.
\end{proof}

Next we show the following intermediate result in order to prove the
second claim of Theorem~\ref{thm:main-body-expressivity}.

\begin{restatable}{lemma}{lemmaalwaysfinite} \label{lemma:always-finite}
  Let $\Sigma$ and $\Gamma$ be alphabets,
  and 
  let $S = \tuple{X, U, f, x^0, Y, g}$ be a MinMax Recurrent Cascade.
  For every function $F: \Sigma^+ \to \Gamma$ that is implemented by $S$ with
  encoder $\mathrm{enc}: \Sigma \to U$ and decoder $\mathrm{dec}: Y \to \Gamma$,
  there exist finite subsets 
  $\tilde{X} \subseteq X$, $\tilde{U} \subseteq U$, $\tilde{Y} \subseteq Y$
  such that the MinMax Recurrent Cascade 
  $\tilde{S} = \tuple{\tilde{X}, \tilde{U}, f, x^0, \tilde{Y}, g}$
  implements $F$ with the same encoder $\mathrm{enc}$ and decoder $\mathrm{dec}$.
\end{restatable}
\begin{proof}
  Let us consider 
  a finite input alphabet $\Sigma$
  and 
  a finite output alphabet $\Gamma$.
  Let $S = \tuple{X, U, f, x^0, Y, g}$ be a MinMax RNC,
  and
  let $F: \Sigma^+ \to \Gamma$ be a function that is implemented by $S$ with encoder 
  $\mathrm{enc}: \Sigma \to U$ and decoder $\mathrm{dec}: Y \to \Gamma$.

  Since $\Sigma$ is finite, the image of $\mathrm{enc}$ is a finite set 
  $\tilde{U} \defeq \{ u_1, \ldots, u_m \} \subseteq U$.

  Let the dynamics of $S$ be of the form
  \begin{align*}
    C = 
    \big( D_{1,1} \parallelop \cdots \parallelop D_{1,m_1} \big) 
    \ltimes \cdots \ltimes 
    \big( D_{d,1} \parallelop \cdots \parallelop D_{d,m_d} \big),
  \end{align*}
  with $D_{i,j} = \tuple{X_{i,j}, U_i, f_{i,j} \mid R_{i,j}, s_{i,j}}$.
  Letting $X_i \defeq X_{i,1} \times \cdots \times X_{i,m_i}$, we have
  that $U_i = U \times X_1 \times \cdots \times X_{i-1}$,
  and the state space of $C$ is $X = X_1 \times \cdots \times X_d$.

  We show there exists a finite $\tilde{X}$ with $\{x^0 \} \subseteq \tilde{X} \subseteq X$ such that,
  for every input string $\sigma_1 \cdots \sigma_\ell \in \Sigma^+$,
  we have that $C(x^0, \mathrm{enc}(\sigma_1 \cdots \sigma_\ell)) \in \tilde{X}$;
  it will follow that the states of $C$ can be restricted to $\tilde{X}$,
  and that $Y$ can be restricted to 
  $\tilde{Y} \defeq \{ g(x,u) \in Y \mid x \in \tilde{X}, u \in \tilde{U} \}$.

  For every $i \in \iival{1}{d}$ and every $j \in \iival{1}{m_i}$,
  let $n_{i,j}$ be the state dimension of $D_{i,j}$,
  and 
  let us define 
  \begin{align}
    \label{eq:always-finite-0}
    \tilde{X}^0_{i,j} 
    & \defeq 
    \Big\{ \big(x^0_{i,j}\big)_p \Bigm| p \in \iival{1}{n_{i,j}} \Big\},
    \\[.5em]
    \label{eq:always-finite-1}
    \tilde{X}_{i,j} 
    & \defeq 
    \tilde{X}^0_{i,j}
    \cup
    \Big\{ \big(R_{i,j}(u)\big)_{p,q}, s_{i,j}(u)\big)_p \Bigm| u \in \big(\tilde{U} \times
    \tilde{\mathbf{X}}_{i-1}\big),\, p,q \in \iival{1}{n_{i,j}} \Big\},
    \\[.5em]
    \label{eq:always-finite-2}
    \tilde{X}_i 
    & \defeq 
    \big(\tilde{X}_{i,1}\big)^{n_{i,1}} \times \cdots \big(\tilde{X}_{i,m_i}\big)^{n_{i,m_i}},
    \\[.5em]
    \label{eq:always-finite-3}
    \tilde{\mathbf{X}}_i & \defeq \tilde{X}_1 \times \cdots \tilde{X}_i,
    \\[.5em]
    \label{eq:always-finite-4}
    \tilde{X} & \defeq \tilde{\mathbf{X}}_d.
  \end{align}

  For an arbitrary input string $\sigma_1, \ldots, \sigma_\ell \in \Sigma^*$,
  its encoding is in $\tilde{U}^*$.

  We proceed by induction on $d$, showing that $\tilde{\mathbf{X}}_d$ includes
  the state of $C$ upon reading any sequence in $\tilde{U}^*$.

  In the base case $d=1$.
  The input elements to the units $D_{1,j}$ are from $\tilde{U}$,
  and 
  the the image of the dynamics function of $D_{1,j}$ consist of $n_{1,j}$-vectors with components
  that are necessarily 
  \begin{inlineenum}
  \item 
    a component $\big(x^0_{1,j}\big)_p$ of its initial state $x^0_{1,j}$,
  \item 
    an entry $\big(R_{1,j}(u)\big)_{p,q}$ of the matrix $R_{1,j}(u)$ returned by its reset
    function $R_{1,j}$ on an input $u \in \tilde{U}$,
  \item 
    a component $\big(s_{1,j}(u)\big)_p$ of the vector $s_{1,j}(u)$ returned by its set
    function $s_{1,j}$ on an input $u \in \tilde{U}$.
  \end{inlineenum}
  The three cases are included in the definition of $\tilde{X}_{i,j}$
  (Eq.~\eqref{eq:always-finite-1}), and hence the set of possible states of $D_{1,j}$ is contained
  in $\big(\tilde{X}_{1,j}\big)^{n_{1,j}}$.
  Then the set of possible states of the cascade (a single layer in this case) is contained in the
  cross-product $\tilde{X}_1$ (Eq.~\eqref{eq:always-finite-2}),
  and we have $\tilde{\mathbf{X}}_1 = \tilde{X}_1$ (Eq.~\eqref{eq:always-finite-3}).

  In the inductive case $d>1$,
  and we assume by induction that 
  $\tilde{\mathbf{X}}_{d-1}$ includes the state of the first $d-1$ layers of $C$ upon reading any
  sequence in $\tilde{U}^*$.
  Hence,
  the input elements to each unit $D_{d,j}$ are contained in the set 
  $\tilde{U} \times \tilde{\mathbf{X}}_{d-1}$.
  The image of the dynamics function of $D_{d,j}$ consist of $n_{d,j}$-vectors with components
  that are necessarily 
  \begin{inlineenum}
  \item 
    a component $\big(x^0_{d,j}\big)_p$ of its initial state $x^0_{d,j}$,
  \item 
    an entry $\big(R_{d,j}(u)\big)_{p,q}$ of the matrix $R_{d,j}(u)$ returned by its reset
    function $R_{d,j}$ on an input $u \in \tilde{U} \times \tilde{\mathbf{X}}_{d-1}$,
  \item 
    a component $\big(s_{d,j}(u)\big)_p$ of the vector $s_{d,j}(u)$ returned by its set
    function $s_{d,j}$ on an input $u \in \tilde{U} \times \tilde{\mathbf{X}}_{d-1}$.
  \end{inlineenum}
  The three cases are included in the definition of $\tilde{X}_{d,j}$
  (Eq.~\eqref{eq:always-finite-1}), and hence the set of possible states of $D_{d,j}$ is contained
  in $\big(\tilde{X}_{d,j}\big)^{n_{d,j}}$.
  Then the set of possible states of layer $d$ of the cascade is contained in the
  cross-product $\tilde{X}_d$ (Eq.~\eqref{eq:always-finite-2}).
  By the inductive hypothesis,
  we obtain that the possible sates of $C$ is contained in the cross-product
  $\tilde{\mathbf{X}}_d = \tilde{X}_1 \times \cdots \times \tilde{X}_d$ (Eq.~\eqref{eq:always-finite-3}).

  Then, by definition of $\tilde{X} = \tilde{\mathbf{X}}_d$ (Eq.~\eqref{eq:always-finite-4}),
  we obtain that $\tilde{X}$ contains the possible sates of $C$.
\end{proof}

The next theorem corresponds to the second claim of Theorem~\ref{thm:main-body-expressivity}.

\begin{restatable}{theorem}{thnotregularcannotimplement}\label{thm:main-body-expressivity-2}
  Every function (language) that is not regular cannot be implemented (recognised) by a MinMax
  Recurrent Cascade.
\end{restatable}
\begin{proof}
  Let $F$ be a function that can be implemented by a MinMax Recurrent Cascade $S$.
  By Lemma~\ref{lemma:always-finite},
  $F$ can also be implemented by a MinMax Recurrent Cascade $S'$ that has a finite input and state
  space, and hence trivially \etafinite{}. 
  It follows that $F$ is regular by Theorem~\ref{thm:canon-implement}.
  The proof applies to languages as well, since recognising a language means implementing its
  characteristic function.
\end{proof}

We proceed towards proving the third claim of Theorem~\ref{thm:main-body-expressivity}.
We employ the notions of $\eta$-convergence and aperiodicity from MAT, 
introduced in Section~\ref{sec:prelim-aperiodicity}.

\begin{lemma} \label{lemma:aperiodic-parallel}
  Let $D = D_1 \parallel \cdots \parallel D_n$ be a parallel composition of \etafinite{} aperiodic
  dynamics.  Then $D$ is aperiodic.
\end{lemma}
\begin{proof}
  We have that each $D_i$ is of the form $D_i = \tuple{X_i, U, f_i}$.
  Then,
  we have that $D$ is of the form $D = \tuple{X_1 \times \cdots \times X_n, U, f}$
  with $f(\tuple{x_1, \ldots, x_n}, u) = \tuple{f_1(x_1, u), \ldots, f_n(x_n, u)}$.

  Consider an input sequence $\mathbf{u} = (u_t)_{t \geq 1}$ that is $\eta$-convergent in $U$.
  For each $i \in \iival{1}{n}$,
  let $x^0_i \in X_i$.
  For each $i \in \iival{1}{n}$,
  we define the state sequence $(x_i,t)_{t \geq 1}$ with 
  $x_{i,t} = D_i(x^0_i, u_1 \cdots u_t)$ the state of $D_n$ on input $u_1 \cdots u_t$
  when started from state $x^0_i$;
  and we define the state sequence $(x_t)_{t \geq 1}$ with 
  $x_t = D(\tuple{x^0_1,\ldots,x^0_n}, u_1 \cdots u_t)$ the state of $D$ on input $u_1 \cdots u_t$
  when started from state $\tuple{x^0_1,\ldots,x^0_n}$.

  By the form of the dynamics function of $D$ noted above,
  we have that the state sequence $(x_t)_{t \geq 1}$ of the given composition $D$ equals the state
  sequence $(\tuple{x_{1,t}, \ldots, x_{n,t}})_{t\geq 1}$.

  Now, for every $i\in \iival{1}{n}$, we have that the sequence $(x_{i,t})_{t \geq 1}$ is
  $\eta$-convergent since $(u_t)_{t \geq 1}$ is $\eta$-convergent by assumption, and $D_i$ is
  aperiodic by assumption.
  Therefore, the sequence $(\tuple{x_{1,t}, \ldots, x_{n,t}})_{t\geq 1}$ is $\eta$-convergent,
  and hence the state sequence $(x_t)_{t \geq 1}$ is $\eta$-convergent, showing that $D$ is
  aperiodic.
\end{proof}

\begin{lemma} \label{lemma:aperiodic-cascade}
  Let $C = D_1 \ltimes \cdots \ltimes D_n$ be a cascade of \etafinite{} aperiodic dynamics.
  Then $C$ is aperiodic.
\end{lemma}
\begin{proof}
  We proceed by induction on $n$.
  In the base case $n=1$ and the claim holds by assumption.
  In the inductive case $n>1$ and we assume by induction that the cascade 
  $C_{n-1} = D_1 \ltimes \cdots \ltimes D_{n-1}$ is aperiodic.
  We have that $C_{n-1}$ and $D_n$ are of the form
  \begin{align*}
    C_{n-1} = \tuple{X_{n-1}, U, f_{n-1}},
    \quad
    D_n     = \tuple{X_n, U \times X_{n-1}, f_n}.
  \end{align*}
  Then, we have that $C$ is of the form
  \begin{align*}
    C = \tuple{X_{n-1} \times X_n, U, f},
    \qquad
    f(\tuple{x_{n-1}, x_n}, u) = \tuple{f_{n-1}(x_{n-1}, u), f_n(x_n, \tuple{u, x_{n-1}})}
  \end{align*}
  Consider an input sequence $\mathbf{u} = (u_t)_{t \geq 1}$ that is $\eta$-convergent in $U$.
  Let $x^0_{n-1} \in X_{n-1}$ and let $x^0_n \in X_n$.
  We define the state sequence $(x_{n-1,t})_{t \geq 1}$ with 
  $x_{n-1,t} = C_{n-1}(x^0_{n-1}, u_1 \cdots u_t)$ the state of $C_{n-1}$ on input $u_1 \cdots u_t$
  when started from state $x^0_{n-1}$.
  Then we define the state sequence $(x_{n,t})_{t \geq 1}$ with 
  $x_{n,t} = D_n(x^0_n, \tuple{u_1,x_{n-1,1}} \cdots \tuple{u_t,x_{n-1,t-1}})$ the state of $D_n$ on
  input $\tuple{u_1,x_{n-1,1}} \cdots \tuple{u_t,x_{n-1,t-1}}$ when started from state $x^0_n$.
  Finally, 
  we define the state sequence $(x_t)_{t \geq 1}$ with 
  $x_t = C(\tuple{x^0_{n-1},x^0_n}, u_1 \cdots u_t)$ the state of $C$ on input $u_1 \cdots u_t$
  when started from state $\tuple{x^0_{n-1},x^0_n}$.

  By the form of the dynamics function of $C$ noted above,
  we have that the state sequence $(x_t)_{t \geq 1}$ of the given cascade $C$ equals the state
  sequence $(\tuple{x_{n-1,t}, x_{n,t}})_{t\geq 1}$.

  Now, we have that the sequence $(x_{n,t})_{t \geq 1}$ is $\eta$-convergent since 
  $(u_t)_{t \geq 1}$ is $\eta$-convergent by assumption, and $C_{n-1}$ is aperiodic by the inductive
  hypothesis.
  In turn, this implies that the state sequence $(x_{n,t})_{t \geq 1}$ is $\eta$-convergent,
  since $D_n$ is aperiodic by assumption.
  Therefore, the sequence $(\tuple{x_{n-1,t}, x_{n,t}})_{t\geq 1}$ is $\eta$-convergent,
  and hence the state sequence $(x_t)_{t \geq 1}$ is $\eta$-convergent, showing that $C$ is
  aperiodic.
\end{proof}

\begin{lemma} \label{lemma:minmax-aperiodic}
  Every finite MinMax Recurrent Unit of state dimension one is aperiodic.
\end{lemma}
\begin{proof}
  Let $D = \tuple{X,U,f \mid R,s}$ be a finite MinMax Recurrent Unit of state dimension one.
  Let $(u_t)_{t \geq 1}$ be an $\eta$-convergent sequence in $U$.
  Since $U$ is finite, its $\eta$-components are singleton, and hence all inputs $u_t$ are the same
  input $u \in U$.
  Let $x \in X$, and let $\tau(x) \defeq f(x,u)$.
  We have 
  \begin{align*}
    \tau(\tau(x)) 
    & = f(f(x, u),u) 
    \\[.3em]
    & = \Big(R(u) \odot \big((R(u) \odot x) \oplus s(u)\big)\Big) \oplus s(u)
    \\[.3em]
    & = \big(R(u) \odot R(u) \odot x) \oplus (R(u) \odot s(u))\big) \oplus s(u)
    \\[.3em]
    & = (R(u) \odot x) \oplus (R(u) \odot s(u)) \oplus s(u)
    \\[.3em]
    & = \big(R(u) \odot (x \oplus s(u))\big) \oplus s(u)
  \end{align*}
  and hence
  \begin{align*}
    \tau(\tau(\tau(x)))
    & = R(u) \odot \Big( \big(R(u) \odot (x \oplus s(u))\big) \oplus s(u)\Big) \oplus s(u)
    \\[.3em]
    & = \Big( \big(R(u) \odot R(u) \odot (x \oplus s(u))\big) \oplus R(u) \odot s(u)\Big) \oplus s(u)
    \\[.3em]
    & = \Big( \big(R(u) \odot (x \oplus s(u))\big) \oplus R(u) \odot s(u)\Big) \oplus s(u)
    \\[.3em]
    & = \big(R(u) \odot (x \oplus s(u) \oplus s(u))\big) \oplus s(u)
    \\[.3em]
    & = \big(R(u) \odot (x \oplus s(u))\big) \oplus s(u)
    \\[.3em]
    & = \tau(\tau(x)).
  \end{align*}
  It follows that $\tau^n = \tau^{n+1}$ for all $n \geq 2$.

  Let us consider the state sequence $(x_t)_{t \geq 1}$ with $x_t = D(x, u_1 \cdots u_t)$ the state
  of $D$ on input $u_1\cdots u_t$ when started from state $x$.
  Since every $u_t$ equals $u$,
  we have that $x_t = \tau^t(x)$.
  Since $\tau^t = \tau^{t+1}$ for all $t \geq 2$,
  we have that $x_t = \tau^2(x)$ for all $t \geq 2$, and hence the sequence $(x_t)_{t \geq 1}$ is
  $\eta$-convergent, showing that $D$ is aperiodic.
\end{proof}

The following theorem corresponds to the third claim of Theorem~\ref{thm:main-body-expressivity}.

\begin{theorem} \label{thm:main-body-expressivity-3}
  Let $S$ be a MinMax RNC of state degree one.
  Every regular function $F$ that is not group-free cannot be implemented by $S$.
\end{theorem}
\begin{proof}
  Let $F$ be a function implemented by a MinMax Recurrent Cascade of state degree one.
  By Lemma~\ref{lemma:always-finite},
  we have that $F$ can be implemented by a MinMax Recurrent Cascade $S$ having state degree one, and 
  a finite input and state space, and hence it is trivially \etafinite{}.
  Then,
  by Lemma~\ref{lemma:minmax-aperiodic}, we have that each MinMax Recurrent Unit of $S$ is
  aperiodic,
  and hence the dynamics of $S$ are aperiodic by
  Lemmas~\ref{lemma:aperiodic-parallel}~and~\ref{lemma:aperiodic-cascade}.
  By Theorem~\ref{thm:aperiodic}, it follows that $\ceta{S}$ is group-free.
  Then $F$ is group-free since, by Theorem~\ref{thm:canon-implement}
  we have that $\ceta{S}$ can implement $F$.
  The proof applies to languages as well, since recognising a language means implementing its
  characteristic function.
\end{proof}

\thmmainbodyexpressivity*
\begin{proof}
  The three claims of the theorem correspond to 
  Theorems~\ref{thm:main-body-expressivity-1},~\ref{thm:main-body-expressivity-2}, 
  and~\ref{thm:main-body-expressivity-3} respectively.
\end{proof}

Next we provide the floating-point counter-part of the theorem above.

\begin{theorem}\label{th:language-expressivity-overall-float}
  Every regular function can be implemented by a floating-point MinMax RNC of state degree equal to the degree
  of the function. 
  Any function that is not regular cannot be implemented by a floating-point MinMax RNC.
  Any function that is not group-free cannot be implemented by a floating-point MinMax RNC of state
  degree one.
\end{theorem}
\begin{proof}
  The three claims of the theorem correspond to 
  Theorems~\ref{thm:main-body-expressivity-1},~\ref{thm:main-body-expressivity-2}, 
  and~\ref{thm:main-body-expressivity-3} respectively.
\end{proof}

\begin{corollary} \label{cor:language-expressivity-overall-float}
  The functions that can be implemented by floating-point MinMax RNCs are the regular functions,
  and the group-free functions if restricted to state degree one.
  The languages that can be recognised by floating-point MinMax RNCs are the regular languages,
  and the star-free languages if restricted to state degree one.
\end{corollary}

\clearpage

\subsection{Positive-margin expressivity of MinMax RNCs}
\label{subsec:positive-margin-minmax-rnc-expressivity}

\todo{}

\begin{closingtheorem}[Positive-margin expressivity of MinMax RNCs]
  \label{thm:positive-margin-minmax-rnc-expressivity}
  The positive-margin expressivity of MinMax RNCs is exactly the class of
  regular languages.
\end{closingtheorem}
\begin{proof}
  We show the two inclusions separately.

  By
  Theorem~\ref{thm:main-body-expressivity},
  every regular language can be recognised by an
  $\eta$-finite MinMax RNC.
  Hence,
  by
  Corollary~\ref{cor:eta-expressivity-implies-positive-margin},
  the positive-margin expressivity of MinMax RNCs includes the class of
  regular languages.

  Conversely,
  recognition with positive margin is,
  in particular,
  recognition.
  Therefore,
  the positive-margin expressivity of MinMax RNCs is contained in their
  ordinary formal-language expressivity.
  Then the second inclusion follows by
  Theorem~\ref{thm:main-body-expressivity}.
\end{proof}

\clearpage{}%

\clearpage{}%

\section{Algorithms for MinMax RNCs}
\label{appendix:algorithms}

\todo{}

We present the algorithms to evaluate MinMax RNCs---their pseudocodes 
follow right below.
For sequential evaluation, we simply follows the recurrence.
For parallel evaluation,
we first note that any two MinMax affine operators (Definition~\ref{def:minmax-affine-operator}),
given by their matrix vector pairs
$\tuple{A_1,b_1},\tuple{A_2,b_2}$, can be composed via the following associative operator:
  $$\tuple{A_1,b_1} \minmaxcomp \tuple{A_2,b_2} \defeq \tuple{A_2 \otimes A_1, (A_2 \otimes
  b_1) \oplus b_2}.$$
Then we apply parallel scan \citep{blelloch1990prefix,blellochmaggs10}.

Thus, the algorithms in this appendix are mainly a basis for building rigorous proofs of the
theorems in Appendix~\ref{appendix:proofs-complexity}, where we show correctness and complexity of
all the present algorithms, and overall of MinMax Recurrence and MinMax RNCs.

\subsection{Preliminary remarks on our Pseudocode}

We use common instructions with a notation that should be self-clear.

\paragraph{Parallel for.}
To describe parallel algorithms, we use the \emph{parallel for} instruction, 
cf.~\citep{blellochmaggs10}. 
    $$\textbf{in parallel for } \VarI \gets e_1 \textbf{ to } e_2 \textbf{ do }
    \textit{instructions}$$
It corresponds to executing the block $\textit{instructions}$ in parallel providing to each
thread a different index $\VarI$ ranging from the value of expression  $e_1$ to the value of
expression $e_2$. Furthermore, we avoid concurrent writes, thus satisfying the 
\emph{Concurrent Read Exclusive Write (CREW)} constraint.

\paragraph{0-indexed tensors.}
In this appendix and in Appendix~\ref{appendix:proofs-complexity} we adopt 0-indexed real-valued 
tensors. We use $\mathsf{shape}(A)$ to obtain the shape of a tensor. We use $A \parallel B$ to
concatenate tensors only when they are vectors, making its meaning straightforward.

\paragraph{No side effects.}
Although we adopt an imperative pseudocode with side effects,
we avoid to assign twice any variable or entry of a tensor.
This makes the correctness proofs easier to develop.

\subsection{Auxiliary Algorithms}

In this subsection we provide the pseudocode of the auxiliary algorithms for parallel evaluation.

\begin{itemize}
  \item
    Algorithm~\ref{alg:parallel-min} is an immediate application of the parallel sum algorithm 
    (cf.~\citep{blellochmaggs10}) by replacing $+$ with $\odot$, since the algorithm works for all
    associative operators, including $\odot$.

  \item
    Algorithm~\ref{alg:parallel-max} is analogous to the above Algorithm~\ref{alg:parallel-max}, except
    that now $+$ is replaced with $\oplus$.

  \item
    Algorithm~\ref{alg:parallel-sum} is a straightforward computation entry-wise.

  \item
    Algorithm~\ref{alg:parallel-mult} is a straightforward algorithm for parallel matrix
    multiplication, with arithmetic addition and multiplication replaced by $\{ \oplus, \odot \}$.

  \item
    Algorithm~\ref{alg:affine-comp} 
    uses the above algorithms to evaluate the $\minmaxcomp$ operator, to compose
    a pair of MinMax recurrences, used by parallel scan to compute multi-step compositions.
  \item
    Algorithm~\ref{alg:parallel-apply} 
    applies a MinMax affine operator to a given vector
    (specifically, the given vector will be a state vector). 
    It is used as the last step after calling parallel scan.
\end{itemize}

\subsection{Sequential evaluation}

We describe the sequential evaluation algorithm.
\begin{itemize}
  \item \ref{alg:sequential-evaluation} is the algorithm for sequential evaluation of MinMax Recurrence.
  It is a straightforward algorithm that computes in order following the recurrence.
\item
  \ref{alg:sequential-evaluation-layer} is the algorithm for sequential evaluation of a MinMax
  Recurrent Layer.
  It evaluates the neurons' input functions by assuming a subroutine to evaluate MLPs,
  and by calling \ref{alg:sequential-evaluation} on the result of their their evaluation.
  Finally, it evaluates the layer's output function, again by calling the subroutine for MLPs.
\item
  \ref{alg:sequential-evaluation-rnc} is the algorithm for sequential evaluation of a MinMax
  RNC. It simply evaluates layers iteratively by calling \ref{alg:sequential-evaluation}.
\end{itemize}

\subsection{Parallel Evaluation}

We present the parallel evaluation a algorithm.

\subsubsection{MinMax Parallel Scan}

Parallel scan \citep{parallelscan,blelloch1990prefix,kogge2009parallel} 
is the core of the parallel evaluation algorithm.

We compose the steps of a MinMax recurrence with the MinMax affine composition of
Appendix~\ref{subsec:minmax-affine-operators}, which we recall.

\defminmaxaffineoperator*

Then, parallel scan is used to evaluate all (exclusive) prefixes of an expression of the form:
\begin{equation} \label{eq:algs-monoidal-expression}
\tuple{A_1,b_1} \minmaxcomp  \cdots \minmaxcomp \tuple{A_t,b_t}.
\end{equation}
The MinMax affine composition $\minmaxcomp$ is associative and, on entries between $\vmin$ and
$\vmax$, it admits an identity element (Proposition~\ref{prop:minmax-affine-monoid}), which is
the reason why we can use parallel scan---see Appendix~\ref{appendix:proofs-complexity} for its
analysis.

First, parallel scan requires the expression in~\eqref{eq:algs-monoidal-expression} to be a monoidal
expression. Namely, the operator $\minmaxcomp$ must be associative, and there must be an identity 
element, which is\footnote{allowing ourselves not to be entirely precise in this informal explanation for
the sake for providing the intuition} an element $\tuple{E,e}$ such that 
$$\tuple{E,e} \minmaxcomp \tuple{A,b} = \tuple{A,b} \minmaxcomp \tuple{E,e} = \tuple{A,b}$$
for all elements $\tuple{A,b}$ that can be generated by any intermediate composition
$$\tuple{A,b} = \tuple{A_{t_1},b_{t_1}} \minmaxcomp  \cdots \minmaxcomp \tuple{A_{t_2},b_{t_2}}.$$

\paragraph{Algorithms.}
We describe parallel scan and its subroutine for the identity element.
\begin{itemize}

  \item 
    Algorithm \ref{alg:compute-identity} generates the identity element $\tuple{E,e}$ out of two
    given values $\vmin$ and $\vmax$, which in the overall algorithm will be the largest and
    smallest scalars returned by unit's input function on its entire input sequence, also including
    the entries of the initial state. Intuitively, $(\vmin,\vmax)$ will mimick $(-\infty,+\infty)$ 
    which act as $(0,1)$ in the MinMax semiring
    $(\oplus,\odot,-\infty,+\infty)$.\footnote{Potentially we could have used the MinMax semiring
    $(\oplus,\odot,-\infty,+\infty)$ throughout the paper, instead of using only the MinMax algebra 
  $(\oplus,\odot)$. However, we have preferred not to introduce infinities.} Thus, $E$ is
  similar to the identity matrix $I_{n,n}$ (replacing $1$'s with $+\infty$ and $0$'s with $-\infty$)
  and $e$ is similar to the null vector $0_{n}$ (replacing $0$'s with $-\infty$).
  \item 
    Algorithm \ref{alg:parallel-scan} follows the standard parallel scan algorithm (e.g.,
    see~\citep{blellochmaggs10}), combining pairs of elements of the given expression, with some
    careful handling of the odd and even cases. It calls Algorithm~\ref{alg:compute-identity} in the
    case case, as it computes exclusive prefixes, hence identity on a given expression of length 1.
\end{itemize}

\subsubsection{Parallel Evaluation}

Parallel evaluation of MinMax Recurrence, and then of layers and cascades, is 
straightforward once parallel scan computes prefixes of recurrences.
\begin{itemize}
  \item \ref{alg:parallel-general-evaluation} is the algorithm for parallel evaluation of MinMax Recurrence.
    It computes $\vmin,\vmax$ that are required by parallel scan.
    Then, it calls parallel scan (Algorithm~\ref{alg:parallel-scan}) to compute the sequence of matrices
    corresponding to prefixes of compositions of MinMax Recurrence. 
    Then, in parallel for every
    sequence step $t$, it includes the recurrence step at time $t$ 
    (since parallel scan returns exclusive prefixes), and finally it applies the overall
    composition to the initial state, hence obtaining the state at time $t$.
\item
  \ref{alg:parallel-evaluation-layer} is the algorithm for parallel evaluation of a MinMax
  Recurrent Layer.  It is as \ref{alg:sequential-evaluation-layer}, except that it handles each
  neuron in parallel, it calls \ref{alg:parallel-general-evaluation}, and then computes the output's
  layer in parallel wrt sequence steps.
\item
  \ref{alg:parallel-evaluation-rnc} is the algorithm for parallel evaluation of a MinMax RNC. 
  It is as 
  \ref{alg:parallel-evaluation-rnc}, except that it calls \ref{alg:parallel-evaluation-layer} to
  evaluate layers.
\end{itemize}

\clearpage
\FloatBarrier

\begin{algorithm}
  \SetKwFunction{ParallelMin}{ParallelMin}

  \Input{Array $A$ of length $n\geq 1$}
  \Output{$\bigodot_{i=0}^{n-1} \idx{A}{i}$}

  \Fn{\ParallelMin{$A$}}{
    \If{$n = 1$}{
      \Return $\idx{A}{0}$\;
    }
    \Else{
      $m \gets \lfloor n/2 \rfloor$\;
      \Allocate array $B$ of length $m$\;

      \ParFor{$i \in \iival{0}{m-2}$}{
        $\idx{B}{i} \gets \idx{A}{2i} \odot \idx{A}{2i+1}$\;
      }
      \If{$n$ is even}{
        $\idx{B}{m-1} \gets \idx{A}{n-2} \odot \idx{A}{n-1}$\;
      }
      \Else{
        $\idx{B}{m-1} \gets \idx{A}{n-3} \odot \idx{A}{n-2} \odot \idx{A}{n-1}$\;
      }

      \Return \ParallelMin{$B$}\;
    }
  }
  \caption{Minimum element of an array computed in parallel}
  \label{alg:parallel-min}
\end{algorithm}

\begin{algorithm}
  \SetKwFunction{ParallelMax}{ParallelMax}

  \Input{Array $A$ of length $n$}
  \Output{$\bigoplus_{i=0}^{n-1} \idx{A}{i}$}

  \Fn{\ParallelMax{$A$}}{
    \If{$n = 1$}{
      \Return $\idx{A}{0}$\;
    }
    \Else{
      $m \gets \lfloor n/2 \rfloor$\;
      \Allocate array $B$ of length $m$\;

      \ParFor{$i \in \iival{0}{m-2}$}{
        $\idx{B}{i} \gets \idx{A}{2i} \oplus \idx{A}{2i+1}$\;
      }
      \If{$n$ is even}{
        $\idx{B}{m-1} \gets \idx{A}{n-2} \oplus \idx{A}{n-1}$\;
      }
      \Else{
        $\idx{B}{m-1} \gets \idx{A}{n-3} \oplus \idx{A}{n-2} \oplus \idx{A}{n-1}$\;
      }
      \Return \ParallelMax{$B$}\;
    }
  }
  \caption{Maximum element of an array computed in parallel}
  \label{alg:parallel-max}
\end{algorithm}

\begin{algorithm}
  \SetKwFunction{ParallelAdd}{ParallelAdd}
  \Input{Matrices $A,B \in \reals^{m \times n}$}
  \Output{Matrix $C \in \reals^{m \times n}$ such that $C = A \oplus B$}

  \Fn{\ParallelAdd{$A,B$}}{
    allocate matrix $C$ of shape $(m,n)$\;

    \ParFor{$i \gets 0$ \KwTo $m-1$}{
      \ParFor{$j \gets 0$ \KwTo $n-1$}{
        $\idx{C}{i,j} \gets \idx{A}{i,j} \oplus \idx{B}{i,j}$\;
      }
    }
    \Return{$C$}\;
  }
  \caption{Parallel matrix addition}
  \label{alg:parallel-sum}
\end{algorithm}

\begin{algorithm}
  \SetKwFunction{ParallelMult}{ParallelMult}
  \Input{Matrix $A \in \mathbb{R}^{m \times p}$, matrix $B \in \mathbb{R}^{p \times n}$}
  \Output{Matrix $C \in \mathbb{R}^{m \times n}$ such that $C = A \otimes B$}

  \Fn{\ParallelMult{$A,B$}}{
    allocate matrix $C$ of shape $(m,n)$\;
    allocate tensor $T$ of shape $(m,n,p)$\;
    \ParFor{$i \gets 0$ \KwTo $m-1$}{
        \ParFor{$j \gets 0$ \KwTo $n-1$}{
            \ParFor{$k \gets 0$ \KwTo $p-1$}{
              $\idx{T}{i,j,k} \gets \idx{A}{i,k} \odot \idx{B}{k,j}$\;
            }
            $\idx{C}{i,j} \gets \ParallelMax(\idx{T}{i,j})$\;
        }
    }

    \Return{$C$}\;
  }
  \caption{Parallel matrix multiplication}
  \label{alg:parallel-mult}
\end{algorithm}

\begin{algorithm}
  \SetKwFunction{ParallelComp}{ParallelComp}
  \Input{$\ptuple{A_1, B_1} \in \reals^{n \times n}\times \reals^n$, $\ptuple{A_2,B_2} \in \reals^{n \times n}\times \reals^n$}
  \Output{$\ptuple{A,B}\in \reals^{n \times n}\times \reals^n$ such that $\tuple{A,B} =
  \tuple{A_1,B_1} \oplustimes \tuple{A_2,B_2}$}

  \Fn{\ParallelComp{$A_1,B_1,A_2,B_2$}}{
    \Allocate tensor $A$ of shape $(n,n)$\;
    \Allocate arrays $B,Z$ of length $n$\;
    $A \gets \ParallelMult(A_2, A_1)$\;
    $Z \gets \ParallelMult(A_2, B_1)$\;
    $B \gets \ParallelAdd(Z, B_2)$\;
    \Return{$\ptuple{A,B}$}\;
  }
  \caption{Parallel MinMax Composition}
  \label{alg:affine-comp}
\end{algorithm}

\begin{algorithm}
  \SetKwFunction{ParallelApply}{ParallelApply}
  \Input{tensor $A$ of shape $(N,N)$, array $B$ of length $N$, array $X$ of length $N$}
  \Output{tensor $Y$ such that $Y = (A \otimes X) \oplus B$}

  \vspace{.3em}
  \Fn{\ParallelApply{$A,B,X$}}{
    \Allocate arrays $Z,Y$ of length $N$\;
    $Z \gets \ParallelMult(A,X)$\;
    $Y \gets \ParallelAdd(Z,B)$\;
    $\Return\, Y$\;
  }
  \caption{Parallel MinMax Operator Application}
  \label{alg:parallel-apply}
\end{algorithm}

\begin{algorithm}
  \SetKwFunction{ComputeIdentity}{ParallelIdentity}

  \caption{Compute identity element in parallel}
\label{alg:compute-identity}
\Input{scalars $\vmin,\vmax \in \reals$ and $n \in \posnaturals$}
\Output{identity $(E,e)$ for MinMax monoid $\mathcal{M}_{[\vmin,\vmax],n}$}

\vspace{.3ex}

\Fn{\ComputeIdentity{$\vmin,\vmax,n$}}{

  \Allocate tensor $E$ of shape $(N,N)$\; 
  \ParFor{$i \in \iivalro{0}{N-1}$}{
    \ParFor{$j \in \iivalro{0}{N}$}{
      \lIf{$i=j$}{
        $\idx{E}{i,j} \gets \vmax$
      }
      \lElse{
        $\idx{E}{i,j} \gets \vmin$
      }
    }
  }

  \Allocate array $e$ of length $N$\;
  \ParFor{$i \in \iivalro{0}{N}$}{
    $\idx{e}{i} \gets \vmin$\;
  }

  \Return{$(E,e)$}\;
}
\end{algorithm}

\begin{algorithm}
  \SetKwFunction{ParallelScan}{ParallelScan}
  \Input{tensor $A$ of shape $(T,N,N)$, tensor $B$ of shape $(T,N)$, 
    scalars $\vmin,\vmax$;}
  \Output{tensors $(C,D)$ with 
  $(\idx{C}{t},\idx{D}{t}) = (\idx{A}{0},\idx{B}{0}) \oplustimes \cdots \oplustimes
(\idx{A}{t-1},\idx{B}{t-1}) \oplustimes \mathcal{E}$ for all $t \in \iivalro{0}{T}$,
with $\mathcal{E}$ the identity of the MinMax monoid $\mathcal{M}_{[\vmin,\vmax],N}$}

  \vspace{.3em}
  \Fn{\ParallelScan{$A,B,\vmin,\vmax$}}{

    $(T,N) \gets \mathsf{shape}(B)$\;

    \If{$T=1$}{
      \Return{$\ComputeIdentity(\vmin,\vmax,N)$}\;
    }

    $m \gets \lfloor T/2 \rfloor$; 
    $\ell \gets \lceil T/2 \rceil$\;
    \Allocate tensor $X$ of shape $(\ell, N, N)$\; 
    \Allocate tensor $Y$ of shape $(\ell, N)$\;

    \ParFor{$t \in \iival{0}{m-1}$}{
      $(\idx{X}{t},\idx{Y}{t}) \gets 
      \ParallelComp(\idx{A}{2t},\idx{B}{2t},\idx{A}{2t+1},\idx{B}{2t+1})$\;
    }
    \If{$T$ is odd}{
      $(\idx{X}{\ell-1},\idx{Y}{\ell-1}) \gets (\idx{A}{T-1},\idx{B}{T-1})$\;
    }

    \vspace{0.3em}

    $(Z,W) \gets \ParallelScan(X, Y, \vmin,\vmax)$\;

    \vspace{0.3em}

    \Allocate tensor $C$ of shape $(T,N,N)$\; 
    \Allocate tensor $D$ of shape $(T,N)$\;
    \ParFor{$t \in \iival{0}{T-1}$}{
      \If{$t$ is even}{
        $(\idx{C}{t},\idx{D}{t}) \gets (\idx{Z}{t/2},\idx{W}{t/2})$\;
      }
      \Else{
        $(\idx{C}{t},\idx{D}{t}) \gets \ParallelComp(\idx{Z}{(t-1)/2}, \idx{W}{(t-1)/2}, \idx{A}{t-1},\idx{B}{t-1})$\;
      }
    }
    \Return{$(C,D)$}\;
  }
  \caption{Parallel Scan}
  \label{alg:parallel-scan}
\end{algorithm}

\begin{function}
  \caption{SeqRecEval(${X_\mathrm{init}, A, B}$)}
\label{alg:sequential-evaluation}
\vspace{0.6ex}
\Input{tensors $\tuple{X_\mathrm{init}, A, B}$ with $X_\mathrm{init} \in \reals^N$, $A \in \reals^{T \times N \times
N}$, $B \in \reals^{T \times N}$;}
\vspace{1.0ex}
\Output{tensor $X$ with $\idx{X}{0} = \minmax(X_\mathrm{init}, \idx{A}{0}, \idx{B}{0})$ and 
$\idx{X}{t} = \minmax(\idx{X}{t-1}, \idx{A}{t}, \idx{B}{t})$ for all $t \in \iivalro{1}{T}$;}
\vspace{1.6ex}
$(T, N) \gets \operatorname{shape}(B)$\;
$\Allocate$ tensor $X$ of shape $(T,N)$\;
$\Allocate$ tensor $Z$ of shape $(T+1,N+1,N)$\;
$\idx{Z}{0,N} \gets X_\mathrm{init}$\;
\For{$t \gets 1$ \KwTo $T$}{
  \For{$i \gets 1$ \KwTo $N$}{
    $\idx{Z}{t,0,i-1} \gets \idx{B}{t-1,i-1}$\;
    \For{$j \gets 1$ \KwTo $N$}{
      $\idx{Z}{t,j,i-1} \gets \idx{Z}{t,j-1,i-1} \oplus \big(\idx{A}{t-1,i-1,j-1} \odot
      \idx{Z}{t-1,N,j-1}\big)$\;
    }
  }
  $\idx{X}{t-1} \gets \idx{Z}{t,N}$\;
}
\Return{$X$}\;
\end{function}

\begin{function}
  \caption{SeqLayerEval(${W,U}$)}
\label{alg:sequential-evaluation-layer}
  \SetKwFunction{EvaluateMLP}{EvaluateMLP}
  \Input{$(W,U)$ where $W$ are the weights of a MinMax Layer $S$ and 
  $U \in \reals^{T \times d_\mathrm{in}}$ for $d_\mathrm{in}$ the input dimension of $S$;}
  \Output{$Y \in \reals^{T \times d_\mathrm{out}}$ for $d_\mathrm{out}$ the output dimension of $S$,
  such that $Y=S(U)$;}

\BlankLine

$(T,d_\mathrm{in}) \gets \mathsf{shape}(U)$;\,
$(n_\mathrm{units},d_\mathrm{state}) \gets \mathsf{shape}(X^\mathrm{init})$;\,
$(d_\mathrm{out},\_) \gets \mathsf{shape}(W^\mathrm{out})$\;
$(W^\mathrm{R}, W^\mathrm{s}, W^\mathrm{out}, X^\mathrm{init}) \gets W$\;

\Allocate tensor $X$ of shape $(T+1,\, n_\mathrm{units} \cdot d_\mathrm{state})$\;
\For{$i \gets 0$ \KwTo $n_\mathrm{units}-1$}{
  \Allocate tensor $A_i$ of shape $(T,d_\mathrm{state},d_\mathrm{state})$\;
  \Allocate tensor $B_i$ of shape $(T,d_\mathrm{state})$\;
  \For{$t \gets 0$ \KwTo $T-1$}{
    $\idx{A_i}{t} \gets \EvaluateMLP(\idx{W^\mathrm{R}}{i}, \idx{U}{t})$\;
    $\idx{B_i}{t} \gets \EvaluateMLP(\idx{W^\mathrm{s}}{i}, \idx{U}{t})$\;
  }
  $p_i \gets i \cdot d_\mathrm{state}$\;
  $q_i \gets (i+1)\cdot d_\mathrm{state}$\;
  $\idx{X}{0,\, p_i: q_i} \gets \idx{X^\mathrm{init}}{i}$\;
  $\idx{X}{1:\,,\, p_i: q_i} \gets \SeqRecEval{$\idx{X^\mathrm{init}}{i},A_i,B_i$}$\;
}
\Allocate tensor $Y$ of shape $(T,d_\mathrm{out})$\;
\For{$t \gets 0$ \KwTo $T-1$}{
  $\idx{Y}{t} \gets \EvaluateMLP(W^\mathrm{out},\, \idx{X}{t} \parallel \idx{X}{t+1} \parallel \idx{U}{t})$\;
}
\Return{$Y$}\;
\end{function}

\begin{function}
  \caption{SeqCascEval(${W,U}$)}
\label{alg:sequential-evaluation-rnc}
\Input{$\tuple{W,U}$ where $W$ are the weights of an RNC $S$, and $U \in \reals^{T \times
d_\mathrm{in}}$ for $d_\mathrm{in}$ the input dimension of $S$;}
\Output{$Y \in \reals^{T \times d_\mathrm{out}}$ for $d_\mathrm{out}$ the output dimension of $S$, 
such that $Y = S(U)$;}
\BlankLine
$n_\mathrm{layers} \gets \mathsf{length}(W)$\;
$Y \gets U$\;
\For{$i \gets 0$ \KwTo $n_\mathrm{layers}-1$}{
  $Y \gets \SeqLayerEval{$\idx{W}{i}, Y$}$\;
}
\Return{$Y$}\;
\end{function}

\begin{function}
  \caption{ParallelRecEval(${X_\mathrm{init}, A, B}$)}
  \label{alg:parallel-general-evaluation}
\Input{array $X_\mathrm{init}$ of length $N$, tensor $A$ of shape $(T,N,N)$, tensor $B$ of shape
$(T,N)$;}
\Output{tensor $X$ of shape $(T,N)$ such that  
$\idx{X}{0} = \minmax(X_\mathrm{init}, \idx{A}{0}, \idx{B}{0})$
and
$\idx{X}{t} = \minmax(\idx{X}{t-1}, \idx{A}{t}, \idx{B}{t})$ for every $t \in \iivalo{0}{T}$;}

$(T,N) \gets \mathsf{shape}(B)$\;
$\vmin \gets \ParallelMin(X_\mathrm{init} \parallel \operatorname{flat}(A) \parallel \operatorname{flat}(B))$\;
$\vmax \gets \ParallelMax(X_\mathrm{init} \parallel \operatorname{flat}(A) \parallel \operatorname{flat}(B))$\;
$(C,D) \gets \ParallelScan(A, B, \vmin,\vmax)$\;

\Allocate tensors $X,W$ of shape $(T,N)$\;
\Allocate tensor $Z$ of shape $(T,N,N)$\;
\ParFor{$t \gets 0$ \KwTo $T-1$}{
  $\ptuple{\idx{Z}{t},\idx{W}{t}} \gets
  \ParallelComp(\idx{C}{t},\idx{D}{t},\idx{A}{t},\idx{B}{t})$\;
  $\idx{X}{t} \gets \ParallelApply(\idx{Z}{t},\idx{W}{t},X_\mathrm{init})$\;
}

\Return{$X$}\;
\end{function}

\begin{function}
  \caption{ParLayerEval(${W,U}$)}
\label{alg:parallel-evaluation-layer}
\vspace{0.6ex}
  \Input{$(W,U)$ where $W$ are the weights of a MinMax Layer $S$ and 
  $U \in \reals^{T \times d_\mathrm{in}}$ for $d_\mathrm{in}$ the input dimension of $S$;}
\Output{$Y$ such that $Y=S(U)$;}

\BlankLine

$(T,d_\mathrm{in}) \gets \mathsf{shape}(U)$;\,
$(n_\mathrm{units},d_\mathrm{state}) \gets \mathsf{shape}(X^\mathrm{init})$;\,
$(d_\mathrm{out},\_) \gets \mathsf{shape}(W^\mathrm{out})$\;
$(W^\mathrm{R}, W^\mathrm{s}, W^\mathrm{out}, X^\mathrm{init}) \gets W$\;

\Allocate tensor $X$ of shape $(T+1,\, n_\mathrm{units} \cdot d_\mathrm{state})$\;
\ParFor{$i \gets 0$ \KwTo $n_\mathrm{units}-1$}{
  \Allocate tensor $A_i$ of shape $(T,d_\mathrm{state},d_\mathrm{state})$\;
  \Allocate tensor $B_i$ of shape $(T,d_\mathrm{state})$\;
  \ParFor{$t \gets 0$ \KwTo $T-1$}{
    $\idx{A_i}{t} \gets \EvaluateMLP(\idx{W^\mathrm{R}}{i}, \idx{U}{t})$\;
    $\idx{B_i}{t} \gets \EvaluateMLP(\idx{W^\mathrm{s}}{i}, \idx{U}{t})$\;
  }
  $p_i \gets i \cdot d_\mathrm{state}$\;
  $q_i \gets (i+1) \cdot d_\mathrm{state}$\;
  $\idx{X}{0,\, p_i: q_i} \gets \idx{X^\mathrm{init}}{i}$\;
  $\idx{X}{1:\,,\, p_i: q_i} \gets \ParallelRecEval{$\idx{X^\mathrm{init}}{i},A_i,B_i$}$\;
}
\Allocate tensor $Y$ of shape $(T,d_\mathrm{out})$\;
\ParFor{$t \gets 0$ \KwTo $T-1$}{
  $\idx{Y}{t} \gets \EvaluateMLP(W^\mathrm{out},\, \idx{X}{t} \parallel \idx{X}{t+1} \parallel \idx{U}{t})$\;
}
\Return{$Y$}\;
\end{function}

\begin{function}
  \caption{ParCascEval(${W,U}$)}
\label{alg:parallel-evaluation-rnc}
\Input{$\tuple{W,U}$ where $W$ are the weights of an RNC $S$, and $U \in \reals^{T \times
d_\mathrm{in}}$ for $d_\mathrm{in}$ the input dimension of $S$;}
\vspace{1.0ex}
\Output{$Y \in \reals^{T \times d_\mathrm{out}}$ for $d_\mathrm{out}$ the output dimension of $S$, 
such that $Y = S(U)$;}
\BlankLine
$n_\mathrm{layers} \gets \mathsf{length}(W)$\;
$Y \gets U$\;
\For{$i \gets 0$ \KwTo $n_\mathrm{layers}-1$}{
  $Y \gets \ParLayerEval{$\idx{W}{i}, Y$}$\;
}
\Return{$Y$}\;
\end{function}

\clearpage
\FloatBarrier

\clearpage{}%

\clearpage{}%

\section{Computational Complexity of MinMax RNCs}
\label{appendix:proofs-complexity}

\todo{}

In this section we prove the theorems of Section~\ref{sec:main-body-complexity}.
The theorems are proved in Section~\ref{subsec:complexity-proofs}, after discussing the setup in
which the results are proved (Section~\ref{subsec:models}), and after formally defining the relevant 
computation problems (Section~\ref{subsec:problems}).

\subsection{Model of computation} \label{subsec:models}

The complexity results are established considering the following models of computation.

\paragraph{Sequential case.}
The sequential complexity results are proven under the assumptions of 
\emph{Random-Access Machine (RAM)} on real-valued arithmetic, cf.~\citep{BCSS98}.
A RAM has a finite number of registers and, in one computation step, it may perform one
of the following operations:
\begin{itemize}
  \item 
    write to a register the scalar value $a \circ b$, where $a,b$ are scalar values read from some
    registers, for any operation $\circ\in\{+,-,\times,\div,\odot,\oplus\}$ and $b\neq 0$ in the
    case of division;
  \item 
    write to a register the scalar value $\lceil a \rceil$ or $\lfloor a \rfloor$ for $a$ a scalar
    value read from a register;
  \item 
    compare the value of two registers and branch accordingly.
\end{itemize}
The instructions in the pseudocode we use are in line with RAM, while providing a higher-level
specification language (e.g., registers correspond to variables and entries of a tensor,
branching is involved in if-then-else and in for loops). Importantly, in our pseudocode we only use
the elementary arithmetic operations mentioned above, whose cost is assumed to be $1$. 

\paragraph{Parallel case.}
The parallel complexity results are proven considering 
\emph{Parallel RAM on real-valued arithmetic with Concurrent Read and Exclusive Write (PRAM-CREW)},
cf.~\citep{blellochmaggs10}.
A PRAM-CREW has an arbitrary number of processors that carry out the computation proceeding by
rounds. In each round, the processors simultaneously perform one of the operations described above
for the RAM model, under the condition that two processors never write to the same register in the
same round.
The \emph{work} is the total number of operations performed by the processors,
and
the \emph{depth} is the number of rounds.
The instructions in our pseudocode are in line with the PRAM-CREW, while providing a higher-level
specification language. Compared to the sequential case, we now have the `parallel for' instruction,
which allows us to describe simultaneous computations carried out by the processors.
Brent's Theorem translates work and depth to execution time on PRAM-CREW.

\begin{theorem}[Brent’s Theorem, cf.~Theorem~25.1 in \citep{blellochmaggs10}]
  Any algorithm having work $W$ and depth $D$ can be executed in $O(W/P + D)$ steps in the CREW-PRAM
  model.
\end{theorem}

\subsection{Computation problems} \label{subsec:problems}

We define the relevant compuation problems, formalising the tasks for which we want to show their
complexity.

\subsubsection{Background}

\begin{definition}
  A \emph{computation problem over $\reals$} is a triple $\tuple{I,O,f}$ 
  where 
  $I$ is a set of tuples of tensors called \emph{valid inputs} or simply \emph{inputs}, 
  $O$ is a set of tuples of tensors called \emph{possible outputs} or simply \emph{outputs}, 
  $f: I \to O$ is a function that maps every input element $x \in I$ to an output element 
  $f(x) \in O$ called \emph{the expected output for input $x$}.
  The \emph{size of an input $x \in I$} is the number of scalar values in the tensors of $x$.
\end{definition}

\begin{definition}
  A program $\mathcal{P}$ \emph{solves} a computation problem $\tuple{I,O,f,s}$ if,
  for every input $x \in I$,
  we have that $x$ is a valid input for program $\mathcal{P}$ and
  the output $\mathcal{P}(x)$ of the program equals $f(x)$.
\end{definition}

\begin{definition}
  For $T : \naturals \to \naturals$ a monotonic function,
  \emph{a sequential program $\mathcal{P}$ solves a computation problem $\tuple{I,O,f}$ in time $T$}
  if, for every input $x \in I$,
  we have that
  $x$ is a valid input for $\mathcal{P}$,
  the the output $\mathcal{P}(x)$ of the program equals $f(x)$,
  and 
  the execution time of $\mathcal{P}$ on input $x$ is at most $T(n)$ for $n$ the size of input $x$.
\end{definition}

\begin{definition}
  For $W,D : \naturals \to \naturals$ monotonic functions,
  \emph{a parallel program $\mathcal{P}$ solves a computation problem $\tuple{I,O,f}$ with depth $D$
    and work $W$}
  if, for every input $x \in I$, 
  we have that
  $x$ is a valid input for $\mathcal{P}$,
  the output of the program $\mathcal{P}(x)$ equals $f(x)$,
  the depth of the execution of $\mathcal{P}$ on $x$ is at most $D(n)$ for $n$ the size $x$,
  and
  the work of the execution of $\mathcal{P}$ on $x$ is at most $W(n)$.
\end{definition}

\begin{definition}
  For $T,P : \naturals \to \naturals$ monotonic functions,
  \emph{a parallel program $\mathcal{P}$ solves a computation problem $\tuple{I,O,f}$ in time $T$
  having $P$ processors available}
  if, for every input $x \in I$, 
  we have that
  $x$ is a valid input for $\mathcal{P}$,
  the output of the program $\mathcal{P}(x)$ equals $f(x)$,
  and
  the execution time of program $\mathcal{P}$ on input $x$ having $P(n)$ processors available is at
  most $T(n)$ for $n$ is the size of input $x$.
\end{definition}

\begin{definition}
  We say that a computation problem can be solved sequentially in time $T$ if there exists a
  sequential program that solves it in time $T$.
  We say that a computation problem can be solved in parallel with work $W$ and depth $D$ 
  if there exists a parallel program that solves it with work $W$ and depth $D$.
  We say that a computation problem can be solved in parallel in time $T$ having $P$ processors
  available if there exists a parallel program that solves it in time $T$ having $P$ processors
  available.
\end{definition}

\subsubsection{Definition of the considered computation problems}

\begin{closingdefinition}[MinMax Recurrence Evaluation]
  \label{def:minmax-recurrence-evaluation-problem}
  \emph{MinMax Recurrence Evaluation} is the computation problem where inputs and outputs
  are respectively given by
  \begin{align*}
    I & \defeq \bigcup_{T=1}^\infty \bigcup_{N=1}^\infty I_{T,N}
      && \text{with }\;
    I_{T,N} \defeq 
    \big\{ 
      \tuple{x_\mathrm{init}, A, b} \mid x_\mathrm{init} \in \reals^N,\, A \in \reals^{T \times N \times N},\,
      b \in \reals^{T \times N}
    \big\}, 
    \\[.1em]
    O & \defeq \bigcup_{T=1}^\infty \bigcup_{N=1}^\infty O_{T,N}
      && \text{with }\;
    O_{T,N} \defeq
    \big\{ X \mid X \in \reals^{T \times N} \big\}, 
  \end{align*}
  and
  the expected output for input $\tuple{x_\mathrm{init}, A, b} \in I_{T,N}$
  is $X \in O_{T,N}$ satisfying
  \begin{equation*}
    \idx{X}{0} = \minmax\big(x_\mathrm{init}, \idx{A}{0}, \idx{b}{0}\big),\quad
    \idx{X}{t} = \minmax\big(\idx{X}{t-1}, \idx{A}{t}, \idx{b}{t}\big) \quad 
    \forall t \in \iivalro{1}{T}.
  \end{equation*}
  For inputs in $I_{T,N}$, 
  we call $T$ the \emph{length} or \emph{number of steps},
  and 
  we call $N$ the \emph{state dimension}.
\end{closingdefinition}

\begin{closingdefinition}[Representation of MinMax RNCs] \label{def:rnc-representation}
  The size of a linear function is $\mathbf{s} = \tuple{d_\mathrm{in}, d_\mathrm{out}}$.
  The weights for a linear function of size $\mathbf{s}$ are
  \begin{align*}
    \mathbf{W}^\mathrm{lin}_\mathbf{s} 
    \defeq \{ \tuple{A, b} \mid 
      A \in \reals^{d_\mathrm{out} \times d_\mathrm{in}},\,
      b \in \reals^{d_\mathrm{out}}
    \}
  \end{align*}
  The linear function having weights 
  $W = \tuple{A,b} \in
  \mathbf{W}^\mathrm{linear}_\mathbf{s}$
  is the function $f: \reals^{d_\mathrm{in}} \to \reals^{d_\mathrm{out}}$ given by
  \begin{align*}
    f(x) = Ax + b,
  \end{align*}
  and denoted by $\operatorname{Lin}(W) \defeq f$.

  The size of an MLP is $\mathbf{s} = \tuple{d_\mathrm{in},d_\mathrm{out},d_\mathrm{h},n_\mathrm{h}}$
  with
  input dimension $d_\mathrm{in}$,
  output dimension $d_\mathrm{out}$,
  hidden dimension $d_\mathrm{h}$,
  and
  number of hidden layers $n_\mathrm{h}$.
  The weights for an MLP of size $\mathbf{s}$ are
  \begin{align*}
    \mathbf{W}^\mathrm{mlp}_\mathbf{s} 
    \defeq \{ \tuple{W^\mathrm{in},
      \tuple{W_0^\mathrm{h}, \ldots, W_{n_\mathrm{h}-1}^\mathrm{h}}, W^\mathrm{out}} \mid 
      W^\mathrm{in} \in \mathbf{W}^\mathrm{lin}_{d_\mathrm{in},d_\mathrm{h}},\,
      W^\mathrm{h}_i \in \mathbf{W}^\mathrm{lin}_{d_\mathrm{h}, d_\mathrm{h}},\,
      W^\mathrm{out} \in \mathbf{W}^\mathrm{lin}_{d_\mathrm{h}, d_\mathrm{out}}
    \}
  \end{align*}
  The MLP having weights 
  $W = \tuple{W^\mathrm{in}, W^\mathrm{h},W^\mathrm{out}} \in
  \mathbf{W}^\mathrm{mlp}_\mathbf{s}$
  is the function $\reals^{d_\mathrm{in}} \to \reals^{d_\mathrm{out}}$ given by
  \begin{align*}
    \operatorname{MLP}(W) \defeq
    \operatorname{Lin}(W^\mathrm{in}) \circ \phi \circ 
    \operatorname{Lin}(\idx{W^\mathrm{h}}{0}) \circ \phi \circ 
    \cdots \circ 
    \operatorname{Lin}(\idx{W^\mathrm{h}}{n_\mathrm{h}-1}) \circ \phi \circ 
    \operatorname{Lin}(W^\mathrm{out}),
  \end{align*}
  where $\phi: \reals \to \reals$ is the activation function applied element-wise.

  The size of a MinMax Neural Layer is
  $\mathbf{s} = \tuple{d_\mathrm{in},d_\mathrm{mlp},n_\mathrm{mlp},d_\mathrm{out},n_\mathrm{units},
  d_\mathrm{state}}$
  with
  input dimension $d_\mathrm{in}$,
  hidden dimension of MLPs $d_\mathrm{mlp}$,
  number of hidden layers of MLPs $n_\mathrm{mlp}$,
  output dimension $d_\mathrm{out}$,
  number of units $n_\mathrm{units}$,
  and
  state dimension of each unit $d_\mathrm{state}$.

  The weights of the input functions of a MinMax Neural Layer of size $\mathbf{s}$ are 
  \begin{align*}
    \mathbf{W}^\mathrm{layer,R}_\mathbf{s} 
    & \defeq 
    \{
      \tuple{W_0^\mathrm{R}, \ldots, W^\mathrm{R}_{n_\mathrm{units}-1}} 
      \mid
      W_i^\mathrm{R} \in
      \mathbf{W}^\mathrm{mlp}_{\mathbf{s}_\mathrm{R}}
    \}
    && \text{ with }
    \mathbf{s}_\mathrm{R} 
    = \tuple{{d_\mathrm{in},\,  d_\mathrm{state}^2,\, \lceil d_\mathrm{mlp}/n_\mathrm{units}\rceil,\, n_\mathrm{mlp}}},
    \\[.2em]
    \mathbf{W}^\mathrm{layer,s}_\mathbf{s} 
    & \defeq 
    \{
      \tuple{W_0^\mathrm{s}, \ldots, W^\mathrm{s}_{n_\mathrm{units}-1}} 
      \mid
      W_i^\mathrm{s} \in
      \mathbf{W}^\mathrm{mlp}_{\mathbf{s}_\mathrm{s}}
    \}
    && \text{ with }
    \mathbf{s}_\mathrm{s} 
    = \tuple{{d_\mathrm{in},\,  d_\mathrm{state},\, \lceil d_\mathrm{mlp}/n_\mathrm{units}\rceil,\, n_\mathrm{mlp}}},
  \end{align*}
  and
  the weights of a MinMax Neural Layer of size $\mathbf{s}$ are 
  \begin{align*}
    \mathbf{W}^\mathrm{layer}_\mathbf{s} 
    \defeq 
    \{
      \tuple{W^\mathrm{R}, W^\mathrm{s}, W^\mathrm{out}, X^\mathrm{init}} 
      \mid {}
      & 
      W^\mathrm{R} \in
      \mathbf{W}^\mathrm{layer,R}_{\mathbf{s}},
      \\
      & 
      W^\mathrm{s} \in
      \mathbf{W}^\mathrm{layer,s}_{\mathbf{s}},
      \\
      & 
      W^\mathrm{out} \in
      \mathbf{W}^\mathrm{mlp}_{\mathbf{s}_\mathrm{out}},
      \\
      & 
      X^\mathrm{init} \in \reals^{n_\mathrm{units} \times d_\mathrm{state}} 
    \}
  \end{align*}
  where
  \begin{align*}
    \mathbf{s}_\mathrm{out} 
    & = \tuple{d_\mathrm{state}\cdot n_\mathrm{units},\,
      d_\mathrm{out},\,d_\mathrm{mlp},\,n_\mathrm{mlp}}.
  \end{align*}

  The MinMax Layer $\operatorname{Layer}(W)$ with weights 
  $W = \tuple{W^\mathrm{R}, W^\mathrm{s}, W^\mathrm{out}, X^\mathrm{init}} \in
  \mathbf{W}^\mathrm{layer}_\mathbf{s}$
  is the MinMax Layer
  where the output function is $\operatorname{MLP}(W^\mathrm{out})$,
  the reset function of unit $i$ is $\operatorname{MLP}(W^\mathrm{R}_i)$, 
  and
  the set function of unit $i$ is $\operatorname{MLP}(W^\mathrm{R}_i)$.

  The set of possible sizes of a MinMax RNC is 
  $\mathbf{S}^\mathrm{rnc} \defeq (\posnaturals)^8$---the set of $8$-tuples of non-zero natural
  numbers.
  A size $\mathbf{s}  = \tuple{d_\mathrm{in}, d_\mathrm{out}, d_\mathrm{model}, d_\mathrm{state},
  n_\mathrm{units}, n_\mathrm{layers}, d_\mathrm{mlp}, n_\mathrm{mlp}} \in \mathbf{S}^\mathrm{rnc}$
  is to be interpreted as follows:
  \begin{itemize}
    \item
      $d_\mathrm{in}$ is the input dimension, hence the input dimension of the first layer;
    \item
      $d_\mathrm{out}$ is the output dimension, hence the output dimension of the last layer;
    \item
      $d_\mathrm{model}$ is the dimension of the output of each layer except the last one, and also
      the input of all layers except the first one;
    \item
      $d_\mathrm{state}$ is the state dimension of each MinMax Recurrent Unit, hence the state
      degree of the RNC;
    \item
      $n_\mathrm{units}$ is the number of MinMax Recurrent Units in each layer;
    \item
      $n_\mathrm{layers}$ is the number of MinMax Layers;
    \item
      $d_\mathrm{mlp}$ is the dimension of the hidden layers of each MLP;
    \item
      $n_\mathrm{mlp}$ is the number of hidden layers of each MLP.
  \end{itemize}

  The weights of a MinMax RNC of size $\mathbf{s}$ are
  \begin{align*}
    \mathbf{W}^\mathrm{rnc}_\mathbf{s}
    \defeq \{
      \tuple{W_0, \ldots, W_{n_\mathrm{layers}-1}} \mid W_i \in \mathbf{W}^\mathrm{layer}_{\mathbf{s}_i}
    \},
  \end{align*}
  where
  \begin{align*}
    \mathbf{s}_0 & = \tuple{d_\mathrm{in},d_\mathrm{model}, d_\mathrm{state},
    n_\mathrm{units},d_\mathrm{mlp},n_\mathrm{mlp}},
    \\
    \mathbf{s}_i & = \tuple{d_\mathrm{model},d_\mathrm{model}, d_\mathrm{state},
    n_\mathrm{units},d_\mathrm{mlp},n_\mathrm{mlp}},
    \quad \forall i \in \iival{1}{n_\mathrm{layers}-2},
    \\
    \mathbf{s}_{n_\mathrm{layers}-1} & = \tuple{d_\mathrm{model},d_\mathrm{out}, d_\mathrm{state},
    n_\mathrm{units},d_\mathrm{mlp},n_\mathrm{mlp}}.
  \end{align*}
  The MinMax RNC having weights $W = \tuple{W_0, \ldots, W_{n_\mathrm{layers}-1}} \in
  \mathbf{W}^\mathrm{rnc}_\mathbf{s}$, denoted by $\operatorname{RNC}(W)$,
  is the MinMax RNC $S_0 \ltimes \cdots \ltimes S_{n_\mathrm{layers}-1}$ where  
  $S_i = \operatorname{Layer}(W_i)$.
\end{closingdefinition}

\begin{closingdefinition}
  The computation problem of \emph{evaluating MinMax RNCs}
  has valid inputs and possible outputs respectively given by
  \begin{align*}
    I & \defeq \bigcup_{\mathbf{s} \in \mathbf{S}^\mathrm{rnc}} \bigcup_{T=1}^\infty I_{\mathbf{s},T}
    \qquad \text{ with }\;
    I_{\mathbf{s},T} \defeq \{
      \tuple{W,U} \mid 
      W \in \mathbf{W}^\mathrm{rnc}_{\mathbf{s}},\, 
      U \in \reals^{T \times d_\mathrm{in}}
    \},
    \\
    O & \defeq \bigcup_{d_\mathrm{out}=1}^\infty \bigcup_{T=1}^\infty
    \qquad \text{ with }\;
    O_{d_\mathrm{out},T} \defeq \{ Y \mid 
      Y \in \reals^{T \times d_\mathrm{out}}
    \},
  \end{align*}
  and the expected output for input $\tuple{W,U} \in I_{\mathbf{s},T}$ is the sequence $Y$
  satisfying the following, for $S = \operatorname{RNC}(W)$,
  \begin{equation*}
    \idx{Y}{t} = S(\idx{U}{0} \cdots \idx{U}{t}) \quad \forall t \in \iival{0}{T-1}.
  \end{equation*}
  For inputs in $I_{\mathbf{s},T}$, we call $\mathbf{s}$ the \emph{RNC size} and $T$ the
   \emph{input-sequence length} or \emph{number of steps}.
\end{closingdefinition}

\subsection{Proofs}
\label{subsec:complexity-proofs}

In this section we prove the theorems of 
Section~\ref{sec:main-body-complexity}.
\begin{itemize}
  \item
Theorem~\ref{thm:complexity-minmax-recurrence-main-body} 
is given by
Lemma~\ref{lemma:sequential-evaluation-complexity}
for the sequential case,
and by Lemma~\ref{lemma:parallel-evaluation-complexity} for the parallel case.

\item
Theorem~\ref{thm:complexity-main-sequential} is given by its more formal variant
Theorem~\ref{th:complexity-sequential-evaluation}.

\item Theorem~\ref{thm:complexity-main-parallel}
  is given by its more formal and more general variant
  Theorem~\ref{theorem:parallel-evaluation-complexity-cascade}, which also show the complexity
  without collapsing any quantity describing the size of the RNC, also without fixing the number of
  hidden layers of MLPs.
\end{itemize}

\subsubsection{Sequential}

\begin{lemma} \label{lemma:sequential-evaluation-complexity}
  \ref{alg:sequential-evaluation} solves
  \nameref{def:minmax-recurrence-evaluation-problem} in
  time $O(TN^2)$ for $T$ the number of steps and $N$ the state dimension.
\end{lemma}
\begin{proof}
  Let $\tuple{X_\mathrm{init}, A, B}$ be an input to the problem.

  \paragraph{Correctness.}
  We show that \ref{alg:sequential-evaluation} solves the problem.

  To distinguish variables of the algorithm from mathematical variables,
  let us add a hat to the variables of the algorithm. In particular,
  the input variables are $\tuple{\hat{X}_\mathrm{init}, \hat{A}, \hat{B}}$,
  and the other variables are $\hat{T},\hat{N},\hat{X},\hat{Z}$ (along with the loop indices).
  We note that the input variables 
  $\tuple{\hat{X}_\mathrm{init}, \hat{A}, \hat{B}}$ are never modified by the algorithm,
  and hence we can write $\hat{X}_\mathrm{init}$, $\hat{A}$, $\hat{B}$ to denote their 
  values, with no ambiguity.
  Furthermore,
  the scalar variables $\hat{T},\hat{N}$ as well as
  the entries of tensors $\hat{X}$ and $\hat{Z}$ are written at most once and never read
  before they are written, and hence
  we can write $\hat{T},\hat{N},\idx{\hat{X}}{t,i},\idx{\hat{Z}}{t,i,j}$ to denote the values held
  by the variables, with no ambiguity when such values have been assigned. 

  Let us define $X$ as the tensor of shape $(T+1,N,N)$ 
  satisfying $\idx{X}{0} = X_\mathrm{init}$ and 
  $\idx{X}{t} = \minmax(\idx{X}{t-1}, \idx{A}{t-1}, \idx{B}{t-1})$
  for every $t \in \iival{1}{T}$.

  It suffices to show that $\idx{\hat{X}}{0:T-1} = \idx{X}{1:T}$ when Line~11 is executed.

  We have that the value of the input variables
  $\tuple{\hat{X}_\mathrm{init}, \hat{A}, \hat{B}}$ is 
  $\tuple{X_\mathrm{init}, A, B}$.
  Then, after executing Line~1, the value of $\hat{T}$ is $T$ and the value of $\hat{N}$ is $N$.

  We first show by induction on $t$ from $0$ to $T$ that after having executed Line~1--4 and after
  $t$ iterations of the loop of
  Line~5 the equalities $\idx{\hat{Z}}{0:t,N} = \idx{X}{0:t}$ and 
  $\idx{\hat{X}}{0:t-1} = \idx{X}{1:t}$ hold.
  The base case $t=0$ holds since, after having executed Line~1--4 and before executing any iteration of
  the loop, the equality $\idx{\hat{Z}}{0,N} = \idx{X}{0} = X_\mathrm{init}$ holds by Line~4,
  and the equality $\idx{\hat{X}}{0:-1} = \idx{X}{1:0}$ holds vacously.
  In the inductive case we have $t>0$, hence the loop has been executed $t$ times, and we assume by
  induction that the equalities
  $\idx{\hat{Z}}{0:t-1,N} = \idx{X}{0:t-1}$ and 
  $\idx{\hat{X}}{0:t-2} = \idx{X}{1:t-1}$ hold.
  Letting $i \in \iival{1}{N}$, we show that
  that the $i$-th iteration of the loop of Line~6 ensures that
  the equality $\idx{\hat{Z}}{t,N,0:i-1} = \idx{X}{t,0:i-1}$ holds.
  We have that right after executing Line~7, $\idx{\hat{Z}}{t,0,i-1} = \idx{B}{t-1,i-1}$.
  We proceed by a second induction on $j$ from $0$ to $N$,
  showing that that after $j$ iterations of the loop of Line~8, 
  the following equality holds: 
  \begin{align*}
    \idx{\hat{Z}}{t,j,i-1} =  \idx{B}{t-1,i-1} \oplus \bigoplus_{\ell=0}^{j-1} \big(
    \idx{A}{t-1,i-1,\ell} \odot \idx{X}{t-1,\ell}\big).
  \end{align*}
  In the base case we have $j=0$, and the equality holds by the assignment of Line~7.
  The inductive case $j>0$ holds since
  \begin{align*}
    & \idx{\hat{Z}}{t,j,i-1} 
    \\[.3em]
    & = 
    \idx{\hat{Z}}{t,j-1,i-1} \oplus 
    \big(\idx{A}{t-1,i-1,j-1} \odot \idx{\hat{Z}}{t-1,N,j-1}\big)
    \\[.3em]
    & =
    \idx{\hat{Z}}{t,j-1,i-1} \oplus 
    \big(\idx{A}{t-1,i-1,j-1} \odot \idx{X}{t-1,j-1}\big)
    \mathsidecomment{first inductive hypothesis}
    \\[.3em]
    & =
    \idx{B}{t-1,i-1} \oplus \left(\bigoplus_{\ell=0}^{j-2} \big(
    \idx{A}{t-1,i-1,\ell} \odot \idx{X}{t-1,\ell}\big) \right)
    \oplus
    \big(\idx{A}{t-1,i-1,j-1} \odot \idx{X}{t-1,j-1}\big)
    \mathsidecomment{second inductive hypothesis}
    \\[.3em]
    & =
    \idx{B}{t-1,i-1} \oplus \bigoplus_{\ell=0}^{j-1} \big(
    \idx{A}{t-1,i-1,\ell} \odot \idx{X}{t-1,\ell}\big)
    \mathsidecomment{associativity of $\oplus$}
  \end{align*}

  Therefore,
  \begin{align*}
    \idx{\hat{Z}}{t,N}
    & = 
    \begin{pmatrix}
      \idx{B}{t-1,0}
      \\
      \vdots
      \\
      \idx{B}{t-1,N-1}
    \end{pmatrix}
    \oplus 
    \begin{pmatrix}
      \bigoplus_{j=0}^{N-1} \big(\idx{A}{t-1,0,j} \odot \idx{X}{t-1,j} \big)
      \\
      \vdots
      \\
      \bigoplus_{j=0}^{N-1} \big(\idx{A}{t-1,N-1,j} \odot \idx{X}{t-1,j} \big) 
    \end{pmatrix}
    \\[.3em]
    & 
    =  
    \idx{B}{t-1}
    \oplus 
    (\idx{A}{t-1} \otimes \idx{X}{t-1})
    \\[.3em]
    & 
    = 
    (\idx{A}{t-1} \otimes \idx{X}{t-1})
    \oplus 
    \idx{B}{t-1}
    \\[.3em]
    & 
    = 
    \idx{X}{t}.
  \end{align*}
  Hence, by Line~10, we have $\idx{\hat{X}}{t-1} = \idx{\hat{Z}}{t,N} = \idx{X}{t}$,
  which proves the induction on $t$.
  Therefore, $\idx{\hat{X}}{t-1} = \idx{X}{t}$ for every $t \in \iival{1}{T}$, as required.

  \paragraph{Execution time.}
  Since reading and writing a real number, as well as executing an elementary operation on real
  numbers, take constant time in the assumed model of computation,
  we have that the execution time is $O(\ell)$ for $\ell$ the total number of times Line~9 is
  executed, and hence it is $O(TN^2)$ as claimed.
\end{proof}

\begin{lemma} \label{lemma:sequential-evaluation-complexity-layer}
  On every input $\tuple{W,U}$ for $W \in \mathbf{W}^\mathrm{layer}_\mathbf{s}$ of size
  $\mathbf{s} = \tuple{d_\mathrm{in},d_\mathrm{mlp},n_\mathrm{mlp},d_\mathrm{out},n_\mathrm{units},
  d_\mathrm{state}}$, and $U \in \reals^{T \times d_\mathrm{in}}$,
  the output of \ref{alg:sequential-evaluation-layer} is
  $S(U)$ for $S = \operatorname{Layer}(W)$,
  and its execution time is 
  \begin{align*}
      O\Big( T \cdot \big( 
      n_\mathrm{units} \cdot d_\mathrm{state}^2 +
      d_\mathrm{mlp} \cdot (d_\mathrm{in} +  d_\mathrm{out} + n_\mathrm{mlp} d_\mathrm{mlp} +
      d_\mathrm{state}^2 + n_\mathrm{units} d_\mathrm{state}) \big) \Big)
  \end{align*}
\end{lemma}
\begin{proof}
  We first show correctness and then complexity.
  We have that the dynamics $D$ of $S$ are given by a parallel composition
  $D_1 \parallel \cdots \parallel D_n$ with each $D_i$ a MinMax Recurrent Unit.

  \paragraph{Correctness.}
  After Line~1, the values of $T,d_\mathrm{in}, n_\mathrm{units}, d_\mathrm{state}, d_\mathrm{out}$
  are correctly set according to the shape of $U$ and the definition of 
  $W \in \mathbf{W}^\mathrm{layer}_\mathbf{s}$ (Definition~\ref{def:rnc-representation}).
  After Line~2, the values of $W^\mathrm{R},W^\mathrm{s},W^\mathrm{out},X^\mathrm{init}$
  are correctly set according to the definition of 
  $W \in \mathbf{W}^\mathrm{layer}_\mathbf{s}$ (Definition~\ref{def:rnc-representation}).
  Then the algorithm proceeds to computing the states of the MinMax Layer $S$ on the given input
  $U$.
  Each MinMax Unit of the layer has state dimension $d_\mathrm{state}$, and hence
  the tensor allocated in Line~3 suffices to store the initial state and the subsequent states of
  all units.
  Let us now consider an iteration of the loop of Lines~4--13 for index 
  $i \in \iival{0}{n_\mathrm{units}-1}$.
  Each iteration is independent from the others, as it does not read values written in other
  iterations,
  nor assigns a value to a memory location that is also written in some other iteration.
  After executing Lines~5--9, we have that, for every $t \in \iival{0}{T-1}$, 
  the value of 
  $\idx{A_i}{t}$ is $R_i(\idx{U}{t})$ for $R_i = \operatorname{MLP}(\idx{W^\mathrm{R}}{i})$ the
  reset function of component $i$ of the layer,
  and
  the value of 
  $\idx{B_i}{t}$ is $s_i(\idx{U}{t})$ for $s_i = \operatorname{MLP}(\idx{W^\mathrm{s}}{i})$ the
  set function of component $i$ of the layer.
  After executing Lines~10--12,
  we have that 
  $\idx{X}{0,\, i\cdot d_\mathrm{state} : (i+1)\cdot d_\mathrm{state}} =
  \idx{X^\mathrm{init}}{i}$,
  and 
  after executing Line~13,
  by Theorem~\ref{lemma:sequential-evaluation-complexity}
  we have that 
  $\idx{X}{t+1,\, i\cdot d_\mathrm{state} : (i+1)\cdot d_\mathrm{state}} =
  D_i(\idx{X^\mathrm{init}}{i}, \idx{U}{0:t})$ for every $t \in \iival{0}{T-1}$.
  Thus, after the execution of the loop of Lines~4--13 completes,
  we have that 
  $\idx{X}{t+1} = D(X^\mathrm{init}, \idx{U}{0:t})$ for every $t \in \iival{0}{T-1}$.

  Next, let us consider an iteration of the loop of Lines~15--16 for $t \in \iival{0}{T-1}$, noting
  that each iteration is independent from the others.
  We have that 
  \begin{align*}
  \idx{X}{t} \parallel \idx{X}{t+1} \parallel \idx{U}{t} = 
  \tuple{D(X^\mathrm{init}, \idx{U}{0:t-1}),\, D(X^\mathrm{init}, \idx{U}{0:t}),\, \idx{U}{t}},
  \end{align*}
  and hence after executing Line~16, we have 
  \begin{align*}
    \idx{Y}{t} = 
    h\big(D(X^\mathrm{init}, \idx{U}{0:t-1}),\, D(X^\mathrm{init}, \idx{U}{0:t}),\, \idx{U}{t}\big)
  \end{align*}
  for $h = \operatorname{MLP}(W^\mathrm{out})$ the output function of $S$.
  Thus, after the execution of the loop of Lines~15--16 completes,
  we have that $Y = S(U)$ as required.

  \paragraph{Complexity.}
  The execution time of the algorithm is asymptotically bounded by the sum of the following
  execution times:
  \begin{itemize}
    \item
      Line~8: $O(n_\mathrm{units} \cdot T)$ executions of $\mathtt{EvaluateMLP}$,
      where the MLP has 
      input dimension $d_\mathrm{in}$,
      output dimension $d_\mathrm{state}^2$,
      and
      $n_\mathrm{mlp}$ hidden layers of dimension $d_\mathrm{mlp,u} \defeq \lceil
      d_\mathrm{mlp}/\sqrt{n_\mathrm{units}}\rceil$;
      the time of each evaluation is $O(d_\mathrm{in} d_\mathrm{mlp,u} + n_\mathrm{mlp}
        d_\mathrm{mlp,u}^2 + d_\mathrm{mlp,i} d_\mathrm{state}^2)$;
    \item
      Line~9: $O(n_\mathrm{units} \cdot T)$ executions of $\mathtt{EvaluateMLP}$ 
      where the MLP has 
      input dimension $d_\mathrm{in}$,
      output dimension $d_\mathrm{state}$,
      and
      $n_\mathrm{mlp}$ hidden layers of dimension $d_\mathrm{mlp,u} \defeq \lceil
      d_\mathrm{mlp}/n_\mathrm{units}\rceil$;
      the time of each evaluation is $O(d_\mathrm{in} d_\mathrm{mlp,u} + n_\mathrm{mlp}
        d_\mathrm{mlp,u}^2 +
      d_\mathrm{mlp,u} d_\mathrm{state})$;
    \item
      Line~13: $O(n_\mathrm{units})$ executions of \ref{alg:sequential-evaluation} on inputs of
      dimension $d_\mathrm{state}$ and length $T$;
      its execution time is $O(T d_\mathrm{state}^2)$ by
      Theorem~\ref{lemma:sequential-evaluation-complexity};
    \item
      Line~16: $O(T)$ executions of $\mathtt{EvaluateMLP}$,
      where the MLP has 
      input dimension $2 n_\mathrm{units} d_\mathrm{state} + d_\mathrm{in}$,
      output dimension $d_\mathrm{out}$,
      and
      $n_\mathrm{mlp}$ hidden layers of dimension $d_\mathrm{mlp}$;
      the time of each evaluation is 
      $O\big((n_\mathrm{units} d_\mathrm{state} + d_\mathrm{in}) \cdot d_\mathrm{mlp} + 
        n_\mathrm{mlp} d_\mathrm{mlp}^2 + d_\mathrm{mlp} d_\mathrm{out})$.
  \end{itemize}
  Thus, the execution time is

  \begin{align*}
    & O(n_\mathrm{units} \cdot T) 
    \cdot
    O(d_\mathrm{in} d_\mathrm{mlp,u} + n_\mathrm{mlp} d_\mathrm{mlp,u}^2 +
    d_\mathrm{mlp,u} d_\mathrm{state}^2) + {}
      \\
      &
    O(n_\mathrm{units} \cdot T)
    \cdot
    O(d_\mathrm{in} d_\mathrm{mlp,u} + n_\mathrm{mlp} d_\mathrm{mlp,u}^2 +
    d_\mathrm{mlp,u} d_\mathrm{state}) + {}
      \\
      &
    O(n_\mathrm{units})
    \cdot
    O(T d_\mathrm{state}^2) + {}
      \\
       &
      O(T)
      \cdot
      O\big((n_\mathrm{units} d_\mathrm{state} + d_\mathrm{in}) \cdot d_\mathrm{mlp} + 
      n_\mathrm{mlp} d_\mathrm{mlp,u}^2 + d_\mathrm{mlp} d_\mathrm{out}\big)
      \\
      & = 
      \\
      & O\Big(n_\mathrm{units} \cdot T \cdot
      (d_\mathrm{in} d_\mathrm{mlp,u} + n_\mathrm{mlp} d_\mathrm{mlp,u}^2 +
      d_\mathrm{mlp,u} d_\mathrm{state}^2) \Big) + {}
      \\
       & 
      O\Big( n_\mathrm{units} \cdot T
      \cdot
      (d_\mathrm{in} d_\mathrm{mlp,u} + n_\mathrm{mlp} d_\mathrm{mlp,u}^2 +
      d_\mathrm{mlp,u} d_\mathrm{state}) \Big) + {}
      \\
      & 
      O\Big(n_\mathrm{units} \cdot T \cdot d_\mathrm{state}^2\Big) + {}
      \\
      & 
    O\Big(T \cdot \big((n_\mathrm{units} d_\mathrm{state} + d_\mathrm{in}) \cdot d_\mathrm{mlp} + 
    n_\mathrm{mlp} d_\mathrm{mlp}^2 + d_\mathrm{mlp} d_\mathrm{out} \big) \Big)
      \\
      & = 
      \\
      & O\Big(n_\mathrm{units} \cdot T \cdot
      (d_\mathrm{in} d_\mathrm{mlp,u} + n_\mathrm{mlp} d_\mathrm{mlp,u}^2 +
      d_\mathrm{mlp,u} d_\mathrm{state}^2) \Big) + {}
      \\
      & 
      O\Big(n_\mathrm{units} \cdot T \cdot d_\mathrm{state}^2\Big) + {}
      \\
      & 
    O\Big(T \cdot \big((n_\mathrm{units} d_\mathrm{state} + d_\mathrm{in}) \cdot d_\mathrm{mlp} + 
    n_\mathrm{mlp} d_\mathrm{mlp}^2 + d_\mathrm{mlp} d_\mathrm{out} \big) \Big)
      \\
      & = 
      \\
      & O\Big(T \cdot
      (d_\mathrm{in} d_\mathrm{mlp} + n_\mathrm{mlp} d_\mathrm{mlp}^2 +
      d_\mathrm{mlp} d_\mathrm{state}^2) \Big) + {}
      \\
      & 
      O\Big(n_\mathrm{units} \cdot T \cdot d_\mathrm{state}^2\Big) + {}
      \\
      & 
    O\Big(T \cdot \big((n_\mathrm{units} d_\mathrm{state} + d_\mathrm{in}) \cdot d_\mathrm{mlp} + 
    n_\mathrm{mlp} d_\mathrm{mlp}^2 + d_\mathrm{mlp} d_\mathrm{out} \big) \Big)
      \\
      & = 
      \\
      & O\Big(T \cdot
      (d_\mathrm{in} d_\mathrm{mlp} + n_\mathrm{mlp} d_\mathrm{mlp}^2 +
      d_\mathrm{mlp} d_\mathrm{state}^2) + {}
      \\
      & 
      \qquad n_\mathrm{units} \cdot T \cdot d_\mathrm{state}^2 + {}
      \\
      & 
    \qquad T \cdot \big((n_\mathrm{units} d_\mathrm{state} + d_\mathrm{in}) \cdot d_\mathrm{mlp} + 
    n_\mathrm{mlp} d_\mathrm{mlp}^2 + d_\mathrm{mlp} d_\mathrm{out} \big) \Big)
      \\
      & = 
      \\
      & O\Big( T \cdot \big( 
      d_\mathrm{in} d_\mathrm{mlp} + n_\mathrm{mlp} d_\mathrm{mlp}^2 +
      d_\mathrm{mlp} d_\mathrm{state}^2 +
       n_\mathrm{units} \cdot d_\mathrm{state}^2 + {}
    n_\mathrm{units} \cdot d_\mathrm{mlp} d_\mathrm{state} + d_\mathrm{mlp} d_\mathrm{out} \big) \Big)
      \\
      & = 
      \\
      & O\Big( T \cdot \big( 
      n_\mathrm{units} \cdot d_\mathrm{state}^2 +
      d_\mathrm{mlp} \cdot (d_\mathrm{in}  + n_\mathrm{mlp} d_\mathrm{mlp} +
      d_\mathrm{state}^2 +  d_\mathrm{out}+ n_\mathrm{units}  d_\mathrm{state})    \big) \Big)
  \end{align*}
\end{proof}

\begin{lemma} \label{lemma:sequential-evaluation-complexity-cascade}
  On input $\tuple{W,U}$ for $W \in \mathbf{W}^\mathrm{rnc}_\mathbf{s}$ of size
  $$\mathbf{s} = \tuple{d_\mathrm{in}, d_\mathrm{out}, d_\mathrm{model}, d_\mathrm{state},
  n_\mathrm{units}, n_\mathrm{layers}, d_\mathrm{mlp}, n_\mathrm{mlp}},$$ 
  and $U \in \reals^{T \times d_\mathrm{in}}$,
  the output of \ref{alg:sequential-evaluation-rnc} is
  $S(U)$ for $S = \operatorname{RNC}(W)$,
  and its execution time is 
  \begin{align*}
    O\Big( T \cdot n_\mathrm{layers} \cdot \big( 
      n_\mathrm{units} \cdot d_\mathrm{state}^2 +
      d_\mathrm{mlp} \cdot (d_\mathrm{in} + d_\mathrm{model} + d_\mathrm{out} + n_\mathrm{mlp}
      d_\mathrm{mlp} + d_\mathrm{state}^2 + n_\mathrm{units} d_\mathrm{state}) \big) \Big),
  \end{align*}
  and for fixed $n_\mathrm{mlp}$ and $d \defeq \max\{ d_\mathrm{in},d_\mathrm{out},d_\mathrm{mlp},
  n_\mathrm{units} \cdot d_\mathrm{state}\}$,
  \begin{align*}
      O\Big( T \cdot  n_\mathrm{layers} \cdot d \cdot \big( d + d_\mathrm{state}^2 ) \big) \Big).
  \end{align*}
\end{lemma}
\begin{proof}
  We first show correctness and then complexity.
  We have $W = \tuple{W_1, \ldots, W_{n_\mathrm{layers}-1}}$
  and that the RNC $S$ is given by the cascade composition
  $S_0 \ltimes \cdots \ltimes S_{n_\mathrm{layers}-1}$ where each layer is 
  $S_i = \operatorname{Layer}(W_i)$.

  \paragraph{Correctness.}
  After Line~1, the value of $n_\mathrm{layers}$
  is correctly set according to the definition of 
  $W \in \mathbf{W}^\mathrm{rnc}_\mathbf{s}$ (Definition~\ref{def:rnc-representation}).
  Right after Line~2, the value of $Y$ is $U$.
  Thus, after the $i$-th iteration of the loop,
  the value of $Y$ is $S_{0:i}(U)$ for $S_{0:i} = S_0 \ltimes \cdots \ltimes S_i$,
  by Theorem~\ref{lemma:sequential-evaluation-complexity-layer}.
  Therefore, when the execution of the loops completes, 
  the value of $Y$ is $S(U)$ as required.

  \paragraph{Complexity.}
  The execution time is given by $O(n_\mathrm{layers})$ calls of $\mathtt{SeqLayerEval}$
  to evaluate a layer having 
  input dimension $d_\mathrm{in}$ and output dimension $d_\mathrm{model}$ for $i = 0$,
  input and output dimension $d_\mathrm{model}$ for each $i \in \iival{1}{n_\mathrm{layers}-2}$,
  and
  input dimension $d_\mathrm{model}$ and output dimension $d_\mathrm{out}$ for $i =
  n_\mathrm{layers}-1$.
  For the rest, the size of each layer is described by
  $\tuple{d_\mathrm{state}, n_\mathrm{units}, d_\mathrm{mlp}, n_\mathrm{mlp}}$.

  By Theorem~\ref{lemma:sequential-evaluation-complexity-layer}, 
  the execution time for the first layer is
  \begin{align*}
      O\Big( T \cdot \big( 
      n_\mathrm{units} \cdot d_\mathrm{state}^2 +
      d_\mathrm{mlp} \cdot (d_\mathrm{in} +  d_\mathrm{model} + n_\mathrm{mlp} d_\mathrm{mlp} +
      d_\mathrm{state}^2 + n_\mathrm{units} d_\mathrm{state}) \big) \Big),
  \end{align*}
  the execution time for each layer $i \in \iival{1}{n_\mathrm{layers}-2}$
  is
  \begin{align*}
      O\Big( T \cdot \big( 
      n_\mathrm{units} \cdot d_\mathrm{state}^2 +
      d_\mathrm{mlp} \cdot (d_\mathrm{model} +  d_\mathrm{model} + n_\mathrm{mlp} d_\mathrm{mlp} +
      d_\mathrm{state}^2 + n_\mathrm{units} d_\mathrm{state}) \big) \Big),
  \end{align*}
  and
  the execution time for layer $i = n_\mathrm{layers}-1$ is
  \begin{align*}
      O\Big( T \cdot \big( 
      n_\mathrm{units} \cdot d_\mathrm{state}^2 +
      d_\mathrm{mlp} \cdot (d_\mathrm{model} +  d_\mathrm{out} + n_\mathrm{mlp} d_\mathrm{mlp} +
      d_\mathrm{state}^2 + n_\mathrm{units} d_\mathrm{state}) \big) \Big).
  \end{align*}
  Thus, overall we have
  \begin{align*}
    O\Big( T \cdot n_\mathrm{layers} \cdot \big( 
      n_\mathrm{units} \cdot d_\mathrm{state}^2 +
      d_\mathrm{mlp} \cdot (d_\mathrm{in} + d_\mathrm{model} + d_\mathrm{out} + n_\mathrm{mlp}
      d_\mathrm{mlp} + d_\mathrm{state}^2 + n_\mathrm{units} d_\mathrm{state}) \big) \Big).
  \end{align*}
  For fixed $n_\mathrm{mlp}$ and $d \defeq \max\{ d_\mathrm{in},d_\mathrm{out},d_\mathrm{mlp},
  n_\mathrm{units} \cdot d_\mathrm{state}\}$,
  \begin{align*}
    &
    O\Big( T \cdot n_\mathrm{layers} \cdot \big( 
      n_\mathrm{units} \cdot d_\mathrm{state}^2 + d \cdot (d + d_\mathrm{state}^2)  \big) \Big)
      \\
      & =
      O\Big( T \cdot n_\mathrm{layers} \cdot \big( 
      d \cdot d_\mathrm{state} + d \cdot (d + d_\mathrm{state}^2)  \big) \Big)
      \\
      & =
      O\Big( T \cdot n_\mathrm{layers} \cdot d \cdot \big( d + d_\mathrm{state}^2 \big) \Big).
  \end{align*}
\end{proof}

\begin{restatable}{theorem}{thcomplexitysequentialevaluation}
  \label{th:complexity-sequential-evaluation}
  \nameref{def:minmax-recurrence-evaluation-problem} can be solved sequentially in time $O(TN^2)$ 
  for $T$ the number of steps and $N$ the state dimension.
\end{restatable}
\begin{proof}
  Immediate by Lemma~\ref{lemma:sequential-evaluation-complexity}.
\end{proof}

\subsubsection{Auxiliary}

\begin{theorem} \label{thm:complexity-parallel-min}
  On any given input $A \in \reals^n$,
  the output of $\ParallelMin$ (Algorithm~\ref{alg:parallel-min}) is
  $\bigodot_{i=0}^{n-1} \idx{A}{i}$,
  its depth is $O(\log n)$, 
  and 
  its work is $O(n)$.
\end{theorem}
\begin{proof}
  We first show correctness and then complexity.

  \paragraph{Correctness.}
  We proceed by induction on $n$.
  In the base case $n=1$ and we have that the function returns $\idx{A}{0}$ as required
  (Lines~2--3).
  In the inductive case we have $n>1$ and we assume that the 
  output of function $\ParallelMin$ on input $B$ is $\bigodot_{i=0}^{m-1} \idx{B}{i}$ for every
  input $B$ of length $m \leq n-1$.
  We consider two cases separately, according to whether $n$ is even or odd.

  In the first case, $n$ is even. 
  Right after Line~5, we have
  \begin{align*}
    m & = \lfloor n/2 \rfloor = n/2.
  \end{align*}
  We have that right after Lines~7--8, for every $i \in \iival{0}{m-2}$, 
  the equality $\idx{\hat{B}}{i} = \idx{A}{2i} \odot \idx{A}{2i+1}$ holds.
  Then, in Lines~9--12, we have that Line~10 is executed,
  and hence $\idx{\hat{B}}{m-1} = \idx{A}{n-2} \odot \idx{A}{n-1}$.
  Let us define the following sets of indices
  \begin{align*}
    I_0 & \defeq
    \{ 2i \mid i \in \iival{0}{m-2} \} = \{ 0, 2, \ldots, 2m-4 \} = \{ 0,2,\ldots,n-4\},
    \\[.3em]
    I_1 & \defeq
    \{ 2i+1 \mid i \in \iival{0}{m-2} \} = \{ 1, 3, \ldots, 2m-3 \} = \{ 1,3,\ldots,n-3\}.
  \end{align*}
  We note that $I_0 \cap I_1 = \emptyset$ and $I_0 \cup I_1 = \iival{0}{n-3}$.
  Then, by Line~13, the value returned by $\ParallelMin(A)$ is the value returned by 
  $\ParallelMin(\hat{B})$, which is as follows by the inductive hypothesis since the length of
  $\hat{B}$ is
  smaller than $n$,
  \begin{align*}
    \bigodot_{i = 0}^{m-1} \idx{\hat{B}}{i}
    & =
    \idx{\hat{B}}{m-1} \odot \bigodot_{i = 0}^{m-2} \idx{\hat{B}}{i}
    \\[.3em]
    & =
    \idx{A}{n-2} \odot \idx{A}{n-1} \odot \bigodot_{i = 0}^{m-2} \idx{A}{2i} \odot \idx{A}{2i+1}
    \\[.3em]
    & = 
    \idx{A}{n-2} \odot \idx{A}{n-1} \odot 
    \left(\bigodot_{i = 0}^{m-2} \idx{A}{2i}\right) \odot \left(\bigodot_{i = 0}^{m-2} \idx{A}{2i+1}\right)
    \\[.3em]
    & =
    \idx{A}{n-2} \odot \idx{A}{n-1} \odot 
    \left(\bigodot_{i\in I_0} \idx{A}{i}\right) 
    \odot 
    \left(\bigodot_{i\in I_1} \idx{A}{i}\right) 
    \\[.3em]
    & =
    \bigodot_{i=0}^{n-1} \idx{A}{i},
  \end{align*}
  as required.

  In the second case $n$ is odd, hence $n = n'+1$ with $n'$ even.
  Right after Line~5,
  we have
  \begin{align*}
    m & = \lfloor n/2 \rfloor = \lfloor n'/2 + 1/2 \rfloor = n'/2.
  \end{align*}
  We have that right after Lines~7--8, for every $i \in \iival{0}{m-2}$, 
  the equality $\idx{\hat{B}}{i} = \idx{A}{2i} \odot \idx{A}{2i+1}$ holds.
  Then, in Lines~9--12, we have that Line~12 is executed,
  and hence $\idx{\hat{B}}{m-1} = \idx{A}{n-3} \odot \idx{A}{n-2} \odot \idx{A}{n-1}$.
  Let us define the following sets of indices
  \begin{align*}
    I_0 & \defeq
    \{ 2i \mid i \in \iival{0}{m-2} \} = \{ 0, 2, \ldots, 2m-4 \} 
    = \{ 0,2,\ldots,n'-4\}
    = \{ 0,2,\ldots,n-5\},
    \\[.3em]
    I_1 & \defeq
    \{ 2i+1 \mid i \in \iival{0}{m-2} \} = \{ 1, 3, \ldots, 2m-3 \} 
    = \{ 1,3,\ldots,n'-3\}
    = \{ 1,3,\ldots,n-4\}.
  \end{align*}
  We note that $I_0 \cap I_1 = \emptyset$ and $I_0 \cup I_1 = \iival{0}{n-4}$.
  Then, by Line~13, the value returned by $\ParallelMin(A)$ is the value returned by 
  $\ParallelMin(\hat{B})$, which is as follows by the inductive hypothesis since the length of $B$ is
  smaller than $n$,
  \begin{align*}
    \bigodot_{i = 0}^{m-1} \idx{\hat{B}}{i}
    & = 
    \idx{\hat{B}}{m-1} \odot \bigodot_{i = 0}^{m-2} \idx{\hat{B}}{i}
    \\[.3em]
    & = 
    \idx{A}{n-3} \odot \idx{A}{n-2} \odot \idx{A}{n-1} \odot \bigodot_{i = 0}^{m-2} \idx{A}{2i} \odot \idx{A}{2i+1}
    \\[.3em]
    & = 
    \idx{A}{n-3} \odot \idx{A}{n-2} \odot \idx{A}{n-1} \odot 
    \left(\bigodot_{i = 0}^{m-2} \idx{A}{2i}\right) 
    \odot 
    \left(\bigodot_{i = 0}^{m-2} \idx{A}{2i+1}\right)
    \\[.3em]
    & =
    \idx{A}{n-3} \odot \idx{A}{n-2} \odot \idx{A}{n-1} \odot 
    \left(\bigodot_{i\in I_0} \idx{A}{i}\right) 
    \odot 
    \left(\bigodot_{i\in I_1} \idx{A}{i}\right) 
    \\[.3em]
    & =
    \bigodot_{i=0}^{n-1} \idx{A}{i},
  \end{align*}
  as required.

  \paragraph{Complexity.}
  The depth is given by
  \begin{align*}
    \operatorname{depth}(n) & =
    \begin{cases}
      O(1) & \text{ if } n=1,
      \\
      \operatorname{depth}(\lfloor n/2 \rfloor) + O(1) & \text{ otherwise}.
    \end{cases}
  \end{align*}
  The work is given by
  \begin{align*}
    \operatorname{work}(n) & =
    \begin{cases}
      O(1) & \text{ if } n=1,
      \\
      \operatorname{work}(\lfloor n/2 \rfloor) + O(n) & \text{ otherwise}.
    \end{cases}
  \end{align*}
  Therefore,
  $\operatorname{depth}(n) = O(\log n)$
  and
  $\operatorname{work}(n) = O(n)$ by the Master Theorem, cf.~\citep{cormen2022introduction}.
\end{proof}

\begin{theorem} \label{thm:complexity-parallel-max} \label{thm:complexity-parallel-sum}
  On any given input $A \in \reals^n$,
  the output of $\ParallelMax$ (Algorithm~\ref{alg:parallel-max}) is
  $\bigoplus_{i=0}^{n-1} \idx{A}{i}$,
  its depth is $O(\log n)$, 
  and 
  its work is $O(n)$.
\end{theorem}
\begin{proof}
  Omitted as it is analogous to $\ParallelMin$ (Theorem~\ref{thm:complexity-parallel-min}).
\end{proof}

\begin{theorem} \label{thm:complexity-parallel-add}
  On any given input $A,B \in \reals^{m \times n}$,
  the output of $\ParallelAdd$ (Algorithm~\ref{alg:parallel-sum}) is
  $C = A \oplus B$,
  its depth is $O(1)$, 
  and 
  its work is $O(m \cdot n)$.
\end{theorem}
\begin{proof}
  Correctness is immediate as the algorithm closely follows the definition of $A \oplus B$.
  Depth is $O(1)$ since the algorithm allocates $C$ (Line~2) which has constant cost,
  and then it executes Line~{5} in parallel of each value of $i,j$.
  Work is $O(m \cdot n)$ since the algorithm allocates $C$ (Line~2) which has constant cost,
  and then it executes Line~{5} for each value of $i \in \iival{0}{m-1}$ and $j \in \iival{0}{n-1}$,
  and hence it executes Line~{5} $m\cdot n$ times overall.
\end{proof}

\begin{theorem}\label{thm:complexity-parallel-mult}
  On any given input $A \in \reals^{m \times p}, B \in \reals^{p \times n}$,
  the output of $\ParallelMult$ (Algorithm~\ref{alg:parallel-mult}) is
  $C = A \otimes B$,
  its depth is $O(\log p)$, 
  and 
  its work is $O(m\cdot n \cdot p)$.
\end{theorem}
\begin{proof}
  We first show correctness and the complexity.

  \paragraph{Correctness.}
  We have that, for 
  each $i \in \iival{0}{m-1}$, 
  each $j \in \iival{0}{n-1}$,
  and
  each $k \in \iival{0}{p-1}$,
  after executing Line~7,
  it holds that
  $$\idx{T}{i,j,k} = \idx{A}{i,k} \odot \idx{B}{k,j}.$$
  Thus, by Theorem~\ref{thm:complexity-parallel-sum},
  for each $i \in \iival{0}{m-1}$
  and
  each $j \in \iival{0}{n-1}$,
  after executing Line~8,
  it holds that 
  \begin{align*}
    \idx{C}{i,j} 
    = \bigoplus_{k=0}^{p-1} \idx{T}{i,j,k} 
    = \bigoplus_{k=0}^{p-1} (\idx{A}{i,k} \odot \idx{B}{k,j}),
  \end{align*}
  which is the required value for $\idx{C}{i,j}$.

  \paragraph{Complexity.}
  The depth is $O(1)$ plus the maximum depth of $\ParallelMax$ on an input $\idx{T}{i,j}$.
  Since each $\idx{T}{i,j}$ has length $p$,
  by Theorem~\ref{thm:complexity-parallel-sum} we have that the depth of each call of $\ParallelMax$
  has depth $O(\log p)$.
  Thus, the algorithm has depth $O(\log p)$ overall.

  The work is $O((m \cdot n) \cdot (c + w))$ for $c$ the cost of Line~7 (which is executed $m
  \cdot n$ times) and for $w$ the maximum work of a call of $\ParallelMax$ on
  input $\idx{T}{i,j}$.
  Since each $\idx{T}{i,j}$ has length $p$,
  by Theorem~\ref{thm:complexity-parallel-sum} 
  we have that the work of each call of $\ParallelMax$ is $O(p)$.
  Thus, since $c$ is constant, the algorithm has work $O(m \cdot n \cdot p)$ overall.
\end{proof}

\begin{theorem} \label{thm:complexity-affine-comp}
  On any given input 
  $\ptuple{A_1, B_1} \in \reals^{n \times n}\times \reals^n$, $\ptuple{A_2,B_2} \in \reals^{n \times
  n}\times \reals^n$,
  the output of $\ParallelComp$ (Algorithm~\ref{alg:affine-comp}) is
  $\tuple{A,B} = \tuple{A_1,B_1} \minmaxcomp \tuple{A_2,B_2}$,
  its depth is $O(\log n)$, 
  and 
  its work is $O(n^3)$.
\end{theorem}
\begin{proof}
  We first show correctness and the complexity.

  \paragraph{Correctness.}
  By Theorem~\ref{thm:complexity-parallel-mult},
  after executing Line~4
  it holds that $A = A_2 \otimes A_1$,
  and
  after executing Line~5
  it holds that $Z = A_2 \otimes B_1$.
  Then,
  by Theorem~\ref{thm:complexity-parallel-add},
  after executing Line~6
  it holds that $B = Z \oplus B_2 = (A_2 \otimes B_1) \oplus B_2$.
  Thus, the returned pair $(A,B)$ is as required.

  \paragraph{Complexity.}
  The depth is $O(d_1 + d_2 + d_3)$ for 
  $d_1$ the depth of $\ParallelMult(A_2, A_1)$,
  $d_2$ the depth of $\ParallelMult(A_2, B_1)$,
  and
  $d_3$ the depth of $\ParallelAdd(Z, B_2)$.
  By Theorem~\ref{thm:complexity-parallel-mult},
  we have that $d_1$ and $d_2$ are $O(\log n)$.
  By Theorem~\ref{thm:complexity-parallel-add},
  we have that $d_3$ is $O(1)$.
  Therefore, the algorithm has depth $O(\log n + \log n + 1) = O(\log n)$.

  The work is $O(w_1 + w_2 + w_3)$ for 
  $w_1$ the work of $\ParallelMult(A_2, A_1)$,
  $w_2$ the work of $\ParallelMult(A_2, B_1)$,
  and
  $w_3$ the work of $\ParallelAdd(Z, B_2)$.
  By Theorem~\ref{thm:complexity-parallel-mult},
  we have that 
  $w_1$ is $O(n^3)$
  and
  $w_2$ is $O(n^2)$.
  By Theorem~\ref{thm:complexity-parallel-add},
  we have that $w_3$ is $O(n)$.
  Therefore, the algorithm has work $O(n^3 + n^2 + n) = O(n^3)$.
\end{proof}

\begin{theorem} \label{thm:complexity-parallel-apply}
  On any given input 
  $\ptuple{A, B, X} \in \reals^{n \times n} \times \reals^n \times \reals^n$,
  the output of $\ParallelApply$ (Algorithm~\ref{alg:parallel-apply}) is
  $C = (A \otimes X) \oplus B$,
  its depth is $O(\log n)$, 
  and 
  its work is $O(n^2)$.
\end{theorem}
\begin{proof}
  We first show correctness.
  After Line~2 we have that $Z = A \otimes X$ by Theorem~\ref{thm:complexity-parallel-mult}.
  Hence by Theorem~\ref{thm:complexity-parallel-add},
  after Line~3 $Y = Z \oplus B = (A \otimes X) \oplus$ as required.

  Next we analyse its complexity.
  The depth is $O(\log n)$ since it is the sum of the following depths:
  \begin{itemize}
    \item
      the depth of $\ParallelMult$ on input $(A,X)$, which is $O(\log n)$ by
      Theorem~\ref{thm:complexity-parallel-mult};
    \item
      the depth of $\ParallelAdd$ on input $(Z,B)$, which is $O(1)$ by
      Theorem~\ref{thm:complexity-parallel-add}.
  \end{itemize}
  The work is $O(n^2)$ since it is the sum of the following work:
  \begin{itemize}
    \item
      the work of $\ParallelMult$ on input $(A,X)$, which is $O(n^2)$ by
      Theorem~\ref{thm:complexity-parallel-mult};
    \item
      the work of $\ParallelAdd$ on input $(Z,B)$, which is $O(n)$ by
      Theorem~\ref{thm:complexity-parallel-add}.
  \end{itemize}
\end{proof}

\subsubsection{Parallel}

\paragraph{MinMax affine monoid.}
\label{par:complexity-minmax-affine-monoid}
We use the MinMax affine operators and their composition~$\minmaxcomp$
(Definition~\ref{def:minmax-affine-operator}), and, for $V \subseteq \reals$ with a minimum
and a maximum, the MinMax affine monoid $\mathcal{M}_{V,n}$ with identity
$\mathcal{E} = \tuple{E,e}$ and the maps $\tau_M = \restr{\actfun{M}}{V^n}$ of
Proposition~\ref{prop:minmax-affine-monoid}; we also write
$\tau(A,b,x) \defeq \tau_{\tuple{A,b}}(x)$.
By that proposition, $\tau_{M_1 \minmaxcomp M_2} = \tau_{M_2} \circ \tau_{M_1}$, and
$\tau_{\mathcal{E}}$ is the identity function on~$V^n$.

\begin{theorem}\label{thm:complexity-compute-identity}
  On input $(\vmin,\vmax,N)$, the output of 
  $\ComputeIdentity$ (Algorithm~\ref{alg:compute-identity})
  is the identity element $\mathcal{E}$ of the MinMax monoid $\mathcal{M}_{[\vmin,\vmax],N}$,
  its depth is $O(1)$ and its work is $O(N^2)$.
\end{theorem}
\begin{proof}
  Correctness is immediate as the algorithm closely follows the characterisation of $\mathcal{E}$
  as given in Proposition~\ref{prop:minmax-affine-monoid}.
  The depth is $O(1)$ as the algorithm consists of two subsequent parallel blocks, at Lines~3--6 and
  Lines~7--9, respectively; and each parallel block consists of $O(1)$ instructions executed in
  parallel.
  The work is $O(N^2)$ as the algorithm performs work $O(N^2)$ to assign values to the matrix $E$ of
  shape $(N,N)$ (Lines~3--6) and work $O(N)$ to assign values to the array $e$ of length $N$
  (Lines~7--9).
\end{proof}

\begin{theorem}\label{thm:complexity-parallel-scan}
  Let $\vmin,\vmax \in \reals$ with $\vmin \leq \vmax$,
  let $V = [\vmin,\vmax]$,
  let $A \in V^{T \times N \times N}$,
  and
  let $B \in V^{T \times N}$.
  On input $\tuple{A,B,\vmin,\vmax}$,
  the output of $\ParallelScan$ (Algorithm~\ref{alg:parallel-scan}) is
  $(C,D)$ with 
  $$(\idx{C}{t},\idx{D}{t}) = (\idx{A}{0},\idx{B}{0}) \minmaxcomp \cdots \minmaxcomp
  (\idx{A}{t-1},\idx{B}{t-1}) \minmaxcomp \mathcal{E},
  \quad \forall t \in \iivalro{0}{T},$$
  for $\mathcal{E}$ the identity element of the MinMax monoid $\mathcal{M}_{V,N}$;
  furthermore,
  its depth is $O(\log T \cdot \log N)$, 
  and 
  its work is $O(TN^3)$.
\end{theorem}
\begin{proof}
  We show correctness and then complexity.

  \paragraph{Correctness.}
  We proceed by induction on $T$.
  In the base case $T=1$ and the algorithm executes Line~4 returning the identity element of the
  MinMax monoid $\mathcal{M}_{[\vmin,\vmax],N}$ by Theorem~\ref{thm:complexity-compute-identity}.

  In the inductive case we have $T>1$ and we assume that $\ParallelScan(X,Y,\vmin,\vmax)$
  returns the expected output on all inputs $\tuple{X,Y,\vmin,\vmax}$ of length at most $T-1$.

  We consider two cases separately, depending on whether $T$ is even or odd.

  In the first case we assume that $T$ is even, and hence $m = \ell = T/2$ after executing Line~5.
  Let $M_t = \tuple{\idx{A}{t},\idx{B}{t}}$ for each $t \in \iival{0}{T-1}$.
  We have that, after executing Line~8--9, for every $t \in \iival{0}{m-1}$,
  by Theorem~\ref{thm:complexity-affine-comp},
  it holds that
  \begin{align} \label{eq:thm:complexity-parallel-scan-1}
    \tuple{\idx{X}{t}, \idx{Y}{t}} = M_{2t} \minmaxcomp M_{2t+1}.
  \end{align}
  Lines~10--11 are not executed since $T$ is even.
  After executing Line~12, by the inductive hypothesis (IH) we have that,
  for every $t \in \iival{0}{m-1} = \iival{0}{(T/2)-1}$,
  \begin{align} 
    \tuple{\idx{Z}{t},\idx{W}{t}} 
    & = \tuple{\idx{X}{0},\idx{Y}{0}} \minmaxcomp \cdots \minmaxcomp
    \tuple{\idx{X}{t-1},\idx{Y}{t-1}} \minmaxcomp \mathcal{E}
    \mathsidecomment{Line~12 and IH}
    \\[.3em]
    & = 
    \big(M_0 \minmaxcomp M_1\big) 
    \minmaxcomp \cdots \minmaxcomp
    \big( M_{2(t-1)} \minmaxcomp M_{2(t-1)+1} \big) 
    \minmaxcomp \mathcal{E}
    \mathsidecomment{Eq.~\eqref{eq:thm:complexity-parallel-scan-1}}
    \\[.3em]
    \label{eq:thm:complexity-parallel-scan-2}
    & = 
    M_0 \minmaxcomp \cdots \minmaxcomp M_{2t-1} \minmaxcomp \mathcal{E}.
    \mathsidecomment{associativity}
  \end{align}

  Then, after executing Lines~13--19,
  for each even $t \in \iival{0}{T-2}$, it holds that
  \begin{align}
    \tuple{\idx{C}{t},\idx{D}{t}} 
    & = \tuple{\idx{Z}{t/2},\idx{W}{t/2}} 
    \mathsidecomment{Line~17}
    \\[.3em]
    & = 
    M_0 \minmaxcomp \cdots \minmaxcomp M_{2(t/2)-1} \minmaxcomp \mathcal{E}
    \mathsidecomment{Eq.~\eqref{eq:thm:complexity-parallel-scan-2} as $t/2 \in \iival{0}{(T/2)-1}$ }
    \\[.3em]
    \label{eq:thm:complexity-parallel-scan-3}
    & = 
    M_0 \minmaxcomp \cdots \minmaxcomp M_{t-1} \minmaxcomp \mathcal{E},
    \mathsidecomment{$2(t/2)-1 = t-1$}
  \end{align}
  and for each odd $t \in \iival{1}{T-1}$, it holds that
  \begin{align}
    \tuple{\idx{C}{t},\idx{D}{t}} 
    & = 
    \tuple{\idx{Z}{(t-1)/2},\idx{W}{(t-1)/2}} \minmaxcomp M_{t-1}
    \mathsidecomment{Line~19 and Thm.~\ref{thm:complexity-affine-comp}}
    \\[.3em]
    & = 
    M_0 \minmaxcomp \cdots \minmaxcomp M_{2((t-1)/2)-1} \minmaxcomp \mathcal{E} \minmaxcomp M_{t-1}
    \mathsidecomment{Eq.~\eqref{eq:thm:complexity-parallel-scan-2} as $(t-1)/2 \in \iival{0}{(T/2)-1}$}
    \\[.3em]
    & = 
    M_0 \minmaxcomp \cdots \minmaxcomp M_{t-2} \minmaxcomp \mathcal{E} \minmaxcomp M_{t-1}
    \mathsidecomment{$2((t-1)/2)-1 = t-2$}
    \\[.3em]
    & = 
    M_0 \minmaxcomp \cdots \minmaxcomp M_{t-2} \minmaxcomp M_{t-1}
    \mathsidecomment{As $\mathcal{E}$ is identity and $t-1 \geq 0$}
    \\[.3em]
    \label{eq:thm:complexity-parallel-scan-4}
    & = 
    M_0 \minmaxcomp \cdots \minmaxcomp M_{t-1} \minmaxcomp \mathcal{E}.
    \mathsidecomment{As $\mathcal{E}$ is identity and $t-1 \geq 0$}
  \end{align}
  Therefore, at the return Line~20, by Equations~\eqref{eq:thm:complexity-parallel-scan-3}
  and~\eqref{eq:thm:complexity-parallel-scan-4}, for each $t \in \iival{0}{T-1}$, it holds that
  \begin{align*}
    \tuple{\idx{C}{t},\idx{D}{t}} 
    & = 
    M_0 \minmaxcomp \cdots \minmaxcomp M_{t-1} \minmaxcomp \mathcal{E},
  \end{align*}
  as required.

  In the second case we have that $T$ is odd.
  After executing Line~5, it holds that $m = \lfloor T/2 \rfloor$ and
  $\ell = \lceil T/2 \rceil$.
  We have that $T' \defeq T-1$ is even,
  and hence
  \begin{align*}
    m 
    & 
    = \lfloor T/2 \rfloor 
    = \lfloor (T'+1)/2 \rfloor 
    = \lfloor T'/2+1/2 \rfloor 
    = T'/2
    = (T-1)/2,
    \\[.3em]
    \ell 
    & 
    = \lceil T/2 \rceil 
    = \lceil (T'+1)/2 \rceil 
    = \lceil T'/2+1/2 \rceil 
    = T'/2 + 1
    = (T-1)/2 + 1.
  \end{align*}
  In particular, we have
  \begin{align} \label{eq:thm:complexity-parallel-scan-5}
    m = (T-1)/2,
    \qquad
    m-1 = (T-3)/2,
    \qquad
    m = \ell-1.
  \end{align}
  Let $M_t = \tuple{\idx{A}{t},\idx{B}{t}}$ for each $t \in \iival{0}{T-1}$.
  We have that, after executing Lines~8--9, for every $t \in \iival{0}{m-1} = \iival{0}{(T-3)/2}$,
  by Theorem~\ref{thm:complexity-affine-comp},
  it holds that
  \begin{align} \label{eq:thm:complexity-parallel-scan-7}
    \tuple{\idx{X}{t}, \idx{Y}{t}} = M_{2t} \minmaxcomp M_{2t+1}.
  \end{align} 
  Then, since $T$ is odd, we have that Line~11 is executed, yielding
  \begin{align} \label{eq:thm:complexity-parallel-scan-8}
    \tuple{\idx{X}{m}, \idx{Y}{m}} = M_{T-1}. 
  \end{align}
  since $m = \ell-1$ as noted in Equation~\eqref{eq:thm:complexity-parallel-scan-5}.
  Then, after executing Line~12, by the inductive hypothesis~(IH) we have that,
  for every $t \in \iival{0}{m} = \iival{0}{(T-1)/2}$,
  \begin{align}
    & \tuple{\idx{Z}{t},\idx{W}{t}} 
    \\[.3em]
    & = \tuple{\idx{X}{0},\idx{Y}{0}} \minmaxcomp \cdots \minmaxcomp
    \tuple{\idx{X}{t-1},\idx{Y}{t-1}} \minmaxcomp \mathcal{E}
    \mathsidecomment{Line~12 and IH}
    \\[.3em]
    & = 
    \big(M_0 \minmaxcomp M_1\big) 
    \minmaxcomp \cdots \minmaxcomp
    \big( M_{2(t-1)} \minmaxcomp M_{2(t-1)+1} \big) 
    \minmaxcomp \mathcal{E}
    \mathsidecomment{Eq.~\eqref{eq:thm:complexity-parallel-scan-7} as $\iival{0}{t-1} \subseteq
      \iival{0}{(T-3)/2}$}
    \\[.3em]
    \label{eq:thm:complexity-parallel-scan-9}
    & = 
    M_0 \minmaxcomp \cdots \minmaxcomp M_{2t-1} \minmaxcomp \mathcal{E}.
    \mathsidecomment{associativity}
  \end{align}

  Then, after executing Lines~13--19,
  for each even $t \in \iival{0}{T-1}$, 
  it holds that
  \begin{align}
    \tuple{\idx{C}{t},\idx{D}{t}} 
    & = \tuple{\idx{Z}{t/2},\idx{W}{t/2}} 
    \mathsidecomment{Line~17}
    \\[.3em]
    & = 
    M_0 \minmaxcomp \cdots \minmaxcomp M_{2(t/2)-1} \minmaxcomp \mathcal{E}
    \mathsidecomment{Eq.~\eqref{eq:thm:complexity-parallel-scan-9} as $t/2 \in \iival{0}{(T-1)/2}$}
    \\[.3em]
    \label{eq:thm:complexity-parallel-scan-10}
    & = 
    M_0 \minmaxcomp \cdots \minmaxcomp M_{t-1} \minmaxcomp \mathcal{E},
    \mathsidecomment{$2(t/2)-1 = t-1$}
  \end{align}
  and for each odd $t \in \iival{1}{T-2}$, 
  it holds that
  \begin{align}
    \tuple{\idx{C}{t},\idx{D}{t}} 
    & = 
    \tuple{\idx{Z}{(t-1)/2},\idx{W}{(t-1)/2}} \minmaxcomp M_{t-1}
    \mathsidecomment{Line~19 and Thm.~\ref{thm:complexity-affine-comp}}
    \\[.3em]
    & = 
    M_0 \minmaxcomp \cdots \minmaxcomp M_{2((t-1)/2)-1} \minmaxcomp \mathcal{E} \minmaxcomp M_{t-1}
    \mathsidecomment{Eq.~\eqref{eq:thm:complexity-parallel-scan-9} as $(t-1)/2 \in \iival{0}{(T-3)/2}$}
    \\[.3em]
    & = 
    M_0 \minmaxcomp \cdots \minmaxcomp M_{t-2} \minmaxcomp \mathcal{E} \minmaxcomp M_{t-1}
    \mathsidecomment{$2((t-1)/2)-1 = t-2$}
    \\[.3em]
    & = 
    M_0 \minmaxcomp \cdots \minmaxcomp M_{t-1}
    \mathsidecomment{As $\mathcal{E}$ is identity and $t > 0$}
    \\[.3em]
    \label{eq:thm:complexity-parallel-scan-11}
    & = 
    M_0 \minmaxcomp \cdots \minmaxcomp M_{t-1} \minmaxcomp \mathcal{E}.
    \mathsidecomment{As $\mathcal{E}$ is identity and $t > 0$}
  \end{align}
  Therefore, at the return instruction of Line~20, by
  Equations~\eqref{eq:thm:complexity-parallel-scan-10}
  and~\eqref{eq:thm:complexity-parallel-scan-11}, for each $t \in \iival{0}{T-1}$, it holds that
  \begin{align*}
    \tuple{\idx{C}{t},\idx{D}{t}} 
    & = 
    M_0 \minmaxcomp \cdots \minmaxcomp M_{t-1} \minmaxcomp \mathcal{E},
  \end{align*}
  as required.
  This completes the correctness proof.

  \paragraph{Complexity.}
  We first analyse the depth.
  When $T=1$, the depth is given by the depth of $\ComputeIdentity(\vmin,\vmax,N)$
  which is $O(1)$ by Theorem~\ref{thm:complexity-compute-identity}.
  When $T > 1$, the depth is asymptotically bounded
  by the depth of the recursive call at Line~12 on inputs of length $\lceil T/2 \rceil$
  and dimension $N$,
  plus 
  the depth of $\ParallelComp$ on inputs of dimension $N$, which is $O(\log N)$ by
  Theorem~\ref{thm:complexity-affine-comp}.
  Thus, there exist constants $a_1,c,c_1$ such that the depth is bounded as follows:
  \begin{align} \label{eq:thm:complexity-parallel-scan-1-depth-1}
    \operatorname{depth}(T,N) \leq
    \begin{cases}
      c & \text{ if } T = 1,
      \\[.1em]
      \operatorname{depth}(\lceil T/2 \rceil,N) + c_1 + a_1 \log_2 (N+1)
      & \text{ otherwise}.
    \end{cases}
  \end{align}
  We show $\operatorname{depth}(T,N) \in O(\log T \cdot \log N)$
  by showing that, for $a \defeq 3(a_1+c_1)$,
  it holds that 
  \begin{equation} \label{eq:thm:complexity-parallel-scan-1-depth-2}
    \operatorname{depth}(T,N) \leq c + a \cdot \log_2 T \cdot \log_2 (N+1).
  \end{equation}
  The base case $T=1$ is proved as follows:
  \begin{align*}
    \operatorname{depth}(T,N)
    & \leq c
    \mathsidecomment{by Eq.~\eqref{eq:thm:complexity-parallel-scan-1-depth-1}}
    \\[.2em]
    & = c  + a \cdot \log_2 T \cdot \log_2 (N+1).
    \mathsidecomment{$\log_2 T = 0$ as $T=1$}
  \end{align*}

  In the inductive case we have $T>1$ and we assume that, for all $T' \leq T-1$,
  \begin{equation} \label{eq:thm:complexity-parallel-scan-1-depth-3}
    \operatorname{depth}(T',N) \leq c + a \cdot \log_2 T' \cdot \log_2 (N+1).
  \end{equation}
  We consider two cases separately, according to whether $T$ is even or odd.
  Let us first consider the case when $T$ is even.
  We have
  \begin{align*}
    \operatorname{depth}(T,N)
    & \leq
    \operatorname{depth}(\lceil T/2 \rceil,N) + c_1 + a_1 \log_2 (N+1)
    \mathsidecomment{Eq.~\eqref{eq:thm:complexity-parallel-scan-1-depth-1}}
    \\[.2em]
    & \leq
    \operatorname{depth}(T/2,N) + c_1 + a_1 \log_2 (N+1)
    \mathsidecomment{$T$ is even}
    \\[.2em]
    & \leq
    c + a \cdot \log_2 (T/2) \cdot \log_2 (N+1) + c_1 + a_1 \log_2 (N+1).
    \mathsidecomment{Eq.~\eqref{eq:thm:complexity-parallel-scan-1-depth-3}}
  \end{align*}
  Thus, it suffices to show that the following inequality holds:
  \begin{align*}
    c + a \cdot \log_2 (T/2) \cdot \log_2 (N+1) + c_1 + a_1 \log_2 (N+1)
    \leq
    c + a \cdot \log_2 T \cdot \log_2 (N+1).
  \end{align*}
  We have
  \begin{align*}
    & c + a \cdot \log_2 (T/2) \cdot \log_2 (N+1) + c_1 + a_1 \log_2 (N+1)
    \leq
    c + a \cdot \log_2 T \cdot \log_2 (N+1)
    \\[.2em]
    & \text{if } \quad
    a \cdot \log_2 (T/2) \cdot \log_2 (N+1) + c_1 + a_1 \log_2 (N+1)
    \leq
    a \cdot \log_2 T \cdot \log_2 (N+1)
    \\[.2em]
    & \text{if } \quad
    c_1 + a_1 \log_2 (N+1)
    \leq
    a \cdot \log_2 T \cdot \log_2 (N+1)
    -
    a \cdot \log_2 (T/2) \cdot \log_2 (N+1) 
    \\[.2em]
    & \text{if } \quad
    c_1 + a_1 \log_2 (N+1)
    \leq
    a \cdot \log_2 T \cdot \log_2 (N+1)
    -
    a \cdot (\log_2 T - 1) \cdot \log_2 (N+1) 
    \\[.2em]
    & \text{if } \quad
    c_1 + a_1 \log_2 (N+1)
    \leq
    a \cdot \log_2 T \cdot \log_2 (N+1)
    +
    a \cdot \log_2 (N+1) 
    -
    a \cdot \log_2 T \cdot \log_2 (N+1) 
    \\[.2em]
    & \text{if } \quad
    c_1 + a_1 \log_2 (N+1)
    \leq
    a \cdot \log_2 (N+1) 
    \\[.2em]
    & \text{if } \quad
    \frac{c_1}{\log_2 (N+1)} + a_1
    \leq
    a 
    \\[.2em]
    & \text{if } \quad
    a \geq \frac{c_1}{\log_2 (N+1)} + a_1
    \\[.2em]
    & \text{if } \quad
    a \geq c_1 + a_1,
    \qquad \text{| $\log_2 (N+1) \geq 1$ as $N \geq 1$}
  \end{align*}
  and the inequality above holds by definition of $a$.
  Let us now consider the case when $T$ is odd.
  We have
  \begin{align*}
    \operatorname{depth}(T,N)
    & \leq
    \operatorname{depth}(\lceil T/2 \rceil,N) + c_1 + a_1 \log_2 (N+1)
    \mathsidecomment{Eq.~\eqref{eq:thm:complexity-parallel-scan-1-depth-1}}
    \\[.2em]
    & \leq
    \operatorname{depth}((T+1)/2,N) + c_1 + a_1 \log_2 (N+1)
    \mathsidecomment{$\lceil T/2 \rceil = (T+1)/2$ as $T$ is odd}
    \\[.2em]
    & \leq
    c + a \cdot \log_2 ((T+1)/2) \cdot \log_2 (N+1) + c_1 + a_1 \log_2 (N+1).
    \mathsidecomment{Eq.~\eqref{eq:thm:complexity-parallel-scan-1-depth-3}}
  \end{align*}
  Thus, it suffices to show that the following inequality holds:
  \begin{align*}
    c + a \cdot \log_2 ((T+1)/2) \cdot \log_2 (N+1) + c_1 + a_1 \log_2 (N+1)
    \leq
    c + a \cdot \log_2 T \cdot \log_2 (N+1).
  \end{align*}
  \begin{align*}
    & c + a \cdot \log_2 ((T+1)/2) \cdot \log_2 (N+1) + c_1 + a_1 \log_2 (N+1)
    \leq
    c + a \cdot \log_2 T \cdot \log_2 (N+1)
    \\[.2em]
    & \text{if } \quad
    a \cdot \log_2 ((T+1)/2) \cdot \log_2 (N+1) + c_1 + a_1 \log_2 (N+1)
    \leq
    a \cdot \log_2 T \cdot \log_2 (N+1)
    \\[.2em]
    & \text{if } \quad
    c_1 + a_1 \log_2 (N+1)
    \leq
    a \cdot \log_2 T \cdot \log_2 (N+1)
    -
    a \cdot \log_2 ((T+1)/2) \cdot \log_2 (N+1) 
    \\[.2em]
    & \text{if } \quad
    c_1 + a_1 \log_2 (N+1)
    \leq
    a \cdot \log_2 T \cdot \log_2 (N+1)
    -
    a \cdot (\log_2 (T+1) - 1) \cdot \log_2 (N+1) 
    \\[.2em]
    & \text{if } \quad
    c_1 + a_1 \log_2 (N+1)
    \leq
    a \cdot \log_2 T \cdot \log_2 (N+1)
    +
    a \cdot \log_2 (N+1) 
    -
    a \cdot \log_2 (T+1) \cdot \log_2 (N+1) 
    \\[.2em]
    & \text{if } \quad
    c_1 + a_1 \log_2 (N+1)
    \leq
    a \cdot (\log_2 T + 1 - \log_2 (T+1)) \cdot \log_2 (N+1)
    \\[.2em]
    & \text{if } \quad
    c_1 + a_1 \log_2 (N+1)
    \leq
    a \cdot \log_2 (2T/(T+1)) \cdot \log_2 (N+1)
    \\[.2em]
    & \text{if } \quad
    \frac{1}{\log_2 (2T/(T+1))} \cdot \left(\frac{c_1}{ \log_2 (N+1)} + a_1\right)
    \leq
    a 
    \\[.2em]
    & \text{if } \quad
    a \geq \frac{1}{\log_2 (2T/(T+1))} \cdot \left(\frac{c_1}{ \log_2 (N+1)} + a_1\right)
    \\[.2em]
    & \text{if } \quad
    a \geq \frac{1}{\log_2 (4/3)} \cdot \left(\frac{c_1}{ \log_2 (N+1)} + a_1\right)
    \qquad \text{|~$2T/(T+1) \leq 4/3$ shown below and $\log_2$ monotonic}
    \\[.2em]
    & \text{if } \quad
    a \geq \frac{1}{\log_2 (4/3)} \cdot \left(c_1 + a_1\right)
    \qquad \text{|~$\log_2(N+1) \geq 1$}
    \\[.2em]
    & \text{if } \quad
    a \geq 3 \cdot \left(c_1 + a_1\right),
    \qquad \text{|~$1/\log_2(4/3) \leq 3$}
  \end{align*}
  and the above holds by definition of $a$.
  It remains to show $2T/(T+1) \leq 4/3$.
  The associated equation is
  \begin{align*}
    & 2T/(T+1) = 4/3
    \\
    & \Leftrightarrow\quad 
    2T = 4/3(T+1)
    \\
    & \Leftrightarrow\quad 
    2T = 4T/3+ 4/3
    \\
    & \Leftrightarrow\quad 
    2T-4T/3 = 4/3
    \\
    & \Leftrightarrow\quad 
    (2-4/3)T = 4/3
    \\
    & \Leftrightarrow\quad 
    (2/3)T = 4/3
    \\
    & \Leftrightarrow\quad 
    T = 4/3\cdot 3/2
    \\
    & \Leftrightarrow\quad 
    T = 2,
  \end{align*}
  and hence the inequality holds for all $T \geq 2$.

  Next we analyse the work.
  When $T=1$, the work is given by the work of $\ComputeIdentity(\vmin,\vmax,N)$
  which is $O(N^2)$ by Theorem~\ref{thm:complexity-compute-identity}.
  When $T>1$,
  the work is asymptotically bounded by sum of the work of the following:
  \begin{itemize}
    \item
      the work of Lines~8--9,
      which is $O(TN^3)$ since the lines perform $m = O(T)$ calls of $\ParallelComp$ on inputs of
      dimension $N$ (Lines~8--9), with each call having work that is $O(N^3)$ by
      Theorem~\ref{thm:complexity-affine-comp};
    \item
      the work of Line~12, which is a call of $\ParallelScan$ on an input of length $\ell = \lceil
      T/2 \rceil$ and dimension $N$;
    \item
      the work of Lines~15--19,
      which is $O(TN^3)$ since the lines perform for $T$ times either an assignment of a previously
      computed value (Line~17) which
      is $O(1)$ or a call of $\ParallelComp$ on an input of
      dimension $N$ (Lines~19), with each call having work that is $O(N^3)$ by
      Theorem~\ref{thm:complexity-affine-comp}.
  \end{itemize}
  Thus,
  there exist constants $a_0,a_1,c_0,c_1$ such that
  \begin{equation} \label{eq:thm:complexity-parallel-scan-1-work-1}
    \operatorname{work}(T,N) \leq
    \begin{cases}
      c_0 + a_0 \cdot N^2 & \text{ if } T = 1,
      \\[.1em]
      \operatorname{work}(\lceil T/2 \rceil ,N) + c_1 + a_1 \cdot TN^3
      & \text{ otherwise}.
    \end{cases}
  \end{equation}
  We show $\operatorname{work}(T,N) \in O(TN^3)$ by showing that,
  for $c = c_0$ and $a = \max(a_0, 4(a_1 + c_1))$,
  the following bound holds
  \begin{align}
    \operatorname{work}(T,N) \leq c + a \cdot TN^3.
  \end{align}
  We proceed by induction on $T$.
  In the base case $T=1$,
  and we have
  \begin{align*}
    \operatorname{work}(T,N) 
    & \leq 
    c_0 + a_0 \cdot N^2 
    \mathsidecomment{Eq.~\eqref{eq:thm:complexity-parallel-scan-1-work-1}}
    \\
    & = 
    c + a_0 \cdot N^2 
    \mathsidecomment{$c = c_0$ by def.}
    \\
    & \leq 
    c + a \cdot N^2 
    \mathsidecomment{$a_0 \leq a$ by def.~of $a$}
    \\
    & =
    c + a \cdot TN^2 
    \mathsidecomment{$T=1$}
    \\
    & \leq 
    c + a \cdot TN^3,
    \mathsidecomment{$N^2 \leq N^3$}
  \end{align*}
  as required.

  In the inductive case $T>1$, and we assume by induction that,
  for all $T' \leq T-1$,
  \begin{equation} \label{eq:thm:complexity-parallel-scan-1-work-2}
    \operatorname{work}(T',N) \leq c + a \cdot T'N^3.
  \end{equation}

  We consider two separate cases, according to whether $T$ is even or odd.
  Let us first consider the case when $T$ is even.
  We have
  \begin{align*}
    \operatorname{work}(T,N) 
    & \leq 
    \operatorname{work}(\lceil T/2 \rceil ,N) + c_1 + a_1 TN^3
    \mathsidecomment{Eq.~\eqref{eq:thm:complexity-parallel-scan-1-work-1}}
    \\
    & =
    \operatorname{work}(T/2 ,N) + c_1 + a_1 TN^3
    \mathsidecomment{$T$ is even}
    \\
    & \leq 
    c + a (T/2)N^3 + c_1 + a_1 TN^3.
    \mathsidecomment{Eq.~\eqref{eq:thm:complexity-parallel-scan-1-work-2}}
  \end{align*}
  Thus, it suffices to show that the following inequality holds:
  \begin{align*}
    c + a (T/2)N^3 + c_1 + a_1 TN^3
    & \leq 
    c + a TN^3.
  \end{align*}
  We have
  \begin{align*}
    & c + a (T/2)N^3 + c_1 + a_1 TN^3 \leq c + a TN^3
    \\[.2em]
    & \text{if } \quad 
    a (T/2)N^3 + c_1 + a_1 TN^3 \leq a TN^3
    \\[.2em]
    & \text{if } \quad 
    a (1/2)TN^3 + c_1 + a_1 TN^3 \leq a TN^3
    \\[.2em]
    & \text{if } \quad 
    c_1 + a_1 TN^3 \leq a TN^3 - a (1/2)TN^3
    \\[.2em]
    & \text{if } \quad 
    c_1 + a_1 TN^3 \leq a (1/2) TN^3 
    \\[.2em]
    & \text{if } \quad 
    \frac{2(c_1 + a_1 TN^3)}{TN^3} \leq a
    \\[.2em]
    & \text{if } \quad 
    2\left(\frac{c_1}{TN^3} + a_1\right) \leq a
    \\[.2em]
    & \text{if } \quad 
    a \geq 2\left(\frac{c_1}{TN^3} + a_1\right) 
    \\[.2em]
    & \text{if } \quad 
    a \geq 2\left(c_1 + a_1\right) 
    \mathsidecomment{$T,N > 1$}
  \end{align*}
  and the above inequality holds by definition of $a$.
  Let us now consider the case when $T$ is odd.
  We have
  \begin{align*}
    \operatorname{work}(T,N) 
    & \leq 
    \operatorname{work}(\lceil T/2 \rceil ,N) + c_1 + a_1 TN^3
    \mathsidecomment{Eq.~\eqref{eq:thm:complexity-parallel-scan-1-work-1}}
    \\
    & =
    \operatorname{work}\big((T+1)/2 , N\big) + c_1 + a_1 TN^3
    \mathsidecomment{$\lceil T/2 \rceil = (T+1)/2$ as $T$ is odd}
    \\
    & \leq 
    c + a \cdot \big((T+1)/2\big)N^3 + c_1 + a_1 TN^3.
    \mathsidecomment{Eq.~\eqref{eq:thm:complexity-parallel-scan-1-work-2}}
  \end{align*}
  Thus, it suffices to show that the following inequality holds:
  \begin{align*}
    c + a \cdot \big((T+1)/2\big)N^3 + c_1 + a_1 TN^3
    & \leq 
    c + a TN^3.
  \end{align*}
  We have
  \begin{align*}
    & c + a \cdot \big((T+1)/2\big)N^3 + c_1 + a_1 TN^3 \leq c + a TN^3
    \\[.2em]
    & \text{if } \quad 
    a \cdot \big((T+1)/2\big)N^3 + c_1 + a_1 TN^3 \leq a TN^3
    \\[.2em]
    & \text{if } \quad 
    c_1 + a_1 TN^3 \leq a TN^3 - a \cdot \big((T+1)/2\big)N^3
    \\[.2em]
    & \text{if } \quad 
    c_1 + a_1 TN^3 \leq a TN^3 - a \cdot \frac{T+1}{2T} \cdot TN^3
    \\[.2em]
    & \text{if } \quad 
    c_1 + a_1 TN^3 \leq a \cdot \left( 1 - \frac{T+1}{2T} \right) \cdot TN^3
    \\[.2em]
    & \text{if } \quad 
    c_1 + a_1 TN^3 \leq a \cdot \frac{2T - T - 1}{2T} \cdot TN^3
    \\[.2em]
    & \text{if } \quad 
    c_1 + a_1 TN^3 \leq a \cdot \frac{T-1}{2T} \cdot TN^3
    \\[.2em]
    & \text{if } \quad 
    \frac{c_1 + a_1 TN^3}{TN^3} \leq a \cdot \frac{T-1}{2T}
    \\[.2em]
    & \text{if } \quad 
    \frac{2T}{T-1} \cdot \frac{c_1 + a_1 TN^3}{TN^3} \leq a 
    \\[.2em]
    & \text{if } \quad 
    a \geq \frac{2T}{T-1} \cdot \frac{c_1 + a_1 TN^3}{TN^3}
    \\[.2em]
    & \text{if } \quad 
    a \geq \frac{2T}{T-1} \cdot \left(\frac{c_1}{TN^3} + a_1\right)
    \\[.2em]
    & \text{if } \quad 
    a \geq \frac{2T}{T-1} \cdot \left(c_1 + a_1\right)
    \mathsidecomment{$T,N \geq 1$}
    \\[.2em]
    & \text{if } \quad 
    a \geq 4 \cdot \left(c_1 + a_1\right),
    \mathsidecomment{$2T/(T-1) \leq 4$ shown below}
  \end{align*}
  and the above holds by definition of $a$.
  It remains to show $2T/(T-1) \leq 4$.  
  The associate equation is
  \begin{align*}
    & 2T/(T-1) = 4
    \\
    & \Leftrightarrow\quad
    2T = 4(T-1)
    \\
    & \Leftrightarrow\quad
    T = 2(T-1)
    \\
    & \Leftrightarrow\quad
    T = 2T-2
    \\
    & \Leftrightarrow\quad
    2 = T,
  \end{align*}
  and hence the inequality holds for all $T > 1$.
\end{proof}

\todo{}

\begin{lemma} \label{lemma:parallel-evaluation-complexity}
  \ref{alg:parallel-general-evaluation} solves
  \nameref{def:minmax-recurrence-evaluation-problem} 
  with depth
  $O(\log N \cdot \log T)$ and work $O(TN^3)$ for $T$ the number of steps and $N$ the state
  dimension.
  Thus, it runs in time $O(\log N \cdot \log T + TN^3/P)$ given $P$ processors.
\end{lemma}
\begin{proof}
  Let $(X_\mathrm{init}, A, B)$ be an input of \nameref{def:minmax-recurrence-evaluation-problem}.
  We first show correctness and then complexity.

  \paragraph{Correctness.}
  Let $Y \in \reals^{T \times N}$ be such that 
  $\idx{Y}{0} = \minmax(X_\mathrm{init}, \idx{A}{0}, \idx{B}{0})$
  and
  $\idx{Y}{t} = \minmax(\idx{Y}{t-1}, \idx{A}{t}, \idx{B}{t})$ for every $t \in \iival{1}{T-1}$.
  It suffices to show that the variable $X$ of the algorithm satisfies 
  $X = Y$ when the return instruction of Line~10 is executed.

  After executing Line~2, we have that $\vmin$ is the minimum scalar value occurring in the input,
  by Theorem~\ref{thm:complexity-parallel-min}.
  After executing Line~3, we have that $\vmax$ is the maximum scalar value occurring in the input,
  by Theorem~\ref{thm:complexity-parallel-max}.
  After executing Line~4,
  since each $\tuple{\idx{A}{t},\idx{B}{t}}$ belongs to the MinMax monoid
  $\mathcal{M}_{[\vmin,\vmax],N}$,
  by Theorem~\ref{thm:complexity-parallel-scan}
  we have that $(C,D)$ is such that
  \begin{equation} \label{eq:lemma:parallel-evaluation-complexity-1}
    (\idx{C}{t},\idx{D}{t}) = (\idx{A}{0},\idx{B}{0}) \minmaxcomp \cdots \minmaxcomp
    (\idx{A}{t-1},\idx{B}{t-1}) \minmaxcomp \mathcal{E},
    \quad \forall t \in \iival{0}{T-1},
  \end{equation}
  for $\mathcal{E}$ the identity element of the MinMax monoid $\mathcal{M}_{[\vmin,\vmax],N}$.
  Let us now consider Lines~7--9 for $t \in \iival{0}{T-1}$.
  After executing Line~8, it holds that
  \begin{align}
    & 
    (\idx{Z}{t},\idx{W}{t}) 
    \\[.2em]
    & = 
    (\idx{C}{t},\idx{D}{t}) \minmaxcomp (\idx{A}{t},\idx{B}{t}) 
    \mathsidecomment{Line~8 and Theorem~\ref{thm:complexity-affine-comp}}
    \\[.2em]
    & = 
    \big((\idx{A}{0},\idx{B}{0}) \minmaxcomp \cdots \minmaxcomp
    (\idx{A}{t-1},\idx{B}{t-1}) \minmaxcomp \mathcal{E} \big)
    \minmaxcomp (\idx{A}{t},\idx{B}{t}) 
    \mathsidecomment{Eq.~\eqref{eq:lemma:parallel-evaluation-complexity-1}}
    \\[.2em]
    & = 
    (\idx{A}{0},\idx{B}{0}) \minmaxcomp \cdots \minmaxcomp
    (\idx{A}{t-1},\idx{B}{t-1}) \minmaxcomp \mathcal{E}
    \minmaxcomp (\idx{A}{t},\idx{B}{t}) 
    \mathsidecomment{associativity}
    \\[.2em]
    & = 
    (\idx{A}{0},\idx{B}{0}) \minmaxcomp \cdots \minmaxcomp
    (\idx{A}{t-1},\idx{B}{t-1}) 
    \minmaxcomp (\idx{A}{t},\idx{B}{t}) 
    \mathsidecomment{$\mathcal{E}$ is identity}
    \\[.2em]
    \label{eq:lemma:parallel-evaluation-complexity-2}
    & = 
    (\idx{A}{0},\idx{B}{0}) 
    \minmaxcomp \cdots \minmaxcomp
    (\idx{A}{t},\idx{B}{t}). 
  \end{align}
  After executing Line~9, by Theorem~\ref{thm:complexity-parallel-apply} it holds that
  \begin{align} \label{eq:lemma:parallel-evaluation-complexity-3}
    \idx{X}{t} = (\idx{Z}{t} \otimes X_\mathrm{init}) \oplus \idx{W}{t}.
  \end{align}
  We conclude the proof by showing that the r.h.s.\ above equals $\idx{Y}{t}$ as required.
  Let us recall the maps $\tau_M$ of the monoid $\mathcal{M}_{[\vmin,\vmax],N}$
  (\S\ref{par:complexity-minmax-affine-monoid}),
  and
  let us introduce the notation $\tau_i(x) \defeq \tau(\idx{A}{i},\idx{B}{i}, x)$ for each $i \in
  \iival{0}{T-1}$,
  and then the notation
  $\tau_{0:i} \defeq (\tau_i \circ \cdots \circ \tau_0)$ for each $i \in \iival{0}{T-1}$.
  By the definition of~$Y$,
  we have
  \begin{align} \label{eq:lemma:parallel-evaluation-complexity-4}
    \idx{Y}{t} = \tau_{0:t}(X_\mathrm{init}),
  \end{align}
  and by Eq.~\eqref{eq:minmax-affine-monoid-action}, applied $t$~times,
  we have 
  \begin{align} \label{eq:lemma:parallel-evaluation-complexity-5}
    \tau_{0:t} = \tau_{M_t}
    \quad \text{ for }
    M_t \defeq (\idx{A}{0},\idx{B}{0}) \minmaxcomp \cdots \minmaxcomp (\idx{A}{t},\idx{B}{t}).
  \end{align}
  Then,
  \begin{align}
    \idx{Y}{t} 
    & = \tau_{0:t}(X_\mathrm{init}) 
    \mathsidecomment{Eq.~\eqref{eq:lemma:parallel-evaluation-complexity-4}}
    \\[.2em]
    & = 
    \tau_{M_t}(X_\mathrm{init}) 
    \mathsidecomment{Eq.~\eqref{eq:lemma:parallel-evaluation-complexity-5}}
    \\[.2em]
    & = 
    \tau(Z[t],W[t],X_\mathrm{init}) 
    \mathsidecomment{$M_t = (\idx{Z}{t},\idx{W}{t})$ by Eq.~\eqref{eq:lemma:parallel-evaluation-complexity-2}}
    \\[.2em]
    & = 
    (\idx{Z}{t} \otimes X_\mathrm{init}) \oplus \idx{W}{t}
    \mathsidecomment{def.~of $\tau$}
    \\[.2em]
    & = 
    \idx{X}{t},
    \mathsidecomment{Eq.~\eqref{eq:lemma:parallel-evaluation-complexity-3}}
  \end{align}
  as required.

  \paragraph{Complexity.}
  The depth is given by the sum of:
  \begin{itemize}
    \item
      the sum of the depths of
      $\ParallelMin$ and $\ParallelMax$ (Lines~2--3) on input of length $N + TN^2 + TN$, each having 
      depth $O(\log(N + TN^2 + TN))$ by
      Theorems~\ref{thm:complexity-parallel-min},\ref{thm:complexity-parallel-max}, and hence
      the sum of their depths is $2 \cdot O(\log(N + TN^2 + TN)) = O(\log(TN)) = O(\log T + \log N)$;
    \item
      the depth of $\ParallelScan$ (Line~4) on an input having $T$ steps and dimension $N$, which is 
      $O(\log T \cdot \log N)$ by Theorem~\ref{thm:complexity-parallel-scan};
    \item
      the maximum for $t \in \iival{0}{T-1}$ of
      \begin{itemize}
        \item
          the depth of $\ParallelComp$ (Line~8) on input of dimension $N$,
          which is 
          $O(\log N)$ by Theorem~\ref{thm:complexity-affine-comp};
        \item
          the depth of $\ParallelApply$ (Line~9) on input of dimension $N$, which is
          $O(\log N)$ by Theorem~\ref{thm:complexity-parallel-apply}.
      \end{itemize}
  \end{itemize}
  Thus the depth is $O(\log T + \log N) + O(\log T \cdot \log N) + O(\log N)$,
  which is $O(\log T \cdot \log N)$ as claimed.

  The work is given by the sum of:
  \begin{itemize}
    \item
      the sum of the work of
      $\ParallelMin$ and $\ParallelMax$ (Lines~2--3) on input of length $N + TN^2 + TN$, each 
      performing work $O(N + TN^2 + TN)$ by
      Theorems~\ref{thm:complexity-parallel-min},\ref{thm:complexity-parallel-max}, and hence
      their total work is $2 \cdot O(N + TN^2 + TN) = O(TN^2)$;
    \item
      the work of $\ParallelScan$ (Line~4) on an input having $T$ steps and dimension $N$, which is 
      $O(TN^3)$ by Theorem~\ref{thm:complexity-parallel-scan};
    \item
      the sum of the work of the following for $t \in \iival{0}{T-1}$:
      \begin{itemize}
        \item
          the work of $\ParallelComp$ (Line~8) on input of dimension $N$,
          which is 
          $O(N^3)$ by Theorem~\ref{thm:complexity-affine-comp};
        \item
          the work of $\ParallelApply$ (Line~9) on input of dimension $N$, which is
          $O(N^2)$ by Theorem~\ref{thm:complexity-parallel-apply}.
      \end{itemize}
  \end{itemize}
  Thus the work is $O(TN^2) + O(TN^3) + T \cdot (O(N^3) + O(N^2))$,
  which is $O(TN^3)$ as claimed.
\end{proof}

\begin{restatable}{theorem}{thcomplexityparallelevaluation}
  \label{th:complexity-parallel-evaluation}
  \nameref{def:minmax-recurrence-evaluation-problem} can be solved in parallel 
  with depth
  $O(\log N \cdot \log T)$ and work $O(TN^3)$ for $T$ the number of steps and $N$ the state
  dimension. 
  Thus, it can be solved in time $O(\log N \cdot \log T + TN^3/P)$ given $P$ processors,
  and hence in $O(\log N \cdot \log T)$ if $P \in \Omega(TN^3)$.
\end{restatable}
\begin{proof}
  Immediate by Lemma~\ref{lemma:parallel-evaluation-complexity}.
\end{proof}

\begin{lemma} \label{lemma:parallel-evaluation-complexity-layer}
  On input $\tuple{W,U}$ for $W \in \mathbf{W}^\mathrm{layer}_\mathbf{s}$ with size
  $\mathbf{s} = \tuple{d_\mathrm{in},d_\mathrm{mlp},n_\mathrm{mlp},d_\mathrm{out},n_\mathrm{units},
  d_\mathrm{state}}$, and $U \in \reals^{T \times d_\mathrm{in}}$,
  the output of \ref{alg:parallel-evaluation-layer} is
  $S(U)$ for $S = \operatorname{Layer}(W)$,
  its depth is 
  \begin{align*}
    O\Big(\log T \cdot \log d_\mathrm{state} 
    + n_\mathrm{mlp} \cdot \log d_\mathrm{mlp}
    +
    \log(n_\mathrm{units} d_\mathrm{state} + d_\mathrm{in}) \Big)
  \end{align*}
  and its work is 
  \begin{align*}
      O\Big( T \cdot \big( 
      n_\mathrm{units} \cdot d_\mathrm{state}^3 +
      d_\mathrm{mlp} \cdot (d_\mathrm{in}  + n_\mathrm{mlp} d_\mathrm{mlp} +
      d_\mathrm{state}^2 +  d_\mathrm{out}+ n_\mathrm{units}  d_\mathrm{state})    \big) \Big)
  \end{align*}
\end{lemma}
\begin{proof}
  We note that the algorithm is obtained from the sequential 
  algorithm~\ref{alg:sequential-evaluation-layer} in a straighforward manner.
  Thus, correctness follows by the same arguments as in the proof of
  Lemma~\ref{lemma:sequential-evaluation-complexity-layer}.
  In particular, $\mathtt{SeqRecEval}$ is now replaced by $\mathtt{ParRecEval}$,
  which yields the same result, by Lemma~\ref{alg:parallel-general-evaluation};
  and the loops in $\mathtt{SeqLayerEval}$ have independent iterations and hence their iterations
  can be executed in parallel.

  Next we analyse the complexity.
  The depth is asymptotically bounded by the sum of the following depths:
  \begin{itemize}
    \item
      Line~8: the depth of $\mathtt{EvaluateMLP}$,
      where the MLP has 
      input dimension $d_\mathrm{in}$,
      output dimension $d_\mathrm{state}^2$,
      and
      $n_\mathrm{mlp}$ hidden layers of dimension $d_\mathrm{mlp,u} \defeq \lceil
      d_\mathrm{mlp}/n_\mathrm{units}\rceil$;
      its depth is $O(\log d_\mathrm{in} + n_\mathrm{mlp} \cdot \log d_\mathrm{mlp})$;
    \item
      Line~9: the depth of $\mathtt{EvaluateMLP}$ 
      where the MLP has 
      input dimension $d_\mathrm{in}$,
      output dimension $d_\mathrm{state}$,
      and
      $n_\mathrm{mlp}$ hidden layers of dimension $d_\mathrm{mlp,u} \defeq \lceil
      d_\mathrm{mlp}/n_\mathrm{units}\rceil$;
      its depth is $O(\log d_\mathrm{in} + n_\mathrm{mlp} \cdot \log d_\mathrm{mlp})$;
    \item
      Line~13: the depth \ref{alg:sequential-evaluation} on inputs of
      dimension $d_\mathrm{state}$ and length $T$;
      its depth is $O(\log T \cdot \log d_\mathrm{state})$ by
      Theorem~\ref{lemma:parallel-evaluation-complexity};
    \item
      Line~16: the depth of $\mathtt{EvaluateMLP}$,
      where the MLP has 
      input dimension $2 n_\mathrm{units} d_\mathrm{state} + d_\mathrm{in}$,
      output dimension $d_\mathrm{out}$,
      and
      $n_\mathrm{mlp}$ hidden layers of dimension $d_\mathrm{mlp}$;
      its depth is 
      $O(\log(n_\mathrm{units} d_\mathrm{state} + d_\mathrm{in}) + n_\mathrm{mlp} \cdot 
      \log d_\mathrm{mlp})$;
  \end{itemize}
  Thus, the depth is
  \begin{align*}
    &
    O(\log d_\mathrm{in} + n_\mathrm{mlp} \cdot \log d_\mathrm{mlp})
    +
    O(\log T \cdot \log d_\mathrm{state})
    +
    O(\log(n_\mathrm{units} d_\mathrm{state} + d_\mathrm{in}) + n_\mathrm{mlp} \cdot 
      \log d_\mathrm{mlp})
    \\[.2em]
    & 
    =
    O\Big(\log d_\mathrm{in} + n_\mathrm{mlp} \cdot \log d_\mathrm{mlp}
    +
    \log T \cdot \log d_\mathrm{state}
    +
    \log(n_\mathrm{units} d_\mathrm{state} + d_\mathrm{in}) + n_\mathrm{mlp} \cdot 
      \log d_\mathrm{mlp}\Big)
    \\[.2em]
    & 
    =
    O\Big(n_\mathrm{mlp} \cdot \log d_\mathrm{mlp}
    +
    \log T \cdot \log d_\mathrm{state}
    +
    \log(n_\mathrm{units} d_\mathrm{state} + d_\mathrm{in}) \Big)
  \end{align*}

  The work is asymptotically bounded by the sum of the following work:
  \begin{itemize}
    \item
      Line~8: the work of $O(n_\mathrm{units} \cdot T)$ executions of $\mathtt{EvaluateMLP}$,
      where the MLP has 
      input dimension $d_\mathrm{in}$,
      output dimension $d_\mathrm{state}^2$,
      and
      $n_\mathrm{mlp}$ hidden layers of dimension $d_\mathrm{mlp,u} \defeq \lceil
      d_\mathrm{mlp}/n_\mathrm{units}\rceil$;
      the time of each evaluation is $O(d_\mathrm{in} d_\mathrm{mlp,u} + n_\mathrm{mlp}
        d_\mathrm{mlp,u}^2 + d_\mathrm{mlp,i} d_\mathrm{state}^2)$;
    \item
      Line~9: the work of $O(n_\mathrm{units} \cdot T)$ executions of $\mathtt{EvaluateMLP}$ 
      where the MLP has 
      input dimension $d_\mathrm{in}$,
      output dimension $d_\mathrm{state}$,
      and
      $n_\mathrm{mlp}$ hidden layers of dimension $d_\mathrm{mlp,u} \defeq \lceil
      d_\mathrm{mlp}/n_\mathrm{units}\rceil$;
      the time of each evaluation is $O(d_\mathrm{in} d_\mathrm{mlp,u} + n_\mathrm{mlp}
        d_\mathrm{mlp,u}^2 +
      d_\mathrm{mlp,u} d_\mathrm{state})$;
    \item
      Line~13: the work of $O(n_\mathrm{units})$ executions of \ref{alg:sequential-evaluation} on inputs of
      dimension $d_\mathrm{state}$ and length $T$;
      its execution time is $O(T d_\mathrm{state}^3)$ by
      Theorem~\ref{lemma:sequential-evaluation-complexity};
    \item
      Line~16: the work of $O(T)$ executions of $\mathtt{EvaluateMLP}$,
      where the MLP has 
      input dimension $2 n_\mathrm{units} d_\mathrm{state} + d_\mathrm{in}$,
      output dimension $d_\mathrm{out}$,
      and
      $n_\mathrm{mlp}$ hidden layers of dimension $d_\mathrm{mlp}$;
      the time of each evaluation is 
      $O\big((n_\mathrm{units} d_\mathrm{state} + d_\mathrm{in}) \cdot d_\mathrm{mlp} + 
        n_\mathrm{mlp} d_\mathrm{mlp}^2 + d_\mathrm{mlp} d_\mathrm{out})$.
  \end{itemize}
  Thus, the work is
  \begin{align*}
    & O(n_\mathrm{units} \cdot T) 
    \cdot
    O(d_\mathrm{in} d_\mathrm{mlp,u} + n_\mathrm{mlp} d_\mathrm{mlp,u}^2 +
    d_\mathrm{mlp,u} d_\mathrm{state}^2) + {}
      \\
      &
    O(n_\mathrm{units} \cdot T)
    \cdot
    O(d_\mathrm{in} d_\mathrm{mlp,u} + n_\mathrm{mlp} d_\mathrm{mlp,u}^2 +
    d_\mathrm{mlp,u} d_\mathrm{state}) + {}
      \\
      &
    O(n_\mathrm{units})
    \cdot
    O(T d_\mathrm{state}^3) + {}
      \\
       &
      O(T)
      \cdot
      O\big((n_\mathrm{units} d_\mathrm{state} + d_\mathrm{in}) \cdot d_\mathrm{mlp} + 
      n_\mathrm{mlp} d_\mathrm{mlp,u}^2 + d_\mathrm{mlp} d_\mathrm{out}\big)
      \\
      & = 
      \\
      & O\Big(n_\mathrm{units} \cdot T \cdot
      (d_\mathrm{in} d_\mathrm{mlp,u} + n_\mathrm{mlp} d_\mathrm{mlp,u}^2 +
      d_\mathrm{mlp,u} d_\mathrm{state}^2) \Big) + {}
      \\
       & 
      O\Big( n_\mathrm{units} \cdot T
      \cdot
      (d_\mathrm{in} d_\mathrm{mlp,u} + n_\mathrm{mlp} d_\mathrm{mlp,u}^2 +
      d_\mathrm{mlp,u} d_\mathrm{state}) \Big) + {}
      \\
      & 
      O\Big(n_\mathrm{units} \cdot T \cdot d_\mathrm{state}^3\Big) + {}
      \\
      & 
    O\Big(T \cdot \big((n_\mathrm{units} d_\mathrm{state} + d_\mathrm{in}) \cdot d_\mathrm{mlp} + 
    n_\mathrm{mlp} d_\mathrm{mlp}^2 + d_\mathrm{mlp} d_\mathrm{out} \big) \Big)
      \\
      & = 
      \\
      & O\Big(n_\mathrm{units} \cdot T \cdot
      (d_\mathrm{in} d_\mathrm{mlp,u} + n_\mathrm{mlp} d_\mathrm{mlp,u}^2 +
      d_\mathrm{mlp,u} d_\mathrm{state}^2) \Big) + {}
      \\
      & 
      O\Big(n_\mathrm{units} \cdot T \cdot d_\mathrm{state}^3\Big) + {}
      \\
      & 
    O\Big(T \cdot \big((n_\mathrm{units} d_\mathrm{state} + d_\mathrm{in}) \cdot d_\mathrm{mlp} + 
    n_\mathrm{mlp} d_\mathrm{mlp}^2 + d_\mathrm{mlp} d_\mathrm{out} \big) \Big)
      \\
      & = 
      \\
      & O\Big(T \cdot
      (d_\mathrm{in} d_\mathrm{mlp} + n_\mathrm{mlp} d_\mathrm{mlp}^2 +
      d_\mathrm{mlp} d_\mathrm{state}^2) \Big) + {}
      \\
      & 
      O\Big(n_\mathrm{units} \cdot T \cdot d_\mathrm{state}^3\Big) + {}
      \\
      & 
    O\Big(T \cdot \big((n_\mathrm{units} d_\mathrm{state} + d_\mathrm{in}) \cdot d_\mathrm{mlp} + 
    n_\mathrm{mlp} d_\mathrm{mlp}^2 + d_\mathrm{mlp} d_\mathrm{out} \big) \Big)
      \\
      & = 
      \\
      & O\Big(T \cdot
      (d_\mathrm{in} d_\mathrm{mlp} + n_\mathrm{mlp} d_\mathrm{mlp}^2 +
      d_\mathrm{mlp} d_\mathrm{state}^2) + {}
      \\
      & 
      \qquad n_\mathrm{units} \cdot T \cdot d_\mathrm{state}^3 + {}
      \\
      & 
    \qquad T \cdot \big((n_\mathrm{units} d_\mathrm{state} + d_\mathrm{in}) \cdot d_\mathrm{mlp} + 
    n_\mathrm{mlp} d_\mathrm{mlp}^2 + d_\mathrm{mlp} d_\mathrm{out} \big) \Big)
      \\
      & = 
      \\
      & O\Big( T \cdot \big( 
      d_\mathrm{in} d_\mathrm{mlp} + n_\mathrm{mlp} d_\mathrm{mlp}^2 +
      d_\mathrm{mlp} d_\mathrm{state}^2 +
       n_\mathrm{units} \cdot d_\mathrm{state}^3 + {}
    n_\mathrm{units} \cdot d_\mathrm{mlp} d_\mathrm{state} + d_\mathrm{mlp} d_\mathrm{out} \big) \Big)
      \\
      & = 
      \\
      & O\Big( T \cdot \big( 
      n_\mathrm{units} \cdot d_\mathrm{state}^3 +
      d_\mathrm{mlp} \cdot (d_\mathrm{in}  + n_\mathrm{mlp} d_\mathrm{mlp} +
      d_\mathrm{state}^2 +  d_\mathrm{out}+ n_\mathrm{units}  d_\mathrm{state})    \big) \Big)
  \end{align*}
\end{proof}

\begin{theorem} \label{theorem:parallel-evaluation-complexity-cascade}
  On input $\tuple{W,U}$ for $W \in \mathbf{W}^\mathrm{rnc}_\mathbf{s}$ with size
  $$\mathbf{s} = \tuple{d_\mathrm{in}, d_\mathrm{out}, d_\mathrm{model}, d_\mathrm{state},
  n_\mathrm{units}, n_\mathrm{layers}, d_\mathrm{mlp}, n_\mathrm{mlp}},$$ 
  and $U \in \reals^{T \times d_\mathrm{in}}$,
  the output of \ref{alg:parallel-evaluation-rnc} is
  $S(U)$ for $S = \operatorname{RNC}(W)$,
  its depth is
  \begin{align*}
    O\Big(n_\mathrm{layer} \cdot \big(\log T \cdot \log d_\mathrm{state} 
    + n_\mathrm{mlp} \cdot \log d_\mathrm{mlp}
    +
    \log(n_\mathrm{units} d_\mathrm{state} + d_\mathrm{in} + d_\mathrm{model}) \big) \Big),
  \end{align*}
  and the work is
  \begin{align*}
      O\Big( T \cdot \big( 
      n_\mathrm{units} \cdot d_\mathrm{state}^3 +
      d_\mathrm{mlp} \cdot (d_\mathrm{in} +  + n_\mathrm{mlp} d_\mathrm{mlp} +
      d_\mathrm{state}^2 +  d_\mathrm{model} + d_\mathrm{out} + n_\mathrm{units}
      d_\mathrm{state})  \big) \Big).
  \end{align*}
  Furthermore,
  for fixed $n_\mathrm{mlp}$ and $d \defeq \max\{ d_\mathrm{in},d_\mathrm{out},d_\mathrm{mlp},
  n_\mathrm{units} \cdot d_\mathrm{state}\}$,
  the depth is
  \begin{align*}
    O\Big( n_\mathrm{layer} \cdot \big(\log T \cdot \log d_\mathrm{state} + \log d \big) \Big),
  \end{align*}
  and the work is
  \begin{align*}
      O\Big( T \cdot n_\mathrm{layers} \cdot d \cdot \big( d + d_\mathrm{state}^2 \big) \Big).
  \end{align*}
\end{theorem}
\begin{proof}
  We note that the algorithm is obtained from the sequential 
  algorithm~\ref{alg:sequential-evaluation-rnc} in a straighforward manner.
  Thus, correctness follows by the same arguments as in the proof of
  Lemma~\ref{lemma:sequential-evaluation-complexity-cascade}.
  In particular, $\mathtt{SeqLayerEval}$ is now replaced by $\mathtt{ParLayerEval}$,
  which yields the same result, by Lemma~\ref{lemma:parallel-evaluation-complexity-layer};
  and the loops in $\mathtt{SeqCascEval}$ have independent iterations and hence their iterations
  can be executed in parallel.

  Then, the depth and work are given by the sum of depth and work, respectively, of the calls to
  $\mathtt{ParLayerEval}$,
  to evaluate a layer having 
  input dimension $d_\mathrm{in}$ and output dimension $d_\mathrm{model}$ for $i = 0$,
  input and output dimension $d_\mathrm{model}$ for each $i \in \iival{1}{n_\mathrm{layers}-2}$,
  and
  input dimension $d_\mathrm{model}$ and output dimension $d_\mathrm{out}$ for $i =
  n_\mathrm{layers}-1$.
  For the rest, the size of each layer is described by
  $\tuple{d_\mathrm{state}, n_\mathrm{units}, d_\mathrm{mlp}, n_\mathrm{mlp}}$.
  By Lemma~\ref{lemma:parallel-evaluation-complexity-layer}
  the depth of the execution for the first layer is
  \begin{align*}
    O\Big(\log T \cdot \log d_\mathrm{state} 
    + n_\mathrm{mlp} \cdot \log d_\mathrm{mlp}
    +
    \log(n_\mathrm{units} d_\mathrm{state} + d_\mathrm{in}) \Big)
  \end{align*}
  the depth for each layer $i \in \iival{1}{n_\mathrm{layers}-1}$
  is
  \begin{align*}
    O\Big(\log T \cdot \log d_\mathrm{state} 
    + n_\mathrm{mlp} \cdot \log d_\mathrm{mlp}
    +
    \log(n_\mathrm{units} d_\mathrm{state} + d_\mathrm{model}) \Big)
  \end{align*}
  Thus, overall the depth is
  \begin{align*}
    O\Big(n_\mathrm{layer} \cdot \big(\log T \cdot \log d_\mathrm{state} 
    + n_\mathrm{mlp} \cdot \log d_\mathrm{mlp}
    +
    \log(n_\mathrm{units} d_\mathrm{state} + d_\mathrm{in} + d_\mathrm{model}) \big) \Big).
  \end{align*}
  For fixed $n_\mathrm{mlp}$ and $d \defeq \max\{ d_\mathrm{in},d_\mathrm{out},d_\mathrm{mlp},
  n_\mathrm{units} \cdot d_\mathrm{state}\}$,
  \begin{align*}
    O\Big( n_\mathrm{layer} \cdot \big(\log T \cdot \log d_\mathrm{state} + \log d \big) \Big).
  \end{align*}

  By Lemma~\ref{lemma:parallel-evaluation-complexity-layer}
  the work of the execution for the first layer is
  \begin{align*}
      O\Big( T \cdot \big( 
      n_\mathrm{units} \cdot d_\mathrm{state}^3 +
      d_\mathrm{mlp} \cdot (d_\mathrm{in}  + n_\mathrm{mlp} d_\mathrm{mlp} +
      d_\mathrm{state}^2 +  d_\mathrm{model}+ n_\mathrm{units}  d_\mathrm{state})    \big) \Big)
  \end{align*}
  the work for each layer $i \in \iival{1}{n_\mathrm{layers}-2}$
  is
  \begin{align*}
      O\Big( T \cdot \big( 
      n_\mathrm{units} \cdot d_\mathrm{state}^3 +
      d_\mathrm{mlp} \cdot (d_\mathrm{model}  + n_\mathrm{mlp} d_\mathrm{mlp} +
      d_\mathrm{state}^2 +  d_\mathrm{model}+ n_\mathrm{units}  d_\mathrm{state})    \big) \Big)
  \end{align*}
  and for the last layer, when $i = n_\mathrm{layers}-1$,
  \begin{align*}
      O\Big( T \cdot \big( 
      n_\mathrm{units} \cdot d_\mathrm{state}^3 +
      d_\mathrm{mlp} \cdot (d_\mathrm{model}  + n_\mathrm{mlp} d_\mathrm{mlp} +
      d_\mathrm{state}^2 +  d_\mathrm{out}+ n_\mathrm{units}  d_\mathrm{state})    \big) \Big)
  \end{align*}
  Thus, overall the work is
  \begin{align*}
      O\Big( T \cdot \big( 
      n_\mathrm{units} \cdot d_\mathrm{state}^3 +
      d_\mathrm{mlp} \cdot (d_\mathrm{in} +  + n_\mathrm{mlp} d_\mathrm{mlp} +
      d_\mathrm{state}^2 +  d_\mathrm{model} + d_\mathrm{out} + n_\mathrm{units}
      d_\mathrm{state})  \big) \Big)
  \end{align*}
  For fixed $n_\mathrm{mlp}$ and $d \defeq \max\{ d_\mathrm{in},d_\mathrm{out},d_\mathrm{mlp},
  n_\mathrm{units} \cdot d_\mathrm{state}\}$,
  \begin{align*}
    &
      O\Big( T \cdot n_\mathrm{layers} \cdot \big( 
      n_\mathrm{units} \cdot d_\mathrm{state}^3 + d \cdot (d + d_\mathrm{state}^2)  \big) \Big)
      \\
      & =
      O\Big( T \cdot n_\mathrm{layers} \cdot \big( 
      d \cdot d_\mathrm{state}^2 + d \cdot (d + d_\mathrm{state}^2)  \big) \Big)
      \\
      & =
      O\Big( T \cdot n_\mathrm{layers} \cdot d \cdot \big( d + d_\mathrm{state}^2 \big) \Big).
  \end{align*}
\end{proof}

\clearpage{}%

\clearpage{}%

\section{BIBS and BIBO Stability of MinMax RNCs}
\label{sec:appendix:stability}

\localtableofcontents

This appendix proves the stability result stated as
Theorem~\ref{th:stability-simpler}.
More generally, we establish \emph{uniform bounded-input bounded-state~(uBIBS)} and
\emph{uniform bounded-input bounded-output~(uBIBO)} properties for parametric MinMax
RNCs, together with their weaker variants, as defined in
Appendix~\ref{sec:properties}.

The key observation is that a MinMax update
\begin{align*}
  x_t
  \,=\,
  \minmaxop{x_{t-1}}{R_t}{s_t}
  \,=\,
  (R_t\otimes x_{t-1})
  \oplus
  s_t
\end{align*}
satisfies a one-step bound depending only on the current reset and set values,
and not on the magnitude of the previous state.
Thus, whenever the reset and set functions are bounded on bounded input and
parameter sets, the state remains uniformly bounded independently of the time
horizon.

We first formalise this observation and use it to prove uBIBS for individual
MinMax recurrent units.
We then lift the property to arbitrary MinMax RNC dynamics using the general
closure results for recurrent neural circuits.
Finally, boundedness of the analytic output map yields uBIBO, from which the
corresponding weak BIBS/BIBO properties follow.
Theorem~\ref{th:stability-simpler} is then obtained directly from these stronger
uniform results.

\subsection{A one-step MinMax state-value bound}
\label{subsec:a-onestep-minmax-value-bound}

The following elementary bound is used to establish uBIBS of MinMax recurrent units. 

\begin{lemma}[Value bound for a MinMax update]
  \label{lem:minmax-update-value-bound}
  \label{lem:minmax-sdv-update-bound}
  Let $x\in\reals^n$,
  let $R\in\reals^{n\times n}$,
  let $s\in\reals^n$,
  and 
  let $F \defeq \minmaxop{x}{R}{s}$.
  Then
  \begin{equation*}
    \label{eq:minmax-update-value-bound}
    \|F\|_\infty
    \leq
    \max
    \left\{
      \|s\|_\infty,
      \|R\|_\mathrm{max}
    \right\}.
  \end{equation*}
\end{lemma}
\begin{proof}
  Let $M \defeq \max \left\{ \|s\|_\infty, \|R\|_\mathrm{max} \right\}$.
  We have
  \begin{align*}
    \getentry{F}{i}
    \,=\,
    \max
    \left\{
      \getentry{s}{i},
      \max_{j\in\iival{1}{n}}
      \min\big(\getentry{R}{i,j},\getentry{x}{j}\big)
    \right\}
    \qquad\;
    \forall i\in\iival{1}{n}.
  \end{align*}
  Since
  \begin{align*}
    \min\big(\getentry{R}{i,j},\getentry{x}{j}\big) 
    \,\leq\, 
    \getentry{R}{i,j}
    \,\leq\, 
    M 
    \qquad\text{and}\qquad
    \getentry{s}{i} 
    \,\leq\,
    M 
    \qquad\quad 
    \forall i,j \in \iival{1}{n},
  \end{align*}
  we have 
  \begin{align*}
    \getentry{F}{i} 
    \leq 
    M 
    \qquad\quad 
    \forall i \in \iival{1}{n}.
  \end{align*}

  Furthermore, since $\getentry{F}{i}$ is a maximum including
  $\getentry{s}{i}$ and
  $|\getentry{s}{i}|\leq\|s\|_\infty\leq M$, we have
  \begin{align*}
    \getentry{F}{i}
    \,\geq\,
    \getentry{s}{i}
    \,\geq\,
    -M
    \qquad\quad
    \forall i\in\iival{1}{n}.
  \end{align*}

  Hence 
  \begin{align*}
    |\getentry{F}{i}| 
    \leq 
    M
    \qquad\quad 
    \forall i \in \iival{1}{n},
  \end{align*}
  proving the claim.
\end{proof}

\subsection{Uniform-BIBS of MinMax recurrent units and of MinMax RNCs}

\begin{proposition}[uBIBS of parametric MinMax neurons]
  \label{prop:minmax-unit-ubibs}
  The class $\classparneuralunits{MinMax}$ satisfies uBIBS.
\end{proposition}
\begin{proof}
  Let us consider an arbitrary
  $r\in\nonnegreals$
  and an arbitrary parametric MinMax neuron $D \in \classparneuralunits{MinMax}$ 
  of state dimension $n$
  given by
  \begin{equation*}
    D
    =
    \mctuple{X,U,f}{\Theta}{R,s}.
  \end{equation*}
  By
  Definition~\ref{def:parametric-minmax-recurrent-unit},
  the joint reset--set function
  $\rho \defeq \ptuple{R,s}$ is a parametric analytic neural map.
  Therefore,
  Corollary~\ref{cor:restricted-feedforward-neural-map-boundedness}
  yields
  $M_r\in\nonnegreals$
  such that
  \begin{equation*}
    \|\fpinst{\rho}{u}{\theta}\|
    \leq
    M_r
    \qquad\quad
    \forall u\in\truncation{U}{r},\,
    \forall\theta\in\truncation{\Theta}{r}.
  \end{equation*}
  Since
  $\|\fpinst{R}{u}{\theta}\|_\mathrm{max}
  \leq \|\fpinst{\rho}{u}{\theta}\|$
  and
  $\|\fpinst{s}{u}{\theta}\|_\infty
  \leq \|\fpinst{\rho}{u}{\theta}\|$,
  it follows that
  \begin{equation*}
    \max
    \left\{
      \|\fpinst{R}{u}{\theta}\|_\mathrm{max},
      \|\fpinst{s}{u}{\theta}\|_\infty
    \right\}
    \leq
    M_r.
  \end{equation*}

  Let us fix arbitrary
  $x_0\in\truncation{X}{r}$,
  $u\in\infseqs{\truncation{U}{r}}$,
  and
  $\theta\in\truncation{\Theta}{r}$.
  Let
  \begin{equation*}
    R_t
    \defeq
    \fpinst{R}{u_t}{\theta},
    \qquad
    s_t
    \defeq
    \fpinst{s}{u_t}{\theta}.
    \qquad\quad \forall t\in\timedomain.
  \end{equation*}
  By
  \eqref{eq:minmax-update-value-bound},
  \begin{equation*}
    \|x_t\|_\infty
    \leq
    M_r
    \qquad\quad
    \forall t\in\timedomain.
  \end{equation*}
  Hence, using
  $\|x\|\leq\sqrt{n} \cdot \|x\|_\infty$ and the assumption $\|x_0\| \leq r$,
  \begin{equation*}
    \sup_{t\in\naturals}
    \|x_t\|
    \leq
    \max\big\{r,\,\sqrt{n} \cdot M_r\big\}.
  \end{equation*}
  Since $x_0$, $u$, and $\theta$ are arbitrary,
  this bound is uniform over the bounded input sequence, initial state, and
  parameter value.
  Thus $D$ satisfies uBIBS.
  Since $D$ is arbitrary in $\classparneuralunits{MinMax}$, 
  we conclude that the class $\classparneuralunits{MinMax}$ satisfies uBIBS.
\end{proof}

\begin{corollary}[uBIBS of parametric MinMax RNC dynamics]
  \label{cor:minmax-rnc-ubibs}
  The class $\rncdynop\big(\classparneuralunits{MinMax}\big)$ satisfies uBIBS.
\end{corollary}
\begin{proof}
  By
  Proposition~\ref{prop:minmax-unit-ubibs},
  the class
  $\classparneuralunits{MinMax}$
  satisfies uBIBS.
  The claim follows from
  Theorem~\ref{thm:rnc-ubibs-closure}.
\end{proof}

\subsection{BIBS and BIBO of MinMax RNCs}

Since uBIBS implies wBIBS,
Proposition~\ref{prop:minmax-unit-ubibs}
and
Corollary~\ref{cor:minmax-rnc-ubibs}
also yield wBIBS for parametric MinMax neurons and MinMax RNC dynamics.

\begin{theorem}[uBIBS and uBIBO of parametric MinMax RNCs]
  \label{theorem:minmax-rnc-bibs-bibo}
  The class
  $\rncop\big(\classparneuralunits{MinMax}\big)$
  satisfies uBIBS and uBIBO, and hence also wBIBS and wBIBO.
\end{theorem}
\begin{proof}
  Let
  $S \in \rncop\big(\classparneuralunits{MinMax}\big)$
  be arbitrary.
  There exist
  $D \in \rncdynop\big(\classparneuralunits{MinMax}\big)$,
  an initial state $x_0$ of $D$,
  and a parametric analytic map $h$
  such that
  \begin{equation*}
    S
    =
    \ptuple{D,x_0}
    \triangleright
    h.
  \end{equation*}

  By
  Corollary~\ref{cor:minmax-rnc-ubibs},
  $D$
  satisfies uBIBS.
  Hence, by
  Definition~\ref{def:parametric-ubibs-ubibo},
  $S$
  satisfies uBIBS and therefore also wBIBS.

  It remains to prove uBIBO.
  Fix
  $r\in\nonnegreals$
  and set
  \begin{equation*}
    \hat r
    \defeq
    \max\{
      r,
      \|x_0\|
    \}.
  \end{equation*}
  Since
  $D$
  satisfies uBIBS,
  there exists
  $B^X_{\hat r}\in\nonnegreals$
  such that, for every
  \begin{equation*}
    \theta\in\truncation{\Theta}{r},
    \qquad
    u\in\infseqs{\truncation{U}{r}},
  \end{equation*}
  the state sequence
  $(x_t)_{t\in\timedomain}$
  of
  $\pinst{D}{\theta}$
  starting from the fixed state
  $x_0$
  satisfies
  \begin{equation*}
    \sup_{t\in\timedomain}
    \|x_t\|
    \leq
    B^X_{\hat r}.
  \end{equation*}
  Indeed,
  $\theta\in\truncation{\Theta}{\hat r}$,
  $u\in\infseqs{\truncation{U}{\hat r}}$,
  and
  $x_0\in\truncation{X}{\hat r}$.

  Consequently, the output contexts
  \begin{equation*}
    p_t
    \defeq
    \ptuple{
      x_{t-1},
      u_t,
      x_t
    }
    \qquad\quad
    \forall t\in\posnaturals
  \end{equation*}
  are uniformly bounded.
  Hence there exists
  $\rho_r\in\nonnegreals$,
  depending only on
  $r$,
  $x_0$,
  and
  $B^X_{\hat r}$,
  such that
  \begin{equation*}
    \|p_t\|
    \leq
    \rho_r
    \qquad\quad
    \forall t\in\posnaturals.
  \end{equation*}

  Let
  $\theta_h$
  be the parameter block of the output map.
  Since
  $\|\theta_h\|\leq\|\theta\|\leq r$,
  setting
  \begin{equation*}
    \bar r
    \defeq
    \max\{
      r,
      \rho_r
    \},
  \end{equation*}
  Corollary~\ref{cor:restricted-feedforward-neural-map-boundedness}
  applied to
  $h$
  yields
  $B^Y_r\in\nonnegreals$
  such that
  \begin{equation*}
    \|y_t\|
    =
    \left\|
      \fpinst{h}{p_t}{\theta_h}
    \right\|
    \leq
    B^Y_r
    \qquad\quad
    \forall t\in\posnaturals,
  \end{equation*}
  uniformly over all parameter values and input sequences bounded by
  $r$.
  Thus
  $S$
  satisfies uBIBO, and therefore also wBIBO.
  Since
  $S$
  was arbitrary, the claim follows.
\end{proof}

We restate Theorem~\ref{th:stability-simpler} equivalently---but more precisely---in the notation of
Appendix~\ref{sec:properties} as follows.

\begin{theorem}[Equivalent formulation of Theorem~\ref{th:stability-simpler}]
  \label{th:stability-precise}
  Let
  $S = \ctuple{X,U,f,x_0,Y,h}{\Theta}$
  be a parametric MinMax RNC.
  For every
  $r\in\nonnegreals$,
  there exists
  $B_r\in\nonnegreals$
  such that, for every
  \begin{align*}
    \theta\in\truncation{\Theta}{r},
    \qquad
    u=(u_t)_{t=1}^T\in\truncation{U}{r}^+,
  \end{align*}
  the corresponding state sequence
  $(x_t)_{t=1}^T$
  and output sequence
  $(y_t)_{t=1}^T$,
  given by
  \begin{equation*}
    x_t
    \defeq
    \repval{\dynop{\pinst{S}{\theta}}}
    (x_0,\seq{u}{1}{t}),
    \qquad
    y_t
    \defeq
    \repval{\pinst{S}{\theta}}
    (\seq{u}{1}{t})
    \qquad\quad
    \forall t\in\iival{1}{T},
  \end{equation*}
  satisfy
  \begin{equation*}
    \max_{t\in\iival{1}{T}}\|x_t\|
    \leq
    B_r,
    \qquad
    \max_{t\in\iival{1}{T}}\|y_t\|
    \leq
    B_r.
  \end{equation*}
\end{theorem}
\begin{proof}
  By
  Theorem~\ref{theorem:minmax-rnc-bibs-bibo},
  there exist uniform state and output bounds
  $B^X_r,B^Y_r\in\nonnegreals$.
  The claim follows with
  $B_r \defeq \max\{ B^X_r, B^Y_r \}$.
\end{proof}

\clearpage{}%

\clearpage{}%

\todo{}

\section{Gradients of MinMax RNCs}
\label{app:gradient}
\label{sec:minmax-selected-intensional-derivative-bounds}

\localtableofcontents

This appendix provides the full proofs of the gradient results stated in
Section~\ref{sec:main-body-gradient}.
The results have two complementary goals.

First, we prove
Theorem~\ref{thm:gradient-bounded-main},
showing that the derivatives relevant to training parametric MinMax RNCs remain
uniformly bounded on bounded sets, independently of the input horizon.
More precisely, we show that, for every admissible MinMax-algebraic computation
graph selected for the loss,
the corresponding MinMax selected pseudo-derivative is uniformly bounded;
under the appropriate intensional-admissibility assumptions, it is an
intensional derivative, and the same bound consequently holds for the ordinary
loss derivative almost everywhere.

The main difficulty is that a MinMax recurrence may be represented by different
computation-graph topologies.
In particular, sequential, balanced-tree, parallel-scan, and mixed evaluation
strategies may induce different selected derivatives at nondifferentiable
points.
We therefore prove a stronger, topology-independent result.
We first show that selected pseudo-derivatives of scalar minimum and maximum
nodes are non-expansive in a suitable row norm.
We then introduce MinMax-algebraic recurrence computation graphs and prove that
the selected derivative of any such graph is controlled only by the selected
derivatives of its reset and set source graphs, independently of both the
recurrence horizon and the internal graph topology.
Combining this property with uniform derivative bounds for the analytic
parts of the architecture yields
graph-uniform selected uBIBSD for complete MinMax RNC dynamics.
Finally, we propagate this state-derivative bound through the output and loss
maps to obtain the bounded-loss-derivative statement of
Theorem~\ref{thm:gradient-bounded-main}.

Second, we prove
Theorem~\ref{thm:gradient-nonvanishing}.
which show that bounded state derivatives need not vanish with the sequence
length.

The appendix is therefore organised around these two aspects of gradient
behaviour:
uniform boundedness over arbitrary horizons and admissible computation-graph
topologies, and the existence of MinMax recurrences whose state derivatives
remain non-vanishing over arbitrarily long finite horizons.

\subsection{Preliminaries}

\todo{}

\subsection{Relevant Class of Functions}

\begin{closingdefinition}[Analytic-MinMax primitive functions]
  \label{def:analytic-minmax-primitive-functions}
  The class
  $\minmaxprimitivefuncclass$
  of \emph{analytic-MinMax primitive functions}
  consists of:
  \begin{itemize}
    \item
      all analytic Euclidean functions;
    \item
      all MinMax reductions $\minreduction{n}$ and $\maxreduction{n}$ for $n\in\naturals$ with 
      $n \geq 2$ as per Definition~\ref{def:minmax-scalar-reductions}.
  \end{itemize}
\end{closingdefinition}

\begin{closingdefinition}[Analytic-MinMax functions]
  \label{def:analytic-minmax-functions}
  The class
  $\minmaxfuncclass$
  of \emph{analytic-MinMax functions (AMM-functions)}
  is the closure of $\minmaxprimitivefuncclass$.\footnote{The notion of \emph{closure} is
  introduced in Definition~\ref{def:function-class-closure}.}
  Namely,
  it is the smallest class such that:
  (i)~it contains $\minmaxprimitivefuncclass$,
  (ii)~it is closed under function tupling,
  and
  (iii)~it is closed under function composition.
\end{closingdefinition}

\begin{closingremark}[Terminology]
  \label{rem:analytic-minmax-terminology}
  The adjective
  \emph{analytic-MinMax}
  refers to the fact that the class includes two kinds of primitive functions:
  \begin{itemize}
    \item
      analytic Euclidean functions;
    \item
      MinMax reductions.
  \end{itemize}
  In particular,
  \emph{analytic-MinMax}
  does not mean that every analytic-MinMax function is analytic.
  In fact, Min and Max reductions are generally non-differentiable at
  points where an extremum is attained by more than one argument.
\end{closingremark}

We introduce the notion of 
\emph{active-variable index function}, which we will use to define pseudo-derivatives of MinMax
functions.
\begin{definition} \label{def:active-variable-index-function}
  Let $f : \reals^n \to \reals$ be a MinMax reduction.
  The \emph{active-variable index function} of $f$ is the function
  $I : \reals^n \to \powersetplus{\iival{1}{n}}$
  given by
  \begin{equation*}
    I(x_1, \ldots, x_n) \defeq \big\{ i \in \iival{1}{n} \mid x_i = f(x_1, \ldots, x_n) \big\}
    \qquad
    \forall \ptuple{x_1, \ldots, x_n} \in \reals^n.
  \end{equation*}
\end{definition}

\subsection{Pseudo-differentiating Min and Max Functions}
\label{subsec:gradient-prel}

The derivative of min and max functions is is not guaranteed to exist everywhere.

To provide some intuition, we discuss the case of the max function $\max_2: \reals^2 \to \reals$
given by $\max_2(x_1, x_2) = x_1 \oplus x_2$.
Its gradient is well-defined at 
$(\bar{x}_1,\bar{x}_2)$ when $\bar{x}_1 \neq \bar{x}_2$, since in this case 
its change is given 
by the change of $\bar{x}_1$ if $\bar{x}_1 > \bar{x}_2$,
and
by the change of $\bar{x}_2$ if $\bar{x}_2 > \bar{x}_1$.
However, 
when $\bar{x}_1 = \bar{x}_2$
the gradient of $\max_2$ at $(\bar{x}_1,\bar{x}_2)$ does not exist.

To describe the change at all points,
one can employ a \emph{subgradient} of $\max_2$, a generalisation of the gradient for convex
functions.
In fact, the subgradient is not unique, and one refers to the set of all subgradients as
the \emph{subdifferential}.
One subgradient of $\max_2$ at $(\bar{x}_1,\bar{x}_2)$ when $\bar{x}_1 = \bar{x}_2$
is given by $(1/2, 1/2)$, which essentially says that the change of $\max_2$ is the average change
of its inputs.
This is in fact the subgradient of minimum (Euclidean) norm, which is the one employed in 
\citep{pytorchautograd}.

The issue with subgradients is that they are only defined for convex functions.
For concave functions such as min, one can employ the dual notion of supergradient.
However, compositions of min and max functions will not be
convex nor concave, nor compositions that will additionally include other functions. 
In particular, the loss functions of MinMax RNCs are not convex nor concave in general.

To address the issue, we focus on the strategy adopted in \citep{pytorchautograd} to be able to 
always compute a \emph{pseudo-gradient}, with two main benefits.
First, we are able to prove properties of the actual gradient, holding almost everywhere, such as
its existence and boundedness.
Second, we are able to prove properties for the pseudo-gradient computed by \cite{pytorchautograd};
notably, the fact that it coincides with the gradient almost everywhere, and that it is always  
bounded, even at points where the actual gradient does not exist.

The strategy adopted in \citep{pytorchautograd} is to compute a pseudo-gradient 
over a computation graph via the chain rule, computing locally at each node:
\begin{itemize}
  \item
    the gradient of the function for the node when such function is differentiable,
  \item
    the \emph{subgradient} of minimum norm in the case of a convex function,
    and
  \item
    the \emph{supergradient} of minimum norm in the case of a concave function.
\end{itemize}
The result of such computation is not guaranteed to be a gradient.
Furthermore, the outcome may be different for different computation graphs---even though such
computation graphs are equivalent in the sense that they represent the same function.
Formally, this corresponds to the graph-induced pseudo-derivative of
Definition~\ref{def:graph-induced-pseudo-derivative}.

We employ the theory of 
\emph{piecewise analytic under analytic partition (PAP)} functions \citep{lee2020correctness},
to show that the loss functions of MinMax RNCs
are differentiable almost everywhere,
and that the pseudo-gradients computed in the above manner agree with the gradient almost everywhere.
In particular, we show that the loss functions of MinMax RNCs are PAP; and also that, for all 
\emph{algebraic computation graphs}, 
the corresponding pseudo-gradient is an \emph{intensional derivative} as defined
by~\cite{lee2020correctness}. 

We introduce subgradients and supergradients, cf.\ \citep{rockafellar1997convex}.

\begin{definition}[Subgradient]
  Let $X = \reals^n$,
  let $f: X \to \reals$ be a convex function,
  and 
  let $x \in X$.
  A vector $g \in \reals^n$ is a \emph{subgradient} of $f$ at $x$ if, for all $z \in X$,
  \begin{align*}
    f(z) \geq f(x) + g^\transposeop (z-x).
  \end{align*}
  Then,
  the \emph{subdifferential} $\subdiff f(x)$ is the set consisting of each subgradient of 
  $f$ at $x \in X$.
\end{definition}

\begin{definition}[Supergradient]
  Let $X = \reals^n$,
  let $f: X \to \reals$ be a concave function,
  and 
  let $x \in X$.
  A vector $g \in \reals^n$ is a \emph{supergradient} of $f$ at $x$ if 
  $-g \in \subdiff (-f)(x)$.
  Then,
  the \emph{superdifferential} $\supdiff f(x)$ is the set consisting of
  each supergradient of $f$ at $x \in X$.
\end{definition}

\begin{restatable}{proposition}{propmaxsubgradienthull}
  \label{prop:max-subgradient-hull}
  Let $f: \reals^n \to \reals$ be the function $f(x) = \bigoplus_{i=1}^n x_i$,
  let $I$ be the active-variable 
  index function of~$f$~(Definition~\ref{def:active-variable-index-function}),
  and
  let $x \in \reals^n$.
  Then, the subdifferential of $f$ at $x$ is given by
  \begin{align*}
    \subdiff f(x)
    =
    \left\{
      g\in\reals^n \;\middle|\;
      \textstyle\bigwedge_{i=1}^n g_i\ge 0,\
      \textstyle\sum_{i\in I(x)} g_i=1
    \right\}.
  \end{align*}
  Furthermore, the unique subgradient $g^\star \in \partial_\circ f(x)$ of minimum (Euclidean) norm
  is given by
  \begin{align*}
    g^\star = (g^\star_1, \ldots, g^\star_n)
    \qquad
    \text{ where }\;
    g_i^\star
    =
    \begin{cases}
      1/|I(x)|
      & \text{ if }\, i\in I(x),
      \\
      0 & \text{ otherwise}, 
    \end{cases}
  \end{align*}
  and its norm is $\|g^\star\| = 1/\sqrt{|I(x)|}$.
\end{restatable}
\begin{proof}
  Deferred to Section~\ref{appendix:sec:gradient-aux}.
\end{proof}

\begin{restatable}{proposition}{propminsubgradienthull}
  \label{prop:min-subgradient-hull}
  Let $f: \reals^n \to \reals$ be the function $f(x) = \bigodot_{i=1}^n x_i$,
  let $I$ be the active-variable index function of $f$~(Definition~\ref{def:active-variable-index-function}),
  and
  let $x \in \reals^n$.
  Then, the superdifferential of $f$ at $x$ is given by
  \begin{align*}
    \supdiff f(x)
    =
    \left\{
      g\in\reals^n \;\middle|\;
      \textstyle\bigwedge_{i=1}^n g_i\ge 0,\
      \textstyle\sum_{i\in I(x)} g_i=1
    \right\}.
  \end{align*}
  Furthermore, the unique supergradient $g^\star \in \partial_\circ f(x)$ of minimum (Euclidean)
  norm is given by
  \begin{align*}
    g^\star = (g^\star_1, \ldots, g^\star_n)
    \qquad
    \text{ where }\;
    g_i^\star
    =
    \begin{cases}
      1/|I(x)|
      & \text{ if }\, i\in I(x),
      \\
      0 & \text{ otherwise}, 
    \end{cases}
  \end{align*}
  and its norm is $\|g^\star\| = 1/\sqrt{|I(x)|}$.
\end{restatable}
\begin{proof}
  Deferred to Section~\ref{appendix:sec:gradient-aux}.
\end{proof}

\begin{closingdefinition}[Analytic-MinMax pseudo-derivative operator]
  \label{def:minmax-pseudoderivative-operator}
  The
  \emph{analytic-MinMax pseudo-derivative operator},
  abbreviated as \emph{AMM-PD-operator},
  is the pseudo-derivative operator $\minmaxsdv$ applicable to $\minmaxprimitivefuncclass$
  such that:
  \begin{itemize}
    \item
      when $f$ is analytic,
      we have $\minmaxsdv f \defeq \dvop{f}$;
    \item
      when
      $f:\reals^n\to\reals$
      is the Max reduction
      \begin{equation*}
        f(x)
        =
        \bigoplus_{i=1}^n x_i,
      \end{equation*}
      the pseudo-derivative
      $\minmaxsdv f(x)$
      is the unique subgradient of
      $f$
      at
      $x$
      having minimum Euclidean norm;

    \item
      when
      $f:\reals^n\to\reals$
      is the Min reduction 
      \begin{equation*}
        f(x)
        =
        \bigodot_{i=1}^n x_i,
      \end{equation*}
      the pseudo-derivative
      $\minmaxsdv f(x)$
      is the unique supergradient of
      $f$
      at
      $x$
      having minimum Euclidean norm.
  \end{itemize}
\end{closingdefinition}

\begin{proposition} \label{prop:max-partition}
  Let $f: \reals^n \to \reals$ be the function $f(x) = \bigoplus_{i=1}^n x_i$, 
  and 
  let $I$ be its active variable index function~(Definition~\ref{def:active-variable-index-function}).
  Let $\{ I_1, \ldots, I_k \} = \im(I)$,
  and
  let $A = \{ A_1, \ldots, A_k \}$ where $A_i = \{ x \in \reals^n \mid I(x) = I_i \}$.
  Then $A$ is a partition of $\reals^n$.
\end{proposition}
\begin{proof}
  First, we show that each $A_i$ is non-empty. 
  Since $I_i \in \im(I)$, 
  there exists 
  $x\in \reals^n$ such that $I(x)=I_i$,
  hence $x\in A_i$, 
  and 
  hence $A_i \neq \emptyset$.

  Second, we show $\bigcup_{i=1}^k A_i = \reals^n$. 
  Let $x\in \mathbb{R}^n$. 
  Since $I(x) \in \im(I)$, there exists $i \in \iival{1}{k}$ such that
  \begin{align*}
    I(x) = I_i.
  \end{align*}
  Thus $x \in A_i \subseteq \bigcup_{i=1}^k A_i$.

  Third, we show that the sets of $A$ are pair-wise disjoint. 
  Suppose that $x \in A_i \cap A_j$.
  Then
  \begin{align*}
    I(x) = I_i,
    \qquad
    I(x) = I_j,
  \end{align*}
  and hence
  \begin{align*}
    I_i = I_j.
  \end{align*}
  Since $I_1, \ldots, I_k$ are distinct, it follows that $i=j$, and hence $A_i = A_j$.

  Thus the sets of $A$ are non-empty, pair-wise disjoint, and their union is $\reals^n$.
  Therefore $\mathcal{A}$ is a partition of $\reals^n$.
\end{proof}

\paragraph{Required background.}
See \S\ref{subsec:pap-intensional-preliminaries}
for the definition of 
\emph{piecewise analytic under analytic partition (PAP) function}
and
\emph{intensional derivatives} of a PAP function.

\begin{proposition} \label{prop:max-pap}
  The function $f(x) = \bigoplus_{i=1}^n x_i$ is PAP, and 
  $\minmaxsdv f$ is an intensional derivative of $f$.
\end{proposition}
\begin{proof}
  We show a PAP representation $\gamma$ of $f$.
  Let $I$ be the active-variable index function of $f$,
  and
  let $\{I_1, \ldots, I_k\} = \powersetplus{\iival{1}{n}}$.
  For each $i \in \iival{1}{k}$,
  let $\rho_i$ be an arbitrary index in $I_i$.
  For each $i \in \iival{1}{k}$ and each $j \in \iival{1}{n}$,
  let us define 
  the functions $g_{i,j}^{+},g_{i,j}^{-} : \reals^n \to \reals$ as 
  \begin{align*}
    g_{i,j}^{+}(x) 
    \defeq 
    \begin{cases}
      1 & \text{ if } j \in I_i,
      \\
      x_{\rho_i} - x_j & \text{ otherwise},
    \end{cases}
    \quad\qquad
    g_{i,j}^{-}(x) 
    \defeq 
    \begin{cases}
      (x_{\rho_i} - x_j)^2 & \text{ if } j \in I_i,
      \\
      0 & \text{ otherwise}.
    \end{cases}
  \end{align*}
  For each $i \in \iival{1}{k}$,
  Then, we define $\gamma \defeq \{ \tuple{A_i, f_i} \mid i \in \iival{1}{k} \}$ where
  \begin{align*}
    f_i(x) 
    \defeq 
    \frac{1}{|I_i|} \sum_{j \in I_i} x_j,
    \qquad
    A_i 
    \defeq 
    \left\{ 
      x \in \reals^n \,\middle|\, 
      \textstyle\bigwedge_{j=1}^n g_{i,j}^{+}(x) > 0 \land 
      \textstyle\bigwedge_{j=1}^n g_{i,j}^{-}(x) \leq 0 
    \right\}.
  \end{align*}

  We first show that $\{ A_i \mid i \in \iival{1}{k} \}$ is a partition of $\reals^n$.
  By Proposition~\ref{prop:max-partition},
  it suffices to show that
  each $A_i$ is given by $A_i = \{x \in \reals^n \mid I(x) = I_i \}$.
  Namely, it suffices to show that the following equality holds for every $i \in \iival{1}{k}$,
  \begin{gather}
    \label{eq:prop:max-pap-1}
    \left\{ 
      x \in \reals^n \,\middle|\, 
      \textstyle\bigwedge_{j=1}^n g_{i,j}^{+}(x) > 0 \land 
      \textstyle\bigwedge_{j=1}^n g_{i,j}^{-}(x) \leq 0 
    \right\} 
    \,=\,
    \{x \in \reals^n \mid I(x) = I_i \}.
  \end{gather}
  For every $i \in \iival{1}{k}$,
  and 
  every $x \in \reals^n$,
  we have
  \begin{align*}
    & \bigwedge_{j=1}^n g_{i,j}^{+}(x) > 0
    \\
    & \Leftrightarrow\;
    \bigwedge_{j\in I_i} 1 > 0
    \land
    \bigwedge_{j \notin I_i} x_{\rho_i} - x_j > 0
    \\
    & \Leftrightarrow\;
    \bigwedge_{j \notin I_i} x_{\rho_i} > x_j
  \end{align*}
  and
  \begin{align*}
    & \bigwedge_{j=1}^n g_{i,j}^{-}(x) \leq 0
    \\
    & \Leftrightarrow\;
    \bigwedge_{j\in I_i} (x_{\rho_i} - x_j)^2 \leq 0
    \land
    \bigwedge_{j \notin I_i} 0 \leq 0
    \\
    & \Leftrightarrow\;
    \bigwedge_{j\in I_i} (x_{\rho_i} - x_j)^2 \leq 0
    \\
    & \Leftrightarrow\;
    \bigwedge_{j\in I_i} x_{\rho_i} = x_j 
  \end{align*}
  and hence overall
  \begin{align*}
    \bigwedge_{j=1}^n g_{i,j}^{+}(x) > 0 \land 
    \bigwedge_{j=1}^n g_{i,j}^{-}(x) \leq 0 
    \quad\Leftrightarrow\quad
    \bigwedge_{j \notin I_i} x_{\rho_i} > x_j
    \land 
    \bigwedge_{j\in I_i} x_{\rho_i} = x_j 
    \quad\Leftrightarrow\quad
    I_i = I(x).
  \end{align*}
  Therefore, 
  for every $i \in \iival{1}{k}$,
  the characterisations of the two sets in~\eqref{eq:prop:max-pap-1} are equivalent,
  and hence the two sets are equal.

  Next,
  we show that $\gamma$ is a representation of $f$; namely, we show that,
  for every $x \in \reals^n$,
  it holds that $f_i(x) = f$ where $i$ is the (unique) index such that $x \in A_i$.
  Let $x \in \reals^n$ and let $i$ be its partition index.
  We have that $f(x) = x_j$ for all $j \in I(x)$.
  We have shown above that $I_i = I(x)$.
  Then,
  \begin{align*}
    f_i(x) = \frac{1}{|I_i|} \sum_{j \in I_i} x_j = \frac{1}{|I_i|} \sum_{j \in I_i} f(x) = f(x).
  \end{align*}

  Finally, the intensional derivative $\dv \gamma$ is the function such that,
  for every $x \in \reals^n$ and for $i$ its partition index,
  satisfies $\dv\gamma(x)  = \dv f_i(x)$,
  and hence
  \begin{align*}
    \dv \gamma(x)  
    =
    \dv f_i(x)  
    =
    \begin{pmatrix}
      a_1(x) & \cdots & a_n(x)
    \end{pmatrix}
    \qquad a_j(x) =
    \begin{cases}
      1/|I_i| & \text{ if } j \in I_i = I(x),
      \\
      0 & \text{ otherwise}.
    \end{cases}
  \end{align*}
  Therefore, 
  the selected intensional derivative $\minmaxsdv f$
  is in fact the intensional derivative $\dv \gamma$.
\end{proof}

\begin{proposition} \label{prop:min-pap}
  The function $f(x) = \bigodot_{i=1}^n x_i$ is PAP, and 
  $\minmaxsdv f$ is an intensional derivative of $f$.
\end{proposition}
\begin{proof}
  Let $g(x) = \bigoplus_{i=1}^n x_i$.
  We have $f(x) = - g(-x)$, and hence $f$ is PAP since it is the composition of PAP functions.
  Then,
  by the chain rule for PAP functions (Proposition~\ref{prop:pap-intensional-chain-rule}), we have
  that 
  \begin{align*}
  -\minmaxsdv g(-x) \cdot \dv (-x) 
  \,=\, 
  -\minmaxsdv g(-x) \cdot (-\dv x) 
  \,=\, 
  \minmaxsdv g(-x)
  \end{align*}
  is an intensional derivative of $f$.
  Let $I_f$ and $I_g$ be the active variable index functions of $f$ and $g$, respectively,
  and let us note that 
  \begin{align*}
    I_f(x) 
    = 
    \Big\{ i \in \iival{1}{n} \mid \textstyle\bigodot_{i=1}^n x_i = x_i \Big\}
    =
    \Big\{ i \in \iival{1}{n} \mid \textstyle\bigoplus_{i=1}^n -x_i = -x_i \Big\}
    =
    I_g(-x) 
    \qquad \forall x \in \reals^n.
  \end{align*}
  Hence, we conclude that $\minmaxsdv g(-x) = \minmaxsdv f(x)$,
  and hence $\minmaxsdv f$ is an intensional derivative of $f$.
\end{proof}

\begin{proposition}[Intensional-admissibility of the AMM-PD-operator]
  \label{prop:minmax-pseudoderivative-intensionally-admissible}
  Let $G$ be a $\minmaxprimitivefuncclass$-computation graph.
  Then,
  the pseudo-derivative operator $\minmaxsdv$ is intensionally-admissible for $G$.
\end{proposition}
\begin{proof}
  By
  Definition~\ref{def:intensionally-admissible-pseudo-derivative},
  it suffices to show that
  $\minmaxsdv$
  is admissible for
  $G$
  and that,
  for every computation node
  $\cgnode{v}$
  of
  $G$,
  it holds that
  $\minmaxsdvop{\cglabel{v}}$
  is an intensional derivative of
  $\cglabel{v}$.

  Since every computation-node label of
  $G$
  belongs to
  $\minmaxprimitivefuncclass$,
  Definition~\ref{def:minmax-pseudoderivative-operator}
  implies immediately that
  $\minmaxsdv$
  is applicable to every computation-node label of
  $G$.
  Hence
  $\minmaxsdv$
  is admissible for
  $G$.

  Let
  $\cgnode{v}$
  be an arbitrary computation node of
  $G$.
  By the definition of
  $\minmaxprimitivefuncclass$,
  its label
  $\cglabel{v}$
  is one of the following:
  \begin{itemize}
    \item
      an analytic Euclidean function;
    \item
      a point-wise maximum function;
    \item
      a point-wise minimum function.
  \end{itemize}

  Suppose first that
  $\cglabel{v}$
  is analytic.
  Then
  $\cglabel{v}$
  is PAP,
  using the trivial PAP representation consisting of a single analytic piece,
  and its ordinary derivative
  $\dvop{\cglabel{v}}$
  is an intensional derivative of
  $\cglabel{v}$.
  By
  Definition~\ref{def:minmax-pseudoderivative-operator},
  \begin{equation*}
    \minmaxsdvop{\cglabel{v}}
    =
    \dvop{\cglabel{v}}.
  \end{equation*}
  Therefore,
  $\minmaxsdvop{\cglabel{v}}$
  is an intensional derivative of
  $\cglabel{v}$.
  This case includes zero-arity computation-node labels:
  by the convention for zero-arity functions, such a label is an analytic
  constant function
  \begin{equation*}
    \cglabel{v}
    :
    \reals^0
    \to
    \reals,
  \end{equation*}
  and
  \begin{equation*}
    \minmaxsdvop{\cglabel{v}}
    =
    \dvop{\cglabel{v}}
    =
    0_{1,0}.
  \end{equation*}

  Suppose next that
  $\cglabel{v}$
  is a point-wise maximum function.
  By
  Proposition~\ref{prop:max-pap},
  \begin{equation*}
    \minmaxsdvop{\cglabel{v}}
  \end{equation*}
  is an intensional derivative of
  $\cglabel{v}$.

  Finally,
  suppose that
  $\cglabel{v}$
  is a point-wise minimum function.
  By
  Proposition~\ref{prop:min-pap},
  \begin{equation*}
    \minmaxsdvop{\cglabel{v}}
  \end{equation*}
  is an intensional derivative of
  $\cglabel{v}$.

  Hence,
  for every computation-node label
  $\cglabel{v}$
  of
  $G$,
  the function
  $\minmaxsdvop{\cglabel{v}}$
  is an intensional derivative of
  $\cglabel{v}$.
  Therefore,
  $\minmaxsdv$
  is intensionally-admissible for
  $G$.
\end{proof}

\subsection{Analytic-MinMax Computation Graphs}
\label{subsec:minmax-algebraic-graphs}

\begin{closingdefinition}[MinMax-algebraic computation-graph extensions]
  \label{def:minmax-algebraic-extension}
  Let
  $m\in\posnaturals$,
  let
  $Q_1,\ldots,Q_m$
  be computation graphs with common Euclidean input space
  $\Lambda$,
  and let
  \begin{equation*}
    Q^\sqcup
    \defeq
    Q_1
    \sqcup_{\mathrm{in}}
    \cdots
    \sqcup_{\mathrm{in}}
    Q_m
  \end{equation*}
  be their common-input union as per
  Definition~\ref{def:computation-graph-common-input-union}.
  Let
  $V_\mathrm{src}$
  denote the set of scalar output nodes inherited from
  $Q_1,\ldots,Q_m$,
  called the \emph{source-output nodes}.

  A computation graph
  $G$
  is a \emph{MinMax-algebraic extension} of
  $\ptuple{Q_1,\ldots,Q_m}$
  if it is obtained from
  $Q^\sqcup$
  by adjoining fresh scalar computation nodes $W = \{ \cgnode{w}_1,\ldots,\cgnode{w}_N \}$ 
  while satisfying the following:
  \begin{itemize}
    \item
      for every $\ell\in\iival{1}{N}$:
      \begin{itemize}
        \item
          node $\cgnode{w}_\ell$ has label $\cglabel{w}_\ell \in \minmaxclassprimitive$;
        \item
          the predecessor node-vector $\cgnode{v}[\ell]$ of $\cgnode{w}_\ell$
          satisfies
          \begin{equation*}
            \entries{\cgnode{v}[\ell]}
            \subseteq
            V_\mathrm{src}
            \cup
            \{
              \cgnode{w}_1,\ldots,\cgnode{w}_{\ell-1}
            \};
          \end{equation*}
      \end{itemize}
    \item
      for some
      $d_\mathrm{out}\in\posnaturals$,
      the output node-vector of $G$ is 
      $\cgnode{v}^\mathrm{out}
      \in
      ( V_\mathrm{src} \cup W)^{d_\mathrm{out}}$.
  \end{itemize}

  We call
  $V_\mathrm{new} \defeq W$ the \emph{new nodes},
  and a node $\cgnode{w} \in W$ is a
  \emph{minimum node} or \emph{maximum node}
  if its label is a Min or Max reduction, respectively.

  We write $\minmaxextensionclass[Q_1,\ldots,Q_m]$ for the class of MinMax-algebraic extensions of
  $\ptuple{Q_1,\ldots,Q_m}$.
\end{closingdefinition}

\begin{closingdefinition}[MinMax-algebraic virtual computation-graph extensions]
  \label{def:minmax-algebraic-virtual-extension}
  Let
  $m\in\posnaturals$,
  let
  $Q_1,\ldots,Q_m$
  be computation graphs with common Euclidean input space
  $\Lambda$,
  and let
  \begin{equation*}
    Q^\sqcup
    \defeq
    Q_1
    \sqcup_{\mathrm{in}}
    \cdots
    \sqcup_{\mathrm{in}}
    Q_m
  \end{equation*}
  be their common-input union as per
  Definition~\ref{def:computation-graph-common-input-union}.
  Let
  $V_\mathrm{src}$
  denote the set of scalar output nodes inherited from
  $Q_1,\ldots,Q_m$,
  called the \emph{source-output nodes}.

  A virtual-computation graph
  $G$
  is a \emph{MinMax-algebraic virtual extension} of
  $\ptuple{Q_1,\ldots,Q_m}$
  if it is obtained from
  $Q^\sqcup$
  by adjoining fresh virtual computation nodes
  $W = \{ \cgnode{w}_1,\ldots,\cgnode{w}_N \}$
  in topological order
  while satisfying the following:
  \begin{itemize}
    \item
      for every $\ell\in\iival{1}{N}$:
      \begin{itemize}
        \item
          node $\cgnode{w}_\ell$ has label
          $\cglabel{w}_\ell \in \minmaxclassprimitive$;
        \item
          the predecessor node-vector $\cgnode{v}[\ell]$ of $\cgnode{w}_\ell$
          satisfies
          \begin{equation*}
            \entries{\cgnode{v}[\ell]}
            \subseteq
            V_\mathrm{src}
            \cup
            \{
              \cgnode{w}_1,\ldots,\cgnode{w}_{\ell-1}
            \};
          \end{equation*}
      \end{itemize}
    \item
      for some
      $d_\mathrm{out}\in\posnaturals$,
      the output node-vector of $G$ is
        $\cgnode{v}^\mathrm{out}
        \in
        \big(
          V_\mathrm{src}
          \cup
          W
        \big)^{d_\mathrm{out}}$.
  \end{itemize}

  We call
  $V_\mathrm{new}^{\mathrm{v}}
    \defeq
    W$
  the \emph{virtual new nodes},
  and a node
  $\cgnode{w}\in W$
  is a
  \emph{minimum node} or \emph{maximum node}
  if its label is a Min or Max reduction, respectively.

  We write
  $\minmaxextensionclass^\mathrm{v}[Q_1,\ldots,Q_m]$
  for the class of MinMax-algebraic virtual extensions of
  $\ptuple{Q_1,\ldots,Q_m}$.
\end{closingdefinition}

\begin{closingproposition}[Materialisation of MinMax-algebraic virtual extensions]
  \label{prop:minmax-virtual-extension-materialisation}
  Let $Q = \ptuple{Q_1,\ldots,Q_m}$ be computation graphs with the same input arity,
  let $G \in \minmaxextensionclass^\mathrm{v}[Q]$,
  and 
  let $\mu$ be a materialisation map for $G$,
  and
  let $G^\mu \defeq \matgraph{G}{\mu}$.
  Then
    $\matgraph{G}{\mu}
    \in
    \minmaxextensionclass[Q]$
    and
    $\cgval{G}^\mu
    =
    \cgval{G}$.
\end{closingproposition}
\begin{proof}
  Let
    $W
    =
    \{
      \cgnode{w}_1,\ldots,\cgnode{w}_N
    \}$
  be the virtual new nodes of
  $G$.
  Since materialisation is the identity on the concrete nodes of
  $G$,
  it leaves the common-input union of
  $Q_1,\ldots,Q_m$
  unchanged.
  By materialisation,
  \begin{equation*}
    W^\mu
    \defeq
    \{
      \cgnode{w}_1^\mu,\ldots,\cgnode{w}_N^\mu
    \},
    \qquad
    \cgnode{w}_\ell^\mu
    \defeq
    \mu(\cgnode{w}_\ell),
  \end{equation*}
  consists of fresh scalar computation nodes in the same topological order as
  $\cgnode{w}_1,\ldots,\cgnode{w}_N$,
  with
  \begin{equation*}
    \cglabel{w}_\ell^\mu
    =
    \cglabel{w}_\ell
    \in
    \minmaxclassprimitive
    \qquad
    \forall\ell\in\iival{1}{N}.
  \end{equation*}
  For every
  $\ell\in\iival{1}{N}$,
  let
  $\cgnode{v}[\ell]$
  be the predecessor node-vector of
  $\cgnode{w}_\ell$.
  By
  Definition~\ref{def:minmax-algebraic-virtual-extension},
  \begin{equation*}
    \entries{\cgnode{v}[\ell]}
    \subseteq
    V_\mathrm{src}
    \cup
    \{
      \cgnode{w}_1,\ldots,\cgnode{w}_{\ell-1}
    \}.
  \end{equation*}
  Since materialisation is the identity on
  $V_\mathrm{src}$
  and it maps predecessor node-vectors entry-wise,
  the predecessor node-vector of
  $\cgnode{w}_\ell^\mu$
  is
  $\bar\mu(\cgnode{v}[\ell])$,
  and
  \begin{equation*}
    \entries{\bar\mu(\cgnode{v}[\ell])}
    \subseteq
    V_\mathrm{src}
    \cup
    \{
      \cgnode{w}_1^\mu,\ldots,\cgnode{w}_{\ell-1}^\mu
    \}.
  \end{equation*}
  Similarly,
  the materialised output node-vector belongs to
  $(
    V_\mathrm{src}
    \cup
    W^\mu
  )^{d_\mathrm{out}}$.
  Hence
  $\matgraph{G}{\mu}$
  satisfies
  Definition~\ref{def:minmax-algebraic-extension},
  and therefore
  $\matgraph{G}{\mu} \in \minmaxextensionclass[Q]$.

  Finally,
  Proposition~\ref{prop:virtual-graph-materialisation-correctness}
  gives
  $\cgval{G}^\mu = \cgval{G}$.
\end{proof}

\begin{closingdefinition}[Analytic-MinMax computation graphs]
  \label{def:analytic-minmax-computation-graphs}
  An \emph{analytic-MinMax computation graph (AMM-graph)}
  is a
  $\minmaxprimitivefuncclass$-computation graph.
  We denote by $\minmaxgraphclass$ the class of all AMM-graphs.
  An \emph{AMM-graph class} is a subset of $\minmaxgraphclass$.

  \todo{}
  For every AMM-function $q$,
  we define
  \begin{equation*}
    \minmaxgraphclass[q]
    \defeq
    \big\{
      G
      \in
      \minmaxgraphclass
      \,\bigm|\,
      \cgval{G}
      =
      q
    \big\};
  \end{equation*}
  and we call $\minmaxgraphclass[q]$
  the \emph{AMM-graph family representing $q$}
  whenever $\minmaxgraphclass[q] \neq \emptyset$.
\end{closingdefinition}

\begin{closingproposition}[MinMax extensions of AMM-graphs]
  \label{proposition:minmax-extension-alc}
  Let $Q_1,\ldots,Q_m$ be AMM-graphs.
  Then $\minmaxextensionclass[ Q_1,\ldots,Q_m ]$ is an AMM-graph class.
\end{closingproposition}
\begin{proof}
  By
  Definition~\ref{def:analytic-minmax-primitive-functions},
  every constant function and every finite minimum or maximum reduction belongs
  to
  $\minmaxprimitivefuncclass$.
  Hence $\minmaxclassprimitive \subseteq \minmaxprimitivefuncclass$.
  The computation-node labels inherited from
  $Q_1,\ldots,Q_m$
  belong to
  $\minmaxprimitivefuncclass$
  because these graphs are AMM-graphs,
  while every newly introduced node has a label in
  $\minmaxclassprimitive$.
  Therefore every computation-node label of every
  $G \in \minmaxextensionclass[ Q_1,\ldots,Q_m ]$
  belongs to
  $\minmaxprimitivefuncclass$,
  so $G$ is an AMM-graph.
\end{proof}

\begin{corollary}
  Let $G \in \minmaxextensionclass[Q_1,\ldots,Q_m]$ be a MinMax-algebraic extension of AMM-graphs
  $Q_1,\ldots,Q_m$,
  and
  let $q$ be a Euclidean function.
  If
  $G$ represents $q$,
  then $G \in \minmaxgraphclass[q]$.
\end{corollary}
\begin{proof}
  Immediate consequence of Proposition~\ref{proposition:minmax-extension-alc}.
\end{proof}

\begin{proposition}[Characterisation of AMM-functions by AMM-graphs]
  \label{prop:analytic-minmax-functions-computation-graphs}
  For every function $q$,
  \begin{equation*}
    q
    \in
    \minmaxfuncclass
    \qquad\Longleftrightarrow\qquad
    \minmaxgraphclass[q]
    \neq
    \emptyset.
  \end{equation*}
\end{proposition}
\begin{proof}
  Suppose first that $q \in \minmaxfuncclass$.
  By
  Definition~\ref{def:analytic-minmax-functions},
  $q$
  is obtained by finitely many applications of function tupling and
  composition to functions in
  $\minmaxprimitivefuncclass$.
  These operations can be represented by a finite computation graph whose
  computation-node labels belong to
  $\minmaxprimitivefuncclass$.
  Hence there exists an AMM-graph
  $G$
  such that $\cgval{G} = q$.
  Therefore, $G \in \minmaxgraphclass[q]$,
  and hence $\minmaxgraphclass[q] \neq \emptyset$.

  Conversely,
  suppose that $G \in \minmaxgraphclass[q]$.
  By
  Definition~\ref{def:analytic-minmax-computation-graphs},
  every computation-node label of
  $G$
  belongs to
  $\minmaxprimitivefuncclass$.
  Since
  $G$
  is finite,
  its value function is obtained by finitely many applications of function
  tupling and composition to functions in
  $\minmaxprimitivefuncclass$.
  Hence,
  by
  Definition~\ref{def:analytic-minmax-functions},
  we have $q = \cgval{G} \in \minmaxfuncclass$.
\end{proof}

\subsection{Virtual Graph-Extension Notation for Analytic-Minax Graphs}
\label{subsec:virtual-graph-extension-notation-for-amm-graphs}

\begin{closingdefinition}[MinMax vector and matrix notation]
  \label{def:minmax-vector-presentation}
  Let
  \begin{equation*}
    G
    =
    \tuple{
      V,
      V_\mathrm{v},
      E,
      \cgnode{v}[][\mathrm{in}],
      \cgnode{v}[][\mathrm{c}],
      \cgnode{v}[][\mathrm{out}],
      \lambda
    }
  \end{equation*}
  be a virtual-computation graph.

  We introduce notation for compactly specifying virtual-node extensions of
  $G$
  using node-vector and node-matrix presentations as per
  Definition~\ref{def:virtual-node-presentations}.

  A
  \emph{presentation pattern}
  is, recursively,
  a scalar node,
  a node-vector presentation,
  a node-matrix presentation,
  or a finite tuple of presentation patterns.
  This is only syntactic notation for structured collections of scalar nodes
  and does not introduce an additional kind of graph object.

  We write
  \begin{equation*}
    \vgraphext
    {G}
    {
      \mathcal{E}_1,
      \ldots,
      \mathcal{E}_q
    }
    {G'}
  \end{equation*}
  when
  $G'$
  is the virtual-node extension of
  $G$
  obtained by successively applying the expressions
  \begin{equation*}
    \mathcal{E}_1,
    \ldots,
    \mathcal{E}_q
  \end{equation*}
  from left to right, according to the rules below.
  Thus, every expression may use nodes introduced by the preceding
  expressions in the same extension.

  Every scalar node introduced by this notation is a fresh virtual computation
  node, and all virtual nodes introduced in a single extension are pair-wise
  distinct and distinct from the nodes already present in
  $G$.

  The expressions are interpreted as follows.

  \begin{itemize}
    \item
      \emph{Scalar copy.}~
      For every node
      $\cgnode{u}$
      available in the current virtual-computation graph,
      an expression
      \begin{equation*}
        \cgnode{w}
        \defeq
        \cgnode{u}
      \end{equation*}
      denotes the virtual-node extension from
      Definition~\ref{def:virtual-node-extension}
      obtained by adjoining
      $\cgnode{w}$
      with predecessor node-vector
      $\ptuple{\cgnode{u}}$
      and label
      $\id[\reals]$.

    \item
      \emph{Vector and matrix copies.}~
      If
      $\cgnode{u}$
      and
      $\cgnode{w}$
      are node-vector presentations of the same dimension, then
      \begin{equation*}
        \cgnode{w}
        \defeq
        \cgnode{u}
      \end{equation*}
      is interpreted entrywise by the scalar-copy rule.
      Likewise, if
      $\cgnode{A}$
      and
      $\cgnode{B}$
      are node-matrix presentations of the same dimensions, then
      \begin{equation*}
        \cgnode{B}
        \defeq
        \cgnode{A}
      \end{equation*}
      is interpreted entrywise by the scalar-copy rule.

    \item
      \emph{Tuple copies.}~
      If
      \begin{equation*}
        \mathcal{P}
        =
        \ptuple{
          \mathcal{P}_1,
          \ldots,
          \mathcal{P}_r
        }
      \end{equation*}
      is a presentation pattern, then an expression
      \begin{equation*}
        \mathcal{Q}
        \defeq
        \mathcal{P},
      \end{equation*}
      where
      $\mathcal{Q}$
      has the same tuple, vector, and matrix structure as
      $\mathcal{P}$,
      is interpreted recursively by applying the scalar-, vector-, and
      matrix-copy rules to its components.
      In particular,
      \begin{equation*}
        \cgnode{M}
        \defeq
        \ptuple{
          \cgnode{A},
          \cgnode{b}
        }
      \end{equation*}
      introduces fresh virtual copies of all scalar entries of
      $\cgnode{A}$
      and
      $\cgnode{b}$,
      arranged with the same matrix--vector structure.

    \item
      \emph{Fresh vector and matrix presentations.}~
      An expression consisting only of a node-vector or node-matrix symbol,
      such as
      \begin{equation*}
        \cgnode{z}
        \qquad\text{or}\qquad
        \cgnode{A},
      \end{equation*}
      denotes a fresh virtual node-vector or node-matrix presentation,
      respectively,
      provided that the surrounding specification gives, for every scalar
      entry, its predecessor node-vector and scalar label.
      The corresponding scalar virtual nodes are adjoined successively
      according to
      Definition~\ref{def:virtual-node-extension}.

    \item
      \emph{Minimum and maximum reductions.}~
      Let
      \begin{equation*}
        \cgnode{u}
        =
        \ptuple{
          \cgnode{u}[1],
          \ldots,
          \cgnode{u}[r]
        }
      \end{equation*}
      be a node-vector presentation available in the current graph,
      with
      $r\geq2$.
      Then
      \begin{equation*}
        \cgnode{w}
        \defeq
        \minreductionsymb(\cgnode{u})
      \end{equation*}
      denotes the virtual-node extension obtained by adjoining
      $\cgnode{w}$
      with predecessor node-vector
      $\cgnode{u}$
      and label
      $\minreduction{r}$.
      Analogously,
      \begin{equation*}
        \cgnode{w}
        \defeq
        \maxreductionsymb(\cgnode{u})
      \end{equation*}
      denotes the extension obtained by adjoining
      $\cgnode{w}$
      with predecessor node-vector
      $\cgnode{u}$
      and label
      $\maxreduction{r}$.

    \item
      \emph{Component-wise maximum.}~
      Let
      \begin{equation*}
        \cgnode{x},
        \cgnode{y}
        \in
        (V\cup V_\mathrm{v})^n
      \end{equation*}
      be node-vector presentations available in the current graph.
      Then
      \begin{equation*}
        \cgnode{z}
        \defeq
        \cgnode{x}
        \oplus
        \cgnode{y}
      \end{equation*}
      denotes the successive introduction, for every
      $i\in\iival{1}{n}$,
      of a fresh virtual node
      $\cgnode{z}[i]$
      with predecessor node-vector
      \begin{equation*}
        \ptuple{
          \getentry{\cgnode{x}}{i},
          \getentry{\cgnode{y}}{i}
        }
      \end{equation*}
      and label
      $\maxreduction{2}$.

    \item
      \emph{MinMax matrix--vector product.}~
      Let
      \begin{equation*}
        \cgnode{A}
        \in
        (V\cup V_\mathrm{v})^{m\times n},
        \qquad
        \cgnode{x}
        \in
        (V\cup V_\mathrm{v})^n
      \end{equation*}
      be node-matrix and node-vector presentations available in the current
      graph.
      Then
      \begin{equation*}
        \cgnode{z}
        \defeq
        \cgnode{A}
        \otimes
        \cgnode{x}
      \end{equation*}
      denotes the following successive virtual-node extensions:
      \begin{itemize}
        \item
          for every
          $i\in\iival{1}{m}$
          and
          $j\in\iival{1}{n}$,
          introduce a fresh virtual node
          $\cgnode{w}[i,j]$
          with predecessor node-vector
          \begin{equation*}
            \ptuple{
              \getentry{\cgnode{A}}{i,j},
              \getentry{\cgnode{x}}{j}
            }
          \end{equation*}
          and label
          $\minreduction{2}$;

        \item
          for every
          $i\in\iival{1}{m}$,
          introduce a fresh virtual node
          $\cgnode{z}[i]$
          with predecessor node-vector
          \begin{equation*}
            \ptuple{
              \cgnode{w}[i,1],
              \ldots,
              \cgnode{w}[i,n]
            }
          \end{equation*}
          and label
          $\maxreduction{n}$.
        \end{itemize}

    \item
      \emph{MinMax matrix--matrix product.}~
      Let
      \begin{equation*}
        \cgnode{A}
        \in
        (V\cup V_\mathrm{v})^{m\times n},
        \qquad
        \cgnode{B}
        \in
        (V\cup V_\mathrm{v})^{n\times p}
      \end{equation*}
      be node-matrix presentations available in the current graph.
      Then
      \begin{equation*}
        \cgnode{C}
        \defeq
        \cgnode{A}
        \otimes
        \cgnode{B}
      \end{equation*}
      denotes the following successive virtual-node extensions:
      \begin{itemize}
        \item
          for every
          $i\in\iival{1}{m}$,
          $j\in\iival{1}{p}$,
          and
          $r\in\iival{1}{n}$,
          introduce a fresh virtual node
          $\cgnode{w}[i,j,r]$
          with predecessor node-vector
          \begin{equation*}
            \ptuple{
              \getentry{\cgnode{A}}{i,r},
              \getentry{\cgnode{B}}{r,j}
            }
          \end{equation*}
          and label
          $\minreduction{2}$;

        \item
          for every
          $i\in\iival{1}{m}$
          and
          $j\in\iival{1}{p}$,
          introduce a fresh virtual node
          $\cgnode{C}[i,j]$
          with predecessor node-vector
          \begin{equation*}
            \ptuple{
              \cgnode{w}[i,j,1],
              \ldots,
              \cgnode{w}[i,j,n]
            }
          \end{equation*}
          and label
          $\maxreduction{n}$.
        \end{itemize}
      \end{itemize}

  Whenever the right-hand side of an expression contains several MinMax
  operations, they are expanded recursively in the order determined by the
  expression, with every intermediate scalar result represented by a fresh
  virtual computation node.

  In particular, several expressions may be combined in a single extension,
  as in
  \begin{equation*}
    \vgraphext
    {G}
    {
      \cgnode{v}[\minreductionsymb]
      \defeq
      \minreductionsymb(\cgnode{\mathcal V}),
      \;
      \cgnode{v}[\maxreductionsymb]
      \defeq
      \maxreductionsymb(\cgnode{\mathcal V}),
      \;
      \cgnode{E},
      \;
      \cgnode{e},
      \;
      \cgnode{M}[\id]
      \defeq
      \ptuple{
        \cgnode{E},
        \cgnode{e}
      }
    }
    {G'},
  \end{equation*}
  where component-wise predecessor node-vectors and labels for
  $\cgnode{E}$
  and
  $\cgnode{e}$
  may be specified separately.
\end{closingdefinition}

\begin{closingfact}[Value functions of MinMax presentations]
  \label{fact:minmax-vector-presentation-values}
  Assume the notation of
  Definition~\ref{def:minmax-vector-presentation},
  and let
  \begin{equation*}
    \vgraphext
    {G}
    {
      \mathcal{E}_1,
      \ldots,
      \mathcal{E}_q
    }
    {G'}
  \end{equation*}
  be an extension specified by that notation.
  Then every scalar, node-vector, and node-matrix presentation introduced by
  an expression
  $\mathcal{E}_j$
  has the value function obtained by evaluating its right-hand side
  point-wise.

  In particular,
  for every
  $u\in\reals^{d_\mathrm{in}}$,
  the following hold whenever the corresponding expressions occur in the
  extension:
  \begin{align*}
    \cgnode{z}
    \defeq
    \cgnode{x}
    &\qquad\Longrightarrow\qquad
    \cgval{z}(u)
    =
    \cgval{x}(u),
    \\[.4em]
    \cgnode{z}
    \defeq
    \minreductionsymb(\cgnode{x})
    &\qquad\Longrightarrow\qquad
    \cgval{z}(u)
    =
    \minreduction{n}\bigl(\cgval{x}(u)\bigr),
    \\[.4em]
    \cgnode{z}
    \defeq
    \maxreductionsymb(\cgnode{x})
    &\qquad\Longrightarrow\qquad
    \cgval{z}(u)
    =
    \maxreduction{n}\bigl(\cgval{x}(u)\bigr),
    \\[.4em]
    \cgnode{z}
    \defeq
    \cgnode{x}
    \oplus
    \cgnode{y}
    &\qquad\Longrightarrow\qquad
    \cgval{z}(u)
    =
    \cgval{x}(u)
    \oplus
    \cgval{y}(u),
    \\[.4em]
    \cgnode{z}
    \defeq
    \cgnode{A}
    \otimes
    \cgnode{x}
    &\qquad\Longrightarrow\qquad
    \cgval{z}(u)
    =
    \cgval{A}(u)
    \otimes
    \cgval{x}(u),
    \\[.4em]
    \cgnode{C}
    \defeq
    \cgnode{A}
    \otimes
    \cgnode{B}
    &\qquad\Longrightarrow\qquad
    \cgval{C}(u)
    =
    \cgval{A}(u)
    \otimes
    \cgval{B}(u).
  \end{align*}
  Here, in the second and third lines,
  $\cgnode{x}$
  has dimension
  $n$.

  The first implication applies component-wise to node-vector and node-matrix
  copies.
  More generally,
  the value of every structured presentation introduced by the notation is
  obtained component-wise from the values of the presentations occurring on
  its right-hand side.

  If several expressions occur in the same extension, the implications above
  apply successively in their order of occurrence.
  In particular, later expressions may use presentations introduced by
  earlier expressions, and all corresponding value identities hold
  simultaneously in
  $G'$.
\end{closingfact}

\begin{closingremark}[MinMax notation and materialisation]
  \label{rem:minmax-vector-presentation-materialisation}
  The notation of
  Definition~\ref{def:minmax-vector-presentation}
  is shorthand for successive virtual-node extensions in the sense of
  Definition~\ref{def:virtual-node-extension}.
  In particular, an expression
  \begin{equation*}
    \vgraphext
    {G}
    {
      \mathcal{E}_1,
      \ldots,
      \mathcal{E}_q
    }
    {G'}
  \end{equation*}
  specifies the extended virtual-computation graph
  $G'$
  together with all scalar, node-vector, and node-matrix presentations
  introduced by the expressions
  $\mathcal{E}_1,\ldots,\mathcal{E}_q$.

  Moreover, suppose that, for computation graphs
  $Q_1,\ldots,Q_m$,
  \begin{equation*}
    G
    \in
    \minmaxextensionclass^\mathrm{v}
    [
      Q_1,\ldots,Q_m
    ].
  \end{equation*}
  Since every scalar virtual computation node introduced by the notation has
  a label admitted by
  Definition~\ref{def:minmax-algebraic-virtual-extension},
  and its predecessors belong to the source nodes or to previously introduced
  extension nodes,
  we also have
  \begin{equation*}
    G'
    \in
    \minmaxextensionclass^\mathrm{v}
    [
      Q_1,\ldots,Q_m
    ].
  \end{equation*}

  Consequently,
  by
  Proposition~\ref{prop:minmax-virtual-extension-materialisation},
  for every materialisation map
  $\mu$
  of
  $G'$,
  \begin{equation*}
    \matgraph{G'}{\mu}
    \in
    \minmaxextensionclass
    [
      Q_1,\ldots,Q_m
    ],
    \qquad
    \cgval{\matgraph{G'}{\mu}}
    =
    \cgval{G'}.
  \end{equation*}
  In particular, materialisation preserves component-wise the value functions
  of every node-vector and node-matrix presentation introduced by the
  notation.

  Finally, if
  $Q_1,\ldots,Q_m$
  are AMM-graphs, then
  Proposition~\ref{proposition:minmax-extension-alc}
  implies that
  $\matgraph{G'}{\mu}$
  is an AMM-graph;
  consequently, by
  Proposition~\ref{prop:analytic-minmax-functions-computation-graphs},
  its value function is an AMM-function.
\end{closingremark}

\subsection{Properties of the Analytic-MinMax Pseudo-Derivative Operator}

\begin{closingdefinition}[AMM computation-graph family]
  \label{def:analytic-minmax-computation-graph-family}
  \todo{}
  Let
  $\mathcal{Q}\subseteq\minmaxfuncclass$.
  We define the
  \emph{AMM computation-graph family for $\mathcal{Q}$}
  by
  \begin{equation*}
    \minmaxgraphclass[\mathcal{Q}]
    \defeq
    \big(
      \minmaxgraphclass[q]
    \big)_{q\in\mathcal{Q}}.
  \end{equation*}
\end{closingdefinition}

\begin{closingfact}
  \label{fact:analytic-minmax-computation-graph-family}
  For every
  $\mathcal{Q}\subseteq\minmaxfuncclass$,
  the family
  $\minmaxgraphclass[\mathcal{Q}]$
  is a
  $\minmaxprimitivefuncclass$-computation-graph family for $\mathcal{Q}$.
\end{closingfact}
\begin{proof}
  Let
  $q\in\mathcal{Q}$.
  Proposition~\ref{prop:analytic-minmax-functions-computation-graphs}
  ensures $\minmaxgraphclass[q] \neq \emptyset$.
  Furthermore,
  by definition,
  every
  graph $G\in\minmaxgraphclass[q]$ is a $\minmaxprimitivefuncclass$-computation graph
  representing $q$.
  Hence
  $\minmaxgraphclass[\mathcal{Q}]$
  is a
  $\minmaxprimitivefuncclass$-computation-graph family for $\mathcal{Q}$.
\end{proof}

\begin{theorem}[SID-operators for AMM-functions]
  \label{lemma:minmax-selected-pseudoderivatives-are-sids}
  \label{lemma:analytic-minmax-selected-pseudoderivatives-are-sids}
  \label{theorem:minmax-selected-pseudoderivatives-are-sids}
  \label{theorem:analytic-minmax-selected-pseudoderivatives-are-sids}
  Let
  $\mathcal{Q}\subseteq\minmaxfuncclass$,
  let
  $\minmaxsdv$
  be the analytic-MinMax pseudo-derivative operator,\footnote{The operator is introduced in
  Definition~\ref{def:minmax-pseudoderivative-operator}.}
  and let
  $\selgraphsymb$
  be a graph selector for
  $\mathcal{Q}$
  over
  $\minmaxgraphclass[\mathcal{Q}]$.
  Then
  the graph-induced pseudo-derivative operator
  $\minmaxsdv[\selgraphsymb]$
  is a selected intensional-derivative operator on $\mathcal{Q}$.
\end{theorem}
\begin{proof}
  Since
  $\selgraphsymb$
  is a graph selector for
  $\mathcal{Q}$
  over
  $\minmaxgraphclass[\mathcal{Q}]$,
  we have
  \begin{equation*}
    \selgraphsymb(q)
    \in
    \minmaxgraphclass[q]
    \subseteq
    \minmaxgraphclass
    \qquad
    \forall q\in\mathcal{Q}.
  \end{equation*}
  Hence,
  by
  Definition~\ref{def:analytic-minmax-computation-graphs},
  every computation-node label of every graph in
  $\im(\selgraphsymb)$
  belongs to
  $\minmaxprimitivefuncclass$.
  Proposition~\ref{prop:minmax-pseudoderivative-intensionally-admissible}
  therefore implies that
  $\minmaxsdv$
  is intensionally-admissible for every graph in
  $\im(\selgraphsymb)$,
  and hence for
  $\im(\selgraphsymb)$.
  The claim follows from
  Proposition~\ref{prop:intensional-admissibility-implies-sid}.
\end{proof}

\begin{lemma}[Main convexity lemma]
  \label{lem:minmax-algebraic-source-convexity}
  Let
  $\minmaxsdv$
  be the pseudo-derivative operator of Definition~\ref{def:minmax-pseudoderivative-operator},
  and let
  \begin{equation*}
    G
    \in
    \minmaxextensionclass[Q_1,\ldots,Q_m]
  \end{equation*}
  be a MinMax-algebraic extension with source-output node set
  $V_\mathrm{src}$
  and new-node set
  $V_\mathrm{new}$.
  For every $i \in \iival{1}{m}$,
  let $d_i$ be the output arity of $Q_i$.
  and 
  let $Q_{i,j} \defeq \graphindexrestr{Q_i}{j}$ be the output-restriction of $Q_i$ to coordinate
  $j \in \iival{1}{d_i}$.

  Then,
  for every
  $\cgnode{v} \in V_\mathrm{src} \cup V_\mathrm{new}$
  and 
  every $\lambda\in\Lambda$,
  there exist non-negative coefficients
  \begin{equation*}
    \alpha_{i,j}^{\cgnode{v}}(\lambda)
    \in
    \nonnegreals
    \qquad\;
    \text{ for }\, i\in\ival{m}, j\in\ival{d_i},
  \end{equation*}
  such that
  \begin{align}
    \label{eq:minmax-algebraic-convex-coefficient-and-derivative-identity}
    \sum_{i=1}^m
    \sum_{j=1}^{d_i}
    \alpha_{i,j}^{\cgnode{v}}(\lambda)
    =
    1
    \qquad\text{and}\qquad
    \minmaxsdvop{\cgnode{v}}(\lambda)
    =
    \sum_{i=1}^m
    \sum_{j=1}^{d_i}
    \alpha_{i,j}^{\cgnode{v}}(\lambda)
    \cdot
    \minmaxsdv{Q_{i,j}}(\lambda)\,.
  \end{align}
  Consequently,
  \begin{align}
    \label{eq:minmax-algebraic-row-bound}
    \big\|
      \minmaxsdvop{G}(\lambda)
    \big\|_\mathrm{row}
    \leq
    \max_{i\in\iival{1}{m}}
    \big\|
      \minmaxsdvop{Q_i}(\lambda)
    \big\|_\mathrm{row}.
  \end{align}
\end{lemma}
\begin{proof}
  We first consider the source-output nodes.
  Let
  $\cgnode{v} \in V_\mathrm{src}$
  be the $j$-th scalar output node inherited from $Q_i$,
  and define
  \begin{equation*}
    \alpha_{i',j'}^{\cgnode{v}}(\lambda)
    \defeq
    \begin{cases}
      1
      & \text{ if }
      (i',j')=(i,j),
      \\
      0
      & \text{ if }
      (i',j')\neq(i,j).
    \end{cases}
  \end{equation*}
  Then
  \begin{equation*}
    \sum_{i'=1}^m
    \sum_{j'=1}^{d_{i'}}
    \alpha_{i',j'}^{\cgnode{v}}(\lambda)
    =
    1,
  \end{equation*}
  and, since
  $\cgnode{v}$
  is the corresponding scalar output node of
  $Q_i$,
  \begin{equation*}
    \minmaxsdvop{\cgnode{v}}(\lambda)
    =
    \minmaxsdvop{Q_{i,j}}(\lambda).
  \end{equation*}
  Hence
  \eqref{eq:minmax-algebraic-convex-coefficient-and-derivative-identity}
  holds for every source-output node.

  We now proceed by induction over the topological ordering over the new nodes
  \begin{equation*}
    \cgnode{w}_1,
    \ldots,
    \cgnode{w}_N
    \in
    V_\mathrm{new}.
  \end{equation*}
  Let
  $\ell\in\iival{1}{N}$,
  and suppose that the claim holds for every
  $\cgnode{w}_r$
  with
  $r\in\iival{1}{\ell-1}$.
  Let
  $\cgnode{v}\defeq\cgnode{w}_\ell$
  have arity $k$ and ordered predecessor tuple
  \begin{equation*}
    \ptuple{
      \cgnode{v}_1,
      \ldots,
      \cgnode{v}_k
    }.
  \end{equation*}
  By
  Definition~\ref{def:minmax-algebraic-extension},
  every predecessor satisfies
  \begin{equation*}
    \cgnode{v}_r
    \in
    V_\mathrm{src}
    \cup
    \{
      \cgnode{w}_1,
      \ldots,
      \cgnode{w}_{\ell-1}
    \},
    \qquad
    r\in\iival{1}{k}.
  \end{equation*}
  Hence, by the source-output case or the induction hypothesis,
  for every
  $r\in\iival{1}{k}$
  there exist coefficients
  \begin{equation*}
    \alpha_{i,j}^{\cgnode{v}[r]}(\lambda)
    \in
    \nonnegreals
  \end{equation*}
  satisfying
  \begin{gather}
    \label{eq:minmax-predecessor-convex-coefficients}
    \sum_{i=1}^m
    \sum_{j=1}^{d_i}
    \alpha_{i,j}^{\cgnode{v}[r]}(\lambda)
    =
    1
  \end{gather}
  and
  \begin{gather}
    \label{eq:minmax-predecessor-convex-derivative}
    \minmaxsdvop{\cgnode{v}[r]}(\lambda)
    =
    \sum_{i=1}^m
    \sum_{j=1}^{d_i}
    \alpha_{i,j}^{\cgnode{v}[r]}(\lambda)
    \cdot
    \minmaxsdvop{Q_{i,j}}(\lambda).
  \end{gather}

  Since
  $\cgnode{v}$
  is either a minimum node or a maximum node,
  Definition~\ref{def:minmax-pseudoderivative-operator}
  and
  Propositions~\ref{prop:max-subgradient-hull}
  and~\ref{prop:min-subgradient-hull}
  imply that the local pseudo-derivative of its label is a convex
  combination of the coordinate derivatives.
  Hence there exist
  \begin{equation*}
    \beta_1(\lambda),
    \ldots,
    \beta_k(\lambda)
    \in
    \nonnegreals
  \end{equation*}
  such that
  \begin{align}
    \label{eq:minmax-local-convex-coefficients-sum}
    \sum_{r=1}^k
    \beta_r(\lambda)
    &=
    1.
  \end{align}
  Therefore, by the definition of graph-induced pseudo-derivatives in
  Definition~\ref{def:graph-induced-pseudo-derivative},
  \begin{align}
    \label{eq:minmax-local-convex-derivative}
    \minmaxsdvop{\cgnode{v}}(\lambda)
    &=
    \sum_{r=1}^k
    \beta_r(\lambda)
    \cdot
    \minmaxsdvop{\cgnode{v}[r]}(\lambda).
  \end{align}

  Substituting
  \eqref{eq:minmax-predecessor-convex-derivative}
  into
  \eqref{eq:minmax-local-convex-derivative}
  gives
  \begin{align*}
    \minmaxsdvop{\cgnode{v}}(\lambda)
    &=
    \sum_{r=1}^k
    \beta_r(\lambda)
    \cdot
    \sum_{i=1}^m
    \sum_{j=1}^{d_i}
    \alpha_{i,j}^{\cgnode{v}[r]}(\lambda)
    \cdot
    \minmaxsdvop{Q_{i,j}}(\lambda)
    \\[.4em]
    &=
    \sum_{i=1}^m
    \sum_{j=1}^{d_i}
    \left(
      \sum_{r=1}^k
      \beta_r(\lambda)
      \cdot
      \alpha_{i,j}^{\cgnode{v}[r]}(\lambda)
    \right)
    \cdot
    \minmaxsdvop{Q_{i,j}}(\lambda).
  \end{align*}
  Define
  \begin{equation*}
    \alpha_{i,j}^{\cgnode{v}}(\lambda)
    \defeq
    \sum_{r=1}^k
    \beta_r(\lambda)
    \cdot
    \alpha_{i,j}^{\cgnode{v}[r]}(\lambda)
    \qquad
    \forall i\in\iival{1}{m},\;
    \forall j\in\iival{1}{d_i}.
  \end{equation*}
  These coefficients are non-negative, and
  by
  \eqref{eq:minmax-predecessor-convex-coefficients}
  and
  \eqref{eq:minmax-local-convex-coefficients-sum},
  \begin{align*}
    \sum_{i=1}^m
    \sum_{j=1}^{d_i}
    \alpha_{i,j}^{\cgnode{v}}(\lambda)
    \;=\;
    \sum_{r=1}^k
    \beta_r(\lambda)
    \cdot
    \sum_{i=1}^m
    \sum_{j=1}^{d_i}
    \alpha_{i,j}^{\cgnode{v}[r]}(\lambda)
    \;=\;
    \sum_{r=1}^k
    \beta_r(\lambda)
    \;=\;
    1.
  \end{align*}
  Thus
  Eq.~\eqref{eq:minmax-algebraic-convex-coefficient-and-derivative-identity}
  holds for
  $\cgnode{w}_\ell$.
  Hence, by induction over
  $\ell\in\iival{1}{N}$,
  it holds for every node in
  $V_\mathrm{new}$.
  Together with the source-output case proved above,
  it therefore holds for every node in
  $V_\mathrm{src}\cup V_\mathrm{new}$.

  Finally, let
  $\cgnode{v}$
  be a scalar output node of
  $G$.
  By
  Definition~\ref{def:minmax-algebraic-extension},
  we have
  $\cgnode{v}
  \in
  V_\mathrm{src}
  \cup
  V_\mathrm{new}$.
  Hence,
  by
  Eq.~\eqref{eq:minmax-algebraic-convex-coefficient-and-derivative-identity}
  and
  the triangle inequality,
  we obtain
  \begin{align*}
    \big\|
    \minmaxsdvop{\cgnode{v}}(\lambda)
    \big\|
    \;\leq\;
    \sum_{i=1}^m
    \sum_{j=1}^{d_i}
    \alpha_{i,j}^{\cgnode{v}}(\lambda)
    \big\|
    \minmaxsdvop{Q_{i,j}}(\lambda)
    \big\|
    \;\leq\;
    \max_{\substack{
      i\in\iival{1}{m}\\
      j\in\iival{1}{d_i}
    }}
    \big\|
    \minmaxsdvop{Q_{i,j}}(\lambda)
    \big\|.
  \end{align*}
  Taking the maximum over the output nodes of $G$ gives
  \begin{align*}
    \big\|
      \minmaxsdvop{G}(\lambda)
    \big\|_\mathrm{row}
    \;\leq\;
    \max_{\substack{
      i\in\iival{1}{m}\\
      j\in\iival{1}{d_i}
    }}
    \big\|
    \minmaxsdvop{Q_{i,j}}(\lambda)
    \big\|
    \;=\;
    \max_{i\in\iival{1}{m}}
    \big\|
      \minmaxsdvop{Q_i}(\lambda)
    \big\|_\mathrm{row},
  \end{align*}
  proving
  \eqref{eq:minmax-algebraic-row-bound}.
\end{proof}

\subsection{MinMax Recurrence Computation Graphs}

\begin{closingdefinition}[MinMax-algebraic computation graphs for a MinMax recurrence]
  \label{def:minmax-algebraic-recurrence-graphs}
  Let
  $x_0\in\reals^n$,
  and
  let $T\in\posnaturals$.
  For every
  $t\in\iival{1}{T}$,
  let us consider AMM-functions $R_t, s_t \in \minmaxfuncclass$ with
  \begin{equation*}
    R_t
    :
    \Lambda
    \to
    \reals^{n\times n},
    \qquad
    s_t
    :
    \Lambda
    \to
    \reals^n,
  \end{equation*}
  and corresponding AMM-graphs 
  $\cgnode{R}_t \in \minmaxgraphclass[R_t]$ and $\cgnode{s}_t \in \minmaxgraphclass[s_t]$.

  Let $x_0 : \Lambda \to \reals^n$ 
  be the constant function $x_0(\lambda) \defeq x_0$
  with canonical graph $\cgnode{x}_0 \defeq \constpargraph{x_0}{\Lambda}$,
  and 
  let $(x_t)_{t=1}^T$ be the sequence given recursively by
  \begin{equation*}
    x_t(\lambda)
    \defeq
    \big(
      R_t(\lambda)
      \otimes
      x_{t-1}(\lambda)
    \big)
    \oplus
    s_t(\lambda)
    \qquad
    \forall t \in \iival{1}{T}.
  \end{equation*}

  The corresponding
  \emph{MinMax-algebraic recurrence computation-graph family}
  is
  \begin{equation*}
    \minmaxrecgraphclass
    \big(
      x_0;
      \seq{\cgnode{R}}{1}{T},
      \seq{\cgnode{s}}{1}{T}
    \big)
    \defeq
    \big\{
      G
      \in
      \minmaxextensionclass\big[\cgnode{x}_0, \seq{\cgnode{R}}{1}{T}, \seq{\cgnode{s}}{1}{T}\big]
      \bigm|
      \cgval{G}
      =
      x_T^u
    \big\}.
  \end{equation*}
\end{closingdefinition}

\begin{closingremark}
  The qualification
  \emph{MinMax-algebraic}
  of the recurrence graphs in
  Definition~\ref{def:minmax-algebraic-recurrence-graphs}
  is meant to emphasise that they are obtained through
  MinMax-algebraic extensions.
  In particular, we do not introduce a separate notion of
  \emph{MinMax-algebraic computation graph}
  or
  \emph{MinMax-algebraic computation-graph class}.

  Instead, we show below in
  Proposition~\ref{prop:minmax-algebraic-recurrence-graphs-al}
  that, whenever the source graphs are Analytic-MinMax,
  the resulting MinMax-algebraic recurrence graphs are themselves
  Analytic-MinMax computation graphs.
  In particular, this applies to the recurrence graphs arising in MinMax RNCs,
  where the functions used outside the MinMax recurrence are analytic.
\end{closingremark}

\begin{proposition}[Analytic-MinMax closure of MinMax-algebraic recurrence graphs]
  \label{prop:minmax-algebraic-recurrence-graphs-al}
  Assume the context of
  Definition~\ref{def:minmax-algebraic-recurrence-graphs}.
  Then
  \begin{equation*}
    \minmaxrecgraphclass
    \big(
      x_0;
      \seq{\cgnode{R}}{1}{T},
      \seq{\cgnode{s}}{1}{T}
    \big)
    \subseteq
    \minmaxgraphclass[x_T].
  \end{equation*}
\end{proposition}
\begin{proof}
  Let
  \begin{equation*}
    G
    \in
    \minmaxrecgraphclass
    \big(
      x_0;
      \seq{\cgnode{R}}{1}{T},
      \seq{\cgnode{s}}{1}{T}
    \big).
  \end{equation*}
  By
  Definition~\ref{def:minmax-algebraic-recurrence-graphs},
  we have
  \begin{equation*}
    G
    \in
    \minmaxextensionclass
    \big[
      \cgnode{x}_0,
      \seq{\cgnode{R}}{1}{T},
      \seq{\cgnode{s}}{1}{T}
    \big]
    \qquad\text{and}\qquad
    \cgval{G} = x_T.
  \end{equation*}
  For every
  $t\in\iival{1}{T}$,
  we have
  $\cgnode{R}_t
  \in
  \minmaxgraphclass[R_t]$
  and
  $\cgnode{s}_t
  \in
  \minmaxgraphclass[s_t]$,
  and hence
  $\cgnode{R}_t$
  and
  $\cgnode{s}_t$
  are AMM-graphs.
  Furthermore,
  $\cgnode{x}_0$
  is a constant computation graph,
  whose computation-node labels are constant and therefore analytic functions;
  hence
  $\cgnode{x}_0$
  is also an AMM-graph.

  Therefore,
  by
  Proposition~\ref{proposition:minmax-extension-alc},
  $G$
  is an AMM-graph.
  Since
  $\cgval{G}=x_T$,
  we conclude that
  $G \in \minmaxgraphclass[x_T]$.
  Since
  $G$
  was arbitrary,
  the claimed inclusion follows.
\end{proof}

\begin{theorem}[Topology-independent derivative bound for MinMax recurrence evaluation]
  \label{theorem:minmax-algebraic-recurrence-graph-bound}
  Assume the context of
  Definition~\ref{def:minmax-algebraic-recurrence-graphs},
  and let
  \begin{equation*}
    G
    \in
    \minmaxrecgraphclass
    \big(
      x_0;
      \seq{\cgnode{R}}{1}{T},
      \seq{\cgnode{s}}{1}{T}
    \big).
  \end{equation*}
  Then,
  for every
  $\lambda\in\Lambda$,
  \begin{align*}
    \big\|
      \minmaxsdvop{G}(\lambda)
    \big\|_\mathrm{row}
    &\leq
    \max_{t\in\iival{1}{T}}
    \max
    \left\{
      \big\|
        \minmaxsdvop{\cgnode{R}_t}(\lambda)
      \big\|_\mathrm{row},
      \big\|
        \minmaxsdvop{\cgnode{s}_t}(\lambda)
      \big\|_\mathrm{row}
    \right\}.
  \end{align*}
\end{theorem}
\begin{proof}
  By
  Definition~\ref{def:minmax-algebraic-recurrence-graphs},
  \begin{equation*}
    G
    \in
    \minmaxextensionclass
    \big[
      \cgnode{x}_0,
      \seq{\cgnode{R}}{1}{T},
      \seq{\cgnode{s}}{1}{T}
    \big].
  \end{equation*}
  Hence,
  Lemma~\ref{lem:minmax-algebraic-source-convexity}
  gives
  \begin{align*}
    \big\|
      \minmaxsdvop{G}(\lambda)
    \big\|_\mathrm{row}
    \leq
    \max
    \Bigg\{
      \big\|
        \minmaxsdvop{\cgnode{x}_0}(\lambda)
      \big\|_\mathrm{row},\;
      \max_{t\in\iival{1}{T}}
      \big\|
        \minmaxsdvop{\cgnode{R}_t}(\lambda)
      \big\|_\mathrm{row},\;
      \max_{t\in\iival{1}{T}}
      \big\|
        \minmaxsdvop{\cgnode{s}_t}(\lambda)
      \big\|_\mathrm{row}
    \Bigg\}.
  \end{align*}

  Since
  $\cgnode{x}_0$
  represents a constant function of
  $\lambda$,
  Lemma~\ref{lem:parameter-independent-zero-derivative}
  gives
  \begin{equation*}
    \minmaxsdvop{\cgnode{x}_0}(\lambda)
    =
    0.
  \end{equation*}
  Therefore,
  \begin{align*}
    \big\|
      \minmaxsdvop{G}(\lambda)
    \big\|_\mathrm{row}
    & \;\leq\;
    \max
    \left\{
      \max_{t\in\iival{1}{T}}
      \big\|
        \minmaxsdvop{\cgnode{R}_t}(\lambda)
      \big\|_\mathrm{row},
      \;
      \max_{t\in\iival{1}{T}}
      \big\|
        \minmaxsdvop{\cgnode{s}_t}(\lambda)
      \big\|_\mathrm{row}
    \right\}
    \\[.3em]
    &\;=\;
    \max_{t\in\iival{1}{T}}
    \max
    \left\{
      \big\|
        \minmaxsdvop{\cgnode{R}_t}(\lambda)
      \big\|_\mathrm{row},
      \big\|
        \minmaxsdvop{\cgnode{s}_t}(\lambda)
      \big\|_\mathrm{row}
    \right\},
  \end{align*}
  proving the claim.
\end{proof}

\subsection{MinMaxRNC State Computation Graphs}

\begin{closingdefinition}[Identity parameter computation graph]
  \label{def:identity-computation-graph}
  Let
  $d_\Theta \in \naturals$,
  and 
  let
  \begin{equation*}
    \cgnode{\theta}
    =
    \ptuple{
      \cgnode{\theta}_1,
      \ldots,
      \cgnode{\theta}_{d_\Theta}
    }
  \end{equation*}
  be a node-vector.

  The
  \emph{identity parameter computation graph}
  $\mathring{I}[\cgnode{\theta}]$
  is the computation graph having no computation nodes and satisfying
  \begin{equation*}
    \cgnode{v}^{\mathrm{in}}
    =
    \cgnode{v}^{\mathrm{out}}
    =
    \cgnode{\theta}.
  \end{equation*}
\end{closingdefinition}

\begin{closingfact}
  In the context of Definition~\ref{def:identity-computation-graph},
  we have that
  $\mathring{I}[\cgnode{\theta}]$
  represents the identity function on $\reals^{d_\Theta}$.
\end{closingfact}

\begin{closingdefinition}[MinMax-algebraic RNC state computation-graph class]
  \label{def:minmax-algebraic-rnc-state-graphs}
  We define MinMax-algebraic computation graphs for the states of MinMax RNCs.

  \par\smallskip
  \textbf{Context.}~
  Let us consider
  RNC dynamics 
  $D \in \rncdynop\big(\classparneuralunits{MinMax}\big)$
  having
  state space $X$,
  input space $U$,
  parameter space $\Theta\subseteq\reals^{d_\Theta}$,
  and 
  cascade decomposition
  \begin{equation*}
    D
    =
    D_1
    \ltimes_{g_1}
    \cdots
    \ltimes_{g_{L-1}}
    D_L,
    \qquad
    D_\ell
    =
    D_{\ell,1}
    \parallelop
    \cdots
    \parallelop
    D_{\ell,n_\ell}.
  \end{equation*}

  Let
  $u\in\infseqs{U}$,
  $x_0\in X$,
  and
  $T\in\timedomain$.

  \par\smallskip
  \textbf{Auxiliary.}~
  Let
  \begin{equation*}
    \cgnode{\theta}
    =
    \ptuple{
      \cgnode{\theta}_1,
      \ldots,
      \cgnode{\theta}_{d_\Theta}
    }
  \end{equation*}
  be a node-vector.
  All computation graphs introduced below have $\cgnode{\theta}$
  as their input node-vector.
  In particular, all source graphs combined by common-input unions or
  smooth graph extensions have this same input node-vector.
  The smooth extensions below use no additional parameter input,
  since the parameter value is supplied explicitly by
  $\mathring{I}[\cgnode{\theta}]$.

  For every
  $\ell\in\iival{1}{L}$,
  let
  $X_\ell$
  and
  $U_\ell$
  denote the state and input spaces of
  $D_\ell$, 
  with $U_1 = U$,
  and, for every
  $k\in\iival{1}{n_\ell}$,
  let
  $X_{\ell,k}$
  be the state space of
  $D_{\ell,k}$.
  We identify
  \begin{equation*}
    X_\ell
    =
    X_{\ell,1}
    \times
    \cdots
    \times
    X_{\ell,n_\ell},
    \qquad
    X
    =
    X_1
    \times
    \cdots
    \times
    X_L.
  \end{equation*}

  For every
  $\ell\in\iival{1}{L}$
  and
  $k\in\iival{1}{n_\ell}$,
  \begin{itemize}
    \item
      let $\rho_{\ell,k}$ be the function 
      \begin{equation*}
        \rho_{\ell,k}
        :
        U_\ell
        \times
        \Theta
        \to
        \reals^{d_R^{\ell,k}+d_s^{\ell,k}},
        \qquad\;
        \rho_{\ell,k}
        \big(
          u,
          \theta
        \big)
        \defeq
        \ptuple{
        R_{\ell,k}
        \big(
          u;
          \theta
        \big),
        s_{\ell,k}
        \big(
          u;
          \theta
        \big)
        },
      \end{equation*}
      where
      $d_R^{\ell,k}$
      and
      $d_s^{\ell,k}$
      are the output arities of
      $R_{\ell,k}$
      and
      $s_{\ell,k}$,
      respectively;
    \item
      let $J_R^{\ell,k}$ and $J_s^{\ell,k}$ be the sets of indices
      \begin{align*}
        J_R^{\ell,k}
        \defeq
        \ptuple{
          1,
          \ldots,
          d_R^{\ell,k}
        },
        \qquad\;
        J_s^{\ell,k}
        \defeq
        \ptuple{
          d_R^{\ell,k}+1,
          \ldots,
          d_R^{\ell,k}+d_s^{\ell,k}
        }.
      \end{align*}
  \end{itemize}

  Let us write
  \begin{equation*}
    x_0
    =
    \ptuple{
      x_0^1,
      \ldots,
      x_0^L
    },
    \qquad\quad
    x_0^\ell
    =
    \ptuple{
      x_0^{\ell,1},
      \ldots,
      x_0^{\ell,n_\ell}
    }.
  \end{equation*}

  \par\smallskip
  \textbf{Definition.}~
  For
  $T=0$,
  we define
  \begin{equation*}
    \minmaxstategraphclass{D}[0][u][x_0]
    \defeq
    \big\{
      \constpargraph{x_0}{\cgnode{\theta}}
    \big\}.
  \end{equation*}

  For
  $T\in\posnaturals$,
  the
  \emph{MinMax-algebraic RNC state computation-graph class}
  $\minmaxstategraphclass{D}[T][u][x_0]$
  is the class of computation graphs
  \begin{equation*}
    G
    =
    \cgnode{X}_T^1
    \sqcup_{\mathrm{in}}
    \cdots
    \sqcup_{\mathrm{in}}
    \cgnode{X}_T^L
  \end{equation*}
  for which there exist computation graphs
  \begin{equation*}
    \cgnode{U}_t^\ell,
    \quad
    \cgnode{R}_t^{\ell,k},
    \quad
    \cgnode{s}_t^{\ell,k},
    \quad
    \cgnode{H}_t^{\ell,k},
    \quad
    \cgnode{X}_t^{\ell,k},
    \quad
    \cgnode{X}_t^\ell,
  \end{equation*}
  called \emph{constituent graphs} of $G$,
  such that:
  \begin{align*}
    \cgnode{X}_0^\ell & = \constpargraph{x_0^\ell}{\cgnode{\theta}}
    && 
    \forall \ell\in\iival{1}{L},
    \\[.3em]
    \cgnode{U}_t^1 & = \constpargraph{u_t}{\cgnode{\theta}}
    && 
    \forall t\in\iival{1}{T},
    \\[.3em]
    \cgnode{H}_t^{\ell,k}
    & =
    \ptuple{
      \cgnode{U}_t^\ell,
      \mathring{I}[\cgnode{\theta}]
    }
    \triangleright
    \rho_{\ell,k}
    && 
    \forall \ell\in\iival{1}{L},\,
    \forall k\in\iival{1}{n_\ell},\,
    \forall t\in\iival{1}{T},
    \\[.3em]
    \cgnode{R}_t^{\ell,k}
    & =
    \big(
    \cgnode{H}_t^{\ell,k}
    \big)_{J_R^{\ell,k}},
    &&
    \forall \ell\in\iival{1}{L},\,
    \forall k\in\iival{1}{n_\ell},\,
    \forall t\in\iival{1}{T},
    \\[.3em]
    \cgnode{s}_t^{\ell,k}
    & =
    \big(
    \cgnode{H}_t^{\ell,k}
    \big)_{J_s^{\ell,k}}
    &&
    \forall \ell\in\iival{1}{L},\,
    \forall k\in\iival{1}{n_\ell},\,
    \forall t\in\iival{1}{T},
    \\[.3em]
    \cgnode{X}_t^{\ell,k}
    & \in
    \minmaxrecgraphclass
    \big(
    x_0^{\ell,k};
    \seq{\cgnode{R}^{\ell,k}}{1}{t},
    \seq{\cgnode{s}^{\ell,k}}{1}{t},
    \big)
    &&
    \forall \ell\in\iival{1}{L},\,
    \forall k\in\iival{1}{n_\ell},\,
    \forall t\in\iival{1}{T},
    \\[.3em]
    \cgnode{X}_t^\ell
    & =
    \cgnode{X}_t^{\ell,1}
    \sqcup_{\mathrm{in}}
    \cdots
    \sqcup_{\mathrm{in}}
    \cgnode{X}_t^{\ell,n_\ell}
    &&
    \forall \ell\in\iival{1}{L},\,
    \forall t\in\iival{1}{T}
    \\[.3em]
    \cgnode{U}_t^{\ell+1}
    & =
    \bigptuple{
      \cgnode{X}_{t-1}^\ell,
      \cgnode{U}_t^\ell,
      \cgnode{X}_t^\ell,
      \mathring{I}[\cgnode{\theta}]
    }
    \triangleright
    g_\ell
    &&
    \forall \ell\in\iival{1}{L-1},\,
    \forall t\in\iival{1}{T}.
  \end{align*}

  Finally, we define the following aggregate classes:
  \begin{align*}
    \minmaxstategraphclass{D}[T][u]
    &\defeq
    \bigcup_{x_0\in X}
    \minmaxstategraphclass{D}[T][u][x_0],
    \\
    \minmaxstategraphclass{D}[T]
    &\defeq
    \bigcup_{u\in\infseqs{U}}
    \minmaxstategraphclass{D}[T][u],
    \\
    \minmaxstategraphclass{D}
    &\defeq
    \bigcup_{T\in\timedomain}
    \minmaxstategraphclass{D}[T].
  \end{align*}
\end{closingdefinition}

\begin{closingremark}[Choices of constituent state graphs]
  \label{rem:minmax-rnc-state-graph-construction-choices}
  A graph
  \begin{equation*}
    G
    \in
    \minmaxstategraphclass{D}[T][u][x_0]
  \end{equation*}
  need not uniquely determine the constituent computation graphs satisfying
  Definition~\ref{def:minmax-algebraic-rnc-state-graphs}.
  In particular, for each unit and each horizon,
  the definition permits an arbitrary recurrence graph from the corresponding
  class
  $\minmaxrecgraphclass$.
  Consequently, different admissible choices of recurrence graphs may give
  rise to the same terminal computation graph
  $G$.

  This non-uniqueness is a feature of the general
  MinMax-algebraic state-graph class and should be distinguished from the
  specific sequential and parallel-scan constructions introduced below.
  As discussed in
  Remarks~\ref{rem:minmax-sequential-recurrence-graph-determinacy}
  and~\ref{rem:minmax-parallel-scan-graph-determinacy},
  once their source graphs are fixed, these constructions determine their
  recurrence computation graphs, subject in the parallel-scan case to the
  fixed choice of materialisation.
\end{closingremark}

\begin{proposition}[Closure of MinMax RNC state graphs under analytic-MinMax computation]
  \label{prop:minmax-rnc-state-graphs-analytic-minmax}
  Assume the context of
  Definition~\ref{def:minmax-algebraic-rnc-state-graphs},
  and let
  \begin{equation*}
    G
    \in
    \minmaxstategraphclass{D}[T][u][x_0].
  \end{equation*}
  Then
  $G$
  is an
  AMM-graph.

  Furthermore,
  if
  $T\in\posnaturals$,
  then, for every choice of constituent computation graphs
  \begin{equation*}
    \cgnode{U}_t^\ell,
    \qquad
    \cgnode{R}_t^{\ell,k},
    \qquad
    \cgnode{s}_t^{\ell,k},
    \qquad
    \cgnode{H}_t^{\ell,k},
    \qquad
    \cgnode{X}_t^{\ell,k},
    \qquad
    \cgnode{X}_t^\ell
  \end{equation*}
  satisfying the defining conditions of
  Definition~\ref{def:minmax-algebraic-rnc-state-graphs}
  for
  $G$,
  every such constituent graph is an
  AMM-graph.
\end{proposition}
\begin{proof}
  Suppose first that
  $T=0$.
  By
  Definition~\ref{def:minmax-algebraic-rnc-state-graphs},
  \begin{equation*}
    G
    =
    \constpargraph{x_0}{\cgnode{\theta}}.
  \end{equation*}
  Its computation-node labels are constant and therefore analytic,
  so
  $G$
  is an
  AMM-graph.

  Suppose now that
  $T\in\posnaturals$,
  and fix any choice of constituent computation graphs satisfying the
  defining conditions of
  Definition~\ref{def:minmax-algebraic-rnc-state-graphs}
  for
  $G$.
  We proceed inductively over the layers.

  The identity graph
  \begin{equation*}
    \mathring{I}[\cgnode{\theta}]
  \end{equation*}
  has no computation nodes and is therefore an
  AMM-graph.

  For every
  $t\in\iival{1}{T}$,
  Definition~\ref{def:minmax-algebraic-rnc-state-graphs}
  gives
  \begin{equation*}
    \cgnode{U}_t^1
    =
    \constpargraph{u_t}{\cgnode{\theta}},
  \end{equation*}
  and hence
  $\cgnode{U}_t^1$
  is an
  AMM-graph.

  Let
  $\ell\in\iival{1}{L}$
  and suppose that
  $\cgnode{U}_t^\ell$
  is an
  AMM-graph
  for every
  $t\in\iival{1}{T}$.
  For every
  $k\in\iival{1}{n_\ell}$,
  the function
  \begin{equation*}
    \rho_{\ell,k}
    :
    U_\ell
    \times
    \Theta
    \to
    \reals^{d_R^{\ell,k}+d_s^{\ell,k}}
  \end{equation*}
  is analytic by the definition of MinMax RNCs.
  Therefore,
  by
  Fact~\ref{fact:analytic-extension-primitive-class},
  for every
  $t\in\iival{1}{T}$,
  \begin{equation*}
    \cgnode{H}_t^{\ell,k}
    =
    \ptuple{
      \cgnode{U}_t^\ell,
      \mathring{I}[\cgnode{\theta}]
    }
    \triangleright
    \rho_{\ell,k}
  \end{equation*}
  is an
  AMM-graph.

  Output restriction introduces no new computation nodes.
  Hence
  \begin{equation*}
    \cgnode{R}_t^{\ell,k}
    =
    \big(
      \cgnode{H}_t^{\ell,k}
    \big)_{J_R^{\ell,k}},
    \qquad
    \cgnode{s}_t^{\ell,k}
    =
    \big(
      \cgnode{H}_t^{\ell,k}
    \big)_{J_s^{\ell,k}}
  \end{equation*}
  are
  AMM-graphs
  for every
  $t\in\iival{1}{T}$.

  Furthermore,
  for every
  $t\in\iival{1}{T}$,
  \begin{equation*}
    \cgnode{X}_t^{\ell,k}
    \in
    \minmaxrecgraphclass
    \big(
      x_0^{\ell,k};
      \seq{\cgnode{R}^{\ell,k}}{1}{t},
      \seq{\cgnode{s}^{\ell,k}}{1}{t}
    \big).
  \end{equation*}
  Hence,
  by
  Proposition~\ref{prop:minmax-algebraic-recurrence-graphs-al},
  every
  $\cgnode{X}_t^{\ell,k}$
  is an
  AMM-graph.

  Since common-input union introduces no new computation-node labels,
  every
  \begin{equation*}
    \cgnode{X}_t^\ell
    =
    \cgnode{X}_t^{\ell,1}
    \sqcup_{\mathrm{in}}
    \cdots
    \sqcup_{\mathrm{in}}
    \cgnode{X}_t^{\ell,n_\ell}
  \end{equation*}
  is an
  AMM-graph.

  If
  $\ell<L$,
  then
  \begin{equation*}
    \cgnode{X}_0^\ell
    =
    \constpargraph{x_0^\ell}{\cgnode{\theta}}
  \end{equation*}
  is an
  AMM-graph.
  Furthermore,
  $g_\ell$
  is analytic by the definition of the RNC cascade composition.
  Thus,
  by
  Fact~\ref{fact:analytic-extension-primitive-class},
  for every
  $t\in\iival{1}{T}$,
  \begin{equation*}
    \cgnode{U}_t^{\ell+1}
    =
    \bigptuple{
      \cgnode{X}_{t-1}^\ell,
      \cgnode{U}_t^\ell,
      \cgnode{X}_t^\ell,
      \mathring{I}[\cgnode{\theta}]
    }
    \triangleright
    g_\ell
  \end{equation*}
  is an
  AMM-graph.
  This closes the induction over layers.

  Finally,
  by
  Definition~\ref{def:minmax-algebraic-rnc-state-graphs},
  \begin{equation*}
    G
    =
    \cgnode{X}_T^1
    \sqcup_{\mathrm{in}}
    \cdots
    \sqcup_{\mathrm{in}}
    \cgnode{X}_T^L.
  \end{equation*}
  Since each
  $\cgnode{X}_T^\ell$
  is an
  AMM-graph
  and common-input union introduces no new computation-node labels,
  $G$
  is an
  AMM-graph.

  The argument simultaneously establishes that every constituent
  computation graph in the fixed choice is an
  AMM-graph.
  Since the choice was arbitrary,
  the stronger claim follows.
\end{proof}

\begin{closingremark}[Scope of MinMax-algebraic RNC state-graph constructions]
  \label{rem:minmax-algebraic-graph-scope}
  Definition~\ref{def:minmax-algebraic-rnc-state-graphs}
  allows, for every recurrent unit and every relevant time horizon, an
  arbitrary recurrence graph from the corresponding class
  $\minmaxrecgraphclass$.

  Since
  $\minmaxrecgraphclass$
  permits arbitrary MinMax-algebraic evaluation topology,
  a graph in
  $\minmaxstategraphclass{D}[T][u][x_0]$
  may use, in particular,
  sequential recurrence evaluation,
  parallel-scan recurrence evaluation,
  or mixtures of such evaluation topologies across units,
  layers,
  and time horizons.

  The construction does not permit arbitrary algebraically equivalent smooth
  reparameterisations of the recurrence.
  Smooth computation nodes arise only from the smooth graph extensions by
  the functions
  $\rho_{\ell,k}$
  and
  $g_\ell$
  prescribed by the RNC architecture.
  The identity graph
  $\mathring{I}[\cgnode{\theta}]$
  introduces no computation nodes.
  Apart from these smooth graph extensions,
  constant parametric graphs,
  common-input unions,
  output restrictions,
  and the graph constructions used to combine them,
  recurrence evaluation is obtained through MinMax-algebraic extensions and
  therefore introduces only the scalar minimum and maximum computation nodes
  prescribed by the MinMax primitive-function family.

  The constituent computation graphs satisfying
  Definition~\ref{def:minmax-algebraic-rnc-state-graphs}
  need not be uniquely determined by the terminal graph
  \begin{equation*}
    G
    \in
    \minmaxstategraphclass{D}[T][u][x_0].
  \end{equation*}
  In particular, different choices of recurrence graphs from
  $\minmaxrecgraphclass$
  may yield the same terminal computation graph.
  All subsequent topology-independent bounds hold for every choice of
  constituent computation graphs satisfying the defining conditions and
  hence do not depend on such a choice.
  No separate construction record is attached to
  $G$.
\end{closingremark}

\begin{theorem}[Topology-independent derivative bound for MinMax RNC state evaluation]
  \label{theorem:minmax-algebraic-rnc-state-graph-bound}
  Assume the context of
  Definition~\ref{def:minmax-algebraic-rnc-state-graphs},
  let
  $\theta\in\Theta$,
  and
  let
  \begin{equation*}
    G
    \in
    \minmaxstategraphclass{D}[T][u][x_0],
  \end{equation*}
  Then,
  \begin{align*}
    \big\|
      \minmaxsdvop{G}(\theta)
    \big\|_\mathrm{row}
    &\leq
    \max_{
      \substack{
        \ell\in\iival{1}{L}\\
        k\in\iival{1}{n_\ell}\\
        t\in\iival{1}{T}
      }
    }
    \max
    \left\{
      \big\|
        \minmaxsdvop{\cgnode{R}_t^{\ell,k}}(\theta)
      \big\|_\mathrm{row},
      \;
      \big\|
        \minmaxsdvop{\cgnode{s}_t^{\ell,k}}(\theta)
      \big\|_\mathrm{row}
    \right\}.
  \end{align*}
\end{theorem}
\begin{proof}
  Suppose first that
  $T=0$.
  By
  Definition~\ref{def:minmax-algebraic-rnc-state-graphs},
  \begin{equation*}
    G
    =
    \constpargraph{x_0}{\cgnode{\theta}}.
  \end{equation*}
  Hence
  $G$
  is parameter-independent and therefore,
  by
  Lemma~\ref{lem:parameter-independent-zero-derivative},
  \begin{equation*}
    \minmaxsdvop{G}(\theta)
    =
    0.
  \end{equation*}

  Suppose now that
  $T\in\posnaturals$,
  and fix any choice of constituent computation graphs satisfying the
  defining conditions of
  Definition~\ref{def:minmax-algebraic-rnc-state-graphs}
  for
  $G$.

  By definition,
  \begin{equation*}
    G
    =
    \cgnode{X}_T^1
    \sqcup_{\mathrm{in}}
    \cdots
    \sqcup_{\mathrm{in}}
    \cgnode{X}_T^L,
  \end{equation*}
  and, for every
  $\ell\in\iival{1}{L}$,
  \begin{equation*}
    \cgnode{X}_T^\ell
    =
    \cgnode{X}_T^{\ell,1}
    \sqcup_{\mathrm{in}}
    \cdots
    \sqcup_{\mathrm{in}}
    \cgnode{X}_T^{\ell,n_\ell}.
  \end{equation*}

  By
  Proposition~\ref{prop:minmax-rnc-state-graphs-analytic-minmax},
  all these graphs are
  AMM-graphs,
  so their
  $\minmaxsdvop$-pseudo-derivatives
  are well defined.

  Since common-input union concatenates the output node-vectors of its
  constituent graphs,
  the rows of
  $\minmaxsdvop{G}(\theta)$
  are precisely the rows of
  \begin{equation*}
    \minmaxsdvop{\cgnode{X}_T^{\ell,k}}(\theta),
    \qquad
    \ell\in\iival{1}{L},
    \quad
    k\in\iival{1}{n_\ell}.
  \end{equation*}
  Hence, by the definition of the row norm,
  \begin{equation}
    \label{eq:minmax-rnc-state-bound-unit-reduction}
    \big\|
      \minmaxsdvop{G}(\theta)
    \big\|_\mathrm{row}
    =
    \max_{
      \substack{
        \ell\in\iival{1}{L}\\
        k\in\iival{1}{n_\ell}
      }
    }
    \big\|
      \minmaxsdvop{\cgnode{X}_T^{\ell,k}}(\theta)
    \big\|_\mathrm{row}.
  \end{equation}

  Fix
  $\ell\in\iival{1}{L}$
  and
  $k\in\iival{1}{n_\ell}$.
  By
  Definition~\ref{def:minmax-algebraic-rnc-state-graphs},
  \begin{equation*}
    \cgnode{X}_T^{\ell,k}
    \in
    \minmaxrecgraphclass
    \big(
      x_0^{\ell,k};
      \seq{\cgnode{R}^{\ell,k}}{1}{T},
      \seq{\cgnode{s}^{\ell,k}}{1}{T}
    \big).
  \end{equation*}
  Therefore,
  by
  Theorem~\ref{theorem:minmax-algebraic-recurrence-graph-bound},
  \begin{equation*}
    \big\|
      \minmaxsdvop{\cgnode{X}_T^{\ell,k}}(\theta)
    \big\|_\mathrm{row}
    \leq
    \max_{t\in\iival{1}{T}}
    \max
    \left\{
      \big\|
        \minmaxsdvop{\cgnode{R}_t^{\ell,k}}(\theta)
      \big\|_\mathrm{row},
      \;
      \big\|
        \minmaxsdvop{\cgnode{s}_t^{\ell,k}}(\theta)
      \big\|_\mathrm{row}
    \right\}.
  \end{equation*}

  Since
  $\ell$
  and
  $k$
  were arbitrary,
  taking the maximum over
  $\ell\in\iival{1}{L}$
  and
  $k\in\iival{1}{n_\ell}$,
  and using
  \eqref{eq:minmax-rnc-state-bound-unit-reduction},
  yields
  \begin{align*}
    \big\|
      \minmaxsdvop{G}(\theta)
    \big\|_\mathrm{row}
    &\leq
    \max_{
      \substack{
        \ell\in\iival{1}{L}\\
        k\in\iival{1}{n_\ell}\\
        t\in\iival{1}{T}
      }
    }
    \max
    \left\{
      \big\|
        \minmaxsdvop{\cgnode{R}_t^{\ell,k}}(\theta)
      \big\|_\mathrm{row},
      \;
      \big\|
        \minmaxsdvop{\cgnode{s}_t^{\ell,k}}(\theta)
      \big\|_\mathrm{row}
    \right\}.
  \end{align*}
\end{proof}

\subsection{Sequential Computation Graphs for MinMax RNCs}
\label{subsec:sequential-computation-graphs-for-minmax-rncs}

\begin{closingdefinition}[Sequential computation graph]
  \label{def:minmax-canonical-recurrence-graphs}
  We define the computation graph for evaluating a MinMax recurrence
  sequentially.

  \par\smallskip
  \textbf{Context.}~
  Assume the context of
  Definition~\ref{def:minmax-algebraic-recurrence-graphs}.

  For every
  $t\in\iival{1}{T}$,
  let
  $\cgnode{R}[t][\mathrm{out}]$
  and
  $\cgnode{s}[t][\mathrm{out}]$
  be the node-matrix and node-vector output presentations of
  $\cgnode{R}[t]$
  and
  $\cgnode{s}[t]$,
  respectively,
  and let
  $\cgnode{x}[0][\mathrm{out}]$
  be the output node-vector of
  $\cgnode{X}[0]$.
  For brevity, in the construction below we write
  \begin{equation*}
    \cgnode{x}[0]
    \defeq
    \cgnode{x}[0][\mathrm{out}].
  \end{equation*}

  \par\smallskip
  \textbf{Auxiliary.}~
  Let us define the common-input union
  \begin{equation*}
    G^\sqcup
    \defeq
    \cgnode{X}[0]
    \sqcup_{\mathrm{in}}
    \cgnode{R}[1]
    \sqcup_{\mathrm{in}}
    \cdots
    \sqcup_{\mathrm{in}}
    \cgnode{R}[T]
    \sqcup_{\mathrm{in}}
    \cgnode{s}[1]
    \sqcup_{\mathrm{in}}
    \cdots
    \sqcup_{\mathrm{in}}
    \cgnode{s}[T],
  \end{equation*}
  regarded as a virtual-computation graph with no virtual nodes.

  We set
  \begin{equation*}
    G^\mathrm{seq}_0
    \defeq
    G^\sqcup,
  \end{equation*}
  and, for every
  $t\in\iival{1}{T}$,
  define the virtual-node extension
  \begin{equation}
    \label{eq:minmax-sequential-extension}
    \vgraphext
    {G^\mathrm{seq}_{t-1}}
    {
      \cgnode{x}[t]
      \defeq
      \big(
        \cgnode{R}[t][\mathrm{out}]
        \otimes
        \cgnode{x}[t-1]
      \big)
      \oplus
      \cgnode{s}[t][\mathrm{out}]
    }
    {G^\mathrm{seq}_t}.
  \end{equation}
  Here the nested MinMax expression is interpreted according to
  Definition~\ref{def:minmax-vector-presentation};
  in particular, the required intermediate virtual nodes are introduced as
  part of the extension.

  \par\smallskip
  \textbf{Main definition: graph instances.}~
  The
  \emph{MinMax sequential virtual-computation graph of horizon $T$ given
  initial-state graph
  $\cgnode{X}[0]$
  and reset and set graphs
  $\seq{\cgnode{R}}{1}{T},\seq{\cgnode{s}}{1}{T}$}
  is the virtual-computation graph
  \begin{equation*}
    \minmaxseqvgraph{
      \cgnode{X}[0]
    }{
      \seq{\cgnode{R}}{1}{T}
    }{
      \seq{\cgnode{s}}{1}{T}
    }
    \defeq
    G^\mathrm{seq}_T,
  \end{equation*}
  having output node-vector
  $\cgnode{x}[T]$.

  The
  \emph{MinMax sequential computation graph of horizon $T$ given
  initial-state graph
  $\cgnode{X}[0]$
  and reset and set graphs
  $\seq{\cgnode{R}}{1}{T},\seq{\cgnode{s}}{1}{T}$},
  denoted by
  \begin{equation*}
    \minmaxseqgraph{
      \cgnode{X}[0]
    }{
      \seq{\cgnode{R}}{1}{T}
    }{
      \seq{\cgnode{s}}{1}{T}
    },
  \end{equation*}
  is the computation graph given by the materialisation of
  \begin{equation*}
    \minmaxseqvgraph{
      \cgnode{X}[0]
    }{
      \seq{\cgnode{R}}{1}{T}
    }{
      \seq{\cgnode{s}}{1}{T}
    }
  \end{equation*}
  under an arbitrary but fixed materialisation map
  $\mu_T^\mathrm{seq}$.

  \par\smallskip
  \textbf{Main definition: graph classes.}~
  The
  \emph{MinMax sequential virtual-computation graph family of horizon $T$
  given initial state
  $x_0$
  and reset and set functions
  $\seq{R}{1}{T},\seq{s}{1}{T}$},
  denoted by
  \begin{equation*}
    \minmaxseqvgraphclass{
      x_0
    }{
      \seq{R}{1}{T}
    }{
      \seq{s}{1}{T}
    },
  \end{equation*}
  is the set consisting of each graph
  \begin{equation*}
    \minmaxseqvgraph{
      \cgnode{X}[0]
    }{
      \seq{\cgnode{R}}{1}{T}
    }{
      \seq{\cgnode{s}}{1}{T}
    }
  \end{equation*}
  for
  $\cgnode{X}[0]$
  the constant canonical graph for
  $x_0$,
  \begin{equation*}
    \cgnode{R}[t]
    \in
    \minmaxgraphclass[R_t],
    \qquad
    \cgnode{s}[t]
    \in
    \minmaxgraphclass[s_t]
    \qquad
    \forall t\in\iival{1}{T}.
  \end{equation*}

  The
  \emph{MinMax sequential computation graph family of horizon $T$
  given initial state
  $x_0$
  and reset and set functions
  $\seq{R}{1}{T},\seq{s}{1}{T}$},
  denoted by
  \begin{equation*}
    \minmaxseqgraphclass{
      x_0
    }{
      \seq{R}{1}{T}
    }{
      \seq{s}{1}{T}
    },
  \end{equation*}
  is the set consisting of each graph
  \begin{equation*}
    \minmaxseqgraph{
      \cgnode{X}[0]
    }{
      \seq{\cgnode{R}}{1}{T}
    }{
      \seq{\cgnode{s}}{1}{T}
    }
  \end{equation*}
  for
  $\cgnode{X}[0]$
  the constant canonical graph for
  $x_0$,
  \begin{equation*}
    \cgnode{R}[t]
    \in
    \minmaxgraphclass[R_t],
    \qquad
    \cgnode{s}[t]
    \in
    \minmaxgraphclass[s_t]
    \qquad
    \forall t\in\iival{1}{T}.
  \end{equation*}
\end{closingdefinition}

\begin{closingremark}[Determination of sequential recurrence graphs]
  \label{rem:minmax-sequential-recurrence-graph-determinacy}
  In contrast with the general state-graph construction discussed in
  Remark~\ref{rem:minmax-rnc-state-graph-construction-choices},
  the sequential recurrence graph is determined once its source graphs
  \begin{equation*}
    X_0^G,
    \qquad
    R_1^G,\ldots,R_T^G,
    \qquad
    s_1^G,\ldots,s_T^G
  \end{equation*}
  are fixed.
  Indeed,
  Definition~\ref{def:minmax-canonical-recurrence-graphs}
  recursively determines the node-vectors
  \begin{equation*}
    \cgnode{z}_t
    =
    \cgnode{R}_t\otimes\cgnode{x}_{t-1},
    \qquad
    \cgnode{x}_t
    =
    \cgnode{z}_t\oplus\cgnode{s}_t,
    \qquad
    t\in\iival{1}{T},
  \end{equation*}
  together with their prescribed scalar expansions.

  In particular,
  $\cgnode{x}_1,\ldots,\cgnode{x}_{T-1}$
  are intermediate node-vectors of the single horizon-$T$ graph
  $G_T^\mathrm{seq}$;
  they are not independently chosen recurrence graphs of smaller horizons.
  Thus no additional choice of recurrence topology remains after the source
  graphs have been fixed.
\end{closingremark}

\begin{closingremark}
  In
  Definition~\ref{def:minmax-canonical-recurrence-graphs},
  the source graphs
  \begin{equation*}
    \cgnode{X}[0],
    \qquad
    \cgnode{R}[1],\ldots,\cgnode{R}[T],
    \qquad
    \cgnode{s}[1],\ldots,\cgnode{s}[T]
  \end{equation*}
  retain all their concrete nodes,
  multi-edges,
  and computation-node labels.

  Every node added to evaluate the recurrence is a virtual scalar computation
  node introduced by the MinMax notation of
  Definition~\ref{def:minmax-vector-presentation}.
  Materialisation replaces these virtual nodes by fresh concrete computation
  nodes while preserving their predecessor structure,
  labels,
  and value functions.
\end{closingremark}

\begin{proposition}[Correctness of the sequential computation graph]
  \label{prop:correctness-minmax-canonical-recurrence-graphs}
  Assume the context of
  Definition~\ref{def:minmax-canonical-recurrence-graphs}.

  \todo{}

  For every
  $\lambda\in\Lambda$
  and
  $t\in\iival{0}{T}$,
  \begin{equation}
    \label{eq:minmax-sequential-state-correctness}
    \cgval{x}[t](\lambda)
    =
    x_t(\lambda).
  \end{equation}
  Consequently,
  for every choice of source graphs admitted by
  Definition~\ref{def:minmax-canonical-recurrence-graphs},
  the graph
  \begin{equation*}
      \minmaxseqgraph{
        \cgnode{X}[0]
      }{
        \seq{\cgnode{R}}{1}{T}
      }{
        \seq{\cgnode{s}}{1}{T}
      }
  \end{equation*}
  represents $\cgval{x}_T$.
\end{proposition}

\begin{proof}
  We prove
  Equation~\eqref{eq:minmax-sequential-state-correctness}
  by induction on
  $t$.

  For
  $t=0$,
  the claim follows from the definition of
  $\cgnode{X}[0]$:
  \begin{equation*}
    \cgval{x}[0](\lambda)
    =
    \cgval{x}[0][\mathrm{out}](\lambda)
    =
    x_0
    =
    x_0(\lambda).
  \end{equation*}

  Let now
  $t\in\iival{1}{T}$
  and suppose that
  \begin{equation*}
    \cgval{x}[t-1](\lambda)
    =
    x_{t-1}(\lambda).
  \end{equation*}
  By
  Equation~\eqref{eq:minmax-sequential-extension},
  Fact~\ref{fact:minmax-vector-presentation-values},
  and the defining values of the source presentations,
  \begin{align*}
    \cgval{x}[t](\lambda)
    &=
    \big(
      \cgval{R}[t][\mathrm{out}](\lambda)
      \otimes
      \cgval{x}[t-1](\lambda)
    \big)
    \oplus
    \cgval{s}[t][\mathrm{out}](\lambda)
    \\
    &=
    \big(
      R_t(\lambda)
      \otimes
      x_{t-1}(\lambda)
    \big)
    \oplus
    s_t(\lambda)
    \\
    &=
    x_t(\lambda).
  \end{align*}
  This proves
  Equation~\eqref{eq:minmax-sequential-state-correctness}.

  In particular,
  the final output presentation
  $\cgnode{x}[T]$
  of
  \begin{equation*}
    \minmaxseqvgraph{
      \cgnode{X}[0]
    }{
      \seq{\cgnode{R}}{1}{T}
    }{
      \seq{\cgnode{s}}{1}{T}
    }
  \end{equation*}
  represents
  $x_T$.
  By
  Proposition~\ref{prop:virtual-graph-materialisation-correctness},
  materialisation preserves its value function, and hence the value of the materialised graph
  \begin{equation*}
      \minmaxseqgraph{
        \cgnode{X}[0]
      }{
        \seq{\cgnode{R}}{1}{T}
      }{
        \seq{\cgnode{s}}{1}{T}
      }
  \end{equation*}
  is $x_T$.
\end{proof}

\begin{proposition}[Sequential graphs are MinMax-algebraic]
  \label{prop:minmax-sequential-graphs-algebraic}
  Assume the context of
  Definition~\ref{def:minmax-canonical-recurrence-graphs}.

  For every choice of source graphs admitted by that definition,
  \begin{equation}
    \label{eq:minmax-sequential-virtual-extension-class}
    \minmaxseqvgraph{
      \cgnode{X}[0]
    }{
      \seq{\cgnode{R}}{1}{T}
    }{
      \seq{\cgnode{s}}{1}{T}
    }
    \in
    \minmaxextensionclass^{\mathrm{v}}[
      \cgnode{X}[0],
      \cgnode{R}[1],\ldots,\cgnode{R}[T],
      \cgnode{s}[1],\ldots,\cgnode{s}[T]
    ],
  \end{equation}
  and
  \begin{equation}
    \label{eq:minmax-sequential-extension-class}
    \minmaxseqgraph{
      \cgnode{X}[0]
    }{
      \seq{\cgnode{R}}{1}{T}
    }{
      \seq{\cgnode{s}}{1}{T}
    }
    \in
    \minmaxextensionclass[
      \cgnode{X}[0],
      \cgnode{R}[1],\ldots,\cgnode{R}[T],
      \cgnode{s}[1],\ldots,\cgnode{s}[T]
    ].
  \end{equation}

  Consequently,
  \begin{equation}
    \label{eq:minmax-sequential-recurrence-class}
    \minmaxseqgraphclass{
      x_0
    }{
      \seq{R}{1}{T}
    }{
      \seq{s}{1}{T}
    }
    \subseteq
    \minmaxrecgraphclass
    \big(
      x_0;
      \seq{R}{1}{T},
      \seq{s}{1}{T}
    \big).
  \end{equation}
\end{proposition}

\begin{proof}
  By construction,
  \begin{equation*}
    \minmaxseqvgraph{
      \cgnode{X}[0]
    }{
      \seq{\cgnode{R}}{1}{T}
    }{
      \seq{\cgnode{s}}{1}{T}
    }
  \end{equation*}
  is a virtual-node extension of the common-input union
  $G^\sqcup$
  of
  \begin{equation*}
    \cgnode{X}[0],
    \qquad
    \cgnode{R}[1],\ldots,\cgnode{R}[T],
    \qquad
    \cgnode{s}[1],\ldots,\cgnode{s}[T].
  \end{equation*}

  Every extension
  in
  Equation~\eqref{eq:minmax-sequential-extension}
  is specified by the MinMax notation of
  Definition~\ref{def:minmax-vector-presentation}.
  Hence every scalar virtual computation node introduced by the sequential
  recurrence construction has a label admitted by
  Definition~\ref{def:minmax-algebraic-virtual-extension}.
  It follows that
  \begin{equation*}
    \minmaxseqvgraph{
      \cgnode{X}[0]
    }{
      \seq{\cgnode{R}}{1}{T}
    }{
      \seq{\cgnode{s}}{1}{T}
    }
    \in
    \minmaxextensionclass^{\mathrm{v}}[
      \cgnode{X}[0],
      \cgnode{R}[1],\ldots,\cgnode{R}[T],
      \cgnode{s}[1],\ldots,\cgnode{s}[T]
    ],
  \end{equation*}
  proving
  Equation~\eqref{eq:minmax-sequential-virtual-extension-class}.

  By
  Proposition~\ref{prop:minmax-virtual-extension-materialisation},
  its materialisation belongs to the corresponding ordinary
  MinMax-algebraic extension class.
  Thus
  \begin{equation*}
    \minmaxseqgraph{
      \cgnode{X}[0]
    }{
      \seq{\cgnode{R}}{1}{T}
    }{
      \seq{\cgnode{s}}{1}{T}
    }
    \in
    \minmaxextensionclass[
      \cgnode{X}[0],
      \cgnode{R}[1],\ldots,\cgnode{R}[T],
      \cgnode{s}[1],\ldots,\cgnode{s}[T]
    ],
  \end{equation*}
  proving
  Equation~\eqref{eq:minmax-sequential-extension-class}.

  Proposition~\ref{prop:correctness-minmax-canonical-recurrence-graphs}
  gives
  that the graph
  \begin{equation*}
      \minmaxseqgraph{
        \cgnode{X}[0]
      }{
        \seq{\cgnode{R}}{1}{T}
      }{
        \seq{\cgnode{s}}{1}{T}
      }
  \end{equation*}
  represents $\cgval{x}_T$.
  Therefore,
  by
  Definition~\ref{def:minmax-algebraic-recurrence-graphs},
  every such materialised sequential graph belongs to
  \begin{equation*}
    \minmaxrecgraphclass
    \big(
      x_0;
      \seq{R}{1}{T},
      \seq{s}{1}{T}
    \big).
  \end{equation*}

  Since the source graphs were arbitrary members of the graph classes
  specified in
  Definition~\ref{def:minmax-canonical-recurrence-graphs},
  Equation~\eqref{eq:minmax-sequential-recurrence-class}
  follows.
\end{proof}

\subsection{Parallel-Scan Computation Graphs for MinMax RNCs}
\label{subsec:parallel-scan-computation-graphs-for-minmax-rncs}

\begin{closingdefinition}[Parallel-scan computation graph]
  \label{def:minmax-parallel-scan-graph}
  We define the computation graph for evaluating MinMax recurrence via 
  \emph{exclusive parallel-scan}.

  \par\smallskip
  \textbf{Context.}~
  Assume the context of
  Definition~\ref{def:minmax-algebraic-recurrence-graphs}.

  For every
  $t\in\iival{1}{T}$,
  let
  $\cgnode{R}[t][\mathrm{out}]$
  and
  $\cgnode{s}[t][\mathrm{out}]$
  be the node-matrix and node-vector output presentations of
  $\cgnode{R}[t]$
  and
  $\cgnode{s}[t]$,
  respectively,
  and let
  $\cgnode{x}[0][\mathrm{out}]$
  be the output node-vector of
  $\cgnode{X}[0]$.
  For some $n \in \posnaturals$,
  we have that each 
  $\cgnode{R}[t][\mathrm{out}]$
  has dimension $n^2$,
  and each $\cgnode{s}[t][\mathrm{out}]$
  has dimension $n$.

  \par\smallskip
  \textbf{Auxiliary.}~
  Let us define the common-input union
  \begin{equation*}
    G^\sqcup
    \defeq
    \cgnode{X}[0]
    \sqcup_{\mathrm{in}}
    \cgnode{R}[1]
    \sqcup_{\mathrm{in}}
    \cdots
    \sqcup_{\mathrm{in}}
    \cgnode{R}[T]
    \sqcup_{\mathrm{in}}
    \cgnode{s}[1]
    \sqcup_{\mathrm{in}}
    \cdots
    \sqcup_{\mathrm{in}}
    \cgnode{s}[T]
  \end{equation*}
  regarded as a virtual-computation graph with no virtual nodes.

  Let
  $\cgnode{\mathcal{V}}$
  be the node-vector obtained by concatenating all scalar entries of
  \begin{equation*}
    \cgnode{x}[0][\mathrm{out}],
    \quad
    \cgnode{R}[1][\mathrm{out}],\ldots,\cgnode{R}[T][\mathrm{out}],
    \quad
    \cgnode{s}[1][\mathrm{out}],\ldots,\cgnode{s}[T][\mathrm{out}].
  \end{equation*}

  We extend $G^\sqcup$ as follows:
  \begin{gather*}
    \vgraphext
    {G^\sqcup}
    {
      \cgnode{v}[\minreductionsymb]
      \defeq 
      \minreductionsymb(\cgnode{\mathcal{V}}),
      \;
      \cgnode{v}[\maxreductionsymb]
      \defeq 
      \maxreductionsymb(\cgnode{\mathcal{V}}),
      \;
      \cgnode{E},
      \;
      \cgnode{e},
      \;
      \cgnode{M}[\id]
      \defeq
      \ptuple{
        \cgnode{E},
        \cgnode{e}
      }
    }
    {G^\mathrm{src}_0}
  \end{gather*}
  where the node-matrix $\cgnode{E}$ and the node-vector $\cgnode{e}$ have components as follows:
  \begin{align*}
    \qquad
    \getentry{\cgnode{E}}{i,j}
    \defeq
    \begin{cases}
      \cgnode{v}[\maxreductionsymb]
      &\text{if } i=j,
      \\[.2em]
      \cgnode{v}[\minreductionsymb]
      &\text{if } i\neq j,
    \end{cases}
    \quad\quad
    \getentry{\cgnode{e}}{i}
    \defeq
    \cgnode{v}[\minreductionsymb],
    \qquad\qquad \forall i,j \in \iival{1}{n}.
  \end{align*}
  Then we define the extensions
  \begin{align*}
    \vgraphext
    {G^\mathrm{src}_{t-1}}
    {
      \cgnode{M}[t]
      \defeq
      \ptuple{
        \cgnode{R}[t][\mathrm{out}],
        \cgnode{s}[t][\mathrm{out}]
      }
    }
    {G^\mathrm{src}_t}
    \qquad\quad
    \forall t \in \iival{1}{T}.
  \end{align*}

  \par\smallskip
  \textbf{Exclusive-scan virtual-computation graph.}~
  For every
  $t\in\posnaturals$,
  we inductively define the \emph{exclusive-scan virtual-computation graph}
  \begin{equation*}
    \mathrm{xscan}_t
    \big(
      \cgnode{M}[\id]; \seq{\cgnode{M}}{1}{t} \,|\, G^\mathrm{src}_t 
    \big)
    \quad\text{ having output node-vector }\;\; \bigptuple{\cgnode{P}[1],\ldots,\cgnode{P}[t]},
  \end{equation*}
  as follows:
  \begin{itemize}
    \item
      when $t=1$,
      the exclusive-scan virtual-computation graph is given by the following extension:
      \begin{align*}
        \vgraphext
        {G^\mathrm{src}_1}
        {
          \cgnode{P}[1] \defeq \cgnode{M}[\id]
        }
        {
          \mathrm{xscan}_t
          \big(
          \cgnode{M}[\id]; \seq{\cgnode{M}}{1}{t} \,|\, G^\mathrm{src}_t 
          \big)
        }
      \end{align*}
    \item
      when $t>1$,
      we define $m \defeq \left\lceil \frac{t}{2} \right\rceil$, and then we proceed as follows:
      \begin{itemize}
        \item
          we first define the extension
          \begin{align*}
            \begin{cases}
              \vgraphext
              {G^\mathrm{src}_t}
              {
                \big(
                \cgnode{C}[\tau]
                \defeq
                \cgnode{M}[2\tau- 1]
                \adaoplustimes
                \cgnode{M}[2\tau]
                \,{\textstyle : }\, 
                \tau \in \iival{1}{t/2}
                \big)
              }
              {G^{\mathrm{c}}_t}
              &
              \text{ if $t$ is even},
              \\[.3em]
              \vgraphext
              {G^\mathrm{src}_t}
              {
                \big(
                \cgnode{C}[\tau]
                \defeq
                \cgnode{M}[2\tau- 1]
                \adaoplustimes
                \cgnode{M}[2\tau]
                \,{\textstyle : }\, 
                \tau \in \iival{1}{\lfloor t/2\rfloor}
                \big),\;
                \cgnode{C}[m]
                \defeq
                \cgnode{M}[t]
              }
              {G^{\mathrm{c}}_t}
              &
              \text{ if $t$ is odd};
            \end{cases}
          \end{align*}
        \item
          the induction yields the exclusive-scan virtual-computation graph for $m < t$,
          \begin{align*}
            G^\mathrm{rec}_m
            & 
            \defeq 
            \mathrm{scan}_m
            \big(
            \cgnode{M}[\id]; \cgnode{C}[1],\ldots,\cgnode{C}[m]
            \bigm|
            G^\mathrm{c}_t
            \big),
            \; \text{ with output node-vector }\,
            \bigptuple{\cgnode{Q}[1],\ldots,\cgnode{Q}[m]},
          \end{align*}
        \item
          then, we define the extension
          \begin{align*}
            \begin{cases}
              \vgraphext
              {G^\mathrm{rec}_t}
              {
                \big(
                \cgnode{P}[2\tau- 1]
                \defeq
                \cgnode{Q}[\tau],\;
                \cgnode{P}[2\tau]
                \defeq
                \cgnode{Q}[\tau]
                \oplustimes
                \cgnode{M}[2\tau- 1]
                \,{\textstyle : }\, 
                \tau \in \iival{1}{t/2}
                \big)
              }
              {G^\mathrm{scan}_t}
              &
              \text{ if $t$ is even},
              \\[1em]
              \vgraphext
              {G^\mathrm{rec}_t}
              {
                \big(
                \cgnode{P}[2\tau- 1]
                \defeq
                \cgnode{Q}[\tau],\;
                \cgnode{P}[2\tau]
                \defeq
                \cgnode{Q}[\tau]
                \oplustimes
                \cgnode{M}[2\tau- 1]
                \,{\textstyle : }\, 
                \tau \in \iival{1}{\lfloor t/2\rfloor}
                \big),
                \;
                \cgnode{P}[t]
                \defeq
                \cgnode{Q}[m]
              }
              {G^\mathrm{scan}_t}
              &
              \text{ if $t$ is odd};
            \end{cases}
          \end{align*}
      \end{itemize}
    \item
      finally, we define the exclusive-scan virtual-computation graph as the graph
      \begin{equation*}
        \mathrm{xscan}_t
        \big(
          \cgnode{M}[\id]; \seq{\cgnode{M}}{1}{t} \,|\, G^\mathrm{src}_t 
        \big)
        \defeq 
        G^\mathrm{scan}_t.
      \end{equation*}
  \end{itemize}

  \par\smallskip
  \textbf{Inclusive virtual-computation graphs.}~
  We extend the exclusive scan graphs to compute states:
  \begin{align*}
    \vgraphext
    {
      \mathrm{xscan}_t
      \big(
      \cgnode{M}[\id]; \seq{\cgnode{M}}{1}{t} \,|\, G^\mathrm{src}_t 
      \big)
    }
    {
      \ptuple{
        \cgnode{A}[1:t],
        \cgnode{b}[1:t]
      }
      \defeq
      \cgnode{P}[t]
      \oplustimes
      \cgnode{M}[t],
      \;
      \;
      \cgnode{x}[t]
      \defeq
      (
      \cgnode{A}[1:t]
      \otimes
      \cgnode{x}[0]
      )
      \oplus
      \cgnode{b}[1:t]
    }
    {
      \mathrm{scan}_t
      \big(
      \cgnode{M}[\id]; \seq{\cgnode{M}}{1}{t} \,|\, G^\mathrm{src}_t 
      \big)
    }
  \end{align*}
  for every $t \in \iival{1}{T}$.

  \par\smallskip
  \textbf{Main definition: graph instances.}~
  The
  \emph{MinMax parallel-scan virtual-computation graph of horizon $T$ given initial state graph 
  $\cgnode{X}_0$ and reset and set graphs $\seq{\cgnode{R}}{1}{T},\seq{\cgnode{s}}{1}{T}$} is the
  virtual-computation graph
  \begin{align*}
    \mathrm{scan}_{\scriptscriptstyle\adaoplustimes,\mathrm{v}}\big(\cgnode{X}_0, \seq{\cgnode{R}}{1}{T},\seq{\cgnode{s}}{1}{T}, \big) 
    \defeq \mathrm{scan}_t
      \big(
      \cgnode{M}[\id]; \seq{\cgnode{M}}{1}{t} \,|\, G^\mathrm{src}_t 
      \big)
  \end{align*}
  having output node-vector $\cgnode{x}[T]$.

  The \emph{MinMax parallel-scan computation graph of horizon $T$ given initial state graph 
  $\cgnode{X}_0$ and reset and set graphs $\seq{\cgnode{R}}{1}{T},\seq{\cgnode{s}}{1}{T}$} 
  denoted by
  \begin{align*}
    \mathrm{scan}_{\scriptscriptstyle\adaoplustimes}
    \big(\cgnode{X}_0, \seq{\cgnode{R}}{1}{T},\seq{\cgnode{s}}{1}{T}, \big)
  \end{align*}
  is the computation graph given by the materialisation of 
  \begin{align*}
    \mathrm{scan}_{\scriptscriptstyle\adaoplustimes,\mathrm{v}}\big(\cgnode{X}_0, \seq{\cgnode{R}}{1}{T},\seq{\cgnode{s}}{1}{T}, \big)
  \end{align*}
  under an arbitrary but fixed materialisation map $\mu_T^\mathrm{scan}$.

  \par\smallskip
  \textbf{Main definition: graph classes.}~
  The \emph{MinMax parallel-scan virtual-computation graph family of horizon $T$ given initial state  
  $x_0$ and reset and set functions $\seq{R}{1}{T},\seq{s}{1}{T}$} 
  denoted by
  \begin{align*}
    \minmaxgraphclass^{\mathrm{scan},\mathrm{v}}\big(x_0, \seq{R}{1}{T},\seq{s}{1}{T}, \big) 
  \end{align*}
  is the set consisting of each graph
  \begin{align*}
    \mathrm{scan}_{\scriptscriptstyle\adaoplustimes,\mathrm{v}}\big(\cgnode{X}_0,
    \seq{\cgnode{R}}{1}{T},\seq{\cgnode{s}}{1}{T}, \big) 
  \end{align*}
  for 
  $\cgnode{X}_0$ the constant canonical graph for $x_0$,
  $\cgnode{R}_t \in \minmaxgraphclass[R_t]$,
  and
  $\cgnode{s}_t \in \minmaxgraphclass[s_t]$.

  The \emph{MinMax parallel-scan computation graph family of horizon $T$ given initial state  
  $x_0$ and reset and set functions $\seq{R}{1}{T},\seq{s}{1}{T}$},
  denoted by
  \begin{align*}
    \minmaxgraphclass^{\mathrm{scan}}\big(x_0, \seq{R}{1}{T},\seq{s}{1}{T}, \big) 
  \end{align*}
  is the set consisting of each graph
  \begin{align*}
    \mathrm{scan}_{\scriptscriptstyle\adaoplustimes}\big(\cgnode{X}_0, \seq{\cgnode{R}}{1}{T},\seq{\cgnode{s}}{1}{T}, \big) 
  \end{align*}
  for 
  $\cgnode{X}_0$ the constant canonical graph for $x_0$,
  $\cgnode{R}_t \in \minmaxgraphclass[R_t]$,
  and
  $\cgnode{s}_t \in \minmaxgraphclass[s_t]$.
\end{closingdefinition}

\begin{closingremark}[Determination of parallel-scan recurrence graphs]
  \label{rem:minmax-parallel-scan-graph-determinacy}
  As in the sequential case, and in contrast with the general state-graph
  construction of
  Remark~\ref{rem:minmax-rnc-state-graph-construction-choices},
  the parallel-scan construction determines its recurrence graph once the
  source graphs
  \begin{equation*}
    X_0^G,
    \qquad
    R_1^G,\ldots,R_T^G,
    \qquad
    s_1^G,\ldots,s_T^G
  \end{equation*}
  and the materialisation map used in
  Definition~\ref{def:minmax-parallel-scan-graph}
  are fixed.

  In particular, the presentations
  $\cgnode{x}[t]$,
  for
  $t\in\iival{1}{T}$,
  are determined within the single parallel-scan construction and are not
  independently chosen recurrence graphs of the corresponding horizons.
  Hence, after fixing the source graphs and the materialisation,
  no additional choice of recurrence topology remains.
\end{closingremark}

\begin{fact}[Values and data-generated closure of the parallel scan]
  \label{fact:minmax-parallel-scan-source-and-closure}
  Assume the context of
  Definition~\ref{def:minmax-parallel-scan-graph}.

  For every
  $\lambda\in\Lambda$,
  let
  \begin{equation*}
    \mathcal{A}_\lambda
    \defeq
    \minmaxdatadomain[\lambda]^{n\times n}
    \times
    \minmaxdatadomain[\lambda]^n.
  \end{equation*}

  For every
  $t\in\iival{1}{T}$,
  \begin{equation}
    \label{eq:minmax-parallel-scan-source-values}
    \cgval{x}[0][\mathrm{out}](\lambda)
    =
    x_0,
    \qquad
    \cgval{R}[t][\mathrm{out}](\lambda)
    =
    R_t(\lambda),
    \qquad
    \cgval{s}[t][\mathrm{out}](\lambda)
    =
    s_t(\lambda).
  \end{equation}

  Furthermore,
  \begin{equation}
    \label{eq:minmax-parallel-scan-reduction-values}
    \cgval{v}[\minreductionsymb](\lambda)
    =
    \vmin[\lambda],
    \qquad
    \cgval{v}[\maxreductionsymb](\lambda)
    =
    \vmax[\lambda].
  \end{equation}

  The node-matrix and node-vector presentations
  $\cgnode{E}$
  and
  $\cgnode{e}$
  represent parametric functions
  \begin{equation*}
    E
    :
    \Lambda
    \to
    \reals^{n\times n},
    \qquad
    e
    :
    \Lambda
    \to
    \reals^n
  \end{equation*}
  given by
  \begin{align}
    \label{eq:minmax-parallel-scan-identity-values}
    \getentry{E(\lambda)}{i,j}
    &=
    \begin{cases}
      \vmax[\lambda],
      &\text{if } i=j,
      \\[.2em]
      \vmin[\lambda],
      &\text{if } i\neq j,
    \end{cases}
    &
    \getentry{e(\lambda)}{i}
    &=
    \vmin[\lambda].
  \end{align}
  Hence
  $\cgnode{M}[\id]$
  represents
  \begin{equation*}
    M_{\id}
    :
    \Lambda
    \to
    \reals^{n\times n}
    \times
    \reals^n,
    \qquad
    M_{\id}(\lambda)
    \defeq
    \ptuple{
      E(\lambda),
      e(\lambda)
    }.
  \end{equation*}
  Furthermore,
  $M_{\id}(\lambda)$
  is the identity element under
  $\oplustimes$
  on
  $\mathcal{A}_\lambda$.

  For every
  $t\in\iival{1}{T}$,
  \begin{equation}
    \label{eq:minmax-parallel-scan-source-M-values}
    \cgval{M}[t](\lambda)
    =
    M_t(\lambda)
    \defeq
    \ptuple{
      R_t(\lambda),
      s_t(\lambda)
    }
    \in
    \mathcal{A}_\lambda.
  \end{equation}

  Finally,
  $\mathcal{A}_\lambda$
  is closed under
  $\oplustimes$.
  Consequently,
  every MinMax affine operator represented by a node presentation generated
  during the recursive exclusive scan or the subsequent inclusive step belongs
  to
  $\mathcal{A}_\lambda$.
\end{fact}

\begin{proof}
  Equation~\eqref{eq:minmax-parallel-scan-source-values}
  follows from the definitions of
  $\cgnode{X}[0]$,
  $\cgnode{R}[t]$,
  and
  $\cgnode{s}[t]$.

  By
  Equation~\eqref{eq:minmax-parallel-scan-source-values},
  the values at
  $\lambda$
  of the entries of
  $\cgnode{\mathcal{V}}$
  are precisely the elements used to generate
  $\minmaxdatadomain[\lambda]$.
  Therefore,
  by the minimum- and maximum-reduction clauses of
  Definition~\ref{def:minmax-vector-presentation}
  and
  Fact~\ref{fact:minmax-vector-presentation-values},
  \begin{equation*}
    \cgval{v}[\minreductionsymb](\lambda)
    =
    \vmin[\lambda],
    \qquad
    \cgval{v}[\maxreductionsymb](\lambda)
    =
    \vmax[\lambda],
  \end{equation*}
  proving
  Equation~\eqref{eq:minmax-parallel-scan-reduction-values}.

  By the definitions of the components of
  $\cgnode{E}$
  and
  $\cgnode{e}$,
  together with the copy clauses of
  Definition~\ref{def:minmax-vector-presentation}
  and
  Fact~\ref{fact:minmax-vector-presentation-values},
  Equation~\eqref{eq:minmax-parallel-scan-reduction-values}
  gives
  Equation~\eqref{eq:minmax-parallel-scan-identity-values}.
  The same copy-value property applied to
  \begin{equation*}
    \cgnode{M}[\id]
    \defeq
    \ptuple{
      \cgnode{E},
      \cgnode{e}
    }
  \end{equation*}
  gives
  \begin{equation*}
    \cgval{M}[\id](\lambda)
    =
    \ptuple{
      E(\lambda),
      e(\lambda)
    }
    =
    M_{\id}(\lambda).
  \end{equation*}
  Proposition~\ref{prop:minmax-data-generated-identity}
  then implies that
  $M_{\id}(\lambda)$
  is the identity element under
  $\oplustimes$
  on
  $\mathcal{A}_\lambda$.

  For every
  $t\in\iival{1}{T}$,
  the copy-value property applied to
  \begin{equation*}
    \cgnode{M}[t]
    \defeq
    \ptuple{
      \cgnode{R}[t][\mathrm{out}],
      \cgnode{s}[t][\mathrm{out}]
    }
  \end{equation*}
  and
  Equation~\eqref{eq:minmax-parallel-scan-source-values}
  gives
  \begin{equation*}
    \cgval{M}[t](\lambda)
    =
    \ptuple{
      R_t(\lambda),
      s_t(\lambda)
    }.
  \end{equation*}
  By definition of
  $\minmaxdatadomain[\lambda]$,
  every entry of
  $R_t(\lambda)$
  and
  $s_t(\lambda)$
  belongs to
  $\minmaxdatadomain[\lambda]$.
  Hence
  Equation~\eqref{eq:minmax-parallel-scan-source-M-values}
  follows.

  Finally,
  Proposition~\ref{prop:minmax-data-generated-closure}
  implies that
  $\mathcal{A}_\lambda$
  is closed under
  $\oplustimes$.
  Since
  $M_{\id}(\lambda)$
  and every
  $M_t(\lambda)$
  belong to
  $\mathcal{A}_\lambda$,
  every MinMax affine operator obtained from them by the
  $\oplustimes$
  operations appearing in the recursive exclusive scan and in the inclusive
  step also belongs to
  $\mathcal{A}_\lambda$.
\end{proof}

\begin{proposition}[Correctness of the parallel-scan computation graph]
  \label{prop:minmax-parallel-scan-graph-correctness}
  Assume the context of
  Definition~\ref{def:minmax-parallel-scan-graph}.

  For every
  $\lambda\in\Lambda$
  and
  $t\in\iival{1}{T}$,
  let
  \begin{equation*}
    M_t(\lambda)
    \defeq
    \ptuple{
      R_t(\lambda),
      s_t(\lambda)
    },
  \end{equation*}
  and let
  \begin{equation*}
    M_{1:t}(\lambda)
    \defeq
    M_1(\lambda)
    \oplustimes
    \cdots
    \oplustimes
    M_t(\lambda).
  \end{equation*}

  Then the exclusive-scan presentation
  $\cgnode{P}[t]$
  satisfies
  \begin{equation}
    \label{eq:minmax-parallel-scan-exclusive-correctness}
    \cgval{P}[t](\lambda)
    =
    \begin{cases}
      M_{\id}(\lambda),
      &t=1,
      \\[.3em]
      M_{1:t-1}(\lambda),
      &t>1.
    \end{cases}
  \end{equation}

  Consequently,
  the inclusive presentations introduced by
  $\mathrm{scan}_t$
  satisfy
  \begin{equation}
    \label{eq:minmax-parallel-scan-inclusive-correctness}
    \ptuple{
      \cgval{A}[1:t](\lambda),
      \cgval{b}[1:t](\lambda)
    }
    =
    M_{1:t}(\lambda).
  \end{equation}

  Writing
  \begin{equation*}
    M_{1:t}(\lambda)
    =
    \ptuple{
      A_{1:t}(\lambda),
      b_{1:t}(\lambda)
    },
  \end{equation*}
  we furthermore have
  \begin{equation}
    \label{eq:minmax-parallel-scan-state-correctness}
    \cgval{x}[t](\lambda)
    =
    \big(
      A_{1:t}(\lambda)
      \otimes
      x_0
    \big)
    \oplus
    b_{1:t}(\lambda)
    =
    x_t(\lambda).
  \end{equation}

  In particular,
  for every choice of source graphs as in
  Definition~\ref{def:minmax-parallel-scan-graph},
  the graph
  \begin{equation*}
      \minmaxscangraph{
        \cgnode{X}[0]
      }{
        \seq{\cgnode{R}}{1}{T}
      }{
        \seq{\cgnode{s}}{1}{T}
      }
  \end{equation*}
  represents $\cgval{x}_T$.
\end{proposition}

\begin{proof}
  Fix
  $\lambda\in\Lambda$
  and let
  \begin{equation*}
    \mathcal A_\lambda
    \defeq
    \minmaxdatadomain[\lambda]^{n\times n}
    \times
    \minmaxdatadomain[\lambda]^n.
  \end{equation*}

  By
  Fact~\ref{fact:minmax-parallel-scan-source-and-closure},
  \begin{equation}
    \label{eq:minmax-parallel-scan-correctness-source-M-values}
    \cgval{M}[t](\lambda)
    =
    M_t(\lambda)
    \in
    \mathcal A_\lambda
    \qquad
    \forall t\in\iival{1}{T},
  \end{equation}
  $M_{\id}(\lambda)$
  is the identity element under
  $\oplustimes$
  on
  $\mathcal A_\lambda$,
  and
  $\mathcal A_\lambda$
  is closed under
  $\oplustimes$.

  We first establish correctness of the recursive exclusive scan.
  More generally,
  consider an invocation
  \begin{equation*}
    \mathrm{xscan}_m
    \big(
      \cgnode{M}[\id];
      \cgnode{N}[1],\ldots,\cgnode{N}[m]
      \bigm|
      H
    \big),
  \end{equation*}
  where
  \begin{equation*}
    \cgval{N}[1](\lambda)
    =
    N_1,
    \ldots,
    \cgval{N}[m](\lambda)
    =
    N_m
    \in
    \mathcal A_\lambda,
  \end{equation*}
  and let
  $\cgnode{P}[1],\ldots,\cgnode{P}[m]$
  be its exclusive-scan presentations.
  We claim that
  \begin{equation}
    \label{eq:minmax-parallel-scan-general-exclusive}
    \cgval{P}[r](\lambda)
    =
    \begin{cases}
      M_{\id}(\lambda),
      &r=1,
      \\[.3em]
      N_1
      \oplustimes
      \cdots
      \oplustimes
      N_{r-1},
      &r>1,
    \end{cases}
    \qquad
    \forall r\in\iival{1}{m}.
  \end{equation}

  We prove the claim by induction on
  $m$.
  If
  $m=1$,
  the base-case extension gives
  \begin{equation*}
    \cgnode{P}[1]
    \defeq
    \cgnode{M}[\id].
  \end{equation*}
  Therefore,
  by
  Fact~\ref{fact:minmax-vector-presentation-values},
  \begin{equation*}
    \cgval{P}[1](\lambda)
    =
    M_{\id}(\lambda).
  \end{equation*}

  Suppose now that
  $m>1$
  and that the claim holds for every smaller scan length.
  Let
  \begin{equation*}
    \tilde m
    \defeq
    \left\lceil
      \frac{m}{2}
    \right\rceil.
  \end{equation*}

  For every
  $j\in\iival{1}{\lfloor m/2\rfloor}$,
  the contraction step introduces
  \begin{equation*}
    \cgnode{C}[j]
    \defeq
    \cgnode{N}[2j-1]
    \oplustimes
    \cgnode{N}[2j].
  \end{equation*}
  By
  Fact~\ref{fact:minmax-vector-presentation-values},
  it represents
  \begin{equation}
    \label{eq:minmax-parallel-scan-paired-values}
    C_j
    \defeq
    N_{2j-1}
    \oplustimes
    N_{2j}.
  \end{equation}
  If
  $m$
  is odd,
  then the contraction step additionally introduces
  \begin{equation*}
    \cgnode{C}[\tilde m]
    \defeq
    \cgnode{N}[m],
  \end{equation*}
  and therefore
  \begin{equation*}
    C_{\tilde m}
    =
    N_m.
  \end{equation*}

  By closure of
  $\mathcal A_\lambda$
  under
  $\oplustimes$,
  every
  $C_j$
  belongs to
  $\mathcal A_\lambda$.
  Since
  $\tilde m<m$,
  the induction hypothesis applies to the recursive invocation
  \begin{equation*}
    \mathrm{xscan}_{\tilde m}
    \big(
      \cgnode{M}[\id];
      \cgnode{C}[1],\ldots,\cgnode{C}[\tilde m]
      \bigm|
      G_m^{\mathrm c}
    \big).
  \end{equation*}
  Its exclusive-scan presentations
  $\cgnode{Q}[1],\ldots,\cgnode{Q}[\tilde m]$
  therefore satisfy
  \begin{equation}
    \label{eq:minmax-parallel-scan-Q-values}
    \cgval{Q}[j](\lambda)
    =
    \begin{cases}
      M_{\id}(\lambda),
      &j=1,
      \\[.3em]
      C_1
      \oplustimes
      \cdots
      \oplustimes
      C_{j-1},
      &j>1.
    \end{cases}
  \end{equation}

  By
  Equation~\eqref{eq:minmax-parallel-scan-paired-values}
  and associativity of
  $\oplustimes$,
  for every
  $j>1$,
  \begin{equation}
    \label{eq:minmax-parallel-scan-Q-expanded}
    \cgval{Q}[j](\lambda)
    =
    N_1
    \oplustimes
    \cdots
    \oplustimes
    N_{2j-2}.
  \end{equation}

  For every
  $j\in\iival{1}{\lfloor m/2\rfloor}$,
  the expansion step introduces
  \begin{equation*}
    \cgnode{P}[2j-1]
    \defeq
    \cgnode{Q}[j].
  \end{equation*}
  Hence,
  by the copy-value property,
  if
  $j=1$,
  \begin{equation*}
    \cgval{P}[1](\lambda)
    =
    M_{\id}(\lambda),
  \end{equation*}
  while,
  for
  $j>1$,
  \begin{equation*}
    \cgval{P}[2j-1](\lambda)
    =
    N_1
    \oplustimes
    \cdots
    \oplustimes
    N_{2j-2}.
  \end{equation*}

  Furthermore,
  \begin{equation*}
    \cgnode{P}[2j]
    \defeq
    \cgnode{Q}[j]
    \oplustimes
    \cgnode{N}[2j-1].
  \end{equation*}
  Therefore,
  by
  Fact~\ref{fact:minmax-vector-presentation-values},
  Equation~\eqref{eq:minmax-parallel-scan-Q-expanded}
  for
  $j>1$,
  and the identity property for
  $j=1$,
  \begin{equation*}
    \cgval{P}[2j](\lambda)
    =
    N_1
    \oplustimes
    \cdots
    \oplustimes
    N_{2j-1}.
  \end{equation*}

  Finally,
  if
  $m$
  is odd,
  then
  \begin{equation*}
    m
    =
    2\tilde m-1,
    \qquad
    \cgnode{P}[m]
    \defeq
    \cgnode{Q}[\tilde m].
  \end{equation*}
  Hence,
  by the copy-value property and
  Equation~\eqref{eq:minmax-parallel-scan-Q-expanded},
  \begin{equation*}
    \cgval{P}[m](\lambda)
    =
    N_1
    \oplustimes
    \cdots
    \oplustimes
    N_{m-1}.
  \end{equation*}

  This proves
  Equation~\eqref{eq:minmax-parallel-scan-general-exclusive}.

  Applying the general claim to
  \begin{equation*}
    \cgnode{N}[t]
    =
    \cgnode{M}[t],
    \qquad
    N_t
    =
    M_t(\lambda),
    \qquad
    t\in\iival{1}{T},
  \end{equation*}
  gives
  Equation~\eqref{eq:minmax-parallel-scan-exclusive-correctness}.

  By the definition of the inclusive extension,
  \begin{equation*}
    \ptuple{
      \cgnode{A}[1:t],
      \cgnode{b}[1:t]
    }
    \defeq
    \cgnode{P}[t]
    \oplustimes
    \cgnode{M}[t].
  \end{equation*}
  Hence,
  by
  Fact~\ref{fact:minmax-vector-presentation-values},
  for
  $t=1$,
  \begin{align*}
    \ptuple{
      \cgval{A}[1:1](\lambda),
      \cgval{b}[1:1](\lambda)
    }
    &=
    M_{\id}(\lambda)
    \oplustimes
    M_1(\lambda)
    \\
    &=
    M_1(\lambda),
  \end{align*}
  while, for
  $t>1$,
  \begin{align*}
    \ptuple{
      \cgval{A}[1:t](\lambda),
      \cgval{b}[1:t](\lambda)
    }
    &=
    M_{1:t-1}(\lambda)
    \oplustimes
    M_t(\lambda)
    \\
    &=
    M_{1:t}(\lambda).
  \end{align*}
  This proves
  Equation~\eqref{eq:minmax-parallel-scan-inclusive-correctness}.

  Write
  \begin{equation*}
    M_{1:t}(\lambda)
    =
    \ptuple{
      A_{1:t}(\lambda),
      b_{1:t}(\lambda)
    }.
  \end{equation*}
  By
  Definition~\ref{def:minmax-affine-operator}
  and associativity of
  $\oplustimes$,
  $M_{1:t}(\lambda)$
  represents the composition,
  in temporal order,
  of the MinMax affine maps
  \begin{equation*}
    x
    \longmapsto
    \big(
      R_\tau(\lambda)
      \otimes
      x
    \big)
    \oplus
    s_\tau(\lambda),
    \qquad
    \tau\in\iival{1}{t}.
  \end{equation*}
  Applying this composed operator to
  $x_0$
  yields
  \begin{equation*}
    x_t(\lambda)
    =
    \big(
      A_{1:t}(\lambda)
      \otimes
      x_0
    \big)
    \oplus
    b_{1:t}(\lambda).
  \end{equation*}

  By the definition of
  $\cgnode{x}[t]$,
  Equation~\eqref{eq:minmax-parallel-scan-source-values},
  Equation~\eqref{eq:minmax-parallel-scan-inclusive-correctness},
  and
  Fact~\ref{fact:minmax-vector-presentation-values},
  the right-hand side equals
  $\cgval{x}[t](\lambda)$.
  This proves
  Equation~\eqref{eq:minmax-parallel-scan-state-correctness}.

  In particular,
  the final output presentation
  $\cgnode{x}[T]$
  of
  \begin{equation*}
    \minmaxscanvgraph{
      \cgnode{X}[0]
    }{
      \seq{\cgnode{R}}{1}{T}
    }{
      \seq{\cgnode{s}}{1}{T}
    }
  \end{equation*}
  represents
  $x_T$.
  By
  Proposition~\ref{prop:minmax-virtual-extension-materialisation},
  materialisation preserves its value function, and hence the value of the materialised graph
  \begin{equation*}
      \minmaxscangraph{
        \cgnode{X}[0]
      }{
        \seq{\cgnode{R}}{1}{T}
      }{
        \seq{\cgnode{s}}{1}{T}
      }
  \end{equation*}
  is $x_T$.
\end{proof}

\begin{proposition}[Parallel-scan graphs are MinMax-algebraic]
  \label{prop:minmax-parallel-scan-graphs-algebraic}
  Assume the context of
  Definition~\ref{def:minmax-parallel-scan-graph}.

  For every choice of
  \begin{equation*}
    \cgnode{X}[0],
    \qquad
    \seq{\cgnode{R}}{1}{T},
    \qquad
    \seq{\cgnode{s}}{1}{T}
  \end{equation*}
  admitted by that definition,
  \begin{equation}
    \label{eq:minmax-parallel-scan-virtual-extension-class}
    \minmaxscanvgraph{
      \cgnode{X}[0]
    }{
      \seq{\cgnode{R}}{1}{T}
    }{
      \seq{\cgnode{s}}{1}{T}
    }
    \in
    \minmaxextensionclass^{\mathrm{v}}[
      \cgnode{X}[0],
      \cgnode{R}[1],\ldots,\cgnode{R}[T],
      \cgnode{s}[1],\ldots,\cgnode{s}[T]
    ],
  \end{equation}
  and
  \begin{equation}
    \label{eq:minmax-parallel-scan-extension-class}
    \minmaxscangraph{
      \cgnode{X}[0]
    }{
      \seq{\cgnode{R}}{1}{T}
    }{
      \seq{\cgnode{s}}{1}{T}
    }
    \in
    \minmaxextensionclass[
      \cgnode{X}[0],
      \cgnode{R}[1],\ldots,\cgnode{R}[T],
      \cgnode{s}[1],\ldots,\cgnode{s}[T]
    ].
  \end{equation}

  Consequently,
  \begin{equation}
    \label{eq:minmax-parallel-scan-recurrence-class}
    \minmaxscangraphclass{
      x_0
    }{
      \seq{R}{1}{T}
    }{
      \seq{s}{1}{T}
    }
    \subseteq
    \minmaxrecgraphclass
    \big(
      x_0;
      \seq{R}{1}{T},
      \seq{s}{1}{T}
    \big).
  \end{equation}
\end{proposition}

\begin{proof}
  By construction,
  \begin{equation*}
    \minmaxscanvgraph{
      \cgnode{X}[0]
    }{
      \seq{\cgnode{R}}{1}{T}
    }{
      \seq{\cgnode{s}}{1}{T}
    }
  \end{equation*}
  is a virtual-node extension of the common-input union
  $G^\sqcup$
  of
  \begin{equation*}
    \cgnode{X}[0],
    \qquad
    \cgnode{R}[1],\ldots,\cgnode{R}[T],
    \qquad
    \cgnode{s}[1],\ldots,\cgnode{s}[T].
  \end{equation*}

  The virtual nodes
  $\cgnode{v}[\minreductionsymb]$
  and
  $\cgnode{v}[\maxreductionsymb]$
  are labelled by
  \begin{equation*}
    \minreduction{|\cgnode{\mathcal V}|}
    \qquad\text{and}\qquad
    \maxreduction{|\cgnode{\mathcal V}|},
  \end{equation*}
  respectively.

  The scalar virtual nodes introduced by the node-matrix, node-vector, and
  tuple-copy presentations
  \begin{equation*}
    \cgnode{E},
    \qquad
    \cgnode{e},
    \qquad
    \cgnode{M}[\id],
    \qquad
    \cgnode{M}[1],\ldots,\cgnode{M}[T]
  \end{equation*}
  are identity-labelled virtual computation nodes, as prescribed by
  Definition~\ref{def:minmax-vector-presentation}.
  These labels are admitted by
  Definition~\ref{def:minmax-algebraic-virtual-extension}.

  Every occurrence of
  $\oplustimes$
  in the contraction, recursive expansion, and inclusive stages is expanded,
  by
  Definition~\ref{def:minmax-affine-operator},
  into MinMax matrix and vector expressions.
  By
  Definition~\ref{def:minmax-vector-presentation},
  every scalar virtual computation node introduced by these expressions is
  labelled by an identity, minimum, or maximum operation admitted by
  Definition~\ref{def:minmax-algebraic-virtual-extension}.
  The same holds for the final state presentations
  $\cgnode{x}[t]$,
  for
  $t\in\iival{1}{T}$.

  Therefore,
  \begin{equation*}
    \minmaxscanvgraph{
      \cgnode{X}[0]
    }{
      \seq{\cgnode{R}}{1}{T}
    }{
      \seq{\cgnode{s}}{1}{T}
    }
    \in
    \minmaxextensionclass^{\mathrm{v}}[
      \cgnode{X}[0],
      \cgnode{R}[1],\ldots,\cgnode{R}[T],
      \cgnode{s}[1],\ldots,\cgnode{s}[T]
    ],
  \end{equation*}
  proving
  Equation~\eqref{eq:minmax-parallel-scan-virtual-extension-class}.

  By
  Proposition~\ref{prop:minmax-virtual-extension-materialisation},
  every materialisation of this graph belongs to the corresponding ordinary
  MinMax-algebraic extension class.
  In particular,
  \begin{equation*}
    \minmaxscangraph{
      \cgnode{X}[0]
    }{
      \seq{\cgnode{R}}{1}{T}
    }{
      \seq{\cgnode{s}}{1}{T}
    }
    \in
    \minmaxextensionclass[
      \cgnode{X}[0],
      \cgnode{R}[1],\ldots,\cgnode{R}[T],
      \cgnode{s}[1],\ldots,\cgnode{s}[T]
    ],
  \end{equation*}
  proving
  Equation~\eqref{eq:minmax-parallel-scan-extension-class}.

  Proposition~\ref{prop:minmax-parallel-scan-graph-correctness}
  gives
  that the graph
  \begin{equation*}
      \minmaxscangraph{
        \cgnode{X}[0]
      }{
        \seq{\cgnode{R}}{1}{T}
      }{
        \seq{\cgnode{s}}{1}{T}
      }
  \end{equation*}
  represents $\cgval{x}_T$.
  Therefore,
  by
  Definition~\ref{def:minmax-algebraic-recurrence-graphs},
  every such materialised parallel-scan graph belongs to
  \begin{equation*}
    \minmaxrecgraphclass
    \big(
      x_0;
      \seq{R}{1}{T},
      \seq{s}{1}{T}
    \big).
  \end{equation*}

  Since the source graphs were arbitrary members of the graph classes
  specified in
  Definition~\ref{def:minmax-parallel-scan-graph},
  Equation~\eqref{eq:minmax-parallel-scan-recurrence-class}
  follows.
\end{proof}

\subsection{BIBSD: Uniform Selected uBIBSD for MinMax RNCs}
\label{subsec:minmax-graph-uniform-ubibsd}

\todo{}

\todo{}

We now lift the topology-independent AMM-pseudo-derivative bound of
Theorem~\ref{theorem:minmax-algebraic-rnc-state-graph-bound}
from individual MinMax recurrence graphs to complete MinMax RNC dynamics.

The required uniform state-value bound is already provided by
Corollary~\ref{cor:minmax-rnc-ubibs}.
Thus, no additional stability argument is needed in this section:
the only remaining task is to propagate bounded values and bounded selected
derivatives through the finitely many analytic maps occurring
between recurrent layers.

\begin{context}
  For this section,
  we consider arbitrary but fixed
  parametric MinMax RNC dynamics
  \begin{equation*}
    D \in \rncdynop\big(\classparneuralunits{MinMax}\big),
    \qquad\quad
    D = \ctuple{X,U,f}{\Theta},
  \end{equation*}
  and
  we write
  $\customIndexedcompgraphfamily{\mathcal{G}_{\scriptscriptstyle\adaoplustimes}}$
  for the corresponding state computation-graph family of $D$
  as per Definition~\ref{def:minmax-algebraic-rnc-state-graphs}.
  We note that the class
  $\customIndexedcompgraphfamily{\mathcal{G}_{\scriptscriptstyle\adaoplustimes}}$
  is an AMM-computation-graph family, 
  by Proposition~\ref{prop:minmax-rnc-state-graphs-analytic-minmax}.
\end{context}

\begin{closingremark}[Graph choices in the MinMax BIBSD result]
  \label{rem:minmax-bibsd-graph-choices}
  The BIBSD result below is stated for the general state computation-graph class
  $\customIndexedcompgraphfamily{\mathcal{G}_{\scriptscriptstyle\adaoplustimes}}$, 
  which provides great flexibility in
  choosing the subgraphs of a state computation graph---as discussed in
  Remark~\ref{rem:minmax-rnc-state-graph-construction-choices}.
  The bound below is uniform over all such admissible choices and
  does not depend on selecting a particular recurrence-evaluation topology.

  The sequential and parallel-scan constructions yield specific graphs included in 
  $\customIndexedcompgraphfamily{\mathcal{G}_{\scriptscriptstyle\adaoplustimes}}$.
  For such graphs, once the corresponding source graphs are fixed, the recurrence
  computation is determined as described in
  Remarks~\ref{rem:minmax-sequential-recurrence-graph-determinacy}
  and~\ref{rem:minmax-parallel-scan-graph-determinacy},
  respectively.
\end{closingremark}

\begin{theorem}[Graph-uniform selected uBIBSD of parametric MinMax RNC dynamics]
  \label{thm:minmax-rnc-ubibs-selected-ubibsd}
  Parametric MinMax RNC dynamics~$D$ satisfy
  $\customIndexedcompgraphfamily{\mathcal{G}_{\scriptscriptstyle\adaoplustimes}}
  \text{-}\minmaxsdv\text{--uBIBSD}$.
  Specifically,
  for every
  $r\in\nonnegreals$,
  there exists
  $B_r\in\nonnegreals$
  such that
  \begin{equation*}
    \big\|
      \minmaxsdvop{G}(\theta)
    \big\|_\mathrm{row}
    \leq
    B_r
    \qquad
    \forall
    u       \in \infseqs{\truncation{U}{r}},
    \,
    \forall
    \initialstate{x}     \in \truncation{X}{r},
    \,
    \forall
    T       \in \timedomain,
    \,
    \forall
    \theta  \in \truncation{\Theta}{r},
    \,
    \forall
    G       \in \minmaxstategraphclass{D}[T][u][\initialstate{x}]\,.
  \end{equation*}
\end{theorem}
\begin{closingremark}
  \label{remark:minmax-rnc-ubibs-selected-ubibsd}
  The bounding constant $B_r$ is independent 
  of 
  $u$,
  $\initialstate{x}$,
  $T$,
  and of the graph $G$---in particular, it is independent of its 
  MinMax-algebraic recurrence-evaluation topology and the chosen constituent computation-graphs.
\end{closingremark}
\begin{proof}
  Let $r\in\nonnegreals$ be arbitrary.
  By
  Corollary~\ref{cor:minmax-rnc-ubibs},
  $D$
  satisfies uBIBS.
  Hence there exists
  $B^X_r\in\nonnegreals$
  such that, 
  for all
  $u\in\infseqs{\truncation{U}{r}}$,
  $\initialstate{x}\in\truncation{X}{r}$,
  $\theta\in\truncation{\Theta}{r}$,
  the corresponding state sequence $(x_t)_{t \in \timedomain}$ satisfies
  \begin{equation}
    \label{eq:minmax-gradient-rnc-state-value-bound}
    \sup_{t\in\timedomain}
    \|x_t\|
    \leq
    B^X_r.
  \end{equation}

  Since the RNC has finitely many layers and every layer-state space is
  finite-dimensional, after enlarging
  $B^X_r$
  by a fixed norm-equivalence factor if necessary, we use the same symbol for
  a uniform bound on every layer-state block.

  Let us fix an abitrary
  input sequence $u\in\infseqs{\truncation{U}{r}}$,
  initial state $\initialstate{x}\in\truncation{X}{r}$,
  horizon $T\in\timedomain$,
  and 
  computation graph $G \in \minmaxstategraphclass{D}[T][u][\initialstate{x}]$.

  By
  Proposition~\ref{prop:minmax-rnc-state-graphs-analytic-minmax},
  $G$
  is an AMM-graph.
  Hence
  Proposition~\ref{prop:minmax-pseudoderivative-intensionally-admissible}
  implies that
  $\minmaxsdv$
  is intensionally-admissible for
  $G$.
  Since
  $G$
  was arbitrary,
  $\minmaxsdv$
  is intensionally-admissible for the complete family
  $\customIndexedcompgraphfamily{\mathcal{G}_{\scriptscriptstyle\adaoplustimes}}$.

  If
  $T=0$,
  then
  $G = \constpargraph{\initialstate{x}}{\cgnode{\theta}}$, 
  and therefore
  $\minmaxsdvop{G}(\theta) = 0$.
  Thus the claim is immediate.

  Hence, suppose now that $T\in\posnaturals$.
  Fix arbitrary constituent computation graphs
  \begin{equation*}
    \cgnode{U}_t^\ell,
    \quad
    \cgnode{R}_t^{\ell,k},
    \quad
    \cgnode{s}_t^{\ell,k},
    \quad
    \cgnode{H}_t^{\ell,k},
    \quad
    \cgnode{X}_t^{\ell,k},
    \quad
    \cgnode{X}_t^\ell
  \end{equation*}
  satisfying
  Definition~\ref{def:minmax-algebraic-rnc-state-graphs} for $G$.

  We prove inductively over
  $\ell\in\iival{1}{L}$
  that there exist constants
  \begin{equation*}
    B^U_\ell,
    \qquad
    B^{U'}_\ell,
    \qquad
    M_\ell
    \in
    \nonnegreals
  \end{equation*}
  depending only on
  $r$
  and the fixed RNC architecture such that the following bounds hold uniformly
  over
  $u$,
  $\initialstate{x}$,
  $T$,
  $\theta\in\truncation{\Theta}{r}$,
  $G$,
  the constituent graph choices,
  and the relevant time indices:
  \begin{gather}
    \label{eq:minmax-layer-input-value-derivative-bound}
    \begin{aligned}
      \big\|
      \cgval{U}_t^\ell(\theta)
      \big\|
      \leq
      B^U_\ell,
      \qquad\qquad
      \big\|
      \minmaxsdvop{\cgnode{U}_t^\ell}(\theta)
      \big\|_\mathrm{row}
      \leq
      B^{U'}_\ell,
    \end{aligned}
    \\[1em]
    \label{eq:minmax-prefix-source-derivative-bound}
    \max
    \left\{
      \big\|
        \minmaxsdvop{\cgnode{R}_t^{j,k}}(\theta)
      \big\|_\mathrm{row},
      \;
      \big\|
        \minmaxsdvop{\cgnode{s}_t^{j,k}}(\theta)
      \big\|_\mathrm{row}
    \right\}
    \leq
    M_\ell
    \qquad
    \forall j\in\iival{1}{\ell}.
  \end{gather}

  For the first layer,
  Definition~\ref{def:minmax-algebraic-rnc-state-graphs}
  gives
  \begin{equation*}
    \cgnode{U}_t^1
    =
    \constpargraph{u_t}{\cgnode{\theta}}.
  \end{equation*}
  Therefore
  \begin{equation*}
    \big\|
      \cgval{U}_t^1(\theta)
    \big\|
    =
    \|u_t\|
    \leq
    r,
    \qquad
    \minmaxsdvop{\cgnode{U}_t^1}(\theta)
    =
    0.
  \end{equation*}
  Hence we may take
  \begin{equation*}
    B^U_1
    =
    r,
    \qquad
    B^{U'}_1
    =
    0.
  \end{equation*}

  The identity parameter graph satisfies
  \begin{equation*}
    \cgval{I}[][][\cgnode{\theta}](\theta)
    =
    \theta,
    \qquad
    \minmaxsdvop{\mathring{I}[\cgnode{\theta}]}(\theta)
    =
    I_{d_\Theta}.
  \end{equation*}
  Consequently,
  on
  $\truncation{\Theta}{r}$,
  both its value and its selected derivative are uniformly bounded.

  Let now
  $\ell\in\iival{1}{L}$
  and suppose that
  Equation~\eqref{eq:minmax-layer-input-value-derivative-bound}
  holds at layer
  $\ell$,
  and, when
  $\ell>1$,
  that
  Equation~\eqref{eq:minmax-prefix-source-derivative-bound}
  holds through layer
  $\ell-1$.

  Fix
  $k\in\iival{1}{n_\ell}$.
  By
  Definition~\ref{def:minmax-algebraic-rnc-state-graphs},
  \begin{equation*}
    \cgnode{H}_t^{\ell,k}
    =
    \ptuple{
      \cgnode{U}_t^\ell,
      \mathring{I}[\cgnode{\theta}]
    }
    \analyticextension
    \rho_{\ell,k},
  \end{equation*}
  where
  $\rho_{\ell,k}$
  is analytic.

  The families of source graphs
  \begin{equation*}
    \big\{
      \cgnode{U}_t^\ell
    \big\},
    \qquad
    \big\{
      \mathring{I}[\cgnode{\theta}]
    \big\}
  \end{equation*}
  have uniformly bounded values and uniformly bounded selected derivatives.
  Therefore,
  by
  Lemma~\ref{lem:graph-uniform-selected-feedforward-extension-bounds}, \todo{}
  the graphs
  $\cgnode{H}_t^{\ell,k}$
  have uniformly bounded values and uniformly bounded
  AMM-pseudo-derivatives.

  The reset and set graphs are output restrictions:
  \begin{equation*}
    \cgnode{R}_t^{\ell,k}
    =
    \big(
      \cgnode{H}_t^{\ell,k}
    \big)_{J_R^{\ell,k}},
    \qquad
    \cgnode{s}_t^{\ell,k}
    =
    \big(
      \cgnode{H}_t^{\ell,k}
    \big)_{J_s^{\ell,k}}.
  \end{equation*}
  Output restriction only removes rows of the selected Jacobian and therefore
  cannot increase the row norm.
  Since layer
  $\ell$
  contains finitely many units,
  there exists
  $\hat M_\ell\in\nonnegreals$
  such that
  \begin{equation}
    \label{eq:minmax-current-layer-source-derivative-bound}
    \max
    \left\{
      \big\|
        \minmaxsdvop{\cgnode{R}_t^{\ell,k}}(\theta)
      \big\|_\mathrm{row},
      \;
      \big\|
        \minmaxsdvop{\cgnode{s}_t^{\ell,k}}(\theta)
      \big\|_\mathrm{row}
    \right\}
    \leq
    \hat M_\ell
  \end{equation}
  uniformly over
  $k$,
  $t$,
  and all admissible choices.

  Define
  \begin{equation*}
    M_\ell
    \defeq
    \begin{cases}
      \hat M_1,
      &\ell=1,
      \\[.2em]
      \max
      \left\{
        M_{\ell-1},
        \hat M_\ell
      \right\},
      &\ell>1.
    \end{cases}
  \end{equation*}
  Then
  Equation~\eqref{eq:minmax-prefix-source-derivative-bound}
  holds through layer
  $\ell$.

  For every
  $t\in\iival{1}{T}$
  and every
  $k\in\iival{1}{n_\ell}$,
  Definition~\ref{def:minmax-algebraic-rnc-state-graphs}
  gives
  \begin{equation*}
    \cgnode{X}_t^{\ell,k}
    \in
    \minmaxrecgraphclass
    \big(
      \initialstate{x}^{\ell,k};
      \seq{\cgnode{R}^{\ell,k}}{1}{t},
      \seq{\cgnode{s}^{\ell,k}}{1}{t}
    \big).
  \end{equation*}
  Therefore,
  Theorem~\ref{theorem:minmax-algebraic-recurrence-graph-bound}
  and
  Equation~\eqref{eq:minmax-current-layer-source-derivative-bound}
  yield
  \begin{equation}
    \label{eq:minmax-layer-unit-state-derivative-bound}
    \big\|
      \minmaxsdvop{\cgnode{X}_t^{\ell,k}}(\theta)
    \big\|_\mathrm{row}
    \leq
    \hat M_\ell
    \leq
    M_\ell.
  \end{equation}

  Since
  \begin{equation*}
    \cgnode{X}_t^\ell
    =
    \cgnode{X}_t^{\ell,1}
    \sqcup_{\mathrm{in}}
    \cdots
    \sqcup_{\mathrm{in}}
    \cgnode{X}_t^{\ell,n_\ell},
  \end{equation*}
  common-input union concatenates the corresponding output rows.
  Consequently,
  \begin{equation}
    \label{eq:minmax-layer-state-derivative-bound}
    \big\|
      \minmaxsdvop{\cgnode{X}_t^\ell}(\theta)
    \big\|_\mathrm{row}
    \leq
    M_\ell.
  \end{equation}

  If
  $\ell=L$,
  the layer induction is complete.
  Suppose therefore that
  $\ell<L$.

  By
  Equation~\eqref{eq:minmax-gradient-rnc-state-value-bound},
  the graphs
  $\cgnode{X}_{t-1}^\ell$
  and
  $\cgnode{X}_t^\ell$
  represent uniformly bounded values.
  Their selected derivatives are uniformly bounded by
  Equation~\eqref{eq:minmax-layer-state-derivative-bound},
  with the time-zero state having selected derivative zero.
  Together with
  Equation~\eqref{eq:minmax-layer-input-value-derivative-bound}
  and the identity-parameter graph bounds,
  the source graphs
  \begin{equation*}
    \cgnode{X}_{t-1}^\ell,
    \quad
    \cgnode{U}_t^\ell,
    \quad
    \cgnode{X}_t^\ell,
    \quad
    \mathring{I}[\cgnode{\theta}]
  \end{equation*}
  therefore have uniformly bounded values and selected derivatives.

  The next-layer input graph is
  \begin{equation*}
    \cgnode{U}_t^{\ell+1}
    =
    \bigptuple{
      \cgnode{X}_{t-1}^\ell,
      \cgnode{U}_t^\ell,
      \cgnode{X}_t^\ell,
      \mathring{I}[\cgnode{\theta}]
    }
    \analyticextension
    g_\ell,
  \end{equation*}
  where
  $g_\ell$
  is analytic.
  Lemma~\ref{lem:graph-uniform-selected-feedforward-extension-bounds} \todo{}
  therefore yields constants
  \begin{equation*}
    B^U_{\ell+1},
    \qquad
    B^{U'}_{\ell+1}
    \in
    \nonnegreals
  \end{equation*}
  satisfying
  Equation~\eqref{eq:minmax-layer-input-value-derivative-bound}
  at layer
  $\ell+1$.
  This closes the induction.

  At the final layer,
  Equation~\eqref{eq:minmax-prefix-source-derivative-bound}
  gives a uniform bound
  $M_L$
  on every reset and set source graph occurring in the construction of
  $G$.
  Hence,
  by
  Theorem~\ref{theorem:minmax-algebraic-rnc-state-graph-bound},
  \begin{equation*}
    \big\|
      \minmaxsdvop{G}(\theta)
    \big\|_\mathrm{row}
    \leq
    M_L.
  \end{equation*}
  Thus, the claimed bound holds with $B_r \defeq M_L$.

  The bound is uniform over every graph in
  $\customIndexedcompgraphfamily{\mathcal{G}_{\scriptscriptstyle\adaoplustimes}}$.
  Finally, since all spaces are finite-dimensional, norm equivalence converts
  the row-norm bound into a bound in the norm appearing in
  Definition~\ref{def:graph-uniform-selected-ubibsd}.
  Therefore
  $D$
  satisfies
  $\customIndexedcompgraphfamily{\mathcal{G}_{\scriptscriptstyle\adaoplustimes}}$-$\minmaxsdv$--uBIBSD.
\end{proof}

\begin{corollary}[Selected uBIBSD for MinMax state-graph selectors]
  \label{cor:minmax-sequential-parallel-scan-ubibsd}
  Let
  $\selgraphsymb$
  be a graph selector with
  \begin{equation*}
    \selgraphsymb[
      \indexedStateToStateFuncfamily{D}[\initialstate{x}][u][t]
    ]
    \in
    \minmaxstategraphclass{D}[t][u][\initialstate{x}]
    \qquad\quad
    \forall u\in\infseqs{U},
    \;
    \forall \initialstate{x}\in X,
    \;
    \forall t\in\timedomain.
  \end{equation*}
  Then,
  the parametric MinMax RNC dynamics~$D$
  satisfy
  $\minmaxsdv[\selgraphsymb]\text{--uBIBSD}$.
  In particular,
  this applies to graph selectors over the subclass of 
  $\customIndexedcompgraphfamily{\mathcal{G}_{\scriptscriptstyle\adaoplustimes}}$
  where recurrent-unit evaluations use 
  sequential or parallel-scan of MinMax-algebraic recurrence graphs---including mixtures thereof.
\end{corollary}
\begin{proof}
  Theorem~\ref{thm:minmax-rnc-ubibs-selected-ubibsd}
  states that
  $D$
  satisfies
  $\customIndexedcompgraphfamily{\mathcal{G}_{\scriptscriptstyle\adaoplustimes}}
  \text{-}\minmaxsdv\text{--uBIBSD}$
  Therefore,
  Proposition~\ref{prop:graph-uniform-implies-selector-ubibsd}
  gives
  $\minmaxsdv[\selgraphsymb]$-uBIBSD,
  with a constant independent of the selector.

  Proposition~\ref{prop:minmax-sequential-graphs-algebraic}
  shows that sequential recurrence graphs belong to the corresponding
  MinMax-algebraic recurrence-graph classes, while
  Proposition~\ref{prop:minmax-parallel-scan-graphs-algebraic}
  gives the same conclusion for parallel-scan recurrence graphs.
  Hence selectors based on either construction, or on admissible mixtures of
  them, satisfy the hypothesis.
\end{proof}

\begin{closingremark}[Topology independence]
  Theorem~\ref{thm:minmax-rnc-ubibs-selected-ubibsd}
  is stronger than a comparison between particular sequential and parallel
  implementations.
  For fixed bounded input, initial-state, and parameter sets,
  the same constant applies to every graph in
  \begin{equation*}
    \minmaxstategraphclass{D}[t][u][\initialstate{x}],
  \end{equation*}
  uniformly over the time horizon.
  Thus the selected derivative may depend on the recurrence-evaluation graph
  at non-differentiable points, but its boundedness does not.
\end{closingremark}

\subsection{BIBLG: Bounded Loss Gradients of MinMax RNCs}
\label{subsec:minmax-bounded-loss-gradients}

We now propagate the graph-uniform state-derivative bound through the analytic
analytic output map and the admissible loss.
The value bound required in this propagation is supplied by uBIBS from
Corollary~\ref{cor:minmax-rnc-ubibs};
the derivative bound is supplied by
Theorem~\ref{thm:minmax-rnc-ubibs-selected-ubibsd}.

\begin{context}
  For this section, we fix an arbitrary parametric MinMax RNC
  \begin{equation*}
    S \in \rncop\big(\classparneuralunits{MinMax}\big),
    \quad
    S = \ctuple{X, U, f, \initialstate{x}, Y, h}{\Theta^f,\Theta^h}
    \;\;\text{with}\;\;
    (\Theta^f \!\times \Theta^h)
    \eqdef 
    \Theta 
    \subseteq 
    \reals^{d_f}
    \!\times
    \reals^{d_h}.
  \end{equation*}
\end{context}

\begin{closingremark}[Analytic functions of MinMax RNCs]
  We recall that the output function $h$ is analytic by the definition of MinMax RNCs, and the same
  is true for all functions of $S$ beyond the Min and Max operations used in its recurrence---e.g., 
  the set and reset functions of each unit are analytic.
\end{closingremark}

\begin{closingdefinition}[MinMax-algebraic objective computation graphs]
  \label{def:minmax-algebraic-objective-graphs}
  We define MinMax-algebraic objective computation graphs and then the resulting class.

  \begin{definitionpart}[light]
    Say we are given:
    \begin{itemize}
      \item
        an admissible loss function $\lossfn \in \admissiblelossfunctions[S]$,%
        \footnote{%
          See~Definition~\ref{def:sample-space-loss-function}.%
        }
        having analytic ambient extension $\ambientExtension{\lossfn}$;

      \item
        a sample $z = \ptuple{u,y} \in \samplespacep{S}$ for $S$,%
        \footnote{%
          See~Definition~\ref{def:sample-space-loss-function}.%
        }
        with $u = (u_t)_{t=1}^T$ for $T\in\timedomain$;

      \item
        the infinite extension $uu_T^\infseqOperatorSymb\in\infseqs{U}$ of the finite input sequence
        $u$ given by 
        \begin{equation*}
          uu_T^\infseqOperatorSymb 
          \;\defeq\; 
          u u_T, u_T, \ldots, u_T, \ldots 
          \;=\;
          u_1, u_2, \ldots, u_T, u_T, \ldots, u_T, \ldots 
        \end{equation*}
        namely, it is the input sequence obtained from $u$ by 
        repeating the last input element $u_T$ indefinitely after step $T$.
    \end{itemize}

    Let $\cgnode{\theta}^f$ and $\cgnode{\theta}^h$ be node-vectors of the form
    \begin{equation*}
      \cgnode{\theta}^f
      =
      \bigptuple{
      \cgnode{\theta}^f_1,
      \ldots,
      \cgnode{\theta}^f_{d_f}
      },
      \qquad\quad
      \cgnode{\theta}^h
      =
      \bigptuple{
      \cgnode{\theta}^h_1,
      \ldots,
      \cgnode{\theta}^h_{d_h}
      }.
    \end{equation*}
    All state graphs below have 
    $\cgnode{\theta} \defeq \bigptuple{\cgnode{\theta}^f, \cgnode{\theta}^h}$ 
    as their input node-vector.

    The
    \emph{MinMax-algebraic objective computation-graph class} for 
    given $S$, $z$ and $\lossfn$ is denoted by
    \begin{equation*}
      \minmaxobjectivegraphclass{S}[z][\lossfn]
    \end{equation*}
    and it consists of every computation graph $G$
    for which there exist computation graphs
    \begin{equation*}
      \cgnode{X}^-,
      \;
      \cgnode{X}^+,
      \;
      \cgnode{U}_T,
      \;
      \cgnode{P},
      \;
      \cgnode{H},
      \;
      \cgnode{Y},
    \end{equation*}
    called \emph{constituent graphs} of $G$, 
    such that:%
    \hspace{.1em}\footnote{Analytic extension of computation graphs are introduced in
    Definition~\ref{def:analytic-extension-selected-graph}.}$^\text{,}$%
    \footnote{Canonical constant computation graphs are introduced in
    Definition~\ref{def:parametric-canonical-constant-computation-graph}.}
    \begin{align*}
      \cgnode{X}^-
      & =
      \constpargraph{\initialstate{x}}{\cgnode{\theta}^f}
      & \text{when $T = 1$};
      \\
      \cgnode{X}^-
      & \in
      \minmaxstategraphclass{D}[T-1][uu_T^\infseqOperatorSymb][\initialstate{x}]
      & \text{when $T\geq 2$};
      \\[.7em]
      \cgnode{X}^+
      & \in
      \minmaxstategraphclass{D}[T][uu_T^\infseqOperatorSymb][\initialstate{x}];
      \\[.3em]
      \cgnode{U}_T 
      & =
      \constpargraph{u_T}{\cgnode{\theta}^f};
      \\[.3em]
      \cgnode{P}
      & = 
      \cgnode{X}^- \sqcup_{\mathrm{in}} \cgnode{U}_T \sqcup_{\mathrm{in}} \cgnode{X}^+;
      \\[.3em]
      \cgnode{H}
      & =
      \cgnode{P} \analyticextension h; 
      \\[.3em]
      \cgnode{Y}
      & = 
      \constpargraph{y}{\cgnode{\theta}};
      \\[.3em]
      G
      & =
      \ptuple{ \cgnode{H}, \cgnode{Y} } \analyticextension \ambientExtension{\lossfn}.
    \end{align*}
  \end{definitionpart}

  \begin{definitionpart}[title={Joint graph class},light]
    The 
    \emph{MinMax-algebraic objective computation-graph class} for the given MinMax RNC $S$ 
    is
    \begin{equation*}
      \minmaxobjectivegraphclass{S}
      \defeq
      \left(
        \minmaxobjectivegraphclass{S}[z][\lossfn]
      \right)_{
      \lossfn\in\admissiblelossfunctions[S],\,
      z\in\samplespacep{S}
      }\,.
    \end{equation*}
  \end{definitionpart}

  \begin{definitionpart}[title={Indexed graph family},light]
    The 
    \emph{MinMax-algebraic objective indexed computation-graph family} 
    for the given MinMax RNC $S$ 
    is
    \begin{equation*}
      \qquad\qquad
      \customIndexedcompgraphfamily{\mathcal{S}_{\scriptscriptstyle\adaoplustimes}}[%
        \Lossfn{S,z}[\lossfn]%
      ]
      \defeq\,
      \minmaxobjectivegraphclass{S}[z][\lossfn]
      \qquad\qquad
      \forall \lossfn\in\admissiblelossfunctions[S],
      \;
      \forall z\in\samplespacep{S}
      \,.
    \end{equation*}
  \end{definitionpart}
\end{closingdefinition}

\begin{closingremark}[Graph choices in objective-graph constructions]
  \label{rem:minmax-objective-graph-witnesses}
  Objective-graph constructions inherit the possible non-uniqueness of the
  state graphs from which they are built.
  In particular, by
  Remark~\ref{rem:minmax-rnc-state-graph-construction-choices},
  membership of a state--computation-graph class 
  $\indexedcompgraphfamily_{D;T,u,\initialstate{x}}$
  in the general class
  $\minmaxstategraphclass{D}[T][u][\initialstate{x}]$
  does not in general determine a unique choice of the constituent recurrence
  subgraphs of a graph
  \begin{equation*}
    G
    \in
    \indexedcompgraphfamily_{D;T,u,\initialstate{x}}
    \in
    \minmaxobjectivegraphclass{S}[z][\lossfn];
  \end{equation*}
  therefore, such graph $G$ need not be uniquely determined by its class
  $\indexedcompgraphfamily_{D;T,u,\initialstate{x}}$.

  The above does not affect the correctness result below
  (Theorem~\ref{thm:minmax-algebraic-objective-graphs-correct}).
  In fact,
  every admissible construction uses state graphs representing the same
  corresponding RNC states, and hence represents the same objective function.
  When the state computations are restricted to the sequential or
  parallel-scan constructions, the corresponding recurrence graphs are
  determined once their source graphs are fixed---as described in
  Remark~\ref{rem:minmax-sequential-recurrence-graph-determinacy}
  and Remark~\ref{rem:minmax-parallel-scan-graph-determinacy}.
\end{closingremark}

\begin{closingremark}[Irrelevance of the specific infinite extension of the input sequence]
  The specific extension $uu_T^\infseqOperatorSymb$ 
  of the finite sequence $u = u_1, \ldots, u_T$ is immaterial, 
  since the state computation-graph class at horizon $t \in \iival{1}{T}$ 
  depends on the input sequence only through the $t$-step prefix $u_1,\ldots,u_t$ of $u$, 
  which is also the $t$-step prefix of the extension $uu_T^\infseqOperatorSymb$.
\end{closingremark}

\begin{theorem}[Correctness of MinMax-algebraic objective graphs]
  \label{thm:minmax-algebraic-objective-graphs-correct}
  Let 
  $\lossfn \admissiblelossfunctions[S]$, 
  let $\ambientExtension{\lossfn}$ be an analytic ambient extension of $\lossfn$,
  let $z = \ptuple{u,y} \in \samplespacep{S}$,
  and
  let $uu_T^\infseqOperatorSymb\in\infseqs{U}$ be the infinite extension of $u$.%
  \footnote{Same context as Definition~\ref{def:minmax-algebraic-objective-graphs}.}
  Then,
  every graph $G \in \minmaxobjectivegraphclass{S}[z][\lossfn]$
  is an AMM-graph representing $\Lossfn{S,z}$.
  Consequently,
  \begin{equation*}
    \minmaxobjectivegraphclass{S}[z][\lossfn]
    \,\subseteq\,
    \customIndexedcompgraphfamily{\mathcal{S}_{\scriptscriptstyle\adaoplustimes}}[%
      \Lossfn{S,z}[\lossfn]%
    ]\,.
  \end{equation*}
\end{theorem}
\begin{proof}
  Let $G \in \minmaxobjectivegraphclass{S}[z][\lossfn]$
  and fix arbitrary constituent graphs satisfying
  Definition~\ref{def:minmax-algebraic-objective-graphs}.

  Let us introduce 
  the parameter-to-state function for $S$ 
  and 
  on the fixed input sequence 
  $uu^\infseqOperatorSymb$, 
  given by
  \begin{align*}
    \statefuncSymb : \Theta^f \times \timedomain \to X\,,
    \qquad\quad
    \statefuncSymb(\theta_f; t) 
    \,\defeq \,
    \indexedParamToStateFuncfamily{S}[uu^\infseqOperatorSymb][t][\theta_f]\,.
  \end{align*}

  By
  Proposition~\ref{prop:minmax-rnc-state-graphs-analytic-minmax},
  $\cgnode{X}^+$
  is an AMM-graph representing
  \begin{equation*}
    \theta^f
    \longmapsto
    \statefuncSymb(\theta^f; T).
  \end{equation*}
  When
  $T\geq2$,
  the same proposition shows that
  $\cgnode{X}^-$
  is an AMM-graph representing
  \begin{equation*}
    \theta^f
    \longmapsto
    \statefuncSymb(\theta^f; T\mkern-2mu - \mkern-2mu 1).
  \end{equation*}
  When
  $T=1$,
  $\cgnode{X}^-$
  is the canonical constant graph for
  $\initialstate{x}$
  and is therefore analytic.

  Similarly,
  $\cgnode{U}_T$
  is a constant analytic graph.
  Hence their common-input union
  $\cgnode{P}$
  is an AMM-graph representing
  \begin{equation*}
    \theta^f
    \longmapsto
    \bigptuple{
      \statefuncSymb(\theta^f; T\mkern-2mu - \mkern-2mu 1),
      u_T,
      \statefuncSymb(\theta^f; T)
    }.
  \end{equation*}

  Since
  $h$
  is analytic,
  the analytic extension
  \begin{equation*}
    \cgnode{H}
    =
    \ptuple{
      \cgnode{P}
    }
    \analyticextension
    h
  \end{equation*}
  is an AMM-graph.
  By correctness of analytic graph extension
  (Proposition~\ref{prop:correctness-analytic-extension-selected-graph}),
  \begin{align*}
    \cgval{H}(\theta^f,\theta^h)
    \;=\;
    \Bigfpinst{h}{
      \bigptuple{
        \statefuncSymb(\theta^f; T\mkern-2mu - \mkern-2mu 1),\,
        u_T,\,
        \statefuncSymb(\theta^f; T)
      }
    }{\theta^h}
    \;=\;
    \fpinst{\repval{S}}{u}{\theta^f,\theta^h}\,.
  \end{align*}

  The target graph
  $\cgnode{Y}$
  is constant and hence analytic.
  Since
  $\ambientExtension{\lossfn}$
  is analytic,
  \begin{equation*}
    G
    =
    \ptuple{
      \cgnode{H},
      \cgnode{Y}
    }
    \analyticextension
    \ambientExtension{\lossfn}
  \end{equation*}
  is an AMM-graph and
  \begin{align*}
    \cgval{G}(\theta^f,\theta^h)
    \;=\;
    \ambientExtension{\lossfn}
    \left(
    \cgval{H}(\theta^f,\theta^h),
      y
    \right)
    \;=\;
    \lossfn
    \Big(
    \fpinst{\repval{S}}{u}{(\theta^f,\theta^h)},\,
      y
    \Big)
    \;=\;
    \Lossfn{S,z}(\theta^f,\theta^h)\,.
  \end{align*}
  Thus $G \in \minmaxgraphclass[\Lossfn{S,z}]$, 
  as required.
\end{proof}

The following is a more precise restatement of Theorem~\ref{thm:gradient-bounded-main}.

\begin{theorem}[Uniform bounded loss derivatives of MinMax RNCs]
  \label{thm:gradient-bounded-precise}
  Let us consider 
  an admissible loss function 
  $\lossfn \in \admissiblelossfunctions[S]$
  and 
  a constant $r\in\nonnegreals$.

  Then, there exists a constant $B_{\lossfn,r}\in\nonnegreals$ such that the following holds:
  \begin{itemize}
    \item
      for every sample $z = \ptuple{u,y} \in \truncsamplespacep{S}{r}$,
      the following claims hold:
      \begin{enumerate}
        \item
          \label{thm:gradient-bounded-precise:item-graph}
          for every graph
          $G \in \minmaxobjectivegraphclass{S}[z][\lossfn]$,
          \begin{equation}
            \label{eq:minmax-objective-graph-uniform-derivative-bound}
            \sup_{\theta\in\truncation{\Theta}{r}}
            \big\|
            \minmaxsdvop{G}(\theta)
            \big\|
            \leq
            B_{\lossfn,r}\,;
          \end{equation}

        \item
          \label{thm:gradient-bounded-precise:item-id}
          for every graph
          $G \in \minmaxobjectivegraphclass{S}[z][\lossfn]$,
          the function
          $\minmaxsdvop{G}$
          is an intensional-derivative function of the objective function~$\Lossfn{S,z}[\lossfn]$; namely,
          \begin{equation*}
            \minmaxsdvop{\Lossfn{S,z}[\lossfn]} \in \intensionalderivatives \Lossfn{S,z}[\lossfn]\,;
          \end{equation*}

        \item
          \label{thm:gradient-bounded-precise:item-ae}
          there exists a full-measure subset
          $\hat\Theta_{r,z}
          \subseteq
          \truncation{\Theta}{r}$
          such that
          \begin{equation}
            \label{eq:minmax-objective-ae-derivative-bound}
            \hspace{-2em}
            \Lossfn{S,z}[\lossfn] \
            \text{is differentiable on}\
            \hat\Theta_{r,z}
            \quad\text{ and }\quad
            \sup_{\theta\in\hat\Theta_{r,z}}
            \big\|
            \dvop{\Lossfn{S,z}[\lossfn]}(\theta)
            \big\|
            \leq
            B_{\lossfn,r}\,;
          \end{equation}
      \end{enumerate}

    \item
      consequently,
      for every graph selector
      $\selgraphsymb$
      satisfying
      \begin{equation*}
        \selgraphsymb[\Lossfn{S,z}[\lossfn]]
        \in
        \minmaxobjectivegraphclass{S}[z][\lossfn]
        \qquad\quad
        \forall z\in\samplespacep{S}\,,
      \end{equation*}
      and for every sample $z\in\samplespacep{S}$,
      the function
      $\minmaxsdv[\selgraphsymb]\Lossfn{S,z}[\lossfn]$
      is a selected intensional derivative of $\Lossfn{S,z}[\lossfn]$,
      namely: 
      \begin{equation*}
        \minmaxsdvop[\selgraphsymb]{\Lossfn{S,z}[\lossfn]} \in \intensionalderivatives
        \Lossfn{S,z}[\lossfn]\,;
      \end{equation*}
    \item
      furthermore,
      \begin{equation}
        \label{eq:minmax-selector-loss-gradient-bound}
        \sup_{\theta\in\truncation{\Theta}{r}}
        \big\|
        \minmaxsdv[\selgraphsymb]
        \Lossfn{S,z}[\lossfn](\theta)
        \big\|
        \leq
        B_{\lossfn,r}
        \qquad\quad
        \forall
        z\in\truncsamplespacep{S}{r}\,.
      \end{equation}
  \end{itemize}
\end{theorem}

\begin{remark}[Uniformity of the bound]
  The bounding constant $B_{\lossfn,r}$ only depends on the considered 
  MinMax RNC $S$, 
  loss function $\lossfn$,
  and 
  constant $r$ bounding samples and parameters.
  In particular, the bound holds uniformly 
  for every
  admissible graph-selector $\selgraphsymb$ 
  and 
  every sequence length $T$.
\end{remark}

\begin{proof}[Proof of Theorem~\ref{thm:gradient-bounded-precise}]
  We can write $S$ as an analytic extension, as follows:
  \begin{equation*}
    S
    =
    \ptuple{D,\initialstate{x}}
    \analyticextension
    h,
    \qquad
    \Theta
    =
    \Theta^f
    \times
    \Theta^h.
  \end{equation*}

  Fix
  an admissible
  $\lossfn$
  and
  $r\in\nonnegreals$,
  and let
  \begin{equation*}
    z
    =
    \ptuple{u,y}
    \in
    \truncsamplespacep{S}{r},
    \qquad
    u
    =
    u_1\cdots u_T.
  \end{equation*}
  Set
  \begin{equation*}
    \rho_r
    \defeq
    \max
    \left\{
      r,
      \|\initialstate{x}\|
    \right\}.
  \end{equation*}
  The continuation
  $uu_T^\infseqOperatorSymb$
  of
  Definition~\ref{def:minmax-algebraic-objective-graphs}
  satisfies
  \begin{equation*}
    uu_T^\infseqOperatorSymb
    \in
    \infseqs{\truncation{U}{\rho_r}}.
  \end{equation*}

  We first prove
  Point~\ref{thm:gradient-bounded-precise:item-graph}.
  By
  Corollary~\ref{cor:minmax-rnc-ubibs}
  and
  Lemma~\ref{lem:uniform-output-context-state-bounds},
  the output-context parameter functions are uniformly bounded on
  $\truncation{\Theta^f}{r}$,
  uniformly over
  $z\in\truncsamplespacep{S}{r}$.
  Thus there exists
  $B^P_r\in\nonnegreals$
  such that every context graph
  $\cgnode{P}$
  occurring in
  Definition~\ref{def:minmax-algebraic-objective-graphs}
  satisfies
  \begin{equation}
    \label{eq:gradient-proof-context-value-bound}
    \sup_{\theta^f\in\truncation{\Theta^f}{r}}
    \big\|
      \cgval{\cgnode{P}}(\theta^f)
    \big\|
    \leq
    B^P_r.
  \end{equation}

  Furthermore,
  Theorem~\ref{thm:minmax-rnc-ubibs-selected-ubibsd},
  applied at radius
  $\rho_r$,
  uniformly bounds the
  $\minmaxsdvop$-pseudo-derivatives of every admissible choice of
  $\cgnode{X}^-$
  and
  $\cgnode{X}^+$.
  Since
  $\cgnode{U}_T$
  is parameter-independent and common-input union concatenates output rows,
  there exists
  $B^{P'}_r\in\nonnegreals$
  such that every admissible context graph
  $\cgnode{P}$
  satisfies
  \begin{equation}
    \label{eq:gradient-proof-context-derivative-bound}
    \sup_{\theta^f\in\truncation{\Theta^f}{r}}
    \big\|
      \minmaxsdvop{\cgnode{P}}(\theta^f)
    \big\|
    \leq
    B^{P'}_r.
  \end{equation}
  Here and below,
  finite-dimensional norm equivalence is used whenever passing from the row norm
  of
  Theorem~\ref{thm:minmax-rnc-ubibs-selected-ubibsd}
  to the operator norm.

  The
  $\Theta^h$-parameter input of the output graph is the canonical coordinate
  projection from
  $\Theta=\Theta^f\times\Theta^h$.
  Hence its values and derivatives are uniformly bounded on
  $\truncation{\Theta}{r}$,
  as in
  Corollary~\ref{cor:selected-feedforward-coordinate-projection-bounds}. \todo{}
  Therefore,
  applying
  Lemma~\ref{lem:graph-uniform-selected-feedforward-extension-bounds} \todo{}
  to the analytic output map
  $h$,
  with the index set ranging over all samples and all admissible constituent
  graph choices,
  yields constants
  \begin{equation*}
    B^H_r,
    \qquad
    B^{H'}_r
    \in
    \nonnegreals
  \end{equation*}
  such that every admissible output graph
  $\cgnode{H}$
  satisfies
  \begin{align}
    \label{eq:gradient-proof-output-value-bound}
    \sup_{\theta\in\truncation{\Theta}{r}}
    \big\|
      \cgval{\cgnode{H}}(\theta)
    \big\|
    &\leq
    B^H_r,
    \\
    \label{eq:gradient-proof-output-derivative-bound}
    \sup_{\theta\in\truncation{\Theta}{r}}
    \big\|
      \minmaxsdvop{\cgnode{H}}(\theta)
    \big\|
    &\leq
    B^{H'}_r.
  \end{align}

  The target graph
  $\cgnode{Y}$
  satisfies
  \begin{equation*}
    \|y\|
    \leq
    r,
    \qquad
    \minmaxsdvop{\cgnode{Y}}(\theta)
    =
    0.
  \end{equation*}
  Since
  $\ambientExtension{\lossfn}$
  is the designated analytic ambient extension of
  $\lossfn$,
  a second application of
  Lemma~\ref{lem:graph-uniform-selected-feedforward-extension-bounds}, \todo{}
  now to the analytic extension of
  $\ptuple{\cgnode{H},\cgnode{Y}}$
  by
  $\ambientExtension{\lossfn}$,
  yields
  $B_{\lossfn,r}\in\nonnegreals$
  such that every
  \begin{equation*}
    G
    \in
    \minmaxobjectivegraphclass{S}[z][\lossfn]
  \end{equation*}
  satisfies
  \begin{equation*}
    \sup_{\theta\in\truncation{\Theta}{r}}
    \big\|
      \minmaxsdvop{G}(\theta)
    \big\|
    \leq
    B_{\lossfn,r}.
  \end{equation*}
  The constant
  $B_{\lossfn,r}$
  is uniform over
  $z$
  and over all admissible objective-graph and constituent-graph choices.
  This proves
  Point~\ref{thm:gradient-bounded-precise:item-graph}.

  We next prove
  Point~\ref{thm:gradient-bounded-precise:item-id}.
  Let
  \begin{equation*}
    G
    \in
    \minmaxobjectivegraphclass{S}[z][\lossfn].
  \end{equation*}
  By
  Theorem~\ref{thm:minmax-algebraic-objective-graphs-correct},
  \begin{equation*}
    G
    \in
    \minmaxgraphclass[\Lossfn{S,z}[\lossfn]],
  \end{equation*}
  and in particular
  $\Lossfn{S,z}[\lossfn]\in\minmaxfuncclass$.
  Consider the singleton class
  \begin{equation*}
    \mathcal Q_z
    \defeq
    \{
      \Lossfn{S,z}[\lossfn]
    \},
  \end{equation*}
  and the graph selector
  $\selgraphsymb_G$
  defined by
  \begin{equation*}
    \selgraphsymb_G[\Lossfn{S,z}[\lossfn]]
    \defeq
    G.
  \end{equation*}
  By
  Theorem~\ref{theorem:minmax-selected-pseudoderivatives-are-sids},
  \begin{equation*}
    \minmaxsdvop{G}
    =
    \minmaxsdv[\selgraphsymb_G]
    \Lossfn{S,z}[\lossfn]
  \end{equation*}
  is an intensional derivative of
  $\Lossfn{S,z}[\lossfn]$.
  This proves
  Point~\ref{thm:gradient-bounded-precise:item-id}.

  We finally prove
  Point~\ref{thm:gradient-bounded-precise:item-ae}.
  Fix any
  \begin{equation*}
    G
    \in
    \minmaxobjectivegraphclass{S}[z][\lossfn].
  \end{equation*}
  By
  Point~\ref{thm:gradient-bounded-precise:item-id},
  \begin{equation*}
    \minmaxsdvop{G}
    \in
    \partial_\bullet
    \Lossfn{S,z}[\lossfn].
  \end{equation*}
  Hence,
  by
  Proposition~\ref{prop:pap-intensional-derivative-ae-correct},
  there exists a full-measure subset
  \begin{equation*}
    \hat\Theta_z
    \subseteq
    \Theta
  \end{equation*}
  such that
  \begin{equation*}
    \Lossfn{S,z}[\lossfn]
    \text{ is differentiable at every }
    \theta\in\hat\Theta_z,
    \qquad
    \dvop{\Lossfn{S,z}[\lossfn]}(\theta)
    =
    \minmaxsdvop{G}(\theta)
    \quad
    \forall\theta\in\hat\Theta_z.
  \end{equation*}
  Let
  \begin{equation*}
    \hat\Theta_{r,z}
    \defeq
    \hat\Theta_z
    \cap
    \truncation{\Theta}{r}.
  \end{equation*}
  By
  Fact~\ref{fact:full-measure-restriction},
  $\hat\Theta_{r,z}$
  is a full-measure subset of
  $\truncation{\Theta}{r}$.
  Furthermore,
  Point~\ref{thm:gradient-bounded-precise:item-graph}
  gives
  \begin{equation*}
    \Lossfn{S,z}[\lossfn] \
    \text{ is differentiable at every }\
    \theta\in\hat\Theta_{r,z}
    \qquad\text{ and }\qquad
    \sup_{\theta\in\hat\Theta_{r,z}}
    \big\|
    \dvop{\Lossfn{S,z}[\lossfn]}(\theta)
    \big\|
    \leq
    B_{\lossfn,r}.
  \end{equation*}
  This proves
  Point~\ref{thm:gradient-bounded-precise:item-ae}.

  Finally,
  let
  $\selgraphsymb$
  satisfy
  \begin{equation*}
    \selgraphsymb[\Lossfn{S,z}[\lossfn]]
    \in
    \minmaxobjectivegraphclass{S}[z][\lossfn]
    \qquad
    \forall z\in\samplespacep{S}.
  \end{equation*}
  For every
  $z\in\samplespacep{S}$,
  Point~\ref{thm:gradient-bounded-precise:item-id}
  applied to
  \begin{equation*}
    G
    =
    \selgraphsymb[\Lossfn{S,z}[\lossfn]]
  \end{equation*}
  shows that
  \begin{equation*}
    \minmaxsdv[\selgraphsymb]
    \Lossfn{S,z}[\lossfn]
  \end{equation*}
  is an intensional derivative of
  $\Lossfn{S,z}[\lossfn]$.
  Furthermore,
  Point~\ref{thm:gradient-bounded-precise:item-graph}
  gives
  \begin{equation*}
    \sup_{\theta\in\truncation{\Theta}{r}}
    \big\|
      \minmaxsdv[\selgraphsymb]
      \Lossfn{S,z}[\lossfn](\theta)
    \big\|
    \leq
    B_{\lossfn,r}
    \qquad
    \forall
    z\in\truncsamplespacep{S}{r}.
  \end{equation*}
  The constant
  $B_{\lossfn,r}$
  is independent of
  $\selgraphsymb$
  because
  Point~\ref{thm:gradient-bounded-precise:item-graph}
  holds uniformly over the complete objective computation-graph class.
\end{proof}

\begin{theorem}[BIBLG properties of MinMax RNCs]
  \label{theorem:minmax-rnc-biblg}
  For every loss function $\ell \in \admissiblelossfunctions[S]$,
  the following holds:
  \begin{itemize}
    \item
      $S$ satisfies $\minmaxobjectivegraphclass{S} \text{-}\minmaxsdv\text{-BIBLG}$;
    \item
      for every graph selector
      $\selgraphsymb$
      satisfying
      \begin{equation*}
        \selgraphsymb[\Lossfn{S,z}[\lossfn]]
        \in
        \minmaxobjectivegraphclass{S}[z][\lossfn]
        \qquad
        \forall z\in\samplespacep{S},
      \end{equation*}
      we have that $S$ satisfies
      $\minmaxsdv[\selgraphsymb]\text{-BIBLG}$,
      with constants independent of
      $\selgraphsymb$;
    \item
      $S$ satisfies ae-BIBLG.
    \end{itemize}
\end{theorem}
\begin{proof}
  By
  Theorem~\ref{thm:minmax-algebraic-objective-graphs-correct},
  every graph in
  $\minmaxobjectivegraphclass{S}$
  represents the corresponding objective function and is an AMM-graph.
  Furthermore,
  Points~\ref{thm:gradient-bounded-precise:item-graph}
  and~\ref{thm:gradient-bounded-precise:item-id}
  of
  Theorem~\ref{thm:gradient-bounded-precise}
  give, respectively, the graph-uniform derivative bound and the
  intensional-derivative property required by
  Definition~\ref{def:graph-uniform-selected-biblg}.
  Hence
  $S$
  satisfies
    $\minmaxobjectivegraphclass{S}
    \text{-}\minmaxsdv\text{-BIBLG}$.

  The second claim follows from
  Proposition~\ref{prop:graph-uniform-implies-selector-biblg},
  which also shows that the corresponding constants can be chosen independently
  of
  $\selgraphsymb$.

  Finally,
  Point~\ref{thm:gradient-bounded-precise:item-ae}
  of
  Theorem~\ref{thm:gradient-bounded-precise}
  is precisely the full-measure differentiability and uniform derivative bound
  required by
  Definition~\ref{def:ae-biblg}.
  Hence
  $S$
  satisfies ae-BIBLG.
\end{proof}

\begin{closingremark}[Relation to Theorem~\ref{thm:gradient-bounded-main}]
  Theorem~\ref{thm:gradient-bounded-precise}
  gives the graph-uniform formulation underlying
  Theorem~\ref{thm:gradient-bounded-main}.
  In particular, the derivative bound is established uniformly over the complete
  objective computation-graph family before any graph selector is chosen.
  Consequently, every admissible graph selector inherits the same bounding
  constant.
\end{closingremark}

\subsection{Non-necessarily Vanishing State-to-State Derivatives}
\label{subsec:minmax-nonvanishing-state-derivatives}

We prove Theorem~\ref{thm:gradient-nonvanishing} uniformly over
MinMax-algebraic \emph{recurrence-evaluation graphs}.  In this subsection,
reset and set coefficients are fixed once an input sequence has been chosen;
the initial state is the variable with respect to which derivatives are taken.
The admissible graphs therefore contain the initial-state coordinates and
these coefficients as their only scalar sources, with only MinMax reductions
introduced during recurrence evaluation.  This includes sequential,
composed-pair, and parallel-scan evaluations, and their MinMax-algebraic
rearrangements.  It does not mean all analytic-MinMax graphs representing the
same state-to-state function.

\begin{definition}[Initial-state-variable MinMax recurrence graphs]
  \label{def:nnvsd-initial-state-variable-graphs}
  Let $D=\mtuple{X,U,f}{R,s}$ be an $n$-dimensional MinMax
  recurrent unit, let $t\in\posnaturals$, and fix
  $u_1,\ldots,u_t\in U$.  Write $u\defeq(u_1,\ldots,u_t)$ and let
  $q_t:X\to X$ be the corresponding state-to-state map.
  Let $\bar q_t:\reals^n\to\reals^n$ be its canonical ambient
  extension, obtained by evaluating the same MinMax recurrence
  formula on arbitrary initial vectors in $\reals^n$.
  Thus $\bar q_t|_X=q_t$, and derivatives of $q_t$ below
  are understood through this ambient extension.
  Let $C_{u,t}$ be the set of scalar entries of
  $R(u_k)$ and $s(u_k)$ for $1\leq k\leq t$.
  Let $Q^{\mathrm{id}}$ be the identity computation graph on the
  initial-state variable $x\in\reals^n$.  For each $k$, let
  $Q_k^R$ and $Q_k^s$ be canonical constant computation graphs
  representing $R(u_k)$ and $s(u_k)$ on the same input space.
  The \emph{initial-state-variable MinMax recurrence-graph family}
  for the ambient extension of $q_t$ is
  \begin{align*}
    \mathcal G^x_{D,u,t}
    \defeq
    \big\{
      G\in\minmaxextensionclass[
        Q^{\mathrm{id}},Q_1^R,\ldots,Q_t^R,
        Q_1^s,\ldots,Q_t^s]
      \;\bigm|\;\cgval{G}=\bar q_t
    \big\}.
  \end{align*}
  The source-output nodes of such a graph are the initial-state
  coordinates and the fixed recurrence coefficients; its new nodes
  are min/max reductions.  A graph selector is called
  \emph{recurrence-selected} if, at every horizon $t$, it selects
  a graph from $\mathcal G^x_{D,u,t}$.
\end{definition}

\begin{lemma}[Graph-independent routing of initial-state derivatives]
  \label{lem:nnvsd-graph-independent-routing}
  Fix $D,u,t$ as above and $x_0\in X$.  Suppose that
  the coordinates $(x_0)_1,\ldots,(x_0)_n$ are pairwise distinct,
  that none belongs to $C_{u,t}$, and that each coordinate of
  $q_t(x_0)$ is one of the coordinates of $x_0$.
  Then there exists a unique map
  $\pi_t:\{1,\ldots,n\}\to\{1,\ldots,n\}$ such that
  \begin{align}
    \label{eq:nnvsd-routing-value}
    (q_t(x_0))_i=(x_0)_{\pi_t(i)}
    \qquad\forall i\in\{1,\ldots,n\}.
  \end{align}
  The map $q_t$ is ordinarily differentiable at $x_0$, and,
  for every recurrence-selected MinMax graph selector~$\selgraphsymb$,
  \begin{align}
    \label{eq:nnvsd-graph-independent-Jacobian}
    \minmaxsdvop[\selgraphsymb]{q_t(x_0)}
    =\dvop{q_t(x_0)}
    =P_{\pi_t},
    \qquad
    (P_{\pi_t})_{ij}\defeq\ind{j=\pi_t(i)}.
  \end{align}
  If $\pi_t$ is a permutation, then
  $\|P_{\pi_t}\|=1$ in the Euclidean operator norm.
\end{lemma}
\begin{proof}
  Fix a graph $G\in\mathcal G^x_{D,u,t}$.
  Every scalar source of $G$ is either a coordinate function
  $x\mapsto x_j$, with derivative $e_j^\transposeop$,
  or a constant function with value in $C_{u,t}$, with
  derivative zero.  Every other scalar node applies a finite
  minimum or maximum to predecessor values.  Hence, at $x_0$,
  the value of every scalar node belongs to
  \begin{equation}
    \label{eq:nnvsd-source-values}
    C_{u,t}\cup\{(x_0)_1,\ldots,(x_0)_n\}.
  \end{equation}

  Fix an output coordinate $i$.  By assumption its value is
  $(x_0)_{\pi_t(i)}$, and the index $\pi_t(i)$ is unique.
  At any min/max node, the MinMax pseudo-derivative puts
  nonnegative weights summing to one on the active predecessors,
  i.e., those whose values equal the node's value.
  Following an active predecessor therefore preserves the
  scalar value.  Starting at output coordinate $i$ and
  following active predecessors to the sources, every reached
  source must consequently have value $(x_0)_{\pi_t(i)}$.
  It cannot be a coefficient source, since
  $(x_0)_{\pi_t(i)}\notin C_{u,t}$; nor can it be another
  initial-state coordinate, since the coordinates of $x_0$
  are pairwise distinct.  All active source-to-output paths
  therefore have derivative $e_{\pi_t(i)}^\transposeop$.
  By the chain rule for the selected MinMax pseudo-derivative,
  taking convex combinations of these identical rows yields
  \begin{align*}
    \big(\minmaxsdvop{G}(x_0)\big)_{i,:}
    =e_{\pi_t(i)}^\transposeop.
  \end{align*}
  This argument allows ties between distinct computational paths
  as long as they carry the same initial-state coordinate.
  Since $G$ was arbitrary, the selected Jacobian is
  $P_{\pi_t}$ for every recurrence-selected graph.

  We also establish ordinary differentiability.  Because the graph
  is finite, it has only finitely many scalar sources, each either
  constant or a coordinate projection.  At $x_0$, all different
  initial-state coordinates are separated from one another and
  from every coefficient value.  Thus there is an open
  neighbourhood $V$ of $x_0$ on which the order of any two
  source functions that have different values at $x_0$ does not
  change.  Source functions agreeing at $x_0$ are either
  identical coordinate projections or constant functions;
  if they are identical projections, their values also agree
  throughout $V$.  Consequently, by induction through the
  finite min/max graph, each output coordinate agrees on $V$
  with the same initial-state coordinate projection selected
  at $x_0$.  Hence
  \begin{align*}
    q_t(x)=P_{\pi_t}x\qquad\forall x\in V,
  \end{align*}
  proving $\dvop{q_t(x_0)}=P_{\pi_t}$.
  If $\pi_t$ is a permutation, then $P_{\pi_t}$ is orthogonal,
  so $\|P_{\pi_t}\|=1$.
\end{proof}

\subsubsection{Proof of Theorem~\ref{thm:gradient-nonvanishing}}

\thmgradientnonvanishing*

\begin{proof}
  Fix $\epsilon\in\posreals$.  We construct the classes
  $(\mathcal D_n)_{n\geq1}$ using the identity-reset and
  permutation realisations of
  Theorems~\ref{theorem:expressivity-main-ir}
  and~\ref{theorem:expressivity-main-perm}, choosing their
  state regions and coefficient margins as specified below.
  In particular, the choices are made so that every neuron
  in the classes has a hypercube of side length $\epsilon$
  and a fixed infinite input witnessing the asserted
  derivative identities.

  \paragraph{Scalar identity-reset realisations.}
  Let $A=(Q,\Sigma,\delta)$ be an arbitrary identity-reset
  semiautomaton with $Q=\{q_1,\ldots,q_m\}$.
  Choose pairwise-disjoint closed intervals
  \begin{align*}
    I_j=[c_j-\epsilon/2,c_j+\epsilon/2]
    \qquad (1\leq j\leq m),
  \end{align*}
  with positive gaps between them, and set
  $X\defeq\bigcup_{j=1}^m I_j$.
  Choose $L,H\in\reals$ with
  \begin{align}
    \label{eq:nnvsd-identity-guards}
    L<\min X<\max X<H.
  \end{align}
  Assign distinct input components to the symbols of $\Sigma$.
  If necessary, adjoin one extra input component, represented
  by an input $u_\star$, whose transformation is the identity;
  this extra component need not represent a symbol of $A$.
  On an identity component prescribe the scalar coefficients
  $(R,s)=(H,L)$; on a component resetting to $q_j$
  prescribe $(R,s)=(L,c_j)$.
  The resulting MinMax unit has
  \begin{align*}
    f(x,u)
    &=\max\{\min\{R(u),x\},s(u)\}
    \\
    &=\begin{cases}
      x,&\text{on an identity component},\\
      c_j,&\text{on a component resetting to $q_j$},
    \end{cases}
    \qquad x\in X.
  \end{align*}
  Thus its separated state components $I_j$ realise $A$,
  and $u_\star$ acts as the identity if it was adjoined.

  In passing to a MinMax Neuron, choose MLP coefficient
  functions approximating the prescribed functions uniformly
  on the finitely many compact input components, with error
  small compared with the strict gaps in
  \eqref{eq:nnvsd-identity-guards} and with the distance
  from $c_j$ to the boundary of $I_j$.
  The realised neuron $D$ then still satisfies, on each
  identity component,
  \begin{align}
    \label{eq:nnvsd-identity-coefficients}
    R(u)>\max X,
    \qquad s(u)<\min X,
  \end{align}
  and, on a component resetting to $q_j$,
  \begin{align*}
    R(u)<\min X,
    \qquad s(u)\in\operatorname{int}(I_j).
  \end{align*}
  Hence the identity transitions remain \emph{exactly}
  identity, and each reset still reaches its prescribed state
  component.  Such neurons realise $A$ as required by
  Theorem~\ref{theorem:expressivity-main-ir}.

  Include all chosen scalar neurons in $\mathcal D_1$,
  retaining one identity input $u_\star$ for each.
  For each such $D$, choose
  \begin{align*}
    \tilde X_D\defeq I_1,
    \qquad
    u=(u_\star,u_\star,\ldots).
  \end{align*}
  Equation~\eqref{eq:nnvsd-identity-coefficients} implies
  $f(x,u_\star)=x$ on the open interval
  $(s(u_\star),R(u_\star))$, which contains $X$.
  Therefore, for every $t\in\posnaturals$,
  \begin{align}
    \label{eq:nnvsd-identity-prefix}
    q_t(x)=x
    \qquad\text{on a neighbourhood of $\tilde X_D$}.
  \end{align}
  In particular, $\dvop{q_t(x_0)}=[1]$ for all
  $x_0\in\tilde X_D$.
  All recurrence-coefficient values along this input
  are strictly outside $X$ and hence outside
  $\tilde X_D$.  Lemma~\ref{lem:nnvsd-graph-independent-routing}
  (with $n=1$) shows that, for every recurrence-selected
  graph selector,
  \begin{align}
    \label{eq:nnvsd-identity-selected}
    \minmaxsdvop[\selgraphsymb]{q_t(x_0)}
    =\dvop{q_t(x_0)}=[1],
    \qquad
    \big\|\minmaxsdvop[\selgraphsymb]{q_t(x_0)}\big\|=1.
  \end{align}
  This verifies the required assertion for every
  $D\in\mathcal D_1$.  In addition, degree-one permutation
  semiautomata are identity semiautomata and are covered by
  these scalar realisations.

  \paragraph{Permutation realisations in dimensions $n\geq2$.}
  Fix $n\geq2$ and let $A$ be an arbitrary permutation
  semiautomaton of degree $n$.  As in
  Theorem~\ref{theorem:expressivity-main-perm},
  we may represent its input transformations by coordinate
  permutations $\pi_\sigma$ on $\{1,\ldots,n\}$.
  Choose pairwise-disjoint closed intervals
  \begin{align*}
    I_j=[c_j-\epsilon/2,c_j+\epsilon/2]
    \qquad(1\leq j\leq n),
  \end{align*}
  with positive gaps, put
  $K\defeq\bigcup_{j=1}^n I_j$ and
  $X\defeq K^n$, and choose
  \begin{align}
    \label{eq:nnvsd-permutation-guards}
    L<\min K<\max K<H.
  \end{align}
  For a symbol $\sigma\in\Sigma$ and its corresponding
  input component, prescribe
  \begin{align}
    \label{eq:nnvsd-permutation-coefficients}
    R(u)_{ij}
    &\defeq\begin{cases}
      H,&j=\pi_\sigma(i),\\
      L,&j\ne\pi_\sigma(i),
    \end{cases}
    &
    s(u)_i&\defeq L.
  \end{align}
  Then for all $x\in X$,
  \begin{align}
    \label{eq:nnvsd-permutation-transition}
    f(x,u)_i
    &=\max\!\left\{
      \max_{1\leq j\leq n}\min\{R(u)_{ij},x_j\},
      s(u)_i
    \right\}
    =x_{\pi_\sigma(i)}.
  \end{align}
  This is exactly the coordinate-permutation realisation
  used in Theorem~\ref{theorem:expressivity-main-perm}.

  As in the scalar case, choose MLP coefficient functions
  approximating the prescribed coefficients uniformly, with
  sufficiently small error that every coefficient formerly
  equal to $H$ remains strictly greater than $\max K$,
  and every coefficient formerly equal to $L$ remains
  strictly smaller than $\min K$.  Equation
  \eqref{eq:nnvsd-permutation-transition} then holds
  \emph{exactly} for the resulting MinMax Neuron,
  not merely approximately.  Let $\mathcal D_n$ consist
  of these chosen neurons, one for each degree-$n$
  permutation semiautomaton, and choose for each one
  \begin{align}
    \label{eq:nnvsd-permutation-cube}
    \tilde X_D\defeq I_1\times\cdots\times I_n
    \subseteq X.
  \end{align}
  This is an $n$-dimensional hypercube of side length
  $\epsilon$; every point of it has pairwise-distinct
  coordinates.

  Choose any infinite input sequence
  $u=(u_1,u_2,\ldots)$ using the input components
  of the constructed neuron.  For each $k$, let $P_k$
  denote the permutation matrix of the $k$-th transition.
  The equality in
  \eqref{eq:nnvsd-permutation-transition} holds on a
  neighbourhood of $X$ (by the strict guard margins),
  and $X$ is invariant under coordinate permutations.
  Consequently, for every $t\in\posnaturals$,
  \begin{align}
    \label{eq:nnvsd-permutation-prefix}
    q_t(x)=P_tP_{t-1}\cdots P_1x
    \qquad\text{on a neighbourhood of $\tilde X_D$},
  \end{align}
  and hence
  \begin{align}
    \label{eq:nnvsd-permutation-ordinary}
    \dvop{q_t(x_0)}=P_t\cdots P_1
    \qquad\forall x_0\in\tilde X_D.
  \end{align}
  The product $P_t\cdots P_1$ is a permutation matrix,
  so its Euclidean operator norm is one.

  Finally, every scalar coefficient of $R(u_k)$ and
  $s(u_k)$ is strictly below $\min K$ or strictly above
  $\max K$, whereas each coordinate of every
  $x_0\in\tilde X_D$ belongs to $K$.
  The coordinates of $x_0$ are pairwise distinct, and
  \eqref{eq:nnvsd-permutation-prefix} shows that each
  coordinate of $q_t(x_0)$ is one of those coordinates.
  Lemma~\ref{lem:nnvsd-graph-independent-routing}
  therefore applies to every horizon $t$ and every
  recurrence-selected selector, giving
  \begin{align}
    \label{eq:nnvsd-permutation-selected}
    \minmaxsdvop[\selgraphsymb]{q_t(x_0)}
    &=\dvop{q_t(x_0)}=P_t\cdots P_1,\\
    \big\|\minmaxsdvop[\selgraphsymb]{q_t(x_0)}\big\|
    &=1
    \qquad\forall x_0\in\tilde X_D.
  \end{align}

  \paragraph{Conclusion.}
  Every member of $\mathcal D_n$ is a MinMax Neuron
  with an $n$-dimensional hypercube of side length
  $\epsilon$ in its state space.
  By construction, $\mathcal D_1$ realises every
  identity-reset semiautomaton and also covers degree-one
  permutation semiautomata; for each $n\geq2$,
  $\mathcal D_n$ realises every degree-$n$ permutation
  semiautomaton.  Equations
  \eqref{eq:nnvsd-identity-selected} and
  \eqref{eq:nnvsd-permutation-selected} establish the
  stated graph-uniform derivative identity and unit norm
  for every $t\in\posnaturals$ and every
  $x_0\in\tilde X_D$, completing the proof.
\end{proof}

\clearpage{}%

\clearpage{}%

\subsection{Deferred Proofs for MinMax Pseudo-Derivatives}
\label{appendix:sec:gradient-aux}

\begin{proposition} \label{prop:subdifferential-max-limit}
  Let $f: X \to \reals$ be the function $f(x) = \bigoplus_{i=1}^n x_i$.
  For every $x \in \reals^n$ and every $d\in\reals^n$,
  \begin{align*}
    \lim_{t\downarrow 0}
    \frac{f(x + t d)-f(x)}{t}
    =
    \bigoplus_{i\in I(x)} d_i.
  \end{align*}
\end{proposition}
\begin{proof}
  Let
  \begin{align*}
    d_*
    &=
    \bigoplus_{i\in I(x)} d_i.
  \end{align*}

  We show the two following inequalities separately
  \begin{align} 
    \label{eq:prop:subdifferential-max-limit-1}
    \liminf_{t\downarrow 0}
    \frac{f(x+td)-f(x)}{t}
    &\geq
    d_*\, ,
    \\[.3em]
    \label{eq:prop:subdifferential-max-limit-2}
    \liminf_{t\downarrow 0}
    \frac{f(x+td)-f(x)}{t}
    &\leq
    d_*\,.
  \end{align}

  We first show~\eqref{eq:prop:subdifferential-max-limit-1}.
  For every $i\in I(x)$, 
  we have $x_i=f(x)$ and hence
  \begin{align*}
    f(x + t d)
    \;=\;
    \bigoplus_{j \in \iival{1}{n}}(x_j+td_j)
    \;\geq\;
    x_i+td_i
    \;=\;
    f(x)+td_i
    \qquad \forall t \in \reals.
  \end{align*}
  Thus,
  \begin{align*}
    \frac{f(x+td)-f(x)}{t}
    &\geq
    d_i
    \qquad \forall i\in I(x),\, \forall t \in \posreals,
  \end{align*}
  hence
  \begin{align*}
    \frac{f(x+td)-f(x)}{t}
    &\ge
    d_*
    \qquad \forall t \in \posreals,
  \end{align*}
  and hence~\eqref{eq:prop:subdifferential-max-limit-1} follows immediately.

  We now prove~\eqref{eq:prop:subdifferential-max-limit-2}. 
  If $I(x)= \iival{1}{n}$,
  then 
  \begin{align*}
    f(x + td) 
    =
    \bigoplus_{j=1}^n x_j+td_j
    =
    \bigoplus_{j=1}^n f(x)+td_j
    =
    f(x)+t \cdot \bigoplus_{j=1}^n d_j
    =
    f(x)+t \cdot\bigoplus_{i \in I(x)} d_i
    =
    f(x)+td_*
    \quad \forall t \in \posreals,
  \end{align*}
  and hence
  \begin{align*}
    \frac{f(x+td)-f(x)}{t}
    =
    \frac{f(x)+td_*-f(x)}{t}
    =
    \frac{td_*}{t}
    =
    d_*
    \qquad \forall t \in \posreals,
  \end{align*}
  which immediately implies~\eqref{eq:prop:subdifferential-max-limit-2}.

  Next, let us consider the complementary case when $I(x) \subset \iival{1}{n}$,
  and hence let
  \begin{align*}
    J(x)
    & \defeq \iival{1}{n} \setminus I(x)
  \end{align*}
  Since $J(x)$ is finite and $x_j<f(x)$ for every
  $j\in J(x)$, we have
  \begin{align*}
    \delta
    & \defeq
    \min_{j\in J(x)}(f(x)-x_j) > 0.
  \end{align*}
  Let us define
  \begin{align*}
    C
    & \defeq
    1+\bigoplus_{j\in J(x)}|d_j-d_*| > 0.
  \end{align*}
  For every $j\in J(x)$ and every $t$ satisfying
  $0<t<\delta/C$, we have
  \begin{align*}
    x_j+td_j
    &=
    f(x)-(f(x)-x_j)+td_j\\
    &\le
    f(x)-\delta+td_j\\
    &=
    f(x)+td_*-\delta+t(d_j-d_*)\\
    &\le
    f(x)+td_*-\delta+tC\\
    &<
    f(x)+td_*-\delta+\frac{\delta}{C}C\\
    &=
    f(x)+td_*-\delta+\delta\\
    &=
    f(x)+td_*,
  \end{align*}
  overall,
  \begin{align*}
    x_j+td_j < f(x)+td_* \qquad \forall t \in (0,\delta/C).
  \end{align*}

  On the other hand, by definition of $d_*$, there exists some $i^\star\in I(x)$
  such that $d_{i^\star}=d_*$, and hence
  \begin{align*}
    x_{i^\star}+td_{i^\star}
    &=
    f(x)+td_* 
    \qquad \forall t \in \posreals.
  \end{align*}
  Thus, for all $t \in (0,\delta/C)$, 
  we have
  \begin{align*}
    f(x + td) = \bigoplus_{i \in I(x)} x_i+td_i
    \qquad \forall t \in \posreals,
  \end{align*}
  hence
  \begin{align*}
    f(x + td) 
    = \bigoplus_{i \in I(x)} x_i+td_i
    = \bigoplus_{i \in I(x)} f(x)+td_i
    = f(x) + t \cdot \bigoplus_{i \in I(x)} d_i
    =
    f(x)+td_*,
  \end{align*}
  and hence
  \begin{align*}
    \frac{f(x+td)-f(x)}{t}
    =
    \frac{f(x)+td_*-f(x)}{t}
    =
    \frac{td_*}{t}
    =
    d_*.
  \end{align*}
  Then \eqref{eq:prop:subdifferential-max-limit-2} follows immediately.
\end{proof}

\propmaxsubgradienthull*
\begin{proof}
  By definition, $g\in\partial_\circ f(x)$ if and only if
  \begin{align*}
    f(y)
    &\ge
    f(x)+g^\top(y-x)
    \qquad \text{for all } y\in\reals^n.
  \end{align*}

  Equivalently,
  \begin{align*}
    \bigoplus_{j=1}^n y_j
    &\ge
    f(x)+g^\top(y-x)
    \qquad \text{for all } y\in\reals^n.
  \end{align*}

  Now set $y=x+td$, where $t>0$ and $d\in\reals^n$. Then the subgradient
  inequality becomes
  \begin{align*}
    \bigoplus_{j=1}^n(x_j+td_j)
    &\ge
    f(x)+t g^\top d.
  \end{align*}

  Subtracting $f(x)$ and dividing by $t>0$, we obtain
  \begin{align*}
    \frac{\bigoplus_{j=1}^n(x_j+td_j)-f(x)}{t}
    &\ge
    g^\top d.
  \end{align*}

  By Proposition~\ref{prop:subdifferential-max-limit}
  we have
  \begin{align*}
    \lim_{t \downarrow 0} \frac{\bigoplus_{j=1}^n(x_j+td_j)-f(x)}{t}
    = 
    \bigoplus_{i \in I(x)} d_i,
  \end{align*}
  and hence
  \begin{align*}
    \lim_{t \downarrow 0} \frac{\bigoplus_{j=1}^n(x_j+td_j)-f(x)}{t}
    &\geq
    \lim_{t \downarrow 0} g^\top d
  \end{align*}
  yields
  \begin{align*}
    \bigoplus_{i \in I(x)} d_i
    &\geq
    g^\top d
    \qquad \forall d\in\reals^n.
  \end{align*}

  Therefore every subgradient $g\in\partial_\circ f(x)$ satisfies
  \begin{align*}
    g^\top d
    &\leq
    \bigoplus_{i\in I(x)} d_i
    \qquad \forall d\in\reals^n.
  \end{align*}

  We now derive the consequences of this inequality.
  First, take $d=\mathbf{1}$, where $\mathbf{1}=(1,\dots,1)$. Then
  \begin{align*}
    g^\top \mathbf{1}
    &\le
    \bigoplus_{i\in I(x)} 1
    =
    1.
  \end{align*}
  Taking $d=-\mathbf{1}$ gives
  \begin{align*}
    -g^\top \mathbf{1}
    &\le
    \bigoplus_{i\in I(x)}(-1)
    =
    -1.
  \end{align*}
  Hence
  \begin{align*}
    g^\top \mathbf{1}
    &\ge
    1.
  \end{align*}
  Combining the two inequalities, we get
  \begin{align*}
    \sum_{j=1}^n g_j
    &=
    1.
  \end{align*}

  Next, let $j\notin I(x)$. Choose $d$ such that
  \begin{align*}
    d_\ell &= 0 \qquad \forall \ell \in \iival{1}{n} \setminus \{j\},
  \end{align*}
  while $d_j$ is arbitrary. Then
  \begin{align*}
    \bigoplus_{i\in I(x)} d_i
    &=
    0.
  \end{align*}
  The inequality $g^\top d = g_jd_j\le 0$ must hold for every value of $d_j \in \reals$. 
  In particular, the inequalities 
  $g_j \leq 0$ (where $d_j = 1$)
  and 
  $g_j \geq 0$ (where $d_j = -1$) must hold, and hence $g_j = 0$.
  As the above applies to each $j\notin I(x)$, we conclude that
  \begin{align*}
    g_j
    &=
    0
    \qquad \forall j\notin I(x).
  \end{align*}

  Finally, let $i\in I(x)$. If $|I(x)|\geq 2$, choose $d$ such that
  \begin{align*}
    d_i &= -1,\\
    d_\ell &= 0 \qquad \text{for each } \ell\in I(x),\ \ell\neq i,\\
    d_j &= 0 \qquad \text{for each } j\notin I(x).
  \end{align*}

  Then
  \begin{align*}
    \bigoplus_{\ell\in I(x)} d_\ell
    &=
    0.
  \end{align*}

  The inequality $g^\top d\le 0$ gives
  \begin{align*}
    -g_i
    &\le
    0,
  \end{align*}
  and therefore
  \begin{align*}
    g_i
    &\ge
    0.
  \end{align*}

  If $|I(x)|=1$, say $I(x)=\{i\}$, then since $g_j=0$ for all $j\neq i$ and
  $\sum_{j=1}^n g_j=1$, we get
  \begin{align*}
    g_i
    &=
    1
    \ge
    0.
  \end{align*}

  Thus every subgradient satisfies
  \begin{align*}
    g_i &\geq 0 \qquad \text{for each } i\in I(x),\\
    g_i &= 0 \qquad \text{for each } i\notin I(x),\\
    \sum_{i\in I(x)} g_i &= 1.
  \end{align*}

  Conversely, suppose $g\in\reals^n$ satisfies
  \begin{align*}
    g_i &\geq 0 \qquad \text{for each } i\in I(x),\\
    g_i &= 0 \qquad \text{for each } i\notin I(x),\\
    \sum_{i\in I(x)} g_i &= 1.
  \end{align*}

  Then, for any $y\in\reals^n$,
  \begin{align*}
    g^\top(y-x)
    &=
    \sum_{i\in I(x)} g_i(y_i-x_i).
  \end{align*}

  Since $x_i=f(x)$ for every $i\in I(x)$, this becomes
  \begin{align*}
    g^\top(y-x)
    &=
    \sum_{i\in I(x)} g_i(y_i-f(x)).
  \end{align*}

  Also, for every $i\in I(x)$,
  \begin{align*}
    y_i
    &\le
    \bigoplus_{j=1}^n y_j.
  \end{align*}

  Therefore,
  \begin{align*}
    y_i-f(x)
    &\le
    \bigoplus_{j=1}^n y_j-f(x).
  \end{align*}

  Using $g_i\geq 0$ and $\sum_{i\in I(x)}g_i=1$, we obtain
  \begin{align*}
    g^\top(y-x)
    &\le
    \sum_{i\in I(x)} g_i
    \left(
    \bigoplus_{j=1}^n y_j-f(x)
    \right)\\
    &=
    \bigoplus_{j=1}^n y_j-f(x).
  \end{align*}

  Hence
  \begin{align*}
    f(x)+g^\top(y-x)
    &\le
    \bigoplus_{j=1}^n y_j.
  \end{align*}

  Since $f(y) = \bigoplus_{j=1}^n y_j$, this is exactly
  \begin{align*}
    f(y)
    &\ge
    f(x)+g^\top(y-x).
  \end{align*}

  Thus $g\in\partial_\circ f(x)$. 

  It remains to find the subgradient of minimum (Euclidean) norm. Since
  $g_i=0$ for every $i\notin I(x)$, we need to solve
  \begin{align*}
    \min \sum_{i\in I(x)} g_i^2
  \end{align*}
  subject to
  \begin{align*}
    \bigwedge_{i=1}^n g_i &\geq 0,\\
    \sum_{i\in I(x)} g_i &= 1.
  \end{align*}

  By the Cauchy--Schwarz inequality,
  \begin{align*}
    \left(\sum_{i\in I(x)} g_i\right)^2
    &\le
    |I(x)|\sum_{i\in I(x)} g_i^2.
  \end{align*}

  Since $\sum_{i\in I(x)}g_i=1$, we get
  \begin{align*}
    1
    &\le
    |I(x)|\sum_{i\in I(x)} g_i^2.
  \end{align*}

  Therefore,
  \begin{align*}
    \sum_{i\in I(x)} g_i^2
    &\geq
    \frac{1}{|I(x)|}.
  \end{align*}

  Equality in Cauchy--Schwarz holds if and only if all the active components
  $g_i$, $i\in I(x)$, are equal. Since they sum to $1$, equality holds precisely
  when
  \begin{align*}
    g_i
    &=
    \frac{1}{|I(x)|}
    \qquad \text{for all } i\in I(x).
  \end{align*}

  Thus the minimum norm subgradient is
  \begin{align*}
    g_i^\star
    =
    \begin{cases}
      \dfrac{1}{|I(x)|}, & i\in I(x),\\[6pt]
      0, & i\notin I(x).
    \end{cases}
  \end{align*}

  Finally,
  \begin{align*}
    \|g^\star\|
    =\sqrt{\sum_{i \in I(x)} \frac{1}{|I(x)|^2}}
    =\sqrt{|I(x)|\frac{1}{|I(x)|^2}}
    = \sqrt{\frac{1}{|I(x)|}}
    = \frac{1}{\sqrt{|I(x)|}}.
  \end{align*}

  This concludes the proof.
\end{proof}

\propminsubgradienthull*
\begin{proof}
  Let $\bar{f} : \reals^n \to \reals$ be the function 
  $\bar{f}(x_1, \ldots, x_n) = \bigoplus_{i=1}^n x_i$.
  Let $I_f$ be the active variable index function of $f$,
  and
  let $\bar{I}$ be the active variable index function of $\bar{f}$.
  Let $x = (x_1, \ldots, x_n) \in \reals^n$.
  By the definition of superdifferential, 
  we have
  \begin{align} \label{eq:prop:min-subgradient-hull-1}
    g \in \supdiff f(x)  \quad\Longleftrightarrow\quad -g \in \subdiff (-f)(x)
    \quad\qquad \forall g \in \reals^n.
  \end{align}

  Then,
  \begin{align*}
    -f(x)
    = 
    -\bigodot_{i=1}^n x_i
    = 
    \bigoplus_{i=1}^n -x_i
    =
    \bar{f}(-x),
  \end{align*}
  and,
  by Proposition~\ref{prop:max-subgradient-hull},
  \begin{align} \label{eq:prop:min-subgradient-hull-2}
    \subdiff \bar{f}(-x) = 
    \left\{
      g\in\reals^n \;\middle|\;
      \textstyle\bigwedge_{i=1}^n g_i\ge 0,\
      \textstyle\sum_{i\in \bar{I}(-x)} g_i=1
    \right\}.
  \end{align}
  We have
  \begin{align*}
    \bar{I}(-x)
    =
    \Big\{ i \in \iival{1}{n} \mid \textstyle\bigoplus_{i=1}^n -x_i = -x_i \Big\}
    =
    \Big\{ i \in \iival{1}{n} \mid \textstyle\bigodot_{i=1}^n x_i = x_i \Big\}
    = 
    I(x),
  \end{align*}
  and hence, from \eqref{eq:prop:min-subgradient-hull-2}, we obtain
  \begin{align*}
    \subdiff \bar{f}(-x) = 
    \left\{
      g\in\reals^n \;\middle|\;
      \textstyle\bigwedge_{i=1}^n g_i\ge 0,\
      \textstyle\sum_{i\in I(x)} g_i=1
    \right\}.
  \end{align*}
  By the calculus of subdifferentials,
  we have $\subdiff (-\bar{f})(x) = -\subdiff \bar{f}(x)$,
  meaning that
  \begin{align*}
    g \in \subdiff (-\bar{f})(x) \quad\Longleftrightarrow\quad  -g \in \subdiff \bar{f}(x)
    \qquad\qquad \forall g \in \reals^n,
  \end{align*}
  and hence
  \begin{align*}
    \subdiff (-f)(x) 
    \,=\,
    \subdiff (-\bar{f})(-x) 
    \,=\,
    -\subdiff \bar{f}(-x) 
    \,=\,
    \left\{
      -g\in\reals^n \;\middle|\;
      \textstyle\bigwedge_{i=1}^n g_i\ge 0,\
      \textstyle\sum_{i\in I(x)} g_i=1
    \right\}.
  \end{align*}
  Thus,
  by \eqref{eq:prop:min-subgradient-hull-1},
  we have
  \begin{align*}
    \supdiff f(x) = 
    \left\{
      g\in\reals^n \;\middle|\;
      \textstyle\bigwedge_{i=1}^n g_i\ge 0,\
      \textstyle\sum_{i\in I(x)} g_i=1
    \right\},
  \end{align*}
  which is our first claim.
  Then, our second claim, on the supergradient of minimum norm, follows by
  Proposition~\ref{prop:max-subgradient-hull}.
\end{proof}

\clearpage{}%

\clearpage{}%

\section{Piecewise-Affine Structure of MinMax Recurrences}
\label{sec:minmax-piecewise-linear}

\todo{}

\localtableofcontents

This appendix shows that MinMax recurrences have piecewise-affine structure.

We first provide a characterisation of MinMax dynamics functions where they are seen as producing
updated states by selecting individual components of the given
(i)~state vector $x$ 
and 
(ii)~input vectors $R(u),s(u)$.
We refer to any component of $x,R(u),s(u)$ as a \emph{source variable},
and to any source variable that is selected within an update as an \emph{active variable}.
Then, the above selections induce a finite 
partition of the joint state-input space, 
which we call \emph{active-variable partition} since its regions are characterised by sets of active
variables, with such set being constant across each region.

On every active-variable region
$Z\in\partitionstyle{Z}$,
the MinMax dynamics admits a state-affine representation
\begin{align*}
  f(x,u)
  \,=\,
  A_Z\cdot x+b_Z(u),
\end{align*}
with a highly constrained coefficient-offset pair:
every row of
$A_Z$
is either zero or a standard basis row,
and state propagation and input-dependent reset are mutually exclusive
coordinatewise.
We refer to this structure as
\emph{propagate-reset}.
Thus, whenever the recurrence is iterated on a forward-invariant state space,
its evolution along a trajectory can be written as a switched time-varying
affine recurrence
\begin{align*}
  x_t
  \,=\,
  A_t\cdot x_{t{-}1}+b_t
  \qquad\qquad
  \forall t\in\timedomain,
\end{align*}
whose active coefficients satisfy the propagate-reset constraints.

We develop several consequences of this representation.
First, the propagate-reset structure is closed under composition:
each state coordinate at a later time is obtained either from a single earlier
state coordinate or from a single input-dependent source value introduced at
one intermediate time point.
In particular, input-dependent contributions do not accumulate additively.
This provides a structural explanation for the bounded propagation properties
of MinMax recurrences without requiring strict contraction.
At points where the active-variable pattern is locally constant,
the active matrices also coincide with the state Jacobians;
products of these Jacobians remain uniformly bounded, while active propagation
paths may preserve derivatives of magnitude one over arbitrarily long
horizons.

The same structure yields regularity properties.
For every fixed input,
the state map is piecewise affine;
if the reset and set functions are themselves piecewise affine,
then the joint MinMax dynamics is piecewise affine.
Likewise, suitable PAP regularity of the source functions is inherited by the
MinMax dynamics.
We also distinguish the deterministic active-variable routing used in the
piecewise representation from the selected derivatives used at switching
boundaries, where several source variables may be simultaneously active.

Finally, we study the geometry of the active-variable partition.
For a fixed input,
changes of active-variable pattern are controlled by a finite family of
coordinate-equality and coordinate-threshold hyperplanes.
Consequently, every fixed-input section of an active-variable region is the
intersection of the state space with a semilinear set.
The possible hyperplane orientations are restricted to the coordinate
directions and coordinate-difference directions, whereas their threshold
offsets are determined by the fixed-input source values.
Already in the scalar case, the resulting active-variable regions can realise
arbitrary bounded interval interiors, showing that their sizes and placements
are not uniformly constrained by the structural form of the MinMax recurrence.

------------------------------

\todo{}

This section develops a linear-system interpretation of MinMax recurrences that
helps relate their stability and gradient properties to those of linear RNNs.
The key observation is that MinMax is piecewise state-affine: on every
active-variable region, each state coordinate is either copied from one coordinate
of the previous state or replaced by a current input-dependent value.
Thus, along a state sequence $(x_t)_{t \in \timedomain}$ starting from a given state $x_0$, MinMax
behaves as a
switched time-varying affine recurrence
\begin{align*}
  x_t\,=\,A_t \cdot x_{t{-}1}+b_t
  \qquad\qquad
  \forall t \in \timedomain,
\end{align*}
but with highly constrained recurrent matrices and mutually exclusive state
propagation and input injection.
This structure permits unit-gain information propagation while preventing
additive accumulation, thereby explaining both state stability and bounded,
non-necessarily-vanishing state gradients.

\subsection{Active Variables of MinMax Functions}
\label{subsec:active-variables-of-minmax-functions}

\begin{closingdefinition}[MinMax primitives and functions]
  \label{def:minmax-functions}
  For
  $d\in\posnaturals$
  and
  $j\in\iival{1}{d}$,
  let
  \begin{equation*}
    \pi_j^d:
    \reals^d
    \to
    \reals,
    \qquad\quad
    \pi_j^d(x)
    \defeq
    \getentry{x}{j},
  \end{equation*}
  be the $j$-th coordinate projection.

  The class of
  \emph{MinMax primitives}
  is
  \begin{equation*}
    \pureprimitiveminmaxfuncclass
    \defeq
    \left\{
      \pi_j^d
      \;\middle|\;
      d\in\posnaturals,\;
      j\in\iival{1}{d}
    \right\}
    \cup
    \left\{
      \minreduction{d},\maxreduction{d}
      \;\middle|\;
      d\in\posnaturals
    \right\},
  \end{equation*}

  The class
  $\pureminmaxfuncclass$
  of
  \emph{MinMax functions}
  is
  the closure of $\pureprimitiveminmaxfuncclass$.
\end{closingdefinition}

\begin{closingdefinition}[Closure derivation]
  \label{def:function-closure-derivation}
  Let $\mathcal{P}$ be a class of functions.

  A \emph{closure derivation} relative to the class $\mathcal{P}$
  is defined inductively as follows:
  \begin{itemize}
    \item
      for $f\in\mathcal{P}$,
      the tuple
      $\tuple{\mathsf{primitive},f}$,
      called the \emph{primitive derivation} of $f$;

    \item
      for $f = \ptuple{f_1,\ldots,f_k} \in \functionclassclosure{\mathcal{P}}$,
      a tuple
      $\tuple{\mathsf{tupling},\delta_1,\ldots,\delta_k}$
      where
      $\delta_1,\ldots,\delta_k$
      are closure derivations of
      $f_1,\ldots,f_k$,
      respectively;

    \item
      for $f = (h\circ g) \in \functionclassclosure{\mathcal{P}}$,
      a tuple 
      $\tuple{\mathsf{comp},\delta_h,\delta_g}$
      where
      $\delta_h$
      and
      $\delta_g$
      are closure derivations of
      $h$
      and
      $g$,
      respectively.
  \end{itemize}

  A \emph{MinMax closure derivation}
  is a closure derivation relative to
  $\pureprimitiveminmaxfuncclass$.
\end{closingdefinition}

\begin{closingdefinition}[Deterministic tie-breaking scheme]
  \label{def:deterministic-tie-breaking-scheme}
  For
  $d\in\posnaturals$,
  a
  \emph{(deterministic) tie-breaking selector of arity $d$}
  is a function
  \begin{equation*}
    \tiebreakingselector[d]:
    \powersetplus{\iival{1}{d}}
    \to
    \iival{1}{d}
  \end{equation*}
  such that
  \begin{sidealign*}
    \quanteq{
      \tiebreakingselector[d][J] \in J
    }{
      \forall J\in\powersetplus{\iival{1}{d}}.
    }
  \end{sidealign*}

  A
  \emph{(deterministic) tie-breaking scheme}
  is a family
  \begin{equation*}
    \rho
    =
    \big(
      \tiebreakingselector[d]
    \big)_{d\in\posnaturals}
  \end{equation*}
  where
  $\tiebreakingselector[d]$
  is a tie-breaking selector of arity $d$
  for every
  $d\in\posnaturals$.
  We write
  \begin{sidealign*}
    \quanteq{
      \tiebreakingselectorscheme[d]
      \defeq
      \tiebreakingselector[d]
    }{
      \forall d\in\posnaturals.
    }
  \end{sidealign*}
\end{closingdefinition}

\begin{closingdefinition}[Active variables of MinMax functions]
  \label{def:minmax-function-active-variable}
  Let $f \in \pureminmaxfuncclass$ be a MinMax function,
  \begin{equation*}
    f:
    \reals^{d_\mathrm{in}}
    \to
    \reals^{d_\mathrm{out}},
  \end{equation*}
  let $\delta$ be a MinMax closure derivation of $f$,
  and 
  let $\rho$ be a tie-breaking scheme.

  \par\medskip
  \textbf{Selected active-variable index map.}~
  We inductively define  
  the \emph{selected active-variable index map} of $f$
  relative to its closure derivation $\delta$ under the tie-breaking scheme $\rho$,
  denoted by
  \begin{equation*}
    \selectedvarmap[\delta][\rho]:
    \reals^{d_\mathrm{in}}
    \to
    \iival{1}{{d_\mathrm{in}}}^{{d_\mathrm{out}}},
  \end{equation*}
  as follows:
  \begin{itemize}
    \item
      if 
      $f = \pi_j^{d_\mathrm{in}}$ is the $j$-th coordinate projection,
      \begin{equation*}
        \selectedvarmap[\delta][\rho][x] \defeq j;
      \end{equation*}
    \item
      if 
      $f$ is a Min or Max reduction,
      then 
      \begin{equation*}
        \selectedvarmap[\delta][\rho][x] \defeq \rho_d\big(\activevarmap{f}[x]\big),
      \end{equation*}
      where
      $\activevarmap{f}$
      is 
      the active-variable index map of $f$ as per
      Definition~\ref{def:active-variable-index-function};
    \item
      if 
      $f = \ptuple{f_1, \ldots, f_k}$
      and
      $\delta = \tuple{\mathsf{tupling}, \delta_1, \ldots, \delta_k}$,
      then
      \begin{equation*}
        \selectedvarmap[\delta][\rho][x]
        \defeq
        \bigptuple{
          \sigma_{\delta_1}^\rho(x),
          \ldots,
          \sigma_{\delta_k}^\rho(x)
        };
      \end{equation*}
    \item
      if 
      $f = h \circ g$ 
      and
      $\delta = \tuple{\mathsf{comp}, \delta_h, \delta_g}$,
      then
      \begin{sidealign*}
        \sideeq{
          \biggetentry{\selectedvarmap[\delta][\rho][x]}{i}
          \,\defeq\,
          \biggetentry{
            \sigma_{\delta_g}^\rho(x)
          }{
            \phi(x;i) 
          }
        }{
          for
        }{
          \phi(x;i) \defeq \biggetentry{\sigma_{\delta_h}^\rho(g(x))}{i}
        }[
          \forall i\in\iival{1}{{d_\mathrm{out}}}\,.
        ]
      \end{sidealign*}
  \end{itemize}
  We may write 
  \begin{sidealign*}
    \quanteq{
      \selectedvarmap[\delta][\rho][x; i] 
      \defeq 
      \biggetentry{\selectedvarmap[\delta][\rho][x]}{i}
    }{
      \forall i \in \iival{1}{d_\mathrm{out}}.
    }
  \end{sidealign*}

  \par\medskip
  \textbf{Active map.}~
  The \emph{active map} of $\delta$ under the tie-breaking scheme $\rho$
  is 
  the map $\activevar[\delta][\rho]$ given by
  \begin{sidealign*}
    \sideeq{
      \biggetentry{\activevar[\delta][\rho][x]}{i}
      \,\defeq\,
      \getentry{x}{j}
      }{
        for
      }{
      j \defeq \selectedvarmap[\delta][\rho][x; i]
      }[
        \forall i \in \iival{1}{{d_\mathrm{out}}}\,.
    ]
  \end{sidealign*}
  We may write
  \begin{sidealign*}
    \quanteq{
      \activevar[\delta][\rho][x; i]
      \,\defeq\,
      \biggetentry{\activevar[\delta][\rho][x]}{i}
      }{
      \forall i \in \iival{1}{{d_\mathrm{out}}};
    }
  \end{sidealign*}
  and then, when ${d_\mathrm{out}}=1$, we may also write
  \begin{equation*}
    \activevar[\delta][\rho](x) \defeq \activevar[\delta][\rho](x;1)
    \,.
  \end{equation*}

  When 
  a deterministic tie-breaking scheme $\rho$
  and 
  a MinMax derivation $\delta$
  are clear from the context, 
  we may omit them and hence simply write 
  $\selectedvarmap$
  and
  $\activevar$ 
  in place of 
  $\selectedvarmap[\delta][\rho]$
  and
  $\activevar[\delta][\rho]$, respectively.
\end{closingdefinition}

\begin{proposition}[Active-variable property of MinMax functions]
  \label{prop:minmax-function-active-variable}
  Let
  $f:\reals^{d_\mathrm{in}}\to\reals^{d_\mathrm{out}}$
  be a MinMax function,
  let
  $\delta$
  be a MinMax derivation of $f$,
  and let
  $\rho$
  be a deterministic tie-breaking scheme.
  Then
  \begin{sidealign*}
    \quanteq{
      \getentry{f(x)}{\ell}
      =
      \activevar[\delta][\rho][x;\ell]
    }{
      \forall
      x\in\reals^{d_\mathrm{in}},\;
      \forall
      \ell\in\iival{1}{{d_\mathrm{out}}}.
    }
  \end{sidealign*}
\end{proposition}
\begin{proof}
  The claim follows by induction on the MinMax derivation
  $\delta$.

  For coordinate projections, it holds by definition.
  For Min and Max reductions, it follows from
  Definition~\ref{def:active-variable-index-function}.
  The tupling case follows coordinatewise.

  Finally, suppose
  $f=h\circ g$,
  with corresponding derivations
  $\delta_h$
  and
  $\delta_g$.
  For every output coordinate
  $\ell \in \iival{1}{{d_\mathrm{out}}}$,
  let
  \begin{equation*}
    j
    \defeq
    \getentry{
      \sigma_{\delta_h}^\rho(g(x))
    }{\ell}.
  \end{equation*}
  By the induction hypothesis for $h$,
  \begin{equation*}
    \getentry{f(x)}{\ell}
    =
    \getentry{h(g(x))}{\ell}
    =
    \getentry{g(x)}{j}.
  \end{equation*}
  Applying the induction hypothesis for
  $g$
  to its
  $j$-th coordinate gives
  \begin{equation*}
    \getentry{g(x)}{j}
    =
    \activevar[\delta_g][\rho](x;j).
  \end{equation*}
  By the definition of the active-variable function for composition,
  \begin{equation*}
    \activevar[\delta_g][\rho](x;j)
    =
    \activevar[\delta][\rho][x;\ell],
  \end{equation*}
  which proves the claim.
\end{proof}

\subsection{Piecewise Propagate-Reset Structure of MinMax Recurrences}
\label{subsec:piecewise-copyset-representations}

\subsubsection{Context and Structural Notions}
\label{subsubsec:piecewise-copyset-representations--context}

Let us consider
spaces $X \subseteq \reals^n,\, U \subseteq \reals^m$,
and
functions $R : U \to \reals^{n \times n},\, s : U \to \reals^n$.
Then,
we introduce the corresponding MinMax dynamics function
\begin{equation}
  \label{eq:minmax-recurrence}
  f: X \times U \to \reals^n,
  \qquad\qquad
  f(x,u)
  \,\defeq\,
  \big(R(u)\otimes x\big)\oplus s(u),
\end{equation}
which can be written coordinate-wise as
\begin{equation*}
  \biggetentry{f(x,u)}{i}
  \,=\,
  \left(\,
    \bigoplus_{j=1}^n
    \biggetentry{R(u)}{i,j} \odot \biggetentry{x}{j}
  \right)
  \oplus
  \biggetentry{s(u)}{i}
  \qquad\quad 
  \forall i \in \iival{1}{n}.
\end{equation*}

\begin{closingdefinition}[State-affine dynamics function]
  \label{def:state-affine-dynamics-function}
  The considered dynamics function
  $f$
  is
  \emph{state-affine}
  if there exist 
  $A: U \to \reals^{n\times n}$
  and
  $b:U\to\reals^n$
  such that
  \begin{sidealign*}
    \quanteq{
      f(x,u)
      =
      A(u) \cdot x+b(u)
    }{
      \forall x\in X,\;
      \forall u\in U.
    }
  \end{sidealign*}
  Furthermore,
  it is
  \emph{state-affine with input-independent linear coefficient}
  if $A$ is constant;
  in this case, we call
  $\tuple{A,b}$
  a
  \emph{linear-coefficient--offset-function pair for $f$},
  or simply a
  \emph{coefficient-offset pair for $f$}.
\end{closingdefinition}

\begin{closingdefinition}[Propagate-reset state-affine map]
  \label{def:propagate-reset-state-affine}
  \label{def:copy-reset-state-affine}
  The dynamics function
  $f$
  is
  \emph{propagate-reset}
  if it is state-affine with input-independent linear coefficient
  and additionally, for $\tuple{A,b}$ its coefficient-offset pair,
  the following consditions hold jointly:
  \begin{enumerate}
    \item
      $A$
      is deterministic row-substochastic;%
      \footnote{%
        The notion of
        \emph{deterministic row-substochastic}
        matrix is introduced in
        Definition~\ref{def:deterministic-row-substochastic}.%
      }
    \item
      the following implications hold:
      \begin{sidealign*}
        \quanteq{
          \getentry{A}{i,j}=1
          \quad\Longrightarrow\quad
          \getentry{b(u)}{i}=0
        }{
          \forall i,j\in\iival{1}{n},\;
          \forall u\in U.
        }
      \end{sidealign*}
  \end{enumerate}
\end{closingdefinition}

\begin{closingremark}
  Every output coordinate produced by a propagate-reset map
  is either propagated from a previous state coordinate or set to the
  corresponding coordinate of its offset function.
  In particular, the two contributions are mutually exclusive:
  whenever
  $\getrow{A}{i}\neq0_n^\top$,
  we have
  $\getentry{b(u)}{i}=0$.
\end{closingremark}

\subsubsection{Source Variables and Active-Variable Regions}
\label{subsubsec:minmax-source-variables}

\begin{closingdefinition}[Flat MinMax recurrence function]
  \label{def:minmax-coordinate-function}
  The
  \emph{flat MinMax recurrence function} 
  is
  \begin{align}
    \label{eq:minmax-coordinate-minmax-function}
    \flatminmaxfunc:\reals^{2n{+}1}\to\reals,
      \qquad\qquad
      \flatminmaxfunc(z)
      \,\defeq\,
      \getentry{z}{1}
      \oplus
      \bigoplus_{j=1}^{n}
      \big(
      \getentry{z}{j{+}1}
        \odot
        \getentry{z}{n{+}j{+}1}
      \big).
  \end{align}
\end{closingdefinition}

\begin{closingdefinition}[Source-variable label map]
  \label{def:minmax-source-variable-label-map}
  For every
  $i\in\iival{1}{n}$,
  we define the
  \emph{$i$-th source-variable label map}
  \begin{equation*}
    \sourcevar_i
    \;:\;
    \iival{1}{2n {+} 1}
    \;\to\;
    \mBig\{
      \getentry{s}{i}\,,\,
      \getentry{R}{i,1}\,,
      \ldots,\,
      \getentry{R}{i,n}\,,\,
      \getentry{x}{1}\,,\,
      \ldots,\,
      \getentry{x}{n}
    \mBig\}
  \end{equation*}
  by
  \begin{equation*}
    \sourcevar_i(k)
    \defeq
    \begin{cases}
      \,\getentry{s}{i}
      &\, \text{ if }\, k=1,
      \\[.3em]
      \getentry{R}{i,j}
      &\, \text{ if }\, k = 1 {+} j \,\text{ and }\, j\in\iival{1}{n},
      \\[.3em]
      \,\getentry{x}{j}
      &\, \text{ if }\, k = 1 {+} n {+} j \,\text{ and }\, j\in\iival{1}{n}.
    \end{cases}
  \end{equation*}
\end{closingdefinition}

\begin{closingdefinition}[MinMax source-variable maps]
  \label{def:minmax-source-variable-maps}
  For every
  $i\in\iival{1}{n}$,
  the \emph{$i$-th source-variable map} 
  is
  the function
  $\sourceValue_i:X\times U\to\reals^{2n {+} 1}$
  given by
  \begin{align*}
    \sourceValue_i(x,u)
    \defeq
    \begin{pmatrix}
      \biggetentry{s(u)}{i} 
      &
      \biggetentry{R(u)}{i,1}
      &
      \cdots
      &
      \biggetentry{R(u)}{i,n}
      &
      \getentry{x}{1}
      &
      \cdots
      &
      \getentry{x}{n}
    \end{pmatrix}.
  \end{align*}
  Then,
  the \emph{source-variable maps} for the considered MinMax dynamics function $f$ are 
  $(\sourceValue_i)_{i=1}^n$.
\end{closingdefinition}

\begin{closingdefinition}[Active-variable pattern and partition of a MinMax recurrence]
  \label{def:minmax-active-variable-partition}
  Let $\flatminmaxfunc$ be the flat MinMax recurrence function of Definition~\ref{def:minmax-coordinate-function},
  let $(\sourceValue_i)_{i=1}^n$ be the source-variable maps of
  Definition~\ref{def:minmax-source-variable-maps},
  let $\delta$ be the MinMax closure derivation corresponding to the expression
  in~\eqref{eq:minmax-coordinate-minmax-function},
  and 
  let $\rho$ be a tie-breaking scheme.
  Let us write
  $\selectedvarmap \defeq \selectedvarmap[\delta][\rho]$.

  We define the
  \emph{active-variable pattern map}
  \begin{equation*}
    \alpha
    :
    X\times U
    \to
    \iival{1}{2n {+} 1}^{n},
    \quad\;
    \alpha(x,u)
    \defeq
    \Bigptuple{
      \selectedvarmap\bigl(\sourceValue_1(x,u)\bigr),
      \ldots,
      \selectedvarmap\bigl(\sourceValue_n(x,u)\bigr)
    }.
  \end{equation*}
  Then,
  writing 
  \begin{align*}
    \hat{\alpha}(v) \defeq \alpha^{{-}1}(V_v) \quad\text{ for }\quad V_v \defeq \{ v\},
  \end{align*}
  we define the corresponding \emph{active-variable partition}
  \begin{equation*}
    \partitionstyle{Z}
    \defeq
    \mBig\{
      \hat{\alpha}(v)
      \bigm|
      v \in\iival{1}{2n {+} 1}^{n},\;
      \hat{\alpha}(v) \neq \emptyset
    \mBig\}.
  \end{equation*}
  We call every $Z\in\partitionstyle{Z}$ 
  an \emph{active-variable region} of $f$ induced by $\delta$ and $\rho$.
\end{closingdefinition}

\begin{closingfact}[Finiteness of the active-variable partition]
  \label{fact:minmax-active-variable-partition-finite}
  The set $\partitionstyle{Z}$ of
  Definition~\ref{def:minmax-active-variable-partition}
  is a finite partition of $X\times U$.
  In particular, its cardinality is bounded as
  \begin{equation*}
    \big|
      \partitionstyle{Z}
    \big|
    \leq
    (2n {+} 1)^n.
  \end{equation*}
\end{closingfact}
\begin{proof}
  The active-variable pattern map
  \begin{equation*}
    \alpha
    :
    X\times U
    \to
    \iival{1}{2n {+} 1}^{\,n}
  \end{equation*}
  has finite codomain of cardinality
  $(2n {+} 1)^n$.
  Its nonempty fibres are pairwise disjoint and cover
  $X\times U$.
  By definition, these nonempty fibres are precisely the elements of
  $\partitionstyle{Z}$.
\end{proof}

\begin{closingdefinition}[Regional active source variable]
  \label{def:minmax-regional-active-source-variable}

  Let $\alpha$ and $\partitionstyle{Z}$ 
  be the 
  active-variable pattern map
  and
  active-variable partition of
  Definition~\ref{def:minmax-active-variable-partition}, respectively.

  Let $Z\in\partitionstyle{Z}$.

  Since $Z$ is an active-variable region,
  there exists a unique
  \begin{equation*}
    a^Z\in\iival{1}{2n {+} 1}^{n}
  \end{equation*}
  such that
  \begin{sidealign*}
    \quanteq{
      \alpha(x,u)
      =
      a^Z
    }{
      \forall (x,u)\in Z.
    }
  \end{sidealign*}

  For every
  $i\in\iival{1}{n}$,
  we define
  \begin{equation*}
    a_i^Z
    \defeq
    \getentry{a^Z}{i}
  \end{equation*}
  and the
  \emph{active source variable of $f$ for coordinate $i$ on $Z$}
  \begin{equation*}
    \activeSource(f;Z,i)
    \defeq
    \sourcevar_i(a_i^Z).
  \end{equation*}
\end{closingdefinition}

\begin{closingfact}[Coordinate representation of MinMax recurrences]
  \label{fact:minmax-coordinate-representation}
  Let
  $\flatminmaxfunc$
  be the flat MinMax recurrence function of
  Definition~\ref{def:minmax-coordinate-function},
  and let
  $(\sourceValue_i)_{i=1}^n$
  be the source-variable maps of
  Definition~\ref{def:minmax-source-variable-maps}.
  Then,
  \begin{align*}
    \biggetentry{f(x,u)}{i}
    \,=\,
    \flatminmaxfunc\bigl(\sourceValue_i(x,u)\bigr)
    \qquad\quad
    \forall i\in\iival{1}{n},
    \;
    \forall (x,u)\in X\times U.
  \end{align*}
\end{closingfact}
\begin{proof}
  By
  Definitions~\ref{def:minmax-coordinate-function}
  and~\ref{def:minmax-source-variable-maps},
  for every $i \in \iival{1}{n}$,
  \begin{align*}
    \flatminmaxfunc\bigl(\sourceValue_i(x,u)\bigr)
    \;=\;
    \getentry{s(u)}{i}
    \oplus
    \bigoplus_{j=1}^{n}
    \left(
      \getentry{R(u)}{i,j}
      \odot
      \getentry{x}{j}
    \right)
    \;=\;
    \getentry{f(x,u)}{i},
  \end{align*}
  where the last equality follows from
  Equation~\eqref{eq:minmax-recurrence}.
\end{proof}

\subsubsection{Piecewise Propagate-Reset Representation}
\label{subsubsec:minmax-piecewise-propagate-reset}

\begin{proposition}[Piecewise propagate-reset representation of MinMax recurrences]
  \label{prop:minmax-piecewise-copyset}
  Let
  $\partitionstyle{Z}$
  be the active-variable partition of
  Definition~\ref{def:minmax-active-variable-partition}.

  There exists a piecewise representation
  $\gamma = \{ \tuple{Z,f_Z} \mid Z\in\partitionstyle{Z} \}$
  on
  $(X\times U,\reals^n)$
  such that:
  \begin{enumerate}
    \item
      $\gamma$
      has input partition
      $\partitionstyle{Z}$;

    \item
      $\gamma$
      represents
      $f$;

    \item
      $\gamma$
      is affine in the first argument;

    \item
      every piece
      $f_Z$
      is a propagate-reset map.
  \end{enumerate}

  Equivalently, there exists a coefficient-based piecewise representation
  $\hat{\gamma} = \{ \tuple{Z,A_Z,b_Z} \mid Z\in\partitionstyle{Z} \}$
  on
  $(X,U,\reals^n)$
  that is affine in the first argument,
  represents
  $f$,
  and satisfies the following property:
  for every
  $Z\in\partitionstyle{Z}$,
  the coefficient-offset pair
  $\tuple{A_Z,b_Z}$
  defines a propagate-reset map
  \begin{equation}
    \label{eq:minmax-regional-linear}
    f_Z(x,u)
    \defeq
    A_Z\cdot x+b_Z(u).
  \end{equation}

  The representation
  $\gamma$
  can be chosen as the canonical piecewise representation induced by
  $\hat{\gamma}$;
  in particular,
  \begin{equation*}
    \repval{\gamma}
    =
    \repval{\hat{\gamma}}
    =
    f.
  \end{equation*}
\end{proposition}
\begin{proof}
  By
  Fact~\ref{fact:minmax-active-variable-partition-finite},
  $\partitionstyle{Z}$
  is a finite partition of
  $X\times U$.

  Fix
  $Z\in\partitionstyle{Z}$.
  For every
  $i\in\iival{1}{n}$,
  let
  $a_i^Z$
  be the regional active-variable index of
  Definition~\ref{def:minmax-regional-active-source-variable}.

  By
  Definitions~\ref{def:minmax-active-variable-partition}
  and~\ref{def:minmax-regional-active-source-variable},
  \begin{sidealign*}
    \quanteq{
      \selectedvarmap[\delta][\rho]
      \bigl(
        \sourceValue_i(x,u)
      \bigr)
      =
      a_i^Z
    }{
      \forall (x,u)\in Z.
    }
  \end{sidealign*}

  By
  Fact~\ref{fact:minmax-coordinate-representation}
  and
  Proposition~\ref{prop:minmax-function-active-variable},
  we have
  \begin{align}
    \label{eq:minmax-coordinate-active-variable}
    \getentry{f(x,u)}{i}
    &=
    \flatminmaxfunc
    \bigl(
      \sourceValue_i(x,u)
    \bigr)
    \notag
    \\
    &=
    \activevar[\delta][\rho]
    \bigl(
      \sourceValue_i(x,u);1
    \bigr)
    \notag
    \\
    &=
    \getentry{
      \sourceValue_i(x,u)
    }{
      a_i^Z
    }
    \qquad
    \forall (x,u)\in Z.
  \end{align}

  We now define
  \begin{equation*}
    A_Z
    \in
    \reals^{n\times n}
  \end{equation*}
  and
  \begin{equation*}
    b_Z:U\to\reals^n
  \end{equation*}
  row by row.
  For every
  $i\in\iival{1}{n}$,
  set
  \begin{equation}
    \label{eq:minmax-regional-propagate-reset-definition}
    \ptuple{
      \getrow{A_Z}{i},
      \getentry{b_Z(u)}{i}
    }
    \defeq
    \begin{cases}
      \ptuple{
        0_n^\top,
        \getentry{s(u)}{i}
      },
      &
      a_i^Z=1,
      \\[.4em]
      \ptuple{
        0_n^\top,
        \getentry{R(u)}{i,j}
      },
      &
      a_i^Z\,=\, {1{+}j},
      \quad
      j\in\iival{1}{n},
      \\[.4em]
      \ptuple{
        e_j^\top,
        0
      },
      &
      a_i^Z \,=\, {1{+}n{+}j},
      \quad
      j\in\iival{1}{n}.
    \end{cases}
  \end{equation}
  Since
  \begin{equation*}
    a_i^Z
    \in
    \iival{1}{2n{+}1},
  \end{equation*}
  exactly one of these cases applies.

  Define
  \begin{equation*}
    f_Z:X\times U\to\reals^n,
    \qquad
    f_Z(x,u)
    \defeq
    A_Z\cdot x+b_Z(u).
  \end{equation*}
  By
  Equation~\eqref{eq:minmax-coordinate-active-variable}
  and
  Definition~\ref{def:minmax-source-variable-maps},
  the
  $i$-th coordinate of
  $f_Z(x,u)$
  is precisely
  \begin{equation*}
    \getentry{
      \sourceValue_i(x,u)
    }{
      a_i^Z
    }
  \end{equation*}
  for every
  $(x,u)\in Z$.
  Hence,
  coordinatewise,
  \begin{sidealign*}
    \quanteq{
      f(x,u)
      =
      f_Z(x,u)
    }{
      \forall (x,u)\in Z.
    }
  \end{sidealign*}

  Therefore,
  $\hat{\gamma} = \{ \tuple{Z,A_Z,b_Z} \mid Z\in\partitionstyle{Z} \}$
  is a coefficient-based piecewise representation on
  $(X,U,\reals^n)$
  that is affine in the first argument.
  By its semantics,
  \begin{equation*}
    \repval{\hat{\gamma}}
    =
    f.
  \end{equation*}

  By
  Fact~\ref{fact:piecewise-uniformly-affine-coefficient-representation}.2,
  $\hat{\gamma}$
  induces the canonical piecewise representation
  $\gamma = \{ \tuple{Z,f_Z} \mid Z\in\partitionstyle{Z} \}$
  which is affine in the first argument and satisfies
  \begin{equation*}
    \repval{\gamma}
    =
    \repval{\hat{\gamma}}
    =
    f.
  \end{equation*}

  It remains to show that every
  $f_Z$
  is propagate-reset.
  By
  Equation~\eqref{eq:minmax-regional-propagate-reset-definition},
  every row of
  $A_Z$
  is either
  $0_n^\top$
  or a standard basis row
  $e_j^\top$.
  Hence
  $A_Z$
  is deterministic row-substochastic.

  Furthermore, if
  \begin{equation*}
    \getentry{A_Z}{i,j}
    =
    1,
  \end{equation*}
  then the third case of
  Equation~\eqref{eq:minmax-regional-propagate-reset-definition}
  applies to row
  $i$,
  and therefore
  \begin{equation*}
    \getentry{b_Z(u)}{i}
    =
    0
    \qquad
    \forall u\in U.
  \end{equation*}
  Thus
  $\tuple{A_Z,b_Z}$
  satisfies the propagate-reset conditions of
  Definition~\ref{def:propagate-reset-state-affine}.

  Therefore every piece
  $f_Z$
  is a propagate-reset map,
  completing the proof.
\end{proof}

\subsubsection{State-Coordinate Behaviours}
\label{subsubsec:minmax-state-behaviours}

\begin{closingdefinition}[State-propagation behaviour]
  \label{def:minmax-state-propagation-behaviour}
  Let
  $Z\in\partitionstyle{Z}$
  be an active-variable region,
  and let
  $i\in\iival{1}{n}$.
  We say that the
  $i$-th state coordinate exhibits
  \emph{state-propagation behaviour through $f$ on $Z$}
  if there exists
  $j\in\iival{1}{n}$
  such that
  \begin{equation*}
    \activeSource(f;Z,i)
    =
    \getentry{x}{j}.
  \end{equation*}
  In this case, the
  $j$-th previous state coordinate is the
  \emph{propagated state coordinate}.
\end{closingdefinition}

\begin{closingfact}[Characterisation of state-propagation behaviour]
  \label{fact:minmax-state-propagation-behaviour}
  Consider the coefficient-based piecewise representation
  $\hat{\gamma} = \{ \tuple{Z,A_Z,b_Z} \mid Z\in\partitionstyle{Z} \}$
  of
  Proposition~\ref{prop:minmax-piecewise-copyset}.
  Suppose that the
  $i$-th state coordinate exhibits
  state-propagation behaviour through
  $f$
  on
  $Z$
  with propagated state coordinate
  $j\in\iival{1}{n}$.

  \begin{itemize}
    \item
      First,
      we have
      \begin{align*}
        \getentry{f(x,u)}{i}
        =
        \getentry{x}{j}
        \qquad\quad
        \forall (x,u)\in Z.
      \end{align*}

    \item
      Second,
      we have
      \begin{align*}
        \getrow{A_Z}{i}
        =
        e_j^\top
        \qquad\;\text{ and }\qquad\;
        \getentry{b_Z(u)}{i}
        =
        0
        \qquad
        \forall u\in U.
      \end{align*}

    \item
      Third,
      the propagated state value satisfies
      \begin{align}
        \label{eq:minmax-copied-value-bracketing}
        \getentry{s(u)}{i}
        \leq
        \getentry{x}{j}
        \leq
        \getentry{R(u)}{i,j}
        \qquad\quad
        \forall (x,u)\in Z.
      \end{align}
  \end{itemize}
\end{closingfact}
\begin{proof}
  By
  Definition~\ref{def:minmax-regional-active-source-variable},
  \begin{equation*}
    \activeSource(f;Z,i)
    =
    \sourcevar_i(a_i^Z).
  \end{equation*}
  Hence, by
  Definition~\ref{def:minmax-source-variable-label-map},
  \begin{equation*}
    \activeSource(f;Z,i)
    =
    \getentry{x}{j}
    \qquad\Longrightarrow\qquad
    a_i^Z
    \,=\,
    {1{+}n{+}j}.
  \end{equation*}

  Therefore,
  Equation~\eqref{eq:minmax-coordinate-active-variable}
  gives
  \begin{sidealign*}
    \quanteq{
      \getentry{f(x,u)}{i}
      =
      \getentry{x}{j}
      }{
      \forall (x,u)\in Z.
    }
  \end{sidealign*}
  Furthermore,
  Equation~\eqref{eq:minmax-regional-propagate-reset-definition}
  gives
  \begin{align*}
    \getrow{A_Z}{i}
    & =
    e_j^\top\,,
    \\[.2em]
    \getentry{b_Z(u)}{i}
    & =
    0
    \qquad\qquad
    \forall u\in U.
  \end{align*}

  It remains to prove
  Equation~\eqref{eq:minmax-copied-value-bracketing}.
  Since
  $\getentry{x}{j}$
  is the active source variable,
  it is selected by the inner Min operation
  \begin{equation*}
    \getentry{R(u)}{i,j}
    \odot
    \getentry{x}{j}.
  \end{equation*}
  Hence
  \begin{equation*}
    \getentry{x}{j}
    \leq
    \getentry{R(u)}{i,j}.
  \end{equation*}
  The resulting value is subsequently selected by the outer Max operation,
  whose arguments include
  $\getentry{s(u)}{i}$.
  Hence
  \begin{equation*}
    \getentry{s(u)}{i}
    \leq
    \getentry{x}{j}.
  \end{equation*}
  Combining the two inequalities proves
  Equation~\eqref{eq:minmax-copied-value-bracketing}.
\end{proof}

\begin{closingdefinition}[State-reset behaviour]
  \label{def:minmax-state-reset-behaviour}
  Let
  $Z\in\partitionstyle{Z}$
  be an active-variable region,
  and let
  $i\in\iival{1}{n}$.

  We say that the
  $i$-th state coordinate exhibits
  \emph{state-reset behaviour through $f$ on $Z$}
  if
  \begin{equation*}
    \activeSource(f;Z,i)
    \in
    \mBig\{
      \getentry{s}{i},
      \getentry{R}{i,1},
      \ldots,
      \getentry{R}{i,n}
    \mBig\}.
  \end{equation*}
\end{closingdefinition}

\begin{closingfact}[Characterisation of state-reset behaviour]
  \label{fact:minmax-state-reset-behaviour}
  Consider the coefficient-based piecewise representation
  $\hat{\gamma} = \{ \tuple{Z,A_Z,b_Z} \mid Z\in\partitionstyle{Z} \}$
  of
  Proposition~\ref{prop:minmax-piecewise-copyset}.

  Suppose that the
  $i$-th state coordinate exhibits
  state-reset behaviour through
  $f$
  on
  $Z$.
  Then
  \begin{equation*}
    \getrow{A_Z}{i}
    =
    0_n^\top,
  \end{equation*}
  and exactly one of the following two cases holds:
  \begin{itemize}
    \item
      we have
      \begin{equation*}
        \activeSource(f;Z,i)
        =
        \getentry{s}{i},
      \end{equation*}
      in which case
      \begin{align*}
        \getentry{f(x,u)}{i}
        &=
        \getentry{s(u)}{i}
        \qquad\quad
        \forall (x,u)\in Z,
        \\
        \getentry{b_Z(u)}{i}
        &=
        \getentry{s(u)}{i}
        \qquad\quad
        \forall u\in U;
      \end{align*}

    \item
      there exists
      $j\in\iival{1}{n}$
      such that
      \begin{equation*}
        \activeSource(f;Z,i)
        =
        \getentry{R}{i,j},
      \end{equation*}
      in which case
      \begin{align*}
        \getentry{f(x,u)}{i}
        &=
        \getentry{R(u)}{i,j}
        \qquad\quad
        \forall (x,u)\in Z,
        \\
        \getentry{b_Z(u)}{i}
        &=
        \getentry{R(u)}{i,j}
        \qquad\quad
        \forall u\in U.
      \end{align*}
  \end{itemize}
\end{closingfact}
\begin{closingremark}
  Thus, under state-reset behaviour,
  the updated state coordinate is independent of the previous state
  and is set directly to an input-dependent source value.
\end{closingremark}
\begin{proof}
  By
  Definition~\ref{def:minmax-regional-active-source-variable},
  \begin{equation*}
    \activeSource(f;Z,i)
    =
    \sourcevar_i(a_i^Z).
  \end{equation*}

  Suppose first that
  \begin{equation*}
    \activeSource(f;Z,i)
    =
    \getentry{s}{i}.
  \end{equation*}
  By
  Definition~\ref{def:minmax-source-variable-label-map},
  we have
  \begin{equation*}
    a_i^Z
    =
    1.
  \end{equation*}
  Therefore,
  Equations~\eqref{eq:minmax-coordinate-active-variable}
  and~\eqref{eq:minmax-regional-propagate-reset-definition}
  yield
  \begin{align*}
    \getentry{f(x,u)}{i}
    &=
    \getentry{s(u)}{i}
    \qquad
    \forall (x,u)\in Z,
    \\
    \getrow{A_Z}{i}
    &=
    0_n^\top,
    \qquad
    \getentry{b_Z(u)}{i}
    =
    \getentry{s(u)}{i}
    \qquad
    \forall u\in U.
  \end{align*}

  Otherwise,
  state-reset behaviour implies that there exists
  $j\in\iival{1}{n}$
  such that
  \begin{equation*}
    \activeSource(f;Z,i)
    =
    \getentry{R}{i,j}.
  \end{equation*}
  By
  Definition~\ref{def:minmax-source-variable-label-map},
  \begin{equation*}
    a_i^Z
    \,=\,
    {1{+}j}.
  \end{equation*}
  Therefore,
  Equations~\eqref{eq:minmax-coordinate-active-variable}
  and~\eqref{eq:minmax-regional-propagate-reset-definition}
  yield
  \begin{align*}
    \getentry{f(x,u)}{i}
    &=
    \getentry{R(u)}{i,j}
    \qquad\qquad
    \forall (x,u)\in Z,
    \\[.4em]
    \getrow{A_Z}{i}
    &=
    0_n^\top\,,
    \\[.2em]
    \getentry{b_Z(u)}{i}
    & =
    \getentry{R(u)}{i,j}
    \qquad\qquad
    \forall u\in U.
  \end{align*}
\end{proof}

\begin{closingproposition}[State propagation--reset dichotomy]
  \label{prop:minmax-state-propagation-reset-dichotomy}
  \label{thm:minmax-state-propagation-reset-dichotomy}
  For every
  $Z\in\partitionstyle{Z}$
  and every
  $i\in\iival{1}{n}$,
  exactly one of the following two behaviours occurs:
  \begin{enumerate}
    \item
      the
      $i$-th state coordinate exhibits
      state-propagation behaviour through
      $f$
      on
      $Z$;

    \item
      the
      $i$-th state coordinate exhibits
      state-reset behaviour through
      $f$
      on
      $Z$.
  \end{enumerate}
\end{closingproposition}
\begin{proof}
  Fix
  $Z\in\partitionstyle{Z}$
  and
  $i\in\iival{1}{n}$.

  By
  Definition~\ref{def:minmax-regional-active-source-variable},
  \begin{equation*}
    \activeSource(f;Z,i)
    =
    \sourcevar_i(a_i^Z).
  \end{equation*}
  By
  Definition~\ref{def:minmax-source-variable-label-map},
  exactly one of the following holds:
  \begin{equation*}
    \activeSource(f;Z,i)
    \in
    \mBig\{
      \getentry{s}{i},
      \getentry{R}{i,1},
      \ldots,
      \getentry{R}{i,n}
    \mBig\},
  \end{equation*}
  or there exists
  $j\in\iival{1}{n}$
  such that
  \begin{equation*}
    \activeSource(f;Z,i)
    =
    \getentry{x}{j}.
  \end{equation*}

  By
  Definitions~\ref{def:minmax-state-reset-behaviour}
  and~\ref{def:minmax-state-propagation-behaviour},
  these alternatives are precisely
  state-reset behaviour and
  state-propagation behaviour,
  respectively.
\end{proof}

\subsubsection{Regularity Consequences}
\label{subsubsec:minmax-regularity-consequences}

\begin{closingfact}[Piecewise-affinity of MinMax state transformations]
  \label{fact:minmax-state-transformation-piecewise-affine}
  For every $u\in U$,
  the state transformation $f_u(x) \defeq f(x,u)$ is piecewise affine.
\end{closingfact}
\begin{proof}
  By
  Proposition~\ref{prop:minmax-piecewise-copyset},
  the dynamics function
  $f$
  admits a piecewise representation that is affine in its first argument.
  The claim therefore follows directly from
  Fact~\ref{fact:piecewise-uniformly-affine-fixed-second-argument}.
\end{proof}

\begin{proposition}[Piecewise-affine regularity of MinMax recurrences]
  \label{prop:minmax-piecewise-affine}
  Assume that
  the reset $R$
  and 
  set $s$ 
  functions are piecewise affine.

  Then, the MinMax dynamics function $f$ is piecewise affine;
  namely, for
  $\hat{\gamma} = \{ \tuple{Z,A_Z,b_Z} \mid Z\in\partitionstyle{Z} \}$
  the coefficient-based piecewise propagate-reset representation of
  Proposition~\ref{prop:minmax-piecewise-copyset},
  there exists a piecewise-affine coefficient-based representation
  \begin{equation*}
    \hat{\gamma}_f
    \defeq
    \big\{
      \tuple{Q,C_Q,c_Q}
    \big\}_{Q\in\partitionstyle{Q}}
  \end{equation*}
  representing
  $f$
  such that
  $\partitionstyle{Q} \preceq \partitionstyle{Z}$.

  For every
  $Q\in\partitionstyle{Q}$,
  let
  $Z_Q\in\partitionstyle{Z}$
  be the unique active-variable region satisfying
  $Q\subseteq Z_Q$.
  Then there exists
  $B_Q\in\reals^{n\times m}$
  such that
  \begin{equation*}
    C_Q
    =
    \begin{pmatrix}
      A_{Z_Q} & B_Q
    \end{pmatrix},
  \end{equation*}
  and
  \begin{align*}
    f(x,u)
    \,=\,
    A_{Z_Q}\cdot x
    +
    B_Q\cdot u
    +
    c_Q
    \qquad\quad
    \forall (x,u)\in Q.
  \end{align*}
\end{proposition}
\begin{closingremark}
  Thus, all regions $Q$ refining the same active-variable region $Z$
  share the same state-linear coefficient $A_Z$.
\end{closingremark}
\begin{proof}
  Since
  $R$
  and
  $s$
  are piecewise affine,
  each of their finitely many scalar coordinate functions
  \begin{equation*}
    u\mapsto\getentry{s(u)}{i},
    \qquad
    u\mapsto\getentry{R(u)}{i,j}
  \end{equation*}
  is piecewise affine.
  Taking a common refinement of the corresponding finite input partitions
  yields a finite partition
  $\partitionstyle{P}$
  of
  $U$
  such that all these scalar functions are affine on every
  $P\in\partitionstyle{P}$.

  Define
  \begin{equation*}
    \partitionstyle{Q}
    \defeq
    \left\{
      Z\cap(X\times P)
      \;\middle|\;
      Z\in\partitionstyle{Z},
      \;
      P\in\partitionstyle{P},
      \;
      Z\cap(X\times P)\neq\emptyset
    \right\}.
  \end{equation*}
  This is a finite partition of
  $X\times U$
  satisfying
  $\partitionstyle{Q}\preceq\partitionstyle{Z}$.

  Fix
  \begin{equation*}
    Q
    =
    Z\cap(X\times P)
    \in
    \partitionstyle{Q}.
  \end{equation*}
  By
  Proposition~\ref{prop:minmax-piecewise-copyset},
  \begin{sidealign*}
    \quanteq{
      f(x,u)
      =
      A_Z\cdot x+b_Z(u)
    }{
      \forall (x,u)\in Z.
    }
  \end{sidealign*}

  By
  Equation~\eqref{eq:minmax-regional-propagate-reset-definition},
  every coordinate of
  $b_Z$
  is either identically zero or equal to one of
  \begin{equation*}
    u\mapsto\getentry{s(u)}{i},
    \qquad
    u\mapsto\getentry{R(u)}{i,j}.
  \end{equation*}
  Hence
  $b_Z$
  is affine on
  $P$.
  Therefore there exist
  \begin{equation*}
    B_Q\in\reals^{n\times m},
    \qquad
    c_Q\in\reals^n
  \end{equation*}
  such that
  \begin{sidealign*}
    \quanteq{
      b_Z(u)
      =
      B_Q\cdot u+c_Q
    }{
      \forall u\in P.
    }
  \end{sidealign*}

  Consequently,
  \begin{sidealign*}
    \quanteq{
      f(x,u)
      =
      A_Z\cdot x
      +
      B_Q\cdot u
      +
      c_Q
    }{
      \forall (x,u)\in Q.
    }
  \end{sidealign*}

  Define
  \begin{equation*}
    C_Q
    \defeq
    \begin{bmatrix}
      A_Z & B_Q
    \end{bmatrix}.
  \end{equation*}
  Then
  \begin{equation*}
    \hat{\gamma}_f
    \defeq
    \big\{
      \tuple{Q,C_Q,c_Q}
    \big\}_{Q\in\partitionstyle{Q}}
  \end{equation*}
  is a piecewise-affine coefficient-based representation of
  $f$.
  By
  Fact~\ref{fact:piecewise-affine-coefficient-representation}.2,
  it induces a canonical piecewise-affine representation of
  $f$.
  Hence
  $f$
  is piecewise affine.

  Finally,
  \begin{equation*}
    C_Q
    =
    \begin{bmatrix}
      A_Z & B_Q
    \end{bmatrix}
  \end{equation*}
  shows that all refined regions contained in the same
  $Z$
  share the same state-linear coefficient
  $A_Z$.
\end{proof}

\begin{proposition}[PAP regularity of MinMax recurrences]
  \label{prop:minmax-pap}
  Suppose that
  the reset $R$
  and 
  set $s$ 
  functions are PAP.
  Then the dynamics function $f$ is PAP.
  In particular, 
  if
  $R$ and $s$ are analytic, 
  then 
  $f$ is PAP.
\end{proposition}
\begin{proof}
  Let
  $\flatminmaxfunc$
  be the flat MinMax recurrence function of
  Definition~\ref{def:minmax-coordinate-function},
  and let
  $(\sourceValue_i)_{i=1}^n$
  be the source-variable maps of
  Definition~\ref{def:minmax-source-variable-maps}.

  Coordinate projections are analytic and hence PAP.
  Since PAP functions are closed under pointwise minimum and maximum,
  Equation~\eqref{eq:minmax-coordinate-minmax-function}
  implies that
  $\flatminmaxfunc$
  is PAP.

  Fix
  $i\in\iival{1}{n}$.
  By
  Definition~\ref{def:minmax-source-variable-maps},
  \begin{equation*}
    \sourceValue_i(x,u)
    =
    \Bigptuple{
      \biggetentry{s(u)}{i},
      \biggetentry{R(u)}{i,1},
      \ldots,
      \biggetentry{R(u)}{i,n},
      \getentry{x}{1},
      \ldots,
      \getentry{x}{n}
    }.
  \end{equation*}
  Its state-coordinate components are analytic and hence PAP,
  while its remaining components are PAP by hypothesis.
  Closure under finite tupling therefore implies that
  $\sourceValue_i$
  is PAP.

  Since PAP maps are closed under composition,
  $\flatminmaxfunc\circ\sourceValue_i$
  is PAP.
  By
  Fact~\ref{fact:minmax-coordinate-representation},
  \begin{equation*}
    \getentry{f(x,u)}{i}
    =
    \flatminmaxfunc\bigl(\sourceValue_i(x,u)\bigr),
  \end{equation*}
  so every coordinate of
  $f$
  is PAP.
  Closure under finite tupling then implies that
  $f$
  is PAP.

  Finally, every analytic map is PAP,
  which proves the last claim.
\end{proof}

\subsubsection{Stability Consequences}
\label{subsubsec:minmax-stability-consequences}

\begin{theorem}[State non-expansiveness]
  \label{thm:minmax-nonexpansive}
  Let
  $\hat{\gamma} = \{ \tuple{Z,A_Z,b_Z} \mid Z\in\partitionstyle{Z} \}$
  be the coefficient-based piecewise propagate-reset representation of
  Proposition~\ref{prop:minmax-piecewise-copyset}.
  Then,
  \begin{align*}
    \|A_Z\|_\infty
    \leq
    1
    \qquad\quad
    \forall Z\in\partitionstyle{Z}.
  \end{align*}
  Furthermore, 
  \begin{align}
    \label{eq:minmax-global-nonexpansive}
    \|f(x,u)-f(x',u)\|_\infty
    \leq
    \|x-x'\|_\infty
    \qquad\quad
    \forall u\in U,\;
    \forall x \in X,\;
    \forall x'\in X.
  \end{align}
\end{theorem}
\begin{proof}
  By
  Proposition~\ref{prop:minmax-piecewise-copyset},
  every
  $A_Z$
  is deterministic row-substochastic.
  Hence
  \begin{equation*}
    \|A_Z\|_\infty
    \leq
    1.
  \end{equation*}

  For the global bound, fix
  $u\in U$.
  For every
  $i,j\in\iival{1}{n}$,
  the scalar map
  \begin{equation*}
    z
    \mapsto
    \min
    \left\{
      \getentry{R(u)}{i,j},
      z
    \right\}
  \end{equation*}
  is
  $1$-Lipschitz.
  Furthermore, the maximum of finitely many scalar arguments is
  $1$-Lipschitz with respect to the
  $\ell_\infty$
  norm of those arguments.
  Applying these facts coordinatewise to
  Equation~\eqref{eq:minmax-recurrence}
  gives
  Equation~\eqref{eq:minmax-global-nonexpansive}.
\end{proof}

\begin{closingremark}
  Proposition~\ref{prop:minmax-state-propagation-reset-dichotomy}
  and
  Equation~\eqref{eq:minmax-copied-value-bracketing}
  explain the strong one-step bound used in the uBIBS analysis---see
  \S\ref{subsec:a-onestep-minmax-value-bound}:
  \begin{itemize}
    \item
      under
      \emph{state-reset behaviour},
      the updated coordinate is set directly to a current
      input-dependent source value;

    \item
      under
      \emph{state-propagation behaviour},
      the propagated state value is bounded above and below by current
      input-dependent source values.
  \end{itemize}
  Thus, in either behaviour, the updated state can be controlled in terms of
  current input-dependent quantities rather than through additive
  accumulation of the previous state.
\end{closingremark}

\subsection{Composition and Relation to Linear Recurrences}

The propagate-reset coefficient structure is preserved under composition.

\begin{proposition}[Closure of propagate-reset structure under composition]
  \label{prop:copy-reset-composition}
  For $k\in\{1,2\}$, let
  \begin{equation*}
    F_k(x) \,=\, A_k\cdot x + b_k
  \end{equation*}
  be an affine map such that $A_k$ is deterministic row-substochastic and
  \begin{align*}
    \getentry{A_k}{i,j}=1
    \quad\implies\quad
    \getentry{b_k}{i}=0
    \qquad\qquad
    \forall i,j\in\iival{1}{n}.
  \end{align*}
  Then,
  \begin{align*}
    F_2\circ F_1(x)
    \,=\,
    A \cdot x+b
    \qquad\text{ for }\qquad
    A\,\defeq\, A_2 \cdot A_1,
    \quad
    b\,\defeq\, A_2 \cdot b_1+b_2,
  \end{align*}
  satisfies the same propagate-reset coefficient conditions.
\end{proposition}
\begin{proof}
  Fix row $i$.
  If $(A_2)_{i,\cdot}=0_n^\top$, then
  $A_{i,\cdot}=0_n^\top$ and $b_i=(b_2)_i$.
  Otherwise $(A_2)_{i,\cdot}=e_k^\top$, and hence $(b_2)_i=0$.
  If $(A_1)_{k,\cdot}=0_n^\top$, then
  $A_{i,\cdot}=0_n^\top$ and $b_i=(b_1)_k$;
  if $(A_1)_{k,\cdot}=e_j^\top$, then $(b_1)_k=0$ and
  $A_{i,\cdot}=e_j^\top$, $b_i=0$.
\end{proof}

\begin{closingdefinition}[Active-variable regions and coefficients along a MinMax trajectory]
  \label{def:minmax-active-coefficients-along-trajectory}
  Let $\partitionstyle{Z}$ be the partition of
  Definition~\ref{def:minmax-active-variable-partition},
  and
  let
  $\hat{\gamma} = \{ \tuple{Z,A_Z,b_Z} \mid Z\in\partitionstyle{Z} \}$
  be the coefficient-based piecewise propagate-reset representation of $f$
  given by
  Proposition~\ref{prop:minmax-piecewise-copyset}.

  Assume
  \begin{align*}
    f(x,u) \in X
    \qquad\quad
    \forall \ptuple{x,u} \in X \times U.
  \end{align*}

  Let $u=(u_t)_{t\in\timedomain}\in\infseqs{U}$,
  let $x_0\in X$,
  and 
  let $(x_t)_{t\in\timedomain}$
  be the corresponding MinMax trajectory,
  \begin{align*}
    x_t
    \,\defeq\,
    f(x_{t{-}1},u_t)
    \qquad\quad
    \forall t\in\timedomain.
  \end{align*}
  For every $t\in\timedomain$,
  let $Z_t\in\partitionstyle{Z}$ be the unique active-variable region such that
  $(x_{t{-}1},u_t)\in Z_t$.
  We call
  $(Z_t)_{t\in\timedomain}$
  the \emph{active-region sequence} along the trajectory
  $(x_t)_{t\in\timedomain}$ induced by $u$.
  For every $t \in\timedomain$,
  we define
  \begin{align*}
    A_t
    \defeq
    A_{Z_t},
    \qquad\quad
    b_t
    \defeq
    b_{Z_t}(u_t).
  \end{align*}
  We call
  $(A_t)_{t\in\timedomain}$
  and
  $(b_t)_{t\in\timedomain}$
  the \emph{active coefficient sequences} along the trajectory
  $(x_t)_{t\in\timedomain}$ induced by $u$.
\end{closingdefinition}
\begin{closingfact}
  In the context of Definition~\ref{def:minmax-active-coefficients-along-trajectory},
  we have
  \begin{align*}
    x_t
    \,=\,
    A_t\cdot x_{t{-}1}+b_t
    \qquad\quad
    \forall t\in\timedomain.
  \end{align*}
\end{closingfact}

\begin{proposition}[Single-source propagation and no additive accumulation]
  \label{prop:minmax-no-accumulation}
  Let
  $u=(u_t)_{t\in\timedomain}\in\infseqs{U}$,
  let $x_0\in X$,
  and let
  $(x_t)_{t\in\timedomain}$,
  $(A_t)_{t\in\timedomain}$,
  and
  $(b_t)_{t\in\timedomain}$
  be the corresponding state and active coefficient sequences of
  Definition~\ref{def:minmax-active-coefficients-along-trajectory}.
  Let $s,t\in\posnaturals$ with $s<t$.
  Then, the following hold.
  \begin{itemize}
    \item
      There exists $\tuple{A_{t:s},b_{t:s}}$ such that
      \begin{align*}
        x_t
        \,=\,
        A_{t:s}\cdot x_s+b_{t:s},
      \end{align*}
      and
      $\tuple{A_{t:s},b_{t:s}}$
      satisfies the propagate-reset coefficient conditions
      \begin{align*}
        & A_{t:s}
        \in \{0,1\}^{n\times n},
        \\[.3em]
        & A_{t:s}\cdot 1_n
        \,\leq\, 1_n,
        \\[.3em]
        & \getentry{A_{t:s}}{i,j}=1
        \;\implies\;
        \getentry{b_{t:s}}{i}=0
        \qquad\quad
        \forall i,j\in\iival{1}{n}.
      \end{align*}

    \item
      Every coordinate of $b_{t:s}$ is either zero or a single
      coordinate of an active offset introduced at some time point
      $\tau\in\iival{s{+}1}{t}$.
      Namely, for every $i\in\iival{1}{n}$,
      at least one of
      Equations~\eqref{eq:prop:minmax-no-accumulation-4}
      and~\eqref{eq:prop:minmax-no-accumulation-5}
      holds:
      \begin{align}
        \label{eq:prop:minmax-no-accumulation-4}
        \getentry{b_{t:s}}{i}
        &=
        0,
        \\[.2em]
        \label{eq:prop:minmax-no-accumulation-5}
        \getentry{b_{t:s}}{i}
        &\in
        \entries{b_\tau}
        \qquad
        \text{for some }\,
        \tau\in\iival{s{+}1}{t}.
      \end{align}

    \item
      Every coordinate of $x_t$ comes from
      \begin{romanenumerate}
        \item
          a single coordinate of $x_s$, or
        \item
          a single input-dependent source value introduced at some
          time point $\tau\in\iival{s{+}1}{t}$;
      \end{romanenumerate}
      namely,
      for every $i\in\iival{1}{n}$,
      at least one of the conditions given by
      Equations~\eqref{eq:prop:minmax-no-accumulation-1}--%
      \eqref{eq:prop:minmax-no-accumulation-3}
      holds:
      \begin{align}
        \label{eq:prop:minmax-no-accumulation-1}
        \getentry{x_t}{i}
        &\in
        \entries{x_s},
        \\[.2em]
        \label{eq:prop:minmax-no-accumulation-2}
        \getentry{x_t}{i}
        &\in
        \entries{R(u_\tau)}
        \qquad
        \text{for some }\,
        \tau\in\iival{s{+}1}{t},
        \\[.2em]
        \label{eq:prop:minmax-no-accumulation-3}
        \getentry{x_t}{i}
        &\in
        \entries{s(u_\tau)}
        \qquad
        \,\,\text{for some }\,
        \tau\in\iival{s{+}1}{t}.
      \end{align}

    \item
      State values admit the bound
      \begin{align*}
        \|x_t\|_\infty
        \;\leq\;
        \max\big\{\,
          \|x_s\|_\infty,\,
          \|b_{s{+}1}\|_\infty,\,
          \|b_{s{+}2}\|_\infty,\,
          \ldots,\,
          \|b_t\|_\infty
        \,\big\}.
      \end{align*}
  \end{itemize}
\end{proposition}
\begin{proof}
  By Definition~\ref{def:minmax-active-coefficients-along-trajectory},
  for every $\tau\in\timedomain$,
  \begin{align*}
    A_\tau
    \,=\,
    A_{Z_\tau},
    \qquad
    b_\tau
    \,=\,
    b_{Z_\tau}(u_\tau),
  \end{align*}
  where
  $\tuple{A_{Z_\tau},b_{Z_\tau}}$
  are the coefficients of the piecewise propagate-reset representation
  fixed in that definition.
  Hence,
  $\tuple{A_\tau,b_\tau}$
  satisfies the corresponding propagate-reset coefficient conditions.

  Repeated application of
  Proposition~\ref{prop:copy-reset-composition}
  to the coefficient pairs
  \begin{equation*}
    \tuple{A_{s{+}1},b_{s{+}1}},
    \ldots,
    \tuple{A_t,b_t}
  \end{equation*}
  yields a pair
  \begin{equation*}
    \tuple{A_{t:s},b_{t:s}}
    \in
    \reals^{n\times n}\times\reals^n
  \end{equation*}
  satisfying the propagate-reset coefficient conditions and such that
  \begin{align*}
    x_t
    \,=\,
    A_{t:s}\cdot x_s+b_{t:s}.
  \end{align*}
  This proves the first claim.

  Furthermore,
  Proposition~\ref{prop:copy-reset-composition}
  implies that, under each composition, every coordinate of the
  accumulated offset either remains zero, propagates a single
  previously introduced offset coordinate, or is replaced by a single
  coordinate of the newly composed offset.
  Consequently, no two offset coordinates introduced at distinct time
  points are ever added.
  It follows inductively that, for every $i\in\iival{1}{n}$,
  \begin{align*}
    \getentry{b_{t:s}}{i}
    =0
    \qquad\text{or}\qquad
    \getentry{b_{t:s}}{i}
    \in
    \entries{b_\tau}
  \end{align*}
  for some
  $\tau\in\iival{s{+}1}{t}$.
  This proves the second claim.

  By the form of the active regional affine representation, for every
  $\tau\in\iival{s{+}1}{t}$ and every $i\in\iival{1}{n}$,
  either the $i$-th row of $A_\tau$ contains exactly one entry equal to one
  and
  \begin{align*}
    \getentry{b_\tau}{i}
    \,=\,
    0,
  \end{align*}
  or the $i$-th row of $A_\tau$ is zero and
  \begin{align*}
    \getentry{b_\tau}{i}
    \in
    \entries{R(u_\tau)}
    \,\cup\,
    \entries{s(u_\tau)}.
  \end{align*}
  Hence, at each time step and in each coordinate, the update either
  propagates a single state coordinate or replaces it by a single
  input-dependent source value.

  Since this propagate-reset structure is preserved under composition,
  for every $i\in\iival{1}{n}$,
  the coordinate $\getentry{x_t}{i}$ is either a single coordinate of
  $x_s$, or a single input-dependent source value introduced at some
  time
  $\tau\in\iival{s{+}1}{t}$.
  Therefore,
  \begin{align*}
    \getentry{x_t}{i}
    \in
    \entries{x_s}
    \,\cup\,
    \bigcup_{\tau=s{+}1}^{t}
    \left(
      \entries{R(u_\tau)}
      \,\cup\,
      \entries{s(u_\tau)}
    \right).
  \end{align*}
  Combining this observation with the preceding characterization gives
  Equations~\eqref{eq:prop:minmax-no-accumulation-1}--%
  \eqref{eq:prop:minmax-no-accumulation-3}.

  Finally, every coordinate of $x_t$ is therefore equal to either a
  coordinate of $x_s$ or a coordinate of some
  $b_\tau$ with
  $\tau\in\iival{s{+}1}{t}$.
  Hence
  \begin{align*}
    \|x_t\|_\infty
    \;\leq\;
    \max\big\{\,
      \|x_s\|_\infty,\,
      \|b_{s{+}1}\|_\infty,\,
      \ldots,\,
      \|b_t\|_\infty
    \,\big\},
  \end{align*}
  which proves the last claim.
\end{proof}

\begin{closingremark}[Comparison with linear recurrences]
  Proposition~\ref{prop:minmax-no-accumulation}
  sharply contrasts MinMax with a generic linear recurrence
  \begin{align*}
    x_t
    \,=\,
    A_t\cdot x_{t{-}1}+b_t.
  \end{align*}
  Indeed, unrolling a generic linear recurrence gives
  \begin{align*}
    x_t
    \;=\;
    A_t\cdots A_1\cdot x_0
    \,+\,
    \sum_{\tau=1}^{t}
    A_t\cdots A_{\tau{+}1}\cdot b_\tau.
  \end{align*}
  Hence, bounded additive contributions may accumulate along
  non-contracting directions.
  For example, the recurrence
  \begin{align*}
    x_t=x_{t{-}1}+1
  \end{align*}
  is non-expansive in the state variable but is not BIBS stable.
  In contrast,
  Proposition~\ref{prop:minmax-no-accumulation}
  shows that, for MinMax, the corresponding additive expansion
  collapses coordinatewise to a single surviving source value:
  contributions introduced at different time points do not add
  coordinatewise.

  The comparison with contractive linear recurrences is also instructive.
  A condition such as
  \begin{align*}
    \|A_t\|\leq\alpha<1
  \end{align*}
  ensures stability through geometric attenuation of past contributions.
  MinMax instead only guarantees
  \begin{align*}
    \|A_t\|_\infty\leq1
  \end{align*}
  and may therefore preserve a state value with unit gain indefinitely.
  Its stability mechanism is not geometric attenuation, but
  single-source propagation and replacement without additive
  accumulation, together with the bracketing property
  in~\eqref{eq:minmax-copied-value-bracketing}.

  The analogy with linear recurrences is nevertheless only local.
  The active coefficient matrix $A_t$ depends on the active-variable
  region containing $(x_{t{-}1},u_t)$ and therefore generally depends on
  the current state.
  Thus, MinMax is a switched piecewise state-affine system rather than
  a globally linear recurrent system.
\end{closingremark}

\clearpage{}%

\appendixpart{part:experiments}{Implementation and Experiments}

\clearpage{}%

\section{Implementation of MinMax RNCs}
\label{sec:minmax-implementation}

Our implementation of MinMax RNCs is released as a Python package \citep{minmaxpip}, with source
code available at \citep{minmaxgithub}.
We describe version \minmaxversion{}.
In the implementation, the \emph{state dimension} of single neurons 
(i.e., the \emph{state degree} of the MinMax RNC) is the hyperparameter (\code{d_state}), and
neurons of state dimension one are called \emph{scalar-state neurons}.

The overall architecture is shown below.
\code{MinMaxNeuron} and \code{MinMaxNeuronArray} implement the MinMax neurons of a layer in the
formal description; \code{MinMaxLayer} equips them with a convolution, a feedforward network,
normalisation, and a residual connection.

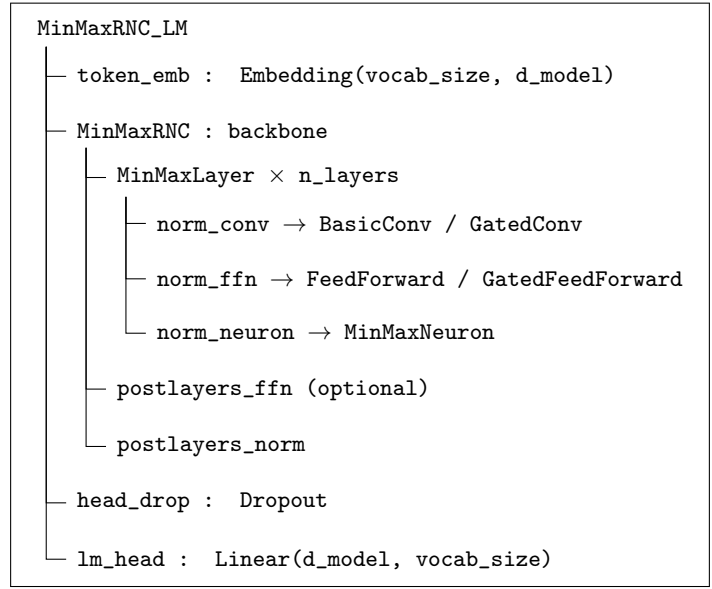
\begin{center}
\fbox{
\begin{forest}
  for tree={
    font=\ttfamily\small,
    grow'=0,
    child anchor=west,
    parent anchor=south,
    anchor=west,
    calign=first,
    inner xsep=4pt,
    edge path={
      \noexpand\path [draw, \forestoption{edge}]
        (!u.south west) +(7.5pt,0) |- (.child anchor)\forestoption{edge label};
    },
    before typesetting nodes={
      if n=1
        {insert before={[,phantom]}}
        {}
    },
    fit=band,
    before computing xy={l=15pt},
  }
  [{MinMaxRNC\_LM}
    [{token\_emb : Embedding(vocab\_size, d\_model)}]
    [{MinMaxRNC : backbone}
      [{MinMaxLayer $\times$ n\_layers}
        [{norm\_conv $\to$ BasicConv / GatedConv}]
        [{norm\_ffn $\to$ FeedForward / GatedFeedForward}]
        [{norm\_neuron $\to$}
          [{MinMaxNeuron (scalar\_neurons, optional)}]
          [{MinMaxNeuronArray $\times$ len(neurons)}]
        ]
      ]
      [{postlayers\_ffn (optional)}]
      [{postlayers\_norm}]
    ]
    [{head\_drop : Dropout}]
    [{lm\_head : Linear(d\_model, vocab\_size)}]
  ]
\end{forest}
\vspace{1em}
}
\end{center}

\subsection{MinMax Neurons}
\label{sec:minmaximpl-neurons}

A MinMax neuron is the recurrent cell. It projects the input to the parameters of its
recurrence, runs the MinMax recurrence as a parallel scan, and projects the resulting states back
to the residual-stream dimension.

\subsubsection{Scalar-state neurons}
A layer of $D$ scalar-state neurons has state ${x}_t \in \reals^D$, updated element-wise as
\begin{align*}
  {x}_t \;=\; ({r}_t \odot {x}_{t-1}) \oplus {s}_t,
  \quad\qquad\text{i.e.,}\quad
  \getentry{x_t}{i} \;=\; \big(\getentry{r_t}{i} \odot \getentry{x_{t-1}}{i}\big) \oplus
  \getentry{s_t}{i} 
  \qquad \forall\mkern2mu i \in \ival{D}\mkern2mu,
\end{align*}
where ${r}_t$ and ${s}_t$ are affine projections of the input
${u}_t \in \reals^{d_\mathrm{in}}$ (with ${d_\mathrm{in}} = \code{d_model}$):
\begin{align*}
  {r}_t \,=\, W_\mathrm{r}\cdot {u}_t + {b}_\mathrm{r}\mkern2mu,
  \qquad\qquad
  {s}_t \,=\, W_\mathrm{s}\cdot {u}_t + {b}_\mathrm{s}\mkern2mu.
\end{align*}
They are implemented by \code{MinMaxNeuron}; their number per layer is the field
\code{scalar_neurons} of the configuration (or its legacy name \code{d_state}).

\subsubsection{Neurons of any state degree}
A neuron of state degree $d$ has state ${x}_t \in \reals^d$, updated by MinMax Recurrence:
\begin{align*}
  {x}_t \;=\; ({R}_t \oplus {x}_{t-1}) \oplus {s}_t,
  \quad\qquad\text{i.e.,}\quad
  \getentry{x_t}{i} \;=\; \left(\bigoplus_{j=1}^d \getentry{R_t}{i,j} \odot
  \getentry{x_{t-1}}{j}\right) \oplus \getentry{s_t}{i}
  \qquad \forall\mkern2mu i \in \ival{d}\mkern2mu,
\end{align*}
where $R_t \in \reals^{d \times d}$ and ${s}_t \in \reals^d$ are affine projections of the
input: ${s}_t$ as above, and the $d^2$ entries of $R_t$ given by an affine projection
reshaped to a $d \times d$ matrix, with no further constraint:
\begin{align*}
  R_t \,=\, \code{reshape}\big(W_\mathrm{r}\cdot {u}_t + {b}_\mathrm{r},\, \ptuple{d,d}\big)
  \mkern2mu,
  \qquad\qquad
  {s}_t \,=\, W_\mathrm{s}\cdot {u}_t + {b}_\mathrm{s}
  \mkern2mu.
\end{align*}
For $d = 1$ this is the recurrence of a scalar-state neuron.

Neurons are grouped into \emph{arrays}: an array of $K$ neurons of the same state degree $d$, 
all fed the same input, implemented by \code{MinMaxNeuronArray}.
Its state is the concatenation of the $K$ neuron states, with $D = K \cdot d$ coordinates.

\code{MinMaxRNCConfig.neurons} lists the arrays of every layer, each given as
\code{NeuronArray(d_state, num_neurons)}; for instance, \code{neurons=[NeuronArray(3, 16)]} gives
16 neurons of state degree 3 per layer.
A layer may combine scalar-state neurons and any number of arrays, and has at least one neuron.
\code{MinMaxNeuron} is the array of \code{scalar_neurons} neurons of state degree one.

\subsubsection{Output}
The concatenated states ${x}_t \in \reals^D$ of an array are projected back to the
residual-stream dimension:
\begin{align*}
  {y}_t =
  \begin{cases}
    W_\mathrm{o} \cdot {x}_t + {b}_\mathrm{o} & \text{if \code{output_gate=False}}, \\[4pt]
    W_\mathrm{o}\cdot \left({x}_t * \mathrm{sigmoid}(W_\mathrm{g}\cdot u_t
      + {b}_\mathrm{g})\right) + {b}_\mathrm{o}
    & \text{if \code{output_gate=True}},
  \end{cases}
\end{align*}
where $*$ is the element-wise product. This is the sum of the projections of the individual
neurons' states. The outputs of all arrays of a layer are summed.
The learnable parameters are the weights 
$W_\mathrm{r}$, $W_\mathrm{s}$, $W_\mathrm{o}$, $W_\mathrm{g}$, 
the corresponding biases 
$b_\mathrm{r}$, $b_\mathrm{s}$, $b_\mathrm{o}$, $b_\mathrm{g}$, 
and
optionally the initial state ${x}_0$ when \code{train_init} is set to \code{True}---%
otherwise ${x}_0 = \mathbf{0}$.
For an array of $K$ neurons of degree $d$, the projection for $R_t$ has 
${d_\mathrm{in}} \cdot K \cdot d^2$ weights,
and 
those for ${s}_t$, the output and the gate have 
${d_\mathrm{in}} \cdot K \cdot d$ weights each.

\subsubsection{Evaluation strategy and computational cost}
All states for a sequence of length $T$ are computed simultaneously by a 
\emph{Hillis-Steele parallel prefix scan}%
  \footnote{%
    Hillis-Steele parallel prefix scan~\citep{hillis1986data} is less work-efficient than Blelloch's 
    parallel prefix scan, introducing an extra $O(\log T)$ term, but has smaller depth by a constant
    factor, hence requiring fewer kernel launches in practice.%
}
of depth $O(\log T)$, which composes the per-step update maps of the MinMax Recurrence. 
For scalar-state neurons, the scan is element-wise.
For state degree $d > 1$, the $K$ neurons of an array are evaluated as one batched scan, and each
composition costs $O(d^3)$; the scan's work grows as $O(K \cdot T \cdot \log T \cdot d^3)$, so $d$
should be kept small and the size scaled through $K$.

At inference time the recurrence runs sequentially, with $O(K \cdot d^2)$ compute and $O(K \cdot d)$
memory per token for each array.

\subsection{MinMax Layer}

A \code{MinMaxLayer} is a residual block. 
The data flow uses pre-norm; only the neuron output is added back to the residual stream:
\begin{align*}
  \mathit{conv}   &= \code{Conv}\bigl(\mathrm{norm}(u)\bigr), \\
  \mathit{ffn}    &= \code{FFN}\bigl(\mathrm{norm}(u + \mathit{conv})\bigr), \\
  \mathit{neuron} &= \code{Neurons}\bigl(\mathrm{norm}(u + \mathit{conv} + \mathit{ffn})\bigr), \\
  \mathit{output} &= u + \mathit{neuron},
\end{align*}
where $\mathrm{Neurons}$ is the sum of the outputs of all neuron arrays of the layer, all fed the
same input.
The convolution~\code{Conv}'s and feedfoward network~\code{FFN}'s outputs feed the neurons' input
but are not added to the residual stream independently. 
The convolution~\code{Conv} provides short-range context (one previous token). 
The feedfoward network~\code{FFN} mixes features. 
The neurons integrate information over arbitrarily long ranges via the recurrence.
The three norms are separate, of the same type (\code{norm}).

\subsubsection{Convolution Variants}

The implementation provides two convolution variants:
\begin{itemize}
  \item
    \code{BasicConv} (default): a learned affine projection of the concatenation
    $[u_{t-1},\, u_t]$ back to $\reals^{d_\mathrm{in}}$, with 
    $2{d_\mathrm{in}^2} + {d_\mathrm{in}}$ parameters.

  \item
    \code{GatedConv}: a learned per-feature gate logit ${g} \in \reals^{d_\mathrm{in}}$, 
    with ${d_\mathrm{in}}$ parameters, interpolates between the previous and current token:
    \begin{align*}
      \mathrm{out}_t \mkern-1mu
      \;=\; \mathrm{sigmoid}({g}) * u_{t-1} \mkern-1mu
      \,+\, \bigl(1 - \mathrm{sigmoid}({g})\bigr) * u_t
      \mkern2mu .
    \end{align*}
\end{itemize}
In both, $u_0 = \mathbf{0}$ at the start of a sequence.

\subsection{MinMax RNC}

The overall architecture \code{MinMaxRNC} is obtained by stacking \code{n_layers} MinMaxLayers.
An optional post-layers FFN and a normalisation are applied after the final layer.
The forward pass takes an argument \code{unroll_steps}: the sequence is split into chunks of
\code{unroll_steps} tokens, each chunk is evaluated by the parallel scan, and the recurrent state is
carried from one chunk to the next. This bounds the memory of the scan, without changing the result.
The forward pass can also start from, and return, the recurrent state, for sequential inference.

All architecture hyperparameters are specified through a single flat
\code{MinMaxRNCConfig} dataclass, described in Table~\ref{tab:minmax-config}.

\subsection{Language Model Wrapper}

\code{MinMaxRNC_LM} wraps \code{MinMaxRNC} adding:
\begin{itemize}
  \item a token embedding ($\mathrm{vocab\_size} \times d_{\mathrm{model}}$)
    initialised with \code{small_init_} (see Section~\ref{sec:minmaximpl-initialisers} below);
  \item a linear \code{LM_head} without bias ($d_{\mathrm{model}} \times \mathrm{vocab\_size}$),
    optionally tied to the embedding weights;
  \item a dropout before the \code{LM_head}.
\end{itemize}

Its hyperparameters are specified through the
\code{MinMaxRNCLMConfig} dataclass described in Table~\ref{tab:minmax-lm-config}.
Its fields \code{output_gate}, \code{conv_type} and \code{ffn_dropout} always replace those of the
backbone configuration, so these three are set on \code{MinMaxRNCLMConfig}; 
a warning is raised when one of them is set on the backbone to a non-default value that differs from
the one of \code{MinMaxRNCLMConfig}.

\subsection{Initialisation}
\label{sec:minmaximpl-initialisers}

\begin{table}[t]
  \small
  \center
  \caption{Initialisation schemes for the $s$ and $r$ projections.}
  \label{tab:s-r--initialisation-schemes}
  \vspace{.2em}
  \renewcommand{\arraystretch}{1.25}
  \begin{tabular}{lllcc}
    \toprule
    \textbf{Initialisation} & $W_\mathrm{s}$ \textbf{weights} & $W_\mathrm{r}$ \textbf{weights} &
    $b_\mathrm{s}$ & $b_\mathrm{r}$ \\
    \midrule
    \code{`small_init'} & \code{small_init_} & \code{small_init_} & $0$ & $0$ \\
    \code{`kaiming'}     & Kaiming uniform ($a=\sqrt{5}$) & Kaiming uniform ($a=\sqrt{5}$) & $0$ & $0$ \\
    \code{`asymmetric'}  & Kaiming uniform ($a=\sqrt{5}$) & \code{small_init_} & $1$ & $0$ \\
    \bottomrule
  \end{tabular}
\end{table}

The following initialisation scheme is employed.
\begin{itemize}
  \item
    \code{small_init_}~\citep{nguyen2019transformers} is applied to all \emph{input-side}
    projections, with $\mathrm{dim}$ their input dimension: token
    embedding, \code{LM_head} (when untied), \code{BasicConv}, \code{FFN} up-projection, and the
    neuron $W_\mathrm{s}$, $W_\mathrm{r}$, and output-gate~$W_\mathrm{g}$ projections.
    It keeps activations small at initialisation in deep networks.
    \begin{align*}
    \code{small_init_}(z, \mathrm{dim}): \qquad
      z \sim \mathcal{N}(0,\sigma^2),\;\;\sigma = \sqrt{2 / (5 \cdot \mathrm{dim})}
    \end{align*}
  \item
    \code{wang_init_}~\citep{radford2019language,beck2024xlstm} is applied to the two
    \emph{residual output} projections:
    the neuron output $W_\mathrm{o}$, with $\mathrm{dim}$ the number $D$ of state coordinates of
    the whole layer (summed over all its neuron arrays), and the FFN down-projection, with
    $\mathrm{dim}$ its hidden dimension.
    It additionally divides by $n_\mathrm{layers}$ (the number of residual blocks) so the total
    variance contributed by all residual branches to the stream stays $O(1/n_\mathrm{layers})$
    regardless of depth---following the GPT-2 scaled-init strategy.
    Taking $\mathrm{dim}$ over the whole layer keeps the layer's contribution the same however
    its state is split across neuron arrays.
    \begin{align*}
      \code{wang_init_}(z, \mathrm{dim}, n_\mathrm{layers}): \qquad
      z \sim \mathcal{N}(0,\sigma^2),\;\;\sigma = 2 / (n_\mathrm{layers} \cdot \sqrt{\mathrm{dim}})
    \end{align*}
  \item
    All biases are initialised to zero, except ${b}_\mathrm{s}$ under the \code{asymmetric}
    scheme below; the gate logits of GatedConv are initialised to \code{conv_init_val}.

  \item
    For the $s$ and $r$ projections, three initialisation types are possible, as described in
    Table~\ref{tab:s-r--initialisation-schemes}. 
    For neurons of state degree $d > 1$, they apply to all $d^2$ entries of $R_t$.
    The \code{asymmetric} scheme addresses an initialisation dead zone: with
    ${x}_0 = \mathbf{0}$, the recurrence
    $({r}_t \odot {x}_{t-1}) \oplus {s}_t$ keeps a hidden unit at 0
    whenever $s < 0 < r$, preventing state writes early in training.
    Setting $b_\mathrm{s} = 1$ makes $s > 0$ for most units at initialisation, so writes happen from
    the first token; keeping $W_\mathrm{r}$ small prevents $r$ from immediately
    clipping the written values back towards zero.
\end{itemize}

\subsection{Feedforward Networks}

The implementation provides two following variants of feedforward networks.
\begin{itemize}
  \item
    \textbf{Gated variant} (default): the gated variant is the default (\code{ffn_type=`gated'}).
    With activation $\phi=$ReLU it is ReGLU (default); with $\phi=$Swish it is SwiGLU.
  \item
    \textbf{Basic variant}: one-hidden layer MLP (\code{ffn_type=`basic'}).
\end{itemize}
In both variants, the hidden dimension is $\code{ffn_proj_factor} \cdot d_\mathrm{model}$ rounded up
to an even number, the projections have biases, and dropout is applied to the output.

\begin{center}
\footnotesize
\renewcommand{\arraystretch}{1.25}
\begin{tabular}{llc}
  \toprule
  \textbf{Class} & \textbf{Formula} & \textbf{Cf.} \\
  \midrule
  \code{FeedForward}
  & $W_2\cdot\phi(W_1 \cdot {x} + b_1) + b_2$ 
  & \citep{vaswanispujgkp17} 
  \\[.1em]
  \code{GatedFeedForward} 
  & $W_2 \cdot \big(\phi(\mathit{gate}) * \mathit{value}\big) + b_2 
  \;\,\text{ with }\;\,\mkern1mu
  [\mathit{gate} \;\mkern2mu \mathit{value}] \,=\, W_1 \cdot {x} + b_1$
  & \citep{dauphin2017gated,shazeer2020glu} \\
  \bottomrule
\end{tabular}
\end{center}

\subsection{Numerical Precision}
\label{sec:minmaximpl-precision}

Two settings control the precision, and both default to leaving it to the caller.
\begin{itemize}
  \item
    \code{amp_dtype} runs the whole forward pass, including embedding and \code{LM_head}, under PyTorch's
    automatic mixed precision in bf16 or fp16.
  \item
    \code{recurrence_dtype} pins the dtype of the MinMax recurrence---$R_t$, ${s}_t$, the
    states, the state carried across chunks, and their gradients---independently of the surrounding
    precision: the recurrence then runs in that dtype with autocast disabled, while the input and
    output projections follow the surrounding precision.
\end{itemize}
The recurrence itself introduces no rounding error, since $\max$ and $\min$ return one of their
arguments: in a given dtype, the computed states are exactly those of the recurrence on $R_t$ and
${s}_t$ as represented in that dtype.
Reduced precision affects only the projections and the representation of their outputs.

\begin{table}[p]
\centering
\footnotesize
\caption{Description of the \code{MinMaxRNCConfig} dataclass. Every layer needs at least one
neuron: \code{scalar_neurons} (or \code{d_state}) $> 0$, or a non-empty \code{neurons}.}
\renewcommand{\arraystretch}{1.25}
\begin{tabular}{lccl}
  \toprule
  \textbf{Field} & \textbf{Type} & \textbf{Default} & \textbf{Description} \\
  \midrule
  \code{d_model}           & int   & --              & Residual-stream / embedding width \\
  \code{n_layers}          & int   & --              & Number of MinMaxLayers \\
  \midrule
  \code{scalar_neurons}    & int   & \code{0}         & Number of scalar-state neurons per layer \\
  \code{d_state}           & int   & \code{0}
    & Legacy name of \code{scalar_neurons} \\
  \code{neurons}           & tuple & \code{()}
    & Neuron arrays per layer (Section~\ref{sec:minmaximpl-neurons}) \\
  \midrule
  \code{norm}               & str   & \code{`layernorm'}
    & Pre-norm inside each layer:
  \code{`layernorm'}, \code{`rmsnorm'}, \code{`none'} \\
  \code{postlayers_norm}   & str   & \code{`layernorm'}
    & Norm applied after the final layer \\
  \midrule
  \code{ffn_type}          & str   & \code{`gated'}
                            & \code{`gated'} (ReGLU/SwiGLU) or \code{`basic'} (MLP) \\
  \code{ffn_proj_factor}  & float & \code{1.3}
    & FFN hidden-dim expansion factor \\
  \code{ffn_act_fn}       & str   & \code{`relu'}
                            & Activation: \code{`relu'}, \code{`relu^2'}, \code{`gelu'},
  \code{`swish'}, \code{`sigmoid'}, \code{`selu'} \\
  \code{ffn_dropout}       & float & \code{0.1}
    & Dropout inside the FFN \\
  \code{ffn_init}          & str   & \code{`scaled'}
                            & \code{`basic'} (PyTorch default) or
  \code{`scaled'} (\code{small_init} + \code{wang_init}) \\
  \midrule
  \code{output_gate}       & bool  & \code{True}
    & Gate the neuron output by a learned projection of the input \\
  \code{train_init}        & bool  & \code{False}
    & Make the initial hidden state ${x}_0$ a learned parameter \\
  \code{neuron_dropout}    & float & \code{0.0}
    & Dropout on the neuron input \\
  \code{s_r_init}         & str   & \code{`small_init'}
                          & Init for $W_\mathrm{s}$, $W_\mathrm{r}$:
  \code{`small_init'}, \code{`kaiming'}, \code{`asymmetric'} \\
  \midrule
  \code{conv_type}         & str   & \code{`basic'}
                            & \code{`gated'} (per-feature gate) or \code{`basic'} (linear mix) \\
  \code{conv_init_val}    & float & \code{0.0}
    & Initial gate logit for GatedConv \\
  \midrule
  \code{prelayers_dropout} & float & \code{0.0}
    & FFN dropout override for the first layer only \\
  \code{use_postlayers_ffn} & bool & \code{False}
    & Add an FFN after all layers \\
  \midrule
  \code{recurrence_dtype}  & str   & \code{None}
    & Recurrence dtype (\code{`fp32'}, \code{`bf16'}, \code{`fp16'}) \\
  \code{amp_dtype}         & str   & \code{None}
    & Autocast dtype of the forward pass (\code{`bf16'}, \code{`fp16'}) \\
  \bottomrule
\end{tabular}
\label{tab:minmax-config}
\end{table}

\begin{table}[p]
\centering
\footnotesize
\caption{Description of the \code{MinMaxRNCLMConfig} dataclass.}
\vspace{.2em}
\renewcommand{\arraystretch}{1.27}
\begin{tabular}{lccl}
  \toprule
  \textbf{Field} & \textbf{Type} & \textbf{Default} & \textbf{Description} \\
  \midrule
  \code{backbone}      & \code{MinMaxRNCConfig} & --
    & Backbone configuration. \\
  \code{head_dropout} & float & \code{0.0}
                      & Dropout before the \code{LM_head}. \\
  \code{tie_weights}  & bool  & \code{True}
    & Share embedding and LM-head weights. \\
  \code{output_gate}  & bool  & \code{True}
                      & Gate each neuron output by $\mathrm{sigmoid}(W_\mathrm{g} \cdot u)$;
  \\[-0.3em] & & & 
      replaces \code{backbone.output_gate}. \\
  \code{conv_type}    & str   & \code{`basic'}
    & Short-range conv variant; 
  \\[-0.3em] & & & 
  replaces \code{backbone.conv_type}. \\
  \code{ffn_dropout}  & float & \code{0.1}
    & FFN dropout for all layers except the first;
  \\[-0.3em] & & & 
  uses \code{backbone.prelayers_dropout};
  \\[-0.3em] & & & 
      replaces \code{backbone.ffn_dropout}. \\
  \bottomrule
\end{tabular}
\label{tab:minmax-lm-config}
\end{table}
\clearpage{}%

\clearpage{}%

\section{Details of the Experimental Evaluation on Group-free Benchmarks}
\label{sec:experimental_setup}
\label{app:experiments}

This appendix provides further details of our empirical evaluation on group-free benchmarks.

\paragraph{Reproducing the experiments.}
The datasets, configurations, and training and evaluation protocols of the experiments on
MinMax RNCs in this appendix are available in our benchmarks
repository~\citep{minmaxbenchmarksgithub} (version~\code{0.2.0}), which runs them with the
\code{minmaxrnc}~package~\citep{minmaxgithub}.
The repository includes only MinMax RNCs; the other models of this appendix must be plugged
into it, following the instructions of its documentation, and are then trained and evaluated
under the same protocols.
The following commands install the repository, run the experiments on \code{Latching},
\code{Sequences} and \code{InductionHeads} (generating the datasets, training and evaluating
the models), and run the throughput measurements:
\begin{commands}
git clone https://github.com/minmaxrnc/benchmarks && cd benchmarks
pip install -r requirements.txt
python main.py reproduce main
python main.py reproduce throughput
\end{commands}

\localtableofcontents

\subsection{Benchmarks}

We consider three token-based synthetic benchmark tasks:
\dlatching{n}, \dsequences{n}, and \dheads{n}.
Each task is parametrised along a difficulty parameter $n$,
and it consists in predicting output tokens given a sequence of input tokens.

The input sequences to the models are obtained by encoding the finite vocabulary
$\mathcal{V} = \{ v_1, \ldots, v_m \}$ of a task into integers $\iival{1}{m}$
via a bijection $\enc : \mathcal{V} \to \iival{1}{m}$.
Models are required to return logits over the vocabulary $\mathcal{V}$, and the predicted token
is the one having maximum logit (with ties broken arbitrarily).

Next we describe the benchmark tasks.
For each task, we first define a general parametrisation, and then we define the parametrisation in
the difficulty parameter $n$ by defining some parameters as functions of $n$ and fixing the other
parameters.

\paragraph{Latching.}
The \code{Latching} task is parametrised by
the total number $V \in \posnaturals$ of tokens,
and
the number $M \in \posnaturals$ of tokens that can appear in the first position (with $V-M \geq 1$).

The vocabulary is the disjount union
$\mathcal{V} = \mathcal{V}_\mathrm{mem} \cup \mathcal{V}_\mathrm{other}$ with
\begin{align*}
  \mathcal{V}_\mathrm{mem} = \{ \mathit{mem}_i \mid i \in \iival{1}{M}  \},
  \qquad
  \mathcal{V}_\mathrm{other} = \{ \mathit{other}_i \mid i \in \iival{1}{V-M}  \}.
\end{align*}

On input $s = \sigma_1 \cdots \sigma_T$,
the expected output sequence is $\gamma_1 \cdots \gamma_T$ where
$\gamma_t = \sigma_1$ for every $t \in \iival{1}{T}$.

To define datasets, we introduce parameters specifying the minimum and maximum sequence length:
$T_\mathrm{min},T_\mathrm{max} \in \naturals$ satisfying $T_\mathrm{min} \leq T_\mathrm{max}$.
Then, an input sequence $\rv{U}_1 \cdots \rv{U}_\rv{T}$ is generated according to the following
distributions:
\begin{align*}
  \rv{T} \sim \operatorname{Unif}(\iival{T_\mathrm{min}}{T_\mathrm{max}}),
  \qquad
  \rv{U}_1 \sim \operatorname{Unif}(\mathcal{V}_\mathrm{mem}),
  \qquad
  \rv{U}_t \sim \operatorname{Unif}(\mathcal{V}_\mathrm{other})
  \quad
  \forall t \in \iival{2}{\rv{T}}.
\end{align*}

For the experiments, we adopt a single-number parametrisation \dlatching{n} where increasing the
value of $n \in \posnaturals$ increases the difficulty of the task as it is the
number of candidate tokens to memorise.
Table~\ref{tab:latching_datasets} reports the details of the parameterisation in $n$,
and also the values of the other parameters (derived or fixed) for the values of $n$ that we consider
in the experiments.

\begin{table}[htb]
\small
  \centering
  \caption{Details of \dlatching{n}.}
  \vspace{.3em}
  \label{tab:latching_datasets}
  \fbox{
    \begin{minipage}{\linewidth}
      \vspace{1.0em}
      \centering
      \begin{tabular}{l@{\hspace{2.5em}}rrrrcr@{\hspace{2em}}c}
        \multicolumn{7}{l}{\textbf{Training}} \\[.2em]
        \toprule
        & \multicolumn{4}{c}{\emph{Parameter values}} \\
        \cmidrule(lr){2-5}
        \emph{Parametrisation} & $V$  & $M$ & $T_\mathrm{min}$ & $T_\mathrm{max}$ & & $|\mathcal{V}|$ & Size  \\
        \midrule
        \dlatching{n}   & $5n$ & $n$ & 256       & 512  &   & $5n$    & 20k \\
        \midrule
        \dlatching{4}   &   20 &   4 & 256       & 512  &   &   20    & 20k \\
        \dlatching{8}   &   40 &   8 & 256       & 512  &   &   40    & 20k \\
        \dlatching{16}  &   80 &  16 & 256       & 512  &   &   80    & 20k \\
        \dlatching{32}  &  160 &  32 & 256       & 512  &   &  160    & 20k \\
        \dlatching{64}  &  320 &  64 & 256       & 512  &   &  320    & 20k \\
        \dlatching{128} &  640 & 128 & 256       & 512  &   &  640    & 20k \\
        \dlatching{256} & 1280 & 256 & 256       & 512  &   & 1280    & 20k \\
        \dlatching{512} & 2560 & 512 & 256       & 512  &   & 2560    & 20k \\
        \bottomrule
      \end{tabular}

      \vspace{1.5em}
      \begin{tabular}{l@{\hspace{2.5em}}rrrrcr@{\hspace{2em}}c}
        \multicolumn{7}{l}{\textbf{Validation}} \\[.2em]
        \toprule
        & \multicolumn{4}{c}{\emph{Parameter values}} \\
        \cmidrule(lr){2-5}
        \emph{Parametrisation} &$V$ &$M$ &$T_\mathrm{min}$ &$T_\mathrm{max}$&&$|\mathcal{V}|$&Size \\
        \midrule
        \dlatching{n}   & $5n$ & $n$ & 1024             & 2048     &   &      $5n$ & 1k   \\
        \midrule
        \dlatching{4}   &   20 &   4 & 1024             & 2048     &   &        20 & 1k   \\
        \dlatching{8}   &   40 &   8 & 1024             & 2048     &   &        40 & 1k   \\
        \dlatching{16}  &   80 &  16 & 1024             & 2048     &   &        80 & 1k   \\
        \dlatching{32}  &  160 &  32 & 1024             & 2048     &   &       160 & 1k   \\
        \dlatching{64}  &  320 &  64 & 1024             & 2048     &   &       320 & 1k   \\
        \dlatching{128} &  640 & 128 & 1024             & 2048     &   &       640 & 1k   \\
        \dlatching{256} & 1280 & 256 & 1024             & 2048     &   &      1280 & 1k   \\
        \dlatching{512} & 2560 & 512 & 1024             & 2048     &   &      2560 & 1k   \\
        \bottomrule
      \end{tabular}

      \vspace{1.5em}
      \begin{tabular}{l@{\hspace{2.5em}}rrrrcr@{\hspace{2em}}c}
        \multicolumn{7}{l}{\textbf{Evaluation}} \\[.2em]
        \toprule
        & \multicolumn{4}{c}{\emph{Parameter values}} \\
        \cmidrule(lr){2-5}
        \emph{Parametrisation} & $V$  & $M$ & $T_\mathrm{min}$ & $T_\mathrm{max}$ & &
        $|\mathcal{V}|$ & Size  \\
        \midrule
        \dlatching{n}   & $5n$ & $n$ & $2^{10}$         & $2^{10}$  &  & $5n$ & 1k  \\
        \midrule
        \dlatching{4}   &   20 &   4 & $2^{10}$         & $2^{10}$  &  &   20 & 1k  \\
        \dlatching{8}   &   40 &   8 & $2^{10}$         & $2^{10}$  &  &   40 & 1k  \\
        \dlatching{16}  &   80 &  16 & $2^{10}$         & $2^{10}$  &  &   80 & 1k  \\
        \dlatching{32}  &  160 &  32 & $2^{10}$         & $2^{10}$  &  &  160 & 1k  \\
        \dlatching{64}  &  320 &  64 & $2^{10}$         & $2^{10}$  &  &  320 & 1k  \\
        \dlatching{128} &  640 & 128 & $2^{10}$         & $2^{10}$  &  &  640 & 1k  \\
        \dlatching{256} & 1280 & 256 & $2^{10}$         & $2^{10}$  &  & 1280 & 1k  \\
        \dlatching{512} & 2560 & 512 & $2^{10}$         & $2^{10}$  &  & 2560 & 1k  \\
        \bottomrule
      \end{tabular}
    \vspace{1.0em}
    \end{minipage}
  }
\end{table}

\paragraph{Sequences.}
The \code{Sequences} task is parametrised by
the number $N \in \posnaturals$ of distinct sequences to match,
the length $L \in \posnaturals$ of the sequences to match,
the number $M \in \posnaturals$ of distinct tokens for each event type,
and
the number $I \in \posnaturals$ of irrelevant tokens.

The vocabulary consists of input tokens from the disjoint sets $\mathcal{V}_\mathrm{set},
\mathcal{V}_\mathrm{reset}, \mathcal{V}_\mathrm{others}$ defined as follows:
\begin{align*}
  \mathcal{V}_{\mathrm{set},n,\ell,m}
  & = \{ \mathit{set}_{n,\ell,m} \mid m \in \iival{1}{M} \},
  &
  \mathcal{V}_{\mathrm{set},n}
  & = \bigcup_{\ell=1}^L \mathcal{V}_{\mathrm{set},n,\ell},
  &
  \mathcal{V}_\mathrm{set}
  & = \bigcup_{n=1}^N \mathcal{V}_{\mathrm{set},n};
  \\
  \mathcal{V}_{\mathrm{reset},n}
  & = \{ \mathit{reset}_{n,m} \mid m \in \iival{1}{M} \},
  &
  \mathcal{V}_\mathrm{reset}
  & = \bigcup_{n=1}^N \mathcal{V}_{\mathrm{reset},n};
  \\
  \mathcal{V}_\mathrm{other}
  & = \{ \mathit{other}_i \mid i \in \iival{1}{I}  \},
\end{align*}
and it also contains output tokens representing the subsets of $\iival{1}{N}$.

For an input sequence $s = \sigma_1 \cdots \sigma_T$, and for
$n \in \iival{1}{N}$ and $t \in \iival{1}{T}$,
we say that \emph{the $n$-th sequence is matched in $s$ at time $t$} if there are
$t_1 < \cdots < t_L \leq t$ such that $\sigma_{t_\ell} \in \mathcal{V}_{\mathrm{set},n,\ell}$ and there
is no $t'$ with $t_1 < t' \leq t$ such that $\sigma_{t'} \in \mathcal{V}_{\mathrm{reset},n}$.
Then, the expected output sequence $\gamma_1 \cdots \gamma_T$ is the one where
$\gamma_t$ is the token representing the set of indices $n$ such that the $n$-th sequence is
matched in $s$ at time $t$.

To define datasets, we introduce the following parameters:
$p_\mathrm{set}, p_\mathrm{reset}, p_\mathrm{other} \in [0,1]$ satisfying
$p_\mathrm{set} + p_\mathrm{reset} + p_\mathrm{other} = 1$,
and
$T_\mathrm{min},T_\mathrm{max} \in \naturals$ satisfying $T_\mathrm{min} \leq T_\mathrm{max}$.
Then, an input sequence $\rv{U}_1 \cdots \rv{U}_\rv{T}$ is generated by first drawing
$\rv{T} \sim \mathrm{Unif}(\iival{T_\mathrm{min}}{T_\mathrm{max}})$ and then drawing tokens
according to the following probabilities:
\begin{gather*}
  \probop(\rv{U}_t = \mathit{set}_{n,\ell,m}) =
  \frac{p_\mathrm{set}}{|\mathcal{V}_\mathrm{set}|},
  \qquad
  \probop(\rv{U}_t = \mathit{reset}_{n,\ell,m}) =
  \frac{p_\mathrm{reset}}{|\mathcal{V}_\mathrm{reset}|},
  \qquad
  \probop(\rv{U}_t = \mathit{other}_i) =
  \frac{p_\mathrm{other}}{|\mathcal{V}_\mathrm{other}|}.
\end{gather*}

For the experiments, we adopt a single-number parametrisation \dsequences{n} where increasing the value of
$n \in \posnaturals$ increases the difficulty of the task as it corresponds to
the length of the sequences to match.
Table~\ref{tab:sequences_datasets} reports the details of the parameterisation in $n$,
and also the values of the other parameters (derived or fixed) for the values of $n$ that we
consider in the experiments.

\begin{table}[htb]
\small
  \centering
  \caption{Details of \dsequences{n}.}
  \vspace{.2em}
  \label{tab:sequences_datasets}
  \fbox{
    \begin{minipage}{\linewidth}
      \vspace{.8em}
      \centering
      \begin{tabular}{l@{\hspace{2.5em}}rrrrrrrrcc@{\hspace{2em}}r}
        \multicolumn{11}{l}{\textbf{Training}} \\[.2em]
        \toprule
        & \multicolumn{8}{c}{\emph{Parameter values}} \\
        \cmidrule(r){2-9}
        \emph{Parametrisation} & $N$ & $L$ & $M$ & $I$ & $p_\mathrm{set}$ & $p_\mathrm{reset}$ &
        $T_\mathrm{min}$ & $T_\mathrm{max}$ && $|\mathcal{V}|$ & Size \\
        \midrule
        \dsequences{n}  &  2  & $n$ &  2  &  10 & 0.5 & 0.02  & 256  &  512 && 4$n{+}$19   & 20k \\
        \midrule
        \dsequences{2}  &  2  &  2  &  2  &  10 & 0.5 & 0.02  & 256  &  512 &&  27         & 20k \\
        \dsequences{3}  &  2  &  3  &  2  &  10 & 0.5 & 0.02  & 256  &  512 &&  31         & 20k \\
        \dsequences{4}  &  2  &  4  &  2  &  10 & 0.5 & 0.02  & 256  &  512 &&  35         & 20k \\
        \dsequences{8}  &  2  &  8  &  2  &  10 & 0.5 & 0.02  & 256  &  512 &&  51         & 20k \\
        \dsequences{16} &  2  & 16  &  2  &  10 & 0.5 & 0.02  & 256  &  512 &&  83         & 20k \\
        \bottomrule
      \end{tabular}

      \vspace{1em}
      \centering
      \begin{tabular}{l@{\hspace{2.5em}}rrrrrrrrcc@{\hspace{2em}}r}
        \multicolumn{11}{l}{\textbf{Validation}} \\[.2em]
        \toprule
        & \multicolumn{8}{c}{\emph{Parameter values}} \\
        \cmidrule(r){2-9}
        \emph{Parametrisation} & $N$ & $L$ & $M$ & $I$ & $p_\mathrm{set}$ & $p_\mathrm{reset}$ &
        $T_\mathrm{min}$ & $T_\mathrm{max}$ && $|\mathcal{V}|$ & Size \\
        \midrule
        \dsequences{n}  &  2  & $n$ &  2  &  10 & 0.3  & 0.002  &  1024 &  2048 && 4$n{+}$19 & 1k  \\
        \midrule
        \dsequences{2}  &  2  &  2  &  2  &  10 & 0.3  & 0.002  &  1024 &  2048 &&  27       & 1k  \\
        \dsequences{3}  &  2  &  3  &  2  &  10 & 0.3  & 0.002  &  1024 &  2048 &&  31       & 1k  \\
        \dsequences{4}  &  2  &  4  &  2  &  10 & 0.3  & 0.002  &  1024 &  2048 &&  35       & 1k  \\
        \dsequences{8}  &  2  &  8  &  2  &  10 & 0.3  & 0.002  &  1024 &  2048 &&  51       & 1k  \\
        \dsequences{16} &  2  & 16  &  2  &  10 & 0.3  & 0.002  &  1024 &  2048 &&  83       & 1k  \\
        \bottomrule
      \end{tabular}

      \vspace{1em}
      \centering
      \begin{tabular}{l@{\hspace{2.5em}}rrrrrrrrcc@{\hspace{2em}}r}
        \multicolumn{11}{l}{\textbf{Evaluation}} \\[.2em]
        \toprule
        & \multicolumn{8}{c}{\emph{Parameter values}} \\
        \cmidrule(r){2-9}
        \emph{Parametrisation} & $N$ & $L$ & $M$ & $I$ & $p_\mathrm{set}$ & $p_\mathrm{reset}$ &
        $T_\mathrm{min}$ & $T_\mathrm{max}$ && $|\mathcal{V}|$ & Size \\
        \midrule
        \dsequences{n}  & 2 & $n$ & 2 & 10 & $10^{-5}$ & $10^{-6}$ & $2^{10}$ & $2^{10}$ && 4$n{+}$19&1k \\
        \midrule
        \dsequences{2}  & 2 &  2  & 2 & 10 & $10^{-5}$ & $10^{-6}$ & $2^{10}$ & $2^{10}$ && 27      & 1k \\
        \dsequences{3}  & 2 &  3  & 2 & 10 & $10^{-5}$ & $10^{-6}$ & $2^{10}$ & $2^{10}$ && 31      & 1k \\
        \dsequences{4}  & 2 &  4  & 2 & 10 & $10^{-5}$ & $10^{-6}$ & $2^{10}$ & $2^{10}$ && 35      & 1k \\
        \dsequences{8}  & 2 &  8  & 2 & 10 & $10^{-5}$ & $10^{-6}$ & $2^{10}$ & $2^{10}$ && 51      & 1k \\
        \dsequences{16} & 2 & 16  & 2 & 10 & $10^{-5}$ & $10^{-6}$ & $2^{10}$ & $2^{10}$ && 83      & 1k \\
        \bottomrule
      \end{tabular}
    \vspace{.8em}
    \end{minipage}
  }
\end{table}

\paragraph{Induction Heads.}
The \code{InductionHeads} task is parametrised by the number $M \in \posnaturals$ of candidate tokens to recall.

The vocabulary is $\mathcal{V} = \mathcal{V}_\mathrm{mem} \cup \{ \mathit{recall}, \mathit{pad} \}$
where $\mathcal{V}_\mathrm{mem}$ is a set of
$M$ tokens,
$\mathit{recall}$ is the recall token (not included in $\mathcal{V}$),
and
$\mathit{pad}$ is the padding token (not included in $\mathcal{V}$).

Input sequences are of the form
\begin{align*}
  s =
  \sigma_1,\ldots,\sigma_{t_0-1},\mathit{recall},\sigma_{t_0+1},\ldots,\sigma_T,\mathit{recall}
  \qquad \text{ where }\quad \sigma_t \in \mathcal{V}_\mathrm{mem},
\end{align*}
and the expected output is $\sigma_{t_0+1}$.

To define datasets, we introduce the parameters
$T^0_\mathrm{min},T^0_\mathrm{max},T_\mathrm{min},T_\mathrm{max} \in \naturals$ satisfying
$T^0_\mathrm{min} \leq T^0_\mathrm{max} \leq T_\mathrm{min} \leq T_\mathrm{max}$.
Then, an input sequence $\rv{U}_1 \cdots \rv{U}_\rv{T}$ is generated by first drawing
\begin{align*}
  \rv{T}_0 \sim \mathrm{Unif}(\iival{T^0_\mathrm{min}}{T^0_\mathrm{max}}),
  \qquad
  \rv{T} \sim \mathrm{Unif}(\iival{T_\mathrm{min}}{T_\mathrm{max}}),
\end{align*}
then setting $\rv{U}_{\rv{T}_0} = \rv{U}_{\rv{T}} = \mathit{recall}$, and finally drawing tokens for
the other positions:
\begin{align*}
  \rv{U}_t \sim \mathrm{Unif}(\mathcal{V})
  \qquad\forall t \in \iivalro{1}{\rv{T}_0} \cup \iivalo{\rv{T}_0}{\rv{T}}.
\end{align*}

For the experiments, we adopt a single-number parametrisation \dheads{n} where increasing the value
of $n \in \posnaturals$ increases the difficulty of the task as it is the number
of candidate tokens to recall.
Table~\ref{tab:inductionheads_datasets} reports the details of the parameterisation in $n$, and also
the values of the other parameters (derived or fixed) for the values of $n$ that we consider in the
experiments.

\begin{table}[htb]
\small
  \centering
  \caption{Details of \dheads{n}.}
  \vspace{.2em}
  \label{tab:inductionheads_datasets}
  \fbox{
    \begin{minipage}{\linewidth}
      \vspace{0.8em}
      \centering
      \begin{tabular}{l@{\hspace{2.5em}}rrrrrcc@{\hspace{2em}}r}
        \multicolumn{8}{l}{\textbf{Training}} \\[.2em]
        \toprule
        & \multicolumn{5}{c}{\emph{Parameter values}} \\
        \cmidrule(r){2-6}
        \emph{Parametrisation} & $M$ & $T^0_\mathrm{min}$ & $T^0_\mathrm{max}$ & $T_\mathrm{min}$ &
        $T_\mathrm{max}$ && $|\mathcal{V}|$ & Size \\
        \midrule
        \dheads{n}   & $n$  &  0  &  30 & 35       &  512       && $n{+}$2 & 20k  \\
        \midrule
        \dheads{16}  &  16  &  0  &  30 & 35       &  512       && 18      & 20k   \\
        \dheads{30}  &  30  &  0  &  30 & 35       &  512       && 32      & 20k   \\
        \bottomrule
      \end{tabular}

      \vspace{1em}
      \centering
      \begin{tabular}{l@{\hspace{2.5em}}rrrrrcc@{\hspace{2em}}r}
        \multicolumn{8}{l}{\textbf{Validation}} \\[.2em]
        \toprule
        & \multicolumn{5}{c}{\emph{Parameter values}} \\
        \cmidrule(r){2-6}
        \emph{Parametrisation} & $M$ & $T^0_\mathrm{min}$ & $T^0_\mathrm{max}$ & $T_\mathrm{min}$ &
        $T_\mathrm{max}$ && $|\mathcal{V}|$ & Size\\
        \midrule
        \dheads{n}   & $n$  &  0  &  50 &  1024       &  2048      && $n{+}$2 & 1k  \\
        \midrule
        \dheads{16}  &  16  &  0  &  50 &  1024       &  2048      && 18      & 1k  \\
        \dheads{30}  &  30  &  0  &  50 &  1024       &  2048      && 32      & 1k  \\
        \bottomrule
      \end{tabular}

      \vspace{1em}
      \centering
      \begin{tabular}{l@{\hspace{2.5em}}rrrrrcc@{\hspace{2em}}r}
        \multicolumn{8}{l}{\textbf{Evaluation}} \\[.2em]
        \toprule
        & \multicolumn{5}{c}{\emph{Parameter values}} \\
        \cmidrule(r){2-6}
        \emph{Parametrisation} & $M$ & $T^0_\mathrm{min}$ & $T^0_\mathrm{max}$ & $T_\mathrm{min}$ &
        $T_\mathrm{max}$ && $|\mathcal{V}|$ & Size \\
        \midrule
        \dheads{n}   & $n$ &  0  & 50 & $2^{10}$ & $2^{10}$ && $n{+}$2 & 1k   \\
        \midrule
        \dheads{16}  & 16  &  0  & 50 & $2^{10}$ & $2^{10}$ && 18      & 1k   \\
        \dheads{30}  & 30  &  0  & 50 & $2^{10}$ & $2^{10}$ && 32      & 1k   \\
        \bottomrule
      \end{tabular}
    \vspace{.8em}
    \end{minipage}
  }
\end{table}

\FloatBarrier
\subsection{Models}

\begin{table}[p]
  \small
\centering
\caption{Model configurations used in the experiments, along with dimensions and parameter count.
\emph{Legend:} $V$ is the vocabulary size, and $L$ is the number of layers.}
\vspace{.4em}
\label{tab:models_summary}
\begin{tabular}{l@{\hspace{2.5em}}l@{\hspace{2.7em}}rrrc@{\hspace{2em}}l}
\toprule
& & \multicolumn{3}{c}{\hspace{-1em}\textbf{Dimensions}} \\
\cmidrule(r){3-5}
\textbf{Model}&\hspace{-.1em}\textbf{Config} &
$d_\mathrm{model}$ & $d_\mathrm{state}$& $d_\mathrm{conv}$ &&
\hspace{-.3em}\textbf{Trainable Parameters} \\
\midrule
MinMax &\tsc{s} & 90~~& 40~~& 90~~&& 64k${} \cdot L$ $\;+\,$ \phantom{1}90${} \cdot V$ $\,+$ 180 \\
MinMax &\tsc{m} & 90~~& 90~~& 90~~&& 82k${} \cdot L$ $\;+\,$ \phantom{1}90${} \cdot V$ $\,+$ 180 \\
\midrule
Mamba    &\tsc{s} &  95~~& 3040~~& 570~~&& 67k${} \cdot L$ $\;+\,$ \phantom{1}95${}  \cdot V$        \\
Mamba    &\tsc{m} & 100~~& 3200~~& 600~~&& 74k${} \cdot L$ $\;+\,$ 100${} \cdot V$                   \\
\midrule
sLSTM    &\tsc{s} &  96~~&  384~~& 288~~&& 56k${} \cdot L$ $\;+\,$ 192${} \cdot V$ $\,+$ \phantom{1}96 \\
sLSTM    &\tsc{m} & 100~~&  400~~& 300~~&& 79k${} \cdot L$ $\;+\,$ 200${} \cdot V$ $\,+$ 100           \\
\midrule
mLSTM    &\tsc{s} &  96~~& 9412~~& 576~~&& 64k${} \cdot L$ $\;+\,$ 192${} \cdot V$ $\,+$ \phantom{1}96 \\
mLSTM    &\tsc{m} &  97~~&16644~~& 768~~&& 86k${} \cdot L$ $\;+\,$ 194${} \cdot V$ $\,+$ \phantom{1}97 \\
\midrule
xLSTM-L &\tsc{s} & 72~~& 688~~& 0~~&& 63k${} \cdot L$ $\;+\,$ 144${} \cdot V$ $\,+$ \phantom{1}72 \\
\midrule
DeltaNet &\tsc{s} & 90~~& 2700~~& 810~~&& 64k${} \cdot L$ $\;+\,$ \phantom{1}90${} \cdot V$ $\,+\,$ \phantom{1}90 \\
\bottomrule
\end{tabular}
\end{table}

\begin{table}[p]
  \footnotesize
  \centering
  \caption{Number of trainable parameters for each model on each benchmark. \emph{Legend:} $V$ is
    the vocabulary size; $L$ is the number of layers. The range reported for MinMax aggregates the
  counts for different configurations obtained via the optional flags \tsc{og} and \tsc{gc}.}
  \vspace{.2em}
  \fbox{
    \begin{minipage}{\linewidth}
      \vspace{.7em}
      \centering
      \begin{tabular}{rrcr@{\hspace{2.5em}}rrrrrr}
        \multicolumn{10}{c}{\dlatching{n}} \\[.1em]
        \toprule
        $n$ & $V$ & Config & $L$ & MinMax & Mamba & xLSTM-L & mLSTM & sLSTM & DeltaNet \\
        \midrule
        4 & 20 & \tsc{s} & 2 & 129k & 136k & 129k & 131k & 117k & 129k \\
        8 & 40 & \tsc{s} & 2 & 131k & 138k & 132k & 135k & 121k & 131k \\
        16 & 80 & \tsc{s} & 2 & 135k & 142k & 138k & 143k & 128k & 135k \\
        32 & 160 & \tsc{s} & 2 & 142k & 149k & 149k & 158k & 144k & 142k \\
        64 & 320 & \tsc{s} & 2 & 156k & 165k & 172k & 189k & 174k & 156k \\
        128 & 640 & \tsc{s} & 2 & 185k & 195k & 218k & 250k & 236k & 185k \\
        256 & 1280 & \tsc{s} & 2 & 243k & 256k & 310k & 373k & 359k & 243k \\
        512 & 2560 & \tsc{s} & 2 & 358k & 377k & 495k & 619k & 605k & 358k \\
        \bottomrule
      \end{tabular}
      \vspace{1.5em}
      \begin{tabular}{rrcr@{\hspace{2.5em}}rrrrrr}
        \multicolumn{10}{c}{\dsequences{n}} \\[.1em]
        \toprule
        $n$ & $V$ & Config & $L$ & MinMax & Mamba & xLSTM-L & mLSTM & sLSTM & DeltaNet \\
        \midrule
        2 & 27 & \tsc{s} & 2 & 130k & 137k & 130k & 133k & 118k & 130k \\
        3 & 31 & \tsc{s} & 3 & 194k & 204k & 194k & 197k & 175k & 194k \\
        4 & 35 & \tsc{s} & 4 & 258k & 271k & 257k & 261k & 233k & 258k \\
        8 & 51 & \tsc{s} & 8 & 514k & 541k & 511k & 519k & 461k & 514k \\
        16 & 83 & \tsc{s} & 16 & 1026k & 1079k & 1020k & 1035k & 919k &1026k \\
        \bottomrule
      \end{tabular}
      \vspace{1.5em}
      \begin{tabular}{rrcr@{\hspace{2.5em}}rrrrrr}
        \multicolumn{10}{c}{\dheads{n}} \\[.1em]
        \toprule
        $n$ & $V$ & Config & $L$ & MinMax & Mamba & xLSTM-L & mLSTM & sLSTM & DeltaNet \\
        \midrule
        16 & 18 & \tsc{s} & 2 & 129k & 136k & 129k & 131k & 116k & 129k \\
        30 & 32 & \tsc{s} & 2 & 130k & 137k & 131k & 134k & 119k & 130k \\
        \midrule
        16 & 18 & \tsc{m} & 2 & 165k & 150k & --- & 175k & 161k & --- \\
        30 & 32 & \tsc{m} & 2 & 167k & 151k & --- & 178k & 164k & --- \\
        \bottomrule
      \end{tabular}

      \vspace{1em}
    \end{minipage}
  }
  \label{tab:concrete_model_sizes}
\end{table}

The experiments include the following models.

\begin{table}[tbh]
  \centering
  \small
  \caption{%
    Hyperparameters of the models that are fixed across benchmarks; 
    all the others are left to their default values. The size of each model is scaled by varying 
    its number $L$ of layers 
    and 
    its width (\code{d_model}, \code{embedding_dim} or \code{hidden_size}),
    as described in the text.%
  }
  \vspace{.4em}
  \label{tab:model-hparams}
  \begin{tabular}{l@{\hspace{2em}}l@{\hspace{2em}}c}
    \toprule
    \textbf{Model} & \textbf{Hyperparameter} & {\textbf{Value}} \\
    \midrule
    MinMax & \code{output_gate} & {True} \\
           & \code{conv_type} & {basic} \\
    \midrule
    Mamba & \code{d_state} & 16 \\
          & \code{d_conv} & 4 \\
          & \code{expand} & 2 \\
          & \code{dt_rank} & {$\lceil \code{d_model} / 16 \rceil$} \\
          & \code{dropout} & 0  \\
    \midrule
    sLSTM & \code{num_heads} & 4 \\
          & \code{conv1d_kernel_size} & 4 \\
          & \code{proj_factor} & 1.3 \\
          & \code{act_fn} & {GeLU} \\
          & \code{dropout} & 0.1 \\
    \midrule
    mLSTM & \code{num_heads} & 4 \\
          & \code{qkv_proj_blocksize} & 4 \\
          & \code{conv1d_kernel_size} & 4 \\
          & \code{proj_factor} & 2 \\
          & \code{dropout} & 0.1 \\
    \midrule
    xLSTM-L & \code{num_heads} & 4 \\
            & \code{chunk_size} & 32 \\
            & \code{qk_dim_factor} & 0.5 \\
            & \code{v_dim_factor} & 1 \\
            & \code{ffn_proj_factor} & 2.667 \\
            & \code{gate_soft_cap} & 15 \\
            & \code{output_logit_soft_cap} & 30 \\
    \midrule
    DeltaNet & \code{num_heads} & 3 \\
             & \code{intermediate_size} & 110 \\
             & \code{expand_k}, \code{expand_v} & 1 \\
             & \code{use_short_conv} & {True} \\
             & \code{conv_size} & 4 \\
             & \code{use_gate} & {False} \\
             & \code{allow_neg_eigval} & {False} \\
    \bottomrule
  \end{tabular}
\end{table}

\begin{description}[
  style=sameline,
  labelindent=0pt,
  labelwidth=\widthof{\bfseries DeltaNet},
  leftmargin=!,
]
  \item[MinMax]
    We use the implementation described in
    Appendix~\ref{sec:minmax-implementation}, with scalar-state neurons only.
    We scale its size by varying \code{d_model}, the number of scalar-state neurons
    per layer, and the number $L$ of stacked layers: \code{d_model} = 90 with 40
    neurons per layer (\tsc{s}) or 90 neurons per layer (\tsc{m}).
    Every layer has the output gate and the basic convolution
    (Table~\ref{tab:model-hparams}).
    Layers are pre-normed with LayerNorm, the input and output embeddings are tied,
    and no dropout is applied to the neurons.
    All other parameters are left to their default values.
  \item[Mamba]
    Introduced in \citep{gu2024mamba}, as a state-space model whose parameters depend on
    the input (selective state spaces), with a hardware-aware parallel scan.
    We use the Mamba block (\code{mamba_ssm.Mamba}) of the official PyTorch
    implementation \citep{mambaimpl} version \code{2.2.5}.
    We scale its size by varying \code{d_model} and the number $L$ of stacked Mamba
    blocks, while setting all the other parameters to the values specified in
    \citep{gu2024mamba} (Table~\ref{tab:model-hparams}).
    Blocks are pre-normed with RMSNorm, with residual connections between blocks; the
    output of the last block is passed to the output layer directly, without adding the
    residual stream and without a final normalisation.
    The input and output embeddings are tied.
    Weights are initialised as in \citep{gu2024mamba}: embeddings with standard deviation
    0.02, zero biases, and output projections rescaled by $1/\sqrt{L}$.
  \item[sLSTM]
    By sLSTM we mean xLSTM consisting of sLSTM blocks only \citep{beck2024xlstm}, that build on
    LSTM cells~\citep{hochreiters97} by introducing exponential gating, normaliser and stabiliser
    states, and block-diagonal (multi-head) recurrent connections.
    We use the official PyTorch implementation~\citep{xlstmimpl} version~\code{2.0.5}
    (\code{xLSTMLMModel} with \code{slstm_at} covering all blocks), with its CUDA
    kernel for the sLSTM cell.
    We scale its size by varying \code{embedding_im} and the number~$L$ of stacked
    blocks, while setting the other parameters as in Table~\ref{tab:model-hparams}: the
    values specified in~\citep{beck2024xlstm}, where \code{proj_factor} and
    \code{act_fn} are those of the gated feed-forward network (its size rounded up to
    a multiple of~64), and a dropout of~0.1.
    Blocks are pre-normed with LayerNorm and followed by a final LayerNorm; the input
    and output embeddings are not tied.
  \item[mLSTM]
    By mLSTM we mean xLSTM consisting of mLSTM blocks only \citep{beck2024xlstm}, that have 
    a matrix memory updated by a covariance (outer-product) rule with
    exponential gating, and no hidden-to-hidden connections, so that it can be computed
    in parallel.
    We use the official PyTorch implementation~\citep{xlstmimpl} version~\code{2.0.5}
    (\code{xLSTMLMModel} without sLSTM blocks).
    We scale its size by varying~\code{embedding_dim} and the number~$L$ of stacked
    blocks, while setting the other parameters as in Table~\ref{tab:model-hparams}: the
    values specified in~\citep{beck2024xlstm}, where \code{proj_factor} is the factor of
    the up-projection (its size rounded up to a multiple of~64), and a dropout of~0.1.
    mLSTM blocks have no separate feed-forward network.
    Blocks are pre-normed with LayerNorm and followed by a final LayerNorm; the input
    and output embeddings are not tied.
  \item[xLSTM-L]
    Introduced in~\citep{beck2025xlstm} as the architecture of xLSTM~7B: mLSTM~blocks
    redesigned for training and inference efficiency, with soft-capped gates and
    logits, and fused Triton kernels.
    We use the official PyTorch implementation~\citep{xlstmimpl} version~\code{2.0.5}
    (\code{xLSTMLarge}), with its chunkwise Triton kernels.
    We scale its size by varying~\code{embedding_dim} and the number~$L$ of stacked
    blocks, while setting \code{num_heads} and \code{chunk_size} as in
    Table~\ref{tab:model-hparams}, and leaving the other parameters to their default
    values, also reported there (the feed-forward size is rounded up to a multiple of~64); there is
    no bias and no dropout.
    Blocks are pre-normed with RMSNorm and followed by a final RMSNorm; the input and
    output embeddings are not tied.
  \item[DeltaNet]
    Based on the delta-rule update of linear Transformers introduced in
    \citep{schlagis21}, and improved in \citep{yang2024parallelizing}, which makes its
    training parallel over the sequence length through a chunkwise algorithm, and
    modernises the architecture with SiLU and L2 normalisation of queries and keys,
    short convolutions, and output normalisation.
    We use the reference implementation of the flash-linear-attention
    library~\citep{flashlinearattention} version~\code{0.5.1}.
    We scale its size by varying \code{hidden_size} and the number~$L$ of stacked
    blocks, while setting the parameters in Table~\ref{tab:model-hparams}, with tied
    input and output embeddings, and leaving the other parameters to their default
    values.
    The size of the feed-forward layers~(\code{intermediate_size}) is set so that the
    number of parameters is comparable to the one of MinMax.
    Its kernels require~\code{bf16}, hence it always runs under~\code{bf16} automatic mixed precision.
\end{description}

Table~\ref{tab:models_summary} reports the model configurations we consider in the experiments.
For each model we can have a smaller variant~($\tsc{s}$) and a larger variant~($\tsc{m}$).
The table reports the formula to compute the number of trainable parameters as a function of the
number of layers~$L$ and vocabulary size~$V$, and it also reports the following dimensions
for each model:
\begin{itemize}
  \item
    $d_\mathrm{model}$ that is \code{d_model} for MinMax and Mamba, it is
    \code{embedding_dim} for sLSTM, mLSTM, xLSTM-L, and it is \code{hidden_size} for DeltaNet;
  \item
    $d_\mathrm{state}$ that is the total number of scalar values in the state of a single layer;
  \item
    $d_\mathrm{conv}$ that is the number of past input values that the convolutions of a
    single layer need to keep, i.e., the number of channels times the kernel size minus one
    (zero if the layer has no convolution).
\end{itemize}
The above dimensions allow for a closer comparison of the models rather than simply comparing the
total number of trainable parameters.
Interestingly, the table makes it evident that the state dimension $d_\mathrm{state}$ of MinMax
RNCs is significantly smaller than the one of the other models, even though all models have
comparable $d_\mathrm{model}$ and parameter count.
We have first chosen $d_\mathrm{model}=90$ for MinMax RNCs as an educated guess for a reasonable
value,
and then we have chosen $d_\mathrm{model}$ for the other models in order to obtain models having a
comparable number of trainable parameters; for DeltaNet, we have kept $d_\mathrm{model} = 90$ and
chosen the size of its feed-forward layers (\code{intermediate_size}) instead.

Table~\ref{tab:concrete_model_sizes} reports parameter counts for the model configurations
considered in the experiments.

\FloatBarrier
\subsection{Trainers}

We introduce the notion of a \emph{trainer} to group the choice of the optimiser, learning rate
schedule, and set of batch sizes.
We use a dedicated trainer for each family of models.
They are defined in Table~\ref{tab:trainers}, and further described and discussed below.

\begin{table}[ht!]
  \caption{Trainers.}
  \vspace{.1em}
  \footnotesize
  \begin{tabular}{lc@{\hspace{1em}}c@{\hspace{1.3em}}c@{\hspace{.8em}}lc@{\hspace{.5em}}lcc@{\hspace{.7em}}c}
\toprule
& \multicolumn{4}{c}{\textbf{Optimiser}} & & \multicolumn{2}{c}{\hspace{-1.0em}\textbf{Learning Rate}} \\
\cmidrule(lr){2-5} \cmidrule(r){7-8}
\textbf{Trainer} & Type & $\lambda$ & $\beta_1$ & ~~$\beta_2$ &&~~Peak &
\multicolumn{1}{c}{~~~~~Schedule}
                 && \textbf{Batch sizes} \\
\midrule
\code{trainer_minmax} &
Adam  & $10^{\smallminus 4}$ & 0.9 & 0.999 &&~~$10^{\smallminus 3}$ & Step ($\gamma{=}0.5$, step$=$5)
      && 64
\\
\code{trainer_minmax_bs8} &
Adam  & $10^{\smallminus 4}$ & 0.9 & 0.999 &&~~$10^{\smallminus 3}$ & Step ($\gamma{=}0.5$, step$=$5)
      && 8
\\
\code{trainer_xlstm} &
AdamW & $0.1$    & 0.9 & 0.99              &&~~$10^{\smallminus 3}$ & Linear warmup + cosine && 64,~128
\\
\code{trainer_xlstm_bs8} &
AdamW & $0.1$    & 0.9 & 0.99              &&~~$10^{\smallminus 3}$ & Linear warmup + cosine && 8
\\
\code{trainer_mamba} &
Adam  & no decay & 0.9 & 0.999             &&~~$10^{\smallminus 3}$ & None && 64
\\
\code{trainer_mamba_bs8} &
Adam  & no decay & 0.9 & 0.999             &&~~$10^{\smallminus 3}$ & None && 8
\\
\code{trainer_deltanet} &
AdamW & $0.1$ & 0.9 & 0.95 &&~~$3{\cdot}10^{\smallminus 4}$
& Linear warmup + cosine && 64
\\
\bottomrule
\end{tabular}
\label{tab:trainers}
\end{table}

\paragraph{MinMax.}
The training configuration for MinMax is Adam optimiser with weight decay $\lambda = 10^{-4}$,
learning rate $10^{-3}$, and a step learning-rate schedule that halves the learning rate every 5
epochs ($\gamma = 0.5$);
batch size 64 for the default trainer, and batch size 8 for the dedicated trainer.
The rationale is to provide a rather standard and basic training setup.

\paragraph{xLSTM.}
The training configuration for sLSTM, mLSTM, and xLSTM-L is AdamW optimiser with $(\beta_1,
\beta_2) = (0.9, 0.99)$, weight decay $0.1$, $\varepsilon = 10^{-5}$, gradient clipping at 1.0;
learning rate $10^{-3}$ with a linear warmup over the first 10\% of total training steps
followed by cosine annealing; batch size ranges over $\{64, 128\}$ for the default trainer,
and we include a dedicate trainer using batch size 8.
The configuration is in line with the one used in \citep{beck2024xlstm}
for the experiments on synthetic benchmarks based on formal languages.

\paragraph{Mamba.}
The training configuration for Mamba is Adam optimiser without weight decay, learning rate
$10^{-3}$, no learning-rate schedule, gradient clipping at 1.0.
For batch size 8, it is the configuration used in \citep{gu2024mamba} for the InductionHeads
benchmark.
We also include the batch-size-64 variant for a closer comparison with the other
models---also noting that this configuration is in fact used in \citep{gu2024mamba} for
benchmarks other than InductionHeads (e.g., Selective Copying).

\paragraph{DeltaNet.}
The training configuration for DeltaNet is AdamW optimiser with $(\beta_1, \beta_2) = (0.9, 0.95)$,
weight decay $0.1$, $\varepsilon = 10^{-5}$; learning rate $3 \cdot 10^{-4}$ with a linear warmup
over the first 10\% of total training steps followed by cosine annealing to 10\% of the peak; batch
size 64.  The model runs under \code{bf16} automatic mixed precision, as required by the kernels of the
implementation of~\citet{yang2024parallelizing}.

\FloatBarrier
\subsection{Training Protocol}

\begin{table}[tb]
  \caption{Overview of the training protocol.}
  \vspace{.1em}
  \fbox{
  \footnotesize
  \begin{tabular}{ccccc@{\hspace{.7em}}c@{\hspace{.8em}}cc}
    \multicolumn{4}{c}{} & \multicolumn{3}{c}{\textbf{CI}} \\
    \cmidrule(lr){5-7}
    \textbf{Loss}          &
    \textbf{Metric}        &
    \textbf{Epochs}        &
    \textbf{Early stop}    &
    \textbf{Confidence} &
    \textbf{Margin} &
    \textbf{Method} &
    \textbf{Time limit}    \\
    \midrule
    cross-entropy & token accuracy & 30 & patience-based  & 99\%  & 0.1\% & bootstrap & 4h
  \end{tabular}
}
\label{tab:training_protocol}
\end{table}

Our training protocol is summarised in Table~\ref{tab:training_protocol}.

The loss function is \emph{cross-entropy} over all non-masked token positions.
The metric is \emph{token accuracy}: the number of correctly predicted tokens over the total
number of tokens to predict.

Each model-trainer pair defines a separate experiment.
For each batch size specified by the trainer,
we run independent trainings for different random seeds until we reach 99\% confidence that no
training run will produce a model that has validation accuracy 0.1\% higher than the one of the best
model found so far---trying at least 3 seeds to avoid poor statistical estimates, and no more than
10 seeds to limit the amount of compute.

We estimate 99$\%$ confidence intervals (CIs) for accuracies using the \emph{nonparametric
percentile bootstrap} over seeds with 10k resamples.

We use independently-generated training and validation datasets (per seed), having the
characteristics reported in
Tables~\ref{tab:latching_datasets},~\ref{tab:sequences_datasets},~\ref{tab:inductionheads_datasets}.
In particular, training datasets consist of 20k examples, and validation datasets consist of
1k~examples.

We use epoch-based training, where each epoch amounts to a full pass over the training dataset.
Models are trained for a maximum of \emph{30~epochs with early stopping}.
Specifically, we employ
patience-based early stopping on the validation loss. The
criterion monitors a sliding window of 3 epochs and triggers if no improvement
greater than $\delta = 10^{-5}$ is observed for 5 consecutive windows (patience = 5).

We set a \emph{time limit} of 4 hours for each model-trainer pair,
in order to contain the total computation time to carry out the experiments.

The \emph{best model} resulting from a full training run is the one that has highest validation
accuracy across all models (for all batch sizes and considered seeds); for a training run that
times out, it is the best model obtained when the run times out.

Although the training protocol is independent of the specific datasets considered, we recall size
and sequence length of the datasets we consider in our experiments (as described in
Tables~\ref{tab:latching_datasets},~\ref{tab:sequences_datasets},~\ref{tab:inductionheads_datasets}):
\begin{itemize}
  \item
    training datasets consist of 20k labelled sequences of length
    $\iival{256}{512}$ for \code{Latching} and \code{Sequences}, and of length
    $\iival{35}{512}$ for \code{InductionHeads};
  \item
    all validation datasets consist of 1k labelled sequences of length $\iival{1024}{2048}$.
\end{itemize}

\FloatBarrier
\subsection{Evaluation Protocol}

After training, the best models are evaluated on long sequences length $2^{10}$ ($\approx$ 1M).
We set a time limit of 20 hours to evaluate a single model.
We measure accuracy at each non-masked sequence steps,
also computing the average across non-masked steps and the minimum accuracy at any non-masked
step.
We recall that on \code{Latching} and \code{Sequences} no step is masked, whereas on
\code{InductionHeads} all steps are masked except the last one---hence accuracy is the one at
the last step.
We repeat the evaluation for different random seeds (with an independently-generated dataset for
each seed) until the 99\%-CIs for the step-average accuracy and step-minimum accuracy both satisfy
the following (with CIs computed as in training):
\begin{itemize}
  \item
    if the mean is in the range 0--50\%, then the error margin is below 20\%;
  \item
    if the mean is in the range 50\%--80\%, then the error margin is below 10\%;
  \item
    if the mean is in the range 80\%--90\%, then the error margin is below 5\%;
  \item
    if the mean is in the range 90\%--100\%, then the error margin is below 1\%.
\end{itemize}

Although the evaluation protocol is independent of the specific datasets considered, we recall that
the details of the datasets for our experiments are described in
Tables~\ref{tab:latching_datasets},~\ref{tab:sequences_datasets},~\ref{tab:inductionheads_datasets}.
In particular, each dataset consists of 1k labelled sequences of length $2^{10}$ ($\approx$ 1M).

\FloatBarrier
\subsection{Overview of the Included Experiments}

Table~\ref{tab:overview_experiments} summarises our experiments on the synthetic benchmarks.
For every benchmark, value of $n$, and included model configuration, we carry out training and
evaluation for the listed model-trainer pairs; also including the trainer variants using batch
size~8 when the table specifies ``+ \code{bs8}''.

\begin{table}[tbh]
  \centering
  \caption{Overview of the experiments.}
  \vspace{.1em}
  \label{tab:overview_experiments}
  \small
  \begin{tabular}{l@{}ccrl}
    \toprule
    \textbf{Benchmark} & $n$ & \textbf{Configurations} & \textbf{Model} & \textbf{Trainer} \\
    \midrule
    \multirow{6}{*}{\dlatching{2^n}} &
    \multirow{6}{*}{$\{2,3,\ldots, 9 \}$} &
    \multirow{6}{*}{small (\tsc{s})} &
    MinMax & \code{trainer_minmax}\\
           &&& Mamba & \code{trainer_mamba} \phantom{\code{x}}+ \code{bs8} \\
           &&& sLSTM & \code{trainer_xlstm} \\
           &&& mLSTM & \code{trainer_xlstm} \\
           &&& xLSTM-L & \code{trainer_xlstm} \\
           &&& DeltaNet & \code{trainer_deltanet} \\
    \midrule
    \multirow{6}{*}{\dsequences{n}} &
    \multirow{6}{*}{$\{2, 3, 4, 8, 16\}$} &
    \multirow{6}{*}{small (\tsc{s})} &
    MinMax & \code{trainer_minmax}\\
           &&& Mamba & \code{trainer_mamba} \\
           &&& sLSTM & \code{trainer_xlstm} \\
           &&& mLSTM & \code{trainer_xlstm} \\
           &&& xLSTM-L & \code{trainer_xlstm} \\
           &&& DeltaNet & \code{trainer_deltanet} \\
    \midrule
    \multirow{6}{*}{\dheads{n}} &
    \multirow{6}{*}{$\{16, 30\}$} &
    \multirow{6}{*}{small (\tsc{s}), medium (\tsc{m})} &
    MinMax & \code{trainer_minmax} + \code{bs8} \\
           &&& Mamba & \code{trainer_mamba} \phantom{\code{x}}+ \code{bs8} \\
           &&& sLSTM & \code{trainer_xlstm} \phantom{\code{x}}+ \code{bs8} \\
           &&& mLSTM & \code{trainer_xlstm} \phantom{\code{x}}+ \code{bs8} \\
           &&& xLSTM-L & \code{trainer_xlstm} \phantom{\code{x}}+ \code{bs8} \\
           &&& DeltaNet & \code{trainer_deltanet} \\
    \bottomrule
  \end{tabular}
\end{table}

\FloatBarrier
\subsection{Results: Accuracy}

Table~\ref{tab:experiments-all} summarises the results of our evaluation.
For each benchmark instance, the table reports the best evaluation accuracy of each model across the
considered configurations---i.e., model size and trainer.
For \dlatching{n} and \dsequences{n} the reported accuracy is the step-average accuracy;
and for \dheads{n} is the last-step accuracy.

Tables~\ref{tab:experiments-latching},~\ref{tab:experiments-sequences}, and~\ref{tab:experiments-inductionheads}
provide additional details for the evaluation on \dlatching{n}, \dsequences{n}, and \dheads{n},
respectively---they report
the train, validation, and evaluation accuracy for each model-trainer pair.

\begin{table}[tbh]
  \centering
  \caption{Summary of the evaluation: \emph{mean $\pm$ error margin} of $99\%$ CIs for
  evaluation accuracy (\%). \emph{Legend:} \cmark{} denotes perfect accuracy on all seeds.}
  \vspace{.3em}
  \label{tab:experiments-all}
  \setlength{\tabcolsep}{4pt}
  \renewcommand{\arraystretch}{1.15}
  \begin{tabular}{l @{\hspace{1.0em}} U%
      @{\hspace{1.3em}} U%
      @{\hspace{1.3em}} U%
      @{\hspace{1.3em}} U%
      @{\hspace{1.3em}} U%
      @{\hspace{1.3em}} U%
    }
    \toprule
    & \multicolumn{6}{c}{\textbf{Accuracy (\%)}} \\
    \cmidrule(r){2-7}
    \textbf{Benchmark} & {MinMax} & {xLSTM-L} & {mLSTM} & {sLSTM} & {Mamba} & {DeltaNet} \\
    \midrule
    \dlatchingSmall{4}   & \cmark & 25.4(.5) & 25.5(1.0) & \cmark    & 52.3(2.3) & 26.4(0.6) \\
    \dlatchingSmall{8}   & \cmark & 12.5(.8) & 13.5(0.8) & 62.8(1.4) & 70.5(0.6) & 13.3(0.1) \\
    \dlatchingSmall{16}  & \cmark & 6.1(.4)  & 7.8(0.3)  & 40.4(0.2) & 42.6(1.5) & 7.8(0.6) \\
    \dlatchingSmall{32}  & \cmark & 4.4(.5)  & 3.7(0.4)  & 4.6(0.3)  & 12.9(0.3) & 4.8(0.2) \\
    \dlatchingSmall{64}  & \cmark & 3.3(.3)  & 2.0(0.3)  & 5.1(0.2)  & 99.9(0.1) & 3.8(0.1) \\
    \dlatchingSmall{128} & \cmark & 5.9(.3)  & 2.2(0.1)  & 3.4(0.2)  & 90.6(1.0) & 2.8(0.1) \\
    \dlatchingSmall{256} & \cmark & 7.5(.5)  & 1.8(0.2)  & 2.7(0.4)  & 64.1(1.3) & 2.8(0.1) \\
    \dlatchingSmall{512} & \cmark & 0.1(.1)  & 2.4(0.2)  & 6.6(0.2)  & 65.2(1.7) & 5.3(0.1) \\
    \midrule
    \dsequencesSmall{2} & \cmark & 47 (1.1) & 47 (1.2) & 47.5(1.0) & 47.5(1.0) & 18.5(1.6) \\
    \dsequencesSmall{3} & \cmark & 63.1(0.9) & 47.1(0.2) & 63.1(0.9) & 63.1(0.9) & 7.8(1.2) \\
    \dsequencesSmall{4} & \cmark & 74.1(1.0) & 74.1(1.0) & 74.1(1.0) & 74.1(1.0) & 59.0(0.8) \\
    \dsequencesSmall{8} & \cmark & 90.5(0.5) & 90.5(0.5) & 90.5(0.5) & 90.5(0.5) & 1.6(0.2) \\
    \dsequencesSmall{16} & \cmark & 97 (0.3) & 97 (0.3) & 96.8(0.2) & 97.0(0.3) & 0.6(0.1) \\
    \midrule
    \dheadsSmall{16} & \cmark & 6.2(0.7) & 6.5(1.0) & 6.2(0.2) & \cmark & 5.3(0.5) \\
    \dheadsSmall{30} & \cmark & 3.1(0.9) & 3.0(0.6) & 3.0(0.6) & 3.5(0.2) & 3.6(0.6) \\
    \bottomrule
  \end{tabular}
\end{table}

\newcommand{\benchSpace}{\hspace{1.6em}}%
\newcommand{\benchInnerSpace}{\hspace{.5em}}%
\newcommand{\benchFirstSpace}{\hspace{0.7em}}
\newcommand{\benchSecondSpace}{\hspace{1.2em}}
\newcommand{\benchTrim}{2.1em}%

\begin{table}[p]
\centering
\caption{Results for \dlatching{n}: training accuracy of the best model, and \emph{mean~$\pm$~error
  margin} of $99\%$ CIs for validation and evaluation
  accuracy (\%). \emph{Legend:} \cmark{}~perfect accuracy on all seeds; superscript~\tout{} denotes
  time-out at training time.}
\vspace{.2em}
\label{tab:experiments-latching}
\fbox{%
  \begin{minipage}{\dimexpr\linewidth-2\fboxsep-2\fboxrule\relax}
\centering
\footnotesize
\vspace{.5em}
\setlength{\tabcolsep}{0pt}%
\newcolumntype{G}{S[table-format=2.1]@{\benchInnerSpace}S[table-format=2.1]@{\benchInnerSpace}S[table-format=2.1(2)]}%
\newcolumntype{H}{S[table-format=2.1]@{\benchInnerSpace}S[table-format=2.1(2)]@{\benchInnerSpace}S[table-format=2.1(2)]}%
\renewcommand{\arraystretch}{1.15}
\begin{tabular}{@{}l@{}}%
\begin{tabular}{@{}l@{\benchFirstSpace}l@{\benchSecondSpace}G@{\benchSpace}G@{\benchSpace}G@{}}
&
  \multicolumn{9}{c}{\textbf{Accuracy (\%)}} & \multicolumn{1}{c}{\emph{(Part~I)}} \\[.1em]
\cmidrule(lr){3-11}
&
& \multicolumn{3}{c@{\benchSpace}}{\dlatching{4}}
& \multicolumn{3}{c@{\benchSpace}}{\dlatching{8}}
& \multicolumn{3}{c}{\dlatching{16}} \\
\cmidrule(l{.5em}r{\benchTrim}){3-5}\cmidrule(l{.5em}r{\benchTrim}){6-8}\cmidrule(lr){9-11}
\textbf{Model}
& \textbf{Config}
& {Train} & {~Val~} & {Evaluation}
& {Train} & {~Val~} & {Evaluation}
& {Train} & {~Val~} & {Evaluation} \\
\midrule
MinMax &small&  \cmark & \cmark & \cmark   &\cmark&\cmark&\cmark   & \cmark & \cmark & \cmark    \\
xLSTM-L&small&  \cmark & \cmark & 25.4(.5) &\cmark&\cmark&12.5(.8) & \cmark & \cmark &  6.1(.4)  \\
xLSTM  &small&  \cmark & \cmark & 25.5(1.0)&\cmark&\cmark&13.5(0.8)& \cmark & \cmark &  7.8(0.3) \\
sLSTM  &small&  \cmarko & \cmarko & \cmark   &\cmarko&\cmarko&62.8(1.4)& \cmarko & \cmarko & 40.4(0.2) \\
Mamba  &small&  \cmark & \cmark & 52.3(2.3)&\cmark&\cmark&38.8(1.5)& \cmark & \cmark & 42.6(1.5) \\
Mamba&small,bs8&\cmark & \cmark & 27.7(1.2)&\cmark&\cmark&70.5(0.6)& \cmark & \cmark & 36.0(1.3) \\
DeltaNet&small& \cmark & \cmark & 26.4(0.6)&\cmark&\cmark&13.3(0.1)& \cmark & \cmark & 7.8(0.6) \\
\bottomrule
\end{tabular}
\\
\noalign{\vspace{\bigskipamount}}
\begin{tabular}{@{}l@{\benchFirstSpace}l@{\benchSecondSpace}G@{\benchSpace}G@{\benchSpace}G@{}}
&
  \multicolumn{9}{c}{\hspace{0em}\textbf{Accuracy (\%)}} & \multicolumn{1}{c}{\emph{(Part~II)}} \\[.1em]
\cmidrule(lr){3-11}
&
& \multicolumn{3}{c@{\benchSpace}}{\dlatching{32}}
& \multicolumn{3}{c@{\benchSpace}}{\dlatching{64}}
& \multicolumn{3}{c}{\dlatching{128}} \\
\cmidrule(l{.5em}r{\benchTrim}){3-5}\cmidrule(l{.5em}r{\benchTrim}){6-8}\cmidrule(lr){9-11}
\textbf{Model}
& \textbf{Config}
& {Train} & {~Val~} & {Evaluation}
& {Train} & {~Val~} & {Evaluation}
& {Train} & {~Val~} & {Evaluation} \\
\midrule
MinMax &small  &\cmark&\cmark& \cmark  &\cmark&\cmark& \cmark  &\cmark&\cmark& \cmark   \\
xLSTM-L&small  & 99.9 &\cmark& 4.4(.5) & 99.9 &\cmark& 3.3(.3) & 99.9 &\cmark& 5.9(.3)  \\
xLSTM  &small  &\cmark&\cmark& 3.7(0.4)&\cmark&\cmark& 2.0(0.3)&\cmark&\cmark& 2.2(0.1) \\
sLSTM  &small  &\cmarko&\cmarko& 4.6(0.3)&\cmarko&\cmarko& 5.1(0.2)&\cmarko&\cmarko& 3.4(0.2) \\
Mamba  &small  &\cmark&\cmark&12.9(0.3)&\cmark&\cmark&14.4(0.9)&\cmark&\cmark&87.3(1.6) \\
Mamba&small,bs8&\cmark&\cmark&13.0(1.3)&\cmark&\cmark&99.9(0.0)&\cmark&\cmark&90.6(1.0) \\
DeltaNet&small &\cmark&\cmark& 4.8(0.2)&\cmark&\cmark& 3.8(0.1)&\cmark&\cmark& 2.8(0.1) \\
\bottomrule
\end{tabular}
\\
\noalign{\vspace{\bigskipamount}}
\begin{tabular}{@{}l@{\benchFirstSpace}l@{\benchSecondSpace}H@{\benchSpace}H@{}}
&
  \multicolumn{6}{c}{\hspace{2em}\textbf{Accuracy (\%)}} & \multicolumn{1}{c}{\emph{(Part~III)}} \\[.1em]
\cmidrule(lr){3-8}
&
& \multicolumn{3}{c@{\benchSpace}}{\dlatching{256}}
& \multicolumn{3}{c}{\dlatching{512}} \\
\cmidrule(l{.5em}r{\benchTrim}){3-5}\cmidrule(lr){6-8}
\textbf{Model}
& \textbf{Config}
& {Train} & {~Val~} & {Evaluation}
& {Train} & {~Val~} & {Evaluation} \\
\midrule
MinMax  &small&     \cmark & \cmark    & \cmark    & \cmark & \cmark  & \cmark    \\
xLSTM-L &small&      99.9  &  99.8(0.1)&  7.5(.5)  &  0.2   & 0.4(0.1)&  0.1(.1)  \\
xLSTM   &small&     \cmark & \cmark    &  1.8(0.2) & \cmark & \cmark  &  2.4(0.2) \\
sLSTM   &small&     \cmarko & \cmarko    &  2.7(0.4) & \cmarko & \cmarko  &  6.6(0.2) \\
Mamba   &small&     \cmark & \cmark    & 64.1(1.3) & \cmark & \cmark  &  6.5(0.5) \\
Mamba   &small,bs8& \cmark & \cmark    & 31.5(1.5) & \cmark & \cmark  & 65.2(1.7) \\
DeltaNet&small& \cmark & \cmark & 2.8(0.1) & \cmark & \cmark & 5.3(0.1) \\
\bottomrule
\end{tabular}
\end{tabular}
\vspace{1.0em}
\end{minipage}%
}
\end{table}

\begin{table}[p]
\centering
\caption{Results for \dsequences{n}: training accuracy of the best model, and \emph{mean~$\pm$~error
  margin} of $99\%$ CIs for validation and evaluation accuracy (\%). \emph{Legend:} \cmark{}~perfect
accuracy on all seeds; superscript~\tout{} denotes time-out at training time.}
\vspace{.2em}
\label{tab:experiments-sequences}
\fbox{%
\begin{minipage}{\dimexpr\linewidth-2\fboxsep-2\fboxrule\relax}
\centering
\footnotesize
\vspace{.3em}
\setlength{\tabcolsep}{0pt}%
\newcolumntype{G}{S[table-format=2.1]@{\benchInnerSpace}S[table-format=2.1]@{\benchInnerSpace}S[table-format=2.1(2)]}%
\newcolumntype{H}{S[table-format=2.1]@{\benchInnerSpace}S[table-format=2.1(2)]@{\benchInnerSpace}S[table-format=2.1(2)]}%
\renewcommand{\arraystretch}{1.15}
\begin{tabular}{@{}l@{}}%
\begin{tabular}{@{}l@{\benchFirstSpace}l@{\benchSecondSpace}G@{\benchSpace}G@{\benchSpace}G@{}}
  & \multicolumn{9}{c}{\hspace{0em}\textbf{Accuracy (\%)}} & \multicolumn{1}{c}{\emph{(Part~I)}} \\[.1em]
\cmidrule(lr){3-11}
&
& \multicolumn{3}{c@{\benchSpace}}{\dsequences{2}}
& \multicolumn{3}{c@{\benchSpace}}{\dsequences{3}}
& \multicolumn{3}{c}{\dsequences{4}} \\
\cmidrule(l{.5em}r{\benchTrim}){3-5}\cmidrule(l{.5em}r{\benchTrim}){6-8}\cmidrule(lr){9-11}
\textbf{Model}
& \textbf{Config}
& {Train} & {~Val~} & {Evaluation}
& {Train} & {~Val~} & {Evaluation}
& {Train} & {~Val~} & {Evaluation} \\
\midrule
MinMax &small&\cmark &\cmark &\cmark   & \cmark & \cmark & \cmark   & \cmark & \cmark & \cmark   \\
xLSTM-L&small&\cmark &\cmark &47  (1.1)& \cmark & \cmark & 63.1(0.9)& \cmark & \cmark & 74.1(1.0)\\
xLSTM  &small&\cmarko&\cmarko&47  (1.2)& \cmarko& \cmarko& 47.1(0.2)& \cmarko& \cmarko& 74.1(1.0)\\
sLSTM  &small& 99.9  & 99.9  &47.5(1.0)& \cmark &  99.9  & 63.1(0.9)& \cmark &  99.9  & 74.1(1.0)\\
Mamba  &small&\cmark &\cmark &47.5(1.0)& \cmark & \cmark & 63.1(0.9)& \cmark & \cmark & 74.1(1.0)\\
DeltaNet&small&\cmark &\cmark &18.5(1.6)& \cmark & \cmark & 7.8(1.2)& \cmark & \cmark & 59.0(0.8)\\
\bottomrule
\end{tabular}
\\
\noalign{\vspace{\bigskipamount}}
\begin{tabular}{@{}l@{\benchFirstSpace}l@{\benchSecondSpace}G@{\benchSpace}G@{}}
  & \multicolumn{6}{c}{\hspace{0em}\textbf{Accuracy (\%)}} & \multicolumn{1}{c}{\emph{(Part~II)}} \\[.1em]
\cmidrule(lr){3-8}
&
& \multicolumn{3}{c@{\benchSpace}}{\dsequences{8}}
& \multicolumn{3}{c}{\dsequences{16}} \\
\cmidrule(l{.5em}r{\benchTrim}){3-5}\cmidrule(lr){6-8}
\textbf{Model}
& \textbf{Config}
& {Train} & {~Val~} & {Evaluation}
& {Train} & {~Val~} & {Evaluation} \\
\midrule
MinMax &small& \cmark  & \cmark & \cmark   & \cmark & \cmark     & \cmark   \\
xLSTM-L&small&  99.9   & \cmark & 90.5(0.5)& \cmark & \cmark     & 97  (0.3)\\
xLSTM  &small& \cmarko & \cmarko& 90.5(0.5)& \cmarko&  99.9\tout & 97  (0.3)\\
sLSTM  &small&  99.9   &  99.9  & 90.5(0.5)&  99.9  &  99.9      & 96.8(0.2)\\
Mamba  &small& \cmark  & \cmark & 90.5(0.5)& \cmark & \cmark     & 97.0(0.3)\\
DeltaNet&small& \cmark & \cmark & 1.6(0.2)& \cmark & \cmark & 0.6(0.1)\\
\bottomrule
\end{tabular}
\end{tabular}
\vspace{1.0em}
\end{minipage}%
}
\end{table}

\begin{table}[p]
\footnotesize
\centering
\caption{Results for \dheads{n}: training accuracy of the best model, and
  \emph{mean~$\pm$~error margin} of $99\%$ CIs for validation and evaluation accuracy (\%).
  \emph{Legend:}
  L2/L3 denotes 2/3 layers;
  \cmark{}~perfect accuracy on all seeds;
  superscript~\tout{} denotes time-out at training time;
  entry~\notexecuted{} when not executed (because of low validation accuracy).%
}
\vspace{.2em}
\label{tab:experiments-inductionheads}
\fbox{%
\begin{minipage}{\dimexpr\linewidth-2\fboxsep-2\fboxrule\relax}
\centering
\footnotesize
\vspace{.3em}
\setlength{\tabcolsep}{0pt}%
\newcolumntype{G}{S[table-format=2.1]@{\benchInnerSpace}S[table-format=2.1]@{\benchInnerSpace}S[table-format=2.1(2)]}%
\newcolumntype{H}{S[table-format=2.1]@{\benchInnerSpace}S[table-format=2.1(2)]@{\benchInnerSpace}S[table-format=2.1(2)]}%
\renewcommand{\arraystretch}{1.15}
\begin{tabular}{@{}l@{}}%
\begin{tabular}{@{}l@{\benchFirstSpace}l@{\benchSecondSpace}G@{\benchSpace}G@{}}
  & & \multicolumn{6}{c}{\hspace{0em}\textbf{Accuracy (\%)}} \\[.1em]
\cmidrule(lr){3-8}
&
& \multicolumn{3}{c@{\benchSpace}}{\dheads{16}}
& \multicolumn{3}{c}{\dheads{30}} \\
\cmidrule(l{.5em}r{\benchTrim}){3-5}\cmidrule(lr){6-8}
\textbf{Model}
& \textbf{Config}
& {Train} & {~Val~} & {Evaluation}
& {Train} & {~Val~} & {Evaluation}
\\
\midrule
MinMax  & L2,small     & \cmark & \cmark & \cmark   &\cmark   &\cmark  &\cmark     \\
Mamba   & L2,small     &   8.5  &   7.5  & 5.5(1.0) &    5.1  &   4.1  &  3.5(0.4) \\
xLSTM   & L2,small     & \cmark & \cmark & 6.5(1.0) & \cmark  & \cmark &  3.0(0.6) \\
sLSTM   & L2,small     &\cmarko &  87.2$\tout$  & 6.2(0.2)$\tout$ &\cmarko  &  88.3  &  3.0(0.6) \\
xLSTM-L & L2,small     &   7.2  &   8.3  & 6.2(0.7) &    3.5  &   4.7  &  3.1(0.9) \\
DeltaNet& L2,small & \cmark & \cmark & 5.3(0.5) & \cmark & \cmark & 3.6(0.6) \\[1em]
MinMax  & L2,small,bs8 & \cmark & \cmark & \cmark   & \cmark  & \cmark & \cmark    \\
Mamba   & L2,small,bs8 & \cmark & \cmark & \cmark   &    4.7  &   5.2  &  3.5(0.2) \\
xLSTM   & L2,small,bs8 & \cmark & \cmark & 6.1(1.3) & \cmark  & \cmark &  3.0(0.2) \\
sLSTM   & L2,small,bs8 &  99.9$\tout$  &  65.4$\tout$  & 5.9(0.8) &   99.9$\tout$  &
80.8$\tout$  &  3.7(1.4) \\
xLSTM-L & L2,small,bs8 &   7.1  &   6.6  & 6.7(1.1) &    4.1  &   4.6  &  3.0(0.2) \\[1em]
MinMax  & L2,medium    & \cmark & \cmark & \cmark   & \cmark  & \cmark & \cmark    \\
Mamba   & L2,medium    &   8.0  &   8.0  & 6.2(1.0) &    3.8  &   4.0  &  \notexecuted    \\
xLSTM   & L2,medium    & \cmark & \cmark & 6.0(1.0) & \cmark  & \cmark &  3.4(0.2) \\
sLSTM   & L2,medium    &\cmarko &  99.2$\tout$  & 5.7(1.0) &\cmarko  &  83.3$\tout$  &  3.7(0.5) \\[1em]
MinMax  & L3,small     & \cmark & \cmark & \cmark   & \cmark  & \cmark & \cmark    \\
Mamba   & L3,small     &   7.3  &   7.0  & 6.0(0.7) &    4.5  &   4.3  &  \notexecuted    \\
sLSTM   & L3,small     &  99.0$\tout$  &  61.3$\tout$  & 6.5(0.9) &\cmarko  &
79.1$\tout$  &  3.7(0.5) \\
xLSTM   & L3,small     & \cmark & \cmark & 6.5(1.0) & \cmark  & \cmark &  3.4(0.6) \\
xLSTM-L & L3,small     &   6.6  &   8.1  & 5.8(0.2) &    4.0  &   4.9  &  \notexecuted    \\%
\bottomrule
\end{tabular}
\end{tabular}
\vspace{1.0em}
\end{minipage}%
}
\end{table}

\FloatBarrier
\subsection{Results: Throughput}
\label{sec:experiments-throughput}

We complement the accuracy evaluation with an empirical assessment of computational
efficiency.
We measure \emph{training throughput} and \emph{evaluation throughput} in processed
tokens per second, using the same model configurations and single A100 GPU employed
in the experiments above.
For each model, we report throughput averaged across the considered benchmark
instances; the uncertainty is the standard error (SE), expressed as a percentage of
the mean.

For MinMax, in addition to the implementation used for the experiments above, we
consider an optimised implementation employing \code{torch.compile} and PyTorch
automatic mixed precision (AMP).
We rerun the MinMax experiments with these optimisations and obtain the same
accuracy on every benchmark.
Thus, the resulting throughput improvement does not come at the cost of predictive
performance.

Table~\ref{tab:experiments-throughput} reports the results.
The ratios are relative to MinMax with compilation and AMP.

\begin{table}[htb]
\centering
\caption{%
Training and evaluation throughput (tokens/s).
Each value is the mean over all runs of the model, benchmarks, configurations, and seeds; it is
followed by the standard error of the mean (SE), as a percentage of the mean.
Since the pooled runs differ in benchmark and configuration, the SE mostly reflects
this variation rather than measurement noise.
Ratios are relative to MinMax with compilation and AMP.%
}
\vspace{.4em}
\label{tab:experiments-throughput}
\newcommand{\pctpm}[1]{{\color{black!55}\hspace{0.1em}\scalebox{0.8}{$\pm$}\hspace{0.1em}#1\%}}%
\begin{tabular}{lcrrcrr}
\toprule
&
&
\multicolumn{2}{c}{\textbf{Training}}
&
&
\multicolumn{2}{c}{\textbf{Evaluation}}
\\
\cmidrule(lr){3-4}
\cmidrule(lr){5-6}
\textbf{Model}
&
& \textbf{tok/s}
& \textbf{Ratio}
&
& \textbf{tok/s}
& \textbf{Ratio}
\\
\midrule
MinMax (compiled+AMP)
&
& $1{,}136{,}600$\pctpm{7} & $1.00\times$
&
& $10{,}853$\pctpm{9}      & $1.00\times$
\\
MinMax
&
& $206{,}127$\pctpm{4}     & $0.18\times$
&
& $1{,}528$\pctpm{5}       & $0.14\times$
\\
Mamba
&
& $724{,}930$\pctpm{5}     & $0.64\times$
&
& $595$\pctpm{1}           & $0.05\times$
\\
xLSTM-L
&
& $1{,}606{,}865$\pctpm{3} & $1.41\times$
&
& $9{,}210$\pctpm{5}       & $0.85\times$
\\
mLSTM
&
& $206{,}524$\pctpm{2}     & $0.18\times$
&
& $249$\pctpm{3}           & $0.02\times$
\\
sLSTM
&
& $9{,}323$\pctpm{7}       & $0.01\times$
&
& $207$\pctpm{3}           & $0.02\times$
\\
DeltaNet
&
& $591{,}412$\pctpm{4}     & $0.52\times$
&
& $10{,}372$\pctpm{6}      & $0.96\times$
\\
\bottomrule
\end{tabular}
\end{table}

Compilation and AMP increase MinMax throughput from $206$k to $1.14$M tokens/s
during training, a $5.5\times$ speedup, and from $1.53$k to $10.9$k tokens/s during
evaluation, a $7.1\times$ speedup.
With these optimisations, MinMax has the second-highest training throughput among
the considered models: xLSTM-L is $1.41\times$ faster, while MinMax is
$1.57\times$ faster than Mamba and $1.92\times$ faster than DeltaNet.
During evaluation, MinMax has the highest measured throughput, exceeding DeltaNet
by $1.05\times$, xLSTM-L by $1.18\times$, and Mamba by $18.2\times$.

These measurements indicate that the favourable parallel complexity of MinMax RNCs
is reflected in practical execution time.
In particular, MinMax is competitive with the fastest considered parallel recurrent
architectures in both training and evaluation, while the compiled+AMP variant  
retains the accuracy reported in the preceding experiments.

\clearpage{}%

\clearpage{}%

\section{Details of the Experimental Evaluation on Long-Range Dependencies}
\label{app:experiments-longrange}

This appendix provides the details of the experiment assessing the propagation of state gradients
across long-range dependencies (Section~\ref{subsec:experiments-synth--star-free}), where we train on
a variant of \dsequences{2} with dependencies spanning tens of thousands of steps.

The experiments use the same setup as Appendix~\ref{app:experiments} except for the differences we
describe.

\paragraph{Reproducing the experiments.}
The datasets, configurations, and training and evaluation protocols of the experiment on
MinMax RNCs in this appendix are available in our benchmarks
repository~\citep{minmaxbenchmarksgithub}.
DeltaNet and Mamba are not included, and must be plugged into the repository as described in
Appendix~\ref{app:experiments}.
From the repository, installed as in Appendix~\ref{app:experiments}, the following commands
generate the datasets and train and evaluate the model, and then compute the majority-class
baseline of \S\ref{subsec:longrange-baseline} and the statistics of
Table~\ref{tab:longrange_stats}:
\begin{commands}
python main.py reproduce longrange
python scripts/majority_baseline.py
python scripts/sequences_gap_stats.py
\end{commands}

\localtableofcontents

\subsection{Benchmark}

We consider a long-range variant~\dsequencesLongRange{n} of the benchmark~\dsequences{n}---the
one described in Appendix~\ref{app:experiments}.
The benchmark~\dsequencesLongRange{n} has the same structure as \dsequences{n}:
$N = 2$ sequences to match, each of length $L = n$, with $M = 2$ distinct tokens
for each event type, and $I = 10$ irrelevant tokens; hence $|\mathcal{V}| = 4n + 19$.
The difference with \dsequences{n} lies in the parameters used to generate the datasets, which do
not depend on~$n$: we make the relevant tokens more rare, and the sequences longer, so that the
events to match are farther apart.
In the training and validation datasets, all sequences have length $2^{17} = 131{,}072$ and
$p_\mathrm{set} = 2^{-16}$, so that each sequence contains two relevant tokens in expectation.
Table~\ref{tab:longrange_datasets} reports the details of the datasets.
In our experiments, we consider the benchmark~\dsequencesLongRange{2}, for which
$|\mathcal{V}| = 27$.

\begin{table}[htb]
\small
  \centering
  \caption{Details of~\dsequencesLongRange{n}.
  The parameters $N = 2$, $L = n$, $M = 2$, $I = 10$ are as in \dsequences{n},
  hence $|\mathcal{V}| = 4n + 19$.
  In training and validation, $p_\mathrm{set} = 2^{-16}$.
  All sequences of a dataset have the same length~$T$.}
  \vspace{.3em}
  \label{tab:longrange_datasets}
  \sisetup{round-mode=none, group-separator={,}, group-minimum-digits=4}
  \begin{tabular}{
    l@{\hspace{2em}}
    r@{}l@{\hspace{2em}}
    r@{}l@{\hspace{2em}}
    S[table-format=7.0]@{\hspace{2em}}
    S[table-format=5.0]}
    \toprule
    & \multicolumn{2}{c@{\hspace{2em}}}{$p_\mathrm{set}$}
    & \multicolumn{2}{c@{\hspace{2em}}}{$p_\mathrm{reset}$}
    & {$T$} & {Size} \\
    \midrule
    \textbf{Training}   & $1$ & $.53 \cdot 10^{-5}$ & $0$ &                     & 131072  & 20000 \\
    \textbf{Validation} & $1$ & $.53 \cdot 10^{-5}$ & $0$ &                     & 131072  &  1000 \\
    \textbf{Evaluation} & $1$ & $.00 \cdot 10^{-5}$ & $1$ & $.00 \cdot 10^{-6}$ & 1024000 &  1000 \\
    \bottomrule
  \end{tabular}
\end{table}

\paragraph{Event gap and retention span.}
Table~\ref{tab:longrange_stats} reports two statistics that quantify how far apart the
relevant steps are.
For a match of one of the sequences, completed at step~$t$, we call \emph{event gap} the
number of steps $t - t_0$ since the latest first event~$t_0$ of that sequence before~$t$, and
\emph{retention span} the number of steps $T - 1 - t$ from the match to the end of the input
sequence.
The two play different roles in training.
The outputs from step~$t$ on depend on the input at step~$t_0$; hence the error signal that
teaches a model to store the first event arises at step~$t$ or later, and reaches step~$t_0$
only through the product of at least $t - t_0$ state-to-state derivatives.
Before a model has learnt to store the first event, nothing else rewards storing it, so the
event gap is the distance across which state gradients must not vanish.
The retention span, instead, measures how long a model has to keep a match in memory, since a
match is never reset ($p_\mathrm{reset} = 0$ in training and validation): the error signal
that teaches a model to keep the match arises at every step after the match, including the
next few, so it does not need to propagate across the retention span, and keeping the match
across it is a property of the forward dynamics.
In the training datasets of~\dsequences{2}, the event gap is 4~steps on average, and the
retention span 367~steps; in~\dsequencesLongRange{2}, the event gap is 37k~steps on average,
and only 1.9\% of the matches have an event gap below 1{,}000~steps.
The evaluation datasets of the two benchmarks have the same parameters, with event gaps of
165k~steps on average: in~\dsequences{2}, long event gaps occur only at evaluation, where they
test whether the forward dynamics retain the first event; in~\dsequencesLongRange{2}, they
occur already in training, where they test the propagation of state gradients.

\begin{table}[htb]
\small
  \centering
  \caption{Event gap and retention span in the datasets of \dsequences{2} and
  \dsequencesLongRange{2}: mean, median, and 10th--90th percentile range over the matches;
  Matched is the fraction of input sequences with at least one match.
  The statistics are those of the distribution of the data generator, estimated on
  20{,}000 sampled sequences (5{,}000 for evaluation); the evaluation datasets of the two
  benchmarks have the same parameters.}
  \vspace{.3em}
  \label{tab:longrange_stats}
  \footnotesize
  \setlength{\tabcolsep}{3.5pt}
  \begin{tabular}{ll@{\hspace{1em}}r@{\hspace{1em}}rrr@{\hspace{1em}}rrr}
    \toprule
    & & & \multicolumn{3}{c}{\textbf{Event gap}} & \multicolumn{3}{c}{\textbf{Retention span}} \\
    \cmidrule(lr){4-6}\cmidrule(lr){7-9}
    \textbf{Benchmark} & \textbf{Datasets} & \textbf{Matched}
    & Mean & Median & 10--90\% & Mean & Median & 10--90\% \\
    \midrule
    \dsequences{2}          & training
      & 100\%  & 4.0   & 3     & 1--9        & 367   & 367   & 265--471    \\
    \dsequencesLongRange{2} & training, validation
      & 16.8\% & 37.1k & 30.6k & 5.6k--80.3k & 47.7k & 43.6k & 8.3k--93.6k \\
    both                    & evaluation
      & 92.1\% & 165k  & 121k  & 19k--372k   & 502k  & 511k  & 134k--846k  \\
    \bottomrule
  \end{tabular}
\end{table}

\FloatBarrier
\subsection{Models}

We compare MinMax RNCs with DeltaNet and Mamba, all with two layers, as for \dsequences{2} in
Appendix~\ref{app:experiments}.
\begin{itemize}
  \item
    \emph{MinMax.} The MinMax RNC has the `small' configuration (\tsc{s}) of Appendix~\ref{app:experiments},
    with three differences: it has no output gate (\code{output_gate=False}); the set and reset
    projections have the asymmetric initialisation (\code{s_r_init=`asymmetric'}); and it runs
    under \code{bf16} automatic mixed precision, with the MinMax recurrence computed in \code{fp16}
    (\code{amp_dtype=`bf16'}, \code{recurrence_dtype=`fp16'}).
  \item
    \emph{DeltaNet.} DeltaNet has the `small' configuration of Appendix~\ref{app:experiments},
    with version \code{0.5.2} of the \code{fla} implementation.
  \item
    \emph{Mamba.} Mamba has the `small' configuration of Appendix~\ref{app:experiments}.
\end{itemize}

\paragraph{Model sizes.}
The number of trainable parameters are 122{,}842  for MinMax, 129{,}840 for DeltaNet, and 136{,}515
for Mamba.

\FloatBarrier
\subsection{Trainers}

All models are trained with the same recipe, the one of the MinMax trainer of
Appendix~\ref{app:experiments} (\code{trainer_default}): Adam with weight decay
$\lambda = 10^{-4}$, learning rate $10^{-3}$ halved every 5~epochs, batch size~64, and no gradient
clipping.
To fit sequences of length 131{,}072 in the memory of a single NVIDIA A100 GPU (40GB), we use
two trainers that implement this recipe with different memory settings, which leave the
computed gradients unchanged up to floating-point rounding:
\begin{itemize}
  \item
    \code{trainer_default_longseq}, for MinMax and DeltaNet, accumulates gradients over
    2~micro-batches of 32~sequences, and computes the forward pass in chunks of 65{,}536 steps
    with gradient checkpointing;
  \item
    \code{trainer_mamba_longseq}, for Mamba, processes each batch as a whole, in a single
    forward pass over the full sequence length.
\end{itemize}

\FloatBarrier
\subsection{Training Protocol}

The training protocol is the one of Appendix~\ref{app:experiments}, with the following
differences.
\begin{itemize}
  \item
    \emph{Datasets.} Training and validation sequences have length 131{,}072
    (Table~\ref{tab:longrange_datasets}); in particular, validation is on sequences of the same
    length as training.
  \item
    \emph{Early stopping.} Early stopping is as in Appendix~\ref{app:experiments} but with
    patience~3.
  \item
    \emph{Time limit.} The time limit is 12~hours per model-trainer pair, checked before starting
    a new seed. It is reached by MinMax after 3~seeds (13.6~hours in total); DeltaNet completes
    3~seeds in 5.4~hours.
  \item
    \emph{Compilation.} MinMax is compiled with \code{torch.compile}; DeltaNet and Mamba are
    not.
\end{itemize}

\subsection{Majority-Class Baseline}
\label{subsec:longrange-baseline}

We compare the models with the \emph{majority-class baseline}: the predictor that ignores the
input and outputs, at every step, the most frequent label of the dataset.
In the datasets of~\dsequencesLongRange{2}, the most frequent label is the empty set, i.e., that
no sequence is matched: matches are rare, and the steps from a match to the end of its input
sequence are only from 5.40\% to 6.58\% of the steps of the validation datasets of the three seeds.
Hence the accuracy of the baseline is 93.70\% on the training dataset of seed~1, and from
93.42\% to 94.60\% on the validation datasets; it is the accuracy of a model that has not learnt
to recognise the matches.
Since the accuracy of the baseline depends on the dataset, we also report the \emph{gain} of a
model over the baseline of the same dataset,
\begin{equation*}
  \frac{\mathit{acc} - \mathit{acc}_\mathrm{base}}{100\% - \mathit{acc}_\mathrm{base}} \,,
\end{equation*}
where $100\%$ is perfect accuracy and $0\%$ is no better than the baseline.

\FloatBarrier
\subsection{Results}

Table~\ref{tab:longrange_results} reports, for each model, the validation accuracy of the best
model of each seed, aggregated across the three seeds, together with its gain over the
majority-class baseline of the same seed.
The gain removes the variation of the baseline across the validation datasets of the seeds, which
dominates the confidence intervals of the accuracies.

\emph{MinMax} learns the task for all three seeds, with a gain of 99.96\% over the baseline and a
validation accuracy of at least 99.993\%; its validation accuracy exceeds 99.9\% after 3, 6 and
11~epochs, respectively.

\emph{DeltaNet} and \emph{Mamba} do not learn the task: for every seed, their validation accuracy
stays within 0.1~points of the baseline, and their gain is not distinguishable from zero; their
training accuracy is constant across epochs, and equal to the accuracy of the baseline on the
training dataset (93.70\% for seed~1).
Since the task is solved by MinMax with the same training recipe and a comparable number of
parameters, this indicates that the error signal of the matches does not propagate across the
event gaps in DeltaNet and Mamba, whereas it does in MinMax, in line with the non-vanishing
gradient results of Theorem~\ref{thm:gradient-nonvanishing}.

\begin{table}[tbh]
  \centering
  \caption{Results on~\dsequencesLongRange{2}: \emph{mean~$\pm$~error margin} of 99\% CIs
  across three seeds of the validation accuracy of the best model of each seed, and of its gain
  over the majority-class baseline (Section~\ref{subsec:longrange-baseline}).}
  \vspace{.3em}
  \label{tab:longrange_results}
  \small
  \sisetup{round-mode=uncertainty, round-precision=2}
  \begin{tabular}{l@{\hspace{2em}}S[table-format=2.3(2)]@{\hspace{2em}}S[table-format=2.2(2)]}
    \toprule
    \textbf{Model} & {\textbf{Validation accuracy (\%)}} & {\textbf{Gain (\%)}} \\
    \midrule
    Baseline & 94.0717(34435) &     \notexecuted   \\
    \midrule
    MinMax   & 99.9977(218)   & 99.960(376)    \\
    DeltaNet & 94.0840(30636) & 0.136(6652)    \\
    Mamba$^{\circ}$ & 93.420  & 0.0            \\
    \bottomrule
  \end{tabular}
\end{table}

\clearpage{}%

\clearpage{}%

\section{Details of the Experimental Evaluation on Group-based Benchmarks}
\label{app:experiments-groups}

This appendix provides the details of our empirical evaluation on the \emph{word problems} of the
symmetric groups $S_3$ and $S_4$, which require MinMax neurons of state degree greater than one.
We compare MinMax RNCs with DeltaProduct~\citep{siems2025deltaproduct}, a linear RNN designed
for state tracking, in the configuration of its original evaluation.

The experiments use the same the setup as Appendix~\ref{app:experiments} except for the differences
we describe.

\paragraph{Reproducing the experiments.}
The datasets, configurations, and training and evaluation protocols of the experiments on
MinMax RNCs in this appendix are available in our benchmarks
repository~\citep{minmaxbenchmarksgithub}.
DeltaProduct is not included, and must be plugged into the repository as described in
Appendix~\ref{app:experiments}.
From the repository, installed as in Appendix~\ref{app:experiments}, the following command
generates the datasets of \dword{S_3} and \dword{S_4}, and trains and evaluates the MinMax
configurations:
\begin{commands}
python main.py reproduce groups
\end{commands}

\localtableofcontents

\subsection{Benchmarks}

For a finite permutation group~$G$, the benchmark task~\dword{G} is the \emph{word problem of~$G$},
as in the evaluation of DeltaProduct in~\citep{siems2025deltaproduct}.

We focus on \dword{S_3} and \dword{S_4}.

The vocabulary $\mathcal{V} = G$ consists of the group elements, hence $|\mathcal{V}| = |G|$,
with $|S_3| = 6$ and $|S_4| = 24$; they are encoded into integers as in
Appendix~\ref{app:experiments}, with two additional special tokens (padding and undefined), so
models have $|G| + 2$ input and output tokens.

Let us write $\star$ for the operator of $G$.

On an input $(\sigma_t)_{t=1}^T \in \allnonemptyseqs{G}$, the expected output sequence is
$(\gamma_t)_{t=1}^T$ where
\begin{align*}
  \gamma_t \,=\, \sigma_t \star \sigma_{t-1} \star \cdots \star \sigma_1
  \qquad
  \forall t \in \iival{1}{T},
\end{align*}
i.e., $\gamma_t$ is the group element obtained by multiplying the first $t$ input elements in
order.
The output is required at every step, so no step is masked.

To define datasets, we use the parameters $T_\mathrm{min},T_\mathrm{max} \in \naturals$ as in
Appendix~\ref{app:experiments}; an input sequence $\rv{U}_1 \cdots \rv{U}_\rv{T}$ is generated as
\begin{align*}
  \rv{T} \sim \operatorname{Unif}(\iival{T_\mathrm{min}}{T_\mathrm{max}}),
  \qquad
  \rv{U}_t \sim \operatorname{Unif}(G)
  \quad
  \forall t \in \iival{1}{\rv{T}}.
\end{align*}
Table~\ref{tab:groups_datasets} reports the details of the datasets; all of them have fixed
length~$T$ (i.e., $T_\mathrm{min} = T_\mathrm{max} = T$).

\begin{table}[htb]
\small
  \centering
  \caption{Details of the datasets of~\dword{G}.}
  \vspace{.3em}
  \label{tab:groups_datasets}
  \begin{tabular}{l@{\hspace{2em}}c@{\hspace{2em}}r@{\hspace{2.5em}}r@{}p{1.5em}@{}r@{\hspace{2.5em}}r@{}p{1.5em}@{}r@{\hspace{2.5em}}r}
    \toprule
    & & \multicolumn{2}{c}{\textbf{Training}}
    & & \multicolumn{2}{c}{\textbf{Validation}}
    & & \multicolumn{2}{c}{\textbf{Evaluation}} \\
    \cmidrule(lr){3-4} \cmidrule(lr){6-7} \cmidrule(lr){9-10}
    \emph{Benchmark} & $|G|$
      & $T$ & Size & & $T$ & Size & & $T$ & Size \\
    \midrule
    \dword{S_3} &  6 & 128 &  20k & & 512 & 1k & & 1{,}024{,}000 & 1k \\
    \dword{S_4} & 24 & 128 & 200k & & 512 & 1k & & 1{,}024{,}000 & 1k \\
    \bottomrule
  \end{tabular}
\end{table}

\FloatBarrier
\subsection{Models}

We use MinMax RNCs whose layers contain a single array of $K = 16$ MinMax neurons of state
degree $d \in \{3, 4, 5\}$, and no scalar-state neurons; i.e., in the implementation, where the state degree is called \emph{state dimension}
(Appendix~\ref{sec:minmax-implementation}), \code{neurons=[NeuronArray(d_state=}$d$\code{,
num_neurons=16)]} and \code{scalar_neurons=0}.
A neuron of state degree $d$ suffices to capture the permutations of $d$ elements, hence
$d = 3$ suffices for $S_3$ and $d = 4$ for $S_4$ (Theorem~\ref{thm:main-body-expressivity});
we also include $d = 5$, and models with two layers.
All models have $d_\mathrm{model} = 90$, and use the \emph{asymmetric} initialisation of the
set and reset projections (\code{s_r_init=`asymmetric'}): the set projection is initialised
with Kaiming-uniform weights and bias~$1$, and the reset projection with small-init weights and
bias~$0$, so that the neurons write to their state from the first step.
All other parameters are left to their default values; in particular, \code{output_gate=True}
and \code{conv_type=`basic'}.

We write \mmconf{L}{d} for the configuration with $L$ layers of neurons of state degree~$d$; for
instance, \mmconf{1}{3} has one layer of 16~neurons of state degree~3.
Table~\ref{tab:groups_models} reports the model configurations, with the dimensions defined as in
Table~\ref{tab:models_summary}; in particular, $d_\mathrm{state} = K \cdot d$ is the total number
of scalar values in the state of a layer.

\begin{table}[htb]
  \small
  \centering
  \caption{Model configurations used in the experiments on group-based benchmarks.
  \emph{Legend:} \mmconf{L}{d} has $L$ layers, each with $K = 16$ neurons of state degree~$d$; $V$ is the vocabulary
  size, i.e., $|G| + 2$ including the two special tokens, and $L$ is the number of layers. The last two columns report the number of trainable
  parameters for the one-layer models.}
  \vspace{.4em}
  \label{tab:groups_models}
  \setlength{\tabcolsep}{4pt}
  \begin{tabular}{c@{\hspace{1em}}rrr@{\hspace{1em}}l@{\hspace{1em}}rr}
    \toprule
    & \multicolumn{3}{c}{\textbf{Dimensions}} & & \multicolumn{2}{c}{\textbf{$L = 1$}} \\
    \cmidrule(lr){2-4} \cmidrule(lr){6-7}
    \textbf{Configuration} &
    $d_\mathrm{model}$ & $d_\mathrm{state}$ & $d_\mathrm{conv}$ &
    \textbf{Trainable Parameters} & \dword{S_3} & \dword{S_4} \\
    \midrule
    \mmconf{L}{3} & 90 & 48 & 90 & \phantom{1}75k${} \cdot L$ $\,+\,$ 90${} \cdot V$ $\,+\,$ 180
      &  76k &  78k \\
    \mmconf{L}{4} & 90 & 64 & 90 & \phantom{1}90k${} \cdot L$ $\,+\,$ 90${} \cdot V$ $\,+\,$ 180
      &  91k &  92k \\
    \mmconf{L}{5} & 90 & 80 & 90 & 107k${} \cdot L$ $\,+\,$ 90${} \cdot V$ $\,+\,$ 180
      & 108k & 110k \\
    \bottomrule
  \end{tabular}
\end{table}

\paragraph{DeltaProduct.}
DeltaProduct~\citep{siems2025deltaproduct} updates its state at each step with a product of
$n_h$ generalised Householder transformations, whose eigenvalues we allow in $[-1, 1]$, as
required by permutations.
We use the configuration of the evaluation of~\citet{siems2025deltaproduct} on these benchmarks:
one layer with $d_\mathrm{model} = 128$, 12~heads of dimension~32, $n_h = 2$ on~\dword{S_3} and
$n_h = 3$ on~\dword{S_4}, no forget gate, no short convolution, no output gate, and a feed-forward
block of expansion~4.
We write \dpconf{n_h} for the configuration with $n_h$ Householder steps; \dpconf{2} and \dpconf{3}
have 497k and 601k trainable parameters, respectively.
We use the kernels of the flash-linear-attention library~\citep{flashlinearattention}
version~\code{0.6.0}: the parameters are in \code{fp32}, and the kernels run in \code{bf16}.

\FloatBarrier
\subsection{Trainers}

We use two trainers, defined in Table~\ref{tab:groups_trainers}, which differ only in the batch
size; DeltaProduct is trained with \code{trainer_default}, the recipe of its original evaluation: Adam without weight decay, with $(\beta_1, \beta_2) = (0.9, 0.999)$, peak learning rate
$10^{-3}$, cosine annealing of the learning rate to $0$ over the training epochs, and no gradient
clipping.

\begin{table}[h!]
  \centering
  \caption{Trainers used in the experiments on group-based benchmarks.}
  \vspace{.1em}
  \footnotesize
  \setlength{\tabcolsep}{4pt}
  \begin{tabular}{lc@{\hspace{1em}}c@{\hspace{1.3em}}c@{\hspace{.8em}}lc@{\hspace{.5em}}lcc@{\hspace{.7em}}c}
\toprule
& \multicolumn{4}{c}{\textbf{Optimiser}} & & \multicolumn{2}{c}{\hspace{-1.0em}\textbf{Learning Rate}} \\
\cmidrule(lr){2-5} \cmidrule(r){7-8}
\textbf{Trainer} & Type & $\lambda$ & $\beta_1$ & ~~$\beta_2$ &&~~Peak &
\multicolumn{1}{c}{~~~~~Schedule}
                 && \textbf{Batch size} \\
\midrule
\code{trainer_default} &
Adam  & no decay & 0.9 & 0.999 &&~~$10^{\smallminus 3}$ & Cosine && 64
\\
\code{trainer_bs8} &
Adam  & no decay & 0.9 & 0.999 &&~~$10^{\smallminus 3}$ & Cosine && 8
\\
\bottomrule
\end{tabular}
\label{tab:groups_trainers}
\end{table}

\FloatBarrier
\subsection{Training Protocol}

The training protocol is the one of Appendix~\ref{app:experiments}, with the following
differences; the first four concern the MinMax RNCs.
\begin{itemize}
  \item
    \emph{Epochs.} MinMax RNCs are trained for 30~epochs on \dword{S_3} and for 100~epochs on \dword{S_4}.
  \item
    \emph{Early stopping.} On \dword{S_3}, early stopping is as in Appendix~\ref{app:experiments} but
    with patience~3; on \dword{S_4}, early stopping is disabled.
  \item
    \emph{Seeds.} In addition to the stopping criterion of Appendix~\ref{app:experiments}, no
    further seeds are tried once the upper bound of the 99\% confidence interval of the validation
    accuracy is below 50\%.
  \item
    \emph{Time limit.} The time limit of 4~hours per model-trainer pair is checked before starting
    a new seed. On \dword{S_4} it is reached by both configurations: \mmconf{1}{4} completes
    3~seeds, and \mmconf{1}{5} completes only 2~seeds, an exception to the minimum of
    3~seeds of Appendix~\ref{app:experiments}.
  \item
    \emph{DeltaProduct.} DeltaProduct is trained for 3~seeds on the same datasets, at most
    30~epochs on~\dword{S_3}, with early stopping on the validation loss (patience~3), and at
    most 100~epochs on~\dword{S_4}, stopping once 99\% of the validation sequences are predicted
    entirely correctly, as in the evaluation of~\citet{siems2025deltaproduct}.
\end{itemize}

\FloatBarrier
\subsection{Evaluation Protocol}

The evaluation protocol is the one of Appendix~\ref{app:experiments}, on datasets of 1k sequences
of length 1{,}024{,}000 (Table~\ref{tab:groups_datasets}). Since no step is masked, we report both
the step-average accuracy and the step-minimum accuracy.
DeltaProduct is evaluated on the same datasets, for each of its 3~seeds, and also with the metric
of~\citet{siems2025deltaproduct}: the fraction of 10k test sequences of length~512 whose outputs
are all correct.

\FloatBarrier
\subsection{Overview of the Included Experiments}

Table~\ref{tab:groups_overview} summarises the experiments: for each benchmark, we carry out training
and evaluation for the listed model-trainer pairs.

\begin{table}[tbh]
  \centering
  \caption{Overview of the experiments on group-based benchmarks.}
  \vspace{.2em}
  \label{tab:groups_overview}
  \small
  \begin{tabular}{l@{\hspace{1em}}ll@{\hspace{1em}}l}
    \toprule
    & \multicolumn{2}{c}{\textbf{Configurations}} & \\
    \cmidrule(r){2-3}
    \textbf{Benchmark} & Name & Parameter values & \textbf{Trainers} \\
    \midrule
    \dword{S_3} & \mmconf{L}{d} & $d \in \{3, 4, 5\}$, $\mkern1mu L \in \{1, 2\}$
                & \code{trainer_default}, $\mkern2mu$\code{trainer_bs8} \\
                & \dpconf{n_h} & $n_h = 2$ & \code{trainer_default} \\
    \addlinespace[2pt]
    \dword{S_4} & \mmconf{1}{d} & $d \in \{4, 5\}$ & \code{trainer_default} \\
                & \dpconf{n_h} & $n_h = 3$ & \code{trainer_default} \\
    \bottomrule
  \end{tabular}
\end{table}

\FloatBarrier

\subsection{Results}

Table~\ref{tab:groups_results} reports, for each model-trainer pair, the train, validation, and
step-minimum and step-average evaluation accuracies of the best model, and the mean validation
accuracy across seeds with its 99\% confidence interval.

\par\smallskip
\textbf{MinMax}~learns to solve both tasks with seamless generalisation on longer sequences.
\begin{itemize}
  \item
    On~\dword{S_3}, every configuration reaches perfect validation accuracy for some seed and batch
    size, and the best models of all the evaluated model-trainer pairs retain perfect accuracy on
    sequences of length 1{,}024{,}000, with one exception: the best model of \mmconf{2}{3} with batch
    size~8 drops to 33\% accuracy (chance level being $1/6$) despite perfect validation accuracy.
    Results depend on the seed, and for one layer on the batch size: with batch size~64,
    \mmconf{1}{3} and \mmconf{1}{4} do not learn the task for any seed, whereas with batch size~8
    they do for two seeds out of three.
    \mmconf{2}{4} with batch size~8 learns the task for all three seeds.

  \item
    On~\dword{S_4}, \mmconf{1}{4} reaches perfect validation accuracy for two seeds out of three,
    and its best model retains perfect accuracy on sequences of length 1{,}024{,}000, whereas
    \mmconf{1}{5} does not learn the task.
\end{itemize}

\par\smallskip
\textbf{DeltaProduct}~reaches perfect validation accuracy on both tasks, but its evaluation accuracy
deteriorates as sequence length increases---%
as shown in Figure~\ref{fig:groups_eval} for~\dword{S_3}.
\begin{itemize}
  \item
    On~\dword{S_3}, \dpconf{2} reaches a validation accuracy of~99.97\% and~99.96\% for two seeds,
    and~96.2\% for the third seed, whose training diverges after the second epoch; on test sequences
    of length~512, the two best seeds predict entirely correctly 97\% and 95\% of the sequences, in
    line with the results of~\citet{siems2025deltaproduct}.
  \item
    On~\dword{S_4}, \dpconf{3} reaches a validation accuracy of~99.99\%,~99.91\%, and~99.90\%, and
    predicts entirely correctly 98.5\%, 91.1\% and 90.1\% of the test sequences of length~512.
    On sequences of length 1{,}024{,}000, the accuracy within chunks of 1024~steps falls from
    86--99\% (on~\dword{S_3}) and 93--98\% (on~\dword{S_4}) in the first chunk to 51--86\% and
    42--66\% in the second, and to chance level ($1/6$ and $1/24$) from step~$10^5$ and
    step~$10^4$ on, respectively; the step-average accuracy ranges across seeds from 16.9\% to
    17.6\% on~\dword{S_3}, and from 4.3\% to 4.4\% on~\dword{S_4}, against 100\% for the best
    models of MinMax on both benchmarks.
\end{itemize}

\begin{table}[tbh]
  \centering
  \caption{Results on group-based benchmarks. \emph{Legend:} for the best model of each
  model-trainer pair, `Train' and `Val' are its train and validation accuracies, and `Eval min'
  and `Eval avg' its step-minimum and step-average evaluation accuracies on sequences of length
  1{,}024{,}000 (their 99\%~confidence intervals have width below $10^{-3}$); `Val mean' is the
  mean validation accuracy across seeds, followed by its 99\%~confidence interval;
  `\notexecuted{}' denotes that the evaluation was not carried out. All values are in~\%.}
  \vspace{.1em}
  \label{tab:groups_results}
  \footnotesize
  \setlength{\tabcolsep}{3.5pt}
  \sisetup{round-mode=none}
  \newcommand{\ci}[2]{{\color{black!55}\scalebox{0.8}{$[#1, #2]$}}}%
  \begin{tabular}{l@{\hspace{1em}}lS[table-format=2.0]@{\hspace{1.2em}}
    S[table-format=3.1]S[table-format=3.2]S[table-format=3.1]S[table-format=3.1]@{\hspace{1.5em}}
    S[table-format=3.1]@{\hspace{.3em}}l}
    \toprule
    & & & \multicolumn{4}{c}{\textbf{Best model}} & \multicolumn{2}{c}{\textbf{Across seeds}} \\
    \cmidrule(lr){4-7} \cmidrule(l){8-9}
    \textbf{Benchmark} & \textbf{Configuration} & {\textbf{Batch}} &
    {Train} & {Val} & {Eval min} & {Eval avg} & \multicolumn{2}{c}{Val mean} \\
    \midrule
    \dword{S_3}
      & \mmconf{1}{3} &  8 & 100.0 & 100.0 & 100.0 & 100.0 &  78.8 & \ci{36.4}{100.0} \\
      &               & 64 &  36.7 &  34.3 & {\notexecuted} & {\notexecuted} &  28.8 & \ci{17.7}{34.3} \\
    \addlinespace[2pt]
      & \mmconf{1}{4} &  8 & 100.0 & 100.0 & 100.0 & 100.0 &  78.6 & \ci{35.9}{100.0} \\
      &               & 64 &  42.7 &  35.9 & {\notexecuted} & {\notexecuted} &  24.1 & \ci{18.1}{35.9} \\
    \addlinespace[2pt]
      & \mmconf{1}{5} &  8 & 100.0 & 100.0 & 100.0 & 100.0 &  78.2 & \ci{34.7}{100.0} \\
      &               & 64 & 100.0 & 100.0 & {\notexecuted} & {\notexecuted} &  45.7 & \ci{18.2}{100.0} \\
    \addlinespace[2pt]
      & \mmconf{2}{3} &  8 & 100.0 & 100.0 &  32.9 &  33.3 &  78.9 & \ci{36.8}{100.0} \\
      &               & 64 & 100.0 & 100.0 & 100.0 & 100.0 &  79.1 & \ci{37.4}{100.0} \\
    \addlinespace[2pt]
      & \mmconf{2}{4} &  8 & 100.0 & 100.0 & 100.0 & 100.0 & 100.0 & \ci{100.0}{100.0} \\
      &               & 64 & 100.0 & 100.0 & 100.0 & 100.0 &  74.3 & \ci{22.8}{100.0} \\
    \addlinespace[2pt]
      & \mmconf{2}{5} &  8 & 100.0 & 100.0 & 100.0 & 100.0 &  80.0 & \ci{40.0}{100.0} \\
      &               & 64 & 100.0 & 100.0 & 100.0 & 100.0 &  80.2 & \ci{40.7}{100.0} \\
    \addlinespace[2pt]
      & \dpconf{2}    & 64 &  99.4 &  99.97 &  16.5 &  17.6 &  98.7 & \ci{86.3}{100.0} \\
    \midrule
     \dword{S_4}
      & \mmconf{1}{4} & 64 & 100.0 & 100.0 & 100.0 & 100.0 &  75.3 & \ci{26.0}{100.0} \\
    \addlinespace[2pt]
      & \mmconf{1}{5} & 64 &  10.7 &   5.8 & {\notexecuted} & {\notexecuted} &   5.1 & \ci{4.4}{5.8} \\
    \addlinespace[2pt]
      & \dpconf{3}    & 64 & 100.0 &  99.99 &   3.6 &   4.4 &  99.9 & \ci{99.7}{100.0} \\
    \bottomrule
  \end{tabular}
\end{table}

\begin{figure}[tbh]
  \centering
  \tikzsetnextfilename{plot__groups_eval__s3__mean_ci}
  \begin{tikzpicture}
    \tikzpicturedependsonfile{data/evaluations_groups/s3/minmax_d4_l2_bs8.csv}
    \tikzpicturedependsonfile{data/evaluations_groups/s3/deltaproduct.csv}
    \begin{axis}[
        width=0.95\textwidth, height=5cm,
        xmin=1024, xmax=1024000,
        scaled x ticks=false,
        xtick={200000,400000,600000,800000,1000000},
        ymin=0, ymax=105, ytick={0,25,50,75,100},
        xlabel={Step}, ylabel={Accuracy (\%)},
        grid=major,
        label style={font=\footnotesize}, tick label style={font=\scriptsize},
        legend cell align=left, legend columns=3,
        legend style={font=\footnotesize, draw=black!50, line width=0.4pt, rounded corners=1pt,
          inner xsep=4pt, inner ysep=2pt, at={(0.5,1)}, anchor=south, yshift=12pt,
          /tikz/every even column/.append style={column sep=0.4cm}},
      ]
      \addplot[name path=dphi, draw=none, forget plot]
        table[x=step, y=acc_high, col sep=comma] {data/evaluations_groups/s3/deltaproduct.csv};
      \addplot[name path=dplo, draw=none, forget plot]
        table[x=step, y=acc_low, col sep=comma] {data/evaluations_groups/s3/deltaproduct.csv};
      \addplot[fill=plotorange!25, draw=none, forget plot] fill between[of=dphi and dplo];

      \addplot[cyan, line width=1.0pt, mark=none]
        table[x=step, y=acc, col sep=comma]
        {data/evaluations_groups/s3/minmax_d4_l2_bs8.csv};
      \addlegendentry{MinMax}
      \addplot[plotorange, line width=1.0pt, mark=none]
        table[x=step, y=acc_mean, col sep=comma]
        {data/evaluations_groups/s3/deltaproduct.csv};
      \addlegendentry{DeltaProduct}
      \addplot[black, dashed, line width=0.6pt, domain=1024:1024000, samples=2] {100/6};
      \addlegendentry{Chance level}
    \end{axis}
  \end{tikzpicture}
  \caption{Evaluation accuracy on~\dword{S_3} over the steps of sequences of length 1{,}024{,}000:
  accuracy within each chunk of 1024~steps of \dpconf{2}, mean across its 3~seeds with its 99\%
  confidence interval (shaded), and of the best model of \mmconf{2}{4} with batch size~8 (mean over
  its evaluation runs).}
  \label{fig:groups_eval}
\end{figure}

\clearpage{}%

\clearpage{}%

\section{Details of Next-Token Prediction on OpenWebText with MinMax RNCs}
\label{app:details-of-next-token-prediction-on-openwebtext}

This appendix provides further details on the next-token prediction experiment.

\textbf{Overview.} We train a MinMax RNC as a causal language model on the OpenWebText corpus using the GPT-2
tokeniser.
The model has \textasciitilde112M trainable parameters.
The run completed 100k gradient steps in 14~days and 7~hours
on a single A100~GPU.

\subsection{Setup}

Details of the setup are provided in Table~\ref{tab:gpt_setup}.
The loss function is cross-entropy over the next token. %
The computed validation loss is a Monte Carlo estimate: it is obtained by averaging over 2,000 batches
drawn randomly with replacement from the held-out validation set (each batch has 4 sequences of
1,024 tokens).
Letting $\mu$ and $\sigma$ be the mean and standard deviation of the validation loss,
and letting $\hat\mu$ be the mean of $n = 2000$ i.i.d.\ batch losses,
by the Central Limit Theorem we have
  $\hat\mu \approx \mathcal{N}(\mu, \sigma^2/n)$, and hence
a 95\% CI is $\hat\mu \pm 1.96\,\sigma/\sqrt{n}$.
As we do not save individual batch losses,
we estimate the standard error $\mathrm{SE} = \sigma/\sqrt{n}$ from the residual standard deviation
of $k$ consecutive losses after linear detrending.

\begin{table}[h]
  \caption{Setup.}
  \vspace{.1em}
\fbox{
  \begin{minipage}{\textwidth}
    \small

    \vspace{0.8em}

    \centering

    \begin{tabular}{r@{ }l}
      \multicolumn{2}{c}{\textbf{Model}} \\[.1em]
      \toprule
      \multicolumn{2}{c}{\rule{0pt}{1em}\code{minmaxrnc v0.1.0}~~\citep{minmaxpipgpt}} \\[.2em]
      \midrule
      \rule{0pt}{1em}
      \code{d_model}      = & 728             \\[.1em]
      \code{d_state}      = & 1,456           \\[.1em]
      \code{n_layers}     = & 12              \\[.1em]
      \code{conv_type}    = & \code{gated}  \\[.1em]
      \code{gated_output} = & \code{True}   \\[.1em]
      \midrule
      \rule{0pt}{1em}
      \emph{Total parameters:}       & 112,467,448    \\[.1em]
      \bottomrule
    \end{tabular}
    \hspace{2.0em}
    \begin{tabular}{r@{\hspace{1em}}l}
      \multicolumn{2}{c}{\textbf{Training}} \\[.1em]
      \toprule
      \rule{0pt}{1em}
      Optimiser:                      & AdamW (fused)      \\[.3em]
      $\beta_1, \beta_2$:             & 0.9,\ 0.95         \\[.1em]
      $\epsilon$:                     & $10^{-8}$          \\[-.1em]
      Base learning rate:             & $6\times10^{-4}$   \\[.2em]
      \multirow{2}{*}{Weight decay:}  & 0.1 until step 94k \\[-.1em]
                                      & 0.0 after step 94k \\[.1em]
      \midrule
      \rule{0pt}{1em}
      Scheduler:              & Linear warmup $+$ linear decay \textit{\footnotesize(*)}  \\[.2em]
      Total steps:            & 200k steps                                                \\[.1em]
      Warm-up ratio:          & 1\% of total steps (2k steps)                             \\
      Final LR ratio:         & 1\% of peak LR ($6\times10^{-6}$)                         \\[.1em]
      \multicolumn{2}{c}{%
        \footnotesize\it%
        (*) Manual downward adjustments when observing long plateaus}                    \\[.1em]
      \midrule
      \rule{0pt}{1em}
      Micro batch size:                         & 16 sequences \\[.2em]
      \multirow{2}{*}{Gradient accumulation:}   & 32 micro-steps until step 36k \\[-.1em]
                                                & 64 micro-steps after step 36k \\[.1em]
      \midrule
      Precision:               & \code{bfloat16} \\[.1em]
      \code{torch.compile}:  & enabled \\
      \bottomrule
    \end{tabular}

    \vspace{1.4em}

    \begin{tabular}{r@{\hspace{1em}}l}
      \multicolumn{2}{c}{\textbf{Data}} \\[.1em]
      \toprule
      \textit{Dataset:} & OpenWebText (downloaded via Hugging Face \code{datasets}) \\[.2em]
      \textit{Tokeniser:} & GPT-2 BPE tokeniser (vocabulary size 50,257) \\[.2em]
      \textit{Context length:} & 1,024 tokens \\[.2em]
      \textit{Validation split:} & 5,000 documents held out from training \\
      \bottomrule
    \end{tabular}

    \vspace{1.4em}

    \begin{tabular}{r@{\hspace{1em}}l}
      \multicolumn{2}{c}{\textbf{Loss and Stats}} \\[.1em]
      \toprule
      \rule{0pt}{1em}
      \textit{Training:} & loss, learning rate, and throughput logged every 50 steps \\[.2em]
      \textit{Validation:} & loss computed every 1k steps over 2k randomly-sampled batches of
      size~4  \\[.1em]
      \bottomrule
    \end{tabular}
    \vspace{.9em}
  \end{minipage}
}
  \label{tab:gpt_setup}
\end{table}

\FloatBarrier
\subsection{Results}
Figure~\ref{fig:curves} shows the training cross-entropy loss (raw and
exponential moving average with $\beta=0.99$), the validation loss, and the
validation perplexity over the course of training.

The next table reports the key metrics at the start, end, and best
checkpoint of the run.

\begin{center}
  \small
\begin{tabular}{rcc@{\hspace{2em}}l}
\toprule
\textbf{Step} & \textbf{Val.\ loss} & \textbf{Val.\ perplexity} & \textbf{Checkpoint} \\
\midrule
 1,000   & 4.558 & 95.4  & first evaluation \\
 50,000  & 3.254 & 25.9  & mid-training \\
 99,500  & \textbf{3.124} & \textbf{22.7}  & \textbf{best checkpoint} \\
 99,750  & 3.131 & 22.9  & last evaluation \\
\bottomrule
\end{tabular}
\end{center}

The average throughput was \textasciitilde95,500 tokens/s.

\paragraph{Discussion.}
The training loss decreased from \textasciitilde9.59 (step 50) to \textasciitilde3.09
(step 99,975).  The validation loss converged steadily throughout training,
reaching its lowest value of 3.124 (perplexity 22.73) at
step 99,500.
The validation loss of 3.124 translates to the following CI.
\begin{align*}
  \tikz[baseline=(X.base)]{
    \node[
      draw,
       inner xsep=8pt,
       inner ysep=0pt,
       text height=2.7ex, %
       text depth=1.5ex   %
      ] (X) {\ensuremath{\text{true best loss} = 3.124 \pm 0.006 \ ,
    \qquad
    \text{with probability at least }  0.95 \ .}};
}
\end{align*}
The CI is obtained as follows.
With $k=9$, for training steps after 97k,
we obtain $\widehat{\mathrm{SE}} \approx 0.0025$.
The pivot is $t_t$ since we have $k-2 = 7$ residual degrees of freedom,
and hence the 95\% CI is:
$$\hat\mu \;\pm\; t_{7,0.025}\cdot\widehat{\mathrm{SE}} \;=\; \hat\mu \pm 0.006\ .$$

The impact of doubling gradient accumulation from step~36k is clearly visible in the loss and
perplexity curves. Also the impact of disabling weight decay in the last training phase (from
step~94k) is visible in the loss and perplexity curves.

Although we used an LR schedule that would decay linearly in 200k training steps without our manual
downwards adjustments, the results suggest that a linear or cosine decay in 100k steps would yield
a similar outcome without any mid-training intervention---noting that they would have a similar
trend to the learning rate curve in our run (see bottom plot in Figure~\ref{fig:curves}).

\begin{figure}[p]
\centering
\begin{tikzpicture}
\begin{groupplot}[
  group style={group size=1 by 3, vertical sep=2.2cm},
  width=0.88\textwidth,
  height=5.8cm,
  xmin=0, xmax=100000,
  xtick={0,20000,40000,60000,80000,100000},
  xticklabels={0,20K,40K,60K,80K,100K},
  grid=both,
  grid style={line width=0.25pt, draw=gray!25},
  tick label style={font=\small},
  label style={font=\small},
  title style={font=\small\bfseries, yshift=-2pt},
  legend style={font=\footnotesize, draw=none, fill=white, fill opacity=0.8,
                text opacity=1, inner sep=2pt},
  legend cell align=left,
  unbounded coords=jump,
  scaled x ticks=false,
]

\nextgroupplot[
  xlabel={Training step},
  ylabel={Cross-entropy loss},
  title={Training \& Validation Loss},
  legend pos=north east,
  ymin=2.5,
]
\addplot[color=plotgreen, line width=0.6pt, opacity=0.6]%
  table[col sep=comma, x=step, y=loss]{data/gptlike/data_train.csv};
\addlegendentry{Train}
\addplot[color=cyan, line width=1.4pt]
  table[col sep=comma, x=step, y=ema_loss]{data/gptlike/data_train.csv};
\addlegendentry{Train (EMA $\beta{=}0.99$)}
\addplot[color=plotorange, line width=1.4pt]
  table[col sep=comma, x=step, y=val_loss]{data/gptlike/data_val.csv};
\addlegendentry{Validation}

\nextgroupplot[
  xlabel={Training step},
  ylabel={Perplexity},
  title={Validation Perplexity},
]
\addplot[color=plotpurple, line width=1.2pt]
  table[col sep=comma, x=step, y=ppl]{data/gptlike/data_val.csv};

\nextgroupplot[
  xlabel={Training step},
  ylabel={Learning rate ($\times 10^{-4}$)},
  title={Learning Rate},
]
\addplot[color=plotblue, line width=1.2pt]
  table[col sep=comma, x=step, y expr=\thisrow{lr}*10000]{data/gptlike/data_train.csv};

\end{groupplot}
\end{tikzpicture}
\caption{\emph{Top:} training loss (EMA, $\beta=0.99$) and validation loss.
  \emph{Middle:} validation perplexity. \emph{Bottom:} learning rate
         with linear warmup, linear decay, and manual downward adjustments.}
\label{fig:curves}
\end{figure}

\todo{}

\clearpage{}%

\appendixpart{part:related}{Technical Analysis of Related Architectures}

\clearpage{}%

\section{Analysis of Linear RNNs}
\label{appendix:linear-rnns}

\todo{}

\localtableofcontents

\subsection{Linear RNNs}

\begin{closingdefinition}[Linear Recurrent Units]
  Given
  an input space $U \subseteq \reals^{d_\mathrm{in}}$,
  parameter space 
  $\Theta_\mathrm{in} = \Theta_A \times \Theta_B \subseteq \reals^{d_A} \times \reals^{d_B}$,
  and parametric functions
  \begin{align*}
    A : U \times \Theta_A \to \reals^{d_\mathrm{state} \times d_\mathrm{state}},
    \qquad
    B : U \times \Theta_B \to \reals^{d_\mathrm{state}},
  \end{align*}
  a \emph{parametric Linear Recurrent Unit} with input functions $\ptuple{A,B}$
  is parametric dynamics
  $D = \mctuple{X,U,f}{\Theta}{A,B}$
  having
  state space $X \subseteq \reals^{d_\mathrm{state}}$
  and
  dynamics function given by the following equation:
  \begin{equation} \label{eq:linear-dynamics-function-parametric}
    \fpinst{f}{x,u}{\theta_A,\theta_B} 
    \,=\, 
    \fpinst{A}{u}{\theta_A} \cdot x + \fpinst{B}{u}{\theta_B}\, .
  \end{equation}

  In the above Equation~\eqref{eq:linear-dynamics-function-parametric}, 
  the matrix
  $\fpinst{A}{u}{\theta_A} \in \reals^{d_\mathrm{state} \times d_\mathrm{state}}$ 
  is the \emph{state-update matrix} on input $u$ with parameter values $\theta_A$
  and
  the vector
  $\fpinst{B}{u}{\theta_B} \in \reals^{d_\mathrm{state}}$
  is the \emph{additive term} on input $u$ with parameter values $\theta_B$.

  A \emph{Linear Recurrent Unit} is an instance $\pinst{D}{\theta}$ for $\theta \in \Theta$.

  We denote by 
  $\mathfrak{D}^\mathrm{p}_\mathrm{Linear}$ 
  and
  $\mathfrak{D}_\mathrm{Linear}$ 
  the class of parametric Linear Recurrent Units
  and
  the class of Linear Recurrent Units, respectively.
\end{closingdefinition}

\begin{closingdefinition}[Linear Recurrent Layers]
  \emph{Linear Recurrent Layer dynamics} are 
  recurrent layer $\mathfrak{D}_\mathrm{Linear}$-dynamics.

  \emph{Parametric Linear Recurrent Layer dynamics} are 
  recurrent layer $\mathfrak{D}^\mathrm{p}_\mathrm{Linear}$-dynamics.

  Given a class $\mathfrak{G}$ of functions,
  A \emph{Linear Recurrent Layer of type $\mathfrak{G}$} is 
  a recurrent layer of type $(\mathfrak{D}_\mathrm{Linear},\mathfrak{G})$.

  Given a class $\mathfrak{G}^\mathrm{p}$ of parametric functions,
  a \emph{Parametric Linear Recurrent Layer of type $\mathfrak{G}$} is 
  a recurrent layer of type $(\mathfrak{D}^\mathrm{p}_\mathrm{Linear},\mathfrak{G}^\mathrm{p})$.
\end{closingdefinition}

\begin{closingdefinition}[Linear Recurrent Cascades]
  Given a class $\mathfrak{G}$ of functions,
  \emph{Linear Recurrent Cascade dynamics} are 
  recurrent $\mathfrak{G}$-cascade $\mathfrak{D}_\mathrm{Linear}$-dynamics,
  and 
  a \emph{Linear Recurrent Cascade of type $\mathfrak{G}$} is a system having 
  recurrent $\mathfrak{G}$-cascade $\mathfrak{D}_\mathrm{Linear}$-dynamics
  and output function belonging to $\mathfrak{G}$.

  Given a class $\mathfrak{G}^\mathrm{p}$ of parametric functions,
  \emph{Parametric Linear Recurrent Cascade dynamics} are 
  recurrent $\mathfrak{G}^\mathrm{p}$-cascade $\mathfrak{D}^\mathrm{p}_\mathrm{Linear}$-dynamics,
  and
  a \emph{Parametric Linear Recurrent Cascade of type $\mathfrak{G}$} is a system having 
  recurrent $\mathfrak{G}^\mathrm{p}$-cascade $\mathfrak{D}^\mathrm{p}_\mathrm{Linear}$-dynamics
  and output function belonging to $\mathfrak{G}^\mathrm{p}$.
\end{closingdefinition}

\begin{closingdefinition}[Linear Recurrent Neural Cascades]
  \emph{Linear Recurrent Neural Cascade dynamics},
  abbreviated as \emph{Linear RNC dynamics},
  are $\mathfrak{D}_\mathrm{Linear}$-RNC dynamics.

  \emph{Parametric Linear Recurrent Neural Cascade dynamics},
  abbreviated as \emph{Parametric Linear RNC dynamics},
  are $\mathfrak{D}^\mathrm{p}_\mathrm{Linear}$-RNC dynamics.

  A \emph{Linear Recurrent Neural Cascade},
  abbreviated as \emph{Linear RNC},
  is
  a $\mathfrak{D}_\mathrm{Linear}$-RNC.

  A \emph{Parametric Linear Recurrent Neural Cascade},
  abbreviated as \emph{Parametric Linear RNC},
  is
  a $\mathfrak{D}^\mathrm{p}_\mathrm{Linear}$-RNC.
\end{closingdefinition}

\begin{remark}
  We refer to 
  \emph{Linear RNCs} as \emph{Linear RNNs}, 
  and to \emph{Parametric Linear RNCs} as \emph{Parametric Linear RNNs}. 
  The rationale is that, both in the research literature and in practice, the existing recurrent
  neural architectures are referred to as RNNs even though they are RNCs.
\end{remark}

\subsection{General Linear Recurrent Units}

We show next that a general version of Linear Recurrent Units can be expressed as a Linear Recurrent
Unit as per the definitions above, while preserving the norm of state updates.

\begin{closingdefinition}[Linear operators]
  \todo{}
  We denote by
  $\linop \bigl( \reals^{m \times n}, \reals^{m \times n} \bigr)$
  the vector space of linear operators from
  $\reals^{m \times n}$ to $\reals^{m \times n}$.
\end{closingdefinition}

\begin{closingdefinition}[Frobenius norm of a linear operator]
  \label{def:linear-operator-frobenius-norm}
  Let $L \in \linop \big( \reals^{m\times n}, \reals^{p\times q} \big)$.

  For
  $i\in\iival{1}{m}$
  and
  $j\in\iival{1}{n}$,
  let $E_{i,j} \in \reals^{m\times n}$ denote the matrix 
  having 
  entry $1$ at position $(i,j)$ 
  and 
  entry $0$ at every other position.

  The \emph{Frobenius norm} of
  $L$
  is
  \begin{equation*}
    \|L\|_{\mathrm{F}}
    \defeq
    \sqrt{
      \sum_{i=1}^{m}
      \sum_{j=1}^{n}
      \|
        L(E_{i,j})
      \|_{\mathrm{F}}^2
    }.
  \end{equation*}
\end{closingdefinition}

\begin{closingdefinition}[General Linear Recurrent Units]
  Given
  an input space $U \subseteq \reals^{d_\mathrm{in}}$
  and functions
  \begin{align*}
    L : U \to \linop\bigl(\reals^{d_\mathrm{state}\times n_\mathrm{heads}}\bigr),
    \qquad
    B : U \to \reals^{d_\mathrm{state} \times n_\mathrm{heads}},
  \end{align*}
  a \emph{General Linear Recurrent Unit} with input functions $\ptuple{L,B}$
  is dynamics
  \begin{align*}
    D = \mtuple{X,U,f}{L,B},
    \qquad\quad
    X \subseteq \reals^{d_\mathrm{state} \times n_\mathrm{heads}}
  \end{align*}
  with function of the form
  \begin{equation} \label{eq:linear-dynamics-function}
    f(x,u) \,=\, L_u(x) + B(u)
    \qquad\text{with}\qquad 
    L_u(x) \defeq L(u)\,,
  \end{equation}
  The linear operator $L_u$ is called 
  the \emph{state-update linear operator} on input $u$,
  and
  the matrix $B(u)$ is called the \emph{additive term} on input $u$.
\end{closingdefinition}

\begin{closingdefinition}[Parametric General Linear Recurrent Units]
  Given Euclidean subspaces
  \begin{align*}
    U \subseteq \reals^{d_\mathrm{in}},
    \quad
    \Theta_L \subseteq \reals^{d_{\Theta_L}},
    \quad
    \Theta_B \subseteq \reals^{d_{\Theta_B}},
  \end{align*}
  and parametric functions
  \begin{align*}
    L
    : \big(U \times \Theta_L\big)
    \to
    \linop\bigl(\reals^{d_\mathrm{state}\times n_\mathrm{heads}}\bigr),
    \qquad
    B
    : \big(U \times \Theta_B\big)
    \to
    \reals^{d_\mathrm{state} \times n_\mathrm{heads}},
  \end{align*}
  a \emph{Parametric General Linear Recurrent Unit}
  with parametric input functions $\ptuple{L,B}$
  is parametric dynamics
  \begin{equation*}
    D = \ctuple{X,U,f}{\Theta_L, \Theta_B},
    \qquad\quad
    X \subseteq \reals^{d_\mathrm{state} \times n_\mathrm{heads}},
  \end{equation*}
  with parametric dynamics function of the form
  \begin{equation}
    \label{eq:parametric-linear-dynamics-function}
    \fpinst{f}{x,u}{\theta_L,\theta_B}
    \,=\,
    \fpinst{L_u}{x}{\theta_L} + \fpinst{B}{u}{\theta_B}
    \qquad\text{with}\qquad
    \fpinst{L_u}{x}{\theta_L} \defeq \fpinst{L}{u}{\theta_L}\,.
  \end{equation}
  The linear operator $\fpinst{L_u}{x}{\theta_L}$ is called 
  the \emph{state-update linear operator} on input $u$ with parameter values $\theta_L$,
  and
  the matrix $\fpinst{B}{u}{\theta_B}$ is called the \emph{additive term} on input $u$
  with parameter values $\theta_B$.
\end{closingdefinition}

\begin{closingdefinition}[Left-multiplicative Linear Recurrent Units]
  \label{def:left-multiplicative-linear-recurrent-unit}
  Given
  an input space $U \subseteq \reals^{d_\mathrm{in}}$,
  and 
  functions
  \begin{align*}
    A : U \to \reals^{d_\mathrm{state}\times d_\mathrm{state}}, 
    \qquad\quad
    B : U \to \reals^{d_\mathrm{state}\times n_\mathrm{heads}},
  \end{align*}
  a \emph{Left-Multiplicative Linear Recurrent Unit}
  with input functions $\ptuple{A,B}$
  is dynamics
  \begin{equation*}
    D = \mtuple{X,U,f}{A,B},
    \qquad\qquad
    X \subseteq \reals^{d_\mathrm{state}\times n_\mathrm{heads}},
  \end{equation*}
  with dynamics function of the form
  \begin{equation*}
    f(H,u)
    \,\defeq\,
    A(u)\cdot H
    +
    B(u).
  \end{equation*}
  The matrix $A(u)$ is called the \emph{(left-multiplicative) state-update matrix} on input $u$,
  and
  the matrix $B(u)$ is called the \emph{additive term} on input $u$.
\end{closingdefinition}

\begin{closingdefinition}[Parametric left-multiplicative linear recurrent units]
  \label{def:parametric-left-multiplicative-linear-recurrent-unit}
  Given
  Euclidean subspaces
  \begin{align*}
    U \subseteq \reals^{d_\mathrm{in}},
    \quad
    \Theta_A \subseteq \reals^{d_{\Theta_A}},
    \quad
    \Theta_B \subseteq \reals^{d_{\Theta_B}},
  \end{align*}
  and 
  parametric functions
  \begin{align*}
    A
    : \big(U \times \Theta_A\big)
    \to
    \reals^{d_\mathrm{state}\times d_\mathrm{state}},
    \qquad
    B
    : \big(U \times \Theta_B\big)
    \to
    \reals^{d_\mathrm{state} \times n_\mathrm{heads}},
  \end{align*}
  a \emph{parametric left-multiplicative linear recurrent unit}
  with input functions
  $\ptuple{A,B}$
  is dynamics
  \begin{equation*}
    D
    =
    \mctuple{X,U,f}{A,B}{\Theta_A, \Theta_B},
    \qquad\qquad
    X
    \subseteq
    \reals^{d_\mathrm{state}\times n_\mathrm{heads}},
  \end{equation*}
  with parametric dynamics function of the form
  \begin{equation*}
    \fpinst{f}{H,u}{\theta_A, \theta_B}
    \,\defeq\,
    \fpinst{A}{u}{\theta_A} \cdot H
    +
    \fpinst{B}{u}{\theta_B}.
  \end{equation*}
  The matrix
  $\fpinst{A}{u}{\theta_A}$
  is called the
  \emph{(left-multiplicative) state-update matrix} on input $u$ for parameter values $\theta_A$,
  and
  the matrix $\fpinst{B}{u}{\theta_B}$
  is called the
  \emph{additive term}
  on input $u$ for parameter values $\theta_B$.
\end{closingdefinition}

\subsubsection{Facts about General Linear Recurrent Units}
\label{subsubsec:facts-about-general-linear-recurrent-units}

We next provide facts about the above definitions.

\begin{closingproposition}[Correctness of instantiation of Parametric General Linear Recurrent Units]
  \label{prop:parametric-general-linear-instance-is-general-linear}
  Let $D$ be a Parametric General Linear Recurrent Unit,
  and 
  let $\Theta$ be its joint parameter space.
  For every $\theta \in \Theta$,
  we have that
  $\pinst{D}{\theta}$ is a General Linear Recurrent Unit.
\end{closingproposition}
\begin{proof}
  Let
  \begin{equation*}
    D
    =
    \ctuple{X,U,f}{\Theta}
  \end{equation*}
  be a Parametric General Linear Recurrent Unit,
  with parametric input functions
  \begin{equation*}
    L
    :
    U\times\Theta_L
    \to
    \linop
    \bigl(
      \reals^{d_\mathrm{state}\times n_\mathrm{heads}}
    \bigr)
    \qquad\text{and}\qquad
    B
    :
    U\times\Theta_B
    \to
    \reals^{d_\mathrm{state}\times n_\mathrm{heads}},
  \end{equation*}
  where
  \begin{equation*}
    \Theta
    =
    \Theta_L\times\Theta_B.
  \end{equation*}

  Let
  \begin{equation*}
    \theta
    =
    \ptuple{
      \theta_L,
      \theta_B
    }
    \in
    \Theta.
  \end{equation*}
  The instance
  $\pinst{D}{\theta}$
  has instantiated input functions
  \begin{align*}
    \pinst{L}{\theta_L}(u)
    &\defeq
    L(u,\theta_L),
    \\
    \pinst{B}{\theta_B}(u)
    &\defeq
    B(u,\theta_B),
  \end{align*}
  and instantiated dynamics function
  \begin{align*}
    \pinst{f}{\theta}(x,u)
    &\defeq
    f(x,u;\theta)
    \\
    &=
    L(u,\theta_L)(x)
    +
    B(u,\theta_B)
    \\
    &=
    \pinst{L}{\theta_L}(u)(x)
    +
    \pinst{B}{\theta_B}(u).
  \end{align*}

  Furthermore,
  \begin{equation*}
    \pinst{L}{\theta_L}
    :
    U
    \to
    \linop
    \bigl(
      \reals^{d_\mathrm{state}\times n_\mathrm{heads}}
    \bigr)
  \end{equation*}
  and
  \begin{equation*}
    \pinst{B}{\theta_B}
    :
    U
    \to
    \reals^{d_\mathrm{state}\times n_\mathrm{heads}}.
  \end{equation*}
  Hence
  $\pinst{D}{\theta}$
  satisfies the definition of a General Linear Recurrent Unit.
\end{proof}

\begin{closingproposition}[Correctness of instantiation of parametric left-multiplicative linear recurrent units]
  \label{prop:parametric-left-multiplicative-instance-is-left-multiplicative}
  Let $D$ be a parametric Left-Multiplicative Linear Recurrent Unit,
  and 
  let $\Theta$ be its joint parameter space.
  For every $\theta \in \Theta$,
  we have that
  $\pinst{D}{\theta}$ is a Left-Multiplicative Linear Recurrent Unit.
\end{closingproposition}
\begin{proof}
  Let
  \begin{equation*}
    D
    =
    \mctuple{X,U,f}{A,B}{\Theta_A,\Theta_B}
  \end{equation*}
  be a parametric Left-Multiplicative Linear Recurrent Unit,
  where
  \begin{equation*}
    \Theta
    =
    \Theta_A
    \times
    \Theta_B
  \end{equation*}
  is its joint parameter space, and
  \begin{align*}
    A
    &:
    U\times\Theta_A
    \to
    \reals^{d_\mathrm{state}\times d_\mathrm{state}},
    \\
    B
    &:
    U\times\Theta_B
    \to
    \reals^{d_\mathrm{state}\times n_\mathrm{heads}}
  \end{align*}
  are its parametric input functions.

  Let
  \begin{equation*}
    \theta
    =
    \ptuple{
      \theta_A,
      \theta_B
    }
    \in
    \Theta.
  \end{equation*}
  The instance
  $\pinst{D}{\theta}$
  has instantiated input functions
  \begin{align*}
    \pinst{A}{\theta_A}(u)
    &\defeq
    A(u,\theta_A),
    \\
    \pinst{B}{\theta_B}(u)
    &\defeq
    B(u,\theta_B),
  \end{align*}
  and instantiated dynamics function
  \begin{align*}
    \pinst{f}{\theta}(H,u)
    &\defeq
    \fpinst{f}{H,u}{\theta_A,\theta_B}
    \\
    &=
    A(u,\theta_A)\cdot H
    +
    B(u,\theta_B)
    \\
    &=
    \pinst{A}{\theta_A}(u)\cdot H
    +
    \pinst{B}{\theta_B}(u).
  \end{align*}

  Furthermore,
  \begin{equation*}
    \pinst{A}{\theta_A}
    :
    U
    \to
    \reals^{d_\mathrm{state}\times d_\mathrm{state}}
  \end{equation*}
  and
  \begin{equation*}
    \pinst{B}{\theta_B}
    :
    U
    \to
    \reals^{d_\mathrm{state}\times n_\mathrm{heads}}.
  \end{equation*}
  Hence
  $\pinst{D}{\theta}$
  satisfies the definition of a
  Left-Multiplicative Linear Recurrent Unit.
\end{proof}

\begin{closingfact}[Left-multiplicative Linear Recurrent Units are General Linear Recurrent Units]
  \label{fact:left-multiplicative-is-general-linear}
  We have that (parametric) left-multiplicative Linear Recurrent Units are (parametric, resp.)
  General Linear Recurrent Units, as more explicitly stated below.

  \par\medskip
  \textbf{Fact~\ref{fact:left-multiplicative-is-general-linear}.1: Non-parametric case.}~
  Every Left-Multiplicative Linear Recurrent Unit is a General Linear Recurrent Unit.
  Specifically,
  every Left-Multiplicative Linear Recurrent Unit 
  \begin{align*}
    D = \mtuple{X,U,f}{A,B}, 
    \qquad\quad
    X \subseteq \Omega_X = \reals^{d_\mathrm{state} \times n_\mathrm{heads}},
  \end{align*}
  can be equivalently expressed as the 
  General Linear Recurrent Unit
  \begin{equation*}
    \tilde{D} \defeq \mtuple{X,U,f}{L^A,B}
  \end{equation*}
  where the function $L^A : U \to \linop(\Omega_X)$ is given by
  \begin{align} \label{eq:left-multiplicative-induced-linear-operator}
    &
    \qquad\qquad
    L^A(u) 
    \defeq 
    L^A_u 
    \;\quad\text{with}\quad\;
    L^A_u(H) 
    \defeq
    A(u)\cdot H
    &&
    \forall\,
    u \!\in\! U,\,
    H \!\in\! \Omega_X.
  \end{align}

  \par\medskip
  \textbf{Fact~\ref{fact:left-multiplicative-is-general-linear}.2: Parametric case.}~
  Every Parametric Left-Multiplicative Linear Recurrent Unit is a
  Parametric General Linear Recurrent Unit.
  Specifically,
  every Parametric Left-Multiplicative Linear Recurrent Unit 
  \begin{align*}
    D = \mctuple{X,U,f}{A,B}{\Theta_A, \Theta_B}, 
    \qquad\quad
    X \subseteq \Omega_X = \reals^{d_\mathrm{state} \times n_\mathrm{heads}},
  \end{align*}
  can be equivalently expressed as the 
  Parametric General Linear Recurrent Unit
  \begin{equation*}
    \tilde{D} \defeq \mctuple{X,U,f}{L^A,B}{\Theta_A, \Theta_B}
  \end{equation*}
  where 
  the parametric function 
  $L^A : \big(U \times \Theta_A\big) \to \linop(\Omega_X)$ 
  is given by
  \begin{align} \label{eq:parametric-left-multiplicative-induced-linear-operator}
    &
    \fpinst{L^A}{u}{\theta_A}
    \defeq 
    \pinst{L^A_u}{\theta_A}
    \;\;\text{with}\;\;
    \fpinst{L^A_u}{H}{\theta_A}
    \defeq
    \fpinst{A}{u}{\theta_A}\cdot H
    &&
    \quad
    \forall\, 
    \theta_A \!\in\! \Theta_A,
    u \!\in\! U,
    H \!\in\! \Omega_X.
  \end{align}
\end{closingfact}
\begin{proof}[%
    Proof of Fact~\ref{fact:left-multiplicative-is-general-linear}.1: Non-parametric case%
  ]
  Let
  \begin{equation*}
    D
    =
    \mtuple{X,U,f}{A,B}
  \end{equation*}
  be a Left-Multiplicative Linear Recurrent Unit, with
  \begin{equation*}
    X
    \subseteq
    \Omega_X
    =
    \reals^{d_\mathrm{state}\times n_\mathrm{heads}}.
  \end{equation*}
  Let
  \begin{equation*}
    L^A
    :
    U
    \to
    \linop(\Omega_X)
  \end{equation*}
  be defined as in
  Equation~\eqref{eq:left-multiplicative-induced-linear-operator}.

  We first verify that
  $L^A$
  is well defined.
  Fix
  $u\in U$.
  For every
  $H_1,H_2\in\Omega_X$
  and
  $\alpha,\beta\in\reals$,
  \begin{align*}
    L^A_u
    \big(
      \alpha H_1+\beta H_2
    \big)
    &=
    A(u)
    \cdot
    \big(
      \alpha H_1+\beta H_2
    \big)
    \\
    &=
    \alpha
    A(u)\cdot H_1
    +
    \beta
    A(u)\cdot H_2
    \\
    &=
    \alpha
    L^A_u(H_1)
    +
    \beta
    L^A_u(H_2).
  \end{align*}
  Hence
  $L^A_u$
  is a linear operator on
  $\Omega_X$.
  Since
  $u\in U$
  was arbitrary,
  \begin{equation*}
    L^A
    :
    U
    \to
    \linop(\Omega_X)
  \end{equation*}
  is well defined.

  It remains to verify that
  $\tilde D$
  has the same dynamics as
  $D$.
  Since
  $D$
  is a Left-Multiplicative Linear Recurrent Unit,
  for every
  $H\in X$
  and
  $u\in U$,
  \begin{align*}
    f(H,u)
    &=
    A(u)\cdot H
    +
    B(u)
    \\
    &=
    L^A_u(H)
    +
    B(u)
    \\
    &=
    L^A(u)(H)
    +
    B(u).
  \end{align*}
  Therefore,
  \begin{equation*}
    \tilde D
    =
    \mtuple{X,U,f}{L^A,B}
  \end{equation*}
  satisfies the definition of a
  General Linear Recurrent Unit,
  and has the same state space, input space, and dynamics function as
  $D$.
\end{proof}
\begin{proof}[%
    Proof of Fact~\ref{fact:left-multiplicative-is-general-linear}.2: Parametric case%
  ]
  Let
  \begin{equation*}
    D
    =
    \mctuple{X,U,f}{A,B}{\Theta_A,\Theta_B}
  \end{equation*}
  be a Parametric Left-Multiplicative Linear Recurrent Unit, with
  \begin{equation*}
    X
    \subseteq
    \Omega_X
    =
    \reals^{d_\mathrm{state}\times n_\mathrm{heads}}.
  \end{equation*}
  Let
  \begin{equation*}
    L^A
    :
    U\times\Theta_A
    \to
    \linop(\Omega_X)
  \end{equation*}
  be defined as in
  Equation~\eqref{eq:parametric-left-multiplicative-induced-linear-operator}.

  We first verify that
  $L^A$
  is well defined.
  Fix
  $u\in U$
  and
  $\theta_A\in\Theta_A$.
  For every
  $H_1,H_2\in\Omega_X$
  and
  $\alpha,\beta\in\reals$,
  \begin{align*}
    \fpinst{L^A_u}{
      \alpha H_1+\beta H_2
    }{\theta_A}
    &=
    \fpinst{A}{u}{\theta_A}
    \cdot
    \big(
      \alpha H_1+\beta H_2
    \big)
    \\
    &=
    \alpha
    \fpinst{A}{u}{\theta_A}
    \cdot H_1
    +
    \beta
    \fpinst{A}{u}{\theta_A}
    \cdot H_2
    \\
    &=
    \alpha
    \fpinst{L^A_u}{H_1}{\theta_A}
    +
    \beta
    \fpinst{L^A_u}{H_2}{\theta_A}.
  \end{align*}
  Hence
  $\pinst{L^A_u}{\theta_A}$
  is a linear operator on
  $\Omega_X$.
  Since
  $u\in U$
  and
  $\theta_A\in\Theta_A$
  were arbitrary,
  \begin{equation*}
    L^A
    :
    U\times\Theta_A
    \to
    \linop(\Omega_X)
  \end{equation*}
  is well defined.

  It remains to verify that
  $\tilde D$
  has the same parametric dynamics as
  $D$.
  Since
  $D$
  is a Parametric Left-Multiplicative Linear Recurrent Unit,
  for every
  $H\in X$,
  $u\in U$,
  $\theta_A\in\Theta_A$,
  and
  $\theta_B\in\Theta_B$,
  \begin{align*}
    \fpinst{f}{H,u}{\theta_A,\theta_B}
    &=
    \fpinst{A}{u}{\theta_A}
    \cdot H
    +
    \fpinst{B}{u}{\theta_B}
    \\
    &=
    \fpinst{L^A_u}{H}{\theta_A}
    +
    \fpinst{B}{u}{\theta_B}
    \\
    &=
    \fpinst{L^A}{u}{\theta_A}(H)
    +
    \fpinst{B}{u}{\theta_B}.
  \end{align*}
  Therefore,
  \begin{equation*}
    \tilde D
    =
    \mctuple{X,U,f}{L^A,B}{\Theta_A,\Theta_B}
  \end{equation*}
  satisfies the definition of a
  Parametric General Linear Recurrent Unit,
  and has the same state space, input space, parameter spaces, and
  parametric dynamics function as
  $D$.
\end{proof}

\subsubsection{Reduction to Linear Recurrent Units}
\label{subsubsec:reduction-to-linear-recurrent-units-}

\todo{}

\begin{closingdefinition} \label{def:flattening}
  Let $m,n,p,q \in \posnaturals$.

  \begin{definitionpart}[indent=body]
    The \emph{flattening operator} on matrices is the linear bijection
    $\flattenop
    :
    \reals^{m \times n}
    \to
    \reals^{mn}$
    that maps every matrix to the vector obtained by stacking its columns; namely, it is the map
    \begin{align*}
      \begin{pmatrix}
        x_{1,1} & \cdots & x_{1,n} \\
        \vdots  & \ddots & \vdots  \\
        x_{m,1} & \cdots & x_{m,n}
      \end{pmatrix}
      \quad\mapsto\quad
      \begin{pmatrix}
        x_{1,1} \\
        \vdots  \\
        x_{m,1} \\
        \vdots  \\
        x_{1,n} \\
        \vdots  \\
        x_{m,n}
      \end{pmatrix}.
    \end{align*}
  \end{definitionpart}

  \begin{definitionpart}[indent=body]
    For every subset
    $X \subseteq \reals^{m \times n}$,
    we define its flattening element-wise as
    \begin{align*}
      \flattenop(X)
      \defeq
      \bigl\{
        \flattenop(x)
        \mid
        x \in X
      \bigr\}
      \subseteq
      \reals^{mn}.
    \end{align*}
  \end{definitionpart}

  \begin{definitionpart}[indent=body]
    The flattening operator on linear operators is the map
    \begin{align*}
      \flattenop
      :
      \linop
      \bigl(
      \reals^{m \times n},
      \reals^{p \times q}
      \bigr)
      \to
      \reals^{(pq) \times (mn)}
    \end{align*}
    such that, for every linear operator
    \begin{align*}
      L
      \in
      \linop
      \bigl(
      \reals^{m \times n},
      \reals^{p \times q}
      \bigr),
    \end{align*}
    we have that $\flattenop(L)$ is the unique matrix satisfying
    \begin{align*}
      \flattenop\bigl(L(X)\bigr)
      \,=\,
      \flattenop(L)
      \cdot
      \flattenop(X)
      \qquad\qquad
      \forall X \in \reals^{m \times n}.
    \end{align*}
  \end{definitionpart}
\end{closingdefinition} 

\begin{closingdefinition}[State flattening induced by recurrent cascade presentations]
  \label{def:flattening-rnc}
  Let us consider a parametric recurrent cascade presentation
  \begin{equation*}
    \presentationSty{C}
    \,=\,
    {
      \repinputprecomposition{
        g_1
      }{
        \presentationSty{L}_1
      }
      \repcascade_{g_2}
      \presentationSty{L}_2
      \repcascade_{g_3}
      \cdots
      \repcascade_{g_\ell}
      \presentationSty{L}_\ell
    }
    \qquad\text{where}\qquad
    \presentationSty{L}_i
    \,=\,
    {
      D_{i,1}
      \replayer
      \cdots
      \replayer
      D_{i,m_i}
    }
    \quad\;
    \forall
    i\in\iival{1}{\ell}
    \,.
  \end{equation*}

  \begin{definitionpart}[indent=body]
    Let $i \in \iival{1}{\ell}$ be a layer index,
    let $j \in \iival{1}{m_i}$ be an index of a unit within the $i$-th layer,
    and
    let $X_{i,j}$ the state space of unit $D_{i,j}$.

    Assume that the ambient space $\Omega_{X_{i,j}}$ of the state space $X_{i,j}$
    is
    either
    $\reals^{d_{i,j}}$
    or
    $\reals^{r_{i,j}\times c_{i,j}}$.

    The
    \emph{flattened ambient state space}
    of
    unit
    $D_{i,j}$
    is
    \begin{equation*}
      \tilde\Omega_{X_{i,j}}
      \,\defeq\,
      \begin{cases}
        \reals^{d_{i,j}}
        &
        \text{if }\,
        \Omega_{X_{i,j}}
        \!=
        \reals^{d_{i,j}},
        \\[.3em]
        \reals^{r_{i,j}c_{i,j}}
        &
        \text{if }\,
        \Omega_{X_{i,j}}
        \!=
        \reals^{r_{i,j}\times c_{i,j}}\,.
      \end{cases}
    \end{equation*}
    The
    \emph{state flattening} 
    for unit $D_{i,j}$
    is the linear bijection
    \begin{equation*}
      \phi_{i,j}
      :
      \Omega_{X_{i,j}}
      \to
      \tilde\Omega_{X_{i,j}},
      \qquad\qquad
      \phi_{i,j}
      \,\defeq\,
      \begin{cases}
        \operatorname{id}_{\reals^{d_{i,j}}}
        &
        \text{if }\,
        \Omega_{X_{i,j}}
        \!=
        \reals^{d_{i,j}},
        \\[.3em]
        \flattenop
        &
        \text{if }\,
        \Omega_{X_{i,j}}
        \!=
        \reals^{r_{i,j}\times c_{i,j}}
        \,.
      \end{cases}
    \end{equation*}
    The
    \emph{flattened state space} 
    of unit $D_{i,j}$
    is
    \begin{equation*}
      \tilde X_{i,j}
      \,\defeq\,
      \phi_{i,j}(X_{i,j})
      \,.
    \end{equation*}
  \end{definitionpart}

  \begin{definitionpart}[indent=body]
    Let $i\in\iival{1}{\ell}$ be a layer index.

    The \emph{state flattening at layer $i$}
    induced by $\presentationSty{C}$
    is
    \begin{equation*}
      \phi_i
      \,\defeq\,
      \bigptuple{
      \phi_{i,1},
      \ldots,
      \phi_{i,m_i}
      }
      \,.
    \end{equation*}

    For $i \geq 2$,
    the
    \emph{prefix state flattening up to layer $i-1$}
    induced by $\presentationSty{C}$
    is
    \begin{equation*}
      \phi_{<i}
      \,\defeq\,
      \bigptuple{
      \phi_1,
      \ldots,
      \phi_{i-1}
      }\,.
    \end{equation*}
  \end{definitionpart}

  \begin{definitionpart}[indent=body]
    The 
    \emph{state flattening induced by $\presentationSty{C}$}
    is
    \begin{equation*}
      \phi_\presentationSty{C}
      \,\defeq\,
      \bigptuple{
        \phi_1,\,
        \ldots,\,
        \phi_\ell
      },
    \end{equation*}
    and
    the \emph{flattened state space induced by $\presentationSty{C}$}
    is
    \begin{equation*}
      \tilde X
      \,\defeq\,
      \phi_\presentationSty{C}(X)
    \end{equation*}
    for $X$ the state space of $\presentationVal{\presentationSty{C}}$.
  \end{definitionpart}
\end{closingdefinition}

\begin{closingfact}[Invertibility of presentation-induced state flattening]
  \label{fact:flattening-rnc-bijection}
  In the context of
  Definition~\ref{def:flattening-rnc},
  the maps
  $(\phi_i)_{i=1}^\ell$,
  $(\phi_{<i})_{i=2}^\ell$,
  and
  $\phi_\presentationSty{C}$
  are linear bijections.
  In particular,
  \begin{equation*}
    \phi_\presentationSty{C}^{-1}
    \,=\,
    \bigptuple{
      \phi_1^{-1},\,
      \ldots,\,
      \phi_\ell^{-1}
    }
    \qquad\text{where}\qquad
    \phi_i^{-1}
    \,=\,
    \bigptuple{
      \phi_{i,1}^{-1},\,
      \ldots,\,
      \phi_{i,m_i}^{-1}
    }
    \quad\;\forall i \in \iival{1}{\ell}.
  \end{equation*}
\end{closingfact}
\begin{proof}
  Each
  $\phi_{i,j}$
  is either an identity map or the matrix flattening map of
  Definition~\ref{def:flattening},
  and hence is a linear bijection.
  Finite Cartesian products of linear bijections are linear bijections,
  with inverse given component-wise.
\end{proof}

\begin{closingdefinition}[Canonical Linear Recurrent Unit representation]
  \label{def:linear-rnn-representation}
  \label{part:linear-rnn-representation--units}
  \label{part:canonical-linear-rnn--units}
  Let us consider a Parametric General Linear Recurrent Unit
  \begin{equation*}
    D
    =
    \ctuple{
      X,
      U,
      f
    }{
      \Theta_L,
      \Theta_B
    },
  \end{equation*}
  let its state space be 
  $X \subseteq \Omega_X = \reals^{d_\mathrm{state}\times n_\mathrm{heads}}$,
  and 
  let its parametric input functions be
  \begin{align*}
    L
    :
    U\times\Theta_L
    \to
    \linop
    \big(
      \Omega_X,
      \Omega_X
    \big)
    \qquad\text{and}\qquad
    B
    :
    U\times\Theta_B
    \to
    \Omega_X
    .
    \quad
  \end{align*}
  Let us define
  \begin{equation*}
    d
    \,\defeq\,
    d_\mathrm{state}
    \cdot
    n_\mathrm{heads}
    \,,
    \qquad\quad
    \phi
    \,\defeq\,
    \flattenop
    :
    \Omega_X
    \to
    \reals^d
    \,,
    \qquad\qquad
    \tilde X 
    \,\defeq\, 
    \phi(X)
    \,.
  \end{equation*}

  The
  \emph{canonical Linear Recurrent Unit representation of $D$}
  is the Parametric Linear Recurrent Unit
  \begin{equation*}
    \tilde D
    \;\defeq\;
    \mctuple{
      \tilde X,
      U,
      \tilde f
    }{
      \Theta_L\times\Theta_B
    }{
      \tilde A,
      \tilde B
    }
  \end{equation*}
  whose parametric input functions are 
  \begin{align*}
    \fpinst{\tilde A}{u}{\theta_L}
    \,\defeq\,
    \flattenop
    \left(
      \fpinst{L}{u}{\theta_L}
    \right)
    \qquad\text{and}\qquad
    \fpinst{\tilde B}{u}{\theta_B}
    \,\defeq\,
    \flattenop
    \left(
      \fpinst{B}{u}{\theta_B}
    \right).
  \end{align*}
\end{closingdefinition}

\begin{closingremark}
  In the context of Definition~\ref{def:linear-rnn-representation},
  the dynamics function $\tilde f$ is given by
  \begin{equation*}
    \bigfpinst{\tilde f}{
      \tilde x,
      u
    }{
      \theta_L,
      \theta_B
    }
    \,=\,
    \fpinst{\tilde A}{u}{\theta_L}
    \cdot
    \tilde x
    +
    \fpinst{\tilde B}{u}{\theta_B}.
  \end{equation*}
\end{closingremark}

\begin{closingproposition}[Correctness of the canonical Linear Recurrent Unit representation]
  \label{prop:general-linear-unit-flattening}
  In the context of
  Definition~\ref{def:linear-rnn-representation},
  we have
  \begin{equation*}
    \phi
    \left(
      \bigfpinst{f}{
        x,
        u
      }{
        \theta_L,
        \theta_B
      }
    \right)
    \,=\,
    \bigfpinst{\tilde f}{
      \phi(x),
      u
    }{
      \theta_L,
      \theta_B
    }
    \qquad
    \forall x\in X,
    \;
    \forall u\in U,
    \;
    \forall \theta_L\in\Theta_L,
    \;
    \forall \theta_B\in\Theta_B.
  \end{equation*}
  Hence
  $D$
  and
  $\tilde D$
  are isomorphic under the state isomorphism
  $\phi$.
\end{closingproposition}
\begin{proof}
  By the definition of a Parametric General Linear Recurrent Unit,
  \begin{align*}
    \phi
    \left(
      \bigfpinst{f}{
        x,
        u
      }{
        \theta_L,
        \theta_B
      }
    \right)
    &=
    \flattenop
    \left(
      \fpinst{L}{u}{\theta_L}(x)
      +
      \fpinst{B}{u}{\theta_B}
    \right)
    \\[.2em]
    &=
    \flattenop
    \left(
      \fpinst{L}{u}{\theta_L}(x)
    \right)
    +
    \flattenop
    \left(
      \fpinst{B}{u}{\theta_B}
    \right).
  \end{align*}
  By
  Definition~\ref{def:flattening},
  \begin{equation*}
    \flattenop
    \left(
      \fpinst{L}{u}{\theta_L}(x)
    \right)
    =
    \flattenop
    \left(
      \fpinst{L}{u}{\theta_L}
    \right)
    \cdot
    \flattenop(x).
  \end{equation*}
  Therefore,
  \begin{align*}
    \phi
    \left(
      \bigfpinst{f}{
        x,
        u
      }{
        \theta_L,
        \theta_B
      }
    \right)
    &=
    \fpinst{\tilde A}{u}{\theta_L}
    \cdot
    \phi(x)
    +
    \fpinst{\tilde B}{u}{\theta_B}
    \\[.2em]
    &=
    \bigfpinst{\tilde f}{
      \phi(x),
      u
    }{
      \theta_L,
      \theta_B
    },
  \end{align*}
  proving the fact.
\end{proof}

\begin{closingdefinition}[Linear RNN representations induced by RNC presentations]
  \label{def:canonical-linear-rnn}
  Let us consider a parametric RNC
  \begin{equation*}
    S
    =
    \ctuple{
    X,
    U_S,
    f_S,
    x^\mathrm{init},
    Y,
    h
    }{
    \Theta_f,
    \Theta_h
    },
  \end{equation*}
  and a parametric analytic output extension presentation of $S$ given by
  \begin{equation*}
    \presentationSty{S}
    \,\defeq\,
    \repoutoutextension{
    \presentationSty{C}
    }{
    x^\mathrm{init}
    }{
    h
    }.
  \end{equation*}

  Suppose that the above parametric recurrent cascade presentation 
  $\presentationSty{C}$ 
  is the form
  \begin{equation*}
    \presentationSty{C}
    \;=\;
    {
    \repinputprecomposition{
    g_1
    }{
    \presentationSty{L}_1
    }
    \repcascade_{g_2}
    \presentationSty{L}_2
    \repcascade_{g_3}
    \cdots
    \repcascade_{g_\ell}
    \presentationSty{L}_\ell
    }
  \end{equation*}
  where, in turn, each parametric recurrent layer presentation
  $\presentationSty{L}_i$ %
  is of the form
  \begin{equation*}
    \presentationSty{L}_i
    \;=\;
    {
    D_{i,1}
    \replayer
    \cdots
    \replayer
    D_{i,m_i}
    }
  \end{equation*}
  with units $D_{i,1}, \ldots, D_{i,m_i}$ being Parametric General Linear Recurrent Units.

  Let
  $(\phi_i){i=1}^\ell$,
  $(\phi_{<i})_{i=2}^\ell$,
  and
  $\phi_\presentationSty{C}$
  be the state flattenings induced by
  $\presentationSty{C}$
  according to
  Definition~\ref{def:flattening-rnc}.

  \begin{definitionpart}[indent=body]
    \label{part:def:linear-rnn-representation--rnc}
    \label{part:def:canonical-linear-rnn--rnc}
    Let $i \in \iival{1}{\ell}$ be a layer index.
    For every $j \in \iival{1}{m_i}$,
    let us write $\tilde D_{i,j}$ for the canonical Linear Recurrent Unit representation of
    $D_{i,j}$ according to Definition~\ref{def:linear-rnn-representation}.

    We define
    \emph{flattened layer presentation for layer $i$}
    as
    \begin{equation*}
      \presentationSty{\tilde L}_i
      \;\defeq\;
      {
      \tilde D_{i,1}
      \replayer
      \cdots
      \replayer
      \tilde D_{i,m_i}
      }
      \,.
    \end{equation*}
  \end{definitionpart}

  \begin{definitionpart}[indent=body]
    Let $i \in \iival{2}{\ell}$ be a layer index---note that $i \geq 2$.

    We define the \emph{flattened $i$-th connecting function} as
    \begin{equation}
      \label{eq:flattened-connecting-function}
      \bigfpinst{\tilde g_i}{
      \tilde x_\mathrm{pre},
      u,
      \tilde x_\mathrm{cur}
      }{
      \theta_{g_i}
      }
      \,\defeq\,
      \mBigfpinst{g_i}{
      \phi_{<i}^{-1}(\tilde x_\mathrm{pre}),\;
      u,\;
      \phi_{<i}^{-1}(\tilde x_\mathrm{cur})
      }{
      \theta_{g_i}
      }.
    \end{equation}
  \end{definitionpart}

  \begin{definitionpart}[indent=body]
    Given the above intermediate definitions, we define the following.

    The
    \emph{flattened recurrent cascade presentation}
    associated with
    $\presentationSty{C}$
    is
    \begin{equation*}
      \presentationSty{\tilde C}
      \,\defeq\,
      \repinputprecomposition{
      g_1
      }{
      \presentationSty{\tilde L}_1
      }
      \repcascade_{\tilde g_2}
      \presentationSty{\tilde L}_2
      \repcascade_{\tilde g_3}
      \cdots
      \repcascade_{\tilde g_\ell}
      \presentationSty{\tilde L}_\ell.
    \end{equation*}

    The \emph{flattened initial state}
    associated with
    $\presentationSty{C}$
    is
    \begin{equation*}
      \tilde x^\mathrm{init}
      \,\defeq\,
      \phi_\presentationSty{C}\big(x^\mathrm{init}\big)
      \,.
    \end{equation*}

    The \emph{flattened output function} 
    associated with
    $\presentationSty{C}$
    is
    \begin{equation}
      \label{eq:flattened-rnc-output-function}
      \mBigfpinst{\tilde h}{
      \tilde x_\mathrm{pre},\,
      u,\,
      \tilde x_\mathrm{cur}
      }{
      \theta_h
      }
      \;\defeq\;
      \mBigfpinst{h}{
      \phi_\presentationSty{C}^{-1}\big(\tilde x_\mathrm{pre}\big),\;
      u,\;
      \phi_\presentationSty{C}^{-1}\big(\tilde x_\mathrm{cur}\big)
      }{
      \theta_h
      }
      \,.
    \end{equation}

    The \emph{Linear RNN presentation} 
    associated with $\presentationSty{S}$
    is the output extension presentation
    \begin{equation*}
      \presentationSty{\tilde S}
      \,\defeq\,
      \repoutoutextension{
      \presentationSty{\tilde C}
      }{
      \tilde x^\mathrm{init}
      }{
      \tilde h
      }.
    \end{equation*}

    The \emph{Linear RNN representation of $S$} 
    associated with $\presentationSty{S}$
    is the parametric system
    \begin{equation*}
      \tilde S
      \,\defeq\,
      \presentationVal{
      \presentationSty{\tilde S}
      }.
    \end{equation*}
  \end{definitionpart}
\end{closingdefinition}

\begin{closingproposition}[Correctness of presentation-induced Linear RNN representations]
  \label{prop:linear-rnn-representation-correctness}
  Let
  \begin{equation*}
    S
    =
    \ctuple{
      X,
      U,
      f,
      x^\mathrm{init},
      Y,
      h
    }{
      \Theta_f,
      \Theta_h
    }
  \end{equation*}
  be a parametric RNC,
  and let
  \begin{equation*}
    \presentationSty{S}
    \,\defeq\,
    \repoutoutextension{
      \presentationSty{C}
    }{
      x^\mathrm{init}
    }{
      h
    }
  \end{equation*}
  be a parametric analytic output extension presentation of
  $S$
  whose recurrent cascade presentation is of the form
  \begin{equation*}
    \presentationSty{C}
    \,=\,
    \repinputprecomposition{
      g_1
    }{
      \presentationSty{L}_1
    }
    \repcascade_{g_2}
    \presentationSty{L}_2
    \repcascade_{g_3}
    \cdots
    \repcascade_{g_\ell}
    \presentationSty{L}_\ell
    \qquad\text{ where }\qquad
    \presentationSty{L}_i
    \,=\,
    D_{i,1}
    \replayer
    \cdots
    \replayer
    D_{i,m_i}
    \qquad
    \forall i\in\iival{1}{\ell},
  \end{equation*}
  and 
  each $D_{i,j}$ is a Parametric General Linear Recurrent Unit.

  Let 
  $\presentationSty{\tilde C}$
  be the flattened recurrent cascade presentation,
  let 
  $\tilde x^\mathrm{init}$
  be the flattened initial state,
  let
  $\tilde h$
  be the flattened output function, 
  let 
  $\presentationSty{\tilde S}$,
  be the associated Linear RNN presentation,
  and
  let
  $\tilde S$
  be the associated Linear RNN representation---all induced by $\presentationSty{S}$ according to
  Definition~\ref{def:canonical-linear-rnn}.

  Let 
  $\phi_\presentationSty{C}$ 
  be the state flattening induced by $\presentationSty{C}$---according to
  Definition~\ref{def:flattening-rnc}.

  Let us write
  \begin{equation*}
    \presentationVal{
      \presentationSty{C}
    }
    =
    \ctuple{
      X,
      U,
      f
    }{
      \Theta_f
    },
    \qquad\qquad
    \presentationVal{
      \presentationSty{\tilde C}
    }
    =
    \ctuple{
      \tilde X,
      U,
      \tilde f
    }{
      \Theta_f
    }.
  \end{equation*}

  Then:
  \begin{itemize}
    \item
      $\tilde X = \phi_\presentationSty{C}(X)$;

    \item
      for
      every
      $x\in X$,
      every
      $u\in U$,
      and every
      $\theta_f\in\Theta_f$,
      \begin{equation}
        \label{eq:linear-rnn-whole-state-conjugacy}
        \phi_\presentationSty{C}
        \left(
          \fpinst{f}{
            x,
            u
          }{
            \theta_f
          }
        \right)
        =
        \bigfpinst{\tilde f}{
          \phi_\presentationSty{C}(x),
          u
        }{
          \theta_f
        }.
      \end{equation}
  \end{itemize}

  Furthermore:
  \begin{itemize}
    \item
      $\tilde x^\mathrm{init} = \phi_\presentationSty{C}(x^\mathrm{init})$;

    \item
      for
      every
      $x_\mathrm{pre},x_\mathrm{cur}\in X$,
      every
      $u\in U$,
      and every
      $\theta_h\in\Theta_h$,
      \begin{equation}
        \label{eq:linear-rnn-output-conjugacy}
        \mbigfpinst{\tilde h}{
          \phi_\presentationSty{C}(x_\mathrm{pre}),
          u,
          \phi_\presentationSty{C}(x_\mathrm{cur})
        }{
          \theta_h
        }
        =
        \bigfpinst{h}{
          x_\mathrm{pre},
          u,
          x_\mathrm{cur}
        }{
          \theta_h
        }.
      \end{equation}
  \end{itemize}

  Consequently,
  $S$
  and
  $\tilde S$
  have the same input-output behaviour under the same parameter values;
  namely,
  \begin{equation*}
    \repval{\pinst{S}{\theta}}(u)
    \,=\,
    \repval{\pinst{\tilde S}{\theta}}(u)
    \qquad\qquad
    \forall
    \theta\in\Theta_f\times\Theta_h,
    \quad
    \forall
    u\in\allnonemptyseqs{U}.
  \end{equation*}
\end{closingproposition}
\begin{proof}
  We first prove the claims concerning the cascade dynamics.

  For every
  $i\in\iival{1}{\ell}$,
  let
  $X_i$
  and
  $\tilde X_i$
  denote the state spaces of the dynamics presented by
  $\presentationSty{L}_i$
  and
  $\presentationSty{\tilde L}_i$,
  respectively,
  and let
  $f_i$
  and
  $\tilde f_i$
  denote their respective parametric dynamics functions.

  By
  Proposition~\ref{prop:general-linear-unit-flattening},
  for every
  $j\in\iival{1}{m_i}$,
  the unit state flattening
  $\phi_{i,j}$
  maps the state space of
  $D_{i,j}$
  onto the state space of
  $\tilde D_{i,j}$
  and conjugates their parametric dynamics functions.
  Since parallel composition operates component-wise,
  it follows that
  \begin{equation}
    \label{eq:proof-linear-rnn-layer-state-space}
    \tilde X_i
    =
    \phi_i(X_i)
  \end{equation}
  and that
  \begin{equation}
    \label{eq:proof-linear-rnn-layer-conjugacy}
    \phi_i
    \left(
      \fpinst{f_i}{
        x_i,
        v
      }{
        \theta_i
      }
    \right)
    =
    \bigfpinst{\tilde f_i}{
      \phi_i(x_i),
      v
    }{
      \theta_i
    }
  \end{equation}
  for every admissible
  $x_i$,
  $v$,
  and corresponding layer parameter value
  $\theta_i$.

  We now prove the corresponding claims for the recurrent cascade
  by induction over its layers.

  For
  $i\in\iival{1}{\ell}$,
  let
  $\presentationSty{C}_{\leq i}$
  and
  $\presentationSty{\tilde C}_{\leq i}$
  denote, respectively,
  the prefixes through layer
  $i$
  of
  $\presentationSty{C}$
  and
  $\presentationSty{\tilde C}$.
  Furthermore, define
  \begin{equation*}
    \phi_{\leq i}
    \,\defeq\,
    \bigptuple{
      \phi_1,
      \ldots,
      \phi_i
    }.
  \end{equation*}
  Thus,
  for
  $i\geq2$,
  \begin{equation*}
    \phi_{<i}
    =
    \phi_{\leq i-1},
  \end{equation*}
  and
  \begin{equation*}
    \phi_{\leq\ell}
    =
    \phi_\presentationSty{C}.
  \end{equation*}

  We show inductively that,
  for every
  $i\in\iival{1}{\ell}$,
  the state space
  $\tilde X_{\leq i}$
  presented by
  $\presentationSty{\tilde C}_{\leq i}$
  satisfies
  \begin{equation}
    \label{eq:proof-linear-rnn-prefix-state-space}
    \tilde X_{\leq i}
    =
    \phi_{\leq i}(X_{\leq i}),
  \end{equation}
  where
  $X_{\leq i}$
  is the state space presented by
  $\presentationSty{C}_{\leq i}$,
  and that their parametric dynamics functions
  $f_{\leq i}$
  and
  $\tilde f_{\leq i}$
  satisfy
  \begin{equation}
    \label{eq:proof-linear-rnn-prefix-conjugacy}
    \phi_{\leq i}
    \left(
      \fpinst{f_{\leq i}}{
        x,
        u
      }{
        \theta_{\leq i}
      }
    \right)
    =
    \bigfpinst{\tilde f_{\leq i}}{
      \phi_{\leq i}(x),
      u
    }{
      \theta_{\leq i}
    }
  \end{equation}
  for every admissible state
  $x$,
  input
  $u$,
  and corresponding parameter value
  $\theta_{\leq i}$.

  For
  $i=1$,
  the two prefix presentations are
  \begin{equation*}
    \repinputprecomposition{
      g_1
    }{
      \presentationSty{L}_1
    }
    \qquad\text{and}\qquad
    \repinputprecomposition{
      g_1
    }{
      \presentationSty{\tilde L}_1
    }.
  \end{equation*}
  They use the same input-precomposition function
  $g_1$.
  Therefore
  Equations~\eqref{eq:proof-linear-rnn-prefix-state-space}
  and~\eqref{eq:proof-linear-rnn-prefix-conjugacy}
  for
  $i=1$
  follow directly from
  Equations~\eqref{eq:proof-linear-rnn-layer-state-space}
  and~\eqref{eq:proof-linear-rnn-layer-conjugacy}.

  Let now
  $i\in\iival{2}{\ell}$,
  and suppose that
  Equations~\eqref{eq:proof-linear-rnn-prefix-state-space}
  and~\eqref{eq:proof-linear-rnn-prefix-conjugacy}
  hold for
  $i-1$.

  Consider an arbitrary state of the original prefix through layer
  $i$,
  written as
  \begin{equation*}
    x
    =
    \ptuple{
      x_{<i},
      x_i
    },
  \end{equation*}
  where
  $x_{<i}\in X_{\leq i-1}$
  and
  $x_i\in X_i$.
  Its corresponding flattened state is
  \begin{equation*}
    \tilde x
    =
    \ptuple{
      \phi_{<i}(x_{<i}),
      \phi_i(x_i)
    }
    =
    \phi_{\leq i}(x).
  \end{equation*}

  Fix an input
  $u$,
  and let
  \begin{equation*}
    x_{<i}'
    \,\defeq\,
    \fpinst{f_{\leq i-1}}{
      x_{<i},
      u
    }{
      \theta_{<i}
    }
  \end{equation*}
  be the updated state of the original prefix through layer
  $i-1$.
  By the induction hypothesis,
  the corresponding updated flattened prefix state is
  \begin{equation*}
    \tilde x_{<i}'
    =
    \phi_{<i}(x_{<i}').
  \end{equation*}

  Let
  \begin{equation*}
    v_i
    \,\defeq\,
    \bigfpinst{g_i}{
      x_{<i},
      u,
      x_{<i}'
    }{
      \theta_{g_i}
    }
  \end{equation*}
  be the input presented to layer
  $i$
  in the original cascade.
  By
  Equation~\eqref{eq:flattened-connecting-function}
  and the preceding state correspondences,
  the input presented to the flattened layer
  $i$
  is
  \begin{align*}
    \bigfpinst{\tilde g_i}{
      \phi_{<i}(x_{<i}),
      u,
      \phi_{<i}(x_{<i}')
    }{
      \theta_{g_i}
    }
    &=
    \mBigfpinst{g_i}{
      \phi_{<i}^{-1}
      \big(
        \phi_{<i}(x_{<i})
      \big),
      u,
      \phi_{<i}^{-1}
      \big(
        \phi_{<i}(x_{<i}')
      \big)
    }{
      \theta_{g_i}
    }
    \\
    &=
    \bigfpinst{g_i}{
      x_{<i},
      u,
      x_{<i}'
    }{
      \theta_{g_i}
    }
    \\
    &=
    v_i.
  \end{align*}

  Hence the original and flattened versions of layer
  $i$
  receive the same input.
  Applying
  Equation~\eqref{eq:proof-linear-rnn-layer-conjugacy},
  their updated states satisfy
  \begin{equation*}
    \phi_i
    \left(
      \fpinst{f_i}{
        x_i,
        v_i
      }{
        \theta_i
      }
    \right)
    =
    \bigfpinst{\tilde f_i}{
      \phi_i(x_i),
      v_i
    }{
      \theta_i
    }.
  \end{equation*}

  Consequently,
  the complete updated states of the two prefixes correspond under
  \begin{equation*}
    \phi_{\leq i}
    =
    \bigptuple{
      \phi_{<i},
      \phi_i
    },
  \end{equation*}
  proving
  Equation~\eqref{eq:proof-linear-rnn-prefix-conjugacy}
  for
  $i$.
  Furthermore,
  by the induction hypothesis and
  Equation~\eqref{eq:proof-linear-rnn-layer-state-space},
  \begin{align*}
    \tilde X_{\leq i}
    &=
    \tilde X_{\leq i-1}
    \times
    \tilde X_i
    \\
    &=
    \phi_{<i}(X_{\leq i-1})
    \times
    \phi_i(X_i)
    \\
    &=
    \phi_{\leq i}
    \big(
      X_{\leq i-1}
      \times
      X_i
    \big)
    \\
    &=
    \phi_{\leq i}(X_{\leq i}),
  \end{align*}
  proving
  Equation~\eqref{eq:proof-linear-rnn-prefix-state-space}
  for
  $i$.

  The induction is therefore complete.
  Taking
  $i=\ell$,
  we obtain
  \begin{equation*}
    \tilde X
    =
    \phi_\presentationSty{C}(X),
  \end{equation*}
  together with
  \begin{equation*}
    \phi_\presentationSty{C}
    \left(
      \fpinst{f}{
        x,
        u
      }{
        \theta_f
      }
    \right)
    =
    \bigfpinst{\tilde f}{
      \phi_\presentationSty{C}(x),
      u
    }{
      \theta_f
    },
  \end{equation*}
  for every
  $x\in X$,
  $u\in U$,
  and
  $\theta_f\in\Theta_f$.
  This proves
  Equation~\eqref{eq:linear-rnn-whole-state-conjugacy}
  and the first state-space claim.

  We next consider the output extension.
  By
  Definition~\ref{def:canonical-linear-rnn},
  \begin{equation*}
    \tilde x^\mathrm{init}
    =
    \phi_\presentationSty{C}
    \big(
      x^\mathrm{init}
    \big),
  \end{equation*}
  proving the claimed relation between the initial states.
  Moreover,
  by
  Equation~\eqref{eq:flattened-rnc-output-function},
  for every
  $x_\mathrm{pre},x_\mathrm{cur}\in X$,
  every
  $u\in U$,
  and every
  $\theta_h\in\Theta_h$,
  \begin{align*}
    \mbigfpinst{\tilde h}{
      \phi_\presentationSty{C}(x_\mathrm{pre}),
      u,
      \phi_\presentationSty{C}(x_\mathrm{cur})
    }{
      \theta_h
    }
    &=
    \mBigfpinst{h}{
      \phi_\presentationSty{C}^{-1}
      \big(
        \phi_\presentationSty{C}(x_\mathrm{pre})
      \big),
      u,
      \phi_\presentationSty{C}^{-1}
      \big(
        \phi_\presentationSty{C}(x_\mathrm{cur})
      \big)
    }{
      \theta_h
    }
    \\
    &=
    \bigfpinst{h}{
      x_\mathrm{pre},
      u,
      x_\mathrm{cur}
    }{
      \theta_h
    }.
  \end{align*}
  This proves
  Equation~\eqref{eq:linear-rnn-output-conjugacy}.

  It remains to prove equality of the represented input--output behaviours.
  Fix arbitrary
  \begin{equation*}
    \theta
    =
    \ptuple{
      \theta_f,
      \theta_h
    }
    \in
    \Theta_f
    \times
    \Theta_h
  \end{equation*}
  and
  \begin{equation*}
    u
    =
    (u_t)_{t\in\ival{T}}
    \in
    \allnonemptyseqs{U}.
  \end{equation*}
  Let
  \begin{equation*}
    (x_t)_{t\in\iival{0}{T}}
    \qquad\text{and}\qquad
    (\tilde x_t)_{t\in\iival{0}{T}}
  \end{equation*}
  be the state sequences of
  $\pinst{S}{\theta}$
  and
  $\pinst{\tilde S}{\theta}$,
  respectively.

  We claim that
  \begin{equation}
    \label{eq:proof-linear-rnn-state-sequence-conjugacy}
    \tilde x_t
    =
    \phi_\presentationSty{C}(x_t)
    \qquad
    \forall
    t\in\iival{0}{T}.
  \end{equation}
  For
  $t=0$,
  this follows from
  \begin{equation*}
    \tilde x_0
    =
    \tilde x^\mathrm{init}
    =
    \phi_\presentationSty{C}
    \big(
      x^\mathrm{init}
    \big)
    =
    \phi_\presentationSty{C}(x_0).
  \end{equation*}
  Suppose now that
  Equation~\eqref{eq:proof-linear-rnn-state-sequence-conjugacy}
  holds at time
  $t-1$.
  Then,
  by
  Equation~\eqref{eq:linear-rnn-whole-state-conjugacy},
  \begin{align*}
    \tilde x_t
    &=
    \bigfpinst{\tilde f}{
      \tilde x_{t-1},
      u_t
    }{
      \theta_f
    }
    \\
    &=
    \bigfpinst{\tilde f}{
      \phi_\presentationSty{C}(x_{t-1}),
      u_t
    }{
      \theta_f
    }
    \\
    &=
    \phi_\presentationSty{C}
    \left(
      \fpinst{f}{
        x_{t-1},
        u_t
      }{
        \theta_f
      }
    \right)
    \\
    &=
    \phi_\presentationSty{C}(x_t).
  \end{align*}
  Hence
  Equation~\eqref{eq:proof-linear-rnn-state-sequence-conjugacy}
  follows by induction.

  Let
  $(y_t)_{t\in\ival{T}}$
  and
  $(\tilde y_t)_{t\in\ival{T}}$
  be the respective output sequences.
  By
  Equation~\eqref{eq:linear-rnn-output-conjugacy}
  and
  Equation~\eqref{eq:proof-linear-rnn-state-sequence-conjugacy},
  for every
  $t\in\ival{T}$,
  \begin{align*}
    \tilde y_t
    &=
    \mbigfpinst{\tilde h}{
      \tilde x_{t-1},
      u_t,
      \tilde x_t
    }{
      \theta_h
    }
    \\
    &=
    \mbigfpinst{\tilde h}{
      \phi_\presentationSty{C}(x_{t-1}),
      u_t,
      \phi_\presentationSty{C}(x_t)
    }{
      \theta_h
    }
    \\
    &=
    \bigfpinst{h}{
      x_{t-1},
      u_t,
      x_t
    }{
      \theta_h
    }
    \\
    &=
    y_t.
  \end{align*}

  Therefore,
  \begin{equation*}
    \repval{\pinst{S}{\theta}}(u)
    =
    \repval{\pinst{\tilde S}{\theta}}(u).
  \end{equation*}
  Since
  $\theta\in\Theta_f\times\Theta_h$
  and
  $u\in\allnonemptyseqs{U}$
  were arbitrary,
  the final claim follows.
\end{proof}

\begin{closingproposition}[The associated representation is a Linear RNN]
  \label{prop:linear-rnn-representation-is-linear-rnn}
  Let $S$ be a parametric RNC,
  and 
  let
  \begin{equation*}
    \presentationSty{S}
    \,\defeq\,
    \repoutoutextension{
    \presentationSty{C}
    }{
    x^\mathrm{init}
    }{
    h
    }
  \end{equation*}
  be a parametric analytic output extension presentation of $S$.
  Suppose that the parametric recurrent cascade presentation $\presentationSty{C}$
  is of the form
  \begin{equation*}
    \presentationSty{C}
    \,=\,
    \repinputprecomposition{
      g_1
    }{
      \presentationSty{L}_1
    }
    \repcascade_{g_2}
    \presentationSty{L}_2
    \repcascade_{g_3}
    \cdots
    \repcascade_{g_\ell}
    \presentationSty{L}_\ell
    \qquad\text{ where }\qquad
    \presentationSty{L}_i
    \,=\,
    D_{i,1}
    \replayer
    \cdots
    \replayer
    D_{i,m_i}
    \quad
    \forall
    i\in\iival{1}{\ell}
  \end{equation*}
  with
  each connecting function $g_i$ satisfying $g_i \in \analyticparamfuncclass$,
  and 
  each unit $D_{i,j}$ being a Parametric General Linear Recurrent Unit.
  Let
  \begin{equation*}
    \presentationSty{\tilde S}
    \,\defeq\,
    \repoutoutextension{
    \presentationSty{\tilde C}
    }{
    \tilde x^\mathrm{init}
    }{
    \tilde h
    }
  \end{equation*}
  be the Linear RNN presentation associated with
  $\presentationSty{S}$
  according to
  Definition~\ref{def:canonical-linear-rnn}.

  Then,
  the Linear RNN representation
  $\presentationVal{\presentationSty{\tilde S}}$
  is a Parametric Linear RNN.
\end{closingproposition}
\begin{proof}
  By
  Definition~\ref{def:linear-rnn-representation},
  for every
  $i\in\iival{1}{\ell}$
  and
  $j\in\iival{1}{m_i}$,
  the canonical Linear Recurrent Unit representation
  $\tilde D_{i,j}$
  of
  $D_{i,j}$
  is a Parametric Linear Recurrent Unit.
  Hence
  \begin{equation*}
    \tilde D_{i,j}
    \in
    \classpardynamics{Linear}
    \qquad
    \forall
    i\in\iival{1}{\ell},
    \quad
    \forall
    j\in\iival{1}{m_i}.
  \end{equation*}

  Therefore,
  for every
  $i\in\iival{1}{\ell}$,
  the flattened layer presentation
  \begin{equation*}
    \presentationSty{\tilde L}_i
    =
    \tilde D_{i,1}
    \replayer
    \cdots
    \replayer
    \tilde D_{i,m_i}
  \end{equation*}
  is a parametric recurrent layer presentation of type
  $\classpardynamics{Linear}$.

  We next consider the connecting functions.
  By assumption,
  \begin{equation*}
    g_i
    \in
    \analyticparamfuncclass
    \qquad
    \forall
    i\in\iival{1}{\ell}.
  \end{equation*}
  In particular,
  the first connecting function
  $g_1$
  of
  $\presentationSty{\tilde C}$
  is parametric analytic.

  For every
  $i\in\iival{2}{\ell}$,
  Definition~\ref{def:canonical-linear-rnn}
  gives
  \begin{equation*}
    \mBigfpinst{\tilde g_i}{
      \tilde x_\mathrm{pre},
      u,
      \tilde x_\mathrm{cur}
    }{
      \theta_{g_i}
    }
    =
    \mBigfpinst{g_i}{
      \phi_{<i}^{-1}
      \big(
        \tilde x_\mathrm{pre}
      \big),
      u,
      \phi_{<i}^{-1}
      \big(
        \tilde x_\mathrm{cur}
      \big)
    }{
      \theta_{g_i}
    }.
  \end{equation*}
  By
  Fact~\ref{fact:flattening-rnc-bijection},
  $\phi_{<i}^{-1}$
  is a linear map.
  Hence it is analytic.
  Consequently,
  $\tilde g_i$
  is obtained from the parametric analytic map
  $g_i$
  by composition with analytic coordinate maps and linear maps.
  By closure of parametric analytic Euclidean maps under finite tupling and composition,
  \begin{equation*}
    \tilde g_i
    \in
    \analyticparamfuncclass
    \qquad
    \forall
    i\in\iival{2}{\ell}.
  \end{equation*}

  It follows that
  \begin{equation*}
    \presentationSty{\tilde C}
    =
    \repinputprecomposition{
      g_1
    }{
      \presentationSty{\tilde L}_1
    }
    \repcascade_{\tilde g_2}
    \presentationSty{\tilde L}_2
    \repcascade_{\tilde g_3}
    \cdots
    \repcascade_{\tilde g_\ell}
    \presentationSty{\tilde L}_\ell
  \end{equation*}
  is a parametric recurrent
  $\analyticparamfuncclass$-cascade
  $\classpardynamics{Linear}$-presentation.

  We finally consider the output function.
  Since
  $\presentationSty{S}$
  is a parametric analytic output extension presentation,
  \begin{equation*}
    h
    \in
    \analyticparamfuncclass.
  \end{equation*}
  By
  Definition~\ref{def:canonical-linear-rnn},
  \begin{equation*}
    \mBigfpinst{\tilde h}{
      \tilde x_\mathrm{pre},
      u,
      \tilde x_\mathrm{cur}
    }{
      \theta_h
    }
    =
    \mBigfpinst{h}{
      \phi_\presentationSty{C}^{-1}
      \big(
        \tilde x_\mathrm{pre}
      \big),
      u,
      \phi_\presentationSty{C}^{-1}
      \big(
        \tilde x_\mathrm{cur}
      \big)
    }{
      \theta_h
    }.
  \end{equation*}
  By
  Fact~\ref{fact:flattening-rnc-bijection},
  $\phi_\presentationSty{C}^{-1}$
  is linear and hence analytic.
  Thus,
  again by closure under finite tupling and composition,
  \begin{equation*}
    \tilde h
    \in
    \analyticparamfuncclass.
  \end{equation*}

  Therefore,
  \begin{equation*}
    \presentationSty{\tilde S}
    =
    \repoutoutextension{
      \presentationSty{\tilde C}
    }{
      \tilde x^\mathrm{init}
    }{
      \tilde h
    }
  \end{equation*}
  is a parametric analytic output extension presentation
  of parametric recurrent cascade having type
  $\ptuple{\classpardynamics{Linear},\analyticparamfuncclass}$.
  Consequently,
  \begin{equation*}
    \tilde S
    =
    \presentationVal{
      \presentationSty{\tilde S}
    }
  \end{equation*}
  is a
  $\classpardynamics{Linear}$-RNC,
  that is,
  a Parametric Linear RNC,
  and hence a Parametric Linear RNN.
\end{proof}

\begin{closingremark}[Dependence on the chosen RNC presentation]
  \label{rem:linear-rnn-representation-presentation-dependence}
  The canonical Linear Recurrent Unit representation of
  Definition~\ref{def:linear-rnn-representation}
  is determined by the fixed flattening convention.

  At the level of a complete RNC, however,
  the construction of
  Definition~\ref{def:canonical-linear-rnn}
  is relative to a chosen output extension presentation
  $\presentationSty{S}$,
  and in particular to its underlying recurrent cascade presentation
  $\presentationSty{C}$.
  Indeed,
  the presentation
  $\presentationSty{C}$
  determines the layer and unit decomposition,
  the corresponding product decomposition and ordering of the state space,
  and therefore the presentation-induced state flattening
  $\phi_\presentationSty{C}$.
  It consequently also determines the flattened connecting functions,
  flattened output function,
  and the resulting Linear RNN presentation
  $\presentationSty{\tilde S}$.

  Thus,
  different presentations of the same parametric RNC
  may induce different Linear RNN representations.
  The construction is therefore canonical relative to the chosen presentation,
  but need not be canonical as a function of the presented parametric system alone.
\end{closingremark}

\subsection{Contractive Linear RNNs}

\begin{closingdefinition} \label{def:contractive}
  Euclidean dynamics $D=\tuple{X,U,f}$ are:
  \begin{itemize}
    \item
      \emph{contractive} if, for every $u\in U$, there exists
      $\alpha_u\in[0,1)$ such that
      \begin{equation*}
        \|f(x,u)-f(x',u)\|
        \,\leq\,
        \alpha_u \cdot \|x-x'\|
        \qquad\qquad 
        \forall\, x  \in X,\;
        \forall\, x' \in X;
      \end{equation*}
    \item
      \emph{contractive on bounded input sets (b-contractive)} if,
      for every $r\in\nonnegreals$, there exists
      $\alpha_r\in[0,1)$ such that
      \begin{equation*}
        \|f(x,u)-f(x',u)\|
        \,\leq\,
        \alpha_r \cdot \|x-x'\|
        \qquad\qquad 
        \forall\, u  \in \truncation{U}{r},\;
        \forall\, x  \in X,\; 
        \forall\, x' \in X\,; 
      \end{equation*}
    \item
      \emph{uniformly contractive (u-contractive)} if there exists
      $\alpha\in[0,1)$ such that
      \begin{equation*}
        \|f(x,u)-f(x',u)\|
        \,\leq\,
        \alpha \cdot \|x-x'\|
        \qquad\qquad 
        \forall\, u  \in U,\;
        \forall\, x  \in X,\; 
        \forall\, x' \in X\,.
      \end{equation*}
  \end{itemize}

  Then,
  \begin{itemize}
    \item
      a class $\mathcal{D}$ of dynamics is 
      \emph{contractive}, \emph{b-contractive}, or \emph{u-contractive} 
      if
      all dynamics $D \in \mathcal{D}$ are contractive, b-contractive, and u-contractive,
      respectively;
    \item
      parametric dynamics $D$ are  
      \emph{contractive}, \emph{b-contractive}, or \emph{u-contractive} 
      if
      all instances $\pinst{D}{\theta}$ are contractive, b-contractive, and u-contractive,
      respectively;
    \item
      a system $S$ is 
      \emph{contractive}, \emph{b-contractive}, or \emph{u-contractive} 
      if
      its dynamics $\dynop{S}$ are contractive, b-contractive, and u-contractive, respectively;
    \item
      a parametric system $S$ is 
      \emph{contractive}, \emph{b-contractive}, or \emph{u-contractive} 
      if
      all instances $\pinst{S}{\theta}$ are contractive, b-contractive, and u-contractive,
      respectively;
    \item
      a class $\mathcal{S}$ of systems is 
      \emph{contractive}, \emph{b-contractive}, or \emph{u-contractive} 
      if
      every system $S \in \mathcal{S}$ is contractive, b-contractive, and u-contractive,
      respectively.
  \end{itemize}
\end{closingdefinition}

\begin{proposition}
  Let $D=\mtuple{X,U,f}{A,B}$ be a Linear Recurrent Unit, with
  $X\subseteq\reals^{d_\mathrm{state}}$.
  Let us define
  \begin{equation*}
    \operatorname{Lip}_X(M)
    \defeq
    \sup_{\substack{x,x'\in X\\x\neq x'}}
    \frac{\|M \cdot x-M \cdot x'\|}{\|x-x'\|}
    \qquad\qquad
    \text{for }
    M\in\reals^{d_\mathrm{state}\times d_\mathrm{state}}.
  \end{equation*}
  Then:
  \begin{enumerate}
    \item
      $D$ is contractive if and only if
      \begin{equation*}
        \operatorname{Lip}_X(A(u))<1
        \qquad\qquad
        \forall u\in U;
      \end{equation*}
    \item
      $D$ is b-contractive if and only if,
      for every $r\in\nonnegreals$, there exists
      $\alpha_r\in[0,1)$ such that
      \begin{equation*}
        \sup_{u\in\truncation{U}{r}}
        \operatorname{Lip}_X(A(u))
        \leq
        \alpha_r;
      \end{equation*}
    \item
      $D$ is u-contractive if and only if there exists
      $\alpha\in[0,1)$ such that
      \begin{equation*}
        \sup_{u\in U}
        \operatorname{Lip}_X(A(u))
        \leq
        \alpha.
      \end{equation*}
  \end{enumerate}
\end{proposition}
\begin{proof}
  For every $x,x'\in X$ and $u\in U$,
  \begin{align*}
    f(x,u)-f(x',u)
    &=
    A(u)\cdot x+B(u)
    -
    A(u)\cdot x'-B(u)
    \\
    &=
    A(u)\cdot x-A(u)\cdot x'.
  \end{align*}
  Hence,
  \begin{align*}
    \sup_{\substack{x,x'\in X\\x\neq x'}}
    \frac{\|f(x,u)-f(x',u)\|}{\|x-x'\|}
    \;=\;
    \sup_{\substack{x,x'\in X\\x\neq x'}}
    \frac{\|A(u)\cdot x-A(u)\cdot x'\|}{\|x-x'\|}
    \;=\;
    \operatorname{Lip}_X(A(u)).
  \end{align*}
  Thus, for every $u\in U$,
  $\operatorname{Lip}_X(A(u))$ is precisely the Lipschitz constant
  of the state map $f(\cdot,u)$ on $X$.

  Therefore, $D$ is contractive if and only if
  \begin{equation*}
    \operatorname{Lip}_X(A(u))<1
    \qquad\qquad
    \forall u\in U,
  \end{equation*}
  proving the first claim.

  Similarly, $D$ is b-contractive if and only if,
  for every $r\in\nonnegreals$, the Lipschitz constants of the state
  maps $f(\cdot,u)$ admit a common bound $\alpha_r<1$ over
  $u\in\truncation{U}{r}$.
  By the equality above, this is equivalent to the existence of
  $\alpha_r\in[0,1)$ such that
  \begin{equation*}
    \sup_{u\in\truncation{U}{r}}
    \operatorname{Lip}_X(A(u))
    \leq
    \alpha_r.
  \end{equation*}

  Finally, $D$ is u-contractive if and only if the Lipschitz constants
  of the state maps admit a common bound $\alpha<1$ over all $u\in U$.
  Again, by the equality above, this is equivalent to
  \begin{equation*}
    \sup_{u\in U}
    \operatorname{Lip}_X(A(u))
    \leq
    \alpha
  \end{equation*}
  for some $\alpha\in[0,1)$.
\end{proof}

\begin{proposition}
  A Linear Recurrent Unit $D=\mtuple{X,U,f}{A,B}$ 
  having a full-Euclidean state space
  $X = \reals^{d_\mathrm{state}}$ 
  is:
  \begin{itemize}
    \item
      contractive if and only if
      \begin{equation*}
        \|A(u)\|<1
        \qquad\qquad
        \forall u\in U;
      \end{equation*}

    \item
      b-contractive if, for every $r\in\nonnegreals$,
      there exists $\alpha_r\in[0,1)$ such that
      \begin{equation*}
        \sup_{u\in\truncation{U}{r}}\|A(u)\|
        \leq\alpha_r;
      \end{equation*}

    \item
      u-contractive if there exists $\alpha\in[0,1)$ such that
      \begin{equation*}
        \sup_{u\in U}\|A(u)\|
        \leq\alpha.
      \end{equation*}
  \end{itemize}
\end{proposition}
\begin{proof}
  Since $D$ is a Linear Recurrent Unit,
  \begin{equation*}
    f(x,u)=A(u)\cdot x+B(u).
  \end{equation*}
  Therefore, for every $x,x'\in X$ and every $u\in U$,
  \begin{align*}
    \|f(x,u)-f(x',u)\|
    &=
    \|A(u)\cdot x+B(u)-A(u)\cdot x'-B(u)\|
    \\[.3em]
    &=
    \|A(u)\cdot(x-x')\|.
  \end{align*}

  Since $X=\reals^{d_\mathrm{state}}$, every nonzero vector
  $v\in\reals^{d_\mathrm{state}}$ can be written as
  $v=x-x'$ for some $x,x'\in X$.
  Hence, by the definition of the induced operator norm,
  \begin{align*}
    \sup_{\substack{x,x'\in X\\x\neq x'}}
    \frac{\|f(x,u)-f(x',u)\|}{\|x-x'\|}
    &=
    \sup_{\substack{x,x'\in X\\x\neq x'}}
    \frac{\|A(u)\cdot(x-x')\|}{\|x-x'\|}
    \\[.3em]
    &=
    \sup_{v\in\reals^{d_\mathrm{state}}\setminus\{\mathbf{0}\}}
    \frac{\|A(u)\cdot v\|}{\|v\|}
    \\[.3em]
    &=
    \|A(u)\|.
  \end{align*}
  Thus, for every $u\in U$, the Lipschitz constant of the state map
  $f(\cdot,u)$ is exactly $\|A(u)\|$.
  The three claims then follow directly from the definitions of
  contractivity, b-contractivity, and u-contractivity.
\end{proof}

\begin{closingdefinition}[Contractive Linear RNNs]
  \label{def:contractive-linear-rnn}
  A Linear RNN is 
  \emph{(unitwise-)contractive},
  \emph{(unitwise-)b-contractive},
  or
  \emph{(unitwise-)u-contractive}
  if
  all its units are
  contractive,
  b-contractive,
  and 
  u-contractive, respectively.
\end{closingdefinition}

\subsection{Smoothness up to the Closure}

\begin{closingdefinition}[Closure]
  \todo{}
  Let $X$ be a metric space having ambient space $\Omega_X$.
  The \emph{closure} of $X$ in $(\Omega_X)$, denoted by
    $\closureopp{X}{\Omega_X}$,
  is the set of all points $x\in \Omega_X$ such that every open ball centred at $x$
  intersects $X$; namely,
  \begin{equation*}
    \closureopp{X}{\Omega_X}
    \defeq
    \big\{
      x\in X
      \mid
      \forall\epsilon\in\posreals,\;
      \openballopp{x,\epsilon}{\Omega_X}\cap X\neq\emptyset
    \big\}.
  \end{equation*}

  Equivalently, $\closureopp{X}{\Omega_X}$ is the smallest closed subset of $\Omega_X$
  containing $X$.

  When the ambient space $\Omega_X$ is clear from the context,
  we simply write $\closureop{X}$ in place of $\closureopp{X}{\Omega_X}$.
\end{closingdefinition}

\begin{closingdefinition}[Continuous and differentiable extension to the closure]
  \label{def:closure-regularity}
  Let $Z\subseteq\reals^d$ and $Y\subseteq\reals^q$.

  A function $g:Z\to Y$ is:
  \begin{itemize}
    \item
      \emph{continuous up to the closure of its domain}
      if there exists a continuous function
        $\widetilde g:
        \closureop{Z}
        \to
        \reals^q$
        such that
        $\restr{\widetilde g}{Z} = g$;

    \item
      \emph{continuously-differentiable up to the closure of its domain}
      if there exist an open set
        $O\subseteq\reals^d$
        with
        $\closureop{Z}\subseteq O$
      and a continuously-differentiable function
        $\widetilde g:
        O
        \to
        \reals^q$
        such that
        $\restr{\widetilde g}{Z} = g$.
  \end{itemize}
\end{closingdefinition}

\begin{lemma}[Boundedness on bounded subsets]
  \label{lemma:closure-continuous-bounded}
  Every function between finite-dimensional Euclidean spaces that is
  continuous up to the closure of its domain is boundedness-preserving.
\end{lemma}
\begin{proof}
  Let $g:Z\to Y$ be continuous up to the closure of its domain,
  and let
  $K\subseteq Z$
  be bounded.

  The closure $\closureop K$ is closed and bounded in $\reals^d$, and hence compact.
  Also, it holds that $\closureop K \subseteq \closureop Z$.

  Let $\widetilde g: \closureop Z \to \reals^q$ be a continuous extension of $g$.
  Then $\widetilde g(\closureop K)$
  is compact and therefore bounded.

  Since $g(K) \subseteq \widetilde g(\closureop K)$, we conclude that $g(K)$ is bounded.
\end{proof}

\begin{lemma}[Bounded derivatives on bounded subsets]
  \label{lemma:closure-cd-bounded-derivative}
  Let $g:Z\to Y$ 
  be continuously-differentiable up to the closure of its domain.
  Then, for every bounded
  $K\subseteq Z$,
  there exists
  $M_K\in\nonnegreals$
  such that
  \begin{equation*}
    \|\dvop{g}(z)\|
    \leq
    M_K
    \qquad\qquad
    \forall z\in K.
  \end{equation*}
\end{lemma}
\begin{proof}
  Let
    $\widetilde g:O\to\reals^q$
  be a continuously-differentiable extension of $g$ to an open
  neighbourhood $O$ of $\closureop Z$.

  Since $K$ is bounded,
  $\closureop K$ is compact.
  The map
  \begin{equation*}
    z
    \mapsto
    \|\dvop{\widetilde g}(z)\|
  \end{equation*}
  is continuous and therefore attains a finite maximum on
  $\closureop K$.
  The claim follows by restriction to $K$.
\end{proof}

\begin{lemma}[Uniform strictness on bounded subsets]
  \label{lemma:closure-strict-uniform}
  Let $A: Z \to \reals^{d\times d}$ be continuous up to the closure of its domain,
  and 
  let $\widetilde A: \closureop Z \to \reals^{d\times d}$
  be a continuous extension satisfying
  \begin{equation*}
    \|\widetilde A(z)\|
    <
    1
    \qquad\qquad
    \forall z\in\closureop Z.
  \end{equation*}

  Then, for every bounded set
  $K\subseteq Z$,
  there exists
  $\alpha_K\in[0,1)$
  such that
  \begin{equation*}
    \|A(z)\|
    \leq
    \alpha_K
    \qquad\qquad
    \forall z\in K.
  \end{equation*}
\end{lemma}
\begin{proof}
  Since $K$ is bounded,
  we have that
  $\closureop K$ is compact and $\closureop K \subseteq \closureop Z$.

  The function $z \mapsto \|\widetilde A(z)\|$
  is continuous and therefore attains its maximum on
  $\closureop K$.
  Let us define
  \begin{equation*}
    \alpha_K
    \defeq
    \max_{z\in\closureop K}
    \|\widetilde A(z)\|.
  \end{equation*}

  By hypothesis,
  every value of
  $\|\widetilde A(z)\|$
  on $\closureop K$
  is strictly smaller than one.
  It follows that $\alpha_K<1$,
  and hence the claim.
\end{proof}

\subsubsection{Smooth Unitwise-contractive Linear RNNs}

\begin{closingdefinition}[Smooth unitwise-contractive Linear RNNs]
  \label{def:smooth-contractive-linear-rnn}
  A parametric Linear RNN $S$ is
  \emph{smooth (unitwise-)contractive}
  if the following conditions hold:
  \begin{itemize}
    \item
      for every constituent parametric Linear Recurrent Unit
      \begin{equation*}
        D_{i,j}
        =
        \mctuple{
        X_{i,j},
        U_i,
        f_{i,j}
        }{
        \Theta^f_{i,j}
        }{
        A_{i,j},
        B_{i,j}
        },
      \end{equation*}
      with 
      parametric input functions
      \begin{align*}
        A_{i,j}
        :
        U_i\times\Theta^f_{i,j}
        \to
        \reals^{d_{i,j}\times d_{i,j}},
        B_{i,j}
        :
        U_i\times\Theta^f_{i,j}
        \to
        \reals^{d_{i,j}},
      \end{align*}
      the following two conditions hold jointly:
      \begin{enumerate}
        \item
          the input functions $A_{i,j}$ and $B_{i,j}$ are continuously-differentiable up to the
          closures of their domains;
        \item
          the parametric input function $A_{i,j}$ admits an extension
          \begin{equation*}
            \widetilde A_{i,j}
            :
            \closureop{
            U_i\times\Theta^f_{i,j}
            }
            \to
            \reals^{d_{i,j}\times d_{i,j}}
          \end{equation*}
          satisfying
          \begin{equation}
            \label{eq:closure-contraction}
            \|
            \widetilde A_{i,j}(v,\theta)
            \|
            <
            1
            \qquad\qquad
            \forall
            \ptuple{v,\theta}
            \in
            \closureop{
            U_i\times\Theta^f_{i,j}
            };
          \end{equation}
      \end{enumerate}
    \item
      every layer's output function $h_i$
      is continuously-differentiable up to the closure of its domain.
  \end{itemize}
\end{closingdefinition}

\begin{closingremark}
  Condition~\eqref{eq:closure-contraction} implies that every constituent
  Linear Recurrent Unit is contractive.
  Indeed, for every
  $x,x'\in X_{i,j}$,
  $u\in U_i$,
  and
  $\theta\in\Theta^f_{i,j}$,
  \begin{align*}
    \|
      \fpinst{f_{i,j}}{x,u}{\theta}
      -
      \fpinst{f_{i,j}}{x',u}{\theta}
    \|
    &=
    \|
      \fpinst{A_{i,j}}{u}{\theta}
      \cdot
      (x-x')
    \|
    \\
    &\leq
    \|
      \fpinst{A_{i,j}}{u}{\theta}
    \|
    \cdot
    \|x-x'\|
    \\
    &<
    \|x-x'\|.
  \end{align*}
\end{closingremark}

\begin{closingremark}[Why contraction on the closure is required]
  It is not sufficient to assume only that
  \begin{equation*}
    \|A(u)\|<1
    \qquad
    \forall u\in U
  \end{equation*}
  together with continuous extendability.

  For example, for
  \begin{equation*}
    U=(0,1)
    \qquad\text{and}\qquad
    A(u)=u,
  \end{equation*}
  $A$ extends continuously to $[0,1]$ and satisfies
  $|A(u)|<1$ for every $u\in U$, but
  \begin{equation*}
    \sup_{u\in U}|A(u)|
    =
    1.
  \end{equation*}

  Requiring strict contraction also on the closure rules out this
  boundary degeneration.
\end{closingremark}

\begin{closinglemma}[Bounded reachability of smooth contractive Linear RNNs]
  \label{lemma:contractive-bounded-reachability}
  Let $S$ be a smooth contractive parametric Linear RNN.

  Let
  \begin{equation*}
    K_X\subseteq X,
    \qquad
    K_U\subseteq U,
    \qquad
    K_\Theta\subseteq\Theta
  \end{equation*}
  be bounded sets.

  Then there exists
  $R\in\nonnegreals$
  such that, for every
  \begin{equation*}
    x_0\in K_X,
    \qquad
    \theta\in K_\Theta,
    \qquad
    u\in\infseqs{K_U},
  \end{equation*}
  the corresponding state sequence satisfies
  \begin{equation*}
    \|x_t^{x_0,u,\theta}\|
    \leq
    R
    \qquad\qquad
    \forall t\in\naturals.
  \end{equation*}

  Furthermore, the state, input, and output sequences of every constituent
  layer are uniformly bounded over these choices.
\end{closinglemma}
\begin{proof}
  Let
  \begin{equation*}
    \dynop S
    =
    S_1
    \ltimes
    \cdots
    \ltimes
    S_m
  \end{equation*}
  be the cascade decomposition of the dynamics of $S$.
  We proceed inductively over its layers.

  Consider first layer $S_1$.
  Its input set $K_U$ is bounded.

  For every constituent Linear Recurrent Unit
  $D_{1,j}$,
  let
  \begin{equation*}
    K_{\Theta^f_{1,j}}
  \end{equation*}
  denote the projection of $K_\Theta$ onto
  $\Theta^f_{1,j}$.
  This set is bounded.

  Hence
  \begin{equation*}
    K_U
    \times
    K_{\Theta^f_{1,j}}
  \end{equation*}
  is bounded.
  By
  Lemma~\ref{lemma:closure-strict-uniform},
  there exists
  $\alpha_{1,j}\in[0,1)$
  such that
  \begin{equation*}
    \|
      \fpinst{A_{1,j}}{v}{\theta}
    \|
    \leq
    \alpha_{1,j}
  \end{equation*}
  for every
  \begin{equation*}
    v\in K_U,
    \qquad
    \theta\in K_{\Theta^f_{1,j}}.
  \end{equation*}

  Moreover,
  by
  Lemma~\ref{lemma:closure-continuous-bounded},
  there exists
  $\beta_{1,j}\in\nonnegreals$
  such that
  \begin{equation*}
    \|
      \fpinst{B_{1,j}}{v}{\theta}
    \|
    \leq
    \beta_{1,j}
  \end{equation*}
  on the same bounded set.

  Since $K_X$ is bounded, the initial-state component associated with
  $D_{1,j}$ is uniformly bounded over $x_0\in K_X$.
  Thus there exists
  $R^\mathrm{init}_{1,j}\in\nonnegreals$
  such that
  \begin{equation*}
    \|x_{1,j,0}\|
    \leq
    R^\mathrm{init}_{1,j}.
  \end{equation*}

  Therefore,
  \begin{align*}
    \|x_{1,j,t}\|
    &\leq
    \alpha_{1,j}
    \cdot
    \|x_{1,j,t-1}\|
    +
    \beta_{1,j}
    \\
    &\leq
    R^\mathrm{init}_{1,j}
    +
    \frac{
      \beta_{1,j}
    }{
      1-\alpha_{1,j}
    }
    \qquad
    \forall t\in\naturals.
  \end{align*}

  Hence the state sequence of every unit of the first layer is
  uniformly bounded.
  Since the number of units is finite, the state sequence of $S_1$ is
  uniformly bounded.

  The arguments supplied to the output function $h_1$ consist of:
  \begin{itemize}
    \item
      the previous state of $S_1$;
    \item
      the input to $S_1$;
    \item
      the updated state of $S_1$;
    \item
      the corresponding output-function parameters.
  \end{itemize}
  All these quantities range over bounded sets.

  Since $h_1$ is continuously-differentiable up to the closure of its
  domain, it is in particular continuous up to the closure.
  Hence, by
  Lemma~\ref{lemma:closure-continuous-bounded},
  the output sequence of $S_1$ is uniformly bounded.

  Now let
  $i\in\iival{2}{m}$
  and suppose inductively that the output sequence of
  $S_{i-1}$ is uniformly bounded.
  By the definition of cascade composition, this output sequence is the
  input sequence of $S_i$.
  Therefore the input sequence of $S_i$ ranges over a bounded set.

  Repeating the preceding argument for every unit
  $D_{i,j}$ gives:
  \begin{itemize}
    \item
      a uniform contraction constant
      $\alpha_{i,j}<1$ on the bounded set of reachable layer inputs and
      parameters;
    \item
      a uniform bound
      $\beta_{i,j}$ on the corresponding additive terms;
    \item
      consequently, a uniform bound on every unit state;
    \item
      consequently, a uniform bound on the state of $S_i$;
    \item
      by closure-continuity of $h_i$, a uniform bound on its output.
  \end{itemize}

  Thus the induction holds for every layer.

  Since the complete state of the Linear RNN is the finite tuple of the
  states of its layers, there exists
  $R\in\nonnegreals$
  such that
  \begin{equation*}
    \|x_t^{x_0,u,\theta}\|
    \leq
    R
  \end{equation*}
  uniformly over
  $x_0\in K_X$,
  $\theta\in K_\Theta$,
  $u\in\infseqs{K_U}$,
  and
  $t\in\naturals$.
\end{proof}

\begin{closinglemma}[State-transition decay on bounded sets]
  \label{lemma:contractive-state-transition}
  Let $S$ be a smooth contractive Linear RNN.

  For every bounded
  \begin{equation*}
    K_X\subseteq X,
    \qquad
    K_U\subseteq U,
    \qquad
    K_\Theta\subseteq\Theta,
  \end{equation*}
  there exist constants
  \begin{equation*}
    C\in\posreals,
    \qquad
    \rho\in[0,1)
  \end{equation*}
  such that
  \begin{equation*}
    \left\|
      J_t^{x_0,u,\theta}
      \cdot
      J_{t-1}^{x_0,u,\theta}
      \cdots
      J_{s+1}^{x_0,u,\theta}
    \right\|
    \leq
    C\cdot\rho^{t-s}
  \end{equation*}
  for every
  \begin{equation*}
    x_0\in K_X,
    \qquad
    \theta\in K_\Theta,
    \qquad
    u\in\infseqs{K_U},
  \end{equation*}
  and every
  $s,t\in\naturals$
  satisfying
  $s<t$.
\end{closinglemma}
\begin{proof}
  By
  Lemma~\ref{lemma:contractive-bounded-reachability},
  all layer states and layer inputs reachable from the considered
  bounded sets are uniformly bounded.

  For every unit $D_{i,j}$,
  its input and parameter arguments therefore range over a bounded set.
  By
  Lemma~\ref{lemma:closure-strict-uniform},
  there exists
  $\alpha_{i,j}<1$
  such that
  \begin{equation*}
    \|
      \fpinst{A_{i,j}}{v}{\theta}
    \|
    \leq
    \alpha_{i,j}
  \end{equation*}
  uniformly over all reachable arguments.

  Since there are finitely many units, we have
  \begin{equation*}
    \alpha
    \defeq
    \max_{\substack{
      i\in\iival{1}{m}\\
      j\in\iival{1}{n_i}
    }}
    \alpha_{i,j},
  \end{equation*}
  satisfying $\alpha<1$.

  The one-step state Jacobian is block lower triangular with respect to
  the cascade layers:
  \begin{equation*}
    J_t
    \,=\,
    \begin{pmatrix}
      A_{1,t}
      & 0
      & \cdots
      & 0
      \\
      C_{2,1,t}
      & A_{2,t}
      & \cdots
      & 0
      \\
      \vdots
      & \vdots
      & \ddots
      & \vdots
      \\
      C_{m,1,t}
      & C_{m,2,t}
      & \cdots
      & A_{m,t}
    \end{pmatrix}.
  \end{equation*}

  Every diagonal block $A_{i,t}$ is block diagonal with blocks given by
  the state-update matrices of the units in layer $i$.
  Hence
  \begin{equation*}
    \|A_{i,t}\|
    \leq
    \alpha
    \qquad
    \forall i,t.
  \end{equation*}

  The off-diagonal blocks $C_{i,j,t}$ are finite combinations of:
  \begin{itemize}
    \item
      values and derivatives of the functions $A_{i,j}$ and $B_{i,j}$;
    \item
      values and derivatives of the output functions $h_i$;
    \item
      reachable state and input values.
  \end{itemize}

  All their arguments range over bounded sets.
  Since all the relevant functions are continuously-differentiable up
  to the closures of their domains,
  Lemma~\ref{lemma:closure-cd-bounded-derivative}
  and
  Lemma~\ref{lemma:closure-continuous-bounded}
  imply that these quantities are uniformly bounded.

  Since the architecture is finite, there exists
  $M\in\nonnegreals$
  such that
  \begin{equation*}
    \|C_{i,j,t}\|
    \leq
    M
    \qquad
    \forall j<i,\;
    \forall t.
  \end{equation*}

  Fix any
  \begin{equation*}
    \rho\in(\alpha,1).
  \end{equation*}

  Let $h$ be an arbitrary state perturbation and define
  \begin{equation*}
    z_\ell
    \defeq
    J_{s+\ell}
    \cdots
    J_{s+1}
    \cdot h,
    \qquad
    z_0\defeq h.
  \end{equation*}
  Write
  \begin{equation*}
    z_\ell
    =
    \tuple{
      z_\ell^{(1)},
      \ldots,
      z_\ell^{(m)}
    }
  \end{equation*}
  according to the cascade layers.

  For the first layer,
  \begin{equation*}
    z_\ell^{(1)}
    =
    A_{1,s+\ell}
    \cdot
    z_{\ell-1}^{(1)},
  \end{equation*}
  and hence
  \begin{equation*}
    \|z_\ell^{(1)}\|
    \leq
    \alpha^\ell
    \|h\|
    \leq
    \rho^\ell
    \|h\|.
  \end{equation*}

  Suppose inductively that, for every $j<i$,
  there exists $K_j\in\posreals$ such that
  \begin{equation*}
    \|z_\ell^{(j)}\|
    \leq
    K_j
    \rho^\ell
    \|h\|.
  \end{equation*}

  Then
  \begin{equation*}
    z_\ell^{(i)}
    =
    A_{i,s+\ell}
    z_{\ell-1}^{(i)}
    +
    \sum_{j<i}
    C_{i,j,s+\ell}
    z_{\ell-1}^{(j)}.
  \end{equation*}
  Consequently,
  \begin{align*}
    \|z_\ell^{(i)}\|
    &\leq
    \alpha
    \|z_{\ell-1}^{(i)}\|
    +
    M
    \sum_{j<i}
    K_j
    \rho^{\ell-1}
    \|h\|.
  \end{align*}

  Iterating gives
  \begin{align*}
    \|z_\ell^{(i)}\|
    &\leq
    \alpha^\ell\|h\|
    +
    M
    \sum_{j<i}K_j
    \sum_{q=1}^{\ell}
    \alpha^{\ell-q}
    \rho^{q-1}
    \|h\|
    \\
    &=
    \alpha^\ell\|h\|
    +
    M
    \sum_{j<i}K_j
    \frac{
      \rho^\ell-\alpha^\ell
    }{
      \rho-\alpha
    }
    \|h\|
    \\
    &\leq
    \left(
      1
      +
      \frac{
        M\sum_{j<i}K_j
      }{
        \rho-\alpha
      }
    \right)
    \rho^\ell
    \|h\|.
  \end{align*}

  Thus the induction holds for
  \begin{equation*}
    K_i
    \defeq
    1
    +
    \frac{
      M\sum_{j<i}K_j
    }{
      \rho-\alpha
    }.
  \end{equation*}

  Since the number of layers is finite,
  \begin{equation*}
    C
    \defeq
    \sqrt{
      \sum_{i=1}^mK_i^2
    }
  \end{equation*}
  is finite and
  \begin{equation*}
    \|z_\ell\|
    \leq
    C\rho^\ell\|h\|.
  \end{equation*}

  Taking
  $\ell=t-s$
  and then the supremum over
  $h\neq0$
  proves the claim.
\end{proof}

\subsection{Positive-Margin Expressivity of Smooth Contractive Linear RNNs}
\label{subsec:positive-margin-smooth-contractive-linear-rnns}

The next lemma isolates the analytic property used by the expressivity proof.
For a finite input alphabet, smooth unit-wise contractivity
(Definition~\ref{def:smooth-contractive-linear-rnn}) implies that two
computations processing the same sufficiently long suffix forget their
possibly different preceding prefixes uniformly.

\begin{closinglemma}[Uniform suffix forgetting]
  \label{lemma:positive-margin-contractive-uniform-suffix-forgetting}
  Let
  $S$
  be an instance of a smooth contractive Linear RNN in the sense of
  Definition~\ref{def:smooth-contractive-linear-rnn},
  and let $\enc : \Sigma \to U$
  be an encoder from a finite alphabet
  $\Sigma$.
  Then, for some sequence
  \begin{equation*}
    (\varepsilon_n)_{n\in\posnaturals}
    \in
    \infseqs{\nonnegreals}
    \qquad\quad\text{with}\qquad\quad
    \varepsilon_n \to 0 \quad\text{as}\quad n\to\infty,
  \end{equation*}
  we have:
  \begin{equation}
    \label{eq:positive-margin-suffix-forgetting-bound}
    \qquad
    d_Y
    \mBig(
      S\big(\enc(vz)\big),\,
      S\big(\enc(wz)\big)
    \mBig)
    \,\leq\,
    \varepsilon_n
    \qquad\quad
    \forall n\in\posnaturals,\;
    \forall v,w\in\Sigma^*,\;
    \forall z\in\Sigma^n.
  \end{equation}
\end{closinglemma}
\begin{proof}
  Since
  $\Sigma$
  is finite,
  the encoded input set $\enc(\Sigma)$ is bounded.
  Applying
  Lemma~\ref{lemma:contractive-bounded-reachability} \todo{}
  with the fixed initial states and fixed parameter values of
  $S$
  shows that the input, state, and output sequences of every constituent
  layer are uniformly bounded over all encoded words.

  For every constituent recurrent unit
  $D_{i,j}$,
  the set of input-parameter pairs encountered in these computations is
  bounded.
  Hence the strict contraction-on-the-closure requirement of
  Definition~\ref{def:smooth-contractive-linear-rnn},
  together with
  Lemma~\ref{lemma:closure-strict-uniform},
  yields a constant
  \begin{equation*}
    \alpha_{i,j}
    \in
    [0,1)
  \end{equation*}
  satisfying
  \begin{equation}
    \label{eq:positive-margin-smooth-uniform-contraction}
    \|A_{i,j}(u)\|
    \leq
    \alpha_{i,j}
  \end{equation}
  for every layer input
  $u$
  occurring while processing encoded words.
  Furthermore, continuous differentiability up to the closure implies that
  each coefficient function
  $A_{i,j}$
  and
  $B_{i,j}$,
  and each layer output function
  $h_i$,
  is uniformly continuous on the closure of every bounded set of arguments
  encountered in these computations.

  Consider now two computations that have processed arbitrary prefixes
  $v,w\in\Sigma^*$
  and subsequently process the same suffix
  \begin{equation*}
    z
    =
    z_1\cdots z_n.
  \end{equation*}
  We prove inductively over the layers that the corresponding layer states
  and layer outputs converge uniformly to one another as the length of the
  common suffix increases.

  Consider first a recurrent unit
  $D_{1,j}$
  in the first layer.
  During the common suffix the two computations receive exactly the same
  input
  $\enc(z_t)$
  at every time
  $t$.
  If
  $x_t$
  and
  $x'_t$
  denote the corresponding unit states after the first
  $t$
  symbols of the common suffix, then
  Equation~\eqref{eq:positive-margin-smooth-uniform-contraction}
  gives
  \begin{align*}
    \|x_t-x'_t\|
    \;\leq\;
    \alpha_{1,j}
    \cdot
    \|x_{t-1}-x'_{t-1}\|
    \;\leq\;
    \alpha_{1,j}^t
    \cdot
    \|x_0-x'_0\|.
  \end{align*}
  Uniform boundedness of the reachable states therefore implies that this
  difference converges uniformly to zero over the prefixes
  $v,w$
  and common suffixes.
  Since the first layer contains finitely many recurrent units, the same
  conclusion holds for its complete state.
  Uniform continuity of
  $h_1$
  on the corresponding bounded argument set then implies uniform convergence
  of the first-layer outputs.

  Let now
  $i\in\iival{2}{m}$,
  and suppose inductively that the outputs of layer
  $S_{i-1}$
  in the two computations converge uniformly to one another during an
  increasingly long common suffix.
  These outputs are the inputs
  $u_t$
  and
  $u'_t$
  to layer
  $S_i$,
  and therefore
  \begin{equation}
    \label{eq:positive-margin-smooth-layer-input-convergence}
    \|u_t-u'_t\|
    \to
    0
  \end{equation}
  uniformly over the prefixes and common suffixes as
  $t\to\infty$.

  Consider a recurrent unit
  $D_{i,j}$
  in this layer and write
  \begin{align*}
    x_t
    &
    \,=\,
    A_{i,j}(u_t)\cdot x_{t-1}
    +
    B_{i,j}(u_t),
    \\[.3em]
    x'_t
    &
    \,=\,
    A_{i,j}(u'_t) \cdot x'_{t-1}
    +
    B_{i,j}(u'_t).
  \end{align*}
  Subtracting gives
  \begin{align}
    \label{eq:positive-margin-smooth-layer-difference}
    x_t-x'_t
    = & \
    A_{i,j}(u_t)
    \cdot
    (x_{t-1}-x'_{t-1})
    \\[.2em]
    & 
    +
    \big(
      A_{i,j}(u_t)-A_{i,j}(u'_t)
    \big) 
    \cdot x'_{t-1}
    \\[.2em]
    & 
    +
    B_{i,j}(u_t)-B_{i,j}(u'_t).
  \end{align}

  The states
  $x'_t$
  are uniformly bounded.
  Together with
  Equation~\eqref{eq:positive-margin-smooth-layer-input-convergence}
  and the uniform continuity of
  $A_{i,j}$
  and
  $B_{i,j}$
  on the relevant bounded argument sets,
  Equation~\eqref{eq:positive-margin-smooth-layer-difference}
  yields a sequence
  $(e_t)_{t\in\posnaturals}$
  satisfying
  \begin{equation*}
    e_t
    \to
    0
  \end{equation*}
  uniformly over prefixes and common suffixes, such that
  \begin{equation}
    \label{eq:positive-margin-smooth-stable-forced-recurrence}
    \|x_t-x'_t\|
    \,\leq\,
    \alpha_{i,j}
    \cdot
    \|x_{t-1}-x'_{t-1}\|
    +
    e_t.
  \end{equation}
  Iterating gives
  \begin{equation*}
    \|x_t-x'_t\|
    \,\leq\,
    \alpha_{i,j}^t
    \cdot
    \|x_0-x'_0\|
    +
    \sum_{s=1}^t
    \alpha_{i,j}^{t-s} \cdot e_s.
  \end{equation*}
  The first term converges uniformly to zero by boundedness of the states.
  Since
  $\alpha_{i,j}<1$
  and
  $e_s\to0$
  uniformly, the convolution term also converges uniformly to zero.
  Hence the unit states converge uniformly to one another.
  Since the layer contains finitely many units, the same holds for its full
  state, and uniform continuity of
  $h_i$
  implies uniform convergence of the layer outputs.

  Proceeding inductively through the finitely many layers yields
  \begin{equation*}
    \sup_{\substack{
      v,w\in\Sigma^*
      \\
      z\in\Sigma^n
    }}
    d_Y
    \big(
      S(\enc(vz)),
      S(\enc(wz))
    \big)
    \to 0 
    \quad\text{as}\quad 
    n\to\infty.
  \end{equation*}
  Defining this supremum to be
  $\varepsilon_n$
  proves the claim.
\end{proof}

\begin{closingtheorem}[Positive-margin expressivity of smooth contractive Linear RNNs]
  \label{thm:positive-margin-smooth-contractive-linear-rnn-expressivity}
  The positive-margin expressivity of smooth contractive Linear RNNs is
  contained in the definite languages.
\end{closingtheorem}
\begin{proof}
  Let
  $S$
  be a smooth contractive Linear RNN that recognises a language
  $L$
  over a finite alphabet
  $\Sigma$
  with positive margin through an encoder
  $\enc$
  and decoder
  $\dec$.
  Let
  $\gamma\in\posreals$
  be a corresponding decoder margin.

  By
  Lemma~\ref{lemma:positive-margin-contractive-uniform-suffix-forgetting},
  choose
  $N\in\posnaturals$
  such that
  \begin{equation*}
    \varepsilon_N
    <
    \gamma.
  \end{equation*}
  Then, for every
  $v,w\in\Sigma^*$
  and
  $z\in\Sigma^N$,
  \begin{equation*}
    d_Y
    \big(
      S(\enc(vz)),
      S(\enc(wz))
    \big)
    <
    \gamma.
  \end{equation*}
  By the decoder-margin condition,
  \begin{equation*}
    \dec
    \big(
      S(\enc(vz))
    \big)
    =
    \dec
    \big(
      S(\enc(wz))
    \big),
  \end{equation*}
  and hence
  \begin{equation*}
    vz\in L
    \quad\Longleftrightarrow\quad
    wz\in L.
  \end{equation*}
  Lemma~\ref{lemma:positive-margin-definite-suffix-characterisation}
  therefore implies that
  $L$
  is definite.
\end{proof}

\subsection{BIBS of Smooth Contractive Linear RNNs}

\begin{closingtheorem}
  \label{thm:contractive-bibs}
  Every smooth contractive Linear RNN satisfies BIBS.
\end{closingtheorem}
\begin{proof}
  Let us consider a smooth contractive Linear RNN
  \begin{equation*}
    S
    =
    \ctuple{
      X,U,f,x^\mathrm{init},Y,h
    }{
      \Theta^f,\Theta^h
    }.
  \end{equation*}

  Let us fix arbitrary
  \begin{equation*}
    \theta\in\Theta,
    \qquad
    x_0\in X,
    \qquad
    u=(u_t)_{t\in\timedomain}\in\binfseqs{U}.
  \end{equation*}

  Let
  \begin{equation*}
    K_X
    \defeq
    \{x_0\},
    \qquad
    K_\Theta
    \defeq
    \{\theta\},
    \qquad
    K_U
    \defeq
    \{u_t\mid t\in\timedomain\}.
  \end{equation*}
  These sets are bounded.

  Hence,
  by
  Lemma~\ref{lemma:contractive-bounded-reachability},
  there exists
  $B\in\nonnegreals$
  such that
  \begin{equation*}
    \big\|
      x_t^{x_0,u,\theta}
    \big\|
    \leq
    B
    \qquad
    \forall t\in\naturals.
  \end{equation*}

  Therefore
  \begin{equation*}
    \sup_{t\in\timedomain}
    \big\|
      x_t^{x_0,u,\theta}
    \big\|
    \leq
    B.
  \end{equation*}

  Since
  $\theta$,
  $x_0$,
  and
  $u$
  were arbitrary,
  $S$ satisfies BIBS.
\end{proof}

\subsection{BIBSD and nnVSD of Smooth Contractive Linear RNNs}

\begin{closingtheorem}
  \label{thm:contractive-bibsd}
  Every smooth contractive Linear RNN satisfies BIBSD.
\end{closingtheorem}
\begin{proof}
  Let $r\in\nonnegreals$,
  let $x_0\in\truncation{X}{r}$,
  and 
  let $u = (u_t)_{t\in\timedomain} \in \infseqs{\truncation{U}{r}}$ 
  be arbitrary.

  Since the parametric input functions
  $A_{i,j}$
  and
  $B_{i,j}$
  of every Linear Recurrent Unit
  and the output functions
  $h_i$
  of every Linear Recurrent Layer
  are continuously-differentiable up to the closures of their domains,
  they are differentiable at every point of their original domains.
  Therefore, by finite composition and repeated application of the chain
  rule, for every $t \in \timedomain$,
  the parameter-to-state function $\statefunc_t^{u,x_0}$ is differentiable at every
  $\theta\in\truncation{\Theta}{r}$.
  This establishes the differentiability condition in the definition of
  BIBSD.

  For every
  $\theta\in\truncation{\Theta}{r}$,
  let
  \begin{equation*}
    x_t^\theta
    \defeq
    \begin{cases}
      x_0
      &
      \text{if }t=0,
      \\
      \fpinst{f}{x_{t-1}^\theta,u_t}{\theta}
      &
      \text{otherwise},
    \end{cases}
  \end{equation*}
  and define the one-step state and parameter Jacobians
  \begin{align*}
    J_t^\theta
    \defeq
    \dvop{f^\theta_{u_t}}
    \big(
      x_{t-1}^\theta
    \big),
    \qquad
    G_t^\theta
    \defeq
    \dvop{
      f^{x_{t-1}^\theta}_{u_t}
    }(\theta),
  \end{align*}
  where
  \begin{align*}
    f^\theta_{u_t}(x)
    \defeq
    \fpinst{f}{x,u_t}{\theta},
    \qquad
    f^x_{u_t}(\vartheta)
    \defeq
    \fpinst{f}{x,u_t}{\vartheta}.
  \end{align*}

  By
  Lemma~\ref{lemma:contractive-state-transition},
  there exist
  \begin{equation*}
    C_r\in\posreals,
    \qquad
    \rho_r\in[0,1),
  \end{equation*}
  which we may take with
  $C_r\geq1$,
  such that
  \begin{equation}
    \label{eq:bibsd-transition-bound}
    \left\|
      J_t^\theta
      \cdot
      J_{t-1}^\theta
      \cdots
      J_{s+1}^\theta
    \right\|
    \leq
    C_r
    \cdot
    \rho_r^{t-s}
  \end{equation}
  for every $\theta\in\truncation{\Theta}{r}$ and every $s,t\in\naturals$ satisfying $s<t$.

  We next uniformly bound the one-step parameter Jacobians.
  By
  Lemma~\ref{lemma:contractive-bounded-reachability},
  there exists
  $R_r\in\nonnegreals$
  such that
  \begin{equation*}
    \|x_t^\theta\|
    \leq
    R_r
    \qquad\qquad
    \forall\, \theta\in\truncation{\Theta}{r},\;
    \forall\, u\in\infseqs{\truncation{U}{r}},\;
    \forall\, x_0\in\truncation{X}{r},\;
    \forall\, t\in\naturals.
  \end{equation*}
  Thus all states, inputs, and parameters occurring in the one-step
  parameter Jacobians range over bounded sets.
  Each block of
  $G_t^\theta$
  is obtained, by the chain rule, from finitely many values and
  derivatives of the parametric input functions
  $A_{i,j}$ and $B_{i,j}$
  and of the layer output functions $h_i$.
  Since all their arguments range over bounded sets and these functions
  are continuously-differentiable up to the closures of their domains,
  Lemmas~\ref{lemma:closure-continuous-bounded}
  and~\ref{lemma:closure-cd-bounded-derivative}
  imply that all such values and derivatives are uniformly bounded.

  Since the architecture contains finitely many layers, units, and
  parameters, there exists
  \begin{equation*}
    M_r\in\nonnegreals
  \end{equation*}
  such that
  \begin{equation}
    \label{eq:bibsd-parameter-step-bound}
    \|G_t^\theta\|
    \leq
    M_r
    \qquad\qquad
    \forall
    \ptuple{\theta,t}
    \in
    \truncation{\Theta}{r}
    \times
    \timedomain.
  \end{equation}

  For every
  $\theta\in\truncation{\Theta}{r}$
  and
  $t\in\naturals$,
  define the parameter sensitivity
  \begin{equation*}
    Z_t^\theta
    \defeq
    \dvop{
      \statefunc_t^{u,x_0}
    }(\theta).
  \end{equation*}

  Since the initial state $x_0$ is independent of the parameters,
  \begin{equation*}
    Z_0^\theta=0
    \qquad\qquad
    \forall
    \theta\in\truncation{\Theta}{r}.
  \end{equation*}

  By the chain rule,
  \begin{equation*}
    Z_t^\theta
    \,=\,
    J_t^\theta
    \cdot
    Z_{t-1}^\theta
    +
    G_t^\theta
    \qquad\qquad
    \forall
    \ptuple{\theta,t}
    \in
    \truncation{\Theta}{r}
    \times
    \timedomain.
  \end{equation*}

  Iterating this recurrence gives
  \begin{equation*}
    Z_t^\theta
    \,=\,
    G_t^\theta
    +
    \sum_{k=1}^{t-1}
    J_t^\theta
    \cdot
    J_{t-1}^\theta
    \cdots
    J_{k+1}^\theta
    \cdot
    G_k^\theta.
  \end{equation*}

  Hence,
  by
  Equations~\eqref{eq:bibsd-transition-bound}
  and~\eqref{eq:bibsd-parameter-step-bound},
  and using
  $C_r\geq1$,
  \begin{align*}
    \|Z_t^\theta\|
    \;\leq\;
    M_r
    +
    C_r
    \cdot
    M_r
    \sum_{k=1}^{t-1}
    \rho_r^{t-k}
    \;\leq\;
    C_r
    \cdot
    M_r
    \sum_{k=0}^{t-1}
    \rho_r^k
    \;\leq\;
    \frac{
      C_r
      \cdot
      M_r
    }{
      1-\rho_r
    }.
  \end{align*}

  Let us define
  \begin{equation*}
    B_r
    \defeq
    \frac{
      C_r
      \cdot
      M_r
    }{
      1-\rho_r
    }.
  \end{equation*}
  Then
  \begin{equation*}
    \left\|
      \dvop{
        \statefunc_t^{u,x_0}
      }(\theta)
    \right\|
    \leq
    B_r
    \qquad\qquad
    \forall\, \theta\in\truncation{\Theta}{r},\;
    \forall\, t\in\timedomain.
  \end{equation*}

  Hence both conditions in the definition of BIBSD hold.
  Since
  $r$,
  $x_0$,
  and
  $u$
  were arbitrary,
  $S$ satisfies BIBSD.
\end{proof}

\begin{closingtheorem}[Vanishing state derivatives of smooth contractive Linear RNNs]
  \label{thm:contractive-vanishing-state-derivative}
  Let $S$ be a smooth contractive Linear RNN
  having
  input space $U$
  and
  initial state $x^\mathrm{init}$,
  and
  let $\Theta$ be the parameter space of its dynamics $\dynop S$.

  For
  every $\theta\in\Theta$
  and
  every $u\in\binfseqs{U}$,
  \begin{equation*}
    \lim_{t\to\infty}
    \left\|
      \dvop{\statefunc_t^{u,\theta}}(x^\mathrm{init})
    \right\|
    =
    0,
  \end{equation*}
  where
  $(\statefunc_t^{u,\theta})_{t\in\timedomain}$
  are the state-to-state functions of $\dynop S$.

  In particular, $S$ does not satisfy nnVSD.
\end{closingtheorem}
\begin{proof}
  Let us consider arbitrary
    $\theta\in\Theta$ and
    $u=(u_t)_{t\in\timedomain}\in\binfseqs{U}$.

  Since $u$ is bounded, there exists
  $r_u\in\nonnegreals$
  such that
  \begin{equation*}
    u_t\in\truncation{U}{r_u}
    \qquad
    \forall t\in\timedomain.
  \end{equation*}
  Furthermore, there exist
  $r_x,r_\theta\in\nonnegreals$
  such that
  \begin{equation*}
    x^\mathrm{init}\in\truncation{X}{r_x}
    \qquad\text{and}\qquad
    \theta\in\truncation{\Theta}{r_\theta}.
  \end{equation*}

  Let $r \defeq \max\{r_x,r_u,r_\theta\}$.
  Then
  \begin{equation*}
    x^\mathrm{init}\in\truncation{X}{r},
    \qquad
    \theta\in\truncation{\Theta}{r},
    \qquad
    u\in\infseqs{\truncation{U}{r}}.
  \end{equation*}

  Since the parametric input functions $A_{i,j}$ and $B_{i,j}$ of
  every Linear Recurrent Unit and the output functions $h_i$ of every
  Linear Recurrent Layer are continuously-differentiable up to the
  closures of their domains, they are differentiable at every point of
  their original domains.
  Therefore the dynamics function $f$ of the Linear RNN is
  differentiable in its state argument.
  Hence
  $\statefunc_t^{u,\theta}$
  is differentiable at
  $x^\mathrm{init}$
  for every
  $t\in\timedomain$.

  Let
  \begin{equation*}
    J_t^{u,\theta}
    \defeq
    \dvop{f^\theta_{u_t}}
    \big(
      x_{t-1}^{u,\theta}
    \big)
  \end{equation*}
  denote the one-step state Jacobian along the corresponding trajectory,
  where
  \begin{equation*}
    f^\theta_{u_t}(x)
    \defeq
    \fpinst{f}{x,u_t}{\theta}
    \qquad\text{and}\qquad
    x_t^{u,\theta} \defeq 
    \begin{cases}
      x_0 & \text{if } t = 0,
      \\
      f^\theta_{u_t}(x_{t-1}^{u,\theta}) & \text{otherwise}.
    \end{cases}
  \end{equation*}

  By repeated application of the chain rule,
  \begin{equation*}
    \dvop{\statefunc_t^{u,\theta}}
    (x^\mathrm{init})
    \,=\,
    J_t^{u,\theta}
    \cdot
    J_{t-1}^{u,\theta}
    \cdots
    J_1^{u,\theta}
    \quad\qquad
    \forall t\in\timedomain.
  \end{equation*}

  By
  Lemma~\ref{lemma:contractive-state-transition},
  there exist
  $C_r\in\posreals$ and
  $\rho_r\in[0,1)$
  such that
  \begin{equation*}
    \left\|
      J_t^{u,\theta}
      \cdot
      J_{t-1}^{u,\theta}
      \cdots
      J_1^{u,\theta}
    \right\|
    \,\leq\,
    C_r
    \cdot
    \rho_r^t
    \quad\qquad
    \forall t\in\timedomain.
  \end{equation*}
  Therefore,
  \begin{equation*}
    \left\|
      \dvop{\statefunc_t^{u,\theta}}
      (x^\mathrm{init})
    \right\|
    \,\leq\,
    C_r
    \cdot
    \rho_r^t
    \quad\qquad
    \forall t\in\timedomain.
  \end{equation*}

  Since
  $\rho_r<1$,
  it follows that
  \begin{equation*}
    \lim_{t\to\infty}
    \left\|
      \dvop{\statefunc_t^{u,\theta}}
      (x^\mathrm{init})
    \right\|
    \,=\,
    0.
  \end{equation*}

  Since
  $\theta\in\Theta$
  and
  $u\in\binfseqs{U}$
  are arbitrary,
  the state derivatives vanish for every parameter instance and every
  bounded input sequence.

  Suppose, by contradiction, that $S$ satisfies nnVSD.
  Then there exist
  $\theta\in\Theta$,
  $u\in\binfseqs{U}$,
  and
  $B\in\posreals$
  such that
  \begin{equation*}
    \left\|
      \dvop{\statefunc_t^{u,\theta}}
      (x^\mathrm{init})
    \right\|
    \geq
    B
    \qquad
    \forall t\in\timedomain.
  \end{equation*}
  This contradicts the convergence to zero established above.

  Therefore $S$ does not satisfy nnVSD.
\end{proof}

\clearpage{}%

\clearpage{}%

\section{Analysis of DeltaNet}
\label{app:delta-net}

\todo{}

\localtableofcontents

\todo{}

\subsection{Definition of DeltaNet}
\label{subsec:definition-of-deltanet}

\begin{closingdefinition}[Generic DeltaNet Recurrent Units]
  \label{def:deltanet-unit}
  Let $c \in \{1,2\}$,
  let
  $d_\mathrm{k},d_\mathrm{v} \in \posnaturals$,
  and
  let $d_\mathrm{kv} \defeq (d_\mathrm{k},d_\mathrm{v})$.

  The \emph{admissible DeltaNet input space} of key-value dimensions $d_\mathrm{kv}$ 
  is
  \begin{equation*}
    \mathcal U_{\mathrm{kv}}
    \,\defeq\,
    [0,1]
    \times
    (0,1)
    \times
    \mathbb{S}^{d_\mathrm{k}}
    \times
    \reals^{d_\mathrm{v}}
    \quad\text{ where }\quad
    \mathbb{S} \defeq \{ z \in \reals : \|z\| = 1 \}.
  \end{equation*}

  A \emph{Generic DeltaNet Recurrent Unit} 
  is a Left-Multiplicative Linear Recurrent Unit
  \begin{equation*}
    D = \mtuple{X, U, f}{A, B}
    \qquad\qquad
    X \subseteq \reals^{d_\mathrm{k}\times d_\mathrm{v}}\,,
    \quad
    U \subseteq \mathcal U_{\mathrm{kv}}\,,
  \end{equation*}
  where
  the dynamics function is as follows,
  for every state $H\in X$ and input $u = \ptuple{ g, \beta, k, v } \in U$:
  \begin{align*}
    f
    \big(
      H,
      \ptuple{g,\beta,k,v}
    \big)
    \,\defeq\;
    A(g,\beta,k) \cdot H
    \,+\,
    B(\beta,k,v)
  \end{align*}
  with state-update matrix and additive term given respectively by
  \begin{align*}
    A(g,\beta,k)
    \,\defeq\;
    g 
    \cdot 
    \left(
      I_{d_\mathrm{k}}
      -
      c \cdot \beta \cdot k\cdot k^\transposeop 
    \right)
    \qquad\text{and}\qquad
    B(\beta,k,v)
    \,\defeq\,
    \beta \cdot k \cdot v^\transposeop\,.
  \end{align*}
  By expanding the definitions of $A$ and $B$,
  the dynamics function can be equivalently written as
  \begin{align*}
    f
    \big(
    H,
    \ptuple{g,\beta,k,v}
    \big)
    \,\defeq\;
    g 
    \cdot 
    \left(
      I_{d_\mathrm{k}}
      -
      c \cdot \beta \cdot k\cdot k^\transposeop 
    \right)
    \cdot H
    \,+\,
    \beta \cdot k \cdot v^\transposeop\,.
  \end{align*}

  The following terminology applies:
  \begin{itemize}
    \item
      $c$ is the \emph{update hyperparameter $c$},
    \item
      $d_\mathrm{k}$ is the \emph{key dimension},
    \item
      $d_\mathrm{v}$ is the \emph{value dimension},
    \item
      $d_\mathrm{kv}$ are the \emph{key-value dimensions}.
  \end{itemize}

  We say that $D$ is \emph{non-gated} when
  every input $\ptuple{ g, \beta, k, v } \in U$
  satisfies $g=1$.
\end{closingdefinition}

\begin{closingdefinition}[DeltaNet and Gated DeltaNet, cf.~\citep{yang2025gated,yang2024parallelizing}]
  \label{def:deltanet-rnc}

  We first define Gated DeltaNet, and then we define DeltaNet as a special case.

  Let $c \in \{1,2\}$,
  let $d_\mathrm{in},d_\mathrm{k},d_\mathrm{v}\in\posnaturals$,
  let $d_\mathrm{kv} \defeq \ptuple{ d_\mathrm{k}, d_\mathrm{v} }$,
  and
  let
  \begin{equation*}
    \hat D = \mtuple{ X, \hat U, \hat f }{ \hat A, \hat B }
  \end{equation*}
  be a Generic DeltaNet Recurrent Unit  with updated hyperparameter $c$ and key-value dimensions $d_\mathrm{kv}$.%
  \footnote{%
    \emph{Generic DeltaNet Recurrent Unit} are introduced in Definition~\ref{def:deltanet-unit}.%
  }

  A \emph{Gated DeltaNet Recurrent Unit} with \emph{update hyperparameter $c$}
  is a Parametric Left-Multiplicative Linear Recurrent Unit
  \begin{equation*}
    D = \mctuple{X,U,f}{A,B}{\Theta}
  \end{equation*}
  having components as described next.

  \begin{definitionpartsimple}[Input space] 
    \label{def:deltanet-rnc-input-space}
    The input space is $U \subseteq \reals^{d_\mathrm{in}}$.
  \end{definitionpartsimple}

  \begin{definitionpartsimple}[Parameter space] 
    \label{def:deltanet-rnc-parameter-space}
    The joint parameter space $\Theta$ consists of parameter values $\theta$ of the following form:
    \begin{gather*}
      \theta
      =
      \bigptuple{
        a,
        (W_\mathrm{k},b_\mathrm{k}),
        (W_\mathrm{v},b_\mathrm{v}),
        (w_\beta,b_\beta),
        (w_\Delta,b_\Delta)
      }
      \mkern15mu
      \text{for}
      \mkern15mu
      \begin{cases}
        \mkern2mu
        W_\mathrm{k}
        \in
        \reals^{d_\mathrm{k}\times d_\mathrm{in}}
        \mkern1mu ,
        \\%
        \mkern2mu
        W_\mathrm{v}
        \in
        \reals^{d_\mathrm{v}\times d_\mathrm{in}}
        \mkern1mu ,
        \\%
        \mkern5mu 
        w_\beta
        \in
        \reals^{d_\mathrm{in}}
        \mkern1mu ,
        \\%
        \mkern3mu 
        w_\Delta
        \in
        \reals^{d_\mathrm{in}}
        \mkern1mu ,
      \end{cases}
      \text{and}
      \mkern16mu
      \begin{cases}
        \mkern9mu
        a
        \in
        \reals
        \mkern1mu ,
        \\%
        \mkern3mu
        b_\mathrm{k}
        \in
        \reals^{d_\mathrm{k}}
        \mkern1mu ,
        \\%
        \mkern3mu
        b_\mathrm{v}
        \in
        \reals^{d_\mathrm{v}}
        \mkern1mu ,
        \\%
        \mkern2mu
        b_\beta
        \in
        \reals
        \mkern2mu ,
        \\%
        b_\Delta
        \in
        \reals
        \mkern2mu ,
      \end{cases}
    \end{gather*}
    satisfying the \emph{key-admissibility condition} 
    \begin{equation} 
      \label{eq:def:deltanet-rnc-key-admissibility} 
        W_\mathrm{k}\cdot u + b_\mathrm{k} \;\neq\; 0 
        \qquad\quad 
        \forall u\in U. 
    \end{equation}
  \end{definitionpartsimple}

  \begin{definitionpartsimple}[Recurrence input functions]
    \label{def:deltanet-rnc-recurrence-input-functions}
    The recurrence inputs are given by 
    \emph{parametric recurrence-input functions} 
    \begin{align*}
      g
      & :
      U\mkern-2mu \times\Theta
      \to
      [0,1]
      &
      \fpinst{g}{u}{\theta}
      & \defeq
      \exp
      \left(
      {-}
      \fpinst{\Delta}{u}{\theta}
      \cdot
      \exp(a)
      \right)
      && \textit{decay gate}
      \\[.3em]
      \beta
      & :
      U\mkern-2mu \times\Theta
      \to
      (0,1)
      &
      \fpinst{\beta}{u}{\theta}
      & \defeq
      \sigma
      \big(
      \fpinst{\tilde\beta}{u}{\theta}
      \big)
      && \textit{write strength}
      \\[.3em]
      k
      & :
      U\mkern-2mu \times\Theta
      \to
      \mathbb S^{d_\mathrm{k}-1}
      &
      \fpinst{k}{u}{\theta}
      & \defeq
      \fpinst{\tilde k}{u}{\theta}
      \,/\,
      \| \fpinst{\tilde k}{u}{\theta} \|
      && \textit{key}
      \\[.3em]
      v
      & :
      U\mkern-2mu \times\Theta
      \to
      \reals^{d_\mathrm{v}}
      &
      \fpinst{v}{u}{\theta}
      & \defeq
      W_\mathrm{v}
      \cdot u
      +
      b_\mathrm{v}
      && \textit{value}
    \end{align*}
    where the mentioned intermediate functions are
    \begin{align*}
      \Delta
      & :
      U\mkern-2mu \times\Theta
      \to
      \posreals
      &
      \fpinst{\Delta}{u}{\theta}
      & \defeq
      \operatorname{softplus}
      \big(
      \fpinst{\tilde\Delta}{u}{\theta}
      \big)
      && \textit{step size}
      \\[.3em]
      \tilde\Delta
      & :
      U\mkern-2mu \times\Theta
      \to
      \reals
      &
      \fpinst{\tilde\Delta}{u}{\theta}
      & \defeq
      w_\Delta^\transposeop
      \cdot u
      +
      b_\Delta
      && \textit{step-size preactivation}
      \\[.3em]
      \tilde\beta
      & :
      U\mkern-2mu \times\Theta
      \to
      \reals
      &
      \fpinst{\tilde\beta}{u}{\theta}
      & \defeq
      w_\beta^\transposeop
      \cdot u
      +
      b_\beta
      && \textit{write-strength preactivation}
      \\[.3em]
      \tilde k
      & :
      U\mkern-2mu \times\Theta
      \to
      \reals^{d_\mathrm{k}}
      &
      \fpinst{\tilde k}{u}{\theta}
      & \defeq
      W_\mathrm{k}
      \cdot u
      +
      b_\mathrm{k}
      && \textit{key preactivation}\,.
    \end{align*}
  \end{definitionpartsimple}

  \begin{definitionpartsimple}[Dynamics function]
    \label{def:deltanet-rnc-recurrence-dynamics-function}
    The parametric dynamics function is obtained by applying the dynamics function of $\hat D$
    to the corresponding parametrised recurrent input:
    \begin{equation*}
      \label{eq:def:gated-deltanet-rnc--generic-dynamics}
      \fpinst{f}{H,u}{\theta}
      \defeq
      \hat f
      \mBig(
        H,\,
        \ptuple{
        \fpinst{g}{u}{\theta},
        \fpinst{\beta}{u}{\theta},
        \fpinst{k}{u}{\theta},
        \fpinst{v}{u}{\theta}
        }
      \mBig).
    \end{equation*}

  \end{definitionpartsimple}

  \begin{definitionpartsimple}[DeltaNet Recurrent Units with no decay gate]
    \label{def:deltanet-rnc-nongated-deltanet-recurrent-units}
    A \emph{DeltaNet Recurrent Unit with no decay gate} is defined as above except that the decay
    gate function is the constant function $g = 1$; consequently, the unit does not need to include
    the parameters $a$, $w_\Delta$, and $b_\Delta$---since they are exclusive dependencies of $g$.
  \end{definitionpartsimple}

  \begin{definitionpartsimple}[Classes of DeltaNet Recurrent Units]
    \label{def:deltanet-rnc-classes-of-deltanet-recurrent-units}
    We define the following classes:
    \begin{itemize}
      \item
        the \emph{class of DeltaNet Recurrent Units with update hyperparameter $c$} 
        is the set consisting of all DeltaNet recurrent units with update hyperparameter $c$,
        and is denoted by
        $\classunits{DeltaNet}[c]$;
      \item
        the \emph{class of Gated DeltaNet Recurrent Units}
        is the set consisting of all Gated DeltaNet recurrent units,
        and is denoted by
        $\classunits{GatedDeltaNet}[c]$.
    \end{itemize}
  \end{definitionpartsimple}

  \begin{definitionpartsimple}[DeltaNet Models]
    \label{def:deltanet-rnc-deltanet-models}
    A \emph{Gated DeltaNet model} with update hyperparameter $c$ 
    is a parametric $\classunits{GatedDeltaNet}[c]$-RNC.
    A \emph{DeltaNet model} with update hyperparameter $c$ 
    is a parametric $\classunits{DeltaNet}[c]$-RNC.%
  \end{definitionpartsimple}

  \begin{definitionpartsimple}[Classes of DeltaNet Models]
    \label{def:deltanet-rnc-classes-of-deltanet-models}
    We define the following classes:
    \begin{itemize}
      \item
        the \emph{class of DeltaNet Models} with update hyperparameter $c$
        is the set consisting of each DeltaNet model with update hyperparameter $c$,
        and is denoted by
        $\classmodels{DeltaNet}[c]$;
      \item
        the \emph{class of Gated DeltaNet Models} with update hyperparameter $c$
        is the set consisting of each DeltaNet model with update hyperparameter $c$,
        and is denoted by $\classmodels{GatedDeltaNet}[c]$;
      \item
        the \emph{class of DeltaNet Dynamics} with update hyperparameter $c$
        is the set consisting of each $D = \dynop{S}$ for $S \in \classmodels{DeltaNet}[c]$,
        and is denoted by $\classdynamics{DeltaNet}[c]$;
      \item
        the \emph{class of Gated-DeltaNet Dynamics} with update hyperparameter $c$
        is the set consisting of each $D = \dynop{S}$ for $S \in \classmodels{GatedDeltaNet}[c]$,
        and is denoted by $\classdynamics{GatedDeltaNet}[c]$.
    \end{itemize}
  \end{definitionpartsimple}
\end{closingdefinition}

\begin{closingremark}[Decay-gate parametrisation in Gated DeltaNet]
  \label{rem:gated-deltanet-decay-gate-parametrisation}

  The Gated DeltaNet paper~\citep{yang2025gated} specifies that the
  decay gate follows the Mamba2 parametrisation~\citep{dao2024transformers},
  while omitting its explicit form for brevity.
  The standalone official implementation of Gated DeltaNet
  \citep{yang2024gateddeltanetcode}
  and its FLA implementation
  \citep{flashlinearattention}
  make this parametrisation explicit by first forming the log-decay
  \begin{equation*}
    \gamma(u)
    =
    -
    \exp(a)
    \cdot
    \operatorname{softplus}
    \big(
      \tilde\Delta(u)
    \big),
  \end{equation*}
  and then using its exponential as the multiplicative decay gate.
  Consequently,
  in the notation of Definition~\ref{def:deltanet-rnc},
  \begin{align*}
    \fpinst{g}{u}{\theta}
    &=
    \exp
    \mBig(
      -
      \exp(a)
      \cdot
      \operatorname{softplus}
      \big(
        \fpinst{\tilde\Delta}{u}{\theta}
      \big)
    \mBig)
    \\[.1em]
    &=
    \exp
    \left(
      -
      \fpinst{\Delta}{u}{\theta}
      \cdot
      \exp(a)
    \right),
  \end{align*}
  where
  \begin{equation*}
    \fpinst{\Delta}{u}{\theta}
    =
    \operatorname{softplus}
    \big(
      \fpinst{\tilde\Delta}{u}{\theta}
    \big).
  \end{equation*}
  Thus, the parametrisation used in
  Definition~\ref{def:deltanet-rnc}
  is the explicit form of the Mamba2-style decay parametrisation
  adopted by Gated DeltaNet and used in its reference implementations.
  We note that the implementations store the quantity corresponding
  to $\gamma$ as the decay variable, whereas $g$ in
  Definition~\ref{def:deltanet-rnc} denotes the resulting
  multiplicative decay gate $\exp(\gamma)$.
\end{closingremark}

\subsection{Overview of the Analysis}

For DeltaNet,
we show show 
BIBS instability (Theorem~\ref{thm:divergent-delta-rule-beta}),
BIBSD instability (Theorem~\ref{thm:deltanet-does-not-satisfy-wbibsd}),
and
BIBLG instability (Theorem~\ref{thm:deltanet-linear-loss-gradient-growth} and
Theorem~\ref{thm:deltanet-standard-loss-gradient-growth}).

The theoretical results are validated experimentally, 
using the reference implementation of DeltaNet~\citep{flashlinearattention} to instantiate the
unstable instances identified by the theoretical analyses.
In particular, we report figures showing divergent trajectories for 
states, 
state derivatives, 
and 
loss gradients.

\subsection{BIBS Instability of DeltaNet}
\label{sec:deltarule-instability}

We show that DeltaNet dynamics do not satisfy wBIBS. In particular, we show that already 
DeltaNet Units admit divergent state trajectories under mild conditions.
The divergent state sequence of an unstable Delta Unit is shown in 
Figure~\ref{fig:generated-divergent-state-beta09}.

\begin{restatable}[DeltaNet does not satisfy wBIBS]{closingtheorem}{thmdivergentdeltarulebeta}
  \label{thm:divergent-delta-rule-beta}
  Let
  $\beta\in(0,1)$,
  and let us define
  \begin{equation*}
    \tilde U
    \defeq
    \big\{
      \ptuple{
        1,
        \beta,
        k,
        v
      }
      \bigm|
      k\in K,\,
      v\in\{0,1\}
    \big\}
    \quad\text{for}\;\;
    K
    \defeq
    \left\{
      \frac{k}{\|k\|}
      \in
      \reals^2
      \,\middle|\,
      k
      =
      \begin{pmatrix}
        1
        \\
        \alpha
      \end{pmatrix},
      \,
      \alpha\in[0,1/2]
    \right\}.
  \end{equation*}

  Let
  $D=\mtuple{X,U,f}{A,B}$
  be a Generic DeltaNet Recurrent Unit
  with
  update hyperparameter $c=1$
  and
  key-value dimensions
  $d_\mathrm{kv}=\ptuple{2,1}$.

  If the state and input spaces satisfy
  \begin{align*}
    X\neq\emptyset
    \quad\;\;\text{and}\quad\;\;
    \tilde U
    \subseteq
    U,
  \end{align*}
  then 
  the DeltaNet Unit $D$ does not satisfy wBIBS.
\end{restatable}
\begin{proof}
  Deferred to \S\ref{subsec:deltanet--deferred-proofs--bibs}.
\end{proof}

\begin{figure}[ht]
  \centering
  \caption{%
    Trajectory of state $x_t=(x_{t,1},x_{t,2})$ of the divergent DeltaNet dynamics of Lemma~\ref{lemma:divergent-delta-rule-beta}, computed with the official DeltaNet recurrence (flash-linear-attention), with $\beta=0.9$ and $x_0=(0,0)$. Components are shown in~(a) and~(b), the state norm in~(c).%
  }
  \vspace{1em}

  \begin{subfigure}{0.98\textwidth}
    \centering
    \tikzsetnextfilename{plot__divergent_state_beta09__x1}
    \begin{tikzpicture}
    \begin{axis}[
        width=12.5cm,
        height=6.5cm,
        axis lines=middle,
        xlabel={$t$},
        ylabel={$x_{t,1}$},
        xlabel style={xshift=2pt, yshift=2pt, font=\footnotesize},
        ylabel style={xshift=-22pt, yshift=4pt, font=\footnotesize},
        xmin=0, xmax=50.8,
        ymin=0, ymax=0.9897,
        xtick={0,5,10,15,20,25,30,35,40,45,50},
        ytick={0,0.2,0.4,0.6,0.8},
        grid=none,
        legend style={
            at={(0.53,1.02)},
            anchor=north west,
            fill=white,
            draw=black,
            cells={anchor=west},
        },
        every x tick label/.append style={font=\footnotesize},
        every y tick label/.append style={font=\footnotesize},
        yticklabel style={text width=2.8em, align=right},
        tick label style={
          /pgf/number format/fixed,
          /pgf/number format/precision=2
        }
    ]

    \addplot[
        black,
        dashed,
        thin,
        domain=0:52,
        samples=2,
        forget plot
    ] {0.090909};

    \addplot[
        orange,
        thick,
        mark=*,
        mark size=1.3pt
    ] coordinates {
        (0, 0.0000)
        (1, 0.0000)
        (2, 0.8931)
        (3, 0.1009)
        (4, 0.9111)
        (5, 0.1049)
        (6, 0.9144)
        (7, 0.1058)
        (8, 0.9157)
        (9, 0.1059)
        (10, 0.9162)
        (11, 0.1056)
        (12, 0.9164)
        (13, 0.1052)
        (14, 0.9164)
        (15, 0.1048)
        (16, 0.9163)
        (17, 0.1043)
        (18, 0.9161)
        (19, 0.1039)
        (20, 0.9160)
        (21, 0.1034)
        (22, 0.9158)
        (23, 0.1030)
        (24, 0.9156)
        (25, 0.1026)
        (26, 0.9154)
        (27, 0.1023)
        (28, 0.9153)
        (29, 0.1019)
        (30, 0.9151)
        (31, 0.1016)
        (32, 0.9149)
        (33, 0.1013)
        (34, 0.9148)
        (35, 0.1010)
        (36, 0.9146)
        (37, 0.1007)
        (38, 0.9145)
        (39, 0.1005)
        (40, 0.9144)
        (41, 0.1002)
        (42, 0.9142)
        (43, 0.1000)
        (44, 0.9141)
        (45, 0.0998)
        (46, 0.9140)
        (47, 0.0996)
        (48, 0.9139)
        (49, 0.0994)
        (50, 0.9138)
    };
    \addlegendentry{\scriptsize trajectory of $x_{t,1}$}

    \addplot[
        only marks,
        mark=*,
        mark size=1.8pt,
        black,
        forget plot
    ] coordinates {(0,0.0000)};
    \node[black, left] at (axis cs:0,0.0000) {$x_{0,1}$};

    \addplot[
        black,
        dashed,
        thin,
        domain=0:52,
        samples=2
    ] {0.909091};
    \addlegendentry{\scriptsize guides $\frac{1-\beta}{2-\beta},\frac{1}{2-\beta}$}

    \end{axis}
    \end{tikzpicture}
    \caption{Component $x_{t,1}$ remains bounded and approaches an alternating two-cycle near $\frac{1-\beta}{2-\beta}\approx0.0909$ and $\frac{1}{2-\beta}\approx0.9091$.}
  \end{subfigure}

  \vspace{0.4cm}

  \begin{subfigure}{0.98\textwidth}
    \centering
    \tikzsetnextfilename{plot__divergent_state_beta09__x2}
    \begin{tikzpicture}
    \begin{axis}[
        width=12.5cm,
        height=6.5cm,
        axis lines=middle,
        xlabel={$t$},
        ylabel={$x_{t,2}$},
        xlabel style={xshift=2pt, yshift=2pt, font=\footnotesize},
        ylabel style={xshift=-22pt, yshift=4pt, font=\footnotesize},
        xmin=0, xmax=50.8,
        ymin=-0.496, ymax=0.1566,
        xtick={0,5,10,15,20,25,30,35,40,45,50},
        ytick={-0.4,-0.3,-0.2,-0.1,0,0.1},
        grid=none,
        legend style={
            at={(0.53,1.02)},
            anchor=north west,
            fill=white,
            draw=black,
            cells={anchor=west},
        },
        every x tick label/.append style={font=\footnotesize},
        every y tick label/.append style={font=\footnotesize},
        yticklabel style={text width=2.8em, align=right},
        tick label style={
          /pgf/number format/fixed,
          /pgf/number format/precision=2
        }
    ]

    \addplot[
        blue,
        semithick,
        mark=square*,
        mark size=1.1pt
    ] coordinates {
        (0, 0.0000)
        (1, 0.0000)
        (2, 0.1116)
        (3, -0.0468)
        (4, 0.0207)
        (5, -0.0945)
        (6, -0.0439)
        (7, -0.1337)
        (8, -0.0932)
        (9, -0.1668)
        (10, -0.1331)
        (11, -0.1954)
        (12, -0.1665)
        (13, -0.2205)
        (14, -0.1952)
        (15, -0.2429)
        (16, -0.2204)
        (17, -0.2631)
        (18, -0.2428)
        (19, -0.2815)
        (20, -0.2631)
        (21, -0.2984)
        (22, -0.2815)
        (23, -0.3140)
        (24, -0.2983)
        (25, -0.3284)
        (26, -0.3139)
        (27, -0.3420)
        (28, -0.3284)
        (29, -0.3547)
        (30, -0.3420)
        (31, -0.3666)
        (32, -0.3546)
        (33, -0.3779)
        (34, -0.3666)
        (35, -0.3886)
        (36, -0.3779)
        (37, -0.3988)
        (38, -0.3886)
        (39, -0.4084)
        (40, -0.3987)
        (41, -0.4177)
        (42, -0.4084)
        (43, -0.4265)
        (44, -0.4177)
        (45, -0.4350)
        (46, -0.4265)
        (47, -0.4431)
        (48, -0.4350)
        (49, -0.4510)
        (50, -0.4431)
    };
    \addlegendentry{\scriptsize trajectory of $x_{t,2}$}

    \addplot[
        black,
        dashed,
        thin,
        domain=2:52,
        samples=200
    ] {0.347087 - 0.204545*ln(x)};
    \addlegendentry{\scriptsize guide $0.35\hspace{.1em}-\hspace{.1em}\frac{\beta}{4(2-\beta)}\log t$}

    \addplot[
        only marks,
        mark=*,
        mark size=1.8pt,
        black,
        forget plot
    ] coordinates {(0,0.0000)};
    \node[black, left] at (axis cs:0,0.0000) {$x_{0,2}$};

    \end{axis}
    \end{tikzpicture}
    \caption{Component $x_{t,2}$ drifts downward, consistently with $x_{2n,2}={-}\frac{\beta}{4(2-\beta)}\log n+O(1)$.}
  \end{subfigure}

  \vspace{0.4cm}

  \begin{subfigure}{0.98\textwidth}
    \centering
    \tikzsetnextfilename{plot__divergent_state_beta09__norm}
    \begin{tikzpicture}
    \begin{axis}[
        width=12.5cm,
        height=6.5cm,
        axis lines=middle,
        xlabel={$t$},
        ylabel={$\|x_t\|$},
        xlabel style={xshift=2pt, yshift=2pt, font=\footnotesize},
        ylabel style={xshift=-22pt, yshift=4pt, font=\footnotesize},
        xmin=0, xmax=50.8,
        ymin=0, ymax=1.097,
        xtick={0,5,10,15,20,25,30,35,40,45,50},
        ytick={0,0.2,0.4,0.6,0.8,1},
        grid=none,
        legend style={
            at={(0.53,0.27)},
            anchor=north west,
            fill=white,
            draw=black,
            cells={anchor=west},
        },
        every x tick label/.append style={font=\footnotesize},
        every y tick label/.append style={font=\footnotesize},
        yticklabel style={text width=2.8em, align=right},
        tick label style={
          /pgf/number format/fixed,
          /pgf/number format/precision=2
        }
    ]

    \addplot[
        red!70!black,
        thick,
        mark=triangle*,
        mark size=1.3pt
    ] coordinates {
        (0, 0.0000)
        (1, 0.0000)
        (2, 0.9000)
        (3, 0.1112)
        (4, 0.9113)
        (5, 0.1412)
        (6, 0.9154)
        (7, 0.1705)
        (8, 0.9204)
        (9, 0.1976)
        (10, 0.9258)
        (11, 0.2221)
        (12, 0.9314)
        (13, 0.2444)
        (14, 0.9369)
        (15, 0.2646)
        (16, 0.9424)
        (17, 0.2830)
        (18, 0.9478)
        (19, 0.3001)
        (20, 0.9530)
        (21, 0.3158)
        (22, 0.9581)
        (23, 0.3304)
        (24, 0.9630)
        (25, 0.3441)
        (26, 0.9678)
        (27, 0.3569)
        (28, 0.9724)
        (29, 0.3690)
        (30, 0.9769)
        (31, 0.3804)
        (32, 0.9813)
        (33, 0.3912)
        (34, 0.9855)
        (35, 0.4015)
        (36, 0.9896)
        (37, 0.4113)
        (38, 0.9936)
        (39, 0.4206)
        (40, 0.9975)
        (41, 0.4295)
        (42, 1.0013)
        (43, 0.4381)
        (44, 1.0050)
        (45, 0.4463)
        (46, 1.0086)
        (47, 0.4542)
        (48, 1.0122)
        (49, 0.4618)
        (50, 1.0156)
    };
    \addlegendentry{\scriptsize state norm $\|x_t\|$}

    \addplot[
        black,
        dashed,
        thin,
        domain=4.92651:52,
        samples=200
    ] {-0.326174 + 0.204545*ln(x)};
    \addlegendentry{\scriptsize lower-envelope guide $-0.33+\frac{\beta}{4(2-\beta)}\log t$}

    \addplot[
        only marks,
        mark=*,
        mark size=1.8pt,
        black,
        forget plot
    ] coordinates {(0,0.0000)};
    \node[black, left] at (axis cs:0,0.0000) {$\|x_0\|$};

    \end{axis}
    \end{tikzpicture}
    \caption{The state norm $\|x_t\|$ diverges logarithmically; it oscillates with $x_{t,1}$, and its lower envelope (odd $t$) follows $\frac{\beta}{4(2-\beta)}\log t+O(1)$.}
  \end{subfigure}

  \label{fig:generated-divergent-state-beta09}
\end{figure}
\FloatBarrier
\clearpage

\subsection{BIBSD Instability of DeltaNet}
\label{sec:deltarule-bibsd-instability}

\begin{restatable}[%
    DeltaNet does not satisfy wBIBSD%
  ]{%
    closingtheorem%
  }{%
    thmdeltanetdoesnotsatisfywbibsd%
  }
  \label{thm:deltanet-does-not-satisfy-wbibsd}
  \todo{}
  Let
  $c\in\{1,2\}$,
  $\epsilon\in\posreals$,
  and
  $\beta_\star\in(0,1)$.

  Let
  \begin{equation*}
    D_\star
    =
    \mctuple{X,U,f}{A,B}{\Lambda_\star}
  \end{equation*}
  be the specialised parametric DeltaNet Recurrent Unit
  with hyperparameters
  $\ptuple{c,\epsilon,\beta_\star}$
  of
  Construction~\ref{construction:deltanet-finite-alphabet-parameter-variation}.

  Then
  the DeltaNet Unit
  $D_\star$
  does not satisfy wBIBSD,
  and consequently it does not satisfy uBIBSD.
\end{restatable}
\begin{proof}
  Deferred to \S\ref{subsec:deltanet--deferred-proofs--bibsd}.
\end{proof}

\begin{figure}[ht]
  \centering
  \caption{%
    Bounded state and growing parameter sensitivity for the DeltaNet dynamics of Lemma~\ref{lemma:deltanet-finite-alphabet-parameter-sensitivity-growth}, computed with the official DeltaNet recurrence (flash-linear-attention), with $c=1$, $\beta_\star=0.9$, $u_t=(-1)^{t+1}$, $\lambda_\star=0$, and $x_0=(0,0)$. The reference-state trajectory is shown in~(a), and its sensitivity with respect to the scalar parameter coordinate $\lambda$ is shown in~(b).%
  }
  \vspace{1em}

  \begin{subfigure}{0.98\textwidth}
    \centering
    \tikzsetnextfilename{plot__parameter_sensitivity_c1_beta09__state}
    \begin{tikzpicture}
    \begin{axis}[
        width=12.5cm,
        height=8cm,
        axis lines=middle,
        xlabel={$t$},
        ylabel={$H^\star_{t,1}=a_t$},
        xlabel style={xshift=2pt, yshift=2pt, font=\footnotesize},
        ylabel style={xshift=-22pt, yshift=4pt, font=\footnotesize},
        xmin=0, xmax=50.8,
        ymin=-0.9556, ymax=1.037,
        xtick={0,5,10,15,20,25,30,35,40,45,50},
        ytick={-0.5,0,0.5},
        grid=none,
        legend style={
            at={(0.53,1.02)},
            anchor=north west,
            fill=white,
            draw=black,
            cells={anchor=west},
        },
        every x tick label/.append style={font=\footnotesize},
        every y tick label/.append style={font=\footnotesize},
        yticklabel style={text width=2.8em, align=right},
        tick label style={
          /pgf/number format/fixed,
          /pgf/number format/precision=2
        }
    ]

    \addplot[
        black,
        dashed,
        thin,
        domain=0:52,
        samples=2,
        forget plot
    ] {0.818182};

    \addplot[
        orange,
        thick,
        mark=*,
        mark size=1.3pt
    ] coordinates {
        (0, 0.0000)
        (1, 0.9000)
        (2, -0.8100)
        (3, 0.8190)
        (4, -0.8181)
        (5, 0.8182)
        (6, -0.8182)
        (7, 0.8182)
        (8, -0.8182)
        (9, 0.8182)
        (10, -0.8182)
        (11, 0.8182)
        (12, -0.8182)
        (13, 0.8182)
        (14, -0.8182)
        (15, 0.8182)
        (16, -0.8182)
        (17, 0.8182)
        (18, -0.8182)
        (19, 0.8182)
        (20, -0.8182)
        (21, 0.8182)
        (22, -0.8182)
        (23, 0.8182)
        (24, -0.8182)
        (25, 0.8182)
        (26, -0.8182)
        (27, 0.8182)
        (28, -0.8182)
        (29, 0.8182)
        (30, -0.8182)
        (31, 0.8182)
        (32, -0.8182)
        (33, 0.8182)
        (34, -0.8182)
        (35, 0.8182)
        (36, -0.8182)
        (37, 0.8182)
        (38, -0.8182)
        (39, 0.8182)
        (40, -0.8182)
        (41, 0.8182)
        (42, -0.8182)
        (43, 0.8182)
        (44, -0.8182)
        (45, 0.8182)
        (46, -0.8182)
        (47, 0.8182)
        (48, -0.8182)
        (49, 0.8182)
        (50, -0.8182)
    };
    \addlegendentry{\scriptsize trajectory of $H^\star_{t,1}$}

    \addplot[
        only marks,
        mark=*,
        mark size=1.8pt,
        black,
        forget plot
    ] coordinates {(0,0.0000)};
    \node[black, left] at (axis cs:0,0.0000) {$H^\star_{0,1}$};

    \addplot[
        black,
        dashed,
        thin,
        domain=0:52,
        samples=2
    ] {-0.818182};
    \addlegendentry{\scriptsize guides $\pm\frac{\beta_\star}{2-c\beta_\star}$}

    \end{axis}
    \end{tikzpicture}
    \caption{The reference-state component $H^\star_{t,1}=a_t$ remains bounded and approaches the alternating two-cycle $\pm\frac{\beta_\star}{2-c\beta_\star}\approx\pm0.8182$. The second state component is identically zero.}
  \end{subfigure}

  \vspace{0.4cm}

  \begin{subfigure}{0.98\textwidth}
    \centering
    \tikzsetnextfilename{plot__parameter_sensitivity_c1_beta09__sensitivity}
    \begin{tikzpicture}
    \begin{axis}[
        width=12.5cm,
        height=9cm,
        axis lines=middle,
        xlabel={$t$},
        ylabel={$g_t$},
        xlabel style={xshift=2pt, yshift=2pt, font=\footnotesize},
        ylabel style={xshift=-22pt, yshift=4pt, font=\footnotesize},
        xmin=0, xmax=50.8,
        ymin=0, ymax=87.64,
        xtick={0,5,10,15,20,25,30,35,40,45,50},
        ytick={0,20,40,60,80},
        grid=none,
        legend style={
            at={(0.04,0.98)},
            anchor=north west,
            fill=white,
            draw=black,
            cells={anchor=west},
        },
        every x tick label/.append style={font=\footnotesize},
        every y tick label/.append style={font=\footnotesize},
        yticklabel style={text width=2.8em, align=right},
        tick label style={
          /pgf/number format/fixed,
          /pgf/number format/precision=2
        }
    ]

    \addplot[
        blue,
        thick,
        mark=square*,
        mark size=1.1pt
    ] coordinates {
        (0, 0.0000)
        (1, 0.9000)
        (2, 2.6100)
        (3, 4.2390)
        (4, 5.8761)
        (5, 7.5124)
        (6, 9.1488)
        (7, 10.7851)
        (8, 12.4215)
        (9, 14.0578)
        (10, 15.6942)
        (11, 17.3306)
        (12, 18.9669)
        (13, 20.6033)
        (14, 22.2397)
        (15, 23.8760)
        (16, 25.5124)
        (17, 27.1488)
        (18, 28.7851)
        (19, 30.4215)
        (20, 32.0578)
        (21, 33.6942)
        (22, 35.3306)
        (23, 36.9669)
        (24, 38.6033)
        (25, 40.2397)
        (26, 41.8760)
        (27, 43.5124)
        (28, 45.1488)
        (29, 46.7851)
        (30, 48.4215)
        (31, 50.0578)
        (32, 51.6942)
        (33, 53.3306)
        (34, 54.9669)
        (35, 56.6033)
        (36, 58.2397)
        (37, 59.8760)
        (38, 61.5124)
        (39, 63.1488)
        (40, 64.7851)
        (41, 66.4215)
        (42, 68.0578)
        (43, 69.6942)
        (44, 71.3306)
        (45, 72.9669)
        (46, 74.6033)
        (47, 76.2397)
        (48, 77.8760)
        (49, 79.5124)
        (50, 81.1487)
    };
    \addlegendentry{\scriptsize parameter sensitivity $g_t$}

    \addplot[
        black,
        dashed,
        thin,
        domain=0:52,
        samples=2
    ] {0.9*x};
    \addlegendentry{\scriptsize lower bound $\beta_\star t=0.9 t$}

    \addplot[
        only marks,
        mark=*,
        mark size=1.8pt,
        black,
        forget plot
    ] coordinates {(0,0.0000)};
    \node[black, left] at (axis cs:0,0.0000) {$g_0$};

    \end{axis}
    \end{tikzpicture}
    \caption{The parameter sensitivity $g_t\defeq\big\|\pdvop{H_t}{\lambda_\star}(\lambda_\star)\big\|$
    grows without bound and satisfies $g_t\geq\beta_\star t$.}
  \end{subfigure}

  \label{fig:generated-parameter-sensitivity-c1-beta09}
\end{figure}

\begin{figure}[ht]
  \centering
  \caption{%
    Bounded state and growing parameter sensitivity for the DeltaNet dynamics of Lemma~\ref{lemma:deltanet-finite-alphabet-parameter-sensitivity-growth}, computed with the official DeltaNet recurrence (flash-linear-attention), with $c=2$, $\beta_\star=0.9$, $u_t=(-1)^{t+1}$, $\lambda_\star=0$, and $x_0=(0,0)$. The reference-state trajectory is shown in~(a), and its sensitivity with respect to the scalar parameter coordinate $\lambda$ is shown in~(b).%
  }
  \vspace{1em}

  \begin{subfigure}{0.98\textwidth}
    \centering
    \tikzsetnextfilename{plot__parameter_sensitivity_c2_beta09__state}
    \begin{tikzpicture}
    \begin{axis}[
        width=12.5cm,
        height=8cm,
        axis lines=middle,
        xlabel={$t$},
        ylabel={$H^\star_{t,1}=a_t$},
        xlabel style={xshift=2pt, yshift=2pt, font=\footnotesize},
        ylabel style={xshift=-22pt, yshift=4pt, font=\footnotesize},
        xmin=0, xmax=50.8,
        ymin=-5.22, ymax=5.22,
        xtick={0,5,10,15,20,25,30,35,40,45,50},
        ytick={-4,-2,0,2,4},
        grid=none,
        legend style={
            at={(0.53,1.02)},
            anchor=north west,
            fill=white,
            draw=black,
            cells={anchor=west},
        },
        every x tick label/.append style={font=\footnotesize},
        every y tick label/.append style={font=\footnotesize},
        yticklabel style={text width=2.8em, align=right},
        tick label style={
          /pgf/number format/fixed,
          /pgf/number format/precision=2
        }
    ]

    \addplot[
        black,
        dashed,
        thin,
        domain=0:52,
        samples=2,
        forget plot
    ] {4.500000};

    \addplot[
        orange,
        thick,
        mark=*,
        mark size=1.3pt
    ] coordinates {
        (0, 0.0000)
        (1, 0.9000)
        (2, -1.6200)
        (3, 2.1960)
        (4, -2.6568)
        (5, 3.0254)
        (6, -3.3204)
        (7, 3.5563)
        (8, -3.7450)
        (9, 3.8960)
        (10, -4.0168)
        (11, 4.1135)
        (12, -4.1908)
        (13, 4.2526)
        (14, -4.3021)
        (15, 4.3417)
        (16, -4.3733)
        (17, 4.3987)
        (18, -4.4189)
        (19, 4.4351)
        (20, -4.4481)
        (21, 4.4585)
        (22, -4.4668)
        (23, 4.4734)
        (24, -4.4787)
        (25, 4.4830)
        (26, -4.4864)
        (27, 4.4891)
        (28, -4.4913)
        (29, 4.4930)
        (30, -4.4944)
        (31, 4.4955)
        (32, -4.4964)
        (33, 4.4971)
        (34, -4.4977)
        (35, 4.4982)
        (36, -4.4985)
        (37, 4.4988)
        (38, -4.4991)
        (39, 4.4993)
        (40, -4.4994)
        (41, 4.4995)
        (42, -4.4996)
        (43, 4.4997)
        (44, -4.4998)
        (45, 4.4998)
        (46, -4.4998)
        (47, 4.4999)
        (48, -4.4999)
        (49, 4.4999)
        (50, -4.4999)
    };
    \addlegendentry{\scriptsize trajectory of $H^\star_{t,1}$}

    \addplot[
        only marks,
        mark=*,
        mark size=1.8pt,
        black,
        forget plot
    ] coordinates {(0,0.0000)};
    \node[black, left] at (axis cs:0,0.0000) {$H^\star_{0,1}$};

    \addplot[
        black,
        dashed,
        thin,
        domain=0:52,
        samples=2
    ] {-4.500000};
    \addlegendentry{\scriptsize guides $\pm\frac{\beta_\star}{2-c\beta_\star}$}

    \end{axis}
    \end{tikzpicture}
    \caption{The reference-state component $H^\star_{t,1}=a_t$ remains bounded and approaches the alternating two-cycle $\pm\frac{\beta_\star}{2-c\beta_\star}\approx\pm4.5000$. The second state component is identically zero.}
  \end{subfigure}

  \vspace{0.4cm}

  \begin{subfigure}{0.98\textwidth}
    \centering
    \tikzsetnextfilename{plot__parameter_sensitivity_c2_beta09__sensitivity}
    \begin{tikzpicture}
    \begin{axis}[
        width=12.5cm,
        height=9cm,
        axis lines=middle,
        xlabel={$t$},
        ylabel={$g_t$},
        xlabel style={xshift=2pt, yshift=2pt, font=\footnotesize},
        ylabel style={xshift=-22pt, yshift=4pt, font=\footnotesize},
        xmin=0, xmax=50.8,
        ymin=0, ymax=442.3,
        xtick={0,5,10,15,20,25,30,35,40,45,50},
        ytick={0,100,200,300,400},
        grid=none,
        legend style={
            at={(0.04,0.98)},
            anchor=north west,
            fill=white,
            draw=black,
            cells={anchor=west},
        },
        every x tick label/.append style={font=\footnotesize},
        every y tick label/.append style={font=\footnotesize},
        yticklabel style={text width=2.8em, align=right},
        tick label style={
          /pgf/number format/fixed,
          /pgf/number format/precision=2
        }
    ]

    \addplot[
        blue,
        thick,
        mark=square*,
        mark size=1.1pt
    ] coordinates {
        (0, 0.0000)
        (1, 0.9000)
        (2, 3.4200)
        (3, 7.2360)
        (4, 12.0888)
        (5, 17.7710)
        (6, 24.1168)
        (7, 30.9935)
        (8, 38.2948)
        (9, 45.9358)
        (10, 53.8486)
        (11, 61.9789)
        (12, 70.2831)
        (13, 78.7265)
        (14, 87.2812)
        (15, 95.9249)
        (16, 104.6399)
        (17, 113.4119)
        (18, 122.2295)
        (19, 131.0836)
        (20, 139.9669)
        (21, 148.8735)
        (22, 157.7988)
        (23, 166.7390)
        (24, 175.6912)
        (25, 184.6529)
        (26, 193.6223)
        (27, 202.5979)
        (28, 211.5783)
        (29, 220.5626)
        (30, 229.5501)
        (31, 238.5400)
        (32, 247.5320)
        (33, 256.5256)
        (34, 265.5204)
        (35, 274.5163)
        (36, 283.5130)
        (37, 292.5104)
        (38, 301.5083)
        (39, 310.5066)
        (40, 319.5053)
        (41, 328.5042)
        (42, 337.5033)
        (43, 346.5026)
        (44, 355.5021)
        (45, 364.5016)
        (46, 373.5013)
        (47, 382.5010)
        (48, 391.5008)
        (49, 400.5006)
        (50, 409.5004)
    };
    \addlegendentry{\scriptsize parameter sensitivity $g_t$}

    \addplot[
        black,
        dashed,
        thin,
        domain=0:52,
        samples=2
    ] {0.9*x};
    \addlegendentry{\scriptsize lower bound $\beta_\star t=0.9 t$}

    \addplot[
        only marks,
        mark=*,
        mark size=1.8pt,
        black,
        forget plot
    ] coordinates {(0,0.0000)};
    \node[black, left] at (axis cs:0,0.0000) {$g_0$};

    \end{axis}
    \end{tikzpicture}
    \caption{The parameter sensitivity $g_t\defeq\big\|\pdvop{H_t}{\lambda_\star}(\lambda_\star)\big\|$ grows without bound and satisfies $g_t\geq\beta_\star t$.}
  \end{subfigure}

  \label{fig:generated-parameter-sensitivity-c2-beta09}
\end{figure}
\FloatBarrier
\clearpage

\subsection{BIBLG Instability of DeltaNet}
\label{subsec:deltanet--divergent-loss-gradient}

\todo{}

\todo{}

We show that DeltaNet's loss gradients are not bounded in general. 
In particular, we show that already 
DeltaNet Models consisting of a single recurrent unit show such behaviour, and do so already when
their input space consists of two elements only.
The proof makes use of state trajectories built in
Lemma~\ref{lemma:deltanet-finite-alphabet-parameter-sensitivity-growth}
of~\S\ref{subsec:technical-lemmas-for-deltanet-instability}, which are bounded but yield a
parameter-sensitivity that grows at least linearly with sequence length.

The formulation of the results makes use of notions introduced in \S\ref{subsec:properties-biblg}.

\begin{closingdefinition}[Nonzero local derivative condition for loss functions]
  \label{def:nonzero-local-derivative}
  A loss function $\lossfn: Y \times \targetspace{Y} \to \reals$
  with $Y = \reals^d$
  satisfies the \emph{nonzero local derivative condition}
  if there exist
  $a, y_\star\in Y$ and
  $\hat y_\star \in \targetspace{Y}$
  such that
  \begin{equation*}
    a
    \cdot
    {
      \pdvop{%
        \lossfn(y_\star,\hat y_\star)%
      }{%
        y_\star%
      }
    }
    \,\neq\,
    0
    \,.
  \end{equation*}
  In this case,
  we say that the 
  nonzero local derivative condition is \emph{witnessed by}
  $\ptuple{a, y_\star, \hat y_\star}$.
\end{closingdefinition}

\begin{restatable}[%
    Linear loss-gradient growth of DeltaNet (already on finite input spaces)%
  ]{%
    closingtheorem%
  }{%
    thmdeltanetlinearlossgradientgrowth%
  }
  \label{thm:deltanet-linear-loss-gradient-growth}
  Let $c\in\{1,2\}$,
  let $\beta_\star\in(0,1)$,
  and 
  let $\lossfn : Y \times \targetspace{Y} \to \reals$ be an analytic loss function
  with $Y \defeq \reals^{d}$ and $\targetspace{Y} \subseteq \reals^{d_\mathrm{target}}$.

  Suppose that $\lossfn$ satisfies the nonzero local derivative condition as witnessed by
  $\ptuple{a, y_\star, \hat y_\star}$.

  For some $\epsilon \in \posreals$,
  a DeltaNet model 
  $S \in \classmodels{DeltaNet}[c]$,
  with 
  input space $U$, output space $Y = \reals^{d}$, parameter space $\Theta$ satisfying 
  \begin{align*}
    U_\star \defeq \{{-}1,1\} \,\subseteq\, U\,,
    \qquad\;
    \Theta_\star \,\subseteq\,\Theta\,,
  \end{align*}
  where $\Theta_\star$ is as per
  Construction~\ref{construction:deltanet-finite-alphabet-parameter-variation} 
  under the present $\beta_\star$ and $\epsilon$,
  for which
  there exist
  a radius $r\in\posreals$,
  an input sequence $u \in \infseqs{U_\star}$,
  a parameter $\theta_\star \in \Theta_\star$,
  and 
  a constant $C\in\posreals$
  such that, 
  for every $T\in\posnaturals$,
  we have that
  \begin{equation*}
    z_T
    \defeq
    \ptuple{
      \seq{u}{1}{T},
      \hat y_\star
      }
    \end{equation*}
    is an admissible sample for $S$, and additionally:
    \begin{enumerate}
      \item
        the sample $z_T$ and parameter $\theta_\star$ belong to their corresponding truncated sets:
        \begin{equation*}
          z_T
          \in
          \horizontruncsamplespacep{S}{r}{T}\,,
          \qquad\quad
          \theta_\star
          \in
          \truncation{\Theta}{r}\,;
        \end{equation*}
      \item
        the objective function $\Lossfn{S,z_T}[\lossfn]$ is differentiable at $\, \theta_\star$;
      \item
        the loss gradient is bounded from below as follows:
        \begin{equation}
          \label{eq:thm:deltanet-linear-loss-gradient-growth}
          \big\|
          \dvop{\Lossfn{S,z_T}}[\lossfn](\theta_\star)
          \big\|
          \geq
          C
          \cdot
          T\,.
        \end{equation}
    \end{enumerate}
  Consequently,
  for $r$ the above radius,
  the growth of the corresponding loss-gradients norm
  is asymptotically lower-bounded as
  \begin{equation*}
    \LossGrowth{S}{\lossfn}{r}(T)
    \,=\,
    \Omega(T) \,.
  \end{equation*}

  In particular,
  $S$
  does not satisfy BIBLG,
  and hence
  $\classmodels{DeltaNet}[c]$
  does not satisfy BIBLG,
  even when the input space is restricted to a fixed two-symbol alphabet.
\end{restatable}
\begin{proof}
  Deferred to \S\ref{subsec:deltanet--deferred-proofs--biblg}.
\end{proof}

\subsubsection{DeltaNet: Divergent Loss-Gradient for Standard Loss Functions}
\label{subsubsec:deltanet-loss-gradient-divergence-standard-loss-functions}

\begin{closingdefinition}[Standard loss functions]
  \label{def:standard-loss-functions}
  We introduce the following loss functions.
  \begin{definitionpart}[title={Squared error}, light]
    The \emph{squared-error loss} is the function
    \begin{equation*}
      \seloss
      :
      \reals^d
      \times
      \reals^d
      \to
      \nonnegreals
      \,,
      \qquad\qquad
      \seloss(y,\hat y)
      \defeq
      \frac{1}{2}
      \|
      y-\hat y
      \|^2
      \,.
    \end{equation*}
  \end{definitionpart}

  \begin{definitionpart}[title={Binary cross-entropy}, light]
    The \emph{binary cross-entropy loss with logits}, 
    or simply \emph{binary cross-entropy loss},
    is the function%
    \footnote{%
      Note that $\sigma(x) \defeq 1/\big(1+\exp(-x)\big)$ is the logistic sigmoid function.%
    }
    \begin{equation*}
      \bceloss
      :
      \reals
      \times
      \{0,1\}
      \to
      \nonnegreals
      \,,
      \qquad\qquad
      \begin{aligned}
        \bceloss(y,\hat y)
        \defeq
        &
        \log
        \big(
          1+\exp(y)
        \big)
        -
        \hat y \cdot y
        \,.
      \end{aligned}
    \end{equation*}
  \end{definitionpart}

  \begin{definitionpart}[title={Cross-entropy}, light]
    For
    $K\in\posnaturals$ with $K \geq 2$,
    the \emph{$K$-class softmax cross-entropy loss with logits}, 
    or simply \emph{$K$-class cross-entropy loss}, 
    is the function
    \begin{equation*}
      \softmaxceloss[K]
      :
      \reals^K
      \times
      \iival{1}{K}
      \to
      \nonnegreals
      \,,
      \qquad\qquad
      \begin{aligned}
        \softmaxceloss[K](y,k)
        \defeq
        {-}
        \log\left(
          \dfrac{
          \exp(y_k)
          }{
          \sum_{j=1}^{K}
          \exp(y_j)
          }
        \right)
        \,.
      \end{aligned}
    \end{equation*}
  \end{definitionpart}
  We call each of the above loss functions a \emph{standard loss function}.
\end{closingdefinition}

\begin{restatable}[%
    Standard losses satisfy the nonzero local derivative condition%
  ]{%
    closinglemma%
  }{%
    lemmastandardlossesnonzerolocalderivative%
  }
  \label{lemma:standard-losses-nonzero-local-derivative}

  Each of the standard loss functions is analytic and hence admissible for every system having
  matching output space.
  Furthermore,
  each of the standard loss functions satisfies the nonzero local derivative condition:%
  \footnote{%
    The \emph{nonzero local derivative condition} is introduced in
    Definition~\ref{def:nonzero-local-derivative}.%
  }
  \begin{enumerate}
    \item
      the squared-error loss $\seloss : \reals^d \times \reals^d \to \reals$
      satisfies the 
      nonzero local derivative condition
      as witnessed by
      $\ptuple{ a_{\selossName},\, y_{\selossName},\, \hat y_{\selossName}} 
      \defeq \ptuple{ e_1, e_1, 0 }$,
      and specifically
      \begin{equation}
        \label{eq:lemma:standard-losses-se-derivative}
        \pdvop{\seloss}{y_{\selossName}}
        \big(
          y_{\selossName},\, \hat y_{\selossName}
        \big)
        \cdot
        a_{\selossName}
        \,=\,
        1\,;
      \end{equation}

    \item
      the binary cross-entropy loss $\bceloss : \reals \times \{ 0,1 \} \to \reals$
      satisfies the 
      nonzero local derivative condition
      as witnessed by
      $\ptuple{ a_{\bcelossName},\, y_{\bcelossName},\, \hat y_{\bcelossName}} 
      \defeq \ptuple{ 1, 0, 0 }$,
      and specifically
      \begin{equation}
        \label{eq:lemma:standard-losses-bce-derivative}
        \pdvop{\bceloss}{y_{\bcelossName}}
        \big(
          y_{\bcelossName},\, \hat y_{\bcelossName}
        \big)
        \cdot
        a_{\bcelossName}
        \,=\,
        \frac{1}{2}
        \,;
      \end{equation}

    \item
      for every
      $K\in\posnaturals$
      with
      $K\geq2$,
      the $K$-class cross-entropy loss $\softmaxceloss[K] : \reals^K \times \iival{1}{K} \to \reals$
      satisfies the 
      nonzero local derivative condition
      as witnessed by
      $\ptuple{ a_{\softmaxcelossName},\, y_{\softmaxcelossName},\, \hat y_{\softmaxcelossName}} 
      \defeq \ptuple{ e_2, 0, 1 }$,
      and specifically
      \begin{equation}
        \label{eq:lemma:standard-losses-ce-derivative}
        \pdvop{\softmaxceloss[K]}{y_{\softmaxcelossName}}
        \big(
          y_{\softmaxcelossName},\, \hat y_{\softmaxcelossName}
        \big)
        \cdot
        a_{\softmaxcelossName}
        \,=\,
        \frac{1}{K}
        \,.
      \end{equation}
  \end{enumerate}
\end{restatable}
\begin{proof}
  Deferred to \S\ref{subsec:deltanet--deferred-proofs--biblg}.
\end{proof}

\begin{restatable}[%
    Finite-alphabet loss-gradient growth for standard loss functions%
  ]{%
    closingtheorem%
  }{%
    thmdeltanetstandardlossgradientgrowth%
  }
  \label{thm:deltanet-standard-loss-gradient-growth}
  Let
  $c\in\{1,2\}$,
  let $\beta_\star\in(0,1)$,
  and 
  let $U_\star \defeq \{{-}1,1\}$.
  For each of the standard loss functions of Definition~\ref{def:standard-loss-functions},
  there exists a DeltaNet model
  $S \in \classmodels{DeltaNet}[c]$ having input space $U_\star\subseteq U$ and admitting
  \begin{align*}
    \text{radius}\;\,
    r\in\posreals\,,
    \qquad\quad
    \text{parameters}\;\,
    \theta_\star \in \truncation{\Theta}{r}\,,
    \qquad\quad
    \text{sample}\;\,
    z_T \in \horizontruncsamplespacep{S}{r}{T}
    \;\,\text{for each}\; T \in \timedomain\,,
  \end{align*}
  such that:
  \begin{align}
    \big\|
      \dvop{\Lossfn{S,z_T}[\seloss]}(\theta_\star)
    \big\|
    &\,\geq\,
    \beta_\star \cdot T
    &&
    \text{for squared error},
    \label{eq:thm:deltanet-standard-loss-gradient-growth-se}
    \\[.2em]
    \mBig\|
      \dvop{\Lossfn{S,z_T}[\bceloss]}(\theta_\star)
    \big\|
    &\,\geq\,
    \frac{\beta_\star}{2} \cdot T
    &&
    \text{for binary cross-entropy},
    \label{eq:thm:deltanet-standard-loss-gradient-growth-bce}
    \\[.2em]
    \big\|
      \dvop{
        \Lossfn{S,z_T}[\softmaxceloss[K]]
      }(\theta_\star)
    \big\|
    &\,\geq\,
    \frac{\beta_\star}{K} \cdot T
    &&
    \text{for $K$-class cross-entropy, $\forall K \geq 2$}.
    \label{eq:thm:deltanet-standard-loss-gradient-growth-ce}
  \end{align}

  Consequently, 
  \begin{itemize}
    \item
      the loss-gradient growth is 
      \begin{equation*}
        \LossGrowth{S}{\lossfn}{r}(T) = \Omega(T)
        \qquad\qquad
        \forall \lossfn \in \{\seloss, \bceloss \} \cup \{ \softmaxceloss[K] \}_{K \geq 2},
      \end{equation*}
    \item
      the class $\classmodels{DeltaNet}[c]$ does not satisfy BIBLG.
  \end{itemize}
\end{restatable}
\begin{proof}
  Deferred to \S\ref{subsec:deltanet--deferred-proofs--biblg}.
\end{proof}

\begin{closingremark}
  Theorem~\ref{thm:deltanet-standard-loss-gradient-growth} shows that DeltaNet models can exhibit
  an unbounded loss gradient even when their input space consists of two elements,
  for any choice a loss function among the standard ones we have considered.
\end{closingremark}

\begin{figure}[ht]
  \centering
  \caption{%
    Loss-gradient norms $\|\nabla_\theta L_T(\theta_\star)\|$ of the one-unit DeltaNet model of
    Theorem~\ref{thm:deltanet-linear-loss-gradient-growth} on input prefixes $u_{1:T}$,
    $u_t=(-1)^{t+1}$, for the squared-error, binary cross-entropy and $3$-class cross-entropy losses
    (Theorem~\ref{thm:deltanet-standard-loss-gradient-growth}), computed by automatic differentiation through the official DeltaNet recurrence (flash-linear-attention), with $c=1$ and $\beta_\star=0.9$. Dashed lines are the lower bounds $\beta_\star T$, $\frac{\beta_\star}{2}T$, and $\frac{\beta_\star}{3}T$, respectively.%
  }
  \vspace{1em}

  \begin{subfigure}{0.98\textwidth}
    \centering
    \tikzsetnextfilename{plot__loss_gradient_c1_beta09_K3}
    \begin{tikzpicture}
    \begin{axis}[
        width=12.5cm,
        height=9cm,
        axis lines=middle,
        xlabel={$t$},
        ylabel={$\|\nabla_\theta L_T\|$},
        xlabel style={xshift=2pt, yshift=2pt, font=\footnotesize},
        ylabel style={xshift=-22pt, yshift=4pt, font=\footnotesize},
        xmin=0, xmax=50.8,
        ymin=0, ymax=87.65,
        xtick={0,5,10,15,20,25,30,35,40,45,50},
        ytick={0,20,40,60,80},
        grid=none,
        legend style={
            at={(0.06,0.93)},
            anchor=north west,
            fill=white,
            draw=black,
            cells={anchor=west},
        },
        every x tick label/.append style={font=\footnotesize},
        every y tick label/.append style={font=\footnotesize},
        yticklabel style={text width=2.8em, align=right},
        tick label style={
          /pgf/number format/fixed,
          /pgf/number format/precision=2
        }
    ]

    \addplot[
        orange,
        semithick,
        mark=*,
        mark size=1.1pt
    ] coordinates {
        (0, 0.0000)
        (1, 1.2728)
        (2, 2.7328)
        (3, 4.3174)
        (4, 5.9328)
        (5, 7.5568)
        (6, 9.1853)
        (7, 10.8161)
        (8, 12.4484)
        (9, 14.0816)
        (10, 15.7155)
        (11, 17.3499)
        (12, 18.9846)
        (13, 20.6195)
        (14, 22.2547)
        (15, 23.8900)
        (16, 25.5255)
        (17, 27.1611)
        (18, 28.7967)
        (19, 30.4325)
        (20, 32.0683)
        (21, 33.7041)
        (22, 35.3400)
        (23, 36.9760)
        (24, 38.6120)
        (25, 40.2480)
        (26, 41.8840)
        (27, 43.5201)
        (28, 45.1562)
        (29, 46.7923)
        (30, 48.4284)
        (31, 50.0645)
        (32, 51.7007)
        (33, 53.3368)
        (34, 54.9730)
        (35, 56.6092)
        (36, 58.2454)
        (37, 59.8816)
        (38, 61.5178)
        (39, 63.1541)
        (40, 64.7903)
        (41, 66.4265)
        (42, 68.0628)
        (43, 69.6990)
        (44, 71.3353)
        (45, 72.9715)
        (46, 74.6078)
        (47, 76.2440)
        (48, 77.8803)
        (49, 79.5166)
        (50, 81.1529)
    };
    \addlegendentry{\scriptsize $\|\nabla_\theta L_T\|$ for $\ell_{\mathrm{SE}}$}

    \addplot[
        orange,
        dashed,
        thin,
        domain=0:52,
        samples=2
    ] {0.900000*x};
    \addlegendentry{\scriptsize bound $\beta_\star T$}

    \addplot[
        blue,
        semithick,
        mark=square*,
        mark size=1.1pt
    ] coordinates {
        (0, 0.0000)
        (1, 0.6364)
        (2, 1.3664)
        (3, 2.1587)
        (4, 2.9664)
        (5, 3.7784)
        (6, 4.5926)
        (7, 5.4081)
        (8, 6.2242)
        (9, 7.0408)
        (10, 7.8578)
        (11, 8.6749)
        (12, 9.4923)
        (13, 10.3098)
        (14, 11.1274)
        (15, 11.9450)
        (16, 12.7628)
        (17, 13.5805)
        (18, 14.3984)
        (19, 15.2162)
        (20, 16.0341)
        (21, 16.8521)
        (22, 17.6700)
        (23, 18.4880)
        (24, 19.3060)
        (25, 20.1240)
        (26, 20.9420)
        (27, 21.7600)
        (28, 22.5781)
        (29, 23.3961)
        (30, 24.2142)
        (31, 25.0323)
        (32, 25.8503)
        (33, 26.6684)
        (34, 27.4865)
        (35, 28.3046)
        (36, 29.1227)
        (37, 29.9408)
        (38, 30.7589)
        (39, 31.5770)
        (40, 32.3951)
        (41, 33.2133)
        (42, 34.0314)
        (43, 34.8495)
        (44, 35.6676)
        (45, 36.4858)
        (46, 37.3039)
        (47, 38.1220)
        (48, 38.9402)
        (49, 39.7583)
        (50, 40.5764)
    };
    \addlegendentry{\scriptsize $\|\nabla_\theta L_T\|$ for $\ell_{\mathrm{BCE}}$}

    \addplot[
        blue,
        dashed,
        thin,
        domain=0:52,
        samples=2
    ] {0.450000*x};
    \addlegendentry{\scriptsize bound $\frac{\beta_\star}{2}T$}

    \addplot[
        red!70!black,
        semithick,
        mark=triangle*,
        mark size=1.1pt
    ] coordinates {
        (0, 0.0000)
        (1, 0.4243)
        (2, 0.9109)
        (3, 1.4391)
        (4, 1.9776)
        (5, 2.5189)
        (6, 3.0618)
        (7, 3.6054)
        (8, 4.1495)
        (9, 4.6939)
        (10, 5.2385)
        (11, 5.7833)
        (12, 6.3282)
        (13, 6.8732)
        (14, 7.4182)
        (15, 7.9633)
        (16, 8.5085)
        (17, 9.0537)
        (18, 9.5989)
        (19, 10.1442)
        (20, 10.6894)
        (21, 11.2347)
        (22, 11.7800)
        (23, 12.3253)
        (24, 12.8707)
        (25, 13.4160)
        (26, 13.9613)
        (27, 14.5067)
        (28, 15.0521)
        (29, 15.5974)
        (30, 16.1428)
        (31, 16.6882)
        (32, 17.2336)
        (33, 17.7789)
        (34, 18.3243)
        (35, 18.8697)
        (36, 19.4151)
        (37, 19.9605)
        (38, 20.5059)
        (39, 21.0513)
        (40, 21.5968)
        (41, 22.1422)
        (42, 22.6876)
        (43, 23.2330)
        (44, 23.7784)
        (45, 24.3238)
        (46, 24.8693)
        (47, 25.4147)
        (48, 25.9601)
        (49, 26.5055)
        (50, 27.0510)
    };
    \addlegendentry{\scriptsize $\|\nabla_\theta L_T\|$ for $\ell_{\mathrm{CE}_{3}}$}

    \addplot[
        red!70!black,
        dashed,
        thin,
        domain=0:52,
        samples=2
    ] {0.300000*x};
    \addlegendentry{\scriptsize bound $\frac{\beta_\star}{3}T$}

    \end{axis}
    \end{tikzpicture}
    \caption{The loss-gradient norm grows at least linearly in the sequence length $T$, for every standard loss.}
  \end{subfigure}

  \label{fig:generated-loss-gradient-c1-beta09-K3}
\end{figure}
\begin{figure}[ht]
  \centering
  \caption{%
    Loss-gradient norms $\|\nabla_\theta L_T(\theta_\star)\|$ of the one-unit DeltaNet model of
    Theorem~\ref{thm:deltanet-linear-loss-gradient-growth} on input prefixes $u_{1:T}$,
    $u_t=(-1)^{t+1}$, for the squared-error, binary cross-entropy and $3$-class cross-entropy losses
    (Theorem~\ref{thm:deltanet-standard-loss-gradient-growth}), computed by automatic differentiation through the official DeltaNet recurrence (flash-linear-attention), with $c=2$ and $\beta_\star=0.9$. Dashed lines are the lower bounds $\beta_\star T$, $\frac{\beta_\star}{2}T$, and $\frac{\beta_\star}{3}T$, respectively.%
  }
  \vspace{1em}

  \begin{subfigure}{0.98\textwidth}
    \centering
    \tikzsetnextfilename{plot__loss_gradient_c2_beta09_K3}
    \begin{tikzpicture}
    \begin{axis}[
        width=12.5cm,
        height=9cm,
        axis lines=middle,
        xlabel={$t$},
        ylabel={$\|\nabla_\theta L_T\|$},
        xlabel style={xshift=2pt, yshift=2pt, font=\footnotesize},
        ylabel style={xshift=-22pt, yshift=4pt, font=\footnotesize},
        xmin=0, xmax=50.8,
        ymin=0, ymax=442.3,
        xtick={0,5,10,15,20,25,30,35,40,45,50},
        ytick={0,100,200,300,400},
        grid=none,
        legend style={
            at={(0.06,0.93)},
            anchor=north west,
            fill=white,
            draw=black,
            cells={anchor=west},
        },
        every x tick label/.append style={font=\footnotesize},
        every y tick label/.append style={font=\footnotesize},
        yticklabel style={text width=2.8em, align=right},
        tick label style={
          /pgf/number format/fixed,
          /pgf/number format/precision=2
        }
    ]

    \addplot[
        orange,
        semithick,
        mark=*,
        mark size=1.1pt
    ] coordinates {
        (0, 0.0000)
        (1, 1.2728)
        (2, 3.7843)
        (3, 7.5619)
        (4, 12.3773)
        (5, 18.0267)
        (6, 24.3443)
        (7, 31.1968)
        (8, 38.4775)
        (9, 46.1007)
        (10, 53.9983)
        (11, 62.1153)
        (12, 70.4080)
        (13, 78.8413)
        (14, 87.3871)
        (15, 96.0231)
        (16, 104.7313)
        (17, 113.4972)
        (18, 122.3094)
        (19, 131.1586)
        (20, 140.0376)
        (21, 148.9402)
        (22, 157.8620)
        (23, 166.7990)
        (24, 175.7483)
        (25, 184.7074)
        (26, 193.6743)
        (27, 202.6476)
        (28, 211.6259)
        (29, 220.6084)
        (30, 229.5941)
        (31, 238.5824)
        (32, 247.5728)
        (33, 256.5650)
        (34, 265.5585)
        (35, 274.5532)
        (36, 283.5487)
        (37, 292.5450)
        (38, 301.5419)
        (39, 310.5392)
        (40, 319.5370)
        (41, 328.5350)
        (42, 337.5333)
        (43, 346.5319)
        (44, 355.5306)
        (45, 364.5294)
        (46, 373.5284)
        (47, 382.5275)
        (48, 391.5266)
        (49, 400.5259)
        (50, 409.5252)
    };
    \addlegendentry{\scriptsize $\|\nabla_\theta L_T\|$ for $\ell_{\mathrm{SE}}$}

    \addplot[
        orange,
        dashed,
        thin,
        domain=0:52,
        samples=2
    ] {0.900000*x};
    \addlegendentry{\scriptsize bound $\beta_\star T$}

    \addplot[
        blue,
        semithick,
        mark=square*,
        mark size=1.1pt
    ] coordinates {
        (0, 0.0000)
        (1, 0.6364)
        (2, 1.8921)
        (3, 3.7809)
        (4, 6.1887)
        (5, 9.0134)
        (6, 12.1722)
        (7, 15.5984)
        (8, 19.2387)
        (9, 23.0504)
        (10, 26.9991)
        (11, 31.0576)
        (12, 35.2040)
        (13, 39.4206)
        (14, 43.6936)
        (15, 48.0116)
        (16, 52.3656)
        (17, 56.7486)
        (18, 61.1547)
        (19, 65.5793)
        (20, 70.0188)
        (21, 74.4701)
        (22, 78.9310)
        (23, 83.3995)
        (24, 87.8741)
        (25, 92.3537)
        (26, 96.8372)
        (27, 101.3238)
        (28, 105.8130)
        (29, 110.3042)
        (30, 114.7970)
        (31, 119.2912)
        (32, 123.7864)
        (33, 128.2825)
        (34, 132.7793)
        (35, 137.2766)
        (36, 141.7744)
        (37, 146.2725)
        (38, 150.7709)
        (39, 155.2696)
        (40, 159.7685)
        (41, 164.2675)
        (42, 168.7667)
        (43, 173.2659)
        (44, 177.7653)
        (45, 182.2647)
        (46, 186.7642)
        (47, 191.2637)
        (48, 195.7633)
        (49, 200.2629)
        (50, 204.7626)
    };
    \addlegendentry{\scriptsize $\|\nabla_\theta L_T\|$ for $\ell_{\mathrm{BCE}}$}

    \addplot[
        blue,
        dashed,
        thin,
        domain=0:52,
        samples=2
    ] {0.450000*x};
    \addlegendentry{\scriptsize bound $\frac{\beta_\star}{2}T$}

    \addplot[
        red!70!black,
        semithick,
        mark=triangle*,
        mark size=1.1pt
    ] coordinates {
        (0, 0.0000)
        (1, 0.4243)
        (2, 1.2614)
        (3, 2.5206)
        (4, 4.1258)
        (5, 6.0089)
        (6, 8.1148)
        (7, 10.3989)
        (8, 12.8258)
        (9, 15.3669)
        (10, 17.9994)
        (11, 20.7051)
        (12, 23.4693)
        (13, 26.2804)
        (14, 29.1290)
        (15, 32.0077)
        (16, 34.9104)
        (17, 37.8324)
        (18, 40.7698)
        (19, 43.7195)
        (20, 46.6792)
        (21, 49.6468)
        (22, 52.6207)
        (23, 55.5997)
        (24, 58.5828)
        (25, 61.5691)
        (26, 64.5581)
        (27, 67.5492)
        (28, 70.5420)
        (29, 73.5361)
        (30, 76.5314)
        (31, 79.5275)
        (32, 82.5243)
        (33, 85.5217)
        (34, 88.5195)
        (35, 91.5177)
        (36, 94.5162)
        (37, 97.5150)
        (38, 100.5140)
        (39, 103.5131)
        (40, 106.5123)
        (41, 109.5117)
        (42, 112.5111)
        (43, 115.5106)
        (44, 118.5102)
        (45, 121.5098)
        (46, 124.5095)
        (47, 127.5092)
        (48, 130.5089)
        (49, 133.5086)
        (50, 136.5084)
    };
    \addlegendentry{\scriptsize $\|\nabla_\theta L_T\|$ for $\ell_{\mathrm{CE}_{3}}$}

    \addplot[
        red!70!black,
        dashed,
        thin,
        domain=0:52,
        samples=2
    ] {0.300000*x};
    \addlegendentry{\scriptsize bound $\frac{\beta_\star}{3}T$}

    \end{axis}
    \end{tikzpicture}
    \caption{The loss-gradient norm grows at least linearly in the sequence length $T$, for every standard loss.}
  \end{subfigure}

  \label{fig:generated-loss-gradient-c2-beta09-K3}
\end{figure}
\FloatBarrier
\clearpage

\FloatBarrier
\clearpage

\subsection{DeltaNet: Technical Lemmas on Instability}
\label{subsec:technical-lemmas-for-deltanet-instability}

\subsubsection{DeltaNet: Divergent state trajectories}
\label{subsubsec:technical-lemmas-for-deltanet-instability--divergent-state-trajectories}

\begin{restatable}[%
    \textit{Aux.\ for Lemma~\ref{lemma:divergent-delta-rule-beta}:}~%
    Initial-state comparison for nonexpansive affine recurrences%
  ]{%
    closinglemma%
  }{%
    lemmanonexpansiveaffineinitialstatecomparison%
  }
  \label{lemma:nonexpansive-affine-initial-state-comparison}
  Let us consider 
  a sequence of vectors
  $(b_t)_{t \in \timedomain}$
  and
  a sequence of matrices
  $(A_t)_{t \in \timedomain}$ 
  such that, for every $t \in \timedomain$,
  \begin{equation*}
    b_t \in \reals^d,
    \quad\qquad
    A_t \in \reals^{d\times d}
    \quad\text{with}\quad
    \|A_t\|
    \leq
    1.
  \end{equation*}
  Let $x_0,\tilde x_0 \in \reals^d$,
  and
  let the sequences $(x_t)_{t \in \naturals}$ and $(\tilde x_t)_{t \in \naturals}$
  be defined as follows, for every $t \in \timedomain$,
  \begin{align*}
    x_t
    \,\defeq\,
    A_t \cdot x_{t-1}
    +
    b_t,
    \qquad\quad
    \tilde x_t
    \,\defeq\,
    A_t \cdot \tilde x_{t-1}
    +
    b_t.
  \end{align*}
  Then, for every $t \in \naturals$,
  \begin{equation}
    \label{eq:nonexpansive-affine-initial-state-comparison}
    \|x_t-\tilde x_t\|
    \,\leq\,
    \|x_0-\tilde x_0\|.
  \end{equation}
  Consequently, 
  if
  $\|\tilde x_t\| \to \infty$
  as
  $t\to\infty$,
  then
  $\|x_t\| \to \infty$ 
  as
  $t\to\infty$.
\end{restatable}

\begin{restatable}[%
    \textit{Aux.~for~Thm.~\ref{thm:divergent-delta-rule-beta}:}~%
    Divergent state trajectories of DeltaNet%
  ]{%
    closinglemma%
  }{%
    lemmadivergentdeltarulebeta%
  }
  \label{lemma:divergent-delta-rule-beta}
  Let $\beta \in (0,1)$,
  and let us define the following quantities for all $t \geq 1$,
  \begin{align*}
    \beta_t\defeq \beta
    \qquad
    \epsilon_t \defeq
    \begin{cases}
      \frac{1}{t+2} & t \text{ odd},\\[.5em]
      \frac{1}{2(t+2)} & t \text{ even},
    \end{cases}
    \qquad
    k_t \defeq
    \frac{1}{\sqrt{1+\epsilon_t^2}}
    \cdot
    \begin{pmatrix}
      1\\
      \epsilon_t
    \end{pmatrix}
    \qquad
    v_t \defeq
    \begin{cases}
      0, & t \text{ odd},\\
      1, & t \text{ even}.
    \end{cases}
  \end{align*}
  Let $x_0 \defeq (0,0)$ and let $(x_t)_{t \geq 1}$ be the state sequence
  defined by the recurrence
  \begin{align*}
    x_t
    \,\defeq\,
    \big(I-\beta_t \cdot k_t \cdot k_t^\top\big) \cdot x_{t-1}
    \,+\,
    \beta_t \cdot k_t \cdot v_t^\transposeop
    \ .
  \end{align*}
  Then $\|x_t\|\to\infty$ as $t\to\infty$.
  In particular,
  the values $x_{2n} = (x_{2n,1},x_{2n,2})$ at even steps and the values
  $x_{2n-1} = (x_{2n-1,1},x_{2n-1,2})$ at odd steps are as follows,
  \begin{align*}
    x_{2n,1}
    &=
    \frac{1}{2-\beta}
    +
    O\left(\frac{\log n}{n}\right),
    &
    x_{2n-1,1}
    & =
    \frac{1-\beta}{2-\beta}
    +
    O\left(\frac{\log n}{n}\right),
    \\[.5em]
    x_{2n,2}
    &=
    -\frac{\beta}{4(2-\beta)}\log n
    +
    O(1),
    &
    x_{2n-1,2}
    & =
    -\frac{\beta}{4(2-\beta)}\log n
    +
    O(1).
  \end{align*}
  Note that $\frac{\beta}{4(2-\beta)} > 0$.
\end{restatable}

\subsubsection{DeltaNet: Divergent parameter-sensitivity of DeltaNet on finite input spaces}
\label{subsubsec:technical-lemmas-for-deltanet-instability--state-trajectories-for-gradient}

\begin{construction}[%
    \textit{Aux.~for~Lemma~\ref{lemma:deltanet-finite-alphabet-parameter-sensitivity-growth}:}~%
    Specialised DeltaNet parametrisation%
  ]
  \label{construction:deltanet-finite-alphabet-parameter-variation}
  Let 
  $c\in\{1,2\}$
  $\epsilon\in\posreals$,
  and
  $\beta_\star\in(0,1)$.

  We define the \emph{specialised parametric DeltaNet Recurrent Unit}
  with hyperparameters $\ptuple{c, \epsilon, \beta_\star}$
  as the parametric Left-Multiplicative Linear Recurrent Unit
  \begin{equation*}
    D_\star
    \defeq
    \mctuple{X,U,f}{A,B}{\Lambda_\star}
    \qquad\text{with}\qquad
    d_\mathrm{in}
    =
    1\,,
    \quad
    d_\mathrm{k}
    =
    2\,,
    \quad
    d_\mathrm{v}
    =
    1\,,
  \end{equation*}
  and components as follows:
  as follows:
  \begin{itemize}
    \item
      state, input, and parameter spaces given respectively by
      \begin{equation*}
        X
        \defeq
        \reals^{2\times1}
        \cong
        \reals^2
        \,,
        \qquad\quad
        U
        \defeq
        \{-1,1\}
        \,,
        \qquad\quad
        \Lambda_\star
        \,\defeq\,
        \openballopp{\lambda_\star,\epsilon}{\reals}
        \quad\text{with}\quad
        \lambda_\star
        \,=\,
        0
        \,;
      \end{equation*}

    \item
      the parameter values induced by $\lambda\in\Lambda_\star$
      are
      \begin{align*}
        \pinst{W_\mathrm{k}}{\lambda}
        &\defeq
        \begin{pmatrix}
          0
          \\
          \lambda
        \end{pmatrix},
        &
        W_\mathrm{v}
        &\defeq
        1,
        &
        w_\beta
        &\defeq
        0,
        &
        b_\mathrm{k}
        &\defeq
        e_1
        =
        \begin{pmatrix}
          1 
          \\
          0
        \end{pmatrix}
        &
        b_\mathrm{v}
        &\defeq
        0,
        &
        b_\beta
        &\defeq
        \log
        \left(
        \frac{
          \beta_\star
          }{
          1-\beta_\star
        }
        \right)\,;
      \end{align*}

    \item
      the recurrence-input functions of
      are
      \begin{align*}
        \tilde k
        &:
        U\times\Lambda_\star
        \to
        \reals^2
        &
        \fpinst{\tilde k}{u}{\lambda}
        &\defeq
        \pinst{W_\mathrm{k}}{\lambda}
        \cdot
        u
        +
        b_\mathrm{k}
        &&\textit{key preactivation},
        \\[.3em]
        k
        &:
        U\times\Lambda_\star
        \to
        \mathbb S^1
        &
        \fpinst{k}{u}{\lambda}
        &\defeq
        \fpinst{\tilde k}{u}{\lambda}
        /
        \big\|
        \fpinst{\tilde k}{u}{\lambda}
        \big\|
        &&\textit{key},
        \\[.3em]
        \beta
        &:
        U\times\Lambda_\star
        \to
        (0,1)
        &
        \fpinst{\beta}{u}{\lambda}
        &\defeq
        \sigma
        \big(
        w_\beta\cdot u+b_\beta
        \big)
        &&\textit{write strength},
        \\[.3em]
        v
        &:
        U\times\Lambda_\star
        \to
        \reals
        &
        \fpinst{v}{u}{\lambda}
        &\defeq
        W_\mathrm{v}\cdot u+b_\mathrm{v}
        &&\textit{value};
      \end{align*}

    \item
      the state-update matrix and additive term are
      \begin{align*}
        \fpinst{A}{u}{\lambda}
        &\defeq
        I_2
        -
        c
        \cdot
        \fpinst{\beta}{u}{\lambda}
        \cdot
        \fpinst{k}{u}{\lambda}
        \cdot
        \fpinst{k}{u}{\lambda}^{\transposeop}
        \quad\text{and}\quad
        \fpinst{B}{u}{\lambda}
        &\defeq
        \fpinst{\beta}{u}{\lambda}
        \cdot
        \fpinst{k}{u}{\lambda}
        \cdot
        \fpinst{v}{u}{\lambda}\,.
      \end{align*}
  \end{itemize}
  Thus,
  the dynamics function is
  \begin{equation*}
    \fpinst{f}{H,u}{\lambda}
    \,=\,
    \fpinst{A}{u}{\lambda}
    \cdot
    H
    +
    \fpinst{B}{u}{\lambda}\,.
  \end{equation*}
\end{construction}

\begin{restatable}[%
    The specialised parametrisation is DeltaNet%
  ]{%
    closingfact%
  }{%
    factdeltanetfinitealphabetspecialisedparametrisation%
  }
  \label{fact:deltanet-finite-alphabet-specialised-parametrisation}

  The \emph{specialised parametric DeltaNet Recurrent Unit}
  $D_\star$
  with hyperparameters $\ptuple{c, \epsilon, \beta_\star}$
  of
  Construction~\ref{construction:deltanet-finite-alphabet-parameter-variation}
  is a DeltaNet Recurrent Unit with no decay gate and update hyperparameter $c$,
  under the specialised parametrisation 
  \begin{equation*}
    \vartheta 
    \,:\,
    \lambda 
    \longmapsto 
    \mBig(
      \big(
        \pinst{W_\mathrm{k}}{\lambda},
        b_\mathrm{k}
      \big),\;
      \big(
        W_\mathrm{v},
        b_\mathrm{v}
      \big),\;
      \big(
        w_\beta,
        b_\beta
      \big)
    \mBig)
    \qquad\quad\text{for}\quad
    \lambda \in \Lambda_\star\,.
  \end{equation*}
  Specifically,
  the standard DeltaNet Recurrent Unit with no decay gate,
  update hyperparameter~$c$,
  and 
  parameter space
  \begin{equation*}
    \Theta_\star
    \,\defeq\,
    \im(\vartheta)
    \,=\,
    \big\{
      \pinst{\theta}{\lambda}
      :
      \lambda\in\Lambda_\star
    \big\}
  \end{equation*}
  has
  precisely the dynamics of~$D_\star$
  under the parameter identification~$\vartheta$.
\end{restatable}

\begin{closingremark}
  Under the specialised parametrisation of
  Fact~\ref{fact:deltanet-finite-alphabet-specialised-parametrisation},
  the scalar parameter
  $\lambda$
  is precisely the
  $(2,1)$-entry of
  $\pinst{W_\mathrm{k}}{\lambda}$,
  while every remaining standard DeltaNet parameter coordinate is constant.
\end{closingremark}

\begin{restatable}[%
    \textit{Aux.~for~Thms.~\ref{thm:deltanet-does-not-satisfy-wbibsd},%
    ~\ref{thm:deltanet-linear-loss-gradient-growth}:}~%
    Parameter-sensitivity growth of DeltaNet on finite input spaces%
  ]{%
    closinglemma%
  }{%
    lemmadeltanetfinitealphabetparametersensitivitygrowth%
  }
  \label{lemma:deltanet-finite-alphabet-parameter-sensitivity-growth}
  Let
  $c\in\{1,2\}$,
  $\epsilon\in\posreals$,
  and
  $\beta_\star\in(0,1)$,
  and 
  let
  \begin{equation*}
    D_\star
    =
    \mctuple{X,U,f}{A,B}{\Lambda_\star}
  \end{equation*}
  be the specialised parametric DeltaNet Recurrent Unit 
  with hyperparameters
  $\ptuple{c, \epsilon, \beta_\star}$ as per
  Construction~\ref{construction:deltanet-finite-alphabet-parameter-variation}.
  Let
  \begin{equation*}
    \qquad\qquad
    x_0
    \defeq
    \ptuple{0,0}\,,
    \qquad\qquad
    u
    \defeq
    (u_t)_{t\in\timedomain}
    \in
    \infseqs{U}
    \quad\text{with}\quad
    u_t
    \defeq
    (-1)^{t+1}\,.
  \end{equation*}
  For every
  $t\in\timedomain$,
  let us define the parameter-to-state function
  \begin{align*}
    H_t
    :
    \Lambda_\star
    \to
    \reals^2,
    \qquad\quad
    H_t(\lambda)
    \defeq
    \indexedDynParamToStateFuncfamily{D_\star}[x_0][u][t][\lambda][big]\,,
  \end{align*}
  and 
  let $H_t^\star \defeq H_t(\lambda_\star)$ for $\lambda_\star \defeq 0$.

  Then the following holds:
  \begin{itemize}
    \item
      the state sequence $(H_t^\star)_{t\in\timedomain}$
      is bounded:
      \begin{equation}
        \label{eq:lemma:deltanet-finite-alphabet-bounded-reference-state}
        \sup_{t\in\naturals}
        \big\|
          H_t^\star
        \big\|
        <
        \infty\,;
      \end{equation}

    \item
      for every
      $t\in\timedomain$,
      the parameter-to-state function $H_t$ 
      is differentiable at $\lambda_\star$,
      and 
      it satisfies
      \begin{equation}
        \label{eq:lemma:deltanet-finite-alphabet-sensitivity-growth}
        \big\|
          \dvop{H_t}(\lambda_\star)
        \big\|
        \,\geq\,
        \beta_\star
        \cdot
        t\,;
      \end{equation}

    \item
      consequently,
      \begin{equation*}
        \big\|
          \dvop{H_t}(\lambda_\star)
        \big\|
        \to
        \infty
        \quad
        \text{as}
        \quad
        t\to\infty\,.
      \end{equation*}
  \end{itemize}
\end{restatable}

\subsection{DeltaNet: Deferred Proofs}
\label{subsec:deltanet--deferred-proofs}

\subsubsection{BIBS Instability of DeltaNet}
\label{subsec:deltanet--deferred-proofs--bibs}

\lemmanonexpansiveaffineinitialstatecomparison*
\begin{proof}
  \todo{}
  We prove
  Equation~\eqref{eq:nonexpansive-affine-initial-state-comparison}
  by induction on
  $t\in\naturals$.

  For
  $t=0$,
  the claim holds trivially, since
  \begin{equation*}
    \|x_0-\tilde x_0\|
    \leq
    \|x_0-\tilde x_0\|.
  \end{equation*}

  Now let
  $t\in\timedomain$,
  and suppose that
  \begin{equation*}
    \|x_{t-1}-\tilde x_{t-1}\|
    \leq
    \|x_0-\tilde x_0\|.
  \end{equation*}
  By the definitions of
  $x_t$
  and
  $\tilde x_t$,
  \begin{align*}
    x_t-\tilde x_t
    &=
    \big(
      A_t\cdot x_{t-1}
      +
      b_t
    \big)
    -
    \big(
      A_t\cdot \tilde x_{t-1}
      +
      b_t
    \big)
    \\
    &=
    A_t
    \cdot
    \big(
      x_{t-1}-\tilde x_{t-1}
    \big).
  \end{align*}
  Therefore,
  by the defining property of the induced matrix norm,
  the assumption
  $\|A_t\|\leq1$,
  and the induction hypothesis,
  \begin{align*}
    \|x_t-\tilde x_t\|
    &\leq
    \|A_t\|
    \cdot
    \|x_{t-1}-\tilde x_{t-1}\|
    \\
    &\leq
    \|x_{t-1}-\tilde x_{t-1}\|
    \\
    &\leq
    \|x_0-\tilde x_0\|.
  \end{align*}
  This proves
  Equation~\eqref{eq:nonexpansive-affine-initial-state-comparison}
  for every
  $t\in\naturals$.

  For the consequent claim,
  the reverse triangle inequality and
  Equation~\eqref{eq:nonexpansive-affine-initial-state-comparison}
  give
  \begin{align*}
    \|x_t\|
    &\geq
    \|\tilde x_t\|
    -
    \|x_t-\tilde x_t\|
    \\
    &\geq
    \|\tilde x_t\|
    -
    \|x_0-\tilde x_0\|
  \end{align*}
  for every
  $t\in\naturals$.
  Since
  $\|x_0-\tilde x_0\|$
  is independent of
  $t$,
  it follows that
  \begin{equation*}
    \|\tilde x_t\|
    \to
    \infty
    \quad\Longrightarrow\quad
    \|x_t\|
    \to
    \infty
    \qquad
    \text{as }
    t\to\infty.
  \end{equation*}
\end{proof}

\begin{lemma}[Discrete Gronwall estimate]
  \label{lemma:discrete-gronwall}
Let $(a_t)_{t\geq 0}$ be a nonnegative sequence satisfying
\[
a_t\leq u_t a_{t-1}+b_t,
\qquad t\geq 1,
\]
where $u_t\geq 0$ and $b_t\geq 0$. Then
\[
a_t
\leq
a_0\prod_{j=1}^t u_j
+
\sum_{i=1}^t b_i\prod_{j=i+1}^t u_j.
\]
In particular, if
\[
u_t=1+\eta_t,
\qquad \eta_t\geq 0,
\]
then
\[
a_t
\leq
\left(a_0+\sum_{i=1}^t b_i\right)
\exp\left(\sum_{j=1}^t \eta_j\right).
\]
\end{lemma}
\begin{proof}
Iterating the recursion gives
\[
a_t
\leq
a_0\prod_{j=1}^t u_j
+
\sum_{i=1}^t b_i\prod_{j=i+1}^t u_j.
\]
This proves the first claim. If $u_j=1+\eta_j$, then
\[
\prod_{j=i+1}^t u_j
=
\prod_{j=i+1}^t(1+\eta_j)
\leq
\exp\left(\sum_{j=i+1}^t\eta_j\right)
\leq
\exp\left(\sum_{j=1}^t\eta_j\right),
\]
using $1+x\leq e^x$ for $x\geq 0$. Therefore
\[
a_t
\leq
\left(a_0+\sum_{i=1}^t b_i\right)
\exp\left(\sum_{j=1}^t\eta_j\right).
\]
\end{proof}

\lemmadivergentdeltarulebeta*
\begin{proof}
  \todo{}
  The recurrence can be equivalently written as
  \begin{align*}
    x_t
    =
    x_{t-1}
    +
    \beta\cdot k_t\cdot \big(v_t-k_t^\top\cdot x_{t-1}\big)
    \ .
  \end{align*}
  Also, using the quantities
  \begin{align*}
    x_t \defeq
    \begin{pmatrix}
      x_{t,1}\\
      x_{t,2}
    \end{pmatrix},
    \qquad
    k_t \defeq
    \begin{pmatrix}
      c_t\\
      s_t
    \end{pmatrix},
    \qquad
    c_t \defeq \frac{1}{\sqrt{1+\epsilon_t^2}},
    \qquad
    s_t \defeq \frac{\epsilon_t}{\sqrt{1+\epsilon_t^2}},
  \end{align*}
  it can be written as
  \begin{align*}
    x_t
    =
    x_{t-1}
    +
    \beta\cdot
    \begin{pmatrix}
      c_t\\
      s_t
    \end{pmatrix}
    \cdot
    \big(v_t\,-\,c_t\cdot x_{t-1,1}\,-\,s_t\cdot x_{t-1,2}\big),
  \end{align*}
  and component-wise as
  \begin{align*}
    x_{t,1}
    & =
    x_{t-1,1}
    +
    \beta\cdot c_t\cdot
    \big(v_t\,-\,c_t\cdot x_{t-1,1}\,-\,s_t\cdot x_{t-1,2}\big),
    \\[.5em]
    x_{t,2}
    & =
    x_{t-1,2}
    +
    \beta\cdot s_t\cdot
    \big(v_t\,-\,c_t\cdot x_{t-1,1}\,-\,s_t\cdot x_{t-1,2}\big).
  \end{align*}

  \emph{Bound on overall state.}~~We first derive an upper bound on the state.
  Since
  \begin{align*}
    \left\|I-\beta\cdot k_tk_t^\top\right\| \leq 1,
    \qquad
    \|k_t\|=1,
  \end{align*}
  we have
  \begin{align*}
    \|x_t\|
    &\leq
    \|x_{t-1}\|
    +
    \beta\cdot |v_t|
    \leq
    \|x_{t-1}\|
    +
    \beta.
  \end{align*}
  Since $x_0=(0,0)$, this gives
  \begin{align*}
    \|x_t\|
    &\leq
    \beta\cdot t,
  \end{align*}
  and in turn
  \begin{align*}
    |x_{t,1}|
    \leq
    \beta\cdot t,
    \qquad
    |x_{t,2}|
    \leq
    \beta\cdot t.
  \end{align*}

  \emph{Bound on first state component.}~~We now derive a uniform upper bound on the first state component.
  We expand the recurrence for the first state component:
  \begin{align*}
    x_{t,1}
    &=
    \left(1-\beta\cdot c_t^2\right)\cdot x_{t-1,1}
    -
    \beta\cdot c_t\cdot s_t\cdot x_{t-1,2}
    +
    \beta\cdot c_t\cdot v_t.
  \end{align*}
  Taking absolute values and using $0<c_t\leq 1$, $|v_t|\leq 1$, and
  $s_t\leq \epsilon_t\leq \frac{1}{t+2}$, we obtain
  \begin{align*}
    |x_{t,1}|
    &\leq
    \left(1-\beta\cdot c_t^2\right)\cdot |x_{t-1,1}|
    +
    \beta\cdot s_t\cdot |x_{t-1,2}|
    +
    \beta.
  \end{align*}
  Since $\epsilon_t\leq \frac{1}{t+2}\leq \frac{1}{3}$,
  \begin{align*}
    c_t^2
    &=
    \frac{1}{1+\epsilon_t^2}
    \geq
    \frac{1}{1+\frac{1}{9}}
    =
    \frac{9}{10}.
  \end{align*}
  Thus
  \begin{align*}
    s_t\cdot |x_{t-1,2}|
    &\leq
    \frac{1}{t+2}\cdot \beta\cdot (t-1)
    \leq
    \beta,
  \end{align*}
  and hence
  \begin{align*}
    |x_{t,1}|
    &\leq
    \left(1-\frac{9}{10}\cdot \beta\right)\cdot |x_{t-1,1}|
    +
    \beta^2
    +
    \beta.
  \end{align*}
  Iterating this scalar inequality yields
  \begin{align*}
    |x_{t,1}|
    &\leq
    (\beta+\beta^2)\cdot
    \sum_{j=0}^{t-1}\left(1-\frac{9}{10}\cdot \beta\right)^j
    \leq
    (\beta+\beta^2)\cdot
    \frac{1}{\frac{9}{10}\cdot \beta}
    \\
    &=
    \frac{10}{9}\cdot (1+\beta)
    \eqdef M.
  \end{align*}
  Namely, $M$ is a bound on $|x_{t,1}|$ holding uniformly for all time steps $t$.

  \emph{Asymptotic upper bound on second state component.}~~We now show an asymptotic upper bound on the second state component.
  We expand the recurrence for the second state component as follows:
  \begin{align*}
    x_{t,2}
    &=
    x_{t-1,2}
    +
    \beta\cdot s_t\cdot (v_t-c_t\cdot x_{t-1,1})
    -
    \beta\cdot s_t^2\cdot x_{t-1,2}.
  \end{align*}
  Taking absolute values gives
  \begin{align*}
    |x_{t,2}|
    &\leq
    |x_{t-1,2}|
    +
    \beta\cdot s_t\cdot |v_t-c_t\cdot x_{t-1,1}|
    +
    \beta\cdot s_t^2\cdot |x_{t-1,2}|.
  \end{align*}
  Since $|v_t|\leq 1$ and $0<c_t\leq 1$,
  \begin{align*}
    |v_t-c_t\cdot x_{t-1,1}|
    &\leq
    |v_t|+c_t\cdot |x_{t-1,1}|
    \leq
    1+M.
  \end{align*}
  Since
  \begin{align*}
    0<s_t
    =
    \frac{\epsilon_t}{\sqrt{1+\epsilon_t^2}}
    \leq
    \epsilon_t,
  \end{align*}
  we have $s_t^2\leq \epsilon_t^2$. Thus
  \begin{align*}
    |x_{t,2}|
    &\leq
    |x_{t-1,2}|
    +
    \beta\cdot (1+M)\cdot \epsilon_t
    +
    \beta\cdot \epsilon_t^2\cdot |x_{t-1,2}|,
  \end{align*}
  or equivalently
  \begin{align*}
    |x_{t,2}|
    &\leq
    \left(1+\beta\cdot \epsilon_t^2\right)\cdot |x_{t-1,2}|
    +
    \beta\cdot (1+M)\cdot \epsilon_t.
  \end{align*}
  Letting
  \begin{align*}
    C \defeq \max\left(\beta,\,\beta\cdot (1+M)\right),
  \end{align*}
  the above yields
  \begin{align*}
    |x_{t,2}|
    &\leq
    (1+C\cdot \epsilon_t^2)\cdot |x_{t-1,2}|
    +
    C\cdot \epsilon_t.
  \end{align*}
  We apply the discrete Gronwall estimate, Lemma~\ref{lemma:discrete-gronwall}, with
  \begin{align*}
    a_t=|x_{t,2}|,
    \qquad
    u_t=1+C\cdot \epsilon_t^2,
    \qquad
    b_t=C\cdot \epsilon_t.
  \end{align*}
  This gives
  \begin{align*}
    |x_{t,2}|
    &\leq
    \left(|x_{0,2}|+C\cdot \sum_{i=1}^t \epsilon_i\right)
    \cdot
    \exp\left(C\cdot \sum_{i=1}^t \epsilon_i^2\right).
  \end{align*}
  We have
  \begin{align*}
    \sum_{i=1}^t \epsilon_i
    &=
    O\left(\sum_{i=1}^t \frac{1}{i}\right)
    =
    O(\log t),
    \\
    \sum_{i=1}^{\infty} \epsilon_i^2
    &<
    \infty.
  \end{align*}
  Therefore
  \begin{align*}
    |x_{t,2}|=O(\log t),
    \qquad
    \epsilon_t\cdot x_{t,2}
    =
    O\left(\frac{\log t}{t}\right)\to 0.
  \end{align*}

  \emph{Convergence of first component.}~~We now prove
  \begin{align*}
    x_{2n,1}
    =
    \frac{1}{2-\beta}
    +
    O\left(\frac{\log n}{n}\right),
    \qquad
    x_{2n-1,1}
    =
    \frac{1-\beta}{2-\beta}
    +
    O\left(\frac{\log n}{n}\right).
  \end{align*}
  Recall that
  \begin{align*}
    x_{t,1}
    &=
    \left(1-\beta\cdot c_t^2\right)\cdot x_{t-1,1}
    -
    \beta\cdot c_t\cdot s_t\cdot x_{t-1,2}
    +
    \beta\cdot c_t\cdot v_t.
  \end{align*}
  Since
  \begin{align*}
    c_t=1+O(\epsilon_t^2),
    \qquad
    s_t=O(\epsilon_t),
    \qquad
    \epsilon_t=O\left(\frac{1}{t}\right),
  \end{align*}
  and since $|x_{t,1}|\leq M$ and $|x_{t,2}|=O(\log t)$, there exists a sequence
  $(\theta_t)_{t\geq 1}$ and a constant $C'>0$ such that
  \begin{align*}
    x_{t,1}
    &=
    (1-\beta)\cdot x_{t-1,1}
    +
    \beta\cdot v_t
    +
    \theta_t,
    \qquad
    |\theta_t|
    \leq
    C'\cdot \frac{\log t}{t}.
  \end{align*}
  Since $v_{2n-1}=0$ and $v_{2n}=1$, we obtain
  \begin{align*}
    x_{2n-1,1}
    &=
    (1-\beta)\cdot x_{2n-2,1}
    +
    \theta_{2n-1},
    \\
    x_{2n,1}
    &=
    (1-\beta)\cdot x_{2n-1,1}
    +
    \beta
    +
    \theta_{2n}.
  \end{align*}
  Substituting the first relation into the second gives
  \begin{align*}
    x_{2n,1}
    &=
    (1-\beta)^2\cdot x_{2n-2,1}
    +
    \beta
    +
    (1-\beta)\cdot \theta_{2n-1}
    +
    \theta_{2n}.
  \end{align*}
  Letting $y_n\defeq x_{2n,1}$, we have
  \begin{align*}
    y_n
    &=
    (1-\beta)^2\cdot y_{n-1}
    +
    \beta
    +
    \delta_n,
    \qquad
    \delta_n\defeq (1-\beta)\cdot \theta_{2n-1}+\theta_{2n}.
  \end{align*}
  Since $|\delta_n|\leq C'\cdot \frac{\log n}{n}$, and since the affine map
  \begin{align*}
    y \mapsto (1-\beta)^2\cdot y + \beta
  \end{align*}
  has fixed point $\frac{1}{2-\beta}$, the same geometric-series estimate as in the proof of a stable affine recursion gives
  \begin{align*}
    x_{2n,1}
    =
    \frac{1}{2-\beta}
    +
    O\left(\frac{\log n}{n}\right).
  \end{align*}
  Using
  \begin{align*}
    x_{2n-1,1}
    &=
    (1-\beta)\cdot x_{2n-2,1}
    +
    \theta_{2n-1},
  \end{align*}
  we also obtain
  \begin{align*}
    x_{2n-1,1}
    &=
    \frac{1-\beta}{2-\beta}
    +
    O\left(\frac{\log n}{n}\right).
  \end{align*}

  \emph{Divergence of second component.}~~We derive the two-step change for the second state component. Since $v_{2n-1}=0$ and $v_{2n}=1$, we have
  \begin{align*}
    x_{2n,2}-x_{2n-2,2}
    &=
    \beta\cdot s_{2n-1}
    \left(
    0
    -
    c_{2n-1}\cdot x_{2n-2,1}
    -
    s_{2n-1}\cdot x_{2n-2,2}
    \right)
    \\
    &\quad
    +
    \beta\cdot s_{2n}
    \left(
    1
    -
    c_{2n}\cdot x_{2n-1,1}
    -
    s_{2n}\cdot x_{2n-1,2}
    \right).
  \end{align*}
  Using
  \begin{align*}
    c_t=1+O\left(\frac{1}{n^2}\right),
    \qquad
    s_t=\epsilon_t+O\left(\frac{1}{n^3}\right),
  \end{align*}
  for $t\sim n$, together with $x_{t,1}=O(1)$ and $x_{t,2}=O(\log t)$, we obtain
  \begin{align*}
    x_{2n,2}-x_{2n-2,2}
    &=
    -\beta\cdot \epsilon_{2n-1}\cdot x_{2n-2,1}
    +
    \beta\cdot \epsilon_{2n}\cdot
    \left(1-x_{2n-1,1}\right)
    +
    O\left(\frac{\log n}{n^2}\right).
  \end{align*}
  Now substitute
  \begin{align*}
    x_{2n-2,1}
    &=
    \frac{1}{2-\beta}
    +
    O\left(\frac{\log n}{n}\right),
    &
    x_{2n-1,1}
    &=
    \frac{1-\beta}{2-\beta}
    +
    O\left(\frac{\log n}{n}\right).
  \end{align*}
  Thus
  \begin{align*}
    1-x_{2n-1,1}
    &=
    \frac{1}{2-\beta}
    +
    O\left(\frac{\log n}{n}\right),
  \end{align*}
  and hence
  \begin{align*}
    x_{2n,2}-x_{2n-2,2}
    &=
    -\beta\cdot \epsilon_{2n-1}
    \left(
    \frac{1}{2-\beta}
    +
    O\left(\frac{\log n}{n}\right)
    \right)
    \\
    &\quad
    +
    \beta\cdot \epsilon_{2n}
    \left(
    \frac{1}{2-\beta}
    +
    O\left(\frac{\log n}{n}\right)
    \right)
    +
    O\left(\frac{\log n}{n^2}\right).
  \end{align*}
  Since
  \begin{align*}
    \epsilon_{2n-1}
    &=
    \frac{1}{2n+1}
    =
    \frac{1}{2n}
    +
    O\left(\frac{1}{n^2}\right),
    \\
    \epsilon_{2n}
    &=
    \frac{1}{2(2n+2)}
    =
    \frac{1}{4n}
    +
    O\left(\frac{1}{n^2}\right),
  \end{align*}
  we get
  \begin{align*}
    x_{2n,2}-x_{2n-2,2}
    &=
    -\beta\cdot \frac{1}{2n}\cdot \frac{1}{2-\beta}
    +
    \beta\cdot \frac{1}{4n}\cdot \frac{1}{2-\beta}
    +
    O\left(\frac{\log n}{n^2}\right)
    \\
    &=
    -\frac{\beta}{4(2-\beta)}\cdot \frac{1}{n}
    +
    O\left(\frac{\log n}{n^2}\right).
  \end{align*}
  Finally,
  \begin{align*}
    x_{2n,2}
    &=
    x_{2n,2}-x_{0,2}
    \mathsidecomment{$x_{0,2}=0$ by assumption}
    \\
    &=
    \sum_{m=1}^n \big(x_{2m,2}-x_{2(m-1),2}\big)
    \mathsidecomment{telescopic sum}
    \\
    &=
    \sum_{m=1}^n
    \left(
    -\frac{\beta}{4(2-\beta)}\cdot \frac{1}{m}
    +
    O\left(\frac{\log m}{m^2}\right)
    \right)
    \\
    &=
    -\frac{\beta}{4(2-\beta)}\cdot \sum_{m=1}^n \frac{1}{m}
    +
    \sum_{m=1}^n O\left(\frac{\log m}{m^2}\right).
  \end{align*}
  Since
  \begin{align*}
    \sum_{m=1}^n \frac{1}{m}
    &=
    \log n + O(1),
    &
    \sum_{m=1}^{\infty}\frac{\log m}{m^2}
    &<
    \infty,
  \end{align*}
  we obtain
  \begin{align*}
    x_{2n,2}
    &=
    -\frac{\beta}{4(2-\beta)}\log n
    +
    O(1).
  \end{align*}
  In particular, $x_{2n,2}\to -\infty$, and hence $\|x_t\|\to\infty$.

  \emph{Second component at odd steps.}~~From the odd update,
  \begin{align*}
    x_{2n-1,2}-x_{2n-2,2}
    &=
    \beta\cdot s_{2n-1}
    \left(
    0
    -
    c_{2n-1}\cdot x_{2n-2,1}
    -
    s_{2n-1}\cdot x_{2n-2,2}
    \right).
  \end{align*}
  Using the same estimates as above,
  \begin{align*}
    x_{2n-1,2}-x_{2n-2,2}
    &=
    -\beta\cdot \epsilon_{2n-1}\cdot x_{2n-2,1}
    +
    O\left(\frac{\log n}{n^2}\right)
    \\
    &=
    -\beta\cdot
    \left(
    \frac{1}{2n}
    +
    O\left(\frac{1}{n^2}\right)
    \right)
    \left(
    \frac{1}{2-\beta}
    +
    O\left(\frac{\log n}{n}\right)
    \right)
    +
    O\left(\frac{\log n}{n^2}\right)
    \\
    &=
    -\frac{\beta}{2(2-\beta)}\cdot \frac{1}{n}
    +
    O\left(\frac{\log n}{n^2}\right).
  \end{align*}
  Therefore
  \begin{align*}
    x_{2n-1,2}
    &=
    -\frac{\beta}{4(2-\beta)}\log n
    +
    O(1).
  \end{align*}
  This concludes the proof.
\end{proof}

\thmdivergentdeltarulebeta*
\begin{proof}
  Consider the sequences
  $(\beta_t)_{t \in \timedomain}$,
  $(k_t)_{t \in \timedomain}$,
  and
  $(v_t)_{t \in \timedomain}$
  defined in
  Lemma~\ref{lemma:divergent-delta-rule-beta},
  and define the input sequence
  $(u_t)_{t \in \timedomain}$
  by
  \begin{equation*}
    u_t
    \defeq
    \ptuple{
      1,
      \beta_t,
      k_t,
      v_t
    }
    =
    \ptuple{
      1,
      \beta,
      k_t,
      v_t
    }
    \qquad
    \forall t \in \timedomain.
  \end{equation*}

  By the definitions in
  Lemma~\ref{lemma:divergent-delta-rule-beta},
  we have
  \begin{equation*}
    k_t
    =
    \frac{1}{\sqrt{1+\epsilon_t^2}}
    \cdot
    \begin{pmatrix}
      1
      \\
      \epsilon_t
    \end{pmatrix},
    \qquad
    \epsilon_t\in[0,1/2],
    \qquad
    v_t\in\{0,1\},
  \end{equation*}
  and hence
  \begin{equation*}
    k_t\in K
    \qquad\text{and}\qquad
    u_t\in\tilde U
    \subseteq
    U
    \qquad
    \forall t \in \timedomain.
  \end{equation*}
  Therefore,
  $(u_t)_{t \in \timedomain}$
  is an admissible input sequence for
  $D$.
  Moreover, it is bounded, since
  \begin{equation*}
    g_t=1,
    \qquad
    \beta_t=\beta,
    \qquad
    \|k_t\|=1,
    \qquad
    |v_t|\leq1
    \qquad
    \forall t \in \timedomain.
  \end{equation*}

  Since
  $X\neq\emptyset$,
  fix any
  \begin{equation*}
    x_0\in X,
  \end{equation*}
  and let
  $(x_t)_{t \in \timedomain}$
  be the corresponding state sequence under
  $(u_t)_{t \in \timedomain}$,
  \begin{equation*}
    x_t
    \defeq
    f(x_{t-1},u_t)
    \qquad
    \forall t \in \timedomain.
  \end{equation*}

  Since
  $D$
  has update hyperparameter
  $c=1$
  and the decay-gate component of every
  $u_t$
  is
  $g_t=1$,
  Definition~\ref{def:deltanet-unit}
  gives
  \begin{equation}
    \label{eq:divergent-delta-rule-beta-arbitrary-initial-state}
    x_t
    \,=\,
    A_t \cdot x_{t-1}
    +
    b_t,
  \end{equation}
  where
  \begin{equation*}
    A_t
    \,\defeq\,
    I_2
    -
    \beta
    \cdot
    k_t
    \cdot
    k_t^\transposeop,
    \qquad
    b_t
    \defeq
    \beta
    \cdot
    k_t
    \cdot
    v_t^\transposeop.
  \end{equation*}

  Now define an auxiliary sequence
  $(\tilde x_t)_{t \in \naturals}$
  in the ambient state space
  $\reals^2$
  by
  \begin{equation*}
    \tilde x_0
    \defeq
    \ptuple{0,0},
    \qquad
    \tilde x_t
    \,\defeq\,
    A_t \cdot \tilde x_{t-1}
    +
    b_t
    \qquad
    \forall t \in \timedomain.
  \end{equation*}
  Notice that we do not require
  $\tilde x_t\in X$:
  $(\tilde x_t)_{t \in \naturals}$
  is used only as an auxiliary sequence in the ambient state space.

  By its definition,
  $(\tilde x_t)_{t \in \naturals}$
  is exactly the sequence considered in
  Lemma~\ref{lemma:divergent-delta-rule-beta}.
  Hence,
  \begin{equation}
    \label{eq:divergent-delta-rule-beta-auxiliary-divergence}
    \|\tilde x_t\|
    \to
    \infty
    \quad\text{as}\quad
    t\to\infty.
  \end{equation}

  We next observe that the state-update matrices
  $A_t$
  are nonexpansive.
  Indeed, since
  $\|k_t\|=1$,
  every
  $z\in\reals^2$
  can be uniquely written as
  \begin{equation*}
    z
    \,=\,
    a \cdot k_t
    +
    z^\transposeop,
    \qquad\qquad
    k_t^\transposeop \cdot z^\transposeop
    =
    \ptuple{0,0}.
  \end{equation*}
  Consequently,
  \begin{align*}
    A_t \cdot z
    \,=\,
    (1-\beta)
    \cdot
    a \cdot k_t
    +
    z^\transposeop,
  \end{align*}
  and therefore
  \begin{align*}
    \|A_t \cdot z\|^2
    &\,=\,
    (1-\beta)^2 \cdot a^2
    +
    \|z^\transposeop\|^2
    \\[.2em]
    &\,\leq\,
    a^2
    +
    \|z^\transposeop\|^2
    \\[.2em]
    &\,=\,
    \|z\|^2.
  \end{align*}
  Hence,
  \begin{equation*}
    \|A_t\|
    \leq
    1
    \qquad
    \forall t \in \timedomain.
  \end{equation*}

  The sequences
  $(x_t)_{t \in \naturals}$
  and
  $(\tilde x_t)_{t \in \naturals}$
  therefore satisfy the hypotheses of
  Lemma~\ref{lemma:nonexpansive-affine-initial-state-comparison}.
  Since
  $\tilde x_0=\ptuple{0,0}$,
  that lemma gives
  \begin{equation*}
    \|x_t-\tilde x_t\|
    \leq
    \|x_0\|
    \qquad
    \forall t \in \naturals.
  \end{equation*}
  Thus, by the reverse triangle inequality,
  \begin{equation*}
    \|x_t\|
    \geq
    \|\tilde x_t\|
    -
    \|x_0\|.
  \end{equation*}
  Combining this inequality with
  Equation~\eqref{eq:divergent-delta-rule-beta-auxiliary-divergence}
  yields
  \begin{equation*}
    \|x_t\|
    \to
    \infty
    \quad
    \text{as}\quad
    t\to\infty.
  \end{equation*}

  Thus,
  the bounded admissible input sequence
  $(u_t)_{t \in \timedomain}$
  generates an unbounded state sequence from
  $x_0\in X$.
  Hence
  $D$
  does not satisfy BIBS.
\end{proof}

\subsubsection{BIBSD Instability of DeltaNet}
\label{subsec:deltanet--deferred-proofs--bibsd}

\factdeltanetfinitealphabetspecialisedparametrisation*
\begin{proof}
  We first verify that
  $\Theta_\star$
  is a valid parameter space for a standard DeltaNet Recurrent Unit with no
  decay gate.
  Let
  $\lambda\in\Lambda_\star$
  and
  $u\in U$
  be arbitrary.
  By
  Construction~\ref{construction:deltanet-finite-alphabet-parameter-variation},
  the key preactivation is
  \begin{align}
    \label{eq:lemma:deltanet-finite-alphabet-key-preactivation}
    \fpinst{\tilde k}{u}{\lambda}
    \,=\,
    \pinst{W_\mathrm{k}}{\lambda}
    \cdot
    u
    +
    b_\mathrm{k}
    \,=\,
    \begin{pmatrix}
      1
      \\
      \lambda u
    \end{pmatrix}\,.
  \end{align}
  In particular,
  \begin{equation*}
    \fpinst{\tilde k}{u}{\lambda}
    \neq
    0
    \qquad\qquad
    \forall\,
    \lambda\in\Lambda_\star\,,
    \quad
    \forall\,
    u\in U\,.
  \end{equation*}
  Hence every
  $\vartheta(\lambda)\in\Theta_\star$
  satisfies the key-admissibility condition of
  Definition~\ref{def:deltanet-rnc}.
  All remaining components of
  $\vartheta(\lambda)$
  have the dimensions required by that definition.
  Thus,
  $\Theta_\star$
  is a valid parameter space for a DeltaNet Recurrent Unit with no decay gate
  and update hyperparameter
  $c$.

  We next compare its dynamics with those of
  $D_\star$.
  By
  Equation~\eqref{eq:lemma:deltanet-finite-alphabet-key-preactivation},
  \begin{equation}
    \label{eq:lemma:deltanet-finite-alphabet-key}
    \fpinst{k}{u}{\lambda}
    \,=\,
    \frac{
      \begin{pmatrix}
        1
        \\
        \lambda u
      \end{pmatrix}
    }{
      \sqrt{1+\lambda^2}
    }
    \,.
  \end{equation}
  Furthermore, by the parameter choices of
  Construction~\ref{construction:deltanet-finite-alphabet-parameter-variation},
  \begin{align*}
    \fpinst{\beta}{u}{\lambda}
    \,=\,
    \sigma
    \left(
      \log
      \left(
        \frac{
          \beta_\star
        }{
          1-\beta_\star
        }
      \right)
    \right)
    \,=\,
    \beta_\star\,,
    \qquad\quad
    \fpinst{v}{u}{\lambda}
    \,=\,
    u\,.
  \end{align*}

  On the other hand,
  the standard DeltaNet parameter value associated with
  $\lambda$
  is
  \begin{equation*}
    \vartheta(\lambda)
    =
    \mBig(
      \big(
        \pinst{W_\mathrm{k}}{\lambda},
        b_\mathrm{k}
      \big),\;
      \big(
        W_\mathrm{v},
        b_\mathrm{v}
      \big),\;
      \big(
        w_\beta,
        b_\beta
      \big)
    \mBig).
  \end{equation*}
  Therefore,
  Definition~\ref{def:deltanet-rnc}
  gives, for every
  $\lambda\in\Lambda_\star$
  and
  $u\in U$,
  the same key, write strength, and value functions as those of
  $D_\star$:
  \begin{align*}
    \bigfpinst{k}{u}{\vartheta(\lambda)}
    &=
    \fpinst{k}{u}{\lambda}\,,
    &
    \bigfpinst{\beta}{u}{\vartheta(\lambda)}
    &=
    \fpinst{\beta}{u}{\lambda}\,,
    &
    \bigfpinst{v}{u}{\vartheta(\lambda)}
    &=
    \fpinst{v}{u}{\lambda}\,.
  \end{align*}

  Consequently, the corresponding state-update matrix and additive term satisfy
  \begin{align*}
    \bigfpinst{A}{u}{\vartheta(\lambda)}
    \,=\,
    \fpinst{A}{u}{\lambda}\,,
    \qquad\qquad
    \bigfpinst{B}{u}{\vartheta(\lambda)}
    \,=\,
    \fpinst{B}{u}{\lambda}\,.
  \end{align*}
  Hence, for every
  $H\in X$,
  $u\in U$,
  and
  $\lambda\in\Lambda_\star$,
  \begin{equation*}
    \bigfpinst{f}{H,u}{\vartheta(\lambda)}
    \,=\,
    \fpinst{f}{H,u}{\lambda}\,.
  \end{equation*}

  Thus,
  the standard DeltaNet Recurrent Unit with no decay gate,
  update hyperparameter
  $c$,
  and parameter space
  $\Theta_\star$
  has precisely the dynamics of
  $D_\star$
  under the parameter identification
  $\vartheta$.
\end{proof}

\lemmadeltanetfinitealphabetparametersensitivitygrowth*
\begin{proof}
  \todo{}

  We first show the following intermediate claim.

  \par\medskip\noindent
  \emph{Intermediate Claim.}~
  For every $t\in\naturals$,
  there exists a scalar sequence
  $(a_t)_{t\in\naturals}$
  such that, for every $t \in \naturals$,
  \begin{equation}
    \label{eq:lemma:deltanet-finite-alphabet-reference-scalar-recurrence}
    H_t^\star
    \,=\,
    a_t \cdot e_1
    \qquad\text{where}\qquad
    a_t
    \,=\,
    \begin{cases}
      0 
      & \text{ if }\;\,
      t = 0\,,
      \\[.2em]
      (1-c \cdot \beta_\star) \cdot a_{t-1}
      +
      \beta_\star \cdot u_t
      & \text{ if }\;\,
      t \geq 1\,.
    \end{cases}
  \end{equation}

  \par\smallskip\noindent
  \emph{Proof of Intermediate Claim.}~
  We proceed by induction on $t\in\naturals$.

  In the base case we have $t = 0$,
  and hence
  \begin{equation*}
    H_0^\star
    =
    x_0
    =
    0
    =
    a_0 \cdot e_1
  \end{equation*}
  since
  $x_0=\ptuple{0,0}$
  and
  $a_0 = 0$.

  In the inductive case, we have $t \geq 1$,
  and we assume by indunction that
  \begin{equation*}
    H_{t-1}^\star
    =
    a_{t-1}\cdot e_1\,.
  \end{equation*}
  By the dynamics function of $D_\star$,
  together with
  \begin{equation*}
    \fpinst{k}{u_t}{\lambda_\star}
    =
    e_1,
    \qquad
    \fpinst{\beta}{u_t}{\lambda_\star}
    =
    \beta_\star,
    \qquad
    \fpinst{v}{u_t}{\lambda_\star}
    =
    u_t,
  \end{equation*}
  we obtain
  \begin{align*}
    H_t^\star
    &=
    \fpinst{f}{
      H_{t-1}^\star,
      u_t
    }{
      \lambda_\star
    }
    \\[.2em]
    &=
    \left(
      I_2
      -
      c\cdot\beta_\star
      \cdot
      e_1e_1^\transposeop
    \right)
    \cdot
    H_{t-1}^\star
    +
    \beta_\star
    \cdot
    e_1
    \cdot
    u_t
    \\[.2em]
    &=
    \left(
      I_2
      -
      c\cdot\beta_\star
      \cdot
      e_1e_1^\transposeop
    \right)
    \cdot
    \big(
      a_{t-1}\cdot e_1
    \big)
    +
    \beta_\star
    \cdot
    u_t
    \cdot
    e_1
    \\[.2em]
    &=
    \left(
      a_{t-1}
      -
      c\cdot\beta_\star
      \cdot
      a_{t-1}
      \cdot
      e_1^\transposeop e_1
    \right)
    \cdot
    e_1
    +
    \beta_\star
    \cdot
    u_t
    \cdot
    e_1
    \\[.2em]
    &=
    \mBig(
      (1-c\cdot\beta_\star)
      \cdot
      a_{t-1}
      +
      \beta_\star
      \cdot
      u_t
    \mBig)
    \cdot
    e_1,
  \end{align*}
  where we have used
  $e_1^\transposeop \cdot e_1\,=\,1$.
  Thus, defining
  \begin{equation*}
    a_t
    \defeq
    (1-c\cdot\beta_\star)
    \cdot
    a_{t-1}
    +
    \beta_\star
    \cdot
    u_t,
  \end{equation*}
  we have
  $H_t^\star=a_t\cdot e_1$.
  This proves the claim by induction.
  \hfill
  \emph{End of proof of Intermediate Claim.}
  \par

  \medskip
  Since
  \begin{equation*}
    |1-c \cdot \beta_\star|
    \,<\,
    1
    \qquad\text{and}\qquad
    |u_t|
    \,=\,
    1
    \quad
    \forall\,
    t\in\timedomain\,,
  \end{equation*}
  we have
  \begin{equation*}
    \sup_{t\in\timedomain}
    \big\|
      H_t^\star
    \big\|
    \;=\;
    \sup_{t\in\timedomain}
    |a_t|
    \;<\;
    \infty\,,
  \end{equation*}
  proving
  Equation~\eqref{eq:lemma:deltanet-finite-alphabet-bounded-reference-state}.

  We next consider the derivative with respect to the parameter
  $\lambda$
  at
  $\lambda_\star$.
  We have
  \begin{equation}
    \label{eq:lemma:deltanet-finite-alphabet-key-derivative}
    \dvop{
      \fpinst{k}{u}{\cdot}
    }(\lambda_\star)
    =\mkern1mu
    u \cdot e_2
    \qquad
    \forall\,
    u\in U\,,
  \end{equation}
  and
  hence,
  \begin{equation*}
    \dvop{
      \fpinst{\beta}{u}{\cdot}
    }(\lambda_\star)
    \,=\,
    0
    \qquad\text{and}\qquad
    \dvop{
      \fpinst{v}{u}{\cdot}
    }(\lambda_\star)
    \,=\,
    0
    \qquad\quad
    \forall\,
    u\in U\,.
  \end{equation*}

  We claim that there exists a scalar sequence
  $(g_t)_{t\in\naturals}$
  such that
  \begin{equation}
    \label{eq:lemma:deltanet-finite-alphabet-state-derivative-form}
    \dvop{H_t}(\lambda_\star)
    \,=\,
    g_t \cdot e_2\,,
    \qquad\quad
    g_0
    \,=\,
    0\,.
  \end{equation}
  Indeed,
  by
  Construction~\ref{construction:deltanet-finite-alphabet-parameter-variation},
  \begin{align*}
    \begin{aligned}
    H_t(\lambda)
    \,=\,
    &
    \left(
      I_2
      -
      c
      \cdot
      \fpinst{\beta}{u_t}{\lambda}
      \cdot
      \fpinst{k}{u_t}{\lambda}
      \cdot
      \fpinst{k}{u_t}{\lambda}^{\transposeop}
    \right)
    \cdot H_{t-1}(\lambda)
    \\[.3em]
    & 
    +
    \fpinst{\beta}{u_t}{\lambda}
    \cdot
    \fpinst{k}{u_t}{\lambda}
    \cdot
    \fpinst{v}{u_t}{\lambda}
    \end{aligned}
    \qquad\qquad 
    \forall\, \lambda \in \Lambda_\star\,.
  \end{align*}
  Taking the derivative with respect to
  $\lambda$
  at
  $\lambda_\star$,
  and using
  \begin{equation*}
    \fpinst{k}{u_t}{\lambda_\star}
    \,=\,
    e_1,
    \qquad\quad
    H_{t-1}^\star
    \,=\,
    a_{t-1} \cdot e_1\,,
    \qquad\quad
    \dvop{
      \fpinst{k}{u_t}{\cdot}
    }(\lambda_\star)
    \,=\,
    u_t \cdot e_2\,,
  \end{equation*}
  gives
  \begin{align*}
    \dvop{H_t}(\lambda_\star)
    &\,=\,
    \left(
      I_2
      -
      c \cdot \beta_\star  \cdot e_1 \cdot e_1^\transposeop
    \right)
    \cdot
    \dvop{H_{t-1}}(\lambda_\star)
    \\[.2em]
    & \quad
    \, -
    c \cdot \beta_\star
     \cdot \left(
      u_t \cdot e_2 \cdot e_1^\transposeop
      +
      u_t  \cdot e_1 \cdot e_2^\transposeop
    \right)
     \cdot a_{t-1} \cdot e_1
    \\[.2em]
    & \quad
    \, +
    \beta_\star
     \cdot u_t  \cdot e_2
    \cdot
    u_t
    \\[.7em]
    &=
    \left(
      I_2
      -
      c \cdot \beta_\star  \cdot e_1 \cdot e_1^\transposeop
    \right)
     \cdot \dvop{H_{t-1}}(\lambda_\star)
    \\[.2em]
    & \quad
    \, -
    \beta_\star
     \cdot \big(
      1
      -
      c  \cdot u_t  \cdot a_{t-1}
    \big)
     \cdot e_2\,.
  \end{align*}
  Since
  \begin{equation*}
    \left(
      I_2
      -
      c \cdot \beta_\star  \cdot e_1 \cdot e_1^\transposeop
    \right)
     \cdot e_2
    \;=\;
    e_2\,,
  \end{equation*}
  by induction we obtain
  Equation~\eqref{eq:lemma:deltanet-finite-alphabet-state-derivative-form},
  with
  \begin{equation}
    \label{eq:lemma:deltanet-finite-alphabet-sensitivity-recurrence}
    g_t
    \;=\;
    g_{t-1}
    +
    \beta_\star
    \cdot
    \big(
      1
      -
      c \cdot  u_t \cdot  a_{t-1}
    \big)\,.
  \end{equation}

  Let $z_t \defeq u_t \cdot  a_t$.
  Since $u_t = {-}u_{t-1}$,
  we have
  \begin{equation*}
    z_t
    \;=\;
    \beta_\star
    -
    (1-c \cdot \beta_\star) \cdot z_{t-1}\,.
  \end{equation*}
  Writing
  \begin{equation*}
    \rho
    \,\defeq\,
    1-c \cdot \beta_\star
    \,\in\,
    (-1,1)\,,
  \end{equation*}
  and observing that
  $z_0=0$,
  we obtain
  \begin{align*}
    z_t
    \;=\;
    \beta_\star
    \cdot
    \sum_{j=0}^{t-1}
    (-\rho)^j
    \;=\;
    \beta_\star
    \cdot
    \frac{
      1-({-}\rho)^t
    }{
      1+\rho
    }
    \,>\,
    0
    \qquad\quad
    \forall\,
    t\in\timedomain\,.
  \end{align*}
  Therefore,
  for every
  $t\geq2$,
  \begin{equation*}
    u_t \cdot a_{t-1}
    \,=\,
    {-}z_{t-1}
    \,\leq\,
    0\,.
  \end{equation*}
  Substituting this into
  Equation~\eqref{eq:lemma:deltanet-finite-alphabet-sensitivity-recurrence}
  gives
  \begin{equation*}
    g_t
    \,\geq\,
    g_{t-1}
    +
    \beta_\star
    \qquad\quad
    \forall\,
    t\geq 2\,.
  \end{equation*}
  Furthermore,
  Equation~\eqref{eq:lemma:deltanet-finite-alphabet-sensitivity-recurrence},
  together with
  $a_0=0$
  and
  $g_0=0$,
  gives
  \begin{equation*}
    g_1
    \,=\,
    \beta_\star\,.
  \end{equation*}
  Hence,
  by induction we obtain
  \begin{equation*}
    g_t
    \,\geq\,
    \beta_\star
    \cdot
    t
    \qquad\qquad
    \forall\,
    t\in\naturals\,.
  \end{equation*}

  Combining this with
  Equation~\eqref{eq:lemma:deltanet-finite-alphabet-state-derivative-form},
  we obtain
  \begin{equation*}
    \big\|
      \dvop{H_t}(\lambda_\star)
    \big\|
    \,=\,
    |g_t|
    \,\geq\,
    \beta_\star
    \cdot
    t
    \qquad\quad
    \forall\,
    t\in\naturals\,,
  \end{equation*}
  proving
  Equation~\eqref{eq:lemma:deltanet-finite-alphabet-sensitivity-growth}.

  The divergence claim follows immediately.
\end{proof}

\thmdeltanetdoesnotsatisfywbibsd*
\begin{proof}
  Let
  \begin{equation*}
    x_0
    \defeq
    \ptuple{0,0}
    \qquad\text{and}\qquad
    u
    \defeq
    (u_t)_{t\in\timedomain}
    \in
    \infseqs{U}
    \quad\text{with}\quad
    u_t
    \defeq
    (-1)^{t+1}.
  \end{equation*}
  Let moreover
  \begin{equation*}
    \lambda_\star
    \defeq
    0.
  \end{equation*}

  Taking
  $r\defeq1$,
  we have
  \begin{equation*}
    u
    \in
    \infseqs{\truncation{U}{r}},
    \qquad
    x_0
    \in
    \truncation{X}{r},
    \qquad
    \lambda_\star
    \in
    \truncation{(\Lambda_\star)}{r}.
  \end{equation*}

  By
  Lemma~\ref{lemma:deltanet-finite-alphabet-parameter-sensitivity-growth},
  for every
  $t\in\timedomain$,
  the parameter-to-state function
  \begin{equation*}
    H_t
    \defeq
    \indexedDynParamToStateFuncfamily{D_\star}[x_0][u][t]
  \end{equation*}
  is differentiable at
  $\lambda_\star$
  and satisfies
  \begin{equation*}
    \big\|
      \dvop{H_t}(\lambda_\star)
    \big\|
    \geq
    \beta_\star
    \cdot
    t.
  \end{equation*}
  Hence
  \begin{equation*}
    \sup_{t\in\timedomain}
    \big\|
      \dvop{
        \indexedDynParamToStateFuncfamily{D_\star}[x_0][u][t]
      }(\lambda_\star)
    \big\|
    =
    \infty.
  \end{equation*}

  Therefore,
  there exists no constant
  $B_{r,u,x_0}\in\nonnegreals$
  satisfying the derivative bound required by
  Definition~\ref{def:wbibsd}
  for this
  $r$,
  $u$,
  and
  $x_0$.
  Thus,
  $D_\star$
  does not satisfy wBIBSD.

  Since
  uBIBSD implies wBIBSD,
  $D_\star$
  does not satisfy uBIBSD either.
\end{proof}

\subsubsection{BIBLG Instability of DeltaNet}
\label{subsec:deltanet--deferred-proofs--biblg}

\thmdeltanetlinearlossgradientgrowth*
\begin{proof}
  \todo{}
  Apply
  Lemma~\ref{lemma:deltanet-finite-alphabet-parameter-sensitivity-growth}
  for the present values of
  $c$
  and
  $\beta_\star$.
  Let
  $\epsilon\in\posreals$,
  \begin{equation*}
    D
    =
    \mctuple{X,U,f}{A,B}{\Theta},
  \end{equation*}
  and
  \begin{equation*}
    u
    =
    (u_t)_{t\in\posnaturals}
    \in
    \infseqs{U_\star},
    \qquad
    U_\star
    \defeq
    \{-1,1\},
  \end{equation*}
  be witnesses supplied by that lemma.

  Let
  $\lambda_\star$,
  $\Lambda_\star$,
  $\Theta_\star$,
  and
  $\pinst{\theta}{\lambda}$
  be as in
  Construction~\ref{construction:deltanet-finite-alphabet-parameter-variation}
  for the present values of
  $\epsilon$
  and
  $\beta_\star$.
  Thus,
  \begin{equation*}
    \lambda_\star
    =
    0,
    \qquad
    \Theta_\star
    \subseteq
    \Theta.
  \end{equation*}
  Define
  \begin{equation}
    \label{eq:proof:deltanet-finite-alphabet-reference-parameter}
    \theta_\star
    \defeq
    \pinst{\theta}{\lambda_\star}.
  \end{equation}
  Since
  $\lambda_\star\in\Lambda_\star$,
  we have
  \begin{equation*}
    \theta_\star
    \in
    \Theta_\star
    \subseteq
    \Theta.
  \end{equation*}

  Let
  \begin{equation*}
    x_0
    \defeq
    \ptuple{0,0}.
  \end{equation*}
  By
  Lemma~\ref{lemma:deltanet-finite-alphabet-parameter-sensitivity-growth},
  the reference state trajectory
  \begin{equation*}
    H_T^\star
    \defeq
    \statefunc_T^{u,x_0}
    (\theta_\star)
  \end{equation*}
  is bounded:
  \begin{equation}
    \label{eq:proof:deltanet-finite-alphabet-reference-state-bounded}
    \sup_{T\in\naturals}
    \|
      H_T^\star
    \|
    <
    \infty.
  \end{equation}

  We next record the form of the reference states.
  By
  Construction~\ref{construction:deltanet-finite-alphabet-parameter-variation},
  for every
  $\xi\in U$,
  \begin{equation*}
    \fpinst{k}{\xi}{\theta_\star}
    =
    e_1.
  \end{equation*}
  Hence,
  by induction on
  $T$,
  the reference state has the form
  \begin{equation}
    \label{eq:proof:deltanet-finite-alphabet-reference-state}
    H_T^\star
    =
    a_T e_1
  \end{equation}
  for some scalar sequence
  $(a_T)_{T\in\naturals}$.
  By
  Equation~\eqref{eq:proof:deltanet-finite-alphabet-reference-state-bounded},
  this scalar sequence is bounded.

  We also record the direction of the parameter sensitivity.
  Let
  \begin{equation*}
    G_T
    \defeq
    \pdvop{\statefunc_T^{u,x_0}}{\lambda_\star}
    (\theta_\star).
  \end{equation*}
  By the explicit parametrisation in
  Construction~\ref{construction:deltanet-finite-alphabet-parameter-variation},
  for every
  $\xi\in U$,
  \begin{equation*}
    \pdvop{
      \fpinst{k}{\xi}{\cdot}
    }{\lambda_\star}
    (\theta_\star)
    =
    \xi e_2,
  \end{equation*}
  whereas
  $\beta$
  and
  $v$
  are constant with respect to this scalar parameter coordinate.

  Differentiating the DeltaNet recurrence at
  $\theta_\star$
  therefore gives
  \begin{equation*}
    G_T
    =
    \left(
      I_2
      -
      c\beta_\star e_1e_1^\transposeop
    \right)
    G_{T-1}
    +
    \beta_\star
    \big(
      1
      -
      c u_T a_{T-1}
    \big)
    e_2.
  \end{equation*}
  Since
  \begin{equation*}
    G_0
    =
    0
  \end{equation*}
  and
  \begin{equation*}
    \left(
      I_2
      -
      c\beta_\star e_1e_1^\transposeop
    \right)
    e_2
    =
    e_2,
  \end{equation*}
  induction yields
  \begin{equation*}
    G_T
    =
    g_T e_2
  \end{equation*}
  for some scalar sequence
  $(g_T)_{T\in\naturals}$.
  Consequently,
  Equation~\eqref{eq:lemma:deltanet-finite-alphabet-sensitivity-growth}
  gives
  \begin{equation}
    \label{eq:proof:deltanet-finite-alphabet-state-sensitivity}
    \pdvop{\statefunc_T^{u,x_0}}{\lambda_\star}
    (\theta_\star)
    =
    g_T e_2,
    \qquad
    |g_T|
    \geq
    \beta_\star\cdot T
    \qquad
    \forall T\in\posnaturals.
  \end{equation}

  Let
  $S$
  be the one-unit DeltaNet model obtained from
  $D$,
  with initial state
  \begin{equation*}
    H^\mathrm{init}
    \defeq
    x_0,
  \end{equation*}
  output space
  \begin{equation*}
    Y
    =
    \reals^d,
  \end{equation*}
  associated target space
  $\targetspace{Y}$,
  and output function
  \begin{equation}
    \label{eq:proof:deltanet-finite-alphabet-output-function}
    h
    :
    X\times U\times X
    \to
    \reals^d,
    \qquad
    h(H,v,H')
    \defeq
    y_\star
    +
    \big(
      e_2^\transposeop H'
    \big)
    a.
  \end{equation}
  The function
  $h$
  is affine and hence analytic.
  By construction,
  \begin{equation*}
    S
    \in
    \classmodels{DeltaNet}[c],
    \qquad
    U_\star
    \subseteq
    U,
    \qquad
    \Theta_\star
    \subseteq
    \Theta.
  \end{equation*}
  Since
  $\lossfn:Y\times\targetspace{Y}\to\reals$
  is analytic,
  it is an admissible loss function for
  $S$.

  For every
  $T\in\posnaturals$,
  let
  \begin{equation*}
    R_T
    :
    \Theta
    \to
    \reals^d,
    \qquad
    R_T(\theta)
    \defeq
    \fpinst{\repval{S}}{\seq{u}{1}{T}}{\theta}
  \end{equation*}
  be the parameter-to-representation map corresponding to the
  length-$T$ prefix of
  $u$.
  By
  Equations~
  \eqref{eq:proof:deltanet-finite-alphabet-reference-state}
  and~
  \eqref{eq:proof:deltanet-finite-alphabet-output-function},
  \begin{equation}
    \label{eq:proof:deltanet-finite-alphabet-reference-output}
    R_T(\theta_\star)
    =
    y_\star
    \qquad
    \forall T\in\posnaturals,
  \end{equation}
  since
  \begin{equation*}
    e_2^\transposeop
    H_T^\star
    =
    0.
  \end{equation*}

  By the definition of
  $R_T$,
  Equation~\eqref{eq:proof:deltanet-finite-alphabet-output-function},
  and
  Equation~\eqref{eq:proof:deltanet-finite-alphabet-state-sensitivity},
  its named partial derivative with respect to the scalar parameter
  coordinate whose value at
  $\theta_\star$
  is
  $\lambda_\star$
  satisfies
  \begin{align}
    \pdvop{R_T}{\lambda_\star}(\theta_\star)
    &=
    \left(
      e_2^\transposeop
      \pdvop{\statefunc_T^{u,x_0}}{\lambda_\star}
      (\theta_\star)
    \right)
    a
    \notag
    \\
    &=
    g_T a.
    \label{eq:proof:deltanet-finite-alphabet-output-sensitivity}
  \end{align}

  For every
  $T\in\posnaturals$,
  define
  \begin{equation}
    \label{eq:proof:deltanet-finite-alphabet-sample}
    z_T
    \defeq
    \ptuple{
      \seq{u}{1}{T},
      \hat y_\star
    }.
  \end{equation}
  Since
  \begin{equation*}
    \seq{u}{1}{T}
    \in
    U^+
    \qquad\text{and}\qquad
    \hat y_\star
    \in
    \targetspace{Y},
  \end{equation*}
  we have
  \begin{equation*}
    z_T
    \in
    \samplespacep{S},
  \end{equation*}
  so
  $z_T$
  is an admissible sample for
  $S$.

  We next verify differentiability of its objective function.
  By the key-admissibility condition of
  Definition~\ref{def:deltanet-rnc},
  for every
  $\theta\in\Theta$
  and
  $v\in U$,
  \begin{equation*}
    \fpinst{\widetilde k}{v}{\theta}
    \neq
    0.
  \end{equation*}
  Hence the normalisation map
  \begin{equation*}
    q
    \longmapsto
    \frac{q}{\|q\|}
  \end{equation*}
  is smooth at every key preactivation arising from an admissible
  input-parameter pair.
  All remaining recurrence-input functions of DeltaNet are compositions
  of affine maps and smooth functions.
  Therefore,
  for every finite input sequence,
  the corresponding state and representation maps are differentiable
  with respect to the parameters.
  Since
  $h$
  is affine and
  $\lossfn$
  is analytic,
  it follows that
  \begin{equation*}
    \Lossfn{S,z_T}[\lossfn]
  \end{equation*}
  is differentiable at
  $\theta_\star$
  for every
  $T\in\posnaturals$.

  Applying the chain rule to the named partial derivative of
  $\Lossfn{S,z_T}[\lossfn]$
  with respect to the scalar parameter coordinate whose value at
  $\theta_\star$
  is
  $\lambda_\star$,
  and using
  Equations~
  \eqref{eq:proof:deltanet-finite-alphabet-reference-output}
  and~
  \eqref{eq:proof:deltanet-finite-alphabet-output-sensitivity},
  yields
  \begin{align}
    \pdvop{\Lossfn{S,z_T}[\lossfn]}{\lambda_\star}
    (\theta_\star)
    &=
    \pdvop{\lossfn}{y_\star}
    \big(
      y_\star,
      \hat y_\star
    \big)
    \cdot
    \pdvop{R_T}{\lambda_\star}
    (\theta_\star)
    \notag
    \\
    &=
    g_T
    \cdot
    \left(
      \pdvop{\lossfn}{y_\star}
      \big(
        y_\star,
        \hat y_\star
      \big)
      \cdot
      a
    \right).
    \label{eq:proof:deltanet-finite-alphabet-loss-gradient-chain-rule}
  \end{align}

  Define
  \begin{equation*}
    \eta
    \defeq
    \left|
      \pdvop{\lossfn}{y_\star}
      \big(
        y_\star,
        \hat y_\star
      \big)
      \cdot
      a
    \right|.
  \end{equation*}
  By the nonzero local derivative condition,
  \begin{equation*}
    \eta
    >
    0.
  \end{equation*}
  Therefore,
  by
  Equations~
  \eqref{eq:proof:deltanet-finite-alphabet-state-sensitivity}
  and~
  \eqref{eq:proof:deltanet-finite-alphabet-loss-gradient-chain-rule},
  \begin{equation}
    \label{eq:proof:deltanet-finite-alphabet-coordinate-gradient-bound}
    \left|
      \pdvop{\Lossfn{S,z_T}[\lossfn]}{\lambda_\star}
      (\theta_\star)
    \right|
    \geq
    \eta
    \beta_\star
    T
    \qquad
    \forall T\in\posnaturals.
  \end{equation}

  Since
  $\lambda$
  is a scalar parameter coordinate
  and its value at
  $\theta_\star$
  is
  $\lambda_\star$,
  \begin{equation*}
    \pdvop{\Lossfn{S,z_T}[\lossfn]}{\lambda_\star}
    (\theta_\star)
  \end{equation*}
  is the corresponding scalar coordinate of
  \begin{equation*}
    \dvop{
      \Lossfn{S,z_T}[\lossfn]
    }
    (\theta_\star).
  \end{equation*}
  Consequently,
  Equation~\eqref{eq:proof:deltanet-finite-alphabet-coordinate-gradient-bound}
  implies
  \begin{align}
    \big\|
      \dvop{
        \Lossfn{S,z_T}[\lossfn]
      }
      (\theta_\star)
    \big\|
    &\geq
    \left|
      \pdvop{\Lossfn{S,z_T}[\lossfn]}{\lambda_\star}
      (\theta_\star)
    \right|
    \notag
    \\
    &\geq
    \eta
    \beta_\star
    T.
    \label{eq:proof:deltanet-finite-alphabet-full-gradient-bound}
  \end{align}
  Hence
  Equation~\eqref{eq:thm:deltanet-linear-loss-gradient-growth}
  holds with
  \begin{equation*}
    C
    \defeq
    \eta
    \beta_\star
    \in
    \posreals.
  \end{equation*}

  It remains to choose a common truncation radius.
  Since
  $U_\star$
  is finite,
  and
  $\theta_\star$
  and
  $\hat y_\star$
  are fixed finite-dimensional elements,
  there exists
  $r\in\posreals$
  such that
  \begin{equation}
    \label{eq:proof:deltanet-finite-alphabet-fixed-radius}
    U_\star
    \subseteq
    \truncation{U}{r},
    \qquad
    \theta_\star
    \in
    \truncation{\Theta}{r},
    \qquad
    \hat y_\star
    \in
    \truncation{\targetspace{Y}}{r}.
  \end{equation}

  Since
  \begin{equation*}
    u
    \in
    \infseqs{U_\star},
  \end{equation*}
  every input appearing in
  $\seq{u}{1}{T}$
  belongs to
  $\truncation{U}{r}$.
  Together with
  Equation~\eqref{eq:proof:deltanet-finite-alphabet-fixed-radius},
  this gives
  \begin{equation}
    \label{eq:proof:deltanet-finite-alphabet-truncated-sample}
    z_T
    \in
    \horizontruncsamplespacep{S}{r}{T}
    \qquad
    \forall T\in\posnaturals.
  \end{equation}
  Thus all three claims in the first part of the theorem hold for this
  single radius
  $r$.

  By the definition of the corresponding loss-gradient growth function,
  Equations~
  \eqref{eq:proof:deltanet-finite-alphabet-fixed-radius}
  and~
  \eqref{eq:proof:deltanet-finite-alphabet-truncated-sample}
  imply
  \begin{align*}
    \LossGrowth{S}{\lossfn}{r}(T)
    &=
    \sup
    \mBig\{
      \,
      \big\|
        \dvop{
          \Lossfn{S,z}[\lossfn]
        }
        (\theta)
      \big\|
      \;:\;
      z
      \in
      \horizontruncsamplespacep{S}{r}{T},
      \;
      \theta
      \in
      \truncation{\Theta}{r}
    \mBig\}
    \\
    &\geq
    \big\|
      \dvop{
        \Lossfn{S,z_T}[\lossfn]
      }
      (\theta_\star)
    \big\|.
  \end{align*}
  Hence,
  by
  Equation~\eqref{eq:proof:deltanet-finite-alphabet-full-gradient-bound},
  \begin{equation*}
    \LossGrowth{S}{\lossfn}{r}(T)
    \geq
    C T
    \qquad
    \forall T\in\posnaturals.
  \end{equation*}
  Therefore,
  \begin{equation*}
    \LossGrowth{S}{\lossfn}{r}(T)
    =
    \Omega(T).
  \end{equation*}

  In particular,
  the loss-gradient norm is not uniformly bounded over the fixed-radius
  truncated sample and parameter sets as the horizon grows.
  By
  Definition~\ref{def:biblg},
  $S$
  therefore does not satisfy BIBLG.
  Since
  \begin{equation*}
    S
    \in
    \classmodels{DeltaNet}[c],
  \end{equation*}
  the class
  $\classmodels{DeltaNet}[c]$
  does not satisfy BIBLG either.
  Finally,
  the witnessing input sequence takes values entirely in the fixed
  two-symbol set
  \begin{equation*}
    U_\star
    =
    \{-1,1\},
  \end{equation*}
  proving the final claim.
\end{proof}

\lemmastandardlossesnonzerolocalderivative*
\begin{proof}
  We consider the three standard loss functions separately.

  \par\medskip
  \noindent
  \textit{Squared error.}
  The squared-error loss
  \begin{equation*}
    \seloss
    :
    \reals^d
    \times
    \reals^d
    \to
    \nonnegreals
    \,,
    \qquad\qquad
    \seloss(y,\hat y)
    \defeq
    \frac{1}{2}
    \|
    y-\hat y
    \|^2
    \,,
  \end{equation*}
  is polynomial and hence analytic.
  Therefore,
  it is admissible for every system with matching output space
  $Y=\reals^d$.

  Its partial derivative with respect to its input block
  $y_{\selossName}$
  evaluated at 
  $\ptuple{ y_{\selossName}, \hat y_{\selossName} } = \ptuple{ e_1, 0 }$ 
  is
  \begin{equation*}
    \pdvop{\seloss}{y_{\selossName}}
    \big(
      y_{\selossName},
      \hat y_{\selossName}
    \big)
    \;=\;
    \big(
      y_{\selossName}
      -
      \hat y_{\selossName}
    \big)^\transposeop
    \;=\;
    e_1^\transposeop
    \,.
  \end{equation*}
  Since
  $a_{\selossName}=e_1$,
  it follows that
  \begin{align*}
    \pdvop{\seloss}{y_{\selossName}}
    \big(
      y_{\selossName},
      \hat y_{\selossName}
    \big)
    \cdot
    a_{\selossName}
    \;=\;
    e_1^\transposeop
    \cdot
    e_1
    \;=\;
    1
    \;.
  \end{align*}
  This proves
  Equation~\eqref{eq:lemma:standard-losses-se-derivative},
  and hence that
  $\ptuple{
    a_{\selossName},
    y_{\selossName},
    \hat y_{\selossName}
  }$
  witnesses the nonzero local derivative condition for
  $\seloss$.

  \par\medskip
  \noindent
  \textit{Binary cross-entropy.}
  The function
  \begin{equation*}
    \ol\bceloss
    :
    \reals
    \times
    \reals
    \to
    \reals,
    \qquad\quad
    \ol\bceloss(y,\hat y)
    \defeq
    \log
    \big(
      1+\exp(y)
    \big)
    -
    \hat y\cdot y
  \end{equation*}
  is analytic and is an ambient extension of
  $\bceloss$,
  showing that
  $\bceloss$
  is analytic and hence admissible for every system with matching
  output space
  $Y=\reals$.

  For the fixed target
  $\hat y_{\bcelossName}=0$,
  the function
  \begin{equation*}
    y
    \longmapsto
    \bceloss
    \big(
      y,
      \hat y_{\bcelossName}
    \big)
  \end{equation*}
  is the analytic function
  \begin{equation*}
    y
    \longmapsto
    \log
    \big(
      1+\exp(y)
    \big)
    -
    \hat y_{\bcelossName}\cdot y
    \,.
  \end{equation*}
  Hence its derivative at
  $y_{\bcelossName}=0$
  gives
  \begin{align*}
    \pdvop{\bceloss}{y_{\bcelossName}}
    \big(
      y_{\bcelossName},
      \hat y_{\bcelossName}
    \big)
    \,=\,
    \frac{
      \exp(y_{\bcelossName})
    }{
      1+\exp(y_{\bcelossName})
    }
    -
    \hat y_{\bcelossName}
    \,=\,
    \sigma(y_{\bcelossName})
    -
    \hat y_{\bcelossName}
    \,.
  \end{align*}
  Since 
  $y_{\bcelossName} = \hat y_{\bcelossName} = 0$ 
  and 
  $a_{\bcelossName} = 1$,
  it follows that
  \begin{align*}
    \pdvop{\bceloss}{y_{\bcelossName}}
    \big(
      y_{\bcelossName},
      \hat y_{\bcelossName}
    \big)
    \cdot
    a_{\bcelossName}
    \,=\,
    \sigma(0)
    \,=\,
    \frac{1}{2}
    \,.
  \end{align*}
  This proves
  Equation~\eqref{eq:lemma:standard-losses-bce-derivative},
  and hence that
  $\ptuple{
    a_{\bcelossName},
    y_{\bcelossName},
    \hat y_{\bcelossName}
  }$
  witnesses the nonzero local derivative condition for
  $\bceloss$.

  \par\medskip
  \noindent
  \textit{$K$-class cross-entropy.}
  Let
  $K\in\posnaturals$
  with
  $K\geq2$.
  For every
  $k\in\iival{1}{K}$,
  define the polynomial
  \begin{equation*}
    p_k(t)
    \,\defeq\,
    \prod_{
      \substack{
        m=1
        \\[.1em]
        m\neq k
      }
    }^{K}
    \frac{
      t-m
    }{
      k-m
    }
    \;.
  \end{equation*}
  By the
  \emph{Lagrange interpolation theorem}---cf.~Theorem~6.1
  of~\citep{suli2003introduction}---we have
  \begin{equation*}
    p_j(k)
    =
    \begin{cases}
      1
      & \text{if } j=k,
      \\[.2em]
      0
      & \text{if } j\neq k,
    \end{cases}
    \qquad\quad
    \forall j,k\in\iival{1}{K}.
  \end{equation*}

  Consider the function
  \begin{equation*}
    \ol{\softmaxceloss[K]}
    :
    \reals^K
    \times
    \reals
    \to
    \reals,
    \qquad\qquad
    \ol{\softmaxceloss[K]}(y,t)
    \,\defeq\,
    \log
    \left(
      \sum_{j=1}^{K}
      \exp(y_j)
    \right)
    -
    \left(
      \sum_{j=1}^{K}
      p_j(t)\cdot y_j
    \right)
    \,.
  \end{equation*}
  This function is analytic.
  Furthermore,
  for every
  $\hat y\in\iival{1}{K}$,
  \begin{align*}
    \ol{\softmaxceloss[K]}(y,\hat y)
    \;=\;
    \log
    \left(
      \sum_{j=1}^{K}
      \exp(y_j)
    \right)
    -
    y_{\hat y}
    \;=\;
    {-}
    \log
    \left(
      \frac{
        \exp(y_{\hat y})
      }{
        \sum_{j=1}^{K}
        \exp(y_j)
      }
    \right)
    \;=\;
    \softmaxceloss[K](y,\hat y)
    \;.
  \end{align*}
  Hence
  $\ol{\softmaxceloss[K]}$
  is an analytic ambient extension of
  $\softmaxceloss[K]$,
  showing that
  $\softmaxceloss[K]$
  is analytic and hence admissible for every system with matching
  output space
  $Y=\reals^K$.

  For the fixed target
  $\hat y_{\softmaxcelossName}=1$,
  the function
  \begin{equation*}
    y
    \longmapsto
    \softmaxceloss[K]
    \big(
      y,
      \hat y_{\softmaxcelossName}
    \big)
  \end{equation*}
  is the analytic function
  \begin{equation*}
    y
    \longmapsto
    \log
    \left(
      \sum_{j=1}^{K}
      \exp(y_j)
    \right)
    -
    y_{\hat y_{\softmaxcelossName}}
    \,.
  \end{equation*}
  Hence its partial derivative with respect to its input block
  $y_{\softmaxcelossName}$
  evaluated at 
  $\ptuple{ y_{\softmaxcelossName}, \hat y_{\softmaxcelossName} } = \ptuple{ 0, 1 }$
  is
  \begin{equation*}
    \pdvop{\softmaxceloss[K]}{y_{\softmaxcelossName}}
    \big(
      y_{\softmaxcelossName},
      \hat y_{\softmaxcelossName}
    \big)
    =\,
    \frac{1}{K}
    \cdot
    \mathbf 1_K^\transposeop
    -
    e_1^\transposeop
    \;.
  \end{equation*}
  Since
  $a_{\softmaxcelossName} = e_2$
  and
  $K\geq2$,
  it follows that
  \begin{align*}
    \pdvop{\softmaxceloss[K]}{y_{\softmaxcelossName}}
    \big(
      y_{\softmaxcelossName},
      \hat y_{\softmaxcelossName}
    \big)
    \cdot
    a_{\softmaxcelossName}
    \;=\;
    \left(
      \frac{1}{K}
      \mathbf 1_K^\transposeop
      -
      e_1^\transposeop
    \right)
    \cdot
    e_2
    \;=\;
    \frac{1}{K}
    \;.
  \end{align*}
  This proves
  Equation~\eqref{eq:lemma:standard-losses-ce-derivative},
  and hence that
  $\ptuple{
    a_{\softmaxcelossName},
    y_{\softmaxcelossName},
    \hat y_{\softmaxcelossName}
  }$
  witnesses the nonzero local derivative condition for
  $\softmaxceloss[K]$.
\end{proof}

\thmdeltanetstandardlossgradientgrowth*
\begin{proof}
  By
  Lemma~\ref{lemma:standard-losses-nonzero-local-derivative},
  the squared-error,
  binary cross-entropy,
  and
  $K$-class cross-entropy losses
  are admissible and satisfy the nonzero local derivative condition,
  with respective witness derivative magnitudes
  \begin{equation*}
    1,
    \qquad
    \frac{1}{2},
    \qquad
    \frac{1}{K}.
  \end{equation*}
  Therefore,
  Theorem~\ref{thm:deltanet-linear-loss-gradient-growth}
  applies in each of the three cases and yields respectively
  \begin{equation*}
    C
    =
    \beta_\star,
    \qquad
    C
    =
    \frac{\beta_\star}{2},
    \qquad
    C
    =
    \frac{\beta_\star}{K}.
  \end{equation*}
  Hence
  Equations~\eqref{eq:thm:deltanet-standard-loss-gradient-growth-se}--%
  \eqref{eq:thm:deltanet-standard-loss-gradient-growth-ce}
  hold.
  The asymptotic loss-gradient-growth and BIBLG claims follow directly
  from the same theorem.
\end{proof}

\clearpage{}%

\clearpage{}%

\section{Analysis of Mamba}
\label{app:mamba}
\label{subsec:positive-margin-mamba}

\localtableofcontents

\subsection{Definition of Mamba}
\label{subsec:mamba--definition}

\begin{closingdefinition}[Mamba, cf.~\citep{gu2024mamba}]
  \label{def:mamba-rnc--new}
  Let $d_\mathrm{in}, d_\mathrm{s}, n_\mathrm{h} \in \posnaturals$.
  A \emph{Mamba Recurrent Unit}
  of
  \emph{state dimension}~$d_\mathrm{s}$
  and
  \emph{number of heads}~$n_\mathrm{h}$
  is a Parametric General Linear Recurrent Unit
  \begin{equation*}
    D
    =
    \mctuple{X,U,f}{\Theta}{A,B}
  \end{equation*}
  with components as described next.

  \begin{definitionpart}[title={Input and state spaces}, indent=body]
    \label{def:mamba-rnc-input-state-spaces}
    The input and parameter spaces satisfy
    \begin{align*}
      U \subseteq \Omega_U \defeq \reals^{d_\mathrm{in}}
      \qquad\text{ and }\qquad
      X \subseteq \Omega_X \defeq \reals^{d_\mathrm{s}\times n_\mathrm{h}}.
    \end{align*}
  \end{definitionpart}

  \begin{definitionpart}[title={Parameter space}, indent=body]
    \label{def:mamba-rnc-parameter-space}
    The parameter space satisfies $\Theta \subseteq \Omega_\Theta$ 
    for 
    $\Omega_\Theta$ consisting of each parameter value $\theta$ of the form
    \begin{equation} 
      \label{eq:mamba-rnc-parameter-space}
      \theta
      \,=\,
      \mbigptuple{
        \mkern1mu
        a,
        \mkern2mu
        \ptuple{W_\mathrm{v},b_\mathrm{v}},\,
        \ptuple{W_\mathrm{b},b_\mathrm{b}},\,
        \ptuple{W_\Delta,b_\Delta}
        \mkern2mu
      }
    \end{equation}
    with components
    \begin{equation*}
      \begin{tabular}{
        r@{\;}c@{\;}l
        @{\hspace{2em}}
        r@{\;}c@{\;}l
        @{\hspace{5em}}
        l
        }
        $a$
        & $\in$
        & $\reals^{d_\mathrm{s}\times n_\mathrm{h}}$
        & & & 
        &
        \textit{decay parameters}
        \\[.6em]
        $W_\mathrm{v}$
        & $\in$
        & $\reals^{n_\mathrm{h}\times d_\mathrm{in}}$
        &
        $b_\mathrm{v}$
        & $\in$
        & $\reals^{n_\mathrm{h}}$
        &
        \textit{value parameters}
        \\[.6em]
        $W_\mathrm{b}$
        & $\in$
        & $\reals^{d_\mathrm{s}\times d_\mathrm{in}}$
        &
        $b_\mathrm{b}$
        & $\in$
        & $\reals^{d_\mathrm{s}}$
        &
        \textit{state-input parameters}
        \\[.6em]
        $W_\Delta$
        & $\in$
        & $\reals^{n_\mathrm{h}\times d_\mathrm{in}}$
        &
        $b_\Delta$
        & $\in$
        & $\reals^{n_\mathrm{h}}$
        &
        \textit{step-size parameters}.
      \end{tabular}
    \end{equation*}
  \end{definitionpart}

  \begin{definitionpart}[title={Recurrence-input functions}, indent=body]
    \label{def:mamba-rnc-recurrence-input-functions}
    The recurrence inputs are given by the parametric functions
    \begin{align*}
      \Delta
      &:
      U\mkern-2mu\times\Theta
      \to
      (\posreals)^{n_\mathrm{h}}
      & \quad
      \fpinst{\Delta}{u}{\theta}
      &\,\defeq\,
      \operatorname{softplus}
      \big(
        \fpinst{\tilde\Delta}{u}{\theta}
      \big)
      \qquad\qquad
      && \textit{step size}
      \\[.2em]
      \tilde\Delta
      &:
      U\mkern-2mu\times\Theta
      \to
      \reals^{n_\mathrm{h}}
      & \quad
      \fpinst{\tilde\Delta}{u}{\theta}
      &\,\defeq\,
      W_\Delta\cdot u+b_\Delta
      && 
      \textit{step-size preactivation}
      \\[.5em]
      v
      &:
      U\mkern-2mu\times\Theta
      \to
      \reals^{n_\mathrm{h}}
      &
      \fpinst{v}{u}{\theta}
      &\,\defeq\,
      W_\mathrm{v}\cdot u+b_\mathrm{v}
      && \textit{state-input values}
      \\[.3em]
      b
      &:
      U\mkern-2mu\times\Theta
      \to
      \reals^{d_\mathrm{s}}
      &
      \fpinst{b}{u}{\theta}
      &\,\defeq\,
      W_\mathrm{b}\cdot u+b_\mathrm{b}
      && \textit{state-input vector}
    \end{align*}
    where
    $\theta$ is of the form~\eqref{eq:mamba-rnc-parameter-space},
    and
    the function $\operatorname{softplus}$ is applied coordinatewise.
  \end{definitionpart}

  \begin{definitionpart}[title={State-update coefficients}, indent=body]
    \label{def:mamba-rnc-state-update-coefficients}
    The 
    \emph{state-update coefficients} 
    are given by the parametric function
    \begin{align*}
      A
      :
      U\mkern-2mu\times\Theta
      \to
      (0,1)^{d_\mathrm{s}\times n_\mathrm{h}}
      \qquad\qquad
      \fpinst{A}{u}{\theta}
      \,\defeq\,
      \exp
      \mBig(
        {-}\exp(a)
        \cdot
        \operatorname{Diag}
        \big(
          \fpinst{\Delta}{u}{\theta}
        \big)
      \mBig)
    \end{align*}
    where 
    $\theta$ is of the form~\eqref{eq:mamba-rnc-parameter-space},
    and 
    the function $\exp$ is applied coordinatewise.
  \end{definitionpart}

  \begin{definitionpart}[title={Dynamics function}, indent=body]
    \label{def:mamba-rnc-dynamics-function}
    The parametric dynamics function is
    \begin{equation*}
      \fpinst{f}{X,u}{\theta}
      \,\defeq\,
      \fpinst{\Lambda_u}{X}{\theta}
      +
      \fpinst{B}{u}{\theta}
    \end{equation*}
    where the
    state-update linear operator and additive term are given by the parametric functions
    \begin{align*} %
      \fpinst{\Lambda_u}{X}{\theta}
      & \,\defeq\, 
      \fpinst{A}{u}{\theta} \ast X
      &&
      \textit{state-update linear operator}
      \\%
      \fpinst{B}{u}{\theta}
      & \,\defeq\,
      \fpinst{b}{u}{\theta}
      \cdot
      \mbig(
        \fpinst{\Delta}{u}{\theta}
        \odot
        \fpinst{v}{u}{\theta}
        \mbig)^{\raisebox{-.3em}{\ensuremath{\scriptstyle\transposeop}}}
      \;
      &&
      \textit{additive term}
    \end{align*}
    where $\ast$ is element-wise product.
  \end{definitionpart}

  \begin{definitionpart}[indent=body]
    Next we provide definitions
    that build on the intermediate definitions introduced above.

    The \emph{class of Mamba Recurrent Units}
    is denoted by $\classunits{Mamba}$.

    A \emph{Mamba model}
    is a parametric
    $\classunits{Mamba}$-RNC
    with initial states set to zero.
    \label{def:mamba-rnc-classes}

    The \emph{class of Mamba models}
    is denoted by $\classmodels{Mamba}$.

    The \emph{class of Mamba dynamics}
    is the set conssisting of dynamics of each Mamba model,
    \begin{equation*}
      \classdynamics{Mamba} 
      \,\defeq \,
      \big\{\dynop{S} \mid S \in \classmodels{Mamba} \big\}.
    \end{equation*}
  \end{definitionpart}
\end{closingdefinition}

\subsection{Mamba models are smooth unitwise-contractive}
\label{subsec:mamba--smooth-unitwise-contractive}

We show that Mamba models are contractive,
as an intermediate result to show the expressivity expressivity upper bound of
Theorem~\ref{thm:positive-margin-mamba-expressivity}.

Recalling the notion of \emph{smooth unitwise-contractivity} introduced in
Definition~\ref{def:smooth-contractive-linear-rnn},
we show next that 
Mamba models are smooth unitwise-contractive.
We note that smooth unitwise-contractivity is defined for Linear RNNs, 
that are RNCs of Linear Recurrent Units, 
whereas Mamba models are RNCs of Generalised Linear Recurrent Units;
thus, the result below is developed for the canonical Linear RNN associated with a Mamba model, as
introduced in Definition~\ref{def:canonical-linear-rnn}.

\todo{}

\begin{closingtheorem}[%
  Presentation-induced Linear RNN representations of Mamba models are smooth contractive%
  ]
  \label{thm:mamba-is-smooth-unitwise-contractive}
  Let
  $S \in \classmodels{Mamba}$,
  and let
  \begin{equation*}
    \presentationSty{S}
    \,\defeq\,
    \repoutoutextension{
      \presentationSty{C}
    }{
      x^\mathrm{init}
    }{
      h
    }
  \end{equation*}
  be a parametric analytic output extension presentation of
  $S$
  such that
  $\presentationSty{C}$
  is a parametric recurrent cascade presentation of type
  $\ptuple{
  \classunits{Mamba},
  \analyticparamfuncclass
  }$.

  Let
  $\presentationSty{\tilde S}$
  be the Linear RNN presentation associated with
  $\presentationSty{S}$
  according to
  Definition~\ref{def:canonical-linear-rnn},
  and 
  let
  $\tilde S
  \defeq
  \presentationVal{
    \presentationSty{\tilde S}
  }$
  be the corresponding Linear RNN representation of
  $S$.

  Then
  $\tilde S$
  is a smooth unitwise-contractive Linear RNN.
\end{closingtheorem}
\begin{proof}
  For each $i\in\iival{1}{\ell}$,
  let us
  write
  \begin{equation*}
    \presentationSty{C}
    \,=\,
    \repinputprecomposition{
      g_1
    }{
      \presentationSty{L}_1
    }
    \repcascade_{g_2}
    \presentationSty{L}_2
    \repcascade_{g_3}
    \cdots
    \repcascade_{g_\ell}
    \presentationSty{L}_\ell,
    \qquad\text{where}\qquad
    \presentationSty{L}_i
    \,=\,
    D_{i,1}
    \replayer
    \cdots
    \replayer
    D_{i,m_i}
    \,.
  \end{equation*}

  Since
  $\presentationSty{C}$
  is a parametric recurrent cascade presentation of type
  $\ptuple{
    \classunits{Mamba},
    \analyticparamfuncclass
  }$,
  we have that
  every $D_{i,j}$ is a Mamba Recurrent Unit and 
  every parametric connecting function $g_i$ is analytic.

  Furthermore,
  every Mamba Recurrent Unit is a Parametric General Linear Recurrent Unit by
  Definition~\ref{def:mamba-rnc--new}.
  Hence
  Proposition~\ref{prop:linear-rnn-representation-is-linear-rnn}
  applies to
  $\presentationSty{S}$,
  and therefore
  $\tilde S$
  is a Parametric Linear RNN.

  It remains to verify the additional requirements of
  Definition~\ref{def:smooth-contractive-linear-rnn}.

  Let us
  fix arbitrary $i\in\iival{1}{\ell}$ and $j\in\iival{1}{m_i}$,
  and
  let us write the corresponding Mamba Recurrent Unit as
  \begin{equation*}
    D_{i,j}
    \,=\,
    \mctuple{
      X,
      U,
      f
    }{
      \Theta
    }{
      A,
      B
    },
    \qquad\text{with}\quad
    X
    \subseteq
    \reals^{d_\mathrm{s}\times n_\mathrm{h}}.
  \end{equation*}
  Let
  \begin{equation*}
    \tilde D_{i,j}
    \,=\,
    \mctuple{
      \tilde X,
      U,
      \tilde f
    }{
      \Theta
    }{
      \tilde A,
      \tilde B
    }
  \end{equation*}
  be its canonical Linear Recurrent Unit representation,
  part of the flattened layer presentation for layer $i$ 
  under the layer presentation $\presentationSty{\tilde L}_i$.

  By
  Definition~\ref{def:mamba-rnc--new},
  the state-update coefficient matrix is
  \begin{equation}
    \label{eq:prop:mamba-rncs-smooth-contractive-coefficients}
    \fpinst{A}{u}{\theta}
    \,=\,
    \exp
    \mBig(
      {-}\exp(a)
      \cdot
      \operatorname{Diag}
      \big(
        \fpinst{\Delta}{u}{\theta}
      \big)
    \mBig),
  \end{equation}
  where
  \begin{equation}
    \label{eq:prop:mamba-rncs-smooth-contractive-delta}
    \fpinst{\Delta}{u}{\theta}
    \,=\,
    \operatorname{softplus}
    \big(
      W_\Delta\cdot u+b_\Delta
    \big),
  \end{equation}
  and the state-update linear operator satisfies
  \begin{equation}
    \label{eq:prop:mamba-rncs-smooth-contractive-state-update}
    \fpinst{\Lambda_u}{X}{\theta}
    \,=\,
    \fpinst{A}{u}{\theta}
    \ast
    X.
  \end{equation}

  For every
  $X\in\reals^{d_\mathrm{s}\times n_\mathrm{h}}$,
  Equation~\eqref{eq:prop:mamba-rncs-smooth-contractive-state-update}
  gives
  \begin{align*}
    \flattenop
    \left(
      \fpinst{\Lambda_u}{X}{\theta}
    \right)
    &=
    \flattenop
    \left(
      \fpinst{A}{u}{\theta}
      \ast
      X
    \right)
    \\[.2em]
    &=
    \operatorname{Diag}
    \left(
      \flattenop
      \big(
        \fpinst{A}{u}{\theta}
      \big)
    \right)
    \cdot
    \flattenop(X).
  \end{align*}
  Therefore,
  by
  Definition~\ref{def:flattening},
  the matrix representation under flattening of the state-update linear operator is
  \begin{equation*}
    \flattenop
    \left(
      \fpinst{\Lambda_u}{\cdot}{\theta}
    \right)
    \,=\,
    \operatorname{Diag}
    \left(
      \flattenop
      \big(
        \fpinst{A}{u}{\theta}
      \big)
    \right).
  \end{equation*}
  Hence,
  by
  Definition~\ref{def:linear-rnn-representation},
  the state-update matrix of
  $\tilde D_{i,j}$
  is
  \begin{equation}
    \label{eq:prop:mamba-rncs-smooth-contractive-canonical-matrix}
    \fpinst{\tilde A}{u}{\theta}
    \,=\,
    \operatorname{Diag}
    \left(
      \flattenop
      \big(
        \fpinst{A}{u}{\theta}
      \big)
    \right),
  \end{equation}
  and its additive term is
  \begin{equation}
    \label{eq:prop:mamba-rncs-smooth-contractive-canonical-additive}
    \fpinst{\tilde B}{u}{\theta}
    \,=\,
    \flattenop
    \big(
      \fpinst{B}{u}{\theta}
    \big).
  \end{equation}

  \par\medskip
  \emph{Continuous differentiability up to the closure.}
  The functions
  $\Delta$,
  $A$,
  and
  $B$
  of
  Definition~\ref{def:mamba-rnc--new}
  are obtained from affine maps,
  $\exp$,
  $\operatorname{softplus}$,
  linear maps such as
  $\operatorname{Diag}$,
  and finite-dimensional algebraic operations by finite tupling and composition.
  Hence,
  by
  Proposition~\ref{prop:analytic-neural-primitives}
  and
  Proposition~\ref{prop:analytic-maps-closure},
  they are analytic Euclidean maps.

  Since
  $\flattenop$
  and
  $\operatorname{Diag}$
  are linear maps,
  Equations~\eqref{eq:prop:mamba-rncs-smooth-contractive-canonical-matrix}
  and~\eqref{eq:prop:mamba-rncs-smooth-contractive-canonical-additive}
  imply that
  $\tilde A$
  and
  $\tilde B$
  are analytic Euclidean maps as well.
  By
  Fact~\ref{fact:analytic-euclidean-map-smooth-up-to-closure},
  they are continuously-differentiable up to the closures of their domains.

  \par\medskip
  \emph{Strict contractivity on the closure.}
  Let
  $
  \ptuple{ \ol u, \ol \theta }
  \in
  \closureop{ U\!\times\!\Theta }
  $ 
  be arbitrary.

  The functions above are ordinarily analytic,
  and hence their ambient extensions are given by the same defining formulas.
  In particular,
  the ambient extension $\Delta_\mathrm{e}$ of $\Delta$ is such that, 
  for every $h\in\iival{1}{n_\mathrm{h}}$,
  \begin{equation}
    \label{eq:prop:mamba-rncs-smooth-contractive-delta-positive}
    \getentry{
    \fpinst{\Delta_\mathrm{e}}{\ol u}{\ol\theta} 
    }{
      h
    }
    \defeq
    \operatorname{softplus}
    \big(
      \getentry{
      \ol W_\Delta \cdot \ol u + \ol b_\Delta
      }{h}
    \big)
    \,>\,
    0.
  \end{equation}
  Consequently,
  the ambient extension $A_\mathrm{e}$ of $A$ is such that, 
  for 
  every $r\in\iival{1}{d_\mathrm{s}}$
  and 
  every $p\in\iival{1}{n_\mathrm{p}}$,
  \begin{align}
    \label{eq:prop:mamba-rncs-smooth-contractive-entry}
    0
    \,<\,
    \getentry{
      \fpinst{A_\mathrm{e}}{\ol u}{\ol\theta}
    }{r,p}
    =
    \exp
    \big(
      {-}
      \exp \getentry{\ol a}{r,p}
      \cdot
      \getentry{
        \fpinst{\Delta_\mathrm{e}}{\ol u}{\ol\theta}
      }{p}
    \big)
    \,<\,
    1.
  \end{align}

  By
  Equation~\eqref{eq:prop:mamba-rncs-smooth-contractive-canonical-matrix},
  we have that $\fpinst{A_\mathrm{e}}{\ol u}{\ol\theta}$ is diagonal, 
  and therefore
  \begin{align*}
    \left\|
    \fpinst{A_\mathrm{e}}{\ol u}{\ol\theta}
    \right\|
    \,=\,
    \max_{
      \substack{
        r\in\iival{1}{d_\mathrm{s}}
        \\[.1em]
        p\in\iival{1}{n_\mathrm{h}}
      }
    }
    \getentry{
      \fpinst{A_\mathrm{e}}{\ol u}{\ol\theta}
    }{r,p}
    \,<\,
    1,
  \end{align*}
  where the last inequality follows from
  Equation~\eqref{eq:prop:mamba-rncs-smooth-contractive-entry}
  and the fact that the maximum is taken over finitely many entries.

  Since
  $\ptuple{\ol u,\ol\theta}$
  was arbitrary,
  we conclude that
  $\tilde A$
  satisfies
  Equation~\eqref{eq:closure-contraction}
  of
  Definition~\ref{def:smooth-contractive-linear-rnn}.

  Since
  $i\in\iival{1}{\ell}$
  and
  $j\in\iival{1}{m_i}$
  were arbitrary,
  every Parametric Linear Recurrent Unit
  $\tilde D_{i,j}$
  occurring in the flattened recurrent cascade presentation
  $\presentationSty{\tilde C}$
  satisfies both recurrent-unit requirements of
  Definition~\ref{def:smooth-contractive-linear-rnn}.

  It remains to verify the required regularity of the layer output functions.
  Since
  $\presentationSty{C}$
  is a parametric recurrent neural cascade,
  we have that
  every connecting function $g_i$ is analytic.
  Furthermore,
  since
  $\presentationSty{S}$
  is a parametric analytic output extension presentation,
  the parametric output function $h$ is analytic.

  By
  Definition~\ref{def:canonical-linear-rnn},
  the first connecting function $g_1$ is unchanged;
  for every $i\in\iival{2}{\ell}$,
  the corresponding flattened connecting function satisfies
  \begin{equation*}
    \mBigfpinst{\tilde g_i}{
      \tilde x_\mathrm{pre},
      u,
      \tilde x_\mathrm{cur}
    }{
      \theta_{g_i}
    }
    \,=\,
    \mBigfpinst{g_i}{
      \phi_{<i}^{-1}
      \big(
        \tilde x_\mathrm{pre}
      \big),
      u,
      \phi_{<i}^{-1}
      \big(
        \tilde x_\mathrm{cur}
      \big)
    }{
      \theta_{g_i}
    };
  \end{equation*}
  and the flattened output function satisfies
  \begin{equation*}
    \mBigfpinst{\tilde h}{
      \tilde x_\mathrm{pre},
      u,
      \tilde x_\mathrm{cur}
    }{
      \theta_h
    }
    \,=\,
    \mBigfpinst{h}{
      \phi_\presentationSty{C}^{-1}
      \big(
        \tilde x_\mathrm{pre}
      \big),
      u,
      \phi_\presentationSty{C}^{-1}
      \big(
        \tilde x_\mathrm{cur}
      \big)
    }{
      \theta_h
    }.
  \end{equation*}

  By
  Fact~\ref{fact:flattening-rnc-bijection},
  every
  $\phi_{<i}^{-1}$
  and
  $\phi_\presentationSty{C}^{-1}$
  is linear and hence analytic.
  Therefore,
  by
  Proposition~\ref{prop:analytic-maps-closure},
  every connecting function of
  $\presentationSty{\tilde C}$
  and the final output function
  $\tilde h$
  are parametric analytic.
  By
  Fact~\ref{fact:analytic-euclidean-map-smooth-up-to-closure},
  the corresponding layer output functions of
  $\tilde S$
  are continuously-differentiable up to the closures of their domains.

  Thus all conditions of
  Definition~\ref{def:smooth-contractive-linear-rnn}
  hold for
  $\tilde S$.
  Therefore,
  $\tilde S$
  is a smooth unitwise-contractive Linear RNN.
\end{proof}

\subsection{Positive-margin expressivity of Mamba}
\label{subsec:mamba--positive-margin-expressivity}

\begin{closingtheorem}[Positive-margin expressivity of Mamba]
  \label{thm:positive-margin-mamba-expressivity}
  Every language recognised with positive margin by a Mamba model is definite.
\end{closingtheorem}
\begin{proof}
  Let
  $S\in\classmodels{Mamba}$
  recognise a language
  $L$
  with positive margin $\gamma \in \posreals$.

  By
  Definition~\ref{def:mamba-rnc--new},
  we have that $S$ is a parametric $\classunits{Mamba}$-RNC.
  Hence,
  by the definition of parametric RNCs,
  there exists a parametric recurrent cascade presentation
  $\presentationSty{C}$
  of type
  $\bigptuple{
    \classunits{Mamba},
    \analyticparamfuncclass
  }$
  such that
  \begin{equation*}
    \presentationVal{
      \presentationSty{C}
    }
    =
    \dynop{S}.
  \end{equation*}

  Let $x^\mathrm{init}$ be the initial state of $S$,
  and
  let $h$ be its parametric output function.
  Since $S$ is a parametric RNC,
  we have that $h$ is analytic,
  and hence the output extension presentation 
  \begin{equation*}
    \presentationSty{S}
    \,\defeq\,
    \repoutoutextension{
      \presentationSty{C}
    }{
      x^\mathrm{init}
    }{
      h
    }
  \end{equation*}
  is an analytic output extension of $S$.

  Let
  $\presentationSty{\tilde S}$
  be the Linear RNN presentation associated with
  $\presentationSty{S}$ according to
  Definition~\ref{def:canonical-linear-rnn},
  and 
  let 
  \begin{equation*}
    \tilde S
    \,\defeq\,
    \presentationVal{
      \presentationSty{\tilde S}
    }
  \end{equation*}
  be the corresponding Linear RNN representation of $S$.

  By
  Proposition~\ref{prop:linear-rnn-representation-correctness},
  we have that
  $S$
  and
  $\tilde S$
  have the same input-output behaviour under the same parameter values.
  Hence
  $\tilde S$
  recognises
  $L$
  with the same margin $\gamma$.

  By Theorem~\ref{thm:mamba-is-smooth-unitwise-contractive},
  we have that $\tilde S$ is a smooth unitwise-contractive Linear RNN.

  Therefore,
  by Theorem~\ref{thm:positive-margin-smooth-contractive-linear-rnn-expressivity},
  we conclude that $L$ is definite.
\end{proof}

\clearpage{}%

\clearpage{}%

\section{Analysis of mLSTM}
\label{app:mLSTM}

\todo{}

\subsection{Definition of mLSTM}
\label{subsec:definition-of-mlstm}

\begin{closingdefinition}[Generic Stabilised Sigmoid-Forget mLSTM Recurrent Units]
  \label{def:generic-stabilised-sigmoid-forget-mlstm-unit}

  Let
  $d_\mathrm{k},d_\mathrm{v}\in\posnaturals$.
  The
  \emph{admissible stabilised sigmoid-forget mLSTM input space}
  of key-value dimensions
  $\ptuple{d_\mathrm{k},d_\mathrm{v}}$
  is
  \begin{equation*}
    \mathcal U_\mathrm{mLSTM}
    \defeq
    \posreals
    \times
    (0,1)
    \times
    \reals^{d_\mathrm{k}}
    \times
    \reals^{d_\mathrm{v}}.
  \end{equation*}

  A
  \emph{Generic Stabilised Sigmoid-Forget mLSTM Recurrent Unit}
  is dynamics
  \begin{equation*}
    \hat D
    =
    \tuple{
      X,
      \hat U,
      \hat f
    },
    \qquad
    X
    \subseteq
    \reals^{d_\mathrm{v}\times d_\mathrm{k}}
    \times
    \reals^{d_\mathrm{k}}
    \times
    \reals,
    \qquad
    \hat U
    \subseteq
    \mathcal U_\mathrm{mLSTM},
  \end{equation*}
  whose state is written as
  \begin{equation*}
    x
    =
    \bigptuple{
      X_\mathrm{c},
      x_\mathrm{n},
      x_\mathrm{m}
    }
  \end{equation*}
  and whose input is written as
  \begin{equation*}
    \hat u
    =
    \bigptuple{
      \bar g_\mathrm{i},
      \bar g_\mathrm{f},
      k,
      v
    }.
  \end{equation*}

  For every
  $x\in X$
  and
  $\hat u\in\hat U$,
  define
  \begin{align*}
    x'_\mathrm{m}
    &\defeq
    \max
    \left\{
      \log(\bar g_\mathrm{f})+x_\mathrm{m},
      \log(\bar g_\mathrm{i})
    \right\},
    \\[.3em]
    g_\mathrm{f}
    &\defeq
    \exp
    \left(
      \log(\bar g_\mathrm{f})
      +
      x_\mathrm{m}
      -
      x'_\mathrm{m}
    \right),
    \\[.3em]
    g_\mathrm{i}
    &\defeq
    \exp
    \left(
      \log(\bar g_\mathrm{i})
      -
      x'_\mathrm{m}
    \right),
  \end{align*}
  and
  \begin{align*}
    X'_\mathrm{c}
    &\defeq
    g_\mathrm{f}
    \cdot
    X_\mathrm{c}
    +
    g_\mathrm{i}
    \cdot
    v
    \cdot
    k^\transposeop,
    \\[.3em]
    x'_\mathrm{n}
    &\defeq
    g_\mathrm{f}
    \cdot
    x_\mathrm{n}
    +
    g_\mathrm{i}
    \cdot
    k.
  \end{align*}

  Then
  \begin{equation*}
    \hat f
    \left(
      \bigptuple{
        X_\mathrm{c},
        x_\mathrm{n},
        x_\mathrm{m}
      },
      \bigptuple{
        \bar g_\mathrm{i},
        \bar g_\mathrm{f},
        k,
        v
      }
    \right)
    \defeq
    \bigptuple{
      X'_\mathrm{c},
      x'_\mathrm{n},
      x'_\mathrm{m}
    }.
  \end{equation*}

  We call
  $X_\mathrm{c}$,
  $x_\mathrm{n}$,
  and
  $x_\mathrm{m}$
  respectively the
  \emph{memory state},
  \emph{normaliser state},
  and
  \emph{stabiliser state}.
\end{closingdefinition}

\begin{closingdefinition}[Stabilised Sigmoid-Forget mLSTM]
  \label{def:stabilised-sigmoid-forget-mlstm}

  Let
  $d_\mathrm{in},d_\mathrm{k},d_\mathrm{v}
  \in\posnaturals$,
  and let
  \begin{equation*}
    \hat D
    =
    \tuple{
      X,
      \hat U,
      \hat f
    }
  \end{equation*}
  be a Generic Stabilised Sigmoid-Forget mLSTM Recurrent Unit with
  key-value dimensions
  $\ptuple{d_\mathrm{k},d_\mathrm{v}}$.

  A
  \emph{Stabilised Sigmoid-Forget mLSTM Recurrent Unit}
  is parametric dynamics
  \begin{equation*}
    D
    =
    \ctuple{
      X,
      U,
      f
    }{
      \Theta
    },
    \qquad
    U
    \subseteq
    \reals^{d_\mathrm{in}},
  \end{equation*}
  having components as described next.

  \begin{definitionpartsimple}[Parameter space]
    \label{def:stabilised-sigmoid-forget-mlstm-parameter-space}
    The parameter space is a finite-dimensional Euclidean space
    \begin{equation*}
      \Theta
      =
      \Theta_\mathrm{k}
      \times
      \Theta_\mathrm{v}
      \times
      \Theta_\mathrm{i}
      \times
      \Theta_\mathrm{f}.
    \end{equation*}
  \end{definitionpartsimple}

  \begin{definitionpartsimple}[Recurrence-input functions]
    \label{def:stabilised-sigmoid-forget-mlstm-recurrence-input-functions}
    The recurrence-input functions are parametric feedforward neural maps
    \begin{align*}
      k
      &:
      U\times\Theta_\mathrm{k}
      \to
      \reals^{d_\mathrm{k}},
      &
      v
      &:
      U\times\Theta_\mathrm{v}
      \to
      \reals^{d_\mathrm{v}},
      \\[.3em]
      \widetilde g_\mathrm{i}
      &:
      U\times\Theta_\mathrm{i}
      \to
      \reals,
      &
      \widetilde g_\mathrm{f}
      &:
      U\times\Theta_\mathrm{f}
      \to
      \reals,
    \end{align*}
    with raw gate functions
    \begin{align*}
      \bar g_\mathrm{i}
      &:
      U\times\Theta_\mathrm{i}
      \to
      \posreals,
      &
      \fpinst{\bar g_\mathrm{i}}{u}{\theta_\mathrm{i}}
      &\defeq
      \exp
      \left(
        \fpinst{\widetilde g_\mathrm{i}}{u}{\theta_\mathrm{i}}
      \right),
      \\[.3em]
      \bar g_\mathrm{f}
      &:
      U\times\Theta_\mathrm{f}
      \to
      (0,1),
      &
      \fpinst{\bar g_\mathrm{f}}{u}{\theta_\mathrm{f}}
      &\defeq
      \sigma
      \left(
        \fpinst{\widetilde g_\mathrm{f}}{u}{\theta_\mathrm{f}}
      \right).
    \end{align*}
  \end{definitionpartsimple}

  \begin{definitionpartsimple}[Dynamics function]
    \label{def:stabilised-sigmoid-forget-mlstm-dynamics-function}
    For
    \begin{equation*}
      \theta
      =
      \bigptuple{
        \theta_\mathrm{k},
        \theta_\mathrm{v},
        \theta_\mathrm{i},
        \theta_\mathrm{f}
      }
      \in
      \Theta,
    \end{equation*}
    the parametric dynamics function is obtained by applying the dynamics
    function of
    $\hat D$
    to the parametrised recurrence input:
    \begin{align*}
      \fpinst{f}{x,u}{\theta}
      \defeq
      \hat f
      \mBig(
        x,\,
        \bigptuple{
          \fpinst{\bar g_\mathrm{i}}{u}{\theta_\mathrm{i}},
          \fpinst{\bar g_\mathrm{f}}{u}{\theta_\mathrm{f}},
          \fpinst{k}{u}{\theta_\mathrm{k}},
          \fpinst{v}{u}{\theta_\mathrm{v}}
        }
      \mBig).
    \end{align*}
  \end{definitionpartsimple}

  \begin{definitionpartsimple}[Models and classes]
    \label{def:stabilised-sigmoid-forget-mlstm-models}
    An
    \emph{mLSTM$(\exp,\sigma;1,1)$ model}
    is a parametric Recurrent Neural Cascade whose recurrent units are
    Stabilised Sigmoid-Forget mLSTM Recurrent Units and whose recurrent
    initial states are zero.
  \end{definitionpartsimple}
\end{closingdefinition}

\subsection{Positive-Margin Expressivity}
\label{subsec:mlstm-positive-margin-expressivity}

The complete stabilised mLSTM recurrence is not strictly contractive:
when the first branch of the stabiliser maximum is active,
the stabiliser state is propagated with coefficient one.
Nevertheless, sigmoid forget gating yields a stronger property sufficient for
the present expressivity analysis:
dependence on an arbitrarily remote prefix vanishes uniformly.
We establish this first for a single recurrent unit.

\begin{closinglemma}[Fading memory of stabilised sigmoid-forget mLSTM units]
  \label{lemma:mlstm-sigmoid-forget-fading-memory}

  Let
  $D$
  be an instance of a Stabilised Sigmoid-Forget mLSTM Recurrent Unit,
  and let
  $K$
  be a bounded subset of its input space.

  Consider an arbitrary family
  $\mathcal P$
  of pairs of input sequences
  \begin{equation*}
    \ptuple{u,u'}
    \in
    \infseqs{K}
    \times
    \infseqs{K},
  \end{equation*}
  together with initial-state pairs
  $\ptuple{x_0,x'_0}$
  ranging over a bounded set.
  Suppose that
  \begin{equation}
    \label{eq:mlstm-input-tail-convergence}
    \eta_t
    \defeq
    \sup_{
      \ptuple{u,u'}
      \in
      \mathcal P
    }
    \|u_t-u'_t\|
    \to
    0
    \qquad
    \text{as }
    t\to\infty.
  \end{equation}

  Then the corresponding state trajectories satisfy
  \begin{equation}
    \label{eq:mlstm-state-tail-convergence}
    \sup_{
      \ptuple{u,u'}
      \in
      \mathcal P
    }
    \|x_t-x'_t\|
    \to
    0
    \qquad
    \text{as }
    t\to\infty,
  \end{equation}
  uniformly over the bounded set of initial-state pairs.
\end{closinglemma}

\begin{proof}
  Fix the parameter value of the considered instance and define
  \begin{equation*}
    a(u)
    \defeq
    \log
    \big(
      \bar g_\mathrm{f}(u)
    \big),
    \qquad
    b(u)
    \defeq
    \log
    \big(
      \bar g_\mathrm{i}(u)
    \big).
  \end{equation*}
  Since
  \begin{equation*}
    \bar g_\mathrm{f}(u)
    =
    \sigma
    \big(
      \widetilde g_\mathrm{f}(u)
    \big),
  \end{equation*}
  and the recurrence-input functions extend continuously to the compact
  closure of
  $K$,
  there exist constants
  \begin{equation*}
    \delta\in\posreals,
    \qquad
    b_-,b_+\in\reals
  \end{equation*}
  such that
  \begin{equation}
    \label{eq:mlstm-stabiliser-uniform-bounds}
    a(u)
    \leq
    -\delta,
    \qquad
    b_-
    \leq
    b(u)
    \leq
    b_+
    \qquad
    \forall
    u\in K.
  \end{equation}

  Consider first the stabiliser state.
  Its recurrence is
  \begin{equation}
    \label{eq:mlstm-stabiliser-recurrence}
    x_{\mathrm m,t}
    =
    \max
    \left\{
      a(u_t)+x_{\mathrm m,t-1},
      b(u_t)
    \right\}.
  \end{equation}
  Expanding
  Equation~\eqref{eq:mlstm-stabiliser-recurrence}
  gives
  \begin{align}
    x_{\mathrm m,t}
    =
    \max
    \Bigg\{
      &
      x_{\mathrm m,0}
      +
      \sum_{j=1}^t a(u_j),
      \notag
      \\
      &
      b(u_k)
      +
      \sum_{j=k+1}^t a(u_j)
      \;\Bigm|\;
      k\in\iival{1}{t}
    \Bigg\}.
    \label{eq:mlstm-stabiliser-expanded}
  \end{align}

  Let
  $M_\mathrm{m}$
  uniformly bound the initial stabiliser states under consideration.
  Choose
  $N_\mathrm{m}\in\posnaturals$
  sufficiently large that
  \begin{equation}
    \label{eq:mlstm-stabiliser-window-choice}
    M_\mathrm{m}
    -
    N_\mathrm{m}\delta
    <
    b_-,
    \qquad
    b_+
    -
    N_\mathrm{m}\delta
    <
    b_-.
  \end{equation}

  By
  Equations~\eqref{eq:mlstm-stabiliser-uniform-bounds}
  and~\eqref{eq:mlstm-stabiliser-window-choice},
  for every
  $t\geq N_\mathrm{m}$,
  neither the candidate originating from
  $x_{\mathrm m,0}$
  nor a candidate introduced more than
  $N_\mathrm{m}$
  steps earlier can attain the maximum in
  Equation~\eqref{eq:mlstm-stabiliser-expanded}.
  Therefore
  \begin{equation}
    \label{eq:mlstm-stabiliser-finite-window}
    x_{\mathrm m,t}
    =
    \max_{
      k\in
      \iival{
        t-N_\mathrm{m}+1
      }{
        t
      }
    }
    \left\{
      b(u_k)
      +
      \sum_{j=k+1}^t
      a(u_j)
    \right\}.
  \end{equation}

  Thus the stabiliser is not contractive in one step, but after
  $N_\mathrm{m}$
  steps it is independent of its preceding state and depends only on a
  uniformly bounded window of recurrent inputs.

  Since
  $a$
  and
  $b$
  are uniformly continuous on the compact closure of
  $K$,
  Equation~\eqref{eq:mlstm-input-tail-convergence}
  and the finite-window representation
  Equation~\eqref{eq:mlstm-stabiliser-finite-window}
  imply
  \begin{equation}
    \label{eq:mlstm-stabiliser-tail-convergence}
    |x_{\mathrm m,t}-x'_{\mathrm m,t}|
    \to
    0
  \end{equation}
  uniformly over the considered family.

  We next consider the memory and normaliser states.
  Define their rescaled forms by
  \begin{equation*}
    \bar X_{\mathrm c,t}
    \defeq
    \exp(x_{\mathrm m,t})
    \cdot
    X_{\mathrm c,t},
    \qquad
    \bar x_{\mathrm n,t}
    \defeq
    \exp(x_{\mathrm m,t})
    \cdot
    x_{\mathrm n,t}.
  \end{equation*}
  Direct substitution in the dynamics equations gives
  \begin{align}
    \bar X_{\mathrm c,t}
    &=
    \bar g_\mathrm{f}(u_t)
    \cdot
    \bar X_{\mathrm c,t-1}
    +
    \bar g_\mathrm{i}(u_t)
    \cdot
    v(u_t)
    \cdot
    k(u_t)^\transposeop,
    \label{eq:mlstm-rescaled-memory-recurrence}
    \\[.3em]
    \bar x_{\mathrm n,t}
    &=
    \bar g_\mathrm{f}(u_t)
    \cdot
    \bar x_{\mathrm n,t-1}
    +
    \bar g_\mathrm{i}(u_t)
    \cdot
    k(u_t).
    \label{eq:mlstm-rescaled-normaliser-recurrence}
  \end{align}

  Compactness and sigmoid forget gating give a constant
  \begin{equation}
    \label{eq:mlstm-raw-forget-uniform-contraction}
    \alpha
    \in
    [0,1)
    \qquad\text{such that}\qquad
    \bar g_\mathrm{f}(u)
    \leq
    \alpha
    \qquad
    \forall
    u\in K.
  \end{equation}
  Furthermore,
  all additive terms in
  Equations~\eqref{eq:mlstm-rescaled-memory-recurrence}
  and~\eqref{eq:mlstm-rescaled-normaliser-recurrence}
  are bounded and uniformly continuous on the compact closure of
  $K$.
  Hence the rescaled state trajectories are uniformly bounded.

  Stack the rescaled memory and normaliser states into a vector
  $\bar x_t$.
  Comparing two trajectories gives an estimate
  \begin{equation}
    \label{eq:mlstm-rescaled-state-difference}
    \|
      \bar x_t
      -
      \bar x'_t
    \|
    \leq
    \alpha
    \cdot
    \|
      \bar x_{t-1}
      -
      \bar x'_{t-1}
    \|
    +
    e_t,
  \end{equation}
  where
  \begin{equation*}
    e_t
    \to
    0
  \end{equation*}
  uniformly over the considered family, by
  Equation~\eqref{eq:mlstm-input-tail-convergence},
  uniform boundedness of the rescaled states,
  and uniform continuity of the recurrence-input functions.

  Iterating
  Equation~\eqref{eq:mlstm-rescaled-state-difference}
  gives
  \begin{equation*}
    \|
      \bar x_t
      -
      \bar x'_t
    \|
    \leq
    \alpha^t
    \cdot
    \|
      \bar x_0
      -
      \bar x'_0
    \|
    +
    \sum_{s=1}^t
    \alpha^{t-s}
    \cdot
    e_s.
  \end{equation*}
  Since
  $\alpha<1$
  and
  $e_s\to0$
  uniformly,
  both terms converge uniformly to zero.
  Therefore
  \begin{equation}
    \label{eq:mlstm-rescaled-state-tail-convergence}
    \|
      \bar x_t
      -
      \bar x'_t
    \|
    \to
    0.
  \end{equation}

  Finally,
  Equation~\eqref{eq:mlstm-stabiliser-expanded}
  and
  Equation~\eqref{eq:mlstm-stabiliser-uniform-bounds}
  give uniform upper and lower bounds on the stabiliser states.
  Consequently,
  the inverse rescaling
  \begin{equation*}
    \bigptuple{
      \bar X_\mathrm{c},
      \bar x_\mathrm{n},
      x_\mathrm{m}
    }
    \longmapsto
    \bigptuple{
      \exp(-x_\mathrm{m}) \cdot \bar X_\mathrm{c},\,
      \exp(-x_\mathrm{m}) \cdot \bar x_\mathrm{n},\,
      x_\mathrm{m}
    }
  \end{equation*}
  is uniformly continuous on the bounded set under consideration.
  Equations~\eqref{eq:mlstm-stabiliser-tail-convergence}
  and~\eqref{eq:mlstm-rescaled-state-tail-convergence}
  therefore imply
  Equation~\eqref{eq:mlstm-state-tail-convergence}.
\end{proof}

\begin{closinglemma}[Uniform suffix forgetting of stabilised sigmoid-forget mLSTM RNCs]
  \label{lemma:mlstm-positive-margin-uniform-suffix-forgetting}

  Let
  $S$
  be an instance of an
  mLSTM$(\exp,\sigma;1,1)$ model,
  and let
  \begin{equation*}
    \enc
    :
    \Sigma
    \to
    U
  \end{equation*}
  be an encoder from a finite alphabet
  $\Sigma$.

  Then, for some sequence
  \begin{equation*}
    (\varepsilon_n)_{n\in\posnaturals}
    \in
    \infseqs{\nonnegreals}
    \qquad\quad\text{with}\qquad\quad
    \varepsilon_n
    \to
    0
    \quad\text{as}\quad
    n\to\infty,
  \end{equation*}
  we have
  \begin{equation}
    \label{eq:mlstm-positive-margin-uniform-suffix-forgetting}
    d_Y
    \mBig(
      S\big(\enc(vz)\big),
      S\big(\enc(wz)\big)
    \mBig)
    \leq
    \varepsilon_n
  \end{equation}
  for every
  \begin{equation*}
    n\in\posnaturals,
    \qquad
    v,w\in\Sigma^*,
    \qquad
    z\in\Sigma^n.
  \end{equation*}
\end{closinglemma}

\begin{proof}
  Since
  $\Sigma$
  is finite,
  the encoded input set
  $\enc(\Sigma)$
  is bounded.

  Consider the first recurrent layer.
  Lemma~\ref{lemma:mlstm-sigmoid-forget-fading-memory},
  applied with zero initial states and bounded inputs,
  first gives uniform boundedness of its recurrent states over all encoded
  words.
  The corresponding feedforward layer-output maps therefore have bounded
  reachable argument sets and bounded outputs.

  Consider now two computations which have processed arbitrary prefixes
  $v,w\in\Sigma^*$
  and subsequently process the same suffix
  \begin{equation*}
    z
    =
    z_1\cdots z_n.
  \end{equation*}
  During this common suffix, every recurrent unit in the first layer receives
  exactly the same input in the two computations.
  Thus
  Equation~\eqref{eq:mlstm-input-tail-convergence}
  holds with
  $\eta_t=0$.
  Lemma~\ref{lemma:mlstm-sigmoid-forget-fading-memory}
  consequently implies that the first-layer recurrent states converge
  uniformly to one another as the length of the common suffix tends to
  infinity.
  Uniform continuity of the corresponding layer-output map on the bounded
  reachable argument set yields the same conclusion for the first-layer
  outputs.

  Proceed inductively through the finite cascade.
  Suppose that the outputs of one layer in the two computations converge
  uniformly to one another during increasingly long common suffixes.
  These outputs determine the recurrent inputs
  $u_t$
  and
  $u'_t$
  received by the next layer.
  Uniform continuity of the corresponding inter-layer map therefore gives
  \begin{equation*}
    \|u_t-u'_t\|
    \to
    0
  \end{equation*}
  uniformly over prefixes and common suffixes.
  Lemma~\ref{lemma:mlstm-sigmoid-forget-fading-memory}
  then implies uniform convergence of the recurrent states of the next layer,
  and uniform continuity of its output map implies uniform convergence of its
  outputs.

  Since the cascade contains finitely many layers, the final outputs satisfy
  \begin{equation*}
    \sup_{\substack{
      v,w\in\Sigma^*
      \\
      z\in\Sigma^n
    }}
    d_Y
    \mBig(
      S\big(\enc(vz)\big),\,
      S\big(\enc(wz)\big)
    \mBig)
    \to
    0
    \qquad
    \text{as }
    n\to\infty.
  \end{equation*}
  Defining the displayed supremum to be
  $\varepsilon_n$
  proves the claim.
\end{proof}

\begin{closingtheorem}[Positive-margin expressivity of stabilised sigmoid-forget mLSTM RNCs]
  \label{thm:mlstm-positive-margin-expressivity}
  The positive-margin expressivity of mLSTM$(\exp,\sigma;1,1)$ models
  is contained in the definite languages.
\end{closingtheorem}
\begin{proof}
  Let
  $S$
  be an instance of an
  mLSTM$(\exp,\sigma;1,1)$ model that recognises a language
  $L$
  over a finite alphabet
  $\Sigma$
  with positive margin,
  through an encoder
  $\enc$
  and decoder
  $\dec$.
  Let
  $\gamma\in\posreals$
  be a corresponding decoder margin.

  By
  Lemma~\ref{lemma:mlstm-positive-margin-uniform-suffix-forgetting},
  choose
  $N\in\posnaturals$
  such that
  \begin{equation*}
    \varepsilon_N
    <
    \gamma.
  \end{equation*}
  Then, for every
  $v,w\in\Sigma^*$
  and
  $z\in\Sigma^N$,
  \begin{equation*}
    d_Y
    \mBig(
      S\big(\enc(vz)\big),\,
      S\big(\enc(wz)\big)
    \mBig)
    <
    \gamma.
  \end{equation*}
  By the decoder-margin condition,
  \begin{equation*}
    \dec
    \Big(
      S\big(\enc(vz)\big)
    \Big)
    =
    \dec
    \Big(
      S\big(\enc(wz)\big)
    \Big),
  \end{equation*}
  and hence
  \begin{equation*}
    vz\in L
    \quad\Longleftrightarrow\quad
    wz\in L.
  \end{equation*}

  Lemma~\ref{lemma:positive-margin-definite-suffix-characterisation}
  therefore implies that
  $L$
  is definite.
  Since
  $S$
  and
  $L$
  were arbitrary,
  the claim follows.
\end{proof}

\begin{closingremark}[The result does not rely on strict contractivity]
  \label{rem:mlstm-positive-margin-not-contractive}

  Stabilised sigmoid-forget mLSTM dynamics are not strictly contractive in
  their complete state:
  while the first branch of the stabiliser maximum is active, the stabiliser
  propagates its preceding value with coefficient one.

  The preceding result instead follows from two complementary mechanisms.
  The stabiliser has a uniformly bounded memory horizon because every
  continuation of an old candidate accumulates
  \begin{equation*}
    \log(\bar g_\mathrm{f})
    \leq
    -\delta
    <
    0,
  \end{equation*}
  whereas the rescaled memory and normaliser states are uniformly contractive
  because
  \begin{equation*}
    \bar g_\mathrm{f}
    \leq
    \alpha
    <
    1.
  \end{equation*}
  Together these properties yield uniform suffix forgetting and hence the
  definite-language upper bound under positive-margin recognition.
\end{closingremark}

\begin{closingremark}[Relation to the official mLSTM implementation]
  \label{rem:mlstm-official-implementation}

  The stabilised sigmoid-forget mLSTM considered above is in line with the
  recurrent mechanism used by the current official NX-AI xLSTM
  implementation~\citep{xlstmimpl}.
  In particular, its recurrent backend applies a sigmoid forget gate in
  log-space,
  updates the stabiliser according to
  \begin{equation*}
    x'_\mathrm{m}
    =
    \max
    \big\{
      \log(\bar g_\mathrm{f})+x_\mathrm{m},
      \widetilde g_\mathrm{i}
    \big\},
  \end{equation*}
  and uses the corresponding stabilised gates
  \begin{equation*}
    g_\mathrm{f}
    =
    \exp
    \big(
      \log(\bar g_\mathrm{f})
      +x_\mathrm{m}
      -x'_\mathrm{m}
    \big),
    \qquad
    g_\mathrm{i}
    =
    \exp
    \big(
      \widetilde g_\mathrm{i}
      -x'_\mathrm{m}
    \big).
  \end{equation*}
  Thus, in the notation of
  Definition~\ref{def:stabilised-sigmoid-forget-mlstm},
  the implemented recurrent mechanism corresponds to the
  $\mathrm{mLSTM}(\exp,\sigma;1,1)$
  variant considered here.

  The official implementation additionally includes architectural details
  such as multi-head organisation, query-dependent readout, normalisation,
  and optimised parallel or chunkwise kernels.
  These do not change the input-gate, forget-gate, and stabiliser recurrence
  underlying the preceding positive-margin expressivity result.
\end{closingremark}

\clearpage{}%

\end{document}